\documentclass[12pt]{book}

\usepackage{amsmath,amsthm, thmtools}

\usepackage[small,compact]{titlesec}
\usepackage{titletoc,titleps}

\usepackage{kpfonts}
\usepackage{bm,yfonts}
\usepackage[bb=boondox,bbscaled=1.05]{mathalfa}
\usepackage{cleveref}
\usepackage{stackrel}
\usepackage{enumitem}
\usepackage{graphicx}
\usepackage{fancyhdr}
\usepackage{color}
\usepackage[font=small]{caption}
\usepackage[margin=3cm]{geometry}
   
\usepackage[numbers]{natbib}
\usepackage{tikz}
\usetikzlibrary{positioning}
\usetikzlibrary{decorations.pathreplacing,calc} 
\usetikzlibrary{shapes.geometric}
\usetikzlibrary {arrows.meta}
 
\input Headers/newcom
\newcommand{\mA}{{\mathbb A}}

\newcommand{\mS}{{\mathbb S}}
\newcommand{\mT}{{\mathbb T}} 

\newcommand{\mW}{{\mathbb W}}
\newcommand{\defeq}{\stackrel{{\scriptscriptstyle \Delta}}{=}}
\renewcommand{\bfone}{{\boldsymbol 1}}

\newcommand{\dom}{\mathrm{dom}}
\newcommand{\aff}{\mathrm{aff}}
\newcommand{\vaff}{\overrightarrow{\mathrm{aff}}}
\newcommand{\rdom}{\mathrm{ridom}}
\newcommand{\relint}{\mathrm{relint}}
\newcommand{\epi}{\mathrm{epi}}
\newcommand{\proj}{\mathrm{proj}}
\newcommand{\KL}{\mathit{KL}}
\newcommand{\VC}{\textit{VC-dim}}
\newcommand{\Sdim}{\textit{S-dim}}
\newcommand{\Pdim}[1][]{P_{#1}\textit{-dim}}
\newcommand{\Rad}{\textit{Rad}}
\newcommand{\vol}{\text{vol}}
\newcommand{\CTR}{{\textgoth T}}
\newcommand{\hCP}{{\widehat{\mathcal{P}}}}

\newcommand{\vspan}{\mathrm{span}}
\newcommand{\rank}{\mathrm{rank}}
\newcommand{\dsone}{{\mathbb 1}}
\newcommand{\dszero}{{\mathbb 0}}

\newcommand{\pa}[1]{{\mathit{pa}(#1)}}
\newcommand{\cl}[1]{{\mathit{ch}(#1)}}

\renewcommand{\ln}{\log}

\newcommand{\myP}{\boldsymbol{\mathbb P}}
\newcommand{\myE}{\boldsymbol{\mathbb E}}

\newcommand{\wtilde}{\widetilde}
\newcommand{\what}{\widehat}

\usepackage{xcolor}
\usepackage{amsmath,thmtools}
\newcounter{thequestion}
\newcounter{thesubquestion}
\newcounter{thesubsubquestion}
{}{}%

\declaretheoremstyle[
spaceabove=6pt, spacebelow=6pt,
headfont=\normalfont\bfseries,
notefont=\mdseries, notebraces={(}{)},
bodyfont=\rm,
postheadspace=\newline]
{probstyle}

\definecolor{myred}{rgb}{.5,0,0}

\newif\ifsolutions

\newif\ifuselabel

\newcommand{\indp}{{\Bot}}
\newcommand{\cind}[3]{(#1\indp#2\mid #3)}
\newcommand{\twop}[1]{{\boldsymbol{\mathcal P}({#1})}}

\solutionsfalse
\uselabeltrue

\newcommand{\Eq}[1]
{
\[
\displaystyle #1
\]
}

\newcommand{\alert}[1]{{\em #1}}

\declaretheorem[within=chapter]{all2}
\declaretheorem[sibling=all2, style=plain]{corollary,definition,theorem,proposition,
lemma}

\declaretheoremstyle[
spaceabove=6pt, spacebelow=6pt,
headfont={\normalfont\bfseries},
notefont=\mdseries, notebraces={(}{)},
bodyfont=\normalfont,
postheadspace=1em,
qed=\ensuremath{\blacklozenge}]
{myremark}
\declaretheorem[sibling=all2,style=myremark]{remark,example,notation}

\declaretheoremstyle[
spaceabove=6pt, spacebelow=6pt,
headfont=\normalfont\bfseries,
notefont=\mdseries, notebraces={(}{)},
bodyfont=\rm,
postheadhook = {\hspace*{\parindent}},
postheadspace=\newline
]
{algstyle}
\declaretheorem[sibling=all2,style=algstyle,preheadhook=\vskip1cm\hrule,
postfoothook=\hrule\vskip1cm]
{algorithm}

\crefname{equation}{}{}
\Crefname{equation}{Equation}{Equations}
\crefname{definition}{Definition}{Definitions}
\crefname{notation}{Notation}{Notations}
\crefname{algorithm}{Algorithm}{Algorithms}
\Crefname{algorithm}{Algorithm}{Algorithms}
\setlist[enumerate]{label=\arabic*., wide=0.5cm}

\renewcommand{\cite}{\citep}
\graphicspath{{./},{Figures/}}

\setlist[enumerate]{label= (\arabic*), labelsep=*, leftmargin=0.25cm, wide=0cm, itemsep=0cm, topsep=0.1cm, ref=\arabic*}
\setlist[itemize]{label= \textbullet, labelsep=*, leftmargin=0.25cm, wide=0cm, itemsep=0cm, topsep=0.1cm}

\newcommand{\problems}[1]{\newpage
\markright{PROBLEMS}
\section*{Problems}
\addcontentsline{toc}{section}{Problems}
\input #1
}

\renewcommand{\problems}[1]{}

\title{Introduction to Machine Learning}
\author{Laurent Younes}

\begin{document}

\maketitle
\tableofcontents

\setlength{\parskip}{1em}

\chapter*{Preface}
\addcontentsline{toc}{chapter}{Preface}

Machine learning addresses the issue of analyzing, reproducing and predicting various mechanisms and processes observable through experiments and data acquisition. With the impetus of large technological companies in need of leveraging information included in the gigantic datasets that they produced or obtained through user data, with the development of new data acquisition techniques in biology, physics or astronomy, with the improvement of storage capacity and high-performance computing, this field has experienced an explosive growth over the past decades, in terms of scientific production and technological impact.

While it is being recognized in some places as a scientific discipline in itself, machine learning (which has received a few almost synonymic denominations across time, including artificial intelligence, machine intelligence or statistical learning), can also be seen as an interdisciplinary field interfacing techniques from traditional domains such as computer science, applied mathematics, and statistics. From statistics, and more specially nonparametric statistics, it borrows its main formalism, asymptotic results and generalization bounds. It also builds on many classical methods that have been developed for estimation and prediction. From computer science, it involves the construction and implementation of efficient algorithms, programming design and architecture. Finally, machine learning leverages classical methods from linear algebra and functional analysis, as well as from convex and nonlinear optimization, fields within which it had also provided new problems and discoveries. It forms a significant part of the larger field commonly called ``data science,'' which includes methods for storing, sharing and managing data, the development powerful computer architectures for increasingly demanding algorithms, and, importantly, the definition of ethical limits and processes through which data should be used in the modern world.

This book, which originates from lecture notes of a series of graduate course taught in the Department of Applied Mathematics and Statistics at Johns Hopkins University, adopts a  viewpoint (or bias) mainly focused on the mathematical and statistical aspects of the subject. Its goal is to introduce the mathematical foundations and techniques that lead to the development and analysis of many of the algorithms that are used today. It is written with the hope to provide the reader with a deeper understanding of the algorithms made available to her in multiple machine learning packages and software, and that she will be able to assess their prerequisites and limitations, and to extend them and develop new algorithms. Note that, while adopting a presentation with a strong mathematical flavor, we will still make explicit the details of many important machine learning algorithms.

Unsurprisingly, the book will be  more accessible to a reader with some background in mathematics and statistics. It assumes familiarity with basic concepts in linear algebra and matrix analysis, in multivariate calculus and in probability and statistics. We tried to place a limit at the use of measure theoretic tools, that are avoided up to a few exceptions, which are be localized and be accompanied with alternative interpretations allowing for a reading at a more elementary level. 

The book starts with an introductory chapter that describes notation used throughout the book and serve at a reminder of basic concepts in calculus, linear algebra and probability. It also introduces some measure theoretic terminology, and can be used as a reading guide for the sections that use these tools. This chapter is followed by two chapters offering background material on matrix analysis and optimization. The latter chapter, which is relatively long, provides necessary references to many algorithms that are used in the book, including stochastic gradient descent, proximal methods, etc. 

\Cref{chap:intro}, which is also introductory, illustrates the bias-variance dilemma in machine learning through the angle of density estimation and motivates \cref{chap:general} in which basic concepts for statistical prediction are provided. \Cref{chap:higher} provides an introduction to reproducing kernel theory and Hilbert space techniques that are used in many places, before tackling, with \cref{chap:lin.reg,chap:lin.class,chap:nearest.neighbors,chap:trees,chap:neural.nets}, the description of various algorithms for supervised statistical learning, including linear methods, support vector machines, decision trees, boosting, or neural networks.

\Cref{chap:mcmc}, which presents sampling methods and an introduction to the theory of Markov chains, starts a series of chapters on generative models, and associated learning algorithms. Graphical models and described in \cref{chap:mrf,chap:inf.mrf,chap:bayes.net}. \Cref{chap:var.bayes} introduces variational methods for models with latent variables, with applications to graphical models in \cref{chap:learning.graphical}. Generative techniques using deep learning are presented in \cref{chap:generative}.

\Cref{chap:clustering,chap:dim.red,chap:manifold.learning} focus on unsupervised learning methods, for clustering, factor analysis and manifold learning. The final chapter of the book is theory-oriented and discusses concentration inequalities and generalization bounds.

\chapter{General Notation and Background Material}

\section{Linear algebra}

\subsection{Sets and functions}
If $A$ is a set, the set of all subsets of  $A$ is denoted $\twop{A}$.
If $A$ and $B$ are two sets, the notation $B^A$ refers to the set of all functions $f:A\to B$. In particular, $\mR^A$ is the space of real-valued functions, and forms a vector space. When $A$ is finite, this space is finite dimensional and can be identified with $\mR^{|A|}$, where $|A|$ denotes the cardinality (number of elements) of $A$.

The indicator function of a subset $C$ of $A$ will be denoted $\bfone_C: A \to \{0,1\}$, with $\bfone_C(x) = 1$ if $x\in C$ and 0 otherwise. We will sometimes write $\bfone_{x\in C}$ for $\bfone_C(x)$.

\subsection{Vectors}
Elements of the $d$-dimensional Euclidean space $\mR^d$ will be denoted with letters such as $x,y,z$, and their coordinates will be indexed as parenthesized exponents, so that
\[
 x = \begin{pmatrix}
 \pe{x}{1}\\
 \vdots
 \\
 \pe{x}{d}
 \end{pmatrix}
 \]
 (we will always  identify element of $\mR^d$ with column vectors). We will not distinguish in the notation between ``points'' in  $\mR^d$, seen as an  affine space, and ``vectors'' in  $\mR^d$, seen as a vector space. The vectors ${\mathbf 0}_d$ and $\dsone_d$ will denote the $d$-dimensional vectors with all coordinates equal to 0 and 1, respectively. 
 The identity matrix in $\mR^d$ will be denoted $\Id[d]$. The canonical basis of $\mR^d$, provided by the columns of $\Id[d]$ will be denoted $\mathfrak e_1, \ldots, \mathfrak e_d$.

 The Euclidean norm of a vector $x\in \mR^d$ is denoted $|x|$ with 
 \[
 |x| = \big((\pe{x}{1})^2 + \cdots + (\pe{x}{d})^2\big)^{1/2}.
 \]
 It will sometimes be denoted $|x|_2$, identifying it as a member of the family of $\ell^p$ norms
 \begin{equation}
 |x|_p = \big((\pe{x}{1})^p + \cdots + (\pe{x}{d})^p\big)^{1/p}
\label{eq:lp} 
 \end{equation}
 for $p\geq 1$. One can also define $|x|_p$ for $0<p<1$, using \cref{eq:lp}, but in this case one does not get a norm because the triangle inequality $|x+y|_p \leq |x|_p + |y|_p$ is not true in general. The family is interesting, however, because it approximates, in the limit $p\to 0$,  the number of non-zero components of $x$, denoted $|x|_0$, which is a measure of { sparsity}. Note that we  also use the notation $|A|$ to denote the cardinality (number of elements) of a set $A$, hopefully without risk of confusion.
 
While we use single bars ($|x|$) to represent norms of finite-dimensional vectors, we will use double bars ($\|h\|$) for infinite-dimensional objects.
 
 \subsection{Matrices}
 The set of $m\times d$ real matrices with real entries is denoted $\CM_{m,d}(\mR)$, or simply $\CM_{m,d}$ ($\CM_{d,d}$ will also be denoted $\CM_d$).  The set of invertible $d\times d$ matrices will be denoted $\CG\CL_d(\mR)$.

 Given $m$ column vectors $x_1, \ldots, x_m\in \mR^d$, the notation $[x_1, \ldots, x_m]$ refers to the $d$ by $m$ matrix with $j^{\mathrm{th}}$ column equal to $x_j$, so that, for example, $\Id[d] = [\mathfrak e_1, \ldots, \mathfrak e_d]$.

Entry $(i,j)$ in a matrix $A\in \CM_{m,d}(\mR)$ will either be denoted $A(i,j)$ or $\pe {A_j} i$. The rows of $A$ will be denoted $\pe A 1, \ldots,\pe A m$ and the columns $A_1, \ldots, A_m$.

 The operator norm of a matrix $A\in \CM_{m,d}$ is defined by
 \[
 |A|_{\mathrm{op}} = \max\{|Ax|: x\in \mR^d, |x|=1\}.
 \]
 
  The space of  $d \times d$ real symmetric matrices is denoted $\CS_d$, and its subsets containing positive semi-definite  (resp. positive definite) matrices is denoted $\CS^+_d$ (resp. $\CS^{++}_d$). If $m\leq d$, $\CO_{m,d}$ denotes the set of $m\times d$ matrices $A$ such that $AA^T = \Id[m]$, and one writes $\CO_d$ for $\CO_{d,d}$, the space of $d$-dimensional orthogonal matrices. Finally, $\CS\CO_d$ is the subset $\CO_d$ containing orthogonal matrices with determinant 1, i.e., rotation matrices.

 \subsection{Multilinear maps}
 
 A $k$-linear map is a function $a: (x_1, \ldots, x_k) \mapsto a(x_1, \ldots, x_k)$ defined on $(\mR^d)^k$ with values in $\mR^q$ which is linear in each of its variables. The mapping is symmetric if its value is unchanged after any permutation of the variables. If $k=2$ and $q=1$, one also says that $a$ is a bilinear form. The norm of a $k$-linear map is defined as
 \[
 |a| = \max\{a(x_1, \ldots, x_k): |x_j|\leq 1,  j=1, \ldots, k\}
 \]
 so that 
 \[
 |a(x_1, \ldots, x_k)| \leq |a| \, \prod_{j=1}^k |x_j|
 \]
 for all $x_1, \ldots, x_k\in \mR^d$.
 
A symmetric bilinear (i.e., 2-linear) form $a$ is called positive semidefinite if $a(x,x) \geq 0$ for all $x\in \mR^d$, and positive definite if it is positive semi-definite and $a(x,x)=0$ if and only if $x=0$. Symmetric bilinear forms can always be expressed in the form $a(x,y) = x^TAy$ for some symmetric matrix $A$, and $a$ is positive (semi-)definite if and only $A$ is also. Analogous statements hold for negative (semi-)definite forms and matrices. We will use the notation $A \succ 0$ (resp. $\succeq 0$) to indicate that $A$ is positive definite (resp. positive semidefinite). Note that, if $a(x,y) = x^TAy$ for $A\in \CS_d$, then $|a| = |A|_{\mathrm{op}}$.

 \section{Topology}
 
\subsection{Open and closed sets in $\mathbb R^d$}
The open balls in $\mR^d$ will be denoted
\[
B(x, r) = \{y\in \mR^d: |y-x| < r\},
\]
with $x\in \mR^d$ and $r>0$. The closed balls are denoted $\bar B(x,r)$ and contain all $y$'s such that $|y-x|\leq r$. A set $U \subset \mR^d$ is open if and only if for any $x\in U$, there exists $r>0$ such that $B(x,r) \subset U$. A set $\Gamma \subset \mR^d$ is closed if its complement, denoted
\[
\Gamma^c = \{x\in \mR^d: x\not\in\Gamma\}
\]
is open. The topological interior of a set $A\sub \mR^d$ is the largest open set included in $A$. It will be denoted either by $\mathring A$ or $\mathrm{int}(A)$. A point  $x$ belongs to $\mathring A$ if and only if   $B(x,r)\subset A$ for some $r>0$.

The closure of $A$ is the smallest closed set that contains $A$ and will be denoted either $\bar A$ or $\mathrm{cl}(A)$.  A point $x$ belongs to $\bar A$ if and only if $B(x, r) \cap A \neq \emptyset$ for all $r>0$. Alternatively, $x$ belongs to $\bar A$ if and only if there exists a sequence $(x_k)$ that converges to $x$ with $x_k\in A$ for all $k$.

\subsection{Compact sets}
A compact set in $\mR^d$ is a set $\Gamma$ such that  any sequence of points in $\Gamma$ contains a subsequence that converges to some point in $\Gamma$. An alternate definition is that, whenever $\Gamma$ is covered by a collection of open sets, there exists a finite subcollection that still covers $\Gamma$.

One can show that compact subsets of $\mR^d$ are exactly its bounded and closed subsets. 

\subsection{Metric spaces}
 A metric space is a space $\CB$ equipped with a distance, i.e., a function $\rho:\CB\times \CB \to [0, +\infty)$ that satisfies the following three properties.
\begin{subequations}
\begin{equation}
\label{eq:distance.a}
\forall x,y\in \CB: \rho(x,y)=0 \Leftrightarrow x=y,
\end{equation}
\begin{equation}
\label{eq:distance.b}
\forall x,y\in\CB: \rho(x,y) = \rho(y,x),
\end{equation}
\begin{equation}
\label{eq:distance.c}
\forall x,y,z\in\CB: \rho(x,z) \leq \rho(x,y) + \rho(y,z).
\end{equation}
\end{subequations} 
\Cref{eq:distance.c} is called the triangle inequality.
The norm of the difference between two points: $\rho(x,y) = |x-y|$, is a distance on $\mR^d$. The definition of open and closed subsets in metric spaces is the same as above, with $\rho(x,y)$ replacing $|x-y|$, and one says that $(x_n)$ converges to $x$ if and only if $\rho(x_n, x)\to 0$. 

Compact subsets are also defined in the same way, but are not necessarily characterized as bounded and closed.

\section{Calculus}
\subsection{Differentials}
 If $x,y\in \mR^d$, we will denote by $[x,y]$ the closed segment delimited by $x$ and $y$, i.e., the set of all points $(1-t) x + t y$ for $0\leq t \leq 1$. One denotes by $[x,y)$, $(x,y]$ and $(x,y)$ the semi-open or open segments, with appropriate strict inequality for $t$. (Similarly to the notation for open intervals, whether $(x,y)$ denotes an open segment or a pair of points will always be clear from the context.)

 The derivative of  a differentiable function $f: t \mapsto f(t)$ from an interval $I\subset \mR$ to $\mR$ will be denoted by $\prt f$, or $\prt_t f$ if the variable $t$ is well identified. Its value at $t_0\in I$ is denoted either as $\prt f(t_0)$ or $\prt f{|_{t=t_0}}$.  Higher derivatives are denoted as $\prt^k f$, $k\geq 0$, with the usual convention $\prt^0f = f$. Note that notation such as $f', f'', f^{(3)}$ will {\em never} refer to derivatives.

In the following, $U$ is an open subset of $\mR^d$.
If $f$ is a function from $U$ to $\mR^m$, we let $\pe{f}{i}$ denote the $i^{\mathrm{th}}$ component of $f$, so that
\[
f(x) = \begin{pmatrix}
\pe{f}{1}(x)\\
\vdots\\
\pe{f}{m}(x)
\end{pmatrix}
\]
for  $x\in U$. If $d=1$, and $f$ is differentiable, the derivative of $f$ at $x$ is the column vector of the derivatives of its components,   
\[
\prt f(x) = \begin{pmatrix}
\prt \pe{f}{1}(x)\\
\vdots\\
\prt \pe{f}{m}(x)
\end{pmatrix}
\]

For $d\geq 1$ and $j\in \{1,\ldots, d\}$, the $j^{\mathrm{th}}$ partial derivative of $f$ at $x$ 
is
\[
\prt_j f(x) = \prt(t\mapsto f(x+t \mathfrak e_j))|_{t=0} \in \mR^m,
\]
where $\mathfrak e_1, \ldots, \mathfrak e_d$ form  the canonical basis of $\mR^d$.
If the notation for the variables on which $f$ depends is well understood from the context, we will alternatively use $\prt_{x_j} f$. (For example, if $f: (\alpha, \beta) \mapsto f(\alpha, \beta)$, we will prefer $\prt_\alpha f$ to $\prt_1f$.) The differential of $f$ at $x$ is the linear mapping from $\mR^d$ to $\mR^m$ represented by the matrix
\[
d\!f(x) = [\prt_1f(x), \ldots, \prt_d f(x)].
\]
 It is defined so that, for all $h\in \mR^d$
\[
d\!f(x) h = \prt(t\mapsto f(x+th))|_{t=0}
\]
where the right-hand side is the directional derivative of $f$ at $x$ in the direction $h$. Note that, if $f:\mR^d \to \mR$ (i.e., $m=1$), $df(x)$ is a row vector. If $f$ is differentiable on $U$ and $df(x)$ is continuous as a function of $x$, one says that $f$ is continuously differentiable, or $C^1$. 

Differentials obey the product rule and the chain rule. If $f,g: U \to \mR$, then
\[
d(fg)(x) = f(x) dg(x) + g(x) df(x).
\] 
If $f : U \to \mR^m$, $g: \tilde U \subset \mR^k \to U$, then
\[
d(f\circ g)(x) = df(g(x)) dg(x).
\]

If $d=m$ (so that $df(x)$ is a square matrix), we let $\nabla \cdot f (x) = \trace(df(x)) $, the divergence of $f$.

The terms ``derivative'' and ``differential'' are mostly interchangeable, although one often uses ``derivative'' for differentiation with respect to a scalar variable.

The Euclidean gradient of a differentiable function  $f: U \to \mR$ is $\nabla f(x) = df(x)^T$. More generally, one defines the gradient of $f$ with respect to a tensor field $x\mapsto A(x)$ taking values in $\CS^{++}_d$,  as the vector $\nabla_Af(x)$ that satisfies the relation
\[
df(x) h = \nabla_Af(x)^T A(x) h
\]
for all $h\in \mR^d$, so that 
\begin{equation}
\label{eq:gradient.A}
\nabla_Af(x) = A(x)^{-1} df(x)^T.
\end{equation}
In particular, the Euclidean gradient is associated with $A(x) = \Id[d]$ for all $x$.  With some abuse of notation, we will denote $\nabla_A f = A^{-1} \nabla f$ when $A$ is a fixed matrix, therefore identified with the constant tensor field $x\mapsto A$.

\subsection{Important examples} 
\label{sec:diff.examples}
We here compute, as an illustration and because they will be useful later, the differential of the determinant and the inversion in matrix spaces.

Recall that, if $A = [a_1, \ldots, a_d]\in \CM_d$ is a $d$ by $d$ matrix,, with $a_1, \ldots, a_d\in \mR^d$, $\det(A)$ is a $d$-linear form 
$\delta(a_1, \ldots, a_d)$
which vanishes when two columns coincide and such that $\delta(\mathfrak e_1, \ldots, \mathfrak e_d) = 1$. In particular $\delta$ changes signs when two of its columns are inverted. It follows from this that
\begin{multline*}
\partial_{a_{ij}} \det(A) = \delta(a_1, \ldots, a_{i-1}, \mathfrak e_j, a_{j+1}, \ldots, a_d) \\
= (-1)^{i-1} \delta(\mathfrak e_j, a_1, \ldots, a_{i-1}, \ldots, a_d) = (-1)^{i+j} \det \pe A {ij},
\end{multline*}
where $\pe A {ij}$ is the matrix $A$ with row $i$ and column $j$ removed. We therefore find that the differential of $A \mapsto \det(A)$ is the mapping
\begin{equation}
\label{eq:det.der}
H \mapsto \trace(\mathrm{cof}(A)^T H)
\end{equation}
where $\mathrm{cof}(A)$ is the matrix composed of co-factors $ (-1)^{i+j} \det \pe A {ij}$. As a consequence, if $A$ is invertible, then the differential of $\log |\det(A)|$ is the mapping
\begin{equation}
\label{eq:det.log.der}
H \mapsto \trace(\det(A)^{-1}\mathrm{cof}(A)^T H) = \trace(A^{-1} H)
\end{equation}

Consider now the function $\mathfrak I(A) = A\mapsto A^{-1}$ defined on $\CG\CL_d(\mR)$, which is an open subset of $\CM_d(\mR)$. Using $A\mathfrak I(A) = \Id[d]$ and the product rule, we get
\[
A (d\mathfrak I(A) H) + H \mathfrak I(A) = 0
\]
or
\begin{equation}
\label{eq:inv.der}
d\mathfrak I(A) H = - A^{-1} H A^{-1}.
\end{equation}

\subsection{Higher order derivatives}
Higher-order partial derivatives $\prt_{i_k} \cdots \prt_{i_1} f : U \to \mR^m$ are defined by iterating the definition of first-order derivatives, namely 
\[
\prt_{i_k} \cdots \prt_{i_1} f(x) = \prt_{i_k} (\prt_{i_{k-1}} \cdots \prt_{i_1} f)(x)
\]
If all order $k$ partial derivatives of $f$ exist and are continuous, one says that $f$ is $k$-times continuously differentiable, or $C^k$ and, when true, the order in which the derivatives are taken does not matter. In this case,  one typically groups derivatives with the same order using a power notation, writing, for example
\[
\prt_1 \prt_2 \prt_1 f = \prt_1^2 \prt_2 f
\]
for a $C^3$ function.

If $f$ is $C^k$, its $k^{\mathrm{th}}$ differential at $x$ is a symmetric $k$-multilinear map that can also be iteratively defined by (for $h_1, \ldots, h_k\in \mR^d$)
\[
d^k f(x)(h_1, \ldots, h_k) = d(d^{k-1}f(x)(h_1, \ldots, h_{k-1})) h_k  \in \mR^m.
\]
It is related to partial derivatives through the relation:
\[
d^k f(x)(h_1, \ldots, h_k) = \sum_{i_1, \ldots, i_k=1}^d  \pe{h_1}{i_1} \cdots \pe{h_k}{i_k}\,\prt_{i_k} \cdots \prt_{i_1} f(x).
\]

When $m=1$ and $k=2$, one denotes by $\nabla^2 f(x) = (\prt_i\prt_j f(x), i,j=1, \ldots, n)$ the symmetric  matrix formed by partial derivatives of order 2 of $f$ at $x$. It is called the {\em Hessian} of $f$ at $x$ and satisfies
\[
h_1^T \nabla^2 f(x) h_2 =  d^2 f(x)(h_1, h_2).
\]

The Laplacian of $f$ is the trace of $\nabla^2 f$ and denoted $\Delta f$.

\subsection{Taylor's theorem}
Taylor's theorem, in its integral form, generalizes the fundamental theorem of calculus to higher derivatives. It expresses the fact that, if $f$ is $C^k$ on $U$ and $x,y\in U$ are such that the closed segment $[x,y]$ is included in $U$, then, letting $h = y-x$:
\begin{multline}
f(x+h) = f(x) + df(x) h + \frac12 d^2 f(x)(h,h) + \cdots + \frac1{(k-1)!} d^{k-1}f(x)(h, \ldots, h) \\
+ \frac1{(k-1)!} \int_0^1 (1-t)^{k-1}d^kf(x+th)(h, \ldots, h)\, dt
\label{eq:taylor}
\end{multline}

The last term (remainder) can also be written as
\[
\frac1{k!} \frac{\int_0^1 (1-t)^{k-1}d^kf(x+th)(h, \ldots, h)\, dt}{\int_0^1 (1-t)^{k-1}\, dt}.
\]
If $f$ takes scalar values, then $d^kf(x+th)(h, \ldots, h)$ is real and the intermediate value theorem implies that there exists some $z$ in $[x,y]$ such that 
\begin{multline}
\label{eq:taylor.z}
f(x+h) = f(x) + df(x) h + \frac12 d^2 f(x)(h,h) + \cdots + \frac1{(k-1)!} d^{k-1}f(x)(h, \ldots, h) \\
+ \frac1{k!} d^kf(z)(h, \ldots, h).
\end{multline}

This is not true if $f$ takes vector values. However,  for any $M$ such that $|d^kf(z)| \leq M$ for $z \in [x,y]$  (such $M$'s always exist because $f$ is $C^k$), one has 
\[
\frac1{(k-1)!} \int_0^1 (1-t)^{k-1}d^kf(x+th)(h, \ldots, h)\, dt \leq \frac{M}{k!} |h|^k.
\]

\Cref{eq:taylor} can be written as
\begin{multline}
f(x+h) = f(x) + df(x) h + \frac12 d^2 f(x)(h,h) + \cdots + \frac1{k!} d^{k}f(x)(h, \ldots, h) \\
+ \frac1{(k-1)!} \int_0^1 (1-t)^{k-1}(d^kf(x+th)(h, \ldots, h)-d^k f(x)(h,\cdots, h))\, dt\,.
\label{eq:taylor.diff}
\end{multline}
Let 
\[
\epsilon_x(r) = \max\defset{|d^kf(x+h) - d^kf(x)|:|h| \leq r}.
\]
 Since $d^k f$ is continuous, $\epsilon_x(r)$ tends to 0 when $r\to 0$ and we have
 \[
 \int_0^1 (1-t)^{k-1}|d^kf(x+th)(h, \ldots, h)-d^k f(x)(h,\cdots, h)|\, dt \leq \frac{|h|^k}{k}\epsilon_x(|h|).
 \]
This shows that \cref{eq:taylor} implies that 
\begin{align}
\label{eq:taylor.epsilon}
f(x+h) &= f(x) + df(x) h + \frac12 d^2 f(x)(h,h) + \cdots + \frac1{k!} d^{k}f(x)(h, \ldots, h) + \frac{|h|^k}{k!}\epsilon_x(|h|)&\\
&= f(x) + df(x) h + \frac12 d^2 f(x)(h,h) + \cdots + \frac1{k!} d^{k}f(x)(h, \ldots, h)+ o(|h|^k)&
\label{eq:taylor.o}
\end{align}

\section{Probability theory}

\subsection{General assumptions and notation}

When discussing probabilistic concepts, we will make the convenient assumption that all random variables are defined on a fixed probability space $(\Om, \myP)$. This means that $\Omega$ is large enough to include enough randomness to generate all required variables (and implicitly enlarged when needed).

We assume that the reader is familiar with concepts related to discrete random variables (which take values in a discrete or countable space) and their probability mass function (p.m.f.) or continuous variables (with values in  $\mathbb R^d$ for some $d$) and their probability density functions (p.d.f.) when they exist. In particular,  $X: \Om\to \mR^d$ is a random variable with  p.d.f. $f$ if and only if the expectation of $\phi(X)$ is given by
\[
\myE(\phi(X)) = \int_{\mR^d} \phi(x) f(x) dx
\]
for all bounded and continuous functions $\phi: \mR^d \to [0, +\infty)$.
Not all random variables of interest can be categorized as discrete or continuous with a p.d.f., however, and the others are more conveniently handled using measure-theoretic notation as introduced below.

With a few exceptions, we will use capital letters for random variables and small letters for scalars and vectors that  represent realizations of these variables. One of these exceptions will be our notation for training data, defined as an independent and identically distributed (i.i.d.) sample of a given random variable. A realization of such a sample will always be denoted  $T = (x_1, \ldots, x_N)$, which is therefore a series of observations. We will use the notation $\mT = (X_1, \ldots, X_N)$ for the collection of i.i.d. random variables that generate the training set, so that $T = (X_1(\om), \ldots, X_N(\om)) = \mT(\om)$ for some $\om\in \Om$. Another exception will apply to variables denoted using Greek letters, for which we will use boldface fonts (such as $\boldsymbol\alpha, \boldsymbol\beta, \ldots$). 

For a random variable $X$, the notation $[X=x]$, or $[X\in A]$ refers to subsets of $\Omega$, for example,
\[
[X=x] = \defset{\om\in \Om: X(\om) = x}.
\]

\subsection{Conditional probabilities and expectation}
\label{sec:cond.prob.elem}
If $X:\Omega \to \CR_X$ and $Y:\Omega \to \CR_Y$ are discrete random variables, then
\[
\myP(Y=y \mid X=x) = \myP(Y=y, X=x)/\myP(X=x)
\]
if $\myP(X =x) >0$ and is undefined otherwise. Then, if $Y$ is scalar- or vector-valued and discrete, one defines the conditional expectation of $Y$ given $X$, denoted $\myE(Y\mid X)$, by
\[
\myE(Y\mid X)(\om) = \sum_{y\in \CR_Y} y \myP(Y =\eta \mid X=X(\om))
\]
for all $\om$ such that $\myP(X=X(\om))>0$. Note that $\myE(Y\mid X)$ is a random variable, defined over $\Om$. It however only depends on the values of $X$, in the sense that $\myE(Y\mid X)(\om) = \myE(Y\mid X)(\om')$ if $X(\om) = X(\om')$. We will use the notation 
\[
\myE(Y\mid X=x) = \sum_{y\in \CR_Y} y \myP(Y=y \mid X=x), 
\]
which is now a function defined on $\CR_X$, satisfying $\myE(Y\mid X)(\om) = \myE(Y \mid X=X(\om))$.

If $X$ and $Y$ are scalar- or vector-valued and their joint distribution have a p.d.f. $\phi_{X,Y}$, one defines similarly the conditional p.d.f. of $Y$ given $X$ by
\[
\phi_Y(y \mid X=x) = \frac{\phi_{X,Y}(y, x)}{\int_{\CR_Y} \phi_{X,Y}(y', x)\, dy'}
\]
provided that the denominator does not vanish. We will also use the notation $\phi_Y(y\mid X)(\omega) = \phi_Y(y \mid X=X(\omega))$ for $\omega \in \Omega$. One then defines 
\[
E(Y \mid X) (\omega) = \int_{\CR_Y} y \phi_Y(y \mid X=X(\omega)) dy.
\]

In both cases considered above, it is easily checked that the conditional expectation satisfies  the properties
\begin{enumerate}[label= (CE\arabic*)]
\item 
$\myE(Y \mid X)(\omega)$ only depends on $X(\omega)$
\item For all functions $f: \CR_X \to [0, +\infty)$ (continuous in the case of continuous random variables), one has $\myE(\myE(Y\mid X) f(X)) = \myE(Y f(X))$.
\end{enumerate}
The proof that our definition of $\myE(Y \mid X)$ for discrete random variables is the only one satisfying these properties is left to the reader. For continuous random variables, assume that a function $g: \CR_X \to \CR_Y$ satisfies
\[
\myE(g(X) f(X)) = \myE(Y f(X))
\]
for all continuous $f\geq 0$. Then, letting $\phi_X(x) = \int_{\CR_Y} \phi_{X,Y}(x,y) dy$, which is the marginal p.d.f. of $X$,
\[
\int_{\CR_X} g(x) f(x) \phi_X(x) dx = \int_{\CR_X \times \CR_Y} y f(x) \phi_{X,Y}(x,y) dx dy = \int_{\CR_X} \left(\int_{\CR_Y} y \phi_{X,Y}(x,y) dy\right) dx.
\]
If we assume that $\phi_{X,Y}$ is continuous, then this identity being true for all $f$ implies that  
\[
g(x) \phi_X(x) = \int_{\CR_Y} y \phi_{X,Y}(x,y) dy 
\]
so that $g$ is the conditional expectation. If $\phi_{X,Y}$ is not continuous, then the identity holds everywhere except on an exceptional ``negligible'' set (see the measure theoretic introduction below). Properties (CE1) and (CE2) provide the definition of the conditional expectation for general random variables.

Taking $g(x) = 1$ for all $x\in \CR_X$ in (CE2) yields the well-known identity
\[
\myE(\myE(Y\mid X)) = \myE(Y).
\]
Moreover, for any function $g$ defined on $\CR_X$ we have 
\[
\myE(Y g(X)\mid X) = g(X)  \myE(Y\mid X),
\]
 which can be checked by proving that the right-hand side satisfies conditions (i) and (ii).

\subsection{Measure theoretic probability}
 As much as possible---but not always---we will avoid relying on measure theory in our discussions, at the expense of sometimes making not fully rigorous or incomplete statements (that readers familiar with this theory will easily complete). However, there will be situations in which the flexibility of the measure-theoretic formalism is needed for the exposition. The following notions may  help the reader navigate through these situations (basic references in measure theory are \citet{rudin1966real,dudley2018real,billingsley2013convergence}).

 A measurable space is a pair $(S, \boldsymbol{\CS})$ where $S$ is a set and $\boldsymbol{\CS} \subset \twop{S}$ contains $S$, is stable by complementation (if $A\in \boldsymbol{\CS}$, then $A^c = S\setminus A\in \boldsymbol{\CS}$), by countable unions and intersections. Such an $\boldsymbol{\CS}$ is called a $\sigma$-algebra and elements of $\boldsymbol{\CS}$ form the measurable subsets of $S$ (relative to the $\sigma$-algebra).

A (positive) measure $\mu$ on $(S, \boldsymbol{\CS})$ in a mapping from $\boldsymbol{\CS} \to [0, +\infty)$ that associates to $A\in \boldsymbol{\CS}$ its measure $\mu(A)$, such that the measure of a countable union of disjoint sets is the countable sum of their measures. A function $f:\Omega \to \mR^d$ is called measurable if the inverse images by $f$ of open subsets of $\mR^d$ are measurable. More generally, if  $(S, \boldsymbol{\CS})$ and $(S', \boldsymbol{\CS}')$ are measurable spaces, $f: S \to S'$ is measurable if $f^{-1}(A') \in \boldsymbol{\CS}$ for all $A' \in \boldsymbol{\CS}'$.

\begin{remark}
\label{rem:polish}
In these notes, measurable spaces $S$ will always (and without mention) be a complete metric (or ``metrizable'') space with a dense countable subset (also called a Polish space), and $\boldsymbol{\CS}$ the smallest $\sigma$-algebra containing all open subsets of $S$, which is called the Borel $\sigma$-algebra. 

Also without mention, all measures will be assumed to be ``$\sigma$-finite'', which means that there exists a sequence of measurable sets with finite measure whose union is the whole set $S$. 
\end{remark}

If $\mu$ is a measure, a measurable set $A$ is $\mu$-negligible (or negligible for $\mu$) if there exists $B\in \boldsymbol{\CS}$ such that $A\subset B$ and  $\mu(B) = 0$ and events are said to happen almost everywhere if their complements are negligible. A countable union of negligible sets is negligible, but this is not true for non countable unions, which may not even be measurable. It is convenient, and always possible, to extend a $\sigma$-algebra so that it contains all $\mu$-negligible sets.

The integral of a function $f: S \to \mR^d$ with respect to a measure $\mu$ is denoted $\int_S f\,d\mu$ or 
$\int_S f(x) \mu(dx)$. This integral is defined, using a limit argument, as a  function which is linear in $f$ and such that, for all $A\in \boldsymbol\CS$,
\[
\int_A \mu(dx) = \int_S \bfone_A(x) \mu(dx) = \mu(A).
\]

More precisely, this uniquely defines the integral of linear combinations of indicator functions with finite measures (called ``simple functions''), and one then defines $\int_S f\,d\mu$ for $f: S \to [0, +\infty)$ as the supremum of the integrals among all simple functions that are no larger than $f$. After showing that the result is well defined and linear in $f$, one defines the integral of $f: S \to \mR$ as the difference between those of $\max(f,0)$ and $\max(-f, 0)$, which is well define as soon as $\int_S |f|\,d\mu <\infty$, in which case one says that $f$ is $\mu$-integrable. 

The Lebesgue measure, $\CL_d$, on $\mR^d$ provides an important example. For this measure, $\boldsymbol{\CS}$ is the $\sigma$-algebra generated by open subsets of $\mR^d$ (the smallest one that contains all open subsets), and  $\int_{\mR^d} f(x) \CL_d(dx)$ extends the Riemann integral, justifying the alternative notation $\int_{\mR^d} f(x) dx$ that we will preferably use. Another important example, especially when $S$ is finite or countable, is the counting measure, denoted $\mathit{card}$, that returns the number of elements of a set, so that $\mathit{card}(A) = |A|$. If $\CS$ is finite or countable, one generally takes $\boldsymbol{\CS} = \twop{S}$ (every subset of $S$ is measurable) and the integral is simply the sum:
\[
\int_S f(x)\ \mathit{card}(dx) = \sum_{x\in \CF} f(x).
\]

\subsection{Product of measures}
Let $\mu_1$ is a measure on  $(S_1 , \boldsymbol{\CS}_1)$ and $\mu_2$ a measure on  $(S_2 , \boldsymbol{\CS}_2)$, in which we assume the situation described in \cref{rem:polish}. The ``tensor product'' of $\mu_1$ and $\mu_2$ is denoted $\mu_1 \otimes \mu_2$. It is a measure on $S_1\times S_2$ defined by $\mu_1\otimes\mu_2(A_1 \times A_2) = \mu_1(A_1) \mu_2(A_2)$ for $A_1 \in \boldsymbol{\CS}_1$ and $A_2 \in \boldsymbol{\CS}_2$ (the $\sigma$-algebra on $S_1 \times S_2$ is the smallest one that contains all sets $A_1 \times A_2$, $A_1 \in \boldsymbol{\CS}_1$, $A_2 \in \boldsymbol{\CS}_2$). 

The integral, with respect to the product measure, of a function $f: S_1 \times S_2 \to \mR^d$ is denoted
\[
\int_{S_1\times S_2} f(x_1, x_2) \mu_1(dx_1)  \mu_2(dx_2) 
\]
(rather than
\[
\int_{S_1\times S_2} f(x_1, x_2) \mu_1\otimes \mu_2(dx_1, dx_2)).
\]
Fubini's theorem justifies integration by parts: if $f$ is integrable with respect to the product measure, then (integrating first in the second variable)
\[
F(x_1) = \int_{S_1} f(x_1, x_2) mu_2 dx_2
\]
is well defined and 
\[
\int_{S_1\times S_2} f(x_1, x_2) d(\mu_1\otimes\mu_2) = \int_{S_1} F(x_1) d\mu_1.
\]
(And one has a symmetric statement by integrating first in the first variable.)

The tensor product between more that two measures is defined similarly, with notation
\[
\mu_1 \otimes \cdots \otimes \mu_n = \bigotimes_{k=1}^n \mu_k.
\] 

\subsection{Relative absolute continuity and densities}
\label{sec:abs.cont}
If $\mu$ and $\nu$ are measures on $(S , \boldsymbol{\CS})$, one says that $\nu$ is absolutely continuous with respect to $\mu$ and write $\nu \ll \mu$ if,
\begin{equation}
\label{eq:abs.cont}
\forall A\in \boldsymbol{\CS}:
\mu(A) = 0 \Rightarrow \nu(A) = 0.
\end{equation}
Assume that $\mu$ is $\sigma$-finite (\cref{rem:polish}).
 The Radon-Nikodym theorem states that $\nu\ll \mu$ with $\nu(S) < \infty$ if and only if $\nu$ has a {\em density} with respect to $\mu$, i.e., there exists a $\mu$-integrable function $\phi: S\to [0, +\infty)$ such that 
\[
\int_S f(x) \nu(dx) = \int_S f(x) \phi(x) \mu(dx)
\]
for all measurable $f:S \to[0, +\infty)$.

\subsection{Measure-theoretic probability}
\label{sec:meas.prob}
When using measure-theoretic probability, we will therefore assume that the  pair $(\Omega, \myP)$ is completed to a triple $(\Omega, \boldsymbol{\mathcal A}, \myP)$ where $\boldsymbol{\mathcal A}$ is a $\sigma$-algebra and $\myP$ a probability measure, that is a (positive) measure on  $(\Omega, \boldsymbol{\mathcal A})$ such that $\myP(\Omega) = 1$. This triple is called a {\em probability space}. For probability spaces, measurable sets are also called ``events'' and events that happen with probability one are said to happen ``almost surely.'' 

A random variable $X$ must then also take values in a measurable space, say $(\CR_X, \boldsymbol{\CS}_X)$, and must be such that, for all $C\in \boldsymbol{\CS}$, the set $[X\in C]$ belongs to $\boldsymbol{\CS}_X$. This justifies the computation of $\myP(X\in C)$, which is also denoted $P_X(C)$. 

A random variable $X$ taking values in $\mR^d$ has a p.d.f. if and only if $P_X \ll \CL_d$ and the  p.d.f. is the density provided by the Radon-Nikodym theorem. For a discrete random variable (i.e., taking values in a finite or countable set), the p.m.f. of $X$ is also the density of $P_X$ with respect to the counting measure $\mathit{card}$ (every discrete measure is absolutely continuous with respect to $\mathit{card}$).

If $X$ is a random variable with values in $\mR^d$, the integral of $X$ with respect to $\myP$ is the expectation of $X$,  denoted $\myE(X)$. More generally, if $(S, \boldsymbol{\CS}, P)$ is a probability space, we will use the notation
\[
E_P(f) = \int_S f(x) P(dx).
\]
If $P = P_X$ for some random variable $X:\Omega\to S$, we will use $E_X$ rather than $E_{P_X}$.

\subsection{Conditional expectations (general case)}
We use (CE1) and (CE2) as a definition of conditional expectation in the general case.  We assume that $(\CR_X, \boldsymbol{\CS}_X)$ and $(\CR_Y, \boldsymbol{\CS}_Y)$ are measurable spaces.

\begin{definition}
\label{def:cond.exp}
Assume that $\CR_Y = \mR^d$.
Let $X: \Om\to \CR_X$ and $Y:\Om \to \CR_Y$ be two random variables with  $\myE(|Y|) <\infty$. The conditional expectation of $Y$ given $X$ is a random variable $Z: \Om \to \CR_Y$ such that:
\begin{enumerate}[label = (\roman*),wide=0pt]
\item There exists a measurable function $h: \CR_X \to \mR^d$ such that $Z = h 
\circ X$ almost surely.
\item For any measurable function $g: \CR_X \to [0, +\infty)$, one has
\[
\myE(Y g\circ X) = \myE(Z g\circ X).
\]
\end{enumerate}
The variable $Z$ is then denoted $\myE(Y\mid X)$ and the function $h$ in (i) is denoted $\myE(Y \mid X = \cdot)$.
\end{definition}

Importantly, random variables $Z$ satisfying conditions (i) and (ii) always exist and are almost surely unique,  in the sense that if another function $Z'$ satisfies these conditions, then $Z = Z'$ with probability one. One obtains an equivalent definition if one restricts functions $g$ in (ii) to indicators of measurable sets, yielding the condition that, if $A\subset \CR_X$ is measurable,
\[
\myE(Y \bfone_{X\in B} ) = \myE(Z \bfone_{X\in B}).
\]

With this general definition, we still have 
\[
\myE(\myE(Y\mid X)) = \myE(Y)
\]
and, for any function $g$ defined on $\CR_X$, $\myE(Y g\circ X\mid X) = (g\circ X)  \myE(Y\mid X)$.

Conditional expectations share many of the properties of simple expectations. The are linear with respect to the $Y$ variable. Moreover, if $Y \leq Y'$, both taking scalar values, then $\myE(Y\mid X) \leq \myE(Y'\mid X)$ almost surely. Jensen's inequality also holds: if $\gamma: \mR^d\to \mR$ is convex and $\gamma\circ Y$ is integrable, then
\[
\gamma\circ \myE(Y\mid X) \leq \myE(\gamma \circ Y \mid X).
\]
We will discuss convex functions in \cref{chap:optim}, but two important examples for this section are $\gamma(y) = |y|$ and $\gamma(y) = |y|^2$. The first one implies that $|\myE(Y\mid X)| \leq \myE(|Y|\mid X)$ and, taking expectations on both sides:  $\myE(|\myE(Y\mid X|) \leq \myE(|Y|)$, the upper bound being finite by assumption. For the square norm, we find that, if $Y$ is square integrable, then so is $\myE(Y\mid X)$ and 
\[
\myE(|\myE(Y\mid X)|^2) \leq \myE(|Y|^2).
\]

If $Y$ is square integrable, then this inequality shows that $\myE(Y \mid X)$ minimizes $\myE[|Y - Z|^2]$ among all square integrable functions $Z:\Om\to \CR_Y$ that satisfy (i). In other terms, the conditional expectation is the optimal least-square approximation of $Y$ by a function of $X$. To see this, just write
\begin{align*}
\myE(|Y - Z|^2\mid X) &= \myE(|Y|^2\mid X)  - 2\myE(Y^TZ\mid X)
+ |Z|^2\\
&= \myE(|Y|^2\mid X) - 2\myE(Y\mid X)^TZ
+|Z|^2\\
&= \myE(|Y|^2\mid X] - |\myE(Y\mid X)|^2 +|\myE(Y\mid X) - Z|^2\\
&= \myE(|Y - \myE(Y\mid X)|^2\mid X) +|\myE(Y\mid X) - Z|^2\\
&\geq \myE(|Y - \myE(Y\mid X)|^2\mid X)
\end{align*}
and taking expectations on both sides yields the desired result.

\subsection{Conditional probabilities (general case)}
\label{sec:cond.prob.gen}
If $A$ is a measurable subset of $\CR_Y$, then $\bfone_A$ is a random variable and its conditional expectation $\myE(\bfone_A\mid X)$ (resp. $\myE(\bfone_A\mid X = x)$) is denoted  $\myP(Y \in A\mid X)$ (resp. $\myP(Y\in A\mid X = x)$), or $P_Y(A\mid X)$ (resp. $P_Y(A\mid X = x)$). Note that, for each $A$, these conditional probabilities is  defined up to  modifications on sets of probability zero, and it is not obvious that they can be defined for all $A$ together (up to a modification on a common set of probability zero), since there is generally a non-countable number of sets $A$.  This can be done, however, with some mild assumptions on the set $\CR_Y$ and its $\sigma$-algebra (always satisfied in our discussions, see remark \cref{rem:polish}), ensuring that, for all $\omega \in \Omega$,  $A\mapsto P_Y(A \mid X)(\omega)$ is a probability distribution  on $\CR_Y$ such that, for any measurable function $h: \CR_Y \to \mR$ such that $h\circ Y$ is integrable,  
\[
\myE(h(Y)\mid X) = \int_{\CR_Y} h(y) P_Y(dy\mid X).
\] 

Assume now that the the sets $\CR_X$ and $\CR_Y$ are equipped with measures, say $\mu_X$ and $\mu_Y$ such that the joint distribution of $(X,  Y)$ is absolutely continuous with respect to $\mu_X\otimes \mu_Y$, so that there exists a function $\phi:\CR_X\times \CR_Y\to \mR$ (the p.d.f. of $(X, Y)$ with respect to $\mu_X\otimes \mu_Y$) such that 
\[
\myP(X\in A, Y\in B) = \int_{A\times B} \phi(x, y) \mu_X(dx)  \mu_Y(dy).
\]
Then $P_Y(\cdot\mid X)$ is absolutely continuous with respect to $\mu_Y$, with density given by the conditional p.d.f. of $Y$ given $X$, namely,
\begin{equation}
\label{eq:cond.pdf}
\phi(y\mid X): \om \mapsto \frac{\phi(X(\om), y)}{\int_{\CR_Y} \phi(X(\om), y')\, \mu_Y(dy')} = \phi(y\mid X= X(\omega)).
\end{equation}
Note that 
\[
\myP\left\{\omega: \int_{\CR_Y} \phi(Z(\om), y')\, \mu_Y(dy') = 0\right \} = 0
\]
so that the conditional density can be defined arbitrarily when the numerator vanishes\footnote{Letting $\phi_X(x) = \int_{\CR_Y}\phi(x, y')\, \mu_Y(dy')$, which is the marginal p.d.f. of $X$ with respect to $\mu_X$, we have
\[
\myP(\phi_X(X)= 0) = \int_{\CR_X} \bfone_{\phi_X(x)= 0} \phi_X(x) \mu_X(dx) = 0.
\] 
}.

We retrieve here as special cases the definition of conditional probabilities and densities given in \cref{sec:cond.prob.elem}. As an additional example, take $\CR_X = \mR^d$, $\mu_X$ being Lebesgue's measure and assume that $Y$ is discrete (so that $\mu_Y = \mathit{card}$), then
\[
\phi(y \mid X = X(\omega) ) = \dfrac{\phi(X(\om), y)}{\sum_{y' \in \CR_Y} \phi(X(\om), y')}.
\]

\chapter{A Few Results in Matrix Analysis}
\label{chap:linalg}

This chapter collects a few results in linear algebra that will be useful in the rest of this book.

\section{Notation and basic facts}
We  denote by $\CM_{n,d}(\mR)$ the space of all $n\times d$ matrices with real coefficients\footnote{Unless mentioned otherwise, all matrices are assumed to be real.}. For a matrix $A\in \CM_{n,d}(\mR)$ and integer $k\leq n$ and $l\leq d$, we let $A_{\lceil kl\rceil}\in \CM_{k,l}(\mR)$ denote the matrix $A$ restricted to its first $k$ rows and first $l$ columns. The $i,j$ entry of $A$ will be denoted $A(i,j)$ or $\pe A {ij}$.

We assume that the reader is familiar with elementary matrix analysis, including, in particular the fact that symmetric matrices are diagonalizable in an orthonormal basis, i.e., if $A \in \CM_{d,d}(\mR)$ is a symmetric matrix (whose space is denoted $\CS_d$), there exists an orthogonal  matrix $U\in \CO_d$ (i.e., satisfying $U^TU = UU^T = \Id[d]$) and a diagonal matrix $D\in \CM_{d,d}(\mR)$ such that
\[
A = UDU^T.
\]
The identity $AU = UD$ then implies that the columns of $U$ form an orthonormal basis of eigenvectors of $A$. 

If $A\in \CS^+_d$ is positive semi-definite (i.e., $u^T Au \geq 0$ for all $u\in \mR^d$), the entries of  $D$ in the decomposition $A = UDU^T$ are non-negative, and one can define the matrix square root of $A$ as $S = UD^{\odot 1/2} U^T$ where  $D^{\odot 1/2}$ is the diagonal matrix formed taking the square roots of all  coefficients of $D$. We will use the notation $S = A^{1/2}$. Note that $D^{1/2} = D^{\odot 1/2}$ if $D$ is diagonal and positive semi-definite. 

If  $A\in \CS^{++}_d$ is positive definite (i.e., $A$ is positive semi-definite and $u^TAu=0$ implies $u=0$) and $B$ is positive semi-definite, both being $d\times d$ matrices, the generalized eigenvalue problem associated with $A$ and $B$ consists in finding a diagonal matrix $D$ and a matrix $U$ such that $BU = AUD$ and $U^TAU = \Id[d]$. Letting $\tilde U = A^{1/2} U$, the problem is equivalent to solving $A^{-1/2} B A^{-1/2} \tilde U = \tilde U D$ with $\tilde U^T \tilde U = \Id[d]$, i.e., finding the eigenvalue decomposition of the symmetric positive-definite matrix $A^{-1/2} B A^{-1/2}$.

If $A\in \CM_{n,d}(\mR)$, it can be decomposed as 
\[
A = UDV^T
\]
where $U \in \CO_{n}(\mR)$ and  $V\in \CO_{d}(\mR)$) are orthogonal matrices and $D\in \CM_{n,d}(\mR)$ is diagonal (i.e., such that $D(i,j) = 0$ whenever $i\neq j$) with non-negative diagonal coefficients. These coefficients are called the singular values of $A$, and the procedure is called a {\em singular-value decomposition} (SVD) of $A$. An equivalent formulation is that there exist orthonormal bases $u_1, \ldots, u_n$ of $\mR^n$ and $v_1, \ldots, v_d$ of $\mR^d$ (forming the columns of $U$ and $V$) such that
\[
Av_i = \la_i u_i
\]
for $i\leq \min(n,d)$, where $\la_1, \ldots, \la_{\min(n,d)}$ are the singular values. Of course, if $A$ is square and symmetric positive semi-definite, an eigenvalue decomposition of $A$  is also a singular value decomposition (and the singular values coincide with the eigenvalues). More generally, if  $A = UDV^T$, then $AA^T = UDD^TU^T$ and $A^TA = VD^TDV^T$ are eigenvalue decompositions of $AA^T$ and $A^TA$. Singular values are uniquely defined, up to reordering. However, the matrices $U$ and $V$ are not unique up to column reordering in general.
As a consequence, if $A$ is a symmetric matrix and $A = UDV^T$ is a singular value decomposition of $A$, then the columns of both $U$ and $V$ provide eigenvectors of $A^2$, hence of $A$, and $A = UDU^T= VDV^T$.

 If $m = \min(n, d)$, then, forming the matrices $\tilde U = U_{\lceil n,m\rceil}$ (resp. $\tilde V = V_{\lceil d, m\rceil}$) by removing from $U$ (resp. $V$) its last $n-m$ (resp. $d-m$) columns , and $\tilde D=D_{\lceil m,m\rceil}$  by removing from $D$ its $n-m$ rows and $d-m$ columns, one has
\[
A = \tilde U \tilde D \tilde V^T
\]
with $\tilde U$, $\tilde D$ and $\tilde V$ having respectively size $n\times m$, $m\times m$ and $m\times d$, $\tilde U^T \tilde U = \tilde V^T \tilde V = \Id[m]$ and $\tilde D$ diagonal with non-negative coefficients. This representation provides a {\em reduced SVD} of $A$ and one can create a full SVD from a reduced one by completing the missing rows of $\tilde U$ and $\tilde V$ to form orthogonal matrices, and by adding the required number of zeros to $\tilde D$.

\section{The trace inequality}

We now descibe Von Neumann's trace theorem. Its justification follows the proof given in \citet{mirsky1975trace}.
\begin{theorem}[Von Neumann]
\label{th:trace.ineq}
Let $A, B\in \CM_{n,d}(\mR)$ have  singular values $(\la_1, \ldots, \la_m)$ and $(\mu_1, \ldots, \mu_m)$, respectively,  where $m = \min(n,d)$. Assume that these eigenvalues are listed in decreasing order so that
$\la_1\geq \cdots\geq \la_m$ and $\mu_1\geq \cdots\geq \mu_m$. Then,
\begin{equation}
\label{eq:trace.ineq}
\trace(A^TB) \leq \sum_{i=1}^m \la_i\mu_i\,.
\end{equation}

Moreover, if $\trace(A^TB) = \sum_{i=1}^m \la_i\mu_i$, then there exist $n\times n$ and $d\times d$ orthogonal matrices $U$ and $V$ such that $U^TAV$ and $U^TBV$ are both diagonal, i.e., one can find SVDs of $A$ and $B$ in the same bases of $\mR^n$ and $\mR^d$.
\end{theorem}

\begin{proof}
We can assume without loss of generality that $d\leq n$ because, if the result holds for $A$ and $B$, it also holds for $A^T$ and $B^T$. Let $A = U_1 \Lambda V_1^T$ and $B = U_2 M V_2^T$ be the singular values decompositions of $A$ and $B$ (both $\La$ and $M$ are $n\times d$ matrices). Then 
\[
\trace(A^TB) = \trace(V_1 \Lambda^T U_1^TU_2 M V_2) = \trace(\Lambda^T UMV^T)
\]
with $U = U_1^TU_2$ and $V = V_1^T V_2$. Let $u(i,j), 1\leq i,j\leq n$ and $v(i,j), 1\leq i,j\leq d$ be the coefficients of the orthogonal matrices $U$ and $V$. Then
\begin{equation}
\label{eq:trace.ineq.proof.1}
\trace(\Lambda^T UMV^T) = \sum_{i,j=1}^d u(i,j) v(i,j) \la_i\mu_j \leq \frac12 \sum_{i,j=1}^d \la_i \mu_j u(i,j)^2 + \frac12 \sum_{i,j=1}^d \la_i \mu_j v(i,j)^2
 \end{equation}
 Let us consider the first sum in the upper-bound. Let $\xi_d = \la_d$ (resp. $\eta_d = \mu_d$) and $\xi_i = \la_i - \la_{i+1}$ (resp. $\eta_i = \mu_i - \mu_{i+1}$) for $i=1, \ldots, d-1$. Since  singular values are non-increasing, we have $\xi_i, \eta_i\geq 0$  and 
 \[
 \la_i = \sum_{j=i}^d \xi_j, \quad  \mu_i = \sum_{j=i}^d \eta_j
 \]
 for $i=1, \ldots, d$. We have
 \begin{align}
 \nonumber
 \sum_{i,j=1}^d \la_i \mu_j u(i,j)^2 &=  \sum_{i,j=1}^d \sum_{i'=i}^d\xi_{i'}\sum_{j'=j}^d \eta_{j'}  u(i,j)^2 = \sum_{i',j'=1}^d \xi_{i'}\eta_{j'} \sum_{i=1}^{i'}\sum_{j=1}^{j'} u(i,j)^2\\
 &\leq \sum_{i',j'=1}^d \xi_{i'}\eta_{j'} \min(i', j')
 \label{eq:trace.ineq.proof.2}
 \end{align}
where we used the fact that $U$ is orthogonal, which implies that $\sum_{j=1}^{j'} u(i,j)^2$ and $\sum_{i=1}^{i'} u(i,j)^2$ are both less than 1. Notice also that, when $u(i,j) = \de_{ij}$ (i.e., $u(i,j)=1$ if $i=j$ and zero otherwise), then
\[
\sum_{i=1}^{i'}\sum_{j=1}^{j'} u(i,j)^2 = \min(i', j'),
\]
so that the last inequality is an identity, and the chain of equalities leading to \cref{eq:trace.ineq.proof.2} implies
\[
\sum_{i',j'=1}^d \xi_{i'}\eta_{j'} \min(i', j') = \sum_{i=1}^d \la_i\mu_j\,.
\]
We therefore obtain (for any $U$), the fact that 
\[
 \sum_{i,j=1}^d \la_i \mu_j u(i,j)^2 \leq \sum_{i=1}^d \la_i\mu_j.
 \]
The  same identity obviously holds with $v$ in place of $u$, and combining the two yields \cref{eq:trace.ineq}.
\bigskip

We now consider conditions for equality. Clearly, if one can find SVD decompositions of $A$ and $B$ with $U_1=U_2$ and $V_1=V_2$, then $U = \Id[n]$, $V=\Id[d]$ and \cref{eq:trace.ineq} is an identity. We want to prove the converse statement.

For \cref{eq:trace.ineq} to be an equality, we first need \cref{eq:trace.ineq.proof.1} to be an identity, which requires that $u(i,j) = v(i,j)$ as soon as $\la_i\mu_j > 0$. We also need an equality in \cref{eq:trace.ineq.proof.2}, which requires 
\[
\sum_{i=1}^{i'}\sum_{j=1}^{j'} u(i,j)^2 = \min(i',j')
\]
as soon as $\la_{i'} > \la_{i'+1}$ and $\mu_{j'} > \mu_{j'+1}$. The same identity must be true with $v(i,j)$ replacing $u(i,j)$

In view of this, denote by $i_1< \cdots < i_p$ (resp. $j_1 < \cdots < j_q$) the indexes at which the singular values of $A$ (resp. $B$) differ form their successors, with the convention $\la_{d+1} = \mu_{d+1} = 0$. Let, for $k=1, \ldots, p$ and $l=1, \ldots, q$
\[
C(k,l) = \sum_{i=1}^{i_k}\sum_{j=1}^{j_l} u(i,j)^2.
\]
Then, we must have $C(k,l) = \min(i_k, j_l)$ for all $k,l$ and $u(i,j) = v(i,j)$ for $i=1, \ldots, i_p$ and $j=1, \ldots, j_q$. 

If, for all $i,j\leq d$, we let $U_{\lceil ij\rceil}$ be the matrix formed by the first $i$ rows and $j$ columns of $U$, the condition $C_{kl} = \min(i_k, j_l)$ requires that $U_{\lceil i_kj_l\rceil}U_{\lceil i_kj_l\rceil}^T =\Id[i_k]$ if $i_k\leq j_l$ and $U_{\lceil i_kj_l\rceil}^TU_{\lceil i_kj_l\rceil} =\Id[j_l]$ if $j_l\leq i_k$.
This shows that, if $i_k\leq j_l$,  the rows of $U_{\lceil i_kj_l\rceil}$ form an orthonormal family, and necessarily, all elements $u(i,j)$ for $i\leq i_k$ and $j>j_l$ vanish. The symmetric situation holds if $j_l\leq i_k$.

Let $r_k = i_k - i_{k-1}$ and $s_l = j_l - j_{l-1}$ (with $i_0=j_0=0$).
We now consider possible changes in the SVDs of $A$ and $B$. With our notation, the matrix $\La$ takes the form
\[
\Lambda = \begin{pmatrix}
\la_{i_1} \Id[r_1] & 0 & 0 & \cdots & 0 &0 &\ldots& 0 \\
0 & \la_{i_2} \Id[r_2] & 0 & \cdots & 0 &0 &\ldots& 0\\
\vdots &  &\ddots & &\vdots &\vdots&&\vdots\\
0 & 0 & \ldots & \la_{i_p} \Id[r_p] & 0 &0 &\ldots& 0\\
0 &&\ldots && 0&0 &\ldots& 0\\
\vdots &&&&\vdots &\vdots&&\vdots\\
0 &&\ldots && 0 &0 &\ldots& 0
\end{pmatrix}
\]
Let $W, \tilde W$ be  $n\times n$ and $d\times d$ orthogonal matrices taking the form
\[
W = \begin{pmatrix}
W_1 & 0 & 0 & \cdots & 0 \\
0 & W_2 & 0 & \cdots & 0\\
\vdots &  &\ddots & &\vdots \\
0 & 0 & \ldots & W_p & 0 \\
0 &&\ldots && W_{p+1}
\end{pmatrix},\ 
\tilde W = \begin{pmatrix}
W_1 & 0 & 0 & \cdots & 0 \\
0 & W_2 & 0 & \cdots & 0\\
\vdots &  &\ddots & &\vdots \\
0 & 0 & \ldots & W_p & 0 \\
0 &&\ldots && \tilde W_{p+1}
\end{pmatrix}
\]
where $W_1, \ldots, W_p$ are orthogonal with respective sizes $r_1, \ldots, r_p$, $W_{p+1}$ is orthogonal with size $n-i_p$ and $\tilde W_{p+1}$ is orthogonal with size $d-i_p$. Then we have
\[
WD\tilde W = D
\]
proving that $U_1$ can be replaced by $U_1W$ provided that $V_1$ is replaced by $V_1 \tilde  W$.  Similar transformations can be made on $U_2$ and $V_2$, with $U_2$ replaced by $U_2 Z$ and $V_2 $ by $V_2 \tilde Z$ with 
\[
Z = \begin{pmatrix}
Z_1 & 0 & 0 & \cdots & 0 \\
0 & Z_2 & 0 & \cdots & 0\\
\vdots &  &\ddots & &\vdots \\
0 & 0 & \ldots & Z_q & 0 \\
0 &&\ldots && Z_{q+1}
\end{pmatrix},\quad 
\tilde Z = \begin{pmatrix}
Z_1 & 0 & 0 & \cdots & 0 \\
0 & Z_2 & 0 & \cdots & 0\\
\vdots &  &\ddots & &\vdots \\
0 & 0 & \ldots & Z_q & 0 \\
0 &&\ldots && \tilde Z_{q+1}
\end{pmatrix}
\]
with a structure similar to $W$ and $\tilde W$, replacing $r_1, \ldots, r_p$ by $s_1, \ldots, s_q$. As a consequence, $U = U_1^TU_2$ can be replaced by $W^TUZ$ and $V$ by $\tilde W^T V \tilde Z$. To complete the proof, we need to show that, when \cref{eq:trace.ineq} is an equality, these matrices can be chosen so that  $W^TUZ=\Id[n]$ and $\tilde W^T V \tilde Z = \Id[d]$.

Let us consider a first step in this direction, assuming that $i_1 \leq j_1$ so that 
\[
U_{[i_1j_1]} U_{\lceil i_1j_1\rceil}^T = \Id[i_1]. 
\]
Complete $U_{\lceil i_1j_1\rceil}^T$ into a orthogonal matrix $Z_1 = [U_{\lceil i_1j_1\rceil}^T, \tilde U]$. Build a matrix $Z$ as above by taking $Z_2$, \ldots, $Z_{q+1}$ equal to the identity. Then $UZ$ has a first $i_1\times i_1$ block equal to $\Id[i_1]$, which implies that all coefficients on the right and below this block are zeros. If $j_1\leq i_1$, a similar construction can be made on the other side, letting $W_1 =   [U_{\lceil i_1j_1\rceil} \tilde U]$ with the first $j_1\times j_1$ block of the new matrix $U$ equal to the identity. Note that, since $V_{\lceil i_pj_q\rceil } = U_{\lceil i_pj_q\rceil }$, the same result is obtained on $V$ at the same time. 

Pursuing this way (and skipping the formal induction argument, which is a bit tedious), we can progressively introduce identity blocks into $U$ and $V$ and transform them into new matrices (that we still denote by $U$ and $V$) taking the form (letting $k=\min(i_p, j_q)$)
\[
U = \begin{pmatrix}
\Id[k] & 0\\
0 & \bar U
\end{pmatrix}
\text{ and }
V = \begin{pmatrix}
\Id[k] & 0\\
0 & \bar V
\end{pmatrix}
\]

If $k=i_p$ (resp. $k=j_q$), the final reduction can be obtained by choosing $W_{p+1} = \bar U$ and $\tilde W_{p+1} = \bar V$ (resp.  $Z_{p+1} = \bar U^T$ and $\tilde Z_{p+1} = \bar V^T$), leading to SVDs for $A$ and $B$ with identical matrices $U_1=U_2$ and $V_1=V_2$.
\end{proof}

\begin{remark}
Note that, since the singular values of $-A$ and of $A$ coincide, \cref{th:trace.ineq} implies
\begin{equation}
\label{eq:trace.ineq.abs}
\big|\trace(A^TB)\big| \leq \sum_{i=1}^m \la_i\mu_i\,.
\end{equation}
for all matrices $A$ and $B$, with equality if either $A$ and $B$ or $-A$ and $B$ have an SVD using the same bases. 
\end{remark}

\section{Applications}
Let $p$ and $d$ be integers with $p\leq d$.
Let $A\in \CS_{d}(\mR)$, $B\in \CS_{p}(\mR)$ be symmetric matrices.
We consider the following optimization problem: maximize, over  matrices $U\in \CM_{d,p}(\mR)$ such that $U^TU = \Id[p]$, the function
\[
F(U) = \trace(U^TAUB) = \trace(AUBU^T)\,.
\]
We first note that the singular values of $UBU^T$, which is $d\times d$, are the same as the eigenvalues of $B$ completed with zeros. Letting $\la_1 \geq \cdots \geq \la_d$ be the eigenvalues of $A$ and $\mu_1 \geq \cdots \geq \mu_p$ those of $B$, we therefore have, from \cref{th:trace.ineq},
\[
F(U) \leq \sum_{i=1}^p \la_i\mu_i.
\]
Introduce the eigenvalue decompositions of $A$ and $B$ in the form $A = V\Lambda V^T$ and $B = W M W^T$. For $F(U)$ to be equal to its upper-bound, we know that we must arrange $UBU^T$ to take the form
\[
UBU^T = V \begin{pmatrix}
M & 0\\ 0& 0
\end{pmatrix}
V^T.
\]
Use, as before, the notation $V_{\lceil dp\rceil}$ to denote the matrix formed with the $p$ first columns of $V$. Take $U = V_{\lceil dp\rceil}W^T$, which satisfies $U^TU = \Id[p]$. We then have
\[
V_{\lceil dp\rceil} W^T B W V_{\lceil dp\rceil}^T = V_{\lceil dp\rceil} M V_{\lceil dp\rceil}^T =  V \begin{pmatrix}
M & 0\\ 0& 0.
\end{pmatrix}
V^T,
\]
which shows that $U$ is optimal. 
We summarize this discussion in the next theorem.

\begin{theorem}
\label{th:pca.base}
Let $A\in \CS_{d}(\mR)$ and $B\in \CS_{p}(\mR)$ be   symmetric matrices,  with $p\leq d$. Let  eigenvalue decompositions of $A$ and $B$ be  given by $A = V\Lambda V^T$ and  $B= WMW^T$, where the diagonal elements of  $\Lambda$ (resp. $M$) are  $\la_1\geq \cdots \geq \la_d$ (resp. $\mu_1  \geq \cdots \geq \mu_p$).

Define $F(U) = \trace(AUBU^T)$, for $U\in \CM_{d,p}(\mR)$. Then,
\[
\max\left\{F(U): U^TU = \Id[p]\right\} = \sum_{i=1}^p \la_i\mu_i.
\]
This maximum is attained at 
\[
U = V_{\lceil d,p\rceil} W^T.
\]
\end{theorem}

The following corollary applies \cref{th:pca.base} with $B = \mathrm{diag}(\mu_1, \ldots, \mu_p)$.
\begin{corollary}
\label{cor:pca.base}
Let $A\in \CS_{d}(\mR)$ be a symmetric matrix with eigenvalues $\la_1\geq \cdots \geq \la_d$.
For $p\leq d$, let $\mu_1 \geq  \cdots \geq \mu_p > 0$ and define
\[
F(e_1, \ldots, e_p) = \sum_{i=1}^p \mu_i e_i^TAe_i.
\]
Then, the maximum of $F$ over all orthonormal families $e_1, \ldots, e_p$ in $\mR^d$ is $\sum_{i=1}^p \la_i\mu_i$ and is attained when $e_1, \ldots, e_p$ are eigenvectors of $A$ with eigenvalues $\la_1, \ldots, \la_p$. 

The minimum of $F$ over all orthonormal families $e_1, \ldots, e_p$ in $\mR^d$ is $\sum_{i=1}^p \la_{d-i+1}\mu_i$ and is attained when $e_1, \ldots, e_p$ are eigenvectors of $A$ with eigenvalues $\la_d, \ldots, \la_{d-p+1}$. 
\end{corollary}
\begin{proof} 
The statement about the maximum is just a special case of  \cref{th:pca.base}, with $B = \mathrm{diag}(\mu_1, \ldots, \mu_p)$, noting that the $i$th diagonal element of $U^TAU$ is $e_i^TAe_i$ where $e_i$ is the $i$th column of $U$.

The  statement about the minimum is deduced by replacing $A$ by $-A$.
\end{proof}
Applying this corollary with $p=1$, we retrieve the elementary result that $\la_1 = \max\{u^TAu: |u|=1\}$ and $\la_d = \min\{u^TAu: |u|=1\}$. 

\bigskip

To complete this chapter, we quickly state and prove Rayleigh's theorem.
\begin{theorem}
\label{th:rayleigh.base}
Let $A\in \CM_{d,d}(\mR)$ be a symmetric matrix with eigenvalues $\la_1 \geq \cdots \geq \la_d$. Then 
\[
\la_k = \max_{V: \dim(V) = k}\min\{u^TAu, u\in V, |u|=1\} = \min_{V: \dim(V) = d-k+1}\max\{u^TAu, u\in V, |u|=1\}
\]
where the min and max are taken over linear subspaces of $\mR^d$.
\end{theorem}
\begin{proof}
Let $e_1, \ldots, e_d$ be an orthonormal basis of eigenvectors of $A$ associated with $\la_1, \ldots, \la_d$. Let, for $k\leq l$,  $W_{k,l} = \vspan(e_k, \ldots, e_l)$. Let $V$ be  a subspace of dimension $k$. Then $V\cap W_{k,d} \neq \emptyset$ (because the sum of the dimensions of these two spaces is $d+1$). Taking $u_0$ with norm 1 in this intersection, we have  
\[
\min\{u^TAu, u\in V, |u|=1\} \leq u_0^TAu_0 \leq \max\{u^TAu, u\in W_{k,d}, |u|=1\} = \lambda_k,
\]
where the last identity follows by considering the eigenvalues of $A$ restricted to $W_{k,d}$.  So, the maximum of the right-hand side is indeed less than $\lambda_k$, and it is attained for $V = W_{1,k}$. This proves the first identity, and the second one can be obtained by applying the first one to $-A$.
\end{proof}

\section{Some matrix norms}
\label{sec:matrix.norms}

The {\em operator norm} of a matrix $A \in \CM_{n,d}(\mR)$, is defined as
\[
|A|_{\mathrm{op}} = \max \{|Ax|: x\in \mR^d, |x| = 1\}.
\]
It is equal to the square root of the largest eigenvalue of $A^TA$, i.e., to the largest singular value of $A$.

The {\em Frobenius norm } of $A$ is 
\[
|A|_F = \sqrt{\trace(A^TA)} = \sqrt{\sum_{i,j=1}^d A(i,j)^2},
\]
so that
\[
|A|_F = \left(\sum_{k=1}^m \sigma_k^2\right)^{1/2}
\]
where $\sigma_1, \ldots, \sigma_m $ are the singular values of $A$ (and $m  = \min(n,d)$). 

The {\em nuclear norm} of $A$ is defined by
\[
|A|_* = \sum_{k=1}^d \sig_k\,.
\]
One can prove that this is a norm using an equivalent definition, provided by the following proposition.
\begin{proposition}
\label{prop:nuclear}
Let $A$ be an $n$ by $d$ matrix. Then
\[
|A|_* = \max\Big\{\trace(UAV^T): U \in \CM_{n,n} \text{ and } U^TU = \Id,  V \in \CM_{d,d} \text{ and } V^TV = \Id\Big\}\,.
\]
\end{proposition}
\begin{proof}
The fact that $\trace(UAV^T) \leq |A|_*$ for any $U$ and $V$ is a consequence of the trace inequality applied with $B = [\Id, 0]$ or its transpose depending on whether $n\leq d$ or not. The upper-bound being attained when $U$ and $V$ are the matrices forming the singular value decomposition of $A$, the proof is complete.
\end{proof}
The fact that $|A|_*$ is a norm, for which the only non-trivial fact was the triangular inequality, now is an easy consequence of this proposition, because the maximum of the sum of two functions is always less than the sum of their maximums. 
More precisely, we have
\begin{align*}
|A+B|_* &= \max\{\trace(UAV^T) + \trace(UBV^T):\\
& \qquad \qquad \qquad \qquad U^TU = \Id, V^TV = \Id\}\\
&\leq \max\{\trace(UAV^T): U^TU = \Id, V^TV = \Id\}\\
& \quad + \max\{\trace(UBV^T): U^TU = \Id,  V^TV = \Id\}\\
&= |A|_* + |B|_*
\end{align*}


The nuclear norms is also called the Ky Fan norm of order $d$. Ky Fan norms of order $k$ (for $1\leq k\leq d$) associate to a matrix $A$ the quantity
\[
|A|_{(k)} = \lambda_1 + \cdots + \lambda_k,
\]
i.e., the sum of its $k$ largest singular values. One has the following proposition.

\begin{proposition}
\label{prop:ky.fan}
The Ky Fan norms satisfy the triangular inequality. 
\end{proposition}
\begin{proof}
We prove this following the argument suggested in \citet{bhatia2013matrix}. For $A\in\CM_{d,d}$,  and $k=1, \ldots, d$, let
$\trace_{(k)}(A)$ be the sum of the $k$   largest diagonal elements of $A$. Let, for a symmetric matrix $A$, $|A|'_{(k)}$ denote the sum of the $k$ largest eigenvalues of $A$ (it is equal to $|A|_{(k)}$ if $A$ is positive definite, but can also include negative values).

Then, for any  symmetric matrix $A\in \CS_d$, 
\begin{equation}
\label{eq:kf.1}
|A|'_{(k)} = \max\left\{\trace_{(k)}(UAU^T): U\in CO_d\right\}.
\end{equation}
To show this, assume that $V$ in $\CO_d$ diagonalizes $A$, so that $D = VAV^T$ is a diagonal matrix. Assume, without loss of generality, that the coefficients $\lambda_j = D(j,j)$ are non-increasing. Fix $U\in \CO_d$, let $B = UAU^T$ and $W = VU^T$ so that $D = WBW^T$, or $B = W^TDW$. Then, for any $j\leq d$, 
\[
B(j,j) = \sum_{i=1}^d W(i,j)^2 D(i,i).
\]
Then, for any $1\leq j_1 < \cdots < j_k\leq d$
\begin{align*}
\sum_{l=1}^k B(j_l, j_l) &= \sum_{i=1}^d D(i,i) \sum_{l=1}^k W(i, j_l)^2\\
&= \sum_{i=1}^k D(i,i) + \sum_{i=1}^k D(i,i) \big(\sum_{l=1}^k W(i, j_l)^2 - 1\big) + \sum_{i=k+1}^d D(i,i) \sum_{l=1}^k W(i, j_l)^2\\
&= \sum_{i=1}^k D(i,i) + \sum_{i=1}^k (D(i,i) - D(k,k)) \big(\sum_{l=1}^k W(i, j_l)^2 - 1\big) \\
&+ \sum_{i=k+1}^d (D(i,i) - D(k,k)) \sum_{l=1}^k W(i, j_l)^2
+ D(k,k) \left(\sum_{i=1}^n \sum_{j=1}^k W(i,j_l)^2 - k\right).
\end{align*}
Because $W$ is orthogonal, we have  $\sum_{l=1}^k W(i, j_l)^2 \leq 1$ and 
\[
\sum_{i=1}^n \sum_{j=1}^k W(i,j_l)^2 = k.
\]
This shows that the terms after $\sum_{i=1}^k D(i,i)$ in the upper bound are negative or zero, so that
\[
\sum_{l=1}^k B(j_l, j_l) \leq \sum_{i=1}^k D(i,i).
\]
The maximum of the left-hand side is $\trace_{(k)}(B)$. Noting that we get an equality when choosing $U=V$, the proof of \cref{eq:kf.1} is complete.

Using the same argument as that made above for the nuclear norm, one deduces from this that
\[
|A+B|_{(k)}' \leq |A|_{(k)}' + |B|_{(k)}'
\]
for all $A,B \in \CS_d$ and all $k=1, \ldots, d$. 

Now, let $A\in \CM_{n,d}$ and consider the symmetric matrix
\[
\tilde A = \begin{pmatrix}
0 & A^T\\ A & 0
\end{pmatrix}
\in \CS_{n+d}.
\]
Write a vector $u \in \mR^{n+d}$ as $u = \begin{pmatrix}
u_1\\u_2
\end{pmatrix}
$ with $u_1\in \mR^d$ and $u_2\in \mR^n$. Then $u$ is an eigenvector of $\tilde A$ for an eigenvalue $\lambda$ if and only if $A^Tu_2 =\lambda u_1$ and $Au_1 = \lambda u_2$, which implies that $A^TA u_1 = \lambda^2 u_1$ and $\lambda^2 $ is a singular value of $A$.  Conversely, if $\mu$ is a nonzero singular value of $A$, associated with eigenvector $u_1$, then $1 /\sqrt{\mu}$ and $-1 /\sqrt{\mu}$ are eigenvalues of $\tilde A$, associated with eigenvectors $\begin{pmatrix}
u_1 \\ \pm  Au_1 / \sqrt{\mu}
\end{pmatrix} 
$.
It follows from this that $|A|_{(k)} = |\tilde A|'_{(k)}$ for $k \leq \min(n,d)$ and therefore satisfies the triangle inequality. 
\end{proof}

We refer to \cite{bhatia2013matrix} for more examples of matrix norms, including, in particular those provided by taking $p$th powers in Ky Fan's norms, defining
\[
|A|_{(k,p)} = (\lambda_1^p + \cdots + \lambda_k^p)^{1/p}.
\]

\section{Low-rank approximation}
\label{sec:low.rank}
Singular-values decompositions provide optimal approximations of general matrices by low-rank ones in terms of the Frobenius norm. Take $p \leq \min(n,d)$. Optimal rank-$p$ approximations of an $(n,d)$ matrix $X$ in this sense are matrices $Z^*$ that minimize $|X-Z|_F^2$ among all matrices $Z$ of rank $p$.
Rank-$p$ matrices are described in the following proposition.

\begin{proposition}
\label{prop.rank.p}
An $(n,d)$ matrix $Z$ has rank $p$ if and only if it can be written in the form $Z  = AW^T$ where $A$ is $(n, p)$, and $W$ is $(d, p)$ with $W^TW = \Id[p]$, i.e., $W = [e_1, \ldots, e_p]$ where the columns form an orthonormal family of $\mR^d$.
\end{proposition}
\begin{proof}
The ``if'' part is obvious and we focus on the ``only if'' part. Assume that $Z$ has rank $p$.
Take  $W = [e_1, \ldots, e_p]$, where $(e_1, \ldots, e_p)$ is an orthonormal family in $\mathrm{Null}(Z)^\perp$. Letting $e_{p+1}, \ldots, e_d$ denote an orthonormal basis of  $\mathrm{Null}(Z)$, we have $\sum_{i=1}^d e_ie_i^T = \Id[d]$ and
\[
Z = Z\sum_{i=1}^d e_ie_i^T = Z\sum_{i=1}^p e_ie_i^T = Z WW^T
\]
so that one can take $A = Z W$. 
\end{proof}

Taking $Z = AW^T$ as in the proposition, we have
\[
|X-AW^T|^2_F = \trace(XX^T) - 2\trace(XWA^T) + \trace(AA^T).
\]
With fixed $W$, the optimal $A$ is given by $A = XW$, so that $W$ must minimize
\[
\trace(XX^T) - \trace(XWW^TX^T) = \trace(X^TX) - \trace(WW^TX^TX),
\]
i.e., maximize $\trace(WW^TX^TX)$. 
Let $m = \min(n,d)$ and $\lambda_1\geq \cdots\geq \lambda_m$ the singular values of $X$. The singular values of $X^TX$ and $\lambda^2_1\geq \cdots\geq \lambda^2_m$, possibly completed with zeros if $d > m$. Those of $WW^T$ and 1, with multiplicity $p$ and 0, with multiplicity $d-p$. The trace inequality shows that the maximum value of $\trace(WW^TX^TX)$ is $\lambda_1^2 + \cdots + \lambda_p^2$, and is attained when both $X^TX$ and $WW^T$ have singular values decompositions in the same bases, which implies, since these are symmetric matrices, that they have a shared basis of eigenvectors. This shows that taking $W = e_1e_1^T + \cdots + e_pe_p^T$ provides a solutions, where $e_1, \ldots, e_p$ are eigenvectors of $X^TX$ associated with  $\lambda_1^2, \ldots, \lambda_p^2$, or, equivalently, the first column vectors of $V$ in the singular values decomposition $X = UDV^T$. With our notation, this is $W = V_{\lceil d, p\rceil}$ and 
\[
A = XW =  U D V^T V_{\lceil d, p\rceil} = U D_{\lceil n, p\rceil}.
\]
We summarize this in the next proposition.
\begin{proposition}
\label{prop:rank.p.approx}
Let $X$ be an $(n,d)$ matrix and $X = UDV^T$ be a singular value decomposition of $X$. Then an optimal rank-$p$ approximation of $Z$ is provided by the matrix
\[
Z = XV_{\lceil d, p\rceil}V_{\lceil d, p\rceil}^T =  U D_{\lceil n, p\rceil}V_{\lceil d, p\rceil}^T.
\]
\end{proposition}

\chapter{Introduction to Optimization}
\label{sec:optim}
\label{chap:optim}

This chapter summarizes some fundamental concepts in optimization that will be used later in the book. The reader is referred to textbooks, such as \citet{beck2014introduction,eiselt2019nonlinear,nocedal2006nonlinear,boyd2011distributed}  and many others for proofs and deeper results.

\section{Basic Terminology}
\begin{enumerate}[label={\bf \arabic*.}]
\item If $I$ is a subset of $\mR$, a lower bound of $I$ is an element $u\in [-\infty, +\infty]$ such that $u\leq x$ for all $x\in I$. Among these lower bounds, there exists a largest element, denoted $\inf I \in [-\infty, +\infty]$, called the infimum of $I$ (by convention,  the infimum of an empty set is $+\infty$). Similarly, one defines the supremum of $I$, denoted $\sup I$, as the smallest upper bound of $I$ (and the supremum of an empty set is $-\infty$). Every set in $\mR$ has an infimum and a supremum, but these numbers do not necessarily belong to $I$. When they do, they are respectively called minimal and maximal elements of $I$, and are denoted $\min I$ and $\max I$. So, the statement ``$u= \min I$'' means $u\in I$ and $u\leq v$ for all $v\in I$.

\item If $F:\Omega \to \mR$ is a real-valued function defined on a subset $\Omega \subset \mR^d$, the infimum of $F$ over $\Omega$ is defined by
\[
\inf_{\Omega} F = \inf\{F(x): x\in \Omega\} 
\]
and its supremum is
\[
\sup_{\Omega} F = \sup\{F(x): x\in \Omega\}.
\]
As seen above both numbers are well defined, and can take infinite values. One says that $x\in \Omega$ is a (global) minimizer (resp. maximizer) of $F$ if $F(y) \geq F(x)$ (resp. $F(y)\leq F(x)$) for all $y\in \Omega$.  One also says that $F$ reaches its minimum (resp. maximum), or is minimized (resp. maximized) at $x$. Equivalently, $x$ is a minimizer (resp. maximizer) of $F$ if and only if $x\in \Omega$ and
\[
F(x) = \min \{F(y): y\in \Omega\} \ (\text{resp. } \max\{F(y): y\in \Omega\}).
\]
In such cases, one also writes $F(x) = \min_\Omega F$ or $F(x) = \max_\Omega F$. In particular, the
notation $u = \min_\Omega F$ indicates that $u = \inf_\Omega F$ and that there exists an $x$ in $\Omega$ such that $F(x) = u$ (i.e., that the infimum of $F$ over $\Omega$ is realized at some $x\in \Omega$).  Note that the infimum of a function always exists, but not necessarily its minimum. Also note that minimizers, when they exist, are not necessarily unique. We will denote by $\argmin_{\Omega} F$ (resp. $\argmax_\Omega F$) the (possibly empty) set of minimizers (resp. maximizers) of $F$

\item One says that $x$ is a local minimizer (resp. maximizer) of $F$ on $\Omega$ if there exists an open ball $B\subset \mR^d$ such that $x\in B$ and $F(x) = \min_{\Omega \cap B} F$ (resp. $F(x) = \max_{\Omega \cap B} F$). 

\item An optimization problem consists in finding a minimizer or maximizer of an ``objective function'' $F$.
Focusing from now on on minimization problems (statements for maximization problems are symmetric), we will always implicitly assume that a minimizer exists. The following provides some general assumptions on $F$ and $\Omega$ that ensure this fact. 

The sublevel sets of $F$ in $\Omega$  are denoted $[F\leq u]_\Omega$ (or simply $[F\leq u]$ when $\Omega = \mR^d$) for $u\in [-\infty, + \infty]$ with
\[
[F\leq u]_\Omega = \defset{x\in\Omega: F(x) \leq u}.
\] 
Note that 
\[
\argmin_\Omega F = \bigcap_{u>\inf F} [F\leq u]_\Omega\,.
\]

A typical requirement for $F$ is that its sublevel sets are  closed in $\mR^d$,  which means that, if a sequence $(x_n)$ in $\Omega$ satisfies, for some $u\in \mR$, $F(x_n) \leq u$ for all $n$ and converges to a limit $x$, then $x\in \Omega$ and $F(x) \leq u$. 
If this is true, one says that $F$ is {\em lower semi-continuous}, or l.s.c, on $\Omega$. If, in addition to being closed, the sublevel sets of $F$ are bounded (at least for $u$ small enough---larger than $\inf F$), then $\argmin_\Omega F$ is an intersection of nested compact sets, and is therefore not empty (so that  the optimization problem has at least one solution).  

\item Different assumptions on $F$ and $\Omega$ lead to different types of minimization problems, with specific underlying theory and algorithms.
\begin{enumerate}[label= \arabic*.,wide=10pt]
\item If $F$ is $C^1$ or smoother and $\Omega = \mR^d$, one speaks of an unconstrained smooth optimization problem. 
\item For constrained problems, $\Omega$ is often specified by a finite number of inequalities, i.e., 
\[
\Omega = \{x\in \mR^d: \gamma_i(x) \leq 0, i=1, \ldots, q\}.
\]
If $F$ and all functions $\gamma_1, \ldots, \gamma_q$ are $C^1$ one speaks of smooth constrained problems. 
\item If $\Omega$ is a convex set (i.e., $x, y\in \Omega \Rightarrow [x,y]\in \Omega$, where $[x,y]$ is the closed line segment connecting $x$ and $y$) and $F$ is a convex function (i.e., $F((1-t)x + ty) \leq (1-t) F(x) + t F(y)$ for all $x,y\in \Omega$), one speaks of a convex optimization problem. 

\item Non-smooth problems are often considered in data science, and lead to interesting algorithms and solutions. 

\item When both $F$ and $\gamma_1, \ldots, \gamma_q$ are affine functions, one speaks of a linear programming problem (or a linear program). (An affine function is a mapping $x\mapsto b^Tx + \beta$, $b\in \mR^d$, $\beta\in \mR$.) 

If $F$ is quadratic ($F(x) = \frac12 x^TAx - b^T x$), and all $\gamma_i$'s are affine, one speaks of a quadratic programming problem.

\item 
Finally, some machine learning problems are specified over discrete or finite sets $\Omega$ (for example $\mathbb Z^d$, or $\{0,1\}^d$), leading to combinatorial optimization problems. 
\end{enumerate}


\end{enumerate}

\section{Unconstrained Optimization Problems}
\label{sec:unconstrained}

\subsection{Conditions for optimality (general case)}


Consider a function $F: \Om \to \mR$ where $\Om$ is an {\em open} subset
of $\mR^d$. We first discuss
 the unconstrained optimization problem of finding
\begin{equation}
\label{eq:min.eq}
x^* \in \argmin_{\Om} F.
\end{equation}
The following result summarizes (non-identical) necessary and sufficient conditions that are applicable to such a solution.

\begin{theorem}
\label{th:opt.nec}
\begin{enumerate}
\item[{\bf Necessary conditions.}] 
Assume that $F$ is differentiable over $\Om$, and that $x^*$
is a local minimum of $F$. Then
$\nabla F(x^*) = 0$.

If $F$ is $C^2$, then, in addition, $\nabla^2F(x^*)$ must be positive semidefinite.
\item [{\bf Sufficient conditions.}]
\label{th:opt.suff}
Assume that $F\in C^2(\Om)$. If $x^*\in \Om$ is such that $\nabla F(x^*) = 0$
and $\nabla^2F(x^*)$ is positive definite, then $x^*$ is a local minimum of
$F$.
\end{enumerate}
\end{theorem} 

\begin{proof}
Necessary conditions:
Since $\Omega$ is open, it contains an open ball centered at $x^*$, with radius $\epsilon_0$ and therefore all segments $[x^*, x^*+\epsilon h]$ for all $\epsilon\in [0, \epsilon_0]$ and all unit norm vectors $h$. Since $x^*$ is a local minimum, we can choose $\epsilon_0$ so that $F(x^*+\epsilon h) \geq F(x^*)$ for all $h$ with $|h|=1$.

Using Taylor formula, we get (for $\epsilon \in [0, \epsilon_0]$, $|h|=1$)
\[
0\leq F(x^*+\epsilon h) - f(x^*) = \epsilon \int_0^1 dF(x^* + t\epsilon h)h dt\,.
\]
If $dF(x^*)h \neq 0$ for some $h$, then, for small enough $\epsilon$, $dF(x^* + t\epsilon h)h$ cannot change sign for $t\in [0,1]$ and therefore $\int_0^1 dF(x^* + t\epsilon h)h dt$ has the same sign as $dF(x^*)(h)$ which must therefore be positive. But the same argument can be made with $h$ replaced by $-h$, implying that $dF(x^*)(-h) = - dF(x^*)h$ is also positive, and this gives a contradiction. We therefore have $dF(x^*)(h) = 0$ for all $h$, i.e., $\nabla F(x^*) = 0$. 

Assume that $F$ is $C^2$. Then, making a second-order Taylor expansion, one gets
\[
0 \leq F(x^*+\epsilon h) - F(x^*) = \epsilon^2 \int_0^1 (1-t) d^2F(x^* + t\epsilon h)(h, h) dt.
\]
The same argument as above shows that, if $d^2F(x^* )(h, h) \neq 0$, then it must be positive. This shows that $d^2F(x^* )(h, h) \geq 0$ for all $h$ and $d^2F(x^*)$ (or its associated matrix $\nabla^2 F(x^*)$) is positive semidefinite.

Now, assume that $F$ is $C^2$ and $\nabla^2 F(x^*)$ positive definite. One still has
\[
F(x^*+\epsilon h) - F(x^*) = \epsilon^2 \int_0^1 (1-t) d^2F(x^* + t\epsilon h)(h, h) dt
\]
If $\nabla^2F(x^*) \succ 0$, then $\nabla^2F(x^*+t\epsilon h)\succ 0$ for small enough $\epsilon$, showing the the r.h.s. of the identity is positive for $h\neq 0$, and that $F(x^* + \epsilon h) > F(x^*)$.
\end{proof}

Because maximizing $F$ is the same as minimizing $-F$, necessary (resp. sufficient) conditions for optimality in maximization problems are immediately deduced from the above: it suffices to replace positive semidefinite (resp. positive definite) by negative semidefinite (resp. negative definite).

\subsection{Convex sets and functions}

\begin{definition}
\label{def:convex.set}
One says that a set $\Om\sub \mR^d$ is {\em convex} if and only if, for all $x,y\in \Om$, the closed segment $[x,y]$  also belongs to $\Om$. 

A function $F: \mR^d \to (-\infty, +\infty]$ is {\em convex} if, for all $\la\in [0, 1]$ and all $x,y\in \mR^d$, one has
\begin{equation}
\label{eq:conv.1}
F((1-\la)x + \la y) \leq (1-\la) F(x) + \la F(y). 
\end{equation}

If, whenever the lower bound is not infinite,  the inequality above is strict for $\lambda\in (0,1)$, one says that $F$ is strictly convex. 
\end{definition}

Note that, with our definition, convex functions can take the value  $+\infty$ but not  $-\infty$. In order for the upper-bound to make sense when $F$ takes infinite values, one makes the following convention: $a + (+\infty) = +\infty$ for any $a\in (-\infty, +\infty]$; $\la\cdot (+\infty) = +\infty$ for any $\la>0$; $0\cdot (+\infty)$ is not defined but $0\cdot (+\infty) + (+\infty) = +\infty$.

\begin{definition}
\label{def:domain}
The domain of $F$, denoted $\dom(F)$ is the set of $x\in \mR^d$ such that $F(x) < \infty$. One says that  $F$ is {\em proper} if $\dom(F)\neq \emp$. 

We will only consider proper convex functions in our discussions, which will simply be referred to as convex functions for brevity.
\end{definition}

\begin{proposition}
\label{prop:domain}
If $F$ is a convex function, then $\dom(F)$ is a convex subset of $\mR^d$. Conversely, if $\Omega$ is a convex set and $F$ satisfies \eqref{eq:conv.1} for all $x,y\in \Om$ (i.e., $F$ is convex on $\Omega$), then the extension $\hat F$ defined by $\hat F(x) = F(x)$ if $x\in \Omega$ and $\hat F(x) = +\infty$ is a convex function defined on $\mR^d$ (such that $\dom(\hat F) = \Omega$).
\end{proposition}
\begin{proof} The first statement is a direct consequence of  \cref{eq:conv.1}, which implies that $F$ is finite on $[x,y]$ as soon as it is finite at $x$ and at $y$. For the second statement, \cref{eq:conv.1} for $\hat F$ is true for $x,y\in \Omega$, since it is true for $F$, and the uper-bound is $+\infty$ otherwise.
\end{proof}
This proposition shows that there was no real loss of generality in requiring convex functions to be defined on the full space $\mR^d$. Note also that the upper bound in \cref{eq:conv.1} is infinite unless  both $x$ and $y$ belong to $\dom(F)$, so that the inequality only needs to be checked in that case.

One says that a function $F$ is {\em concave} if and only if $-F$ is convex. All definitions and properties made for convex functions then easily transcribe into similar statements for concave functions. We say that a function $f:I \to (-\infty, +\infty]$ (where $I$ is an interval) is non-decreasing if, for all $x,y\in I$, $x< y$ implies $f(x) \leq f(y)$. We say that $f$ is increasing if  if, for all $x,y\in I$, $x< y$ implies $f(x) < f(y)$ if $f(x) < \infty$ and $f(y) = \infty$ otherwise.

Inequality \eqref{eq:conv.1} has important consequences on minimization problems. For example, it implies the following proposition.
\begin{proposition}
\label{prop:conv.2}
Let $F$ be a convex (resp. strictly convex)  function on $\mR^d$.
If $x\in \dom(F)$ and $y\in\mR^d$, the function
\begin{equation}
\label{eq:conv.2}
\la \in (0,1] \mapsto \frac1\la (F((1-\la)x + \la y) - F(x))
\end{equation}
is non-decreasing (resp. increasing). 

Conversely, let $\Om\subset \mR^d$ be a convex set and $F: \Om \to (-\infty, +\infty)$ be a function such that the expression in \cref{eq:conv.2} is non-decreasing (resp. increasing) for all $x\in \dom(F)$ and $y\in\mR^d$. Then, the extension $\hat F$ of $F$ defined in \cref{prop:domain} is convex (resp. strictly convex).
\end{proposition}
\begin{proof}
Let $f(\lambda) = (F((1-\la)x + \la y) - F(x))/\lambda$.
Let $\mu \leq \la$ denote $z_\la = (1-\la)x + \la y$, $z_\mu = (1-\mu)x + \mu y$. One has
$ z_\mu = (1-\nu) x + \nu z_\la$, with $\nu = \mu/\la$, so that
\[
F(z_\mu) \leq (1-\mu/\la) F(x) + (\mu/\la) F(z_\la)\,.
\]
Subtracting $F(x)$ to both sides (which is allowed since $F(x) <\infty$) and dividing by $\mu$ yields
\[
 f(\mu) \leq f(\lambda)\,.
\]
If $F$ is strictly convex, then, either $F(z_\mu) = \infty$, in which case $f(\mu) = f(\lambda) = \infty$, or
\[
F(z_\mu) < (1-\mu/\la) F(x) + (\mu/\la) F(z_\la)\,.
\]
as soon as $0 < \mu < \lambda$, yielding
\[
 f(\mu) < f(\lambda)\,.
\]

Now consider the converse statement. By comparing the expression in \eqref{eq:conv.2} to that obtained with $\la=1$, we find, for all $x,y\in\Om$
\[
\frac1\la (F((1-\la)x + \la y) - F(x)) \leq F(y) - f(x)
\]
which is \cref{eq:conv.1}. Since $\hat F$ satisfies \cref{eq:conv.1} in its domain, it is convex. If the function in \cref{eq:conv.2} is increasing, then the inequality is strict for $0<\lambda < 1$ as soon as the lower bound is finite, and $F$ is strictly convex.
\end{proof}

\begin{corollary}
\label{corr:loc.min}
If $F$ is convex, any  local minimum of $F$ is  a global minimum.
\end{corollary}
\begin{proof}
If $x$ is a local minimum of $F$, then, obviously, $x\in \dom(F)$, and for any $y\in\mR^d$ and small enough $\mu>0$, 
$F(x) \leq F((1-\mu)x + \mu y)$. Using the function in \cref{eq:conv.2} for $\lambda = \mu$ and for  $\lambda=1$, we get
\[
0 \leq \frac1\mu  (F((1-\mu)x + \mu y) - F(x)) \leq F(y) - F(x)
\]
so that $x$ is a global minimum.
\end{proof}

\subsection{Relative interior}

If $\Omega$ is convex, then $\mathring \Omega$ and $\bar \Omega$ (its topological interior and closure) are convex too (the easy proof is left to the reader).  However, topological interiors of interesting convex sets are often empty, and a more adapted notion of {\em relative interior} is preferable. 

Define the  {\em affine hull} of a set $\Omega$, denoted $\mathrm{aff}(\Omega)$,  as the smallest affine subset of $\mR^d$ that contains $\Omega$. The vector space parallel to $\aff(\Omega)$ (generated by all differences $x-y$, $x,y\in \Omega$) will be denoted $\vaff(\Omega)$.
 Their dimension $k$, is the largest integer such that there exist $x_0, x_1, \ldots, x_k \in \Omega$ such that $x_1-x_0, \ldots, x_k-x_0$ are linearly independent. Moreover, given these points, elements of the affine hull are defined through {\em barycentric coordinates}, yielding
\[
\mathrm{aff}(\Omega) = \{ x =  \pe{\lambda}{0} x_0 + \cdots + \pe{\lambda}{k} x_k:, \pe{\lambda}{0}+\cdots + \pe{\lambda}{k} = 1\}\,.
\]
The coordinates $(\pe{\lambda}{0}, \ldots, \pe{\lambda}{k})$ 
are uniquely  associated to $x\in \mathrm{aff}(\Omega)$ and depend continuously on $x$.
They are indeed obtained by solving the linear system
\[
x - x_0=  \pe{\lambda}{1} (x_1-x_0) + \cdots + \pe{\lambda}{k} (x_k-x_0)
\]
which has a unique solution for $x\in \mathrm{aff}(\Omega)$ by linear independence. To see continuity, one can introduce the $k\times k$ matrix $G$ with entries $\pe G {ij}$ given by the inner products $(x_i-x_0)^T(x_j-x_0)$ and the vector $h(x)\in \mR^k$ with entries $\pe{h}{j}(x) = (x-x_0)^T(x_j-x_0)$. Continuity is then clear since $\lambda = G^{-1}h(x)$.

\begin{definition}
\label{def:relint}
If $\Omega$ is a convex set, then its relative interior, denoted $\relint(\Omega)$, is the set of all $x\in \Omega$ such that  there exists $\epsilon >0$ such that $\mathrm{aff}(\Omega) \cap B(x, \epsilon) \subset \Omega$.
\end{definition}

 We have the following important property.
\begin{proposition}
\label{prop:relint.not.empty}
Let $\Omega$ be a nonempty convex set.
If $x\in \relint(\Omega)$ and $y\in \Omega$, then $x_\lambda = (1-\la) x + \lambda y\in \relint(\Omega)$ for all $\lambda\in [0, 1)$. 

Moreover  $\relint(\Omega)$ is a nonempty convex set. 
\end{proposition}
\begin{proof}
Take $\epsilon$ such that $B(x, \epsilon) \cap \mathrm{aff}(\Omega)\subset \Omega$. 
Take any $z\in B(x_\lambda, (1-\lambda)\epsilon) \cap \mathrm{aff}(\Omega)$. Define $\tilde z$ such that $z = (1-\lambda)\tilde  z + \lambda y$, i.e.
\[
\tilde z = \frac{z-\lambda y}{1-\lambda}.
\]
Then $\tilde z \in \mathrm{aff}(\Omega)$ and
\[
|\tilde z - x| = \frac{|z-x_\lambda|}{1-\lambda} < \epsilon
\]
so that $\tilde z$, and therefore $z$ belongs to $\Omega$. This proves that $B(x_\lambda, (1-\lambda)\epsilon) \cap \mathrm{aff}(\Omega) \subset \Omega$ so that $x_\lambda\in \relint(\Omega)$.

If both $x$ and $y$ belong to $\relint(\Omega)$, then $x_\lambda \in \relint(\Omega)$ for $\lambda \in [0,1]$, showing that this set is convex.

We now show that $\relint(\Omega)\neq \emptyset$.
Let $k$ be the dimension of $\mathrm{aff}(\Omega)$, so that there exist $x_0, x_1, \ldots, x_k \in \Omega$ such that $x_1-x_0, \ldots, x_k-x_0$ are linearly independent. Consider the  ``simplex''
\[
S = \{ \pe{\lambda}0 x_0 + \cdots + \pe{\lambda}k x_k:, \pe\lambda 0+\cdots + \pe\lambda k = 1, \pe \lambda j\geq0, j=0, \ldots, k\},
\]
which
is included in $\Omega$. Then the average $x = (x_0 + \cdots +x_k)/(k+1)$ is such that $B(x, \epsilon) \cap \mathrm{aff}(\Omega) \subset S$ for small enough $\epsilon$. Otherwise, there would exist a sequence $y(n) = \pe\lambda 0(n) x_0 + \cdots + \pe\lambda k(n) x_k$ such that $\pe\lambda 0(n) + \cdots +\pe\lambda k(n) = 1$ and at least one $\pe\lambda j(n) < 0$ that converges to $x$. Let $\boldsymbol y_j$ be the set of elements in this sequence such that $\pe\lambda j(n)<0$. This set is infinite for  at least one $j$ and provides a subsequence of $y$ that also converges to $x$.  But this would imply that the $j^\text{th}$ barycentric coordinate, which depends continuously on $x$, is non-positive, which is a contradiction.

We therefore have $x\in \relint(\Omega)$, which completes the proof. 
\end{proof}

The following proposition provides an equivalent definition of the relative interior.
\begin{proposition}
\label{prop:relint.aff}
If $\Om$ is a convex set, then
\begin{equation}
\label{eq:relint}
\relint(\Om) = \defset{x\in \Om:\, \forall y\in \Om, \exists \ep>0 \text{ such that }x - \ep(y-x)\in \Om}.
\end{equation}
\end{proposition}
So $x$ belongs in the relative interior of $\Omega$ if, for all $y\in \Omega$, the segment $[x,y]$ can be extended on the $x$ side and still remain included in $\Omega$.
\begin{proof}
Let $A$ be the set in the r.h.s. of \cref{eq:relint}. The proof that $\relint(\Omega) \subset A$ is straightforward and left to the reader. We consider the reverse inclusion.

Let $x\in A$, and let $y\in \relint(\Omega)$, which is not empty. Then, for some $\epsilon>0$, we have
\[
z = x - \epsilon(y-x)\in \Omega.
\]
Since
\[
x = \frac{1}{1+\ep} (\epsilon y + z),
\]
\cref{prop:relint.not.empty} implies that $x\in \relint(\Omega)$.
 \end{proof}
 
Convex functions have important regularity properties in the relative interior of their domain, that we will denote $\rdom(F)$. Importantly: 
\[
\rdom(F) = \relint(\dom(F))  \neq \mathrm{int}(\dom(F)).
\]
A first such property is provided by the next proposition.
\begin{proposition}
\label{prop:conv.lip}
Let $F$ be a convex function. Then $F$ is locally Lipschitz continuous on $\rdom(F)$, i.e., for every compact subset $C \sub \rdom(F)$, there exists a constant $L>0$ such that $|F(x) - F(y)| \leq L|x-y|$ for all $x,y\in C$.  
\end{proposition}
This implies, in particular, that $F$ is continuous on  $\rdom(F)$. 
\begin{proof}
Take $x\in \rdom(F)$. Let $K = \defset{h\in \vaff(\dom(F)), |h|=1}$. Then,  the segment $[x-ah, x+ah]$ is included in $\rdom(F)$ for small enough $a$ and all $h\in K$. Since $F$ is convex, we have, for $t\leq a$, 
\[
F(x+th) - F(x) \leq \frac{t}{a} (F(x+ah) - F(x))
\]
Writing $x = \lambda (x-ah) + (1-\lambda) (x+th)$ with $\lambda = t/(t+a)$, we also have
\[
F(x) \leq \frac{t}{t+a} (F(x-ah)  + \frac{a}{t+a}  F(x+th))
\]
which can be rewritten as
\[
F(x) - F(x+th) \leq \frac{t}{a} (F(x-ah) - F(x)).
\]
These two inequalities show that $F$ is continuous at $x$ along any direction in $\vaff(\dom(F))$, which implies that $F$ is continuous at $x$. Given this, the differences $F(x+ah) - F(x)$ are bounded over the compact set $C$, by some constant $M$ and, the previous inequalities show that
\[
|F(y) - F(x)| \leq \frac{M}{a} |x-y|
\]
if $y \in \rdom(F)$, $|y-x|\leq a$.
 \end{proof}

\subsection{Derivatives of convex functions and optimality conditions}
 The following theorem provides a stronger version of optimality conditions for the minimization of differentiable convex functions. Note that we have only defined differentiability of functions defined over open sets. 
\begin{theorem}
Let $F$ be a convex function, with $\mathrm{int}({\dom}(F))\neq \emptyset$. Assume that
  $x\in \mathrm{int}({\dom}(F))$ and that $F$ is differentiable at $x$. Then, for all $y\in \mR^d$:
\begin{equation}
\label{eq:conv.3}
\nabla F(x)^T(y-x) \leq F(y) - F(x)\,.
\end{equation}
If $F$ is strictly convex, the inequality is strict for $y\neq x$. 
In particular, $\nabla F(x) = 0$ implies that $x$ is a {\em global minimizer} of $F$. It is the unique minimizer if $F$ is strictly convex. 

Conversely, if $F$ is $C^1$ on an open convex set $\Omega$ and satisfies \cref{eq:conv.3} for all $x, y\in \Omega$, then $F$ is convex. 
\label{th:convex.gradient}
\end{theorem}
\begin{proof}
\Cref{eq:conv.2} implies
\[
\frac1\la (F((1-\la)x + \la y) - F(x)) \leq F(y) - F(x), 0<\lambda\leq 1.
\]
Taking the limit of the lower bound
 for $\la\to 0$, $\lambda >0$ yields \cref{eq:conv.3}. If $F$ is strictly convex, the inequality is strict for $\lambda < 1$ and, since the l.h.s. is increasing in $\lambda$, it remains strict when $\lambda \downarrow 0$.
 
 Conversely, assuming \cref{eq:conv.3} for all $x, y\in \Omega$, the derivative of $\lambda \mapsto \frac1\la (F((1-\la)x + \la y) - F(x))$ is 
\[
\frac{1}{\la^2}(\la\nabla F(x+\la h)^T h - F(x+\la h) + F(x))
\]
with $h = y-x$, which is non-negative by \cref{eq:conv.3}. This proves that  $F$ is convex.  If \cref{eq:conv.3} holds with a strict inequality, then the derivative is positive and $\frac1\la (F((1-\la)x + \la y) - F(x))$ is increasing.
\end{proof}

The next proposition describes $C^2$ convex functions in terms of their second derivatives. 
\begin{proposition}
\label{prop:convex:c2}
Let $F$ be convex and twice differentiable at $x\in \mathrm{int}(\dom(F))$. Then $\nabla^2 F(x)$ is positive semi-definite.

Conversely, assume that $\Om = \dom(F)$ is an open set and that $F$ is $C^2$ on $\Om$ with a positive semi-definite second derivative. Then $F$ (or, rather, its extension $\hat F$) is convex. If the second derivative is everywhere positive definite, then $F$ is strictly convex. 
\end{proposition}
\begin{proof}
Using Taylor formula \cref{eq:taylor.epsilon} at order 2, we get, for any $h\in \mR^d$ with $|h| = 1$,
\[
\frac12 d^2F(x)(h,h) = \frac{1}{2t^2} d^2F(x)(th,th) = \frac{1}{t^2} (F(x+th) - F(x) - t\nabla F(x)^T h) + \ep(t) \geq \ep(t)
\]
with $\ep(t)\to 0$ when $t\to 0$, the last inequality deriving from \cref{eq:conv.3}. This shows that $d^2F(x)(h,h)\geq 0$.

To prove the second statement, assume that  $F$ is $C^2$ and $\nabla^2 F$ is positive semi-definite everywhere. Then \cref{eq:taylor.z} implies 
\[
F(y) - F(x) - \nabla F(x)^T (y-x)  = \frac12 (y-x)^T \nabla^2 F(z) (y-x)
\]
for some $z \in [x,y]$. Since the r.h.s. is non-negative, \cref{eq:conv.3} holds. If $\nabla^2 F$ is positive definite everywhere, then the r.h.s.\ is positive if $y\neq x$ and \cref{eq:conv.3} holds with a strict inequality.
\end{proof}

If $F$ is $C^2$ and $\nabla^2 F$ is positive definite and strictly convex, then \cref{eq:taylor.z} implies that, for some $z\in [x,y]$,
\[
F(y) - F(x) - \nabla F(x)^T (y-x) = \frac12 (y-x)^T \nabla^2 F(z) (y-x) \geq \frac{\rho_{\mathrm{min}}(\nabla^2 F(z))}{2} |y-x|^2
\]
where $\rho_{\mathrm{min}}(A)$ denotes the smallest eigenvalue of $A$. If this smallest eigenvalue is bounded from below away from zero, there exists a constant $m>0$ such that 
\begin{equation}
F(y) - F(x) - \nabla F(x)^T (y-x)  - \frac m2|y-x|^2 \geq 0.
\label{eq:convex.strong}
\end{equation}
This property is captured by the following definition, which does not require $F$ to be $C^2$.
\begin{definition}
\label{def:convex.strong}
A $C^1$ function $F$ is strongly convex if
\begin{enumerate}[label=\arabic*.]
\item $\mathrm{int}(\dom(F))\neq \emptyset$
\item There exists $m>0$ such that \cref{eq:convex.strong} holds for all $x\in \mathrm{int}(\dom(F))$ and $y\in \mR^d$.
\end{enumerate}
\end{definition}

We have the following proposition. 
\begin{proposition}
\label{prop:strong.strict}
If $F$ is strongly convex, then it is strictly convex, so that, in particular $\argmin F$ has at most one element.

If $\dom (F) = \mR^d$, then $\argmin F$ is not empty.
\end{proposition}
\begin{proof}
The first part is  a direct consequence of  \cref{eq:convex.strong} and \cref{th:convex.gradient}. 

For the second part, \cref{eq:convex.strong} implies that 
\[
F(x) - F(0) \geq \nabla F(0)^T x + \frac{m}{2} |x|^2 \geq |x|\left(\frac{m}{2} |x|^2 - 
|\nabla F(0)|\right)
\]
This shows that $F(x) > F(0)$ if $|x| > 2|\nabla F(0)|/m := r$ so that 
\[
\argmin F =  \argmin_{\bar B(0, r)} F.
\]
The set in the r.h.s. involves the minimization of a continuous function on a compact set, and is therefore not empty.
\end{proof}

We will use the following definition.
\begin{definition}
\label{def:lck}
A function $F: \Omega \to \mR^m$ is $L$-$C^k$, $L$ being a positive number, if it is $C^k$ and 
\[
|d^kF(x) - d^k F(y)| \leq L |x-y|.
\]
\end{definition}
If $F$ is $L$-$C^k$, then Taylor formula (\cref{eq:taylor.diff}) implies
\begin{equation}
\left| f(x+h) - f(x) - df(x) h - \frac12 d^2 f(x)(h,h) - \cdots - \frac1{k!} d^{k}f(x)(h, \ldots, h)\right| \leq 
\frac {L |h|^{k+1}}{(k+1)!}
\label{eq:taylor.L}
\end{equation}
for which we used the fact that
\[
\int_0^1 t(1-t)^{k-1}dt = \int_0^1 (1-t)^{k-1} dt - \int_0^1 (1-t)^{k}dt  = \frac1k - \frac1{k+1} = \frac{1}{k(k+1)}.
\]

If $F$ is strongly convex and is, in addition,  $L$-$C^1$ for some $L$, then using \cref{eq:taylor.L}, one gets the double inequality, for all $x,y\in \mathrm{int}(\dom(F))$:
\begin{equation}
\label{eq:convex.strong.L}
\frac m2 |y-x|^2 \leq F(y) - F(x) - \nabla F(x)^T (y-x) \leq \frac L2 |y-x|^2.
\end{equation} 

The following proposition will be used later.
\begin{proposition}
\label{prop:strong.convex}
Assume that $F$ is strongly convex, satisfying \cref{eq:convex.strong}, and that $\argmin F = \{x^*\}$ with $x^* \in \mathrm{int}(\dom(F))$. Then, for all $x\in \mathrm{int}( \dom(F))$:
\begin{equation}
\label{eq:convex.strong.prop}
\frac m2 |x-x^*|^2 \leq F(x) - F(x^*) \leq \frac1{2m} |\nabla F(x)|^2
\end{equation}
\end{proposition}
\begin{proof}
Since $\nabla F(x^*) = 0$, the first inequality is a consequence of \cref{eq:convex.strong} applied to $x=x^*$. Switching the role of $x$ and $x^*$, we have
\[
F(x^*) - F(x) - \nabla F(x)^T(x^*-x) \geq \frac m2 |x-x^*|^2
\]
so that 
\begin{equation}
0 \leq F(x) - F(x^*) \leq - \nabla F(x)^T(x^*-x) - \frac m2 |x-x^*|^2 \leq |\nabla F(x)| \, |x-x^*| - \frac m2 |x-x^*|^2
\label{eq:convex.strong.prop.1}
\end{equation}
The maximum of the r.h.s. with respect to $|x-x^*|$ is attained at $|\nabla F(x)|/m$, showing that
\[
F(x) - F(x^*) \leq \frac{1}{2m} |\nabla F(x)|^2, 
\]
which is the second inequality.
\end{proof}

%
%
%
%
%
%
%
%


\subsection{Direction of descent and steepest descent}
Gradient-based algorithms for optimization iteratively update the variable $x$, creating a sequence governed by an equation taking the form $x_{t+1} = x_t + \alpha_t h_t$ with $\alpha_t > 0$ and $h_t \in \mR^d$. To ensure that the objective function $F$ decreases at each step, $h_t$ is chosen to be a direction of descent for $F$ at $x_t$, a notion which, as seen below, is closely connected with the direction of $\nabla F(x_t)$.

\begin{definition}
\label{def:dir.desc}
 Let $\Omega$ be open in $\mR^d$ and $F:\Omega  \to \mR$  be a $C^1$ function. A direction of descent for $F$ at $x\in\Omega$ is a vector $h\neq 0 \in\mR^d$ such that there exists $\ep_0>0$ such that $F(x+\ep h) < F(x)$ for all $\ep\in (0, \ep_0]$.  
 \end{definition}

\begin{proposition}
\label{prop:dir.desc}
Assume that $F: \Omega\to \mR$  is  $C^1$ and take $x\in \Omega$. Then any direction $h$ such that $h^T\nabla F(x) < 0$ is a direction of descent for $F$ at $x$. Conversely, if $h$ is a direction of descent, then $h^T\nabla F(x) \leq 0$.
\end{proposition}
\begin{proof}
We have the first-order expansion $F(x+ \ep h) - F(x) = \ep h^T\nabla F(x) + o(\ep)$. If $h^T\nabla F(x) < 0$, the r.h.s. is negative for small enough $\epsilon$ and $h$ is a direction of descent. Similarly, if $h^T\nabla F(x) > 0$, the r.h.s. is positive for small enough $\epsilon$ and $h$ cannot be a direction of descent. 
\end{proof}

In particular,  $h = -\nabla F(x)$ is always a direction of descent. It is called the steepest descent direction because it minimizes $h\mapsto \prt_\alpha F(x+\alpha h)|_{\alpha = 0}$ over all $h$ such that $|h|^2=1$. However, this designation has a character of optimality that may be misleading, because using the Euclidean norm for the condition $|h|^2 = 1$ is not necessarily adapted to the optimization problem at hand. In the absence of additional information on the problem, it does have a canonical nature, as it is (up to rescaling) the only norm invariant to rotations (including permutations) of the coordinates. Such invariance is not necessarily desirable when the variable $x$ has a known structure (e.g., it is organized on a graph) which would be broken by permutation. Also, steepest refers to a local ``greedy'' evaluation, but may not be optimal from a global perspective. A simple example to illustrate this is the case of a quadratic function
\[
F(x) = \frac12 x^T A x - b^T x
\]
where $A \in \CS^{++}_n$ is a positive definite symmetric matrix. Then $\nabla F(x) = Ax - b$, but one may argue that $\nabla_A F(x) = A^{-1} \nabla F(x)$ (defined in \cref{eq:gradient.A}) is a better choice, because it allows the algorithm to reach the minimizer of $F$ in one step, since $x - \nabla_AF(x) = A^{-1} b$ (this statement disregards the cost associated in solving the system $Ax = b$, which can be an important factor in large dimension). Importantly, if $F$ is any $C^1$ function, and $A\in \CS^{++}_n$, the minimizer of $h\mapsto \prt_\alpha F(x+\alpha h)|_{\alpha = 0}$ over all $h$ such that $h^TAh = 1$ is given by $-\nabla_A F(x)$, i.e., $-\nabla_A F(x)$ is the steepest descent for the norm associated with $A$. This yields a  general version of steepest descent methods, iterating
\[
x_{t+1} = x_t - \alpha_t \nabla_{A_t} F(x_t)
\]
with $\alpha_t >0$ and $A_t\in \CS^{++}_n$.

One can also notice that $\nabla_A F(x)$ is also a minimizer of 
\[
F(x) + \nabla F(x)^Th + \frac12 h^T A h.
\]
When $\nabla^2 F(x)$ is positive definite, it is then natural to choose it as the matrix $A$, therefore taking $h = - \nabla^2 F(x)^{-1} \nabla F(x)$. This provides Newton's method for optimization. However, Newton method requires computing second derivatives of $F$, which can be computationally costly. It is, moreover, not a gradient-based method, which is the focus of this discussion.

%

%
%

\subsection{Convergence}
We now consider a descent algorithm
\begin{equation}
x_{t+1} = x_t + \alpha_t h_t
\label{eq:grad.desc}
\end{equation}
where $h_t$ is a direction of descent at $x_t$ for the objective function $F$. To ensure convergence,  suitable choices for the direction of descent and the step must be made at each iteration, and some assumptions on the objective function are needed. 

\begin{subequations}
Regarding the direction of descent, which must satisfy $h_k^T \nabla F(x_k) \leq 0$, we will assume a uniform control away from orthogonality to the gradient, with the condition
\begin{equation}
\label{eq:grad.angle}
-h_t^T \nabla F(x_t) \geq \epsilon |h_t|\, |\nabla F(x_t)|
\end{equation}
for some fixed $\epsilon >0$. Without loss of generality (given that a multiplicative step $\alpha_t$ must also be chosen), we assume that $h_t$ is commensurable to the gradient, namely, that
\begin{equation}
\label{eq:grad.norm}
\gamma_1 |\nabla F(x_t)| \leq |h_t| \leq \gamma_2 |\nabla F(x_t)|
\end{equation}
for fixed $0<\gamma_1 \leq \gamma_2$. If $h_t = \nabla_{A_t} F$, these assumptions are satisfied as soon as the smallest and largest eigenvalues of $A_t$ are controlled along the trajectory.
\end{subequations}

We have the following proposition.
\begin{proposition}
\label{prop:descent.gen}
Assume that $F$ is $L$-$C^1$.
Assume that $x_t$ satisfies \cref{eq:grad.desc} and that \cref{eq:grad.angle} and \cref{eq:grad.norm} hold. Then, there exist constants $\bar \alpha>0$ and $C>0$ that depends on $\gamma_1, \gamma_2$ and $\epsilon$, such that, for $\al_t \leq \bar\alpha$, one has
\begin{equation}
\label{eq:descent.gen}
F(x_{t+1}) - F(x_t) \leq - C \alpha_t |\nabla F(x_t)|^2.
\end{equation}
\end{proposition}
\begin{proof}
Applying \cref{eq:taylor.L} to $x_t$ and $x_{t+1}$, we get
\[
F(x_{t+1}) - F(x_t) - \alpha_t \nabla F(x_t)^T h_t \leq \frac{L}{2} \alpha_t^2 |h_t|^2
\]
Using \cref{eq:grad.desc} and  \cref{eq:grad.angle}, this gives
\[
F(x_{t+1}) - F(x_t) + \alpha_t\epsilon\gamma_1 |\nabla F(x_t)|^2 \leq \frac{L}{2} \alpha_t^2\gamma_2^2 |\nabla F(x_t)|^2
\]
so that
\[
F(x_{t+1}) - F(x_t) \leq - \alpha_t \big(\epsilon\gamma_1 -\alpha_t \gamma_2^2 L/2\big) |\nabla F(x_t)|^2. 
\]
It suffices to take $\bar \alpha = \epsilon\gamma_1/L\gamma_2^2$ and $C = \epsilon\gamma_1/2$ to obtain \cref{eq:descent.gen}.
\end{proof}
Iterating \cref{eq:descent.gen} for $t=1$ to $t=T-1$ yields 
\[
\sum_{t=1}^T \alpha_t |\nabla F(x_t)|^2 \leq \frac1C (F(x_1) - F(x_T)).
\]
If $F$ is bounded from below, and one takes $\alpha_t = \bar\alpha$ for all $t$, one deduces that
\[
\min \left\{|\nabla F(x_t)|^2: t=1, \ldots, T\right\} \leq \frac{F(x_1) - \inf F}{CT \bar\alpha} .
\]
We can deduce from this, for example, that there exists a sequence $t_1 < \cdots < t_n<\cdots$ such that  $\nabla F(x_{t_k}) \to 0$ when $k\to\infty$. In particular, if one runs \cref{eq:grad.desc} until $|\nabla F(x_t)|$ is smaller than a given tolerance level (which is standard), the procedure is guaranteed to terminate in a finite number of steps.

Stronger results may be obtained under stronger assumptions on $F$ and on the algorithm. The first assumption is an inequality similar to \cref{eq:descent.gen} and requires that, for some constant $C>0$,
\begin{equation}
\label{eq:descent.good}
F(x_{t+1}) - F(x_t) \leq - C  |\nabla F(x_t)|^2.
\end{equation}
Such an inequality can be deduced from \cref{eq:descent.gen} under the additional assumption that $\alpha_t$ is bounded from below and we will discuss later line search strategies that ensure its validity. The second assumption is that $F$ is convex.   

\begin{theorem}
\label{th:grad.desc.convex}
Assume that $F$ is convex and finite and that its sub-level set $[F\leq F(x_0)]$ is bounded. Assume that $\argmin F$ is not empty and let $x^*$ be a minimizer of $F$. If \cref{eq:descent.good} is true, 
then
\[
F(x_{t}) - F(x_*) \leq \frac{R^2}{C(t+1)}
\]
with $R = \max\{|x - x_*|: F(x) \leq F(x_0)\}$.
\end{theorem}
\begin{proof}
Note that the algorithm never leaves $[F\leq F(x_0)]$.
We have 
\[
F(x_{t+1}) - F(x^*) \leq F(x_t) - F(x^*) - C |\nabla F(x_t)|^2.
\]
Moreover,  by convexity, $F(x^{*}) - F(x_t) \geq \nabla F(x_t)^T (x^*-x_t)$, so that 
\[
F(x_t) - F(x^*) \leq \nabla F(x_t)^T (x_t-x^*) \leq |\nabla F(x_t)| R.
\]
Combining these two inequalities, we get
\[
F(x_{t+1}) - F(x^*) \leq F(x_t) - F(x^*) - \frac{C}{R^2}  (F(x_t) - F(x^*))^2.
\]
Introducing $\delta_t = (C/R^2) (F(x_t) - F(x^*))$, this inequality implies
\[
\delta_{t+1} \leq \delta_t (1-\delta_t)\,.
\]
Taking inverses, we get
\[
\frac1{\delta_{t+1}} \geq \frac1{\delta_t} + \frac1{1-\delta_t} \geq \frac1{\delta_t} + 1
\]
which implies $\frac{1}{\delta_t} \geq t+1$ or
$\delta_t \leq 1/(t+1)$, which in turn implies the statement of the theorem. 
\end{proof}

A faster convergence rate can be obtained if $F$ is assumed to be strongly convex. Indeed, if \cref{eq:convex.strong} and \cref{eq:descent.good} are satisfied, then (using \cref{prop:strong.convex}),
\begin{align*}
F(x_{t+1}) - F(x^*) &\leq F(x_t) - F(x^*) - C |\nabla F(x_t)|^2 \\
&\leq F(x_t) - F(x^*) - 2Cm (F(x_t) - F(x^*)) 
\\
&= (1-2Cm)  (F(x_t) - F(x^*))\,.
\end{align*}
We therefore get the proposition:
\begin{proposition}
\label{prop:strong.convex.conv}
If $F$ is finite and  satisfies \cref{eq:convex.strong}, and if  the descent algorithm satisfies \cref{eq:descent.good}, then
\[
F(x_{t}) - F(x^*) \leq (1-2Cm)^t (F(x_0) - F(x^*)).
\]
\end{proposition}

\subsection{Line search}
\Cref{prop:descent.gen} states that, to ensure that \cref{eq:descent.good} holds, it suffices to take a small enough step parameter $\alpha$. However, the   values of $\alpha$ that are acceptable depend on properties of the objective function that are rarely known in practice. Moreover, even if a valid choice is determined (this can  sometimes be done in practice by trial and error), setting a fixed value of $\alpha$ for the whole algorithm is often too conservative, as the best $\alpha$ when starting the algorithm may be different from the best one close to convergence.

For this reason, most gradient descent procedures select a parameter $\alpha_t$ at each step using a line search.  Given a current position and direction of descent $h$, a line search explores the values of $F(x + \al h)$, $\alpha\in (0, \alpha_{\mathrm{max}}] $ in order to discover some $\alpha^*$ that satisfies some desirable properties. We will assume in the following that $x$ and $h$ satisfy \cref{eq:grad.angle} and \cref{eq:grad.norm} for fixed $\epsilon, \gamma_1, \gamma_2$.

One possible strategy is to define $\alpha^*$ as a minimizer of  the scalar function 
\[
f_h(\alpha) = F(x + \alpha h)
\] 
over $ (0, \alpha_{\mathrm{max}}]$ for a given upper-bound $\ep_{\mathrm{max}}$. This can be implemented using, e.g.,  binary or ternary search algorithms, but  such algorithms would typically require a large number of number of evaluations of the function $F$, and would be too costly to be run at each iteration of a gradient descent procedure.  

Based on the previous convergence study, we should be happy with a line search procedure that ensures that \cref{eq:descent.good} is satisfied for some fixed value of the constant $C$. One such condition is the so-called Armijo rule that requires (with a fixed, typically small, value of $c_1>0$):
\begin{equation}
f_h(\alpha) \leq f_h(0) + c_1\alpha h^T\nabla f(x)\,.
\label{eq:armijo}
\end{equation}
We know that, under the assumptions of \cref{prop:descent.gen}, this condition can always be satisfied with a small enough value of $\alpha$. Such a value can be determined using a ``backtracking procedure,''  which, given $\alpha_{\mathrm{max}}$ and $\rho \in (0,1)$, takes $\alpha = \rho^k \al_{\mathrm{max}}$ where $k$ is the smallest integer such that \cref{eq:armijo} is satisfied. This value of $k$ is then determined iteratively, trying $\alpha_{\mathrm{max}}$, $\rho\alpha_{\mathrm{max}}$, $\rho^2\alpha_{\mathrm{max}}, \ldots$ until \cref{eq:armijo} is true (this provides the ``backtracking method''). 

A stronger requirement in the line search is to ensure that $\prt f_h(\alpha)$ is not ``too negative'' since one would otherwise be able to further reduce $f_h$ by taking a larger value of $\alpha$. This leads to the weak Wolfe conditions, which combine the Armijo's rule in \cref{eq:armijo} and  
\begin{subequations}
\begin{equation}
\label{eq:weak.wolfe}
\prt f_h(\alpha) = h^T \nabla F(x + \alpha h) \geq c_2  h^T \nabla F(x)
\end{equation}
for some constant $c_2 \in (c_1, 1)$. The strong Wolfe conditions require \cref{eq:armijo} and  
\begin{equation}
|h^T \nabla F(x + \alpha h)| \leq c_2  |h^T \nabla F(x)|.
\label{eq:strong.wolfe}
\end{equation}
(Since $h$ is a direction of descent, \cref{eq:strong.wolfe} requires \cref{eq:weak.wolfe} and the fact that $h^T \nabla F(x + \alpha h)$ does not take too large positive values.)
\end{subequations}
If $F$ is $L$-$C^1$, these conditions, with \cref{eq:grad.angle,eq:grad.norm}, imply \cref{eq:descent.good}. Indeed, \cref{eq:weak.wolfe} and the $L$-$C^1$ condition imply
\[
-(1-c_2) h^T \nabla F(x) \leq h^T (\nabla F(x + \alpha h) - \nabla F(x)) \leq L\alpha |h|^2
\]
and \cref{eq:grad.angle,eq:grad.norm} give
\[
(1-c_2) \epsilon |\nabla F(x)|^2 \leq \alpha L \gamma_2^2 |\nabla F(x)|^2
\]
showing that $\alpha \geq  (1-c_2)\epsilon / ( L \gamma_2^2)$. Moreover
\[
F(x+\alpha h) \leq F(x) + c_1\alpha h^T\nabla f(x) \leq F(x) -  c_1\alpha \epsilon |\nabla F(x)|^2
\]
so that
\[
F(x+\alpha h) \leq F(x) - \frac{c_1(1-c_2)\epsilon^2}{ L \gamma_2^2} |\nabla F(x)|^2.
\]
We have just proved the following proposition.
\begin{proposition}
\label{prop:wolfe.good}
Assume that $F$ is $L$-$C^1$ and that \cref{eq:grad.angle,eq:grad.norm,eq:armijo,eq:weak.wolfe} are satisfied.
Then there exists $C>0$, depending only of $L, \epsilon, \gamma_2, c_1$ and $c_2$such that
\[
F(x+\alpha h) \leq F(x) - C |\nabla F(x)|^2.
\]
\end{proposition}

The Wolfe conditions
can always be satisfied by some $\alpha$ as soon as $F$ is $C^1$ and bounded from below, and $h^T \nabla F(x) < 0$.  
The next proposition shows this result for the weak condition, while providing an algorithm finding an $\alpha$ that satisfies it in a finite number of steps.
\begin{proposition}
\label{prop:weak.wolfe.alg}
Let $f: \alpha \mapsto f(\alpha)$ be a $C^1$ function defined on $[0, +\infty)$ such that $f$ is bounded from below and $\prt_\alpha f(0) < 0$. Let $0 < c_1 < c_2 < 1$. 

Let $\alpha_{0,0} = \alpha_{0,1} = 0$ and $\alpha_0 > 0$. Define recursively sequences $\alpha_{n,0}, \alpha_{n,1}$ and $\alpha_{n}$ as follows. 
\begin{enumerate}[label=(\roman*)]
\item If  $f(\alpha_n) \leq  f(0) + c_1\alpha_n \prt_\alpha f(0)$ and  $\prt f(\alpha_n) \geq c_2 \prt_\alpha f(0)$ stop the construction.
\item If $f(\alpha_n) >  f(0) + c_1\alpha_n \prt_\alpha f(0)$ 
let $\alpha_{n+1} = (\alpha_n+\alpha_{n,0})/2$, $\alpha_{n+1,1} = \alpha_{n}$ and $\alpha_{n+1,0} = \alpha_{n,0}$.
\item If $f(\alpha_n) \leq  f(0) + c_1\alpha_n \prt_\alpha f(0)$ and $\prt f(\alpha_n) < c_2 \prt_\alpha f(0)$:
\begin{enumerate}[label = (\alph*),itemindent=1cm]
\item If $\alpha_{n,1}=0$, let  $\alpha_{n+1} = 2\alpha_n$, $\alpha_{n+1,0} = \alpha_{n}$ and $\alpha_{n+1,1} = \alpha_{n,1}$.
\item If $\alpha_{n,1}>0$,  let 
$\alpha_{n+1} = (\alpha_n+\alpha_{n,1})/2$, $\alpha_{n+1,0} = \alpha_{n}$ and $\alpha_{n+1,1} = \alpha_{n,1}$.
\end{enumerate}
%
\end{enumerate}
Then the sequences are always finite, i.e., the algorithm terminates in a finite number of steps.
\end{proposition}  

\begin{proof}
Assume, to get a contradiction, that the algorithm runs indefinitely, so that case (i) never occurs. If case (ii) never occurs, then one runs  step (iii-a) indefinitely, so that $\alpha_n \to \infty$ with $f(\alpha_n) \leq  f(0) + c_1\alpha_n \prt_\alpha f(0)$, and $f$ cannot be bounded from below, yielding a contradiction. As soon as case (ii) occurs, we have, at every step, $\alpha_{n,0} \geq \alpha_{n-1,0}$, $\alpha_{n,1} \leq \alpha_{n-1,1}$,
$\alpha_n \in [\alpha_{n,0}, \alpha_{n,1}]$, $f(\alpha_{n,1}) >  f(0) + c_1\alpha_{n,1} \prt_\alpha f(0)$, $f(\alpha_{n,0}) \leq  f(0) + c_1\alpha_{n,0} \prt_\alpha f(0)$ and $\prt f(\alpha_{n,0}) < c_2 \prt_\alpha f(0)$. This implies that
\[
f(\alpha_{n,1}) - f(\alpha_{n,0}) > c_1(\alpha_{n,1} - \alpha_{n,0}) \prt_\alpha f(0).
\]
Moreover, the updates imply that $(\alpha_{n+1,1} - \alpha_{n+1, 0}) = (\alpha_{n,1} - \alpha_{n,0})/2$. This requires that the three sequences $\alpha_n, \alpha_{n,0}$ and $\alpha_{n,1}$ converge to the same limit, $\alpha$. We have
\[
\prt_\alpha f(\alpha) = \lim_{n\to \infty}\frac{f(\alpha_{n,1}) - f(\alpha_{n,0})}{\alpha_{n,1} - \alpha_{n,0}} \geq c_1 \prt_\alpha f(0)
\]
and
\[
\prt_\alpha f(\alpha) = \lim_{n\to\infty} \prt_\alpha f(\alpha_{n,0}) \leq c_2 \prt_\alpha f(0)
\]
yielding $c_1 \prt_\alpha f(0) \leq c_2 \prt_\alpha f(0)$ which is impossible since $c_2>c_1$ and  $\prt_\alpha f(0)<0$.
\end{proof}

The existence of $\alpha$ satisfying the strong Wolfe condition is a consequence of the following proposition, which also provides an algorithm.
\begin{proposition}
Let $f: \alpha \mapsto f(\alpha)$ be a $C^1$ function defined on $[0, +\infty)$ such that $f$ is bounded from below and $\prt_\alpha f(0) < 0$. Let $0 < c_1 < c_2 < 1$. 

Let $\alpha_{0,0} = \alpha_{0,1} = 0$ and $\alpha_0 > 0$. Define recursively sequences $\alpha_{n,0}, \alpha_{n,1}$ and $\alpha_{n}$ as follows. 
\begin{enumerate}[label=(\roman*)]
\item If  $f(\alpha_n) \leq  f(0) + c_1\alpha_n \prt_\alpha f(0)$ and  $|\prt_\alpha f(\alpha_n)| \leq c_2 |\prt_\alpha f(0)|$ stop the construction.
\item If $f(\alpha_n) >  f(0) + c_1\alpha_n \prt_\alpha f(0)$ 
let $\alpha_{n+1} = (\alpha_n+\alpha_{n,0})/2$, $\alpha_{n+1,1} = \alpha_{n}$ and $\alpha_{n+1,0} = \alpha_{n,0}$.
\item If $f(\alpha_n) \leq  f(0) + c_1\alpha_n \prt_\alpha f(0)$ and $|\prt_\alpha f(\alpha_n)| > c_2 |\prt_\alpha f(0)|$:
\begin{enumerate}[label = (\alph*),itemindent=1cm]
\item If $\alpha_{n,1}=0$ and $\prt_\alpha f(\alpha_n) > -c_2 \prt_\alpha f(0)$, let  $\alpha_{n+1} = 2\alpha_n$, $\alpha_{n+1,0} = \alpha_{n,0}$ and $\alpha_{n+1,1} = \alpha_{n,1}$.
\item If $\alpha_{n,1}=0$ and $\prt_\alpha f(\alpha_n) < c_2 \prt_\alpha f(0)$, let  $\alpha_{n+1} = 2\alpha_n$, $\alpha_{n+1,0} = \alpha_{n}$ and $\alpha_{n+1,1} = \alpha_{n,1}$.
\item If $\alpha_{n,1}>0$ and $\prt_\alpha f(\alpha_n) > - c_2 \prt_\alpha f(0)$, let 
$\alpha_{n+1} = (\alpha_n+\alpha_{n,0})/2$, $\alpha_{n+1,1} = \alpha_{n}$ and $\alpha_{n+1,0} = \alpha_{n,0}$.
\item If $\alpha_{n,1}>0$ and $\prt_\alpha f(\alpha_n) < c_2 \prt_\alpha f(0)$, let 
$\alpha_{n+1} = (\alpha_n+\alpha_{n,1})/2$, $\alpha_{n+1,0} = \alpha_{n}$ and $\alpha_{n+1,1} = \alpha_{n,1}$.
\end{enumerate}
\end{enumerate} 
Then the sequences are always finite, i.e., the algorithm terminates in a finite number of steps.

\label{prop:line.search}
\end{proposition}

\begin{proof}
Assume that the algorithm runs indefinitely in order to get a contradiction.
If the algorithm never enters case (ii), then $\alpha_{n,1} = 0$ for all $n$, $\alpha_n$ tends to infinity and $f(\alpha_n) \leq  f(0) + c_1\alpha_n \prt_\alpha f(0)$, which contradicts the fact that $f$ is bounded from below. 

As soon as the algorithm enter (ii), we have, for all subsequent iterations:
$\alpha_{n,0}\leq \alpha_n\leq \alpha_{n,1}$, $\alpha_{n+1, 0}\geq \alpha_{n,0}$, $\alpha_{n+1, 1}\leq \alpha_{n,1}$ and $\alpha_{n+1,1} - \alpha_{n+1,0} = (\alpha_{n,1} - \alpha_{n,0})/2$. This implies that both $\alpha_{n,0}$ and $\alpha_{n,1}$ converge to the same limit $\alpha$.

Moreover, we have, at each step:
\[
f(\alpha_{n,1}) >  f(0) + c_1\alpha_{n,1} \prt_\alpha f(0) \text{ or }
\prt_\alpha f(\alpha_{n,1}) > - c_2 \prt_\alpha f(0)
\]
and
\[
f(\alpha_{n,0}) \leq  f(0) + c_1\alpha_{n,0} \prt_\alpha f(0) \text{ and }
\prt_\alpha f(\alpha_{n,0}) \leq c_2 \prt_\alpha f(0)\,.
\]

This implies that, at each step:
\[
\frac{f(\alpha_{n,1}) - f(\alpha_{n,0})}{\alpha_{n,1} - \alpha_{n,0}} > c_1 \prt_\alpha f(0) \text{ or }
\prt_\alpha f(\alpha_{n,1}) > - c_2 \prt_\alpha f(0)
\]
and
\[
\prt_\alpha f(\alpha_{n,0}) \leq c_2 \prt_\alpha f(0)\,.
\]
There inequalities remain satisfied at the limit, and we must have
\[
\prt_\alpha f(\alpha)  > c_1 \prt_\alpha f(0) \text{ or }
\prt_\alpha f(\alpha) > - c_2 \prt_\alpha f(0)
\]
and
\[
\prt_\alpha f(\alpha) \leq c_2 \prt_\alpha f(0)\,,
\]
which is a contradiction since $c_2>c_1$ and  $\prt_\alpha f(0)<0$.
\end{proof}

\section{Stochastic gradient descent}
\label{sec:sgd}
\subsection{Stochastic approximation methods}
In some situations, the computation of $\nabla F$ can be too costly, if not intractable, to run gradient descent updates while a low-cost stochastic approximation is available. For example, if $F$ is an average of a sum of many terms, the approximation may simply be based on averaging over a randomly selected subset of the terms. This leads to  a {\em stochastic approximation algorithm }\cite{robbins1985stochastic,kushner2003stochastic,benveniste2012adaptive,duflo2013random} called stochastic gradient descent (SGD). 
 
A general stochastic approximation algorithm of the Robbins-Monro type updates a parameter, denoted $x\in \mR^d$, using stochastic rules. One associates to each $x$ a probability distribution $(\pi_x)$ on some set $\CS$, and, for some function $H: \mR^d\times \CS \to \mR^d$, considers the sequence of random iterations:
\begin{equation}
\label{eq:sa.0}
\left\{
\begin{aligned}
\boldsymbol\xi_{t+1} &\sim \pi_{X_t}\\
X_{t+1} &= X_t + \al_{t+1} H(X_t, \boldsymbol\xi_{t+1})
\end{aligned}
\right.
\end{equation}
where $\boldsymbol\xi_{t+1}$ is a random variable and the notation $\boldsymbol\xi_{t+1} \sim \pi_{X_t}$ should be interpreted as the more precise statement that the conditional distribution of $\boldsymbol\xi_{t+1}$ given all past random variables $\mathcal U_t = (\boldsymbol\xi_1, X_1, \ldots, \boldsymbol\xi_t, X_t)$ only depends on $X_t$ and is given by $\pi_{X_t}$. 

It is sometimes assumed in the literature that $\pi_x$ does not depend on $x$. This is no real loss of generality because under mild assumptions, a random variable $\boldsymbol\xi$ following $\pi_x$ can be generated as function $U(x, \tilde {\boldsymbol\xi})$ where $\tilde {\boldsymbol\xi}$ follows a fixed distribution (such as that of a family of independent uniformly distributed variables) and one can replace $H(x, \xi)$ by $H(x, U(x, \tilde{\xi}))$. On the other hand, allowing $\pi$ to depend on $x$ brings little additional complication in the notation, and corresponds to the natural form of many applications.

More complex situations can also be considered, in which $\boldsymbol\xi_{t+1}$ is not conditionally independent of the past variables given $X_t$. For example, the conditional distribution of $\boldsymbol\xi_{t+1}$ given the past may also depends on $\boldsymbol\xi_t$, which allows for the combination of stochastic gradient methods with Markov chain Monte-Carlo methods. This situation is studied, for example, in \citet{metivier1987theoremes,benveniste2012adaptive}, and we will discuss an example in \cref{sec:stoc.grad}.
 
 \subsection{Deterministic approximation and convergence study} 
 \label{sec:sgd.convergence}
Introduce the  function
\[
\bar H(x) = E_{\pi_x}(H(x, \cdot))
\]
and write
\[
X_{t+1}  = X_t + \alpha_{t+1} \bar H(X_t)  + \al_{t+1} \boldsymbol\eta_{t+1}
\]
with $\boldsymbol\eta_{t+1} = H(X_t, \boldsymbol\xi_{t+1}) - \bar H(X_t)$ in order to represent the evolution of $X_t$ in \cref{eq:sa.0} as a perturbation
of the deterministic algorithm
\begin{equation}
\label{eq:sa.deter}
\bar x_{t+1}  = \bar x_t + \al_{n+1} \bar H(\bar x_t) 
\end{equation}
by the ``noise term'' $\al_{t+1} \boldsymbol\eta_{t+1}$. In many cases, the deterministic algorithm provides the limit behavior of the stochastic sequence, and one should ensure that this limit is as desired.  By definition, the conditional expectation of $\boldsymbol\eta_{t+1}$ given $\CU_t$ (the past) is zero and one says that $\alpha_{t+1}\boldsymbol\eta_{t+1}$ is a ``martingale increment.'' Then, 
\begin{equation}
\label{eq:sa.M}
M_T = \sum_{t=0}^T \al_{t+1} \boldsymbol\eta_{t+1}
\end{equation}
 is called a ``martingale.'' The theory of martingales offers numerous  tools for controlling the size of $M_T$ and is often a key element in proving the convergence of the method. 
 
 Many convergence results have been provided in the literature and can be found in textbooks or lecture notes such as \citet{benaim1999dynamics,kushner2003stochastic,benveniste2012adaptive}. These  results rely on some smoothness and growth assumptions made on the function $H$, and on the dynamics of the deterministic equation \cref{eq:sa.deter}. Depending on these assumptions, proofs may become quite technical. We will here restrict to a reasonably simple context and assume that
\begin{enumerate}[label=(H\arabic*)]
\item There exists a constant $C$ such that, for all $x\in \mR^d$,
\[
E_{\pi_x}(|H(x, \cdot)|^2) \leq C(1+|x|^2).
\]
\item There exists $x^*\in\mR^d$ and $\mu > 0$ such that, for all $x\in \mR^d$ 
\[
(x - x^*)^T \bar H(x) \leq - \mu |x - x^*|^2.
\]
\end{enumerate}
 
Assuming this, let $A_t = |X_t - x^*|^2$ and  $a_t = \myE(A_t)$. Then, using \cref{eq:sa.0},
\[
A_{t+1} = A_t + 2\al_{t+1} (X_t - x^*)^T H(X_t, \boldsymbol\xi_{t+1}) + \al_{t+1}^2 |H(X_t, \boldsymbol\xi_{t+1})|^2\,.
\]
Taking the conditional expectation given past variables yields
\begin{align*}
\myE(A_{t+1}\mid\CU_t) &= A_t + 2\al_{t+1} (X_t - x^*)^T \bar H(X_t) + \al_{t+1}^2 E_{\pi_{x_t}}(|H(X_t, \cdot)|^2)\\
&\leq A_t - 2\al_{t+1}\mu A_t  + \al_{t+1}^2 C (1+|X_t|^2)\\ 
&\leq (1 - 2\al_{t+1}\mu + C\al_{t+1}^2)  A_t  + \al_{t+1}^2 \tilde C
\end{align*}
with $\tilde C = 1 + |x^*|^2$. Taking expectations on both sides yields
\begin{equation}
\label{eq:sg.upper.bound.alpha}
a_{t+1} \leq (1 - 2\al_{t+1}\mu + C\al_{t+1}^2)  a_t  + \al_{t+1}^2 \tilde C.
\end{equation}

We state the next step in the computation as a lemma.
\begin{lemma}
\label{lem:sg.upper.bound}
Assume that the sequence $a_t$ satisfies the recursive inequality 
\begin{equation}
\label{eq:sg.upper.bound.0}
a_{t+1} \leq (1-\delta_t) a_t + \epsilon_t
\end{equation}
with $0 \leq \delta_t \leq 1$.
Let $v_{k,t} = \prod_{j=k+1}^t (1-\de_{j})$. Then
\begin{equation}
a_t \leq a_0 v_{0,t} + \sum_{k=1}^t \ep_k v_{k,t}.
\label{eq:sg.upper.bound}
\end{equation}
\end{lemma}
\begin{proof}
Letting $b_t = a_t /v_{0,t}$, we get
\[
b_{t+1} \leq b_t + \frac{\ep_{t+1}}{v_{0,t+1}}
\]
so that
\[
b_t \leq b_0 + \sum_{k=1}^t \frac{\ep_{k}}{v_{0,k}},
\]
and
\[
a_t \leq a_0 v_{0,t} + \sum_{k=1}^t \ep_k v_{k,t}.
\]
\end{proof}
Using \cref{eq:sg.upper.bound.alpha}, 
we can apply this lemma with $\ep_t = \tilde C \al_t^2$ and $\de_t = 2\al_{t}\mu - C\al_{t}^2$, making the additional assumption that, for all $t$, $\al_t < \min(\frac1{2\mu}, \frac{2\mu}{C})$, which ensures that $0<\de_t < 1$. 

Starting with a simple case, assume that the steps $\ga_t$ are constant, equal to some value $\ga$ (yielding also constant $\de$ and $\ep$). Then, \cref{eq:sg.upper.bound} gives 
\begin{equation}
\label{eq:sgd.constant}
a_t \leq a_0 (1-\de)^t + \ep \sum_{k=1}^t (1-\de)^{t-k-1} \leq a_0 (1-\de)^t + \frac{\ep}{\de}.
\end{equation}
Returning to the expression of $\de$ and $\ep$ as functions of $\alpha$, this gives
\[
a_t \leq a_0 (1-2\alpha\mu+\alpha^2C)^t + \frac{\alpha \tilde C}{2\mu - \alpha C}.
\]
This shows that $\mathrm{limsup}\, a_t = O(\alpha)$.

Return to the general case in which the steps depend on $t$, we will  use the following simple result, that we state as a lemma for future reference.
\begin{lemma}
\label{lem:sgd.simple}
Assume that the double indexed sequence $w_{st}$, $s\leq t$ of non-negative numbers is bounded and such that, for all $s$, $\lim_{t\to \infty} w_{st} = 0$. Let $\beta_1, \beta_2, \ldots$ be such that
\[
\sum_{t=1}^\infty |\beta_t| < \infty.
\]
Then
\[
\lim_{t \to \infty} \sum_{s=1}^t \beta_s w_{st} = 0.
\]
\end{lemma}
\begin{proof}
For any $t_0$, we have
\[
\left|\sum_{s=1}^t \beta_s w_{st}\right| \leq \max_{s} |\beta_s| \sum_{s=1}^{t_0}  w_{st} + \max_{s,t} |w_{st}| \sum_{s=t_0+1} |\beta_s|
\]
so that
\[
\limsup_{t\to \infty} \left|\sum_{s=1}^t \beta_s w_{st}\right| \leq \max_{s,t} |w_{st}| \sum_{s=t_0+1} |\beta_s|
\]
and since this upper bound can be made arbitrarily small, the result follows. 
\end{proof}

\Cref{lem:sg.upper.bound} implies that 
\[
a_t \leq a_0 v_{0,t} + \tilde C \sum_{s=1}^t \alpha^2_{s+1} v_{s,t}.
\]
Assume that 
\begin{enumerate}[label=(H3)]
\item $\sum_{k=1}^\infty \al_k = \infty$ and $\sum_{k=1}^\infty \al_k^2 < \infty$,
\end{enumerate} 
Then $\lim_{t\to\infty} v_{st} = 0$ for all $s$ and \cref{lem:sgd.simple} implies that $a_t$ tends to zero. So, we have just proved that, if (H1), (H2) and (H3) are true,
the sequence $X_t$ converges in the $L^2$ sense to $x^*$. Actually, under these conditions, one can show that $X_t$ converge to $x^*$ almost surely, and we refer to \citet{benveniste2012adaptive}, Chapter 5, for a proof (the argument above for an $L^2$ convergence follows the one given in \citet{nemirovski_robust_2009}).

\bigskip

Under (H3), one can say much more on the asymptotic behavior of the algorithm by comparing it with an ordinary differential equation. The ``ODE method,'' introduced in \citet{ljung1977analysis}, is indeed a fundamental tool for the analysis of stochastic approximation algorithms. The correspondence between discrete and continuous times is provided by the sequence $\al_t$.  More precisely, let $\tau_0=0$ and $\tau_{t} =  \tau_{t-1} + \alpha_t$, $t\geq 1$. From (H3), $\tau_t \to \infty$ when $t\to\infty$. Define the piecewise linear interpolation $x^\ell(\rho)$ of the sequence $x_t$  by 
\[
X^\ell(\rho) = X_t +  \frac{\rho-\tau_t}{\al_{t+1}} (X_{t+1} - X_t),\quad  \rho\in [\tau_{t}, \tau_{t+1}).
\] 
Switching to continuous time allows us to interpret the average iteration $\bar x_{t+1} = \bar x_t + \al_{t+1} \bar H(\bar x_t) $ as an Euler discretization scheme for the ordinary differential equation (ODE)
\begin{equation}
\label{eq:sa.ode}
\prt_\rho \bar x = \bar H(\bar x).
\end{equation}

Most of the insight on long-term behavior of  stochastic approximations results from the fact that the random process $x$ behaves asymptotically like solutions of this ODE. One has, for example, the following result, for which we introduce some additional notation. 

Assume that \cref{eq:sa.ode} has unique solutions for given initial conditions on any finite interval, and denote by $\phi(\rho, \omega)$ its solution at time $\rho$ initialized with $\bar x(0) = \omega$. Let $\al^c(\rho)$ and $\boldsymbol\eta^c(\rho)$ be piecewise constant interpolations of $(\al_t)$ and $(\boldsymbol\eta_t)$ defined by $\al^c(\rho) = \al_{t+1}$ and $\boldsymbol\eta^c(\rho) = \boldsymbol\eta_{t+1}$ on the interval $[\tau_{t}, \tau_{t+1})$. Finally, let 
\[
\De(\rho, T) = \max_{s\in [\rho, \rho+T]} \left| \int_\rho^s \boldsymbol\eta^c(u) du\right|.
\]
The following proposition (see \cite{benaim1999dynamics}) compares the tails of the process $x^\ell$  (i.e., the functions $x^\ell(\rho+s)$, $s\geq 0$) with the solutions of the ODE over finite intervals. 
\begin{proposition}[Benaim]
\label{prop:sgd.ode}
Assume that $\bar H$ is Lipschitz and bounded. Then, for some constant $C(T)$ that only depends on $T$ and $\bar H$, one has, for all $\rho\geq 0$
\begin{equation}
\label{eq:ode.2}
\sup_{h\in [0,T]} |X^\ell(\rho+h) - \phi(h, X^\ell(\rho))| \leq C(T) \left( \Delta(\rho-1, T+1) + \max_{s\in[\rho, \rho+T]} \al^c(s)\right)\,.
\end{equation}
\end{proposition}
Recall that $\bar H$ being Lipschitz means that there exists a constant $C$ such that
\[
|\bar H (w) - \bar H(w') |\leq C |w-w'|
\]
for all $w, w' \in \mR^p$. 

In the upper-bound in \cref{eq:ode.2}, the term $\Delta(\rho-1, T+1)$ is a random variable. It can be related to the variations
\[
\De'(t, N) = \max_{k = 0, \ldots, N}  |M_{t+k} - M_t|, 
\]
where $M$ is defined in \eqref{eq:sa.M},
because, if $m(\rho)$ is the largest integer $t$ such that $\tau_t \leq \rho$, then
\[
\De'(m(\rho)+1, m(\rho+T) - m(\rho)) \leq \De(\rho, T) \leq  \De'(m(\rho), m(\rho+T) - m(t)+1).
\]
In the case we are considering, one can use martingale inequalities (called Doob's inequalities) to control $\De'$. One has, for example,
\begin{equation}
\label{eq:doob}
P\left(\max_{0\leq k\leq N}|M_{t+k} - M_t| > \la\right) \leq \frac{E(|M_{t+N} - M_t|^2)}{\la^2}.
\end{equation}
Furthermore, using the fact that $E(\eta_{k+1}\eta_{l+1}) = 0$ if $k\neq l$, one has
\[
E(|M_{t+N} - M_t|^2 )= \sum_{k=t}^{t+N} \alpha_{k+1}^2 E(|\boldsymbol\eta_{t+1}|^2).
\]
If we assume (to simplify) that $H$ is bounded and $\sum_{k=1}^\infty \alpha_k^2 <\infty$ then, for some constant $C$, we have
\[
E(|M_{t+N} - M_t|^2) \leq C \sum_{k=t}^{\infty} \alpha_{k+1}^2 \to 0
\]
and inequality \eqref{eq:doob} can then be used in \eqref{eq:ode.2} to control the probability of deviation of the stochastic approximation from the solution of the ODE over finite intervals (a little more work is required under weaker assumptions on $H$, such as (H1)).
\bigskip

Proposition \ref{prop:sgd.ode} cannot be used with $T = \infty$ because the constant $C(T)$ typically grows exponentially with $T$. In order to draw conclusions on the limit of the process $W$, one needs additional assumptions on the stability of the ODE. We refer to \cite{benaim1999dynamics} for a collection of results on the 
relationship between invariant sets and attractors of the ODE and  limit trajectories of the stochastic approximation. We here quote one of these results which is especially relevant for SGD. 
\begin{proposition}
\label{prop:sgd.conv}
Assume that $\bar H = -\nabla E$ is the gradient of a function $E$ and that $\nabla E$ only vanishes at a finite number of points. Assume also that $X_t$ is bounded. Then $X_t$ converges to a point $x^*$ such that  $\nabla E(x^*) = 0$.
\end{proposition}

Some additional conditions on $\bar H$ can ensure that stochastic approximation trajectories remain bounded. The simplest one assumes the existence of a ``Lyapunov function'' that controls the ODE at infinity. The following result is a simplified version of Theorem 17 in \citet{benveniste2012adaptive}.
\begin{theorem}
\label{th:sgd.lyap}
In addition to the hypotheses previously made, 
assume that there exists a $C^2$ function $U$ with bounded second derivatives and $K_0 >0$ such that, for all$x$ such that  $|x| \geq K_0$, 
\begin{align*}
&\nabla U(x)^T\bar H(x) \leq 0,\\
&U(x) \geq \gamma |x|^2, \gamma>0.
\end{align*}
Then, the trajectories $X^\ell(\rho)$ are almost surely bounded.
\end{theorem}
Note that hypothesis (H2) above implies the theorem's assumptions. 


\subsection{The ADAM algorithm}
\label{sec:adam}
ADAM (for adaptive moment estimation \citep{kingma2014adam}) is a popular variant of stochastic gradient descent.
 When dealing with high-dimensional vectors $\CW$, using a single ``gain'' parameter ($\gamma_{n+1}$ in \cref{eq:nn.sgd}) is a limiting assumption since all parameters do not need to scale the same way. This can sometimes be handled by reweighting the components of $H$, i.e., using iterations
 \[
 X_{t+1} = X_t + \alpha_t D_t H(X_t, \boldsymbol\xi_{t+1})
 \]
 where $D_t$ is a (typically diagonal) matrix. The previous theory can be applied to situations in which $D$ may be random, provided it converges almost surely to a fixed matrix.

The ADAM algorithm provides such a construction (without the theoretical guarantees) in which $D_t$ is computed using past iterations of the algorithm. It requires several parameters, namely: $\alpha$: the algorithm gain, taken as constant (e.g., $\alpha=0.001$);
Two parameters $\beta_1$ and $\beta_2$ for moment estimates (e.g. $\be_1 = 0.9$ and $\be_2=0.999$);
A small number $\ep$ (e.g., $\ep = 10^{-8}$) to avoid divisions by 0.
In addition, ADAM defines two vectors: a mean $m$ and a second moment $v$, respectively initialized at $\boldsymbol 0$ and $\dsone$. The ADAM iterations are given below, in which $g^{\otimes 2}$ denotes the vector obtained by squaring each coefficient of a vector $g$.

\begin{algorithm}[ADAM]
\label{alg:adam}
\begin{enumerate}[label=\arabic*., wide=0.5cm]
\item Let $X_t$ be the current state, $m_t$ and $v_t$ the current mean and variance.
\item Generate $\boldsymbol\xi_{t+1}$ and let $g_{t+1} = H(X_t, \boldsymbol\xi_{t+1})$.
\item Update $m_{t+1} = \beta_1 m_t + (1-\beta_1) g_{t+1}$.
\item Update $v_{t+1} = \beta_2 v_t + (1-\beta_2) g^{\odot 2}_{t+1}$.
\item Let $\hat m_{t+1} = m_{t+1} / (1-\beta_1^{t+1})$ and $\hat v_{t+1} = v_{t+1} / (1-\beta_2^{t+1})$
\item
Set 
\[
X_{t+1}  = X_t - \alpha \frac{\hat m_{t+1}}{\sqrt{\hat v_{t+1}} + \ep}
\]
\end{enumerate}
\end{algorithm}
Note that the iteration on $m_t$ and $v_{t}$ correspond to defining 
\[ 
\hat m_t = \frac{\be_1}{1 - \be_1^t} \sum_{k=0}^t (1 - \be_1)^{t-k} g_k
\]
and   
\[
\hat v_t = \frac{\be_2}{1 - \be_2^t} \sum_{k=0}^t (1 - \be_2)^{t-k} g_k^{\odot 2}.
\]

\section{Constrained optimization problems}
\label{sec:opt.constr}
\subsection{Lagrange multipliers}
A constrained optimization problem minimizes a function $F$ over a closed subset $\Omega$ of $\mR^d$, with $\Omega \neq \mR^d$. This restriction invalidates, in a large part, the optimality conditions discussed in \cref{sec:unconstrained}. These conditions indeed apply to minimizers belonging to the interior of $\Omega$, and therefore do not hold when they lie at its boundary, which is a very common situation in practice ($\Omega$  often has an empty interior).

In this section, which follows the discussion given in \citet{wright2022optimization}, we review conditions for optimality for constrained minimization of smooth functions, in two cases. The first one, discussed in this section, is when $\Omega$ is defined by a finite number of smooth constraints, leading, under some assumptions, to the Karush-Kuhn-Tucker (or KKT) conditions. The second one, in the next section, specializes to closed convex $\Omega$.

\subsubsection{KKT conditions}
\label{sec:kkt.smooth}

We introduce some notation.
Let $\ga_i$, for $i\in \CC$, be $C^1$ functions  $\gamma_i: \mR^d \to \mR$, where $\CC$ is a finite set of indices.
We assume that $\CC$ is divided into two non-intersecting parts, $\CC = \CE \cup \CI$ and consider minimization problems searching for
\begin{equation}
\label{eq:min.eq.cons}
x^* \in \argmin_{\Om} F
\end{equation}
where
\begin{equation}
\label{eq:omega.constraints}
\Omega = \{x\in \mR^d: \ga_i(x) = 0, i\in \CE \text{ and } \ga_i(x) \leq 0, i\in \CI\}.
\end{equation}
The set  $\Omega$ of all $x$ that satisfy the constraints is called the {\em feasible set} for the considered problem. We will always assume that  it is non-empty. If $x\in \Omega$, one defines the set $\CA(x)$ of {\em active constraints} at $x$ to be 
\[
\CA(x) = \defset{i\in \CC: \ga_i(x) = 0}.
\]
One obviously has $\CE\subset \CA(x)$ for $x\in \Omega$. 

To be valid, the KKT conditions require some additional assumptions on potential minimizers, called ``constraint qualifications.''  An instance of such assumptions is provided by the next definition.
\begin{definition}
\label{def:mfcq}
A point $x\in \Omega$ satisfies the  Mangasarian-Fromovitz constraint qualifications (MF-CQ) if the following two conditions are satisfied. 
\begin{enumerate}[label=(MF\arabic*)]
\item The vectors $(\nabla \ga_i(x), i\in \CE)$ are linearly independent.
\item There exists  a vector $h\in \mR^d$ such that $h^T \nabla \ga_i(x) = 0$ for all $i\in\CE$ and $h^T \nabla \ga_i(x) < 0$ for all $i\in \CA(x)\cap\CI$. 
\end{enumerate}
\end{definition}
A sufficient (and easier to check) condition for $x$ to satisfy these constraints is when the vectors  $(\nabla \ga_i(x), i\in \CA(x))$ are linearly independent \citep{borwein2010convex}. Indeed, if the latter ``LI-CQ'' condition is true, then any set of values can be assigned to $h^T \nabla \gamma_i(x)$ with the existence of a vector $h$ that achieves them.

We introduce the {\em Lagrangian}
\begin{equation}
\label{eq:lag}
L(x, \la) = F(x) + \sum_{i\in \CC} \la_i \ga_i(x)
\end{equation}
where the real numbers $\la_i$, $ i\in \CC$ are called {\em Lagrange multipliers}.
The following theorem (stated without proof, see, e.g., \citep{nocedal2006nonlinear,bonnans2006numerical}) provides necessary conditions satisfied by solutions of the constrained minimization problem that satisfy the constraint qualifications. 
\begin{theorem}
\label{th:lag.finite}
Assume $x^*\in \Omega$ is a solution of \cref{eq:min.eq.cons}, and that $x^*$ satisfies the MF-CQ conditions. Then there exist Lagrange multipliers $\la_i$, $i\in \CC$, such that
\begin{equation}
\label{eq:opt.kkt}
\left\{
\begin{aligned}
&\prt_x L(x^*, \la)  = 0\\
&\la_i \geq 0 \text{ if } i \in \CI, \text{ with } \la_i = 0 \text{ when } i\not\in \CA(x^*) 
\end{aligned}
\right.
\end{equation}
\end{theorem}
Conditions \eqref{eq:opt.kkt} are the KKT conditions for the constrained optimization problem. The second set of conditions is often called the {\em complementary slackness conditions} and state that $\la_i = 0$ for an inequality constraint unless this constraint is satisfied with an equality. The next section provides examples in which the MF-CQ conditions are not satisfied and Theorem \ref{th:lag.finite} does not hold. However, these conditions are not needed in the special case when the constraints are affine.
\begin{theorem}
\label{th:lag.finite.aff}
Assume that for all $i\in \CA(x^*)$, the functions $\ga_i$ are affine, i.e., $\gamma_i(x) = b_i^T x + \beta_i$ for some $b\in \mR^d$ and $\beta\in \mR$.  Then \cref{eq:opt.kkt} holds at any solution of \cref{eq:min.eq.cons}.
\end{theorem}

\begin{remark}
\label{rem:positive.const}
We have taken the convention to express the inequality constraints as $\ga_i(x)\leq 0$, $i\in\CI$. With the reverse convention, i.e., $\ga_i(x)\geq 0$, $i\in\CI$, one generally defines the Lagrangian as
\[
L(x, \la) = F(x) - \sum_{i\in \CC} \la_i \ga_i(x)
\]
and the KKT conditions remain unchanged.
\end{remark}

\paragraph{Examples.}
Constraint qualifications are important to ensure the validity of the theorem.
Consider a problem with equality constraints only, and replace it by
\[
x^* \in \argmin_{\Om} F
\]
subject to  $\tilde \ga_i(x) = 0$, $i\in \CE$, with $\tilde \ga_i = \ga_i^2$.
We clearly did not change the problem. However, the previous theorem applied to the Lagrangian 
\[
L(x, \la) = F(x) + \sum_{i\in \CC} \la_i \tilde \ga_i(x)
\]
would require an optimal solution to satisfy $\nabla F(x) = 0$, because $\nabla\tilde \ga_i(x) = 2\ga_i(x) \nabla \ga_i(x) = 0$ for any feasible solution. Minimizers of constrained problems do not necessarily satisfy $\nabla F(x) = 0$, however. This is no contradiction with the theorem since $\nabla\tilde \ga_i(x) = 0$ for all $i$ shows that no feasible point satisfies the MF-CQ.

To take a more specific example, still with equality constraints,
let $d=3$, $\CC=\{1,2\}$ with
$F(x,y,z) =  x/2 + y$ and $\ga_1(x,y,z) = x^2 - y^2, \ga_2(x,y,z) = y-z^2$. 
Note that
$\ga_1 = \ga_2 = 0$ implies that $y = |x|$, so that, for a feasible point, $F(x,y,z) = |x| + x/2 \geq 0$ and
vanishes only when $x=y=0$, in which case $z=0$ also. So $(0,0,0)$ is a global minimizer.
We have $dF(0) = (1/2, 1, 0)$, $d\ga_1(0) = (0,0,0)$ and $d\ga_2(0) = (0,1,0)$ so that 0 does not satisfy the MF-CQ.
The equation
\[
dF(0) + \la_1 d\ga_1(0) + \la_2 d\ga_2(0) = 0
\]
has no solution $(\la_1, \la_2)$, so that the conclusion of the theorem does not hold.

\subsection{Convex constraints}

We now consider the case in which $\Omega$ is a closed convex set. 
To specify the optimality conditions in this case, we need the following definition.
\begin{definition}
\label{def:normal.cone}
Let $\Omega\subset \mR^d$ be convex and let $x\in \Omega$. The normal cone to $\Omega$ at $x$ is the set 
\begin{equation}
\CN_\Omega(x) = \{h\in \mR^d: h^T(y-x) \leq 0 \text{ for all } y \in\Omega\}
\label{eq:normal.cone}
\end{equation}
\end{definition}
The normal cone is an example of convex cone. (A convex subset  $\Gamma$  of $\mR^d$ is called a convex cone, if it is such that $\lambda x \in \Gamma$ for all $x\in \Gamma$ and $\lambda \geq 0$, a property obviously satisfied by $\CN_\Omega(x)$.) It should also be clear from the definition that  non-zero vectors in $\CN_\Omega(x)$ always point outside $\Omega$, i.e.,  $x+h \not\in\Omega$ if $h\in\CN_\Omega(x)$, $h\neq 0$.  Here are some examples.
\begin{itemize}[wide]
\item
If $x$ is in the interior of $\Omega$, then $\CN_\Omega(x) = \{0\}$. 
\item Assume that $\Omega$ is a half space, i.e.,  $\Omega = \{x: b^Tx + \beta \leq 0\}$ with $|b|=1$, and take $x \in \prt\Omega$, i.e., $b^Tx + \beta = 0$.  Then
\[
 \CN_\Omega(x) = \{h = \mu b: \mu\geq 0\}\,.
\]

Indeed, any element of $\mR^d$ can be written as $y = x + \lambda b + q$ with $q^T b = 0$, and $y\in \Omega$ if and only if $\lambda \leq 0$. Fix such a $y$ and take $h\in \mR^d$, decomposed as $h = \mu b + r$, with $r^Tb = 0$. We have $h^T(y-x) = \lambda\mu + r^Tq$. Clearly, if $\mu < 0$, or if $r \neq 0$, one can find $\lambda \leq 0$ and $q \perp b$ such that $h^T(y-x) > 0$. One the other hand, if $\mu \leq 0$ and $r=0$, we have $h^T(y-x)\leq 0$ for all $y\in \Omega$, which proves the above statement. 
\item
With a similar argument, if $\Omega = \{x: b^Tx + \beta = 0\}$ is a hyperplane, one finds that
\[
 \CN_\Omega(x) = \{h = \lambda b: \lambda\in \mR \}\,.
\]
\end{itemize}

One can build normal cones to domains associated with multiple inequalities or equalities based on the following theorem.
\begin{theorem}
\label{th:normal.cone.intersect}
Let $\Omega_1$ and $\Omega_2$ be two convex sets with $\relint(\Omega_1) \cap \relint(\Omega_2) \neq \emptyset$. Then, if $x\in \Omega_1\cap \Omega_2$
\[
\CN_{\Omega_1\cap \Omega_2} (x) = \CN_{\Omega_1} (x) + \CN_{\Omega_2} (x) 
\] 
Here, the addition is the standard sum between sets in a vector space:
\[
A+B = \{x+y: x\in A, y\in B\}.
\]
\end{theorem} 
%


Finally, we note that, if $x\in \relint(\Omega)$, then
\begin{equation}
\label{eq:normal.relint}
\CN_\Omega(x) = \{h\in \mR^d: h^T(y-x) = 0, y\in \Omega\}.
\end{equation}
Indeed, if $y\in \Omega$, then $x + \epsilon (y-x)\in \Omega$ for small enough $\epsilon$ (positive or negative).  For $ h \in \CN_\Omega(x)$, the condition $\epsilon h^T(y-x) \leq 0$ for small enough $\epsilon$ requires that $h^T(y-x)=0$.

\bigskip

With this definition in hand, we have the following theorem.
\begin{theorem}
\label{th:optimality.convex.constraints}
Let $F$ be a $C^1$ function and $\Omega$ a closed convex set. If $x^* \in \argmin_\Omega F$, then
\begin{equation}
- \nabla F(x^*) \in \CN_\Omega(x^*).
\label{eq:optimality.convex.constraints}
\end{equation}

If $F$ is convex and \cref{eq:optimality.convex.constraints} holds, then $x^* \in \argmin_\Omega F$.
\end{theorem}
\begin{proof}
Assume that $x^* \in \argmin_\Omega F$. If $y\in \Omega$, then $x^* + t(y-x^*)\in \Omega$ for all $t\in [0,1]$ and the function $f(t) = F(x+ t(y-x^*))$ is $C^1$ on $[0,1]$, with a minimum at $t=0$. This requires that $\prt_t f(0) = \nabla F(x^*)^T(y-x^*) \geq 0$, because, if $\prt_t f(0) <0$, a Taylor expansion  would show that $f(t) < f(0)$ for small enough $t>0$.

If $F$ is convex and \cref{eq:optimality.convex.constraints} holds, we have $F(y) \geq F(x^*) +  \nabla F(x^*)^T(y-x^*)$ by convexity, so that
\[
F(x^*) \leq F(y) + (-\nabla F(x^*))^T(y-x^*) \leq F(y).
\]
\end{proof}

\subsection{Applications}

\noindent{\bf Lagrange multipliers revisited.}
Consider $\Omega$ defined by \cref{eq:omega.constraints}, with the additional assumptions that $\gamma_i(x) = b_i^T x + \beta_i$ for $i\in \CE$ and $\gamma_i$ is convex for $i\in \CI$, which ensure that $\Omega$ is convex. 
Define
\[
\CN'_\gamma(x) = \left\{g = \sum_{i\in \CA(x)} \lambda_i \nabla \gamma_i(x): \lambda_i \geq 0, i\in \CA(x) \cap \CI\right \}.
\]
Then, the KKT conditions in \cref{eq:opt.kkt} can be rewritten as 
\[
- \nabla F(x^*) \in \CN'_\gamma(x^*).
\]
Note that one always have $\CN'_\gamma(x) \subset \CN_\Omega(x)$ since, for $g =  \sum_{i\in \CA(x)} \lambda_i \nabla \gamma_i(x)\in \CN'_\gamma(x)$, one has, for $y\in \Omega$,
\begin{align*}
g^T (y-x) &= \sum_{i\in \CA(x)} \lambda_i \nabla \gamma_i(x)^T(y-x)\\
&= \sum_{i\in \CE} \lambda_i (a_i^T y - a_i^Tx) + \sum_{i\in \CA(x)\cap\CI} \lambda_i(\gamma_i(x) + \nabla \gamma_i(x)^T(y-x))\\
& =  \sum_{i\in \CA(x)\cap\CI} \lambda_i(\gamma_i(x) + \nabla \gamma_i(x)^T(y-x))\\
&\leq \lambda_i \gamma_i(y) \leq 0,
\end{align*}
in which the have used the facts that $a_i^Tx = a_i^T y = -\beta_i$ for $x,y\in\Omega$, $i\in \CE$, $\gamma_i(x) = 0$ for $i\in\CA(x)$ and the convexity of $\gamma_i$. Constraint qualifications such as those considered above are sufficient conditions that ensure the identity between the two sets.

Consider now the situation of \cref{th:lag.finite.aff}, and assume that  all constraints are affine inequalities, $\gamma_i(x) = b_i^T x + \beta \leq 0, i\in \CI$. Then, the statement 
$\CN_\Omega(x) \subset \CN'_\gamma(x)$ can be reexpressed as follows. All $h \in \mR^d$ such that
\[
h^T(y-x) \leq 0
\]
as soon as $b_i^T(y-x)\leq 0$ for all $i\in \CA(x)$ must take the form
\[
h = \sum_{i\in\CA(x)} \lambda_i b_i
\]
with $\pe\lambda i \geq 0$. This property is called  {\em Farkas's lemma} (see, e.g. \citep{rockafellar1970convex}). Note that affine equalities $b_i^T x + \beta=0$ can be included as two inequalities $b_i^T x + \beta\leq 0$, $-b_i^T x - \beta\leq 0$, which removes the sign constraint on the corresponding $\pe\lambda i$ and therefore yields
\cref{th:lag.finite.aff}. 

%
%

\noindent{\bf Positive semi-definite matrices.}
We now take an example in which \cref{th:lag.finite.aff} does not apply directly. Let $\Omega = \CS_n^+$ be the space of positive semidefinite $n\times n$  matrices, considered as a subset of the space $\CM_n$ of $n\times n$ matrices, itself identified with $\mR^{n^2}$. With this identification, the Euclidean inner product between two matrices can be expressed as  $(A,B) \mapsto \trace(A^TB)$.

We have $A\in \CS_n^+$ if and only if, for all $u\in \mR^d$, 
$u^T A u \geq 0$, which provides an infinity of linear inequality constraints on $A$. Elements of $\CN_{\CS^+}(A)$ are matrices $H\in \CM_n$ such that
\[
\trace(H^T (B-A)) \leq 0
\]
for all $B\in \CS_n^+$, and we want to make this normal cone explicit. We first note that, every square matrix $H$ can be decomposed as the sum of a symmetric matrix, $H_s$ and of a skew symmetric one, $H_a$ (namely, $H_s = (H+H^T)/2$ and $H_a = (H-H^T)/2$). We have moreover 
$\trace(H_a^T (B-A)) = 0$, so the condition is only on the symmetric part of $H$.

For any $u\in \mR^d$, one can take $B = A + uu^T$, which belongs to $\CS^+_n$, with
$\trace(H_s^T(B-A)) = u^T H_s u$. This shows that, for $H$ to belong to $\CN_{S^+_n}(A)$, one needs $H_s \preceq 0$.

Now, take an eigenvector $u$ of $A$ with eigenvalue $\rho>0$. Then $B = A - \alpha uu^T$ is also in $\CS_n^+$ as soon as $0 \leq \alpha \leq \rho$, and $\trace(H_s^T (B-A))  = - \alpha u^T H_s u$. So, if $H \in \CN_{\CS^+_n}(A)$, we have  $u^T H_s u \geq 0$, and since $H_s \preceq 0$, this gives $u^T H_s u = 0$. Still because $H_s$ is negative semi-definite, this implies $H_s u = 0$. (This can be shown, for example, using Schwarz's inequality which says that $(u^T H_s v)^2 \leq (u^T H_s u) (v^T H_s v)$ for all $v\in \mR^d$.) Decomposing $A$ with respect to its non-zero eigenvectors, i.e., writing
\[
A = \sum_{k=1}^p \rho_k u_ku_k^T
\]
where $p = \mathrm{rank}(A)$, we get $AH_s = H_s A = 0$. We therefore obtained the proposition
\begin{proposition}
\label{prop:semidefinite.normal}
Let $A \in \CS^+_n$. Then $H\in \CM_{n}$ belongs to $\CN_{\CS^+_n}(A)$ if and only if $-H_s \in \CS_n^+$ and $H_s A = 0$, where $H_s = (H+H^T)/2$.
\end{proposition}
%
%
%
%

Now, if one wants to minimize a function $F$ over positive semidefinite matrices, and $A^*$ is a minimizer, we get the necessary condition that $A^* (\nabla F(A^*))_s = 0 $ with  $(\nabla F(A^*))_s$ positive semidefinite. These conditions are sufficient if $F$ is convex. 

For example, 
take
\begin{equation}
\label{eq:semi.definite.example}
F(A) = \frac{1}{2} \trace(A^2) - \trace(BA)
\end{equation}
with $B\in \CS_n$. Then $(\nabla F(A))_s = A -B$ and the condition is $A(A-B) = 0$ with $A \succeq B$. If $B$ is diagonalized in the form $B = U^T D U$, with $U$ orthogonal and $D$ diagonal, then the solution is $A^* = U^T D^+ U$ where $D^+$ is deduced from $U$ by replacing non-negative entries by zeros.

\noindent{\bf Projection.}
Let $\Omega$ be closed convex, $x_0\in \mR^d$ and $F(x) =\frac12 |x-x_0|^2$. We have 
\[
\min_\Omega F = \min_{\Omega\cap \bar B(0, R)} F
\]
for large enough $R$ (e.g., larger than $F(x)$ for any fixed point in $\Omega$), and since the latter minimization is over a compact set, $\argmin_\Omega F$ is not empty. The function $F$ being strongly convex, its minimizer over $\Omega$ is unique and called the projection of $x_0$ on $\Omega$, denoted $\proj_\Omega(x_0)$. 

Since $\nabla F(x) = x-x_0$, \cref{th:optimality.convex.constraints} implies that $\proj_\Omega(x_0)$ is characterized by $\proj_{\Omega}(x_0) \in \Omega$ and
\begin{equation}
\label{eq:projection.1}
x_0 - \proj_\Omega(x_0)  \in \CN_\Omega(\proj_\Omega(x_0))
\end{equation}
or
\begin{equation}
\label{eq:projection.2}
(x_0 - \proj_\Omega(x_0) )^T(y-  \proj_\Omega(x_0)) \leq 0 \text{ for all } y\in \Omega.
\end{equation}
If $x_0\not\in \Omega$, then $\proj_\Omega(x_0) \in \prt\Omega$, since otherwise we would have $\CN_\Omega(\proj_\Omega(x_0)) = \{0\}$ and $x_0 = \proj_\Omega(x_0)$, a contradiction. Of course, if $x_0\in \Omega$, then $\proj_\Omega(x_0) = x_0$.

Here are some important examples.
\begin{enumerate}[label={\bf\arabic*.}, wide]
\item
Let $\Omega = z_0 + V$, where $z_0\in \mR^d$ and $V$ is a linear space (i.e., $\Omega$ is an affine subset of $\mR^d$). Then $N_\Omega(x) = z_0 + V^\perp = x + V^\perp$ for all $x\in \Omega$, where $V^\perp$ is the vector space of vectors orthogonal to $V$, and $\proj_\Omega(x_0)$ is characterized by $\proj_\Omega(x_0)\in \Omega$ and
\[
(x_0 - \proj_\Omega(x_0) ) \in V^\perp
\]
which is the usual characterization of the orthogonal projection on an affine space (compare to \cref{sec:orth.proj}).
\item
If $\Omega = \bar B(0,1)$, the closed unit sphere, then $N_\Omega(x) = \mR^+ x$ for $x\in \prt\Omega$ (i.e., $|x|=1$).  One can indeed note that, if $h\neq 0$ in normal to $\Omega$ at $x$, then $h/|h| \in \Omega$ so that
\[
h^T\left(\frac{h}{|h|} - x\right) \leq 0
\]
which yields $|h| \leq h^T x$. The Cauchy-Schwartz inequality implying that $h^T x \leq |h|\,|x|=|h|$, we must have equality, $h^T x = |h|\, |x|$, which is only possible when $x$ and $h$ are collinear.

Given $x_0\in \mR^d$ with $x_0 \geq 1$, we see that $\proj_{\Omega}(x_0)$ must satisfy the conditions $|\proj_\Omega(x_0)|  =  1$ (to be in $\prt\Omega$) and $x_0 - \proj_\Omega(x_0) = \lambda x_0$ for some $\lambda \geq 0$, which gives $\proj_{\Omega}(x_0) = x_0/|x_0|$.  

\item If $\Omega = \CS^+_n$ and $B$ (taking the role of $x_0$) is a symmetric matrix, then 
$\proj_\Omega(B)$ was found in the previous section, and is given by $A = U^T D^+ U$ where $U^TDU$ provides  a diagonalization of $B$.
 
\end{enumerate}

The projection has the important property of being 1-Lipschitz.
\begin{proposition}
\label{prop:proj.lip}
Let $\Omega$ be a closed convex subset of $\mR^d$. Then, for all $x,y\in \mR^d$
\begin{equation}
\label{eq:proj.lip}
|\proj_\Omega(x) - \proj_\Omega(y)| \leq |x-y|.
\end{equation}
\end{proposition}
\begin{proof}
This proposition is a special case of \cref{prop:prox.lip} below.
\end{proof}


\subsection{Projected gradient descent}
\label{sec:proj.grad}
The projected gradient descent algorithm minimizes $F$ over $\Omega$ by iterating
\begin{equation}
\label{eq:proj.grad}
x_{t+1} = \proj_\Omega(x_t - \alpha_t \nabla F(x_t)),
\end{equation}
which provides a feasible method when $\proj_\Omega$ is easy to compute. An equivalent formulation is
\begin{equation}
\label{eq:proj.grad.2}
x_{t+1} = \argmin_\Omega F(x_t) + \nabla F(x_t)^T (x-x_t) + \frac1{2\alpha_t} |x-x_t|^2.
\end{equation}
To justify this last statement it suffices to notice that the function in the r.h.s. can be written as
\[
\frac{1}{2\alpha_t} |x - x_t + \alpha_t \nabla F(x_t)|^2 - \frac{\alpha_t}{2} |\nabla F(x_t)|^2 + F(x_t)
\]
and apply the definition of the projection. 

The convergence properties of this algorithm will be discussed in \cref{sec:proximal}, in a more general context.

\section{General convex problems}

\subsection{Epigraphs}
\begin{definition}
\label{def:epi}
Let $F$ be a convex function. The  epigraph of $F$ is the set 
\begin{equation}
\epi(F) = \defset{(x,a)\in \mR^d \times \mR: F(x)\leq a}\,.
\label{eq:epi}
\end{equation}
One says that $F$ is closed if $\epi(F)$
is a closed subset of $\mR^d\times \mR$, that is: if $x = \lim_n x_n$ and $a = \lim_n a_n$ with $F(x_n) \leq a_n$, then $F(x) \leq a$.  
\end{definition}
Clearly, if $(x,a)\in \epi(F)$, then $x\in \dom(F)$. It should also be clear that $\epi(F)$ is always convex when $F$ is convex: If $(x,a), (y,b) \in \epi(F)$, then 
\[
F((1-t) x + t y) \leq (1-t) F(x) + t F(y) \leq (1-t)a + tb
\]
so that $(1-t)(x,a) + t(y,b) \in \epi(F)$. 

%
%
%

To illustrate the definition, consider a simple example.
Let $F$ be the function defined on $\mR$ by $F(x) = |x|$ if $|x| < 1$ and $F(x) = +\infty$ otherwise. It is convex, but not closed, as can be seen by taking the sequence $(1-1/n,1)\in \epi(F)$, with, at the limit, $F(1) = +\infty > 1$. In contrast, the function defined by  $\tilde F(x) = |x|$ if $|x| \leq 1$ and $\tilde F(x) = +\infty$ otherwise is convex and closed.

We have the following proposition. 
\begin{proposition}
\label{prop:closed}
A convex function $F$ is closed if and only if all its sub-level sets 
\[
\Lambda_a(F) = \defset{x\in \mR^d: F(x) \leq a}
\]
are closed subsets of $\mR^d$. 
\end{proposition}
\begin{proof}
If $F$ is closed, then $\Lambda_a(F)$ is the intersection of the set $\{(x,a): x\in \mR^d\}$, which is obviously closed, and of $\epi(F)$. It is therefore a closed set. 

Conversely, assume that all $\Lambda_a(F)$ are closed and  take a sequence $(x_n, a_n)$ in $\epi(F)$ that converges to $(x,a)$. Then, fixing $\epsilon>0$, $x_n \in \Lambda_{a+\epsilon}$ for large enough $n$, and since this set is closed, $F(x) \leq a+\epsilon$. Since this is true for all $\epsilon>0$, we have $F(x)\leq a$ and $(x,a)\in \epi(F)$.
\end{proof}

Note that, if $F$ is continuous, then it is closed, so that closedness generalizes continuity for convex functions, but it also applies to the non-smooth case.

If $\Om$ is a convex subset of $\mR^d$, its indicator function $\sig_\Om$ (such that $\sig_\Om(x)= 0$ for $x\in \Om$ and $\sig_\Om(x) = +\infty$ otherwise)  is closed if and only if $\Om$ is a closed subset of $\mR^d$. This is obvious since $\Lambda_a(\sigma_\Omega) = \Omega$ if $a\geq 0$ and $\emptyset$ otherwise.

\subsection{Subgradients}
Several machine learning problems involve convex functions that are not $C^1$, requiring a generalization of the notion of derivative provided by the following definition.
\begin{definition}
\label{def:subgradient}
If $F$ is a convex function and $x\in \dom(F)$, a vector $g\in \mR^d$ such that 
\begin{equation}
\label{eq:subg}
F(x) + g^T(y-x) \leq F(y) 
\end{equation}
for all $y\in \mR^d$ is called a {\em subgradient} of $F$ at $x$. 

The set of subgradients of $F$ at $x$ is denoted $\prt F(x)$ and called the {\em subdifferential} of $F$ at $x$. 
\end{definition}

If $x\in \mathrm{int}({\dom}(F))$ and $F$ is differentiable at $x$, \cref{eq:conv.3} implies that $\nabla F(x) \in \prt F(x)$.  This is in this case the only element of $\prt F(x)$.

\begin{proposition}
\label{prop:subg.diff}
If $F$ is differentiable at $x\in \mathrm{int}({\dom}(F))$, then  $\prt F(x) = \{\nabla F(x)\}$.
\end{proposition}
\begin{proof}
We need to prove that there is no other subgradient.
Assume that $\nabla F(x)$ exists and take 
$y = x + \ep u$ in \cref{eq:subg}  ($u\in \mR^d$). One gets, for $g\in \prt F(x)$, 
\[
\ep g^Tu \leq F(x+\ep u) - F(x) = \ep \nabla F(x)^Tu + o(\ep)
\]
Dividing by $|\ep|$ and letting $\ep\to 0$ gives (depending on the sign of $\ep$)
\[
 g^Tu \leq  \nabla F(x)^T u\ \text{\color{black}  and } - g^Tu \leq - \nabla F(x)^T u
\]
This is only possible if $g^Tu = \nabla F(x)^T u$ for all $u\in\mR^d$, which itself implies $g=\nabla F$. 
\end{proof}

The next theorem, which is an obvious consequence of \cref{def:subgradient},  characterizes minimizers of convex functions in the general case.
\begin{theorem}
\label{th:subg}
Let $F: \mR^d\to \mR$ be convex. Then $x$ is a (global) minimizer of $F$ if and only if $0\in \prt F(x)$.
\end{theorem}

The following result shows that subgradients exist under generic conditions. We note that $g\in \prt F(x)$ if and only if $\proj_{\vaff(\dom(F))}(g) \in \prt F$, because \cref{eq:subg} is trivial if $F(y) = +\infty$. So $\prt F$ cannot be bounded unless $\mathrm{aff}(\dom(D)) = \mR^d$. However, it is the part of this set that is included in  the $\vaff(\dom(F))$ that is of interest.
\begin{proposition}
\label{prop:subg.exist}
For all $x\in \mR^d$, $\prt F(x)$ is a closed convex set (possibly empty, in particular for $x\not\in \dom(F)$).
If $x\in\rdom(F)$, then $\prt F(x) \neq \emp$ and $\prt F(x) \cap \vaff(\dom(F))$ is compact.
\end{proposition}
\begin{proof}
The convexity and closedness of $\prt F(x)$ is clear from the definition.  If $x\in \rdom(F)$, there exists $\epsilon>0$ such that $x+\epsilon h\in \rdom(F)$ for all $h\in \vaff(\dom(F))$ with $|h|=1$. For all $g \in \prt F(x)\cap \vaff(\dom(F))$, one has
\begin{multline*}
|g| = \max\{g^T h: h\in \vaff(\dom(F)), |h|=1\}\\
 \leq \max((F(x+\epsilon h) - F(x))/\epsilon: h\in \vaff(\dom(F)), |h|=1)
\end{multline*}
and the upper bound is finite because it is the maximum of a continuous function over a bounded set. This shows that $\prt F(x)$ is bounded. We defer the proof that $\prt F(x)\neq \emptyset$ to \cref{sec:separation}.
\end{proof}

Subdifferentials are, under mild conditions, additive. More precisely, we have the following proposition.
\begin{theorem}
\label{th:subg.sum}
Let $F_1$ and $F_2$ be convex functions such that 
\[
\rdom(F_1) \cap \rdom(F_2)\neq \emp\,.
\]
Then,
for all $x\in \mR^d$,  $\prt (F_1+F_2)(x) = \prt F_1(x) + \prt F_2(x)$.
\end{theorem}
Note that the inclusion
\[
\prt F_1(x) + \prt F_2(x) \subset \prt (F_1+F_2)(x)
\]
as can be immediately checked by summing the inequalities satisfied by subgradients. The reverse inclusion requires the use of separation theorems for convex sets (see \cref{sec:separation}).

Another important point is how the chain rule works with compositions with affine functions.
\begin{theorem}
\label{th:subg.affine}
Let $F$ be a convex function on $\mR^d$, $A$ a $d\times m$ matrix and $b\in\mR^d$. Let $G(x) = F(Ax+b)$, $x\in \mR^m$. Assume that there exists $x_0\in \mR^m$ such that  $Ax_0\in \rdom(F)$. Then, for all $x\in \mR^m$, 
\[
\prt G(x) = A^T\prt F(Ax+b).
\]
\end{theorem}
One direction is straightforward and does not require the condition on $\rdom(F)$. If $g \in \prt F(Ax+b)$, then
\[
F(z) - F(Ax+b) \geq g^T (z - Ax -b), z\in \mR^d
\]
and applying this inequality to $z = Ay+b$ for $y\in \mR^m$ yields
\[
G(y) - G(x) \geq g^T A (y-x)
\]
so that $A^Tg\in \prt G$ and   $A^T \prt F \subset \prt G$. The reverse inclusion is proved in \cref{sec:separation}.

Subdifferentials can be seen as generalizations of normal cones. 

\begin{proposition}
\label{prop:subg.cone}
Assume that $\Omega$ is a closed convex subset of $\mR^d$. Then $\sigma_\Omega$ (the indicator function of $\Omega$) has a subdifferential  everywhere on $\Omega$ with 
\[
\prt \sigma_\Omega (x) = \CN_\Omega(x),\ x\in \Omega
\]
\end{proposition}
\begin{proof}
For $x\in \Omega$, \cref{eq:subg} is
\[
g^T(y-x) \leq \sigma_\Omega(y)
\]
for $y\in \mR^d$, but since $\sigma_\Omega(y) = +\infty$ outside of $\Omega$, $g\in \prt \sigma_\Omega(x)$ is equivalent to 
\[
g^T(y-x) \leq 0
\]
for $y\in \Omega$, which is exactly the definition of the normal cone.
\end{proof}
Given this proposition, it is also clear (after noting that $\sigma_{\Omega_1} + \sigma_{\Omega_2} = \sigma_{\Omega_1\cap\Omega_2}$) that \cref{th:subg.sum} is a generalization of \cref{th:normal.cone.intersect}. 

\subsection{Directional derivatives}
From \cref{prop:conv.2}, applied with $y = x+h$, we see that 
\[
t \mapsto \frac{1}{t} (F(x+th) - F(x))
\]
is increasing as a function of $t$. This property allows us to define directional derivatives of $F$ at $x$. 
\begin{definition}
\label{def:directional}
Let $F$ be convex and $x\in \dom(F)$. The directional derivative of $F$ at $x$ in the direction $h\in \mR^d$ is defined by
\begin{equation}
\label{eq:directional}
dF(x, h) = \lim_{t\downarrow 0} \frac{1}{t} (F(x+th) - F(x)),
\end{equation}
and belong to $[-\infty, +\infty]$.
\end{definition}
Note that, still from \cref{prop:conv.2}, one has, for all $x\in \dom(F)$ and $y\in \mR^d$:
\begin{equation}
\label{eq:directional.sub}
F(y) \geq F(x) + dF(x, y-x)
\end{equation}

We have the proposition:
\begin{proposition}
\label{prop:directional.minimum}
If $F$ is convex, then $x^* \in \argmin(F)$ if and only if $dF(x^*,h) \geq 0$ for all $h\in \mR^d$.
\end{proposition}
\begin{proof}
If $dF(x^*,h) \geq 0$, then $F(x^*+th) - F(x^*) \geq 0$ for all $t>0$ and this being true for all $h$ implies that $x^*$ is a minimizer. Conversely, if $x^*$ is a minimizer, $dF(x^*, h)$ is a limit of non-negative numbers and is therefore non-negative. 
\end{proof}

\begin{proposition}
\label{prop:directional.convex}
If $F$ is convex and $x\in\dom(F)$, then $dF(x, h)$ is positively homogeneous and subadditive (hence convex) as a function of $h$, namely
\[
dF(x, \lambda h)  = \lambda dF(x, h), \lambda>0
\]
and
\[
dF(x, h_1+h_2) \leq dF(x, h_1) + dF(x, h_2).
\]
\end{proposition}
\begin{proof}
Positive homogeneity is straightforward and left to the reader. For the second one, we can write
\[
F(x+th_1 + th_2) \leq \frac12 (F(x+th_1/2) + F(x+th_2/2))
\]
by convexity so that
\[
\frac1t(F(x+th_1 + th_2)-F(x)) \leq \frac12 \left(\frac1t (F(x+th_1/2)-F(x)) + \frac1t(F(x+th_2/2) - F(x))\right).
\]
Taking $t\downarrow 0$, 
\[
dF(x;h_1+h_2) \leq \frac12 (dF(x; h_1/2) + dF(x, h_2/2)) = dF(x, h_1) + dF(x, h_2).
\]
\end{proof}

\begin{proposition}
\label{prop:directional.subg}
If $F$ is convex and $x\in \dom(F)$, then
\[
dF(x, h) \geq \sup\{ g^T h, g\in \prt F(x)\}.
\]
If $x\in \rdom(F)$, then
\[
dF(x, h) = \max\{ g^T h, g\in \prt F(x)\}.
\]
\end{proposition}
\begin{proof}
If $g\in \prt F(x)$, then for all $t>0$
\[
F(x + th) - F(x) \geq t g^Th.
\]
Dividing by $t$ and passing to the limit yields $dF(x,h) \geq g^T h$. 

We prove that the maximum is attained at some $g \in \prt F(x)$ when $x\in \rdom(F)$. In this case,  the domain of the convex function  $G: \tilde h \mapsto dF(x,\tilde h)$  is the vector space parallel to $\mathrm{aff}(\dom(F))$, namely
\[
\dom(G) = \{h: x+h \in \mathrm{aff}(\dom(F))\}.
\]
Indeed, for any $h$ in this set, there exists $\epsilon>0$ such that $x+th \in \dom(F)$ for $0<t<\epsilon$ and $dF(x, h) \leq (F(x+th) - F(x))/t < \infty$. Conversely, if $h\in \dom(G)$, then $F(x+th) - F(x)$ must be finite for small enough $t$, so that $x+th\in \dom(F)$ and $x+h \in \mathrm{aff}(\dom(F))$.

As a consequence, for any $h\in\mathrm{aff}(\dom(F))$,  there exists $\hat g \in  \prt G(h)$, which therefore satisfies
\[
dF(x, \tilde h) \geq dF(x, h) + \hat g^T(\tilde h - h)
\]
for any   $\tilde h\in \mR^d$ (the upper bound is infinite if $\tilde h\not\in \dom(G)$). Letting $\tilde h\to 0$, we get $dF(x,h) \leq \hat g^T h$. 

Also, by positive homogeneity, we have
\[
tdF(x, \tilde h) \geq dF(x, h) + \hat g^T(t\tilde h - h)
\]
for all $t>0$, which requires $dF(x,\tilde h) \geq \hat g^T \tilde h$ for all $\tilde h$, and in particular $dF(x,h)  = \hat g^T h$. 

Since 
\[
F(x+\tilde h) - F(x) \geq dF(x, \tilde h) \geq \hat g^T \tilde h
\]
we see that $\hat g \in \prt F(x)$, with $\hat g^T h = dF(x,h)$, which concludes the proof. 
\end{proof}

The next proposition gives a criterion for a vector $g$ to belong to $\prt F(x)$ based on directional derivatives.

\begin{proposition}
\label{prop:subg.directional}
Assume that $x\in \dom(F)$ where $F$ is convex. If $g\in \mR^d$ is such that 
\[
dF(x,h) \geq g^T h
\]
for all $h\in \mR^d$, then $g\in \prt F(x)$.
\end{proposition}
\begin{proof}
Just use the fact that $dF(x, h) \leq F(x+h) - F(x)$.
\end{proof}

\subsection{Subgradient descent}
\label{sec:subg.descent}
When $F$ is a non-differentiable a convex function, directions $g$ such that $-g\in \prt F(x)$ do not always provide directions of descent. Indeed, $g\in\prt F(x)$ implies
\[
F(x-\al g) \geq F(x) - \alpha |g|^2
\]
but the inequality goes in the ``wrong direction.''  However, we know that, for any $h\in \mR^d$, there exists $g_h\in \prt F(x)$ such that 
\[
dF(x, -h) = -g_h^T h \geq -g^T h
\]
for all $g \in \prt F(x)$. As a consequence, any non-vanishing solution of the equation $h = g_h$ will provide a direction of descent. This suggests looking for $h\in \prt F(x)$ such that $h\neq 0$ and $|h|^2 \leq g^Th$ for all $g\in \prt F(x)$. Since $g^T h \leq |g|\, |h|$, this requires that $|h| \leq |g|$ for all $g\in \prt F(x)$, i.e., 
\begin{equation}
\label{eq:argmin.subg}
h = \argmin_{\prt F(x)} (g\mapsto |g|).
\end{equation}
Conversely, if $h$ is the minimal-norm element of $\prt F(x)$ (which is necessarily unique since the norm is strictly convex and $\prt F(x)$ is convex and compact), then $|h|^2 \leq |h + t(g-h)|^2$ for all $g\in \prt F(x)$ and $t\in [0,1]$, and taking the difference yields
\[
2t h^T(g-h) + t^2 |g-h|^2 \geq 0.
\]
The fact that this holds for all $t\geq 0$ requires that $h^T(g-h) \geq 0$ as required. We have therefore proved that $h$ defined by \cref{eq:argmin.subg} is a descent direction for $F$ at $x$ (it is actually the steepest descent direction: see \citep{wright2022optimization} for a proof), justifying the algorithm
\[
x_{t+1} = x_t - \alpha_t \argmin_{\prt F(x)} (g\mapsto |g|)
\]
as subgradient descent iterations. 

\noindent{\bf Example.}
Consider the minimization of
\[
F(x) = \psi(x) + \lambda \sum_{i=1}^n |\pe{x}{i}|
\]
where $\psi$ is a $C^1$ convex function on $\mR^d$.
Let $\CA(x) = \{i: \pe{x}{i} = 0\}$. Then
\[
\prt F(x) = \left\{\nabla \psi(x) + \lambda  \sum_{i\not\in \CA(x)}\sign(\pe x i) + \lambda \sum_{i\in \CA(x)} \rho_i \mathfrak e_i: |\rho_i|\leq 1, i\in \CA(x)\right\}
\]
where $\mathfrak e_i$ is the $i$th vector of the canonical basis of $\mR^d$.

For $g = \nabla \psi(x) +  \lambda  \sum_{i\not\in \CA(x)}\sign(\pe x i) + \lambda \sum_{i\in \CA(x)} \rho_i \mathfrak e_i$, we have
\[
|g|^2 = \sum_{i\not\in \CA(x)} ( \prt_i F(x)+\lambda \sign(\pe x i))^2 + \sum_{i\in \CA(x)} (\prt_i \psi(x) - \lambda\rho_i)^2.
\]
Define 
\[
s(t) = \sign(t) \min(|t|, 1).
\]
Then $h$ satisfying \cref{eq:argmin.subg} is given by
\[
\pe{h}{i} = \left\{
\begin{aligned}
\prt_i \psi(x) - \lambda \sign(\pe x i) & \text{ if } i\not\in \CA(x) \\
\lambda \  s(\prt_i \psi(x)/\lambda) & \text{ if } i\in \CA(x) .
\end{aligned}
\right.
\]

In more complex situations, the extra minimization step at each iteration of the algorithm can be challenging computationally. The following subgradient method uses an averaging approach to minimize $F$ without requiring finding subgradients with minimal norms. It simply defines
\[
x_{t+1} = x_t  - \alpha_t g_t, \quad g_t \in \prt F(x_t)
\]
and computes
\[
\bar x_{t} = \frac{\sum_{j=1}^t \alpha_j x_j}{\sum_{j=1}^t\alpha_j}\,.
\]
We refer to \citep{wright2022optimization} for a proof of convergence of this method.

\subsection{Proximal Methods}
\label{sec:proximal}

\noindent{\bf Proximal operator.}
We start with a few simple facts. Let $F$ be a closed convex function and $\psi$ be convex and differentiable, with $\dom(\psi) = \mR^d$. Let $G = F + \psi$. Then $G$ is a closed convex function. Indeed, consider the sub-level set $\Lambda_a(G) = \{x: G(x) \leq a\}$ and assume that $x_n\to x$ with $x_n \in \Lambda_a(g)$. Then $\psi(x_n) \to \psi(x)$ by continuity, and for all $\epsilon > 0$, we have, for large enough $n$, $F(x_n) \leq a - \psi(x) + \epsilon$. This inequality remains true at the limit because $F$ is closed, yielding $G(x) \leq a+\epsilon$ for all $\epsilon>0$, so that $x\in \Lambda_a(G)$.

We have $\rdom(F) \cap \rdom(\psi) \neq \emptyset$ so that (by \cref{th:subg.sum} and \cref{prop:subg.diff}) $\prt G(x) = \nabla \psi(x) + \prt F(x)$. In particular, $x^*$ is a minimizer of $G$ if and only if $-\nabla \psi(x^*) \in \prt F(x^*)$. 

It one assumes that $\psi$ is strongly convex, so that there exists $m$ and $L$ such that
\[
\frac{m}{2} |y-x|^2 \leq \psi(y) - \psi(x) - \nabla \psi(x)^T(y-x) \leq \frac{L}{2} |y-x|^2
\]
for all $x,y\in \mR^d$, then a minimizer of $G$ exists and is unique. To see this, fix $x_0\in \rdom(F)$ and consider the closed convex set
\[
\Omega_0 = \Lambda_{G(x_0)}(G) = \{x: G(x) \leq G(x_0)\}.
\]
Any minimizer of $G$ must clearly belong to $\Omega_0$. If $x\in \Omega_0$, we have
\[
F(x) + \psi(x_0) + \nabla \psi(x_0)^T(x-x_0) +  \frac{m}{2} |x-x_0|^2 \leq G(x) \leq G(x_0)\,.
\]
Moreover, there exists (from \cref{prop:subg.exist}) an element $g\in\prt F(x_0)$ so that 
$F(x) \geq F(x_0) + g^T (x-x_0)$ for all $x\in \mR^d$. We therefore get
\[
F(x_0) + \psi(x_0) + (g+\nabla \psi(x_0))^T(x-x_0) +  \frac{m}{2} |x-x_0|^2  \leq G(x_0)\,.
\]
for all $x\in \Omega_0$, which shows that $\Omega_0$ must be bounded and therefore compact. There exists a minimizer $x^*$ of $G$ on $\Omega_0$, and therefore on all $\mR^d$. This minimizer is unique, since the sum of a convex function and a strictly convex function is strictly convex.

In particular, for any closed convex $F$, we can apply the previous remarks to 
\[
G:  v \mapsto F(v) + \frac1{2} |x-v|^2
\]
where  $x\in \mR^d$ is fixed. The function $\psi: v\mapsto |v-x|^2/2$ is strongly convex (with $L=m = 1$) and $G$ therefore has a unique minimizer $v^*$. This is summarized in the following definition.
\begin{definition}
\label{def:proximal}
Let $F$ be a closed convex function. The proximal operator associated to $F$  is the mapping $\prox_{F} : \mR^d \to \dom(F)$ defined by
\begin{equation}
\label{eq:proximal}
\prox_{F}(x) = \argmin_{\mR^d} (v\mapsto  F(v) + \frac1{2} |x-v|^2).
\end{equation}
\end{definition}

From the previous discussion, we also deduce
\begin{proposition}
\label{prop:proximal.subg}
Let $F$ be a closed convex function and $\alpha>0$.
We have $x' = \prox_{\alpha F}(x)$ if and only if $x\in x'+\alpha \prt F(x')$. In particular, $x^*$ is a minimizer of $F$ if and only if $x^* = \prox_{\alpha F}(x^*)$ 
\end{proposition}

Let us take a few examples.
\begin{itemize}
\item Let $F(x) = \lambda |x|$, $x\in \mR^d$, for some $\lambda>0$. Then $F$ is differentiable everywhere except at $x=0$ and $\dom(F) = \mR^d$. We have $\prt F(x) = \lambda x/|x|$ for $x\neq 0$. A vector $g$ belongs to $\prt F(0)$ if and only if
\[
g^T x \leq \lambda |x|
\]
for all $x\in \mR^d$, which is equivalent to  $|g|\leq \lambda$ so that $\prt F(0) = \bar B(0, \lambda)$. 

We have $x' = \prox_F(x)$ if and only if $x'\neq 0$ and $x = x' + \lambda x'/|x'|$ or $x'=0$ and $|x|\leq \lambda$. For $|x| > \lambda$, the equation $x = x' + \lambda x'/|x'|$ is solved by
\[
x' = \frac{|x| - \lambda}{|x|} x
\]
yielding
\begin{equation}
\label{eq:proximal.norm}
\prox_F(x) = \left\{
\begin{aligned}
\frac{|x| - \lambda}{|x|} x &\text { if } |x| \geq \lambda\\
0 & \text{ otherwise}
\end{aligned}
\right.
\end{equation}

\item Let $\Omega$ be a closed convex set. Then
$\prox_{\sigma_\Omega} = \proj_\Omega$, the projection operator on $\Omega$, as directly deduced from the definition.  
\end{itemize}
The following proposition can then be compared to \cref{prop:proj.lip}.
\begin{proposition}
\label{prop:prox.lip}
Let $F$ be a closed convex function. Then $\prox_F$ is  1-Lipschitz: for all $x, y\in \mR^d$, 
\begin{equation}
\label{eq:prox.lip}
|\prox_F(x) - \prox_F(y)| \leq |x-y|.
\end{equation}
 \end{proposition}
 \begin{proof}
 Let $x' = \prox_F(x)$ and $y' = \prox_F(y)$. Then, there exists $g\in \prt F(x')$ and $h\in \prt_F(y')$ such that $x = x' + g$ and $y = y' + h$. Moreover,  we have
 \begin{align*}
 F(y') - F(x') &\geq  g^T(y'-x')\\
 F(x') - F(y') &\geq  h^T(x'-y')\\  
 \end{align*}
 from which we deduce $g^T(y'-x') \leq h^T(y'-x')$ or $(h-g)^T(x'-y') \geq 0$. Expressing $g,h$ in terms or $x,x',y,y'$, we get
 $(y - x - y' + x')^T(y' - x') \geq 0$ or
 \[
 |y'-x'|^2 \leq (y-x)^T(y'-x') \leq |y-x|\, |y'-x'|
 \]
 which is only possible if $|y'-x'| \leq [y-x|$.
  \end{proof}

If $F$ is differentiable, then 
$x' = \prox_{\alpha F}(x)$ satisfies 
\[
x' = x - \alpha \nabla F(x')
\]
so that $x \mapsto \prox_{\alpha F}(x)$ can be interpreted as an implicit version of the standard gradient step  $x \mapsto x - \alpha \nabla F(x)$. The iterations $x(t+1) = \prox_{\alpha_t F} (x(t))$ provide an algorithm that converges to a minimizer of $F$ (this will be justified below). This algorithm is rarely practical, however, since the minimization required at each step is not necessarily much easier to perform than minimizing $F$ itself. The proximal operator, however, is especially useful when  combined with splitting methods.

\bigskip

\noindent{\bf Proximal gradient descent.}
Assume that the objective function $F$ takes the form
\begin{equation}
\label{eq:split.1}
F(x) = G(x) + H(x)
\end{equation}
where $G$ is $C^1$ on $\mR^d$ and $H$ is a closed convex function. We first note that 
\[
dF(x, h) = \lim_{t\downarrow 0} \frac{F(x+th)-F(x)}{t}
\]
is well defined (even if $G$ is not convex, because it is smooth), with
\[
dF(x,h) = \nabla G(x)^T h + dH(x,h)
\]
In particular, if  $x^*$ be a minimizer of $F$, then $dF(x,h) \geq 0$ for all $h$ so that $dH(x,h) \geq -\nabla G(x)^T h$ for all $h$. Using \cref{prop:subg.directional}, this shows that $-\nabla G(x) \in \prt H(x)$, which is a necessary condition for optimality for $F$ (which is sufficient if $G$ is convex).

Proximal gradient descent implements the algorithm
\begin{equation}
\label{eq:prox.grad}
x_{t+1} = \prox_{\al_t H} (x_t - \al_t \nabla G(x_t)).
\end{equation}
We note that a stationary point of this algorithm, i.e. a point $x$ such that $x = \prox_{\al_t H} (x - \al_t \nabla G(x))$ must be such that $x-\alpha_t \nabla G(x) \in x + \alpha_t \prt H(x)$, so that $-\nabla G(x) \in \prt H(x)$. This shows that the property of being stationary does not depend on $\alpha_t >0$, and is equivalent to the necessary optimality condition that was just discussed.

We first study this algorithm under the assumption that $G$ is $L$-$C^1$, which implies that, for all $x,y\in \mR^d$. 
\[
G(y) \leq G(x) + \nabla G(x)^T(y-x) + \frac{L}{2} |x-y|^2.
\]
At iteration $t$, we have
\[
x_{t} - \alpha_t \nabla G(x_t) \in x_{t+1} + \alpha_t \prt H(x_{t+1})
\]
which implies, in particular
\[
\alpha_t H(x_t) - \alpha_t H(x_{t+1}) \geq (x_t - x_{t+1} )^T (x_{t} - x_{t+1} - \alpha_t \nabla G(x_t)) = |x_t - x_{t+1}|^2  + \alpha_t \nabla G(x_t)^T(x_{t+1} - x_t)
\]
Dividing by $\alpha_t$ and adding $G(x_t) - G(x_{t+1})$, we get
\begin{align}
\nonumber
F(x_t) - F(x_{t+1}) &\geq \frac{1}{\alpha_t} |x_t - x_{t+1}|^2 + G(x_t) + \nabla G(x_t)^T(x_{t+1} - x_t) - G(x_{t+1})\\
\label{eq:prox.grad.conv.1}
& \geq \left(\frac{1}{\alpha_t} - \frac{L}{2}\right) |x_t - x_{t+1}|^2
\end{align}
so that proximal gradient descent iterations reduce the objective function as soon as $\alpha_t \leq 2/L$. 

%
Assuming that $\alpha_t < 2/L$, \cref{eq:prox.grad.conv.1} can be rewritten as
\[
 \left |\frac{x_{t+1}-x_t}{\alpha_t}\right|^2 \leq \frac2{\alpha_t(2-\alpha_t L)} (F(x_{t}) - F(x_{t+1}).
\]
This inequality should be compared to \cref{eq:grad.desc} in the unconstrained case. It yields, in particular, the inequality
\begin{equation}
\label{eq:pg.convergence.1}
\min\left\{ \left |\frac{x_{t+1}-x_t}{\alpha_t}\right|: t\leq T\right\} \leq \frac{F(x_0) - \min F}{2T \min \{\alpha_t(2-\alpha_t L), t\leq T\}}\,.
\end{equation}
As a consequence, if one runs proximal gradient descent until $|x_{t+1}-x_t|/\alpha_t$ is small enough, the algorithm will terminate in finite time as soon as $\alpha_t$ is bounded from below (and, in particular, if $\alpha_t$ is constant).

%
%
%
%
%

If we assume that $G$ is convex, in addition to being $L$-$C^1$, then we have a stronger result.
Let $x^*$ be a minimizer of $F$. Then, using again $x_{t} - \alpha_t \nabla G(x_t) \in x_{t+1} + \alpha_t \prt H(x_{t+1})$, we have
\[
\alpha_t H(x^*) - \alpha_t H(x_{t+1}) \geq (x^* - x_{t+1} )^T (x_{t} - x_{t+1} - \alpha_t \nabla G(x_t)) 
\]
and
\begin{align*}
\alpha_t F(x^*) - \alpha_t F(x_{t+1}) &\geq (x^* - x_{t+1} )^T (x_{t} - x_{t+1}) - \alpha_t (x^* - x_{t+1} )^T \nabla G(x_t)) + \alpha_t G(x^*) - \alpha_t G(x_{t+1})   
\\
&\geq (x^* - x_{t+1} )^T (x_{t} - x_{t+1}) - \alpha_t (x^* - x_{t} )^T \nabla G(x_t)) + \alpha_t G(x^*)\\
& + \alpha_t (x_{t+1}-x_t)^T \nabla G(x_{t}) - \alpha_t G(x_{t+1})\\
& \geq   (x^* - x_{t+1} )^T (x_{t} - x_{t+1}) - \frac{\alpha_tL}{2} |x_t - x_{t+1}|^2
\end{align*}
Assuming that $\alpha_t L \leq 1$, then
\[
\alpha_t F(x^*) - \alpha_t F(x_{t+1}) \geq (x^* - x_{t+1} )^T (x_{t} - x_{t+1}) - \frac{1}{2} |x_t - x_{t+1}|^2 = \frac12 (|x_{t+1} - x^*|^2 - |x_t - x^*|^2), 
\]
which we rewrite as
\[
\alpha_t (F(x_{t+1}) - F(x^*)) \leq \frac12 (|x_{t} - x^*|^2 - |x_{t+1} - x^*|^2)
\]
Note that, from \cref{eq:prox.grad.conv.1}, we also have
\[
F(x_{t+1}) \leq F(x_t) - \frac{1}{2\alpha_t}|x_t -x_{t+1}|^2
\]
when $\alpha_t L \leq 1$, which shows, in particular that $F(x_t)$ is decreasing. Fixing a time $T$, we have, from these two observations
\[
\alpha_t (F(x_{T}) - F(x^*)) \leq \frac12 (|x_{t} - x^*|^2 - |x_{t+1} - x^*|^2)
\]
for all $t \leq T-1$, and summing over $T$, 
\[
(F(x_T) - F(x_*)) \sum_{t=0}^{T-1} \alpha_t \leq \frac12 (|x_{0} - x^*|^2 - |x_{T} - x^*|^2)
\]
yielding
\begin{equation}
\label{eq:prox.conv.2.0}
F(x_T) - F(x_*) \leq \frac{|x_0 -x^*|^2}{2\sum_{t=0}^{T-1} \alpha_t}.
\end{equation}
We summarize this in the following theorem, specializing to the case of constant step $\alpha_t$.
\begin{theorem}
\label{th:proximal.convergence}
Let $G$ be am $L$-$C^1$ function defined on $\mR^d$ and $H$ be closed convex. Assume that $F = G+H$ has a minimizer $x^*$. Then the algorithm \cref{eq:prox.grad} run with $\alpha_t = \alpha \leq 1/L$ for all $t$ is such that, for all $T>0$,  
\begin{equation}
\label{eq:prox.conv.2}
F(x_T) - F(x_*) \leq \frac{|x_0 -x^*|^2}{2\alpha T}.
\end{equation}
\end{theorem}

Also, when $G = 0$, $F=H$, we retrieve  the proximal iterations algorithm
\begin{equation}
\label{eq:prox.iterations}
x_{t+1} = \prox_{\alpha F}(x_t),
\end{equation}
and we have just proved that it
converges for any $\alpha > 0$ as soon as $F$ is a closed convex function.

One gets a stronger result under the assumption that $G$ is $C^2$, and is such that the eigenvalues of $\nabla^2 G(x)$ are included in a fixed interval $[m, L]$ for all $x\in \mR^d$ with $m>0$. Such a $G$ is strongly convex, which implies that $F$ has a unique minimizer.
We have
\begin{align*}
|x_{t+1} - x^*| &= \left| \prox_{\al_t H}(x_t - \alpha_t \nabla G(x_t)) - \prox_{\alpha_t H} (x^* - \alpha_t \nabla G(x^*)\right| \\
&\leq \left|x_t - x^* - \alpha_t (\nabla G(x_t)) - \nabla G(x^*))\right|.
\end{align*}

Write
\begin{align*}
\left|x_t - x^* - \alpha_t (\nabla G(x_t)) - \nabla G(x^*)\right| &= \left|\int_0^1 (\Id[n] - \alpha_t \nabla^2 G(x^* + t(x_t-x^*)))(x_t - x^*) dt\right| \\
&\leq  \int_0^1 \left|(\Id[n] - \alpha_t \nabla^2 G(x^* + t(x_t-x^*)))(x_t - x^*)\right| dt\\
&\leq \max(|1-\alpha_t m|, |1-\alpha_t L|) |x_t - x^*|
\end{align*}
where we have use the fact that the eigenvalues of $\Id[n] - \alpha_t \nabla^2 G(x)$ are included in $[1-\alpha_t L, 1-\alpha_t m]$ for all $x\in \mR^d$. If one assumes that $\alpha_t \leq 1/L$, so that $\max(|1-\alpha_t m|, |1-\alpha_t L|) \leq 1-\alpha_t m$, one gets
\[
|x_{t+1} - x^*| \leq (1-\alpha_t m) |x_t - x^*|\,.
\]
Iterating this inequality, we get the theorem that we state for constant $\alpha_t$.
\begin{theorem}
\label{th:proximal.convergence.strong}
Let $F = G+H$ where $G$ is a $C^2$ convex function and $H$ is a closed convex function. Assume that the eigenvalues of $\nabla^2 G$ are uniformly included  in $[m, L]$ with $m>0$. Let $x^*\argmin F$.

Let $(x_t)$ satisfy \cref{eq:prox.grad} with constant $\alpha_t = \alpha < 1/L$. Then
\[
|x_t - x^*| \leq (1-\alpha m)^t |x_0 - x^*|.
\]

\end{theorem}

Note that these results also apply to projected gradient descent (\cref{sec:proj.grad}), which is a special case (taking $G = \sigma_\Omega$).

\section{Duality}
\label{sec:convex.optim}

\subsection{Generalized KKT conditions}
A constrained convex minimization problem consists in the minimization of a closed convex function $F$ over a closed convex set $\Omega\subset \rdom(F)$. We have seen in \cref{th:optimality.convex.constraints} that, for smooth $F$, any solution $x^*$ of this problem had to satisfy $-\nabla F(x^*) \in \CN_\Omega(x)$ where
\[
\CN_\Omega(x) = \{h: h^T(y-x) \leq 0 \text{ for all } y\in\Om\}\,.
\]

The next theorem generalizes this property to the non-smooth convex case, for which the necessary optimality condition is also sufficient.
\begin{theorem}
\label{th:optimality.convex.constraints.non-smooth}
Let $F$ be a closed convex function, $\Omega\subset \rdom(F)$ a nonempty closed convex set. Then $x^* \in \argmin_\Omega F$ if and only if
\[
0 \in \prt F(x^*) + \CN_\Omega(x^*)
\]
\end{theorem}
\begin{proof}
Introduce the indicator function $\sigma_\Omega$. Then minimizing $F$ over $\Omega$ is the same as minimizing $G = F + \sigma_\Omega$ over $\mR^d$. The assumptions imply that $\rdom(\sigma_\Omega) = \relint(\Omega) \subset \rdom(F)$ and therefore
\[
\prt G(x) = \prt F(x) + \prt \sigma_\Omega(x)
\]
for all $x\in \Omega$. Since
\[
\prt \sigma_\Omega(x) = \CN_\Omega(x)
\]
the result follows for the characterization of minimum of convex functions. 
\end{proof}

In the following, we will restrict to the situation in which $F$ is finite (i.e., $\dom(F) = \mR^d$) and $\Omega$ is defined through a finite number of equalities and inequalities, taking the form
\[
\Om = \defset{x\in \mR^d: \ga_i(x) = 0, i\in \CE \text{ and } \ga_i(x) \leq 0, i\in\CI}
\]
for functions $(\ga_i, i\in \CC = \CE\cup\CI)$ such that  $\ga_i:x \mapsto b_i^Tx +\beta_t$ is affine 
for all $i\in\CE$ and $\ga_i$ is closed convex for all $i\in \CI$. This is similar to the situation considered in \cref{sec:kkt.smooth}, with additional convexity assumptions, but without assuming smoothness. We recall the definition of active constraints from \cref{sec:kkt.smooth}, namely, for $x\in \Omega$, 
\[
\CA(x) = \{i\in \CC: \gamma_i(x) = 0\}.
\]
Following the discussion in the smooth case, define the set $\CN'_\gamma(x)\subset \mR^d$ by
\[
\CN'_\gamma(x) = \left\{\sum_{i\in \CA(x)} \lambda_i s_i: s_i\in \prt \gamma_i(x), i\in \CA(x), \lambda_i \geq 0, i\in \CA(x) \cap \CI\right \}.
\]
The property $0 \in \prt F(x^*) + \CN'_\gamma(x^*)$ is the expression of the KKT conditions in the non-smooth case. It holds for $x^*\in \argmin_\Omega F$ as soon as $\CN_\Omega(x^*) = \CN'_\gamma(x^*)$, which is true under appropriate constraint qualifications. We here replace the MF-CQ in \cref{def:mfcq} by the following conditions that do not involve gradients.
\begin{definition}
\label{def:slater}
Let $(\ga_i, i\in \CC = \CE\cup\CI)$ be a set of equality and inequality constraints, with $\gamma_i: x\mapsto b_i^T x + \beta_i$, $i, \in\CE$ and $\gamma_i$ closed convex, $i\in\CI$. One says that these constraints satisfy the Slater constraint qualifications (Sl-CQ) if and only if:
\begin{enumerate}[label=(Sl\,\arabic*)]
\item The vectors $(b_i, i\in\CE)$ are linearly independent.
\item There exists $x\in \mR^d$ such that $\ga_i(x) = 0$ for $i\in\CE$ and $\ga_i(x) < 0$ for $i\in\CI$.
\end{enumerate}
\end{definition}
The first constraint is a very mild condition. When it is not satisfied, this means that some $b_i$'s are linear combinations of others, and equality constraints for the latter implies equality constraints for the former. These redundancies can therefore be removed without changing the problem.

Note that (Sl2) can be replaced by the apparently weaker condition that, for all $i\in \CI$, there exists $x_i\in \mR^d$ satisfying all the constraints and $\ga_i(x_i) < 0$.  Indeed, if this is true, then the average, $\bar x$, of $(x_i, i\in\CI)$ also satisfies the equality constraints by linearity, and if $i\in \CI$, 
\[
\ga_i(\bar x) \leq \frac1{|\CI|} \sum_{j\in\CI} \ga_{i}(x^{(j)}) \leq \frac1{|\CI|} \ga_{i}(x^{(i)}) < 0.
\]

The following proposition makes a connection between the Slater conditions and the MF-CQ in \cref{def:mfcq}.
\begin{proposition}
\label{prop:mf.sl}
Assume  that $\gamma_i$, $i\in \CI$ are convex $C^1$ functions. Then, if there exists a feasible point $x^*$  that satisfies the MF-CQ, there exists another point $x$ satisfying the Sl-CQ. Conversely, if there exists $x$ satisfying the Sl-CQ, then every feasible point $x^*$ satisfies the MF-CQ.
\end{proposition}
\begin{proof}
The linear independence conditions on equality constraints are the same in MF-CQ and Sl-CQ, so that we only need to consider inequality constraints.

Let $x^*$ satisfy MF-CQ, and take $h\neq 0$ such that $b_i^Th = 0$ for all $i\in\CE$, and $\nabla \gamma_i(x^*)^Th < 0$, $i\in\CA(x)\cap\CI$. Then $x^* + th$ satisfies the equality constraints for all $t\in \mR$. If $i\in \CI$ is not active, then  $\gamma_i(x^*) < 0$ and this will remain true at $x^*+th$ for small $t$ by continuity. If $i\in \CA(x)\cap\CI$, then a first order expansion gives $\gamma_i(x^*+th) = t\nabla \gamma_i(x^*)^T h + o(|h|)$, which is also negative for small enough $t>0$. So, $x^*+th$ satisfies the Sl-CQ for small enough $t>0$. 

Conversely, let $x$ satisfy the Sl-CQ. Take a feasible point $x^*$. If $x^* = x$, then there is no active inequality constraint and $x^*$ satisfies MF-CQ. Assume $x^*\neq x$ and let $h = x-x^*$. Then $b_i^T h = 0$ for all $i\in\CE$, and if $i\in \CI\cap \CA(x^*)$, 
\[
0 > \gamma_i(x) = \gamma_i(x^* + h) \geq \gamma_i(x^*) + \nabla \gamma_i(x^*)^Th = \nabla \gamma_i(x^*)^Th
\]
so that $x^*$ satisfies MF-CQ.
\end{proof}

The following theorem, that we give without proof, states that the Slater conditions implies that the KKT conditions are satisfied for a minimizer.
\begin{theorem}
\label{th:kkt.convex}
Assume that all the constraints are affine, or that they satisfy the Sl-CQ in \cref{def:slater}. Let $x^*\in \argmin_\Omega F$. 
Then $\CN_\Omega(x^*) = \CN'_\gamma(x^*)$, so that there exist $s_0\in \prt F(x^*)$, $s_i\in \prt\gamma_i(x^*)$, $i\in \CA(x^*)$, $(\lambda_i, i\in \CA(x^*))$ with $\lambda_i \geq 0 $ if $i\in \CI\cap \CA(x^*)$, such that
\begin{equation}
\label{eq:kkt.convex}
s_0 + \sum_{i\in\CA(x)} \lambda_i s_i = 0
\end{equation}
\end{theorem} 

\subsection{Dual problem}
Consider the Lagrangian 
\[
L(x, \lambda) = F(x) + \sum_{i\in \CC} \lambda_i \gamma_i(x)
\]
defined in \cref{eq:lag} and let $D = \{\la: \la_i\geq 0, i\in \CI\}$. Because the functions $\ga_i$ are non-positive on $\Om$, we have
\[
L(x, \la) \leq F(x)
\]
for all $x\in \Om$ and $\la\in D$, which implies that
\[
L^*(\la) = \inf \{L(x,\la): x\in \mR^d\}
\]
is such that $L^*(\la) \leq F(x)$ for all $\la\in D$ and $x\in \Om$. Define 
\[
\hat d = \sup \{L^*(\la): \lambda\in D\}
\]
and
\[
\hat p = \inf \{F(x): x\in \Omega\},
\] 
whose computations respectively represent the {\em dual} and {\em primal} problems. Then, we have $\hat d \leq \hat p$.

We did not need much of our assumptions (not even $F$ to be convex) to reach this conclusion. When the converse inequality is true (so that the {\em duality gap} $\hat p - \hat d$ vanishes),  the dual problem provides important insights on the primal problem, as well as alternative ways to solve it. This is true under the Slater conditions.

\begin{theorem}
\label{th:slater}
The duality gap vanishes when the constraints are all affine, or when they satisfy the Sl-CQ in \cref{def:slater}. In this case, any solution $\lambda^*$ of the dual problem provides Lagrange multipliers in \cref{th:kkt.convex} and conversely. 
\end{theorem}

We justify this statement, as a consequence of \cref{th:kkt.convex} and the following analysis. The Lagrangian
$L(x, \lambda)$ is linear in $\lambda$, and when $\lambda \in D$, is a convex function of $x$. Moreover, one can use subdifferential calculus (\cref{th:subg.sum}) to conclude that, for any $\lambda\in D$, \cref{eq:kkt.convex} expresses the fact that $0\in \prt_x L(x^*, \lambda)$, i.e., that $x^* \in \argmin_{\mR^d} L(\cdot, \lambda)$.

Fixing $x\in \mR^d$, one can also consider the maximization of $L$ in $\lambda\in D$. Clearly, if $x\not \in \Omega$, so that $\gamma_i(x) \neq 0$ for some $i\in \CE$ or $\gamma_i(x) > 0$ for some $i\in \CI$, then $\max_D L(x, \lambda) = +\infty$. If $x\in \Omega$, then the slackness conditions, which require $\pe \lambda i \gamma_i(x)=0$, $i\in \CI$, ensure that $\lambda \in \argmax_D L(x,\cdot)$. 

As a consequence, any pair $x^*\in \Omega$, $\lambda^*\in D$ satisfying the KKT conditions is such that
\begin{equation}
L(x^*, \lambda) \leq L(x^*, \lambda^*) \leq L(x, \lambda^*)
\label{eq:dual.saddle}
\end{equation}
for all $x\in \mR^d$ and $\lambda\in D$. Such a pair $(x^*, \lambda^*)$ is called a {\em saddle point} of the function $L$. Conversely, any saddle point of $L$, i.e., any $(x^*, \lambda^*)\in \mR^d \times D$ satisfying  \cref{eq:dual.saddle}, must be such that $x^*\in \Omega$ (to ensure that $L(x^*, \cdot)$ is bounded), and satisfies the KKT conditions. 

We therefore obtain the equivalence of the two properties, for $(x^*, \lambda^*)\in \mR^d \times D$:
\begin{enumerate}[label=(\roman*)]
\item $x^*\in \Omega$ and $(x^*, \lambda^*)$ satisfies the KKT conditions.
\item \Cref{eq:dual.saddle} holds for all $(x, \lambda) \in \mR^d \times D$.
\end{enumerate}

Consider now the additional condition that
\begin{enumerate}[label=(iii)]
\item $x^*\in \argmin_\Omega F$ and $\lambda^* \in \argmax_D L^*$.
\end{enumerate}
We already know that, if $(x^*, \lambda^*)$ satisfy the KKT conditions, then $x^* \in \argmin_\Omega F$ (because $\CN'_\gamma(x^*) \subset \CN_\Omega(x^*)$). Moreover, if \cref{eq:dual.saddle} holds, then the  inequality $L(x^*, \lambda) \leq L(x^*, \lambda^*)$ implies that $L^*(\lambda) \leq L(x^*, \lambda^*)$ for all $\lambda\in D$. The inequality   $L(x^*, \lambda^*) \leq L(x, \lambda^*)$ for all $x$ implies that $L(x^*, \lambda^*) \leq L^*(\lambda^*)$. We therefore obtain the fact that $\lambda^*\in \argmax L^*(\lambda)$. To summarize, we have
\[
\mathrm{(i)} \Leftrightarrow \mathrm{(ii)} \Rightarrow \mathrm{(iii)}.
\]

To obtain the final equivalence, we need to assume constraints qualifications, such as Slater's conditions, to ensure that $\CN'_\gamma(x^*) = \CN_\Omega(x^*)$. If this holds, then (iii) implies (via \cref{th:kkt.convex}) that there exists $\tilde \lambda$ such that (i) and (ii) are satisfied for $(x^*, \tilde \lambda)$, with $L(x^*, \tilde\lambda) = L^*(\tilde\lambda)$ and $\tilde \lambda\in \argmin_D L^*$. This shows that $L^*(\tilde \lambda) =  L^*(\lambda^*)$. Moreover, from \cref{eq:dual.saddle}, we have
\[
L(x^*, \lambda^*) \leq L(x^*, \tilde \lambda) = L^*(\tilde \lambda),
\]
and, by definition of $L^*$, $L(x^*, \lambda^*) \geq L^*(\lambda^*)$. This shows that $L(x^*, \lambda^*) = L(x^*, \tilde \lambda)$. As a consequence, for all $(x, \lambda)\in \mR^d\times D$:
\[
L(x^*, \lambda) \leq L(x^*, \tilde{\lambda}) = L(x^*, \lambda^*) = L^*(\lambda^*) = \inf_{\mR^d} L(\cdot, \lambda^*) \leq L(x, \lambda^*)
\]
so that $(x^*, \lambda^*)$ satisfies (ii).

%

\subsection{Example: Quadratic programming}
\label{sec:example.svm}
Quadratic programming problems minimize $F(x) = \frac12 x^TAx - b^Tx$, where $A$ is a positive semidefinite matrix and $b\in \mR^d$,  subject to affine constraints $c_i^Tx - d_i = 0$, $i\in \CE$ and $c_i^T x - d_i \leq 0$, $i\in\CI$. 

We here consider the following objective function. Introduce variables $x\in \mR^d$, $x_0\in \mR$ and $\xi \in \mR^N$ and minimize, for a fixed parameter $\gamma$,
\[
F(x, x_0, \xi) = \frac12 |x|^2 + \gamma \sum_{k=1}^N \pe \xi k 
\]
subject to constraints, for $k=1, \ldots, N$ $\pe \xi k \geq 0$ and 
\[
b_k(x_0 + x^Ta_k) + \pe \xi k \geq 1
\]
where $b_k\in \{-1,1\}$ and $a_k\in \mR^n$ respectively denote the $k$th output and input training sample. This algorithm minimizes a quadratic function of the input variables $(x, x_0, \xi)$ subject to linear constraints, and is an instance of a quadratic programming problem (this is actually the support vector machine problem for classification, which will be described in \cref{sec:lin.svm}).

Introduce Lagrange multipliers $\eta_k$ for the constraint $\pe\xi k \geq 0$ and $\alpha_k$ for $b_k(x_0 + x^Ta_k) + \pe \xi k \geq 1$. The Lagrangian then takes the form
\begin{align*}
L(x, x_0, \xi, \alpha, \eta) &= \frac12 |x|^2 + \gamma \sum_{k=1}^N \pe \xi k 
- \sum_{i=1}^N \eta_k \pe \xi k - \sum_{k=1}^N \alpha_k (b_k(x_0 + x^Ta_k) + \pe \xi k - 1)\\
&= \frac12 |x|^2 +  \sum_{k=1}^N (\gamma - \eta_k - \alpha_k)\pe \xi k - x_0 \sum_{k=1}^N \alpha_k b_k - x^T \sum_{k=1}^N \alpha_k b_k a_k + \sum_{k=1}^N \alpha_k .
\end{align*}
We compute the dual Lagrangian $L^*$ by minimizing with respect to the primal variables. We note that $L^*(\alpha, \eta) = -\infty$ when $\sum_{k=1}^N \\alpha_k b_k \neq 0$, so that $\sum_{k=1}^N \alpha_k b_k = 0$ is a constraint for the dual problem. The minimization in $\pe \xi k$ also gives $-\infty$ unless $\gamma - \eta_k - \alpha_k = 0$, which is therefore another constraint. 
Finally, the optimal values of $x$ is 
\[
x = \sum_{k=1}^N \alpha_k b_k a_k
\]
and we obtain the expression of the dual problem, which maximizes
\[
- \frac 12 \sum_{k, l=1}^N \alpha_k \alpha_l b_kb_l a_k^T a_l + \sum_{k=1}^N \alpha_k 
\]
subject to $\eta_k, \alpha_k\geq 0$, $\gamma - \eta_k - \alpha_k = 0$ and $\sum_{k=1}^N \alpha_k b_k = 0$. The conditions on $\eta_k$ and $\alpha_k$ can be rewritten as $0 \leq \alpha_k \leq \gamma$, $\eta_k = \gamma - \alpha_k$, and since the rest of the problem does not depends on $\eta$, the dual problem can be reduced to maximizing
\[
L^*(\alpha) = - \frac 12 \sum_{k, l=1}^N \alpha_k \alpha_l a_k^T a_l 
+ \sum_{k=1}^N \alpha_k
\]
subject to $0 \leq \alpha_k \leq \gamma$ and $\sum_{k=1}^N \alpha_k b_k = 0$.

%

\subsection{Proximal iterations and augmented Lagrangian}
The concave function $L^*$ can be maximized by minimizing $-L^*$ using proximal iterations (\cref{eq:prox.iterations}):
\[
\lambda(t+1) = \prox_{-\alpha_t L^*}(\lambda(t)) = \argmax_D(\lambda \mapsto L^*(\lambda) - \frac{1}{2\alpha_t} |\lambda - \lambda(t)|^2).
\]

Introduce the function
\[
\phi(x, \lambda) = F(x) + \sum_{i\in \CC} \pe\lambda i \gamma_i(x) - \frac{1}{2\alpha_t} |\lambda - \lambda(t)|^2
\]
so that
\[
\lambda(t+1) = \argmax_{ \mu\in D} \inf_{x \in \mR^n} \phi(x, \mu).
\]

The function $\phi$ is convex in $x$ and strongly concave in $\mu$. Results in ``minimax theory'' \cite{bertsekas2009convex} implies that one has the equality 
\begin{equation}
\label{eq:inf.sup}
\max_{\mu\in D} \inf_{x\in \mR^n} \phi(x, \mu) = \inf_{x\in \mR^d} \sup_{\mu\in D} \phi(x, \mu)
\end{equation}
(Note that the left-hand side of this equation is never larger than the right-hand side, but their equality requires additional hypotheses---which are satisfied in our context---in order to hold.)

Importantly, the maximization in $\mu$ in the right-hand side has a closed form solution. It requires to maximize
\[
\sum_{i\in \CC} \Big(\mu_i \gamma_i(x) - \frac{1}{2\alpha_t} (\mu_i - \lambda_i(t))^2\Big)
\]
subject to $\mu_i \geq 0$ for $i\in \CI$, and each $\mu_i$ can be computed separately.
For $i \in \CE$, there is no constraint on $\mu_i$, and one finds
\[
\mu_i = \lambda_i(t) + \alpha_t \gamma_i(x),
\]
and
\[
\mu_i \gamma_i(x) - \frac{1}{2\alpha_t} (\mu_i - \lambda_i (t))^2 = \lambda_i(t) \gamma_i + \frac{\alpha_t}{2} \gamma_i(x)^2\\ = \frac{1}{2\alpha_t} (\lambda_i(t) + \alpha_t \gamma_i(x))^2 - \frac{\lambda_i(t)^2}{2\alpha_t}.
\]
For $i\in \CI$, the solution is 
\[
\mu_i = \max(0, \lambda_i(t) + \alpha_t \gamma_i(x))
\]
and one can check that, in this case:
\[
\mu_i \gamma_i(x) - \frac{1}{2\alpha_t} (\mu_i - \lambda_i(t))^2 =
\frac{1}{2\alpha_t} \max(0, \lambda_i(t) + \alpha_t \gamma_i(x))^2 - \frac{\lambda_i(t)^2}{2\alpha_t}
\]

As a consequence, the right-hand side of \cref{eq:inf.sup} requires to minimize
\begin{multline*}
G(x) = F(x) + \frac{1}{2\alpha_t} \sum_{i\in\CE} (\lambda_i(t) + \alpha_t \gamma_i(x))^2 + \frac{1}{2\alpha_t} \sum_{i\in\CI} \max(0,\lambda_i(t) + \alpha_t \gamma_i(x)))^2\\
-  \frac{1}{2\alpha_t}\sum_{i\in \CC} \lambda_i(t)^2.
\end{multline*}
If we assume that the sub-level sets $\{x\in \Omega: F(x) \leq \rho\}$ are bounded (or empty) for any $\rho\in \mR$, then so are the sets $\{x\in \mR^n: G(x) \leq \rho\}$, and this is a sufficient condition for the existence of a {\em saddle point} for $\phi$, which is a pair $(x^*, \lambda^*)$ such that, for all $(x, \lambda) \in \mR^n \times D$, 
\[
\phi(x^*, \lambda) \leq \phi(x^*, \lambda^*) \leq \phi(x, \lambda^*).
\]
One can then check that this implies that $x^*\in \argmin_{\mR^n} G$ while $\lambda^* = \lambda(t+1)$, so that the latter can be computed as follows:
\begin{equation}
\left\{
\begin{aligned}
&x(t) = \argmin_{x\in \mR^n}\bigg\{ F(x) + \frac{1}{2\alpha_t} \sum_{i\in\CE} (\lambda_i(t) + \alpha_t \gamma_i(x))^2 \\
& \qquad \qquad + \frac{1}{2\alpha_t} \sum_{i\in\CI} \max(0,\lambda_i(t) + \alpha_t \gamma_i(x)))^2\bigg\} \\
&\lambda_i (t+1)= \lambda_i (t) + \alpha_t \gamma_i(x(t)),\ i\in \CE\\
&\lambda_i (t+1) = \max(0, \lambda_i (t) + \alpha_t \gamma_i(x(t))),\ i\in\CI
\end{aligned}
\right.
\label{eq:augmented.lagrangian}
\end{equation}
These iterations define the augmented Lagrangian algorithm. Starting this algorithm  with some $\lambda(0)\in \mR^{|\CC|}$, and constant $\alpha$, $\lambda(t)$ will converge to a solution $\hat \lambda$ of the dual problem. The last two iterations stabilizing imply that $\gamma_i(x(t))$ converges to 0 for $i\in \CE$, and also  for $i\in \CI$ such that $\hat \lambda_i >0$, and that $\limsup \gamma_i(x(t)) = 0$ otherwise. This shows that, if $x(t)$ converges to a limit $\tilde x$, then $G(\tilde x) = F(\tilde x)$.
However, for any $x\in \Omega$, we have
\[
G(x(t)) \leq G(x)\leq F(x)
\]
(the proof being left to the reader),
showing that $\tilde x\in \argmin_\Omega F$.

Note that the augmented Lagrangian method can also be used in non-convex optimization problems \cite{nocedal2006nonlinear}, requiring in that case that $\alpha$ is small enough.

\subsection{Alternative direction method of multipliers}
We return to a situation considered in \cref{sec:proximal} where the function to minimize takes the form $F(x) = G(x) + H(x)$. Here, we do not assume that $G$ or $H$ is smooth,  but we will need their respective proximal operators to be easy to compute. 

The problem can be reformulated as a minimization with equality constraints, namely that of minimizing $\tilde F(x,z) = G(x) + H(z)$ subject to $x=z$. We will actually consider a more general situation, namely the problem minimizing a function $\tilde F(x,z)$ subject to constraints $Ax + Bz = c$ where $A$ and $B$ are respectively $d\times n$ and $d\times m$ matrices, $x\in \mR^n$, $z\in \mR^m$, $c\in \mR^d$. The augmented Lagrangian algorithm applied to this problem leads to iterate (with only equality constraints)
\[
\left\{
\begin{aligned}
x_{t}, z_t &= \argmin_{x\in \mR^n, z\in\mR^m}\{ G(x) + H(z) + \frac{1}{2\alpha_t} |\lambda_t + \alpha_t (Ax + Bz - c)|^2\} \\
\lambda_{t+1} &= \lambda_t + \alpha_t (Ax_t+Bz_t -c)
\end{aligned}
\right.
\]
with $\lambda_t\in \mR^d$. 

One can now consider splitting the first step in two and iterate:
\begin{equation}
\label{eq:admm.2}
\left\{
\begin{aligned}
x_{t} &= \argmin_{x\in \mR^n}\{ G(x) + H(z_{t-1}) + \frac{1}{2\alpha_t} |\lambda_t + \alpha_t (Ax + Bz_{t-1} - c)|^2\} \\
z_t &= \argmin_{z\in \mR^m}\{ G(x_t) + H(z) + \frac{1}{2\alpha_t} |\lambda_t + \alpha_t (Ax_t + Bz - c)|^2\} \\
\lambda_{t+1} &= \lambda_t + \alpha_t (Ax_t+Bz_t -c)
\end{aligned}
\right.
\end{equation}
(Obviously, $H(z_{t-1})$ and $G(x_t)$ are constant in the first and second minimization problems and can be removed from the formulation.)
These iterations constitute the ``alternative direction method of multipliers,'' or ADMM (the method is also sometimes called Douglas-Rachford splitting). It is not equivalent to the augmented Lagrangian algorithm (one would need to iterate a large number of times over the first two steps before applying the third one for this), but still satisfies good convergence properties.  The reader can refer to \citet{boyd2011distributed} for a relatively elementary proof that shows that this algorithm converges, with constant $\alpha$, as soon as, in addition to the hypotheses that were already made, the Lagrangian
\[
L(x,z,\lambda) = G(x) + H(z) + \lambda^T(Ax+Bz-c)
\]
has a saddle point: there exists $x^*, z^*, y^*$ such that
\[
\max_y L(x^*, z^*, \lambda) = L(x^*, z^*, \lambda^*) = \min_{x,z} L(x, z, \lambda^*).
\]

\begin{remark}
\label{rem:admm.alt}
If $\al_t = \alpha$ does not depend on time, \cref{eq:admm.2} can be slightly simplified by letting $u_t = \lambda_t/\alpha$, with the iterations
\begin{equation}
\label{eq:admm.3}
\left\{
\begin{aligned}
x_{t} &= \argmin_{x\in\mR^n} \{ G(x) + \frac{\alpha}{2} |u_t +Ax + Bz_{t-1} - c|^2\} \\
z_t &= \argmin_{z\in \mR^m} \{H(z) + \frac{\alpha}{2} |u_t + Ax_t + Bz - c|^2\} \\
u_{t+1} &= u_t + Ax_t+Bz_t -c,
\end{aligned}
\right.
\end{equation}
in which we have removed the constant additive terms.
\end{remark}

\section{Convex separation theorems and additional proofs}
\label{sec:separation}

We conclude this chapter by completing some of the proofs left aside when discussion convex functions. These proofs use convex separation theorems, stated below (without proof). 
\begin{theorem}[c.f., \citet{rockafellar1970convex}]
\label{th:separation.1}
Let $\Omega_1$ and $\Omega_2$ be two nonempty convex sets with $\relint(\Omega_1)\cap\relint(\Omega_2) = \emptyset$. Then there exists $b\in \mR^d$ and $\beta\in \mR$ such that $b\neq 0$,  
$b^T x  \leq \beta$ for all $x\in \Omega_1$ and  $b^T x  \geq \beta$ for all $x\in \Omega_2$, with a strict inequality for at least one $x\in \Omega_1\cup \Omega_2$.
\end{theorem}

\begin{theorem}
\label{th:separation.2}
Let $\Omega_1$ and $\Omega_2$ be two nonempty convex sets with $\Omega_1\cap\Omega_2 = \emptyset$ and $\Omega_1$ compact. Then there exists $b\in \mR^n$, $\beta\in \mR$ and $\epsilon<0$ such that 
$b^T x  \leq \beta - \epsilon$ for all $x\in \Omega_1$ and  $b^T x \geq \beta + \epsilon$ for all $x\in \Omega_2$.
\end{theorem}


\subsection{Proof of \cref{prop:subg.exist}}

We start with a few general remarks. If $x\in \mR^d$, the set $\{x\}$ is convex and $\relint(\{x\}) = \{x\}$. If $\Omega$ is any convex set such that $x\not\in \relint(\Omega)$, then \cref{th:separation.1} implies that there exist $b\in \mR^d$ and $\beta \in \mR$ such that $b^Ty \geq \beta \geq b^Tx$ for all $y\in \Omega$ (with $b^Ty > b^Tx$ for at least one $y$). If $x$  is in $\Omega\setminus(\relint(\Omega))$ (so that $x$ is a point on the relative boundary of $\Omega$), then, necessarily $b^T x = \beta$ and we can write
 \[
 b^T y \geq b^T x 
 \]
 for all $y\in \Omega$ with a strict inequality for some $y\in \Omega$. One says that $b$ and $\beta$ provide a {\em supporting hyperplane} for $\Omega$ at $x$.
 
 Now, if $F$ is a convex function, with
 \[
 \epi(F) = \{(y,a)\in \mR^d\times \mR: F(y) \leq a\}
 \]
 then 
 \[
 \relint(\epi(F)) = \{(y,a)\in \rdom(F) \times \mR: F(y) < a\}
 \]
 (this simple fact is proved in \cref{lem:relint.epi} below). In particular, if $x\in \dom(F)$, then $(x, F(x))$ must be in the relative boundary of $\epi(F)$. This implies that there exists $(b, b_0)\neq(0,0) \in \mR^d \times \mR$ such that, for all $(y, a) \in \epi(F)$:
 \[
 b^T y + b_0 a \geq b^T x + b_0 F(x)\,.
 \] 

If one assumes that $x\in \rdom(F)$, then, necessarily, $b_0\neq 0$. To show this, assume otherwise, so that $b^T y \geq b^T x$ for all $y\in \dom(F)$, with $b\neq 0$. We get a contradiction using the fact that, for some $\epsilon>0$, $[y, x - \epsilon(y-x)]$ belongs to $\dom(\Om)$, because $b^T(y-x)$ cannot have a constant sign on this segment.

So $b_0\neq 0$ and necessarily $b_0>0$ to ensure that $b^Ty + b_0 a$ is bounded from below for all $a\geq F(y)$. Without loss of generality, we can assume $b_0=1$ and we get, for all $y\in \dom(F)$
\[
F(y) + b^T y \geq F(x) + b^T x
\]
which shows that $-b \in \prt F(x)$, justifying the fact that $\prt F(x)\neq \emptyset$ for $x\in \rdom(F)$.

We now state and prove the result announced above on the relative interior of the epigraph of a convex function.
 \begin{lemma}
 \label{lem:relint.epi}
Let $F$ be a convex function with epigraph
\[
\epi(F) = \{(y,a): y\in \dom(F), F(y) \leq a\}.
\]
Then
\[
\relint(\epi(G)) = \{(y,a): y\in \rdom(F), F(y) < a\}.
\]
\end{lemma}
\begin{proof}
Let $\Gamma = \{(y,a): y\in \rdom(F), F(y) < a\}$. Assume that $(y,a) \in \relint(\epi(F))$. Then $(y,b) \in \epi(F)$ for all $b>a$ and there exists $\ep>0$ such that $(y,a) - \ep ((y,b) - (y,a)) \in \epi(F)$ which  requires that $F(y) \leq a - \epsilon(b-1) < a$. Now, take $x\in \dom(F)$. Then, $(x,F(x))\in \epi(\dom(F))$ and $(y, a) - \ep ((x, F(x)) - (y,a)) \in \epi(F)$ for small enough $\epsilon$, showing that $F(y - \epsilon(x-y)) \leq (1+\epsilon) a -\epsilon F(x)$ and $y - \epsilon(x-y) \in\dom(F)$. This proves that $y\in \rdom(F)$ and the fact that $\relint(\epi(F)) \subset \Gamma$.    

Take $(y,a) \in \Gamma$, and $(x,b) \in \epi(F)$. We need to show that $(y - \ep(x-y), a - \ep(b-a))\in \epi(F)$ for small enough $\epsilon$, i.e., that
\[
F(y - \ep(x-y)) \leq a - \epsilon(b-a)
\]
for small enough $\epsilon$. But this is an immediate consequence of the facts that $F$ is continuous at $y\in\rdom(G)$ and $F(y) < a$.
\end{proof}

\subsection{Proof of \cref{th:subg.sum}}
Assume that there exists $\bar x\in \rdom(F_1) \cap \rdom(F_2)$. Take $x\in \dom(F_1) \cap \dom(F_2)$ and $g \in \prt (F_1+F_2)(x)$. We want to show that $g = g_1 + g_2$ with $g_1\in \prt F_1(x)$ and $g_2\in \prt F_2(x)$. 

By definition, we have
\[
F_1(y) + F_2(y) \geq F_1(x) + F_2(x) + g^T(y-x)
\]
for all $y$. 
We want to decompose $g$ as $g = g_1 + g_2$ with $g_1\in \prt F_1(x)$ and $g_2\in \prt F_2(x)$. Equivalently, we want to find $g_2\in \mR^d$ such that, for all $y\in \mR^d$,
\begin{align*}
F_1(y)& \geq F_1(x) + (g-g_2)^T(y-x)\\
F_2(y)& \geq F_2(x) + g_2^T(y-x)
\end{align*}
First note that we can replace $F_1$ by $y \mapsto F_1(y) - F_1(x) - g^T (y-x)$ and $F_2$ by $y\mapsto F_2(y) - F_2(x)$ and assume with loss of generality that $F_1(x) = F_2(x) = 0$ and $g=0$. Making this assumption, we need to find $g_2$ such that
\begin{align*}
F_1(y)& \geq -g_2^T(y-x)\\
F_2(y)& \geq  g_2^T(y-x)
\end{align*}
for all $y\in\mR^d$ and some $g_2\in \mR^d$, under the assumption that $F_1(y) + F_2(y)\geq 0$ for all $y$.
Introduce the two convex sets in $\mR^d\times \mR$
\begin{align*}
\Omega_1 & =\epi(F_1) = \{(y, a) \in \mR^d\times \mR: F_1(y)  \leq a\}
\\
\Omega_2 &= \{(y, a) \in \mR^d\times \mR:  F_2(y) \leq -a\}\,.
\end{align*}
The set $\Omega_2$ is the image of $\epi(F_2)$ by the transformation $(y,a) \mapsto (y, -a)$. 
We have
\begin{align*}
\relint(\Omega_1) & =\epi(F_1) = \{(y, a) \in \rdom(F_1) \times \mR: F_1(y)  < a\}
\\
\relint(\Omega_2) &= \{(y, a) \in \rdom(F_2) \times \mR:  F_2(y) < -a\}\,.
\end{align*}
Since $F_1+F_2 \geq 0$, $\Omega_1$ and $\Omega_2$ have non-intersecting relative interiors. 
We can apply the first separation theorem, providing  $\bar b = (b, b_0) \in \mR^d \times \mR$ and $\beta\in \mR$ such that $\bar b \neq (0,0)$, 
$b^Ty + b_0a - \beta \leq 0$ for $(y, a)\in \Omega_1$ and $b^Ty + b_0a - \beta \geq 0$ for $(y,b) \in \Omega_2$, with a strict inequality for at least one point in $\Omega_1\cup \Omega_2$. We therefore obtain the fact that, for all $y$ and $a$,
\begin{align*}
F_1(y)  \leq a \Rightarrow b^Ty + b_0 a -\beta \leq 0 \\
F_2(y)  \leq -a \Rightarrow b^Ty + b_0 a -\beta \geq 0.
\end{align*}
We claim that  $b_0 \neq 0$. Indeed, if $b_0=0$, the statement for $F_1$ would imply  that $b^T y -\beta \leq 0$ for all $y\in \dom(F_1)$ and the one on $F_2$ that $b^Ty - \beta \geq 0$ for $y\in \dom(F_2)$.  The point $\bar x \in \relint(\Omega_1) \cap \relint(\Omega_2)$ should then satisfy $b^T \bar x -\beta = 0$. We know that there exists a point $y\in \Omega_1 \cup \Omega_2$ such that $b^T y\neq \beta$. Assume that $y\in \Omega_1$, so that $b^T y -\beta < 0$ and take $\epsilon >0$ such that $\tilde y = \bar{x} - \epsilon(y-\bar x) \in \Omega_1$. Then 
\[
b^T \tilde y - \beta  = - \epsilon (b^T y - \beta) < 0,
\]
which is a contradiction. A similar contradiction is obtained when $y$ belongs to $\Omega_2$, yielding the fact that $b_0$ cannot vanish.

Moreover, we clearly need $b_0<0$ to ensure that $b^Ty + b_0 a -\beta \leq 0$ for all large enough $a$ if $y\in \dom(\Omega_1)$. There is then no loss of generality in assuming $b_0 = -1$ and we get 
\begin{align*}
F_1(y)  \leq a \Rightarrow b^Ty -\beta \leq a \\
F_2(y) \leq -a \Rightarrow b^Ty -\beta \geq a,
\end{align*}
which is equivalent to 
\[
 - F_2(y) \leq b^Ty -\beta \leq F_1(y)
\]
Taking $y=x$ gives $\beta = b^T x$ and we get the desired inequality with $g_2 = -b$.


\subsection{Proof of \cref{th:subg.affine}}
Let $\bar x\in \mR^m$ such that $A\bar x\in \rdom(F)$. 
We need to prove that $\prt G(x) \subset A^T \prt F(Ax+b)$ when $G(x) = F(Ax+b)$. We assume in the following that $b=0$, since the theorem with $G(x) = F(x+b)$ is obvious.
If $g\in \prt G(x)$, we have
\[
F(Ay) \geq F(Ax) + g^T(y-x)
\]
for all $y\in \mR^m$. We want to show that there exists $h\in \mR^d$ such that $g = A^T h$ and, for all $z\in \mR^d$,
\[
F(z) \geq F(Ax) + h^T(z- Ax) = F(Ax) + h^T z - g^T x.
\]

Let $\Omega_1 = \epi(F) = \{(z,a): , z\in \mR^d, F(z) \leq a\}$ and
\[
\Omega_2 = \{(Ay, a): y\in \mR^m, a = g^T(y-x) + G(x)\}\subset \mR^d \times \mR.
\]
Note that $\Omega_2$ is an affine space with $\relint(\Omega_2) = \Omega_2$.
If $(z, a) \in \relint(\Omega_1)\cap \Omega_2$, then $z = Ay$ for some $y\in \mR^m$ and $g^T(y-x) + G(x) > F(z) = G(y)$. This contradicts the fact that $g\in \prt G(x)$ and shows that  $\relint(\Omega_1) \cap \Omega_2 = \emptyset$. As a consequence, there exist $(b,b_0)\neq (0,0)$ and $\beta$ such that 
\begin{align*}
F(z) \leq a \Rightarrow b^T z + b_0 a\leq \beta\\
z=Ay, a = g^T(y-x) + G(x) \Rightarrow b^T z + b_0 a\geq \beta
\end{align*}
Assume, to get a contradiction, that $b_0 = 0$ (so that $b\neq 0$). Then $b^T Ay \geq \beta$ for all $y$, which is only possible if $b$ is perpendicular to the range of $A$ and $\beta\leq 0$. On the other hand, $F(A\bar x) <\infty$ implies that $0 = b^T A\bar x + b_0 F(A\bar x) \leq \beta$, so that $\beta = 0$. Furthermore, we know that one of the inequalities above has to be strict for at least one element of $\Omega_1\cup\Omega_2$, but this cannot be true on $\Omega_2$, so there exists $z\in \dom(F)$ such that $b^T z < 0$. Since $b^TA\bar x = 0$ and $A\bar x\in \rdom(F)$, we have $A\bar x - \epsilon(z - A\bar x) \in \dom(F)$, so that $b^T(-\epsilon z) \leq 0$, yielding a contradiction.

So, we need $b_0\neq 0$, and the first pair of inequalities clearly requires $b_0<0$, so that we can take $b_0 = -1$. This shows that 
\[
b^Tz - \beta \leq F(z)
\]
for all $z$ and
\[
b^T Ay - \beta \geq g^T(y-x) + F(Ax)
\]
for all $y$. Taking $y=x$, $z=Ax$, we find that $\beta = b^TAx - F(Ax)$ yielding
\[
F(z) - F(Ax) \geq b^T(z-x) 
\]
for all $z$ and $b^TA (y - x) \geq g^T (y-x)$ for all $y$. This last inequality implies that $g = A^Tb$ and the first one that $b\in \prt F(Ax)$, therefore concluding the proof. 

\newpage
\markright{PROBLEMS}
\section*{Problems}
\addcontentsline{toc}{section}{Problems}
\input Problems_Optimization




\chapter[Introduction: Bias and Variance]{Introduction: Bias, Variance and Density Estimation}
\label{chap:intro}

In this chapter, we illustrate the bias variance dilemma in the context of density estimation, in which problems are similar to those  encountered in classical parametric or non-parametric statistics \citep{rao1983nonparametric,dgl96,pons2011functional}. 

For density estimation, one assumes that a random variable $X$ is given with unknown p.d.f. $f$ and we want to build an estimator, i.e., a mapping $(x, T) \mapsto \hat f(x;T)$ that provides an estimation of $f(x)$ based on a training set $T = (x_1, \ldots, x_N)$ containing $N$ i.i.d. realizations of $X$ (i.e., $T$ is a realization of $\mT = (X_1, \ldots, X_N)$, $N$ independent copies of $X$). Alternatively, we will say that the mapping $T \mapsto \hf(\,\cdot\,; T)$ is an estimator of the full density $f$. Note that, to further illustrate our notation, $\hf(x;T)$ is a number while $\hf(x; \mT)$ is a random variable.


\section{Parameter estimation and sieves}
\label{sec:sieves} 
Parameter estimation is the most common density estimation method, in which one restrict $\hat f$ to belong to
a finite-dimensional parametric class, denoted $(f_\th, \th\in\Th)$, with $\Th\sub
\mR^p$. For example, $f_\th$ can be a family of Gaussian distributions
on $\mR^d$. With our notation, a parametric model provides estimators taking the form
\[
\hf(x; T) = f_{\hat\th(T)}(x)
\]
and the problem becomes the estimation of the parameter $\hat\th$.

There are several, well-known methods for parameter
estimation, and, since this is not the focus of the book, we only
consider the most common one, maximum likelihood, which consists in
computing $\hth$ that maximizes the log-likelihood
\begin{equation}
\label{eq:log.lik}
C(\th) = \frac{1}{N}\sum_{k=1}^N \log f_\th(x_k)\,.
\end{equation}
The resulting $\hth$ (when it exists) is called the maximum likelihood estimator of $\th$, or m.l.e.

If the true $f $ belongs to the parametric class, so that $f= f_{\th_*}$ for some $\theta_*\in \Th$, standard results in mathematical statistics \cite{bickel2015mathematical,lehmann2006theory} provide sufficient conditions for  $\hth$ to converge to $\th_*$ when $N$ tends to infinity.
However, the fact that the true p.d.f. belongs to the finite
dimensional class $(f_\th)$ is an optimistic assumption that is generally false. In this regard, the standard theorems in parametric statistics may be regarded as analyzing  a ``best case scenario,'' or as performing a ``sanity check,'' in which one asks whether, in the ideal situation in which $f$ actually belongs to the parametric class, the designed estimator has a proper behavior. In non-parametric statistics, a  parametric model can still be a plausible approach in order to approximate the true $f$, but the relevant question should then be whether $\hat f$ provides (asymptotically), the best approximation to $f$ among all  $f_\th$, $\th\in \Th$. The maximum likelihood estimator can be
analyzed from this viewpoint, if one measures the difference between two
density functions by the Kullback-Leibler divergence (also called differential entropy):
\begin{equation}
\label{eq:kl.pdf}
\KL(f \| f_\th) = \int_{\mR^d} \log \frac{f(x)}{f_\th(x)} f(x) dx
\end{equation}
which is positive unless $f=f_\th$ (and may be equal to $+\infty$). 

This expression of the divergence is a simplification of its general mea\-sure-theo\-retic definition, that we now provide for completeness---and future use. Let $\mu$ and $\nu$ be two probability measures on a set $\widetilde \Om$. Recall from \cref{sec:abs.cont} that one says that $\mu$ is absolutely continuous with respect to $\nu$, with notation $\mu\ll\nu$, if, for every (measurable) subset  $A \sub \widetilde \Omega$, $\nu(A) = 0$ implies $\mu(A) = 0$. The Radon-Nikodym theorem then states that $\mu\ll \nu$ is and only if there exists a non-negative function $g=d\mu/d\nu$ (the Radon-Nikodym derivative of $\mu$ with respect to $\nu$) defined on $\widetilde \Omega$ such that 
\[
\mu(A) = \int_{A} g(x) d\nu(x).
\]
In terms of random variables, this says that, if $X: \Omega \to \widetilde \Omega$ and $Y: \Omega \to \widetilde \Omega$ are two random variables with respective distributions $\mu$ and $\nu$, and $\phi: \widetilde \Omega \to \mR$ is measurable, then $E(\phi(X)) = E(g(Y)\phi(Y))$. The general definition of the Kullback-Leibler divergence between $\mu$ and $\nu$ is then:
\begin{equation}
\label{eq:kl.definition}
\KL(\mu \| \nu) = \left\{
\begin{aligned}
&\int_{\tilde\Om} \left(\log \frac{d\mu}{d\nu}\right)  \frac{d\mu}{d\nu} d\nu && \text{ if } \mu \ll \nu\\
&+\infty && \text{ otherwise}
\end{aligned}
\right.
\end{equation}
In the case when $\mu = f \,dx$ and $\nu = \tilde f \,dx$ are both probability measures on $\mR^d$ with respective p.d.f.'s $f$ and $\tilde f$, $\mu \ll \nu$ means that $f/\tilde f$ is well defined everywhere except on a set of $\nu$-probability zero. It is then equal to $d\mu/d\nu$.  If $\mu\ll\nu$, we can therefore write
\[
\KL(\mu\|\nu) = \int_{\mR^d} \frac{f(x)}{\tilde f(x)} \log \bigg(\frac{f(x)}{\tf(x)}\bigg) \tilde f(x) dx = \int_{\mR^d}  \log \bigg(\frac{f(x)}{\tf(x)}\bigg) f(x) dx
\] 
and
we will make the abuse of notation of writing $\KL(f \| \tf)$ for $\KL(f\, dx\| \tf\,dx)$, which gives the expression provided in \cref{eq:kl.pdf}. 

The general definition also gives a simple expression when $\widetilde \Om$ is a finite set, with
\[
\KL(\mu\|\nu) = \sum_{x\in\widetilde\Om} \log \frac{\mu(x)}{\nu(x)} \mu(x),
\]
that we will use later in these notes (if there exists $x$ such that $\mu(x) > 0$ and $\nu(x) =0$, then $\KL(\mu\|\nu)=\infty$).  The most important property for us is that the Kullback-Leibler divergence can be used as a measure of discrepancy between two probability distribution, based on the following proposition.
\begin{proposition}
\label{prop:kl}
Let $\mu$ and $\nu$ be two probability measures on $\widetilde \Om$. Then
  $\KL(\mu\|\nu)\geq 0$ and vanishes if and only if $\mu=\nu$.
  \end{proposition}
\begin{proof}
Assume that $\mu\ll\nu$ since the statement is obvious otherwise and let $g = d\mu/d\nu$. We have $\int_{\widetilde \Om} g d\nu = 1$ (since, by definition, it is equal to $\mu(\widetilde \Omega)$) so that 
\[
\KL(\mu\|\nu) = \int_{\widetilde \Omega} (g\log g + 1 - g) d\nu.
\] 
We have $t\log t + 1-t \geq 0$ with equality if and only $t=1$ (the proof being left to the reader) so that $\KL(\mu\|\nu) = 0$ if and only if $g=1$ with $\nu$-probability one, i.e., if and only if $\mu=\nu$. 
\end{proof}
\bigskip

Minimizing $\KL(f\| f_\th)$ with respect to $\th$ is equivalent to maximizing
$$
E_f(\log f_\th) = \int_{\mR^d} \log f_\th(x) f(x) dx\,,
$$
and
an empirical evaluation of this expectation is
$\frac{1}{N} \sum_{k=1}^N \log f_\th(x_k)$, which provides the maximum
likelihood method. Seen in this context, consistency of the maximum likelihood estimator states that this estimator almost surely converges to a best approximator of
the true $f$ in the class $(f_\th, \th\in\Th)$. More precisely, if one assumes that the function $\theta \mapsto \log f_\th(x)$ is continuous\footnote{Upper-semi continuous is  sufficient.} in $\th$ for almost all $x$ and that, for all $\th \in \Th$, there exists a small enough $\de >0$ such that
\[
\int_{\mR^d} \Big(\sup_{|\th'-\th|<\de} \log f_{\th'}(x)\Big)\, f(x)\, dx  < \infty
\]
then, letting $\Th_*$ denote the set of maximizers of $E_f(\log f_\th)$, and assuming that it is not empty, the maximum likelihood estimator $\hth_N$ is such that, for all $\ep>0$ and all compact subsets $K\sub \Th$, 
\[
\lim_{N\to \infty} \myP\big(d(\hth_N, \Th_*) > \ep \text{ and } \hth_N\in K\big) \to 0
\]
where $d(\hth_N, \Th_*)$ is the Euclidean distance between $\hth_N$ and the set $\Th_*$. 
The interested reader can refer to \citet{van2000asymptotic}, Theorem 5.14, for a proof of this statement. Note that this assertion does not exclude the situation in which $\hth_N$ goes to infinity (i.e., steps out of ever compact subset $K$ in $\Th$), and the boundedness of the m.l.e. is either asserted from additional properties of the likelihood, or by simply restricting $\Th$ to be a compact set.

If $\Th_* = \{\th_*\}$ and the m.l.e. almost surely converges to $\th_*$, the speed of convergence can also be quantified by a central limit theorem (see \citet{van2000asymptotic}, Theorem 5.23) ensuring that, in standard cases $\sqrt N(\hth_N - \th_*)$ converges to a normal distribution.

Even though these results relate our present subject
to classical parametric statistics, they are not sufficient for our purpose,
because, when $f\neq f_{\th_*}$, the convergence of the m.l.e. to the best
approximator in $\Th$ still leaves a gap in the estimation of $f$. This
gap is often called the {\em bias} of the class $(f_\th, \th\in \Th)$. One can reduce it
 by considering larger classes (e.g., with more dimensions), but the larger the class, the less accurate the estimation of the best approximator becomes for a fixed sample size (the estimator has a larger
{\em variance}). This issue is known as the ``bias vs.\! variance dilemma,'' and to address it, it is necessary  to adjust the
class $\Th$ to the sample size in order to optimally balance the two types of error (and all non-parametric estimation methods have at least one mechanism that allows for this). When the ``tuning parameter''  is the dimension of $\Th$, the overall approach is often referred to as the  {\em method of sieves} \cite{grenander1981abstract,geman1982nonparametric}, in which  the dimension of $\Th$ is increased as a function of $N$ in a suitable way.

Gaussian mixture models provide one of the most popular choices with the me\-thod of sieves. Modeling in this setting typically follows some variation of the following construction.
Fix a sequence $(m_N, N\geq 1)$ and let
\begin{multline}
\label{eq:MoG.model}
\Th_N =  \Big\{f: f(x) =  \sum_{j=1}^{m_N} \al_j
\frac{e^{-|x-\mu_j|^2/2\sig^2}}{(2\pi\sig^2)^{d/2}},\\
 \mu_1, \ldots,
\mu_{m_N}\in\mR^d, 
\al_1 + \cdots + \al_{m_N} = 1, \al_1, \ldots, \alpha_{m_N} \in [0, +\infty), \sig>0\Big\}.
\end{multline}
There are therefore $(d+1) m_N$ free parameters in $\Th_N$. The integer $m_N$ allows one to adjust the dimension of $\Th_N$ and therefore controls the bias-variance trade-off. If  $m_N$ tends to infinity ``slowly enough,'' the m.l.e. will converges (almost surely) to the true p.d.f. $f$ \citep{geman1982nonparametric}.  However, determining optimal sequences $N \to m_N$ remains a challenging and largely unsolved  problem.

In practice the computation of  the
m.l.e. in this context uses an algorithm called {\em EM}, for expectation-maximization. This algorithm will be described later in \cref{chap:var.bayes}.

\section{Kernel density  estimation}
Kernel density estimators \cite{parzen1962estimation,sheather1991reliable,silverman1998density} provide alternatives to the method of sieves. They also lend themselves to some analytical developments that provide elementary illustrations of the bias-variance dilemma.

Define a kernel function as a function $K: \mR^d \to [0, +\infty)$ such that
\begin{equation}
\label{eq:kernel.density.def}
\int_{\mR^d} K(x) dx = 1,\quad \int_{\mR^d} |x| K(x)\, dx <\infty, \quad \int_{\mR^d} x K(x)\, dx = 0.
\end{equation}
Note that the third equation is satisfied, in particular, when $K$ is an even function, i.e., $K(-x) = K(x)$. 

Given $K$ and a scalar  $\sig >0$, the rescaled kernel is defined by
\[
K_\sig(x) = \frac{1}{\sig^d} K\left(\frac x \sig\right).
\]
Using the change of variable $y = x/\sig$ (so that $dy =
dx/\sig^d$) one sees that $K_\sig$ satisfies \cref{eq:kernel.density.def} as soon as $K$ does.

Based on a training set $T = (x_1, \ldots, x_N)$, the kernel density estimator defines the family of densities
$$
\hat f_\sig(x; T)  = \frac{1}{N} \sum_{k=1}^N K_\sig(x-x_k)
$$
One has 
\[
\int_{\mR^d} K_\sig(x-x_k)\, dx = 1
\]
so that it is clear that $\hat f_\sig$ is a p.d.f. In addition, 
\[
\int_{\mR^d} xK_\sig(x-x_k)\, dx = \int_{\mR^d} (y+x_k)K_\sig(y)\, dy = x_k
\]
so that 
\[
\int_{\mR^d} x \hf_\sig(x;T)\, dx = \bar x
\]
where $\bar x = (x_1 + \cdots + x_N)/N$.

\begin{figure}
\centering
\begin{tabular}{cc}
\includegraphics[trim=1cm 0cm 1cm 0cm,clip,height=0.2\textheight]{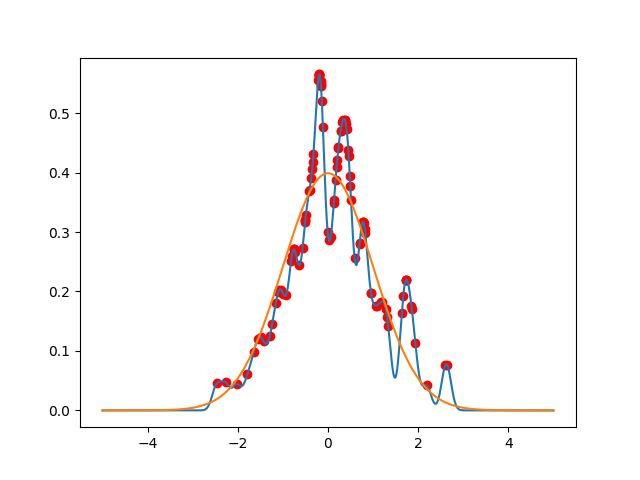}
&
\includegraphics[trim=1cm 0cm 1cm 0cm,clip,height=0.2\textheight]{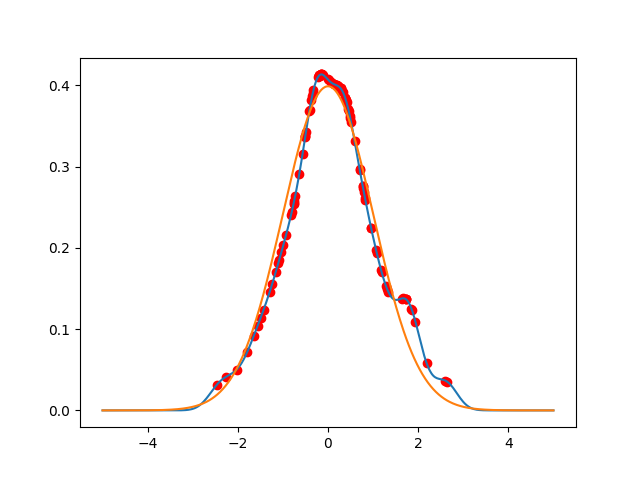}\\
\small $\sig = 0.1$ & \small $\sig = 0.25$\\

\includegraphics[trim=1cm 0cm 1cm 0cm,clip,height=0.2\textheight]{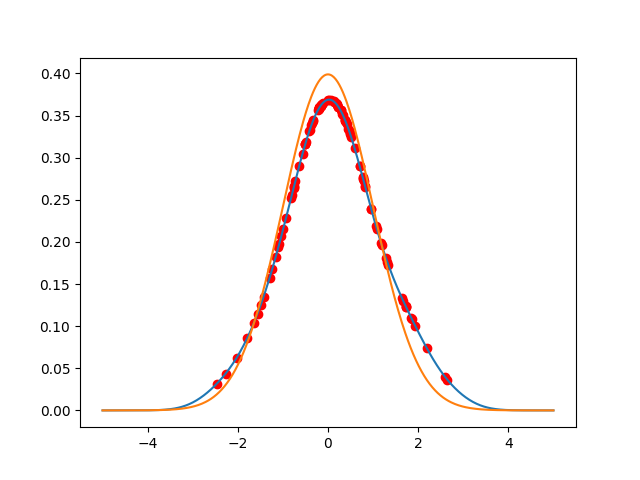}
&
\includegraphics[trim=1cm 0cm 1cm 0cm,clip,height=0.2\textheight]{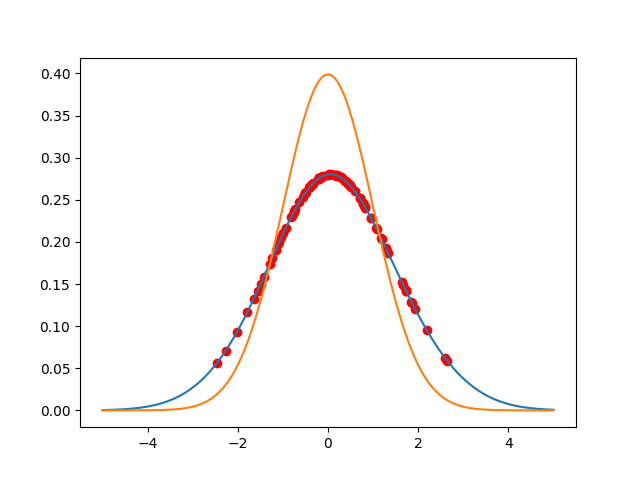}\\
\small $\sig = 0.5$ & \small $\sig = 1.0$
\end{tabular}
\caption{\label{fig:kde.1} Kernel density estimators using a Gaussian kernel and various values of $\sig$ when the true distribution of the data is a standard Gaussian (Orange: true density; Blue: estimated density,  Red dots: training data).}
\end{figure}

\begin{figure}
\centering
\begin{tabular}{cc}
\includegraphics[trim=1cm 0cm 1cm 0cm,clip,height=0.2\textheight]{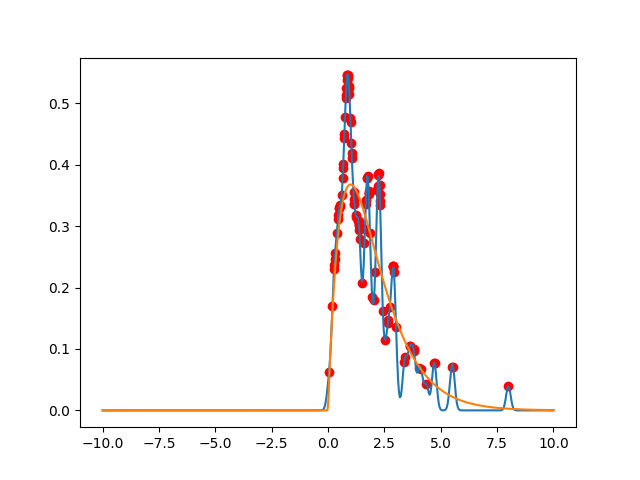}
&
\includegraphics[trim=1cm 0cm 1cm 0cm,clip,height=0.2\textheight]{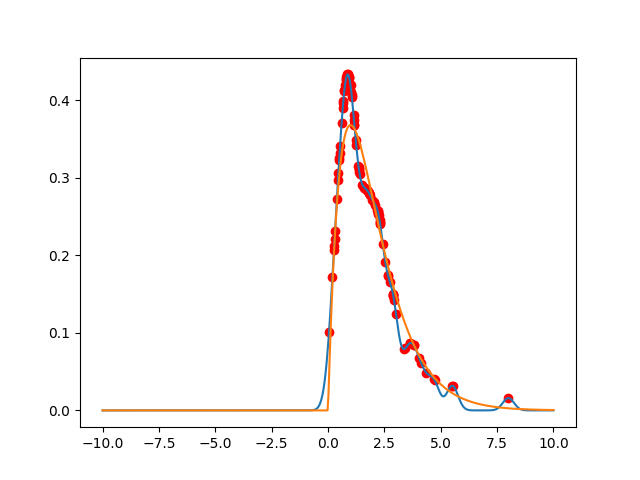}\\
\small $\sig = 0.1$ & \small $\sig = 0.25$
\\
\includegraphics[trim=1cm 0cm 1cm 0cm,clip,height=0.2\textheight]{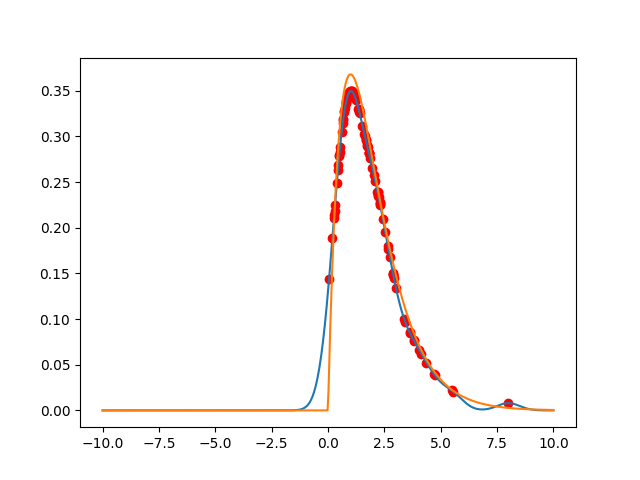}
&
\includegraphics[trim=1cm 0cm 1cm 0cm,clip,height=0.2\textheight]{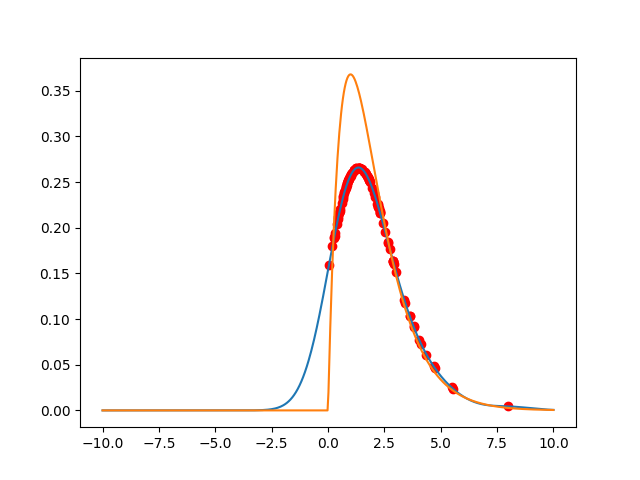}\\
\small $\sig = 0.5$ & \small $\sig = 1.0$
\end{tabular}
\caption{\label{fig:kde.2} Kernel density estimators using a Gaussian kernel and various values of $\sig$ when the true distribution of the data is a Gamma distribution with parameter 2 (Orange: true density; Blue: estimated density, Red dots: training data).}
\end{figure}

A typical choice for $K$ is a Gaussian kernel, $K(y) =
e^{-|y|^2/2}/(2\pi)^{d/2}$. In this case, the estimated density is a sum
of bumps centered at the data points $x_1, \ldots, x_N$. The width of
the bumps is controlled by the parameter $\sig$.  A small $\sig$
implies less rigidity in the model, which will therefore be more
affected by changes in the data: the estimated density will have a
larger variance. The converse is true for large $\sig$, at the cost of
being less able to adapt to variations in the true density: the model
has a larger bias (see \cref{fig:kde.1} and \cref{fig:kde.2}).

As we now show, in order to get a consistent estimator, one needs to let  $\sig = \sig_N$ depend on the size of
the training set. 
We have, taking expectations with respect to training data, 
\begin{eqnarray*}
E(\hf_\sig(x;\mT)) &=& \frac{1}{N\sig^d} \sum_{k=1}^N E\big
(K((x-X_k)/\sig)\big)\\
&=& \frac{1}{\sig^d} \int_{\mR^d} K((x-y)/\sig) f(y) dy\\
&=& \int_{\mR^d} K(z) f(x-\sig z) dz
\end{eqnarray*}
The bias of the estimator, i.e., the average difference between $\hf_\sig(x;\mT)$ and
$f(x)$ is therefore given by
$$
E(\hf_\sig(x;\mT)) - f(x) = \int_{\mR^d} K(z) (f(x-\sig z) - f(x)) dz.
$$
Interestingly, this bias does not depend on $N$, but only on $\sig$, and it is clear that, under mild continuity assumptions on $f$, it will go to zero with $\sig$. 

The variance of $\hf_\sig(x;\mT)$ is given by
$$
\var(\hf_\sig(x;\mT)) = \frac{1}{N\sig^{2d}} \var(K((x-X)/\sig))
$$
with
\begin{eqnarray*}
\frac{1}{N\sig^{2d}} \var(K((x-X)/\sig)) &=& \frac{1}{N\sig^{2d}} \int_\mR^d
K((x-y)/\sig)^2 f(y) dy \\
&& - \frac{1}{N\sig^{2d}} \left(\int_{\mR^d}
K((x-y)/\sig) f(y) dy\right)^2\\
&=& \frac{1}{N\sig^d} \int_{\mR^d} K(z)^2 f(x-\sig z) dz - \frac1N
\left(\int_{\mR^d}
K(z) f(x-\sig z) dz\right)^2
\end{eqnarray*}
The total mean-square error of the estimator is
$$
\myE((\hf_\sig(x) - f(x))^2) =  \var(\hf_\sig(x)) + (\myE(\hf_\sig(x)) -
f(x))^2.
$$
Clearly, this error cannot go to zero unless we allow $\sig = \sig_N$ to depend on $N$. For the bias term to go to zero, we know that we need $\sig_N \to 0$, in which case we can expect the second term in the variance to decrease like $1/N$, while, for the first term to go to zero, we need $N\sig_N^d$ to go to infinity. This illustrates the bias-variance dilemma: $\sig_N$ must go to zero in order to cancel the bias, but not too fast in order to also cancel the variance. There is, for each $N$, an optimal value of $\sig$ that minimizes the error, and we now proceed to a more detailed analysis and make this statement a little more precise.

Let us make a
Taylor expansion of both bias and variance, assuming that $f$ has at least three bounded derivatives and that $\int_{\mR^d} |x|^3 K(x) \,dx < \infty$.
We can write
$$
f(x-\sig z) = f(x) - \sig z^T \nabla f(x) + \frac{\sig^2}{2} z^T \nabla^2f(x) z +
O(\sig^3|z|^3),
$$ 
where $\nabla^2f(x)$ denotes the matrix of second derivatives of $f$ at $x$.
Since $\int zK(z) dz = 0$, this gives
$$
\myE(\hf_\sig(x;\mT)) - f(x) = \frac{\sig^2}{2} M_f (x) + o(\sig^2)
$$
with $M_f = \int K(z)\, z^T \nabla^2f(x) z\, dz$. Similarly, letting $S = \int K^2(z) \, dz$,
$$
\var(\hf_\sig(x)) = \frac{1}{N\sig^d} \big(S f(x) +  o(\sig^d + \sig^2)\big).
$$
Assuming that $f(x)>0$, we can obtain an asymptotically optimal value for
$\sig$ by minimizing the leading terms of the mean square error, namely
$$
\frac{\sig^4}{4} M^2_f  + \frac{S}{N\sig^d} f(x)
$$
which yields $\sig_N = O(N^{-1/(d+4)})$ and 
$$
\myE((\hf_{\sig_N}(x; \mT) - f(x))^2) = O(N^{-4/(d+4)}).
$$

If $f$ has $r+1$
derivatives, and $K$ has $r-1$ vanishing moments
(this excludes the Gaussian kernel) one can reduce this error to  $N^{- \frac{2r}{2r+d}}$.  These rates can be shown to be
``optimal,'' in the ``min-max'' sense, which roughly expresses the fact that,  for any other
estimator, there exists a function $f$ for which the convergence speed
is at least as ``bad'' as the one obtained for kernel density estimation.
\bigskip

This result  says that, in order to obtain a
given accuracy $\ep$ in the worst case scenario, $N$ should be chosen of order $(1/\ep)^{1 + (d/2r)}$
which grows exponentially fast with the dimension. This is the {\em
curse of dimensionality} which essentially states that the issue of
density estimation may be intractable in large dimensions. The
same statement is true also for  most other types of machine learning problems. Since machine learning essentially deals with high-dimensional data, this issue can be problematic. 

Obviously, because the min-max theory is a worst-case analysis, not all situations will be intractable for a given estimator, and some cases that are challenging for one of them may be quite simple for others: even though all estimators are ``cursed,'' the way each of them is cursed differs. Moreover, while many estimators are optimal in the min-max sense, this theory does not give any information on ``how often''  an estimator performs better than its worst case, or how it will perform on a given class of problems. (For kernel density estimation, however, what we found was almost universal with respect to the unknown density $f$, which indicates that this estimator is not a good choice in large dimensions.)

Another important point with this curse of dimensionality is that data may very often appear to be high dimensional while it has a simple, low-dimensional structure, maybe because many dimensions are irrelevant to the problem (they contain, for example, just random noise), or because the data is supported by a non-linear low-dimensional space, such as a curve or a surface. This information is, of course, not available to the analysis, but can sometimes be inferred using some of the dimension reduction methods that will be discussed later in \cref{chap:dim.red}. Sometimes, and this is also important, information on the data structure can be provided by domain knowledge, that is, by elements, provided by experts, that specify how the data has been generated (such as underlying equations) and reasonable hypotheses that are made in the field. This source of information should never be ignored in practice.

\chapter{Prediction: Basic Concepts}
\label{chap:general}

\section{General Setting}
The goal of prediction is to learn, based on training data, an input-output relationship between two random variables $X$ and $Y$, in the sense of finding, for a specified criterion, the best function of the input $X$ that predicts  the output $Y$. (In statistics, $Y$ is often called the dependent variable, and $X$ the independent variable.) We will, as always, assume that all the variables mentioned in this chapter are defined on a fixed probability space $(\Om, \myP)$. We assume that $X: \Om\to \CR_X$, where $\CR_X$ is the input space, and $Y: \Om \to \CR_Y$, where $\CR_Y$ is the output space. The input-output relationship is therefore captured by an unknown function $f:\CR_X\to\CR_Y$, the predictor.

The following two subclasses of prediction problems are important enough to have learned their own names and specific literature.
\begin{enumerate}[label=$\bullet$,wide]
\item Quantitative output: $\CR_Y = \mR^q$ (often with $q=1$). One then speaks of a {\em regression problem}.
\item Categorical output: $\CR_Y = \{g_1, \dots, g_q\}$ is a finite set. One then speaks of a {\em classification problem}.
\end{enumerate}
In most cases, the input space is Euclidean, i.e., $\CR_X = \mR^d$. Note also that, in classification, instead of a function $f: \CR \to \CR_Y$, one sometimes estimates a function $f:\CR_X \to \boldsymbol\Pi(\CR_Y)$, where $\boldsymbol\Pi(\CR_Y)$ is the space of probability distributions on $\CR_Y$. We will return to this in \cref{rem:pred.prob}.

The quality of a prediction is assessed through the definition of a {\em risk function}.
Such a function, denoted $r$, is defined on $\CR_Y \times \CR_Y$, takes values in $[0, +\infty)$ and should be understood as
\begin{equation}
\label{eq:risk.def}
r(\text{True output}, \text{Predicted output}),
\end{equation}
so that  $r(y,y')$ assigns a cost to the situation in which a true $y$ is predicted by $y'$. 
Note that this definition is asymmetric, and there is no requirement that $r(y,y') = r(y',y)$. It is important to remember our convention that the first variable is the true observation and the second one is a place-holder for a prediction.  
Risk functions are also called {\em loss functions}, or  simply {cost functions} and we will use these terms as synonyms.

The goal in prediction is to minimize the {\em expected risk}, also called the {\em generalization error}:
\[
R(f) = \myE(r(Y, f(X))).
\]

We will prove that an optimal $f$ can be easily described based on the joint distribution of $X$ and $Y$ (which is, unfortunately, never available). We will need for this to use conditional expectations and conditional probabilities, as defined in \cref{sec:cond.prob.elem,sec:cond.prob.gen}.

\section{Bayes predictor}
\label{sec:bayes.pred}
\begin{definition}
\label{def:bayes.pred}
A Bayes predictor is a measurable function $f: \CR_X \to \CR_Y$ such that, for all $x\in \CR_X$ (up to a $P_X$ negligible set), 
\[
\myE\big(r(Y, f(x))\mid X=x\big) = \min\big\{\myE\big(r(Y, y')\mid X=x\big): y'\in \CR_Y\big\}.
\]
\end{definition}
There can be multiple Bayes predictors if the minimum in the proposition is not uniquely attained.
Note that, if $f^{*}$ is a Bayes predictor and $\hf$ any other predictor, we have, by definition 
\[
\myE\big(r(Y, f^{*}(X))\mid X\big) \leq \myE\big(r(Y, \hf(X))\mid X\big).
\]
Passing to expectations, this implies $R(f^{*}) \leq R(\hf)$. We therefore have the following result:
\begin{theorem}
\label{th:bayes.pred}
Any Bayes predictor 
 $f^{*}$ is optimal, in the sense that it minimizes the generalization error $R$.
 \end{theorem}

\begin{enumerate}[label= Example \arabic*., wide=0pt, font=\itshape]
\item {\it Regression with mean-square error.} When $\CR_X = \mR^d$ and $\CR_Y = \mR^{q}$, the most common  risk function is the squared norm of the difference, $r(y,y') = |y-y'|^{2}$. 
The resulting generalization error is called the MSE (mean square error) and given by $R(f) = \myE(|Y-f(X)|^{2})$.
The Bayes predictor is such that $f^{*}(x)$ minimizes 
\[
t \mapsto \myE(|Y-t|^{2}\mid X=x).
\]
Let $f^{*}(x) = E(Y\mid X=x)$ and write
\begin{align*}
\myE(|Y-t|^{2}\mid X=x) =& \myE(|Y-f^*(x)|^{2}\mid X=x) + 2 \myE((Y-f^*(x))^T(f^*(x)-t)\mid X=x) \\
&+ |f^*(x) - t|^2\\
=& \myE(|Y-f^*(x)|^{2}\mid X=x) + 2 \myE((Y-f^*(x))^T\mid X=x)(f^*(x)-t) \\
&+ |f^*(x) - t|^2\\
=& \myE(|Y-f^*(x)|^{2}\mid X=x) + |f^*(x) - t|^2.
\end{align*}
This proves that $\myE(Y\mid X=x)$ is the unique Bayes classifier (up to a modification on a set of probability 0).
\label{item:Bayes.MSE}
\item {\it Classification with zero-one loss.}
Let $\CR_X = \mR^d$ and $\CR_Y$ be a finite set. The zero-one loss function is defined by  $r(y,y') = 1$ if $y\neq y'$ and 0 otherwise.
From this, it results that the generalization error is the probability of misclassification $R(f) = P(Y\neq f(X))$ (also called the misclassification error).

The Bayes predictor is such that $f^{*}(x)$ minimizes 
\[
g \mapsto \myP(Y\neq g\mid X=x) = 1 - \myP(Y = g\mid X=x).
\]
It is therefore given by the so-called {\it posterior  mode}:
\[
f^{*}(x) = \mathrm{argmax}_{g} \myP(Y = g\mid X=x).
\]
\end{enumerate}

\begin{remark}
\label{rem:pred.prob}
As mentioned at the beginning of the chapter, one sometimes replaces a pointwise prediction of the output by a probabilistic one, so that $f(x)$ is a probability distribution on $\CR_Y$. If $A$ is a (measurable) subset of $\CR_Y$, we will write $f(x, A)$ rather than $f(x)(A)$. 

In such a case, the loss function, $r$, is defined on $\CR_Y \times \boldsymbol\Pi(\CR_Y)$, and the expected risk is still defined by
$\myE(r(Y, f(X)))$.

It is quite natural to require that $\pi \mapsto r(y, \pi)$ is minimized for $\pi = \delta_y$. For classification problems, where $\CR_Y$ is finite, one can choose
\begin{equation}
\label{eq:risk.mle}
r(y, \pi) = - \log \pi(y),
\end{equation}
which satisfies this property.
The Bayes estimator is then a minimizer of $\pi \mapsto -\myE(\log\pi(Y)\mid X=x)$. The solution is (unsurprisingly) $f(x,y) = \myP(Y=y \mid X=x) $ since we always have
\[
-\myE(\log\pi(Y)\mid X=x) = -\sum_{y\in\CR_Y} \log\pi(y) f(x,y) \geq  -\sum_{y\in\CR_Y} \log f(x,y) f(x,y).
\]
The difference between these terms is indeed
\[
\sum_{y\in\CR_Y} \log \frac{f(x,y)}{\pi(y)} f(x,y) = \KL(f(x, \cdot)\| \pi) \geq 0.
\]

For regression problems, with $\CR_Y = \mR^q$, one can choose
\[
r(y, \pi) = \int_{\mR^q} |z - y|^2 \pi(dz)
\]
which is indeed minimal when $\pi$ is concentrated on $y$. Here, the Bayes estimator minimizes (with respect to $\pi$)
\[
\int_{\mR^q} \int_{\mR^q} |z - y|^2 \pi(dz) P_Y(dy\mid X=x) = \int_{\mR^q} \left(\int_{\mR^q} |z - y|^2 P_Y(dy\mid X=x)\right) \pi(dz) 
\]
where $P_Y(\cdot\mid X=x)$ is the conditional distribution of $Y$ given $X=x$. For any $z$, one has
\[
\int_{\mR^q} |y - z|^2 P_Y(dy\mid X=x) \geq \int_{\mR^q} |y - \myE(Y\mid X=x)|^2 P_Y(dy\mid X=x),
\]
which shows that the Bayes estimator is, in this case, the Dirac measure concentrated at $\myE(Y\mid X=x)$.
\end{remark}

\section{Examples: model-based approach}
Bayes predictors are never  available in practice, because the
true distribution of $(X,Y)$, or that of $Y$ given $X$, are unknown. These distributions can only be inferred from observations, i.e., from a training set: $T = (x_1, y_1,
\ldots, x_N, y_N)$. 

This is the approach followed by {\em model-based}, or {\em generative} methods, namely using training data to approximate the joint distribution of $X$ and $Y$ before using the Bayes estimator derived from this model for prediction. We now illustrate this approach with a few examples.

\subsection{Gaussian models and naive Bayes}
Consider a regression problem with $\CR_Y = \mR$, and model the joint distribution of $(X, Y)$ as a $(d+1)$-dimensional Gaussian distribution with mean $\mu$ and covariance matrix $\Sigma$, which must be estimated from data.
Write $\mu =\begin{pmatrix} m \\\mu_{0}\end{pmatrix}$, with $\mu_0\in\mR$, $m\in \mR^{d}$ and $\Sig$ in the form, for some symmetric matrix $S$ and $d$-dimensional vector $u$,
\[
\Sig = \begin{pmatrix} S  & u\\  u^{T} & \sig^{2}_{00} \end{pmatrix}.
\]
Then,  letting $\Delta = \sig^{2}_{00} - u^{T }S^{-1}u$,
\[
\Sig^{{-1}} = \frac1 \Delta \begin{pmatrix}\Delta S^{-1} +  S^{-1}uu^{T}S^{-1} & -S^{-1}u \\  - u^{T}S^{-1} & 1 \end{pmatrix}\,.
\]

This shows that the joint p.d.f. of $(X,Y)$ is proportional to
\begin{multline*}
\exp\Big(-\frac1{2\Delta} \big((y-\mu_{0})^{2} - 2 u^{T}S^{-1} (x-m)(y-\mu_{0})
+ \text{(\it terms not depending on $y$)} \big)\Big).
\end{multline*}
In particular
\[
\myE(Y\mid X=x) = \mu_{0} + u^{T} S^{-1}(x-m),
\]
which provides the least-square linear regression predictor. (In this expression, $u$ is the covariance between $X$ and $Y$ and $S$ is the covariance matrix of $X$.)

If one restricts the model to having  a diagonal covariance matrix $S$, then
\[
\myE(Y\mid X=x) = \mu_{0} + \sum_{j=1}^{d} \frac{\pe u j }{s_{jj}} (\pe x j  -\pe  m j ).
\]
This predictor is often called the {\em naive Bayes} predictor for regression.

\subsection{Kernel regression}
Let $\CR_X = \mR^{d}$ and $\CR_Y = \mR$.
Let $K_{1}: \mR^{d} \to \mR$ and $K_{2}: \mR \to \mR$ be two kernels, therefore satisfying
\[
\int_{\mR^{d}} K_{1}(x) dx = \int_{\mR} K_{2}(x) dx = 1;\quad \int_{\mR^{d}} xK_{1}(x) dx = \int_{\mR} yK_{2}(y) dy = 0.
\]
Let $K(x,y) = K_{1}(x) K_{2}(y)$ so that
\[
\int_{\mR^{d+1}} K(x,y) dydx = 1, \quad
\int_{\mR^{d+1}} yK(y,x) dydx = 0, \quad
\int_{\mR^{d+1}} xK(y,x) dydx = 0.
\]

The kernel estimator of the  joint p.d.f., $\phi$, of $(X,Y)$ at
scale $\sig$ is, in this case:
$$
\hat\phi(x,y) = \frac{1}{N} \sum_{k=1}^N \frac{1}{\sig^{d+1}}
K_1\left(\frac{x-x_k}{\sig}\right)K_2\left(\frac{y-y_k}{\sig}\right)\, .
$$
Based on $\hat\phi$, the conditional expectation of $Y$ given $X$ is
$$
\hf(x) = \frac{\frac{1}{N} \sum_{k=1}^N \frac{1}{\sig^{d+1}} \int_{\mR} y
K_1\left(\frac{x-x_k}{\sig}\right)K_2\left(\frac{y-y_k}{\sig}\right)
dy}{\frac{1}{N} \sum_{k=1}^N \frac{1}{\sig^{d+1}} \int_{\mR} 
K_1\left(\frac{x-x_k}{\sig}\right)K_2\left(\frac{y-y_k}{\sig}\right) dy}\, .
$$

Using the fact that
$\sig^{-1}\int_{\mR} y
K_2\left(\frac{y-y_k}{\sig}\right) dy = y_k$, we can simplify this expression to obtain
$$
\hf(x) = \frac{ \sum_{k=1}^N  y_k 
K_1\left(\frac{x-x_k}{\sig}\right)}{ \sum_{k=1}^N 
K_1\left(\frac{x-x_k}{\sig}\right)} \,.
$$
This the kernel-density regression estimator \citep{nadaraya1964estimating,watson1964smooth}.

\subsection{A classification example}
Let $\CR_Y = \{0,1\}$ and assume $\CR_X = \mathbb N =  \{0,1,2,\ldots\}$.
Let $p = \myP(Y=1)$ and assume that conditionally to $Y=g$, $X$ follows a Poisson distribution with mean $\mu_{g}$. Assume that $\mu_{0 }< \mu_1$.

The posterior distribution of $Y$ given $X=x$ is\,\footnote{$\propto$ is the notation for ``proportional to.''}
\[
\myP(Y=g\mid X=x) \propto \left\{
\begin{aligned}
(1-p)\mu_{0}^{x}e^{-\mu_0} \text{ if } g=0\\
p\mu_{1}^{x}e^{-\mu_1} \text{ if } g=1
\end{aligned}
\right.
\]

A Bayes classifier is then provided by taking $f(x) = 1$ if
\[
\log p + x\log \mu_1 - \mu_1 \geq \log(1-p) + x\log\mu_0 - \mu_0,
\]
that is:
\[
x \log\frac{\mu_1}{\mu_0} \geq \log \frac{1-p}{p} + \mu_1 - \mu_0.
\]
Since we are assuming that $\mu_1 > \mu_0$, we find that $f(x) = 1$ if\,\footnote{$\lceil x \rceil$ is the smallest integer larger than $x$ (ceiling).}
\[
x \geq \left\lceil \frac{\log ((1-p)/p) +\mu_1 -\mu_0}{\log(\mu_1/\mu_0)}\right\rceil
\]
and 0 otherwise.

\section{Empirical risk minimization}
\label{sec:emp.risk}
\subsection{General principles}
Model-based approaches for prediction are based on the estimation of the joint distribution of the input and output variables, which is arguably a harder problem than prediction \citep{vapnik1998statistical}. Since the goal is to find $f$ minimizing the expected risk $R(f) = \myE(r(Y, f(X))$, one may prefer a direct approach and consider the minimization of an empirical estimate of this risk, based on training data $T  = (x_1, y_1, \ldots, x_N, y_N)$, namely
\[
\hat R(f) = \frac{1}{N} \sum_{k=1}^N r(y_k, f(x_k)).
\]
This strategy is  called {\em empirical risk minimization.}

Importantly, $\hat R$ must be minimized over a restricted class, $\CF$, of predictors to avoid overfitting. 
For example,
with $\CR_Y=\mR$ and $\CR_X = \mR^{d}$, one can take
\[
\CF = \defset{f: f(x) = \be_0 + \sum_{i=1}^d \pe b i  \pe x i: \ \be_0, \pe b 1  
\ldots, \pe b d \in\mR}\,.
\]
Minimizing the empirical mean-square error 
\[
\hat R(f)  = \frac1N\sum_{k=1}^{N} (y_{k} - f(x_{k}))^{2}
\]
over $f\in \CF$ leads to the standard least-square regression estimator.

As another example, one can choose
\[
\CF = \defset{f: f(x) = \sum_{j=1}^p w_j \psi\left(\be_{j0} + \sum_{i=1}^d
\be_{ji} \pe x i \right), w_j, \be_{ji}\in\mR}\,.
\]
with a fixed function $\psi$. This corresponds to a two-layer perceptron model.

As a last example for now (we will see many others in the rest of this book), taking $d=1$, the set
\[
\CF = \defset{f: \int_{\mR} f''(x)^{2} dx < \mu}
\]
(with $\mu>0$) provides an infinite dimensional space of predictors, 
which leads to spline regression.

\subsection{Bias and variance}
We give a further illustration of the bias-variance dilemma in the regression case, using the mean-square error and taking $q=1$ to simplify. Denote the Bayes predictor by $f^*(x) = \myE(Y\mid X=x)$.

Fix a function space $\CF$, and let $\hf^{*}$ be the optimal predictor in $\CF$, in the sense that it minimizes $E(|Y- f(X)|^{2})$ over $f\in \CF$. Then, letting $\hf_N\in \mathcal F$ denote an estimated predictor,
\begin{eqnarray*}
R(\hf_N) &=& \myE(|Y- \hf_N(X)|^2) \\
&=& \myE(|Y- \hf^*(X)|^2) + \myE(|\hf_N(X) - \hf^*(X)|^2) +
2\myE((Y-\hf^*(X))(\hf^*(X) - \hf_N(X)).
\end{eqnarray*}

Let us make the assumption that there exists $\ep>0$ such that $f_{\la} = \hf^{*} + \la(\hf_{N} - \hf^{*})$ belongs to $\CF$ for  $\la \in [-\ep,\ep]$. This happens when $\CF$ is  a linear space, or more generally when $\CF$ is convex and $\hf^*$ is in its relative interior (see \cref{sec:optim}). Let $\psi: \la \mapsto \myE(|Y- f_\lambda(X)|^2)$, which is minimal at $\la =0$.  We have
 \begin{align*}
 \psi(\la) = &  \myE(|Y- \hf^{*}(X) - \la(\hf_{N}(X) - \hf^{*}(X)) |^2)\\
  = &  \myE(|Y- \hf^{*}(X)|^2)  - 2\la \myE((Y- \hf^{*}(X))(\hf_{N}(X) - \hf^{*}(X)) ) + \la^2 \myE(|\hf_{N}(X) - \hf^{*}(X)) |^2)
  \end{align*}
and
\[
0 = \psi'(0) = 2\myE((Y-\hf^*(X))(\hf^*(X) - \hf_{N}(X))).
\]
 We therefore get the identity
\[
R(\hf_N) =  \myE(|Y- \hf^*(X)|^2) + \myE(|\hf_N(X) - \hf^*(X)|^2) = \text{``Bias''} + \text{``Variance''}.
\]

The bias can be further decomposed as 
\[
\myE(|Y- \hf^*(X)|^2) = \myE(|Y- f^*(X)|^2) + \myE(|f^{*}(X) - \hf^*(X)|^2)
\]
because $f^*$ is the conditional expectation.
 As a result, we obtain an expression the generalization error with three contributions, namely,
\[
R(\hf_N) \leq \myE(|Y- f^*(X)|^2) + \myE(|f^{*}- \hf^*(X)|^2) + \myE(|\hf_N(X) - \hf^*(X)|^2).
\]
The first term is the Bayes error. It is fixed by the joint distribution of $X$ and $Y$ and measures how well $Y$ can be approximated by a function of $X$. The second term compares $f^{*}$ to its best approximation in $\CF$, and is therefore reduced by taking larger model spaces. The last term is the
error caused by using the data to estimate $\hf^*$. It increases with the size of $\CF$. This is illustrated in Figure \ref{fig:bias_var}.

\begin{figure}
\centering
\begin{tikzpicture}
\node[style={rectangle, draw=red,  very thick, minimum width=5cm, minimum height=10cm}]  (left) {};
\node[style={rectangle, draw=green, very thick, minimum width=5cm, minimum height=10cm}]  (right) [right=5cm of left] {};
\node[style={circle, minimum size=0.25cm, fill=blue}] [above right = -3cm of left] (pdata) {};
\node[style={circle, minimum size=0.25cm}] [left=0.1cm of pdata] {\Large $\hat P$};
\node[style={circle, minimum size=0.25cm, fill=blue}] [below = 6cm of pdata] (ptrue) {};
\node[style={circle, minimum size=0.25cm}] [left=0.1cm of ptrue] {\Large $P^*$};

\node[style={circle, minimum size=2.5cm, draw=black}] [above left=-3cm and -3 cm of right] (F) {$\CF$} ;
\node[style={circle, minimum size=0.15cm, fill=black}] [left=0cm of F] (hf0) {} ;
\node [above=0cm of hf0] (hf) {$\hf$} ;
\node[style={circle, minimum size=0.15cm, fill=black}] [below left=0cm of F] (hfs0) {} ;
\node [below=0cm of hfs0] (hfs) {$\hf^*$} ;
\node[style={circle, minimum size=0.15cm, fill=black}] [below left=-3cm and -2cm of right] (fs0) {} ;
\node [below=0cm of fs0] (fs) {$f^*$} ;

\node [below = 0cm of left] {\Large Probability space};
\node [below = 0cm of right] {\Large Predictor space};

\draw[red, thick] (pdata.east) -- (hf0.west) {} ;
\draw[red, thick] (ptrue.east) -- (hfs0.west) {} ;
\draw[red, thick] (ptrue.east) -- (fs0.west) {} ;
\end{tikzpicture}
\caption{\label{fig:bias_var} Sources of errors in statistical Learning: When $P^*$ is the distribution of the data, the optimal predictor $f^*$ minimizes the expected loss function. Based on data $Z_1, \ldots, Z_N$, the sample-based distribution is $\hat P = (\de_{Z_1} + \cdots + \de_{Z_N})/N$ and the empirical loss is minimized over a subset $\CS$ of the space of all possible estimators. The expected discrepancy between the resulting estimator and the  one minimizing the true expected loss on the subspace  is the ``variance'' of the method, and the expected discrepancy between this subspace-constrained estimator and  and the optimal one is the ``bias.''
}
\end{figure}

\begin{remark}
\label{rem:bias}
If the assumption made on $\hf^*$ is not valid, one can  write
\[
R(\hf_N)= \myE(|Y- \hf_N(X)|^2) \leq  2\big(\myE(|Y- \hf^*(X)|^2) + \myE(|\hf_N(X) - \hf^*(X)|^2)\big) 
\]
and still obtain a control (as an inequality) of the generalization error by a bias-plus-variance sum.
\end{remark} 

%
%

\section{Evaluating the  error}
\label{sec:error}

\subsection{Generalization error}
Given input and output variables $X:\Om\to \CR_X$ and $Y:\Om\to\CR_Y$ and a risk function $r: \CR_Y\times \CR_Y \to [0. +\infty)$, we have defined the generalization (or prediction) error as
$$
R(f) = \myE(r(Y, f(X)))\,.
$$
Recall that a training set $T = ((x_1, y_1), \ldots, (x_N, y_N))$ is a realization $T = \mT(\om)$ of  the random variable $\mT=((X_1, Y_1), \ldots, (X_N,
Y_N))$, an i.i.d. sample of the joint distribution of $(X,Y)$. A {\em learning algorithm} is a function $T \mapsto \hf_{T}$ defined on the set of training sets, namely, $\bigcup_{N=1}^\infty (\CR_X\times\CR_Y)^{N}$ and taking values in $\CF$.

For a given $T$ and a specific algorithm, one is primarily interested in evaluating $R(\hf_{T})$, the generalization error of the predictor estimated from observed data.  To emphasize the fact that the training set is fixed in this expression, one often writes:
\[
R(\hf_{T}) = \myE(r(Y, \hf_{\,\mT}(X)) \mid \mT = T)
\]

If we  also take the expectation with respect to $T$ (for fixed $N$), we obtain the averaged generalization risk as
\[
\myE(R(\hf_{\,\mT})) = \myE(r(Y, \hf_{\,\mT}(X))),
\]
which provides an evaluation of the average quality of the algorithm when evaluated on random training sets of size $N$.
If $A: T \mapsto \hf_{T}$ denotes the learning algorithm, we will denote $\overline R_{N}(A) = \myE(R(\hf_{\,\mT}))$.

Since their computation requires the knowledge of the joint distribution of $X$ and $Y$, these errors are not available in practice. Given a training set $T$ and a predictor $f$, one can compute the empirical error
\[
\hat R_T(f) = \frac{1}{N} \sum_{k=1}^N r(y_k, f(x_k))\,.
\]
Under the usual moment conditions, the law of large numbers implies that $\hat R_{\,\mT}(f) \to R(f)$ with probability one for any given predictor $f$.  However, the law of large numbers cannot be applied to assess whether the {\em in-sample error},
\[
\CE_T \defeq \hat R_T(\hf_T) = \frac{1}{N} \sum_{k=1}^N r(y_k, \hf_T(x_k)),
\]
is a good approximation of the generalization error $R(\hf_T)$. This is because each term in the sum depends on the full data set, so that $\CE_\mT$ is not a sum of independent terms. The in-sample error typically under-estimates the generalization error, sometimes with a large discrepancy.

When one has enough data, however, it is possible to set some of it aside to form a test set.
Formally, a test set is a collection $T' = (x'_1, y'_1, \ldots, x'_{N'}, y'_{N'})$ considered as a realization  of an i.i.d. sample of $(X,Y)$, $\mT' = (X'_1, Y'_1, \ldots, X'_{N'}, Y'_{N'})$, independent of $\mT$.
 The test set error is then given by
\[
\CE_{T, T'}  = \hat  R_{T'}(\hf_T) = \frac{1}{N'} \sum_{k=1}^{N'} r(y'_k, \hf_T(x'_k)).
\]
The law of large numbers (applied conditionally to $\mT = T$) implies that $\CE_{T,\mT'}$ converges to $R(\hf_{T})$ with probability one when $N'\to \infty$.

However, in many applications, data acquisition is difficult or expensive (e.g., in the medical field) and sparing a part of it in order to form a test set is not a reasonable option. In such situations, cross-validation is generally a preferred alternative.

\subsection{Cross validation}

\subsubsection{Cross-validation error}
The $n$-fold cross-validation method (see, e.g., \citet{stone1974cross}) separates the training set into $n$ non-overlapping
sets of equal sizes, and estimates $n$ predictors by leaving out one
of these subsets as a temporary test set. A generalization error is 
estimated from each test set and averaged over the $n$ results.  

Let us formalize this computation after introducing some notation. We represent training data in the form $T = (z_1, \ldots, z_N)$, a sample of a random variable $Z$. With this notation, we can include supervised problems, such as prediction (taking $Z = (X,Y)$) and unsupervised ones such as density estimation (taking $Z=X$). One tries to estimate a function $f$ within a given class (e.g., a predictor, or a density) and one has a measure  of ``loss'', denoted $\ell(f, z)\geq 0$ measuring how badly  $f$ performs on the data $z$. For prediction, one takes $\ell(f, z) = r(y, f(x))$ with $z = (x,y)$ and for density estimation, e.g., $\ell(f, z) = - \log f(z)$, the negative log-likelihood. One then lets $R(f) = E(\ell(f, Z))$.  For an algorithm $A: T\mapsto \hat f_T$, the loss $\bar R(A)$ is the quantity of interest.

Given another training set $T' = (z'_1, \ldots, z'_{N'})$, the empirical loss on $T'$ is 
\[
\hat R_{T'}(f) = \frac1{N'} \sum_{k=1}^{N'} \ell(f, z_k')
\]
and, using $T$ as a training set and $T'$ as a test set, we let, as above,
\[
\CE_{T,T'} = \hat R_{T'}(\hf_T).
\]

To define an $n$-fold cross-validation estimator of the error, one assumes that the training set $T$ is partitioned into $n$ subsets of equal sizes (up to one element if $N$ is not a multiple of $n$), $T_1$, \ldots, $T_n$, so that $T_i$ and $T_j$ are non-intersecting if $i\neq j$, and $T = \bigcup_{i=1}^n T_i$.  For each $i$, let $T^{(i)} = T\setminus T_i$, which provides the training data with the elements of $T_i$ removed. Then, the $n$-fold cross-validation error is defined by
\[
\CE_{\mathrm{CV}}(T) = \frac1n \sum_{i=1}^{n} \CE_{T^{(i)}, T_i}\,.
\]

Assuming, to simplify, that $N$ is a multiple of $n$, the expectation of the cross-validation error is $E( R (\hat f_{\mT_{N'}}))$, where the average is made over training sets $\mT_{N'}$ of size $N' = N-N/n$. Note that the cross-validation error is an estimate of the average error of the algorithm over random training sets (of fixed size, $N'$), not necessarily that of the current estimator $\hf_T$. It returns an evaluation of the algorithm $A: T\mapsto \hf_T$. When needed, one can emphasize this and write $\bar R_{\mathrm{CV}, T}(A)$. Since $N' \leq N$ and accuracy generally improves with the size of the training set, cross-validation typically over-estimates (on average) the error for the number of available training samples. 

The limit case when $n=N$ is called
leave-one-out (LOO) cross validation. In this case $\CE_{\mathrm{CV}} $ is an almost unbiased estimator of $E( R (\hf_{\boldsymbol T}))$, but, 
because it is an average of functions of the training set that are quite similar (and that will therefore be positively correlated), its variance (as a function of $T$) may be quite large. Conversely, smaller values of $n$ will have smaller variances, but larger biases. In practice, it is difficult to assess which choice of $n$ is optimal, although 5- or 10-fold cross-validation is quite popular. LOO cross-validation is also often used, especially when $N$ is small. 

\subsubsection{Model selection using cross validation}

Because it evaluates the quality of an algorithm, cross-validation is often used to perform model selection. Indeed, many learning algorithms depends on a parameter, that we will denote $\la$. In kernel density estimation,  for example, $\la=\sig$ is the kernel width. For mixtures of Gaussian, $\la =m$ is the number of Gaussian terms in the mixtures. Formally, this means that one has, for every $\la$, an algorithm $A_\la: T \mapsto \hf_{T, \la}$.

Fixing a training set $T$, one can compute, for each $\la$,  the  cross-validation error $e_T(\la) = \bar R_{\mathrm{CV}, T}(A_\la)$. Model selection is then performed by finding 
\[
\la^*(T) = \argmin_\la e_T(\la). 
\]
Once this $\la^*$ is obtained, the final estimator is $\hf_{T, \la^*(T)}$, obtained by rerunning the algorithm one more time on the full training set. 

This defines a new training algorithm, $A^*: T \mapsto \hf_{T, \la^*(T)}$. It is a common mistake to consider that the cross-validation error associated to this algorithm is still given by $e(\la^*(T))$. This is false, because the computation of $\la^*$ uses the full training set. To compute the cross-validation error of $A^*$, one needs to encapsulate this model selection procedure in an other cross-validation loop. So, one needs to compute, using the previous notation,
\[
\CE^*_{\mathrm{CV}}(T) = \frac1n \sum_{i=1}^n \hat R_{T_i}(\hf_{T^{(i)}, \la^*(T^{(i)})})
\]
where each $\hf_{T^{(i)}, \la^*(T^{(i)})}$ is computed by running a cross-validated model selection procedure restricted to $T^{(i)}$. This is often called a double-loop cross-validation procedure (the number of folds in the inner and outer loops do not have to coincide). 
Note that each $\la^*(T^{(i)})$ that does not necessarily coincide with the optimal $\la^*(T)$ obtained with the full training set.

\newpage
\markright{PROBLEMS}
\section*{Problems}
\addcontentsline{toc}{section}{Problems}
\input Problems_Bayes_Rule

\chapter{Inner Products and Reproducing Kernels}
\label{chap:higher}

\section{Introduction}
We will discuss later in this book various methods that specify the prediction is  as a linear function of the input. These methods are often applied after taking transformations of the original variables, in the form $x\mapsto h(x)$ (i.e., the prediction algorithm is applied to $h(x)$ instead of $x$). We will refer to $h$ as a ``feature function,'' which typically maps the initial data $x\in \CR$ to a vector space, sometimes of infinite dimensions, that we will denote $H$ (the ``feature space''). 

The present chapter provides a formal description of this framework, focusing, in particular, on situations in which $H$ has an inner product, as this inner product is often instrumental in the design of linear methods on $H$. Many machine learning methods can indeed be expressed either as functions of the coordinates of the input data in some space, or as functions of the inner products between the input samples. Such methods can bypass the difficulty of using high-dimensional features with the help of the theory of ``reproducing kernels,'' \citep{aronszajn1950theory,wahba1990spline} which ensures that the inner product between special classes of feature functions $h(x)$ and $h(x')$ can be explicitly computed as a function of $x$ and $x'$.

\section{Basic Definitions}

\subsection{Inner-product spaces}
We recall that a real vector space\,\footnote{All vector spaces in these notes will be real, and will therefore only be referred as vector spaces.} is a set, $H$, on which an addition and a scalar product are defined, namely $(h,h') \in H\times H \mapsto h+h'\in H$ and $(\lambda, h) \in \mR \times H \mapsto \la h\in H$, and we assume that the reader is familiar with the theory of finite-dimensional spaces. 

An inner product on a vector space $H$ is a bilinear function, typically denoted $(\xi,\eta) \mapsto \scp{\xi}{\eta}$ such that $\scp{\xi}{\xi} \geq 0$ with $\scp{\xi}{\xi} = 0$ if and only if $\xi=0$. A vector space equipped with an inner product is called an inner-product space. We will often denote the inner product with a subscript referring to the space (e.g., $\scp{\cdot}{\cdot}_H$). Given such a product, the function
\[
\xi \mapsto \|\xi\|_H = \sqrt{\scp{\xi}{\xi}_H}
\]
is a norm, so that $H$ is also a normed space (but not all normed spaces are inner-product spaces)\,\footnote{Note that we are using double bars for the norm in $H$, which, in most applications, is infinite dimensional}.  

When a normed space is {\em complete} with respect to the topology induced by its norm, it is called a Banach space, or a Hilbert space when the norm is associated with an inner product. Completeness means that Cauchy sequences in this space always have a limit, i.e., if the sequence $(\xi_n)$ is such that, for any $\ep>0$, there exists $n_0>0$ such that $\|\xi_n-\xi_m\|_H < \ep$ for all $n,m\geq n_0$, then there exists $\xi$ such that $\|\xi_n - \xi\|_H \to 0$. Completeness is a very natural property. It allows, for example, for the definition of integrals such as $\int h(t) dt$ as limits of Riemann sums for suitable functions $h: \mR \to H$, leading (with more general notions of integrals) to proper definitions of expectations of $H$-valued random variables. Using a standard (abstract) construction, one can prove that any normed space (resp. inner-product) can be extended to a Banach (resp. Hilbert) space within which it is dense. 

Note that finite-dimensional normed spaces are always complete.

\subsection{Feature spaces and kernels}

Now, consider an input set, say $\CR$, and a mapping $h$ from $\CR$ to $H$, where $H$ is an inner product space. For us, $\CR$ is the set over which the original input data is observed, typically $\mR^d$, and $H$ is the feature space.  One can define the function
$K_h: \CR\times\CR \to \mR$ by 
\[
K_h(x,y) = \scp{h(x)}{h(y)}_H.
\]
The function $K_h$ satisfies the following two properties.
\begin{enumerate}[label={[K\arabic*]}, wide=0pt]
\item $K_h$ is symmetric, namely $K_h(x,y) = K_h(y,x)$ for all $x$ and $y$ in $\CR$. 
\item For any $n>0$, for any choice of scalars $\la_1, \dots, \la_n\in \mR$ and any $x_1, \ldots, x_n \in \CR$, one has
\begin{equation}
\label{eq:pos.kernel}
\sum_{i,j=1}^n \lambda_i \lambda_j K_h(x_i, x_j) \geq 0.
\end{equation}
\end{enumerate}
The first property is obvious, and the second one results from the fact that one can write
\begin{equation}
\label{eq:kh.pos.def}
\sum_{i,j=1}^n \lambda_i \lambda_j K_h(x_i, x_j) = \sum_{i,j=1}^n \lambda_i \lambda_j \scp{h(x_i)}{h(x_j)}_H = \Big\|\sum_{i=1}^n \lambda_i h(x_i)\Big\|_H^2 \geq 0.
\end{equation}

This leads us to the following definition.
\begin{definition}
\label{def:pos.kernel}
A function $K: \CR \ti\CR \mapsto \mR$ satisfying  properties [K1] and [K2]
is called a positive kernel.

One says that
the kernel is positive definite if the sum in \cref{eq:pos.kernel} cannot vanish unless (i) $\la_1 = \cdots = \la_n=0$ or (ii) $x_i=x_j$ for some $i\neq j$.
\end{definition}

An equivalent definition of positive kernels can be given using kernel matrices, for which we introduce a notation.
\begin{definition}
\label{def:kernel.mat}
If $K: \CR \ti\CR \mapsto \mR$ is given, we define, for every $x_1, \ldots, x_n\in \CR$, the kernel matrix 
$\CK_K(x_1, \ldots, x_n)$ with entries $K(x_i, x_j)$, for $i,j = 1, \ldots, n$. (If $K$ is understood from the context, we will simply write $\CK(x_1, \ldots, x_n)$ instead of $\CK_K(x_1, \ldots, x_n)$.)
\end{definition}
Given this notation, it is clear that $K$ is a positive kernel if and only if  for all $x_1, \ldots, x_n\in \CR$, the matrix $\CK_K(x_1, \ldots, x_n)$ is symmetric, positive semidefinite. It is a positive definite kernel if $\CK_K(x_1, \ldots, x_n)$ is positive definite as soon as all $x_j$'s are distinct. This latter condition is obviously needed since, if $x_i=x_j$, the $i$th and $j$th columns of $\CK$ coincide and this matrix cannot be full-rank.

\begin{remark}
\label{rem:positive.kernel}
It is important to point out that $K$ being a positive kernel {\em does not require} that $K(x,y) \geq 0$ for all $x,y\in\CR$ (see examples in the next section). However, it does imply that $K(x,x) \geq 0$ for all $x\in \CR$, since diagonal elements of positive semi-definite matrices are non-negative.
\end{remark}

\bigskip

The function $K_h$ defined above is therefore always a positive kernel, but not always positive definite, as seen below. We will also see later that the converse statement is true: any positive kernel $K: \CR \ti\CR \mapsto \mR$ can be expressed as $K_{h}$ for some feature function $h$ between $\CR$ and some feature space $H$.

Given a feature function $h: \CR \to H$, we will denote by $V_h = \vspan(h(x), x\in\CR)$ the vector space generated by the features, which, by definition, is the space of all linear combinations
\[
\xi = \sum_{i=1}^n \la_i h(x_i)
\]
with $\la_1, \ldots, \la_m\in\mR$, $x_1, \ldots, x_n\in \CR$ and $n\geq 0$ (by convention, $\xi=0$ if $n=0$). Then $K_h$ is positive definite if and only if any family $(h(x_1), \ldots, h(x_n))$ with distinct $x_i$'s is linearly independent. This is a direct consequence of  \cref{eq:kh.pos.def}.
\[
\sum_{i,j=1}^n \lambda_i \lambda_j K_h(x_i, x_j) =  \left\|\sum_{i=1}^n \lambda_i h(x_i)\right\|_H^2.
\]
This implies in particular that positive-definite kernels over infinite input spaces $\CR$ can only be associated to infinite-dimensional spaces $H$, since $V_h\subset H$.

%
%
\section{First examples}

\subsection{Inner product}
Clearly, if $\CR$ is an inner product space, it has an associated reproducing kernel, defined by
\[
K(x,y) = \scp{x}{y}_{\CR}\,.
\]
This kernel is equal to $K_h$ with $H = \CR$ and $h=\id$ (the identity mapping). In particular $K(x,y) = x^Ty$ is a positive kernel if $\CR = \mR^d$. This kernel can obviously take positive and negative values.

Notice that this kernel is not positive definite, because the rank of $\CK(x_1, \ldots, x_n)$ is equal to the dimension of $\mathrm{span}(x_1, \dots, x_n)$, which can be less than $n$ even when the $x_i$'s are distinct. 

\subsection{Polynomial Kernels}

Consider $\CR = \mR^d$ and define 
\[
h(x) = (\pe x {i_1} \dots \pe x {i_k}, 1\leq i_1, \ldots, i_k \leq d),
\]
 which contains all products of degree $k$ formed from variables $\pe x 1, \ldots,\pe x d$, i.e., all monomials of degree $k$ in $x$. This function takes its values in the space $H = \mR^{N_k}$, where $N_k = d^k$. Using, in $H$, the inner product $\scp{\xi}{\eta}_H = \xi^T\eta$, we have
\begin{eqnarray*}
K_h(x,y)& =& \sum_{1\leq i_1, \ldots, i_k\leq d} (\pe x {i_1}\pe y {i_1})
\cdots (\pe x {i_k} \pe y {i_k})\\
&=& (x^Ty)^k.
\end{eqnarray*}
This provides  the homogeneous polynomial kernel of order $k$.

If one now takes all monomials of order less than or equal to $k$, i.e.,  
\[
h(x) = (\pe x {i_1} \dots \pe x {i_l}, 1\leq i_1, \ldots, i_l \leq d, 0\leq l\leq k),
\]
 which now takes values in a space of dimension $1 + d + \cdots + d^k$, the corresponding kernel is 
$$
K_h(x,y) = 1 + (x^Ty) + \cdots + (x^Ty)^k =
\frac{(x^Ty)^{k+1} - 1}{x^Ty - 1}\,.
$$
This provides a polynomial kernel of order $k$. It is important to notice here that, even though the dimension of the feature space increases exponentially in $k$, so that the computation of the feature function rapidly becomes intractable, the computation of the kernel itself remains a relatively mild operation. 

One can make variations on this construction. For example, choosing any family $c_0, c_1, \ldots, c_k$ of positive numbers, one can take 
\[
h(x) = (c_l \pe x {i_1} \dots \pe x {i_l}, 1\leq i_1, \ldots, i_l \leq d, 0\leq l\leq k)
\]
yielding
\[
K_h(x,y) = c_0^2 + c_1^2 (x^Ty) + \cdots + c_k^2 (x^Ty)^k.
\]
Taking $c_l = \bin{k}{l}^{1/2} \al^{l}$ for some $\al>0$, we get another form of polynomial kernel, namely,
\[
K_h(x,y) = (1 + \al^2 x^T y)^k.
\]

\subsection{Functional Features}
\label{sec:functional.feat}
We now consider an example in which $H$ is infinite dimensional.
Let $\CR = \mR^d$. We assume that a  function $s: \mR^d \to \mR$ is chosen, such that $s$ is both (absolutely) integrable and square integrable. We also fix a scaling parameter $\rho>0$. Associate to $x\in \mR^d$ the function 
\[
\xi_x: y \mapsto s((y-x)/\rho),
\] 
which is also square integrable (as a function of $y$). We define the feature function $h: x\mapsto \xi_x$ from $\mR^d$ to $H = L^2(\mR^d)$, the space of square integrable functions on $\mR^d$ with inner product 
\[
\scp{\xi}{\eta}_H = \int_{\mR^d} \xi(z)\eta(z) dz.
\]

The resulting kernel is
\Eq{
K_h(x,y) = \int_{\mR^{d}} s(z/\rho - x) s(z/\rho - y)\,dz = \rho^d \int_{\mR^{d}} s(z)s(z - (y-x)/\rho) \, dz.
}
Note that $K_h(x,y)$ is ``translation-invariant,'' which means that it only depends on $x-y$. It takes the form $K_h(x,y) = \rho^d \Gamma((y-x)/\rho)$ where
\Eq{
\Gamma(u) = \int_{\mR^{d}} s(z)s(z-u) \,dz.
}
is  the convolution\footnote{The convolution between two absolutely integrable functions $f$ and $g$ is defined by $f*g(u) = \int_{\mR^d} f(z) g(u-z)\,dz$} of $s$ with $\tilde s: z\mapsto s(-z)$.

Let $\sigma$ be the Fourier transform of $s$, i.e., 
\Eq{
\sigma(\om) = \int_{\mR^d} e^{-2i\pi \om^Tu} s(u) du.
}
Because $s$ is real-valued, we have $\sigma(-\omega) = \bar \sigma(\omega) $, the complex conjugate of $\sigma$. Moreover,
$\bar \sigma$ is also the Fourier transform of $\tilde s$. Using the fact that the Fourier transform of the convolution of two functions is the product of their Fourier transforms, we see that the Fourier transform of $\Gamma = s * \tilde s$ is equal to $|\sig|^2$. 
Applying the inverse transform, we find
\Eq{
\Gamma(u) = \int_{\mR^{d}} e^{2i\pi\om^{T}u } |\sigma(\om)|^{2}  \,d\om = \int_{\mR^{d}} e^{-2i\pi\om^{T}u } |\sigma(-\om)|^{2}  \,d\om\,.
}
This form is (almost) characteristic of translation-invariant kernels.
\bigskip

Let us consider a few examples of kernels that can be obtained in this way.
\begin{enumerate}[label=(\arabic*), wide=0pt]
\item
Take $d=1$ and let $s$ be the indicator  function of the interval $[-\frac12, \frac12]$.  
Then, one finds
\Eq{
\Gamma(t) = \max(1-|t|, 0)\,.
}
In this case, the space $V_h$ is the space of all functions expressed as finite sums
\Eq{
z\mapsto \sum_{j=1}^n \la_j \bfone_{[x_j - \rho/2, x_j+\rho/2]}(z)\,,
}
and therefore is a space of compactly-supported piecewise constant functions. 
Such a function computed with distinct $x_j$'s cannot vanish everywhere unless all $\la_j$'s vanish, so that $K_h$ is positive definite. 
Indeed, let
\[
f(z) = \sum_{j=1}^n \la_j \bfone_{[x_j - \rho/2, x_j+\rho/2]}(z)
\]
and assume without loss of generality that $x_1 < x_2 < \cdots < x_n$ and let $x_{n+1} = \infty$. Let $i_0$ be the smallest index $j$ such that $\la_j \neq 0$, assuming that such an index exists. Then $f(z) = \la_{i_0}>0$ for all $z\in [x_{i_0}-\rho/2, x_{i_0+1} - \rho/2)$ which is a non-empty interval.  So, if $f$ vanishes almost everywhere, we must have $\la_j=0$ for all $j=1, \ldots, n$.
\item Still with $d=1$, let $s(z) = e^{-|z|}$. Then, for $t>0$, 
\begin{align*}
\Ga(t) &= \int_{-\infty}^\infty e^{-|z|} e^{-|z-t|}\, dz\\
&= \int_{-\infty}^0 e^{z} e^{z-t}\, dz + \int_0^t e^{-z} e^{z-t}\, dz + \int_{t}^\infty e^{-z} e^{-z+t}\, dz\\
&= \frac{e^{-t}}2 + t e^{-t} + \frac{e^{-t}}{2}\\
&= (1 + t) e^{-t}
\end{align*}
Using the fact that $\Ga(-t) = \Ga(t)$ (make the change of variable $z\to -z$ in the integral), we get 
\[
\Ga(t) = (1 + |t|) e^{-|t|}.
\]
for all $t$. This shows that 
\[
K(x,y) = (1+|x-y|) e^{-|x-y|}
\] 
is a positive kernel on $\mR^d$.
\item Take $s(z) = e^{-|z|^2/2}$,  $z\in \mR^d$. Then 
\begin{align*}
\Ga(u) &= \int_{\mR^d} e^{-\frac{|z|^2 + |u-z|^2}2}\, dz
= e^{-\frac{|u|^2}4} \int_{\mR^d} e^{- |z-u/2|^2}\, dz\\
&= (4\pi)^{d/2} e^{-\frac{|u|^2}4}.
\end{align*} 
\end{enumerate}
This provides a special case of Gaussian kernel.

%
%
%
%

\subsection{General construction theorems}

\subsubsection{Translation invariance}
As introduced above, a kernel $K$ is translation invariant if it takes the form $K(x,y)  = \Ga(x-y)$ for some continuous function
$\Ga$ defined on $\mR^d$. 
Bochner's theorem \citep{bochner1933vorlesungen} states that such a $K$ is a positive kernel if and only if $\Ga$
is the Fourier transform of a positive measure, namely,
\Eq{
\Ga(x) = \int_{\mR^d} e^{-2i\pi\scp{x}{\om}} d\mu(\om)
}
where $\mu$ is a positive and symmetric (invariant by sign change) measure on $\mR^d$. For
example one can take $d\mu(\om)  = \nu(\om) d\om$, where $\nu$ is a integrable, positive and even function.

This theorem provides an at least numerical,
and sometimes  analytical, method for constructing kernels. The previous section exhibited a special case of translation-invariant kernel for which $\nu = |\sigma|^2$. 

\subsubsection{Radial kernels}
 A radial kernel takes the form 
 $K(x,y)
= \ga(|x-y|^2)$, for some continuous function $\ga$ defined on $[0,
+\infty)$. 
Shoenberg's theorem \citep{schoenberg1938metric} states that, if this function $\ga$ is
universally valid, i.e., $K$ is a kernel for all dimensions $d$, then, it must take the form
\Eq{
\ga(t) = \int_0^{\infty} e^{-\la t} d\mu(\la) 
}
for some positive finite measure $\mu$ on $[0, +\infty)$. 

 For example, when
$\mu$ is a Dirac measure, i.e., $\mu = \de_{(2a)^{-1}}$ for some $a>0$, then
$K(x,y) = \exp(-|x-y|^2/2a)$, which is the Gaussian kernel.
Taking $d\mu = e^{-a\la} d\la$ yields $\ga(t) = 1/(t+a)$, and $d\mu =
\la e^{-a\la}d\lambda$ yields $\ga(t) = 1/(a+t)^2$.

There is also, in \citet{schoenberg1938metric}, a characterization of radial
kernels for a fixed dimension $d$. Such kernels must take the form
$$
\ga(t) = \int_0^{+\infty} \Om_d(t\la) d\mu(\la)
$$
with $\Om_d(t) = \Ga(d/2) (2/t)^{(d-2)/2} J_{(d-2)/2}(t)$ where $J_{(d-2)/2}$ is Bessel's function of the first kind.

\subsection{Operations on kernels}

Kernels can be combined in several ways as described in the next proposition. 
\begin{proposition}
\label{prop:combine.K}
Let  $K_1: \CR\times\CR \to \mR$ and $K_2: \CR\times\CR \to \mR$ be positive kernels. Then the following assertions hold.
\begin{enumerate}[label=(\roman*),leftmargin=1cm]
\item  If  $\la_1, \la_2 >0$, $\la_1K_1+\la_2 K_2$ is a positive kernel. It is positive definite as soon as either $K_1$ or $K_2$ is positive definite.
\item For any function $f: \CR' \to \CR$, $K'_1(x',y') \defeq K_1(f(x'), f(y'))$ is a positive kernel. It is positive definite as soon as $K_1$ is positive definite and $f$ is one-to-one.
\item $K(x,y) = K_1(x,y)K_2(x,y)$ is a positive kernel. It is positive definite as soon as $K_1$ and $K_2$ are positive definite.
\item Let $K_1$ and $K_2$ be translation-invariant with $\CR = \mR^d$, taking the form $K_i(x,y) = \Ga_i(x-y)$, where $\Ga_i$ is continuous  ( $i=1,2$). Assume that one of the two functions $\Ga_1, \Ga_2$ is integrable on $\mR^d$. Then
\[
K(x,y) = \int_{\mR^d} K_1(x, z) K_2(z,y) dz 
\]
is also a positive kernel.
\end{enumerate}
\end{proposition}
\begin{proof}
Point (i) is obvious. Point (ii) is almost as simple, because, for any $\la_1, \ldots, \la_n\in \mR$ and $x'_1, \ldots, x'_n\in \CR'$,
\[
\sum_{i,j=1}^n \la_i\la_j K'_1(x'_i, x'_j) = \sum_{i,j=1}^n \la_i\la_j K_1(f(x'_i), f(x'_j))\geq 0.
\]
If $K_1$ is positive definite, then the latter sum can only vanish if all $\la_i$ are zero, or some of the points in $(f(x'_1), \ldots, f(x'_n))$ coincide. If, in addition, $f$ is one-to-one, then this is equivalent to all $\la_i$ are zero, or some of the points in $(x'_1, \ldots, x'_n)$ coincide, so that $K'_1$ is positive definite.

To prove point  (iii), take $x_1, \ldots, x_N\in \mR^d$ and form the matrices $\CK_i = \CK_i(x_1, \ldots, x_N)$, $i=1,2$, which are, by assumption positive semi-definite. The matrix $\CK = \CK(x_1, \ldots, x_N)$ is the element-wise (or Hadamard) product of $\CK_1$ and $\CK_2$, and the conclusion follows from the linear algebra result stating that the Hadamard product of two positive semi-definite (resp. positive definite) matrices $A = (a(i,j), 1\leq i,j\leq N)$ and $B= (b(i,j), 1\leq i,j\leq N)$ is positive semi-definite (resp. positive definite). This is proved by diagonalizing, say, $A$ in an orthonormal basis $u_1, \ldots, u_N$, with eigenvalues $\la_1, \ldots, \la_N$ and writing
\begin{align*}
\sum_{i,j=1}^N \pe \al i a(i,j)b(i,j)\pe \al j & = \sum_{i,j, k=1}^N \pe \al i \pe {u_i} k \pe {u_j} k \la_k b(i,j)\pe \al j\\
& = \sum_{k=1}^N \la_k \sum_{i,j=1}^N (\pe \al i \pe {u_i} k) (\pe \al j \pe {u_j}k)) b(i,j) \geq 0
\end{align*}
If $B$ is positive definite, then the sum above can be zero only if, for each $k$, either $\la_k=0$ or $\pe \al i \pe {u_i} k = 0$ for all $i$. If $A$ is also positive definite, then the only possibility is $\pe \al i \pe {u_i} k = 0$ for all $i$ and $k$, which implies $\pe \al i=0$ for all $i$ since $u_i\neq 0$. 

To prove point (iv)\,\footnote{This part of the proof uses some measure theory.}, we first note that a translation invariant kernel $K'(x,y) = \Ga'(x-y)$ is always bounded. Indeed, the matrix $\CK'(x,0)$ is positive semi-definite, with determinant $\Ga'(0)^2 - \Ga'(x)^2 >0$, showing that $|\Ga'(x)| < \Ga'(0)$. 
This shows that the integral defining $K(x,y)$ converges as soon as one of the two functions $\Ga_1$ or $\Ga_2$ is integrable. Moreover, we have 
$K(x,y) = \Ga(x-y)$ with
\[
\Ga(x) = \int_{\mR^d} \Ga_1(x-z) \Ga_2(z)\,dz = \int_{\mR^d} \Ga_1(x-u) \Ga_2(u-y)\,du
\]
Using the fact that both $\Ga_1$ and $\Ga_2$ are even, and making the change of variable $z\mapsto -z$, one easily shows that $\Ga(x) = \Ga(-x)$, which implies that $K$ is symmetric. 

We proceed with the assumption that $\Ga_2$ is integrable and use Bochner's theorem to write 
\[
 \Ga_1(y) = \int_{\mR^d} e^{-i\xi^Ty} d\mu_1(\xi)
 \]
 for some positive finite measure $\mu_1$. Then
\begin{align*}
\Ga(x) &= \int_{\mR^d} \left(\int_{\mR^d} e^{-2i\pi\xi^T(x-z)} d\mu_1(\xi)\right) \Ga_2(z)\,dz\\
&= \int_{\mR^d} e^{-2i\pi\xi^Tx} \left(\int_{\mR^d} e^{2i\pi\xi^Tz}  \Ga_2(z)\,dz\right) d\mu_1(\xi)\\
\end{align*}
The shift in the order of the variables $\xi$ and $z$ uses Fubini's theorem. The function
\[
\psi(\xi) = \int_{\mR^d} e^{2i\pi\xi^Tz}  \Ga_2(z)\,dz
\]
is the inverse Fourier transform of $\Ga_2$. Because $\Ga_2$ is bounded and integrable, it is also square integrable, which implies that its inverse Fourier transform is also a square integrable function. Since Bochner's theorem implies that $\Ga_2$ is the Fourier transform of a positive measure $\mu_2$, we find, using the injectivity of the Fourier transform, that $\psi$ is non-negative. So $\Ga$ is the Fourier transform of the finite positive measure $\psi d\mu_1$, which implies that $K$ is a positive kernel.
\end{proof}

Point (iv)  can be related to the following discrete statement on symmetric matrices: assume that $A$ and $B$ are positive semi-definite and that they commute, so that $AB = BA$: then $AB$ is positive semi-definite. 
In the case of kernels, one may consider the symmetric linear operators $\mathbb K_i : f \mapsto \int_{\mR^d} K_i(\cdot,y) f(y) dy$ which maps the space of square integrable functions into itself. Then $\mathbb K_1$ and $\mathbb K_2$ commute and $\mathbb K = \mathbb K_1 \mathbb K_2$. 


\subsection{Canonical Feature Spaces}

Let $K$ be a positive kernel on a set $\CR$. The following construction, which is fundamental, shows that $K$ can always be associated with a feature function $h$ taking values in a suitably chosen inner-product space $H$. 

Associate to each $x\in \CR$ the function $\xi_x: y \mapsto K(y,x)$ (we will also write $\xi_x = K(\cdot, x)$), and let $H_K = \vspan(\xi_x, x\in \CR)$, a subspace of the vector space of all functions from $\CR$ to $\mR$. Define the feature function $h: x\mapsto \xi_x$ from $\CR$ to $H_K$. There is a unique inner product on $H_K$ such that $K = K_h$. Indeed, by definition, this requires
\begin{equation}
\label{eq:reprod}
\scp{K(\cdot, x)}{K(\cdot, y)}_{H_K} = K(x,y)\,.
\end{equation}
Moreover, by linearity, for any $\xi = \sum_{i=1}^n \la_i K(\cdot, x_i)$ and $\eta = \sum_{i=1}^m \mu_i K(\cdot, y_i)$, one needs
\[
\scp{\xi}{\eta}_{H_K} = \sum_{i=1}^n\sum_{j=1}^m \la_i\mu_j K(x_i, y_j)\,,
\]
so that the inner product is uniquely specified on $H_K$.
To make sure that this inner-product is well defined, we must check that there is no ambiguity, in the sense that, if $\xi$ has an alternative decomposition $\xi = \sum_{i=1}^{n'} \la'_i K(\cdot, x'_i)$, then, the value of $\scp{\xi}{\eta}_{H_K}$ remains unchanged. But this is clear, because one can also write
\[
\scp{\xi}{\eta}_{H_K} = \sum_{j=1}^m \mu_j \xi(y_j)\,,
\]
which only depends on $\xi$ and not on its decomposition. The linearity of the product with respect to $\xi$ is also clear from this expression, and the bilinearity by symmetry. 

The Schwartz inequality implies that 
\[
|\scp{\xi}{\eta}_{H_K}| \leq \|\xi\|_{H_K}\,\|\eta\|_{H_K}
\]
From which we deduce that $\|\xi\|_{H_K} = 0$ implies that $\scp{\xi}{\eta}_{H_K}=0$ for $\eta\in H_K$. Since 
$\scp{\xi}{K(\cdot, y)}_{H_K} = \xi(y)$ for all $y$, this also implies that $\xi=0$, completing the proof that $H_K$ is an inner-product space.
\bigskip

Equation \eqref{eq:reprod} is the ``reproducing property'' of the kernel for the inner-product on $H_K$. In functional analysis, the completion, $\hat H_K$, of $H_K$ for the topology associated to its norm is then a Hilbert space, and is referred to as a ``reproducing kernel Hilbert space,'' or RKHS. 

More generally, an inner-product space $H$ of functions $h: \CR \to \mR$ is a reproducing kernel Hilbert space if $H$ is a complete space (which makes it Hilbert) and there exists a positive kernel $K$ such that, 
\begin{enumerate}[label={[RKHS\arabic*]}, wide=0pt]
\item For all $x\in \CR$, $K(\cdot, x)$ belongs to $H$,
\item For all $h\in H$ and $x\in \CR$, 
\[
\scp{h}{K(\cdot, x)}_H = h(x)\,.
\] 
\end{enumerate}


Returning to the example of functional features in \cref{sec:functional.feat}, we have two different representations of the kernel in feature space, namely in $H = L^2(\mR^d)$, or in $H_K$, with a different inner product. There is not a contradiction, and simply shows that the representation of a positive kernel in terms of a feature function is not unique. 

\begin{remark}
\label{rem:rkhs}
RKHS's are defined as function spaces. While feature space representations, provided by functions
$h: \CR \to H$  from $\CR$ to a Hilbert space $H$ are apparently more general, a simple transformation allows for an identification of (a subspace of) $H$ with an RKHS. We will assume that the subset $h(\CR)$ (containing all $h(x)$, $x\in \CR$) is a dense subset of $H$. If not, one can simply replace $H$ by the closure of $h(\CR)$ which is a Hilbert subspace of $H$.

One can always interpret elements $u\in H$ as functions by letting $\phi_u(x) = \scp{u}{h(x)}_H$. The representation $u\mapsto \phi_u$ is not one-to-one under our assumption that $h(\CR)$ is a dense subset of $H$: if $\phi_u = \phi_v$, then $\scp{u-v}{h(x)}_H = 0$ for all $x\in \CR$, and since $h(\CR)$ is dense in $H$, this implies $u=v$. Letting $\wtilde H = \{\phi_u, u\in H\}$ and defining the inner product $\scp{\phi_u}{\phi_{u'}}_{\wtilde H} = \scp{u}{u'}_H$, one obtains a Hilbert space $\wtilde H$ isometric to $H$ with a new feature function $\tilde h(x) = \phi_{h(x)}$. By definition, $\tilde h(x)(y) = \scp{h(x)}{h(y)}_H = K_h(x,y)$ so that $K_h(x, \cdot)$ belongs to $\wtilde H$ and 
$\scp{\tilde h(x)}{\tilde h(y)}_{\wtilde H} = \scp{h(x)}{h(y)}_H = K_h(x,y)$. This shows that $\wtilde H$ is an RKHS with kernel $K_h$.
\end{remark}

\section{Projection on a finite-dimensional subspace}
\label{sec:orth.proj}
 If $H$ is an inner-product space and $V$ is a subspace of $H$, one defines the orthogonal projection of an element $\xi\in H$ on $V$ as its closest point in $V$, that is, the element $\eta^*$ of $V$ minimizing  the function $F: \eta \mapsto \| \eta - \xi\|^2_H$ over all $\eta\in V$. This closest point does not always exist, but it does in the special case in which $V$ is finite dimensional (or, more generally, when $V$ is a closed subspace of $H$; see \citet{yos70}). We state, without proof, some of the properties of this operation.
 
Assuming that $V$ is closed,  this minimizer is unique and will be denoted $\eta^* = \pi_V(\xi)$. Moreover, $\pi_V$ is a linear transformation from $H$ to $V$, and $\eta^*$ is characterized by the properties
 \[
 \left\{ \begin{aligned} &\eta^*\in V \\ &\xi-\eta^* \perp V,\end{aligned}
 \right.
 \]
 the last condition meaning that $\scp{\xi-\eta^*}{\eta}_H = 0$ for all $\eta\in V$.
 
 Because $\|\xi\|_H^2 = \|\pi_V(\xi)\|_H^2 + \|\xi- \pi_V(\xi)\|_H^2$, one always has $\|\pi_V(\xi)\|_H \leq \|\xi\|_H$, with inequality if and only if $\pi_V(\xi) = \xi$, i.e., if and only if $\xi\in V$.

 If $V$ is finite-dimensional and $\eta_1, \ldots, \eta_n$ is a basis of $V$, then $\pi_V(\xi)$ is given by
 \[
 \pi_V(\xi) = \sum_{i=1}^n \pe \alpha i \eta_i
 \]
 with $\alpha$ (considered as a column vector in $\mR^n$) given by
 \[
 \alpha = \mathrm{Gram}(\eta_1, \ldots, \eta_n)^{-1} \la\,,
 \]
 where $\la\in\mR^n$ is the vector with coordinates $\pe \la i = \scp{\xi}{\eta_i}_H$, $i=1, \ldots, n$. The Gram matrix of $\eta_1, \ldots, \eta_n$, denoted $\mathrm{Gram}(\eta_1, \ldots, \eta_n)$, is the $n$ by $n$ matrix with entries $\scp{\eta_i}{\eta_j}_H$ for $i,j = 1, \ldots, n$.

If $A$ is a subset of $H$, the set $A^\perp$ consists of all vectors perpendicular to $A$, namely
\[
A^\perp = \defset{h\in H: \scp{h}{\tilde h}_H = 0 \text{ for all } \tilde h \in A}\,.
\]
If $V$ is a finite-dimensional  (or, more generally, closed) subspace of $H$, then any point in $h$ is decomposed as $h = \pi_V(h) + h- \pi_V(h)$ with $h-\pi_V(h) \in V^\perp$. This shows that $\pi_{V^\perp}$ is well defined and equal to $\mathrm{id}_H - \pi_V$.

\bigskip

Orthogonal projections can be applied to function interpolation in an RKHS. Indeed, assuming that $H$ is an RKHS, as described at the end of the previous section, with a positive-definite kernel.   Given distinct points $x_1, \ldots, x_N\in \CR$ and values $\alpha_1, \ldots, \alpha_N\in \mR$, the interpolation problem consists in finding $h\in H$ with minimal norm satisfying $h(x_k) = \alpha_k$, $k=1, \ldots, N$. Consider the finite dimensional space
\[
V = \vspan\defset{K(\cdot, x_k), k= 1, \ldots N}.
\]
Then there exists an element $h_0\in V$ that satisfies the constraints. Indeed, looking for $h_0$ in the form
\[
h_0(x) = \sum_{l=1}^N K(x, x_l) \lambda_l 
\]
one has
\[
h_0(x_k) = \sum_{l=1}^N K(x_k, x_l) \lambda_l
\]
so that 
\[
\begin{pmatrix}
\lambda_1 \\ \vdots \\ \lambda_N 
\end{pmatrix}
 =
 \CK(x_1, \ldots, x_N)^{-1} \begin{pmatrix}
\alpha_1 \\ \vdots \\ \alpha_N 
\end{pmatrix}
 \]
 
Any other function $h$ satisfying the constraints satisfies $h(x_k) - h_0(x_k) = 0$, which, using RKHS2, is equivalent to $\scp{h-h_0}{K(\cdot, x_k)}_H = 0$, i.e., to $h-h_0\in V^\perp$. This shows that $h_0 = \pi_V(h)$, so that 
$\|h\|_H \geq \|h_0\|_H$ and $h_0$ provides the optimal interpolation. We summarize this in the proposition:
\begin{proposition}
\label{prop:rkhs.interpolation}
Let $H$ is an RKHS with a positive-definite kernel.   Let $x_1, \ldots, x_N\in \CR$ be distinct points  and $\alpha_1, \ldots, \alpha_N\in \mR$. Then the function $h\in H$ with minimal norm satisfying $h(x_k) = \alpha_k$, $k=1, \ldots, N$ takes the form
\begin{subequations}

\begin{equation}
\label{eq:rkhs.interpolation.1}
h(x_k) = \sum_{l=1}^N K(x_k, x_l) \lambda_l
\end{equation}
with 
\begin{equation}
\label{eq:rkhs.interpolation.2}
\begin{pmatrix}
\lambda_1 \\ \vdots \\ \lambda_N 
\end{pmatrix}
 =
 \CK(x_1, \ldots, x_N)^{-1} \begin{pmatrix}
\alpha_1 \\ \vdots \\ \alpha_N 
\end{pmatrix}.
 \end{equation}
\end{subequations}
\end{proposition}

A variation of this problem replaces the constraint by a penalty that complete the minimization associated with the orthogonal projection, namely, minimizing (in $h\in H$)
\[
\|h\|_H^2 + \sigma^2 \sum_{k=1}^N |h(x_k) - \alpha_k|^2.
\]
Letting $h_0 = \pi_V(h)$, so that $h_0(x_k) = h(x_k)$ for all $k$, this expression can be rewritten as
\[
\|h_0\|_H^2 + \|h-h_0\|^2_H + \sigma^2 \sum_{k=1}^N |h_0(x_k) - \alpha_k|^2.
\]
This shows that the optimal $h$ must coincide with its projection on $V$, and therefore belong to that subspace. Looking for $h$ in the form
\[
h(\cdot) = \sum_{l=1}^N K(\cdot, x_l) \lambda_l,
\]
the objective function is rewritten as
\[
\sum_{k,l=1}^N K(x_k, x_l) \lambda_k\lambda_l + \sigma^2 \sum_{k=1}^N \left|\sum_{l=1}^N K(x_k, x_l) \lambda_l - \alpha_k\right|^2, 
\]
which, in vector notation gives, writing $\boldsymbol{\lambda} = \begin{pmatrix}
\lambda_1 \\ \vdots \\ \lambda_N 
\end{pmatrix}$ and $\boldsymbol\alpha = \begin{pmatrix}
\alpha_1 \\ \vdots \\ \alpha_N 
\end{pmatrix}$, 
\[
\boldsymbol\lambda^T
\CK(x_1, \ldots, x_N)\boldsymbol\lambda +
 \sigma^2 \left(\CK(x_1, \ldots, x_N)\boldsymbol\lambda
 - \boldsymbol \alpha\right)^T \left(\CK(x_1, \ldots, x_N)\boldsymbol\lambda
 - \boldsymbol \alpha\right).
\]

The differential of this expression in $\boldsymbol\lambda$ is 
\[
\CK(x_1, \ldots, x_N)\boldsymbol\lambda + 2\sigma^2 \CK(x_1, \ldots, x_N)(\CK(x_1, \ldots, x_N)\boldsymbol\lambda  - \boldsymbol\alpha).
\]
Assuming that $x_1, \ldots, x_N$ are distinct, this vanishes if and only if
\[
\boldsymbol\lambda = (\CK(x_1, \ldots, x_N) + (1/\sigma^2) \Id[N])^{-1}\boldsymbol\alpha.
\] 

We have just proved the proposition:
\begin{proposition}
\label{prop:rkhs.interpolation.soft}
Let $H$ is an RKHS with a positive-definite kernel.   Let $x_1, \ldots, x_N\in \CR$ be distinct points  and $\alpha_1, \ldots, \alpha_N\in \mR$. Then the unique minimizer of 
\[
h \mapsto
\|h\|_H^2 + \sigma^2 \sum_{k=1}^N |h(x_k) - \alpha_k|^2
\]
on $H$ is given by
\begin{subequations}

\begin{equation}
\label{eq:rkhs.interpolation.soft.1}
h(x_k) = \sum_{l=1}^N K(x_k, x_l) \lambda_l
\end{equation}
with 
\begin{equation}
\label{eq:rkhs.interpolation.soft.2}
\begin{pmatrix}
\lambda_1 \\ \vdots \\ \lambda_N 
\end{pmatrix}
 =
( \CK(x_1, \ldots, x_N)+ (1/\sigma^2) \Id[N])^{-1} \begin{pmatrix}
\alpha_1 \\ \vdots \\ \alpha_N 
\end{pmatrix}.
 \end{equation}
\end{subequations}
\end{proposition}

\newpage
\markright{PROBLEMS}
\section*{Problems}
\addcontentsline{toc}{section}{Problems}
\input Problems_Kernels


\chapter[Linear Regression]{Linear Models for Regression}
\label{chap:lin.reg}

In  regression, {\em linear models}  refer to situations in which one tries to predict the dependent variable $Y\in \CR_Y = \mR^{q}$ by a function $\hf(X)$ of the dependent variable $X\in \CR_X$, where  $\hf$ is optimized over a {\em linear space} $\CF$. 
The most common situation is the ``standard linear model,'' for which $\CR_X = \mR^d$ and
\begin{equation}
\label{eq:cf.linear}
\CF = \{ f(x) = {a_0} + b^{T}x: {a_0}\in \mR^{q}, b\in \CM_{d,q}(\mR)\}.  
\end{equation}

More generally, with $q=1$, given a mapping $h : \CR \to H$, where $H$ is an inner-product space, one can take:
\begin{equation}
\label{eq:cf.feature}
\CF = \{ f(x) = {a_0} + \scp{b}{h(x)}_{H}: {a_0}\in \mR, b\in H\}.  
\end{equation}
Note that $h$ can be nonlinear, and $\CF$ can be infinite dimensional. Such sets corresponds to linear models using feature functions, and will be addressed using kernel methods in this chapter.

Note also that, even if the model is linear, the associated training algorithms can be nonlinear, and we
will review in fact several situations in which solving the estimation problem requires nonlinear optimization methods.

\section{Least-Square  Regression}
\subsection{Notation and Basic Estimator}
We denote by $Y$ and $X$ the dependent and independent variables of the regression problem. We will assume that $Y$ takes values in  $\mR^q$ and that $X$ takes values in a set $\CR_X$, which will, by default, be equal to $\mR^d$, except when discussing kernel methods, for which this set can be arbitrary (provided that there is a mapping $h$ from $\CR_X$ to  an inner product space $H$ with an easily computable kernel). 
 
Least-square regression uses the risk function $r(y,y') = |y-y'|^2$. The prediction error is then $R(f) = E(|Y-f(X)|^2)$ for any predictor $f$ and the Bayes predictor is the conditional expectation $x \mapsto E(Y\mid X=x)$ (see \cref{item:Bayes.MSE} in \cref{sec:bayes.pred}). We also start with the standard setting where $\CR_X = \mR^d$ and
$\CF$ given by \eqref{eq:cf.linear}. 

We will use the following notation, which sometimes simplifies the computation. If $x\in \mR^d$, we let $\tilde x = \begin{pmatrix}1\\x\end{pmatrix}$, which belongs to $\mR^{d+1}$. The linear predictor 
$
f(x) = {a_0} + b^T x
$
with ${a_0}\in \mR^{q}, b\in \CM_{d,q}(\mR)$ can then be written as $f(x) = \beta^{T} \tilde x$ with $\beta =
\begin{pmatrix} a_0^{T}\\ b\end{pmatrix} \in \CM_{d+1,q}(\mR)$.  

In a model-based approach, the linear model is a Bayes predictor under the generative assumption that $Y = {a_0} + b^T X + \ep$ where $\ep$ is a residual noise satisfying $E(\ep\mid X) = 0$, which is true, for example, when $\ep$ is centered and independent of $X$. If one further specifies the model so that $\ep$ is Gaussian, centered and independent of $X$, and one  assumes that the distribution of $X$ does not depend on ${a_0}$ and $b$, then the maximum likelihood estimator of these parameters based on a training set $T = ((x_1, y_1), \ldots, (x_N, y_N))$ must minimize the ``residual sum of squares:''
\[
RSS(\be) \defeq N\hat R(f) \defeq \sum_{k=1}^N |y_k - f(x_k)|^2  = \sum_{k=1}^N |y_k -
\be^{T} \tx_k|^2 \,.
\]
In other terms, the model-based approach is identical, under these (standard) assumptions, to empirical risk minimization (\cref{sec:emp.risk}), on which we now focus. (Recall that, even when using a model-based approach, one does not make assumptions on the true distribution of $X$ and $Y$; one rather treats the model as an approximation of these distributions, estimated by maximum likelihood, and uses the Bayes predictor for the estimated model.) 

The computation of the optimal regression parameters is made easier by the introduction of the following matrices.
 Introduce the $N\ti (d+1)$ matrix $\mathcal X$ with rows $\tilde x_1^T, \ldots, \tilde x_N^T$ and the $N\ti q$ matrix 
$\mathcal Y$ with rows $y_1^{T}, \ldots, y_N^{T}$, 
that is:
\[
 \CX = \begin{pmatrix} 1 & \pe{x_1}{1} & \cdots & \pe{x_1}{d} \\ \vdots & \vdots && \vdots \\  1 & \pe{x_N}{1} & \cdots & \pe{x_N}{d} \end{pmatrix}, \quad
 \CY = \begin{pmatrix} \pe{y_1}{1} & \cdots & \pe{y_1}{q} \\ \vdots && \vdots \\  \pe{y_N}{1} & \cdots & \pe{y_N}{q}\end{pmatrix}\,.
\]
With this notation, we have
\[
RSS(\be) = |\mathcal Y - \mathcal X \be|_{2}^2.
\]
with $|A|_{2}^{2} = \trace(A^{T}A)$ for a rectangular matrix $A$. The solution of the problem is then provided by the following theorem.
\begin{theorem}
\label{th:ls.reg}
Assume that the matrix $\mathcal X$ has rank $d+1$. Then the RSS is
minimized for 
\[
\hat \beta = (\CX^T\CX)^{-1} \CX^T \mathcal Y
\]
\end{theorem}
\begin{proof}
We provide two possible proofs of this elementary problem. The first one is an optimization argument noting that
$F(\be) \defeq RSS(\be)$ is a convex function defined on $\CM_{d+1,q}(\mR)$ and with values in $\mR$. Since $F$ is quadratic, we have, for any matrix $h\in \CM_{d+1,q}(\mR)$, 
\[
dF(\be) h = \prt_\ep F(\be+\ep h){|_{\ep=0}} = - 2 \trace( h^{T} \CX^{T} (\CY-\CX\be))
\]
and
\[
dF(\be) = 0 \Leftrightarrow \CX^{T} (\CY-\CX\be) = 0 \Leftrightarrow\be = \hat\be.
\]

One can alternatively proceed with a direct computation.
We have
\begin{align*}
RSS(\be) &= |\CY|^{2}_{2} - 2\trace(\be^{T} \CX^{T}\CY) + \trace(\be^{T}\CX^{T}\CX\be) \\
& = |\CY|^{2}_{2} - 2\trace(\be^{T} \CX^T\CX \hat \be ) + \trace(\be^{T}\CX^{T}\CX\be).
\end{align*}
\item Replacing $\be$ by $\hat\be$ and simplifying yields
\[
RSS(\hat \be) = |\CY|^{2}_{2} - \trace(\hat \be^{T} \CX^T\CX \hat\be )
\]
It follows that 
\begin{align*}
RSS(\be) =& RSS(\hat\be) + \trace(\hat \be^{T} \CX^T\CX\hat\be )-  2\trace(\be^{T} \CX^T\CX \hat\be ) 
+ \trace(\be^{T}\CX^{T}\CX\be)\\
=& RSS(\hat\be) + |\mathcal X(\hat\be - \be)|^2_{2}
\end{align*}
so that the left-hand side is minimized at $\be = \hat\be$.
\end{proof}

\begin{remark}
\label{rem:linear.low.rank}
If $\CX$ does not have rank $d+1$, then optimal solutions exist, but they are not unique. By convexity, the solutions are exactly the vectors $\beta$ at which the gradient vanishes, i.e., those  that satisfy $\CX^T\CX \beta = \CX^T \CY$. The set of solutions can be obtained by introducing  the SVD of $\CX$ in the form $\CX = UDV^T$ and letting $\gamma = V^T\beta$ and $\CZ = U^T \CY$. Then 
\[
\CX^T\CX \beta = \CX^T \CY\Leftrightarrow D^TD \gamma = D^T \CZ.
\]
Letting $\pe d 1 , \ldots, \pe d m$ denote the nonzero diagonal entries of $D$ (so that $m\leq d+1$), we  find $\pe \gamma i  = \pe z i  / \pe d i $ for $i\leq m$ (the other equalities being $0=0$). So, the $d+1-m$ last entries of $\gamma$ can be chosen arbitrarily (and $\beta = V\gamma$). 
\end{remark}

An alternate representation of the solution use a two-step computation that estimates $b$ first, then ${a_0}$. Indeed, for fixed $\hat b$, the minimum of 
\[
\sum_{k=1}^N |y_k - {a_0} - x_k^T \hat b|^2
\]
is attained at $\hat {a_0} = \bar y - \bar x^T \hat b$ with the usual definitions
$$\bar y = \frac{1}{N} \sum_{k=1}^N y_k \text{ and } \bar x =
\frac{1}{N} \sum_{k=1}^N x_k.
$$
This shows that $\hat b$ itself must be a minimizer of 
\[
\sum_{k=1}^N |y_k - \bar y - (x_k-\bar x)^T b|^2.
\]
Denote by $\CY_c$ and $\CX_c$ the matrices 
\[
 \CX_c = \begin{pmatrix} \pe{x_1} 1 - \pe{\bx} 1 & \cdots & \pe {x_1} d - \pe \bx d  \\ \vdots && \vdots \\  \pe {x_N} 1 - \pe \bx 1  & \cdots & \pe{x_N} d - \pe \bx d \end{pmatrix}, \ 
 \CY_c = \begin{pmatrix} \pe {y_1} 1 - \pe \by 1  & \cdots & \pe{y_1}  q - \pe \by q  \\ \vdots && \vdots \\  \pe {y_N} 1 - \pe{\by} 1 & \cdots & \pe{y_N} q - \pe{\by} q \end{pmatrix}\,.
\]
Then $\hat b$ must minimize $|\CY_c - \CX_c b|^2$, yielding
$$
\hat b = (\CX_c^T\CX_c)^{-1} \CX_c^T \mathcal Y_c, \quad \hat {a_0} = \bar y - \bar x^T \hat b.
$$
The reader may want to double-check that 
this solution coincides with the one provided in  \cref{th:ls.reg}. 

\subsection{Limit behavior}

The matrix 
\[
\hat\Sigma_{XX} = \frac1N \CX_c^T\CX_c = \frac1N \sum_{k=1}^N (x_k - \bx)(x_k-\bx)^T
\]
is a sample estimate of the  covariance matrix of $X$, that we will denote $\Sig_{XX}$. Similarly, $\hat \Sig_{XY} = \CX_c^T\CY_c/N$ is a sample estimate of $\Sigma_{XY}$, the covariance between $X$ and $Y$. With this notation, we have
\[
\hat b = \hat \Sig_{XX}^{-1} \hat\Sig_{XY},
\]
which, by the law of large numbers, converges to $b^* =  \Sig_{XX}^{-1} \Sig_{XY}$. 

Let $a_0^*= m_Y - m_X^T b^*$. Then $f^*(x) = a_0^* + (b^{*})^{T}x$ is the least-square optimal approximation of $Y$ by a {\em linear} function of $X$, and
the linear predictor $\hat f(x) = \hat {a_0} + \hat b^{T}x$ converges a.s. to $f^*(x)$. Of course, $f^*$ generally differs from  $f: x\mapsto E(Y\mid X=x)$, which is the least-square optimal approximation of $Y$ by \alert{any} (square-integrable) function of $X$, so that the linear estimator will  have a residual bias. 

\subsection{Gauss-Markov theorem}
If one makes the (unlikely) assumption that the linear model is exact, i.e., $f(x) = f^*(x)$, one has:
\[
\myE(\hat \be) = \myE(\myE(\hat\be \mid \CX)) = \myE((\CX^T\CX)^{-1}\CX^T \myE(\CY\mid \CX)) = \myE((\CX^T\CX)^{-1}\CX^T \CX \be) = \be
\]
and the estimator is ``unbiased.'' Under  this parametric assumption, many other properties of linear estimators  can be proved, among which the well-known Gauss-Markov theorem on the optimality of least-square estimation that we now state and prove. For this theorem, for which we take (for simplicity) $q=1$, we also assume that $\var(Y\mid X=x)$, the variance of $Y$ for its conditional distribution given $X$ does not depend on $x$, and denote it by $\sig^2$. This typically correspond to the standard regression model in which one assumes that $Y = f(X) + \epsilon$ where $\epsilon$ is independent of $X$ with variance $\sig^2$.

Recall that a symmetric matrix $A$ is said to be larger  than or equal to another
symmetric matrix, $B$, writing $A\succeq B$, if and only if $A-B$ is positive semi-definite.
\begin{theorem}[Gauss-Markov]
Assume that an estimator $\tilde \be$ takes the form $\tilde \be = A(\CX)\CY$ (it is
linear) and is unbiased conditionally to $\CX$: $\myE_{\be}(\tilde \be\mid \CX) =
\be$ (for all $\be$).
Then (under the assumptions above) the covariance matrix of $\tilde \be$ cannot be smaller than that
of the least square estimate, $\hat\be$.
\end{theorem}
\begin{proof}
We write $A = A(\CX)$ for short.
The condition
that $\myE(A\CY\mid\CX) = \be$ for all $\be$ yields $A\CX \be = \be$ for all $\be$, 
or $A\CX = \Id_{\mR^{d+1}}$ ($A$ is a $(d+1) \times N$ matrix). Since $\tilde\be$ is unbiased, its covariance matrix is 
\[
\myE(A\CY\CY^TA^T) - \beta\beta^T
\]
and 
\[
\myE(A\CY\CY^TA^T) = \myE(\myE(A\CY\CY^TA^T\mid \CX)) = \sigma^2 E(AA^T).
\]
 For $\tilde\be = \hat\be$, for which $A = (\CX^T\CX)^{-1}\CX^T$, we get $\myE(A\CY\CY^TA^T) =  \sig^2 \myE((\CX^T\CX)^{-1})$. We therefore need to show that $\myE(AA^T) \succeq \myE(\CX^T\CX)$, i.e., that for any $u\in\mR^{d+1}$, 
$$ u^T \myE(AA^T)  u \geq u^T\myE((\CX^T\CX)^{-1}) u$$
as soon as $A\CX = \Id[d+1]$. We in fact have the stronger result (without expectations): 
$$
A\CX = \Id[d+1] \Ria AA^T \succeq (\CX^T\CX)^{-1}.
$$

To see this, fix $u$ and consider the problem of minimizing  $F_u(A) = A \mapsto u^T AA^T
u$ subject to the linear constraint $A\CX = \Id[d+1]$. The Lagrange multipliers for
this affine constraint can be organized in a matrix $C$ and the
Lagrangian is
$$ u^T AA^T u + \trace(C^T (A\CX- \Id[d+1])).$$
Taking the derivative in $A$, we find that optimal solutions must satisfy
$$ 2 u^T AH^T u + \trace(C^T H\CX) = 0$$
for all $H$, which yields $\trace(H^T (2 uu^TA + C \CX^T)) = 0$ for all $H$. This is only possible when $2 uu^TA + C \CX^T = 0$, which
in turn implies that 
$2 uu^TA\CX =   -C \CX^T\CX$. Using the  constraint, we get
$$
C  = -2 uu^T (\CX^T\CX)^{-1}
$$
so that
$uu^TA = uu^T (\CX^T\CX)^{-1} \CX^T$. 
This implies that $A = (\CX^T\CX)^{-1} \CX^T$ (the least-square estimator) is a minimizer of $F_u(A)$ for all
$u$. 

Any other solution that satisfies $uu^TA = uu^T (\CX^T\CX)^{-1} \CX^T$ for all $u$. Taking $u = \mathfrak e_i$ and summing over $i$ (with $\sum_{i=1}^{d+1} \mathfrak e_i \mathfrak e_i^T = \Id[d+1]$) yields $A = (\CX^T\CX)^{-1} \CX^T$.
\end{proof}

\subsection{Kernel Version}
We now  assume that $X$ takes its values in an arbitrary set $\CR_X$, with a representation $h:\CR_X \to H$ into an inner-product space. This representation does not need to be explicit or computable, but the associated kernel $K(x,y) = \scp{h(x)}{h(y)}_H$ is assumed to be known and easy to compute. (Recall that, from \cref{chap:higher}, a positive kernel is always associated with an inner-product space.) In particular, any algorithm in this context should only rely on the kernel, and the function $h$ only has a conceptual role. 

Assume that $q=1$ to lighten the notation, so that the dependent variable is scalar-valued. We here let the space of predictors be
\[
\CF = \{ f(x) = {a_0} +\scp{b}{h(x)}_{H}: {a_0}\in \mR, b\in H\}.  
\]
The residual sum of squares associated with this function space is
\[
RSS({a_0}, b) = \sum_{k=1}^N (y_k - {a_0} - \scp{b}{h(x_k)})^2.
\]

The following result (or results similar to it) is a key step in almost all kernel methods in machine learning. 
\begin{proposition}
\label{prop:proj.kern}
Let $V = \vspan(h(x_1), \ldots, h(x_N))$ be the finite-dimensional subspace of $H$ generated by the feature functions evaluated on training input data. Then
\[
RSS({a_0}, b) = RSS({a_0}, \pi_V(b)).
\]
where $\pi_V$ is the orthogonal projection on $V$.
\end{proposition}
\begin{proof}
The justification is immediate: since $h(x_k) \in V$, we have 
\[
\scp{b}{h(x_k)}_H = \scp{\pi_V(b)}{h(x_k)}_H
\]
 for all $b\in H$.
\end{proof}

This shows that there is no loss of generality in restricting the minimization of the residual sum of squares to $b\in V$. Such a $b$ takes the form
\begin{equation}
\label{eq:b.kern}
b = \sum_{k=1}^N \alpha_k h(x_k)
\end{equation}
and the regression problem can be reformulated as a function of the coefficients $\alpha_1, \ldots, \alpha_N\in \mR$, with 
\[
f(x) = {a_0} + \sum_{k=1}^N \al_k \scp{h(x)}{h(x_k)}_H = {a_0} + \sum_{k=1}^N \al_k K(x,x_k),
\]
which only depends on the kernel. (This reduction is often referred to as the ``kernel trick.'')

However, the solution of the problem is, in this context, not very interesting. Indeed, assume that $K$ is positive definite and that all observations in the training set are distinct. Then the matrix $\CK(x_1, \ldots, x_N)$ formed by the kernel evaluations $K(x_i, x_j)$ is invertible, and one can solve exactly the equations
\[
y_k = \sum_{j=1}^N \al_j K(x_k,x_j), \quad k=1, \ldots, N
\]
to get a zero RSS with ${a_0} = 0$. Unless there is no noise, such a solution will certainly overfit the data. If $K$ is not positive definite, and the dimension of $V$ is less than $N$ (since this would place us in the previous situation otherwise), then it is more efficient to work directly in a basis of $V$ rather than using the over-parametrized kernel representation. We will see however, starting with the next section, that kernel methods become highly relevant as soon as the regression is estimated with some control on the size of the regression coefficients, $b$.

\section{Ridge regression and Lasso}

\subsection{Ridge Regression}
\label{sec:ridge}

\paragraph{Method.}
When the set $\CF$ of possible predictors  is too large, some additional complexity control is needed to reduce the estimation variance.
One simple approach is to limit the number of parameters to be estimated, which, for regression, corresponds to limiting the number of possible predictors. This is related to the methods of Sieves mentioned in \cref{sec:sieves}. In contrast, ridge regression and lasso control the size of the parameters, as captured by their norm. 

In both cases, one assigns a  measure of complexity, denoted $f \mapsto \ga(f) \geq 0$,  to each element $f\in\CF$. 
Given $\ga$, one can either optimize this predictor (using, for example, the RSS) with the constraint that $\ga(f) \leq C$ for some constant $C$, or add a penalty $\la \ga(f)$ to the objective function for some $\la>0$.
In general, the two approaches (constraint or penalty) are equivalent.

In linear spaces,  complexity measures are often associated with a norm, and  ridge regression uses the sum of squares of coefficients of the prediction matrix $b$, minimizing
\begin{equation}
\label{eq:rr.1}
\sum_{k=1}^N |y_k - {a_0} - b^{T} x_k|^2 + \la\, \trace(b^{T}b)\,,
\end{equation}
which can be written in vector form as
\[
|\CY - \CX \be|_2^2 + \la\, \trace(\be^T\De\be),
\]
where $\De = \text{diag}(0, 1, \ldots, 1)$. In the following, we will work with an unspecified $(d+1)\times (d+1)$ symmetric positive semi-definite matrix $\De$. Various choices are indeed possible, for example, $\De = \text{diag}(0, \hat\sig^2(1), \ldots, \hat\sig^2(d))$, where $\hat\sig^2(i)$ is the empirical variance of the $i$th coordinate of $X$ in the training set. This last choice is quite natural, because it ensures that, whenever one of the variable $\pe X i $ is rescaled by a factor $c$, the corresponding optimal $i^{\mathrm{th}}$ row of $b^T$ is rescaled by $1/c$, leaving the predictor unchanged.

Under this assumption, the optimal parameter is
\[
\hat\be^\la = (\CX^T\CX + \la \De)^{-1}\CX^T \CY\,,
\]
with a proof similar to that made for least-square regression. We obviously retrieve the original formula for regression when
$\la=0$. 

Alternatively, assuming that $\De = \begin{pmatrix} 0&0\\0&\De'\end{pmatrix}$, so that no penalty is imposed on the intercept, we have
\begin{equation}
\label{eq:ridge.2.form}
\hat b^\la = (\CX_c^T\CX_c + \la \De')^{-1}\CX_c^T \CY_c
\end{equation}
and $\hat{a_0}^\la = \bar y - (\hat b^\la)^T \bar x$. The proof of these statements is left to the reader.
\bigskip

\paragraph{Analysis in a special case}
To illustrate the impact of the penalty term on balancing bias and variance, we now make a computation in the special case when $Y = \tilde X\be + \ep$, where $\var(\ep) = \sig^2$ and $\ep$ is independent of $X$. In the following computation, we assume that the training set is fixed (or rather, compute 
 probabilities and expectations 
conditionally to it).
Also, to simplify notation, we
denote
$$
S_\la = \CX^T\CX + \la \De = \sum_{k=1}^N \tx_k^T\tx_k + \la\De
$$
and 
$\Sig = E(\tX^T\tX)$ for a single realization of $X$. Finally, we assume that $q=1$, also to simplify the discussion.

The mean-square prediction error is 
\begin{eqnarray*}
R(\la) &=& E((Y - \tX^T \hat\be_\la)^2)\\
& =& E((\tX^T(\be - \hat\be_\la) + \ep)^2)\\
&=& (\hat\be_\la- \be)^T \Sig (\hat\be_\la - \be) + \sig^2.
\end{eqnarray*}
Denote by $\ep_k$ the (true) residual $\ep_k = y_k -
\tx_k^T \be$ on training data and by $\boldsymbol{\ep}$ the vector stacking these residuals.
We have, writing $S_0 = S_\lambda - \lambda\Delta$, 
\begin{eqnarray*}
\hat\be_\la &=& S_\la^{-1}\CX^T\CY \\
&=& S_\la^{-1}S_0 \be + S_\la^{-1}\CX^T\boldsymbol\ep\\
&=& \be - \la S_\la^{-1}\De \be + S_\la^{-1}\CX^T\boldsymbol\ep
\end{eqnarray*}
So we can rewrite
$$
R(\la) = \la^2 \be^T \De S_\la^{-1} \Sig S_\la^{-1} \De \be -2\la \boldsymbol\ep^T\CX
S_\la^{-1} \Sig S_\la^{-1}\De \be + \boldsymbol\ep^T\CX S_\la^{-1} \Sig
S_\la^{-1} \CX^T\boldsymbol\ep + \sig^2.
$$
Let us analyze the quantities that depend on the training set in this
expression. The first one is $S_\la = S_0 + \la \De$. From the law of
large numbers, $S_0/N \to \Sig$ when $N$ tends to infinity, so that, assuming in addition that $\la=\la_N = O(N)$, we have
$S_\la^{-1} = O(1/N)$. The second one is 
$$
\boldsymbol\ep^T\CX = \sum_{k=1}^N \ep_k \tx_k
$$
which, according to the central limit theorem, is such that
\[
N^{-1/2}\boldsymbol\ep^T\CX \sim \CN(0, \sig^2 \text{Var}(\tX))
\]
 when
$N\to\infty$. So, we can expect the coefficient of $\la^2$  in
$R(\la)$ to have order $N^{-2}$, the coefficient of $\la$ to have
order $N^{-3/2}$ and the constant coefficient of have order
$N^{-1}$. This suggests taking $\la = \mu \sqrt{N}$ so that all
coefficients have roughly the same order when expanding in powers of
$\mu$. 

This gives $S_\la = N(S_0/N + \mu\De/\sqrt N) \simeq N\Sig$ and we 
make the approximation, letting $\xi = N^{-1/2} \sig^{-1/2}\boldsymbol\ep \CX^T$
and $\ga = \Sig^{-1/2} \De\be$, that
$$
N(R(\la) - \sig^2) \simeq \mu^2  |\ga|^2 - 2 \mu \xi^T
\ga  + \xi^T \xi.
$$
With this approximation, the optimal $\mu$ should be 
$$
\mu = \frac{\xi^T\ga}{|\ga|^2}.
$$
Of course, this $\mu$ cannot be computed from  data, but we can see
that, since $\xi$ converges to a centered Gaussian random variable,
its value cannot be too large. It is therefore natural to choose $\mu$ to 
be constant and use ridge regression in the form
$$
\sum_{k=1}^N (y_k - \tilde x_k^T \be)^2 + \sqrt N\mu \be^T \De \be.
$$ 
In all cases, the mere fact that we find that the optimal $\mu$ is not 0 shows that, under the simplifying (and optimistic) assumptions that we made for this computation, allowing for a penalty term always reduces the prediction error.  In other terms, introducing some estimation bias in order to reduce the variance is beneficial.

\paragraph{Kernel Ridge Regression}
We now return to the feature-space situation and take $h: \CR_X \to H$ with associated kernel $K$. We still take $q=1$ for simplicity. One formulates the ridge regression problem in this context as the minimization of
$$
\sum_{k=1}^N(y_l - {a_0} - \scp{b}{h(x_l)}_H)^2 + \la \|b\|_H^2
$$
with respect to $\be = ({a_0}, b)$. Introducing the space $V$ generated by the feature function evaluated on the training set, we know from \cref{prop:proj.kern} that replacing $b$ by $\pi_V(b)$ leaves the residual sum of squares invariant. Moreover, one has $\|\pi_V(b)\|_H^2 \leq \|b\|_H^2$ with equality if and only if $b\in V$. This shows that the solution $b$ must belong to $V$ and therefore take the form \eqref{eq:b.kern}. 

Using this expression, one finds that the problem is  reduced to finding the minimum of 
$$
\sum_{k=1}^N\left(y_k - {a_0} - \sum_{l=1}^N K(x_l, x_k)\al_l \right)^2 +
\la \sum_{k,l=1}^N \alpha_k\alpha_l K(x_k, x_l)\,
$$
with respect to ${a_0}, \alpha_1, \ldots, \alpha_N$. 
Recall that we have denoted by $\CK = \CK(x_1, \ldots, x_N)$ the kernel matrix with entries $K(x_i,x_j)$, $i,j=1, \ldots, N$. We will assume in the following that $\CK$ is invertible. 

Introduce the vector $\dsone_N\in \mR^N$  with all coordinates equal to one.  Let 
\[
\tilde \CK = \begin{pmatrix} \dsone_{N} &\CK\end{pmatrix} \text{ and } \CK' = \begin{pmatrix} 0 & 0 \\0 &\CK\end{pmatrix}.
\]

Let $\al\in \mR^N$ be the vector with coefficients $\al_1, \ldots, \al_N$ and $\tilde \al  = \begin{pmatrix}{a_0}\\ \al \end{pmatrix}$. 
With this notation, the function to minimize is
\[
F(\al) = |\CY - \tilde\CK \tilde\al|^2 + \la \tilde \al^T \CK' \tilde \al.
\]
This takes the same form as standard ridge regression, replacing $\be$ by $\tilde \al$, $\CX$ by $\tilde \CK$ and $\De$ by $\CK'$. The solution therefore is
\[
\tilde \al^\la = (\tilde \CK^T\tilde \CK + \la \CK')^{-1} \tilde\CK^T \CY .
\]
Note that $\CK$ being invertible implies that $\tilde \CK^T\tilde \CK + \la \CK'$ is invertible.
\footnote{Indeed, let $u = \begin{pmatrix}
w_0\\w
\end{pmatrix}
$ with $w_0\in \mR$ and $w\in \mR^N$ be such that $u^T(\tilde \CK^T\tilde \CK + \la \CK')u = 0$.
This requires $\tilde \CK u = 0$ and $u^T\CK'u = 0$.
The latter quantity is $w^T\CK w$, which shows that $w=0$ since $\CK$ has rank $N$.
Then $\tilde \CK = \dsone_{N} w_0$ so that $w_0=0$ also. }


To write the equivalent of \cref{eq:ridge.2.form}, we need to use the equivalent of the matrix $\CX_c$, that is, the matrix $\CK$ with the average of the $j$th column subtracted to each $(i,j)$ entry, given by: 
\[
\CK_c = \CK - \frac1N \dsone_N\dsone_N^T\CK.
\]
Introduce the matrix $P = \Id - \dsone_N\dsone_N^T/N$. It is easily checked that $P^2=P$ ($P$ is a projection matrix). Since $\CK_c = P\CK $, we have $\CK_c^T\CK_c = \CK P\CK$. One deduces from this the expression of the optimal vector $\al^\lambda$, namely,
\[
\al^\la = (\CK P\CK + \la\CK)^{-1}\CK P \CY_c = (P\CK +\la \Id[N])^{-1} \CY_c
\]
where we have, in addition, used the fact that $P\CY_c = \CY_c$. Finally, the intercept is given by
\[
{a_0} = \by - \frac1N (\al^\la)^T K\dsone_N.
\]


\subsection{Equivalence of constrained and penalized formulations}
 
\noindent{\bf Case of ridge regression.} Returning to the basic case (without feature space), we now introduce an alternate formulation
of ridge regression. Let $\mathit{ridge}(\la)$ denote
the ridge regression problem that we have considered so far, for some parameter $\la$. Consider now the
following problem, which will be called $\mathit{ridge}'(C)$:
minimize $\sum_{k=1}^N |y_k - \tx_k^T\be|^2$ subject to the constraint
$\be^T\De\be \leq C$. We claim that this problem is
equivalent to the ridge regression problem, in the following sense:
for any $C$, there exists a $\la$ such that the solution of
$\mathit{ridge}'(C)$ coincides with the solution of
$\mathit{ridge}(\la)$ and vice-versa. \bigskip

Indeed, fix a $C>0$. Consider an optimal $\be$ for $\mathit{ridge}'(C)$. Assuming as above that $\De$ is symmetric positive semi-definite, we let $V$ be its null space and $P_V$ the orthogonal projection on $V$. Write $\be = \be_1 + \be_2$ with $\beta_1 = P_V\be$. Let $d_1$ and $d_2$ be the respective dimensions of $V$ and $V^\perp$ so that $d_1+d_2 = d$. Identifying $\mR^d$ with the product space $V \times V^\perp$ (i.e., making a linear change of coordinates), the problem can be rewritten as the minimization of 
\[
|\CY - \CX_1 \be_1 - \CX_2 \be_2|^2
\]
subject to $\be_2^T \De\be_2 \leq C$, where $\CX_1$ (resp. $\CX_2$) is $N\times d_1$ (resp. $N\times d_2$).

The gradient of the constraint $\ga(\be_2) = \be_2^T \De \be_2 - C$ is $\nabla \ga(\be_2) = 2 \De\be_2$. Assume first that $\De \beta_2 \neq 0$. Then the solution must satisfy the KKT conditions, which require that there exists  $\mu \geq 0$ such that $\be$ is a stationary
point of the Lagrangian
$$
|\CY - \CX_1 \be_1 - \CX_2\be_2|^2 + \mu \be_2^T\De\be_2,
$$
with $\mu > 0$ only possible if $\be^T\De\be = C$. This requires that 
\[
\begin{aligned}
&\CX_1^T\CX_1 \be_1 + \CX_1^T\CX_2 \be_2  = \CX^T\CY,\\
&\CX_2^T\CX_1 \be_1 + \CX_2^T\CX_2 \be_2  + \mu \De \be_2 = \CX^T\CY.
\end{aligned}
\]
Since $\De\be_1 = 0$, and using $\CX = (\CX_1, \CX_2)$, we have
\[
\be  = (\CX^T\CX +\mu \De)^{-1} \CX^T Y,
\]
which is the only solution of  $\text{ridge}(\mu)$.

If $\De\be_2 = 0$, then, necessarily, $\be_2 = 0$. Since $C>0$, $\be$ must then be the solution of the unconstrained problem, which is  $\text{ridge}(0)$.

Conversely, any solution $\be$ of
$\mathit{ridge}(\la)$ satisfies the first-order optimality conditions for $\mathit{ridge}'(C)$ for $C =
\be^T \De \be$ (or any $C \geq
\be^T \De \be$ if $\la=0$). This shows the equivalence of the two problems.

\paragraph{General case.}
We now consider this equivalence in a more general setting. Consider a
 penalized optimization problem, denoted $\mathit{var}(\la)$
which consists in minimizing in $\be$ some objective function of the form $U(\be)
+ \la \phi(\be), \la \geq 0$. Consider also the family of problems $\mathit{var}'(C)$, with $C>\inf(\phi)$, which minimize
$U(\be)$ subject to $\phi(\be) \leq C$.

We make the following assumptions.
\begin{enumerate}[label=(\roman*)]
\item  $U$ and $\phi$ are continuous functions from $\mR^n$ to $\mR$.
\item $\phi(\be) \to \infty$ when $\be\to\infty$. 
\item For any $\la\geq 0$, there is a unique solution
of $\mathit{var}(\la)$,  denoted  $\be_\la$. 
\item For any $C$, there is a unique solution
of $\mathit{var}'(C)$.  denoted  $\be'_C$. 
\end{enumerate}
Assumptions (ii) and (iv) are true, in particular, when $U$ is strictly convex, $\phi$ is convex and $U$ has compact level sets.
We show that, with these assumptions,  the
two families of problems are equivalent.

We first discuss the penalized problems and prove the following
proposition, which has its own interest.
\begin{proposition}
\label{prop:pen.prop}
The function $\la \mapsto U(\be_\la)$ is nondecreasing, and
$\la\mapsto \phi(\be_\la)$ is non-increasing, with
\[
\lim_{\la \to\infty} \phi(\be_\la) = \inf(\phi).
\]
Moreover, $\be_\la$ varies continuously as a function of $\la$.
\end{proposition} 
\begin{proof}
Consider two parameters $\la$ and $\la'$. We have
\begin{eqnarray*}
U(\be_\la) + \la \phi(\be_\la) &\leq& U(\be_{\la'}) + \la
\phi(\be_{\la'})\\
\text{and } U(\be_{\la'}) + \la'
\phi(\be_{\la'}) &\leq& U(\be_\la) + \la' \phi(\be_\la) 
\end{eqnarray*}
since both left-hand sides are minimizers. This implies
\begin{equation}
\label{eq:pen.gen}
\la (\phi(\be_\la) - \phi(\be_{\la'})) \leq U(\be_{\la'}) - U(\be_{\la})
\leq \la' (\phi(\be_\la) - \phi(\be_{\la'})).
\end{equation}

In particular: $(\la'-\la) (\phi(\be_\la) - \phi(\be_{\la'})) \geq
0$. Assume that $\la < \la'$. Then this last inequality implies
$\phi(\be_\la) \geq \phi(\be_{\la'})$ and \eqref{eq:pen.gen} then
implies that $U(\be_\la) \leq U(\be_{\la'})$, which proves the first
part of the proposition.

Now assume that there exists $\ep>0$ such that $\phi(\be_\la) > \inf\phi +\ep$ for all $\la\geq 0$. Take $\tilde \be$ such that $\phi(\tilde \be) \leq \inf\phi + \ep/2$. For any $\la>0$, we have
\[
U(\be_\la) + \la \phi(\be_\la) \leq U(\tilde \be) + \la \phi(\tilde\be)
\]
so that $U(\be_\la) < U(\tilde\be) - \la\ep/2$. Since $U(\be_\la) \geq U({a_0})$, we get $U({a_0}) = -\infty$, which is a contradiction. This shows that $\phi(\be_\la)$ tends to $\inf(\phi)$ when $\la$ tends to infinity.

We now prove that $\la \mapsto \be_\la$ is continuous.
Define $G(\be,
\la)  = U(\be) + \la\phi(\be)$. 
Since we assume that $\phi(\be)\to\infty$ when $\be\to \infty$, and we
have just proved that $\phi(\be_\la) \leq \phi({a_0})$ for any $\la$,
we obtain the fact that the set $(|\be_\la|, \la\geq 0)$ is bounded, say by a
constant $B\geq 0$. 

Consider a sequence $\la_n$ that converges to $\la$. We want to prove
that $\be_{\la_n} \to \be_\la$, for which (because $\be_\la$ is bounded)
it suffices to show that if any subsequence of $(\be_{\la_n})$ converges to some
$\tilde \be$, then $\tilde \be = \be_\la$.

So, consider such a converging subsequence, that we will still denote by $\be_{\la_n}$ for convenience.
Since 
$G$ is continuous, one has $G(\be_{\la_n}, \la_n) \to G(\tilde\be , \la)$ when $n$ tends
to infinity. Let us prove that $G(\be_\la, \la)$ is continuous { in $\la$}.
For any pair $\la, \la'$ and any $\be$, we have
$$
G(\be_{\la'}, \la')\leq G(\be_\la, \la') = G(\be_\la, \la) + (\la'-\la) \phi(\be_\la) \leq G(\be_\la, \la) + 
|\la'-\la| \phi({a_0})\,.
$$
This yields, by symmetry,
$|G(\be_{\la'}, \la') - G(\be_\la, \la)| \leq \phi({a_0}) |\la-\la'|$, proving
the continuity in $\la$.

So we must have $G(\tilde\be, \la) = G(\be_\la, \la)$. This implies
that both $\tilde\be$ and $\be_\la$ are solutions of
$\mathit{var}(\la)$, so that $\be_\la = \tilde \be$ because we assume
that the solution is unique. 
\end{proof}

We now prove
that the classes of problems
$\mathit{var}(\la)$ and $\mathit{var}'(C)$ are  equivalent. First,
$\be_\la$ is  a minimizer of $U(\be)$ subject to the
constraint $\phi(\be) \leq C$, with $C =
\phi(\be_\la)$. Indeed, if $U(\be) < U(\be_\la)$ for some $\be$ with $\phi(\be)
\leq \phi(\be_\la)$, then
$U(\be) + \la \phi(\be) < U(\be_\la) + \la \phi(\be_\la)$ which is a
contradiction. So $\be_\la = \be'_{\phi(\be_\la)}$. Using the continuity of $\be_\la$ and $\phi$, this
proves the equivalence of the problems when $C$ is in the interval $(a,
\phi({a_0}))$ where $a = \lim_{\la\to\infty} \phi(\be_\la) = \inf(\phi)$. 

So, it remains to consider the case  $C>
\phi({a_0})$. For such a $C$,  the solution of $\mathit{var}'(C)$
must be ${a_0}$ since it is a solution of the unconstrained
problem, and satisfies the constraint.

\subsection{Lasso regression}
\label{sec:lasso}
\paragraph{Problem statement}
Assume that the output variable is scalar, i.e., $q=1$.
Let $\hat \sig^2(i)$ be the empirical variance of the
$i$th variable $\pe X i$. Then, the lasso estimator is defined as
a minimizer of
$\sum_{k=1}^N (y_k - \tx_k^T\be)^2$ subject to the constraint
$\sum_{i=1}^d \hat \sig(i)|\pe \be i| \leq C$. Compared to ridge regression, the sum of squares for $\be$ is
simply replaced by a weighted sum of absolute values, but we will see that this change may significantly affect the nature of the solutions. 

As we have just seen, the penalized formulation, minimizing
\[
\sum_{k=1}^N (y_k - \tx_k^T\be)^2 + \la \sum_{i=1}^d \hat \sig(i)|\pe \be i|
\]
provides an equivalent family of problems, on which we will focus (because it is easier to analyze). Since one uses a non-Euclidean norm in the penalty, there is no kernel version of the lasso and we only discuss the method in the original input space $\CR = \mR^d$.

For a vector $a\in \mR^k$, we let $|a|_1 = |\pe a 1| + \cdots + |\pe a k|$, the $\ell^1$ norm of $a$. 
Using the previous notation for $\CY$ and $\CX$, the quantity to minimize can be rewritten as
\[
|\CY - \CX\be|^2 + \la |D\be|_1
\]
where $D$ is the $d \times (d+1)$ matrix with $d(i,i+1) = \hat\sig(i)$ for $i=1, \ldots, d$ and all other coefficients equal to 0. This is a convex optimization problem which, unlike ridge regression, does not have a closed form solution.

\paragraph{ADMM.}
The  alternating direction method of multipliers (ADMM)  that was described in  \cref{sec:convex.optim},  \cref{eq:admm.2} is one of the  state-of-the-art algorithm to solve the lasso problem, especially in large dimensions. Other iterative methods include subgradient descent (see the example in \cref{sec:subg.descent}) and proximal gradient descent. Since $x$ has a different meaning here, we change the notation in \eqref{eq:admm.2} by replacing $x, z, u$ by $\beta, \ga, \tau$, and rewrite the lasso problem as the minimization of 
\[
|\CY - \CX\be|^2 + \la |\ga|_1
\]
subject to $D\be - \ga = 0$. Applying \eqref{eq:admm.3} with $A=D$, $B=-\Id$ and $c=0$, the ADMM iterations are
\[
\left\{
\begin{aligned}
&\be(n+1) &&= \mathrm{argmin}_{\be}  \left(|\CY - \CX\be|^2 + \frac \alpha{2} | D\be - \ga(n) +\tau(n)|^2\right)\\
&\ga^{(i)}(n+1) &&= \mathrm{argmin}_t \left(\la |t| + \frac{\alpha}{2} (t - D\be^{(i)}(n+1) - \tau^{(i)}(n))^2 \right), \ i=1, \ldots, d\\
&\tau(n+1) &&=  \tau(n) + D\be(n+1) - \ga(n+1)
\end{aligned}
\right.
\]
The solutions of both minimization problems are explicit, yielding the following algorithm.
\begin{algorithm}[ADMM for lasso]
\label{alg:admm.lasso}
Let $\rho>0$ be chosen. Starting with initial values $\be^{(0)}$, $\ga^{(0)}$ and $\tau^{(0)}$, the ADMM algorithm for lasso iterates: 
\[
\left\{
\begin{aligned}
&\be(n+1) 
&&= \left(\CX^T\CX + \frac{\alpha }{2}D^TD\right)^{-1} \left(\CX^T\CY + \frac{\alpha }{2} D^T(\ga(n) - \tau(n))\right)\\
&\ga^{(i)}(n+1) 
&& = S_{\la/\alpha}\left(D\be^{(i)}(n+1) + \tau^{(i)}(n)\right),\ i=1, \ldots, d\\
&\tau(n+1) &&=  \tau(n) + D\be(n+1) - \ga(n+1)
\end{aligned}
\right.
\]
until the difference between the variables at steps $n$ and $n+1$ is below a small tolerance level. 
Here,  $S_{\mu}$ is the so-called {\em shrinkage operator}
\[
S_{\mu}(v) = \sign(v) \max(|v|-\mu, 0) = \left\{
\begin{aligned}
&v - \mu &&\text{ if } v \geq \mu\\
&0 &&\text{ if } |v| \leq \mu\\
&v + \mu &&\text{ if } v \leq -\mu
\end{aligned}
\right.
\]
\end{algorithm}
Note that the ADMM algorithm makes an iterative approximation of the constraints, so that they are only satisfied at some precision level when the algorithm is stopped.

\paragraph{Exact computation.}

We now provide a more detailed characterization of the solution of the lasso problem and analyze, in particular, how this solution changes when
$\la$ (or $C$) varies. To simplify the exposition, and without loss of generality, we will
assume that the variables have been normalized so that $\hat \sig(i) = 1$
and the penalty simply is the sum of absolute values. 
Let
$$
G_\la(\be) = \sum_{k=1}^N (y_k - {a_0} - x_k^Tb)^2 + \la \sum_{i=1}^d |b(i)|.
$$

The following proposition, in which we let
\[
r_b = \frac1N \sum_{k=1}^N (y_k-{a_0} - x_k^Tb) x_k,
\]
characterizes the solution of the lasso.
\begin{proposition}
\label{prop:lasso}
The pair $({a_0}, b)$ is the optimal solution of the lasso problem with parameter $\la$ if and only if ${a_0} = \bar y - \bar x^T b$ and, for all $i=1, \ldots, d$, 
\begin{equation}
\label{eq:lasso.resid}
|\pe {r_{b}}i| \leq \frac{\la}{2N} 
\end{equation}
with 
\begin{equation}
\label{eq:lasso.resid.2}
\pe{r_{b}}i = \mathrm{sign}(\pe b i) \frac{\la}{2N} \text{ if } \pe b i \neq 0.
\end{equation}
In particular $|\pe {r_{b}} i| < \la/(2N)$ implies $\pe b i = 0$.
\end{proposition}
\begin{proof}
Using the subdifferential calculus in  \cref{th:subg.sum}, one can  compute the subgradients of $G$ by adding the subdifferentials of the terms that compose it. All these terms are differentiable except $|\pe b i|$ when $\pe b i = 0$, and the  subdifferential of $t\mapsto |t|$ at $t=0$ is the interval $[-1,1]$. 

This shows that $g\in \prt G_\lambda(\beta)$ if and only if
\[
g = -2N r_b + \lambda z
\]
with $\pe z i = \sign(\pe b i)$ if $\pe b i \neq 0$ and $|\pe z i| \leq 1$ otherwise. 
\Cref{prop:lasso} immediately follows by taking $g=0$.
\end{proof}

Let $\zeta = \mathrm{sign}(b)$, the  vector formed by the signs of the coordinates of $b$, with $\mathrm{sign}(0) = 0$.
Then \cref{prop:lasso}  uniquely specifies ${a_0}$ and $b$ once $\la$ and $\zeta$ are known. Indeed, let $J = J_\zeta$ denote the ordered subset of indices $j\in\{1, \ldots, d\}$  such that $\pe \zeta j  \neq 0$, and let $b(J)$, $x_k(J)$, $\zeta(J)$, etc., denote the restrictions of vectors to these indices.  \Cref{eq:lasso.resid.2} can be rewritten as (after replacing ${a_0}$ by its optimal value)  
\[
\CX_c(J)^T\CX_c(J) b(J) = \CX_c(J)^T\CY_c - \frac\la2 \zeta(J)
\]
where
\[
\CX_c(J) = \begin{pmatrix} (x_1(J) - \bx(J))^T\\\vdots\\ (x_N(J) - \bx(J))^T\end{pmatrix}.
\]
This yields
\begin{equation}
\label{eq:lasso.b}
b(J) = (\CX_c(J)^T\CX_c(J))^{-1} \big(\CX_c(J)^T\CY_c - \frac\la2 \zeta(J)\big), 
\end{equation}
which fully determine $b$ since $\pe b j = 0$ if $j\not\in J$, by definition.

For given $\la$, only one sign configuration $\zeta$ will provide the correct solution, with correct signs for nonzero values of $b$ above, and correct inequalities on $r_{b}$. Calling this configuration $\zeta_{\la}$, one can note that
if $\zeta_\la$ is known for a given value of $\la$, it  remains valid if we  increase or decrease $\la$ until one of the optimality conditions changes, i.e., either one of the coordinates $\pe b i, i\in J_{\zeta_\la}$, vanishes, or one of the inequalities  for $i\not \in J_{\zeta_\la}$ becomes an equality. Moreover, 
\cref{prop:lasso} shows that between these events both $b$ and therefore $r_{b}$ depend linearly on $\la$, which makes easy the task of determining maximal intervals around a given $\la$ over which $\zeta$ remains unchanged.

Note that solutions are known for $\la=0$ (standard least squares) and for $\la$ large enough (for which $b=0$).
Indeed, for $b=0$ to be a solution, it suffices that
\[
\la > \la_{0} \defeq 2 \max_{i} \left|\sum_{k=1}^N (y_k -\by)
(\pe{x_k}i -\pe \bx i)\right|.
\]

These remarks set the stage for an algorithm computing the optimal solution of the lasso problem for all values of $\la$, starting either from $\la=0$ or $\la > \la_0$. We will describe this algorithm starting for $\la>\la_0$, which has the merit to avoid complications due to underconstrained least squares when $d$ is large. For this purpose, we need a little more  notation.
For a given $\zeta$, let
\[
b_\zeta = (\CX_c(J_\zeta)^T\CX_c(J_\zeta))^{-1} \CX_c(J_\zeta)^T\CY_c
\]
and 
\[
u_\zeta = \frac12  (\CX_c(J_\zeta)^T\CX_c(J_\zeta))^{-1} \zeta(J_\zeta),
\]
so that $b(J_\zeta) = b_\zeta - \la u_\zeta$. 
The residuals then take the form
\[
\begin{aligned}
\pe{r_{b}}i &= \frac1 N \sum_{k=1}^N (y_k - {a_0} - b_\zeta^Tx_k) \pe {x_k}i + \frac\la N \sum_{k=1}^N (x_k^T u_\zeta)
(\pe{x_k}i -\pe \bx i) \\
&= \pe{\rho_{\zeta}}i + \la \pe{d_{\zeta}}i,
\end{aligned}
\]
where the last equation defines $\rho_\zeta$ and $d_\zeta$. 

Assume that one wants to minimize 
$G_{\la^*}$ for some $\la^* > 0$. We need to describe the sequence of changes to the minimizers of $G_\la$ when $\la$ decreases from some value larger than $\la_0$ to the value $\la^*$. 

If $\la^*\geq \la_0$, then the optimal solution is $b=0$, so we can assume that $\la^* < \lambda_0$. 
When $\la$ is slightly smaller than $\la_{0}$, one needs to introduce some non-zero values in $\zeta$.
Those values are at the indexes $i$ such that  
\[
\la_{0} = 2 \Bigg|\sum_{k=1}^N (y_k -\by)
(\pe {x_k} i - \pe{\bx}i)\Bigg|
\]
The sign of $\pe \zeta i$ is also determined since $\mathrm{sign}(\pe b i) = \mathrm{sign}(\pe {r_{b}}i)$ when $\pe b i\neq 0$.

The algorithm will then continue by progressively adding non-zero entries to $\zeta$ when the covariance between some unused variables and the residual becomes too large, or by removing non-zero values when the optimal $b$ crosses a zero. We now describe it in detail.

\begin{algorithm}[Exact minimization for lasso]
\label{alg:lasso.exact}
\begin{enumerate}[label = \arabic*., wide=0.25cm]
\item Initialization: let $\la(0) = 1 +  \la_0$, $\sigma(0) = 0$ and the corresponding values $a_0(0)=\by$ and $b(0) = 0$. 
\item Assume that the algorithm has reached step $n$ with current variables $\la(n)$, $\sigma(n)$, ${a_0}(n)$ and $b(n)$. 
\item Determine the first $\la'<\la(n)$ for which either
\begin{enumerate}[label=(\roman*),wide=.5cm,leftmargin=.5cm]
\item For some $i$, $\zeta^{(i)}(n) \neq 0$ and  $b_{\zeta(n)}^{(i)} - \la' u_{\zeta(n)}^{(i)} = 0$.
\item For some $i$, $\zeta^{(i)}(n) = 0$ and $(1- 2Nd_{\zeta(n)}^{(i)}) \la'- 2N\rho_{\zeta(n)} = 0$.
\item For some $i$, $\zeta^{(i)}(n) = 0$ and $(1+2Nd_{\zeta(n)}^{(i)}) \la'+ 2N\rho_{\zeta(n)} = 0$.
\end{enumerate}
\item 
Then, there are two cases: 
\begin{enumerate}[label=(\alph*), wide=0.5cm,leftmargin=.5cm]
\item If $\la' \geq \la^*$, set $\la(n+1) = \la'$. Let $\zeta^{(i)}(n+1) = \zeta^{(i)}(n)$ if $i$ does not satisfy (i), (ii) or (iii). If $i$ is in case (i), set $\zeta^{(i)}(n+1) = 0$. For $i$ in case (ii) (resp. (iii)), set $\zeta^{(i)}(n+1) = 1$ (resp. $-1$).
\item
If $\la' < \la^*$, terminate the algorithm without updating $\zeta$ and set
\[
\pe b i =\pe{b_{\zeta(n)}} i - \la^* \pe{u_{\zeta(n)}} i, \quad \zeta^{(i)}(n) \neq 0
\]
and ${a_0} = \bar y - b^T\bar x$ to obtain the final solution.
\end{enumerate}
\end{enumerate}
\end{algorithm}

\section{Other Sparsity Estimators}
\subsection{LARS estimator}
\paragraph{Algorithm.}
The LARS algorithm can be seen as a simplification of the previous lasso algorithm in which one always adds active variables at each step.
We assume as above that input variables are normalized such that $\hat\sig(i) = 1$.

Given a current set $J$ of selected variables, the algorithm will decide either to stop  or to add a new variable to $J$ according to a criterion that depends on a parameter $\la>0$. 
Let $b_{(J)}\in \mR^{|J|}$ be the least-square estimator based on variables in $J$
\[
b_{(J)} = (\CX_c(J)^T\CX_c(J))^{-1} \CX_c(J)^T\CY_c.
\]
Let $b_J\in \mR^d$ such that $\pe {b_J}i = \pe{b_{(J)}} i $ for $i\in J$ and 0 otherwise. The covariances between the remaining variables and the residuals are given by
\[
\pe{r_J}i = \frac1N \sum_{k=1}^N (y_k -\by -  (x_k-\bx)^Tb_J) 
(\pe{x_k}i -\pe{\bx}i), \quad i\not\in J.
\]
If, for all $i\in J$, $|\pe{r_J}i| \leq \sqrt{\la/N}$, the procedure is stopped.
Otherwise, one adds to $J$ the variable $i$ such that  $|\pe{r_J}i|$ is largest and continues.
\bigskip

\paragraph{Justification.}
Recall the notation $|b|_0$ for the number of non-zero entries of $b$.
Consider the objective function
\[
L(b) = |\CY_c - \CX_cb|^2 + \la |b|_0.
\]
 Let $J$ be the set currently selected by the algorithm, and $b_J$ defined as above. We consider the problem of adding one non-zero entry to $b$. Fix $i\not\in J$, and let $\tilde b\in \mR^d$ have all coordinates equalt to those of  $b_J$ for all except the $i$th one, which is therefore allowed to be non-zero. 
Then
\[
L(\tilde b) = \sum_{k=1}^N \Big(y_k - \by - \sum_{j\in J} (\pe{x_k}j - \bx) \pe b j - (\pe{x_k} i  - \bx)\pe{\tilde b}i\Big)^2 + \la |J| + \la,
\]
so that (using $\hat\sig(i) = 1$)
\[
L(\tilde b) = L(b_J) - 2 N \pe{r_J}i \pe{\tilde b}i + N(\pe{\tilde b}i)^2 + \la
\]
 Now, $L(\tilde b)$ is an upper-bound for $L(b_{J\cup\{i\}})$, and so is its minimum with respect to $\pe{\tilde b} i$. This yields:
\[
L(b_{J\cup\{i\}}) \leq L(b_J) - N (\pe{r_J}i)^2 + \la
\]
 The LARS algorithm therefore finds the value of $i$ that minimizes this upper-bound, provided that the resulting minimum is less that $L(b_J)$.
\bigskip

\paragraph{Variant.}
 The same argument can be made with $|b|_0$ replaced by $|b|_1$ and one gets
\[
L(\tilde b) = L(b_J) - 2 N \pe{r_J}i \pe{\tilde b}i + N(\pe{\tilde b}i)^2 + \la |\pe{\tilde b}i|
\]
Minimizing this expression with respect to $\pe{\tilde b}i$ yields the upper bound:
\[
L(b_{J\cup\{i\}}) \leq \left\{
\begin{aligned}
&L(b_J) - N \big(|\pe{r_J}i| -\frac\la{2N}\big)^2 && \text { if } |\pe{r_J}i| \geq \frac\la{2N}\\
&L(b_J)  && \text { if } |\pe{r_J}i| \leq \frac\la{2N}
\end{aligned}
\right.
\]

This leads to the following alternate form of LARS.
 Given a current set $J$ of selected variables, compute
\[
\pe{r_J}i = \frac1N \sum_{k=1}^N (y_k -\by -  (x_k-\bx)^Tb_J) 
(\pe{x_k} i -\pe{\bx}i), \quad i\not\in J\,.
\]
If, for all $i\not\in J$, $|\pe{r_J}i| \leq \la/2N$, stop the procedure.
 Otherwise, add to $J$ the variable $i$ such that  $|\pe{r_J}i|$ is largest and continue.
 This form tends to add more variables since the stopping criterion decreases in $1/N$ instead of $1/\sqrt{N}$.
 \bigskip

\paragraph{Why ``least angle''?}
 Let $\mu_{J,k}  = y_k - \by- (x_k -\bx)^T b_J$ denote the residual after regression. 
 The empirical correlation between $\mu$ and $\pe x i $ is equal to the cosine of the angle, say $\pe{\th_J}i$ between $\mu_J\in \mR^N$ and $\pe x i  -\bx$ both considered as vectors in $\mR^N$. This cosine is also equal to
\[
\cos \pe{\th_J}i = \frac{\mu_J^T(\pe x i  -\pe\bx i)}{|\pe x i -\pe \bx i| \,|\mu_J|} =\sqrt N \frac{\pe{r_J} i}{|\mu_J|}
\]
where we have used the fact that $|\pe x i  -\pe \bx i|/\sqrt N = \hat\sig(i) = 1$. 
 Since $|\mu_J|$ does not depend on $i$, looking for the largest value of $|\pe {r_J} i |$ is equivalent to looking for the smallest value of $|\pe{\th_J}i |$, so that we are looking for the unselected variable for which the angle with the
current residual is minimal.

\subsection{The Dantzig selector}

\paragraph{Noise-free case.}

 Assume that one wants to solve the equation
$\CX \be = \CY$ when the dimension, $N$, of $\CY$ is small compared to
number of columns, $d$, in $\CX$. 
 Since the system is under-determined, one needs additional constraints on $\be$ and a natural one is to look for sparse solutions, i.e., find  solutions with a maximum number of zero coefficients. However, this  is numerically challenging, and it is easier to minimize the  $\ell^{1}$ norm  of $\beta$ instead (as seen when discussing
the lasso, using this norm often provides sparse
solutions).
In the following,   we
assume that the empirical variance of each variable is normalized, so
that, denoting $\CX(i)$ the $i$th column of $\CX$, we have $|\CX(i)| = 1$.

The Dantzig selector \cite{ct05} minimizes
$$
\sum_{i=1}^d |\pe \be i|
$$
subject to the constraint $\CX \be = \CY$. This results in  a linear
program (therefore easy to implement). More precisely, introducing slack variables, it is indeed
equivalent to minimize
\[
\sum_{i=1}^d \xi(i) + \sum_{i=1}^d \xi^*(i)
\]
 subject to constraints
$\xi(i) \geq \pe \be i $, $\xi(i)^*\geq -\pe \be i $, $\xi(i)\geq 0$, $\xi^*(i) \geq 0$ and $\CX \be = \CY$. 
\bigskip

\paragraph{Sparsity recovery}

Under some assumptions, this method does recover sparse solutions when they exist. More precisely, let $\hat\be$ be the solution
of the linear programming problem above. Assume that there is a set $J^*\sub\{1,
\ldots, d\}$ such that $\CX\be = \CY$ for some $\be \in \mathbb R^d$
with $\pe \be i  = 0$ if $i\not\in J^*$. 
Conditions under which 
$\hat\be$ is equal to
$\be$ are provided in \citet{ct05} and involve
the correlations between pairs of columns of $\CX$, and the size of $J$.

That the size of $J^*$ must be a factor is
clear, since, for the statement to make sense, there cannot exist two
$\beta$'s satisfying $\CX\be = \CY$ and $\pe \be i  = 0$ for $i\not\in
J^*$. Uniqueness is obviously not true if $|J|>N$, because, even if one knew $J$,  the condition
would be under-constrained for $\be$. Since the set
$J^*$ is not known, and we also want to avoid any other solution associated
to a set of same size. So, there
cannot exist $\be$ and $\tilde\be$ respectively vanishing outside of $J^*$ and
$\tilde J^*$, where $J^*$ and $\tilde J^*$ have same cardinality, such that $\CX\be = \CY = \CX\tilde \be$. The equation
$\CX(\be-\tilde\be) = 0$ would be under-constrained as soon as the number of non-zero
coefficients of $\be-\tilde\be$ is larger than $N$, and since this
number can be as large as $|J^*|+|\tilde J^*| = 2|J^*|$, we see that one should
impose at least $|J^*| \leq N/2$.

Given this restriction, another obvious remark is that, if the set $J$ on which $\be$ does not
vanish is known, with $|J|$ small enough, then $\CX\be = \CY$ is
over-constrained and any solution is (typically) unique. So the issue really is
whether the set $J_\be$ listing the non-zero indexes of a solution $\be$ is equal to y $J^*$. 

As often, precious insight on the solution of this minimization problem  is obtained by considering the dual problem. Introducing
Lagrange multipliers $\la(i)\geq 0, i=1, \ldots, d$ for the constraints $\xi(i) - \pe \be i  \geq 0$,   $\la^*(i)\geq 0$, $i=1, \ldots, d$ for $\xi^*(i) + \pe \be i  \geq 0$, $\ga(i), \ga^*(i)\geq 0 $ for $\xi(i) \geq 0$ and $\xi^*(i) \geq 0$,  and $\al\in \mR^N$ for $\CX \be = \CY$, the Lagrangian is
\[
\CL(\be, \xi, \la, \la^*, \al) = (\dsone_d - \la - \ga)^T\xi + (\dsone_d - \la^* - \ga^*)^T\xi^* + (\la-\la^* + \CX^T\al)^T\be - \al^T\CY.
\]
The KKT conditions require $\ga = \dsone_d - \la$, $\ga^* = \dsone_d - \la^*$, $\CX\al = \la^* - \la$ and the complementary slackness conditions
give $(1-\la(i))\xi(i) = (1-\la^*_i)\xi^*(i) = 0$, $\la(i)(\pe \be i - \xi(i)) = \la^*(i)(\pe\be i  + \xi^*(i)) = 0$. 

The dual problem requires to minimize $\al^T\CY$ subject to the constraints $\CX^T\al = \la^* - \la$ and $0\leq \la(i), \la^*(i) \leq 1$.
Assume that $(\al, \la, \la^*)$ is a solution of this dual problem. One has the following cases.
\begin{enumerate}[label = (\arabic*)]
\item If $\la(i)\in (0,1)$, then $\xi(i) = \pe \be i - \xi(i) = 0$, which implies $\xi(i) = \pe \be i = 0$, and, as a consequence $(1-\la^*(i))\xi^*(i) = \la^*(i)\xi^*(i) = 0$, so that also $\xi^*(i)=0$ . 
\item Similarly, $\la^*(i) \in (0,1)$ implies $\xi(i) = \xi^*(i) = \pe \be i = 0$. 
\item If $\la(i)= \la^*(i) = 1$, then $\pe \be i  - \xi(i) = \pe \be i + \xi(i) =0$ with $\xi(i), \xi^*(i) \geq 0$, so that also $\xi(i) = \xi^*(i) = \pe \be i = 0$. 
\item If $\la(i)= \la^*(i) = 0$, then $\xi(i) = \xi^*(i) = 0$ and since $\pe \be i \leq \xi(i)$ and $\pe \be i \leq - \xi^*(i)$, we get $\pe \be i = 0$.
\item The only remaining situation, in which $\pe \be i $ can be non-zero, is when $\la(i) = 1-\la^*(i) \in \{0,1\}$, or, equivalently, when $|\la(i) - \la^*(i)|=1$.  
\end{enumerate}
This discussion allows one to reconstruct the set $J_\be$ associated with the primal problem given the solution of the dual problem. Note that $|\la(i) - \la^*(i)| = |\al^T\CX(i)|$, so that the set of indexes with  $|\la(i) - \la^*(i)|=1$ is also
\[
I_\al \defeq \defset{i: |\al^T\CX(i)| = 1}.
\]

One has 
\[
\al^T\CY = \al^T\CX\be = \sum_{i=1}^d \pe \be i \al^T\CX(i) \leq \sum_{i\in J_\be} |\pe \be i| \, |\al^T\CX(i) | \leq \sum_{i\in J_\be} |\pe \be i|.
\]
The upper-bound is achieved when $\al^T\CX(i) = \mathrm{sign}(\pe \be i)$ for $i\in
J_\be$. So, if a vector $\al$ can be found such that
\begin{enumerate}[label=(\roman*)]
\item $\al^T\CX(i) = \mathrm{sign}(\pe \be i)$ for $i\in
J^*$,
\item $|\al^T\CX(j)|<1$
for $j\not \in J^*$,
\end{enumerate}
then it is a solution of the dual problem with
$J_\al = J^*$. 
\bigskip

Let $s_J =
(\pe s j, j\in J)$ be defined by  $\pe s j = \mathrm{sign} (\pe \be j)$.
One can always decompose $\al\in \mR^N$ in the form
\[
\al = \CX_{J^*} \rho + w
\]
where $\rho \in \mR^{|J^*|}$ and $w\in \mR^N$ is perpendicular to the columns of $\CX_{J^*}$. From $\CX_{J^*}^T \al = s_J$, we get
\[
\rho = (\CX_{J^*}^T\CX_{J^*})^{-1} s_{J^*}.
\]
Letting $\al_{J^*}$ be the solution with $w=0$, the question is
therefore whether one can find $w$ such that
$$
\left\{\begin{array}{ll}
w^T\CX(j) = 0, & j\in J^*\\
|\al_J^T \CX(k) + w^T \CX(k)| < 1, & k\not \in J^*
\end{array}
\right.
$$

Denote for short $ \Sig_{JJ'} = \CX_J^T\CX_{J'}$. 
One can show that such a solution exists when the matrices
$ \Sig_{JJ}$ are close to the identity as soon as $|J|$ is small enough \cite{ct05}. More precisely, denote, for $q\leq d$
$$
\de(q) =\max_{|J| \leq q} \max(\|\Sig_{JJ}\|, \|\Sig^{-1}_{JJ}\|^{-1}) - 1,
$$
in which one uses the operator norm on matrices,
and
\[
\th(q,q') = \max \defset{z^T \Sig_{TT'} z':  |J|,|J'|\leq q. J\cap J'=\emp, |z| = |z'| = 1} .
\]
 Then, the
following proposition is true.
\begin{proposition}[Candes-Tao]
Let $q = |J^*|$ and $s = (\pe s j, j\in J^*)\in \mR^q$. Assume that $\de(2q)
+ \th(q,2q) <1$. Then there exists $\al\in \mR^N$ such that
$\al^T\CX(j) = \pe s j$ for $j\in J^*$ and
$$
|\al^T\CX(j)| \leq \frac{\th(q,q)}{1-\de(2q) - \th(q,2q)} \text{ if }
j\not\in J^*.
$$
\end{proposition}
So $\al$ has the desired property as soon as $\de(2q) + \th(q,q) +
\th(q,2q) \leq 1$. 
to control subsets of variables of size less than $3q$ to obtain the
conclusion, which is important, of course, when $q$ is small compared to
$d$.
\bigskip

\paragraph{Noisy case}

Consider now the noisy case. We here again introduce quantities that
were pivotal for the lasso and LARS estimators, namely, the covariances
between the variables and the residual error. So, we define, for a
given $\be$
$$
r_\be^{(i)} = \CX(i)^T(\CY - \CX\be)
$$
which depends linearly on $\be$. Then, the {\em Dantzig selector} is defined
by the linear program: Minimize:
$$
\sum_{j=1}^d |\be^{(j)}|
$$
subject to the constraint:
$$
\max\limits_{j=1, \ldots, d} |r_\be^{(j)}| \leq C.
$$
The explicit expression of this problem as a linear program is obtained as
before by introducing slack variables $\xi(j), \xi^*(j), j=1, \ldots, d$ and
minimizing 
$$
\sum_{j=1}^d \xi(j) + \sum_{j=1}^d \xi^*(j)
$$
with constraints $\xi(j), \xi^*(j) \geq 0$, $\xi \geq \be$, $\xi^*\geq -\be$, $
\max\limits_{j=1, \ldots, d} |r_\be^{(j)}| \leq C.
$

Similar to the noise-free case, the Dantzig selector can identify sparse
solutions (up to a small error) if the columns of $\CX$ are nearly orthogonal, with the same
type of conditions \cite{ct07}. Interestingly enough, the accuracy of
this algorithm can be proved to be comparable to that of the lasso in the
presence of a sparse solution \cite{brt08}.

\section{Support Vector Machines for regression}
\label{sec:svm.reg}

\subsection{Linear SVM}

\paragraph{Problem formulation}
We start by discussing support vector machines (SVM) \cite{vapnik1998statistical,vapnik2013nature} with $\CR_X = \mR^d$ equipped with the standard inner product (generally referred to as linear SVM) and will extend the theory to kernel methods in the next section.
SVMs solve a linear regression problem, but replace the least-squares loss function by $(y,y') \mapsto V(y-y')$ with
\begin{equation}
\label{eq:svm.v}
V(t) = \left\{
\begin{aligned}
 &0 && \text{ if } |t| < \ep\\ 
 &|t|-\ep && \text{ if } |t| \geq \ep
 \end{aligned}\right.
\end{equation}

\begin{figure}[h]
\centering \includegraphics[width=0.6\textwidth]{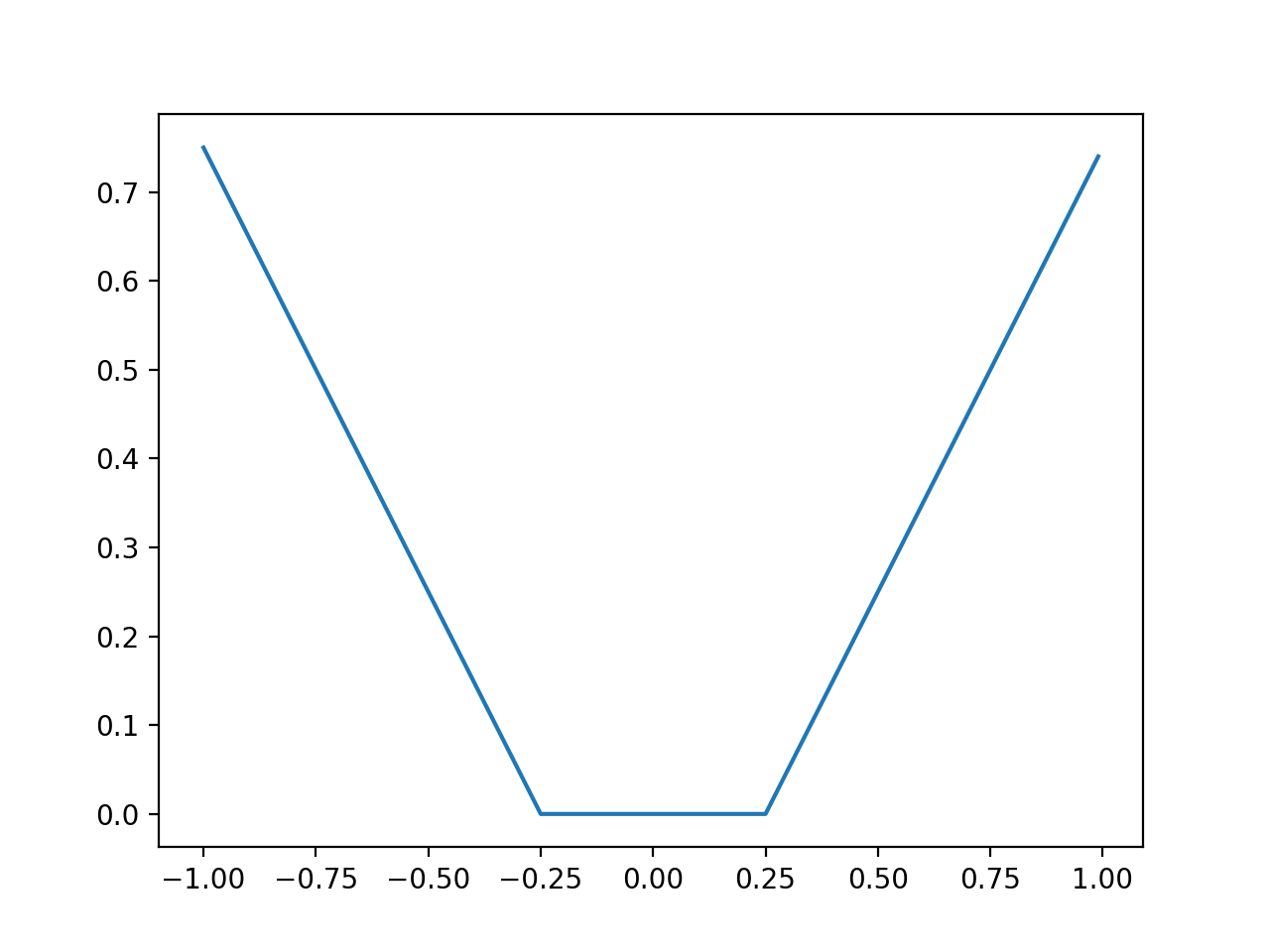}
\caption{\label{fig:SVM.V} The function $V$ defining the SVM risk function.}
\end{figure}

A plot of the function $V$ is provided in \cref{fig:SVM.V}. This function is an example of what is often called a robust
loss function. The quadratic error used in linear regression had the
advantage of providing closed form expressions for the solution, but is
quite sensitive to outliers. For robustness, it is preferable to use loss
functions that, like $V$, increase at most linearly at infinity. One sometimes  choose
them as smooth convex functions, for example $V(t) = (1-\cos\ga
t)/(1-\cos\ga\ep)$ for $|t|<\ep$ and $f(t) = |t|$ for $t \geq \ep$,
where $\ga$ is chosen so that $\ga\ep\sin\ga\ep /(1-\cos\ga\ep) =
1$. In such a case, minimizing
$$
F(\be) = \sum_{k=1}^N V(y_k - {a_0} - x_k^T b)
$$
can be done using gradient descent methods. Using  $V$ in \cref{eq:svm.v} will require a little more work, as we see now. 

The SVM regression problem is generally formulated as the minimization of
$$ 
\sum_{k=1}^N V(y_k - {a_0} - x_k^T b) + \la |b|^2\,,
$$
and we will study a slightly more general problem, minimizing
$$ 
F({a_0}, b) = \sum_{k=1}^N V(y_k - {a_0} - x_k^T b) + \la b^T \De b\,,
$$
where $\De$ is a symmetric positive-definite matrix.
This objective function exhibits the
following features:
\begin{enumerate}[label=$\bullet$,wide]
\item A penalty on the coefficients of $b$, similar to ridge regression.
\item A linear penalty (instead of quadratic) for large errors in the
prediction.
\item An $\ep$-tolerance for small errors, often referred to as {\em the margin} of the regression SVM.
\end{enumerate}

We now  describe the various steps in the analysis and reduction of
the problem. They will lead to simple minimization algorithms, and
possible extensions to nonlinear problems.

\paragraph{Reduction to a quadratic programming problem} 
Introduce  slack
variables $\xi_k, \xi_k^*, k=1, \ldots, N$. The original problem
is equivalent to the minimization, with respect to $({a_0}, b, \xi, \xi^*)$, of
$$
\sum_{k=1}^N (\xi_k+ \xi_k^*) + \la b^T\De b
$$
under the constraints:
\begin{equation}
\label{eq:svm.constr}
\left\{
\begin{aligned} 
&\xi_k, \xi_k^* \geq 0\\ 
&\xi_k - y_k + {a_0} + x_k^Tb +\ep \geq 0\\
&\xi_k^* + y_k - {a_0} - 
x_k^Tb +\ep \geq 0
\end{aligned}
\right.
\end{equation}
The simple proof of this equivalence, which results in a quadratic programming problem, is left to the reader. As often, one gains additional insight by studying the dual problem.

\paragraph{Dual problem}
Introduce $4N$ non-negative Lagrange multipliers for the  $4N$ constraints in the problem, namely,  $\eta_k, \eta_k^*\geq 0$ for the positivity constraints, and $\al_k,
\al_k^*\geq 0$ for the last two in \eqref{eq:svm.constr}. The resulting Lagrangian is
\begin{multline*}
\mathcal L ({a_0}, b , \xi, \xi^*, \alpha, \alpha^*, \eta, \eta^*) = \sum_{k=1}^N (\xi_k+ \xi_k^*) + \la b^T\De b - \sum_{k=1}^N
(\eta_k\xi_k + \eta_k^*\xi_k^*) \\
- \sum_{k=1}^N \al_k (\xi_k - y_k +
{a_0} + x_k^Tb + \ep)  - \sum_{k=1}^N \al_k^* (\xi^*_k + y_k - {a_0} -
x_k^Tb + \ep) .
\end{multline*}
In this formulation,
$({a_0}, b, \xi, \xi^*)$ are the primal variables, and $\alpha, \alpha^*,
\eta, \eta^*$ the dual variables.

The KKT conditions are provided by the system:
\begin{equation}
\label{eq:SVM.KKT}
\left\{\begin{aligned}
&\sum_{k=1}^N (\al_k-\al_k^*) = 0 \\
&2 \la \De b - \sum_{k=1}^N (\al_k-\al_k^*) x_k = 0\\
&1 - \eta_k - \al_k = 0 \\
&1 - \eta_k^* - \al_k^* = 0 \\
&\al_k (\ep + \xi_k - y_k + {a_0} +  x_k^Tb) = 0  \\
&\al_k^* (\ep + \xi_k^* + y_k - {a_0} - x_k^Tb) = 0  \\
&\eta_k\xi_k = \eta_k^*\xi_k^* = 0
       \end{aligned}
\right.
\end{equation}
The first four equations are the derivatives of the Lagrangian with respect to ${a_0}, b, \xi_k, \xi^*_k$ in this order and the last three are the complementary slackness conditions.

The dual problem maximizes the function 
$$
L^*(\al, \al^*, \eta, \eta^*) = \inf_{\be, \xi, \xi^*} \mathcal L.
$$
 under the previous
positivity constraints. 
Since the Lagrangian is linear in ${a_0}$,
$\xi_k$ and $\xi_k^*$, its minimum is $-\infty$ unless the
coefficients vanish. The linear terms must therefore vanish for $L^*$ to be finite. With these conditions plus the fact that  $\prt_b L = 0$, we retrieve the first four equations of system \eqref{eq:SVM.KKT}. 
Using $\eta_k  = 1- \al_k$, $\eta_k^* = 1- \al_k^*$ and 
\begin{equation}
\label{eq:svm.lin.b}
b = \frac1{2\la} \sum_{k=1}^N (\al_k-\al_k^*) \De^{-1} x_k
\end{equation}
one can 
express $L^*$ uniquely as a function of $\al, \al^*$, yielding
$$
L^*(\al, \al^*) = -\frac{1}{4\la} \sum_{k,l=1}^N
(\al_k-\al_k^*)(\al_l-\al_l^*) x_k^T\De^{-1} x_l - \ep \sum_{k=1}^N
(\al_k+\al_k^*) + \sum_{k=1}^N (\al_k-\al_k^*)y_k\,.
$$
This quantity must be maximized subject to the constraints  $0\leq \al_k,
\al_k^*\leq 1$ and $\sum_{k=1}^N (\al_k-\al_k^*) = 0$. This still is
a quadratic programming problem, but it now has nice additional features
and interpretations.

\paragraph{Step 3: Analysis of the dual problem}
The dual problem only depends on the $x_k$'s
through the matrix with coefficients $x_k^T\De^{-1} x_l$, which is the Gram matrix of $x_1, \ldots, x_N$ for the inner product associated with $\De^{-1}$. This property will lead to the the kernel version of SVMs discussed in the next section. The obtained predictor can
also be expressed as a function of these products, since
$$
y = {a_0} + x^T b = {a_0} + \frac{1}{2\la} \sum_{k=1}^N (\al_k-\al_k^*)
(x_k^T\De^{-1} x)\,.
$$
Moreover, the dimension of the dual problem is $2N$, which  allows the method to be used in large (possibly infinite) dimensions with a bounded cost.

We now analyze the solutions   $\al, \al^*$ of the dual problem. The complementary slackness conditions reduce to:
\begin{equation}
\label{eq:SVM.slackness}
\left\{\begin{aligned}
&\al_k (\ep + \xi_k - y_k + {a_0} + x_k^T b) = 0  \\
&\al_k^* (\ep + \xi_k^* + y_k - {a_0} - x_k^T b) = 0  \\
&(1-\al_k)\xi_k = (1-\al_k^*)\xi_k^* = 0
\end{aligned}
\right.
\end{equation}
These conditions have the following consequences,  based on the prediction error made for each training sample.
\begin{enumerate}[label=(\roman*), wide=0.5cm]
\item First consider indexes $k$
such that the error is strictly within the tolerance
margin $\epsilon$: $|y_k - {a_0} - x_k^Tb| < \ep$. Then the terms between parentheses in first two equations of \eqref{eq:SVM.slackness} are strictly positive, which implies that $\al_k = \al_k^* = 0$. The last two equations in \eqref{eq:SVM.slackness} then imply $\xi_k  = \xi_k^* = 0$. 
\item Consider now the case when the prediction is strictly less accurate
than the tolerance margin. Assume that $y_k - {a_0} - x_k^Tb > \ep$. The second and third equations in \eqref{eq:SVM.slackness} imply that $\al_k^* = \xi_k^* = 0$. The assumption also implies that 
\[
\xi_k = y_k - {a_0} - x_k^Tb -\ep >0
\]
and $\al_k = 1$. The case $y_k - {a_0} - x_k^Tb < - \ep$ is symmetric and provides $\al_k = \xi_k = 0$, $\xi_k^*  >0$
and $\al_k^* = 1$.

\item Finally, consider samples for which the prediction error is exactly
at the tolerance margin. If $y_k - {a_0} - x_k^T b = \ep$, we have $\al_k^* = \xi_k = \xi_k^* =0$. The fact that $\al_k^* = \xi_k^*=0$ is clear. To prove that $\xi_k=0$, we note that would have otherwise  $\xi_k - y_k + {a_0} + x_k^Tb +\ep > 0$, which would imply that $\al_k=0$ and we  reach a contradiction with $(1-\al_k)\xi_k=0$. Similarly, $y_k - {a_0} - x_k^T b = -\ep$ implies that $\al_k = \xi_k = \xi_k^* =0$.

The points for which $|y_k - {a_0} - x_k^T b| = \ep$ are called {\em support vectors}.
\end{enumerate}

One important information deriving from this discussion is that the variables $(\al_k, \al_k^*)$ have prescribed values as long as the error $y_k - {a_0} - x_k^T b$ is not exactly $\ep$ in absolute value: $(1,0)$ if the error is larger than $\ep$, $(0,0)$ if it is strictly between $-\ep$ and $\ep$ and $(0,1)$ if it is less than $-\ep$. Also in all cases, at least one of $\al_k$ and $\al_k^*$ must vanish. Only in the case of support vectors does the previous discussion fail to provide a value for one of these variables.

Now, we want to reverse the discussion and assume that the dual problem is solved to see how the variables ${a_0}$ and $b$ of the primal problem can be retrieved. For $b$, this is easy, thanks to \eqref{eq:svm.lin.b}. For ${a_0}$ a direct computation can be made if a support vector is identified, either because $0 < \al_k <1$, which implies that ${a_0} = y_k - x_k^Tb -\ep$, or because 
$0 < \al^*_k <1$, which yields ${a_0} = y_k - x_k^Tb +\ep$.

If no support vector can be identified, ${a_0}$ is not uniquely determined (note that the objective function is not strictly convex in ${a_0}$). However, the coefficients $\al_k, \al_k^*$ provide some information on this intercept, in the form of inequalities. More precisely, let
$J^+ =  \{k : \al_k = 1\}$, $J^- = \{k:\al_k^* = 1\}$ and $J_0 = \{k:\al_k = \al_k^* = 0\}$. Then $k \in J^+$ implies that 
$y_k - {a_0} - b^Tx_k \geq \ep$, so that ${a_0} \leq y_k - b^T x_k -\ep$. Similarly, $k\in J^-$ implies that ${a_0} \geq y_k - b^T x_k +\ep$. Finally, $k\in J_0$ implies that ${a_0} \geq y_k - b^T x_k -\ep$ and ${a_0} \leq y_k - b^T x_k +\ep$. As a consequence, one can take ${a_0}$ to be any point in the interval $[{a_0}^-, {a_0}^+]$, where
\begin{align*}
{a_0}^- &= \max\left(\max_{k\in J^-} (y_k - x_k^Tb + \ep), \max_{k\in J_0} (y_k - x_k^T b - \ep)\right)\\
{a_0}^+ &= \min\left(\min_{k\in J^+} (y_k - x_k^Tb - \ep), \min_{k\in J_0} (y_k - x_k^T b + \ep)\right).
\end{align*}
%
%
%
%

\subsection{The kernel trick and SVMs}
Returning to our feature space notation, let $X$ take values in $\CR_X$ and $h: \CR_X \to H$ be a feature function with values in an inner-product space $H$ with associated kernel $K$. SVMs in feature space must minimize, with ${a_0}\in \mR$ and $b\in H$
$$ 
F({a_0}, b) = \sum_{k=1}^N V\big(y_k - {a_0} - \scp{h(x_k)}{b}_H\big) + \la \|b\|_H^2\,.
$$
Letting as before $V = \vspan(h(x_1), \ldots, h(x_N))$, the same argument as that made for ridge regression works, namely that the first term in $F$ is unchanged if $b$ is replaced by $\pi_V(b)$ and the second one is strictly reduced unless $b\in V$, leading to a finite-dimensional formulation in which 
\[
b = \sum_{k=1}^N c_k h(x_k)
\]
and one minimizes
$$ 
F({a_0}, c) = \sum_{k=1}^N V\Big(y_k - {a_0} - \sum_{l=1}^N K(x_k,x_l) c_l\Big) + \la \sum_{k,l=1}^N K(x_k,x_l) c_kc_l \,.
$$
This function has the same form as the one studied in the linear case with $b$ replaced by $c \in \mR^N$, $x_k$ replaced by the vector with coefficients $K(x_k, x_l), l=1, \ldots, N$, that we will denote $\CK^{(k)}$ and $\De = \CK = \CK(x_1, \ldots, x_N)$. Note that $\CK^{(k)}$ is the $k$th column of $\CK$, so that 
\[
\big(\CK^{(k)}\big)^T \CK^{-1} \CK^{(l)} = K(x_k, x_l).
\]
Using this, we find that the dual problem requires  to maximize
$$
L^*(\al, \al^*) = -\frac{1}{4\la} \sum_{k,l=1}^N
(\al_k-\al_k^*)(\al_l-\al_l^*) K(x_k, x_l) - \ep \sum_{k=1}^N
(\al_k+\al_k^*) + \sum_{k=1}^N (\al_k-\al_k^*)y_k\,.
$$
with
$$
\left\{
\begin{aligned}
&0\leq \al_k \leq 1 \\
&0 \leq  \al_k^* \leq 1 \\
&\sum_{k=1}^N (\al_k-\al_k^*) = 0 \\
\end{aligned}
\right.
$$
The associated vector $c$ satisfies
\[
2 \la c =  \sum_{k=1}^N (\al_k-\al_k^*) \CK^{-1} \CK^{(k)} = \al - \al^*\,.
\]
and the regression function is
\[
f(x) = {a_0} + \scp{b}{h(x)}_H = {a_0} + \frac{1}{2\la} \sum_{k=1}^N (\al_k-\al_k^*)  K(x, x_k)\,.
\]
Finally, the discussions on the values of $\al, \al^*$ and on the computation of ${a_0}$ remain unchanged.

%

\newpage
\markright{PROBLEMS}
\section*{Problems}
\addcontentsline{toc}{section}{Problems}
\input Problems_Linear_Regression

\chapter{Models for linear classification}
\label{chap:lin.class}
In this chapter, $Y$ is categorical and takes values in the finite set $\CR_Y = \defset{g_1,
\ldots, g_q}$. The goal is to predict this class variable from the
input  $X$, taking values in a set $\CR_X$. Using the same progression as in the regression case, we will first discuss basic linear methods, for which $\CR_X = \mR^d$ before extending them, whenever possible, to kernel methods, for which $\CR_X$ can be arbitrary as soon as a feature space representation is available. 

Classifiers will be based on a training set $T = ((x_1, y_1), \ldots, (x_N, y_N))$ with $x_k\in\CR_X$ and $y_k\in \CR_Y$ for $k=1, \ldots, N$. For $g\in \CR_Y$, we will also let $N_g$ denote the number of samples in the training set such that $y_k = g$, i.e., 
\[
N_g = \big|\defset{k: y_k=g}\big| = \sum_{k=1}^N \bfone_{y_k=g}.
\]

\section{Logistic regression} 

\subsection{General Framework}
Logistic regression uses the fact that, in order to apply  Bayes's
rule, only the conditional distribution of the class variables $Y$
given $X$ is needed, and trains a parametric model of this distribution. More precisely, if one denotes by $p(g|x)$  the probability that  $Y=g$ conditional to $X=x$, logistic regression assumes that, for some parameters $({a_0}(g), b(g), g\in \CR_Y)$ with ${a_0}(g) \in \mR$ and $b(g)\in \mR^{d}$, one has $p = p_{{a_0},b}$ with 
\[
\log p_{{a_0},b}(g\mid x) = {a_0}(g) +  x^T b(g) - \log(C({a_0}, b, x)),
\]
where
$C({a_0}, b, x) = \sum_{g\in \CR_Y} \exp({a_0}(g) + x^T b(g)).$

Introduce the functions, defined over mappings $\mu: \CR_Y \to \mR$ (which can be identified with vectors in $\mR^q$)  
\begin{equation}
\label{eq:log.reg.F}
F_g(\mu) = \mu(g) - \log \sum_{g'\in \CR_Y} e^{\mu(g')}\,.
\end{equation}
With this notation, letting $\be(g) = \begin{pmatrix}{a_0}(g)\\ b(g)\end{pmatrix}$ and $\tx = \begin{pmatrix}1\\ x\end{pmatrix}$, one has 
$\log p_{\be}(g|x) = F_g(\beta^T\tx)$, where $\beta^T\tx$ is the function $(g'\mapsto \beta(g')^T\tx)$.  

For any constant function $(g \mapsto \mu_0\in \mR)$ one has
\[
F_g(\mu+\mu_0) = \mu(g) +\mu_0 - \log \sum_{g'\in \CR_Y} e^{\mu(g')+\mu_0} =  \mu(g) +\mu_0 - \mu_0 - \log \sum_{g'\in \CR_Y} e^{\mu(g')} = F_g(\mu).
\]
As a consequence, if one replaces, for all $g$,  $\be(g)$ by $\tilde \be(g) = \be(g) + \ga$, with $\ga\in \mR^{d+1}$, then $\tilde \beta^T\tx = \beta^T\tx + \gamma^T\tx$ and
\[
\log p_{\tilde \be}(g\mid x) =   \log p_{\be}(g\mid x).
\]
This shows that the model is over-parametrized. 
 One therefore needs a $(d+1)$-dimensional constraint to ensure uniqueness, and we will enforce a linear constraint in the form
\[
\sum_{g\in \CR_Y} \rho_g \be(g) = c
\]
with $\sum_g\rho_g \neq 0$.

\subsection{Conditional log-likelihood}
The conditional log-likelihood computed from the training set is:
\[
\ell(\be) = \sum_{k=1}^N \log p_\be(y_k\mid x_k)\,.
\]
 Logistic regression computes a maximizer $\hat \be$ of this log-likelihood.
 The classification rule given a new input $x$ then chooses the class $g$ for which $p_{\hat \be}(g\mid x)$ is largest, or, equivalently, the class $g$ that maximizes $\tx^{T} \be(g)$.
 
\begin{proposition}
\label{prop:log.reg}
Let  $\tm_{g} = \sum_{k:y_{g}=k} x_{k}/N_{g}$.
The conditional log-likelihood $\ell$ is concave with first derivative
\begin{equation}
\label{eq:log.reg.d1}
\prt_{\beta(g)} \ell  = N_{g} \tm^{T}_{g} -\sum_{k=1}^{N} 
\tilde x_{k}^T p_\be(g|x_{k})
\end{equation}
and negative semi-definite second derivative
\begin{equation}
\label{eq:log.reg.d2}
\prt_{\beta(g)}\prt_{\beta(g')} \ell =  -\bfone_{[g=g']}\sum_{k=1}^{N}\tilde x_{k} \tilde x_{k}^T
 p_\be(g|x_{k}) + \sum_{k=1}^{N}
\tilde x_{k} \tilde x^T_{k} p_\be(g|x_{k}) p_\be(g'|x_{k})\,.
\end{equation}
\end{proposition}
\begin{remark}
\label{rem:log.reg.notation}
In this discussion, we consider $\ell$ as a function defined over collections $(\beta(g), g\in \CR_Y)$, or, if one prefers, on the $q(d+1)$-dimensional linear space, $\CF$, of functions $\beta: \CR_Y \to \mR^{q+1}$. With this in mind, the differential $d\ell(\beta)$ is a linear form from $\CF$ to $\mR$, therefore associating to any family $u = (u(g), g\in\CR_Y)$, the expression
\[
d\ell(\beta) \,u = \sum_{g\in\CR_Y} \prt_{\beta(g)} \ell\, u(g).
\]
Similarly, the second derivative is the bilinear form
\[
d^2\ell(\beta)(u,u') = \sum_{g, g'\in\CR_Y} u(g)^T \prt_{\beta(g)}\prt_{\beta(g')}\ell \, u(g').
\]
The last statement in the proposition expresses the fact that $d^2\ell(\beta)(u,u) \leq 0$ for all $u\in\CF$.
\end{remark}
\begin{proof}
First consider the function $F_g$ in \cref{eq:log.reg.F}, so that 
\[
\ell(\beta) = \sum_{k=1}^N F_{y_k}(\beta^T\tx_k).
\]
We have, for $\zeta: \CR_Y\to \mR$, 
\[
dF_g(\mu)\zeta =  \zeta(g) - \frac{\sum_{g'\in \CR_Y} e^{\mu(g')} \zeta(g')}{\sum_{g'\in \CR_Y} e^{\mu(g')}}
\]
as can be easily computed by evaluating the derivative of $F(\mu+\epsilon u)$ at $\epsilon=0$. Introducing the notation
\[
q_\mu(g) = \frac{e^{\mu(g)}}{\sum_{g\in \CR_Y} e^{\mu(g)}}
\]
and
\[
\langle \zeta \rangle_\mu = \sum_{g\in \CR_Y} \zeta(g) q_\mu(g),
\]
we have $dF_g(\mu)\zeta =  \zeta(g) - \langle \zeta \rangle_\mu $. 
Evaluating the derivative of $dF_g(\mu+\epsilon u')(\zeta)$ at $\epsilon=0$, one gets (the computation being left to the reader)
\begin{equation}
\label{eq:d2.Fg}
d^2F_g(\mu)(\zeta, \zeta') =   - \langle \zeta \zeta' \rangle_\mu + \langle \zeta \rangle_\mu \, \langle \zeta' \rangle_\mu.
\end{equation}
Note that $-d^2F_g(\mu)(\zeta, \zeta)$ is the variance of $\zeta$ for the probability mass function $q_\mu$ and is therefore non-negative (so that $F_\mu$ is concave). This immediately shows that $\ell$ is concave as a sum of concave functions.

Using the chain rule, we have, for $u:\CR_Y\to \mR^q$,
\[
d\ell(\beta)u = \sum_{k=1}^N dF_{y_k}(\beta^T\tx_k)\tx_k^Tu(\cdot) = \sum_{k=1}^N \tx_k^Tu(y_k) - \sum_{k=1}^N \langle \tx_k^T u(\cdot) \rangle_{\beta^T\tx_k}.
\]
Reordering the first sum in the right-hand side according to the values of $y_k$ gives
\[
\sum_{k=1}^N u(y_k)^T\tx_k = \sum_{g\in\CR_Y} N_g u(g)^T \tm_g.
\]
Noting that $q_{\beta^T\tx} = p_\beta(\cdot|x)$, we find
\[
d\ell(\beta)(u) =  \sum_{g\in\CR_Y} N_g \tm_g^T u(g) - \sum_{g\in\CR_Y} \tx_k^Tu(g) p_\beta(g|x_k),
\]
yielding \cref{eq:log.reg.d1}. Applying the chain rule again, we have
\begin{equation}
\label{eq:d2.ell.F}
d^2\ell(\beta)(u, u') = \sum_{k=1}^N d^2F_{y_k}(\beta^T\tx_k)(\tx_k^Tu(\cdot), \tx_k^Tu'(\cdot))
\end{equation}
with
\begin{align*}
d^2F_{y_k}(\beta^T\tx_k)(\tx_k^Tu(\cdot), \tx_k^Tu'(\cdot)) =& - \langle u(\cdot)^T \tx_k\tx_k^T u'(\cdot)  \rangle_{\beta^T\tx_k} + \langle \tx_k^Tu(\cdot) \rangle_{\beta^T\tx_k} \, \langle \tx_k^Tu'(\cdot) \rangle_{\beta^T\tx_k}\\
 =& - \sum_{g\in \CR_Y} u(g)^T \tx_k\tx_k^T u'(g) p_\beta(g|x_k) \\
& + \sum_{g,g'\in\CR_Y} u(g)^T \tx_k\tx_k^T u'(g') p_\beta(g|x_k) p_\beta(g'|x_k)
\end{align*}
from which \cref{eq:log.reg.d2} follows.
\end{proof}

\begin{remark}
From $F_g(\mu + \mu_0) = F_g(\mu)$ when $\mu_0$ is constant on $\CR_Y$, one deduces (taking the derivative at $\mu_0 = 0$) that $dF_g(\mu) \boldsymbol 1 = 0$ for all $\mu$, where $\boldsymbol 1$ denotes the constant function equal to 1 on $\CR_Y$. For $h\in \mR^{d+1}$, let $\mathbb c_h$ denote the constant function $\mathbb c_h(g) = h$, $g\in \CR_Y$. We have
\[
d\ell(\beta)\,  \mathbb c_h  =  \sum_{k=1}^N dF_{y_k}(\beta^T\tx_k)\tx_k^T\mathbb c_h =  \sum_{k=1}^N dF_{y_k}(\beta^T\tx_k)(\tx_k^Th \boldsymbol 1) = \sum_{k=1}^N (\tx_k^Th) dF_{y_k}(\beta^T\tx_k) \boldsymbol 1  = 0.
\]
Taking one extra derivative we see that 
\[
d\ell(\beta) (\mathbb c_h , u) = 0
\]
for all functions $u: \CR_Y \to \mR^q$.
\end{remark}

We now discuss whether there are other elements in the null space of the second derivative of $\ell$. We will use notation introduced in the proof of \cref{prop:log.reg}.  From \cref{eq:d2.Fg}, we have $d^2F_g(\mu)(\zeta, \zeta) = 0$ if and only if the variance of $\zeta$ for $q_\mu$ vanishes, which, since $q_\mu >0$, is equivalent to $\zeta$ being constant. So, the null space of $d^2F_g(\mu)$ is one-dimensional, and composed of scalar multiples of $\boldsymbol 1$. Using \cref{eq:d2.ell.F}, we see that $d^2\ell(u,u) = 0$ if and only if , for all $k=1, \ldots, N$, $(g\mapsto \tx_{k}^{T} u(g))$ is a constant function.
 
Assume that  this is true. Then, 
letting $\bar u = \frac1q \sum_{g\in\CR_Y}  u(g)$, one has, for all $g\in\CR_Y$ and $k=1, \ldots, N$,
\[
\tx_{k}^{T} u(g) = \tx_{k}^{T} \bar u
\]
so that $u(g) - \bar u$ is in the null space of the matrix $\CX$. This leads to the following proposition.
\begin{proposition}
\label{prop:log.reg.hess}
Assume that $\CX$ has rank $d+1$. Then the null space of $d^2\ell(\beta)$ is the set of all vectors $u =\mathbb c_h $ for $h\in\mR^{d+1}$. In particular, for any $c\in\mR^{d+1}$, the function $\ell$ restricted to the space
\[
M = \defset{ \be: \sum_{g\in \CR_Y} \rho_g \be(g) = c}
\]
is strictly concave as soon as the scalar coefficients $(\rho_g, g\in\CR_Y)$ are such that $\sum_{g\in\CR_Y} \rho_g \neq 0$.
\end{proposition}
\begin{proof}
From the discussion before the proposition, $u\in \mathrm{Null}(d^2\ell)$ implies that $\CX(u(g) - \bar u) =0$ for all $g$, and since we assume that $\CX$ has rand $d+1$, this requires that $u(g) = \bar u$ for all $g$, i.e., $u = \mathbb c_{\bar u}$. This proves the first point. 

If one restricts $\ell$ to $M$, then we must restrict $d^2\ell(\beta)$ to those  $u$'s such that $\sum_{g\in \CR_Y} \rho_g u(g) = 0$. But if $d^2\ell(\beta) (u,u) = 0$ for such an $u$, then $u = \mathbb c_{\bar u}$ and 
\[
\sum_{g\in \CR_Y} \rho_g u(g) = \Big(\sum_{g\in \CR_Y} \rho_g\Big) \bar u.
\]
Since we assume that $\sum_{g\in \CR_Y} \rho_g\neq 0$, this requires $\bar u=0$, and therefore $u=0$.

This shows that the second derivative of the restriction of $\ell$ to $M$ is negative definite, so this restriction is strictly concave.
\end{proof}

\subsection{Training algorithm}
Given that we have expressed the first and second derivatives of $\ell$ in closed form\footnote{Their computation is feasible unless $N$ is very large, and the matrix inversion in Newton's iteration also requires $d$ to be not too large.}, we can use  Newton-Raphson gradient ascent to maximize $\ell$ over the  affine space:
\[
M = \defset{ \be: \sum_{g\in \CR_Y} \rho_g \be(g) = c}
\]
with $\sum_{g\in\CR_Y} \rho_g \neq 0$. We assume in the following that the matrix $\CX$ has rank $d+1$ so that \cref{prop:log.reg.hess} applies. Since the constraint is affine, it is easy to express one of the parameters $\beta(g)$ as a function of the others and solve the strictly concave problem as a function of the remaining variables. It is not much harder, and arguably more elegant to solve the problem without breaking its symmetry with respect to the class indexes, as described below. 

Let
\[
M_0 = \defset{ \be: \sum_{g\in \CR_Y} \rho_g \be(g) = 0}.
\]
We still have the second order expansion
\[
\ell(\be+ u) = \ell(\be) +  d \ell(\be)u  + \frac12  d^2\ell(\be) (u,u) + o(|u|^2)
\]
and we consider the maximization of  the first three terms, simply restricting to vectors $u\in M_0$. To allow for matrix computation, we use our ordering $\CR_Y = (g_1, \ldots, g_q)$ and identify $a$ with the column vector
\[
\begin{pmatrix}
u(g_1) \\\vdots\\u(g_q)
\end{pmatrix}
\in \mR^{q(d+1)}
\]
Similarly, we let 
\[
\nabla \ell (\beta) = \begin{pmatrix}
\prt_{\beta(g_1)}\ell \\\vdots\\ \prt_{\beta(g_q)}\ell
\end{pmatrix}
\]
and let
$\nabla^2(\ell)(\beta)$ be the block matrix with $i,j$ block given by $\prt_{\beta(g_i)}\prt_{\beta(g_j)}\ell (\beta)$. We let $\hat \rho$ be the $(d+1)\times q(d+1)$ row block matrix 
\[
\begin{pmatrix}
\rho(g_1) \Id[d+1] & \cdots & \rho(g_q) \Id[d+1]
\end{pmatrix}
\]
so that $u\in M_0$ is just $\hat\rho u = 0$ in vector notation. Given this we have
\[
\ell(\be+ u) = \ell(\be) +  \nabla\ell(\beta)^T u  + \frac12 u^T  \nabla^2(\ell)(\beta) u + o(|u|^2).
\]

The maximum of $\ell(\be) +  u^T\nabla \ell(\be)  +  \frac12 u^T \nabla^2(\ell)(\be) u$ subject to $ \hat \rho u = 0$ 
is a stationary point of the Lagrangian
\[
L = \ell(\be) +  u^T\nabla \ell(\be)  +  \frac12 u^T \nabla^2(\ell)(\be) u + \la^T \hat \rho u 
\]
for some $\la\in \mR^{d+1}$ and is characterized by 
\[
\left\{
\begin{aligned}
\nabla^2(\ell)(\be) u + \nabla \ell(\be) + \hat \rho^T \la = 0\\
\hat \rho u = 0
\end{aligned}
\right.
\]
This shows that the Newton-Raphson iterations can be implemented as  
\begin{equation}
\be_{n+1} = \be_n - \ep_{n+1} u_{n+1}
\label{eq:log.reg.nr.1}
\end{equation}
with
\begin{equation}
\label{eq:log.reg.nr.2}
\begin{pmatrix} u_{n+1}\\ \la\end{pmatrix} = \begin{pmatrix}
\nabla^2(\ell)(\be_n) & \hat \rho^T\\ \hat \rho & 0
\end{pmatrix}^{-1} \begin{pmatrix}
\nabla \ell(\be_n) \\ 0
\end{pmatrix}.
\end{equation}

We summarize this discussion in the following algorithm.
\begin{algorithm}[Logistic regression with Newton's gradient ascent]
\label{alg:log.reg.nr}
\begin{enumerate}[label=(\arabic*)]
\item Input: (i) training data $(x_1, y_1, \ldots, x_N, y_N)$ with $x_i\in \mR^d$ and $y_i\in\CR_Y$; (ii) coefficients $\rho_g, g\in \CR_Y$ with non-zero sum and target value $c\in\mR$; (iii) algorithm step $\epsilon$ small enough.
\item Initialize the algorithm with $\beta_0$ such that $\sum_{g} \rho_g \beta_0(g) = c$.
\item At iteration $n$, compute $\nabla \ell(\beta_n)$ and $\nabla^2(\ell)(\beta_n)$ as provided by \cref{prop:log.reg}.
\item
Update $\beta_n$ using \cref{eq:log.reg.nr.1} and \cref{eq:log.reg.nr.2},  with $\epsilon_{n+1} = \epsilon$. Alternatively, optimize  $\epsilon_{n+1}$ using a line search.
\item Stop the procedure if the change in the parameter is below a small tolerance level. Otherwise, return to step 2.
\end{enumerate}
\end{algorithm}

%
%

\subsection{Penalized Logistic Regression}

Logistic regression can be combined with a penalty term, e.g., maximizing 
\begin{equation}
\label{eq:pen.log.reg.1}
\ell_2(\be) = \ell(\be) - \la \sum_{i=1}^d |b^{(i)}|^2
\end{equation}
or
\begin{equation}
\label{eq:pen.log.reg.2}
\ell_1(\be) = \ell(\be) - \la \sum_{i=1}^d |b^{(i)}|
\end{equation}
where $b^{(i)}$ is the $q$-dimensional vector formed with the $i$th coefficients of $b(g)$ for $g\in\CR_Y$. 
Similarly to penalized regression, one generally normalizes the $x$ variables to have unit standard deviation before applying the method.

\paragraph{Maximization with the $\ell_2$ norm}
The problem in \cref{eq:pen.log.reg.1} relates to ridge regression and can be solved using a Newton-Raphson method (\cref{alg:log.reg.nr}) with minor changes. More precisely, letting 
\[
\Delta = \begin{pmatrix}
0 & 0 \\
0 & \Id[d]
\end{pmatrix}
\] 
we have, considering $\beta$ as a $d+1$ by $q$ matrix,
\[
\ell_2(\be) = \ell(\be) - \la \trace(\beta^T \Delta \beta)
\]
and
\[
d\ell_2(\beta) u = d\ell(\beta) u - 2\lambda \trace(\beta^T \Delta u),
\]
\[
d^2\ell_2(\beta) (u, u') = d\ell(\beta) (u, u') - 2\lambda \trace(u^T \Delta u').
\]
In addition, when $\lambda >0$, the problem is over-parametrized only up to the addition of a constant to $(g\mapsto {a_0}(g))$, so that one only needs a single constraint $\sum_g \rho_g {a_0}(g) = c$ and the Lagrange coefficient in \cref{eq:log.reg.nr.2} is one dimensional.

\paragraph{Maximization with the $\ell_1$ norm}
 The maximization in \cref{eq:pen.log.reg.2} can be run using proximal gradient ascent (\cref{sec:proximal}). Let $C$ denote the affine subset of $\mR^{d+1}$ containing all $\beta = \begin{pmatrix}
 a_0\\ b
\end{pmatrix}
$, such that $\sum_g \rho_g {a_0}(g) = c$ and $\sigma_C$ the convex indicator function with $\sigma_C(\beta) = 0$ if $\beta\in C$ and $+\infty$ otherwise.

We want to maximize the objective function 
\[
\ell_1(\beta) = \ell(\beta) - \la \gamma(\beta) 
\]
 with 
 \[
 \gamma({a_0}, b) = \sum_{i=1}^d \sqrt{\sum_{g\in \CR_Y} b^{(i)}(g)^2}
 \sigma_C(\beta).
 \]
Here, $\ell$ is concave and $\gamma$ is convex and the proximal gradient iterations are
\begin{equation}
\label{eq:log.reg.prox}
\be_{n+1} = \prox_{\ep\la \gamma} (\be_n + \ep \nabla \ell(\beta_n)).
\end{equation}
We now compute the proximal operator of $\gamma$, and, since $\gamma$ is the sum of functions depending on $a_0, \pe b1, \ldots, \pe b d$,  it suffices to compute separately the proximal operator of each these functions.

Let $\pe u0, \ldots, \pe ud$ be functions from $\CR_Y$ to $\mR$.
Starting with $a_0$, we know that $\pe h0 = \prox_{\epsilon\lambda\sigma_C}(\pe v0)$ is the projection of $\pe u0$ on $C$, therefore characterized by $\pe h0\in C$ and $(\pe u0 - \pe h0) \perp C$, the latter implying that $\pe h0 = \pe u0 + t \rho$ for some $t\in \mR$ and the former allowing one to identify $t$ as $t = (c - (\pe u0)^T\rho)/|\rho|^2)$ so that
\[
\prox_{\epsilon\lambda\sigma_C}(\pe u0) = \pe u0 + \frac{c - (\pe v0)^T\rho}{|\rho|^2} \rho.
\]

Now consider $i\in \{1, \ldots, d\}$ and compute
\[
\argmin \sqrt{\sum_{g\in \CR_Y} \pe {h}i(g)^2} + \frac{1}{2\la \ep} \sum_{g\in\CR_Y} (\pe ui(g) - \pe{h}i(g))^2.
\]
The function $t \mapsto \sqrt t$ being differentiable everywhere except at $0$, we first search for a solution for which $\pe{h}i\neq 0$ does not vanish.
 If such a solution exists, it must satisfy, for all $g\in\CR_Y$
\[
\frac{\pe {h} i(g)}{\sqrt{\sum_{g'\in \CR_Y} \pe{h}i(g')^2}} + \frac{1}{\la \ep} (\pe{h}i( g) - \pe ui(g)) = 0
\]
Letting $|\pe{h}i| = \sqrt{\sum_{g\in \CR_Y} \pe{h}i(g)^2}$ we get 
\[
\pe{h}i( \cdot) (|\pe{h}i( \cdot)| + \la \ep) = \pe ui(\cdot) |\pe{h} i(\cdot)|
\]
Taking the norm on both sides and dividing by $|\pe hi( \cdot)|$ (which is assumed not to vanish) yields
\[
|\pe{h}i(\cdot)| + \la \ep = | \pe u i(\cdot)|,
\]
which has a positive solution only if
$| \pe u i(\cdot)| > \la\ep$, and gives in that case 
\[
\pe {h}i(\cdot) = \frac{|\pe ui(\cdot)| - \ep\la}{|\pe ui(\cdot)|} \pe ui(\cdot)
\]
If $| \pe ui(\cdot)| \leq \la\ep$, then we must take $\pe{h}i (\cdot) = 0$.
We have therefore obtained:
\[
\prox_{\ep\la g} (u) = h
\]
with 
\begin{subequations}
\begin{equation}
\label{eq:LR.lasso.1}
\pe{h}0( \cdot) = \pe u0 + \frac{c - (\pe v0)^T\rho}{|\rho|^2} \rho
\end{equation}
 and  
\begin{equation}
\label{eq:LR.lasso.2}
\pe{h}i(\cdot) = \max\Big(\frac{|\pe ui(\cdot)| - \ep\la}{|\pe ui(\cdot)|}, 0\Big)\,  \pe ui(\cdot)\,
\end{equation}
\end{subequations}
for $i\geq 1$.  We summarize this discussion in the next algorithm, which should be run with $\epsilon>0$ small enough.

\begin{algorithm}[Logistic lasso]
\label{alg:log.lasso}
\begin{enumerate}[label=(\arabic*)]
\item Input: (i) training data $(x_1, y_1, \ldots, x_N, y_N)$ with $x_k\in \mR^d$ and $y_k\in\CR_Y$; (ii) coefficients $\rho_g, g\in \CR_Y$ with non-zero sum and target value $c\in\mR$; (iii) algorithm step $\epsilon$; (iv) penalty coefficient $\lambda$.
\item Initialize the algorithm with $\beta_0 = ({a_{0,0}}, b_0)$.
\item At iteration $n$, compute $u = \be_n + \ep \nabla \ell(\beta_n)$, with $\beta_n = (a_{0,n}, b_n)$.
\item Let $a_{n+1,0}(\cdot) = \pe h0(\cdot)$ and for $i\geq 1$, 
$\pe{b_{n+1}}i(\cdot) = \pe hi(\cdot)$ where $\pe h0, \ldots, \pe hd$ are given by \cref{eq:LR.lasso.1,eq:LR.lasso.2}.
\item Stop the procedure if the change in the parameter is below a small tolerance level. Otherwise, return to step 2.
\end{enumerate}
\end{algorithm}



\subsection{Kernel logistic regression}
 Let $h: \CR_X  \to H$ be a feature function with values in a Hilbert space $H$ with $K(x,x') = \scp{h(x)}{h(x')}_H$. 
  The kernel version of logistic regression uses the model:
\[
\log p_{{a_0}, b}(g\mid x) = {a_0}(g) +  \scp{h(x)}{b(g)}_H \\
- \log\sum_{\tilde g\in \CR_Y
}\exp({a_0}(\tilde g) + \scp{h(x)}{b(\tilde g)}_H)
\]
with $b(g)\in H$ for $g\in \CR_Y$. 

 Using the usual kernel argument, one sees that, when maximizing the log-likelihood, there is no loss of generality is assuming that each $b(g)$ belongs to $V = \vspan(h(x_1), \ldots, h(x_N))$.
 Taking
\[
b(g) = \sum_{k=1}^N \al_k(g) h(x_k),
\]
we have
\begin{multline*}
\log p_\al(g\mid x) = {a_0}(g) +  \sum_{k=1}^N \al_k(g) K(x, x_k) 
- \log\left(\sum_{\tilde g\in \CR_Y
}\exp({a_0}(\tilde g) + \sum_{k=1}^N \al_k(\tilde g) K(x, x_k))\right).
\end{multline*}
To avoid overfitting, one must include a penalty term in the likelihood, and (in order to take advantage of the kernel), one can take this term proportional to $\sum_g \|b(g)\|_H^2$. The complete learning procedure then
requires to maximize the concave penalized likelihood$$
\ell(\al) = \sum_{k=1}^N \log p_\al(y_k\mid x_k) - \la \sum_{g\in\CR_Y}
\sum_{k,l=1}^N \al_k(g)\al_l(g) K(x_k,x_l).
$$
 The computation of the first and second derivatives of this function is similar to that for the original version, and we skip the details.

\section{Linear Discriminant analysis}
\label{sec:lda}

\subsection{Generative model in classification and LDA}

\paragraph{Generative model}
In  classification, the class variable $Y$ generally has a causal role upon which the variable $X$ is produced. Prediction can therefore be seen as an inverse problem where the cause is deduced from the result. In terms of generative modeling, one should therefore model the distribution of $Y$, followed by the the conditional distribution of $X$ given $Y$.

Taking $\CR_X = \mR^d$, denote by  $f_g$ the conditional p.d.f. of $X$ given $Y=g$ and let $\pi_g =
P(Y=g)$.
The Bayes estimator for the 0--1 loss maximizes the posterior probability
$$
\myP(Y=g\mid X=x) = \frac{\pi_g f_g(x)}{\sum_{g'\in \CR_Y} \pi_{g'} f_{g'}(x)}\,.
$$
Since the denominator does not depend on $g$  the Bayes estimator equivalently maximizes (taking logarithms)
\[
\log f_g(x) + \log \pi_g\,.
\]

One generally speaks of a linear classification method when the prediction is based on the maximization in $g$ of a function $U(g,x)$ where $U$ is affine in $x$. In this sense, logistic regression is linear, and kernel logistic regression is linear in feature space. For the generative approach, this occurs when one uses the following model, which provides the generative form of linear discriminant analysis (LDA). Assume that 
the distributions $f_g$ are all Gaussian  with
mean $m_g$ and {\em common variance} $S$, so that
\begin{equation}
\label{eq:lda.model}
f_g(x) = \frac{1}{\sqrt{(2\pi)^d\det \Sig}} e^{-\frac12 (x-m_g)^T S^{-1}
(x-m_g)}.
\end{equation}
In this case, the optimal predictor must maximize (in $g$)
\[
- \frac12 (x-m_g)^T S^{-1}
(x-m_g) + \log \pi_g\,.
\]
Introduce $m = \myE(X) = \sum_{g\in \CR_Y} \pi_g m_g$. Then the optimal classifier must maximize
\[
- \frac12 (x-m)^T S^{-1}
(x-m) + (x-m)^T S^{-1}
(m_g-m)  - \frac12 (m_g-m)^T S^{-1}
(m_g-m) + \log \pi_g.
\]
Since the first term does not depend on $g$, it is equivalent to maximize
\begin{equation}
\label{eq:lda.minimize}
(x-m)^T S^{-1}
(m_g-m)  - \frac12 (m_g-m)^T S^{-1}
(m_g-m) + \log \pi_g
\end{equation}
with respect to the class $g$, which provides an affine function of $x$. 

\paragraph{Training}
Training for LDA simply consists in estimating the class means and common variance in \cref{eq:lda.model} from data.
We introduce some notation for this purpose (this notation will be reused through the rest of this chapter). 

Recall that $N_g, g\in \CR_Y$ denotes the number of samples with class $g$ in the training set $T = (x_1, y_1, \ldots, x_N, y_N)$. We  let $c_g = N_g/N$ and $C$ be the diagonal matrix with diagonal coefficients $c_{g_1}, \ldots, c_{g_q}$. We  also let $\ze \in \mR^q$  denote the vector with the same coordinates. For $g\in \CR_Y$, $\mu_g$ denotes the class average 
\[
\mu_g = \frac1{N_g} \sum_{k=1}^n x_k \bfone_{y_k = g}
\]
and $\mu$ the global average
\[
\mu = \frac1N \sum_{k=1}^N x_k = \sum_{g\in \CR_Y} c_g\mu_g.
\]
Let $\Sig_{g}$ denote the sample covariance matrix in class $g$, defined by
\[
\Sig_g = \frac1{N_g} \sum_{k=1}^N (x_k-\mu_g)(x_k-\mu_g)^T \bfone_{y_k = g},
\]
and $\Sig_w$ the pooled class covariance (also called \alert{within-class} covariance) defined by
\[
\Sig_w = \frac{1}{N} \sum_{k=1}^N (x_k - \mu_{y_k})(x_k - \mu_{y_k})^T = \sum_{g\in \CR_Y} c_g \Sig_g.
\]
Let, in addition, $\Sig_b$ denotes the ``between-class'' covariance matrix, given by
\[
\Sig_b = \sum_{g\in\CR_Y} c_g (\mu_g- \mu)(\mu_g- \mu)^T
\]
The global covariance matrix, given by, 
\[
\Sig_{XX} = \frac1N \sum_{k=1}^N (x_k-\mu)(x_k-\mu)^T
\]
satisfies $\Sig_{XX} = \Sig_w + \Sig_b$. This identity is proved by noting that, for any $g\in\CR_Y$,
\[
\frac{1}{N_g} \sum_{k=1}^N (x_k- \mu)(x_k- \mu)^T \bfone_{y_k = g} = \Sig_g + (\mu_g- \mu)(\mu_g- \mu)^T.
\]
We will finally denote by $M$ the matrix
\[
M = \begin{pmatrix} (\mu_{g_1}- \mu)^T\\ \vdots\\ (\mu_{g_q}- \mu)^T\end{pmatrix}.
\]
Note that $\Sig_b = M^T C M$.

Given this notation, one can in particular take $\hat m_g = \mu_g$, $m = \mu$ and $\hat S = \Sig_w$ in \cref{eq:lda.minimize}. The class probabilities, $\pi_g$, can be deduced from the normalized frequencies of $y_1, \ldots, y_N$. However, in many applications, one prefers to simply fix $\pi_g=1/q$, in order to balance the importance of each class.

\begin{remark}
If one relaxes the assumption of common class variances, one needs  to use $\Sig_g$ in place of $\Sig_w$ for class $g$. The decision boundaries are not linear in this case, but provided by quadratic equations (and the resulting method if often called quadratic discriminant analysis, or QDA). QDA requires the estimation of $qd(d+3)/2$ coefficients, which may be overly ambitious when the sample size is not large compared to the dimension, in which case QDA is prone to overfitting. (Even LDA, which  involves  $qd + d(d+1)/2$ parameters, may be unrealistic in some cases.) We also note 
a variant of QDA that uses  class covariance
matrices given by
$$
\tilde \Sig_g = \al \Sig_w + (1-\al)\Sig_g.
$$
\end{remark}

\subsection{Dimension reduction}

One of the interests of LDA
is that it can be combined with a rank reduction procedure.
 LDA with $q$ classes can always be seen as a $(q-1)$-dimensional problem after suitable projection on a data-dependent affine space.
 Recall that the classification rule after training requires to maximize w.r.t. $g\in \CR_Y$ the function
\Eq{
(x-\mu)^T \Sig_w^{-1}
(\mu_g-\mu)  - \frac12 (\mu_g-\mu)^T \Sig_w^{-1}
(\mu_g-\mu) + \log \pi_g.
}
Define the ``spherized''  data\,\footnote{In this section only, the notation $\tilde  x$ does not refer to $(1,x^T)^T$.} by  $\tilde x_k = \Sig_w^{-1/2}(x_k-\mu)$, where $\Sig^{1/2}_w$ is the positive symmetric square root of $\Sig_w$. 
Also let  $\tilde\mu_g =  \Sig^{-1/2}_w (\mu_g - \mu)$. 
 
With this notation, the predictor chooses the class $g$ that maximizes 
\[
\tilde x^T \tilde \mu_g - \frac{1}{2} |\tilde \mu_g|^2 + \log \pi_g
\]
with $\tilde x = \Sig_w^{-1/2}(x-\bar\mu)$.

Now, let $V = \vspan\{\tilde \mu_g, g\in
\CR_Y\}$. 
 Since $\sum_g c_g\tmu_g = 0$, this space is at most $(q-1)$-dimensional.
 Let $P_V$ denote the orthogonal projection on $V$. 
 We have
$\tilde x^T z = (P_V\tilde x)^T z$ for any $z\in V$ and $\tilde x\in \mathbb R^d$.

 The classification rule can then be replaced by maximizing
$$
(P_V \tilde x)^T \tilde \mu_g - \frac{1}{2} |\tilde \mu_g|^2 + \log \pi_g
$$
with $\tilde x = \Sig_w^{-1/2}(x-\bar\mu)$.

Recall that  $M = \begin{pmatrix} (\mu_{g_1}-\mu)^T\\ \vdots\\ (\mu_{g_q}-\mu)^T\end{pmatrix}$ and let $\wtilde M = \begin{pmatrix} \tilde \mu^T_{g_1}\\ \vdots\\ \tilde \mu_{g_q}^T\end{pmatrix}$.
The dimension, denoted $r$, of $V$ is equal to the rank of $\wtilde M$.
Let $(\tilde e_1,
\ldots, \tilde e_{r})$ be an orthonormal basis of $V$.  One has
\[
P_V \tilde x = \sum_{j=1}^{r} (\tilde x^T\tilde e_j)\, \tilde e_j.
\]
 Given an input $x$, one must therefore compute the ``scores'' $\ga_j(x) =
\tilde x^T \tilde e_j$ and maximize
\[
\sum_{j=1}^r\ga_j(x)\ga_j(\mu_g) - \frac{1}{2} \sum_{j=1}^r \ga_j(\mu_g)^2 + \log \pi_g\,.
\]

The following proposition is key to the practical implementation of LDA with dimension reduction.
\begin{proposition}
\label{prop:lda.1}
An orthonormal basis of $V = \vspan(\tmu_g, g\in CG)$ is provided by the the first $r$ eigenvectors of $\wtilde M^T C \wtilde M$ associated with eigenvalues $ \la_1 \geq \cdots\geq \la_r >0$ (all other eigenvalues being zero).
\end{proposition}
\begin{proof}
 Indeed, if $\tilde  x$ is perpendicular to $V$, we have 
\[
\tilde M^T C \tilde M \tilde x = \sum_{g\in\CR_Y} c_g (\tilde\mu_g^T\tilde x) \tilde \mu_g = 0
\]
so that $V^\perp \sub \mathrm{Null}(\tilde M^T C \tilde M)$, and both spaces coincide because they have the same dimension ($d-r$).
 This shows that $V = \mathrm{Null}(\wtilde M^T C \wtilde M)^\perp = \mathrm{Range}(\tilde M^T C \wtilde M)$. Since $\wtilde M^T C \wtilde M$ is symmetric, $\mathrm{Null}(\wtilde M^T C \wtilde M)^\perp$ is generated by eigenvectors with non-zero eigenvalues.
 \end{proof}
 \bigskip

\begin{figure}
\centering
\includegraphics[width=0.48\textwidth]{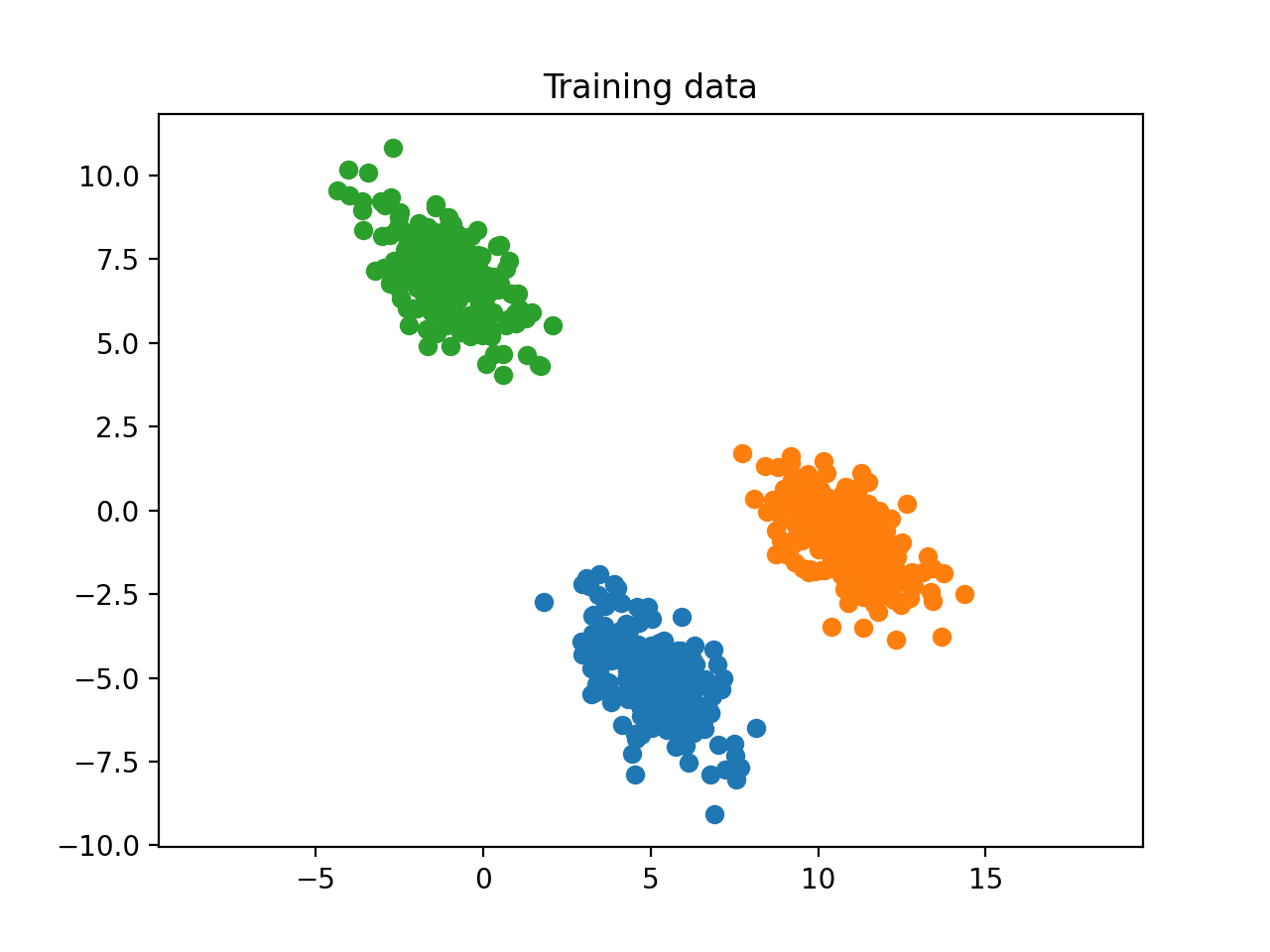}
\includegraphics[width=0.48\textwidth]{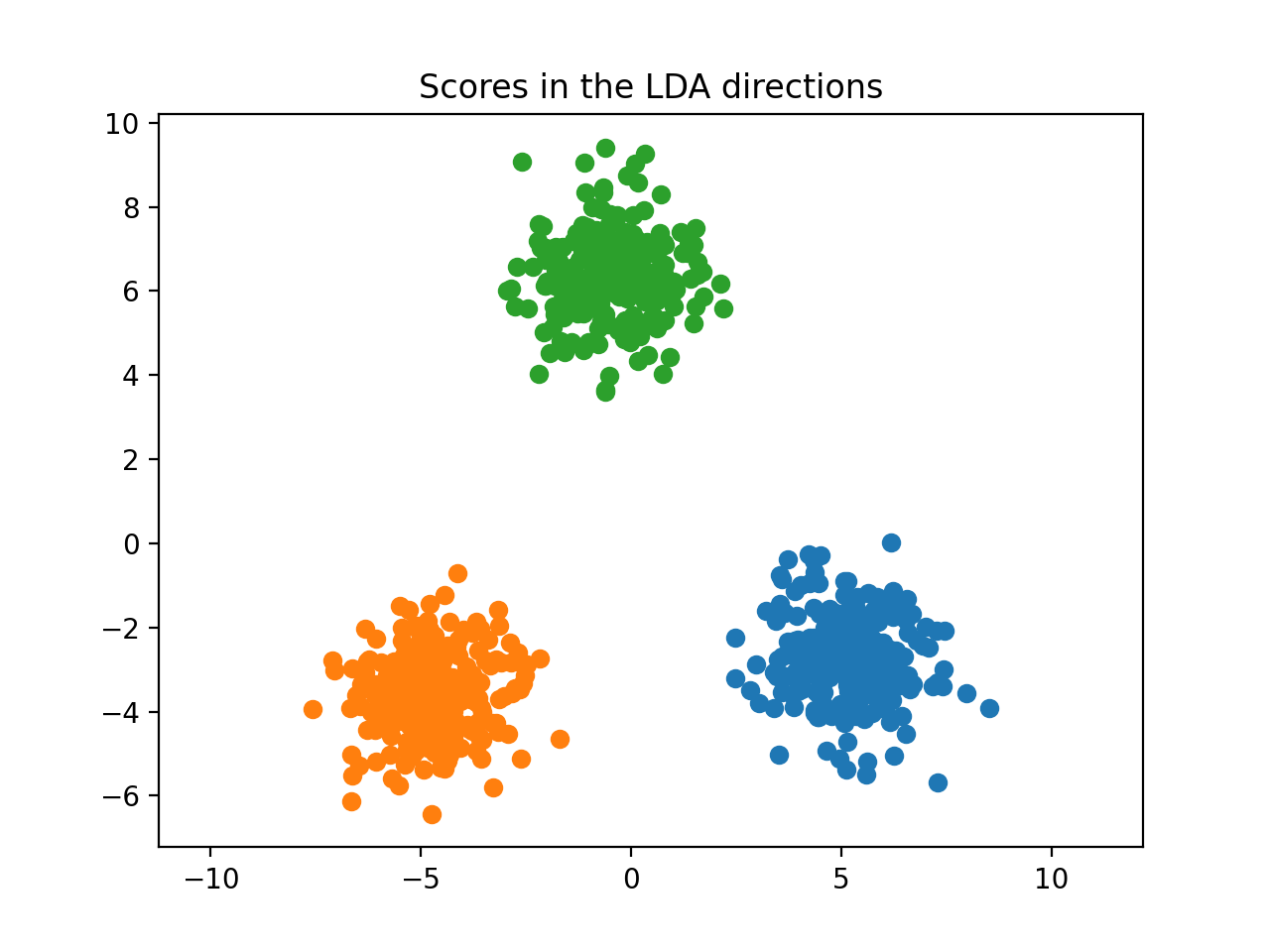}
\caption{\label{fig:lda} Left: Original (training) data with three classes. Right: LDA scores, where the $x$ axis provides $\ga_1$ and the $y$ axis $\ga_2$.}
\end{figure}

 Returning to the original variables, we have 
$\wtilde M = M\Sig_w^{-1/2} $ and $M^TCM = \Sig_b$, the between class covariance matrix. This implies that $\wtilde M^T C\wtilde M = \Sig_w^{-1/2} \Sig_b \Sig_w^{-1/2}$ and each eigenvector $\tilde e_j$ therefore satisfies
\[
\Sig_b \Sig_{w}^{-1/2} \tilde e_j = \la_j \Sig_w^{1/2} \tilde e_j = \la_j \Sig_w (\Sig_w^{-1/2} \tilde e_j).
\]
Therefore, letting $e_j = \Sig_w^{-1/2} \tilde e_j$, $(e_1, \ldots, e_r)$ are the solutions of the generalized eigenvalue problem $\Sig_b e = \la \Sig_w e$ that are associated with non-zero eigenvalues (they are however normalized so that $e_j^T \Sig_w e_j = 1$).  Moreover, the scores are given by
\[
\ga_j(x) = \tilde x^T \tilde e_j = (x - \mu)^T \Sig_w^{-1/2} \tilde e_j = (x-\mu)^T e_j
\]
and can therefore be computed directly from the original data and the vectors $e_1, \ldots, e_r$. An example of training data and its representation in the LDA space (associated with the scores) in provided in \cref{fig:lda}.

We can now describe the LDA learning algorithm with dimension reduction.
\begin{algorithm}[LDA with dimension reduction]
\begin{enumerate}[label = \arabic*.,wide=0.5cm]
\item Compute $\mu_g, g\in \CR_Y$, $\Sig_w$ and $\Sig_b$ from training data.
\item Estimate (if needed) $\pi_g, g\in \CR_Y$
\item Solve the generalized eigenvalue problem $\Sig_b e = \la\Sig_w e$. Let $e_1, \ldots, e_r$ be the eigenvectors associated with non-zero eigenvalues, normalized so that  $e_j^T \Sig_w e_j = 1$. 
\item Choose a reduced dimension $r_0 \leq r$.
\item Precompute mean scores $\ga_j(\mu_g) = (\mu_g- \mu)^T e_j$, $g\in \CR_Y, j=1, \ldots, r_0$.
\item To classify a new example $x$, compute $\ga_j(x) = (x-\mu)^T e_j$ and choose the class that maximizes 
\[
\sum_{j=1}^{r_0}\ga_j(x)\ga_j(\mu_g) - \frac{1}{2} \sum_{j=1}^{r_0} \ga_j(\mu_g)^2 + \log \pi_g\,.
\]
\end{enumerate}
\end{algorithm}


\subsection{Fisher's LDA}
This characterization  leads to the discriminative interpretation of LDA, also called Fisher's LDA. Indeed, the generalized eigenvalue problem $\Sig_b e = \la \Sig_w e$ is directly related to the maximization of the ratio
$e^T\Sig_be$ subject to $e^T\Sig_w e= 1$, which provides directions that have a large between-class variance for within class variance equal to 1. More precisely, $e_1$ is the direction that achieves the maximum; $e_2$ is the second best direction, constrained to being perpendicular to $e_1$, and so on until $e_r$ which is the optimal constrained to be perpendicular to $(e_1, \ldots, e_{r-1})$. We are therefore looking for directions that have the largest ratio of   between-class variance to  within-class variance.

\subsection{Kernel LDA}

\paragraph{Mean and covariance in feature space}
We assume the usual construction where $h: \CR_X \to H$ is a feature function, $H$ a {Hilbert} space with kernel $K(x,x') = \scp{h(x)}{h(x')}_H$. (The assumption that $H$ is a complete space is here required for the expectations below to be meaningful.)

We now discuss the kernel version of LDA by plugging the feature space representation directly in the classification rule. So, consider $h: \CR \to H$. Let $X:\Om \to \CR$ be a random variable such that $\myE(\|h(X)\|_H^2) < \infty$. Then, its mean feature $m = \myE(h(X))$ is well defined as an element of $H$ , and so are the class averages, $m_g = \myE(h(X)\mid Y=g)$. 

In this possibly infinite-dimensional setting, 
the covariance ``matrix'' is  defined as a linear operator $S: H\to H$ such that, for all $\xi, \eta\in H$:
\begin{equation}
\label{eq:klda.0}
\scp{\xi}{S\eta}_H = \myE\left(\scp{h(X)-m}{\xi}_H\scp{h(X)-m}{\eta}_H\right)\,,
\end{equation}
which is equivalent to defining 
\[
S\eta = \myE(\scp{h(X)-m}{\eta}_H (h(X) - m))
\] 
for $\eta\in H$.  This definition generalizes the identity for a random variable $U: \Om\to \mR^d$:
\[
S_U w = \myE((U-E(U))(U-E(U))^T) w = \myE(((U-E(U))^Tw)\,(U-E(U))) 
\]
One can similarly define the covariance matrix in class $g$, $S_g$, by conditioning the right-hand side in \cref{eq:klda.0} by $Y=g$ and replacing $m$ by $m_g$. 
\bigskip

\subsubsection*{LDA in feature space}
Following the LDA model, we assume that the operators $S_g$ are all equal to a fixed operator, the within-class  covariance operator denoted $S$.

Assuming that $S$ is invertible, one can generalize the LDA classification rule to data represented in feature space   by classifying a new input $x$ in class $g$ when 
\begin{equation}
\label{eq:klda.1}
\scp{h(x)- m}{ S^{-1} (m_g - m)}_H - \frac12 \scp{m_g - m}{S^{-1} (m_g- m)}_H +  \log\pi_g
\end{equation}
is maximal over all classes. Notice that this is a transcription of the finite-dimensional Bayes rule, but cannot be derived from a generative model, because the assumption that $h(X)$ is Gaussian is not valid in general. (It would require that $h$ takes values in a $d$-dimensional linear space, which would eliminate all interesting kernel representations.)

Let, as before, $T = (x_1, y_1, \ldots, x_N, y_N)$ be the training set,  $N_g$ denote the number of examples in class $g$ and $c_g = N_g/N$.
When $h$ is known (which, we recall, is not a practical assumption, but we will fix this later), one can estimate the class averages from training data by
\[
\mu_g = \frac1{N_g}\sum_{k=1}^N h(x_k) \bfone_{y_k=g}
\]
and the within-class covariance operator by
\[
\scp{\xi}{\Sig_w\eta}_H = \frac1N \sum_{k=1}^N \scp{h(x_k) - \mu_{y_k}}{\xi}_H \scp{h(x_k) - \mu_{y_k}}{\eta}_H.
\]
Unfortunately, the  resulting variance estimator cannot be directly used in \cref{eq:klda.1}, because it is not invertible if $\mathrm{dim}(H) > N$. Indeed, one has $\Sig_w\eta = 0$ as soon as $\eta$ is perpendicular to $V \defeq \vspan(h(x_1), \ldots, h(x_N))$. 

One way to address the degeneracy of the estimated covariance operator is to add to $\Sig_w$ a small multiple of the identity, say $\rho\Id_H$,\footnote{The operator $A + \rho\Id_H$ is invertible as soon as $A$ is symmetric positive semi-definite.} and let the classification rule maximize in $g$:
\begin{multline}
\label{eq:klda.3}
\scp{h(x)-\mu}{ (\Sig_w + \rho\Id_H)^{-1} (\mu_g -  \mu)}_H 
- \frac12 \scp{\mu_g -  \mu}{(\Sig_w + \rho\Id_H)^{-1} (\mu_g-  \mu)}_H + \log\pi_g\,.
\end{multline}
where $\mu$ is the average of $h(x_1), \ldots, h(x_N)$. 
Taking this option, 
we still need to make this expression computable and remove the dependency in the feature function $h$.

\paragraph{Reduction}
We have $\mu_g\in V$ for all $g\in\CR_Y$ and, since 
\[
\Sig_w\eta = \frac1N \sum_{k=1}^N  \scp{h(x_k) - \mu_{y_k}}{\eta}_H (h(x_k) - \mu_{y_k}),
\]
this operator maps $H$ to $V$, which implies that $\Sig_w + \rho \Id_H$ maps $V$ into itself. 
Moreover,  this mapping is onto: If $v\in V$ and $u = (\Sig_w +\rho \Id_H)^{-1}v$, then, $u\in V$.
Indeed, for any $z\perp V$, we have $\scp{z}{\Sig_w u + \rho u}_H = \scp{z}{v}_H$. We have  $\scp{z}{\Sig_w u}_H = 0$ (because $\Sig_w$ maps $H$ to $V$) and $\scp{z}{v}_H = 0$  (because $v\in V$), so that we can conclude that  $\scp{z}{u}_H = 0$. 
 Since this is true for all $z\perp V$, this requires that $u\in V$.\footnote{One has $(V^\perp)^\perp = V$ for finite-dimensional---or more generally closed---subspaces of $H$}

We now express the classification rule in \cref{eq:klda.3} as a function of the kernel associated with the feature-space representation.
Denote, for any vector $u \in \mR^N$, 
\[
\xi(u) = \sum_{k=1}^N \pe u k h(x_k),
\]
 therefore defining a mapping $\xi$ from $\mR^N$ onto $V$. Letting as usual $\CK = \CK(x_1, \ldots, x_N)$ be the matrix formed by pairwise evaluations of $K$ on training inputs, we have the identity
\[
\scp{\xi(u)}{\xi(u')}_H = u^T \CK u'.\
\]
for all $u,u'\in\mR^N$. For simplicity, we will assume in the rest of the discussion that $\CK$ is invertible.

We have $\mu_g = \xi(\dsone_g/N_g)$, where $\dsone_g\in \mR^N$ is the vector with $k$th coordinate equal to 1 if $y_k = g$ and 0 otherwise. Also $\mu = \xi(\dsone/N)$ (recall that $\dsone$ is the vector with all coordinates equal to 1).

For $u\in \mR^N$, we want to characterize $v\in \mR^N$ such that $\Sig_w \xi(u) = \xi(v)$. Let $\de_k$ denote the vector with 1 at the $k$th entry and 0 otherwise. We have
\begin{align*}
\Sig_w \xi(u) &= \frac1N \sum_{k=1}^N  \scp{\xi(u)}{h(x_k) - \mu_{y_k}}_H (h(x_k) - \mu_{y_k})\\
&= \frac1N \sum_{k=1}^N  \scp{\xi(u)}{\xi(\de_k - \dsone_{y_k}/N_{y_k})}_H \xi(\de_k - \dsone_{y_k}/N_{y_k})\\
&= \frac1N \sum_{k=1}^N  ((\de_k - \dsone_{y_k}/N_{y_k})^T \CK u)\, \xi(\de_k - \dsone_{y_k}/N_{y_k})\\
&= \xi\left( \frac1N \sum_{k=1}^N  ((\de_k - \dsone_{y_k}/N_{y_k})^T \CK u)\, (\de_k - \dsone_{y_k}/N_{y_k})\right)
\end{align*}
so that $\Sig_w \xi(u) = \xi(P\CK u)$ with
\[
P = \frac1N \sum_{k=1}^N  (\de_k - \dsone_{y_k}/N_{y_k})(\de_k - \dsone_{y_k}/N_{y_k})^T
\]
Note that one has
\begin{align*}
\bullet\quad & \sum_{k=1}^N  \de_k \de_k^T = \Id[N],\\ 
\bullet\quad&
\sum_{k=1}^N \left(\frac{\dsone_{y_k}}{N_{y_k}}\right) \de_k^T = \sum_{g\in\CR_Y} \frac{\dsone_{g}}{N_{g}} \sum_{k: y_k=g} \de_k^T = \sum_{g\in\CR_Y} \frac{\dsone_{g}\dsone_{g}^T}{N_{g}} = \sum_{k=1}^N \de_k \left(\frac{\dsone_{y_k}}{N_{y_k}}\right)^T,\\
\bullet\quad&
\sum_{k=1}^N \left(\frac{\dsone_{y_k}}{N_{y_k}}\right) \left(\frac{\dsone_{y_k}}{N_{y_k}}\right)^T = \sum_{g\in\CR_Y} \frac{\dsone_{g}\dsone_{g}^T}{N_{g}}.
\end{align*}
This shows that $P$ can be expressed as
\[
P =\frac1N \left( \Id[N] - \sum_{g\in\CR_Y} \dsone_{g}\dsone_{g}^T/N_g\right).
\] 
We have therefore proved that 
\[
\begin{aligned}
\bullet\quad & (\Sig_w + \rho\Id_H) \xi(u) = \xi\left((P\CK + \rho\Id[N]) u\right)\\
\bullet\quad & (\Sig_w + \rho\Id_H)^{-1} \xi(\tilde u) = \xi\left((P\CK + \rho\Id[N])^{-1} \tilde u \right).
\end{aligned}
\]
 Recall that the feature-space LDA classification rule maximizes
\[
\scp{h(x)-\mu}{ (\Sig_w + \rho\Id_H)^{-1} (\mu_g -  \mu)}_H 
- \frac12 \scp{\mu_g -  \mu}{(\Sig_w + \rho\Id_H)^{-1} (\mu_g-  \mu)}_H + \log\pi_g\,.
\]
 All terms belong to $V$, except $h(x)$, but this term can be replaced by its orthogonal projection on $V$ without changing the result. This projection can be made explicit in terms of the representation $\xi$ as follows. 
For $x\in \CR$, let $\xi(\psi(x))$ denote the orthogonal projection of $h(x)$ on $V$ (this defines the function $\psi$). If $v(x)$ denotes the vector with coordinates $K(x, x_k)$, $k=1, \ldots, N$, then $\psi(x) = \CK^{-1} v(x)$, as can be obtained by identifying the inner products $\scp{h(x)}{h(x_k)}_H$ and $\scp{\xi(\psi(x))}{h(x_k)}_H$.

We are now ready to rewrite the kernel LDA classification rule in terms of quantities that only involve $K$. We have  
\[
\begin{aligned}
&\scp{h(x)-\mu}{ (\Sig_w + \rho\Id_H)^{-1} (\mu_g - \mu)}_H \\
 & = \scp{\xi(\psi(x) - \dsone/N)}{\xi((P\CK + \rho\Id[N])^{-1} (\dsone_g/N_g - \dsone/N))}_H\\
&= (\psi(x) - \dsone/N)^T\CK(P\CK + \rho\Id[N])^{-1} (\dsone_g/N_g - \dsone/N)
\end{aligned}
\]
 Given this, the classification rule must maximize
\begin{multline}
(\psi(x) - \dsone/N)^T\CK (P\CK + \rho\Id[N])^{-1} (\dsone_g/N_g - \dsone/N)\\
 - \frac12 (\dsone_g/N_g - \dsone/N)^T \CK (P\CK + \rho\Id[N])^{-1} (\dsone_g/N_g - \dsone/N) + \log\pi_g.
\label{eq:klda.V}
\end{multline}

\paragraph{Dimension reduction}
Note that $\CK (P\CK + \rho\Id[N])^{-1} = \CK (\CK P\CK + \rho\CK)^{-1} \CK$ is a symmetric matrix. So, the expression in \cref{eq:klda.V} can be written as
\[
(v(x) - \bar\eta )^T R^{-1} (\eta_g - \bar\eta)
 - \frac12 (\eta_g - \bar\eta)^T R^{-1} (\eta_g - \bar\eta) + \log\pi_g.
\]
with $R = \CK P\CK + \rho\CK$, $\eta_g = \CK \dsone_g/N_g$ and $\bar\eta = \CK \dsone /N$. Clearly, if $v_1, \ldots, v_N$ are the column vectors $\CK$, we have
\[
\eta_g = \frac1{N_g} \sum_{k=1}^N v_k \bfone_{y_k=g},
\quad
\bar\eta = \frac1N \sum_{k=1}^N v_k\,.
\]

We therefore retrieve an expression similar to  finite-dimensional LDA, provided that one replaces $x$ by $v(x)$, $x_k$ by $v_k$ and $\Sig_w$ by  $R$. Letting 
\[
Q= \frac1N \sum_{g\in \CR_Y} N_g (\eta_g-\bar\eta)(\eta_g-\bar\eta)^T
\]
be the between-class covariance matrix, 
the discriminant directions are therefore  solutions of the generalized eigenvalue problem
\[
Q f_j = \la_j R f_j
\]
with $f_j^TRf_j = 1$ with $R=(\CK P\CK + \rho \CK)$. Note that 
\[
\CK P \CK = \frac1N \sum_{k=1}^N (v_k - \bar\eta_{y_k})(v_k - \bar\eta_{y_k})^T
\]
is the within-class covariance matrix for the training data $(v_1, y_1, \ldots, v_N, y_N)$.

The following summarizes the kernel LDA classification algorithm.
\begin{algorithm}[Kernel LDA]
\label{alg:k.lda}
\begin{enumerate}[label=(\arabic*)]
\item Select a positive kernel $K$ and a coefficient $\rho >0$.
\item Given $T = (x_1, y_1, \ldots, x_N, y_N)$ and a kernel $K$, compute the kernel matrix $\CK = \CK(x_1, \ldots, x_N)$ and the matrix $R = \CK P\CK + \rho \CK$. Let $v_1, \ldots, v_N$ be the column vectors of $\CK$.
\item Compute, for $g\in \CR_Y$, 
\[
\eta_g = \frac1{N_g} \sum_{k=1}^N v_k \bfone_{y_k=g}\, ,
\quad
\bar\eta = \frac1N \sum_{k=1}^N v_k
\]
and let $Q = \frac1N \sum_{g\in \CR_Y} N_g (\eta_g-\bar\eta)(\eta_g-\bar\eta)^T$.
\item Fix $r_0 \leq q-1$ and compute the eigenvectors $f_1, \ldots, f_{r_0}$ associated with the $r_0$ largest eigenvalues for the generalized eigenvalue problem $Qf = \la Rf$, normalized such that $f_j^T R f_j = 1$.
\item Compute the scores $\ga_{jg} = (\eta_g-\bar\eta)^Tf_j$.
\item Given a new observation $x$, let $v(x)$ be the vector with coordinates $K(x,x_k)$, $k=1, \ldots, N$. Compute the scores $\ga_j(x) = (v(x) - \bar\eta)^T f_j$, $j=1, \ldots, r_0$. Classify $x$ in the class $g$ maximizing 
\begin{equation}
\label{eq:klda.final}
\sum_{i=1}^r \ga_i(x) \ga_{ig} - \frac12 \sum_{i=1}^r \ga_{ig}^2  + \log\pi_g.
\end{equation}
\end{enumerate}
\end{algorithm}

\section{Optimal Scoring}
\label{sec:opt.sc}

It is possible to apply linear regression (\cref{chap:lin.reg}) to solve a classification problem by mapping  the set $\CR_Y$ to a collection of $r$-dimensional row vectors, or ``scores.''  These scores (which have a different meaning from the LDA scores) will be represented by a function $\th : \CR_Y \mapsto \mR^r$. As an example, one can  take $r=q$ and 
\[
\th(g_1) = \begin{pmatrix} 1\\ 0\\0\\ \vdots\\0\end{pmatrix},\ 
\th(g_2) = \begin{pmatrix} 0\\ 1\\0\\ \vdots\\0\end{pmatrix},\,\ldots\,, \ 
\th(g_q) = \begin{pmatrix} 0\\ 0\\ \vdots\\0\\1\end{pmatrix}.
\]
Given a training set $T = (x_1, y_1, \ldots, x_N, y_N)$ and a score function $\th$, a linear
model can then be estimated from  data by minimizing
$$
\sum_{k=1}^N |\th_{y_k} - {a_0} - b^T  x_k|^2
$$ 
where $b$ is a $d \ti
q$ matrix and ${a_0} \in \mR^q$. Letting as before $\be$ be the matrix with $a_0^T$ added as first row to $b$ and $\CX$ the matrix with first row containing only ones and subsequent rows given by $x_1^T, \ldots, x_N^T$, one gets the least square estimator
$\hat \beta = (\CX^T\CX)^{-1}\CX^T\CY$, where $\CY$ is the $N\times q$
matrix of stacked $\th^T_{y_k}$ row vectors. 

Given an input vector $x$, the row vector $\tilde x^T\beta$ will generally not coincide with one of the score vectors.  Assignment to a class can   then be
made by minimizing $|{a_0} + b^T x - \th_{g}|$ over all
$g$ in $\CR_Y$.\bigskip

Since the scores $\th$ are free to choose, one may also try to optimize them, resulting in the \alert{optimal scoring} algorithm. To describe it, we will need the notation already introduced for LDA, plus the following.  We will write, for short,  $\th_j = \th(g_j)$ and introduce the $q\times r$ matrix $\Th = \begin{pmatrix} \th_1^T\\ \vdots\\ \th_q^T\end{pmatrix}$.
We also denote by  $\rho_1, \ldots, \rho_r$ the column vectors of $\Th$, so that $\Th = [\rho_1, \ldots, \rho_r]$. 
Let $u_{g_i}$, for $i=1, \ldots, q$, denote the $q$-dimensional vector with $i$th coordinate equal to 1 and all others equal to 0.
As before, $N_g$ denote the class sizes,  $c_g = N_g/N$, $C$ is the diagonal matrix with coefficients $c_{g_1}, \ldots, c_{g_q}$ and $\ze = \begin{pmatrix} c_{g_1}\\ \vdots \\c_{g_q}\end{pmatrix}$.

The goal of optimal scoring is to minimize, now with respect to $\th$, ${a_0}$ and $b$, the function
\[
F(\th, {a_0}, b) = \sum_{k=1}^N |\th(y_k) - {a_0} - b^T  x_k|^2\,.
\]
Some normalizing conditions are clearly needed, because this problem is under-constrained. (In the  form above, the optimal choice is to take all free parameters equal to 0.) We now discuss the various indeterminacies and redundancies in the model, 

\begin{enumerate}[label = (\alph*), wide=0cm]
\item If $R$ is an $r\times r$ orthogonal matrix, then $F(R\th, R{a_0}, bR^T) = F(\th, {a_0}, b)$, yielding an infinity of possible equivalent solutions (that all lead to the same classification rule).
This implies that there is no loss of generality in assuming that $\Th^T C \Th$ is diagonal (introducing $C$ here will turn out to be convenient). Indeed, given any $(\th, {a_0}, b)$, one can just take $R$ such that $R\Th^T C \Th R^T$ is diagonal and replace $\Th$ by $R\Th$, ${a_0}$ by $R{a_0}$ and $b$ by $bR^T$ to get an equivalent solution satisfying the constraint.
\item Let $D$ be an $r$ by $r$ diagonal matrix with positive entries. Replace $\th$, ${a_0}$ and $b$ respectively by $D\th$, $D{a_0}$ and $bD$. The resulting objective function is 
\[
\begin{aligned}
F(D\th, D{a_0}, bD^T) &= \sum_{k=1}^N |D\th(y_k) - D{a_0} - Db^T  x_k|^2 \\
&= \sum_{j=1}^r \sum_{k=1}^N d_{jj}^2 \Big(\th(y_k, j) - {a_0}(j) - \sum_{i=1}^d b(i,j)  x_k(i)\Big)^2
\end{aligned}
\]
If the coefficient $d_{jj}$ is free to chose, then the objective function can always be reduced by letting $d_{jj} \to 0$, which removes one of the dimensions in $\th$. In order to avoid this, one needs to  fix the diagonal values of  $\Th^T C\Th$, and, by symmetry, it is natural to require  $\Th^T C\Th = \Id[r]$. 
\item Given any $\delta\in \mR^r$, one has $F(\th, {a_0}, b) = F(\th-\de, {a_0}+\de, b)$, with identical classification rule.
One can therefore without loss of generality introduce $r$ linear constraints, and a convenient choice is 
\[
\Th^T\ze = \sum_{g\in \CR_Y} c_g\th_g= 0.
\]
\end{enumerate}
\medskip

Given this reduction, we can now describe the optimal scoring problem as the minimization of 
\[
\sum_{k=1}^N |\th_{y_k} - {a_0} - b^T x_k|^2
\]
subject to $\Th^TC\Th = \Id[r]$ and $\Th^T\ze = 0$. 

The optimal ${a_0}$ is given by
\[
\hat{a_0} = \frac1N \sum_{k=1}^N \th_{y_k} - b^T\mu = -b^T\mu,
\]
so that the problem is reduced to minimizing 
\[
\sum_{k=1}^N |\th_{y_k} - b^T (x_k-\mu)|^2
\]
subject to the same constraints. Using the facts that $\th_{y_k} = \Th^T u_{y_k}$, that
\[
\sum_{k=1}^N u_{y_k}u_{y_k}^T = \sum_{g\in \CR_Y} N_g u_{g}u_{g}^T = NC
\]
and that
\begin{align*}
\sum_{k=1}^N u_{y_k}(x_k-\mu)^T &= \sum_{g\in \CR_Y}^N u_{g} \sum_{k: y_k = g} (x_k-\mu)^T
= \sum_{g\in \CR_Y}^N u_{g} N_g (\mu_g - \mu)^T = NCM,
\end{align*}
one can write 
\[
\begin{aligned}
\sum_{k=1}^N |\th_{y_k} - b^T (x_k-\mu)|^2 &= \sum_{k=1}^N |\Th^T u_{y_k} - b^T (x_k-\mu)|^2 \\
&= \sum_{k=1}^N u_{y_k}^T \Th \Th^T u_{y_k} - 2 \sum_{k=1}^N (x_k-\mu)^T b\Th^Tu_{y_k} + \sum_{k=1}^N (x_k-\mu)^T bb^T (x_k-\mu)\\
&= \sum_{k=1}^N \trace(\Th^T u_{y_k}u_{y_k}^T \Th) - 2 \sum_{k=1}^N \trace(\Th^T u_{y_k} (x_k-\mu)^T b) \\
& \qquad + \sum_{k=1}^N \trace( b^T (x_k-\mu)(x_k-\mu)^T b)\\
&= N\trace(\Th^TC\Th) - 2N\trace(\Th^TCMb) + N\trace(b^T\Sig_{XX}b)\,.
\end{aligned}
\]

Note that, since $\Th^T C \Th= \Id[r]$, then $\trace(\Th^TC\Th)=r$. 
We therefore obtain a concise form of the optimal scoring problem: minimize
\[
- 2\trace(\Th^TCMb) + \trace(b^T\Sig_{XX}b).
\]
subject to $\Th^T C \Th= \Id[r]$ and $\Th^T\ze = 0$.

Given $\Th$,  the optimal $b$ is  $\Sig_{XX}^{-1} M^TC\Th$, and replacing it in the objective function, one finds that $\Th$ must minimize
\[
- 2\trace(\Th^TCM\Sig_{XX}^{-1} M^TC\Th) + \trace(\Th^TCM\Sig_{XX}^{-1} M^TC\Th)
\]
i.e., maximize
\[
\trace(\Th^T CM\Sig_{XX}^{-1} M^T C \Th)
\]
subject to $\Th^T C \Th= \Id[r]$ and $\Th^T\ze = 0$.  We now recall the following linear algebra result (see \cref{chap:linalg}). 
\begin{proposition}
\label{prop:gen.eig}
Let $A$ and $B$ be respectively positive definite and non-negative semi-definite symmetric $q$ by $q$ matrices. Then, the maximum, over all $q$ by $r$ matrices $S$ such that $\trace(S^TAS) = \Id[r]$, of $\trace(S^TBS)$   is attained at $S = [\sig_1, \ldots, \sig_r]$, where the columns vectors $\sig_1, \ldots, \sig_r$ are the solutions of the generalized eigenvalue problem
\[
B\sig = \la A\sig
\]
associated with the largest eigenvalues, normalized so that $\sig_i^T A \sig_i = 1$ for $i=1, \ldots, r$..
\end{proposition}

Given this proposition, let $\rho_1, \ldots, \rho_r$ be the $r$ first  eigenvectors for the problem
\begin{equation}
\label{eq:ge.os}
CM\Sig_{XX}^{-1} M^T C\rho = \lambda C\rho.
\end{equation}
Assume that $r$ is small enough so that the associated eigenvalues are not zero.
Let  $\Th = [\rho_1, \ldots, \rho_r]$. We now prove that $\Th$ is indeed a solution of the optimal scoring problem, and the only point to show to complete the statement is that this $\Th$ satisfies the constraints  $\Th^T\ze = 0$.
 But we have
\[
M^TC\dsone_q = \sum_{g} c_g (\mu_g - \bar\mu) = 0, 
\]
which implies that $\dsone_q$ is a solution of the generalized eigenvalue problem associated with $\la = 0$. This in turn implies that $\dsone_q^T C \rho_i  = \ze^T \rho_i=  0$, which is exactly $\Th^T\ze  = 0$.  
\bigskip

To summarize, we have  found that the solution $\th, b$ minimizing 
\[
- 2\trace(\Th^TCMb) + \trace(b^T\Sig_{XX}b)
\]
subject to $\Th^T C \Th= \Id[r]$ and $\Th^T\ze = 0$ is given by
\begin{enumerate}[label=(\roman*)]
\item $\Th = [\rho_1, \ldots, \rho_r]$ where $\rho_1, \ldots, \rho_r$ are the eigenvectors for the problem
\[
CM\Sig_{XX}^{-1} M^T C\rho = \lambda C\rho
\]
associated with the $r$ largest eigenvalues, normalized so that $\rho^TC\rho = 1$. 
\item $b = \Sig_{XX}^{-1} M^TC\Th$.
\end{enumerate}

The computation can, however, be further simplified.
Let $\la_1, \ldots, \la_r$ be the eigenvalues associated with $\rho_1, \ldots, \rho_r$. Letting $D$ be the associated diagonal matrix, one can write
\[
CM\Sig_{XX}^{-1} M^T C\Th = C\Th D.
\]
This yields
\[
\Th = M\Sig_{XX}^{-1} M^T C\Th D^{-1} = MbD^{-1},
\]
from which we deduce that $\th_g = \Th^T u_g = D^{-1} b^T(\mu_g - \bar\mu)$. So, given a new input vector $x$, the decision rule is to assign it to the class $g$ for which 
\[
|\th_g - b^T(x-\bar\mu)|^2 =  |\Th^T u_g - b^T(x-\bar\mu)|^2 = |D^{-1} b^T(\mu_g - \bar\mu) - b^T(x-\bar\mu)|^2
\]
is minimal. Letting $b_1, \ldots, b_r$ denote the $r$ columns of $b$, this is equivalent to minimizing, in $g$
\begin{equation}
\label{eq:os.rule}
\sum_{j=1}^r (b_j^T (\mu_g - \bar\mu))^2/\la_j^2 - 2 \sum_{j=1}^r (b_j^T (x - \bar\mu))(b_j^T (\mu_g - \bar\mu))/\la_j.
\end{equation}

From $b = \Sig_{XX}^{-1} M^T C \Th$ and $\Th = MbD^{-1}$ we see that 
\[
bD = \Sig_{XX}^{-1} M^T C Mb ,
\]
so that $ \Sig_b b = \Sig_{XX} b D$.
This shows that the columns of $b$ are solution of the eigenvalue problem $\Sig_b u = \la \Sig_{XX} u$. 
Moreover, from $\Th^T C \Th = \Id[r]$, we get $b^T \Sig_b b = D^{2}$. 
Since $b^T   \Sig_b b = b^T\Sig_{XX} b D$, we get that $b$ must be normalized to that $b^T\Sig_{XX} b = D$. 

This shows that the solution of the optimal scoring problem can be reformulated uniquely in terms of $b$: 
if $b_1, \ldots, b_r$ are the $r$ principal solutions of the eigenvalue problem  $\Sig_b u = \la \Sig_{XX} u$, normalized so that $u^T \Sig_{XX} u = \la$, a new input is classified into the class $g$ minimizing
\[
\sum_{j=1}^r \ga_j(\mu_g)^2/\la_j^2 - 2 \sum_{j=1}^r \ga_j(x) \ga_j(\mu_g) /\la_j.
\]
with $\ga_j(x) = b_j^T (x - \bar\mu)$.

\bigskip

\begin{remark}
The following computation shows that optimal scoring is closely related to LDA. 
 Recall the identity $\Sig_{XX} = \Sig_w + \Sig_b$.
It implies that a solution of $\Sig_b u = \la \Sig_{XX} u$ is also a solution of $\Sig_b u = \tilde \la \Sig_{w} u$ with $\tilde \la = \la/(1-\la)$. 
If  $u^T \Sig_{XX} u = \la$, then 
\[
 u^T \Sig_{w} u = \la - u^T \Sig_{b} u = \la - \la^2 = \frac{\tilde \la}{(1+\tilde \la)^2},
 \]
which shows that 
 \[
 \tilde u = \frac{1+\tilde \la}{\sqrt{\tilde\la}} u
 \]
 satisfies $\tilde u^T \Sig_{w} \tilde u = 1$. 
So, 
 \[
 e_j = \frac{1+\tilde \la_j}{\sqrt{\tilde\la_j}} b_j
 \]
 coincide with the LDA directions.
  We have, letting $\tilde \ga_j(x) = e_j^T (x - \bar\mu) = \sqrt{\tilde\la_j}\ga_j(x)/(1+\tilde \la_j)$:
\begin{multline*}
\sum_{j=1}^r \ga_j(\mu_g)^2/\la_j^2 - 2 \sum_{j=1}^r \ga_j(x) \ga_j(\mu_g) /\la_j = 
\sum_{j=1}^r \tilde \ga_j(\mu_g)^2/\tilde \la_j  - 2\sum_{j=1}^r \ga_j(x) \ga_j(\mu_g)/ (1+\tilde \la_j) 
\end{multline*}
which relates the classification rules for the two methods.
\end{remark}

\bigskip

\begin{remark}
Optimal scoring can be modified  by adding a penalty in the form 
\begin{equation}
\label{eq:opt.scoring.penalty}
\ga \sum_{i=1}^r b_i^T\Om b_i = \ga \trace(b^T\Om b)
\end{equation}
where $\Om$ is a weight matrix. This
only modifies the previous discussion by adding $\ga\Om/N$ to both
$\Sig_{XX}$ and $\Sig_w$.
\end{remark}

\subsection{Kernel optimal scoring}
Let $h:\CR_X\to H$ be the feature function and $K$ the associated kernel, as usual.
Optimal scoring in feature space requires to minimize 
$$
\sum_{k=1}^N |\th_{y_k} - {a_0} - \mathbb b(h(x_k))|^2 + \ga \|\mathbb b\|_H^2,
$$
where we have introduced a penalty on $\mathbb b$.
Here, $\mathbb b$ is a linear operator from $H$ to $\mR^r$, therefore taking the form
\[
\mathbb b(h) = \begin{pmatrix} \scp{b_1}{h}_H \\ \vdots\\\scp{b_r}{h}_H
\end{pmatrix}
\]
with $b_1, \ldots, b_r \in H$, and we take
\[
 \|\mathbb b\|_H^2 = \sum_{i=1}^r \|b_i\|_H^2.
 \]
 
It is once again clear (and the argument is left to the reader) that the problem can be reduced to the finite dimensional space $V = \vspan(h(x_1), \ldots, h(x_N))$, and that the optimal $b_1, \ldots, b_r$ must take the form
\[
b_j = \sum_{l=1}^N \al_{li} h(x_l)\,.
\]
Introduce the kernel matrix $\CK = \CK(x_1, \ldots, x_N)$ with $k$th column denoted $\CK^{(k)}$. Let $\al$ be the $N$ by $r$ matrix with entries $\al_{kj}$, $k=1, \ldots, N, j=1, \ldots, r$. Then $\mathbb b(h(x_k))$, which is the vector with coordinates
\[
\scp{b_j}{h(x_k)} = \sum_{l=1}^N \al_{li} K(x_k, x_l),
\]
is equal to  $\al^T \CK^{(k)}$. Moreover 
\[
\|\mathbb b\|_H^2 = \sum_{j=1}^r \sum_{k,l=1}^N \al_{kj} K(x_k, x_l) \al_{lj} = \trace(\al^T \CK\al).
\]
We therefore need to minimize 
\[
\sum_{k=1}^N |\th_{y_k} - {a_0} - \al^T \CK^{(k)}|^2 + \ga \trace(\al^T \CK\al),
\]
so that the problem is reduced to penalized optimal scoring, with $x_k$ replaced by $\CK^{(k)}$, $b$ replaced by $\al$ and the matrix $\Om$ in \cref{eq:opt.scoring.penalty} replaced by $\CK$. Introducing the matrix $P = \Id[N] - \dsone\dsone^T/N$ and
$\CK_c = P\CK$, the covariance matrix $\Sig_{XX}$ becomes $\CK_c^T\CK_c/N = \CK P \CK/N$.

The class averages $\mu_g$ are equal to $\CK \dsone(g)/N_g$ while $\mu = \CK \dsone/N$, so that the matrix $M$ is equal to 
\[
\begin{pmatrix}
\dsone(g_1)^T/N_{g_1} - \dsone^T/N \\
\vdots\\
\dsone(g_q)^T/N_{g_q} - \dsone^T/N
\end{pmatrix}
\CK
\]
which gives $\Sig_b = M^T C M = \CK Q \CK$, where $Q$ is
\[
Q = PCP = \sum_{g\in \CR_Y} \frac{N_g}{N} \left(\frac{\dsone(g)}{N_g} - \frac{\dsone}N\right) \left(\frac{\dsone(g)}{N_g} - \frac{\dsone}N\right)^T \]

So, the columns of $\al$ are the $r$ principal eigenvectors $\rho_1, \ldots, \rho_r$ of the problem 
\[
\CK Q \CK \rho = \frac1 N (\CK P \CK + \ga \CK) \rho.
\]
Given $\al$, one then has, for any $x\in \mR^d$, 
\[
\scp{b_i}{h(x)}_H = \sum_{k=1}^N \al_{ki} K(x,x_k)
\]
and
\[
{a_0}^{(i)} = \frac{1}{N} \sum_{k,l=1}^N \al_{ki} K(x_l, x_k).
\]

\section{Separating hyperplanes and SVMs}

\subsection{One-layer perceptron and margin}
In this whole section, we restrict to  two-class problems, and let $\CR_Y =
\defset{-1,1}$. 
 Given ${a_0}\in \mR$ and $b \neq 0 \in \mR^d$, the equation  ${a_0} + b^T x = 0$ defines a hyperplane in $\mR^d$.
 The function $f(x) = \sign({a_0} + x^T b)$ defines a classifier that attributes a class $\pm1$ to $x$ according to which side of the hyperplane it belongs (we ignore the ambiguity when $x$ is on the hyperplane).
With this
notation, a pair $(x, y)$, where $y$ is the true class, is correctly classified if and only if
$y({a_0} + x^T b) > 0$.

Let $T = (x_1,
y_1, \ldots, x_N, y_N)$ denote, as usual, the training data. A hyperplane, represented by the parameters $({a_0}, b)$ is {\em separating} for $T$ if it correctly classifies all its samples, i.e., if $y_k({a_0} + x_k^T b) > 0$ for $k=1, \ldots, N$. If such a hyperplane exists, one says that $T$ is linearly separable. 
\medskip

Let $\delta$ be a (small) positive number. The perceptron algorithm computes ${a_0}$ and $b$ by minimizing 
\[
L(\be) = \sum_{k=1}^N [\de - y_k({a_0} + x_k^T b)]^+\,.
\]
The problem can be recast as a linear program, i.e., minimize
\[
\sum_{k=1}^N \xi_k
\]
subject to $\xi_k\geq 0, \xi_k + y_k({a_0} + x_k^Tb) -\de \geq
0$ for $k=1, \ldots, N$.

\label{sec:lin.svm}

\begin{figure}[h]
\centering
\includegraphics[height=6.5cm]{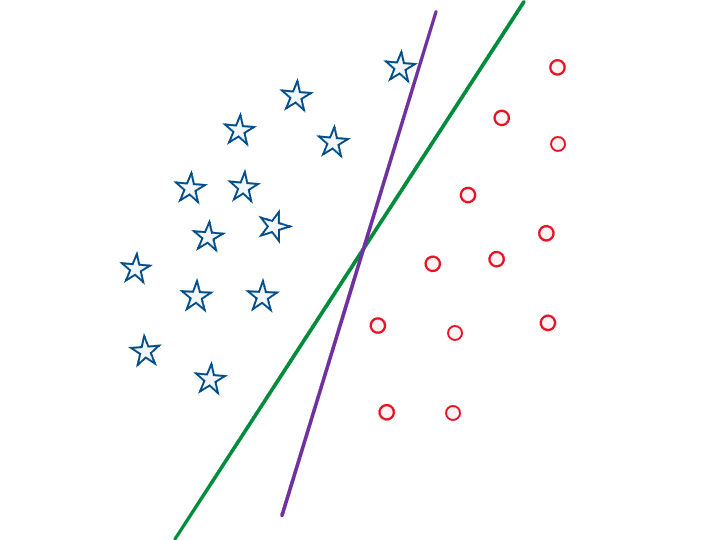}
\caption{\label{fig:svm.1} 
The green line is preferable to the purple one in order to separate the data.
}
\end{figure}

However, when $T$ is linearly separable, separating hyperplanes are not uniquely defined, and there is in general (depending on the choice made for $\de$) an infinity of solutions to the perceptron problem.
Intuitively, one should
prefers a solution that  classifies  the training data with some large
margin, rather than one for which training points may be very close to the
separating boundary (see \cref{fig:svm.1}).

This leads to the maximum margin separating
hyperplane classifier, also called linear SVM, introduced by
Vapnik and Chervonenkis \cite{vapnik1998statistical,vapnik2013nature}.

\subsection{Maximizing the margin}
We will use the following result.
\begin{proposition}
\label{prop:dist.hyp}
The distance of a point $x\in\mR^d$ to the hyperplane $M: {a_0} + b^Tx = 0$ is given by $|{a_0} + x^Tb|/|b|$.
\end{proposition}
\begin{proof}
By definition, $\mathrm{distance}(x, M) = |x - \pi_M(x)|$ where $\pi_M$ is the orthogonal projection on $M$. 
Since $b$ is normal to $M$ and letting $h = \pi_M(x)$, we have $x = \la b + h$ so that  $|\la b| = \mathrm{distance}(x, M)$.
 Writing ${a_0} + b^T h = 0$ in this equation implies ${a_0} + b^T x = \la|b|^2$ so that $|\la|\, |b| = |{a_0} + x^Tb|/|b|$.
\end{proof}

 Assume that $T$ is linearly separable and let $M: {a_0} + b^Tx = 0$ be a separating hyperplane.
 The classification margin is defined as the minimal distance of the input vectors $x_1, \ldots, x_N$ to this hyperplane, i.e., 
\[
m({a_0}, b) = \min\{ |{a_0} + x_k^T b|/|b|: k=1, \ldots, N\}.
\]
 Because the hyperplane is separating, we have $y_k({a_0} + x_k^T b) = |{a_0} + x_k^T b|$ for all $k$, so that we also have
\[
m({a_0}, b) = \min\{ y_k({a_0} + x_k^T b)/|b|: k=1, \ldots, N\}.
\]
 We want to maximize this margin among all separating hyperplanes. This can be expressed as  maximizing, with respect to ${a_0}, b$, the quantity
\[
\min\{ y_k({a_0} + x_k^T b)/|b|: k=1, \ldots, N\}
\]
subject to the constraint that the hyperplane is separating, namely
\[
y_k({a_0} + x_k^T b) \geq 0,\ k=1, \ldots, N.
\]
Introducing a new variable $C$ representing the margin,  the previous problem is equivalent to maximizing $C$ subject to
\[
y_k({a_0} + x_k^T b) \geq C|b|,\ k=1, \ldots, N.
\]

The problem is now overparametrized, and there is no loss of generality in enforcing the additional constraint $C |b| = 1$. Noting that maximizing $C$ is the same as minimizing $|b|^2$, we can now reformulate the maximum margin hyperplane problem as minimizing $|b|^2/2$ subject to
\[
y_k({a_0} + x_k^T b) \geq 1,\ k=1, \ldots, N,
\]
with $C$ (the margin) given by $C = 1/|b|$.
 This results in a quadratic programming problem.

If the data is not separable, there is no feasible point for this problem. To also account for this situation (which is common), 
we can replace the constraint by a penalty and minimize, with respect to ${a_0}$ and  $b$:
\[
\frac{|b|^2}2 + \ga \sum_{k=1}^N (1 - y_k({a_0} + x_k^T b))^+
\]
for some $\ga >0$. (Recall that $x^+ = \max(x, 0)$.) This is equivalent to minimizing the perceptron objective function, with $\de=1$, and with an additional penalty term equal to $|b|^2/(2\ga)$. This minimization problem is equivalent to a quadratic programming problem obtained by introducing  slack variables $\xi_k$, $k=1, \ldots, N$ and  minimizing
\[
\frac{1}{2} |b|^2 + \ga \sum_{k=1}^N \xi_k\,,
\]
subject to the constraints $\xi_k \geq 0$, $y_k({a_0} + x_k^T b) + \xi_k
\geq 1$, for $k=1, \ldots, N$.

\subsection{KKT conditions and dual problem}
Introduce Lagrange multipliers
$\eta_k\geq 0$ for $\xi_k \geq 0$ and $\al_k\geq 0$ for  $y_k({a_0} + x_k^T b) + \xi_k \geq 1$. 
The Lagrangian is then given by
$$
\mathcal L = \frac{1}{2} |b|^2 + \ga \sum_{k=1}^N \xi_k - \sum_{k=1}^N
\eta_k\xi_k -  \sum_{k=1}^N \al_k \big(y_k({a_0} + x_k^T b) +\xi_k - 1\big).
$$
The KKT conditions are
\begin{equation}
\label{eq:svm.kkt.class}
\left\{
\begin{aligned}
&b - \sum_{k=1}^N \al_k y_k x_k = 0\\
&\sum_{k=1}^N \al_k y_k = 0\\
& \ga - \eta_k - \al_k = 0,\quad  k=1, \ldots, N\\
& \xi_k\eta_k = 0, \quad k=1, \ldots, N\\
&  \al_k \big(y_k({a_0} + x_k^T b) +\xi_k - 1\big) = 0, \quad k=1, \ldots, N
\end{aligned}
\right.
\end{equation}

Minimizing $\mathcal L$ with respect to ${a_0}$, $b$ and $\xi_1, \ldots, \xi_N$ and ensuring that the minimum is finite provides the first three KKT conditions.
The resulting dual formulation therefore requires to
maximize
$$
\sum_{k=1}^N \al_k - \frac{1}{2} \sum_{k,l=1}^N \al_k\al_l y_ky_l x_k^Tx_l
$$
subject to the constraints $0\leq \al_k\leq \ga$, $\sum_{k=1}^N \al_k y_k =
0$. 

\bigskip
We now discuss the consequences of the complementary slackness conditions based on the position of training sample relative to the separating hyperplane.
\begin{enumerate}[label=(\roman*),wide=0cm] 
\item First consider indices $k$ such that $(x_k, y_k)$ is correctly classified {\em beyond the margin}, i.e., $y_k({a_0} + x_k^T b ) > 1$.
The last KKT condition and the constraint $\xi_k \geq 0$ require $\al_k = 0$, and the third one then gives $\xi_k=0$.
\item For samples that are misclassified or correctly classified below the margin
\footnote{Note that, even if the training data is linearly separable, there are generally samples that are on the right side of the hyperplane, but at a distance to the hyperplane strictly lower that the ``nominal margin'' $C = 1/|b|$. This is due to our relaxation of the original problem of finding a separating hyperplane with maximal margin.}, i,e.,  $y_k({a_0} + x_k^T b ) < 1$, the constraint $y_k({a_0} + x_k^Tb) +\xi_k \geq 1$ implies $\xi_k >0$, so that $\al_k = \ga$ and $y_k({a_0} + x_k^Tb) +\xi_k = 1$.
\item If $(x_k, y_k)$ is correctly classified exactly at the margin, then $\xi_k=0$ and there is no constrain on $\al_k$ beside belonging to $[0,\ga]$. 
Training samples that lie exactly at the margin are called \alert{support vectors}.
\end{enumerate}

Given a solution $\al_1, \ldots, \al_N$ of the dual problem, one immediately recovers $b$ via the first equation in \cref{eq:svm.kkt.class}. For ${a_0}$, one must, similarly to the regression case, rely on support vectors, which can be identified when $0 < \al_k < \ga$. In this case, one can take ${a_0} = y_k - x_k^T b$. 

If no support vector is found, then ${a_0}$ is not uniquely determined, and can be any value such that $y_k({a_0} + b^Tx_k) \geq 1$ if $\al_k=0$ and $y_k({a_0} + b^Tx_k) \leq 1$ if $\al_k = \ga$.  This shows that ${a_0}$ can be any point in the interval  $[\be^-_{0}, \be^+_{0}]$ with
\begin{align*}
&{a_0}^- =
\max\{ y_k- x_k^Tb: (y_k=1 \text{ and } \al_k=0) \text{ or } (y_k=-1 \text{ and } \al_k=\ga)\}  \\
&{a_0}^+ = \min\{ y_k- x_k^Tb: (y_k=-1 \text{ and } \al_k=0) \text{ or } (y_k=1 \text{ and } \al_k=\ga)\}.
\end{align*}

\subsection{Kernel version}
We make the usual assumptions: $h : \CR_X \to H$ is a feature map with values in an inner-product space with $K(x,y) = \scp{h(x)}{h(y)}_H$.  The predictors take the form $f(x) = \mathrm{sign}({a_0} + \scp{b}{h(x)}_H)$, ${a_0}\in \mR$ and $b\in H$,
and the goal is to minimize
\[
\frac{1}{2} \|b\|_H^2 + \ga \sum_{k=1}^N \xi_k\,,
\]
subject to $\xi_k \geq 0$, $y_k({a_0} + \scp{h(x_k)}{b}_H) + \xi_k \geq 1$, $k=1, \ldots, N$.

 Let $V = \vspan(h(x_1), \ldots, h(x_N))$. The usual projection argument implies that the optimal $b$ must belong to $V$ and therefore take the form
\[
b = \sum_{k=1}^N u_k h(x_k).
\]
We therefore need to minimize
\[
\frac{1}{2} \sum_{k,l=1}^N u_ku_l K(x_k, x_l) + \ga \sum_{k=1}^N \xi_k\,,
\]
subject to
\[
y_k\Big({a_0} + \sum_{l=1}^N K(x_k, x_l) a_l \Big)+ \xi_k
\geq 1
\]
for $k=1, \ldots, N$. Introducing the same Lagrange multipliers as before, the Lagrangian is 
\begin{multline*}
\CL = \frac{1}{2} \sum_{k,l=1}^N u_ku_l K(x_k, x_l) + \ga \sum_{k=1}^N \xi_k \\
- \sum_{k=1}^N
\eta_k\xi_k -  \sum_{k=1}^N \al_k \Big(y_k\Big({a_0} + \sum_{l=1}^N K(x_k, x_l) u_l\Big) +\xi_k - 1\Big)\,.
\end{multline*}
 Using vector notation, we have
\[
\CL = \frac12 u^T \CK u + \xi^T (\ga \dsone - \eta - \al) - {a_0} \al^T y - (\al\odot y)^T\CK u  + \al^T\dsone
\]
where $y\odot\al$ is the vector with coordinates $y_k\al_k$.
 The infimum of $\CL$ is $-\infty$ unless $\ga \dsone - \eta - \al=0$ and $\al^T y=0$.
 If these identities are true, then the optimal $u$ is $u = \al\odot y$ and the minimum of $\CL$ is
\[
- \frac12 (\al\odot y)^T\CK (\al\odot y)  + \al^T\dsone
\]

The dual problem therefore requires to minimize
\[
\frac12 (\al\odot y)^T\CK (\al\odot y)  - \al^T\dsone = \al^T(\CK\odot yy^T) \al  - \al^T\dsone
\]
subject to $\ga \dsone - \eta - \al=0$ and $\al^T y=0$.
\medskip

This is exactly the same problem as the one we obtained in the linear case, up to the replacement of the Euclidean inner products $x_k^Tx_l$ by the kernel evaluations $K(x_k, x_l)$. 
Given the solution of the dual problem, the optimal $b$ is 
\[
b=\sum_{k} u_k h(x_k) = \sum_{k=1}^N \al_k y_k h(x_k).
\]
It is no computable, but the classification rule is explicit and given by
\[
f(x) = \sign\left({a_0} + \sum_{k=1}^N \al_k y_k K(x_k,x)\right).
\]

Similarly to the linear case, the coefficient ${a_0}$ can be identified using a support vector, or is otherwise not uniquely determined. More precisely,  if one of the $\al_k$'s is strictly between $0$ and $\ga$, then ${a_0}$ is given by ${a_0} = y_k - \sum_l \al_l y_l K(x_k,x_l)$. Otherwise, ${a_0}$ is any number between 
\[
a^-_{0} =
\max\defset{ y_k- \sum_l \al_l y_l K(x_k,x_l): (y_k=1 \text{ and } \al_k=0) \text{ or } (y_k=-1 \text{ and } \al_k=\ga)}  
\]
and 
\[
a^+_{0} = \min\defset{ y_k- \sum_l \al_l y_l K(x_k,x_l):  (y_k=-1 \text{ and } \al_k=0)
 \text{ or } (y_k=1 \text{ and } \al_k=\ga)}.
\]

\newpage
\markright{PROBLEMS}
\section*{Problems}
\addcontentsline{toc}{section}{Problems}
\input Problems_Classification

\chapter{Nearest-Neighbor Methods}
\label{chap:nearest.neighbors}

Unlike linear models,  nearest-neighbor methods  are completely non-parametric and
assume no regularity on the decision rule or the regression
function. In their simplest version, they require no training and rely
on the proximity of a new observation to those that belong to the
training set. We will discuss in this chapter how these methods are used for regression and classification, and study some of their theoretical properties.

\section{Nearest neighbors for regression}
\subsection{Consistency}
We let $\CR_X$ denote the input space, and $\CR_Y=\mR^q$ be the output space. We assume that a distance, denoted $\dist$ is defined on $\CR_X$. This means that $\dist: \CR_X\times \CR_X\to [0, +\infty]$ (we allow for infinite values) is a symmetric function such that $\dist(x, x') = 0$ if and only if $x=x'$ and, for all $x,x',x''\in \CR_X$
\[
\dist(x,x') \leq \dist(x,x'') + \dist(x'',x'),
\]
which is the triangle inequality.

Let $T = (x_1, y_1, \ldots, x_N, y_N)$ be the
training set. For $x\in\CR_X$, let 
\[
D_T(x) = (\dist(x, x_k), k=1, \ldots, N)
\]
 be the collection of all distances between $x$ and the inputs in the training set. 
We consider  regression estimators taking the form 
\begin{equation}
\label{eq:nn.general}
\hat f(x) = \sum_{k=1}^N W_k(x) y_k
\end{equation}
where $W_1(x), \ldots, W_N(x)$ is a family of coefficients, or weights, that only depends on $D_T(x)$.

We will, more precisely, use the following construction \cite{stone1977consistent}. Assume that a family of numbers $w_1 \geq w_2 \geq \cdots \geq w_N \geq 0$ is chosen, with $\sum_{j=1}^N w_j = 1$. Given $x\in \mR^d$ and $k\in\{1, \ldots, N\}$, we let 
$r_k^+(x)$ denote the number of indexes $k'$ such  that $\dist(x,x_{k'}) \leq \dist(x, x_k)$ and $r_k^-(x)$ the number of such indexes such  that $d(x,x_{k'}) < d(x, x_k)$. The coefficients defining  $\hat f$ in \cref{eq:nn.general} are then chosen as:
\begin{equation}
\label{eq:wj}
W_k(x) = \frac{\sum_{k' = r^-_k(x)+1}^{r^+_k(x)} w_{k'}}{r^+_k(x) - r^-_k(x)}.
\end{equation}
To emphasize the role of $(w_1, \ldots, w_N)$ is this definition, we will denote the resulting estimation as $\hf_w$.
If there is no tie in the sequence of distances between $x$ and elements of the training set, then $r^+_k(x) = r^-_k(x) + 1$ is the rank of $x_k$ when training data is ordered according to their proximity to $x$, and $W_k(x) = w_{r_k^+(x)}$. In this case, defining $l_1, \ldots, l_N$ such that $d(x, x_{l_1}) < \cdots < d(x, x_{l_N})$, we have
\[
\hf_w(x) = \sum_{j=1}^N w_j y_{l_j}.
\]
In the general case, the weights $w_j$ associated with tied observations are averaged.

If $p$ is an integer,  the $p$-nearest neighbor ($p$-NN) estimator (that we will denote $\hat f_p$) is associated to the weights $w_j = 1/p$ for $j = 1, \ldots, p$ and 0 otherwise. If there is no tie for the definition of the $p$th nearest neighbor of $x$, $W_k(x) = 1/p$ if $k$ is among the $p$ nearest-neighbors and $W_k(x) = 0$ otherwise, so that $\hf_p$ is the average of the output values over these $p$ nearest neighbors. If the $p$th nearest neighbors are tied, their output value is averaged before being used in the sum. For example, assume that $N = 5$ and $p=2$ and let the distances between $x$ and $x_k$ for $k=1, \ldots, 5$ be respectively $9, 3, 2, 4, 6$. Then $\hf_2(x) = (y_2+y_3)/2$. If the distances were $9, 3, 2, 3, 6$, then we would have $\hf_2(x) = (y_2+y_4)/4 + y_3/2$.

When $\CR_X = \mR^d$ and  $d(x,x') = |x-x'|$, the following result is true.
\begin{theorem}[\cite{stone1977consistent}]
\label{th:stone}
Assume that
$\myE(Y^2) < \infty$. Assume that, for each $N$, a sequence $w^{(N)} = w^{(N)}_1 \geq  \cdots \geq w^{(N)}_N \geq 0$ is chosen with $\sum_{j=1}^N w^{(N)}_j =1$. Assume, in addition, that 
\begin{enumerate}[label=(\roman*)]
\item $\lim_{N\to\infty} w^{(N)}_1 = 0 $ 
\item $\lim_{N\to\infty} \sum_{j\geq \al N}w_j^{(N)} \to 0$, for some $\al\in (0,1)$. 
\end{enumerate}
Then the corresponding classifier $\hf_{w^{(N)}}$ converges in the $L^2$ norm to $\myE(Y\mid X)$: 
$$
\myE\left(|\hf_{w^{(N)}}(X) - \myE(Y\mid X)|^2\right) \to 0.
$$
For nearest-neighbor regression, (i) and (ii) mean that the number of nearest neighbors $p_N$ must be chosen such that $p_N \to \infty$ and $p_N/N \to 0$.
\end{theorem}

\begin{proof}
We give a proof under the assumption that $f: x \mapsto \myE(Y\mid X=x)$ is uniformly continuous and
bounded (one can, in fact, prove that it is always possible to reduce to this case).

To lighten the notation, we will not make explicit the dependency on $N$ in of quantities such as $W$ or $w$. One has
\begin{equation}
\label{eq:nn.proof}
\hf_w(X) - \myE(Y\mid X) =
\sum_{k=1}^N
W_{k}(X) (f(X_k) - f(X)) +
  \sum_{k=1}^N W_{k}(X) (Y_k - f(X_k)) 
\end{equation}
and the two sums can be addressed separately. 

%
We start with the first sum  and write, by Schwartz's inequality:
$$
\left(\sum_k W_{k}(X)(f(X_k) - f(X))\right)^2 \leq \sum_k
W_{k}(X)(f(X_k) - f(X))^2.
$$
It therefore suffices to study the limit of 
$E(\sum_k W_{k}(X) (f(X_k)- f(X))^2$. Fix $\ep>0$. By assumption, there
exists $M,a>0$ such that $|f(x)| \leq M$ for all $x$ and $|x-y|\leq a
\Ria |f(x)-f(y)|^2 \leq \ep$.  Then
\begin{align*}
\myE\left(\sum_k W_{k}(X) (f(X_k)- f(X))^2\right) =& \myE\left(\sum_k W_{k}(X) (f(X_k)- f(X))^2\ \bfone_{|X_k-X|\leq
a}\right) \\
& + \myE\left(\sum_k W_{k}(X) (f(X_k)- f(X))^2\ \bfone_{|X_k-X|> a}\right) \\
&\leq \ep^2 + 4M^2 \myE\left(\sum_k W_{k}(X)  \bfone_{|X_k-X|> a}\right).
\end{align*}
Since $\ep$ can be made arbitrarily small, we need to show that, for any positive $a$, the second term in the upper-bound tends to 0  when $N\to\infty$.
We will use the following fact, which requires some minor measure theory argument to prove rigorously. Define
$$
S  = \defset{x: \forall \de > 0, \myP(|X-x|<\de) > 0}.
$$
This set  is called the support of $X$. Then, one can show that $\myP(X\in S) = 1$. This means
that, if $\tilde X$ is independent from $X$ with the same
distribution, then, for any $\de >0$,
$\myP(|X-\tilde X| <\de|X) > 0$ with probability one.
\footnote{This statement is proved  as follows (with the assumption that $X$ is Borel measurable). Let $S^c$ denote the complement of $S$. 
Then $S^c$ is open. 
Indeed if $x\not\in S$, there exists $\de_x>0$ such that, letting $B(x, \de_x)$ denote the open ball with radius $\de_x$, $\myP(X\in {B}(x, \de_x)) = 0$. Then $\myP(X\in B(x', \de_x/3)) = 0$ as soon as $|x-x'|<\de_x/3$, so that $B(x, \de_x/3)\sub S^c$.

If $K \subset S^c$ is compact, then  $K \subset \bigcup_{x\in K}  B(x, \de_x)$ and one can find a finite subset $M\subset K$ such that $K \subset \bigcup_{x\in M}  B(x, \de_x)$, which proves that $\myP(X\in K) = 0$. Since $\myP(X\in S^c) = \max_K \myP(X\in K)$ where the maximum is over all compact subsets of $S^c$, we find $\myP(X\in S^c) = 0$ as required.
}

Let $N_a(x) = |\defset{k: |X_k - x|\leq a} |$. We have, for all $x\in S$ and $a>0$,
and using the law of large numbers,
$$
\frac{N_a(x)}{N} = \frac{1}{N} \sum_{k=1}^N
\bfone_{|X_k-x|\leq a} \to P(|X-x| \leq a) > 0.
$$

If $|X-X_k| > a$, then $r^-_k(X) > N_a(x)$ so that 
\[
\sum_k W_{k}(X)  \bfone_{|X_k-X|> a} \leq \sum_{j\geq N_a(X)} w_j,
\]
and we have, taking $0 < \al < P(|X-x| \leq a)$,  
\[
\myE\left(\sum_{j\geq N_a(X)} w_j\right)\leq
\sum_{j\geq \al N} w_j + \myP(N_a(X) < \al N)
\]
and both terms in the upper bound converge to 0. This shows that the first sum in \eqref{eq:nn.proof} tends to 0.
\bigskip
%

We now consider the second sum in \eqref{eq:nn.proof}. Let $Z_k = Y_k - \myE(Y\mid X_k)$. We have
$\myE(Z_k\mid X_k) = 0$ and $\myE(Z_k^2) <\infty$. We can write
\begin{eqnarray*}
\myE\left(\left|\sum_{k=1}^N W_{k}(X) Z_k\right|^2\right)  &=&\myE\left(\myE\left(\left|\sum_{k=1}^N W_{k}(X)
Z_k\right|^2\, \Big|\,X,X_1, \ldots, X_N\right)\right)\\
&=& \myE\left(\sum_{k=1}^N W_{k}(X)^2 \myE(Z_k^2 \mid X_k)\right) \\
&&+ \sum_{k\neq l=1}^N \myE(W_{k}(X)
W_{l}(X) \myE(Z_kZ_l \mid X_i, X_j))
\end{eqnarray*}
The cross products in the last term vanish because $\myE(Z_k\mid X_k) = 0$ and the samples are
independent. So it only remains to consider
$$
\myE\left(\sum_{k=1}^N W_{k}(X)^2 \myE( Z_k^2 \mid X_k)\right)
$$
The random variable $\myE(Z_k\mid X_k) = \myE(Y_k^2\mid X_k) - \myE(Y_k\mid X_k)^2$ is a fixed non-negative
function of $X_k$, that we will denote $h(X_k)$. 
We have
$$
\myE\left(\sum_{k=1}^N W_{k}(X)^2 h(X_k)\right)\leq w_1
\myE\left(\sum_{k=1}^N W_{k}(X) h(X_i)\right)
$$
with $w_1\to 0$ and the proof is concluded by showing that $\myE\left(\sum_{k=1}^N W_{k}(X) h(X_k)\right)$ is bounded. 

Recall that the weights $W_k$ are functions of $X$ and of the whole training set, and we will need to make this dependency explicit and
write $W_i(X, \mT_X)$ where $\mT_X = (X_1, \ldots, X_N)$. Similarly, the ranks in \eqref{eq:wj} will be written $r^+_j(X, \mT_X)$ and $r^-_j(X, \mT_X)$.

Because $X, X_1, \ldots, X_N$ are i.i.d., we can switch the role of $X$ and $X_k$ in the $k$th term of the sum, yielding
\[
\myE\left(\sum_{k=1}^N W_{k}(X, \mT_X) h(X_k)\right) = \myE\left(\left(\sum_{i=1}^N W_{k}(X_k, \mT_X^{(k)})\right) h(X)\right)
\]
with $\mT_X^{(k)} = (X_1, \ldots, X_{k-1}, X, X_{k+1}, \ldots, X_N)$. We now show that 
$\sum_{k=1}^N W_{k}(X_k, \mT_X^{(k)})$ is bounded independently of $X, X_1, \ldots, X_N$. 

For this purpose, we group $X_1, \ldots, X_N$ according to approximate alignment with $X$. For $u\in\mR^d$ with $|u|=1$ and for $\de \in (0, \pi/4)$, denote by $\Ga(u, \de)$ the cone formed by all vectors $v$ in $\mR^d$ such that $\scp{v}{u} > |v|\cos\de $ (i.e., the angle between $v$ and $u$ is less than $\de$).  Notice that if $v, v'\in \Ga(u, \de)$, then $\scp{v}{v'} \geq \cos(2\de) |v|\,|v'|$ and if $|v'| \leq |v|$, then
\begin{equation}
\label{eq:nn.ineq}
|v|^2 - |v-v'|^2   = |v'|(2|v|\cos(2\de) - |v'|) > 0
\end{equation}
because $\cos(2\de) > 1/2$.

Fixing $\de$, let $C_d(\de)$ be the minimal number of such cones needed to cover $\mR^d$. Choosing such a covering $\Ga(u_1, \de), \ldots, \Ga(u_M, \de)$ where $M = C_d(\de)$, we define the following subsets of $\{1, \ldots, M\}$:
\begin{align*}
I_0 &= \defset{k: X_k = X}\\
I_q &= \defset{k\not \in I_0: X_k-X\in \Ga(u_q, \de)},\quad q=1, \ldots, M
\end{align*}
(these sets may overlap). We have
\[
\sum_{k=1}^N W_{k}(X_k, \mT_X^{(k)}) \leq \sum_{q=0}^M \sum_{k\in I_q} W_{k}(X_k, \mT_X^{(k)})
\]
If $k\in I_0$, then $r^-_k(X_k, \mT_X^{(k)}) = 0$ and $r^+_k(X_k, \mT_X^{(k)}) = c$ with $c = |I_0|$. This implies that, for $k\in I_0$, we have
$W_{k}(X_k, \mT_X^{(k)}) = \sum_{j=1}^c w_j/c$ and 
\[
\sum_{k\in I_0} W_{k}(X_k, \mT_X^{(k)}) = \sum_{j=1}^c w_j.
\]

We now consider $I_q$ with $q\geq 1$. Write $I_q = \{i_1, \ldots, i_r\}$ ordered so that $|X_{i_j} - X|$ is non-decreasing. If $j'<j$, we have (using \eqref{eq:nn.ineq})
$|X_{i_j}-X_{i_{j'}}| < |X-X_{i_j}|$. This implies that $r^-_{i_j}(X_{i_j}, \mT_X^{(i_j)}) \geq j-1$ and $r^+_{i_j}(X_{i_j}, \mT_X^{(i_j)}) - r^-{i_j}(X_{i_j}, \mT_X^{(i_j)}) \geq c+1$. Therefore,
\[
 W_{i_j}(X_{i_j}, \mT_X^{(i_j)}) \leq \frac1{c+1}{\sum_{j'=j}^{c+j} w_{j'}}
 \]
 and
\[
\sum_{k\in I_q} W_{k}(X_k, \mT_X^{(k)}) \leq \frac1{c+1}\sum_{j=1}^N \sum_{j'=j}^{c+j} w_{j'} = \frac1{c+1} \left(\sum_{j'=1}^c j'w_{j'} + (c+1) \sum_{j'=c+1}^N w_{j'}\right).
\]
This yields
\[
\sum_{k=1}^N W_{k}(X_k, \mT_X^{(k)}) \leq \sum_{j=1}^c w_j + C_d(\de) \left(\frac1{c+1}\sum_{j'=1}^c j'w_{j'} + \sum_{j'=c+1}^N w_{j'}\right)
\leq C_d(\de) +1.
\]
We therefore have 
\[
\myE\left(\sum_{k=1}^N W_{k}(X)^2 \myE( Z_k^2 \mid X_k)\right)\leq w_1(C_d(\de) +1) \myE(h(X)) \to 0,
\]
which concludes the proof.
 \end{proof}

Theorem \ref{th:stone} is proved in \citet{stone1977consistent} with weaker hypotheses allowing for more flexibility in the computation of distances, in which, for example, differences $X-X_i$ can be normalized by dividing them by a factor $\sig_i$ that may depend on the training set. These relaxed assumptions slightly complicate the proof, and we refer the reader to \citet{stone1977consistent} for a complete exposition.

\subsection{Optimality}
The NN method can be shown to be optimal over some classes of
functions. Optimality is in the min-max sense, and works as follows. We assume that the regression function $f(x) = \myE(Y\mid X=x)$
belongs to some set $\CF$ of real-valued functions on $\mR^d$. Most of
the time, the estimation methods must be adapted to a given choice of
$\CF$, and various choices have arisen  in the literature: classes of
functions with $r$ bounded derivatives, Sobolev or related spaces,
functions whose Fourier transforms has given properties, etc.

Consider now an estimator of $f$, denoted $\hat f_N$, based on a
training set of size $N$. We can measure the error by, say:
$$
\norm{\hat f_N - f}_2 = \left(\int_{\mR^d} (\hat f_N(x) - f(x))^2
dx\right)^{1/2}
$$

Since $\hat f_N$ is computed from a random sample, this error is a
random variable. One can study, when $b_N\to 0$, the probability
$$
\myP_f\left(\|\hat f_N - f\|^2_2 \geq c b_N\right)
$$
for some constant $c$ and, for example, for the model: $Y = f(X) +
\text{noise}$. Here, the notation $\myP_f$ refers to the model assumption indicating the unobserved function $f$.

The min-max method considers the worst case and computes
$$
M_N(c) = \sup_{f\in \CF}  \myP_f\left(\|\hat f_N - f\|^2_2 \geq c b_N\right).
$$
This quantity now only depends on the estimation algorithm. One defines the notion of 
``lower convergence rate'' as a  sequence $b_N$
such that, for any choice of the estimation algorithm,  $M_N(c)$ can be found arbitrarily close to 1 (i.e., $\|\hat
f_N - f\|^2_2 \geq c b_N$ with arbitrarily high probability for all
$f\in \CF$), for
arbitrarily large $N$ (and for some choice of $c$). The mathematical
statement is
$$
\exists c > 0: \liminf_{N\to\infty} M_N(c) = 1.
$$
So, if $b_N$ is a lower convergence rate, then, for every estimator,
there exists a constant
$c$ such that the accuracy $cb_N$ cannot be achieved. 

On the other hand, one says that $b_N$ is an achievable rate of
convergence if there exists an estimator such that, for some $c'$,
$$
\limsup_{N\to\infty} M_N(c') = 0.
$$
This says that for large $N$, and for some $c'$, the accuracy is
higher than $c'b_N$ for the given estimator. Notice the difference: a
lower rate holds for all estimators, and an achievable rate for at
least one estimator.

The final definition of a min-max optimal rate is that it is both a
lower rate and an achievable rate (obviously for different constants
$c$ and $c'$). And an estimator is optimal in the min-max sense if it
achieves an optimal rate.

One can show that the $p$-NN estimator is optimal (under some assumptions
on the ratio $p_N/N$) when $\CF$ is the class
of Lipschitz functions on $\mR^d$, i.e., the class of functions such that
there exists a constant $K$ with
$$
|f(x) - f(y)|\leq K |x-y|
$$
for all $x,y\in\mR^d$. In this case, the optimal rate is $b_N =
N^{-1/(2+d)}$ (notice again the ``curse of dimensionality'': to achieve
a given accuracy in the worst case, the number of data points must grow exponentially with
the  dimension).

If the function class consists of smoother functions (for example,
several derivatives), the $p$-NN method is not optimal. This is
because the local averaging method is too crude when one knows already
that the function is smooth. But it can be modified (for example by
fitting, using least squares, a polynomial of some degree instead of computing an average)
in order to obtain an optimal rate.

\section{$p$-NN classification}

Let $(x_1,
y_1, \ldots, x_N, y_N)$ be the training set, with $x_i\in\mR^d$ and
$y_i\in \CR_Y$ where $\CR_Y$ is a finite set of classes. Using the same notation as in the previous section, define
\[
\widehat \pi_w(y|x) = \sum_{k=1}^N W_k(x) \bfone_{y_k=y}.
\]
Let the corresponding classifier be 
\[
\hf_w(x) = \argmax_{y\in \CR_Y} \widehat\pi_w(y|x).
\]

\Cref{th:stone} may be applied, for $y\in \CR_Y$, to
the function $f_y(x) = \pi(y\mid x) = \myE(\bfone_{Y=y} \mid X=x)$, which allows
one to interpret the estimator $\what\pi(y \mid x)$ as a nearest-neighbor
predictor of the random variable  $\bfone_{Y=y}$ as a function of $X$. We therefore obtain
the consistency of the estimated posteriors when $N\to\infty$ under the same assumption as those of \cref{th:stone}.
 This implies that, for large $N$, the classification will
be close to Bayes's rule.

An asymptotic comparison with Bayes's rule can already be made  with
$p=1$. Let $\hat y_N(x)$ be the 1-NN  estimator of $Y$
given $x$ and a training set of size $N$, and let $\hat y(x)$ be the
Bayes estimator. We can compute the Bayes error by
\begin{eqnarray*}
\myP(\hat y(X) \neq Y) &=& 1 - \myP(\hat y(X) = Y)\\
&=& 1 - \myE(\myP(\hat y(X) = Y|X))\\
&=& 1 - \myE(\max_{y\in\CR_Y} \pi(y|X))
\end{eqnarray*}

For the 1-NN rule, we have
\begin{eqnarray*}
\myP(\hat y_N(X) \neq Y) &=& 1 - \myP(\hat y_N(X) = Y)\\
&=& 1 - \myE(\myP(\hat y_N(X) = Y|X))\\
\end{eqnarray*}
Let us make the assumption that nearest neighbors are not tied (with probability one). Let $k^*(x, T)$ denote the index of  the nearest neighbor to $x$ in the training set $T$. We have
\begin{eqnarray*}
\myP(\hat y_N(X) = Y\mid X) & = & \myE(\myP(\hat y_N(X) = Y\mid X, \mT))\\
& = & \myE\Big(\sum_{k=1}^N \myP( Y = Y_k\mid X, \mT) \bfone_{k^*(X, \mT) = k}\Big)\\
& = & \myE\Big(\sum_{k=1}^N \myP( Y = Y_k \mid X, X_k) \bfone_{k^*(X, \mT) = k}\Big)\\
& = & \myE\Big(\sum_{k=1}^N \sum_{g\in \CR_Y} \myP( Y = g,  Y_k=g\mid X, X_k) \bfone_{k^*(X, \mT) = k}\Big)\\
& = & \myE\Big(\sum_{k=1}^N \sum_{g\in \CR_Y} \pi( g\mid X) \pi(g\mid X_k) \chi_{k^*(X, \mT) = k}\Big)\\
& = & \myE\Big(\sum_{g\in \CR_Y} \pi( g\mid X) \pi(g\mid  X_{k^*(X,\mT)}) \Big)
\end{eqnarray*}

Now, assume the continuity of $x\mapsto \pi(g\mid x)$ (although the result
can be proved without this
simplifying assumption). We know that $X_{k^*(X,\mT)} \to X$ when
$N\to\infty$ (see the proof of \cref{th:stone}), which implies that
$\pi(g\mid X_{k^*(X,\mT)}) \to \pi(x\mid X)$ and at the limit
$$\myP(\hat y_N(X) = Y\mid X) \to \sum_{g\in \CR_Y} \pi( g\mid X)^2 .$$

This implies that the asymptotic 1-NN misclassification error is always smaller
than 2 times the Bayes error, that is
$$
1- \myE\Big(\sum_{g\in \CR_Y} \pi( g\mid X)^2 \Big) \leq 2(1- \myE(\max_g \pi(
g\mid X) ))
$$
Indeed, the left-hand term is smaller than $1- \myE(\max_g \pi(g|x)^2)$
and the result comes from the fact that for any $t\in\mR$. $1-t^2
\leq 2 - 2t$.

\begin{remark}
\label{rem:nn.comp}
Nearest neighbor methods may require large computation time, since, for a given $x$, the number of
comparisons which are needed is the size of the training set. However,
efficient (tree-based) search algorithms can be used in many cases to
reduce it to a logarithm in the size of the database, which is
acceptable. A reduction of the size of the training set by clustering also
is a possibility for improving the efficiency.

The computation time is also generally proportional to the dimension
$d$ of the input $x$. When $d$ is large, a reduction of dimension is
often a good idea. Principal components (see \cref{chap:dim.red}), or LDA directions (see \cref{chap:lin.class})
can be
used for this purpose.
\end{remark}

\section{Designing the distance}

\paragraph{LDA-based distance}
The most important factor in the design of a NN procedure probably is
the choice of the distance, something we have not discussed
so far. Intuitively, the distance should increase fast in the directions
``perpendicular'' to the regions of constancy of the class variables,
and slowly (ideally not at all) within these regions. The following construction uses  discriminant
analysis \cite{htf03}.

For $g\in \CR_Y$, let $\Sig_g$ be the covariance matrix in class $g$, and $\Sig_w =
\sum_{g\in \CR_Y}\pi_g \Sig_g$ be the within-class variance, where $\pi_g$ is the
frequency of class $g$. Let $\Sig_b$ denote the between-class covariance matrix (see \cref{sec:lda}).

For $x\in\mR^d$, define the spherized vector
$x^* = \Sig_w^{-1/2} x$. The between-class variance computed for spherized
data is $\Sig_b^* = \Sig_w^{-1/2} \Sig_b \Sig_w^{-1/2}$. A direction is discriminant if it
is close to the principal eigenvectors of $\Sig_b^*$. This suggests the
introduction of the norm
$$
|x|_*^2  = (x^*)^T \Sig_b^* x^* = x^T \Sig_w^{-1/2}
(\Sig_w^{-1/2}\Sig_b\Sig_w^{-1/2})\Sig_w^{-1/2}x = x^T \Sig_w^{-1}\Sig_b\Sig_w^{-1}x.
$$
This replaces the standard Euclidean norm (the method can be made more
robust by adding $\ep\Id[d]$ to $\Sig_b^*$.)

\paragraph{Tangent distance}
Designing the distance, however, can sometimes be based on {\it a
priori} knowledge on some invariance properties associated with the classes. A successful example comes from
character recognition, where it is known that transforming images by
slightly rotating, scaling, or translating the character should not
change its class. This corresponds to the following general framework.

For each input $x\in\mR^d$, assume that one can make small transformations without
changing the class of $x$. We model these transformations as
parametrized functions $x \mapsto x_\th = \phi(x, \th)\in\mR^d$, such that $\phi(x,
0) = x$ and $\phi$ is smooth in $\th$, which is a $q$-dimensional parameter. The assumption is that $\phi(x, \th)$
and $x$ should be from the same class, at least for small $\th$. This
will be used to improve on the Euclidean distance on $\mR^d$.

Take $x, x'\in\mR^d$. Ideally, one would like to use the distance $D(x,x') =
\inf_{\th, \th'} \dist(x_\th, x_{\th'})$ where $\th$ and $\th'$ are
restricted to a small neighborhood of 0. A more tractable
expression can be based on first-order approximations
\begin{align*}
&x_\th \simeq x + \nabla_{\th} \phi(x,0) u = x + \sum_{i=1}^q u_i \prt_{\th_i}\phi(x,0)\\
\text{and}\quad & x'_\th \simeq x' + \nabla_{\th} \phi(x',0) u' =x' + \sum_{i=1}^q u'_i \prt_{\th_i}\phi(x',0)
\end{align*}
yielding the approximation (also called the tangent distance)
$$
D(x, x')^2 \simeq \inf_{u,u'\in\mR^q} \norm{x - x' + \nabla_{\th} \phi(x,0) u - \nabla_{\th} \phi(x',0)
u'}^2.
$$
The computation now is a simple least-squares problem, for which the solution is
given by the system
$$
\begin{pmatrix} \nabla_\th \phi(x,0)^T\nabla_\th \phi(x,0) & -\nabla_\th \phi(x,0)^T\nabla_{\th}\phi(x',0)\\
-\nabla_{\th}\phi(x',0)^T\nabla_\th \phi(x,0) & \nabla_{\th}\phi(x',0)^T\nabla_{\th}\phi(x',0)\end{pmatrix} 
\begin{pmatrix}u\\ v\end{pmatrix} = 
\begin{pmatrix} \nabla_\th \phi(x,0)^T (x'-x) \\ \nabla_{\th} \phi(x',0)^T (x-x')\end{pmatrix}.
$$
A slight modification, to ensure that the norms of $u$ and $u'$ are
not too large, is to add a penalty $\la (|u|^2 + |u'|^2)$, which
results in adding $\la \Id[q]$ to the diagonal blocs of the above matrix.


\chapter[Tree-based algorithms]{Tree-based Algorithms, Randomization and Boosting}
\label{chap:trees}

\section{Recursive Partitioning}

 Recursive partitioning methods implement a ``divide and conquer'' strategy to address the prediction problem.
They separate the input space $\CR_X$  into
small regions on which prediction is ``easy,'' i.e., such that the observed values of 
the output variable are (almost) constant for input values in these regions.
The regions  are estimated
by recursive divisions until they become either too small or
homogeneous. These divisions are conveniently represented in the form of binary trees.

\subsection{Binary prediction trees}
 Define a \alert{binary node} to be a structure $\nu$ that contains the following information (note that the definition is recursive):
\begin{enumerate}[label=$\bullet$]
\item A \alert{label} $L(\nu)$ that uniquely identifies the node.
\item A \alert{set of children}, $C(\nu)$, which is either empty or a pair of nodes $(l(\nu), r(\nu))$.
\item A \alert{binary feature}, i.e., a function $\ga_\nu: \CR_X \to \{0,1\}$, which is ``None'' (i.e., irrelevant) if the node has no children.
\item A \alert{predictor}, $f_\nu: \CR_X\to \CR_Y$, which is ``None'' if the node has children.
\end{enumerate}
 A node without children is called a {\em terminal node}, or a leaf.
\medskip

A \alert{binary prediction tree} $\CTR$  is a finite set of nodes, with the following properties:
\begin{enumerate}[label=(\roman*)]
\item Only one node has no parent (the root, denoted $\rho$ or $\rho_{\CTR}$);
\item Each other node has exactly one parent;
\item  No node is a descendent of itself.
\end{enumerate}

\subsection{Training algorithm}
Assume that  a family $\Gamma$ of binary features $\ga: \CR_X \to \{0,1\}$ is chosen, together with 
 a family $\CF$ of predictors $f: \CR_X \to \CR_Y$.
Assume also the existence of two ``algorithms'' as follows:
\begin{enumerate}[label=$\bullet$]
\item Feature selection: Given the feature set $\Gamma$ and a training set $T$, return an optimized binary feature $\what \gamma_{T, \Gamma} \in \Gamma$.
\item Predictor optimization: Given the predictor set $\CF$ and a training set $T$, return an optimized predictor $\hf_{T, \CF}\in \CF$.
\end{enumerate}
 Finally, assume that a stopping rule is defined, as a  function of training sets $\sigma: T \mapsto \sigma(T)\in \{0, 1\}$, where  0 means ``continue'', and 1 means ``stop''.

 Given a training set $T_0$, the algorithm builds a binary tree $\CTR$ using a recursive construction.
 Each node $\nu\in \CTR$ will  be associated to a subset of $T_0$, denoted $T_\nu$.
 We  define below a recursive operation, denoted $\mathrm{Node}(T, j)$ that adds a node $\nu$ to a tree $\CTR$ given a subset $T$ of $T_0$ and a label $j$.
 Starting with $\CTR= \emptyset$, calling $\mathrm{Node}(T_0, 0)$ will then create the desired tree. 

\begin{algorithm}[Node insertion: $\mathrm{Node}(T, j)$]
\begin{enumerate}[label=(\alph*)]
\item Given $T$ and $j$, let $T_\nu = T$ and $L(\nu) = j$.
\item If $\sigma(T) = 1$, let $C(\nu) = \emptyset$, $\ga_\nu =\text{``None''}$ and $f_\nu = \hf_{T, \CF}$.
\item If $\sigma(T) = 0$, let $f_\nu =\text{``None''}$, $\ga_\nu = \what \ga_{T, \Ga}$ and $C(\nu) = (l(\nu), r(\nu))$ with 
\[
l(\nu) = \mathrm{Node}(T_l, 2j+1), \quad r(\nu) = \mathrm{Node}(T_r, 2j+2)
\]
where 
\[
T_l = \{(x,y)\in T: \ga_\nu(x) = 0\}, \quad
T_r = \{(x,y)\in T: \ga_\nu(x) = 1\}
\]
\item Add $\nu$ to $\CTR$ and return.
\end{enumerate}
\end{algorithm}

\begin{remark}
Note that, even though the learning algorithm for prediction trees can be very conveniently described in recursive form as above, efficient computer implementations should avoid recursive calls, which may be inefficient and memory demanding. Moreover, for large trees, it is likely that recursive implementations will reach the maximal number  of recursive calls imposed by compilers.
\end{remark}

\subsection{Resulting predictor}
Once the tree is built, the predictor $x \mapsto \hf_{\CTR}(x)$ is  recursively defined as follows.
 \begin{enumerate}[label=(\alph*)]
 \item Initialize the computation with $\nu=\rho$. 
 \item At a given step of the algorithm,  let $\nu$ be the current node.
 \begin{itemize}[wide=0.5cm]
 \item If $\nu$ has no children: then let $\hf_{\CTR} (x) = f_\nu(x)$.
 \item Otherwise: replace $\nu$ by $l(\nu)$ if $\ga_\nu(x) = 0$ and by $r(\nu)$ if $\ga_\nu(x) = 1$ and go back to (b). 
 \end{itemize}
\end{enumerate}

\subsection{Stopping rule}
 The function $\sigma$, which decides whether a node is terminal or not is generally defined based on very simple rules. Typically, $\sigma(T)=1$  when one the following conditions is satisfied:
\begin{enumerate}[label=$\bullet$]
\item The number of training examples in $T$  is small (e.g., less than 5).
\item The values $y_k$ in $T$ have a small variance (regression) or are constant (classification). 
\end{enumerate}

\subsection{Leaf predictors}
 When one reaches a terminal node $\nu$ (so that $\sig(T_\nu) = 1$), a predictor $f_\nu$ must be determined. 
This function can be optimized within any set $\CF$ of predictors, using any learning algorithm, but in practice, one usually makes this fairly simple and defines $\CF$ to be the family of constant functions taking values in $\CR_Y$. The function
 $\hf_{T, \CF}$ is then defined as:
\begin{enumerate}[label=$\bullet$]
\item the average of the values of $y_k$, for $(x_k, y_k)\in T$ (regression);
\item the mode of the distribution of $y_k$, for $(x_k, y_k)\in T$ (classification).
\end{enumerate}

\subsection{Binary features}
  The space $\Gamma$ of possible binary features must be specified in order to partition non-terminal nodes. A standard choice, used in the CART model \citep{breiman1984classification} with $\CR_X = \mR^d$, is 
\begin{equation}
\label{eq:cart}
\Gamma = \defset{\ga(x) = \bfone_{[\pe x i \geq \th]}, i=1, \ldots, d,
\th \in\mR}
\end{equation}
where $\pe x i$ is the $i$th coordinate of $x$.
 This corresponds to
splitting the space using a hyperplane parallel to one of the coordinate
axes.

 The binary function $\what\ga_{T, \Ga}$ can be optimized over $\Ga$ using a greedy evaluation of the risk, assuming that the prediction is based on the two nodes resulting from the split. 
 For $\ga\in \Gamma$, $f_0, f_1\in \CF$, define 
\[
F_{\ga, f_0, f_1}(x) = 
\left\{
\begin{aligned}
f_0(x) \text{ if } \ga(x) = 0\\
f_1(x) \text{ if } \ga(x) = 1
\end{aligned}
\right.
\]
  Given a risk function $r$, one then evaluates 
\[
\CE_{T}(\ga) = \min_{f_0, f_1\in \CF} \sum_{(x,y)\in T} r(y, F_{\ga, f_0,f_1}(x))
\]
 One then chooses $\what\ga_{T, \Ga} = \argmin_{\ga\in\Ga} (\CE_{T}(\Ga))$.

\begin{example}[Regression]
Consider the regression case, taking squared differences as risk and letting $\CF$ contain only constant functions.
 Then
\[
\CE_{T}(\ga) = \min_{m_0, m_1} \sum_{(x,y)\in T} \left((y-m_0)^2
\bfone_{\ga(x) = 0} + (y-m_1)^2
\bfone_{\ga(x) = 1}\right).
\]
Obviously, the optimal $m_0$ and $m_1$ are the averages 
of the output values, $y$, in each of the subdomains defined by $\ga$.
For CART (see \cref{eq:cart}), this cost must be minimized over all choices $(i, \th)$ with $i=1, \dots, d$ and $\th\in \mR$ where $\ga_{i,\th}(x) = 1$ if $x(i) > \th$ and 0 otherwise.
\end{example}
 
\begin{example}[Classification.]
For classification, one can apply the same method, with the 0/1 loss, letting
\[
\CE_{T}(\ga) = \min_{g_0, g_1} \sum_{(x,y)\in T} \left(\bfone_{y\neq g_0}
\bfone_{\ga(x) = 0} + \bfone_{y\neq g_1}
\bfone_{\ga(x) = 1}\right).
\]
 The optimal $g_0$ and $g_1$ are the majority classes in $T \cap \{\ga = 0\}$ and $T \cap \{\ga = 1\}$.
 \end{example}

\begin{example}[Entropy selection for classification]
For classification trees, other splitting criteria may be used based on  the empirical probability $p_T$ on the set $T$, defined as 
\[
p_T(A) = \frac1 N |\{k: (x_k, y_k) \in A\}|
\]
for $A\subset \CR_X \times \CR_Y$.
 The previous criterion, $\CE_{T}(\ga)$, is proportional to 
\[
p_T(\ga = 0) (1- \max_{g} p_T(g\mid\ga=0)) + p_T(\ga = 1) (1- \max_{g} p_T(g\mid \ga=1)) .
\]
 One can define alternative objectives in the form
\[
p_T(\ga = 0) \mathcal H (p_T(g\mid \ga=0)) + p_T(\ga = 1) \mathcal H (p_T(g\mid \ga=1))
\]
where $\pi \to \mathcal H(\pi)$ associates to a probability distribution $\pi$ a ``complexity measure'' that is minimal when $\pi$ is concentrated on a single class (which is the case for $\pi \mapsto 1 - \max_g \pi(g)$).

 Many such measures exists, and many of them are defined as various forms of entropy designed in information theory.
 The most celebrated   is Shannon's entropy \citep{shannon1949communication}, defined by
\[
\mathcal H(p) =- \sum_{g\in \CR_Y} p(g) \log p(g)\,.
\]
It 
is always positive, and minimal when the distribution is concentrated on a single class. Other entropy measures include:
\begin{enumerate}[label=$\bullet$]
\item The Tsallis entropy: $\mathcal H(p) = \frac1{1-q} \sum_{g\in \CR_Y} (p(g)^q - 1)$, for $q\neq 1$. (Tsallis entropy for $q=2$ is sometimes called the Gini impurity index.)
\item The Renyi entropy: $\mathcal H(p) = \frac1{1-q} \log \sum_{g\in \CR_Y} p(g)^q $, for $q\geq 0, q\neq 1$. 
\end{enumerate}
\end{example}

\subsection{Pruning}

Growing  a decision tree to its maximal depth (given the amount of
available data) generally leads to predictors that overfit the data. The training algorithm is usually  followed by a pruning step that removes some some nodes based on a complexity penalty.

Letting $\tau(\CTR)$ denote the set of terminal nodes in the tree $\CTR$ and $\hat f_{\CTR}$ the associated predictor, pruning is represented as an optimization problem, where one minimizes, given the training set $T$,
\[
U_\lambda(\CTR, T) =  \hat R_T(\hat f_{\CTR}) +\lambda |\tau(\CTR)|
\]
where $\hat R_T$ is as usual the in-sample error measured on the training set $T$. 

To prune a tree, one selects one or more internal nodes and remove all their descendants (so that these nodes become terminal). Associate to each node $\nu$ in $\CTR$ its local in-sample error $\CE_{T_\nu}$ equal to the error made by the optimal classifier estimated from the training data associated with $\nu$. Then, 
\[
U_\lambda(\CTR, T) = \sum_{\nu\in \tau(\CTR)} \frac{|T_\nu|}{|T|} \CE_{T_\nu} + \lambda |\tau(\CTR)|
\]

If $\nu$ is a node in $\CTR$ (internal or terminal), let $\CTR_\nu$ be the subtree of $\CTR$ containing $\nu$ as a root and all its descendants. Let $\CTR^{(\nu)}$ be the tree $\CTR$ will all descendants of $\nu$ removed (keeping $\nu$). Then
\[
U_\lambda(\CTR, T) = U_0(\CTR^{(\nu)}, T) - \frac{|T_\nu|}{|T|} (\CE_{T_\nu} - U_0(\CTR_\nu, T_\nu))  + \lambda (|\tau(\CTR_\nu)|-1).
\]

Note also that, if $\nu$ is internal,  and $\nu'$, $\nu''$ are its children, then
\[
U_0(\CTR_\nu, T_\nu) = \frac{|T_{\nu'}|}{|T_\nu|} U_0(\CTR_{\nu'}, T_{\nu'}) + \frac{|T_{\nu''}|}{|T_\nu|} U_0(\CTR_{\nu''}, T_{\nu''})
\]
This formula can be used to compute $U_0(\CTR_\nu) $ recursively for all nodes,  starting with leaves for which  $U_0(\CTR_\nu)  = \CE(T_\nu)$. (We also have $|\tau(\CTR_\nu)| = |\tau(\CTR_{\nu'})| + |\tau(\CTR_{\nu''})|$.)  The following algorithm converges  to a \alert{global minimizer} of $U_\lambda$. 

\begin{algorithm}[Pruning]
\label{alg:pruning} 
\begin{enumerate}
\item Start with a complete tree $\CTR(0)$ built without penalty. 
\item Compute, for all nodes $U_0(\CTR_\nu)$ and $|\tau(\CTR_\nu)|$. Let 
\[
\psi_\nu = \frac{|T_\nu|}{|T|} (\CE_{T_\nu} - U_0(\CTR_\nu))  - \lambda (|\tau(\CTR_\nu)|-1).
\]
\item Iterate the following steps.
\begin{itemize}
\item If $\psi_\nu <0$ for all internal nodes $\nu$, exit the program and return the current $\CTR(n)$.
\item Otherwise choose an internal node $\nu$ such that $\psi_\nu$ is largest. 
\item Let $\CTR(n+1) = \CTR^{(\nu)}(n)$.  Subtract $ \lambda (|\tau(\CTR_\nu(n))|-1)$ to  $\rho_{\nu'}$ for all $\nu'$ ancestor of $\nu$.
\end{itemize}
\end{enumerate}
\end{algorithm}

 \section{Random Forests}
 \subsection{Bagging}
A random forest \cite{amit1997shape,breiman2001random} is a special case of composite predictors (we will see other examples later in this chapter when describing boosting methods) that train multiple individual predictors under various conditions and combine them, through averaging, or majority voting.  
  With random forests, one generates individual trees by randomizing the parameters of the learning
process. One way to achieve this is to randomly  sample from the training set before running the training algorithm.

Letting as before $T_0 = (x_1, y_1, \ldots, x_N, y_N)$ denote the original set, with size $N$, one can create ``new'' training data by sampling with replacement from $T_0$. More precisely, consider the family of independent random variables 
$\bfxi = (\bfxi_1, \ldots, \bfxi_N)$, with each $\bfxi_j$ following a uniform distribution over $\{1, \ldots, N\}$. One can then form the random training set 
\[
T_0(\bfxi)   = (x_{\bfxi_1},y_{\bfxi_1}, \ldots, x_{\bfxi_N},y_{\bfxi_N}).
\]
Running the training algorithm using $T_0(\bfxi)$ then provides a random tree, denoted $\CTR(\bfxi)$. Now, by sampling $K$ realizations of $\bfxi$, say $\xi^{(1)}, \ldots, \xi^{(K)}$, one obtains a collection of $K$ random trees (a random forest) $\CTR^* = (\CTR_1, \ldots, \CTR_K)$, with $\CTR_j = \CTR(\xi^{(j)})$ that can be combined to provide a final predictor. The simplest way to combine them is to average the predictors returned by each tree (assuming, for classification, that this predictor is a probability distribution on classes), so that
\begin{equation}
\label{eq:rf.mean}
f_{\CTR^*}(x) = \frac 1 K \sum_{j=1}^K f_{\CTR_j}(x).
\end{equation}
For classification, one can alternatively let each individual tree ``vote'' for their most likely class.

Obviously, randomizing training data and averaging the predictors is a general approach that can be applied to any prediction algorithm, not only to decision trees.  In the literature, the approach described above has been called  \alert{bagging} \citep{breiman1996bagging}, which is an acronym for ``bootstrap aggregating'' (bootstrap itself being a general resampling method in statistics that samples training data with replacement to determine some properties of  estimators). Another way to randomize predictors (especially when $d$, the input dimension is large), is to randomize input data by randomly removing some of the coordinates, leading to a similar construction. 

With decision trees one can in addition randomize the binary features use to split nodes, as described next. While bagging may provide some enhancement to predictors, feature randomization for decision trees often significantly improves the performance, and is the typical randomization method used for random forests.

\subsection{Feature randomization}
  When one decides to split a node during the construction of a prediction tree,  one can optimize the binary feature $\ga$ over a random subset of $\Gamma$ rather than exploring the whole set. 
  For CART, for example, one can select a small number of dimensions $i_1, \ldots, i_q\in \{1, \ldots, d\}$ with $q\ll d$, and optimize $\ga$ by thresholding one of the coordinates $x^{(i_j)}$ for $j\in \{1, \ldots, q\}$.
  This results in a randomized version of the node insertion function. 

\begin{algorithm}[Randomized node insertion: $\mathrm{RNode}(T, j)$]
\begin{enumerate}[label=(\alph*), wide = 0.5cm]
\item Given $T$ and $j$, let $T_\nu = T$ and $L(\nu) = j$.
\item If $\sigma(T) = 1$, let $C(\nu) = \emptyset$, $\ga_\nu =\text{``None''}$ and $f_\nu = \hf_{T, CF}$.
\item If $\sigma(T') = 0$, \alert{sample (e.g., uniformly without replacement) a subset $\Ga_\nu$ of $\Ga$} and let  $f_\nu =\text{``None''}$, $\ga_\nu = \hat \ga_{T,\Ga_\nu}$ and $C(\nu) = (l(\nu), r(\nu))$ with 
\[
\begin{aligned}
l(\nu) &= \mathrm{Node}(T_l, 2j+1)\\
r(\nu) &= \mathrm{Node}(T_r, 2j+2)
\end{aligned}
\]
where 
\[
\begin{aligned}
T_l &= \{(x,y)\in T: \ga_\nu(x) = 0\}\\
T_r &= \{(x,y)\in T: \ga_\nu(x) = 1\}
\end{aligned}
\]
\item Add $\nu$ to $\CTR$ and return.
\end{enumerate}
\end{algorithm}

Now, each time the function $\mathrm{RNode}(T_0, 0)$ is run, it returns a different, random,  tree.
  If it is called $K$ times, this results in a random forest  $\CTR^* = (\CTR_1, \ldots \CTR_K)$, with a  predictor $\CF_{\CTR^*}$ given by \eqref{eq:rf.mean}.  Note that trees in random forests are generally not pruned, since this operation has been observed to bring no improvement in the context of randomized tress.

\section{Top-Scoring Pairs}
Top-Scoring Pair (TSP) classifiers were introduced in \citet{geman2004classifying} and can be seen as forests formed with depth-one classification trees in which splitting rules are based on the comparison of pairs of variables. More precisely, define
\[
\ga_{ij}(x) = \bfone_{x^{(i)} > x^{(j)}}.
\]
A decision tree based on these rules only relies on the order between the features, and is therefore  well adapted to situations in which the observations are subject to  increasing transformations, i.e., when the observed variable $X$ is such that $X^{(j)} = \phi(Z^{(j)})$, where $\phi: \mR\to\mR$ is random and increasing and $Z$ is a latent (unobserved) variable. Obviously, in such a case, order-based splitting rules do not depend on $\phi$. Such an assumption is relevant, for example, when experimental conditions (such as temperature) may affect the actual data collection, without changing their order, which is the case when measuring high-throughput biological data, such as microarrays, for which the approach was introduced.

Assuming two classes, a depth-one tree in this context is simply the classifier $f_{ij} = \ga_{ij}$. Given a training set, the associated empirical error is 
\[
\CE_{ij} = \frac1N \sum_{k=1}^N \bfone_{\ga_{ij}(x_k) \neq y_k} =\frac1N  \sum_{k=1}^N |y_k - \ga_{ij}(x_k)|
\]
and the balanced error (better adapted to situations in which one class is observed more often than the other) is
\[
\CE^b_{ij} = \sum_{k=1}^N w_k |y_k - \ga_{ij}(x_k)|
\]
with $w_k = 1/(2N_{y_k})$, where $N_0$, $N_1$ are the number of observations with $y_k=0$, $y_k=1$. Pairs $(i,j)$ with small errors are those for which the order between the features switch with high probability when passing from class 0 to class 1.

In its simplest form, the TSP classifier defines the set
\[
\CP = \argmin_{ij} \CE^b_{ij}
\]
of global minimizers of the empirical error (which may just be a singleton) and predicts the class based on a majority vote among the family of predictors  $(f_{ij}, (i,j) \in \CP)$. Equivalently, selected variables maximize the score $\De_{ij} = 1 - \CE^b_{ij}$, leading to the method's name.

Such classifiers, which are remarkably simple, have been found to be competitive among a wide range of ``advanced'' classification algorithms for large-dimensional problems in computational biology. The method has been refined in \citet{tan2005simple}, leading to the $k$-TSP classifier, which addresses the following remarks.  First, when $j,j'$ are highly correlated, and $(i,j)$ is a high-scoring pair, then $(i,j')$ is likely to be one too, and their associated decision rules will be redundant. Such cases should preferably be pruned from the classification rules, especially if one wants to select a small number of pairs. Second, among pairs of features that switch with the same probability, it is natural to prefer those for which the magnitude of the switch is largest, e.g., when the pair of variables switches from a regime in which one of them is very low and the other very high to the opposite. In \citet{tan2005simple}, a rank-based tie-breaker is introduced, defined as
\[
\rho_{ij} = \sum_{k=1}^N w_k (R_k(i) - R_k(j))(2y_k-1),
\]
where $R_k(i)$ denotes the rank of $x_k^{(i)}$ in $x_k^{(1)}, \ldots, x_k^{(d)}$.
One can now order pairs $(i,j)$ and $(i', j')$ by stating that the former scores higher if (i) $\De_{ij} > \De_{i'j'}$, or (ii) $\De_{ij} =\De_{i'j'}$ and $\rho_{ij} > \rho_{i'j'}$. The $k$-TSP classifier is formed by selecting pairs, starting from the highest scoring one, and use as $l$th pair (for $l\leq k$)  the highest scoring ones among all those that do not overlap with the previously selected ones. In \cite{tan2005simple}, the value of $k$ is optimized using cross-validation. 

\section{Adaboost}

Boosting methods refer to algorithms in which classifiers are enhanced by
recursively making them focus on harder data. We first address the issue
of classification, and describe one of the earliest  algorithms (Adaboost). We will then
interpret it as a greedy gradient descent algorithm, as this interpretation will lead to further extensions. 

\subsection{General set-up}
We first consider binary classification problems, with $\CR_Y=\{-1,1\}$. We want to design a function $x \mapsto F(x) \in \{-1, 1\}$ on the
basis of a training set $T = (x_1, y_1, \ldots, x_N, y_N)$. With the 0-1
loss, minimizing the  empirical error is equivalent to maximizing
$$
\CE_T(F) = \frac{1}{N} \sum_{k=1}^N y_k F(x_k).
$$

Boosting algorithms build the function $F$ as a linear combination of
``base classifiers,'' $f_1, \ldots, f_M$, taking
$$
F(x) = \text{sign}\left(\sum_{j=1}^M \al_j f_j(x)\right)\,.
$$
We assume that each base classifier, $f_j$, takes values in $[-1, 1]$ (the interval). 
\medskip

The sequence of base classifiers is learned by progressively focusing on the hardest examples.  We will therefore assume that the training algorithm for base classifiers takes as input the training set $T$ as well a family of positive weights $W = (w_1, \ldots, w_N)$. More precisely, letting
\[
p_W(k) = \frac{w_k}{\sum_{k=1}^N w_k}, 
\]
the weighted algorithm should implement (explicitly or implicitly) the equivalent of an unweighted algorithm on a simulated training set obtained by sampling with replacement $K\gg N$ elements of $T$  according to $p_W$ (ideally letting $K\to \infty$). Let us take a few examples.
\begin{enumerate}[label=$\bullet$, wide=0.5cm]
\item Weighted LDA: one can use LDA as described in \cref{sec:lda} with
\[
c_g =  \sum_{k: y_k = g} p_W(k),\quad \mu_g = \frac{1}{c_g}\sum_{k: y_k = g} p_W(k) x_k,\quad {\mu} = \sum_{g\in\CR_Y} c_g \mu_g
\]
 and the covariance matrices:
\[
\Sigma_w = \sum_{k=1}^N p_W(k) (x_k - \mu_{y_k})(x_k-\mu_{y_k})^T,
\quad
\Sigma_b = \sum_{g\in\CR_Y} c_g (\mu_g - \bar \mu)(\mu_g - \bar \mu)^T.
\]
\item Weighted logistic regression: just maximize
\[
\sum_{k=1}^N p_W(k) \log \pi_\theta(y_k|x_k)
\]
where $\pi_\theta$ is given by the logistic model.
\item Empirical risk minimization algorithms can  be modified in order to minimize
\[
\hat R_{T, W}(f)  = \sum_{k=1}^N w_k r(y_k, f(x_k)).
\]
\item Of course, any algorithm can be run on a training set resampled using $p_W$.
\end{enumerate}

\subsection{The Adaboost algorithm}
Boosting algorithms keep track of a family of weights and modify it after the $j$th classifier $f_j$ is computed, increasing the importance of misclassified examples, before computing the next classifier. The following algorithm, called Adaboost \cite{schapire1990strength,freund1997decision-theoretic}, describes one such approach.
\begin{algorithm}[Adaboost]
\label{alg:boosting}
\begin{enumerate}[label=$\bullet$]
\item Start with uniform weights, letting  $W(1) = (w_1(1), \ldots, w_N(1))$ with $w_k(1) = 1/N$, $k=1, \ldots, N$. Fix a number $\rho\in (0,1]$ and an integer $M>0$. 
\item Iterate, for $j=1, \ldots, M$:
\begin{enumerate}[wide=0.5cm]
\item Fit a base classifier $f_j$ using the weights $W(j) = (w_1(j), \ldots, w_N(j))$. Let
\begin{subequations}
\begin{align}
\label{eq:sw.1}
S_w^+(j) &= \sum_{k=1}^N w_k(j) (2-|y_k- f_j(x_k)|)\\
\label{eq:sw.2}
S_w^-(j) &= \sum_{k=1}^N w_k(j) |y_k -  f_j(x_k)|
\end{align}
\end{subequations}
and define $\al_j = \rho \log\big(S_w^+(j)/S_w^-(j)\big)$
\item Update the weights by
\[
w_k(j+1) = w_k(j) \exp\big(\al_j |y_k -  f_j(x_k)|/2\big).
\]
\end{enumerate} 
\item Return the classifier:
$$
F(x) = \text{sign}\left(\sum_{j=1}^M \al_j f_j(x)\right)\,.
$$
\end{enumerate}
\end{algorithm}
If $f_j$ is binary, i.e., $f_j(x) \in \{-1,1\}$, then $|y_k- f_j(x_k)| = 2 \bfone_{y_k \neq f_j(x_k)}$, so that $S_W^+/2$ is the weighted number of correct classifications and $S_W^-/2$ is the weighted number of incorrect ones.

For $\al_j$ to be positive, the $j$th classifier must do
better than pure chance on the weighted training set. If not, taking $\al_j \leq 0$ reflects the fact that, in that case, $-f_j$ has better performance on training data.

 Algorithms that  
 do slightly better than chance with high probability are called ``weak learners'' \cite{schapire1990strength}. The following proposition \citep{freund1997decision-theoretic} shows that, if the base classifiers reliably perform strictly better than chance (by a fixed, but not necessarily large, margin), then the boosting algorithm can make the training-set error arbitrarily close to 0.
 
\begin{proposition}
\label{prop:boosting}
Let $\CE_T$ be the training set error of the  estimator $F$ returned by \cref{alg:boosting}, i.e., 
\[
\CE_T = \frac1N  \sum_{k=1}^N \bfone_{y_k\neq F(x_k)}.
\]
 Then
\[
\CE_T \leq \prod_{j=1}^M \Big(\ep_j^\rho(1-\ep_j)^{1-\rho} + \ep_j^{1-\rho}(1-\ep_j)^{\rho}\Big)
\]
where
\[
\ep_j =  \frac{S_W^-(j)}{S_W^+(j)+S_W^-(j)}.
\]
\end{proposition}
\begin{proof}
We note that example $k$ is misclassified by the final classifier if and only if
\[
\sum_{j=1}^M \alpha_j y_kf_j(x_k) \leq 0
\]
or
\[
\prod_{j=1}^M e^{-\alpha_j y_k f_j(x_k)/2} \geq 1
\]
Noting that $|y_k - f_j(x_k)| = 1 - y_k f_j(x_k)$, we see that example $k$ is misclassified when
\[
\prod_{j=1}^M e^{\alpha_j |y_k - f_j(x_k)|/2} \geq \prod_{j=1}^M e^{\alpha_j/2}.
\]
This shows that
\begin{align*}
\CE_T &= \frac1N  \sum_{k=1}^N \bfone_{y_k\neq F(x_k)}\\
&= \frac1N  \sum_{k=1}^N \bfone_{\prod_{j=1}^M e^{\alpha_j |y_k - f_j(x_k)|/2} \geq \prod_{j=1}^M e^{\alpha_j/2}}\\
&\leq \frac1N  \sum_{k=1}^N \prod_{j=1}^M e^{\alpha_j |y_k - f_j(x_k)|/2}  \prod_{j=1}^M e^{-\alpha_j/2}.
\end{align*}
Let, for $q\leq M$,
\[
U_q = \frac1N  \sum_{k=1}^N \prod_{j=1}^q e^{\alpha_j |y_k - f_j(x_k)|/2}.
\]
Since
\[
w_k(q) =  \frac1N \prod_{j=1}^{q-1} e^{\alpha_j |y_k - f_j(x_k)|/2},
\]
we also have $U_q = \sum_{k=1}^N w_k(q+1) = (S_W^+(q+1) + S_W^-(q+1))/2$.

We will use the inequality\,\footnote{This inequality is clear for $\alpha=0$. Assuming $\alpha\neq 0$,
the difference between the upper and lower bound is
\[
q(t) = 1 - e^{\alpha t} - (1 - e^\al)t.
\]
The function $q$ is concave (its second derivative is $-\alpha^2 e^{\al t}$) with $q(0) = q(1) = 0$ and therefore non-negative over $[0,1]$. 
} 
\[
e^{\alpha t} \leq 1 - (1-e^\alpha) t, 
\]
which is true for all $\al\in\mR$ and $t\in [0,1]$, to write
\begin{align*}
U_q &\leq  \frac1N  \sum_{k=1}^N \prod_{j=1}^{q-1} e^{\alpha_j |y_k - f_j(x_k)|/2} (1 - (1-e^{\alpha_q})|y_k - f_q(x_k)|/2)\\
&= \sum_{k=1}^{N} w_k(q) (1 - (1-e^{\alpha_q})|y_k - f_q(x_k)|/2)\\
&= \sum_{k=1}^{N} w_k(q)  - (1-e^{\alpha_q})\sum_{k=1}^{N} w_k(q)|y_k - f_q(x_k)|/2\\
&= U_{q-1} (1 - (1-e^{\alpha_q}) \epsilon_q)
\end{align*}
This gives (using $U_0 = 1$)
\[
U_M \leq \prod_{j=1}^M \Big(1 - (1-e^{\alpha_j}\Big) \epsilon_j)
\]
and
\[
\CE_T \leq \prod_{j=1}^M \Big(1 - (1-e^{\alpha_j}) \epsilon_j\Big)e^{-\alpha_j/2}.
\]
It now suffices to replace $e^{\alpha_j}$ by $(1-\epsilon_j)^\rho\epsilon_j^{-\rho}$ and note that 
\[
\Big(1 - (1-(1-\epsilon_j)^\rho\epsilon_j^{-\rho}) \epsilon_j\Big)(1-\epsilon_j)^{-\rho/2}\epsilon_j^{\rho/2} = \ep_j^\rho(1-\ep_j)^{1-\rho} + \ep_j^{1-\rho}(1-\ep_j)^{\rho}
\]
to conclude the proof.
\end{proof}
For $\ep \in [0,1]$, one has
\[
\ep^\rho(1-\ep)^{1-\rho} + \ep^{1-\rho}(1-\ep)^{\rho} = 1 - (\ep^\rho - (1-\ep)^\rho)(\ep^{1-\rho} - (1-\ep)^{-1-\rho}) \leq 1
\]
with equality if and only if $\ep = 1/2$, so that each term in the upper-bound reduces the error unless the corresponding base classifier does not perform better than pure chance. The parameter $\rho$ determines the level at which one increases the importance of misclassified examples for the next step. Let $\tilde S_W^+(j)$ and $\tilde S^-_W(j)$ denote the expressions in \cref{eq:sw.1,eq:sw.2} with $w_k(j)$ replaced by $w_k(j+1)$. 
Then, in the case when the base classifiers are binary, ensuring that $|y_k - f_j(x_k)|/2 = \bfone_{y_k\neq f_j(x_k)}$, one can easily check that $\tilde S_W^+(j)/\tilde S^-_W(j) = (S_W^+(j)/ S^-_W(j))^{1-\rho}$. So, the ratio is (of course) unchanged if $\rho = 0$, and pushed to a pure chance level if $\rho = 1$. 
We provide below an interpretation of boosting as a greedy optimization procedure that will lead to the value $\rho=1/2$.



\subsection{Adaboost and greedy gradient descent}
We here restrict to the case of binary base classifiers and denote their linear combination  by
\[
h(x) = \sum_{j=1}^M \alpha_j f_j(x).
\]
Whether an observation $x$ is correctly classified in the true class $y$ is associated to the sign of the product $yh(x)$, but the value of this product also has an
important interpretation, since, when it is positive, it can be thought of as a margin with which
$x$ is correctly classified. 

Assume that the function $F$ is evaluated, not only on the basis of
its classification error, but also based on this margin,
using a loss function of the kind
\begin{equation}
\label{eq:boosting.margin}
\Psi(h) = \sum_{k=1}^N \psi(y_kh(x_k))
\end{equation}
where $\psi$ is decreasing. The boosting algorithm can then be interpreted as an classifier which
incrementally improves this objective function.

Let, for $j<M$,
$$
h^{(j)} = \sum_{q=1}^j \al_q f_q\,.
$$
The next combination $h^{(j+1)}$  is equal to $h^{(j)} + \al_{j+1}
f_{j+1}$, and we now consider the problem of minimizing, with respect to   
$f_{j+1}$ and $\al_{j+1}$, the function $\Psi(h^{(j+1)})$, without modifying the previous classifiers (i.e., performing a greedy optimization). So, we want to minimize, with respect
to the base classifier $\tilde f$ and to $\al \geq 0$, the function
$$
U(\al, \tilde f) = \sum_{k=1}^N \psi\left(y_kh^{(j)}(x_k) + \al y_k \tilde
f(x_k)\right)
$$

Using the fact that $\tilde f$ is a binary classifier, this can be written
\begin{align}
U(\al, \tilde f) =& \sum_{k=1}^N \psi(y_kh^{(j)}(x_k) + \al) \bfone_{y_k
= \tilde f(x_k)} + \sum_{k=1}^N \psi(y_kh^{(j)}(x_k) - \al) \bfone_{y_k
\neq \tilde f(x_k)}
\label{eq:boosting.uu}
\\
\nonumber
=& \sum_{k=1}^N (\psi(y_kh^{(j)}(x_k) - \al) - \psi(y_kh^{(j)}(x_k) + \al)) \bfone_{y_k
\neq \tilde f(x_k)} \\
\nonumber
&+ \sum_{k=1}^N \psi(y_kh^{(j)}(x_k) + \al).
\end{align}
This shows that $\alpha$ and $\tilde f$ have  inter-dependent optimality conditions.
For a given $\al$, the best classifier $\tilde f$ must
minimize a weighted empirical error with non-negative weights (since $\psi$ is decreasing)
$$ w_k  = \psi(y_kh^{(j)}(x_k) - \al) - \psi(y_kh^{(j)}(x_k) + \al).$$
Given $\tilde{f}$, $\alpha$ must minimize the expression in \cref{eq:boosting.uu}. One can use an alternative minimization procedure to optimize both $\tilde f$ (as a weighted basic classifier) and $\alpha$. However, for the special choice $\psi(t) = e^{-t}$, this optimization turns out to only require one step.

In this case, we have
\begin{align*}
U(\al, \tilde f) &= \sum_{k=1}^N (e^{\al} - e^{-\al}) e^{-y_kh^{(j)}(x_k)} \bfone_{y_k
\neq \tilde f(x_k)} 
+ e^{-\al} \sum_{k=1}^N e^{-y_kh^{(j)}(x_k)}\\
&=e^{-\al^{(j)}} (e^{\al} - e^{-\al})  \sum_{k=1}^N w_k(j) \bfone_{y_k
\neq \tilde f(x_k)} 
+ e^{-\al^{(j)}} e^{-\al} \sum_{k=1}^N w_k(j)\\
\end{align*}
with 
$w_k(j+1) = e^{\al^{(j)}-y_kh^{(j)}(x_k)} $
and $\al^{(j)} = \al_1 + \cdots + \al_j$. 
This shows that $\tilde{f}$ should minimize
\[
\sum_{k=1}^N w_k(j+1)  \bfone_{y_k \neq \tilde f(x_k)}.
\]
We note that 
\[
w_k(j+1) = w_k(j) e^{\al_j(1-y_k f_{j}(x_k))} = w_k(j) e^{ \al_j |y_k - f_k(x_k)|},
\]
which is identical to the weight updates in algorithm \cref{alg:boosting} (this is the reason why the term $\alpha^{(j)}$ was introduced in the computation). The new value of $\alpha$ must minimize (using the notation of \cref{alg:boosting})
\[
e^{-\al}S_W^+(j)  + e^{\al} S_W^-(j),
\]
which yields $\alpha = \frac12 \log S_W^+(j)/S_W^-(j)$. This is the value $\alpha_{j+1}$ in \cref{alg:boosting} with $\rho=1/2$.

\section{Gradient boosting and regression}
\subsection{Notation}
The boosting idea, and in particular its interpretation as a greedy gradient procedure,  can be extended to non-linear regression problems \citep{friedman2001greedy}. 
Let us denote by $\CF_0$ the set of base predictors, therefore functions from $\CR_X= \mR^d$ to $ \CR_Y = \mR^q$, since we are considering regression problems. The final predictor is  a linear combination
\[
F(x) = \sum_{j=1}^M \alpha_j f_j(x)
\]
with $\alpha_1, \ldots, \alpha_M\in \mR$ and $f_1, \ldots, f_M\in \CF_0$. Note that the 
the coefficients  $\al_j$ are redundant when the class $\CF_0$ is invariant by multiplication by a scalar. Replacing if needed $\CF_0$ by $\defset{f = \alpha g, \alpha\in \mR, g\in \CF_0}$, we will assume that this property holds and therefore remove the coefficients $\alpha_j$ from the problem.

In accordance  with the principle  of performing greedy searches, we let 
\[
F^{(j)}(x) = \sum_{q=1}^j  f_q(x),
\]
and consider the problem of minimizing over $f\in \CF_0$,
\[
U( f) = \sum_{k=1}^N r(y_k, F^{(j)}(x_k) + f(x_k)),
\]
where $T = (x_1, y_1, \ldots, x_N, y_N)$ is the training data and $r$ is the loss function.

\subsection{Translation-invariant loss}
In the case, which is frequent in regression, when $r(y,y')$ only depends on $y-y'$, the problem is equivalent to minimizing
\[
U( f) = \sum_{k=1}^N r(y_k- F^{(j)}(x_k), f(x_k)),
\]
i.e., to let $f_{j+1}$ be the optimal predictor (in $\CF_0$ and for the loss $r$) of the residuals $y_k^{(j)} = y_k- F^{(j)}(x_k)$. In this case, this provides a conceptually very simple algorithm.
\begin{algorithm}[Gradient boosting for regression with translation-invariant loss]
\begin{enumerate}[label=$\bullet$]
\item Let $T = (x_1, y_1, \ldots, x_N, y_N)$ be a training set and $r$ a loss function such that $r(y,y')$ only depends on $y-y'$.
\item Let $\CF_0$ be a function class such that $f\in\CF_0\Rightarrow \alpha f\in \CF_0$ for all $\al\in \mR$.
\item Select an integer $M>0$ and let $F^{(0)} = 0$, $y_k^{(0)} = y_k$, $k=1, \ldots, N$.
\item For $j=1, \ldots, M$:
\begin{enumerate}[wide=0.5cm]
\item Find the optimal predictor $f_{j}\in \CF_0$ for the training set $(x_1, y_1^{(j-1)}, \ldots, x_N, y_N^{(j-1)})$.
\item Let $y_k^{(j)} = y_{k}^{(j-1)} - f_j(x_k)$ 
\end{enumerate}
\item Return $F = \sum_{k=1}^M f_j$.
\end{enumerate}

\end{algorithm}

\begin{remark}
Obviously, the class $\CF_0$ should not be a linear class for the boosting algorithm to have any effect. Indeed, if $f,f'\in\CF_0$ implies $f+f'\in\CF_0$, no improvement could be made to the predictor after the first step. 
\end{remark}

A successful example of this algorithm uses regression trees as base predictors. Recall that the functions 
output by such trees take the form
\[
f(x) = \sum_{A \in \CC} w_A \bfone_{x\in A}
\]
where $\CC$ is a finite partition of $\mR^d$. Each set in the partition is specified by the value taken by a finite number of binary features (denoted by $\gamma$ in our discussion of prediction trees)  and the maximal number of such features is the depth of the tree. We assume that the set $\Gamma$ of binary features is shared by all regression trees in $\CF_0$, and that the depth of these trees is bounded by a fixed constant. These restrictions prevent $\CF_0$ from forming a linear class.\footnote{If $f$ and $g$ are representable as trees, $f+g$ can be represented as a tree whose depth is the sum as those of the original trees,  simply by inserting copies of $g$ below each leaf of $f$.} Note that the maximal depth of tree learnable from a finite training set is always bounded, since such trees cannot have more nodes than the size of the training set (but one may want to restrict the maximal depth of base predictors to be way less than $N$).

\subsection{General loss functions}

We now consider situations in which the loss function is not necessarily a function of the difference between true and predicted output. We are still interested in the problem of minimizing $U(f)$, but we now approximate this problem using the first-order expansion
\[
U(f) = \sum_{k=1}^N r(y_k, F^{(j)}(x_k)) + \sum_{k=1}^N \prt_2 r(y_k, F^{(j)}(x_k))^T f(x_k) + o(f) ,
\]
where $\prt_2 r$ denotes the derivative of $r$ with respect to its second variable. This suggests (similarly to gradient descent) to choose $f$ such that $f(x_k) = -\alpha \prt_2 r(y_k, F^{(j)}(x_k))$ for some $\alpha>0$ and all $k=1, \ldots, N$. However, such an $f$ may not exist in the class $\CF_0$, and the next best choice is to pick $f = \alpha \tilde f$ with $\tilde f$ minimizing
\[
\sum_{k=1}^N |\tilde f(x_k) + \prt_2 r(y_k, F^{(j)}(x_k))|^2
\] 
over all $\tilde f\in \CF_0$. This is similar to projected gradient descent in optimization, and $\alpha$ such that $f = \alpha \tilde f$ should  minimize  
\[
\sum_{k=1}^N r(y_k, F^{(j)}(x_k) + \alpha \tilde f(x_k) ).
\]
This provides a generic ``gradient boosting'' algorithm \citep{friedman2001greedy}, summarized below.
\begin{algorithm}[Gradient boosting]
\label{alg:grad.boost}
\begin{enumerate}[label=$\bullet$]
\item Let $T = (x_1, y_1, \ldots, x_N, y_N)$ be a training set and $r$ a differentiable loss function.
\item Let $\CF_0$ be a function class such that $f\in\CF_0\Rightarrow \alpha f\in \CF_0$ for all $\al\in \mR$.
\item Select an integer $M>0$ and let $F^{(0)} = 0$.
\item For $j=1, \ldots, M$:
\begin{enumerate}[wide=0.5cm]
\item Find  $\tilde f_{j}\in \CF_0$ minimizing
\[
\sum_{k=1}^N |\tilde f(x_k) + \prt_2 r(y_k, F^{(j-1)}(x_k))|^2
\] 
over all $\tilde f\in \CF_0$.
\item Let $f_j = \alpha_j \tilde{f}_j$ where $\alpha_j$ minimizes
\[
\sum_{k=1}^N r(y_k, F^{(j-1)}(x_k) + \alpha \tilde f_j(x_k) ).
\]
\item Let $F^{(j)} = F^{(j-1)} + f_j$.
\end{enumerate}
\item Return $F = F^{(M)}$.
\end{enumerate}
\end{algorithm}

\begin{remark}
Importantly, the fact that $\CF_0$ is stable by scalar multiplication implies that the function $\tilde f_j$ satisfies 
\[
\sum_{k=1}^N \tilde f(x_k)^T \prt_2 r(y_k, F^{(j-1)}(x_k)) \leq 0, 
\] 
that is, excepted in the unlikely case in which the above sum is zero, it is a direction of descent for the function $U$ (because one could otherwise replace $\tilde f_j$ by $-\tilde f_j$ and improve the approximation of the gradient).
%
%
%
%

\end{remark}

\subsection{Return to classification}
A slight modification of this algorithm may also be applied to classification, provided that the classifier $f$ is obtained by learning the conditional distribution, denoted $g \mapsto p(g|x)$, of the output variable (assumed to take values in a finite set $\CR_Y$) given the input (assumed to take values in $\CR_X = \mR^d$). 

Our goal is to estimate an unknown target conditional distribution, $\mu$, therefore taking the form $\mu(g|x)$ for $g\in \CR_Y$ and $x\in\mR^d$. We assume that a family $\mu_k, k=1, \ldots, N$ of distributions on the set $\CR_Y$ is observed, where each $\mu_k$ is assumed to be an approximation of the unknown $\mu(\cdot|x_k)$ (typically, $\mu_k(g) = \bfone_{g=y_k}$, i.e., $\mu_k = \delta_{y_k}$). The risk function must  take the form $r(\mu, \mu')$ where $\mu, \mu'\in \CS(\CR_Y)$, the set of probability distributions on $\CR_Y$. We will work with 
\[
r(\mu, \mu') = - \sum_{g\in\CR_Y} \mu(g) \log\mu'(g).
\]
One can note that 
\[
r(\mu, \mu') = \KL(\mu\|\mu') + r(\mu, \mu),
\]
which is therefore minimal when $\mu'=\mu$.
Moreover, in the special case $\mu_k = \delta_{y_k}$, the empirical risk is 
\[
\hat R(p) = \sum_{k=1}^N r(\mu_k, p(\cdot|x_k)) = - \sum_{k=1}^N \log p(y_k|x_k),
\]
so that minimizing it is equivalent to maximizing the conditional likelihood that was used for  logistic regression.

Before applying the previous algorithm, one must address the issue that probability distributions do not form a vector space, and cannot be added to form new probability distributions. In \citet{friedman2001greedy,htf03}, it is suggested to use the representation, which can be associated with any function $F: (g,x) \mapsto F(g|x)\in \mR$,
\[
p_F(g|x) = \frac{e^{F(g|x)}}{\sum_{h\in\CR_Y} e^{F(h|x)}}.
\]
Because the representation if not unique ($p_F = p_{F'}$ if $F-F'$ only depends on $x$), we will require in addition that
\[
\sum_{h\in\CR_Y} F(h|x) = 0
\]
for all $x\in \mR^d$. The space formed by such functions $F$ is now linear, and we can consider the empirical risk
\[
\hat R(F) = -\sum_{k=1}^N \sum_{g\in \CR_Y} \mu_k(g)\log p_F(g|x_k) = -\sum_{k=1}^N \sum_{g\in \CR_Y} \mu_k(g)F(g|x_k) + \sum_{k=1}^N \log\left(\sum_{g\in \CR_Y}  e^{F(g|x_k)}\right).
\]

One can evaluate the derivative of this risk with respect to a change on $F(g|x_k)$, and a short computation gives
\[
\frac{\partial R}{\partial F(g|x_k)} = - \sum_{k=1}^N (\mu_k(g) - p_F(g|x_k)).
\]

\bigskip

Now assume that a basic space $\CF_0$ of functions $f: (g,x) \mapsto f(g|x)$ is chosen, such that all function in $\CF_0$ satisfy
\[
\sum_{g\in \CR_Y} f(g|x) = 0
\]
for all $x\in \mR^d$.
The gradient boosting algorithm then requires to minimize (in Step (1)):
\[
\sum_{k=1}^N \sum_{g\in \CR_Y} (\mu_k(g) - p_{F^{(j-1)}}(g|x_k) - \tilde f(g|x_k))^2
\]
with respect to all functions $\tilde f \in \CF_0$. Given the optimal $\tilde f_j$, the next step requires to minimize, with respect to $\alpha \in \mR$:
\[
-\alpha \sum_{k=1}^N \sum_{g\in \CR_Y} \mu_k(g)\tilde f_j(g|x_k) + \sum_{k=1}^N \log\left(\sum_{g\in \CR_Y}  e^{F^{(j-1)}(g|x_k) + \alpha \tilde f_j(g|x_k)}\right).
\]
This is a scalar convex problem that can be solved, e.g., using gradient descent.

\subsection{Gradient tree boosting}

We now specialize to the situation in which the set $\CF_0$ contains regression trees. In this situation, the general algorithm can be improved by taking advantage of the fact that the predictors returned by such trees are piecewise constant functions, where the regions of constancy are associated with partitions $\CC$ of $\mR^d$ defined by the leaves of the trees. In particular, $\tilde{f}_j(x)$ in Step (1) takes the form
\[
\tilde f_j(g|x) = \sum_{A\in \CC}^J \tilde f_{j,A}(g) \bfone_{x\in A}.
\]
The final $f$ at Step (2) should therefore take the form
\[
\sum_{A\in\CC} \alpha \tilde f_{j,A}(g) \bfone_{x\in A}
\]
but not much additional complexity is introduced by freely optimizing the values of $f_j$ on $A$, that is, by looking at $f$ in the form 
\[
\sum_{A\in\CC} f_{j,A}(g) \bfone_{x\in A}
\]
where the values $f_{j,A}(g)$ optimize the empirical risk. This risk becomes
\[
- \sum_{k=1}^N \sum_{A\in\CC} \sum_{g\in \CR_Y} \mu_k(g) f_{j,A}(g) \bfone_{x_k\in A} + \sum_{k=1}^N \sum_{A\in\CC} \log\left(\sum_{g\in \CR_Y}  e^{F^{(j-1)}(g|x_k) +   f_{j,A}(g)}\right) \bfone_{x_k\in A}.
\]
The values $f_{j,A}(g), g\in \CR_Y$ can therefore be optimized separately, minimizing
\[
- \sum_{k=1: x_k\in A}  \sum_{g\in \CR_Y} \mu_k(g) f_{j,A}(g)  + \sum_{k: x_k\in A} \log\left(\sum_{g\in \CR_Y}  e^{F^{(j-1)}(g|x_k) +   f_{j,A}(g)}\right) \bfone_{x_k\in A}.
\]
This is still a convex program, which has to be run at every leaf of the optimized tree. If computing time is limited (or for large-scale problems), the determination of $f_{j,A}(g)$ may be restricted to one step of gradient descent starting at $f_{j,A} = 0$. A simple computation indeed shows that the first derivative of the function above with respect to $f_{j,A}(g)$ is
\[
a_A(g) = - \sum_{k: x_k\in A} (\mu_k(g) - p_F(g|x_k)).
\]
The derivative of this expression with respect to $f_{j,A}(g)$ (for the same $g$) is
\[
b_A(g) = \sum_{k: x_k\in A} p_F(g|x_k)(1-p_F(g|x_k)).
\]
The off-diagonal terms in the second derivative are, for $g\neq h$,
\[
- \sum_{k: x_k\in A} p_F(g|x_k)p_F(h|x_k).
\]
In \citet{friedman2000additive}, it is suggested to use an approximate Newton step, where the off-diagonal terms in the second derivative are neglected. This corresponds to minimizing 
\[
\sum_{g\in \CR_Y} a_A(g) f_{j,A}(g) + \frac12 \sum_{g\in\CR_Y}  b_A(g) f_{j,A}(g)^2.
\]
The solution is (introducing a Lagrange multiplier for the constraint $\sum_g f_{j,A}(g) = 0$)
\[
f_{j,A}(g) = -  \frac{a_A(g) - \lambda}{b_A(g)}
\]
with
\[
\lambda = \frac{\sum_{g\in\CR_Y} a_A(g)/b_A(g)}{\sum_{g\in\CR_Y} 1/b_A(g)}.
\]
A small value $\epsilon$ can be added to $b_A$ to avoid divisions by zero. We refer the reader to \citet{friedman2000additive,friedman2001greedy,htf03} for several variations on this basic idea. Note that an approximate but highly efficient implementation of boosted trees, called XGBoost, has been developed in \citet{chen2016xgboost}.


\chapter[Neural Nets]{Iterated Compositions of  Functions and Neural Nets}
\label{chap:neural.nets}

\section{First definitions}
We now discuss a class of methods in which the predictor $f$ is built using iterated compositions, with a main application to neural nets.  We will structure these models using  directed acyclic graphs (DAG). These graphs are  composed with a set of vertexes (or nodes) $\mathcal V = \{0, \ldots, m+1\}$ and a collection $\mathcal L$  of directed edges $i\to j$ between some vertexes.  If  an edge exists between $i$ and $j$, one says that $i$ is a parent of $j$ and $j$ a child of $i$ and  we will use the notation $\pa{i}$ (resp. $\cl{i}$) to denote the set of parents (resp. children) of $i$. The graphs we consider must satisfy the following conditions:
\begin{enumerate}[label = (\roman*),wide=0.5cm]
\item No index is a descendant of itself, i.e., that the graph is acyclic.
\item The only index without parent is $i=0$ and the only one without children in $i=m+1$.
\end{enumerate}

To each node $i$ in the graph, one associates a dimension $d_i$ and a variable $z_i \in \mR^{d_i}$. 
The root node variable,   $z_0 = x$, is the input  and $z_{m+1}$ is the output.
One also associates to each node $i\neq 0$ a function $\psi_i$ defined on the product space $\bigotimes_{j\in\pa{i}} \mR^{d_j}$ and taking values in $\mR^{d_i}$.
The input-output relation is then defined by the family of equations:
\[
z_i = \psi_i(z_{\pa{i}})
\]
where  $z_{\pa{i}} = (z_j, j\in\pa{i})$.
Since there is only one root  and one terminal node, these iterations implement a relationship $y = z_{m+1} = f(x)$, with $z_0=x$. We will refer to the $z_1, \ldots, z_m$ as the {\em latent variables} of the network.

Each function $\psi_i$ is furthermore parametrized by an $s_i$-dimensional vector  $w_i \in \mR^{s_i}$, so that we will write 
\[
z_i = \psi_i(z_{\pa{i}}; w_i).
\]
 We let $\CW$ denote the vector containing all parameters $w_1, \ldots, w_{m+1}$, which therefore has dimension $s = s_1+\cdots + s_{m+1}$.
 The network function $f$ is then parametrized by $\CW$ and we will write $y = f(x;\CW)$.
 
\section{Neural nets}
\subsection{Transitions}
Most neural networks iterate functions taking the form 
\[
\psi_i(z;w) = \rho(bz + \beta_0), z\in\mR^{d_j}
\]
where $b$ is a $d_{i}\times (\sum_{j\in \pa{i}} d_j)$ matrix and $\beta_0\in \mR^{d_{i}}$ (so that $w = (b, \beta_0)$ is $s_i = d_{i}(1+ \sum_{j\in \pa{i}} d_j)$-dimensional);
$\rho$ is defined on and takes values in $\mR$, and we make the abuse of notation, for any $d$ and $u\in \mR^d$
\[
\rho(u) = \begin{pmatrix} \rho(u^{(1)}) \\\vdots\\ \rho(u^{(d)})\end{pmatrix}.
\]

 The most popular choice for $\rho$ is the positive part, or ReLU (for rectified linear unit), given by $\rho(t) = \max(t,0)$.
 Other common choices are $\rho(t) =  1/(1+e^{-t})$ (sigmoid function), or $\rho(t) = \mathrm{tanh} (z)$.

Residual neural networks (or ResNets \cite{he2016deep}) are discussed in \cref{sec:continuous.nn}. They iterate transitions between inputs and outputs of same dimension, taking 
\begin{equation}
\label{eq:resnet}
z_{i+1} = z_i + \psi(z_i;w) .
\end{equation}

\subsection{Output}
The last node of the graph provides the prediction, $y$. Its expression depends on the type of predictor that is learned
\begin{enumerate}[label=$\bullet$, wide=0.5cm]
\item  For regression, $y$ can be chosen as an affine function of is its parents: $z_{m+1} = b z_{\pa{m+1}} + a_0$.
\item For classification, one can also use a linear model  $z_{m+1} = b z_{\pa{m+1}} + a_0$ where $z_{m+1}$ is $q$-dimensional and let the classification be  $\argmax(\pe{z_{m+1}} i, i=1, \ldots, q)$. Alternatively, one  may use a ``softmax'' transformation, taking 
\[
\pe{z_{m+1}} i = \frac{e^{\pe{\zeta_{m+1}}i}}{\sum_{j=1}^q e^{ \pe{\zeta_{m+1}} j }}
\] 
with  $\zeta_{m+1} = b z_{\pa{m+1}} + a_0$.
\end{enumerate}

\subsection{Image data}
 Neural networks have achieved top performance when working with organized structures such as images. 
 A typical problem in this setting is to categorize the content of the image, i.e., return a categorical variable naming its principal element(s).
 Other applications include facial recognition or identification.
 In this case, the transition function can take advantage of the 2D structure, with some special terminology.

Instead of speaking of the total dimension, say, $d$, of the considered variables, writing $z = (z^{(1)}, \ldots, z^{(d)})$, images are better represented with three indices $z(u, v, \lambda)$ where
 $u=1, \ldots, U$ and $U$ is the width of the image,  $v=1, \ldots, V$ and $V$ is the height of the image,  $\la=1, \ldots, \Lambda$ and $\Lambda$ is the depth of the image. (With this notation  $d =UV\Lambda$.)
  Typical images  have length and width of several hundred pixels, and depth $\La=3$ for the three color channels.
 This three-dimensional structure is conserved also for latent variables, with different dimensions. 
 Deep neural networks   often combine compression in width and height with expansion in depth while transitioning  from input to output.

The linear transformation $b$ mapping one layer with dimensions $U_i, V_i, \La_i$ to another with dimensions $U_{i+1}, V_{i+1}, \La_{i+1}$ is then preferably seen as a collection of numbers: $b(u',v',\la', u,v, \la)$ so that the transition from $z_i$ to $z_{i+1}$ is given by
\[
z_{i+1}(u', v', \la') = \rho\left(\be_0(u', v', \la') 
+ \sum_{u=1}^{U_i}\sum_{v=1}^{V_i}\sum_{\lambda=1}^{\La_i} b(u', v', \la', u,v,\la)  z_i(u,v,\la)\right).
\]

For images, it is often preferable to use convolutional transitions, providing convolutional neural networks \citep{lecun1989backpropagation,lecun1995convolutional}, or CNNs.
 If $U_i=U_{i+1}$ and $V_i=V_{i+1}$, such a transition requires that  $b(u', v', \la', u,v,\la)$ only depends on $\la$, $\la'$ and on the differences $u'-u$ and $v-v'$.
In general, one also requires that  $b(u', v', \la', u,v,\la)$ is non-zero only if $|u'-u|$ and $|v'-v|$ are both less than a constant, typically a small number. Also, there is generally little computation across depths: each output at depth $\la'$ only uses values from a single input depth. These restrictions obviously reduce dramatically the number of free parameters involved in the transition.

After one or a few convolutions, the dimension is often reduced by a ``pooling'' operation, dividing the image into small non-overlapping windows and replacing each such window by a single value, either the max (max-pooling) or the average.

\section{Geometry}
In addition to the transitions between latent variables and resulting changes of dimension, the structure of the DAG defining the network is an important element in the design of a neural net.
 The simplest choice is a purely linear structure (as shown in Figure \ref{fig:net.1}), as was, for example, used for image categorization in \cite{krizhevsky2017imagenet}. 
 \begin{figure}
	\begin{center}
		\begin{tikzpicture}[roundnode/.style={rectangle, draw=green!60, fill=green!5, thin, minimum size=20mm}, squarednode/.style={rectangle, draw=black!60, fill=gray!5, very thick},]
		\node[squarednode, minimum height=3cm] (L1) {$x$} ;
		\node[squarednode, minimum height= 2cm, minimum width = 1cm] (L2) [right=.5cm of L1] {$z_1$} ;
		\node[] (dots) [right=.5cm of L2] {$\ldots$} ;
		\node[squarednode] (Lm) [right=.5cm of dots, minimum height= .5cm, minimum width = 4cm] {$z_m$} ;
		\node[right=.5cm of Lm] (O) {$y$} ;
		\draw[blue, thick,->] (L1.east) -- (L2.west) node[midway, above] {};
		\draw[blue, thick,->] (L2.east) -- (dots.west) node[midway, above] {};
		\draw[blue, thick,->] (dots.east) -- (Lm.west) node[midway, above] {};
		\draw[blue, thick,->] (Lm.east) -- (O.west) node[midway, above] {};
		\end{tikzpicture}
		
		\caption{ \label{fig:net.1}
		Linear net with increasing layer depths and decreasing layer width.}
		\end{center}
	\end{figure}	

More complex architectures have been introduced in recent years. Their design is in a large part heuristic and based on an analysis of the kind of computation that should be done in the network to perform a particular task. For example, an architecture used for image segmentation in summarized in \cref{fig:net.2}. 

\begin{figure}

\centering
		\begin{tikzpicture}[roundnode/.style={rectangle, draw=green!60, fill=green!5, thin, minimum size=20mm}, squarednode/.style={rectangle, draw=black!60, fill=gray!5, very thick},]
		\node[squarednode, minimum height=3cm] (L1) {$x$} ;
		\node[squarednode, minimum height= 2cm, minimum width = 1cm] (L2) [below right=.1cm and .1cm of L1] {$z_1$} ;
		\node[squarednode] (Lm) [below right=.5cm and 1cm of L2, minimum height= .5cm, minimum width = 4cm] {$z_m$} ;
		\node[squarednode, minimum height= 2cm, minimum width = 1cm] (L2m-1) [above right=.5cm and 1cm of Lm] {$z_{2m-1}$} ;
		\node[squarednode, minimum height=3cm] (L2m)[above right=.1cm and .1cm of L2m-1] {$y$} ;
		\draw[blue, thick, ->] (L1.east) -- (L2.north) node[midway, above] {};
		\draw[blue, thick, dashed, -] (L2.south east) -- (Lm.north west) node[midway, above] {};
		\draw[blue, thick, dashed, -] (Lm.north east) -- (L2m-1.south west) node[midway, above] {};
		\draw[blue, thick, ->] (L2m-1.north) -- (L2m.west) node[midway, above] {};
		\draw[blue, thick, ->] (L1.east) -- (L2m.west) node[midway, above] {};
		\draw[blue, thick, ->] (L2.east) -- (L2m-1.west) node[midway, above] {};
\end{tikzpicture}
		\caption{ \label{fig:net.2} A sketch of the U-net architecture designed for image segmentation \citep{ronneberger2015u-net}.}
	\end{figure}	

\begin{remark}
\label{rem:attention}
An important feature of neural nets is their modularity, since ``simple'' architectures can be combined (e.g., by placing the output of a network as input of another one) and form a more complex network that still follows the basic structure defined above. One example of such a building block is the ``attention  module.'' Such a module takes as input three sequences of vectors of equal size, say $\pe z q = (\pe {z_k}q)$, $\pe z c=(\pe {z_k}c)$, $\pe z v =(\pe{z_k}v)$, for $k=1, \ldots, n$, which are are typically outputs of previous modules. All three may be identical (self-attention modules), or distinct (encoder-decoder modules in \citep{vaswani2017attention} have $\pe zc = \pe zv \neq \pe zq$). The input vectors are separately linearly transformed into ``query,'' ``key,'' and ``value'' vectors, $q_k = W_q \pe {z_k}q$, $c_k=W_c \pe {z_k}c$ and $v_k = W_v \pe{z_k}v$ (where $W_q, W_c, W_v$ are learned, and $W_q$ and $W_c$ have the same number of rows, say, $d$) and the output of the module is also a sequence of $n$ vectors given by
\[
\pe {z_k}o = \frac{\sum_{l} e^{\tau a(q_k, c_l)} v_l}{\sum_{l} e^{\tau a(q_k, c_l)}}  
\]
where $a(q,c)$ measures the affinity between $q$ and $c$ (e.g, $a(q,c) = q^T c$) and $\tau$ is a fixed constant (e.g., $1/\sqrt d$).  
These attention modules are fundamental components of ``transformer networks'' \citep{vaswani2017attention}, that are used, among other tasks, in natural language processing and large language models. 
\end{remark}

\section{Objective function}
\subsection{Definitions}
 We now return to the general form of  the problem, with variables $z_0, \ldots, z_{m+1}$ satisfying 
\[
z_i = \psi_i(z_{\pa{i}}; w_i).
\]
 Let $T = (x_1, y_1, \ldots, x_N, y_N)$ denote the training data.

For regression problems, the objective function minimized by the algorithm is typically the empirical risk, the simplest choice being the mean square error, which gives
\[
F(\CW) = \frac1N \sum_{k=1}^N |y_k - z_{k,m+1}(\CW)|^2.
\]
with $z_{k;m+1}(\CW) = f(x_k; \CW)$.

For classification, with the dimension of the output variable equal to the number of classes and the decision based on the largest coordinate, one can take (letting  $z_{k,m+1}(i;\CW)$ denote the $i$th coordinate of $z_{k,m+1}(\CW)$):
\[
F(\CW) = \frac1N \sum_{k=1}^N \left(-z_{k,m+1}(y_k;\CW) + \log \Big(\sum_{i=1}^q \exp(z_{k,m+1}(i;\CW))\Big)\right).
\]
 This  objective function is similar to that minimized in logistic regression.

\subsection{Differential}
\label{sec:nn.der}
\paragraph{General computation.}
 The computation of the differential of $F$ with respect to $\CW$ may look daunting, but it has actually a simple structure captured by the back-propagation algorithm.
Even if programming this algorithm can often be avoided by using an automatic differentiation software, it is important to understand how it works, and why the implementation of gradient-descent algorithms remains feasible. 

Consider the general situation of minimizing a function $G(\CW, \boldsymbol z)$ over $\CW\in \mR^s$ and $\boldsymbol z \in \mR^r$, subject to a constraint $\gamma(\CW, \boldsymbol z) = 0$ where $\gamma$ is defined on $\mR^s \times \mR^r$ and takes values in $\mR^r$ (here, it is important that the number of constraints is equal to the dimension of $z$). We will denote below by $\prt_\CW$ and $\prt_{\boldsymbol z}$ the derivatives of these functions with respect to the multi-dimensional variables $\CW$ and $\boldsymbol z$. We make the assumptions that $\prt_{\boldsymbol z}\gamma$, which is an $r\times r$ matrix, is invertible, and that 
the constraints can be solved to express $\boldsymbol z$ as a function of $\CW$, that we will denote $\boldsymbol Z(\CW)$. 

This allows us to define the function $F(\CW) = G(\CW, \boldsymbol Z(\CW))$ and we want to compute the gradient of $F$. (The function $F$ in the previous section satisfies these assumptions, with $\bfz = (z_{kj}, k=1, \ldots, N, j=1, \ldots, m+1)$ ). 
Taking $h\in \mR^s$, we have
\[
dF(\CW) h = \prt_\CW G(\CW, \bfZ(\CW)) h + \prt_{\boldsymbol z} G(\CW, \bfZ(\CW))\, d\boldsymbol Z(\CW) h.
\]
Moreover, since $\gamma(\CW, \boldsymbol Z(\CW)) = 0$ by definition of $\boldsymbol Z$, we have
\[
\prt_{\CW} \gamma(\CW, \boldsymbol Z(\CW)) h + \prt_{\boldsymbol z}\gamma(\CW, \boldsymbol Z(\CW))\, d\boldsymbol Z(\CW) h = 0, 
\]
so that
\[
dF(\CW) h = \prt_\CW G(\CW, \boldsymbol Z(\CW)) h - \prt_{\boldsymbol z} G(\CW, \boldsymbol Z(\CW)) \, \prt_{\boldsymbol z}\gamma(\CW, \boldsymbol Z(\CW))^{-1}\,  \prt_{\CW} \gamma(\CW, \boldsymbol Z(\CW)) h.
\]

Let $\boldsymbol p\in \mR^r$ be the solution of the linear system
\begin{equation}
\label{eq:bp.costate}
\prt_{\boldsymbol z}\gamma(\CW, \boldsymbol Z(\CW))^T \boldsymbol p =  \prt_{\boldsymbol z} G(\CW, \boldsymbol Z(\CW))^T.
\end{equation}
Then,
\[
dF(\CW) h = (\prt_\CW G(\CW, \boldsymbol Z(\CW)) - \boldsymbol p^T \prt_{\CW} \gamma(\CW, \boldsymbol Z(\CW))) h
\]
or 
\begin{equation}
\label{eq:bp.gradient}
\nabla F = \prt_\CW G(\CW, \boldsymbol Z(\CW))^T - \prt_{\CW} \gamma(\CW, \boldsymbol Z(\CW))^T \boldsymbol p.
\end{equation}

Note that, introducing the ``Hamiltonian''
\[
\boldsymbol H(\boldsymbol p, \boldsymbol z, \CW) = \boldsymbol p^T \gamma(\CW, \boldsymbol z) - G(\CW, \boldsymbol z), 
\]
one can summarize the previous computation with the system
\[
\left\{
\begin{aligned}
&\prt_{\boldsymbol p} \boldsymbol H = 0\\
&\prt_{\boldsymbol z}\boldsymbol H = 0 \\
&\nabla F = - \prt_{\CW} \boldsymbol H^T.
\end{aligned}
\right.
\]
\bigskip

\paragraph{Application: back-propagation.}
In our case, we are minimizing a function of the form
\[
\boldsymbol G(\CW, \boldsymbol z_1, \ldots, \boldsymbol z_N) = \frac1N \sum_{k=1}^N r(y_k, z_{k, m+1})
\]
subject to constraints $z_{k,i+1} = \psi_i(z_{k, \pa{i}}; w_i)$, $i=0, \ldots, m$, $z_{k,0} = x_k$. We focus on one of the terms in the sum, therefore fixing $k$, that we will temporarily drop from the notation.

So, we evaluate the gradient of $G(\CW, \boldsymbol z) = r(y, z_{m+1})$ with $z_{i+1} = \psi_i(z_{\pa{i}}; w_i)$, $i=0, \ldots, m$, $z_0 = x$. With the notation of the previous paragraph, we take $\gamma = (\gamma_1, \ldots, \gamma_{m+1})$ with
\[
\gamma_i(\CW, \boldsymbol z) = \psi_{i}(z_{\pa{i}}; w_i) - z_i
\]
These constraints uniquely define $\boldsymbol z$ as a function of $\CW$, which was one of our assumptions. For the derivative, we have, for $\boldsymbol u= (u_1, \ldots, u_{m+1})\in \mR^r$ (with $r = d_1 + \cdots+d_{m+1}$, $u_i\in \mR^{d_i}$), and for $i=1, \ldots, m+1$  
\[
\prt_{\boldsymbol z}\gamma_i(\CW, \boldsymbol z) \boldsymbol u = \sum_{j \in \pa{i}} \prt_{z_j} \psi_i(z_{\pa{i}}; w_i) u_j - u_i
\]
Taking $\boldsymbol p = (p_1, \ldots, p_{m+1}) \in \mR^r$, we get
\begin{align*}
\boldsymbol p^T \prt_{\boldsymbol z}\gamma(\CW, \boldsymbol z) \boldsymbol u &= 
\sum_{i=1}^{m+1}\sum_{j \in \pa{i}} p_i^T\prt_{z_j} \psi_i(z_{\pa{i}}; w_i) u_j - \sum_{i=1}^{m+1} p_i^T u_{i} \\
&= 
\sum_{j=1}^{m+1}\sum_{i \in \cl{j}} p_i^T\prt_{z_j} \psi_i(z_{\pa{i}}; w_i) u_j - \sum_{j=1}^{m+1} p_j^T u_{j} .
\end{align*}
This allows us to identify $\prt_{\boldsymbol z}\gamma(\CW, \boldsymbol z)^T \boldsymbol p$ as the vector $\boldsymbol g = (g_1, \ldots, g_{m+1})$ with 
\[
g_{j} = \sum_{i \in \cl{j}} \prt_{z_j} \psi_i(z_{\pa{i}}; w_i)^T p_i - p_j.
\]
For $j=m+1$ (which has no children), we get $g_{m+1} = -p_{m+1}$, so that the equation $\prt_{\boldsymbol z} \gamma^T p = g$ can be solved recursively by taking $p_{m+1} = -g_{m+1}$ and propagating backward, with
\[
p_{j} = -g_j + \sum_{i \in \cl{j}} \prt_{z_j} \psi_i(z_{\pa{i}}; w_i)^T p_i
\]
for $j=m, \ldots, 1$. 

From \cref{eq:bp.costate}, we need to apply propagation with $g = \prt_{\boldsymbol z} G$. Since $G$ only depends on $z_{m+1}$, we have $g_{m+1} = \prt_{z_{m+1}} r(y, z_{m+1})$ and  $g_j=0$ for $j=1, \ldots, m$. The final computation of the gradient is given by \cref{eq:bp.gradient}, in which $\prt_{\CW} G = 0$ since $G$ does not depend on $\CW$. For the second term in the r.h.s. of \cref{eq:bp.gradient}, we have
\[
\prt_{\CW} \gamma_i = \prt_{w_i} \psi_i(z_{\pa{i}}, w_i),
\]
yielding $\prt_{\CW} \gamma^T p = (\zeta_1, \ldots, \zeta_{m})$ with
\[
\zeta_j = \prt_{w_j} \psi_j(z_{\pa{j}}, w_j)^T p_j.
\]

We can now formulate an algorithm that computes the gradient of $F$ with respect to $\CW$, reintroducing training data indexes in the notation.

\begin{algorithm}[Back-propagation]
\label{alg:backprop} 
Let $(x_1, y_1, \ldots, x_N, y_N)$ be the training set and $R_k(z) = r(y_k, z)$ so that 
\[
F(\CW) = \frac1N \sum_{k=1}^N R_k(z_{k, m+1}(\CW))
\]
 with $z_{k, m+1}(\CW) = f(x_k, \CW)$.
 Let $\CW$ be a family of weights. The following steps compute $\nabla F(\CW)$.
\begin{enumerate}[label=\arabic*., wide = 0.5cm]
\item For all $k=1, \ldots, N$ and all $i=1, \ldots, m+1$, compute $z_{k,i}(\CW)$ (forward computation through the network).
\item Initialize variables $p_{k, m+1} = -\nabla R_k(z_{k,m+1}(\CW))$, $k=1, \ldots, N$.
\item For all $k=1, \ldots, N$ and all $j=1, \ldots, m$, compute $p_{k, j}$ using iterations 
\[
p_{k,j} = \sum_{i\in \cl{j}} \prt_{z_j} \psi_i(z_{k,\pa{i}}, w_i)^Tp_{k,i}.
\]
\item Let 
\[
\nabla F(\CW) = -\frac1N \sum_{k=1}^N \sum_{i=1}^{m+1} D_i^T \prt_{w_i} \psi_i(z_{k,\pa{i}}, w_i)^T p_{k,i},
\]
 where $D_i$ is the $s_i\times s$ matrix such that $D_i h = h_i$.
\end{enumerate}
\end{algorithm}

\subsection{Complementary computations}

 The back-propagation algorithm requires the computation of the gradient of the costs $R_k$ and of the derivatives of the functions $\psi_i$, and this can generally be done in closed form, with relatively simple expressions.
 
 \begin{enumerate}[label=$\bullet$,wide=0.5cm]
\item If $R_k(z) = |y_k - z|^2$ (which is the typical choice for regression models) then $\nabla R_k(z) = 2 (z-y_k)$. 
\item In classification, with $R_k(z) = - z(y_k) + \log \Big(\sum_{i=1}^q \exp(\pe z i)\Big)$, one has
\[
\nabla R_k(z) = - u_{y_k} +   \frac{\exp(z)}{\sum_{i=1}^q \exp(\pe z i)}
\]
where $u_{y_k}\in \mR^d$ is the vector with 1 at position $y_k$ and zero elsewhere, and $\exp(z)$ is the vector with coordinates $\exp(\pe z i)$, $i=1, \ldots, d$.
\item For dense transition functions in the form $\psi(z; w) = \rho(bz+\beta_0)$ with $w = (\beta_0, b)$, then $\prt_z \psi(z, w) = \mathrm{diag}(\rho'(\beta_0 + bz)) b$ so that
\[
\prt_z \psi(z, w)^T p =  b^T \mathrm{diag}(\rho'(\beta_0 + bz))p
\]
\item
Similarly
\[
\prt_w \psi(z, w)^T p =  \left[ \mathrm{diag}(\rho'(\beta_0 + bz))p, \mathrm{diag}(\rho'(\beta_0 + bz))pz^T\right] \,. 
\]
\end{enumerate}
Note that neural network packages implement these functions (and more) automatically.
 
\section{Stochastic Gradient Descent}

\subsection{Mini-batches}
Fix $\ell \ll N$. Consider the set of $B_\ell$ of binary sequences $\xi = (\xi^1, \ldots, \xi^{N})$ such that $\xi^k \in \{0,1\}$ and $\sum_{k=1}^N \xi^k = \ell$.
Define 
\[
H(\CW, \boldsymbol \xi) = \nabla_\CW \left(\frac1\ell \sum_{k=1}^N \boldsymbol\xi^k r(y_k, f(x_k, \CW))\right) = \frac1\ell \sum_{k=1}^N \boldsymbol \xi^k \nabla_{\CW}\, r(y_k, f(x_k, \CW))
\]
where $\boldsymbol \xi$ follows the uniform distribution on $B_\ell$. Consider the stochastic approximation algorithm: 
\begin{equation}
\label{eq:nn.sgd}
\CW_{n+1} = \CW_n - \ga_{n+1}  H(\CW_n, \boldsymbol\xi_{n+1}).
\end{equation}
Because $E(\boldsymbol\xi^k) = \ell/N$, we have $E( H(\CW, \boldsymbol\xi)) = \nabla_\CW \CE_T(f(\cdot, \CW))$ and \eqref{eq:nn.sgd} provides a stochastic gradient descent algorithm to which the  discussion in \cref{sec:sgd}  applies. Such an approach is often referred to as ``mini-batch'' selection in the deep-learning literature, since it correspond to sampling $\ell$ examples from the training set without replacement and only computing the gradient of the empirical loss restricted to these examples.  

\subsection{Dropout}

Introduced for deep learning in \citet{srivastava2014dropout}, ``dropout'' is a learning para\-digm that brings additional robustness (and, maybe, reduces overfitting risks) to massively parametrized predictors.  

Assume that a random perturbation mechanism  of the model parameters has been designed. We will represent it using a random variable $\boldsymbol\eta$ (interpreted as noise) and a transformation $\CW' = \phi(\CW, \eta)$ describing how $\eta$ affects a given weight configuration $\CW$ to form a perturbed one $\CW'$. In order to shorten notation, we will write $\phi(\CW, \eta) = \eta\cdot \CW$, borrowing the notation for a group action from group theory. As a typical example, $\boldsymbol\eta$ can be chosen as a vector of Bernoulli random variables (therefore taking values in \{0,1\}), with same dimension as $\CW$ and one can simply let $\eta\cdot \CW = \eta\odot \CW$ be the pointwise multiplication of the two vectors. This corresponds to replacing some of the parameters by zero (``dropping them out'') while keeping the others unchanged. One generally preserves the parameters of the final layer ($g_m$), so that the corresponding $\eta$'s are equal to one, and let the other ones be independent, with some probability $p$ of  being one, say, $p=1/2$.

Returning to the general case, in which $\boldsymbol\eta$ is simply assumed to be a random variable with known probability distribution, the dropout method replaces the objective function $F(\CW) = \CE_T(f(\cdot, \CW))$ by its expectation over perturbed predictors
$G(\CW) = E(\CE_T(f(\cdot, \boldsymbol\eta\cdot\CW)))$ where the expectation is taken with respect to the random variable $\boldsymbol\eta$.  While this expectation cannot be computed explicitly, its minimization can be performed using stochastic gradient descent, with
\[
\CW_{n+1} = \CW_n - \ga_{n+1}  L(\CW_n, \eta_{n+1}),
\]
where $\eta_1, \eta_2, \ldots$ is a sequence of independent realizations of $\boldsymbol\eta$ and 
\[
L(\CW, \eta) = \nabla_{\CW} \left(\CE_T(f(\cdot, \eta\cdot\CW))\right)\,.
\]
Then, averaging in $\eta$
\[
\bar L(\CW) = E(\nabla_{\CW} F(\eta\odot\CW)) = \nabla F(\CW).
\]

In the special case where $\eta\cdot\CW$ is just pointwise multiplication, then
\[
L(\CW, \eta) = \eta\odot \nabla F(\eta\odot\CW).
\]
As a consequence, this quantity can be evaluated by using back-propagation to compute $\nabla F(\eta\cdot\CW)$ and multiplying the result  by $\eta$ pointwise. Obviously, random weight perturbation can be combined with mini-batch selection in a hybrid stochastic gradient descent algorithm, the specification of which being left to the reader. We also note that stochastic gradient descent in neural networks is often implemented using the ADAM algorithm  (\cref{sec:adam}).

\section{Continuous time limit and dynamical systems}
\label{sec:continuous.nn}
\subsection{Neural ODEs}
\label{sec:neural.ode}
Equation \cref{eq:resnet} expresses the difference of the input and output of a neural transition as a non-linear function $f(z;w)$ of the input. This strongly suggests passing to continuous time and replacing the difference by a derivative, i.e., replacing the neural network by a high-dimensional parametrized dynamical system. The continuous model then takes the form \citep{chen2019neural}
\begin{equation}
\label{eq:neural.ode}
\prt_t z(t) = \psi(z(t); w(t))
\end{equation}
where $t$ varies in a a fixed interval, say, $[0,T]$. The whole process is parametrized by $\CW = (w(t), t\in [0,T])$. We need to assume existence and uniqueness of solutions of \cref{eq:neural.ode}, which usually restricts the domain of admissibility of parameters $\CW$.  

Typical neural transition functions are Lipschitz functions whose constant depend on the weight magnitude, i.e., are such that
\begin{equation}
\label{eq:ode.lip}
|\psi(z,w) - \psi(z', w)| \leq C(w) |z-z'|
\end{equation}
where $C$ is a continuous function of $W$. For example, for $\psi(z, w) = \rho(bz + \beta_0)$, $w = (b, \beta_0)$, one can take
$C(w) = C_\rho |b|_{\mathrm{op}}$. The Caratheodory theorem \citep{barbu2016differential} implies that solutions are well-defined as soon as 
\begin{equation}
\int_0^T C(w(t)) dt < \infty.
\label{eq:ode.carath}
\end{equation}
This is  a relatively mild requirement, on which we will return later. Assuming this, we can consider $z(T)$ as a function of the initial value, $z(0) = x$ and of the parameters, writing $z(T) = f(x, \CW)$.

Given a training set, we consider the problem of minimizing 
\begin{equation}
\label{eq:neural.ode.objective}
F(\CW) = \frac1N \sum_{k=1}^N r(y_k, f(x_k, \CW)).
\end{equation}
The discussion in section \cref{sec:nn.der} applies---formally, at least---to this continuous case, and we can consider the equivalent problem of minimizing
\[
\boldsymbol G(\CW, \bfz_1,\ldots, \bfz_N) = \frac{1}{N} \sum_{k=1}^N r(y_k, \bfz_k(T))
\]
with $\prt_t \bfz_k(t) = \psi(\bfz_k(t); w(t))$, $\bfz_k(0) = x_k$. Once again, we  consider each $k$ separately, which boils down to considering $N=1$ and we drop the index $k$ from the notation, letting $F(\CW) = r(y, f(x, \CW))$
$\boldsymbol G(\CW, \bz) = r(y, \bfz(T))$.

We define 
$\gamma(\CW, \bfz)$ to return the {\em function}
\[
t \mapsto \gamma(\CW, \bfz)(t) = \psi(\bfz(t); w(t)) - \prt_t \bfz(t).
\]
Let $\bfp: [0,T] \to \mR^d$. We want to determine the expression of $\boldsymbol u = \prt_z \gamma^T \bfp$, which satisfies 
\[
\int_0^T \bfu(t)^T \delta \bfz(t) dt = \int_0^T \bfp(t)^T (\prt_z \psi(\bfz(t), w(t))\delta\bfz(t) -\prt_t \delta \bfz(t))  dt
\]
After an integration by parts, the r.h.s. becomes 
\[
-\bfp(T)^T \delta \bfz(T) + \int_0^T \prt_t \bfp(t)^T \delta\bfz(t) dt + \int_0^T \bfp(t)^T \prt_z \psi(\bfz(t), w(t)) \delta\bfz(t)) dt
\]
which gives
\[
\bfu(t) = - \bfp(T) \delta_T + \prt_t \bfp(t) + \prt_z \psi(\bfz(t), w(t))^T \bfp(t).
\]
The equation $\prt_{\boldsymbol z} \gamma^T \boldsymbol p = \prt_{\boldsymbol z} \boldsymbol G^T$
therefore gives
\[
-\bfp(T) \delta_T + \prt_t \bfp(t) + \prt_z \psi(\bfz(t), w(t))^T \bfp(t) = \prt_2 r(y, \bfz(T)) \delta_T,
\]
so that $\bfp$ satisfies $\bfp(T) = -\prt_2 r(y, \bfz(T))$ and 
\begin{equation}
\label{eq:adjoint.p}
\prt_t \bfp(t) = - \prt_z \psi(\bfz(t), w(t))^T \bfp(t).
\end{equation}

We have $\prt_{\CW} \boldsymbol G = 0$ and $\bfv = \prt_{\CW} \gamma^T \bfp$ satisfies
\[
\int_0^T \bfv(t)^T \delta w(t) dt = \int_0^T \bfp(t)^T \prt_w \psi(\bfz(t), w(t)) \delta w(t) dt
\]
so that
\[
\nabla F(\CW) = (t \mapsto - \prt_w \psi(\bfz(t), w(t))^T \bfp(t)).
\]

This informal derivation (more work is needed to justify the existence of various differentials in appropriate function spaces) provides the continuous-time version of the back-propagation algorithm, which is also known as the {\em adjoint method} in the optimal control literature \citep{hocking1991optimal,macki2012introduction}. In that context, $z$ represents the state of the control system, $w$ is the control and $p$ is called the costate, or covector. We summarize the gradient computation algorithm, reintroducing $N$ training samples.
\begin{algorithm}[Adjoint method for neural ODE]
\label{alg:adjoint} 
Let $(x_1, y_1, \ldots, x_N, y_N)$ be the training set and $R_k(z) = r(y_k, z)$ so that 
\[
F(\CW) = \frac1N \sum_{k=1}^N R_k(\bfz_{k}(T, \CW))
\]
 with $\prt_t \bfz_{k} = \psi(\bfz_k, \CW)$, $\bfz_k(0) = x_k$.
 Let $\CW$ be a family of weights. The following steps compute $\nabla F(\CW)$.
\begin{enumerate}[label=\arabic*., wide = 0.5cm]
\item For all $k=1, \ldots, N$ and all $t\in[0,T]$, compute $\bfz_{k}(t, \CW)$ (forward computation through the dynamical system).
\item Initialize variables $\bfp_{k}(T) = -\nabla R_k(\bfz_{k}(T, \CW))/N$, $k=1, \ldots, N$.
\item For all $k=1, \ldots, N$ and all $j=1, \ldots, m$, compute $\bfp_{k}(t)$ by solving (backwards in time)
\[
\prt_t \bfp_k(t) = - \prt_z \psi(\bfz_k(t), w(t))^T \bfp_k(t).
\]
\item Let, for $t\in [0,T]$,
\[
\nabla F(\CW)(t) = -\sum_{k=1}^N \prt_w \psi(\bfz_k(t), w(t))^T \bfp_k(t).
\]
\end{enumerate}
\end{algorithm}

Of course, in numerical applications, the forward and backward dynamical systems need to be discretized, in time, resulting in a finite number of computation steps. This can be done explicitly (for example using basic Euler schemes), or using ODE solvers \citep{chen2019neural} available in every numerical software.

\subsection{Adding a running cost}
Optimal control problems are usually formulated with a ``running cost'' that penalizes the magnitude of the control, which in our case is provided by the function $\CW: t \mapsto w(t)$. Penalties on network weights are rarely imposed with discrete neural networks, but, as discussed above, in the continuous setting, some assumptions on the function $\CW$, such as \cref{eq:ode.carath}, are needed to ensure that the problem is well defined.

It is therefore natural to modify the objective function in \cref{eq:neural.ode.objective} by adding a penalty term ensuring the finiteness of the integral in \cref{eq:ode.carath}, taking, for example, for some $\lambda>0$,
\begin{equation}
\label{eq:neural.ode.objective.2}
F(\CW) = \lambda \int_0^T C(w(t))^2 dt +  \sum_{k=1}^N r(y_k, f(x_k, \CW)).
\end{equation}
The finiteness of the integral of the squared $C(w)^2$ implies, by Cauchy-Schwartz, the integrability of $C(w)$ itself, and usually leads to simpler computations.

If $C(w)$ is known explicitly and is differentiable, the previous discussion and the back-propagation algorithm can be adapted with minor modifications for the minimization of \cref{eq:neural.ode.objective.2}. The only difference appears in Step 4 of \cref{alg:adjoint}, with
\[
\nabla F(\CW)(t) = 2\lambda \nabla C(w(t)) -   \frac1N \sum_{k=1}^N \prt_w \psi(\bfz_k(t), w(t))^T \bfp_k(t).
\]
Computationally, one should still ensure that $C$ and its gradient are not too costly to compute. If $\psi(z, w) = \rho(bz + \beta_0)$, $w = (b, \beta_0)$, the choice
$C(w) = C_\rho |b|_{\mathrm{op}}$ is valid, but not computationally friendly. The simpler choice $C(w) = C_\rho |b|_{2}$ is also valid, but cruder as an upper-bound of the Lipschitz constant. It leads however to straightforward computations.

The addition of a running cost to the objective is important to ensure that any potential solution of the problem leads to a solvable ODE. It does not guarantee that an optimal solution exists, which is a trickier issue in the continuous setting than in the discrete setting. This is an important theoretical issue, since it is needed, for example, to ensure that various numerical discretization schemes lead to consistent approximations of a limit continuous problem. The existence of minimizers is not known in general for ODE networks. It does hold, however, in the following  non-parametric (i.e., weight-free) context that we now describe.   

\bigskip
The function $\psi$ in the r.h.s. of \cref{eq:neural.ode}, is, for any fixed $w$, a function that maps $z\in \mR^d$ to a vector $\psi(z, w)\in \mR^d$. Such functions are called {\em vector fields} on $\mR^d$, and the collection $\psi(\cdot, w), w\in \mR^s$ is a parametrized family of vector fields.

The non-parametric approach replaces this family of functions by a general vector  field, $v$ so that the time-indexed parametrized family of vector fields $(t \mapsto \psi(\cdot, w(t)))$ becomes an unconstrained family $(t \mapsto f(t, \cdot))$. Following the general non-parametric framework in statistics, one needs to define a suitable function space for the vector fields, and use a penalty in the objective function.

We will assume that, at each time, $f(t, \cdot)$ belongs to a reproducing kernel Hilbert space (RKHS), as introduced in \cref{chap:higher}. However, because we are considering a space of vector fields rather than scalar-valued functions, we need work with matrix-valued kernels \cite{alvarez2012kernels}, for which we give a definition that generalizes \cref{def:pos.kernel} (which corresponds to $q=1$ below). 
\begin{definition}
\label{def:pos.kernel.op}
A function $K: \mR^d \ti\mR^d \mapsto \CM_q(\mR)$ satisfying 
\begin{enumerate}[label={[K\arabic*-vec]}, wide=0pt]
\item $K$ is symmetric, namely $K(x,y) = K(y,x)^T$ for all $x$ and $y$ in $\mR^d$. 
\item For any $n>0$, for any choice of vectors $\la_1, \dots, \la_n\in \mR^q$ and any $x_1, \ldots, x_n \in \mR^d$, one has
\begin{equation}
\label{eq:pos.kernel.op}
\sum_{i,j=1}^n \lambda_i^T K(x_i, x_j) \lambda_j \geq 0.
\end{equation}
\end{enumerate}
is called a positive (matrix-valued) kernel.

One says that
the kernel is positive definite if the sum in \cref{eq:pos.kernel} cannot vanish, unless (i) $\la_1 = \cdots = \la_n=0$ or (ii) $x_i=x_j$ for some $i\neq j$.
\end{definition}
If $\kappa$ is a ``scalar kernel'' (satisfying \cref{def:pos.kernel}), then $K(x,y) = \kappa(x,y)\Id[q]$ is a matrix-valued kernel. 

A reproducing kernel Hilbert space of vector-valued functions is a Hilbert space $H$ of functions from $\mR^d$ to $\mR^q$ such that there exists a reproducing kernel $K: \mR^d \ti\mR^d \mapsto \CM_q(\mR)$ with the following properties
\begin{enumerate}[label={[RKHS\arabic*]}, wide=0pt]
\item For all $x\in \mR^d$ and $\lambda\in \mR^q$, $K(\cdot, x)\lambda$ belongs to $H$,
\item For all $h\in H$, $x\in \mR^d$ and $\lambda\in \mR^q$, 
\[
\scp{h}{K(\cdot, x)\lambda}_H = \lambda^T h(x)\,.
\] 
\end{enumerate}

\Cref{prop:rkhs.interpolation} remains valid in the for vector-valued RKHS, with the following modifications: $\lambda_1, \ldots, \lambda_N$ and $\alpha_1, \ldots, \alpha_N$ are $q$-dimensional vectors and the matrix $\CK(x_1, \ldots, x_N)$ is 
now an $Nq\times Nq$ block matrix, with $q\times q$ blocks given by $K(x_k,x_l)$, $k,l= 1, \ldots, N$.

Returning to the specification of the nonparametric control problem, we will assume that a vector-valued RKHS, $H$, has been chosen, with $q=d$ in \cref{def:pos.kernel.op}. We further assume that elements of $H$ are Lipschitz continuous, with
\begin{equation}
\label{eq:lip.rkhs}
|v(z) - v(\tilde z)| \leq C \|v\|_H |z-\tz|
\end{equation}
for some constant $C$ and all $v\in H$. We note that, for every $\lambda\in \mR^d$,
\begin{align*}
|\lambda^T(v(z) - v(\tilde z))|^2 & = |\scp{v}{K(\cdot, z)\lambda - K(\cdot, \tz)\lambda}_H|^2\\
& \leq \|v\|_H^2\, \|K(\cdot, z)\lambda - K(\cdot, \tz)\lambda\|_H^2\\
& = \|v\|_H^2\, (\lambda^T K(z, z)\lambda -2 \lambda^T K(z, \tz)\lambda + \lambda^T K(\tz, \tz))\\
&\leq |\lambda|^2 \|v\|_H^2\, |K(z, z) -2 K(z, \tz) + K(\tz, \tz)|\,.
\end{align*}
This shows that \cref{eq:lip.rkhs} can be derived from regularity properties of the kernel, namely, that
\[
|K(z, z) -2 K(z, \tz) + K(\tz, \tz)| \leq C |z-\tz|^2
\]
for some constant $C$ and all $z, \tz\in \mR^d$. This property is satisfied by most of the kernels that are used in practice.

Let $\eta: t\mapsto \eta(t)$ be a function from $[0,1]$ to $H$. This means that, for each $t$, $\eta(t)$ is a vector field $x \mapsto \eta(t)(x)$ on $\mR^d$, and we will write indifferently $\eta(t)$ and $\eta(t, \cdot)$, with a preference for $\eta(t,x)$ rather than $\eta(t)(x)$. We consider the objective function
\begin{equation}
\label{eq:ode.objective.2}
\bar F(f) = \lambda \int_0^T \|\eta(t)\|_H^2 dt + \frac1N \sum_{k=1}^N r(y_k, z_k(1)),
\end{equation}
with $\prt_t z_k(t) = \eta(t, z_k(t))$, $z_k(0) = x_k$. To compare with \cref{eq:neural.ode.objective.2}, the finite-dimensional $w\in \mR^s$ is now replaced with an infinite-dimensional parameter, $\eta$, and the transition $\psi(z,w)$ becomes $\eta(z)$. 

Using the vector version of \cref{prop:rkhs.interpolation} (or the kernel trick used several times in \cref{chap:lin.reg,chap:lin.class}), one sees that there is no loss of generality in replacing $\eta(t)$ by its projection onto the vector space
\[
V(t) = \defset{\sum_{l=1}^N K(\cdot, z_l(t))w_l: w_1, \ldots, w_N\in \mR^d}.
\]
Noting that, if $\eta(t)$ takes the form 
\[
\eta(t) = \sum_{l=1}^N K(\cdot, z_l(t)) w_l(t),
\]
then
\[
\|\eta(t)\|^2_H = \sum_{k,l=1}^N w_k(t)^T K(z_k(t), z_l(t))w_l(t).
\]
This  allows us to replace the infinite-dimensional parameter $\eta$ by a family $\CW = (w(t), t\in [0,T]$ with $w(t) = (w_k(t), k=1, \ldots, N)$. The minimization of $\bar F$ in \cref{eq:ode.objective.2} can be replaced by that of
\begin{equation}
\label{eq:ode.objective.3}
F(\CW) = \lambda \int_0^T \sum_{k,l=1}^N  w_k(t)^T K(z_k(t), z_l(t))w_l(t) dt + \frac1N \sum_{k=1}^N r(y_k, z_k(1)),
\end{equation}
with 
\[
\prt_t z_k(t) = \sum_{l=1}^N K(z_k(t), z_l(t)) w_l(t).
\]

This optimal control problem has a similar form to that considered in  \cref{eq:neural.ode.objective.2}, where the running cost $C(w)^2$ is replaced by a cost that depends on the control (still denoted $w$) and the state $\bfz$. The discussion in section \cref{sec:neural.ode} can be applied with some modifications.
Let $\CK(\bfz)$ be the $dN \times dN$ matrix formed with $d\times d$ blocks $K(z_k(t), z_l(t))$ and $\bfw(t)$ the $dN$-dimensional vector formed by stacking $w_1, \ldots, w_N$. Let
\[
G(\CW, \bfz) = \lambda \int_0^T \bfw(t)^T \CK(\bfz(t))\bfw(t) dt + \frac1N \sum_{k=1}^N r(y_k, z_k(1))
\]
and
\[
\gamma(\CW, \bfz)(t) = \CK(\bfz(t))\bfw(t)- \prt_t \bfz(t)\,.
\]
The backward ODE in step 3. of \cref{alg:adjoint} now becomes
\[
\prt_t \bfp_k(t) = -\prt_{z_k} (\bfw(t)^T \CK(\bfz(t))\bfp(t)) + \lambda \prt_{z_k} (\bfw(t)^T \CK(\bfz(t))\bfw(t))
\]
for $k=1, \ldots, N$.
Step 4. becomes (for $t\in [0,T]$),
\[
\nabla F(\CW)(t) = \CK(\bfz(t)) (2\lambda  -  \bfp(t)).
\]

The resulting algorithm was introduced in \citep{younes2020diffeomorphic}. It has the interesting property (shared with neural ODE models with smooth controlled transitions) to determine an implicit diffeomorphic transformation of the space, i.e., the function $x\mapsto f(x; \CW, \bfz) = \tilde z(T)$ which returns the solution at time $T$ of the ODE 
\[
\prt_t \tilde z(t) =  \sum_{l=1}^N K(\tilde z(t), z_l(t)) w_l(t)
\]
(or $\prt \tilde z(t) = \psi(\tilde z(t); w(t))$ for neural ODEs) is smooth, invertible, with a smooth inverse.

\newpage
\markright{PROBLEMS}
\section*{Problems}
\addcontentsline{toc}{section}{Problems}
\input Problems_Neural_Networks

\chapter{Comparing probability distributions}
\label{chap:compare.prob}
When discussing, in the next chapters, generative machine learning methods to learn probability distributions, we will need to evaluate the difference between two such distributions, for example, to optimize an algorithm to return a learned distribution close to an observed one. We regroup in this chapter some approaches that are used in machine learning for this purpose. In the following, $\CR$ is taken, as always, as a metric space equipped with its Borel $\sigma$-algebra.

\section{Total variation distance}
\label{sec:total.var}
\begin{definition}
\label{def:total.var}
Let $P$ and $Q$ be two probability distributions on $\CR$.  Their  total variation distance is defined by
\begin{equation}
\label{eq:total.var}
D_{\mathrm{var}}(\mu_1,\mu_2)  = \sup_A(\mu_1(A) - \mu_2(A)).
\end{equation}
where the supremum is taken over all measurable sets $A$. 
\end{definition}

We have the following lemma.
\begin{lemma}
\label{lem:total.var}
There exists a measurable set $A_0$ such that, for all $B$, $P(B\cap A_0) \geq Q(B\cap A_0)$ and $P(B\cap A^c_0) \leq Q(B\cap A^c_0)$. 

Moreover
The supremum in the r.h.s. of \cref{eq:total.var} is achieved at $A = A_0$.
\end{lemma}
\begin{proof}
 If $\CR$ is a finite set, it suffices to let $A_0= \{x\in \CR: P(x) \geq Q(x)\}$. If both $P$ and $Q$ have p.d.f.'s $\psi_1$ and $\psi_2$ with respect to Lebesgue's measure (with $\CR = \mR^d$), then one can take  
$A_0= \{x\in \CB: \psi_1(x) \geq \psi_2(x)\}$. 
In the general case, take $\mu = P+Q$ so that $P, Q\ll \mu$, and let $\psi_i = d\mu_i/d\mu$ and  $A_0= \{x\in \CR: \psi_1(x) \geq \psi_2(x)\}$. (This fact is also a special case of  the Hahn-Jordan decomposition of signed measures \citep{dudley2018real}). 

Now, it is clear that, for any $A\in \CR$,
\begin{align*}
P(A) - Q(A) &= P(A\cap A_0) - Q(A\cap A_0) + P(A\cap A^c_0) - Q(A\cap A^c_0)\\
&\leq P(A\cap A_0) - Q(A\cap A_0) \\
&\leq P(A\cap A_0) - Q(A\cap A_0) + P(A^c\cap A_0) - Q(A^c\cap A_0)\\
&= P(A_0) - Q(A_0)
\end{align*}
showing that
\[
D_{\mathrm{var}}(P,Q)  = P(A_0) - Q(A_0).
\]
\end{proof}

The following proposition lists additional properties.
\begin{proposition}
\label{prop:var.dist}
\begin{enumerate}[label=(\roman*)]
\item If $P, Q$ have densities $\psi_1, \psi_2$ with respect to some positive measure $\mu$ (such as $P+Q$), then
\[
D_{\mathrm{var}}(P,Q)  = \frac12 \int_{\CR} |\psi_1(x) - \psi_2(x)|\mu(dx).
\]
In particular, if $\CB$ is finite
\[
D_{\mathrm{var}}(P,Q)  = \frac12 \sum_{x\in \CB} |P(x) - Q(x)|.
\]

\item For general $\CR$,
\begin{equation}
\label{eq:total.var.2}
D_{\mathrm{var}}(P,Q)  = \sup_f\left(\int_\CR f(x) P(dx) - \int_\CB f(x) Q(dx)\right).
\end{equation}
where the supremum is taken over all measurable functions $f$ taking values in $[0,1]$.
\item If $f: \CR \to \mR$ is bounded, define the maximal oscillation of $f$ by
\[
\mathit{osc}(f) = \sup\{f(x) - f(y): x, y\in \CR\}.
\]
Then 
\[
D_{\mathrm{var}}(\mu_1,\mu_2) = \sup\left\{ \int_{\CR} f(x) P(dx) - \int_{\CR} f(x) Q(dx) : \mathit{osc}(f) \leq 1\right\}
\]  

\item Conversely, for any bounded measurable $f: \CR\to \mR$,
\[
\mathit{osc}(f) = \sup\left\{ \int_{\CR} f(x) \mu_1(dx) - \int_{\CR} f(x) \mu_2(dx) : D_{\mathrm{var}}(\mu_1,\mu_2) \leq 1\right\}
\]
\end{enumerate}
\end{proposition}
\begin{proof}
If one takes $A_0= \{x\in \CB: \psi_1(x) \geq \psi_2(x)\}$, then
\[
D_{\mathrm{var}}(P,Q)  = \int_{A_0} (\psi_1(x) - \psi_2(x))\mu(dx) = \int_{A_0} |\psi_1(x) - \psi_2(x)|\mu(dx).
\]
But, because both $P$ and $Q$ are probability measures,
\[
\int_{\CR} (\psi_1(x) - \psi_2(x))\mu(dx) = 0
\]
so that 
\[
\int_{A_0^c} (\psi_1(x) - \psi_2(x))\mu(dx) = - \int_{A_0} (\psi_1(x) - \psi_2(x))\mu(dx).
\]
However, the l.h.s. is also equal to 
\[
- \int_{A_0^c} |\psi_1(x) - \psi_2(x)|\mu(dx)
\]
so that 
\[
\int_{\CR} |\psi_1(x) - \psi_2(x)|\mu(dx) = 2 \int_{A_0} (\psi_1(x) - \psi_2(x)) = 2 D_{\mathrm{var}}(P,Q),
\]
which proves (i).

To prove (ii), first notice that, for all $A$, 
\[
P(A) - Q(A) = \int_\CR f(x) P(dx) - \int_\CR f(x) Q(dx)
\]
for $f = \bfone_A$, so that 
\[
D_{\mathrm{var}}(P,Q)  \leq \sup_f\left(\int_\CR f(x) P(dx) - \int_\CR f(x) Q(dx)\right).
\]
Conversely, using $A_0$ as above, and taking $f$ with values in $[0,1]$
\begin{align*}
\int_\CR f(x) P(dx) - \int_\CR f(x) Q(dx) &= 
\int_{A_0} f(x) (P-Q)(dx) - 
\int_{A_0^c} f(x) (Q-P)(dx) \\
&\leq 
\int_{A_0} f(x) (P-Q)(dx)\\ 
&\leq 
\int_{A_0} (P-Q)(dx) =  D_{\mathrm{var}}(P,Q)
\end{align*}

This shows (ii). For (iii), one can note that, if $f$ takes values in $[0,1]$, then $\mathit{osc}(f) \leq 1$ so that, using (ii),
\[
D_{\mathrm{var}}(P,Q) \leq \sup\left\{ \int_{\CR} f(x) P(dx) - \int_{\CR} f(x) Q(dx) : \mathit{osc}(f) \leq 1\right\}
\] 
since the maximization on the r.h.s. is done on a larger set than in (ii).

Conversely, take $f$ such that $\mathit{osc}(f) \leq 1$, $\epsilon > 0$ and $y$ such that $f(y) \geq \inf f + \epsilon$. Let $f_\epsilon(x) = (f(x) - f(y) + \epsilon)/(1+\epsilon)$, which takes values in $[0,1]$. Then
\begin{align*}
D_{\mathrm{var}}(P,Q) &\geq 
\int_\CR f_\epsilon(x) P(dx) - \int_\CR f_\epsilon(x) Q(dx)\\
&= \frac{1}{1+\epsilon}\left(
\int_\CR f(x) P(dx) - \int_\CR f(x) Q(dx)\right)
\end{align*}
and since this is true for all $\epsilon>0$, we get
\[
\int_\CR f(x) P(dx) - \int_\CR f(x) Q(dx)\leq D_{\mathrm{var}}(P,Q)
\]
which completes the proof of (iii). 

Using (iii), we find, for any $P, Q$ and any bounded $f$
\[
\int_{\CR} f(x) P(dx) - \int_{\CR} f(x) Q(dx) \leq D_{\mathrm{var}}(P,Q) \mathit{osc}(f)
\]
which shows that
\[
\sup\left\{ \int_{\CR} f(x) P(dx) - \int_{\CR} f(x) Q(dx) : D_{\mathrm{var}}(P,Q) \leq 1\right\} \leq \mathit{osc}(f).
\]
However, taking $P = \delta_x$ and $Q = \delta_y$, so that  $D_{\mathrm{var}}(P,Q) = 0$ if $x=y$ and 1 otherwise, we get
\begin{align*}
f(x) - f(y) &= \int_{\CR} f(x) P(dx) - \int_{\CR} f(x) Q(dx)\\
&\leq  \sup\left\{ \int_{\CR} f(x) P(dx) - \int_{\CR} f(x) Q(dx) : D_{\mathrm{var}}(P,Q) \leq 1\right\} 
\end{align*}
which yields (iv) after taking the supremum with respect to $x$ and $y$.
\end{proof}

\begin{remark}
\label{rem:total.var}
Statements (ii)--(iv) in \cref{prop:var.dist} still hold when the supremums are made over continuous functions.
\end{remark}

\section{Divergences} 
\label{sec:divergence}
We encountered a first definition of divergence in \cref{chap:intro} with 
the Kullback-Leibler (KL) divergence introduced   in equation \cref{eq:kl.definition}). It was shown, in \cref{prop:kl}, to provide a non-negative quantification of the discrepancy between two probability measures $P$ and $P$,  vanishing only when $P=Q$. Equation \cref{eq:kl.definition} can be extended to the notion of $f$-divergence as follows.
\begin{definition}
\label{def:phi.divergence} 
Let $f$ be a non-negative convex function on $(0, +\infty)$ such that $f(1) = 0$ and $f(t) >0$ for $t\neq 1$. Let $P$ and $Q$ be two probability distributions on some space $\CR$, and $\mu$ a measure on $\CR$ such that $P\ll \mu$ and $Q\ll \mu$. Letting $\phi_P = dP/d\mu$ and $\phi_Q = dQ/d\mu$, the $f$-divergence between $\mu$ and $\nu$ is defined by
\begin{equation}
\label{eq:phi.divergence}
D_f(P \ \|\  Q) =\int_{\tilde\Om} \phi_Q f\left( \frac{\phi_P}{\phi_Q}\right)  d\mu 
\end{equation}
with the convention  
\[
f(0) = \lim_{t\to 0}f(t) \text{ and } 0\phi(\phi/0) = \phi \lim_{t\to 0} f^*(t),
\]
where
\[
f^*(t) = tf(1/t).
\]
Note that the limits above may be infinite.
\end{definition}
This definition does not depend on $\mu$ such that $P, Q\ll \mu$. Indeed, if $P, Q \ll\mu$, then $P, Q \ll P+Q \ll \mu$ and letting $\tilde \phi_P = dP/d(P+Q)$, $\tilde \Phi_Q  = dQ/d(P+Q)$ and $\psi = \phi_P+\phi_Q$, one checks that $\phi_P = \tilde \phi_P \psi$, $\phi_Q=\tilde \Phi_Q \psi$ and 
\[
\int_{\tilde\Om} \tilde \phi_Q f\left( \frac{\tilde \phi_P}{\tilde \phi_Q}\right) d(P+Q) =   \int_{\tilde\Om} \phi_Q f\left( \frac{\phi_P}{\phi_Q}\right)  d\mu
\]
with the l.h.s. independent of $\mu$. It is also clear that $D_f(P\ \| \ Q)\geq 0$ and vanishes if and only if $P=Q$. Note that the KL divergence is $D_f$ for $f(t) = t \log t + 1 - t$.

This divergence is, in general, not symmetric, nor does it satisfy the triangular inequality. There are, however, sufficient conditions that ensure that these properties are true \citep{csiszar1967topological,kafka1991powers,vajda2009metric}. Symmetry is captured by 
the ``conjugate'' function $f^*$ in \cref{def:phi.divergence}. It  is, like $f$, non-negative and convex on $(0, +\infty)$ and vanishes only at $t=1$. The only part of this statement that is not obvious is that $f^*$ is convex, but we have, for $\lambda\in (0,1)$, $s,t>0$, 
\begin{align*}
f^*((1-\lambda)s + \lambda t) &= ((1-\lambda)s + \lambda t) f\left(
\frac{1}{(1-\lambda)s + \lambda t}\right)\\
&= ((1-\lambda)s + \lambda t) f\left(\frac{(1-\lambda)s}{(1-\lambda)s + \lambda t} \frac1s + \frac{\lambda t}{(1-\lambda)s + \lambda t} \frac1t\right)\\
&\leq ((1-\lambda)s + \lambda t)\left( \frac{(1-\lambda)s}{(1-\lambda)s + \lambda t}  f\left(\frac1s\right) + \frac{\lambda t}{(1-\lambda)s + \lambda t} f\left( \frac1t\right)\right)\\
&= (1-\lambda) f^*(s) + \lambda f^*(t).
\end{align*}
Clearly $D_{f^*}(P\ \|\  Q) = D_\phi(Q\ \|\ P)$ so that a simple sufficient condition for symmetry is that $f^* = f$. Symmetry can always be obtained by replacing $f$ by $\tilde f = (f + f^*)/2$, yielding
\[
D_{\tilde f}(P\ \|\ Q) = \frac12 ( D_f(P\ \|\ Q) + D_{f}(Q\ \|\ P)).
\]
(The symmetrized KL divergence is called the Jeffreys divergence.)

The validity of the triangle inequality is more challenging to ensure. In  \citet{kafka1991powers}, it is proved that, if $f^* = f$ and the function
\[
h(t) = \frac{|t^\al - 1|^{1/\alpha}}{f(t)}
\]
is well defined on $(0, +\infty)$, non-increasing on $(0, 1)$ and continuous at $t=1$,
then $(P,Q) \mapsto D_f(P, Q)^\alpha$ is symmetric and satisfies the triangle inequality. 

One can, for example, take $f(t) = |t^\al - 1|^{1/\alpha}$, which yields, with the notation of \cref{def:phi.divergence}, 
\[
D_f(P,Q) = \int_{\CR} |f^\alpha - g^\alpha|^{1/\alpha} d\mu.
\]
The case $\alpha=1$ gives two times the total variation distance. The case $\alpha=1/2$ provides the Hellinger distance between $P$ and $Q$
\[
D_{\mathrm{Hellinger}} (P, Q) = \left(\int_{\CR} |\sqrt f - \sqrt g|^{2} d\mu
\right)^{1/2}.
\]

In \cite{vajda2009metric}, it is proved that this condition is satisfied for the family
\[
f_\beta(t) = \frac{\sign(\beta)}{\beta-1} \left((t^{1/\beta}+1)^\beta - 2^{\beta-1}(t+1)\right).
\]
with $\alpha = 1/2$ for $\beta < 2$ and $\alpha = 1/\beta$ for $\beta \geq 2$. The limit cases
\[
f_0(t) = \frac12 |t-1|,
\]
which provides the total variation distance
and
\[
f_1(t) = t\log t - (t+1) \log \left(\frac{t+1}2\right)
\]
are also included. Also, for $\beta = 2$, one retrieves the Hellinger distance.

One can also check that, for $\beta=1$, one has
\[
D_{f_1}(P, Q) = \KL(P\ \| \ (P+Q/2)) + \KL(Q\ \| \ (P+Q/2)).
\] 
The r.h.s. is called the  Jensen-Shannon divergence between $P$ and $Q$.

\section{Monge-Kantorovich distance}
\label{sec:monge}
 The Monge-Kantorovich, also called Wasserstein, and sometimes also called ``earth-mover,'' assigns a transportation cost, say, $\rho(x,y)$, for moving a unit of mass from $x$ to $y$, and evaluates the minimum total cost needed to transform the distribution $P$ into $Q$. Assume that this transport is realized by a function $g: \CR\to \CR$ (so that the mass at $x$ is moved to $y = g(x)$). Then $g$ transforms $P$ into the probability
 \[
 g_\sharp P: A \mapsto P(g^{-1}(A))
 \]
 called the ``push-forward'' of $P$ by $g$ (or the image of $P$ by $g$). One can then formulate a deterministic optimal transport problem that minimizes 
 \begin{equation}
 \label{eq:monge.0}
 \int_\CR \rho(x, g(x)) P(dx)
 \end{equation}
 over all measurable functions $g$ such that $g_{\#} P = Q$.

The Monge-Kantorovitch distance is a relaxation of this problem. Its mathematical definition is
\begin{equation}
\label{eq:monge}
D_{\mathrm{MK}}(P, Q) = \inf_{\pi\in M(P,Q)} \int_{\CR\times\CR} \rho(x,y)  \pi(dx,dy)
\end{equation}
where $M(P,Q)$ is the set of  all joint distributions on $\CR\times\CR$ whose first marginal is $P$ and second marginal $Q$. The interpretation is that $\pi(dx, dy)$ is the infinitesimal fraction of mass moved between the infinitesimal neighborhoods $x+dx$ and $y+dy$. The constraint $\pi\in M(P,Q)$ indicates that $\pi$ displaces the mass distribution $P$ to the mass distribution $Q$.

Note that, if $g$ is such that $g_\# P = Q$, then the distribution $\pi_g$ such that $\pi_g(A\times B) = P(A\cap g^{-1}(B))$ belongs to $M(P,Q)$ and is such that  
\[
\int_{\CR\times\CR} \rho(x,y)  \pi_g(dx,dy) =  \int_\CR \rho(x, g(x)) P(dx).
\]
Indeed, the two marginals of $\pi_g$ are $A  \mapsto P(A\cap g^{-1}(\CR)) = P(A)$ and $B \mapsto P(\CR \cap g^{-1}(B)) = Q(B)$. Moreover, letting $h(x,y) = \bfone_{A\times B}(x,y)$, one has
\[
\int_{\CR\times \CR} h(x,y) \pi_g(dx, dy) = P(A\cap g^{-1}(B)) = \int_\CR h(x,g(x)) P(dx).
\]
This identity can then be extended to any function $h: \CR\to \CR \to [0, +\infty)$ using standard arguments (linearity, passing to limits\ldots) and therefore also holds for $\rho$. \Cref{eq:monge} is therefore a minimization over a larger set than \cref{eq:monge.0}. This relaxation becomes a convex minimization problem (albeit infinite dimensional) since the objective function is linear in $\pi$ and the set $M(P, Q)$ is convex.

One can easily check that $M(\de_{x_1}, \de_{x_2}) = \{\delta_{x_1}\otimes\delta_{x_2}$, so that 
\[
D_{\mathrm{MK}}(\de_{x_1}, \de_{x_2}) = \rho(x_1, x_2).
\]

If we assume that, for some $\beta \geq 1$,  $\sigma = \rho^{1/\beta}$ is a distance on $\CR$, then  $D_{\mathrm{MK}}^{1/\beta}$ is a distance on the space of probability measures on $\CR$ (equipped with the Borel $\sigma$-algebra specified by $\sigma$). For this fact, and the result that follow, the reader can refer to \citet{villani2009optimal,dudley2018real}.

When $\alpha=1$ (so that $\rho=d$ is a distance), one has, furthermore, the following theorem. 
Call a function $f: \CR \to \mR$ $\rho$-contractive if, for all $x,y\in \CR$, one has
\[
|f(x) - f(y)| \leq \rho(x,y).
\]
Define 
\[
D^*_\rho(P, Q) = \max \left(\int f dP - \int f dQ: f \rho\text{-contractive}\right)
\]
Then one has:

\begin{theorem}[Kantorovich-Rubinstein]
\label{th:kant.rub}
One has $D_{\mathit{MK}}(P,Q)  = D^*_\rho(P, Q)$
\end{theorem}
See the reference above for a proof. Further generalizations of this theorem, in particular for the case in which $\rho$ is not equal to a distance, can be found in \cite{villani2009optimal}, chapter 5.

\section{Dual distances}
If $\CF$ is any set of functions from $\CR$ to some finite-dimensional vector space $\CB$ and $\alpha$ is a distance on $\CB$, one can define
\[
\mathfrak D^*_\alpha(P,Q) = \sup\left(\alpha\left(\int_{\CR} f dP, \int_{\CR} fdQ\right): f\in \CF \right).
\]
Here, we do not exclude the possibility that $\mathfrak D^*_\alpha(P,Q) = \infty$.
This function (which can also be defined if $\CB$ is infinite-dimensional, for example, a Banach space) is obviously symmetric in $P$ and $Q$ and satisfies the triangle inequality, as directly derived by applying the triangle inequality for $\alpha$. We have furthermore $\mathfrak D^*_\alpha(P,P) = 0$ for any $P$. The converse statement, $\mathfrak D^*_\alpha(P, Q) = 0 \Rightarrow P=Q$ depends on how large the set $\CF$ is, since it requires that 
\[
\int_{\CR} f dP = \int_{\CR} f dQ
\]
for all $f\in \CF$ implies $P=Q$, so that, in general, $\mathfrak D^*_\alpha$ is a pseudo-distance.

We have met above some examples using this construction, with $\CB = \mR$ and $\CF$ the set of all measurable functions taking values in $[0,1]$, or functions with oscillation bounded by 1 (both leading to the distance in variation), or $\rho$-contractive functions leading to the Monge-Kantorovich distance. Of course, many more similar definitions can be made, with more or less practical or theoretical interest. We here mention one special case that leads to simple computations.

Let $H$ be reproducing kernel Hilbert space (RKHS: see \cref{chap:higher}) of real-valued functions, with associated kernel $K : \CR\times \CR \to \mR$. Assume that $x \mapsto K(x,x)$ is bounded, say by a constant $M$, which implies that $K$ is bounded, since, for all $x,y$, $K(x,y)^2 \leq K(x,x) K(y,y)$. Let $\CF$ be the unit ball in $H$, namely
\[
\CF = \left\{f\in H: \|f\|_H \leq 1\right\}.
\]
Let $\alpha(s,t) = |s-t|$ on $\CR$.
Then, one has the following result.
\begin{proposition}
\label{prop:mmd}
One has (with the notation just introduced)
\[
\mathfrak D^*_\alpha(P,Q) = \left(\scp{P}{P}_{H^*} + \scp{Q}{Q}_{H^*} - 2\scp{P}{Q}_{H^*}\right)^{1/2}
\]
with 
\[
\scp{P}{Q}_{H^*} = \int_{\CR\times \CR} K(x,y) P(dx) Q(dy).
\]
\end{proposition}

\begin{proof}
To prove this statement, we first note that, by definition of a reproducing kernel, one has $|f(x)| = |\scp{f}{K(\cdot, x)}_H| \leq K(x,x) \|f\|_H$. From this, it follows that
\[
\int_\CR f\,dP \leq  M \|f\|_H
\]
for any probability measure on $\CR$. One can now invoke a standard result from functional analysis (Riesz's theorem) to conclude that there exists $u_P \in H$ such that, for all $f\in H$, 
\[
\int_\CR f\,dP = \scp{u_P}{f}_H.
\]
The function $u_P$ can be identified by taking $f = K(\cdot, x)$ in this identity, so that
\[
u_P(x) = \scp{u_P}{K(\cdot, x)}_H = \int_\CR K(y,x)\,P(dy).
\]
It follows that computating of $\mathfrak D_\alpha^*(P,Q)$ requires to maximize 
$\scp{u_P-u_Q}{f}_{H}$ subject to $\|f\|_H = 1$, which yields $\mathfrak D_\alpha^*(P,Q)= \|u_P-u_Q\|_H$. Now, by construction of $u_P$ and $u_Q$, we have
\[
\scp{u_P}{u_Q}_H = \int_\CR u_Q(x)\,P(dx) = \int_{\CR\times \CR} K(x,y) P(dx) Q(dy).
\]
\end{proof}

\begin{remark}
\label{rem:mmd.1}
\begin{enumerate}
\item Our notation, $\scp{P}{Q}_{H^*}$, reflects the fact that $P \mapsto u_P$ is induced  by the isometry between the dual space $H^*$ of $H$ and $H$ itself \citep{brezis2011functional}.
\item One can prove similarly that, given a feature function $h: \CR\to H$ such that $K(x,y) = \scp{h(x)}{h(y)}_H$, one has 
\[
\mathfrak D^*_\alpha(P, Q) = \left\| \int_{\CR} h(x) P(dx) - \int_{\CR} h(x) Q(dx)\right\|_H.
\]
\item This distance (also introduced in \citet{glaunes2004diffeomorphic}) was called {\em maximum mean discrepancy} in \citet{smola2006maximum}.
\end{enumerate}
\end{remark}

\chapter{Monte-Carlo Sampling}
\label{chap:mcmc}

The goal of this section is to describe how, from a basic random
number generator that provides samples from a uniform distribution on $[0,1]$, one
can generate samples that follow, or
approximately follow, complex probability distributions on finite or general spaces. This,
combined with the law of large numbers, permits to 
approximate probabilities or expectations by empirical averages over
a large collection of generated samples.

We assume that as many as needed independent samples of the uniform distribution are
available, which is only an approximation of the truth. In practice,
computer programs are only able to generate pseudo-random numbers,
which are highly chaotic recursive sequences, but still
deterministic. Also, these numbers are generated as integers, which
only provide, after normalization, a distribution on a finite discretization
of the unit interval. We will neglect these facts, however, and work
as if the output of the function {\em random} (or any similar name) in
a computer program is a true realization of the uniform distribution.

\section{General sampling procedures}
\label{sec:gen.samp}
\paragraph{Real-valued variables.}
We will use the following notation for the left limit of a function $F$ at a given point $z$ 
\[
F(z\!-) = \lim_{y\to z, y<z} F(y)
\]
assuming, of course that this limit exists (which is always true, for example when $F$ is non-decreasing). 
Recall that $F$ is left continuous if and only if  $F=F(\ccdot\!-)$. Moreover, it is easy to see that $F(\ccdot\!-)$ is left-continuous\footnote{ For every $z$ and every $\epsilon>0$, there exists $z'<z$ such that for all $z''\in [z', z)$, $|F(z'') - F(z\!-)| < \epsilon$. Moreover, taking any $y\in (z', z)$, there exists $y'<y$ such that for all $y''\in [y', y)$, $|F(y'') - F(y\!-)| < \epsilon$. Without loss of generality, we can assume that $y'\geq z'$, yielding $|F(z\!-) - F(y\!-)| \leq 2\epsilon$, showing the left continuity of $F(\ccdot-0)$ since this is true for all $y\in (z',z)$.}. 
Note also that, if $F$ is non-decreasing, one always has $F(z) \leq F(y\!-)$ whenever $z < y$. The following proposition provides a basic mechanism for Monte-Carlo sampling.

\begin{proposition}
\label{prop:sampling}
Let $Z$ be a real-valued random variable with c.d.f. $F_Z$.
For  $u\in [0,1]$, define
\[
F_X^-(u) = \max\{z: F_Z(z\!-) \leq u\}.
\]
Let $U$ be uniformly distributed over $[0,1]$. Then $F_Z^-(U)$ has the same distribution as $Z$.
\end{proposition}
\begin{proof}
Let $A_z = \{u\in [0,1]: F_Z^-(u) \leq z\}$. We show that
\begin{equation}
\label{eq:sampling.basic}
[0, F_Z(z)) \sub A_z \sub [0, F_Z(z)].
\end{equation}
Showing this will prove the lemma, since one has $\myP(U< F_Z(z)) = \myP(U \leq F_Z(z))= F_Z(z)$, showing that
\[
\myP(F_Z^-(U) \leq z) = \myP(U\in A_z) = F_Z(z).
\]

To prove \cref{eq:sampling.basic}, first assume that $u<F_Z(z)$. Take any $z'$ such that $F_Z(z'\!-) \leq u$. Then, necessarily, $z'\leq z$, since $z'>z$ would imply that $F_Z(z) \leq F_Z(z'\!-)$. This shows that $\max\{z': F_Z(z'\!-) \leq u\} \leq z$, i.e., $u\in A_z$. 

Now, take $u>F_Z(z)$. Because c.d.f.'s are right continuous, there exists $y>z$ such that $u>F_Z(y)$, which implies that $F_Z^-(u) \geq y$ and $u\not \in A_z$. 
\end{proof}

This proposition shows that one can generate random samples of a real-valued random variable $Z$ as soon as one can compute $F_Z^-$ and generate uniformly distributed variables. Note that, if $F_Z$ is strictly increasing, then $F_Z^- = F_Z^{-1}$, the usual function inverse.

\begin{remark}
\label{rem:sampling.real}
\begin{enumerate}
\item The proposition also shows how to sample from random variables taking values in finite sets. Indeed, if $Z$ takes values in $\tOm_Z = \{z_1, \ldots, z_n\}$ with $p_i = \myP(Z=z_i)$, sampling from $Z$ is equivalent to sampling from the integer valued random variable $\widetilde Z$ with $\myP(\widetilde Z = i) = p_i$. For this variable, $F_{\tilde Z}^-(u)$ is the largest $i$ such that $p_1 + \cdots + p_{i-1} \leq u$ (this sum being zero if $i=1$), which provides the standard sampling scheme for discrete probability distributions.
\item Some variables can be simulated more efficiently using special formulas. A well-known example is the Box-Muller formula to generate Gaussian distributions, that states that, if $U$ and $V$ are independent and uniformly distributed on $[0,1]$, then $Z = \sqrt{-2\log U}\cos(2\pi V)$ is standard Gaussian.
\end{enumerate}
\end{remark}

\section{Rejection sampling}
While the previous approach can be generalized to multivariate distributions, it quickly becomes unfeasible when the dimension gets large, excepting simple cases in which the variables are independent, or known functions of independent variables (which includes the Gaussian case, as seen in \cref{rem:sampling.real}). Rejection sampling is a simple algorithm that allows, in some cases, for the generation of samples from a complicated distribution based on repeated sampling of  a simpler one. 

Let us assume that we want to sample from a variable $Z$ taking values in $\CR_Z$, and that there exists a measure $\mu$ on $\CR_Z$ with respect to which the distribution of $Z$ is absolutely continuous, i.e., so that this distribution has a density $f_Z$ with respect to $\mu$. For example, $\CR_Z = \mR^d$, and $f_Z$ is the p.d.f. of $Z$ with respect to Lebesgue's measure. Assume that $g$ is another density function (with respect to $\mu$) from which it is ``easy'' to sample. Consider the following algorithm, which includes a function $a: z\mapsto a(z) \in [0,1]$ that will be specified later.
\begin{algorithm}[Rejection sampling with acceptance function $a$ and base p.d.f. $g$]
\label{alg:rejection}
\begin{enumerate}
\item Sample a realization $z$ of a random variable with p.d.f. $g$.
\item  Generate $b\in \{0,1\}$ with $\myP(b=1) = a(z)$.
\item If $b=1$, return $Z=z$ and exit. 
\item Otherwise, return to step 1. 
\end{enumerate}
\end{algorithm}

We have the following proposition.

\begin{proposition}
\label{prop:rejection}
Let $\rho = \int_{\CR_Z} g(z) a(z) \mu(dz)$ and assume that $\rho>0$. Then
\cref{alg:rejection} exits after a finite number of steps with probability one.  
Moreover, if $\tilde Z$ denote the random variable returned by this algorithm, the distribution of $\tilde Z$ has density (with respect to $\mu$)
\[
\tilde f(z) = \frac{g(z)a(z)}{\rho}.
\]
\end{proposition}
\begin{proof} 
Introduce a sequence of random variables $(Y_1, B_1), (Y_2, B_2), \ldots$ where the pairs $(Y_j, B_j)$ are mutually independent, $Y_j$ follows a distribution with density $g$, and $B_j$ takes values in $\{0,1\}$ and is such that $\myP(B_j=1\mid Y_j=z) = a(z)$. Then, an equivalent definition of \cref{alg:rejection} is
\[
\tilde Z = Y_T
\]
where $T = \min\{j: B_j=1\}$. We note that
\[
P(B_j=1) = \int_{\CR_Z} a(z) g(z) \mu(dz) = \rho
\]
so that $T$ follows a geometric distribution with parameter $\rho$ and is therefore almost surely finite.

For a function $F: \CR_Z \to [0, +\infty)$, one can write
\begin{align*}
\myE(F(\tilde Z)) &= \sum_{k=1}^\infty \myE(F(\tilde Z) \bfone_{T=k})\\
&= \sum_{k=1}^\infty \myE(F(Y_k) \bfone_{B_k=1} \bfone_{B_{k-1} = 0} \ldots \bfone_{B_1=0})\\ 
&= \sum_{k=1}^\infty \myE(F(Y_k) \bfone_{B_k=1}) \prod_{j=1}^{k-1} \myP(B_{j} = 0) \\
& = \sum_{k=1}^{\infty} \myE(F(Y_k) \myP(B_k=1\mid Y_k)) (1-\rho)^{k-1}\\
& = \sum_{k=1}^{\infty} (1-\rho)^{k-1} \int_{\CR_Z} F(z) a(z) g(z) \mu(dz) \\
& = \frac1\rho \int_{\CR_Z} F(z) a(z) g(z) \mu(dz),
\end{align*}
which completes the proof.
\end{proof}

As a consequence, in order to simulate $f_Z$, one must choose $a$ so that $f_Z(z)$ is proportional to $g(z) a(z)$, which (assuming that $g(z) > 0$ whenever $f_Z(z) > 0$) requires that $a(z)$ is proportional to $f_Z(z)/g(z)$. Since $a(z)$ must take values in $[0,1]$, but should otherwise be chosen as large as possible to ensure that fewer iterations are needed, one typically takes
\[
a(z) = \frac{f_Z(z)}{cg(z)}
\]
where $c = \max\{f_Z(z)/g(z): z\in\CR_Z\}$, which must therefore be finite. This fully specifies a rejection sampling algorithm for $f_Z$. Note that $g$ is free to choose (with the restriction that $f_Z(z)/g(z)$ must be bounded), and should be selected so that sampling from it is easy, and the coefficient $c$ above is not too large.

\section{Markov chain sampling}
When dealing with high-dimensional distributions, the constant $c$ in the previous procedure is typically extremely large, and the rejection-sampling algorithm becomes unfeasible, because it keeps rejecting samples for very long times. In such cases, one can use alternative simulation methods that iteratively update the variable $Z$ by making small changes at each step, resulting in a procedure that asymptotically converges to a sample of the target distribution. Such sampling schemes are usually described as Markov chains, leading to the name Markov-chain Monte Carlo (or MCMC) sampling. We therefore start our discussion with some basic results on the theory of Markov chains.


\subsection{Definitions}

We adopt a measure-theoretic formalism in this discussion and refer to \cref{sec:meas.prob} for a short introduction to the terms used here. 
Assume that we want to sample from a random variable that takes values in some (measurable) set $\CB = \CR_X$.\footnote{As always, we  assume that $\CB$ is a complete metric space with a dense countable subset, with the associated Borel $\sigma$-algebra.}

Markov chains generate sequences $X_0, X_1, \ldots$, and can be seen as a stochastic analogue of a recursive sequence
$X_{n+1} = \Phi(X_n)$. Such sequences are fully defined by the function $\Phi:\mathcal B\to \mathcal B$ and the initial value $X_0\in \mathcal B$. For Markov chains, we will need the probability distribution of $X_0$, that we will generally denote $P^0$ with $P^0(x) = \myP(X_0=x)$, and stochastic transition rules between $X_n$ and $X_{n+1}$. More precisely, we have the definition:
\begin{definition}
\label{def:markov.chain}
A stochastic process $X_0, X_1, \ldots$ is a Markov chain if and only if, for any $n\geq 0$,  the conditional distribution of $X_{n+1}$ given previous variables $X_0, \ldots, X_{n}$ only depend on $X_{n}$, i.e.,, for all measurable $A\subset \CB$, 
\begin{equation}
\label{eq:def.mc}
\myP(X_{n+1}\in A\mid X_0, \ldots, X_{n}) = \myP(X_{n+1}\in A\mid X_{n})
\end{equation}
almost surely (recall that both sides of \cref{eq:def.mc} are random variables defined on $\Omega$.)
\end{definition}

The conditional probabilities 
\[
P^{n,n+1}(x, A) = \myP(X_{n+1} \in A \mid X_n =  x),
\]
where $A\subset \CB$ is measurable, are the transitions probabilities of the Markov chain, in accordance with the following definition.
\begin{definition}
\label{def:trans.prob}
Let $F_1$ and $F_2$ be two sets equipped with $\sigma$-algebras $\CA_1$ and $\CA_2$. A transition probability from
$F_1$ to $F_2$ is a function $p:F_1\times \CA_2 \to [0,1]$  such that, for
all $x\in F_1$, the function $A \mapsto p(x, A)$ is a probability on $F_2$ and for all $A\in \CA_2$, the function $x\mapsto p(x, A)$, $x\in F_1$, is measurable.

When $F_2$ is discrete, the probabilities are fully specified by their values on singleton sets, and we will write $p(x,y)$ for $p(x, \{y\})$.
\end{definition}

 When $P^{n,
n+1}(x, \cdot)$ does not depend on $n$, the Markov chain is said to
be homogeneous. To simplify notation, we will restrict to homogeneous
chains (and therefore only write $P(x, A)$) (even though some of the chains used in MCMC sampling may be
non homogeneous\footnote{Non-homogeneous Markov chains can be considered as homogeneous ones
by extending the space $\CB$ on which they are defined. One can, in particular, always extend the chain to $\CB\ti
\mN$, and defining the transition probability
\[
\tilde p\big((x,n), A\times \{r\}\big) = \bfone_{r=n+1} p^{n,n+1}(x, A).
\]
Many of the convergence results considered in this chapter would not apply to this extension, however. Very often,  non-homogeneity results from the existence of a deterministic or stochastic auxiliary process, say, $(\xi_1, \xi_2, \ldots)$ taking values in a space $\tilde \CB$ such that $P^{n, n+1}(x,  A) = \tilde P((x, \xi_n), A)$ where $\tilde P$ is a transition probability from $\CB \times \tilde \CB$ to $\CB$. It is  often the case that the joint process $(X_n, \xi_n)$ can be seen as a homogeneous Markov chain on $\CB \times \tilde \CB$ to which the methods discussed here can apply.
}).

An important special case is when $\CB$ is countable, in which case one only needs to specify transition probabilities for singletons $A = \{y\}$, and we will write
\[
p(x, y) = P(x, \{y\}) = \myP(X_{n+1} =y \mid X_n =  x)
\]
for the p.m.f. associated with $P(x, \cdot)$.

Another simple situation is when $\CB = \mR^d$ and each $P(x, \cdot)$ has a p.d.f. that we will also denote as $p(x, \cdot)$. In this latter case, assuming that $P^0$ also has a p.d.f. that we will denote by $\mu_0$, the joint p.d.f. of $(X_0, \ldots, X_n)$ on $(\mR^d)^{n+1}$ is given by
\begin{equation}
\label{eq:mc.pdf}
f(x_0, x_1, \ldots, x_n) =  \mu_0(x_0) p(x_0, x_1)\cdots p(x_{n-1}, x_n).
\end{equation}
The same expression holds for the joint p.m.f. in the discrete case.

In the general case (invoking measure theory), the joint distribution is also determined by the transition probabilities, and we leave the derivation of  the expression to the reader. An important point is that, in both special cases considered above, and under some very mild assumptions in the general case, these transition probabilities also uniquely define the joint distribution of the infinite process $(X_0, X_1, \ldots)$ on $\CB^\infty$, which gives theoretical support to the consideration of asymptotic properties of Markov chains. 

In this discussion, we are interested in conditions ensuring that the chain asymptotically samples from a target probability distribution $Q$, i.e., that $\myP(X_n \in A)$ converges to $Q(A)$ (one says that $X_n$ converges in distribution to $Q$). In practice, $Q$ is given or modeled, and the goal is to determine the transition probabilities.
Note that the marginal distribution of $X_n$ is computed by integrating (or summing) \cref{eq:mc.pdf} with respect to $x_0, \ldots, x_{n-1}$. This is generally computationally challenging. 

Given a transition probability $P$ on $\CB$, we will use the notation, for a measurable function $f:\CB \to \mR$:
\[
Pf(x) = \int_\CB f(y) P(x, dy).
\]
If $Q$ is a probability distribution on $\CB$, it will also be convenient to write
\[
Qf = \int_\CB f(y) Q(dy).
\]
Also, if $\pi$ is any probability distribution on $\CB$, we will write
\[
E_\pi(f) = \int_\CB f(x) \pi(dx)
\]
for the expectation of $f$ with respect to $\pi$.

\subsection{Convergence}
We will denote $\myP_x(\cdot)$ the conditional distribution $\myP(\cdot \mid X_0 = x)$ and $P^n(x, A) = \myP_x(X_n\in A)$, which is a probability distribution on $\CB$. The goal of Markov Chain Monte Carlo sampling is to design the transition probabilities such that $P^n(x,A)$ converges to $Q(A)$ when $n$ tends to infinity. One furthermore wants to complete this convergence with a law of large numbers, ensuring that
\[
\frac1 n\sum_{k=1}^n f(X_k)\to \int_{\CB} f(x) Q(dx)
\] 
when $n\to \infty$, where $X_n$ is the generated Markov chain and $f$ is $Q$-integrable.

Let $D_{\mathrm{var}}$ be the total variation distance between probability measures introduced in \cref{sec:total.var}.
We will say that the Markov chain with transition $P$ asymptotically samples from $Q$ if 
\begin{equation}
\label{eq:mcmc.goal}
\lim_{n\to \infty} D_{\mathrm{var}}(P^n(x, \cdot),Q) = 0
\end{equation}
for $Q$-almost all $x\in \CB$. As we will see, he chain must satisfy specific conditions for this to be guaranteed.

\subsection{Invariance and reversibility}
\label{sec:mcmc.reversible}

\noindent{\bf Invariant distributions.}
If a Markov chain converges to $Q$, then $Q$ must be an ``invariant distribution,'' in the sense that, if $X_n \sim Q$ for some $n$, then so does $X_{n+1}$ and, as a consequence, all $X_m$ for $n\geq m$. This can be seen by writing
\[
P^{n+1}(x,A) = \myP_x(X_{n+1} \in A) = \myE_x(P(X_n, A)) = E_{P^n(x, \cdot)}(P(\cdot, A))
\]
If $P^n(x, \cdot)$ (and therefore also $P^{n+1}(x, \cdot)$) converges to $Q$, then passing to the limit above yields
\[
Q(A)= E_{Q}(P(\cdot, A))
\]
and this states that, if $X_n\sim Q$, then so does $X_{n+1}$. If $Q$ has a p.d.f. (resp. p.m.f.), say, $q$, this gives
\[
q(y) = \int_{\mR^d} p(x, y)q(x)dx,\quad (\text{resp. } q(y) = \sum_{x\in\CB} p(x, y)q(x)).
\]
So, if one designs a Markov chain with a target asymptotic distribution $Q$, the first thing to ensure is that $Q$ is invariant. However, while invariance leads to an integral equation for $q$, a stronger condition, called reversibility is easier to assess. 

\noindent{\bf Time reversal.}
Assume that $Q$ is invariant by $P$. Make the assumption that $P(x, \cdot)$ has a density $p_*$ with respect to $Q$ (this is, essentially, no loss of generality, see argument below), so that
\[
P(x, A) = \int_{A} p_*(x, y) Q(dy).
\]
 Taking $A= \CB$ above, we have
\[
\int_{\CB} p_*(x, y) Q(dy) = P(x, \CB) = 1
\]
but we also have, because $Q$ is invariant, that 
\[
\int_{\CB} p_*(x, y) Q(dx) = Q(\CB) = 1.
\]
One says that the  density is ``doubly stochastic'' with respect to $Q$.

Conversely, if a transition probability $P$ has a doubly stochastic density $p_*$ with respect to some probability $Q$ on $\CB$, then $Q$ is invariant by $P$, since
\begin{multline*}
\int_{\CB} P(x,A) Q(dx) = \int_{\CB} \int_A p_*(x,y) Q(dy) Q(dx) \\
= \int_{A} \int_{\CB} p_*(x,y) Q(dx) Q(dy) =   \int_A Q(dy) = Q(A).
\end{multline*}

The property of being doubly stochastic can be reinterpreted in terms of time reversal for Markov chains. Let $Q_0$ be an initial distribution for a Markov chain with transition $P$ (not necessarily invariant) so that, for any $n\geq 0$, the distribution of $X_n$ is $Q_n = Q_0P^n$. Fixing any $m>0$, we are interested in the reversed process $\tilde X_k = X_{m-k}$. We first notice that the conditional distribution of $X_n$ given its future $X_{n+1}, \ldots, X_m$ (with $n<m$) only depends on $X_{n+1}$, so that the reversed process is also Markov. Indeed, for any positive function $f: \CB \to \mR$, $g: \CB^{m-n} \to \mR$, one has, using the fundamental properties of conditional expectations and the fact that $(X_n)$ is a Markov chain,
\begin{align*}
\myE(f(X_n) g( X_{n+1}, \ldots, X_m) ) &=  \myE\left(\myE(f(X_n) g( X_{n+1}, \ldots, X_m)\mid X_n, X_{n+1})\right)\\
&=  \myE\left(f(X_n)\myE( g( X_{n+1}, \ldots, X_m)\mid X_n, X_{n+1})\right)\\
&=  \myE\left(f(X_n)\myE( g( X_{n+1}, \ldots, X_m)\mid X_{n+1})\right)\\
&=  \myE\left(\myE(f(X_n)\mid X_{n+1}) \myE( g( X_{n+1}, \ldots, X_m)\mid X_{n+1})\right)\\
&=  \myE\left(\myE(f(X_n)\mid X_{n+1}) g( X_{n+1}, \ldots, X_m)\right).
\end{align*}
This shows that 
\[
\myE(f(X_n) \mid X_{n+1}, \ldots, X_m) = \myE(f(X_n) \mid X_{n+1}),
\]
 which is what we wanted. To identify the conditional distribution of $X_n$ given $X_{n+1}$, we note that for any $x\in \CB$, the transition probability $P(x, \cdot)$ is absolutely continuous with respect to $Q_{n+1}$, since
\[
Q_{n+1}(A) = \int_{\CB} P(x, A) Q_n(dx)
\]
and the r.h.s. is zero only if  $P(x, A) = 0$ $Q_n$-almost everywhere \footnote{The ``almost everywhere'' statement a priori depends on $A$, but can be made independent of it under the mild assumption (that we will always make) that $\CB$ has a countable basis of open sets.}. This shows that there exists a function $r_{n+1}: \CB\times \CB \to \mR$ such that, for all $A$,
\[
P(x,A) = \int_A r_{n+1}(x,y) Q_{n+1}(dy).
\]
Given this point, one can write
\begin{align*}
\myE(f(X_n) g(X_{n+1})) &= \int_{\CB^2} f(x_n) g(x_{n+1}) P(x_n, dx_{n+1}) Q_n(dx_n) \\ 
&= \int_{\CB^2} f(x_n) g(x_{n+1}) r_{n+1}(x_n, x_{n+1}) Q_{n+1}(dx_{n+1}) Q_n(dx_n) \\ 
&= \int_{\CB} \left(\int_{\CB} f(x_n)  r_{n+1}(x_n, x_{n+1}) Q_n(dx_n)\right) g(x_{n+1}) Q_{n+1}(dx_{n+1})  
\end{align*} 
which shows that the conditional distribution of $X_n$ given $X_{n+1} = x_{n+1}$ has density $x_n \mapsto r_{n+1}(x_n, x_{n+1})$ relatively to $Q_{n}$. 

Note that, for discrete probabilities, one has
\[
r_{n+1}(x,y) = \frac{P(x,y)}{Q_{n+1}(y)}
\]
and 
\begin{equation}
\label{eq:reversed.general}
\myP(X_n = x \mid X_{n+1}= y) = \frac{Q_n(x) P(x,y)}{Q_{n+1}(y)}.
\end{equation}
The formula is identical if both $Q_0$ and $P(x, \cdot)$ have p.d.f.'s with respect to a fixed reference measure $\mu$ on $\CB$ (for example, Lebesgue's measure when $\CB = \mR^d$), denoting these p.d.f's by $q_0$ and $p(x, \cdot)$. Then, the p.d.f. of the  distribution of $ X_n$ given $X_{n+1}=y$ is 
\begin{equation}
\label{eq:reversed.pdf.gen}
\tilde p_n(y, x) = \frac{q_n(x) p(x,y)}{q_{n+1}(y)}
\end{equation}
where $q_n$ is the p.d.f. of $Q_n$. Note that the transition probabilities of the reversed Markov chain depend on $n$, i.e., the reversed chain is non-homogeneous in general.

\noindent{\bf Time reversal for stationary chains.}
However, if one assumes that $Q_0 = Q$ is invariant by $P$, then $Q_n = Q$ for all $n$ and therefore $r_n(x,y) = p_*(x,y)$, using the previous notation. In this case, the reversed chain  has transitions independent of $n$ and its transition probability has density
\[
\tilde p_*(x, y) = p_*(y, x)
\] 
with respect to $Q$. In the discrete case, letting $p(x,y) = P(X_{n+1} = y\mid X_n=x)$, we have $ p_*(x,y) = p(x,y) / Q(y)$, so that the reversed transition (call it $\tilde p$) is such that
\[
\frac{\tilde{p}(x,y)}{Q(y)} = \frac{p(y,x)}{Q(x)},
\]
i.e.,
\begin{equation}
\label{eq:reversed.discrete}
Q(y) p(y,x) = Q(x) \tilde p(x,y).
\end{equation}
One retrieves easily the fact that  if $p$ is such that there exists $Q$ and $\tilde p$ such that \cref{eq:reversed.discrete} is satisfied, then (summing the equation over $y$) $Q$ is an invariant probability for $p$. 

\noindent{\bf Reversible Markov chains.}
Let $Q$ be a probability on $\CB$. 
One says that the Markov chain (or the transition probability $p$) is $Q$-reversible if and only if $p(x, \cdot)$ has a density $p_*(x, \cdot)$ with respect to $Q$ such that $p_*(x,y) = p_*(y,x)$ for all $x,y\in \CB$. Since such a density is necessarily doubly stochastic, $Q$ is then invariant by $p$. Reversibility is equivalent to the property that,  whenever  $X_n \sim Q$, the joint distribution of $(X_n, X_{n+1})$ coincides with that of $(X_{n+1}, X_{n})$. Alternatively,
 $Q$-reversibility requires that for all $A, B\subset \CB$, 
\begin{equation}
\label{eq:reversibility}
\int_{A} P(z, B) Q(dz) = \int_{B} P(z, A) Q(dz) .
\end{equation}

In the discrete case, \cref{eq:reversibility} is equivalent to the ``detailed balance'' condition:
\begin{equation}
\label{eq:reversible.discrete}
Q(y) p(y,x) = Q(x) p(x,y).
\end{equation}

While $Q$ can be an invariant distribution for a Markov chain without that chain being $Q$-reversible, the latter property is  easier to ensure when designing transition probabilities, and most sampling algorithms are indeed reversible with respect to their target distribution. 
\begin{remark}
\label{rem:reversible.iterate}
A simple example of non-reversible Markov chain with invariant probability $Q$ is often obtained in practice by alternating two or more  $Q$-reversible transition probabilities. Assume, to simplify, that $\CB$ is discrete and that $p_1$ and $p_2$ are transition probabilities that satisfy \cref{eq:reversible.discrete}. Consider a composite Markov chain for which the transition from $X_n$ to $X_{n+1}$ consists in generating first $Y_n$ according to $p_1(X_n, \cdot)$ and then $X_{n+1}$ according to $p_2(Y_n, \cdot)$. The resulting composite transition probability is 
\[
p(x,y) = \sum_{z\in \CB} p_1(x,z) p_2(z, y).
\]
Trivially, $Q$ is invariant by $p$, since it is invariant by $p_1$ and $p_2$, but $p$ is not $Q$-reversible. Indeed, $p$ satisfies \cref{eq:reversed.discrete} with 
\[
\tilde p(x,y) = \sum_{z\in \CB} p_2(x,z) p_1(z, y).
\]
\end{remark}

\subsection{Irreducibility and recurrence}
While necessary, invariance is not sufficient for a Markov chain to converge to $Q$ in distribution. However, it simplifies the general ergodicity conditions compared to the  general theory of Markov chains \citep{nummelin2004general,revuz2008markov}, as summarized below, following \citep{tierney1994markov} (see also  \citep{athreya1996convergence}). We therefore assume that the transition probability $P$ is such that $Q$ is $P$-invariant.

One says that the Markov chain is $Q$-irreducible (or, simply, irreducible in what follows) if and only if, for all $z\in \CB$ and all (measurable) $B\sub \CB$ such that $Q(B)>0$, there exists $n>0$ with $\myP_z(X_n\in B) > 0$. (Irreducibility implies that $Q$ is the only invariant probability of the Markov chain.)


A Markov chain is called periodic if there exists $m>1$ such that $\CB$ can be covered by disjoint subsets $\CB_0, \ldots, \CB_{m-1}$ that satisfy $P(x, \CB_j) = 1$ for all $x\in \CB_{j-1}$ if $j\geq 1$ and all $x\in \CB_{m-1}$ if $j=0$. In other terms, the chain loops between the sets  $\CB_0, \ldots, \CB_{m-1}$. If such a decomposition does not exists, the chain is called aperiodic. 

A periodic chain cannot satisfy \cref{eq:mcmc.goal}. Indeed, periodicity implies that $P_x(X_n \in \CB_i) = 0$ for all $x\in \CB_i$ unless $i = 0\ (\mathrm{mod}\ d)$. Since the sets $\CB_i$ cover $\CB$, \cref{eq:mcmc.goal} is only possible with $Q=0$.  Irreducibility and aperiodicity are therefore necessary conditions for ergodicity. Combined with the fact that $Q$ is an invariant probability distribution, these conditions are also sufficient, in the sense that \cref{eq:mcmc.goal} is true for  $Q$-almost all $x$. (See \cite{tierney1994markov} for a proof.)

 Without the knowledge that the chain has an invariant probability, showing convergence usually requires showing that the chain is recurrent, which means that, for any set $B$ such that $Q(B) > 0$, the probability that, starting from $x$, $X_n\in B$ for an infinite number of $n$, written as $\myP_x(X_n \in B\ \mathit{i.o.})$ (for infinitely often) is positive for all $x\in E$ and equal to 1 $Q$-almost surely. The fact that irreducibility and aperiodicity combined with $Q$-invariance imply recurrence (or, more precisely, $Q$-positive recurrence \citep{nummelin2004general}) is an important remark that significantly simplifies the theory for MCMC simulation. Note that, by restricting $\CB$ to a suitable set of $Q$-probability 1, one can assume that $\myP_x(X_n \in B \ \mathit{i.o.}) = 1$ for all $x\in \CB$, which is called {\em Harris recurrence}.  If the chain is Harris recurrent, then \cref{eq:mcmc.goal} holds with $\mu_0 = \delta_x$ for all $x\in \CB$. \footnote{Harris recurrence is also associated with the uniqueness of right eigenvectors of $P$, that is functions $h: \CB \to \mR$ such that
\[
Ph(x) \defeq \int_{\CB} P(x, dy) h(y)  = h(x).
\]
Such functions are also called harmonic for $P$.
Because $P$ is a transition probability, constant functions are always harmonic. Harris recurrence, in the current context, is equivalent to the fact that every bounded harmonic function is constant.}

One says that $C\subset \CB$ is a ``small'' set if $Q(C)>0$ and there exists a triple
 $(m_0, \eps, \nu)$, with $\eps>0$ and $\nu$ a probability distribution on $\CB$,  such that
\[
P^{m_0}(x, \cdot) \geq \eps \nu(\cdot)
\]
for all $x\in C$.
A slightly different result, proved in \citep{athreya1996convergence}, replaces irreducibility by the property that there exists a small set  $C\subset \CB$ such that 
\[
\myP_x(\exists n: X_n\in C) > 0
\]
for $Q$-almost all $x\in \CB$.  One then replaces aperiodicity by the similar condition that the greatest common divisor of the set of integers $m$ such that  there exists $\epsilon_m$ with $P^{m}(x, \cdot) \geq \eps_m \nu(\cdot)$
for all $x\in C$ is equal to 1. These two conditions combined with $Q$-invariance also imply that \cref{eq:mcmc.goal} holds for $Q$-almost all $x\in \CB$.

\subsection{Speed of convergence}

It is also important to quantify the speed of convergence in \cref{eq:mcmc.goal}. Efficient algorithms typically have a geometric convergence speed, namely
\begin{equation}
\label{eq:geom.ergod}
D_{\mathrm{var}}(P^n_x, Q) \leq M(x) r^n
\end{equation}
for some $0\leq r < 1$ and some function $M(x)$, or uniformly geometric convergence speed, for which the function $M$ is bounded (or, equivalently, constant).  


A sufficient condition for geometric ergodicity is provided in \citet[Proposition 5.21]{nummelin2004general}.  Assume that the chain is aperiodic, Harris recurrent and that there exist $r>1$,  a small set $C$ and a ``drift function'' $h$ with
\begin{subequations}
\begin{equation}
\label{eq:drift.1}
\sup_{x\not \in C} (r\myE(h(X_{n+1})\mid X_n = x) - h(x)) < 0
\end{equation}
and
\begin{equation}
\label{eq:drift.2}
\sup_{x\in C} \myE(h(X_{n+1}) \bfone_{X_{n+1}\not \in C} \mid X_n = x) <\infty.
\end{equation}
\end{subequations}
Then the Markov chain is geometrically ergodic. Note that $ \myE(h(X_{n+1})\mid X_n = x) = Ph(x)$. \Cref{eq:drift.1,eq:drift.2} can be summarized in a single equation \citep{meyn2012markov}, namely
\begin{equation}
\label{eq:drift.12}
Ph(x) \leq \beta h(x) + M \bfone_C(x)
\end{equation}
with $\beta = 1/r < 1$ and $M\geq 0$.

\subsection{Models on finite state spaces}

Uniform geometric ergodicity is implied by the simple condition that the whole set $\CB$ is small, requiring in a uniform lower bound, for some probability distribution $\nu$,
\begin{equation}
\label{eq:ergodic.unif}
P^{m_0}(x, \cdot) \geq \eps \nu(\cdot)
\end{equation}
for all $x\in \CB$. Such uniform conditions usually require strong restrictions on the space $\CB$, such as compactness or finiteness. 

To illustrate this consider the case in which the set $\CB$ is finite. Assume, to simplify,  that $Q(x)>0$ for all $x\in \CB$ (one can restrict the Markov chain to such $x$'s otherwise).
 Arbitrarily labeling elements of $\CB$ as $\CB = \{x_1, \ldots, x_N\}$, we can consider $p(x,y)$ as the coefficients of a matrix
$\bfP = (p(x_k, x_l), k,l=1, \ldots, N)$. Such a matrix, which has
non-negative entries and  row sums equal to 1, is called a stochastic
matrix.

We will  denote the $n$th power of $\bfP$ as $\bfP^n = (\pe p n(x_k,x_l), k,l=1, \ldots, N)$. One immediately sees that irreducibility is equivalent to the fact that, for all $x, y\in \CB$, there exists $m$ (that may depend of $x$ and $y$) such that $\pe p m(x,y) >0$. One can furthermore 
show that the chain is irreducible and aperiodic if one can choose $m$ independent of $x$ and $y$ above, that is, if there exists $m$ such that $\bfP^m$ has positive coefficients. This condition clearly implies uniformly geometric ergodicity, which is therefore valid for all irreducible and aperiodic Markov chains on finite sets. We emphasize this simple, but important, result in the following proposition.
\begin{proposition}
\label{prop:ergodic.finite}
A Markov chain on a finite state space is uniformly ergodic as soon as it is irreducible and aperiodic.
\end{proposition}

This result can also be deduced from properties of matrices with non-negative or positive coefficients. The Perron-Frobenius theorem \cite{horn2012matrix} states that the eigenvalue 1 (associated with the eigenvector $\dsone$) is the largest, in modulus, eigenvalue of   a stochastic matrix $\tilde{\bfP}$ with positive entries, that it has multiplicity one and that all other eigenvalues have a modulus strictly smaller that one. If $\bfP^m$ has positive entries, this implies that all the eigenvalues of $(\bfP^m - \dsone Q)$ (where $Q$ is considered as a row vector) have modulus strictly less than one. This fact can then be used to prove uniformly geometric ergodicity. 

\subsection{Examples on $\mR^d$}
\label{sec:mcmc.rd}
To take a geometrically ergodic example that is not uniform, consider the simple random walk provided by the iterations
\[
X_{n+1} = \rho X_n + \tau^2 \epsilon_n
\]
where $\epsilon_n \sim \CN(0, \Id[d])$, $\tau^2>0$ and $0<\rho < 1$. One shows easily by induction that the conditional distribution of $X_n$ given $X_0=x$ is Gaussian with mean $m_n = \rho^n x$ and covariance matrix $\sigma_n^2 \Id[d]$  with
\[
\sigma^2_n = \frac{1-\rho^{2n}}{1-\rho^2}\tau^2.
\]
In particular, the distribution $Q = \CN(0, \sigma^2_\infty\Id[d])$, with $\sigma^2_\infty = \tau^2 /(1-\rho^2)$, is invariant. Estimates on the variational distances between Gaussian distributions, such as those provided in  \citet{devroye2018total}, can then be used to show that 
\[
D_{\mathrm{var}}(P^n_x, Q) \leq M(x) \rho^n
\]
where $M$ grows linearly in $x$ but is not bounded.

Situations in which one can, as above, compute the probability distributions of $X_n$ are rare, however, and proving geometric convergence is significantly more difficult than for finite-state chains. For chains on $\mR^d$ (or, more generally, locally compact metric spaces), the drift function criterion \cref{eq:drift.12} can be used. For Feller-continuous chains, this criterion takes a relatively simpler form. Recall that we have defined \[
Ph(x)  = \myE(h(X_{n+1})\mid X_n = x) = \int_{\mR^d} h(y) P(x, dy).
\]

\begin{definition}
\label{def:feller.continuous}
A Markov chain on $\mR^d$  is Feller continuous if $Ph(x)$ is continuous for all continuous $h: \mR^d \to \mR$.
\end{definition}
In this definition, one can replace $\mR^d$ by any topological space.  

Feller continuity is true, for example, if $P(x, \cdot)$ has a p.d.f. with respect to Lebesgue's measure which is continuous in $x$. 

In such a situation, one can check that compact sets are small sets, and \cref{eq:drift.12} can be restated as follows. 
\begin{proposition}
\label{prop:feller.ergodic}
Let $P$ be feller continuous, aperiodic and irreducible.
Assume that there exists: (i)  a function $h: \mR^d\to \mR$ with $h(x) \geq 1$ and such that the sub-level sets $\{x: h(x)\leq c\}$ are compact, (ii) a compact set $C\subset \mR^d$ and (iii) positive constants $\beta< 1$ and $b$, such that, for all $x\in \mR^d$,
\begin{equation}
\label{eq:geom.drift}
P h(x) \leq \beta h(x) + b \bfone_C(x).
\end{equation}
Then, the Markov chain with transitions $P$ is geometrically ergodic.
\end{proposition}

As an example, consider the Markov chain defined by
\begin{equation}
\label{eq:langeving.discrete}
X_{n+1} =X_n  -\delta \nabla H(X_n) + \tau \epsilon_{n+1}
\end{equation}
where $\epsilon_2, \epsilon_2,\ldots$ are i.i.d. standard $d$-dimensional Gaussian variables and $H:\mR^d \to \mR$ is $C^2$. We prove the following result.
\begin{proposition}
\label{prop:langevin.discrete}
Assume that $H$ is $L$-$C^1$ for some $L>0$ (c.f. \cref{def:lck}) and   that $|\nabla H(x)|$ tends to infinity when $x$ tends to infinity.
Then, for small enough $\delta$, the Markov chain defined by \cref{eq:langeving.discrete} is geometrically ergodic.
\end{proposition}

\begin{proof}
This chain is clearly irreducible (with respect to Lebesgue's measure) and aperiodic since there is a positive probability to land in one step in any set of positive Lebesgue's measure. The fact that $|\nabla H(x)|$ tends to infinity implies that the sub-level sets $\{x: |\nabla H(x)| \leq c\} $ are compact for any $c>0$.  We want to show that, if $\delta$ is small enough, \cref{eq:geom.drift} holds for $h(x) = \exp(m H(x))$ and $m$ small enough.

One has
\[
Ph(x) = \frac{1}{(2\pi\tau^2)^{d/2}}\int_{\mR^d} h(y) e^{-\frac{1}{2\tau^2} |y - x  +\delta \nabla H(x)|^2} dy = \frac{1}{(2\pi)^{d/2}}\int_{\mR^d} h(x  -\delta \nabla H(x) + \tau u) e^{-\frac{ |u|^2}2} dy.
\]
We first compute an upper bound of
\[
g(x, u) = m H(x  -\delta \nabla H(x) + \tau u) - \frac{ |u|^2}2.
\]
Using the $L$-$C^1$ property, we have
\begin{align*}
g(x,u) &\leq mH(x) +m (-\delta\nabla H(x) + \tau u)^T \nabla H(x) + \frac{mL}{2}|\delta\nabla H(x) - \tau u|^2 - \frac{ |u|^2}2\\
& = m H(x) - m\delta( 1 - \delta L/2)|\nabla H(x)|^2 + m\tau (1 - \tau L)\nabla H(x)^T u - \frac{1 - mL\tau^2}{2} |u|^2 \\
&= m H(x)  - \frac{1 - mL\tau^2}{2} \left|u - \frac{m\tau (1-\tau L)}{1-mL\tau^2}\nabla H(x)\right|^2\\
& \qquad -m\left(\delta( 1 - \delta L/2) - \frac{m\tau^2 (1-\tau L)^2}{2(1-mL\tau^2)}\right)|\nabla H(x)|^2
\end{align*}
Assume that $mL\tau^2 \leq 1$.
It follows that
\begin{multline*}
Ph(x) = \frac{1}{(2\pi)^{d/2}}\int_{\mR^d} e^{g(x,u)}\, du \leq \frac{h(x)}{(1 - mL\tau^2)^{d/2}}\\
 \exp\left( -m\left(\delta( 1 - \delta L/2) - \frac{m\tau^2 (1-\tau L)^2}{2(1-mL\tau^2)}\right)|\nabla H(x)|^2\right)
 \end{multline*}
Using this upper bound, we see that \cref{eq:geom.drift} will hold if one first chooses $\delta$ such that $\delta L < 2$, then $m$ such that $mL\tau^2 < 1$ and 
\[
\frac{m\tau^2 (1-\tau L)^2}{2(1-mL\tau^2)} < \delta( 1 - \delta L/2),
\]
and finally choose $C = \{x: |\nabla H(x)| \leq c\}$ where $c$ is large enough so that 
\[
\frac{1}{(1 - mL\tau^2)^{d/2}}
 \exp\left( -m\left(\delta( 1 - \delta L/2) - \frac{m\tau^2 (1-\tau L)^2}{2(1-mL\tau^2)}\right) c^2\right) < 1.
 \]
 \end{proof}
 
 Note that this Markov chain is not in detailed balance. Since $P(x, \cdot)$ has a p.d.f., being in detailed balance  requires the ratio $p(x,y)/p(y,x)$ to simplify as a ratio $q(y)/q(x)$ for some function $q$, which does not hold. However, we can identify the invariant distribution approximately with small $\delta$ and $\tau$, that we will assume to satisfy $\tau = a\sqrt{\delta}$ for a fixed $a >0$, with $\delta$ a small number.
 We can write
\begin{align*}
p(x,y) &= \frac{1}{(2\pi\tau^2)^{d/2}}
 \exp\left(-\frac{1}{2\tau^2} |y - x  +\delta \nabla H(x)|^2\right) \\
 &=  \frac{1}{(2\pi\tau^2)^{d/2}}
 \exp\left(-\frac{1}{2\tau^2} |y - x|^2  -\frac{\delta}{\tau^2} (y-x)^T \nabla H(x) - \frac{\delta^2}{2\tau^2} |\nabla H(x)|^2\right).
\end{align*}
 If $q$ is a density, we have
\begin{align*}
qP(y) &= \int_{\mR^d} q(x) p(x,y) dx\\ 
&= \frac{1}{(2\pi)^{d/2}} \int_{\mR^d} q(y +a \sqrt\delta u ) \exp\left(-\frac{1}{2} |u|^2  + \frac{\sqrt\delta}{a} u^T \nabla H(y + a\sqrt\delta  u) - \frac{\delta}{2a^2} |\nabla H(y+a \sqrt\delta  u)|^2\right) du
\end{align*} 
Make the expansions:
\[
q(y +a \sqrt\delta u ) = q(y) + a \sqrt\delta  \nabla q(y)^T u + \frac{a^2\delta}{2} u^T \nabla^2 q(y) u + o(\delta |u|^2)
\]
and
\begin{multline*}
\exp\left( \frac{\sqrt\delta}{a} u^T \nabla H(y + a\sqrt\delta u) - \frac{\delta}{2a^2} |\nabla H(y+a \sqrt\delta u)|^2\right) \\
= 1 + \frac{\sqrt\delta}{a} u^T \nabla H(y) - \frac{\delta}{2a^2} |\nabla H(y)|^2 + \delta u^T \nabla^2 H(u) u + \frac{\delta}{2a^2} (u^T \nabla H(y))^2 + o(\delta |u|^2).
\end{multline*}
Taking the product and using the fact that 
$(2\pi)^{-d/2} \int_{\mR^d} u \exp(-|u|^2/2)du = 0$ and that 
$(2\pi)^{-d/2} \int_{\mR^d} u^TAu \exp(-|u|^2/2)du = \trace(A)$ for any symmetric matrix $A$, we can write:
\[
qP(y) = q(y) + \delta \left(\frac{a^2}{2} \Delta q(y) + \nabla H(y)^T \nabla q(y)  + q(y) \Delta H(y)\right) +o (\delta).
\]  
This indicates that, if $q$ is invariant by $P$, it should satisfy
\[
\frac{a^2}{2} \Delta q(y) + \nabla H(y)^T \nabla q(y)  + q(y) \Delta H(y) = o(1).
\]
The partial differential equation
\begin{equation}
\label{eq:kolm.invariant}
\frac{a^2}{2} \Delta q(y) + \nabla H(y)^T \nabla q(y)  + q(y) \Delta H(y) =0
\end{equation}
is satisfied by the function 
$ y\mapsto e^{-\frac{2H(y)}{a^2}}$. Assuming that this function is integrable, this computation suggests that, for small $\delta$, the Markov chain approximately samples from the probability distribution 
\[
q_0 = \frac{1}{Z}e^{-\frac{2H(x)}{a^2}}.
\]
This is further discussed in the next remark that involves stochastic differential equations. We will also present a correction of this Markov chain that samples from $q_0$ for all $\delta$ in \cref{sec:continuous.met}.

\begin{remark}[Langevin equation]

\label{rem:langevin}
This chain is indeed the Euler discretization \citep{kloeden1992numerical} of the stochastic differential equation,
\begin{equation}
\label{eq:langevin}
dx_t = - \nabla H(x_t) dt + a dw_t
\end{equation}
where $w_t$ is a Brownian motion.
Under general hypotheses, this stochastic diffusion equation, called a {\em Langevin equation}, indeed converges in distribution to $q_0(x)$. 
\footnote{Providing a rigorous account of the  theory of stochastic differential equations is beyond our scope, and we refer the reader to the many textbooks on the subject, such as \citet{mckean1969stochastic,ikeda1981stochastic,ethier1986markov} (see also \citet{berglund2021long} for a short introduction).} 

Such diffusions are continuous-time Markov processes $(X_t, t\geq 0)$, which means that the probability distribution of $X_{t+s}$ given all events  before and including time $s$  only depends on $X_s$ and is provided by a transition probability $P_t$, with
\[
\myP(X_{t+1} \in A \mid X_s = x) = P_t(x, A).
\]
Similarly to deterministic ordinary differential equations, one shows that under sufficient regularity conditions (e.g., $\nabla H$ is $C^1$), equations such as \cref{eq:langevin} have solutions up to some positive (random) explosion time, and that this explosion time is finite under additional conditions that ensure that $|\nabla H(x)|$ does not grow too fast when $x$ tends to infinity. 

If $\phi$ is a smooth enough function (say, $C^2$, with compact support), the function $(t, x) \mapsto P_t\phi(x)$ satisfies the partial differential equation, called Kolmogorov's backward equation,
\[
\partial_t P_t\phi(x) = - \nabla H(x)^T\nabla P_t\phi(x) + \frac {a^2}2 \Delta P_t\phi(x)
\] 
with initial condition $P_0\phi(x) = \phi(x)$. If $P_t(x, \cdot)$ has at all times $t$ a p.d.f. $p_t(x, \cdot)$, then this p.d.f. must satisfy the forward Kolmogorov equation:
\[
\partial_t p_t(x, y) = \nabla_2 \cdot (\nabla H(y) p_t(x,y)) + \frac {a^2}2 \Delta_2 p_t(x,y)
\] 
where $\nabla_2$ and $\Delta_2$ indicate differentiation with respect to the second variable ($y$). (Recall that $\Delta f$ denotes the Laplacian of $f$.)  
Moreover, if $Q$ is an invariant distribution with p.d.f. $q$, it satisfies the equation 
\[
\nabla\cdot (q \nabla H) + \frac {a^2}2 \Delta q(y) = 0.
\]
Noting that $\nabla\cdot (q \nabla H) = \nabla q^T \nabla H + q \Delta H$, we retrieve \cref{eq:kolm.invariant}. Convergence properties (and, in particular, geometric convergence) of the Langevin equation to its limit distribution are studied in \citet{roberts1996exponential}, using methods introduced in \citet{meyn1993stabilityII,meyn1993stabilityIII,meyn2012markov}
\end{remark}

\section{Gibbs sampling}

\subsection{Definition}
The Gibbs sampling algorithm \citep{geman1984stochastic} was introduced to sample from a distribution on large sets for which direct sampling is intractable and rejection samping is inefficient. It generates a Markov chain that converges (under some hypotheses) in distribution to this target probability. A general version of this algorithm is described below. 

Let $Q$ be  a probability distribution on $\CB$. Consider a finite family $U_1, \ldots, U_K$ of random variables defined on $\CB$ with values in measurable spaces $\CB'_1, \ldots, \CB'_K$. Let $Q^i = Q_{U^i}$ denote the image of $Q$ by $U^i$, defined by $Q^i(B_i) = Q(U_i\in B_i)$ for $B_i \subset \mathcal B_i$. Also, assume that there exists, for all $i$, a regular family of conditional probabilities for $Q$ given $U_i$, defined as a collection of transition probabilities $(u_i, A) \mapsto Q_i(u_i, A)$ for $u_i\in \CB_i$ and $A\subset \CB$, that satisfy
\[
\int_A g(U_i(x)) Q(dx) = \int_{\CB_i} Q_i(u_i, A) g(u_i) Q^i(du_i)
\]
for all nonnegative measurable functions $g: \CB_i \to \mR$. In simpler terms, $Q_i(u_i, A)$ determine a consistent set of conditional probabilities $Q(\ccdot \mid U_i = u_i)$. For discrete random variables (resp. variables with p.d.f.'s on $\mR^d$), they are just elementary conditional probabilities.

We then consider the following algorithm.
\begin{algorithm}[Gibbs sampling]
\label{alg:gibbs.sampling}
 Initialize the algorithm with some $z(0) = z_0\in \CB$ and iterate the following two update steps given a current $z(n)\in\CB$:
\begin{enumerate}[label= (\arabic*), wide=0.5cm]
    \item Select $j\in \{1, \ldots, K\}$ according to some pre-defined scheme, i.e., at random according to a probability distribution $\pi^{(n)}$ on the set $\{1, \ldots, K\}$. 
    \item Sample a new value $z(n+1)$ according to the probability distribution $Q_j(U_j(z(n)), \cdot)$.
\end{enumerate}
\end{algorithm}
One typically chooses the probability distribution in step 1 equal to the uniform distribution on $\{1, \ldots, K\}$ (in which case it is independent on $n$), or to $\pi^{(n)} = \de_{j_n}$ where $j_n = 1 + (n \ \mathrm(mod)\  K)$ (periodic scheme). Strictly speaking, Gibbs sampling is a Markov chain if $\pi^{(n)}$ does not depend on $n$, and we will make this simplifying assumption in the rest of our discussion (therefore replacing  $\pi^{(n)}$ by $\pi$). One obvious requirement for the feasibility of the method is that step 2 can be performed efficiently since it must be repeated a very large number of times.

One can see that the Markov chain generated by this algorithm is $Q$-reversible. Indeed, assume that $X_n\sim Q$. For any (measurable) subsets $A$ and $B$ in $\CB$, one has, using the definition of conditional expectations, 
\begin{equation}
\label{eq:gs.theory.1}
\myP(X_n\in A, X_{n+1}\in B) = \sum_{i=1}^K \myE\big(\bfone_{Z\in A} Q_i(U_i(Z), B)\big) \pi(i).
\end{equation}
Now, for any $i$,
\begin{align*}
\myE\big(\bfone_{Z\in A} Q_i(U_i(Z), B)\big) &=
\int_A Q_i(U_i(z), B) Q(dz)\\
&= \int_{\CB_i} Q_i(u_i, A) Q_i(u_i, B) Q^i(du_i)
\end{align*}
which is symmetric in $A$ and $B$.

Note that, in the discrete case
\begin{equation}
\label{eq:gs.discrete}
P(z, \tilde z) = \sum_{i=1}^n \pi (i) \dfrac{Q(\tilde z)\bfone_{U^i(\tilde z) = U^i(z)}}{\sum_{z': U^i(z') = U^i(z)} Q(z')}
\end{equation}
and the relation $Q(z) P(z, \tilde z) = Q(\tilde z) P(\tilde z, z)$ is obvious.

The conditioning variables $U_1, \ldots, U_K$ should ensure, at least, that the associated Markov chain is irreducible and aperiodic. For irreducibility, this requires 
that $Z$ can visits $Q$-almost all elements of $\CB$ by a sequence of steps that leave one of the $U_i$'s invariant.

\begin{remark}
\label{rem:gibbs.sampling.1}
In the standard version of Gibbs sampling, $\CB$ is a product space $\CB_1 \times \cdots \times \CB_K$, and
\[
\CB'_j = \CB_1 \times \cdots \times \CB_{j-1} \times \CB_{j+1} \times \cdots\times \CB_K.
\]
One then takes
$U_j(z^{(1)}, \ldots, z^{(K)}) =  (z^{(1)}, \ldots, z^{(i-1)}, z^{(i+1)}, \ldots, z^{(K)})$. In other terms, step 2 in the algorithm replaces the current value of $z^{(j)}(n)$ by a new one sampled from the conditional distribution of $Z^{(j)}$ given the current values of $z^{(i)}(n)$, $i\neq j$. 
\end{remark}

\begin{remark}
\label{rem:gibbs.sampling.2}
We have considered a fixed number of conditioning variables, $U_1, \ldots, U_K$, for simplicity, but the same analysis can be carried on if one replaces $U_j$ by a function $U: (x, \theta)\mapsto U_\theta (x)$ defined on a product space $\CB\times \Theta$, taking values in some space $\tilde \CB$, where $\Theta$ is a probability space equipped with a probability distribution $\pi$ and $\CB$ is measurable.  The previous discussion corresponds to $\Theta = \{1, \ldots, K\}$ and $\CB = \bigcup_{i=1}^K \{i\}\times \CB_i$ (so that $U_i(x)$ is replaced by $(i, U_i(x))$).

One may then define $Q^\theta$ as the image of $Q$ by $U_\theta$ and let $Q_\theta(u, A)$ provide a version of $Q(A\mid U_\theta = u)$. The only change in the previous discussion (besides using $\theta$ in index) is that \cref{eq:gs.theory.1} becomes 
\[
\myP(X_n\in A, X_{n+1}\in B) = \int_\Theta \myE\big(\bfone_{Z\in A} Q_\theta(U_\theta(Z), B)\big) \pi(d\theta).
\]
\end{remark}

\begin{remark}
\label{rem:gibbs.sampling.3}
Using notation from the previous remark, and allowing $\pi=\pe \pi n$ to depend on $n$, it is possible to allow $\pe\pi n$ to depend on the current state $z(n)$ using the following construction.


For every step $n$, assume that  there exists a subset $\Theta_n$ of $\Theta$ such that  
$ \pe \pi n(z, \Theta_n) = 1$ and that, for all $\theta\in \Theta_n$, $\pe\pi n$ can be expressed in the form
\[
\pe \pi n(z,\cdot) = \pe\psi n_\theta(U_\theta(z), \cdot)
\]
for some transition probability $\pe\psi n_\theta$ from $\CB_\theta$ to $\Theta_n$. The resulting chain remains $Q$-reversible, since
\begin{align*}
\myP(X_n\in A, X_{n+1}\in B) &= \int_{\CB} \int_{\Theta_n} \bfone_{z\in A} Q_\theta(U_\theta(z), B) \pe \pi n(z, d\theta) Q(dz)\\
&= \int_{\Theta_n} \int_{\CB} \bfone_{z\in A} Q_\theta(U_\theta(z), B) \pe {\psi_\theta} n(U_\theta(z), d\theta) Q(dz)\\
&= \int_{\Theta_n} \int_{\tilde \CB} Q_\theta(u, A) Q_\theta(u, B) \pe \psi n_\theta(u, d\theta) Q^\theta(du).
\end{align*}

\end{remark}

\subsection{Example: Ising model}
\label{sec:gibbs.ising}
We will see  several examples of applications of Gibbs sampling in the next few chapters. Here, we consider a special instance of Markov random field (see \cref{chap:mrf}) called the Ising model. For this example,  $\CB = \{0,1\}^L$, and
\[
q(z) = \frac1{C} \exp\left( \sum_{j=1}^L \alpha \pe zj + \sum_{i,j=1, i< j}^L \beta_{ij} \pe z i \pe z j\right).
\]
Note that, although $\CB$ is a finite set, its cardinality, $2^L$, is too large for the enumerative procedure described in \cref{sec:gen.samp} to be applicable as soon as $L$ is, say, larger than 30. In practical applications of this model, $L$ is orders of magnitude larger, typically in the thousands or tens of thousands.

We here apply standard Gibbs sampling, as described in \cref{rem:gibbs.sampling.1}, defining $\CB_j = \{0,1\}$ and 
\[
U_i(z^{(1)}, \ldots, z^{(L)}) =  (z^{(1)}, \ldots, z^{(i-1)}, z^{(i+1)}, \ldots, z^{(L)}).
\]
 The conditional distribution of $Z^{(j)}$ given $U_j(z)$ is a Bernoulli distribution with parameter
\[
q_{Z^{(j)}}(1\mid U_j(z)) = \frac{\exp(\alpha + \sum_{j'=1, j'\neq j}^L \beta_{jj'} \pe z {j'})}{1 + \exp(\alpha + \sum_{j'=1, j'\neq j}^L \beta_{ij} \pe z j)}
\]
(taking  $\beta_{jj'} = \beta_{j'j}$ for $j>j'$).  
Gibbs sampling for this model will generate a sequence of variables $Z(0), Z(1), \ldots$ by fixing $Z(0)$ arbitrarily and, given $Z(n) = z$, applying the two steps:
\begin{enumerate}[label=(\arabic*), wide=0.5cm]
    \item Select $j\in \{1, \ldots, L\}$  at random according to a probability distribution $\pi^{(n)}$ on the set $\{1, \ldots, L\}$. 
    \item Sample a new value $\zeta\in \{0,1\}$ according to the Bernoulli distribution with parameter $q_{Z_n^{(j)}}(1\mid U_j(z))$, and set $Z^{(j)}(n+1) = \zeta$ and $Z^{(j')}(n+1) = Z^{(j')}(n)$ for $j'\neq j$.
\end{enumerate}

Let us now consider the Ising model with fixed total activation, namely the previous distribution conditional to $S(z) \defeq \pe z 1 + \cdots + \pe z L = h$ where $0  < h < L$.  The distribution one wants to sample from now is
\[
q_h(z) = \frac1{C_h} \exp\left( \sum_{j=1}^L \alpha \pe zj + \sum_{i,j=1, i< j}^L \beta_{ij} \pe z i \pe z j\right) \bfone_{S(z) = h}.
\]
In that case, the previous choice for the one-step transitions does not work, because fixing all but one coordinate of $z$ also fixes the last one (so that the chain would not move from its initial value and would certainly not be irreducible). One can however fix all but two coordinates, therefore defining
\[
U_{ij}(z^{(1)}, \ldots, z^{(L)}) =  (z^{(1)}, \ldots, z^{(i-1)}, z^{(i+1)}, \ldots, z^{(j-1)}, z^{(j+1)}, \ldots, z^{(L)})
\]
and $B_{ij} = \{0,1\}^2$. If $U_{ij}(z)$ is fixed, the only acceptable configurations are $z$ itself and the configuration $z'$ deduced from $z$ by switching the value of $\pe z i$ and $\pe z j$. Thus, there is no possible change is $\pe z i = \pe z j$. If $\pe z i \neq \pe z j$, then the probability of flipping the values of $\pe z i$ and $\pe z j$ is
$q_h(z') / (q_h(z) + q_h(z'))$.

\section{Metropolis-Hastings}
\label{sec:met.hast}

\subsection{Definition}
The Metropolis-Hastings algorithm \citep{metropolis1953equation,hastings1970monte} is a generic MCMC method that is defined as follows. Let as before $Q$ denote the distribution to sample from on the set $\CB$. Let $G$ be a transition probability from $\CB$ to itself. Define the distribution $Q\otimes G$ on $\CB\times \CB$ as the law of $(Z, Z')$ where $Z\sim Q$ and the conditional distribution of $Z'$ given $Z=z$ is $G(z, \cdot)$, so that
\[
\myE(f(Z, Z')) = \int_{\CB\times\CB} f(z,z')  (Q\otimes G)(\dz, dz') = \int_\CB \int_\CB f(z, z') G(z, dz') Q(dz).
\]
(In other terms $Q\otimes G$ is the distribution of the first two steps of a Markov chain with initial distribution $Q$ and transition $G$.)

We will assume that $G$ satisfies the following weak symmetry property (with respect to $Q$). Let $\tau: \CB\times \CB \to \CB\times \CB$ be defined by $\tau(z,z') = (z',z)$. We will assume that the image $\tau_{\#} (Q\otimes G) $ absolutely continuous with respect to $Q\otimes G$, i.e., is $\myP( (Z, Z') \in A) = 0$ (with $A\subset \CB\times \CB$), then $\myP( (Z', Z) \in A) = 0$.
This requires that $\tau_{\#} (Q\otimes G)$ has a density with respect to $Q\otimes G$, that we will denote by $\rho$, which satisfies
 \[
\int_{\CB\times\CB} f(z', z) P(z, dz') Q(dz) = 
\int_{\CB\times\CB} f(z, z') \rho(z,z') P(z, dz') Q(dz)
 \]
 for all measurable $f:\CB \to [0, +\infty)$.

 Note that, replacing $f$ by $g(z,z') = f(z, z') \rho(z,z')$ in this identity, one gets
 \begin{align*}
\int_{\CB\times\CB} f(z, z') \rho(z', z) \rho(z,z') P(z, dz') Q(dz)& = \int_{\CB\times\CB} f(z', z) \rho(z, z') P(z, dz') Q(dz) \\
&= 
 \int_{\CB\times\CB} f(z, z') P(z, dz') Q(dz)
 \end{align*}
 showing that $\rho(z', z) \rho(z,z')=1$ $(Q\otimes G)$-almost surely.
 
Define the acceptance function
\[
a(z,z') = \min(1, \rho(z,z')).
\]
The Metropolis-Hastings (MH) algorithm then computes the following sequence of random variables.

\begin{algorithm}[Metropolis-Hastings]
\label{alg:met.hast}
Initialize the algorithm with $Z(0) \sim \mu(0)$ (the initial distribution). Define the iterations
\[
Z_{n+1} = Z_n (1-B_n) + \tilde Z_{n+1} B_{n+1}
\]
where, 
\begin{enumerate}[label=(\roman*)]
\item Conditionally to $Z_n$ (and all previous variables), $\tilde Z_{n+1} \sim G(Z_n, \, \cdot\,)$.
\item Conditionally to $Z_n, \tilde Z_{n+1}$ (and all previous variables), $B_{n+1}$ follows a Bernoulli distribution with parameter $a(Z_n, \tilde Z_{n+1})$. 
\end{enumerate}
\end{algorithm}
In other terms, the transition from  $Z_n$ consists in proposing a new value $\tilde Z_{n+1}$ according to the chain with transitions $G$, and accepting the change with probability $a(Z_n, \tilde Z_{n+1})$. 
Note that the transition probability for the MH algorithm is
\[
P(z, \cdot) = a(z, \cdot) G(z, \cdot)  + \left(1-\bar a(z)\right) \delta_z. 
\]

One then has the proposition.
\begin{proposition}
\label{prop:mh}
The Markov chain defined by \cref{alg:met.hast} is $Q$-reversible.
\end{proposition}
\begin{proof}
One needs to show that for any measurable $f: \CB \times \CB \to [0, +\infty)$, one has 
\[
\myE(f(Z_{n+1}, Z_n) = \myE(f(Z_n, Z_{n+1})
\]
as soon as $Z_n \sim Q$. In the following computation, we use the fact that
\[
\myE(B_{n+1} \mid Z_n, \tilde Z_{n+1}) = a(Z_n, \tilde Z_{n+1})
\]
and 
\[
\myE(B_{n+1} \mid Z_n) = \int_{\CB} a(Z_n, \tilde z) G(Z_n, d\tilde z), 
\]
denoting the r.h.s. by $\bar a(Z_n)$.

We can write
\begin{align*}
\myE(f(Z_{n+1}, Z_n) &= \myE(f(Z_{n}, Z_n)(1-B_{n+1})) + \myE(f(\tilde Z_{n+1}, Z_n)B_{n+1})\\
&= \myE(f(Z_{n}, Z_n)(1-\bar a(Z_n))) + \myE(f(\tilde Z_{n+1}, Z_n)a(Z_n, \tilde Z_{n+1}))).
\end{align*}
Using the fact that $(Z_n, \tilde{Z}_{n+1})\sim Q\otimes G$, we get  
\begin{align*}
\myE(f(Z_{n+1}, Z_n) &= \myE(f(Z_{n}, Z_n)(1-\bar a(Z_n))) + \myE(f(Z_{n}, \tilde Z_{n+1})a(\tilde Z_{n+1}, Z_n)) \rho(Z_n, \tilde Z_{n+1}))\\
&= \myE(f(Z_{n}, Z_n)(1-\bar a(Z_n))) + \myE(f(Z_{n}, \tilde Z_{n+1})a(Z_{n}, \tilde Z_{n+1})))\\
&=\myE(f(Z_n, Z_{n+1})
\end{align*}
\end{proof}

Assume that the distribution $Q$ (resp. the transition $G(x\, \cdot)$  has a density  $q$ (resp. $g(x, \cdot)$ with respect to a measure $\mu$ on  $\CB$. Then the weak symmetry condition states that 
\begin{equation}
\label{eq:weak.sym}
q(z) g(z,z') = 0 \Rightarrow q(z') g(z',z) = 0
\end{equation}
almost everywhere (for the product measure $\mu\otimes \mu$. In other terms the chain with transition $G$ should remain in the support of $Q$ once it reaches it ($q(z)=0$ implies that $g(z',z) =0$ for $Q$-almost all $z'$) and $g(z,z')=0$ implies that $g(z',z)=0$  for $Q$-almost all $z'$.
The acceptance probability is then given by
\begin{equation}
\label{eq:met.hast.update}
a(z,z') = \min\left(1, \frac{q(z')g(z',z)}{q(z)g(z,z')}\right).
\end{equation}

\subsection{Sampling methods for continuous variables}
\label{sec:continuous.met}

\paragraph{Metropolis adjusted Langevin algorithm.}
While the Gibbs sampling and Metro\-polis-Hastings methods were formulated for general variables and probability distributions,  proving that the related chains are ergodic, and checking conditions for  geometric convergence speed is much harder when dealing with general state spaces than with finite or compact spaces (see, e.g., \citep{roberts1994geometric,mengersen1996rates,amit1996convergence,
roberts2004general}).
On the other hand, interesting choices of proposal transitions for Metropolis-Hastings are available when $\CB = \mR^d$ and $\mu$ is Lebesgue's measure, taking advantage, in particular, of differential calculus.  More precisely, assume that $q$ takes the form
\[
q(z)= \frac{1}{C} \exp(- H(z))
\]
for some smooth function $H$ (at least $C^1$), such that $\exp(-H)$ is integrable. We saw in \cref{sec:mcmc.rd} that, under suitable assumptions, the Markov chain
\begin{equation}
\label{eq:langevin.euler}
X_{n+1} = X_n - \frac\delta 2  \nabla H(X_n) + \sqrt{\delta} \epsilon_{n+1}
\end{equation}
with $\epsilon_{n+1} \sim \CN(0, \Id[d])$ has $q$ as invariant distribution in the limit $\delta \to 0$. Its transition probability, such that $g(z, \ccdot)$ is the p.d.f. of $\CN( z - \frac \delta 2 \nabla H(z), \delta \Id[d])$, is therefore a natural choice for a proposal distribution in the Metropolis-Hastings algorithm. The resulting Markov chain is called ``Metropolis adjusted Langevin algorithm,'' or MALA.  It converges the exact target distribution, and can also be proved to satisfy geometric convergence under less restrictive hypotheses than \cref{eq:langevin.euler} \citep{roberts1996exponential}.

\paragraph{Hamiltonian Monte-Carlo.}
Another approach, similar to MALA, is the Hamiltonian Monte-Carlo method (or hybrid Monte-Carlo) \citep{duane1987hybrid,neal2012mcmc}. Inspired by physics, the method introduces a new variable, $m\in \mR^d$, called ``momentum,'' and defines the ``Hamiltonian:''
\[
\CH(z, m) = H(z) +\frac{1}{2} |m|^2.
\]
Introduce the Hamiltonian dynamical system
\begin{equation}
\label{eq:ham.mcmc.system}
\left\{
\begin{aligned}
\partial_t \zeta(t) &= \nabla_m \CH(\zeta(t), \mu(t)) = \mu(t)\\
\partial_t \mu(t) &= -\nabla_z \CH(\zeta(t), \mu(t)) = -\nabla H(\zeta(t))
\end{aligned}
\right.
\end{equation}
and denote by $\Phi_t(z, m) = (\bfz_t(z, m), \bfm_t(z, m))$ the solution $(\zeta(t), \mu(t))$ of the system started with $\zeta(0)= z$ and $\mu(0) = m$. We will use the following facts on Hamiltonian system, that we regroup in the next lemma.
\begin{lemma}
\label{lem:ham.mc}
The Hamiltonian is invariant along solutions of \cref{eq:ham.mcmc.system}, i.e., 
\[
\CH(\bfz_t(z, m), \bfm_t(z, m)) = \CH(z,m)
\]
for all times $t$. Moreover $\Phi_t$ satisfies $\det d\Phi_t(z,m) = 1$ for all $t$, $z$, $m$, i.e., $\Phi_t$ is volume preserving. Finally $\Phi_t$ satisfies the reversibility property
\[
(z', m') = \Phi_t(z,m) \Rightarrow (z, -m) = \Phi_t(z', -m')
\]
i.e., $\Phi_t(\bfz_t(z, m), -\bfm_t(z, m)) = (z, -m)$.
\end{lemma}
\begin{proof}
One has, for a solution of \cref{eq:ham.mcmc.system}, 
\begin{align*}
\partial_t \CH(\zeta(t), \mu(t)) &= \nabla_z\CH(\zeta(t), \mu(t))^T \partial_t \zeta(t) + \nabla_m\CH(\zeta(t), \mu(t))^T \partial_t \mu(t)\\
&=  \nabla_z\CH(\zeta(t), \mu(t))^T \nabla_m\CH(\zeta(t), \mu(t)) - \nabla_m\CH(\zeta(t), \mu(t))^T \nabla_z\CH(\zeta(t), \mu(t)) =0,
\end{align*}
which implies that 
\[
H(\zeta(t)) + \frac{1}{2} |\mu(t)|^2 = H(\zeta(0)) + \frac{1}{2} |\mu(0)|^2 
\]
at all times $t$. 
Applying \cref{eq:det.log.der} and the chain rule, we have
\[
\partial_t \log \det(d\Phi_t( z, m)) = \trace(d\Phi_t( z, m)^{-1} \partial_t d\Phi_t(t,m)).
\]
From
\[\left\{
\begin{aligned}
\partial_t \bfz_t(z, m) &= \bfm_t(z,p)\\
\partial_t \bfp_t(z, m) &= -\nabla H(\bfz_t(z,m))
\end{aligned}
\right.
\]
we get
\begin{align*}
\partial_t d\Phi_t(z, m)& = 
\begin{pmatrix} 
\partial_z \bfm_t(z, m) & \partial_m \bfm_t(z,m) \\
 -\nabla^2 H(\bfz_t(z,m)) \partial_z \bfz_t(z, m) & -\nabla^2 H(\bfz_t(z,m)) \partial_m \bfz_t(z, m)
 \end{pmatrix}\\
 &=   \begin{pmatrix} 0 & \Id[d] \\ 
  -\nabla^2 H(\bfz_t(z,m)) & 0
 \end{pmatrix} d\Phi_t(z,m).
 \end{align*}
 We therefore get 
\[
\partial_t \log \det(d\Phi_t( z, m)) = \trace\begin{pmatrix} 0 & \Id[d] \\ 
 - \nabla^2 H(\bfz_t(z,m)) & 0
 \end{pmatrix} = 0
\]
showing that the determinant is constant.
Since $\Phi_0(z,m) = (z,m)$ by definition, we conclude that $\det(d\Phi_t( z, m)) = 1$ at all times. 

Assume that $(\zeta(\cdot), \mu(\cdot))$ is solution of \cref{eq:ham.mcmc.system}. Fix $t>0$ and let $\tilde \zeta(s) = \zeta(t-s)$ and $\tilde \mu(s) = - \mu(t-s)$. Then
\[
\partial_s  \tilde \zeta(s) = - \partial_t \zeta(t-s) = -\mu(t-s) = \tilde \mu(s)
\]
and
\[
\partial_s \tilde \mu(s) = \partial_t \mu(t-s) = -\nabla H(\zeta(t-s)) = -\nabla H(\tilde\zeta(s)).
\]
This shows that $(\tilde \zeta, \tilde \mu)$ is a  solution of \cref{eq:ham.mcmc.system}, yielding  $(\tilde\zeta(t), \tilde\mu(t)) = \Phi_t(\tilde\zeta(0), \tilde\mu(0))$, or
\[
(\zeta(0), -\mu(0)) = \Phi_t(\zeta(t), -\mu(t))
\]
at all times, proving the reversibility statement.
\end{proof}

We now define the Hamiltonian Monte-Carlo iterations for which we 
fix a time $\theta>0$.
Define a Markov chain $Z_n$ by the stochastic recursion
\begin{equation}
\label{eq:ham.mc.Z}
Z_{n+1} = \bfz_\theta(Z_n, M_n)
\end{equation}
with $M_n \sim \CN(0, \Id[d])$ sampled independently from previous variables. One then has the proposition:
\begin{proposition}
\label{prop:ham.mc}
The distribution $Q$ (with density $q$) is invariant by the Markov chain $(Z_n, n\geq 0)$.
\end{proposition}
\begin{proof} 
Consider the extended system
\begin{equation}
\label{eq:ham.mc.extended}
\left\{
\begin{aligned}
Z_{n+1} &= \bfz_\theta(Z_n, M_n)\\
M_{n+1} &\sim \CN(0, \Id[d])
\end{aligned}
\right.
\end{equation}
Define the p.d.f. on $\mR^d\times \mR^d$ $\bar q(z, m) = q(z) \phi_{\CN}(m)$, where $\phi_{\CN}$ the p.d.f. of the $d$-dimensional standard Gaussian. 

We show that $\bar q$ is invariant by these transitions, that is, if $(Z_n, M_n) \sim \bar q$, then, so does $(Z_{n+1}, M'_{n+1})$, with $M'_{n+1} = \bfm_\theta(Z_n, M_n)$.
 From \cref{lem:ham.mc}, we have
\[
q(\zeta(t)) \phi_{\CN}(\mu(t))  = q(\zeta(0)) \phi_{\CN}(\mu(0)).
\]
Let $\bar q_t$ denote the p.d.f. of $\Phi_t(Z_n,M_n)$ and assume that $\bar q_0(z, m) = q(z) \phi_{\CN}(m)$. We have, using the change of variable formula
\[
\bar q_t(\Phi_t(z,m)) |\det d\Phi_t(z,m)| = q(z) \phi_{\CN}(m).
\]
From \cref{lem:ham.mc}, the r.h.s. is equal to
\[
q(\bfz_t(z,m)) \phi_{\CN}(\bfm_t(z,m)) |\det d\Phi_t(z,m)|
\]
yielding the identification
\[
\bar q_t(z', m') = q(z') \phi_{\CN}(m').
\]
In particular, we find that $Z_{n+1}$ has p.d.f. $q$, which is the statement of the proposition. Note that, since $M_{n+1}$ is an independent sample from $\phi_\CN$ (note that $M'_{n+1}$ is discarded in \cref{eq:ham.mc.extended}), the joint distribution of $(Z_{n+1}, M_{n+1})$ also has density $\bar q$. 
\end{proof}

\begin{remark}[Reversibility] 
\label{rem:ham.mc}
The reversibility of the Hamiltonian system proved in \cref{lem:ham.mc} actually implies that the chain $Z_n$ defined by \cref{eq:ham.mc.Z} is in detailed balance with the distribution $Q$. To prove this, one needs to show that, if $Z_n \sim Q$, then, for any function $f\geq 0$ on $\mR^d\times \mR^d$, 
\[
\myE(f(Z_n, Z_{n+1}) = \myE(f(Z_{n+1}, Z_n)).
\]
We have 
\begin{align*}
\myE(f(Z_{n+1}, Z_{n})) &= 
\int f(\bfz_\theta(z, m), z) \phi_{\CN}(m) q(z) dm  dz.
\end{align*}
Make the change of variables $z' = \bfz_\theta(z, m)$, $m' = \bfm_\theta(z,m)$, which has Jacobian determinant 1, and is such that $z = \bfz_\theta(z', -m')$, $m = -\bfm_\theta(z',-m')$. We get
\begin{align*}
\myE(f(Z_{n+1}, Z_{n}) ) 
& = \int f(z', \bfz_\theta(z', -m')) q(\bfz_\theta(z', -m')) \phi_{\CN}(-\bfm_\theta(z',-m'))  dm' dz'\\
& = \int f(z', \bfz_\theta(z', -m')) q(\bfz_\theta(z', -m')) \phi_{\CN}(\bfm_\theta(z',-m'))  dm' dz'\\
& = \int f(z', \bfz_\theta(z', m')) q(\bfz_\theta(z', m')) \phi_{\CN}(\bfm_\theta(z',m'))  dm' dz'\\
&= 
\int f(z', \bfz_\theta(z', m')) q(z') \phi_{\CN}(m') dm'  dz'\\
&= \myE(f(Z_{n},Z_{n+1}))
\end{align*}
using the symmetry of the Gaussian distribution, the change of variable $m'\to -m'$ and the conservation of $\CH$. 
\end{remark}

\paragraph{Time discretization.}
This simulation scheme can potentially make large moves in the current configuration $z$ while maintaining detailed balance (therefore not requiring an accept/reject step). However, practical implementations require discretizing \cref{eq:ham.mcmc.system}, which breaks the conservation properties that were used in the argument above, therefore requiring a Metropolis-Hastings correction. For example, a second-order Runge Kutta (RK2) scheme with time step $\alpha$ gives 
\[
\left\{
\begin{aligned}
Z_{n+1} &= Z_n + \alpha M_n - \frac{\alpha^2}{2} \nabla H(Z_n)\\
M_{n+1} &= M_n - \frac{\alpha}{2}(\nabla H(Z_n) + \nabla H(Z_n + hM_n))
\end{aligned}
\right.
\]
Only the update for $Z_n$ matters, however, since $M_{n+1}$ is discarded and resampled at each step. Interestingly, if we let $\delta = \sqrt \alpha$ the first equation in the system becomes 
\[
Z_{n+1} = Z_n  - \frac{\delta}{2} \nabla H(Z_n) + \delta M_n
\]
with $M_n \sim \CN(0,1)$, which is exactly \cref{eq:langevin.euler}, showing that the MALA algorithm is a special case of discretized Hamiltonian Monte-Carlo.  Note that one can, in principle, solve \cref{eq:ham.mcmc.system} with more that one discretization step (the continuous equation can be solved for an arbitrary time), but one must then face the challenge of computing the Metropolis correction since the Hamiltonian is not conserved at each step.

One can use discretization schemes that are more adapted to solving Hamiltonian systems \citep{leimkuhler2005simulating}. One of the simplest examples is  the St\"ormer-Verlet scheme, which computes
\[
\left\{
\begin{aligned}
M_{n+1/2} & = M_n - \frac{\alpha}{2} \nabla H(Z_n)\\
Z_{n+1} &= Z_n + \alpha M_{n+1/2}\\
M_{n+1} & = M_{n+1/2} - \frac{\alpha}{2} \nabla H(Z_{n+1})
\end{aligned}
\right.
\]
This scheme computes $\psi_1\circ \psi_2 \circ \psi_1(Z_n, M_n)$ with $\psi_1(z,m) = (z, m - (\alpha/2) \nabla H(z))$ and $\psi_2(z,m) = (z+ \alpha m, m)$. Because both $\psi_1$ and $\psi_2$ have a Jacobian determinant equal to 1, so does their composition. This scheme is also reversible, since we have
\[
\left\{
\begin{aligned}
  -M_{n+1/2} &= - M_{n+1} - \frac{\alpha}{2} \nabla H(Z_{n+1})\\
  Z_{n} &= Z_{n+1} - \alpha M_{n+1/2}\\
-M_{n} & = -M_{n+1/2} - \frac{\alpha}{2} \nabla H(Z_n)
\end{aligned}
\right.
\]
These properties are conserved if one applies the St\"ormer-Verlet scheme more than once at each iteration, that is, fixing some $N>0$ and letting $\Phi_N(z,m) = (\psi_1 \circ \psi_2 \circ \psi_1)^{\circ N}$, then $\Phi_N^{-1} = J\Phi_N \circ J$, with $J(z,m) = (z,-m)$, and $\det d\Phi_N = 1$.

We have the following result.
\begin{proposition}
\label{prop:ham.mc.sv}
The Markov chain defined by
\[
\left\{
\begin{aligned}
M_n &\sim \CN(0, \Id[d])\\
(\tilde Z_{n+1}, \tilde M_{n+1})  &= \Phi_N(Z_n, M_n)\\
B_{n+1} &\sim \mathrm{Bernoulli}(a(Z_n, M_{n}, \tilde Z_{n+1}, \tilde M_{n+1})\\
Z_{n+1} &= B_{n+1} \tilde Z_{n+1} + (1-B_{n+1}) Z_n 
\end{aligned}
\right.
\]
with
\[
a(z, m, \tilde z , \tilde m) = \max\left( 1, \dfrac{q(\tilde z) \phi_\CN(\tilde m)}{q(z)\phi_\CN(m)}\right),
\]
is $Q$-reversible.
\end{proposition}
\begin{proof}
Note that 
\[
a(\tilde z, \tilde m, z, m) q(\tilde z) \phi_\CN(\tilde m) = a(z, m, \tilde z , \tilde m) q(z)\phi_\CN(m).
\]
Writing $\Phi_N(z,m) = (\bfz_N(z,m), \bfm_N(z,m))$, we let 
\[
\bar a(z) = \myE(a(z, M_n, \bfz_N(z, M_n), \bfm_N(z, M_n))
\]
where $M_n\sim \CN(0, \Id[d])$.
We have, assuming that $Z_n \sim Q$:
\[
\myE(f(Z_{n+1}, Z_n)) = \myE(f(Z_n, Z_n) (1-\bar a(Z_n)) + \myE(f(\tilde Z_{n+1}, Z_n) a(Z_n, M_n, \bfz_N(z, M_n), \bfm_N(z, M_n))
\]
and we focus on the second term, which is equal to 
\begin{align*}
&\int_{\mR^d}\int_{\mR^d} f(\bfz_N(z,m), z) a(z, m, \bfz_N(z,m), \bfm_N(z,m)) q(z) \phi_\CN(m) dz dm  \\
&=
\int_{\mR^d}\int_{\mR^d} f(\tilde z, \bfz_N(\tilde z, -\tilde m)) a(\bfz_N(\tilde z, -\tilde m), -\bfm_N(\tilde z, -\tilde m), \tilde z, \tilde m) q(\bfz_N(\tilde z, -\tilde m)) \phi_\CN(-\bfm_N(\tilde z, -\tilde m)) d\tilde z d\tilde m\\
&=
\int_{\mR^d}\int_{\mR^d} f(\tilde z, \bfz_N(\tilde z, -\tilde m)) a(\tilde z, \tilde m, \bfz_N(\tilde z, -\tilde m), -\bfm_N(\tilde z, -\tilde m)) q(\tilde z) \phi_\CN(\tilde m) d\tilde z d\tilde m\\
&=
\int_{\mR^d}\int_{\mR^d} f(\tilde z, \bfz_N(\tilde z, \tilde m)) a(\tilde z, \tilde m, \bfz_N(\tilde z, \tilde m), \bfm_N(\tilde z, \tilde m)) q(\tilde z) \phi_\CN(\tilde m) d\tilde z d\tilde m\\
&=\myE(f(Z_n, \tilde Z_{n+1}) a(Z_n, M_n, \bfz_N(z, M_n), \bfm_N(z, M_n)).
\end{align*}
This shows that $\myE(f(Z_{n+1}, Z_n)) = \myE(f(Z_{n}, Z_{n+1}))$, proving that the Markov chain is $Q$-reversible.
\end{proof}

Note that, because  the St\"ormer-Verlet scheme does not keep the Hamiltonian invariant, the acceptance probability $a$ is needed and is less than 1 in general. But the scheme has good enough stability properties that $a$ will typically be close to 1, even for relatively large $N$.

 \section{Perfect sampling methods}
\label{sec:perfect}
We assume, in this section, that $\CB$ is a finite set.
The  Markov chain simulation methods provided in the previous sections  do not provide exact samples from the distribution
$q$, but only increasingly accurate approximations.
 Perfect sampling algorithms \citep{pp96,pp98,fill98} use Markov chains ``backwards'' to generate exact samples. To describe them, it is easier to describe a
Markov chain as a stochastic recursive equation of the form
\begin{equation}
\label{eq:mc.rec}
X_{n+1} = f(X_n, U_{n+1})
\end{equation}
where $U_{n+1}$ is independent of  $X_n, X_{n-1}, \ldots$, and the $U_k$'s are identically distributed. In the discrete case (assumed in this section), and given a
stochastic matrix $P$, one can take $U_n$
to be the uniformly distributed variable used to sample from
$(p(X_n, x), x\in\CB)$. Conversely, the transition probability
associated to \cref{eq:mc.rec} is
$p(x,y) = P(f(x, U) = y).$

It will be convenient to consider negative times also. For $n>0$,
recursively define
$F_{-n}(x, u_{-n+1}, \ldots, u_0)$ by
$$
F_{-n-1} (x, u_{-n}, \ldots, u_0) = F_{-n}(f(x, u_{-n}), u_{-n+1},
\ldots, u_0)
$$
and $F_{-1}(x,u_0) = f(x, u_0)$. 
Denote, for short, $U_{-n}^0 =
(U_{-n}, \ldots, U_0)$. The function $F_{-n}(x, u_{-n+1}^0)$ provides
the value of $X_0$ when $X_{-n}=x$ and $U_{-n+1}^0 =
u_{-n+1}^0$. 

For an infinite  sequence in the past, $u_{-\infty}^0$, let $\nu(u_{-\infty}^0)$ denote the
first integer $n$ such that $F_{-n}(x, u_{-n+1}^0)$ does not depend on
$x$ (the function
``coalesces''). Then, the following theorem is true:
\begin{theorem}
\label{th:p.s}
Assume that the chain defined by \cref{eq:mc.rec} is
ergodic, with invariant distribution $Q$. Then $\nu = \nu(U_{-\infty}^0)$ is finite with probability 1,
and 
\begin{equation}
\label{eq:p.s}
X_*:= F_{-\nu}(x, U_{-\nu+1}^0)
\end{equation}
(which is independent of $x$) has distribution $Q$.
\end{theorem}
\begin{proof}
Because the  chain is ergodic, we know that there exists an integer
$N$ such that one can pass from any state to any other with positive
probability. So the chain can, starting from anywhere, coalesce with positive probability in
$N$ steps; $\nu$ being infinite would imply that this event never
occurs in an infinite number of trials, and this has probability 0.

For any $k>0$ and any $x\in\CB$, we have
\begin{equation}
\label{eq:inv.past}
X_*= F_{-\nu}(f_{-k}(x, U_{-\nu-k+1}^{-\nu}), U_{-\nu+1}^0) = F_{-\nu-k}(x, U_{-\nu-k+1}^0).
\end{equation}
But, because the chain is ergodic, we have, for any $x\in\CB$
$$
\lim_{k\to\infty} \myP(F_{-k}(x, U_{-k+1}^0) = y) = Q(y).
$$
We can write
\begin{align*}
\myP(F_{-k}(x, U_{-k+1}^0) = y) 
& = \myP(F_{-k}(x, U_{-k+1}^0) = y, \nu\leq k) + \myP(F_{-k}(x, U_{-k+1}^0) = y, \nu> k)\\
& = \myP(X_* = y, \nu\leq k)
+ \myP(F_{-k}(x, U_{-k+1}^0) = y, \nu> k)
\end{align*}
The right-hand side tends to $\myP(X_* = y)$ when $k$ tends to infinity
(because $\myP(\nu>k)$ tends to 0), and the left-hand side tends to
$Q(y)$, which gives the second part of the theorem.
\end{proof}

From  \cref{eq:inv.past}, which is the key step in proving
that $X^*$ follows the invariant distribution, one can see why it is
important to consider sampling that expands backward in time rather
than forward. More specifically, consider the coalescence
time for the forward chain, letting
$\tilde\nu(u_0^\infty)$ be the first index for which 
\[
\tilde X_* := F_{\tilde\nu}(x, u_0^{\tilde\nu})
\]
is independent from the starting point, $x$. For any $k\geq 0$, one
still has the fact that  $F_{\tilde\nu+k}(x, u_0^{\tilde\nu+k})$ does
not depend on $x$, but its value depends on $k$ and will not be equal
to $\tilde X_*$ anymore, which  prevents the rest of the proof of 
 \cref{th:p.s} to carry on.

An equivalent algorithm is described in the next proposition (the
proof is easy and left to the reader).
\begin{proposition}
\label{prop:perf.samp.alg}
Using the same notation as above, the following algorithm generates a
perfect sample, $\xi_*$, of the invariant
distribution of an ergodic Markov chain. 

Assume that an
infinite sample $u_{-\infty}^0$ of $U$ is available. Given
this sequence, the algorithm, starting with $t_0=2$, is:
\begin{itemize}
\item[1.] For all $x\in \CB$, define $\xi^x_{-t}, t=-t_0, \ldots, 0$ by
$\xi^x_{-t_0} = x$ and $\xi_{-t+1}^x = f(\xi_{-t}^x, u_{-t+1})$.
\item[2.] If $\xi_0^x$ is constant (independent of $x$), let $\xi_*$
be equal to this constant value and stop. Otherwise, return to step 1
replacing $t_0$ with $2t_0$.
\end{itemize}
\end{proposition}
In practice, the $u_{-k}$'s are only generated when they are needed. But
it is important to consider the sequence as fixed: once $u_{-k}$ is generated,
it must be stored (or identically regenerated, using the same seed) for further use. It is
important to strengthen the fact that this algorithm works backward in
time, in the sense that the first states of the sequence are not
identical at each iteration, because they are generated
using random numbers with indexes further in the past.

Such an algorithm is not feasible when  $|\CB|$ is too large, since
one would have to consider an intractable number of simulated sequences (one
for each $x\in\CB$). However there are cases in which the constancy of $\xi_0^x$ over all $\CB$ can be
decided from its constancy over a small subset of $\CB$. 

One situation in which this is true is when the Markov chain is
monotone, according to the following definition. Assume that $\CB$ can be
partially ordered, and that $f$ in  \cref{eq:mc.rec} is
increasing in $x$, i.e.,
\begin{equation}
\label{eq:mon.f}
x \leq x' \Ria \forall u, f(x,u) \leq f(x',u).
\end{equation}
Let $\CB_{\mathrm{min}}$ and $\CB_{\mathrm{max}}$ be the set of
minimal and maximal elements in $\CB$. Then the sequence coalesces for
the algorithm above if and only if it coalesces over
$\CB_{\mathrm{min}} \cup \CB_{\mathrm{max}}$. Indeed, any $x\in \CB$ is
smaller than some maximal element, and larger than some minimal
element in $\CB$. By \cref{eq:mon.f}, these inequalities remain true
at each step of the sampling process, which implies that when chains
initialized with extremal elements coalesce, so do the other
ones. Therefore,  it
suffices to run the algorithm with extremal configurations only.

One
can rewrite \cref{eq:mon.f} in terms of transition probabilities
$p(x,y)$, assuming that $U$ follows a uniform distribution on $[0,1]$
and, for all $x\in \CB$, there exists a partition
$(I_{xy}, y\in\CB)$  of $\CB$, such that
\[
f(x,u) = y \Leftrightarrow u\in I_{x,y}
\]
and $I_{xy}$ is
an interval with length $p_{xy}$.
 Condition \cref{eq:mon.f} is then
equivalent to
\[
x \leq x' \Ria \forall y\in \CB, I_{xy} \sub \bigcup_{y'\geq y}
I_{x'y'}.
\]  
This requires in particular that $\sum_{y\geq y_0}p(x,y) \leq
\sum_{y\geq y_0} p(x',y)$ whenever $x\leq x'$ (one says that $p(x, \cdot)$
is stochastically smaller than $p(x', \cdot)$). 

One example in which this reduction works is with the ferromagnetic Ising
model, for which $\CB = \{-1,1\}^L$ and 
$$
q(x) = \frac{1}{C} \exp\big(\sum_{s, t=1, s<t}^L \be_{st} \pe x s
\pe x t\big)
$$
with $\be_{st} \geq 0$ for all $\{s,t\}$. 
Then, the Gibbs sampling algorithm iterates the following steps: take a
random $s\in \{1, \ldots, L\}$ and update $\pe x s$ according to the conditional
distribution
$$
g_s(\pe ys\mid \pe x {s^c}) = \frac{e^{\pe y s v_s(x)}}{e^{-v_s(x)} + e^{v_s(x)}}
$$
with $v_s(x) = \sum_{t\neq s}
\be_{st} \pe x t$. One can order $\CB$ so that $x\leq \tilde x$ if and only if
$\pe x s\leq \pe {\tilde x}s$ for all $s=1, \ldots, L$. The minimal and maximal elements are
unique in this case, with $\pe{x_{\mathrm{min}}}s\equiv -1$ and $\pe {x_{\mathrm{max}}} s\equiv 1$. Moreover,
because all $\be_{st}$ are non-negative, $v_s$ is an increasing function
of $x$ so that, if $x\leq \tilde x$, then $g_s(1\mid \pe x s) \leq
g_s(1\mid \pe {\tilde x}s)$. 

To define the stochastic iterations, first introduce
\[
f_s(x,u) =
\begin{cases}
 \pe 1 s\wedge \pe x {s^c} \text{ if }  u \leq q_s(1\mid \pe x s)\\
\pe{(-1)}s\wedge \pe x {s^c} \text{ if } u > q_s(1\mid \pe x s),
\end{cases}
\]
which satisfies \cref{eq:mon.f}. The whole updating scheme
can then be implemented with the function
$$
f(x, (u,\tilde u)) = \sum_{s=1}^L \de_{I_s}(\tilde u) f_s(x,u)
$$
where $(I_s, s\in V)$ is any partition of $[0,1]$ in intervals of
length $1/L$. This is still monotonic. The algorithm described in
 \cref{prop:perf.samp.alg} can therefore be applied to 
sample exactly, in finite time, from the ferromagnetic Ising model.

\section[Markovian Stochastic Approximation]{Application: Stochastic approximation with Markovian transitions}

Using the material developed in this chapter, we now discuss the convergence of stochastic approximation methods (such as stochastic gradient descent) when the random  variable in the update term follows Markovian transitions. In \cref{sec:sgd}, we considered algorithms in the form
\[
\left\{
\begin{aligned}
\boldsymbol\xi_{t+1} &\sim \pi_{X_t}\\
X_{t+1} &= X_t + \al_{t+1} H(X_t, \boldsymbol\xi_{t+1})
\end{aligned}
\right.
\]
where $\bfxi_t: \Omega \to \CR_\xi$ is a random variable. We now want to address situations in which the random variable $\xi_{t+1}$ is obtained through a transition probability, therefore considering the algorithm
\begin{equation}
\label{eq:sa.markov}
\left\{
\begin{aligned}
\boldsymbol\xi_{t+1} &\sim P_{X_t}(\xi_t, \ccdot)\\
X_{t+1} &= X_t + \al_{t+1} H(X_t, \boldsymbol\xi_{t+1})
\end{aligned}
\right.
\end{equation}
Here  $P_x$ is, for all $x$, a transition probability from $\CR_\xi$ to $\CR_\xi$. We will assume that, for all $x\in \mR^d$, the Markov chain with transition $P_x$ is geometrically ergodic, and we denote by $\pi_x$ its invariant distribution. We let, as in \cref{sec:sgd}, $\bar H(x) = E_{\pi_x}(H(x, \cdot))$. We will use the notation for a function $f: \mR^d \times \CR_\xi \to \mR$
\[
P_xf: (x', \xi)\in \mR^d \times \CR_\xi  \mapsto P_xf(x', \xi)= \int_{\CR_\xi} f(x', \xi') P_x(\xi, d\xi')
\]
and
\[
\pi_x f: x'\in \mR^d \mapsto \pi_x f(x') = \int_{\CR_\xi} f(x', \xi) \pi_x(d\xi).
\]
In particular, $\bar H(x) = \pi_x H(x)$. We also define $h(x, \xi) = H(x, \xi) - \bar H(x)$ and $\tilde h(x, \xi) = P_x h(x, \xi)$. 
We make the following assumptions.
\begin{enumerate}[label=(H\arabic*)]
\item There exists constants $C_0, C_1, c_2$ such that, for all $x, y\in \mR^d$,
\begin{subequations}
\begin{align}
\label{eq:sgdm.h1.a}
&\sup_{\xi\in \CR_\xi}|H(x, \xi)| \leq C_0,\\
\label{eq:sgdm.h1.b}
&\sup_{\xi\in \CR_\xi}|\tilde h(x, \xi)| \leq C_1,\\
\label{eq:sgdm.h1.c}
& \sup_{\xi\in \CR_\xi} |\tilde h(x, \xi) - \tilde h(y, \xi)| \leq C_1 |x-y|,\\
\label{eq:sgdm.h1.d}
& D_{\mathit{var}}(\pi_x, \pi_y) \leq C_2 |x-y|
\end{align}
\end{subequations}
\item There exists $x^*\in\mR^d$ and $\mu > 0$ such that, for all $x\in \mR^d$ 
\begin{equation}
\label{eq:sgdm.h2}
(x - x^*)^T \bar H(x) \leq - \mu |x - x^*|^2.
\end{equation}
\item We assume that there exists a constant $M$ and a non-decreasing function $\rho: [0, +\infty) \to [0, 1)$ such that, for all probability distributions $Q$ and $Q'$ on $\CR_\xi$,
\begin{equation}
\label{eq:sgdm.h3}
D_{\mathrm{var}}(QP_x^n, Q'P_x^n) \leq M\rho(|x|)^n D_{\var}(Q, Q').
\end{equation}
\item The sequence $\alpha_1, \alpha_2, \ldots$ is non-increasing, with 
\begin{subequations}
\begin{equation}
\label{eq:sgdm.h4.a}
\sum_{t=1}^\infty \alpha_t = +\infty \quad \text{ and } \quad
\sum_{t=1}^\infty \alpha^2_t < +\infty.
\end{equation}
Let $\sigma_t = \sum_{s=1}^t \alpha_s$. If $C_1 >0$, we also require that
\begin{equation}
\label{eq:sgdm.h4.b}
\lim_{t\to\infty} \alpha_{t} \sigma_{t} (1-\rho(\sigma_t))^{-1} = 0
\end{equation}
and
\begin{equation}
\label{eq:sgdm.h4.c}
\sum_{s=2}^t \alpha^2_{s} \sigma_{s} (1-\rho(\sigma_s))^{-2} <\infty.
\end{equation}
\end{subequations}

\end{enumerate}

Given this, the following theorem holds.
\begin{theorem}
\label{th:sgdm}
Assuming (H1) to (H4), the sequence defined by \cref{eq:sa.markov} is such that
\[
\lim_{t\to\infty} \myE(|X_t-x_*|^2) =0
\]
\end{theorem}

\begin{remark}
\label{rem:sgdm}
Condition (H1) assumes that $H$ is bounded and uniformly Lipschitz in $x$, which is more restrictive than what was assumed in \cref{sec:sgd.convergence}, but applies, for example, to situations considered in \citet{you88} and later in this book in \cref{sec:stoc.grad}. 

Condition (H3) implies that the Markov chain with transition $P_x$ is uniformly geometrically ergodic, but the ergodicity rate may depend on $x$ and it may, in particular, converge to $1$ when $x$ tends to $\infty$, which is the situation targeted in this theorem.

The reader may refer to \cite{younes1999convergence} for a general discussion of this problem with relaxed hypotheses and almost sure convergence, at the expense of significantly longer proofs. 
\end{remark}

\begin{proof}
We note that, from \cref{eq:sgdm.h1.a}, one has
\begin{equation}
\label{eq:sgd.prior.bound}
|X_t-x_*| \leq C_0\sigma_t  |X_0 - x_*|.
\end{equation}

Similarly to \cref{sec:sgd.convergence}, we let 
$A_t = |X_t - x_*|^2$ and $a_t = \myE(A_t)$. One can then write
\[
A_{t+1} = A_t +  2\alpha_{t+1} (X_t - x_*)^T \bar H(X_t) + 2\alpha_{t+1}  
(X_t - x_*)^T (H(X_t, \bfxi_{t+1}) - \bar H(X_t)) + \alpha_{t+1}^2 |H(X_t, \xi_{t+1})|^2
\]
but we do not have
 \[
\myE( (X_t - x_*)^T (H(X_t, \bfxi_{t+1}) - \bar H(X_t))\mid \CU_t) = 0
\]
anymore, where $\CU_t$ is the $\sigma$-algebra of all past events up to time $t$ (all events depending of $X_s, \bfxi_s$, $s\leq t$). Indeed the Markovian assumption implies that
\begin{align*}
\myE( (X_t - x_*)^T (H(X_t, \bfxi_{t+1}) - \bar H(X_t))\mid \CU_t) &= 
(X_t - x_*)^T \left(\int_{\CR_\xi} H(X_t, \xi) P_{X_t}(\bfxi_t, d\xi)  - \bar H(X_t)\right)\\
&= (X_t - x_*)^T ((P_{X_t} H(X_t, \cdot))(\xi_t) - \bar H(X_t)),
\end{align*}
which does not vanish in general. Following \citet{benveniste2012adaptive}, this can be addressed by introducing the solution $g(x, \ccdot)$ of the ``Poisson equation''
\begin{equation}
\label{eq:sgd.poisson}
 g(x, \ccdot) - P_x g(x, \ccdot) = h(x, \ccdot).
 \end{equation}
(Recall that $h(x, \xi) = H(x, \xi) - \bar H(x)$.) One can then write
 \[
 (X_t - x_*)^T h(X_t, \bfxi_{t+1}) =  
(X_t - x_*)^T ( g(X_t, \bfxi_{t+1}) - P_{X_t} g(X_t, \bfxi_{t+1})\\
\]
and 
\begin{align*}
A_{t+1}  &\leq (1-2\alpha_{t+1}\mu) A_t +  2\alpha_{t+1} (X_t - x_*)^T ( g(X_t, \bfxi_{t+1}) - P_{X_t} g(X_t, \bfxi_{t}))) \\
&+ 2\alpha_{t+1} (X_{t} - x_*)^TP_{X_{t}} g(X_{t}, \bfxi_{t})
 - 2\alpha_{t+1} (X_t- x_*)^TP_{X_t} g(X_t, \bfxi_{t+1})
 +  \alpha_{t+1}^2 |H(X_t, \xi_{t+1})|^2
\end{align*}
Introduce the notation
\[
\eta_{s,t} = \myE((X_{s} - x_*)^TP_{X_{s}} g(X_{s}, \bfxi_{t})).
\]
Using the fact that 
\[
\myE\left((X_t - x_*)^T ( g(X_t, \bfxi_{t+1}) - P_{X_t} g(X_t, \bfxi_{t}))) \mid \CU_t\right) = 0
\]
and 
%
and noting that $|H(X_t, \xi_{t+1})|^2\leq C_0^2$, one finds, after taking expectations,
 \begin{align*}
a_{t+1} & \leq (1-2\alpha_{t+1}\mu)a_t + 2\alpha_{t+1} \eta_{t,t}  - 2\alpha_{t+1} \eta_{t,t+1}
 +  \alpha_{t+1}^2 C_0^2.
\end{align*}
Applying \cref{lem:sg.upper.bound}, and letting $v_{s,t} = \prod_{j=s+1}^t (1-2\alpha_{j+1}\mu)$, one gets 
\[
a_t \leq a_0 v_{0,t} + 2 \sum_{s=1}^t v_{s,t} \alpha_{s+1} (\eta_{s,s}  - \eta_{s,s+1}) + C_0^2\sum_{s=1}^t v_{s,t} \alpha_{s+1}^2.
\]
We now want to ensure that each term in the upper bound converges to 0. Similarly to \cref{sec:sgd.convergence}, \cref{eq:sgdm.h4.a} implies that this holds the first and last terms and we therefore focus on the middle one, writing
\begin{align}
\label{eq:sgd.markov.main.2}
\sum_{s=1}^t v_{s,t} \alpha_{s+1} (\eta_{s,s}  - \eta_{s,s+1})
&= v_{1, t} \alpha_2 \eta_{1,1} - \alpha_{t+1} \eta_{t,t+1} + \sum_{s=2}^t
(v_{s,t} \alpha_{s+1} - v_{s-1,t} \alpha_{s}) \eta_{s,s}  \\
\nonumber
&+ \sum_{s=2}^t v_{s-1,t} \alpha_{s} (\eta_{s,s} - \eta_{s-1,s}) 
\end{align}

We will need the following estimates on the function $g$ in \cref{eq:sgd.poisson}, which is defined by
\[
g(x, \xi) = \sum_{n=0}^\infty P_x^n h(x, \xi) = h(x, \xi) + \sum_{n=0}^\infty P_x^n \tilde h(x, \xi).
\]
\begin{lemma}
\label{lem:sgdm.g}
We have
\begin{subequations}
\begin{align}
\label{eq:sgdm.g.1a}
|g(x, \cdot)| &\leq C_0 + 2C_1M (1- \rho(x))^{-1},\\
\label{eq:sgdm.g.1b}
|P_xg(x, \cdot)| &\leq 2C_1M (1- \rho(x))^{-1}.
\end{align}
\end{subequations}
and, for all $x,y\in \mR^d$ and $\xi\in \CR_\xi$
\begin{equation}
\label{eq:sgdm.g.2}
|P_x g(x, \xi) - P_y(g(y, \xi)| = M^2C_1C_2 (1-\bar\rho)^{-2} + MC_1(1+C_2) (1-\bar \rho)^{-1}.
\end{equation}
with $\bar \rho = \max(\rho(|x|), \rho(|y|))$.

\end{lemma}

Using lemma \cref{lem:sgdm.g} (which is proved at the end of the section), we can control the terms intervening in   \cref{eq:sgd.markov.main.2}. Note that the first term, $v_{1,t} \alpha_2 \eta_{1,1}$, converges to 0 since \cref{eq:sgdm.h4.a} implies that $v_{1,t}$ converges to 0. 

We have,
\[
\alpha_{t+1} |\myE((X_{t} - x_*)^TP_{X_{t}} g(X_{t}, \bfxi_{t+1}))| \leq 2MC_1 \alpha_{t+1} \sigma_t (1-\rho(\sigma_t))^{-1},
\]
so that \cref{eq:sgdm.h4.b} implies that $\alpha_{t+1}\eta_{t,t+1} \to 0$. 

and since $\alpha_{s+1}\leq \alpha_s$, we have
\begin{align*}
\left|\sum_{s=2}^t
(v_{s,t} \alpha_{s+1} - v_{s-1,t} \alpha_{s}) \eta_{ss}\right| &\leq 
\sum_{s=2}^t
|v_{st}\alpha_{s} - v_{s,t} \alpha_{s+1} |\, |\eta_{ss}|\\
&\leq MC_1 \sum_{s=2}^t |v_{s-1,t} \alpha_{s} - v_{s,t} \alpha_{s+1}|\,\alpha_{s+1} \sigma_s (1-\rho(\sigma_s))^{-1}\\
&\leq C \sum_{s=2}^t |v_{s-1,t} \alpha_{s} - v_{s,t} \alpha_{s+1}|
\end{align*}
for some constant $C$, since $\alpha_{s+1} \sigma_s (1-\rho(\sigma_s))^{-1}$ is bounded.
Writing
\[
v_{s,t} \alpha_{s+1} - v_{s-1,t} \alpha_{s} = v_{st} (\alpha_{s+1} - \alpha_s + 2\mu \alpha_s^2),
\]
we get (using $\alpha_{s+1}\leq \alpha_s$)
\[
\sum_{s=2}^t |v_{s-1,t} \alpha_{s} - v_{s,t} \alpha_{s+1}| \leq \sum_{s=2}^t  v_{st} (\alpha_s - \alpha_{s+1}) + \sum_{s=2}^t  v_{st}2\mu \alpha_s^2.
\]
Since both $\sum_s (\alpha_s - \alpha_{s+1})$ and $\sum_{s=2}^t  \alpha_s^2$ converge (the former is just $\alpha_1$), \cref{lem:sgd.simple} implies that 
\[
\sum_{s=2}^t
(v_{s,t} \alpha_{s+1} - v_{s-1,t} \alpha_{s}) \eta_{ss}
\]
tends to zero.
The last term to consider is
\begin{align*}
\sum_{s=2}^t v_{s-1,t} \alpha_{s} (\eta_{ss} - \eta_{s-1,s}) & = 
\sum_{s=2}^t v_{s-1,t} \alpha_{s}
\myE((X_{s} - X_{s-1})^TP_{X_{s}} g(X_{s}, \bfxi_{s}))\\
&+\sum_{s=2}^t v_{s-1,t} \alpha_{s}
\myE((X_{s-1} - x_*)^T(P_{X_{s}} g(X_{s}, \bfxi_{s})) - P_{X_{s-1}} g(X_{s-1}, \bfxi_{s}))).
\end{align*}
We have
\[
\left|\sum_{s=2}^t v_{s-1,t} \alpha_{s}
\myE((X_{s} - X_{s-1})^TP_{X_{s}} g(X_{s}, \bfxi_{s}))\right|
\leq 
2C_0C_1M \sum_{s=2}^t v_{s-1,t} \alpha^2_{s} (1-\rho(\sigma_s))^{-1}
\]
and
\begin{multline*}
\left|\sum_{s=2}^t v_{s-1,t} \alpha_{s}
\myE((X_{s-1} - x_*)^T(P_{X_{s}} g(X_{s}, \bfxi_{s})) - P_{X_{s-1}} g(X_{s-1}, \bfxi_{s})))\right|\\
 \leq 2M^2C_0C_1(1+C_2) |X_0-  x_*| \sum_{s=2}^t v_{s-1,t} \alpha^2_{s} \sigma_{s} (1-\rho(\sigma_s))^{-2} 
\end{multline*}
and \cref{lem:sgd.simple} implies that both terms vanish at infinity. This concludes the proof of \cref{th:sgdm}.
\end{proof}

\begin{proof}[Proof of \cref{lem:sgdm.g}]

Condition (H3) and  \cref{prop:var.dist} and imply that (since $\pi_x \tilde h  = 0$)
\[
|P_x^n \tilde h(x, \xi)| \leq D_{\mathit{var}}(P_x^n(\xi, \cdot), \pi_x) \mathit{osc}(\tilde h(x, \cdot))\leq 2 C_1 M \rho(x)^n
\]
so that $g$ is well defined with 
\begin{align*}
|g(x, \cdot)| &\leq C_0 + 2C_1M (1- \rho(x))^{-1},\\
|P_xg(x, \cdot)| &\leq 2C_1M (1- \rho(x))^{-1}.
\end{align*}
We will also need to control differences of the kind
\[
P_x g(x, \xi) - P_yg(y, \xi).
\]
We consider the $n$th term in the series, writing 
\begin{align*}
P_x^n \tilde h(x, \xi) - P_y^n \tilde h(y, \xi) & = \sum_{k=0}^{n-1} (P_x^{n-k}P_y^k \tilde h(y, \xi)  - (P_x^{n-k-1}P_y^{k+1} \tilde h(y, \xi))\\
&+ P_x^n \tilde h(x, \xi) - P_x^{n}\tilde  h(y, \xi).
\end{align*}
This gives
\begin{align*}
P_x^n \tilde h(x, \xi) - P_y^n \tilde h(y, \xi) & = \sum_{k=0}^{n-1} P_x^{n-k-1}(P_xP_y^k \tilde h(y, \xi) - P_y^{k+1} \tilde h(y, \xi) - \pi_xP_y^k \tilde h(y) + \pi_xP_y^{k+1} \tilde h(y) )\\
&+ \sum_{k=0}^{n-1} (\pi_xP_y^k \tilde h(y) - \pi_xP_y^{k+1} \tilde h(y))+ P_x^n \tilde h(x, \xi) - P_x^{n} \tilde  h(y, \xi)\\
& = \sum_{k=0}^{n-1} P_x^{n-k-1}(P_xP_y^k \tilde h(y, \xi) - P_y^{k+1} \tilde h(y, \xi) - \pi_xP_y^k \tilde h(y) + \pi_xP_y^{k+1} \tilde h(y) )\\
&+ \pi_x \tilde h(y) - \pi_xP_y^{n} \tilde h(y) + P_x^n \tilde h(x, \xi) - P_x^{n} \tilde  h(y, \xi)
\end{align*}
Finally
\begin{align*}
P_x^n h(x, \xi) - P_y^n h(y, \xi) & =
 \sum_{k=0}^{n-1} P_x^{n-k-1}(P_xP_y^k \tilde h(y, \xi) - P_y^{k+1} \tilde h(y, \xi) - \pi_xP_y^k \tilde h(y) + \pi_xP_y^{k+1} \tilde h(y) )\\
&+ P_x^{n} (\tilde h(x, \xi) - \tilde h(y, \xi) + \pi_x \tilde h(y)) - (\pi_x - \pi_y) P_y^{n} \tilde h(y)  
\end{align*}

Using \cref{prop:var.dist}, we can write, letting $\bar \rho = \max(\rho(|x|), \rho(|y|))$,
\begin{align*}
&|P_x^{n-k-1}(P_xP_y^k \tilde h(y, \xi) - P_y^{k+1} \tilde h(y, \xi) - \pi_xP_y^k \tilde h(y, \xi) + \pi_xP_y^{k+1} \tilde h(y, \xi) )| \\
&\leq
M {\bar \rho}^{n-k-1} \mathit{osc} (P_xP_y^k \tilde h(y, \xi) - P_y^{k+1} \tilde h(y, \xi))\\
&\leq C_2M {\bar \rho}^{n-k-1} |x-y| \mathit{osc} (P_y^k \tilde h(y, \xi))\\
&\leq C_2C_1 M^2 {\bar \rho}^{n-1} |x-y|
\end{align*}
We also have
\[
|P_x^n (\tilde h(x, \xi) - \tilde h(y, \xi) + \pi_x \tilde h(y, \xi)) |\leq MC_1 {\bar \rho}^n |x-y|
\]
and
\[
 |(\pi_x - \pi_y) P_y^{n} \tilde h(y, \xi)| \leq MC_2C_1 {\bar \rho}^n |x-y|
 \]
 so that
\[
|P_x^n h(x, \xi) - P_y^n h(y, \xi)| \leq MC_1 \bar\rho^{n-1} (nMC_2 + (1+C_2)\bar \rho)|x-y|
\]
From this, it follows that 
\begin{align*}
|P_x g(x, \xi) - P_y(g(y, \xi)| & \leq MC_1 \sum_{n=1}^\infty \bar\rho^{n-1} (nMC_2 + (1+C_2)|x-y|\\
&= M^2C_1C_2 (1-\bar\rho)^{-2} + MC_1(1+C_2) (1-\bar \rho)^{-1}.
\end{align*}

\end{proof}

\chapter{Markov Random Fields}
\label{chap:mrf}


With this chapter, we start a discussion of large-scale statistical models in data science, starting with graphical models (Markov random fields and Bayesian networks) before discussing more recent approaches using, notably, deep learning. Important textbook references for the present chapter include \citet{pearl1988probabilistic, ancona1990random,winkler1995image,lauritzen1996graphical,
cowell07probabilistic,koller2009probabilistic}.

\section{Independence and conditional independence}

\subsection{Definitions}

We consider random variables $X, Y, Z\ldots $, and denote by $\CR_X, \CR_Y, \CR_Z\ldots$ the sets in which they take their values.
We discuss in this section  concepts of independence and conditional independence between random variables. To simplify the exposition, we will work (unless mentioned otherwise) with discrete random variables ($X$ is discrete if $\CR_X$ is finite or countable)\footnote{In the general case, $\CR_X, \CR_Y, \ldots$ are metric spaces with a countable dense subset with $\sigma$-algebras $\boldsymbol{\CS}_X, \boldsymbol{\CS}_Y, \ldots$}.   
We start with a basic definition.
\begin{definition}
\label{def:indep}

Two discrete random variables $X:\Om \to \CR_X$ and $Y:\Om\to \CR_Y$ are independent if and only if
$$
\forall x\in \CR_X, \forall y\in \CR_Y: \myP(X=x, Y=y) = \myP(X=x)\myP(Y=y).
$$
\end{definition}

The general definition for arbitrary r.v.'s is that 
\[
\myE(f(X)g(Y)) = \myE(f(X))\, \myE(g(Y))
\]
 for any pair of (measurable) non-negative functions $f:\CR_X \to [0, +\infty)$ and $g:\CR_Y\to [0, +\infty)$.


One can easily check that $X$ and $Y$ are independent
if and only if, for any non-negative function $g: \CR_Y \to \mR$, one
has
$$
\myE(g(Y)\mid X) = \myE(g(Y)).
$$

\begin{notation}
\label{not:ind}
Independence is a
property that involves two variables $X$ and $Y$ and an underlying
probability distribution $\myP$. Independence of $X$ and $Y$ relative to
$\myP$ will be denoted $(X\indp Y)_{\myP}$. However we will only write
$X\indp Y$ when there is no ambiguity on $\myP$.  
\end{notation}

More than independence, the concept of conditional independence will
be fundamental in this chapter. It requires three variables, say
$X,Y,Z$. Returning to the discrete case, one says  that $X$ and $Y$ are conditionally independent
given $Z$ is, for any $x\in \CR_X$, $y\in \CR_Y$ and $z\in \CR_Z$ such that $\myP(Z=z) > 0$,
\begin{equation}
\label{eq:cond.ind.discrete}
\myP(X=x, Y=y\mid Z=z) = \myP(X=x\mid Z=z)\, \myP(Y=y\mid Z=z).
\end{equation}
An equivalent statement is that, for any $z$ such that $\myP(Z=z) \neq 0$,  $X$ and $Y$ are independent when $\myP$
is replaced by the conditional distribution $\myP(\cdot\mid Z=z)$.

In the general case conditional independence means that,  for any pair of non-negative measurable functions $f$ and $g$,
\begin{equation}
\label{eq:cond.ind.gen}
\myE(f(X)g(Y)\mid Z) = \myE(f(X)\mid Z)\,\myE(g(Y)\mid Z).
\end{equation}
From now, we restrict our discussion to discrete random variables.

Multiplying both terms in \cref{eq:cond.ind.discrete} by $\myP(Z=z)^2$, we get the equivalent statement:
$X$ and $Y$ are
conditionally independent given $Z$ if and only if, 
\begin{equation}
\label{eq:cond.ind}
\forall x,y,z: \myP(X=x, Y=y, Z=z)\myP(Z=z) = \myP(X=x, Z=z)\, \myP(Y=y, Z=z).
\end{equation}
Note that the identity is meaningful, and always true, for $\myP(Z=z)=0$, so that this case does not need to be excluded anymore.

Conditional independence  can be interpreted by the statement that $X$ brings no more information on $Y$ than what
is already provided by $Z$: one has
\[
\myP(Y=y\mid X=x, Z=z) = \frac{\myP(Y=y, X=x, Z=z)}{\myP(X=x, Z=z)} = \frac{\myP(Y=y,Z=z)}{\myP(Z=z)}
\]
as directly deduced from \cref{eq:cond.ind}. (This computation being valid as soon as $\myP(X=x, Z=z) > 0$.)

\begin{notation}
\label{not:cond.ind}
To indicate
that $X$ and $Y$ are conditionally independent given $Z$ for the
distribution $\myP$, we will write
$(X\indp Y\mid Z)_{\myP}$ or simply $(X\indp Y\mid Z)$.
\end{notation}
So we have the equivalence:
$$
(X\indp Y\mid Z)_{\myP} \Leftrightarrow \big(\forall z: \myP(Z=z) > 0 \Rightarrow (X\indp Y)_{\myP( \mid Z=z)}\big).
$$

Absolute independence is like ``independence conditional to
no variable'', and we will use the notation $\emp$ for the ``empty'' random
variable that contains no information (for example, a set-valued
random variable that always returns the empty set, or any constant
variable). So we have the tautology
$$
X\indp Y \Leftrightarrow \cind{X}{Y}{\emp}.
$$

Note that, dealing with discrete variables, all previous definitions
automatically extend to groups of variables: for example, if $Z_1$,
$Z_2$ are two discrete variables, so is $Z = (Z_1,Z_2)$ and we
immediately obtain a definition for the conditional independence of $X$ and $Y$ given
$Z_1$ and $Z_2$, denoted $(X\indp Y\mid Z_1,Z_2)$. 

\subsection{Fundamental properties}

Proposition \ref{prop:cind} below lists important properties of conditional
independence that will be used
repeatedly in this chapter. 
 Before stating this proposition, we need
the following definition.

\begin{definition} 
\label{def:posit}
One says that the joint distribution of the random variables $(X_1,
\ldots, X_N)$ is positive if there exists subsets $\tilde R_k\subset \CR_{X_k}$, $k=1, \ldots, N$ such that $\myP(X_k\in \tilde R_k) = 1$ and:
\[
\myP(X_1=x_1, \ldots, X_N=x_N) > 0
\]
if $x_k\in \tilde R_k$, $k=1, \ldots, N$.
\end{definition}
Note that the condition implies $\myP(X_k = x_k) >0$ for all $x_k\in \tilde R_k$, so that $\tilde R_k = \{x_k\in \CR_{X_k}: \myP(X_k=x_k) > 0\}$, i.e., $\tilde R_k$ is the support of $P_{X_k}$. One can interpret the definition as expressing the fact that any conjunction of events for different
$X_k$'s has positive probability, as soon as each of them has positive probability (if all events may occur, then they may occur together). 

Note that the sets $\tilde R_k$ depend on $X_1, \ldots, X_N$. However, if this family of variables is fixed, there is no loss in generality in restricting the space $\CR_{X_k}$ to $\tilde R_k$ and there for assume that $\myP(X_1=x_1, \ldots, X_N=x_N) > 0$ everywhere.

\begin{proposition}
\label{prop:cind}
Let $X,Y,Z$ and $W$ be random variables. The following properties are true.
\begin{enumerate}[label=(CI\arabic*)]
\item Symmetry: $(X\indp Y\mid Z) \Rightarrow (Y\indp X\mid Z)$.
\item Decomposition: $(X\indp (Y, W)\mid Z) \Rightarrow (X\indp Y\mid Z)$.
\item Weak union: $(X\indp (Y,W)\mid Z)  \Rightarrow
(X\indp Y\mid (Z,W))$.
\item Contraction: $\cind{X}{Y}{Z} \text{ and } \cind{X}{W}{(Z,Y)} \Rightarrow
\cind{X}{(Y,W)}{Z}$.
\item Intersection: assume that the joint distribution of $W,Y$
  and $Z$ is positive. Then
\[
\cind{X}{W}{(Z,Y)} \text{ and } \cind{X}{Y}{(Z,W)}
\Rightarrow \cind{X}{(Y,W)}{Z}.
\]
\end{enumerate}
\end{proposition}
\begin{proof}
%
%
%
%
%

Properties (CI1) and (CI2) are easily deduced from \cref{eq:cond.ind} and left to the reader. To prove
the last three, we will use the notation $P(x), P(x,y)$ etc. instead of
$\myP(X=x), \myP(X=x, Y=y)$, etc. to save space. Identities are assumed to
hold for all $x,y,z,w$ unless stated otherwise. 

For (CI3), we must prove,
according to  \cref{eq:cond.ind}, that
\begin{equation}
\label{eq:ci.1}
P(x,y,z,w) P(z,w) = P(x,z,w) P(y,z,w)
\end{equation}
whenever $P(x,y,z,w) P(z) = P(x,z) P(y,z,w)$. Summing this last
equation over $y$ (or applying (CI2)) yields $P(x,z,w) P(z) = P(x,z)
P(z,w)$. We can note that all terms in \cref{eq:ci.1} vanish when
$P(z) = 0$, so that the identity is true in this case. When $P(z)\neq
0$, the right-hand side of
\cref{eq:ci.1} becomes
\begin{multline*}
(P(x,z) P(z,w)/P(z))P(y,z,w) = (P(x,z)P(y,z,w)/P(z))P(z,w)
=
P(x,y,z,w)P(z,w),
\end{multline*}
using once again the hypothesis. This proves (CI3).

\medskip

For (CI4), the hypotheses are
\[
\begin{cases}
P(x,y,z)P(z) = P(x, z)P(y,z)\\
P(x,y,z,w)P(y,z) = P(x,y,z)P(y,z,w)
\end{cases}
\]
and the conclusion must be
\begin{equation}
  \label{eq:ci.2}
P(x,y,z,w) P(z) = P(x,z)P(y,z,w).  
\end{equation}
Since \cref{eq:ci.2} is true when $P(y,z) = 0$, we assume that this
probability does not vanish and write
\begin{eqnarray*}
P(x,y,z,w)P(z) &=& P(x,y,z)P(z) P(y,z,w)/P(y,z) \\
&=& P(x,z) P(y,z) P(y,z,w)/P(y,z)\\
&=& P(x,z)P(y,z,w)
\end{eqnarray*}
yielding \cref{eq:ci.2}.
  
\medskip

For (CI5), assuming
\begin{equation}
\label{eq:ci5.assum}
\begin{cases}
P(x,y,z,w)P(y,z) = P(x, y,z)P(y,z,w)\\
P(x,y,z,w)P(z,w) = P(x,z,w)P(y,z,w),
\end{cases}
\end{equation}
we want to show that 
$$
P(x,y,z,w)P(z) = P(x,z)P(y,z,w).
$$
Since this identity is  true when any of the events $W=w,
Y=y$ or $Z=z$ has zero probability, we can assume that their
probabilities are positive, which, by assumption, also implies that all
joint probabilities are positive. From the two identities, we get
\[
P(x,y,z,w)/P(y,z,w) = 
P(x, y,z)/P(y,z)  = P(x,z,w)/P(z,w)
\]
This implies
\[
P(x,y,z) = P(y,z) P(x,z,w)/P(z,w)
\]
that we can sum over $y$ to obtain
\[
P(x, z) = P(z)P(x,z,w)/P(z,w)
\]
We therefore get
\[
P(x,y,z,w)/P(y,z,w) = 
 P(x,z,w)/P(z,w) = P(x,z)/P(z),
\]
which is what we wanted.
\end{proof}

A counter-example of (CI5) when the positivity assumption is not satisfied can be built as follows:
let $X$ be a Bernoulli random variable, and let $Y=W=X$. Let $Z$ be any
Bernoulli random variable, independent from $X$. Given $Z$ and $W$,
$X$ and $Y$ are constant and therefore independent. Similarly, given $Z$ and $Y$,
$X$ and $W$ are constant and therefore independent. However,
given $Z$, $X$ and $(Y,W)$ are not independent (they are equal and non
constant).

\subsection{Mutual independence}

Another concept of interest is the mutual (conditional) independence
of more than two random variables. The random variables
$(X_1, \ldots, X_n)$ are mutually conditionally independent given $Z$
if and only if 
$$
\myE(f_1(X_1)\cdots f_n(X_n)\mid Z) = \myE(f_1(X_1)\mid Z)\cdots \myE(f_n(X_n)\mid Z)
$$
for any non-negative measurable functions $f_1, \ldots, f_n$. In terms of discrete
probabilities, this can be written as
\begin{multline*}
P(X_1=x_1, \ldots, X_n=x_n, Z=z) P(Z=z)^{n-1} = \\
P(X_1=x_1, Z=z) \cdots
P(X_n=x_n, Z=z).
\end{multline*}
This will be summarized with the notation
\[
(X_1 \indp \cdots \indp X_n\mid Z).
\]

We have the proposition
\begin{proposition}
\label{prop:mut.indep}
For variables $X_1, \ldots, X_n$ and $Z$, the following properties are equivalent.
\begin{enumerate}[label=(\roman*)]
\item $(X_1 \indp \cdots \indp X_n\mid Z)$;
\item For all $S, T \sub \defset{1, \ldots, n}$ with $S\cap
  T=\emp$, we have:
$\cind{(X_i, i\in S)}{(X_j, j\in T)}{Z}$;
\item For all $s\in \defset{1, \ldots, n}$, we have: $\cind{X_s}{(X_t,
t\neq s)}{Z}$;
\item For all $s\in \defset{2, \ldots, n}$, we have:
$\cind{X_s}{(X_1,\ldots, X_{s-1})}{Z}$.
\end{enumerate}
\end{proposition}
\begin{proof}
It is clear that $\text{(i)} \Ria \cdots \Ria\text{(iv)}$ so it
suffices to prove that $\text{(iv)} \Ria \text{(i)}$. For this, simply
write (applying (iv) repeatedly to $s = n-1, n-2, \ldots$)
\begin{eqnarray*}
\myE(f_1(X_1)\cdots f_n(X_n)\mid Z) &=& \myE(f_1(X_1)\cdots f_{n-1}(X_{n-1})\mid Z)\, \myE(f_n(X_n) \mid Z) \\
&=& \myE(f_1(X_1)\cdots f_{n-2}(X_{n-2})\mid Z)\, \myE(f_{n-1}(X_{n-1}) \mid Z)\\
&& \quad \myE(f_n(X_n) \mid Z) \\
&\vdots &\\ 
&=& \myE(f_1(X_1)\mid Z)\cdots \myE(f_n(X_n)\mid Z).
\end{eqnarray*}
\end{proof}

\subsection{Relation with Information Theory}

Several concepts in information theory are directly related to
independence between random variables. Recall that the (Shannon) entropy of a discrete
probability distribution over a finite set $\CR$
is defined by 
\begin{equation}
\label{eq:entr.1}
\mathcal H(P) = -\sum_{\om\in \CR} \ln P(\om) P(\om).
\end{equation}
Similarly, the entropy of a random variable $X:\Omega \to \CR_X$ is defined by
\begin{equation}
\label{eq:entr.2}
\mathcal H(X) \defeq \mathcal H(P_X) = -\sum_{x\in \CR_X} \ln P(X=x) P(X=x).
\end{equation}
The entropy is always non-negative, and provides a  measure of the
uncertainty associated to $P$. For a given finite set $\CR$,
it is maximal when $P$ is uniform over $\CR$, and minimal (and vanishes) when $P$ is
supported by a single $\om\in \CR$ (i.e. $P(\om) = 1$). 

One defines the entropy of two or more random variables as the entropy
of their joint distribution, so that, for example,
\[
\mathcal H(X,Y) = -\sum_{(x,y)\in \CR_{X}\times \CR_Y} \ln \myP(X=x, Y=y) \myP(X=x, Y=y).
\]

We  have the proposition:
\begin{proposition}
\label{prop:entr.1} For random variables $X_1, \ldots, X_n$, one has
\[
\mathcal H(X_1, \ldots, X_n)\leq \mathcal H(X_1)+\cdots + \mathcal H(X_n)
\]
 with equality
if and only if 
$(X_1, \ldots, X_n)$ are mutually independent. 
\end{proposition}
\begin{proof}
The proof of this proposition uses  properties of the Kullback-Leibler divergence (c.f. \cref{eq:kl.definition}), given by, for two probability distributions
$\pi$ and $\pi'$ on a finite set $\CB$,
$$
\KL(\pi\|\pi') = \sum_{\om\in \CB} \pi(\om) \ln \frac{\pi(\om)}{\pi'(\om)}.
$$
with the convention $\pi\log(\pi/\pi') = 0$ if $\pi=0$ and $=\infty$ if $\pi>0$ and $\pi'=0$. 
Returning to  \cref{prop:entr.1}, a straightforward
computation (which is left to the reader) shows that 
\[
\mathcal H(X_1)+\cdots + \mathcal H(X_n) - \mathcal H(X_1, \ldots, X_n) = \KL(\pi\|\pi')
\]
with $\pi(x_1, \ldots, x_n) = \myP(X_1=x_1, \ldots, X_n=x_n)$ and
$\pi'(x_1, \ldots, x_n) = \prod_{k=1}^n \myP(X_k=x_k)$. This makes
 \cref{prop:entr.1} a direct consequence of 
\cref{prop:kl}. 
\end{proof}

The mutual information between two random variables $X$ and $Y$ is
defined by
\begin{equation}
\label{eq:mut.inf}
\mathcal I(X,Y) = \mathcal H(X) + \mathcal H(Y) - \mathcal H(X,Y).
\end{equation}
From  \cref{prop:entr.1}, $\mathcal I(X,Y)$ is nonnegative and
vanishes if and only if $X$ and $Y$ are independent. Also from the
proof of  \cref{prop:entr.1}, $\mathcal I(X,Y)$ is equal to
$\KL(P_{(X,Y)}\|P_X\otimes P_Y)$ where the first probability is the
joint distribution of $X$ and $Y$ and the second one the product of
the marginals of $X$ and $Y$, which coincides with $P_{X,Y}$ if and
only if $X$ and $Y$ are independent.

If $X$ and $Y$ are two random variables, and $y\in \CR_Y$
with $\myP(Y=y) > 0$,
the entropy of the conditional probability $x \mapsto \myP(X=x\mid Y=y)$ is
denoted $\mathcal H(X\mid Y=y)$, and is a function of $y$. The conditional entropy of $X$ given $Y$, denoted $\mathcal H(X\mid Y)$ is the
expectation of $\mathcal H(X\mid Y=y)$ for the distribution of $Y$, i.e.,
\begin{multline*}
\mathcal H(X\mid Y) = \sum_{y\in \CR_Y} \mathcal H(X\mid Y=y) \myP(Y=y)\\
 = -\sum_{x\in
  R_X} \sum_{y\in \CR_Y} \ln \myP(X=x\mid Y=y)
\myP(X=x, Y=y).
\end{multline*}
So, we have (with a straightforward proof)
\begin{proposition}
\label{prop:cond.ent}
Given two random variables $X$ and $Y$, we have
\begin{eqnarray}
\label{eq:cond.ent}
\mathcal H(X\mid Y) &=& -E_{X,Y}(\ln \myP(X=\cdot \mid Y=\cdot))\\
\nonumber
&=& \mathcal H(X,Y) - \mathcal H(Y)
\end{eqnarray}
\end{proposition}
This proposition also immediately  yields:
\begin{equation}
\label{eq:cond.ind.mut}
\mathcal I(X,Y) = \mathcal H(X) - \mathcal H(X\mid Y) = \mathcal H(Y) - \mathcal H(Y\mid X).
\end{equation}
The identity $\mathcal H(X,Y) = \mathcal H(X\mid Y) + \mathcal H(Y)$ that is deduced from
\cref{eq:cond.ent} can be generalized to more than two random
variables (the proof being left to the reader), yielding, if $X_1,
\ldots, X_n$ are random variables:
\begin{equation}
\label{eq:data.proc.eq}
\mathcal H(X_1, \ldots, X_n) = \sum_{k=1}^n \mathcal H(X_k\mid X_1, \ldots, X_{k-1}).
\end{equation}

If $Z$ is an additional random variable, the following identity is
obtained by applying the previous one to conditional distributions
given $Z=z$ and taking averages over $z$:
\begin{equation}
\label{eq:data.proc.eq.2}
\mathcal H(X_1, \ldots, X_n\mid Z) = \sum_{k=1}^n \mathcal H(X_k\mid X_1, \ldots, X_{k-1}, Z).
\end{equation}\\

The following proposition characterizes conditional independence in
terms of entropy.
\begin{proposition}
\label{prop:cond.ind.ent}
Let $X,Y$ and $Z$ be three random variables. The following statements
are equivalent.
\begin{enumerate}[label=(\roman*)]
\item $X$ and $Y$ are conditionally independent given $Z$.
\item $\mathcal H(X,Y\mid Z) = \mathcal H(X\mid Z) + \mathcal H(Y\mid Z)$
\item $\mathcal H(X\mid Y,Z) = \mathcal H(X\mid Y)$
\end{enumerate} 
Moreover, when (i) to (iii) are satisfied, we have:
\begin{itemize}
\item[(iv)] $\mathcal I(X,Y) \leq \min(\mathcal I(X,Z), \mathcal I(Y,Z))$.
\end{itemize}
\end{proposition}
\begin{proof}
From  \cref{prop:entr.1}, we have, for any three random
variables $X,Y,Z$, and any $z$ such that $P(Z=z) >0$,
\[
\mathcal H(X,Y\mid Z=z) \leq \mathcal H(X\mid Z=z) + \mathcal H(Y\mid Z=z).
\]
Taking expectations on both sides implies the important inequality
\begin{equation}
\label{eq:cond.ind.ineq}
\mathcal H(X,Y\mid Z) \leq \mathcal H(X\mid Z) + \mathcal H(Y\mid Z)
\end{equation}
and equality occurs if and only if $\myP(X=x, Y=y\mid Z=z) = \myP(X=x\mid Z=z)
\myP(Y=y\mid Z=z)$ whenever $\myP(Z=z)>0$, that is, if and only if $X$ and $Y$
are conditionally independent given $Z$. This proves that (i) and (ii)
are equivalent. The fact that (ii) and (iii) are equivalent comes from
\cref{eq:data.proc.eq.2}, which gives, for any three random variables
\begin{equation}
\label{eq:data.proc.eq.3}
\mathcal H(X, Y\mid Z) = \mathcal H(X\mid Y,Z) + \mathcal H(Y\mid Z).
\end{equation}

To prove that (i)-(iii) implies (iv), we note that
\cref{eq:cond.ind.ineq} and \cref{eq:data.proc.eq.3} imply that, for
any three random variables:
\[
\mathcal H(X\mid Y,Z) \leq \mathcal H(X\mid Y).
\]
If $X$ and $Y$ are conditionally independent given $Z$, then the right-hand side is equal to $\mathcal H(X\mid Z)$ and this yields
\[
\mathcal I(X,Y) = \mathcal H(X) - \mathcal H(X\mid Y) \leq \mathcal H(X) - \mathcal H(X\mid Z) = \mathcal I(X,Z).
\]
By symmetry, we must also have $\mathcal I(X,Y) \leq \mathcal I(Y,Z)$ so that (iv) is
true. 
\end{proof}
Statement (iv) is often called the data-processing inequality, and has
been used to infer conditional independence within gene networks
\cite{margolin2006}.

\section{Models on undirected graphs}
\label{sec:random.fields}
\subsection{Graphical representation of conditional independence}
An undirected graph is a
collection of vertexes and edges, in which edges link pairs of  vertexes
without order. Edges can therefore be identified to {\em subsets} of cardinality two
of the set of vertexes, $V$. This yields the definition:
\begin{definition}
\label{def:undir.grph}
An undirected graph $G$ is a pair $G = (V,E)$ where $V$ is a finite
set of vertexes and elements $e \in E$ are subsets $e=\{s,t\} \sub V$.
\end{definition}
Note that edges in undirected graphs are defined as sets, i.e.,
unordered pairs, which are delimited with braces in these notes. Later
on, we will use parentheses to represent ordered pairs, $(s,t)\neq
(t,s)$. We will write $s\sim_G t$, or simply $s\sim t$ to indicate
that $s$ and $t$ are connected by an edge in $G$ (we also say that $s$
and $t$ are neighbors in $G$).

\begin{definition}
\label{def:path.undir}

A path in an undirected graph $G=(V,E)$ is a finite sequence $(s_0, \ldots,
s_N)$ of vertexes such that $s_{k-1}\sim s_k \in E$. (A sequence,
$(s_0)$, of length 1 is also a path by extension.)

We say that $s$ and $t$ are connected by a path if either $s=t$ or
there exists a path $(s_0, \ldots,
s_N)$ such that $s_0=s$ and $s_N=t$.

A subset $S\sub G$ is connected if any pair of elements in $S$ can be
connected by a path.

A subset $T\sub G$ separates two other subsets $S$ and $S'$ if all
paths between $S$ and $S'$ must  pass in $T$. We will write
$\cind{S}{S'}{T}$ in such a case. 


\end{definition}

One of the goals of this chapter is to relate the notion of
conditional independence within a set of variables to separation in a
suitably chosen undirected graph with vertexes in one-to-one
correspondence with the variables. This will also justifies the similarity of
notation  used for separation and conditional independence.

We have the following simple fact:
\begin{lemma}
\label{lem:simple.1}
Let $G=(V,E)$ be an undirected graph, and $S,S',T\sub V$. Then
\[
\cind{S}{S'}{T} \Ria S\cap S' \sub T.
\]
\end{lemma}
Indeed, if $\cind{S}{S'}{T}$ and $s_0 \in S\cap
S'$, the path $(s_0)$ links $S$ and $S'$ and therefore must pass in
$T$. 

 \Cref{prop:cind} translates into similar
properties for separation:
\begin{proposition}
\label{prop:cind.2}
Let $(V, E)$ be an undirected graph and $S,T,U,R$ be subsets
of $V$. The following properties hold
\begin{enumerate}[label=(\roman*)]
\item $(S\indp T|U) \Leftrightarrow (T\indp S|U)$.
\item $(S\indp T\cup R|U) \Rightarrow (S\indp T|U)$.
\item $(S\indp T\cup R|U)  \Rightarrow
(S\indp T| U\cup R)$.
\item $\cind{S}{T}{U} \text{ and } \cind{S}{R}{U\cup T} \Leftrightarrow
\cind{S}{T\cup R}{U}$.
\item $U\cap R = \emp, \cind{S}{R}{U\cup T} \text{ and } \cind{S}{U}{T\cup R}
\Rightarrow \cind{S}{U\cup R}{T}$. 
\end{enumerate}
\end{proposition}
\begin{proof}
(i) is obvious, and for (ii) (and (iii)), if any path between $S$ and $T\cup R$
must pass by $U$, the same is obviously true for a path between $S$
and $T$. 

For the $\Rightarrow$ part of (iv), if a path links $S$ and
$T\cup R$, then it either links $S$ and $T$ and must pass through $U$
by the first assumption, or link $S$ and $R$ and therefore pass through
$U$ or $T$ by the second assumption. But if the path passes through
$T$, it must also pass through $U$ before by the first assumption. In
all cases, the path passes through $U$. The $\Leftarrow$ part of (iv)
is obvious.

Finally, consider (v) and take a path between two distinct elements in
$S$ and $U\cup R$. Consider the first time the path hits $U$ or $R$,
and assumes that it hits $U$ (the other case being treated similarly
by symmetry). Notice that the path cannot hit both $U$ and $R$ at the
same point since $U\cap R = \emp$. From the assumptions, the path must hit
$T\cup R$ before passing by $U$, and the intersection
cannot be  in $R$, so it is in $T$,
which is the conclusion we wanted.
\end{proof}

To make a connection between separation in graphs and conditional
independence between random variables, we consider a graph $G=(V,E)$
and a family of random variables $(\pe X s, s\in V)$ indexed by $V$. Each variable is assumed to take values in a set $F_s = \CR_{\pe X s}$. The collection of values taken by the random variables will be called configurations, and the sets $F_s, s\in V$ are called the state spaces.

Letting $\mathbb F$ denote the collection $(F_s, s\in V)$, we will denote the set of such configurations as $\CF(V, \mathbb F)$. Then $\mathbb F$ is clear from the context, we will just write $\CF(V)$. If $S \subset V$ and $x\in \CF(V, \mathbb F)$, the restriction of $x$ to $S$ is denoted $\pe x S = (\pe x s, s\in S)$. The set formed by those restrictions will be denoted $\CF(S, \mathbb F)$ (or just $\CF(S)$).

\begin{remark}
\label{rem:configurations}
Some care needs to be given to the definition of the space of configurations, to avoid ambiguities when two sets $F_s$ coincide. The configuration $x = (\pe x s, s\in V)$ should be understood, in an emphatic way, as the collection $\hat x = ((s,\pe x s), s\in V)$, which makes explicit the fact that $\pe x s$ is the value observed at vertex $s$. Similarly the emphatic notation for  $\pe x S \in \CF(V, \mathbb F)$ is $\pe {\hat x} S = ((s,\pe x s), s\in S)$. 

In the following, we will not use the emphatic notation to avoid overly heavy expressions, but its relevance should be clear with the following simple example. Take $V = \{1, 2, 3\}$ and $F_1 = F_2 = F_3 = \{0, 1\}$. Let $\pe x 1 = 0$, $\pe x 2 = 0$ and $\pe x 3 = 1$. Then the sub-configurations $\pe x {\{1,3\}}$ and $\pe x {\{2,3\}}$ both corresponds to values $(0,1)$, but we consider them as distinct. In the same spirit, $\pe x 1 = \pe x2$, but $\pe x {\{1\}} \neq \pe x{\{2\}}$
\end{remark}

If $S, T\subset V$ with $S\cap T=\emptyset$, $\pe x S\in \CF(S, \mathbb F)$, $\pe y T\in \CF(T, \mathbb F)$, we will denote their concatenation by 
$\pe x S \wedge \pe y T$, which is the configuration $z = (z_s, s\in S\cup T)\in \CF(T\cup S, \mathbb F)$ such that $\pe z s=\pe x s$ if $s\in S$ and $\pe z s=\pe y s$ if $s\in T$.

We define a {\em random field} over $V$ as a random configuration $X:\Omega \to \CF(V,  \mathbb F)$, that we will denote for short $X = (\pe X s, s\in V)$. If $S\subset V$, the restriction $\pe X S$ will also be denoted $(\pe X s, s\in
S)$.

We can now write the definition: 
\begin{definition}
\label{def:cond.grph}
Let $G = (V,E)$ be an undirected graph and  $X = (\pe X s, s\in V)$ a
random field over $V$. We say that $X$ is Markov (or
has the Markov property) relative to  $G$ (or is $G$-Markov, or is a
Markov random field on $G$) if and only if, for all $S, T, U \subset V$:
\begin{equation}
\label{eq:cond.grph}
\cind{S}{T}{U} \Rightarrow \cind{\pe X S}{\pe X T}{\pe X U}.
\end{equation}
\end{definition}
Letting the observation over an empty set $S$ be empty, i.e., $X_\emp =
\emp$, this definition includes the statement that, if $S$ and $T$ are
disconnected (i.e., there is no path between them: they are
separated by the empty set), then $\cind{\pe XS}{\pe XT}{\emp}$: $\pe XS$
and $\pe XT$ are independent. 

We will say that a probability distribution $\pi$ on $\CF(V)$ is $G$-Markov if its associated canonical random field $X = (\pe Xs, s\in V)$ defined on $\tilde \Omega = \CF(V)$ by $\pe X s (x) = \pe x s$ is  $G$-Markov.

\subsection{Reduction of the Markov property}

We now proceed, in a series of steps, to a simplification of
 \cref{def:cond.grph} in order to obtain a minimal number of conditional
independence statements. Note that, in its current form, 
\cref{def:cond.grph} requires to check \cref{eq:cond.grph} for any
three subsets of $V$, which provides a huge number of conditions. Fortunately, as we will see, these conditions are not independent, and checking a much smaller number of them will ensure that all of them are true. 

The first step for our reduction is provided by the following lemma.
\begin{lemma}
\label{lem:int.emp}
Let $G = (V,E)$ be an undirected graph and  $X = (X_s, s\in V)$ a
set of random variables indexed by $V$. Then $X$ is $G$-Markov if and
only if, for $S,T,U\sub V$,
\begin{equation}
\label{eq:cond.grph.lem}
S\cap U = T\cap U=\emp \text{ and }\cind{S}{T}{U} \Rightarrow \cind{\pe X S}{\pe X T}{\pe X U}.
\end{equation}
\end{lemma}
\begin{proof}
Assume that \cref{eq:cond.grph.lem} is true, and take any $S,T,U$ with $\cind{S}{T}{U}$.
Let
$A = S\cap U$, $B = T\cap U$ and $C = A\cup B$. Partition $S$ in $S= S_1\cup
A$, $T$ in $T_1\cup B$ and $U$ in $U_1\cup C$. From 
$\cind{S}{T}{U}$, we get $\cind{S_1}{T_1}{U}$. Since $S_1\cap U = T_1\cap
U = \emp$, this implies $\cind{\pe X {S_1}}{\pe X {T_1}}{\pe X U}$. But this implies
$\cind{(\pe X {S_1}, \pe X A)}{(\pe X {T_1}, \pe X B)}{\pe X U}$. Indeed, this property
requires
\begin{multline*}
P_X(\pe x {S_1}\wedge \pe x A\wedge \pe x {T_1}\wedge \pe x B\wedge \pe x {U_1}\wedge \pe y C) P^X(\pe x {U_1}\wedge \pe y C) \\
=  P_X(\pe x {S_1}\wedge \pe x A\wedge \pe x {U_1}\wedge \pe y C)P_X(\pe x {T_1}\wedge \pe x B\wedge \pe x {U_1}\wedge \pe y C)
\end{multline*}
If the configurations $\pe x A, \pe x B, \pe y C$ are not consistent (i.e.,
$\pe x t\neq \pe y t$ for some $t\in C$), then both sides vanish. So we can
assume $\pe x C = \pe y C$ and remove $\pe x A$ and $\pe x B$ from the expression,
since they are redundant. The resulting identity is true since it exactly
states that  $\cind{\pe X {S_1}}{\pe X {T_1}}{\pe X U}$.
\end{proof}

Define the set of neighbors of $s\in V$ (relative to the graph $G$) as
the set of $t\neq s$ such that $\{s,t\}\in E$ and denote this set by
$\mathcal V_s$. For $S\sub V$ define also
$$
\mathcal V_S = S^c\cap \bigcup_{s\in S} \mathcal V_s 
$$
which is the set of neighbors of all vertexes in $S$ that do not belong to
$S$. (Here $S^c$ denotes the complementary set of $S$, $S^c = V\setminus
S$.) Finally, let $\mathcal W_S$ denote the vertexes that are ``remote''
from $S$, $\CW_S = (S\cup \CV_S)^c$.

We have the following important reduction of the condition in  \cref{def:cond.grph}.
\begin{proposition}
\label{prop:mark.prop}
$X$ is Markov relative to $G$ if and only if, for any $S\sub V$,
\begin{equation}
\label{eq:mark.prop}
\cind{\pe X S}{\pe X {\CW_S}}{\pe X {\mathcal V_S}}.
\end{equation}
\end{proposition}
This says that 
$$
\myP(\pe X S = \pe x S \mid \pe X {S^c} = \pe x {S^c})
$$
only depends (when defined) on variables $\pe x t$ for $t\in S\cup \mathcal V_S$. 

\begin{proof} First note that $\cind{S}{\CW_S}{\mathcal V_S}$ is
  always true, since any
path reaching $S$ from $S^c$ must pass through $\mathcal V_S$. This immediately
proves the ``only if'' part of the proposition.

Consider now the ``if'' part. Take $S,T,U$ such that
$\cind{S}{T}{U}$. We want to prove that $\cind{X_S}{X_T}{X_U}$. According to
 \cref{lem:int.emp}, we can assume, without loss of generality, that
$S\cap U = T\cap U = \emp$.

Define $R$ as the set of vertexes $v$
in $V$ such that there exists a path between $S$ and $v$
that
does not pass in $U$. Then:
\begin{enumerate}[label=\arabic*.]
\item $S\sub R$: the path $(s)$ for $s\in S$ does
not pass in $U$ since $S\cap U=\emp$.
\item $U\cap R=\emp$ by definition. 
\item $\CV_R\sub U$: assume that there exists a
point $r$ in $\CV_R$ which is not in $U$. Then $r$ has a neighbor,
say $r'$ in $R$. By definition of $R$, there exists a path from $S$ to
$r'$ that does not hit $U$, and this path can obviously be extended by
adding $r$ at the end to obtain a path that still does not hit
$U$. But this implies that $r\in R$, which contradicts the fact that $\CV_R\cap
R=\emp$. 
\item $T\cap (R\cup \CV_R) = \emp$: if $t\in T$, then $t\not\in R$ from
$\cind{S}{T}{U}$ and $t\not \in \CV_R$ from $T\cap U=\emp$. 
\end{enumerate}

\begin{figure}[h]
\begin{center}
\includegraphics[width=.5\textwidth]{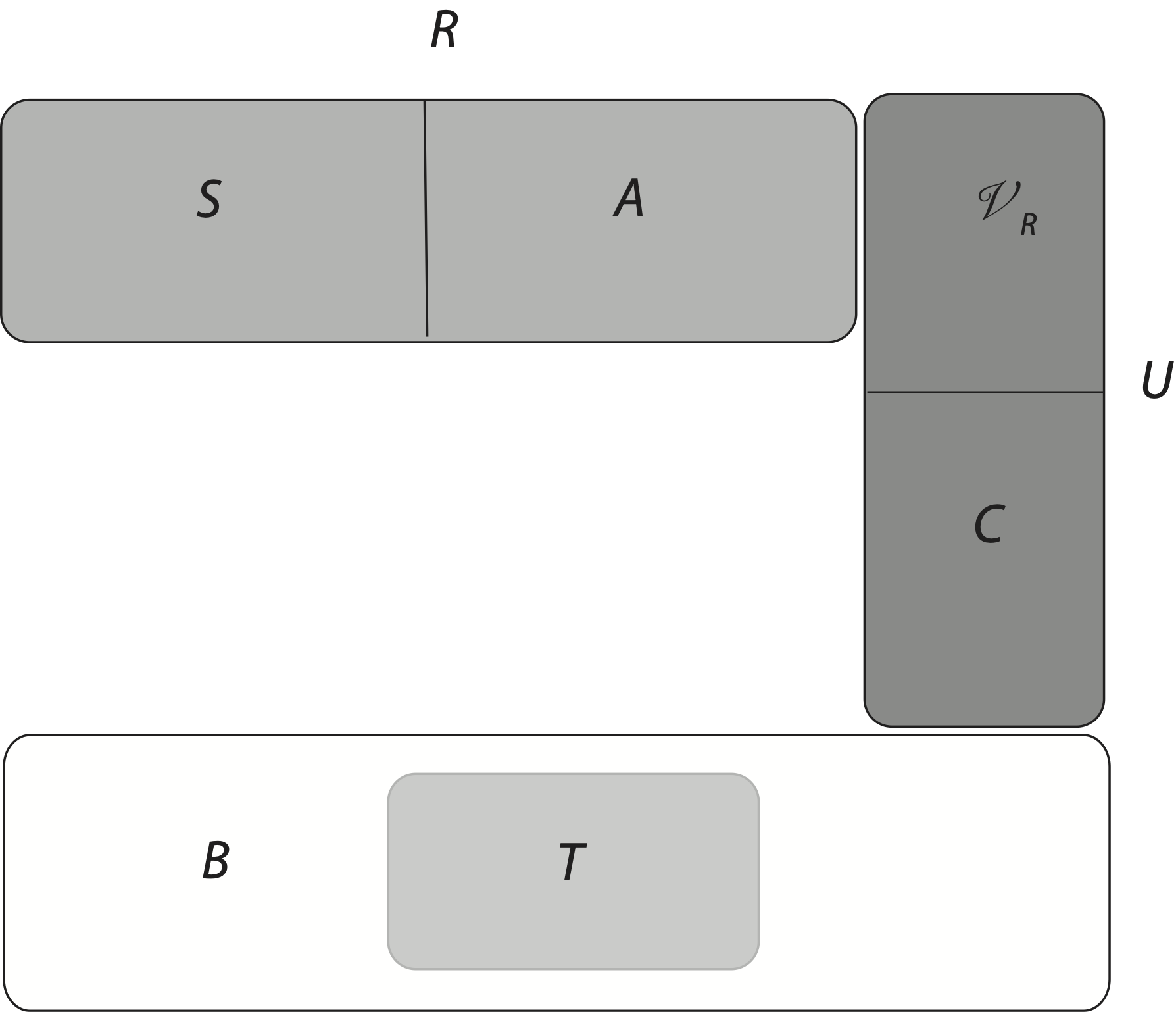}
\caption{\label{fig:prop8}See proof of 
  \cref{prop:mark.prop} for details}
\end{center}
\end{figure}

We can then
write (each decomposition being a partition, implicitly defining the sets $A$,
$B$ and $C$, see Fig. \ref{fig:prop8})
$R = S\cup A$, $U = \CV_R\cup C$, $(R\cup\CV_R)^c = T\cup C\cup B$,
and from $\cind{\pe X R}{\pe X{\CW_R}}{\pe X{\CV_R}}$, we get
$$
\cind{(\pe X S,\pe X A)}{(\pe X T,\pe  X C, \pe X B)}{\pe X {\CV_R}}
$$
which implies
$$
\cind{(\pe X S,\pe X A)}{(\pe X T, \pe X B)}{\pe X {U}}
$$
by (CI3), which finally implies $\cind{\pe X S}{\pe X T}{\pe X U}$ by (CI2).
\end{proof}

For positive probabilities, it suffices to consider singletons in  \cref{prop:mark.prop}.
\begin{proposition}
\label{prop:mark.prop.2}
If the joint distribution of $(\pe X s, s\in V)$ is positive and, for any $s\in V$,
\begin{equation}
\label{eq:mark.prop.2}
\cind{\pe X s}{\pe X {\CW_s}}{\pe X {\mathcal V_s}},
\end{equation}
then
$X$ is Markov relative to $G$. The converse statement is true
without the positivity assumption.
\end{proposition}
\begin{proof}
It suffices to prove that,  if \cref{eq:mark.prop} is true for $S$ and $T\sub V$, with $T\cap S =
\emp$, it is also true for $S\cup
T$. The result will then follow by induction. 

So, let $U = \mathcal V_{S\cup T}$ and $R = \mathcal W_{S\cup T} = V \setminus (S\cup T\cup
U)$. Then, we have
$$\cind{\pe X S}{\pe X {\mathcal W_S}}{\pe X {\mathcal V_{S}}} \Ria
\cind{\pe X S}{\pe X R}{(\pe X U, \pe X T)}
$$
because $R\sub \mathcal \CW_S$ (if $s\in \mathcal V_S$, then it is
either in $U$ or in $T$ and therefore cannot be in $R$). Similarly, 
$\cind{\pe X T}{\pe X R}{(\pe X U, \pe X S)}$, and (CI5) (for which we need $P$ positive) now
implies $\cind{(\pe X T, \pe X S)}{\pe X R}{\pe X U}$.
\end{proof}
To see that the positivity assumption is needed, consider the following
example with six variables $\pe X 1, \ldots, \pe X 6$, and a graph linking
consecutive integers and closing with an edge between 1 and
6. Assume that $\pe X 1 = \pe X 2=\pe X 4= \pe X 5$, and that $\pe X 1, \pe X 3$ and $\pe X 6$ are
independent. Then  \cref{eq:mark.prop.2}
is true, since, for $k=1,2,4,5$, $\pe X k$ is constant given its
neighbors, and $\pe X 3$ (resp. $\pe X 6$) is independent of the rest of the
variables. But $(\pe X 1, \pe X 2)$ is not independent of $(\pe X 4,\pe X 5)$ given
the neighbors $\pe X 3, \pe X 6$.

Finally, another statement equivalent to 
\cref{prop:mark.prop.2} is the following:
\begin{proposition}
\label{prop:mark.prop.3}
If the joint distribution of $(\pe X s, s\in V)$ is positive and, for any $s,t\in V$,
$$
s\not \sim_G t \Ria \cind{\pe X s}{\pe X t}{\pe X {V\setm \{s,t\}}},
$$
then
$X$ is Markov relative to $G$. The converse statement is true
without the positivity assumption.
\end{proposition}
\begin{proof}
Fix $s\in V$ and assume that 
$\cind{\pe X s}{\pe X R}{\pe X {V\setm R}}$ for any $R\sub \CW_s$ with cardinality
at most $k$ (the statement is true for $k=1$ by assumption). Consider
a set $\tilde R\sub \CW_s$ of cardinality $k+1$, that we decompose into
$R\cup\{t\}$ for some $t\in\tilde R$. We have
$\cind{\pe X s}{\pe X t}{\pe X {V\setm\tilde R}, X_R}$ from the initial hypothesis
and 
$\cind{\pe X s}{\pe X R}{\pe X {V\setm\tilde R}, X_t}$ from the induction
hypothesis. Using property (CI5), this yields 
$\cind{\pe X s}{\pe X {\tilde R}}{\pe X {V\setm\tilde R}}$. This proves the
proposition by induction.
\end{proof}

\begin{remark}
It is obvious from the definition of a $G$-Markov process that, if
$X$ is Markov for a graph $G=(V, E)$, it is automatically Markov
for any richer graph, i.e., any graph $\tilde G=(V, \tilde E)$ with
$E\sub \tilde E$. This is because separation in $\tilde G$ implies
separation in $G$. Moreover, any $X$ is $G$-Markov for the {\em complete graph} on $V$, for which $s\sim
t$ for all $s\neq t\in V$. This is because no pair of sets can be
separated in a complete graph. 

Any graph with respect to which $X$ is Markov must be richer than the 
graph $G_X = (V,E_X)$ defined by $s\not\sim_{G_X} t$ if and only
$\cind{\pe X s}{\pe X t}{\pe X {\{s,t\}^c}}$. This is true because, for
any graph $G$ for which $X$ is Markov, we have 
$$
s\not\sim_{G} t\Ria \cind{\pe X s}{\pe X t}{\pe X {\{s,t\}^c}} \Ria
s\not\sim_{G_X} t.
$$
Interestingly,  \cref{prop:mark.prop.3} states that $X$ is
$G_X$-Markov as soon as its joint distribution is positive. This
implies that $G_X$ is the minimal graph over which $X$ is Markov in
this case. 
\end{remark}

\subsection{Restricted graph and partial evidence}
Assume that some variables $\pe X T = (\pe X t, t\in T)$ (with $T\sub V$) have been observed,
with observed values $\pe x T = (\pe x t, t\in T)$. One would like to use this
 partial evidence to get additional information on the
remaining variables, $\pe X S$ where $S = V \setm T$. From the
probabilistic point of view, this means computing the conditional
distribution of $\pe X S$ given $\pe X T=\pe x T$.

One important property of $G$-Markov models is that the Markov
property is essentially conserved when passing to conditional
distributions. We introduce for this the following definitions.
\begin{definition}
\label{def:subg}
If $G = (V,E)$ is an undirected graph, a subgraph of $G$ is a graph
$G'=(V',E')$ with $V'\sub V$ and $E'\sub E$.

If $S\sub V$, the {\em restricted graph}, $G_S$, of $G$ to $S$ is defined by
\begin{equation}
\label{eq:rest.grph}
G_S = (S, E_S) \text{ with } E_S = \defset{e = \{s,t\}: s,t\in S \text {
and } e \in E}.
\end{equation}
\end{definition}

We have the following proposition.
\begin{proposition}
\label{prop:cond.dist.grph}
Let $G = (V,E)$ be an undirected graph and $X$ be $G$-Markov. Let
$S\sub V$ and $T = S^c$. Given a partial evidence $\pe x T$ such that
$P(\pe X T=\pe x T)>0$,  $\pe X S$, conditionally to $\pe X T=\pe x T$, is
$G_S$-Markov.
\end{proposition}
\begin{proof}
The proof is straightforward once it is noticed that 
\[
\cind{A}{B}{C}_{G_S} \Ria \cind{A}{B}{C\cup T}_G
\]
so that
\begin{eqnarray*}
\cind{A}{B}{C}_{G_S} &\Ria& \cind{\pe X A}{\pe X B}{\pe X C, \pe X T}_P\\
&\Ria& \cind{\pe X A}{\pe X B}{\pe X C}_{P(\cdot\mid \pe X T=\pe x T)}
\end{eqnarray*}
\end{proof}

\subsection{Marginal distributions}
The effect of taking marginal distributions for a $G$-Markov model is,
unfortunately, not as much a mild operation as computing conditional
distributions, in the sense that the conditional independence structure
of the marginal distribution may be much more complex than the original one.

Let $G = (V, E)$ be an undirected graph, and let $S$ be a subset of
$V$. Define the graph $G^S = (S, E^S)$ by $\{s, t\}\in E^S$ if and
only if $\{s, t\}\in E$ or there exist $u, u'\in S^c$ such that
$\{s,u\} \in E$, $\{t, u'\}\in E$ and $u$ and $u'$ are connected by a
path in $S^c$. In other terms $E^S$ links all $s,t\in S$ that can be
connected by a path, all but the extremities of which are included in
$S^c$.
  With this notation, the
following proposition holds.

\begin{proposition}
\label{prop:margin}
Let $G = (V, E)$ be an undirected graph, and $S\sub V$. Assume that
$X = (\pe X s, s\in V) $ is a family of random variables which is 
$G$-Markov. Then $\pe X S = (\pe x s, s\in S)$ is $G^S$-Markov.
\end{proposition}
\begin{proof}
It suffices to prove that, for $A,B,C\sub S$,
\[
\cind{A}{B}{C}_{G^S} \Ria \cind{A}{B}{C}_G.
\]
So, assume that $A$ and $B$ are
separated by $C$ in $G^S$. If a path connects $A$ and $B$ in
$G$, we can, by definition of $E^S$, remove from this path any portion
that passes in $S^c$ and obtain a valid path in $G^S$. By assumption,
this path must pass in $C$, and therefore so does the original path.
\end{proof}

The graph $G^S$ can be much more complex than the restricted graph
$G_S$ introduced in the previous section (note that, by definition,
$G^S$ is richer than $G_S$). Take, for example, the graph that
corresponds to ``hidden Markov models,'' for which (cf. \cref{fig:hmm})
\[
V = \{1, \ldots, N\} \times \{0,1\}
\]
and edges $\{s,t\}\in E$ have either $s=(k,0)$ and $t=(l,0)$ with
$|k-l|=1$, or $s = (k,0)$ and $t=(k,1)$. Let $S = \{1, \ldots,
N\}\times\{1\}$. Then, $G_S$ is totally disconnected ($E_S= \emp$),
since no edge in $G$ links two elements of $S$. In contrast, any pair
of elements in $S$ is connected by a path in $S^c$, so that $G^S$ is a
complete graph.  

\begin{figure}[h]
\begin{center}
		\begin{tikzpicture}[roundnode/.style={circle, draw=black!60, fill=gray!50, very thick, minimum size=10mm}, roundnode2/.style={circle, draw=black!60, fill=gray!5, very thick, , minimum size=10mm},]
		
		\node[roundnode](H1){};
		\node[roundnode](H2)[right=1cm of H1]{};
		\node[roundnode](H3)[right=1cm of H2]{};
		\node[roundnode](H4)[right=1cm of H3]{};
		\node[roundnode](H5)[right=1cm of H4]{};
		\node[roundnode](H6)[right=1cm of H5]{};
		
		\node[roundnode2](V1)[below=1cm of H1]{};
		\node[roundnode2](V2)[below=1cm of H2]{};
		\node[roundnode2](V3)[below=1cm of H3]{};
		\node[roundnode2](V4)[below=1cm of H4]{};
		\node[roundnode2](V5)[below=1cm of H5]{};
		\node[roundnode2](V6)[below=1cm of H6]{};

		\draw[black, thick,-] (H1.east) -- (H2.west) node[midway, above] {};
		\draw[black, thick,-] (H2.east) -- (H3.west) node[midway, above] {};
		\draw[black, thick,-] (H3.east) -- (H4.west) node[midway, above] {};
		\draw[black, thick,-] (H4.east) -- (H5.west) node[midway, above] {};
		\draw[black, thick,-] (H5.east) -- (H6.west) node[midway, above] {};
		\draw[black, thick,-] (H1.south) -- (V1.north) node[midway, above]{};
		\draw[black, thick,-] (H2.south) -- (V2.north) node[midway, above]{};
		\draw[black, thick,-] (H3.south) -- (V3.north) node[midway, above]{};
		\draw[black, thick,-] (H4.south) -- (V4.north) node[midway, above]{};
		\draw[black, thick,-] (H5.south) -- (V5.north) node[midway, above]{};
		\draw[black, thick,-] (H6.south) -- (V6.north) node[midway, above]{};
		\end{tikzpicture}
\caption{\label{fig:hmm} In this graph, variables in the lower row
  are conditionally independent given the first row, while their
  marginal distribution requires a completely connected graph.}
\end{center}
\end{figure}

\section{The Hammersley-Clifford theorem}

The Hammersley-Clifford theorem, which will be proved in this section, gives a complete description of positive
Markov processes relative to a given graph, $G$. It states that
positive $G$-Markov models are associated to families of positive local
interactions indexed by cliques in the graph. We now introduce each
of these concepts.

\subsection{Families of local interactions}
\begin{definition}
\label{def:set.int} 
Let $V$ be a set of vertexes and $(F_s, s\in V)$ a collection
of state spaces.  A family of local interactions is a collection of
non-negative functions $\Phi = (\phi_C, C\in \mathcal C)$ indexed over some subset
$\mathcal C$ of $\twop{V}$, such that each $\phi_C$ only depends on configurations restricted to $C$ (i.e., it is defined on $\CF(C)$), with values in $[0, +\infty)$.  (Recall that $\twop{V}$ is the set of all subsets of $V$.)

Such a family has order $p$ if no $C\in\mathcal C$ has cardinality larger
than $p$. A family of local interactions of order 2 is also called a
family of pair interactions.

Such a family is said to be consistent, if there exists an $x\in \CF(V)$
such that
\[
\prod_{C\in \mathcal C} \phi_C(\pe x C) \neq 0.
\]
To a consistent family of local interactions, one associates the
probability distribution $\pi^{\Phi}$ on $\CF(V)$  defined by
\begin{equation}
\label{eq:pi.phi}
\pi^\Phi(x) = \frac1{Z^\Phi} \prod_{C\in \mathcal C} \phi_C(\pe x C) 
\end{equation}
for all $x\in \CF(V)$, where $Z^\Phi$ is a normalizing constant.
\end{definition}

Given  $\mathcal C \sub \twop{V}$, define the  graph
$G_{\CC} = (V, E_{\CC})$ by letting $\{s,t\}\in E_\CC$ if and only if
there exists $C\in \CC$ such that $\{s,t\}\in C$. We then have the
following proposition.
\begin{proposition}
\label{prop:gc}
Let $\Phi = (\phi_C, C\sub \CC)$ be a consistent family of local
interactions, associated to some $\CC\sub \twop{V}$. Then the
associated distribution $\pi^\Phi$ is $G_\CC$-Markov.
\end{proposition}
\begin{proof}
Let $X$ be a random field associated with $\pi = \pi^\Phi$.
According to  \cref{prop:mark.prop}, we must show that, for
any $S\sub V$, one has
\[
\cind{\pe X S}{\pe X {\CW_S}}{\pe X {\mathcal V_S}}
\]
where $\CV_S$ is the set of neighbors of $S$ in $G_{\CC}$ and $\CW_S
= V\setm (\CV_S \cup S)$. Define the set $U_S$ by
\[
U_S = \bigcup_{C\in \CC, S\cap C\neq \emp} C
\]
so that $\CV_S = U_S \setm S$ and $\CW_S = V \setm U_S$. To prove
conditional independence, we need to prove that, for any $x\in F$:
\begin{equation}
\label{eq:gc}
\pi(x)\pi_{\CV_S}(\pe x {\CV_S}) = \pi_{U_S}(\pe x {U_S})\pi_{V\setm S}(\pe x {V\setm S})
\end{equation}
(where we denote $\pi_A$ the marginal distribution of $\pi$ on $\CF(A)$.)

From the
definition of $\pi$, we have
\begin{eqnarray*}
\pi(x) &=& \frac{1}{Z} \prod_{C\in \CC} \phi_C(\pe x C)\\
&=&\frac{1}{Z} \prod_{C:C\cap S\neq\emp} \phi_C(\pe x C) \, \prod_{C:C\cap
  S=\emp} \phi_C(\pe x C).
\end{eqnarray*}
The first term in the last product only depends on $\pe x {U_S}$, and the second one only on
$\pe x {V\setm S}$. 
Introduce the notation
\[
\left\{
\begin{aligned}
\mu_1(\pe x {\CV_S}) &= \sum_{\pe y {U_S}: \pe y {\CV_S} = \pe x {\CV_S}} \prod_{C:C\cap
  S\neq\emp} \phi_C(\pe x C) \\ 
\mu_2(\pe x {\CV_S}) &= \sum_{\pe y {V\setm S}: \pe y {\CV_S} = \pe x {\CV_S}} \prod_{C:C\cap
  S=\emp} \phi_C(\pe x C). 
\end{aligned}
\right.
\] 

With this notation, we have:
\[
\left\{
\begin{aligned}
\pi_{U_S}(\pe x {U_S}) &= (\mu_2(\pe x {\CV_S})/Z) \prod_{C:C\cap S\neq\emp} \phi_C(\pe x C) \\
\pi_{V\setm S}(\pe x {V\setm S}) &= (\mu_1(\pe x {\CV_S})/Z) \prod_{C:C\cap S =
  \emp} \phi_C(\pe x C) \\
\pi_{\CV_S}(\pe x {\CV_S}) &= \mu_1(\pe x {\CV_S}) \mu_2(\pe x {\CV_S})/Z
\end{aligned}
\right.
\]
from which \cref{eq:gc} can be easily obtained.
\end{proof}

We now discuss conditional distributions and marginals for processes
associated with local interactions. If $T\subset V$, we let $\pi_T = \pi_T^\Phi$ denote the marginal distribution of $\pi$ on $T$.

We start with a discussion of conditionals. Let
$\pi$ be associated with $\Phi$, and let
$S\sub V$ and $T = V\setm S$. Assume that a configuration $\pe y T$ is
given, such that $\pi_T(\pe y T) >0$, and consider the conditional
distribution
\begin{equation}
\label{eq:gc.cond}
\pi_{S\mid T}(\pe x S\mid \pe y T) = \pi(\pe x S\wedge \pe y T)/\pi_T(\pe y T).
\end{equation}
 We have the following proposition.
\begin{proposition}
\label{prop:gc.cond}
With the notation above, $\pi_{S\mid T}(\cdot\mid \pe y T)$ is associated to the
family of local interactions $\Phi_{|y_T} = (\phi_{\tilde C\mid \pe y T},
\tilde C\in
\CC_S)$ with
\[
\CC_S = \defset{\tilde C: \tilde C\sub S, \exists C\in\CC: \tilde C =
  C\cap S}
\]
and
\[
\phi_{\tilde C \mid \pe y T}(\pe x {\tilde C}) = \prod_{C\in \CC:C\cap S = \tilde C}
\phi_C(\pe x {\tilde C}\wedge \pe y {C\cap T}).
\]
\end{proposition}
\begin{proof}
From \cref{eq:gc.cond} and the definition of $\pi$, it is easy to sees that
\[
\pi_{S\mid T}(\pe x S\mid \pe y T) = \frac{1}{Z(\pe y T)} \prod_{C:C\cap S\neq \emp}
\phi_C(\pe x {C\cap S} \wedge \pe y {C\cap T}),
\]
where $Z(\pe yT)$ is a constant that only depends on $\pe yT$. The fact that
$\pi_{S\mid T}(\cdot \mid \pe y T)$ is associated to $\Phi_{|\pe yT}$ is then
obtained by reorganizing the product over distinct $S\cap C$'s.
\end{proof}

This result, combined with
 \cref{prop:gc}, is consistent with 
\cref{prop:cond.dist.grph}, in the sense that the restriction to
$G_\CC$ to $S$ coincides with the graph $G_{\CC_S}$. The easy proof is
left to the reader.

We now consider marginals, and more specifically marginals when only
one node is removed, which provides the basis for ``node elimination.'' 
\begin{proposition}
\label{prop:gc.m}
Let $\pi$ be associated to $\Phi = (\phi_C, C\in \CC)$ as above. Let
$t\in V$ and $S = V\setm \{t\}$. Define $\CC_t\in \twop{V}$ as the set
\[
\CC_t = \{C\in \CC: t\not \in C\} \cup \{\tilde C_t\}
\]
with 
\[
\tilde C_t = \bigcup_{C\in \CC: t\in C}C \setm \{t\}.
\]

Define a family of local interactions $\Phi_t = (\tilde \phi_{\tilde C},
\tilde C\in
\CC_t)$ by $\tilde \phi_{\tilde C} = \phi_{\tilde C}$ if $\tilde C\neq
\tilde C_t$ and: 
\begin{itemize}
\item If $\tilde C_t \not\in \CC$:
\[
\tilde \phi_{\tilde C_t}(\pe x{\tilde C_t}) = \sum_{\pe y t\in F_t} \prod_{C\in \CC, t\in C}
\phi_C(\pe x {\tilde C_t}\wedge \pe y t).
\]
\item If $\tilde C_t \in \CC$:
\[
\tilde \phi_{\tilde C_t}(\pe x {\tilde C_t}) = \phi_{C_t}(\pe x {\tilde C_t}) \sum_{\pe y t\in F_t} \prod_{C\in \CC, t\in C}
\phi_C(\pe x {C_t}\wedge \pe y t)
\]
\end{itemize}
Then the marginal, $\pi_S$, of $\pi$ over $S$ is the distribution
associated to $\Phi_t$.
\end{proposition}
The proof is almost straightforward by summing over possible values of
$y_t$ in the expression of $\pi$ and left to the reader. 

\subsection{Characterization of positive $G$-Markov processes}
 
Using families of local interactions is a typical way to build
graphical models in applications. The previous section describes a
graph with respect to which the obtained process is
Markov. Conversely, given a graph $G$, the Hammersley-Clifford theorems
states that families of local interactions over the cliques of $G$
are the only ways to build positive graphical models, which reinforces
the importance of this construction. We now pass to the statement and
proof of this theorem, starting with the following definition.

\begin{definition}
\label{def:cli}
Let $G = (V, E)$ be an undirected graph. A clique in $G$ is a nonempty
subset
$C\sub V$ such that $s\sim_G t$ whenever $s,t\in C$, $s\neq t$.
(In particular, subsets of cardinality one are always cliques.)
Cliques therefore form complete subgraphs of $G$.

The set of cliques of a graph $G$ will be denoted $\mathcal C_G$.

A clique that cannot be strictly included in any other clique is
called a maximal clique, and their set denoted $\mathcal C^*_G$. 

(Note that some authors call cliques what we refer to
as maximal cliques.)

\end{definition}

Given $G=(V,E)$, consider a random field $X= (\pe X s, s\in
V)$. We assume that $\pe X s$ takes values in a finite set $F_s$ with
$\myP(\pe X s=a) > 0$ for any $a\in F_s$ (this is no loss of generality since
one can always restrict $F_s$ to such $a$'s). If $S\sub V$, we denote as before  $\CF(S)$ the set of restrictions of configurations to $S$. With this
notation, $X$ is positive, according to  \cref{def:posit}, if and only if $P(X = x) >0$ for all
$x\in \CF(V)$. We will let $\pi = P_X$ be the probability distribution
of $X$, so that $\pi(x)  = \myP(X = x)$ and use as above the  notation: for $S, T \sub V$
\begin{equation}
\label{eq:px}
\begin{cases}
\pi_S(\pe xS) = \myP(\pe X S = \pe xS) \\
 \pi_{S\mid T}(\pe xS\mid \pe xT) =
\myP(\pe XS=\pe xs\mid \pe XT= \pe xT).
\end{cases}
\end{equation}
(For the first notation, we will simply write $\pi$ if $S=V$.)

We will also need to fix a reference, or ``zero,'' configuration in $\CF(V)$ that we will denote $\dszero = (\pe \dszero s, s\in V)$, with $\pe \dszero s\in F_s$ for all $s$. We can choose it arbitrarily.  Given this, we
have the theorem:
\begin{theorem}[Hammersley-Clifford]
\label{th:ham.clif.1}
With the previous notation, $X$ is a positive $G$-Markov process if
and only if its distribution, $\pi$, is associated to a family of local interactions $\Phi =
(\phi_C, C\sub \CC_G)$ such that $\phi_C(\pe xC) > 0$ for all $\pe xC\in
\CF(C)$.

Moreover, $\Phi$ is uniquely characterized by the additional
constraint: 
$\phi_C(\pe xC) = 1$ as soon as there exists $s \in C$ such that $\pe xs = \pe \dszero s$. 
\end{theorem}

Letting $\la_C = -\ln \phi_C$, we get an equivalent formulation of the
theorem in terms of potentials, where a potential is defined as a
family of functions
\[
\La = (\la_C, C\in \CC)
\]
indexed by a subset $\CC$ of $\twop{V}$, such that $\la_C$ only
depends on $\pe xC$. The distribution
associated to $\La$ is
\begin{equation}
\label{eq:pot.dec}
\pi(x) = \frac{1}{Z_\La} \exp\Big(-\sum_{C\in \CC}
\la_C(\pe xC)\Big).
\end{equation}
With this terminology, we trivially have an equivalent formulation:
\begin{theorem}
\label{th:ham.clif}
$X$ is a positive $G$-Markov process if
and only if its distribution, $\pi$, is associated to a potential $\La =
(\la_C, C\sub \CC_G)$.

Moreover, $\La$ is uniquely characterized by the additional
constraint:
$\la_C(\pe xC) = 0$ as soon as there exists $s \in C$ such that $\pe xs = \pe \dszero s$. 
\end{theorem}
 
We now prove this theorem.
\begin{proof}
Let us start with the ``if'' part.
If $\pi$ is associated to a potential over $\CC_G$, we have already
proved that $\pi$ is $G_{\CC_G}$-Markov, so that it suffices to prove
that $G_{\CC_G} = G$, which is almost obvious: If $s\sim_G t$, then
$\{s,t\}\in \CC_G$ and $s\sim_{G_{\CC_G}} t$ by definition of
$G_{\CC_G}$. Conversely, if $s\sim_{G_{\CC_G}} t$, there exists $C\in
  \CC_G$ such that $\{s,t\}\sub C$, which implies that $s\sim_G t$, by
  definition of a clique.

We now prove the ``only if'' part, which relies on a combinatorial lemma, which is one of
M\"obius's inversion formulas. 
\begin{lemma}
\label{lem:mobius}
Let $A$ be a finite set and $f: \twop{A}\to
\mR$, $B\mapsto f_B$. Then, there is a unique function $\la: \twop{A} \to \mR$ such that
\begin{equation}
\label{eq:mob.1}
\forall B\sub A, \ f_B = \sum_{C\sub B} \la_C,
\end{equation}
and $\la$ is given by
\begin{equation}
\label{eq:mob.2}
\la_C = \sum_{B\sub C} (-1)^{|C|-|B|} f_B.
\end{equation}
\end{lemma}
To prove the lemma, first notice that the space $\bfF$ of functions $f: \twop{A}
\to \mR$ is a vector space of dimension $2^{|A|}$ and that the
transformation $\phi: \la\mapsto f$ with $f_B = \sum_{C\sub B} \la_C$ is
linear. It therefore suffices to prove that, given
any $f$, the function $\la$ given in \cref{eq:mob.2} satisfies
$\phi(\la)=f$, since this proves that $\phi$ is onto from $\bfF$ to $\bfF$
and therefore necessarily one to one.

So consider $f$ and $\la$ given by \cref{eq:mob.2}. Then
\[
\phi(\la)(B) =\sum_{C\sub B} \la_C = \sum_{C\sub B} \sum_{\tilde B \sub C} (-1)^{|C|-|\tilde B|}
f_{\tilde B}
= \sum_{\tilde B \sub B} \left(\sum_{C \supset \tilde B, C\sub B} (-1)^{|C|-|\tilde B|}\right)
f_{\tilde B}
= f_B
\]
The last identity comes from the fact that, for any finite set $\tilde
B\sub B, \tilde
B\neq B$, we have
$$
\sum_{C \supset \tilde B, C\sub B} (-1)^{|C|-|\tilde B|} = 0
$$
(for $\tilde B=B$, the sum is obviously equal to 1). Indeed, if $s\in
B$, $s\not \in \tilde B$, we have
\begin{eqnarray*}
\sum_{C \supset \tilde B, C\sub B} (-1)^{|C|-|\tilde B|} &=&
\sum_{C \supset \tilde B, C\sub B, s\in C} (-1)^{|C|-|\tilde B|} + \sum_{C \supset \tilde B, C\sub B, s\not\in C} (-1)^{|C|-|\tilde
B|} \\
&=&
\sum_{C \supset \tilde B, C\sub B, s\not \in C} ((-1)^{|C\cup
\{s\}|-|\tilde B|} + (-1)^{|C|-|\tilde
B|}) \\
&=&0.
\end{eqnarray*}

So the lemma is proved. We now proceed to proving the existence and
uniqueness statements in 
\cref{th:ham.clif}. Assume that $X$ is $G$-Markov and positive. 
Fix $x \in \CF(V)$ and consider the function, defined on $\twop{V}$ by
$$
f_B(\pe xB) = - \ln\frac{\pi(\pe xB\wedge \pe \dszero {B^c})}{\pi(\dszero)}.
$$
Then, letting
$$
\la_C(\pe xC) = \sum_{B \sub C} (-1)^{|C|-|B|} f_B(\pe xB),
$$
we have
$
f_B(\pe xB) = \sum_{C\sub B} \la_C(\pe xC).
$
In particular, for $B=V$, this gives 
$$
\pi( x) = \frac{1}{Z} \exp\Big(-\sum_{C\sub V}
\la_C(\pe xC)\Big)
$$
with $Z =
P(\dszero)$.
We now prove that $\la_C(\pe xC) = 0$ if $\pe xs = \pe \dszero s$ for some $s\in V$ or if
$C\not \in \mathcal C_G$. This will prove
\cref{eq:pot.dec} and the existence statement in 
\cref{th:ham.clif}. 

So, assume $\pe xs = \pe \dszero s$. Then, for any $B$ such that $s\not\in B$, we
have $f_B (\pe xB)= f_{\{s\}\cup B}(\pe x {\{s\}\cup B})$. Now take $C$ with $s\in C$. We have
\begin{eqnarray*}
\la_C(\pe xC) &=& \sum_{B \sub C, s\in B} (-1)^{|C|-|B|} f_B(\pe xB) + \sum_{B
\sub C, s\not \in B} (-1)^{|C|-|B|} f_B(\pe xB)\\
&=& \sum_{B \sub C, s\not \in B} (-1)^{|C|-|B\cup\{s\}|} f_{B\cup\{s\}}(\pe x {B\cup\{s\}}) + \sum_{B
\sub C, s\not \in B} (-1)^{|C|-|B|} f_B(\pe xB)\\
&=& \sum_{B \sub C, s\not \in B} ((-1)^{|C|-|B\cup\{s\}|} + (-1)^{|C|-|B|}) f_B(\pe xB)\\
&=&\dszero.
\end{eqnarray*}

Now assume that $C$ is not a clique, and let $s\neq t\in C$ such that
$s\not\sim t$. We can write, using
decompositions similar to the above,
$$
\la_C(\pe xC) = \sum_{B \sub C\setm\{s,t\}} (-1)^{|C|-|B|}
\left(f_{B\cup\{s,t\}}(\pe x{B\cup\{s,t\}}) - f_{B\cup\{s\}}(\pe x{B\cup\{s\}}) - f_{B\cup\{t\}}(\pe x{B\cup\{t\}}) +
f_B(\pe xB)\right).
$$
But, for $B \sub C\setm\{s,t\}$, we have
\begin{eqnarray*}
f_{B\cup\{s,t\}}(\pe x{B\cup\{s,t\}}) - f_{B\cup\{s\}}(\pe x{B\cup\{s\}}) &=& -
\ln\frac{\pi(\pe x{B\cup\{s,t\}}\wedge
\pe \dszero{B^c\setm\{s,t\}})}{\pi(\pe x{B\cup\{s\}}\wedge \pe \dszero{B^c\setm\{s\}})}
\\
&=& \ln \frac{\pi_t(\pe xt\mid \pe x{B\cup\{s\}}\wedge
\pe \dszero{B^c\setm\{s,t\}})}{\pi_t(\pe \dszero t\mid \pe x{B\cup\{s\}}\wedge
\pe \dszero {B^c\setm\{s,t\}})}
\end{eqnarray*}
and 
\begin{eqnarray*}
f_{B\cup\{t\}}(\pe x{B\cup\{t\}}) - f_{B}(\pe xB) &=& -
\ln\frac{\pi(\pe x{B\cup\{t\}}\wedge
\pe \dszero {B^c\setm\{t\}})}{\pi(\pe x{B}\wedge \pe \dszero{B^c})}
\\
&=& \ln \frac{\pi_t(\pe x t\mid \pe x{B}\wedge
\pe \dszero {B^c\setm\{t\}})}{\pi_t(\pe \dszero t\mid \pe x{B}\wedge
\pe \dszero {B^c\setm\{t\}})}.
\end{eqnarray*}
So, we can write
$$
\la_C(\pe xC) = \sum_{B \sub C\setm\{s,t\}} (-1)^{|C|-|B|}
\ln \frac{\pi_t(\pe x t\mid \pe x {B\cup\{s\}}\wedge
\pe \dszero {B^c\setm\{s,t\}})\pi_t(\pe \dszero t\mid \pe x{B}\wedge
\pe \dszero{B^c\setm\{t\}})}{\pi_t(\pe \dszero t\mid x_{B\cup\{s\}}\wedge
\pe \dszero {B^c\setm\{s,t\}})\pi_t(\pe xt\mid \pe x{B}\wedge
\pe \dszero {B^c\setm\{t\}})}
$$
which vanishes, because
$$
\pi_t(\pe xt\mid \pe x{B\cup\{s\}}\wedge
\pe \dszero{B^c\setm\{s,t\}}) = \pi_t(\pe xt\mid \pe x{B}\wedge
\pe \dszero {B^c\setm\{t\}})
$$
when $s\not \sim t$.

To prove uniqueness, note that, for any zero-normalized $\La$ satisfying
\cref{eq:pot.dec}, we must have $\pi(\dszero) = 1/Z$ and therefore, for any
$x$,
$$
-\ln \frac{\pi(\pe xB\wedge \pe \dszero {B^c})}{\pi(0)} = \sum_{C\sub B} \la_C(\pe x C)
$$
(extending $\La$ so that $\la_C=0$ for $C\not\in \CC_G$).
But, from  \cref{lem:mobius}, this uniquely defines $\La$.
\end{proof}

The exponential form of the distribution in the Hammersley-Clifford
theorem is related to what is called a Gibbs distribution in statistical
mechanics. More precisely:
\begin{definition}
\label{def:gibbs}
Let $\mathcal F$ be a finite set and $W:\mathcal F \to \mR$ be a scalar
function. The Gibbs distribution with energy $W$ at temperature $T>0$
is defined by
$$\pi(x) = \frac{1}{Z_T}e^{-\frac{W(x)}{T}}, \ x\in \CF$$
The normalizing constant $Z_T = \sum_{y\in\CF} \exp(-W(y)/T)$ is
called the partition function.

If $\La = (\la_C, C\sub V)$ is a potential then its associated energy is
$$
W(x) = \sum_{C\sub V} \la_C(\pe x C).
$$
\end{definition}
So the Hammersley-Clifford theorem implies that any positive
$G$-Markov model is associated to a unique zero-normalized
potential defined over the cliques of $G$. This representation can also be used to
provide an alternate proof of  \cref{prop:mark.prop.3},
which is left to the reader. Finally, one can restate 
\cref{prop:gc.cond} in terms of potentials, yielding:
\begin{proposition}
Let $P$ be a Gibbs distribution associated with a zero-normalized potential $\la = (\la_C,
C\sub V)$. Let $S\sub V$ and $T = S^c$. Then the conditional
distribution of $\pe X S$ given $\pe X T = \pe x T$ is the Gibbs distribution
associated with the zero-normalized potential $\tilde \la = (\tilde \la_C,
C\sub S)$ where 
$$
\tilde \la_C(\pe y C) = \sum_{C'\sub V, C'\cap S = C} \la_{C'}(\pe yC\wedge
\pe x {T\cap C'}).
$$
\end{proposition}

\section{Models on acyclic graphs}

\subsection{Finite Markov chains}
We now review a few important examples of Markov processes $X$
associated to specific graphs $G=(V,E)$. We will always denote by
$F_s$ the space in which $\pe X s$ takes his values, for $s\in V$.

The simplest example of $G$-Markov process (for any graph $G$) is the
case when $X= (\pe X s, s\in V)$ is a collection of independent random
variables. In this case, we can take $G_X=(V, \emp)$, the totally
disconnected graph on $V$. Another simple fact is that, as already
remarked, any $X$ is Markov for the
complete graph $(V, \boldsymbol{\CP}_2(V))$ where $\boldsymbol{\CP}_2(V)$ contains
all subsets of $V$ with cardinality 2.

Beyond these trivial (but nonetheless
important) cases, the simplest graph-Markov processes are those
associated with linear graphs, providing finite Markov
chains. For this, we let $V$ be a finite ordered set, say,
$$
V = \defset{0, \ldots, N}.
$$
We say that $X$ is a finite Markov chain if, for any $k=1,
\ldots, N$
$$
\cind{\pe X k}{(\pe X 0, \ldots, \pe X{k-2})}{\pe X{k-1}}.
$$
So we have the identity
\begin{multline*}
\myP(\pe X0=\pe x0, \ldots, \pe Xk=\pe x k)P(\pe X{k-1}=\pe x{k-1})\\
 = \myP(\pe X0= \pe x0,
\ldots, \pe X{k-1}= \pe x{k-1})\myP(\pe X{k-1}=\pe x{k-1}, \pe Xk=\pe xk).
\end{multline*}
The distribution of a Markov chain is therefore fully specified by
$\myP(\pe X0 = \pe x0), x_0\in F_0$ (the initial distribution) and  the conditional probabilities
\begin{equation}
\label{eq:trans.mc}
p_k( \pe x{k-1}, \pe xk) = \myP(\pe Xk=\pe xk \mid \pe X{k-1} = \pe x{k-1})
\end{equation}
(with an arbitrary choice when $\myP(\pe X{k-1}=\pe x{k-1}) = 0$). Indeed,
assume that $\myP(\pe X0=\pe x0,
\ldots, \pe X{k-1}=\pe x{k-1})$ is known (for all $\pe x0, \ldots,
\pe x{k-1}$). Then, either:
\begin{enumerate}[label = (\roman*)]
\item $\myP(\pe X0= \pe x0,
\ldots, \pe X{k-1}=\pe x{k-1})=0$, in which case 
\[
\myP(\pe X0=\pe x0,
\ldots, \pe X{k}=\pe x{k})=0
\]
 for any $\pe xk$, or:
 \item $\myP(\pe X0=\pe x0,
\ldots, \pe X{k-1}=\pe x{k-1})>0$, in which case, necessarily, $\myP(\pe X{k-1}=\pe x{k-1}) > 0$, and
$$
\myP(\pe X0=\pe x0, \ldots, \pe Xk=\pe xk) = p_k(\pe x {k-1}, \pe xk) \myP(\pe X0=\pe x0,
\ldots, \pe X{k-1}=\pe x{k-1}).
$$
\end{enumerate}
Note that $p_k$ in \cref{eq:trans.mc} is a transition probability (according to \cref{def:trans.prob}) between $F_{k-1}$ and $F_k$.

We have the following identification of a finite Markov chain with a
graph-Markov process:
\begin{proposition}
\label{prop:m.chain}
Let $X=(\pe X0, \ldots, \pe XN)$ be a finite Markov chain, such that $X$ is positive. Then $X$ is $G$-Markov for the linear graph $G=(V,E)$ with
\begin{eqnarray*}
V &=& \defset{1, \ldots, N}\\
E &=& \defset{\{1,2\}, \ldots, \{N-1, N\}}.
\end{eqnarray*}
The converse is true without the positivity assumption: a $G$-Markov
process for the graph above is always a finite Markov chain. 
\end{proposition}
\begin{proof}
We prove the direct statement (the converse one being obvious).
Let $s$ and $t$ be nonconsecutive distinct integers, with, say,
$s<t$. From the Markov chain assumption, we have
$$
\cind{\pe Xt}{(\pe Xs, \pe X{\defset{1,t-2}\setm\{s,\}})}{\pe X{t-1}},
$$
 which, using
(CI3), yields $\cind{\pe Xt}{\pe Xs}{\pe X{\defset{1, \ldots,
t-1}\setm\{s\}}}$. Define $\pe Yu = \pe X{\defset{1, \ldots,
u}\setm\{s,t\}}$: what we have proved is
$\cind{\pe Xt}{\pe Xs}{\pe Yt}$.

We now proceed by induction and assume that $\cind{\pe Xt}{\pe Xs}{\pe Yu}$ for some
$u\geq t$. Then, we have
$\cind{\pe X{u+1}}{(\pe Xs, \pe Xt, \pe Y{u-1})}{\pe Xu}$, which implies (from (CI3))
$\cind{\pe X{u+1}}{\pe Xt}{\pe Xs,\pe Yu}$. Applying (CI4) to 
$\cind{\pe Xt}{\pe Xs}{\pe Yu}$ and $\cind{\pe X{t}}{\pe X{u+1}}{\pe Xs,\pe Yu}$, we obtain
$\cind{\pe Xt}{(\pe Xs, \pe X{u+1})}{\pe Yu}$ and finally,
$\cind{\pe Xt}{\pe Xs}{\pe Y{u+1}}$. By induction, this gives
$\cind{\pe Xt}{\pe Xs}{\pe YN}$ and therefore 
\cref{prop:mark.prop.3} now implies that  $X$ is $G$-Markov.

(The proposition can also be proved as a consequence of the
decomposition
\[
\myP(\pe X0=\pe x0, \ldots, \pe XN=\pe xN) = \myP(\pe X0= \pe x0) p_1(\pe x0, \pe x1)\ldots
p_N(\pe x{N-1}, \pe xN).)
\]

\end{proof}

\subsection{Undirected acyclic graph models and trees}
The situation with acyclic graphs is only slightly more complex than
with linear graphs, but will require a few new definitions, including those of
directed graphs and trees.

The difference between directed and undirected graphs is that the
edges of
the former are ordered pairs, namely:
\begin{definition}
\label{def:dir.grph}
A (finite) directed graph $G$ is a pair $G=(V,E)$ where $V$ is a
finite set of vertexes and $E$ is a subset of 
\[V\ti V \setm
\defset{(s,s), s\in V},
\]
which satisfies, in addition,
\[
(s,t)\in E\Ria (t,s)\not\in E.
\]
\end{definition}
So, for directed graphs, edges $(s,t)$ and $(t,s)$ have different meanings, and we allow at most one of them in $E$. We say that the edge $e=(s,t)$ stems from $s$ and points
to $t$. The {\em parents} of a vertex $s$ are the vertexes $t$ such
that $(t,s)\in E$, and its children are the vertexes $t$ such that
$(s,t)\in E$. We will also use the notation $s\to_G t$ to indicate that $(s,t)
\in E$ (compare to $s\sim_G t$ for undirected graphs).

\begin{definition}
\label{def:path.dir}
A path
in a directed graph $G=(V,E)$ is a sequence $(s_0, \ldots, s_N)$ such that, for
all $k=1, \ldots, N$, $s_k\to_G  s_{k+1}$ (this includes the ``trivial'', one-vertex, paths $(s_0)$). 
(The definition was the same for undirected graph, replacing  $s_k\to_G  s_{k+1}$ by  $s_k\sim_G  s_{k+1}$.) 
For both directed and undirected cases, one says that a path is closed if $s_0 = s_N$.

In an undirected graph, a path is folded if it can be written as $(s_0, \ldots, s_{N-1}, s_N, s_{N-1}, \ldots, s_0)$. 

If $G=(V,E)$ is directed, one says that $t\in V$ is a descendant of $s\in V$ (or that $s$ is an ancestor of $t$) if there exists a path starting at $s$ and ending at $t$. In particular, every vertex is both a descendant and an ancestor of itself.
\end{definition}

We finally define acyclic graphs.
\begin{definition}
\label{def:acyc}
A loop in a  directed (resp. an undirected) graph $G$ is a  path $(s_0, s_1,
\ldots, s_{N})$, with $N\geq 3$,  such that $s_{N} = s_0$, which passes only once through $s_0, \ldots,
s_{N-1}$ (no self-intersection except at the end).

A (directed or  undirected) graph $G$ is acyclic if it contains no
loop.
\end{definition}

The following property will be useful.
\begin{proposition}
\label{prop:loop}
In a directed graph, any non-trivial closed path 
 contains a loop (i.e., one
can delete vertexes from it to finally obtain a loop.)

In an undirected graph, any non-trivial  closed path which is not a union of folded paths contains a loop.
\end{proposition}
\begin{proof}
Take $\ga = (s_0, s_1,
\ldots, s_{N})$ with $s_N=s_0$. The path being non-trivial means $N> 1$.

First take the case of a directed graph. Clearly, $N\geq 3$ since a two-vertex path cannot be closed in an directed graph. Consider the first
occurrence of a repetition, i.e., the first index for which 
\[
s_j\in \{s_0, \ldots, s_{j-1}\}.
\]
Then there is a unique $j'\in \{0, \ldots, j-1\}$ such that $s_{j'} = s_j$, and the path $(s_{j'},
\ldots, s_{j-1})$ must be a loop (any repetition in the sequence would contradict the fact that $j$ was the first occurrence. This proves the result in the directed case.

Consider now an undirected graph. We can recursively remove all folded subpaths, by keeping everything but their initial point, since each such operation still provide a path at the end. Assume that this is done, still denoting the  remaining path $(s_0, s_1, \ldots, s_{N})$, which therefore has no folded subpath. We must have $N\geq 3$ since $N=1$ implies that the original path was a union of folded paths, and $N=2$ provides a folded path. 
Let, $0\leq j' < j$ be as in the directed case. Note that one must have $j' < j-2$, since $j'=j-1$ would imply an edge between $j$ and itself and $j'=j-2$ induces a folded subpath. But this implies that  $(s_{j'},
\ldots, s_{j-1})$ is a loop. 
\end{proof}

Directed acyclic graphs (DAG) will be important for us, because they
are associated with Bayesian networks that we will discuss later. For now, we are
interested with undirected acyclic graphs and their relation to
trees, which form a
subclass of directed acyclic graphs, defined as follows.
\begin{definition}
\label{def:tree}
A forest is a directed acyclic graph with the additional
requirement that each of its vertexes has at most one parent.

A root in a forest is a vertex that has no parent. A
forest with a single root is called a tree.
\end{definition}

It is clear that a forest has at least one root, since one could
otherwise describe a nontrivial loop by starting from a any
vertex and passing to its parent until the sequence self-intersects
(which must happen since $V$ is finite). We will use the following definition.

\begin{definition}
\label{def:flattened}
If $G=(V,E)$ is a directed graph, its flattened graph, denoted $G^\flat=(V,E^\flat)$ is the  undirected graph obtained by forgetting the edge ordering,
namely
$$
\{s,t\}\in E^\flat \Leftrightarrow (s,t)\in E \text{ or } (t,s)\in E.
$$
\end{definition}

The following proposition relates forests and
undirected acyclic graphs.
\begin{proposition}
\label{prop:undir.ac}
If $G$ is a forest, then $G^\flat$ is an undirected acyclic graph.

Conversely, if $G$ is an undirected acyclic graph, there exists a forest $\tilde
G$ such that $\tilde G^\flat = G$.
\end{proposition}
\begin{proof}
Let $G=(V,E)$ be a forest and, in order to reach a contradiction, assume that $G^\flat$ has a loop, $s_0,
\ldots, s_{N-1}, s_N=s_0$. Assume that $(s_0, s_1) \in E$; then, also
$(s_1, s_2)\in E$ (otherwise $s_1$ would have two parents), and this
propagates to all $(s_k,
s_{k+1})$ for $k=0, \ldots, N-1$. But, since $s_N=s_0$, this provides
a loop in $G$ which is not possible. This proves thet $G^\flat$ has no
loop since the case $(s_1,s_0)\in E$ is treated similarly.

Now, let $G$ be an undirected acyclic graph. Fix a vertex $s\in V$ and
consider the following procedure, in which we recursively define sets $S_k$ of processed vertexes, and $\tilde E_k$ of oriented edges, $k\geq 0$, initialized with $S_0 =
\{s\}$ and $\tilde E_0 = \emp$.
\begin{enumerate}[label = --]
\item At step $k$ of the procedure,  assume that vertexes in $S_k$ have been processed and edges in $\tilde E_k$
have been oriented so that $(S_k, \tilde E_k)$ is a forest, and  that $\tilde
E_k^\flat$ is the set of edges $\{s,t\}\in E$ such that $s,t\in S_k$ (so,
oriented edges at step $k$ can only involve processed vertexes). 
\item If
$S_k = V$: stop, the proposition is proved.
\item Otherwise,
apply the  following construction. Let $F_k$ be the set of edges
in $E$ that contain exactly one element of $S_k$.
\begin{enumerate}
\item If $F_k = \emp$,  take any $s\in V\setm S_k$ as a new root
and let $S_{k+1} = S_k\cup \{s\}$, $\tilde E_{k+1}=\tilde E_k$.
\item Otherwise, add to $\tilde E_k$ the oriented edges $(s,t)$ such that
$s\in S_k$ and $\{s,t\}\in F_k$, yielding $\tilde E_{k+1}$, and add to $S_k$
the corresponding children ($t$'s) yielding $S_{k+1}$. 
\end{enumerate}
\end{enumerate}

We need to justify the fact that $\tilde G_{k+1} = (S_{k+1}, \tilde
E_{k+1})$ above is
still a forest. This is
obvious after Case (1), so consider Case (2). First $\tilde G_{k+1}$ is
acyclic, since any oriented loop is a fortiori an unoriented loop and
$G$ is acyclic. So we need to prove that no vertex in $S_{k+1}$ has
two parents. Since we did not add any parent to the vertexes in $S_k$
and, by assumption, $(S_k, \tilde E_k)$ is a forest, the only possibility for
a vertex to have two parents in $S_{k+1}$ is the existence of $t$ such
that there exists $s,s'\in S_k$ with $\{s,t\}$ and $\{s', t\}$ in
$E$. But, since $s$ and $s'$ have unaccounted edges containing them,
they cannot have been introduced in $S_k$ before the previously
introduced root has been added, so they are both connected to this
root: but the two connections to $t$ would create a loop in $G$ which
is impossible.

So the procedure carries on, and must end with $S_k=V$ at some point
since we keep adding points to $S_k$ at each step.
\end{proof}

Note that the previous proof shows there is more than one possible orientation of a connected
undirected tree into a tree is not unique, although uniquely specified
once a root is chosen. The proof is constructive, and provides an algorithm building a forest from an undirected acyclic graph.

We now define graphical models supported by trees, which constitute
our first Markov models associated with directed graphs. Define the
depth of a vertex in a tree $G = (V,E)$ to be the number of edges in the unique
path that links it to the root. We will denote by $G_d$ the set of
vertexes in $G$ that are at depth $d$, so that $G_0$
contains only the root, $G_1$ the children of the root and so on. Using
this, we have the definition: 
\begin{definition}
\label{def:tree.x}
Let $G = (V,E)$ be a tree. A process $X = (\pe Xs, s\in V)$ is
$G$-Markov if and only, for each $d\geq 1$, and for each $s\in G_d$,
we have
\begin{equation}
\label{eq:tree.cond}
\cind{\pe Xs}{(\pe X{G_d\setm\{s\}}, \pe X{G_q\setm\{\pa{s}\}}, q<d)}{\pe X{\pa{s}}}
\end{equation}
where $\pa{s}$ is the parent of $s$.
\end{definition}
So, conditional to its parent, $\pe Xs$ is independent from all other
variables at depth smaller or equal to the depth of $s$. 
 
Note that, from (CI3), \cref{def:tree.x} implies that, for all $s\in G_d$,
\[
\cind{\pe Xs}{\pe X{G_d\setm\{s\}}}{\pe X{G_q}, q<d},
\]
which, using 
\cref{prop:mut.indep}, implies that the variables $(\pe Xs, s\in G_d)$ are
mutually independent given $\pe X{G_q}, q<d$. This implies that, for
$d=1$ (letting $s_0$ denote the root in $G$):
$$
\myP(\pe X{G_1}=\pe x{G_1}, \pe X{s_0}=\pe x{s_0}) = \myP(\pe X{s_0}= \pe x{s_0})\prod_{s\in
G_1} \myP(\pe Xs = \pe xs\mid \pe X{s_0}= \pe x{s_0}).
$$
(If $\myP(\pe X{s_0}= \pe x{s_0})=0$, the choice for the conditional
probabilities can be made arbitrarily without changing the left-hand
side which vanishes.)
More generally, we have, letting $G_{<d} = G_0\cup
\cdots\cup G_{d-1}$,
$$
\myP(\pe X{G_{\leq d}} = \pe x{G_{\leq d}}) = \prod_{s\in G_d} \myP(\pe Xs= \pe xs \mid
\pe X{\pa{s}} = \pe x{\pa{s}}) \myP(\pe X{G_{< d}} = \pe x{G_{< d}})
$$
(with again an arbitrary choice for conditional probabilities that
are not defined) so that, we obtain, by induction, for $x\in \CF(V)$
\begin{equation}
\label{eq:tree.prob}
\myP(X = x) = \myP(\pe X{s_0} = \pe x{s_0}) \prod_{s\neq s_0} p_s(\pe x{\pa{s}}, \pe xs)
\end{equation}
where $p_s(\pe x{\pa{s}}, \pe xs) \defeq \myP(\pe Xs=\pe xs \mid
\pe X{\pa{s}} = \pe x{\pa{s}})$ are the tree transition probabilities between a
parent and a child. So we have the following proposition.
\begin{proposition}
\label{prop:tree.rep}
A process $X$ is Markov relative to a tree $G=(V,E)$ if and only
if there exists a probability distribution $p_0$ on $F_{s_0}$ and a family $(p_{st}, (s,t)\in E)$
such that $p_{st}$ is a transition probability from $F_s$ to $F_t$ and
\begin{equation}
\label{eq:tree.prob.2}
P_X(x) = p_0(x_{s_0})\prod_{(s,t)\in E} p_{st}(\pe xs, \pe xt), \ x\in \CF(V).
\end{equation}
\end{proposition}
We only have proved the ``only if'' part, but the ``if'' part is obvious from
\cref{eq:tree.prob.2}. Another property that becomes obvious with
this expression is the first part of the following proposition.
\begin{proposition}
\label{prop:tree.undir}
If a process $X$ is Markov relative to a tree $G=(V,E)$ then it is
 $G^\flat$ Markov.
Conversely, if $G=(V,E)$ is an undirected acyclic graph and $X$ is
$G$-Markov, then $X$ is Markov relative to any tree $\tilde G$ such
that $\tilde G^\flat = G$.  
\end{proposition}
\begin{proof}
To prove the converse part,  assume that $G=(V,E)$ is undirected acyclic and that $X$ is
$G$-Markov. Take $\tilde G$ such
that $\tilde G^\flat = G$. For $s\in V$ and its parent $\pa{s}$ in
$\tilde G$, the sets $\{s\}$ and  $\tilde G_{\leq
d}\setm \{s, \pa{s}\}$ are separated by $\pa{s}$ in $G$. To see this,  assume that there
exists a $t\in \tilde G_{\leq
d}\setm \{s, \pa{s}\}$ with a path from $t$ to $s$ that does not pass
through $\pa{s}$. Then we can complete this path with the path from $t$
to the first common ancestor (in $\tilde G$) of $t$ and $s$ and back
to $s$ to create a  path from
$s$ to $s$ that passes only once through $\{\pa{s}, s\}$  
and therefore contains a loop by  \cref{prop:loop}. 

The $G$-Markov property now implies 
$$
\cind{\pe Xs}{(\pe X{\tilde G_d\setm\{s\}}, \pe X{\tilde G_q\setm\{\pa{s}\}}, q<d)}{\pe X{\pa{s}}}
$$
which proves that $X$ is $\tilde G$-Markov.
\end{proof}

\begin{remark}
\label{rem:causality}
We see that there is no real gain in generality with passing from
undirected to directed graphs when working with trees. This is an
important remark, because directionality in graphs is often
interpreted as causality. For example, there is a natural causal order in the
statements
\begin{center}
(it rains) $\to$ (car windshields get wet) $\to$ (car wipers are on)
\end{center}
in the sense that each event can be seen as a logical precursor to the
next one. However, because one can pass from this directed chain to an
equivalent undirected chain and then back to a equivalent directed tree by choosing any of
the three variables as roots, there is no way to infer, from the
observation of the joint distribution of the three events (it rains, car windshields get wet, wipers are on), any causal relationship between them: the joint distribution
cannot resolve whether wipers are on because it rains, or whether
turning wipers on automatically wets windshields which in turn
triggers a shower~!

To infer causal relationships, one needs a different kind of
observation, that would modify the distribution of the system. Such an
operation (called an intervention), can be done, for example, by
preventing the windshields from being wet (doing, for example, the
observation in a parking garage), or forcing them to be wet (using a hose). Then, one can compare observations
made with these new conditions, and those made with the original
system, and check, for example, whether they modified the probability
that rain occurs outside. The answer (likely to be negative~!) would refute
any causal relationship from ``windshields are wet'' to ``it rains.'' On the
other hand, the intervention might modify how wipers are used,
which would indicate a possible causal relationship from ``windshields are wet'' to
``wipers are on.''
\end{remark}

\section{Examples of general ``loopy'' Markov random fields}

We will see that acyclic models have very nice computational
properties that make them attractive in designing
distributions. However, the absence of loops is a very restrictive
constraint, which is not realistic in many practical situations.
Feedback effects are often needed, for example. Most models in statistical physics are
supported by a lattice, in which natural translation/rotation
invariance relations forbid using any non-trivial acyclic model. As an
example, we now consider the 2D Ising model on a finite grid, which is a
model for (anti)-ferromagnetic interaction in a spin system.

Let $G = (V,E)$. A (positive) $G$-Markov model is said to have only pair interactions if and only if can be written in the form
$$
\pi(x) = \frac{1}{Z} \exp\Big(-\sum_{s\in G} h_s(\pe xs) -
\sum_{\defset{s,t}\in E} h_{\defset{s,t}}(\pe x{s,t})\Big).
$$
Relating to  \cref{th:ham.clif}, this says that $\pi$ is
associated to a potential involving cliques of order 2 at
most (note that this does not mean that the cliques of the associated
graph have order 2 at most; there can be higher-order cliques,
which would then have a
zero potential). The functions in the potential are indexed
by sets, as they should be from the general definition. However, models with pair interactions are often written in the form
$$
\pi(x) = \frac{1}{Z} \exp\Big(-\sum_{s\in G} h_s(\pe xs) -
\sum_{\defset{s,t}\in E} \tilde h_{st}(\pe x{s}, \pe xt)\Big)
$$
with $\tilde h_{st}(\la,\mu) = \tilde h_{ts}(\mu, \la)$ (which is
equivalent, taking $\tilde h = h/2$). 

The {\em Ising model} is a special case of models with pair interactions, for
which the state space, $F_s$, is equal to $\defset{-1,1}$ for all $s$
and 
$$
h_s(\pe xs) = \al_s \pe xs,\ \   h_{\{s,t\}}(\pe xs, \pe xt) = \be_{st} \pe xs \pe xt.
$$
In fact, for binary variables, this is the most general pair
interaction model. 

\begin{figure}[h]
\begin{center}
		\begin{tikzpicture}

\foreach \x in {0,1,2,..., 10}
\foreach \y in {0,1,2,..., 10}
\draw (\x, \y) circle (0.2cm) ;

\foreach \x in {0,1,2,..., 9}{
\foreach \y in {0,1,2,..., 10}{
\draw (\x+0.2, \y) -- (\x+0.8, \y) ;
}}

\foreach \x in {0,1,2,..., 10}{
\foreach \y in {0,1,2,..., 9}{
\draw (\x, \y+0.2) -- (\x, \y+0.8) ;
}}
\end{tikzpicture}
\caption{\label{fig:grid} Graph forming a two-dimensional regular
  grid.}
\end{center}
\end{figure}
The Ising model is moreover usually defined on a regular lattice,
which, in two dimensions, implies that $V$ is a finite
rectangle in $\mZ^2$, for example $V = \defset{-N,\ldots, N}^2$. The simplest
choice of a translation- and 90-degree rotation-invariant graph is the
nearest-neighbor graph for which $\defset{(i,j), (i',j')}\in E$ if and
only if $|i-i'| + |j-j'| =1$ (see \cref{fig:grid}). With this graph, one can furthermore
simplify the model to obtain the {\em isotropic Ising model} given by
$$
\pi(x) = \frac{1}{Z} \exp\Big( - \al \sum_{s\in V} \pe xs - \be
\sum_{s\sim t} \pe xs \pe xt\Big) .
$$
When $\be <0$, the model is {\em ferromagnetic}: each pair of
neighbors with identical signs brings a negative contribution to the
energy, making the configuration more likely (since lower energy
implies higher probability). 

The Potts model generalizes the Ising model  to finite, but non-necessarily binary, state spaces, say, $F_s
=F= \defset{1, \ldots, n}$. Define the function $\de(\la, \mu) = 1$ if
$\la=\mu$ and $(-1)$ otherwise. Then the Potts model is  given by
\begin{equation}
\label{eq:potts.model}
\pi(x) = \frac{1}{Z} \exp\Big( - \al \sum_{s\in V} h(\pe xs) - \be
\sum_{s\sim t} \de(\pe xs, \pe xt)\Big) 
\end{equation}
for some function $h$ defined on $F$.

\section{General state spaces}
\label{sec:mrf.pdf}

Our discussion of Markov random fields on graphs was done under the assumption of finite state spaces, which notably simplifies many of the arguments and avoids relying too much on measure theory. 
While this situation does cover a large range of application, there are cases in which one wants to consider variables taking values in continuous spaces, or in countable (infinite) spaces.

The results obtained for discrete variables can most of the time be extended to variables whose distribution has a p.d.f. with respect to a product of measures on the sets in which they take their values. For example, let $X, Y, Z$ takes values in $\CR_X, \CR_Y, \CR_Z$, equipped with $\sigma$-algebras $\boldsymbol{\CS}_X$, $\boldsymbol{\CS}_Y$, $\boldsymbol{\CS}_Z$ and measures $\mu_X$, $\mu_Y$, $\mu_Z$. Assume that $P_{X,Y,Z}$ is absolutely continuous with respect to $\mu_X\otimes \mu_Y\otimes \mu_Z$, with density $\phi_{XYZ}$. In such a situation, \cref{eq:cond.ind} remains valid, in that $X$ is conditionally independent of $Y$ given $Z$ if and only if
\begin{equation}
\label{eq:cond.ind.pdf}
\phi_{XYZ}(x,y,z) \phi_Z(z) = \phi_{XZ}(x,z) \phi_{YZ}(y,z)
\end{equation}
almost everywhere (relative to $\mu_X\otimes \mu_Y\otimes \mu_Z$). Here, $\phi_{XZ}, \phi_{YZ}, \phi_Z$ are marginal densities of the indexed random variables. The only difficulty in the argument, provided below for the interested reader, is dealing properly with sets of measure zero. 
\begin{proof}[Proof of \cref{eq:cond.ind.pdf}]
 Introduce the conditional densities
\[
\phi_{XY\mid Z}(x,y\mid z) = \frac{\phi_{XYZ}(x,y,z)}{\phi_Z(z)}
\]
and similarly $\phi_{X\mid Z}$ and $\phi_{Y\mid Z}$, which are defined when $z\not \in M_Z = \{z\in \CR_Z: \phi_Z(z) = 0\}$.  By definition of conditional independence, we have, for all $A \in \boldsymbol{\CS}_X$, $B\in \boldsymbol{\CS}_X$
\[
\int_{A\times B} \phi_{XY\mid Z}(x,y\mid z) \mu_X(dx)\mu_Y(dy) = 
\int_{A\times B} \phi_{X\mid Z}(x\mid z)\phi_{Y\mid Z}(y\mid z) \mu_X(dx)\mu_Y(dy) 
\]
for all $z\not \in M_Z$, which implies that, for all $z\not \in M_Z$, there exists a set $N_z\subset R_X\times R_Y$ such that $\mu_X\times \mu_Y(N_z) = 0$ and
\[
\phi_{XY\mid Z}(x,y\mid z) = \phi_{X\mid Z}(x\mid z)\phi_{Y\mid Z}(y\mid z)
\]
for all $z\not \in M_Z$ and $(x,y)\not \in N_z$. This immediately implies \cref{eq:cond.ind.pdf} for those $(x,y,z)$.  

If $z\in M_Z$, then 
\[
0 = \phi_Z(z) = \int_{\CR_X} \phi_{XZ}(x,z) \mu_X(dx) = \int_{\CR_Y} \phi_{YZ}(x,z) \mu_Y(dy)
\]
implying that $\phi_{XZ}(x,z) = \phi_{YZ}(y, z) = 0$  excepted on some set $N_z$ such that $\mu_{X}\otimes \mu_Y(N_z) = 0$, and \cref{eq:cond.ind.pdf} is therefore also true outside of this set. Now, letting $N = \{(x,y,z): (x,y) \in N_z\}$, we find that \cref{eq:cond.ind.pdf} is true for all $(x,y,z) \not \in N$ and
\[
\mu_X\otimes \mu_Y \otimes \mu_Z(N) = \int_{\CR_X\times \CR_Y\times \CR_Z} \bfone_{(x,y)\in N_z} \mu_X(dx)\mu_Y(dy)\mu_Z(dz) = \int_{\CR_Z}\mu_X\otimes \mu_Y(N_z)\mu_Z(dz) = 0.
\] 
(This argument involves Fubini's theorem \cite{rudin1966real}.)
\end{proof} 
With this definition, the proof of \cref{prop:cind} can be caried on without change, with the positivity condition expressing the fact that there exists $\tilde R_X \subset \CR_X$, $\tilde R_Y \subset \CR_Y$ and $\tilde R_Z \subset \CR_X$ such that $\phi_{XYZ}(x,y,z) > 0$ for all $x,y,z\in \tilde R_X \times \tilde R_Y \times \tilde R_Z$. (This proposition is actually valid in full generality, with a proper definition of positivity.)

When considering random fields with general state spaces, we will restrict to the similar situation in which each state space $F_s$ is equipped with a $\sigma$-algebra $\boldsymbol{\CS}_s$ and a measure $\mu_s$, and the joint distribution, $P_X$ of the random field $X  = (X_s, s\in V)$  is absolutely continuous with respect to $\mu\defeq \bigotimes_{s\in V} \mu_s$, denoting by $\pi$ the corresponding p.d.f. We will says that $\pi$ is positive if there exists $\tilde F = (\tilde F_s, s\in V)$ with measurable $\tilde F_s \subset F_s$ such that $\pi(x) > 0$ for all $x\in \CF(V, \tilde F)$. Without loss of generality unless one considers multiple random fields with different supports, we will assume that $\tilde F_s = F_s$ for all $s$.

The definition of consistent families of local interactions (\cref{def:set.int}) must be modified by adding the condition that
\begin{equation}
\label{eq:local.int.bounded}
\int_{\CF(V)} \prod_{C\in \CC}\phi_C(\pe x C) \mu(dx) < \infty.
\end{equation}
This requirement is obviously needed to ensure that the normalizing constant in \cref{eq:pi.phi} is finite. \Cref{prop:gc} is then true (with sums replaced by integrals in the proof) and so are \cref{prop:gc.cond,prop:gc.m}. Finally, the Hammersley-Clifford theorem (\cref{th:ham.clif.1}) extends to this context.

Even though it is a natural requirement, condition \eqref{eq:local.int.bounded} may be hard to assess with general families of local interactions. In the case of Gaussian distributions, however, one can provide relatively simple conditions. Assume that $F_s = \mR$ for all $s\in V$, and condider a potential $\Lambda = (\lambda_C, C\in \CC)$ with only univariate and bivariate interactions, such that, for some vector $a\in \mR^d$ (with $d = |V|$) and symmetric matrix $b \in \CS_d$,
\[
\left\{
\begin{aligned}
\lambda_{\{s\}}(\pe x {\{s\}}) &= -\pe as \pe xs + \frac12 b_{ss} (\pe xs)^2\\
\lambda_{\{s,t\}}(\pe x {\{s,t\}}) &=  b_{st} \pe xs\pe xt
\end{aligned}
\right.
\]
Then, considering $x\in \CF(V)$ as a $d$-dimensional vector, we have
\[
\pi(x) = \frac{1}{Z} \exp\left(a^T x - \frac12 x^T b x\right),
\]
with the integrability requirement that $b \succ 0$ (positive definite). The random field then follows a Gaussian distribution with mean $m = b^{-1} a$ and covariance matrix $\Sigma = b^{-1}$. The normalizing constant, $Z$, is given by
\[
Z = \frac{e^{-\frac12 a^Tba}(2\pi)^{d/2}}{\sqrt{\det{b}}}.
\]

This Markov random field parametrization of Gaussian distributions emphasizes the conditional  structure of the variables rather than their covariances. It is useful when the associated graph, represented by the matrix $b$ is sparse. In particular, the conditional distribution of $\pe Xs$ given the other variables is Gaussian, with mean $(\pe a s - \sum_{t\neq s} b_{st} \pe x t)/b_{ss}$ and variance $1/b_{ss}$.

\chapter[Probabilistic Inference for MRF]{Probabilistic Inference for  Random Fields}
\label{chap:inf.mrf}

Once the joint distribution of a family of variables has been modeled as a random field,  this model can be used to estimate the probabilities
 of specific events, or the expectations of random variables of interest. 
 For
example, if the modeled variables relate to  a medical condition, in
which variables such as diagnosis, age, gender, clinical evidence can
interact, one may want to compute, say, the probability  of someone having a disease given other observable
factors. Note that, being able to
compute expectations of the modeled variables for $G$-Markov processes also ensures that one can
compute conditional expectations of some modeled variables given others, since, by 
\cref{prop:cond.dist.grph}, conditional $G$-Markov distributions are
Markov over restricted graphs.

We assume  that $X$ is $G$-Markov for a graph $G = (V,E)$ and restrict (unless specified otherwise) to finite state spaces. We condider the basic problem to compute $\myP(\pe XS =\pe x S)$ when $S\sub V$, starting with one-vertex marginals, $\myP(\pe Xs= \pe xs)$. 

The Hammersley-Clifford theorem provides a generic form for general
positive $G$-Markov processes, in the form
\begin{equation}
\label{eq:gibbs.inference}
\myP(X=x) = \pi(x) =  \frac{1}{Z} \exp\Big(-\sum_{C\in \mathcal C_G}
h_C(\pe xC)\Big).
\end{equation}
So, formally, marginal distributions are given by the ratio 
\[
\myP(\pe XS = \pe xS) = \frac{\sum_{y\in \CF(V), \pe yS= \pe xS} \exp\Big(-\sum_{C\in \mathcal C_G}
h_C(\pe yC)\Big)}{\sum_{y\in \CF(V)} \exp\Big(-\sum_{C\in \mathcal C_G}
h_C(\pe yC)\Big)}.
\]
The problem is that the sums involved in this ratio involve a number
of terms
that grows exponentially with the size of $V$. Unless $V$ is very
small, a direct computation of these sums is intractable. An exception
to this is the case of acyclic graphs, as we will see in \cref{sec:inf.acyc}. But for
general, loopy, graphs, the sums can only be approximated, using, for example,
Monte-Carlo sampling, as described in the next section.

\section{Monte Carlo sampling}

Markov chain Monte Carlo methods are well adapted to sampling from Markov random fields, because conditional distributions used in Gibbs sampling, or, more generally, ratios of probabilities used in the Metropolis-Hastings algorithm do not in require the computation of the normalizing constant $Z$ in \cref{eq:gibbs.inference}. The simplest use of Gibbs sampling generalizes the Ising model example of \cref{sec:gibbs.ising}. Using the notation of \cref{alg:gibbs.sampling}, one lets $\CB'_s = \CF(s^c)$ (with the notation $s^c = V \setminus \{s\}$) and $U_s(x) = \pe x{s^c}$. The conditional distribution given $U_s$ is 
\[
Q_s(U_s(x), y) = \myP(\pe Xs = \pe ys\mid \pe X {s^c} = \pe x {s^c}) \bfone_{\pe y {s^c} = \pe x {s^c}}.
\]
The conditional probability in the r.h.s. of this equation takes the form 
\[
\pi_s(\pe ys\mid \pe x {s^c}) \defeq \myP(\pe Xs = \pe ys\mid \pe X {s^c} = \pe x {s^c}) = 
 \frac{1}{Z_s(\pe x {s^c})} \exp\left (- \sum_{C\in \CC, s\in \in C} h_C(\pe y s \wedge \pe x {C\cap s^c})\right)
\]
with
\[
Z_s(\pe x {s^c}) = \sum_{\pe z s\in F_s} \exp\left (- \sum_{C\in \CC, s\in \in C} h_C(\pe z s \wedge \pe x {C\cap s^c})\right).
\]

The Gibbs sampling algorithm samples from $Q_s$ by visiting all $s\in V$ infinitely often, as described in \cref{alg:gibbs.sampling}. Metropolis-Hastings schemes are implemented similarly, the most common choice using a local update scheme in \cref{alg:met.hast} such that $g(x, \cdot)$ only changes one coordinate, chosen at random, so that
\[
g(x, y) = \frac{1}{|V|}\sum_{s\in V} \bfone_{ \pe y{s^c} = \pe x{s^c}} g_s(\pe ys)
\]
where $g_s$ is some probability distribution on $F_s$. The acceptance probability $a(x,y)$ is equal to 1 when $y=x$. If $y \neq x$ and $g(x,y) >0$,  there is a unique $s$ for which $\pe y{s^c} = \pe x{s^c}$ and 
\[
a(x,y) = \min\left(1, \frac{\pi(y)g(y,x)}{\pi(x)g(x,y)}\right)
\]
with 
\[
\frac{\pi(y)g(y,x)}{\pi(x)g(x,y)} = \frac{\pi_s(\pe ys\mid \pe x{s^c})g_s(\pe xs)}{\pi_s(\pe xs\mid \pe x{s^c})g_s(\pe ys)}.
\]
Note that the latter equation avoids the computation of the local normalizing constant $Z_s(\pe x {s^c})$, which simplifies in the ratio.

Both algorithms have a transition probability $P$ that satisfies $P^m(x,y) >0$ for all $x,y\in \CF(V)$, with $m=|V|$ (for Metropolis-Hastings, one must assume that $g_s(\pe ys) > 0$ for all $\pe ys\in F_s$. This ensures that the chain is uniformly geometrically ergodic, i.e., \cref{eq:geom.ergod} is satisfied with a constant $M$ and some $\rho<1$. However, in many practical cases (especially for strongly structured distributions and large sets $V$), the convergence rate, $\rho$ can be very close to 1, resulting in a slow convergence. 

Acceleration strategies have been designed to address this issue, which is often due to the existence of multiple configurations that are local modes of the probability $\pi$. Such configurations are isolated from other high-probability configurations because local updating schemes need to make multiple low-probability changes to access them from the local mode. The following two approaches provide examples designed to address this issue.

\begin{enumerate}[label={\bf \alph*.}]
\item {\bf Cluster sampling.} 
To facilitate escaping from such local modes, it is sometimes possible to augment the state space by introducing a new configuration space, with variable denoted $\boldsymbol \xi$, and designing a joint distributions $\hat\pi(\xi, x)$ such that the marginal distribution on $\CF(V)$ (summing over $\xi$) is the targeted $\pi$. The additional variable can create high-probability bridges between local modes for $\pi$, and accelerate convergence.

To take an example, assume that all sets $F_s$ are identical (letting $F=F_s$, $s\in V$)  and that the auxiliary variable $\boldsymbol \xi$ takes values in the set of functions from $E$ to $\{0,1\}$, that we will denote $\CB(E)$, i.e., that it takes the form $(\pe {\bfxi}{st}, \{s,t\}\in E)$, with $\pe \bfxi {st} \in \{0,1\}$. For $x\in \CF(V)$, introduce the set $\CB_x$ containing all $\xi \in \CB(E)$, such that  for all $\{s,t\}\in E$, 
\[
\pe xs \neq \pe xt \Rightarrow \pe \xi {st} = 1.
\]
Assume that the conditional distribution of $\boldsymbol\xi$ given $x$ is supported by $\CB_x$, such that, for $\xi\in \CB_x$
\[
\myP(\boldsymbol \xi = \xi \mid X = x) = \hat\pi(\xi\mid x) = \frac1{\zeta(x)} \exp\left(- \sum_{\{s,t\}\in E} \mu_{st} \pe \xi {st}\right) .
\]
The coefficients $\mu_{st}$ are free to choose (and one possible choice is to take $\mu_{st} = 0$ for all $\{s,t\}\in E$). For this distribution, all $\pe {\boldsymbol\xi} {st}$ are independent conditionally to $X=x$, with $\pe{\boldsymbol\xi}{st} = 1$ with probability 1 if $\pe xs\neq \pe xt$, and
\begin{equation}
\label{eq:sw.update}
P(\pe {\boldsymbol \xi}{st} = 1\mid X=x) = \frac{e^{-\mu_{st}}}{1 + e^{-\mu_{st}}}
\end{equation}
if $\pe x s = \pe x t$.
This conditional distribution is, as a consequence, very easy to sample from. Moreover, the normalizing constant $\zeta(x)$ has closed form and is given by
\[
\zeta(x) = \prod_{\{s,t\}\in E}\left(\bfone_{\pe xs = \pe xt} + e^{-\mu_{st}}\right) = \exp\left(\sum_{\{s,t\}\in E} \log\left(1+e^{-\mu_{st}}\right) + \sum_{\{s,t\}\in E} \log(1 +e^{\mu_{st}}) \bfone_{\pe xs\neq\pe xt}\right).
\]

Now consider the conditional probability that $X = x$ given $\boldsymbol\xi = \xi$. For this distribution, one has, with probability 1, $\pe Xs = \pe Xt$ when $\pe{\xi}{st} = 0$. This implies that $X$ is constant on the connected components of the subgraph $(V, E_\xi)$ of $(V,E)$, where $\{s,t\} \in E_\xi$ if and only if $\pe \xi {st} = 0$. Let $V_1, \ldots, V_m$ denote these connected components (these components and their number depend on $\xi$). The conditional distribution of $X$ given $\xi$ is therefore supported by the configurations such that there exists $c_1, \ldots, c_m\in F$ such that $\pe x s= c_j$ if and only if $s\in V_j$, that we will denote, with some abuse of notation: $\pe{c_1}{V_1} \wedge \cdots \wedge \pe{c_m}{V_m}$.

Given this remark, the conditional distribution of $X$ given $\boldsymbol \xi = \xi$ is equivalent to a distribution on $F^m$, which may be feasible to sample from directly if $|F|$ and $m$ are not too large. To sample from $\pi$, one now needs to alternate between sampling $\boldsymbol\xi$ given $X$ and the converse, yielding the following first version of cluster-based sampling.

\begin{algorithm}[Cluster-based sampling: Version 1]
\label{alg:sw.1}
This algorithm samples from \cref{eq:gibbs.inference}.
\begin{enumerate}
\item Initialize the algorithm some configuration $x \in \CF(V)$.
\item Loop over the following steps:
\begin{enumerate}[label=\alph*.,left=0.5cm]
\item Generate a configuration $\xi \in \CB_x$ such that $\pe \xi {st} = 1$ with  probability given by \cref{eq:sw.update} when $\pe x s = \pe x t$.
\item Determine the connected components, $V_1, \ldots, V_m$, of the graph $G_\xi = (V, E_\xi)$ with edges given by pairs $\{s,t\}$ such that $\pe \xi{st} = 1$.
\item Sample values $c_1, \ldots, c_m\in F$ according to the distribution
\[
q(c_1, \ldots, c_m) \propto \frac{\pi(\pe{c_1}{V_1} \wedge \cdots \wedge \pe{c_m}{V_m})}{\zeta(\pe{c_1}{V_1} \wedge \cdots \wedge \pe{c_m}{V_m})}.
\] 
\item Set $x = \pe{c_1}{V_1} \wedge \cdots \wedge \pe{c_m}{V_m}$.
\end{enumerate}
\end{enumerate}
\end{algorithm}
Step (2.c) takes a simple form in the special case when $\pi$ is a non-homogeneous Potts model (\cref{eq:potts.model}) with positive interactions, that we will write as
\[
\pi(x) = \exp\left(- \sum_{s\in V} \alpha_s \pe xs - \sum_{\{s,t\}\in E} \beta_{st} \bfone_{\pe xs\neq \pe xt}\right)
\]
with $\beta_{st} \geq 0$.
Then
\[
\frac{\pi(x)}{\zeta(x)} \propto \exp\left(- \sum_{s\in V} \alpha_s \pe xs - \sum_{\{s,t\}\in E} (\beta_{st} - \beta'_{st})\bfone_{x_s\neq s_t}\right)
\]
with $\beta'_{st} = \log(1 +e^{\mu_{st}})$. If one chooses $\mu_{st}$ such that $\beta'_{st} = \beta_{st}$ (which is possible since $\beta_{st}\geq 0$), then the interaction term disappears and the probability $q$ in (2.c) is proportional to 
\[
\prod_{j=1}^m \exp\left(-\sum_{s\in V_j} \alpha_s\right)
\]
so that $c_1, \ldots, c_m$ can be generated independently. The resulting algorithm is the Swendsen-Wang sampling algorithm for the Potts model \citep{swendsen1987nonuniversal}. The presentation given here adapts the one introduced in \citet{barbu2005generalizing}.

For more general models, step (2.c) can be computationally costly, especially if the number of connected components is large. In this case, this step can be replaced by a Gibbs sampling step for one of the $c_j's$ conditional to the others (and $\xi$) that we summarize in the following variation of \cref{alg:sw.1}.
\begin{algorithm}[Cluster-based sampling: Version 2]
\label{alg:sw.2}
This algorithm samples from \cref{eq:gibbs.inference}.
\begin{enumerate}
\item Initialize the algorithm some configuration $x \in \CF(V)$.
\item Loop over the following steps:
\begin{enumerate}[label=\alph*.,left=0.5cm]
\item Generate a configuration $\xi \in \CB_x$ such that $\pe \xi {st} = 1$ with  probability given by \cref{eq:sw.update} when $\pe x s = \pe x t$.
\item Determine the connected components, $V_1, \ldots, V_m$, of the graph $G_\xi = (V, E_\xi)$ with edges given by pairs $\{s,t\}$ such that $\pe \xi{st} = 1$. Note that $x$ is constant on each of these connected components, i.e., there exists $c_1, \ldots, c_m\in F$ such that $x = \pe{c_1}{V_1} \wedge \cdots \wedge \pe{c_m}{V_m}$.  
\item Select at random one of the components, say, $j_0 \in \{1, \ldots, m\}$.
\item Sample the value $\tilde c_{j_0} \in F$ according to the distribution
\[
q(\tilde c_{j_0}) \propto \frac{\pi(\pe{\tilde c_1}{V_1} \wedge \cdots \wedge \pe{\tilde c_m}{V_m})}{\zeta(\pe{\tilde c_1}{V_1} \wedge \cdots \wedge \pe{\tilde c_m}{V_m})}.
\] 
with $\tilde c_j = c_j$ if $j\neq j_0$.
\item Set $\pe x s = \tilde c_{j_0}$ for $s\in V_{j_0}$.
\end{enumerate}
\end{enumerate}
\end{algorithm}
Unlike single-variable updating schemes, these algorithms can update large chunks of the configurations at each step, and may result in significantly faster convergence of the sampling procedure. Note that step (2.d) in \cref{alg:sw.2} can be replaced by a Metropolis-Hastings update with a proper choice of proposal probability \cite{barbu2005generalizing}.

\item {\bf Parallel tempering.}
We now consider a different kind of  extension in which we allow $\pi$ depends continuously on a parameter $\beta>0$, writing $\pi_\beta$ and, the goal is to sample from $\pi_1$. For example,  one can extend \cref{eq:gibbs.inference} by the  family of probability distributions
\[
\pi_\beta(x) =  \frac{1}{Z_\beta} \exp\left(-\beta \sum_{C\in \mathcal C_G}
h_C(\pe xC)\right)
\]
for $\beta\geq 0$. For small $\beta$, $\pi_\beta$ gets close to the uniform distribution on $\CF(V)$ (achieved for $\beta=0$), so that it becomes easier to  move from local mode to local mode. This implies that sampling with small $\beta$ is more efficient and the associated Markov chain moves more rapidly in the configuration space.

Assume given, for all $\beta$,  two ergodic transition probabilities on $\CF(V)$, $q_\beta$ and $\tilde q_\beta$ such that \cref{eq:reversed.discrete} is satisfied with $\pi_\beta$ as invariant probability, namely
\begin{equation}
\label{eq:tempered.neal.1}
\pi_\beta(y) q_\beta(y,x) = \pi_\beta(x) \tilde q_\beta(x,y)
\end{equation}
for all $x,y \in \CF(V)$ (as seen in \cref{eq:reversed.discrete}, $\tilde q_\beta$ is the transition probability for the reversed chain).   The basic idea is that $q_\beta$ provides a Markov chain that converges rapidly for small $\beta$ and slowly when $\beta$ is closer to 1. Parallel tempering (this algorithm was introduced in \citet{neal1996sampling} based on ideas developed in \citet{marinari1992simulated}) leverages this fact (and the continuity of $\pi_\beta$ in $\beta$) to accelerate the simulation of $\pi_1$ by introducing intermediate steps sampling at low $\beta$ values.

The algorithm specifies a sequence of parameters $0\leq \beta_1\leq \cdots \leq \beta_m = 1$. One simulation steps goes down, then up this scale, as described in the following algorithm.
\begin{algorithm}[Parallel Tempering]
Start with an initial configuration $x_0\in \CF(V)$. This configuration is then updated  at each step, using the following sequence of operations.
\begin{enumerate}[label=(\arabic*)]
\item For $j=1, \ldots, m$, generate a configuration $x_j$ according to $\tilde q_{\beta_{j}}(x_{j-1}, \cdot)$.
\item Generate a configuration $z_{m-1}$ according to $q_{\beta_m}(x_m, \cdot)$. 
\item For $j = m-1, \ldots, 1$, generate a configuration $z_{j-1}$ according to $q_{\beta_j}(z_j, \cdot)$.
\item Set $x_0 = z_0$ with probability
\[
\min\left(1, \frac{\pi_{\beta_0}(z_0)}{\pi_{\beta_0}(x_0)}
\left(\prod_{j=1}^{m-1}  \frac{\pi_{\beta_j}(x_{j-1})}{\pi_{\beta_{j}}(x_{j})}\right) \frac{\pi_{\beta_{m}}(x_{m-1})}{\pi_{\beta_{m}}(z_{m-1})}
\left(
\prod_{j=1}^{m-1}  \frac{\pi_{\beta_{j}}(z_{j})}{\pi_{\beta_{j}}(z_{j-1})}
\right)\right).
\]
(Otherwise, keep $x_0$ unchanged).
\end{enumerate}
\end{algorithm}

Importantly, the acceptance probability at step (4) only involves ratios of $\pi_\beta's$ and therefore no normalizing constant.
We now show that this algorithm is $\pi_{\beta_0}$-reversible. Let $p(\cdot, \cdot)$ denote the transition probability of the chain. If $z_0\neq x_0$,  $p(x_0, z_0)$ corresponds to steps (1) to (3), with acceptance at step(4), and is therefore given by the sum, over all  $x_1, \ldots, x_m$ and $z_1, \ldots, z_m$, of products 
\begin{multline*}
\tilde q_{\beta_{1}}(x_{0}, x_1)\cdots \tilde q_{\beta_{m}}(x_{m-1}, x_m) q_{\beta_{m}}(x_{m}, z_{m-1}) \cdots q_{\beta_{1}}(z_{1}, z_0) \\
\min\left(1, \frac{\pi_{\beta_0}(z_0)}{\pi_{\beta_0}(x_0)}
\left(\prod_{j=1}^{m-1}  \frac{\pi_{\beta_j}(x_{j-1})}{\pi_{\beta_{j}}(x_{j})}\right) \frac{\pi_{\beta_{m}}(x_{m-1})}{\pi_{\beta_{m}}(z_{m-1})}
\left(
\prod_{j=1}^{m-1}  \frac{\pi_{\beta_{j}}(z_{j})}{\pi_{\beta_{j}}(z_{j-1})}
\right)\right)
\end{multline*}
Applying \cref{eq:tempered.neal.1}, this is equal to
\begin{multline*}
\min\Big(
\tilde q_{\beta_{1}}(x_{0}, x_1)\cdots \tilde q_{\beta_{m}}(x_{m-1}, x_m) q_{\beta_{m}}(x_{m}, z_{m-1}) \cdots q_{\beta_{1}}(z_{1}, z_0),\\ \frac{\pi_{\beta_0}(z_0)}{\pi_{\beta_0}(x_0)}
q_{\beta_{1}}(x_{0}, x_1)\cdots  q_{\beta_{m}}(x_{m-1}, x_m) \tilde q_{\beta_{m}}(x_{m}, z_{n-1}) \cdots \tilde q_{\beta_{1}}(z_{1}, z_0) \Big)
\end{multline*}
So,
\begin{multline*}
\pi_{\beta_0}(x_0) p(x_0, z_0) = \sum \min\Big(\pi_{\beta_0}(x_0)
\tilde q_{\beta_{1}}(x_{0}, x_1)\cdots \tilde q_{\beta_{m}}(x_{m-1}, x_m) q_{\beta_{m}}(x_{m}, z_{m-1}) \cdots q_{\beta_{1}}(z_{1}, z_0),\\ \pi_{\beta_0}(z_0)
q_{\beta_{1}}(x_{1}, x_0)\cdots  q_{\beta_{m}}(x_{m}, x_{m-1}) \tilde q_{\beta_{m}}(z_{m-1}, x_m) \cdots \tilde q_{\beta_{1}}(z_{0}, z_1) \Big)
\end{multline*}
where the sum is over all $x_1, \ldots, x_m, z_1, \ldots, z_{m-1}\in \CF(V)$. The sum is, of course, unchanged if one renames $x_1, \ldots, x_m, z_1, \ldots, z_{m-1}$ to $z_1, \ldots, z_m, x_1, \ldots, x_{m-1}$, but doing so provides the expression of  $\pi_{\beta_0}(z_0) p(z_0, x_0)$, proving the reversibility of the chain with respect to $\pi_{\beta_0}$.

\end{enumerate}

\section{Inference with acyclic graphs}
\label{sec:inf.acyc}

We now switch to deterministic methods to compute, or approximate, marginal probabilities of Markov random fields. In this section, we consider a directed acyclic graph $G = (V,E)$.
As we have seen, Markov processes for acyclic graphs are also Markov
for any tree structure associated with the graph. Introducing such a
tree, $\tilde G = (V, \tilde E)$ with $\tilde G^\flat = G$, we know
that a Markov process on $G$ can be written in the form (letting $s_0$
denote the root in $\tilde G$):
\begin{equation}
\label{eq:tree.assume}
\pi(x) = p_{s_0}(\pe x {s_0}) \prod_{(s,t)\in \tilde E} p_{st}(\pe xs, \pe xt)
\end{equation}
where $p_{s_0}$ is a probability and $p_{st}$ a transition
probability.

We now show how to compute marginal probabilities of configurations
$\pe xS$, denoted $\pi_S(\pe xS)$,  for a set $S\sub V$, starting with singletons $S = \{s\}$. The computation can be done by propagating down the tree as
follows. For $s=s_0$, the probability is known, with
$\pi_{s_0} = p_{s_0}$. Now take an arbitrary $s\neq s_0$ and let $\pa{s}$
be its parent. Then
\begin{multline*}
\pi_s(\pe xs) = \myP(\pe Xs = \pe xs) = \sum_{\pe y{\pa{s}}\in F_{\pa{s}}} P(\pe Xs =
\pe xs \mid \pe X{\pa{s}} = \pe y{\pa{s}}) P(\pe x{\pa{s}} = \pe y{\pa{s}}) \\
=
\sum_{\pe y{\pa{s}}\in F_{\pa{s}}} \pi_{\pa{s}}(\pe y{\pa{s}}) p_{\pa{s}}(y_{\pa{s}}, \pe xs)
  \end{multline*}
so that the marginal probability at any $s\neq s_0$ can be computed
given the marginal probability of its parent. We can propagate the
computation down the tree, with a total cost for computing $\pi_s$
proportional to $\sum_{k=1}^n |F_{t_{k-1}}|\,|F_{t_k}|$ where $t_0=s_0, t_1,
\ldots, t_n=s$ is the unique path between $s_0$ and $s$. This is
linear in the depth of the tree, and quadratic (not exponential) in
the sizes of the state spaces.
The
computation of all singleton marginals requires 
an order of $\sum_{(s,t)\in E} |F_s|\,|F_t|$ operations. 

Now, assume that probabilities of singletons have been computed and consider an arbitrary set $S \subset V$. Let $s\in V$ be  an ancestor of every vertex in $S$, maximal in the sense that none of its children also satisfy this property. Consider the subtrees
of $\tilde G$ starting from each of the children of $s$, denoted
$\tilde G_1,
\ldots, \tilde G_n$ with $\tilde G_k = (V_k, \tilde E_k)$. Let $S_k = S\cap V_k$. From the
conditional independence,
\begin{eqnarray*}
\pi_S(\pe xS) &=& \sum_{\pe ys\in F_s} P(\pe X{S\setm \{s\}} =
\pe x{S\setm_\{s\}}\mid \pe Xs = \pe ys) \pi_s(\pe ys)\\
&=& \sum_{\pe ys\in F_s} \prod_{k=1, S_k\neq \emp}^n P(\pe X{S_k}=\pe x{S_k}\mid \pe Xs = \pe ys)
\pi_s(y_s)
\end{eqnarray*}
Now, for all $k=1, \ldots, n$, we have $|S_k| < |S|$: this is obvious
if $S$ is not completely included in one of the $V_k$'s. But if $S\sub
V_k$ then the root, $s_k$, of $V_k$ is an ancestor of all the elements
in $S$ and is a child of $s$, which contradicts the assumption that
$s$ is maximal. So we have reduced the computation of $\pi_S(x_S)$ to
the computations of $n$ probabilities of smaller sets, namely
$P(\pe X{S_k}= \pe x{S_k}\mid \pe Xs = \pe ys)$ for $S_k\neq \emp$. Because the
distribution of $\pe X{V_k}$ conditioned at $s$ is a $\tilde G_k$-Markov model, we
can reiterate the procedure until only sets of cardinality one remain,
for which we know how to explicitly compute probabilities.

\bigskip
This provides a  feasible algorithm to compute
marginal probabilities with trees, at least when its distribution is given 
in tree-form, like in \cref{eq:tree.assume}. We now address the situation in which one starts with  a probability distribution associated with pair interactions (cf. 
\cref{def:set.int}) over the acyclic graph $G$
\begin{equation}
\label{eq:pi.tree}
\pi(x) = \frac{1}{Z} \prod_{s\in V} \phi_s(\pe xs) \prod_{\defset{s,t}\in E} \phi_{st}(\pe xs, \pe xt).
\end{equation}
We assume these local interactions to be consistent, still
allowing for some vanishing $\phi_{st}(\pe xs, \pe xt)$.

Putting $\pi$ in the form \cref{eq:tree.assume} is equivalent to computing  all joint probability distributions
$\pi_{st}(\pe xs, \pe xt)$ for $\defset{s,t}\in E$, and we now describe this computation. Denote
$$
U(x)  = \prod_{s\in V} \phi_s(\pe xs) \prod_{\defset{s,t}\in E}
\phi_{st}(\pe xs, \pe xt)
$$
so that $Z=\sum_{y\in \CF(V)} U(y)$. For the tree $\tilde G = (V, \tilde E)$, and $t\in V$,
we let $\tilde G_t = (V_t, \tilde E_t)$ be the subtree of $G$ rooted at $t$ (containing $t$ and all its descendants). For
$S\sub V$,  define 
$$
U_S(\pe xS) = \prod_{s\in S} \phi_s(\pe xs) \prod_{\defset{s,s'}\in E,
s,s'\in D}
\phi_{ss'}(\pe xs, \pe x{s'})
$$
and
$$
Z_t(\pe xt) = \sum_{\pe y{V^*_t}\in \CF(V^*_t)} U_{V_t}(\pe xt\wedge \pe y{V^*_t}).
$$
with $V_t^* = V_t\setm \{t\}$.

\begin{lemma}
\label{lem:acyc.tree}
Let $G = (V, E)$ be a directed acyclic graph and $\pi = P^X$ be the $G$-Markov
distribution given by \cref{eq:pi.tree}. With the notation above, we
have
\begin{equation}
\label{eq:acyc.tree.1}
\pi_{s_0}(\pe x{s_0}) =\frac{Z_{s_0}(\pe x{s_0})}{\sum_{\pe y{s_0}\in F_{s_0} }
Z_{s_0} (\pe y{s_0})}
\end{equation}
and, for $(s,t)\in \tilde E$,
\begin{equation}
\label{eq:acyc.tree.2}
p_{st}(\pe xs, \pe xt) = P(\pe Xt=\pe xt\mid \pe Xs=\pe xs)  = \frac{\phi_{st}(\pe xs, \pe xt)
Z_t(\pe xt)}{\sum_{\pe yt\in F_t} \phi_{st}(\pe xs, \pe yt) Z_t(\pe yt)}
\end{equation}
\end{lemma}
\begin{proof}
Let  $W_t = V\setm V_t$. 
Clearly, $Z = \sum_{\pe x0\in F_{s_0}} Z_{s_0} (\pe x0)$
and $\pi_{s_0}(\pe x0) = Z_{s_0}(\pe x0)/Z$ which gives \cref{eq:acyc.tree.1}. Moreover, if $s\in V$, we have
$$
\myP(\pe X{V_s^*} = \pe x{V_s^*}\mid \pe Xs=\pe xs) = \frac{\sum_{\pe y{W_s}} U(\pe x{V_s}\wedge
\pe y{W_s})}{\sum_{\pe y{V_s^*}, \pe y{W_s}} U(\pe xs\wedge \pe y{V_s^*} \wedge
\pe y{W_s})}.
$$
We can write
$$
U(\pe xs\wedge \pe y{V_s^*} \wedge
\pe y{W_s}) = U_{V_s}(\pe xs\wedge \pe y{V_s^*}) U_{\{s\}\cup W_s}(\pe xs\wedge
\pe y{W_s}) \phi_s(\pe xs)^{-1}
$$
yielding the simplified expression
\begin{align*}
\myP(\pe X{V_s^*} = \pe x{V_s^*}\mid \pe Xs=\pe xs) &=
 \frac{U_{V_s}(\pe x{V_s})
\phi_s(\pe xs)^{-1} \sum_{y_{W_s}} U_{\{s\}\cup W_s}(\pe xs\wedge
\pe y{W_s})}{\phi_s(\pe xs)^{-1} \big(\sum_{\pe y{V_s^*}} U_{V_s}(\pe xs\wedge
\pe y{V_s^*})\big)\big(\sum_{\pe y{W_s}} U_{\{s\}\cup W_s}(\pe xs\wedge
\pe y{W_s})\big)
}\\
&= \frac{U_{V_s}(\pe x{V_s})
}{Z_s(\pe xs) }
\end{align*}
Now, if $t_1, \ldots, t_n$ are the children of $s$, we have
$$
U_{V_s}(\pe x{V_s}) = \phi_s(\pe xs) \prod_{k=1}^n \phi_{st_k}(\pe xs,\pe x{t_k}) \prod_{k=1}^n U_{V_{t_k}}(\pe x{V_{t_k}}),
$$
so that
\begin{align*}
&\myP(\pe X{t_k} = \pe x{t_k}, k=1, \ldots, n \mid \pe Xs = \pe xs) 
\\ &=
\frac{1}{Z_s(\pe xs)}\sum_{\pe y{V_{t_k}^*}, k=1, \ldots, n} \phi_s(\pe xs)
\prod_{k=1}^n \phi_{st_k}(\pe xs, \pe x{t_k}) \prod_{k=1}^n
U_{V_{t_k}}(\pe x{t_k} \wedge \pe y{V^*_{t_k}})\\
&= \frac{\phi_s(\pe xs) \prod_{k=1}^n \phi_{st_k}(\pe xs,\pe x{t_k})
\prod_{k=1}^n Z_{t_k}(\pe x{t_k})}{Z_s(\pe xs)}
\end{align*}
This implies that  the transition
probability needed for the tree model,
$p_{st_1}(\pe xs, \pe x{t_1})$, must be
proportional to  
$\phi_{st_1}(\pe xs, \pe x{t_1}) Z_{t_1}(\pe x{t_1})$ which proves the lemma.
\end{proof}

This lemma reduces the computation of the transition probabilities
to the computation of $Z_s(\pe xs)$, for $s\in V$. This can be done
efficiently, going upward in the tree (from terminal vertexes to the
root). Indeed, if $s$ is terminal, then $V_s  = \{s\}$ and 
$Z_s(\pe xs) = \phi_s(\pe xs)$. Now, if $s$ is non-terminal and $t_1,
\ldots, t_n$ are its children, then, it is easy to see that
\begin{align}
\nonumber
Z_s(\pe xs) &= \phi_s(\pe xs) \sum_{\pe x{t_1}\in F_{t_1}, \ldots, \pe x{t_n}\in
F_{t_n}} \prod_{k=1}^n \phi_{st_k}(\pe xs, \pe x{t_k}) Z_{t_k}(\pe x{t_k}) \\
&= 
\phi_s(\pe xs) \prod_{k=1}^n \Big(\sum_{\pe x{t_k}\in F_{t_k}} \phi_{st_k}(\pe xs, \pe x{t_k}) Z_{t_k}(\pe x{t_k})\Big)
\label{eq:z.induc}
\end{align}
So, $Z_s(\pe xs)$ can be easily computed once the $Z_t(\pe xt)$'s are known
for the children of $s$.

 \Cref{eq:acyc.tree.1,eq:acyc.tree.2,eq:z.induc} therefore provide the necessary relations in order
to compute the singleton and edge marginal probabilities on the
tree. It is important to note that these relations are valid for any
tree structure consistent with the acyclic graph we started with. We
now rephrase them with notation that only depend on this graph and not
on the selected orientation.

Let $\defset{s,t}$ be an edge in $E$. Then $s$ separates the
graph $G\setm\{s\}$ into two components. Let $V_{st}$ be the component
that contains $t$, and  $V_{st}^* = V_{st}\setm t$. Define
$$
Z_{st}(x_t) = \sum_{\pe y{V^*_{st}}\in \CF(V^*_{st})} U_{V_{st}}(\pe xt\wedge \pe y{V^*_{st}}).
$$
This $Z_{st}$ coincides with the previously introduced $Z_t$, computed with any tree
in which the edge $\{s,t\}$ is oriented from $s$ to $t$. 
\Cref{eq:z.induc} can be rewritten with this new notation in the form:
\begin{equation}
\label{eq:z.induc.2}
Z_{st}(\pe xt) =
\phi_t(\pe xt) \prod_{t'\in \CV_t\setm \{s\}} \Big(\sum_{\pe x{t'}\in F_{t'}} \phi_{tt'}(\pe xt, \pe x{t'}) Z_{tt'}(\pe x{t'})\Big).
\end{equation}
This equation is usually written in terms of ``messages'' defined
by
$$
m_{ts}(\pe xs) = \sum_{\pe x{t}\in F_{t}} \phi_{st}(\pe xs, \pe x{t}) Z_{st}(\pe x{t})
$$
which yields
$$
Z_{st}(\pe xt) = \phi_t(\pe xt) \prod_{t'\in\CV_t\setm \{s\}} m_{t't}(\pe xt)
$$
and the message consistency relation 
\begin{equation}
\label{eq:bp.mes}
m_{ts}(\pe xs) = \sum_{\pe x{t}\in F_{t}} \phi_{st}(\pe xs, \pe x{t}) \phi_t(\pe xt) \prod_{t'\in\CV_t\setm \{s\}} m_{t't}(\pe xt).
\end{equation}

Also, because one can start building a tree from $G^\flat$ using 
any vertex as a root,  \cref{eq:acyc.tree.1} is valid
for any $s\in V$, in the form (applying \cref{eq:z.induc} to the root)
\begin{equation}
\label{eq:bp.bel}
\pi_{s}(\pe x{s}) =\frac{1}{\zeta_s}  \phi_s(\pe xs) \prod_{t\in \CV_s} m_{ts}(\pe xs)
\end{equation}
where $\zeta_s$ is chosen to ensure that the sum of probabilities is
1. (In fact, looking at  \cref{lem:acyc.tree}, we have $Z_s=Z$,
independent of $s$.)

Similarly,  \cref{eq:acyc.tree.2} can be written
\begin{equation}
\label{eq:bp.trans}
p_{st}(\pe xs, \pe xt)  = m_{ts}(\pe xs)^{-1}  \phi_{st}(\pe xs, \pe xt)
\phi_t(\pe xt) \prod_{t'\in\CV_t\setm \{s\}} m_{t't}(\pe xt)
\end{equation}
which provides the edge transition probabilities. Combining this with
\cref{eq:bp.bel}, we get the edge marginal probabilities:
\begin{equation}
\label{eq:bp.edge}
\pi_{st}(\pe xs, \pe xt) 
=  \frac{1}{\zeta} \phi_{st}(\pe xs, \pe xt) \phi_s(\pe xs)
\phi_t(\pe xt) \prod_{t'\in\CV_t\setm \{s\}} m_{t't}(\pe xt) \prod_{s'\in\CV_s\setm \{t\}} m_{s's}(\pe xs).
\end{equation}

\begin{remark}
\label{rem:bp.normalize}
We can modify \cref{eq:bp.mes} by multiplying the right-hand side by
an arbitrary constant $q_{ts}$ without changing the resulting
estimation of probabilities: this only multiplies the messages by a
constant, which cancels after normalization. This remark can be useful
in particular to avoid numerical overflow; one can,
for example, define $q_{ts} = 1/\sum_{x_s\in F_s} m_{ts}(x_s)$ so
that the messages always sum to 1. This is also useful when applying
belief propagation (see next section) to loopy networks, for which \cref{eq:bp.mes} may
diverge while the normalized version converges. 
\end{remark}

The following summarizes this message passing algorithm.

\begin{algorithm}[Belief propagation on acyclic graphs]
\label{alg:message.tree}
Given a family of interactions $\phi_s: F_s \to [0, +\infty)$, $\phi_{st}: F_s \times F_t \to [0, +\infty)$,
\begin{enumerate}
\item Initialize functions (messages) $m_{ts}: F_s \to \mathbb R$, e.g., taking $m_{ts}(\pe x s) = 1/|F_s|$.
\item Compute unnormalized messages 
\[
\tilde m_{ts}(\cdot) = \sum_{\pe x{t}\in F_{t}} \phi_{st}(\cdot, \pe x{t}) \phi_t(\pe xt) \prod_{t'\in\CV_t\setm \{s\}} m_{t't}(\pe xt)
\]
and  let $m_{ts}(\cdot) = q_{ts} \tilde m_{ts}(\cdot)$, 
for some choice of constant $q_{ts}$, which must be a fixed function of $\tilde m_{ts}(\cdot)$, such as
\[
q_{ts} = \left(\sum_{\pe xs\in F_s} \tilde m_{ts}(\pe xs)\right)^{-1}.
\]
\item Stop the algorithm when the messages  stabilize (which happens after a finite number of updates). Compute the edge marginal distributions using \cref{eq:bp.edge}. 
\end{enumerate}
\end{algorithm}
It should be clear, from the previous analysis that messages stabilize in finite time, starting from the outskirts of the acyclic graph. Indeed,  messages starting from a terminal $t$ (a
vertex with only one neighbor) are
automatically set to their correct value in \cref{eq:bp.mes},
$$
m_{ts}(x_s) = \sum_{x_{t}\in F_{t}} \phi_{st}(x_s, x_{t}) \phi_t(x_t),
$$
at the first update. These values then propagate to provide messages
that satisfy \cref{eq:bp.mes}
starting from the next-to-terminal vertexes (those that have only one
neighbor left when the terminals are removed) and so on.

\section{Belief propagation and free energy approximation}
\subsection{BP stationarity}
It is possible to run \cref{alg:message.tree} on graphs that are not acyclic, since nothing in its formulation requires this property. However, while the method stabilizes in finite time for acyclic graphs, this property, or even the convergence of the messages is not guaranteed for general, loopy, graphs. Convergence, however, has been observed in a large number of applications, sometimes with very good approximations of the true marginal distributions. 


We will refer to stable solutions of \cref{alg:message.tree} as BP-stationary points, as formally stated in the next
definition, which allows for a possible normalization of messages,
which is particularly important with loopy networks. 

\begin{definition}
\label{def:BP.stat}
Let $G=(V, E)$ be an undirected graph and $\Phi = (\phi_{st},
\defset{s,t}\in E, \phi_s, s\in V)$ a consistent family of pair interactions. We say that a family of joint
probability distributions $(\pi'_{st}, \defset{s,t}\in E)$ is
BP-stationary for $(G,\Phi)$ if there exists messages $x_t\in F_t
\mapsto m_{st}(x_t)$, constants $\zeta_{st}$ for $t\sim s$ and $\al_s$ for
$s\in V$  
satisfying
\begin{equation}
\label{eq:bp.mes.1}
m_{ts}(\pe xs) = \frac{\al_s}{\ze_{ts}} \sum_{\pe x{t}\in F_{t}} \phi_{st}(\pe xs, \pe x{t}) \phi_t(\pe xt) \prod_{t'\in\CV_t\setm \{s\}} m_{t't}(\pe xt)
\end{equation}
 such that
\begin{equation}
\label{eq:bp.edge.1}
\pi'_{st}(\pe xs, \pe xt) 
=  \frac{1}{\ze_{st}} \phi_{st}(\pe xs, \pe xt) \phi_s(\pe xs)
\phi_t(\pe xt) \prod_{t'\in\CV_t\setm \{s\}} m_{t't}(\pe xt) \prod_{s'\in\CV_s\setm \{t\}} m_{s's}(\pe xs).
\end{equation}
\end{definition}
There is no loss  of generality in the specific form chosen for the normalizing
constants in \cref{eq:bp.mes.1,eq:bp.edge.1}, in the
sense that, if the messages satisfy \cref{eq:bp.edge.1} and
\[
m_{ts}(\pe xs) = q_{ts} \sum_{\pe x{t}\in F_{t}} \phi_{st}(\pe xs, \pe x{t}) \phi_t(\pe xt) \prod_{t'\in\CV_t\setm \{s\}} m_{t't}(\pe xt)
\]
for some constants $q_{ts}$, then
 \begin{eqnarray*}
\ze_{st} &=& \sum_{\pe xs \in F_s, \pe xt\in F_t} \phi_{st}(\pe xs, \pe xt) \phi_s(\pe xs)
\phi_t(\pe xt) \prod_{t'\in\CV_t\setm \{s\}} m_{t't}(\pe xt)
\prod_{s'\in\CV_s\setm \{t\}} m_{s's}(\pe xs)\\
 &=& \frac{1}{q_{ts}} \sum_{\pe xs\in F_s} 
\phi_s(\pe xs) \prod_{s'\in\CV_s} m_{s's}(\pe xs)
\end{eqnarray*}
so that $\ze_{st}q_{ts}$ (which has been denoted $\al_s$) does not
depend on $t$.
Of course, the relevant questions regarding BP-stationarity is whether
the collection of pairwise probability $\pi'_{st}$ exists, how to compute them, and whether $\pi'_{st}(\pe xs, \pe xt)$
provides a good approximation of the marginals of the probability
distribution $\pi$ that is associated to $\Phi$, namely 
\[
\pi(x) = \frac{1}{Z} \prod_{s\in V} \phi_s(\pe xs) \prod_{\defset{s,t}\in E} \phi_{st}(\pe xs, \pe xt).
\]

A reassuring statement for BP-stationarity is that it
is not affected when the functions in $\Phi$ are multiplied by
constants, which does not affect the underlying probability $\pi$. This is stated in the next proposition.
\begin{proposition}
\label{prop:bp.ok}
Let $\Phi$ be as above a family of edge and vertex interactions. 
Let $c_{st}, \defset{s,t}\in E$, $c_s, s\in V$ be  families of positive constants, and
define $\tilde \Phi = (\tilde \phi_{st}, \tilde \phi_s)$ by $\tilde
\phi_{st} = c_{st}\phi_{st}$ and $\tilde \phi_s = c_s\phi_s$. Then,
\[
\pi' \text{ is  BP-stationary for } (G,\Phi) \\
\Leftrightarrow \pi' \text{ is
 BP-stationary for } (G,\tilde \Phi).
\]
\end{proposition}
\begin{proof}
Indeed, if \cref{eq:bp.mes.1} and \cref{eq:bp.edge.1} are true for
$(G, \Phi)$, it suffices to replace $\al_s$ by $\al_s c_s$ and $\zeta_{st}$
by $\zeta_{st}c_{st} c_t$ to obtain \cref{eq:bp.mes.1} and \cref{eq:bp.edge.1} for
$(G, \tilde\Phi)$.  
\end{proof}

It is also important to
notice that, if $G$ is acyclic,  \cref{def:BP.stat} is no
more general than the message-passing rule we had considered
earlier. More precisely, we have (see \cref{rem:bp.normalize}),
\begin{proposition}
\label{prop:bp.tree}
Let $G=(V, E)$ be undirected acyclic and  $\Phi = (\phi_{st},
\defset{s,t}\in E, \phi_s, s\in V)$ a consistent family of pair
interactions. Then, the only BP-stationary distributions are the
marginals of the distribution $\pi$ associated to $\Phi$. 
\end{proposition}

%
%

\subsection{Free-energy approximations}

A partial justification of the good behavior of BP with general graphs
has been provided in terms of a quantity introduced in statistical
mechanics, called the Bethe free energy. We let $G = (V,E)$ be an
undirected graph and assume that a consistent
family of pair interactions is given (denoted $\Phi = (\phi_s, s\in V,
\phi_{st}, \{s,t\}\in E)$) and consider the associated distribution,
$\pi$, on $\CF(V)$, given by
\begin{equation}
\label{eq:pi.tree.2}
\pi(x) = \frac{1}{Z} \prod_{s\in V} \phi_s(\pe xs) \prod_{\defset{s,t}\in E} \phi_{st}(\pe xs, \pe xt).
\end{equation}
It will also be convenient to use the
function 
\[
\psi_{st}(\pe xs,\pe xt) = \phi_s(\pe xs)\phi_t(\pe xt) \phi_{st}(\pe xs, \pe xt)
\]
such that
\begin{equation}
\label{eq:pi.tree.3}
\pi(x) = \frac{1}{Z} \prod_{s\in V} \phi_s(\pe xs)^{1-|\CV_s|} \prod_{\defset{s,t}\in E} \psi_{st}(\pe xs, \pe xt).
\end{equation}

We will consider approximations $\pi'$ of $\pi$ that minimize the
Kullback-Leibler divergence, $\KL(\pi'\|\pi)$ (see  \cref{eq:kl.definition}), subject to some
constraints. 
We can write
\begin{eqnarray*}
\KL(\pi'\|\pi) &=& -E_{\pi'} (\ln\pi) - H(\pi')\\
&=& - \ln Z -  \sum_{s\in V} (1-|\CV_s|)E_{\pi'} (\ln \phi_s) -
\sum_{\defset{s,t}\in E} E_{\pi'} (\ln \psi_{st}) - \CH(\pi')
\end{eqnarray*}
(where $\CH(\pi')$ is the entropy of $\pi'$).
Introduce the one- and two-dimensional marginals of $\pi'$, denoted
$\pi'_s$ ad $\pi'_{st}$. Then
\begin{multline*}
\KL(\pi'\|\pi) = - \ln Z -  \sum_{s\in V} (1-|\CV_s|)E_{\pi'} (\ln\frac{\phi_s}{\pi'_s}) -
\sum_{\defset{s,t}\in E} E_{\pi'} (\ln \frac{\psi_{st}}{\pi'_{st}}) \\
+
\sum_{s\in V} (1-|\CV_s|) \CH(\pi'_s) + \sum_{\defset{s,t}\in E}
\CH(\pi'_{st}) -  \CH(\pi').
\end{multline*}
The Bethe free energy is the function $\mathbb F_\be$ defined by
\begin{equation}
\label{eq:bfe}
\mathbb F_\be(\pi') = -  \sum_{s\in V} (1-|\CV_s|)E_{\pi'} (\ln \frac{\phi_s}{\pi'_s} )-
\sum_{\defset{s,t}\in E} E_{\pi'} (\ln \frac{\psi_{st}}{\pi'_{st}})\,;
\end{equation}
so that
$$
\KL(\pi'\|\pi) = \mathbb F_\be(\pi') -\ln Z +
\De_G(\pi')
$$
with
$$
\De_G(\pi') = \sum_{s\in V} (1-|\CV_s|) \CH(\pi'_s) + \sum_{\defset{s,t}\in E}
\CH(\pi'_{st}) - \CH(\pi').
$$

Using this computation, one can consider the approximation problem:
find $\hat \pi'$ that minimizes $\KL(\pi'\|\pi)$ over a class of
distributions $\pi'$ for which the computation of the first and second
order marginals is easy. This problem has an explicit solution when
the distribution $\pi'$ is such that all variables are
independent, leading to what is called the {\em mean-field
  approximation} of $\pi$. Indeed, in this case, we have
$$
\De_G(\pi') =  \sum_{\defset{s, t} \in G} (\CH(\pi'_{s}) + \CH(\pi'_t))
+ \sum_{s\in S} (1-|\CV_s|) \CH(\pi'_s) -  \sum_{s\in S} \CH(\pi'_s) = 0
$$
and
$$
\mathbb F_\be(\pi') = -  \sum_{s\in V} (1-|\CV_s|)E_{\pi'} (\ln \frac{\phi_s}{\pi'_s}) -
\sum_{\defset{s,t}\in E} E_{\pi'} ( \ln
\frac{\psi_{st}}{\pi'_{s}\pi'_t})\, .
$$
$\mathbb F_\be$ must be minimized with respect to the variables $\pi'_s(\pe xs), s\in S,
x_s\in F_S$ subject to the constraints $\sum_{x_s\in F_s} \pi'_s(\pe xs) =
1$. The corresponding necessary optimality conditions  equations provide the mean-field
consistency equations, described in the following definition.
\begin{proposition}
\label{prop:mean.f}
A local minimum of $\mathbb F_\be(\pi')$ over all probability distributions
$\pi'$ of the form
$$
\pi'(x) = \prod_{s\in V}\pi'_s(\pe xs)
$$
must satisfy the mean field consistency equations:
\begin{equation}
\label{eq:mf.cons}
\pi_s(\pe xs) = \frac{1}{Z_s}\phi_s(\pe xs)^{1-|\mathcal V_s|}\prod_{t\sim s} \exp\left(E_{\pi_t} (\ln\psi_{st}(\pe x, .))\right).
\end{equation}
\end{proposition}
\begin{proof}
Since all constraints are affine, we can use Lagrange multipliers, denoted $(\la_s, s\in S)$ for each of the
constraints, to obtain necessary conditions for a minimizer, yielding
$$
\pdr{\mathbb F_\be}{\pi_s(x_s)} -\la_s = 0, \ \ s\in S, x_s\in F_s.
$$
This gives:
\[
-(1-|\CV_s|)\left(\ln \frac{\phi_s(x_s)}{\pi_s(x_s)} -1\right) 
-\sum_{t\sim s} \sum_{x_t\in F_t}\left(\ln\frac{\psi_{st}(x_s,
x_t)}{\pi_s(x_s)\pi_t(x_t)}-1\right)\pi_t(x_t) = \la_s.
\]
Solving this with respect to $\pi_s(x_s)$ and regrouping all constant
terms (independent from $x_s$) in the normalizing constant $Z_s$ yields
\cref{eq:mf.cons}.
\end{proof}

The mean field consistency equations can be solved using a
root-finding algorithm or by directly solving the minimization
problem. We will retrieve this method, with more details, in our discussion of variational approximations in \cref{chap:var.bayes}.

In the particular case in which $G$
is acyclic and the approximation is made by $G$-Markov
processes, the Kullback-Leibler distance is
minimized with $\pi'=\pi$ (since $\pi$ belongs to the approximating
class). A slightly non-trivial remark is that $\pi$ is optimal also
for the minimization of the Bethe free energy $F_\be$, because this
energy coincides, up to the constant term $\ln Z$, with the
Kullback-Leibler divergence, as proved by the following proposition.
\begin{proposition}
\label{prop:bethe.tree}
If $G$ is acyclic and $\pi'$ is $G$-Markov, then $\De_G(\pi')  = 0$.
\end{proposition}
This proposition is a consequence of the following lemma that has
its own interest:
\begin{lemma}
\label{lem:tree.acyc}
If $G$ is acyclic and $\pi$ is a $G$-Markov distribution, then
\begin{equation}
\label{eq:tree.acyc}
\pi(x) = \prod_{s\in V} \pi_s(\pe xs)^{1-|\CV_s|}
\prod_{\defset{s,t}\in E}
\pi_{st}(\pe xs, \pe xt).
\end{equation}
\end{lemma}
\begin{proof}[of  \cref{lem:tree.acyc}]
We know that, if $\tilde G = (V, \tilde E)$ is a tree such that
$\tilde G^\flat = G$, we have, letting $s_0$ be the root in $\tilde
G$
\begin{eqnarray*}
\pi(x) &=& \pi_{s_0}(\pe x{s_0}) \prod_{(s,t)\in \tilde E}
p_{st}(\pe xs, \pe xt)\\
&=& \pi_{s_0}(\pe x {s_0}) \prod_{(s,t)\in \tilde E} (\pi_{st}(\pe xs, \pe xt)
\pi(\pe xs)^{-1}).
\end{eqnarray*}
Each vertex $s$ in $V$ has $|\CV_s|-1$ children in $\tilde G$, except
$s_0$ which has $|\CV_{s_0}|$ children. Using this, we get
\begin{eqnarray*}
\pi(x) &=& \pi_{s_0}(\pe x{s_0}) \pi_{s_0}(\pe x{s_0})^{-|\CV_{s_0}|}
\prod_{s\in V\setm \{s_0\}} \pi_s(\pe xs)^{1-|\CV_s|} \prod_{(s,t)\in \tilde E} \pi_{st}(\pe xs, \pe xt)\\
&=& \prod_{s\in V} \pi_s(\pe xs)^{1-|\CV_s|} \prod_{\defset{s,t}\in  E} \pi_{st}(\pe xs, \pe xt).
\end{eqnarray*}
\end{proof}
\begin{proof}[of  \cref{prop:bethe.tree}]
If $\pi'$ is given by \cref{eq:tree.acyc}, then
\begin{eqnarray*}
H(\pi') &=& - E_{\pi'} \ln \pi'\\
&=& - \sum_{s\in V} (1-|\CV_s|) E_{\pi'} \ln\pi'_s -
\sum_{\defset{s,t}\in E} E_{\pi'}\ln \pi'_{st}\\
&=& \sum_{s\in V} (1-|\CV_s|) H(\pi'_s) +
\sum_{\defset{s,t}\in E} H(\pi'_{st})
\end{eqnarray*}
which proves that $\De_G(\pi') = 0$.
\end{proof}

In view of this, it is tempting to ``generalize'' the mean field
optimization procedure and minimize $\mathbb F_\be(\pi')$ over all possible consistent
singletons and pair marginals ($\pi'_s$ and $\pi'_{st}$), then use the
optimal ones as an approximation of $\pi_s$ and $\pi_{st}$. What we
have just proved is that this procedure 
provides the exact expression of the marginals when $G$ is acyclic. For loopy graphs,
however, it is not justified, and is at best an approximation. A very
interesting fact is that this procedure provides the same consistency
equations as belief propagation. To see this, we first start with the
characterization of minimizers of $\mathbb F_\be$.
\begin{proposition}
\label{prop:bethe.loopy}
Let $G=(V,E)$ be an undirected graph and $\pi$ be given by \cref{eq:pi.tree.2}. Consider the problem of minimizing the
Bethe free energy $\mathbb F_\be$ in \cref{eq:bfe} with respect to all
possible choices of probability distributions $(\pi'_{st},
\defset{s,t}\in E)$, $(\pi'_s, s\in V)$ with the
constraints
$$
\pi'_{s}(\pe xs) = \sum_{\pe xt\in F_t} \pi'_{st}(\pe xs,\pe xt), \forall \pe xs\in
F_s \text{ and } t\sim s.
$$
Then a local minimum of this problem must take the form
\begin{equation}
\label{eq:bethe.loopy.1}
\pi'_{st}(\pe xs, \pe xt) = \frac{1}{Z_{st}}\psi_{st}(\pe xs, \pe xt) \mu_{st}(\pe xt) 
\mu_{ts}(\pe xs)
\end{equation}
where the functions $\mu_{st}: F_t\to [0, +\infty)$ are defined for all $ (s,t)$
such that $\defset{s,t}\in E$ and satisfy the consistency conditions:
\begin{equation}
\label{eq:bethe.loopy.2}
\mu_{ts}(\pe xs)^{-(|\CV_s|-1)} \prod_{s'\sim
s}\mu_{s's}(\pe xs) 
= \left(\frac{e}{Z_{st}}
\sum_{\pe xt\in F_t} \psi_{st}(\pe xs,\pe xt) \phi_t(\pe xt) \mu_{st}(\pe xt)\right)^{|\CV_s|-1}.
\end{equation}
\end{proposition}
\begin{proof}
We introduce Lagrange multipliers:
$\la_{ts}(\pe xs)$ for the constraint 
$$
\pi'_{s}(\pe xs) = \sum_{\pe xt\in F_t} \pi'_{st}(\pe xs, \pe xt)
$$
and $\ga_{st}$ for 
$$
\sum_{\pe xs, \pe xt} \pi'_{st}(\pe xs, \pe xt) = 1,
$$
which covers all constraints associated to the minimization
problem. The associated Lagrangian is
\begin{multline*}
\mathbb F_\be(\pi') - \sum_{s\in V}\sum_{\pe xs\in F_s} \sum_{t\sim s}
\la_{ts}(\pe xs) \left(\sum_{\pe xt\in F_t} \pi'_{st}(\pe xs, \pe xt) -
\pi'_s(\pe xs)\right)\\
 - \sum_{\{s,t\} \in E} \ga_{st}\left(\sum_{\pe xs\in F_s ,\pe xt\in F_t} \pi'_{st}(\pe xs, \pe xt)
- 1\right).
\end{multline*}
The derivative with respect to $\pi'_{st}(\pe xs, \pe xt)$ yields the
condition
$$
\ln\pi'_{st}(\pe xs, \pe xt) - \ln\psi_{st}(\pe xs, \pe xt) +1 -\la_{ts}(\pe xs)
 - \la_{st}(\pe xt) -\ga_{st}=0.
$$
which implies
\[
\pi'_{st}(\pe xs, \pe xt) = \phi_{st}(\pe xs, \pe xt)\exp(\ga_{st}-1) \exp(\la_{ts}(\pe xs)
 + \la_{st}(\pe xt) ).
 \]
 We let $Z_{st} = \exp(1-\ga_{st})$, with $\ga_{st}$ chosen so that $\pi'_{st}$ is a probability.
The derivative with respect to $\pi'_s(\pe xs)$ gives
$$
(1-|\CV_s|)(\ln\pi'_s(\pe xs) - \ln\phi_s(\pe xs) +1)  + \sum_{t\sim
s}\la_{ts}(\pe xs)  = 0.
$$
Combining this with the expression just obtained for $\pi_{st}'$, we
get, for $t\sim s$,
\begin{multline*}
(1-|\CV_s|) \ln\sum_{\pe xt\in F_t} \psi_{st}(\pe xs, \pe xt) e^{\la_{st}(\pe xt)}
+ (1-|\CV_s|) \la_{ts}(\pe xs) \\ + (1-|\CV_s|) (1 -\ln Z_{st} - \ln\phi_s(\pe xs)) 
+ \sum_{s'\sim
s}\la_{s's}(\pe xs) =0,
\end{multline*}
which gives \cref{eq:bethe.loopy.2} with $\mu_{st} =
\exp(\la_{st})$. 
\end{proof}

A family $\pi'_{st}$ satisfying conditions \cref{eq:bethe.loopy.1,eq:bethe.loopy.2}  of 
\cref{prop:bethe.loopy} will be called Bethe-consistent. A very
interesting remark states that Bethe-consistency is
equivalent to BP-stationarity, as stated below.
\begin{proposition}
\label{prop:bp.be}
Let $G= (V,E)$ be an undirected graph and  $\Phi = (\phi_{st},
\defset{s,t}\in E, \phi_s, s\in V)$ a consistent family of pair
interactions.  Then a family $\pi'$ of joint probability distributions
is BP-stationary if and only if it is Bethe-consistent.
\end{proposition}
\begin{proof}
First assume that $\pi'$
is BP-stationary with messages $m_{st}$, so that \cref{eq:bp.mes.1,eq:bp.edge.1} are satisfied. Take
\[
\mu_{st} = a_{t} \prod_{t'\in \CV_t, t'\neq s} m_{t't}(\pe xt)
\] 
for some constant $a_{t}$ that will be determined later. Then, the
left-hand side of \cref{eq:bethe.loopy.2} is  
\begin{align*}
\mu_{ts}(\pe xs)^{-(|\CV_s|-1)} \prod_{s'\in\CV_s}\mu_{s's}(\pe xs) &= a_s \left(\prod_{s'\in \CV_s, s'\neq t}
  m_{s's}(\pe xs)\right)^{-(|\CV_s|-1)} \prod_{s'\in\CV_s}
\prod_{s''\in\CV_s, s''\neq s'} m_{s''s}(\pe xs)\\
&= a_s m_{ts}(\pe xs)^{|\CV_s|-1}.
\end{align*}
The right-hand side is equal to (using \cref{eq:bp.mes.1})
\[
\left(\frac{e a_t \zeta_{st}}{Z_{st}\alpha_s}
  m_{ts}(\pe xs)\right)^{|\CV_s| -1}, 
\] 
so that we need to have
\[
a_s = \left(\frac{e a_t \zeta_{st}}{Z_{st}\alpha_s}\right)^{|\CV_s|
  -1}.
\]
We also need 
\[
Z_{st} = \sum_{\pe xs, \pe xt} \psi_{st}(\pe xs, \pe xt) \mu_{st}(\pe xt)
\mu_{ts}(\pe xs) = a_s a_t \ze_{st}.
\]
Solving these equations, we find that \cref{eq:bethe.loopy.1} and
\cref{eq:bethe.loopy.2} are satisfied with
\[
\begin{cases}
a_s = (e/\al_s)^{(|\CV_s|-1)/|\CV_s|}\\
Z_{st} = \ze_{st} a_s a_t
\end{cases}
\]
which proves that $\pi'$ is Bethe-consistent.

Conversely, take a Bethe-consistent $\pi'$, and $\mu_{st}, Z_{st}$
satisfying \cref{eq:bethe.loopy.1} and
\cref{eq:bethe.loopy.2}. For $s$ such that $|\CV_s| > 1$, define, for $t\in\CV_s$,
\begin{equation}
\label{eq:be.to.bp}
m_{ts}(\pe xs) = \mu_{ts}(\pe xs)^{-1} \prod_{s'\sim s}
\mu_{s's}(\pe xs)^{1/(|\CV_s|-1)}.
\end{equation}
Define also, for $|\CV_s| > 1$,
\[
\rho_{ts}(\pe xs) = \prod_{s'\in \CV_s, s'\neq t} m_{s's}(\pe xs).
\]
(If $|\CV_s| = 1$, take $\rho_{ts} \equiv 1$.)
Using \cref{eq:be.to.bp}, we find
$\rho_{ts} = \mu_{ts}$ when $|\CV_s| > 1$, and this identity is still valid when $|\CV_s| =1$, since in this case,
\cref{eq:bethe.loopy.2} implies that $\mu_{ts}(\pe xs) = 1$. 

We need to find constants $\al_t$ and $\ze_{st}$ such that
\cref{eq:bp.mes.1} and \cref{eq:bp.edge.1} are satisfied. 
 But \cref{eq:bp.edge.1} implies
\[
\ze_{ts} = \sum_{x_t, x_s} \psi_{st}(\pe xs, \pe xt) \rho_{st}(\pe xt)
\rho_{ts}(\pe xs)
\]
and \cref{eq:bethe.loopy.1} implies $\ze_{ts} =  Z_{ts}$.

We now consider \cref{eq:bp.mes.1}, which requires
\[
m_{ts}(\pe xs) = \frac{\al_s}{\ze_{st}} \sum_{\pe xt} \phi_{st}(\pe xs, \pe xt)
\phi_t(\pe xt) \rho_{st}(\pe xt). 
\]
It is now easy to see that this identity to the power $|\CV_s| - 1$
coincides with \cref{eq:bethe.loopy.2} as soon as one takes
$\al_s = e$. 
\end{proof}

\section{Computing the most likely configuration}
We now address the problem of finding a configuration that maximizes
$\pi(x)$ (mode determination). This problem turns out to be very similar to the
computation of marginals, that we have considered so far, and we will
obtain similar algorithms.

Assume that  $G$ is undirected and acyclic
and that $\pi$ can be written as
\[
\pi(x) = \frac{1}{Z} \prod_{\defset{s,t}\in E} \phi_{st}(\pe xs,\pe xt)
\prod_{s\in V} \phi_s(\pe xs).
\]

Maximizing $\pi(x)$ is equivalent  to maximizing
\begin{equation}
\label{eq:u.max}
U(x) = \prod_{\defset{s,t}\in E} \phi_{st}(\pe xs,\pe xt)
\prod_{s\in V} \phi_s(\pe xs).
\end{equation}
Assume that a root has been chosen in $G$, with the resulting edge orientation yielding a tree $\tilde G = (V, \tilde
E)$ such that $\tilde G^\flat = G$. We partially order the vertexes according
to $\tilde G$, writing $s\leq t$ if there exists a path from $s$ to $t$
in $\tilde G$ ($s$ is an ancestor of $t$). Let $V_s^+$ contain all
$t\in V$ with $t\geq s$, and define
\[
U_s(\pe x{V_s^+}) =  \prod_{\defset{t,u}\in E_{V_s^+}} \phi_{tu}(\pe xt, \pe xu)
\prod_{t > s} \phi_t(\pe xt)
\]
and
\begin{equation}
\label{eq:us*}
U^*_s(\pe xs) = \max\defset{U_s(\pe y{V_s^+}), \pe ys = \pe xs}.
\end{equation}
Since we can write
\begin{equation}
\label{eq:us.dec}
U_s(\pe x{V_s^+}) =  \prod_{t\in s^+} \phi_{st}(\pe xs, \pe xt) \phi_t(\pe xt)  U_t(\pe x{V_t^+}),
\end{equation}
we have
\begin{eqnarray}
\nonumber
U^*_s(\pe x{s}) &=& \max_{\pe xt, t\in s^+} \left( \prod_{t\in s^+} \phi_t(\pe xt)\phi_{st}(\pe xs, \pe xt) U^*_t(\pe x{t})\right)\\
\label{eq:max.prod.1}
&=&   \prod_{t\in s^+} \max_{x_t\in F_t}(\phi_t(\pe xt)\phi_{st}(\pe xs, \pe xt) U^*_t(\pe x{t})).
\end{eqnarray}

This provides a
method to compute $U^*_{s}(\pe x{s})$ for all $s$, starting with the leaves and
progressively updating the parents. (When $s$ is a leaf, $U_s^*(\pe xs) = 1$, by definition.)

Once all $U^*_s(\pe xs)$ have been computed, it is possible to obtain a
configuration $x_*$ that maximizes $\pi$. This is because an
optimal configuration must satisfy
$U^*_s(\pe xs_*) = U_s(\pe x{V^+_s}_*)$ for all $s\in V$, i.e., $\pe x{V^+_s\setm\{s\}}_*$ must
solve the maximization problem in \cref{eq:us*}. But because of
\cref{eq:us.dec}, we can separate this problem over the children of
$s$ and obtain the fact that, it $t\in s^+$,
$$
\pe xt_* = \mathop{\mathrm{argmax}}_{\pe xt} \left(\phi_t(\pe xt) \phi_{st}(\pe xs_*,\pe xt)U_t^*(\pe xt)\right).
$$

This procedure can be rewritten in a slightly different form using messages similar to the belief propagation algorithm. It $s \in t^+$, define 
\[
\mu_{st}(\pe xt) = \max_{x_s\in F_s}(\phi_t(\pe xt)\phi_{ts}(\pe xt, \pe xs) U^*_s(\pe x{s}))
\]
and
\[
\xi_{st}(\pe xt) = \mathop{\mathrm{argmax}}_{\pe xs\in F_s}(\phi_t(\pe xt) \phi_{ts}(\pe xt, \pe xs) U^*_s(\pe x{s})).
\]
Using \cref{eq:max.prod.1}, we get
\begin{eqnarray*}
\mu_{st}(\pe xt) &=& \max_{\pe xs\in F_s}\left(\phi_{ts}(\pe xt,\pe xs)\phi_s(\pe xs) \prod_{u\in s^+} \mu_{us}(\pe xs)\right),\\
\xi_{st}(x_t) &=& \mathop{\mathrm{argmax}}_{\pe xs\in F_s}\left(\phi_{ts}(\pe xt, \pe xs)\phi_s(\pe xs) \prod_{u\in s^+} \mu_{us}(\pe xs)\right).
\end{eqnarray*}
An optimal configuration can now be computed using $\pe xt_* = \xi_{ts}(\pe xs_*)$, with $s \in \pa{t}$.

This resulting algorithm therefore first operates upwards on the tree
(from leaves to root) to compute the $\mu_{st}$'s and $\xi_{st}$'s, then downwards to
compute $x_*$. This is summarized in the following algorithm.
\begin{algorithm}
\label{alg:dyn.prog}
A most likely configuration for 
$$
\pi(x) = \frac{1}{Z} \prod_{\defset{s,t}\in E} \phi_{st}(x_s,x_t)
\prod_{s\in V} \phi_s(x_s).
$$
can be computed after iterating the following updates, based on any acyclic orientation of $G$:
\begin{enumerate}
\item Compute, from leaves to root: 
\[
\displaystyle \mu_{st}(\pe xt) = \max_{\pe xs\in F_s}\left(\phi_{ts}(\pe xt, \pe xs)\phi_s(\pe xs) \prod_{u\in s^+} \mu_{us}(\pe xs)\right)
\]
 and $\displaystyle \xi_{st}(\pe xt) = \mathop{\mathrm{argmax}}_{\pe xs\in F_s}\left(\phi_{ts}(\pe xt,\pe xs)\phi_s(\pe xs) \prod_{u\in s^+} \mu_{us}(\pe xs)\right)$.
\item Compute, from root to leaves: $\pe xt_* = \xi_{ts}(\pe xs_*)$, with $s = \pa{t}$.
\end{enumerate}
\end{algorithm}

Similar to the computation of marginals, this
algorithm can be rewritten in an orientation-independent form. The main remark is that the value of $\mu_{st}(\pe xt)$ does not depend on the  tree orientation, as long as it is chosen such that $s\in t^+$, i.e., the edge $\{s,t\}$ is oriented from $t$ to $s$. This is because such a choice uniquely prescribes the orientation of the edges of the descendants of $s$ for any such tree, and $\mu_{st}$ only depends on this structure. Since the same remark holds for $\xi_{st}$, this provides a definition of these two quantities for any pair $s,t$ such that $\{s,t\}\in E$. 
 The updating rule now
becomes
\begin{eqnarray}
\label{eq:max.prod.final}
\mu_{st}(\pe xt) &=& \max_{\pe xs\in F_s}\left(\phi_{ts}(\pe xt,\pe xs)\phi_s(\pe xs) \prod_{u\in \CV_s\setm \{t\}} \mu_{us}(\pe xs)\right),\\
\label{eq:max.prod.final.2}
\xi_{st}(\pe xt) &=& \mathop{\mathrm{argmax}}_{\pe xs\in F_s}\left(\phi_{ts}(\pe xt,\pe xs)\phi_s(\pe xs) \prod_{u\in \CV_s\setm \{t\}} \mu_{us}(\pe xs)\right)
\end{eqnarray}
with $\pe xt_* = \xi_{ts}(\pe xs_*)$ for any pair $s\sim t$.
Like with the $m_{ts}$ in the previous section, looping over updating
all $\mu_{ts}$ in
any order will finally stabilize to their correct values, although, if an orientation is given,
going from leaves to roots is obviously more efficient.

The previous analysis is not valid for loopy graphs but  \cref{eq:max.prod.final} and \cref{eq:max.prod.final.2}
provide well defined iterations when $G$ is an arbitrary undirected
graph, and can therefore be used as such, without any guaranteed behavior.

\section{General sum-prod and max-prod algorithms}
\subsection{Factor graphs}
The expressions we obtained for message updating with belief
propagation and with mode determination respectively took
the form
\[
m_{ts}(\pe xs) \leftarrow \sum_{\pe x{t}\in F_{t}} \phi_{st}(\pe xs, \pe x{t}) \phi_t(\pe xt) \prod_{t'\in\CV_t\setm \{s\}} m_{t't}(\pe xt)
\]
and
$$
\mu_{ts}(\pe xs) \leftarrow \max_{\pe x{t}\in F_{t}}\left( \phi_{st}(\pe xs, \pe x{t}) \phi_t(\pe xt) \prod_{t'\in\CV_t\setm \{s\}} \mu_{t't}(\pe xt)\right).
$$

They first one is often referred to as the ``sum-prod'' update rule,
and the second as the ``max-prod''. In our construction, the sum-prod
algorithm provided us with a method computing
\[
\sig_s(\pe xs) = \sum_{\pe y{V\setm\defset{s}}} U(\pe xs\wedge \pe y{V\setm \{s\}})
\]
with 
\[
U(x) = \prod_{s} \phi_s(\pe xs) \prod_{\defset{s,t}\in E}
\phi_{st}(\pe xs, \pe xt).
\]
Indeed, we have, according to \cref{eq:bp.bel}
\[
\sig_s(\pe xs) = \phi_s(\pe xs) \prod_{t\in \CV_s} m_{ts}(\pe xs).
\]

Similarly, the max-prod algorithm computes
\[
\rho_s(\pe xs) = \max_{y_{V\setm\defset{s}}} U(\pe xs\wedge \pe y{V\setm \{s\}})
\]
via the relation
\[
\rho_s(\pe xs) = \phi_s(\pe xs) \prod_{t\in \CV_s} \mu_{ts}(\pe xs).
\]

We now discuss generalizations of these algorithms to situations in
which the function $U$ does not decompose as a product of bivariate
functions. More precisely, let $\CS$ be a subset of $\twop{V}$,
and assume the decomposition
$$
U(x) = \prod_{C\sub \CS} \phi_C(x_C). 
$$
The previous algorithms can be generalized  using the concept of {\em factor
graphs} associated with the decomposition. The vertexes of this graph
are either indexes $s\in V$ or sets $C\in \CS$, and the only edges link  indexes and sets that contain them. The formal definition is
as follows.
\begin{definition}
\label{def:fact.grph}
Let $V$ be a finite set of indexes and $\CS$ a subset of $\twop{V}$. The
factor graph associated to $V$ and $\CS$ is the graph $G=(V\cup
\CS,E)$, $E$ being constituted of all pairs $\defset{s,C}$ with $C\in
\CS$ and $s\in C$.
\end{definition}
We assign the variable
$\pe xs$ to a vertex $s\in V$ of the factor graph, and  the function
$\phi_C$ to $C\in\CS$. With this in mind,
the sum-prod and max-prod algorithms are extended to factor graphs as
follows.

\begin{definition}
\label{def:maxsumprod}
Let $G=(V\cup
\CS,E)$ be a factor graph, with associated functions
$\phi_C(x_C)$. The sum-prod algorithm on $G$ updates messages
$m_{sC}(x_s)$ and $m_{Cs}(x_s)$ according to the rules
\begin{equation}
\label{eq:sumprod}
\left\{
\begin{aligned}
&m_{sC}(\pe xs) \leftarrow \prod_{\tilde C, s\in \tilde C,  \tilde C\neq C} m_{\tilde C s}(\pe xs)\\
&m_{Cs}(\pe xs) \leftarrow \sum_{y_{C}: \pe ys = \pe xs} \phi_C(\pe y{C}) \prod_{t\in C\setm \{s\}} m_{tC}(\pe yt)
\end{aligned}
\right.
\end{equation}

Similarly, the max-prod algorithm iterates
\begin{equation}
\label{eq:maxprod}
\left\{
\begin{aligned}
&\mu_{sC}(\pe xs) \leftarrow \prod_{\tilde C, s\in \tilde C,  \tilde C\neq C} \mu_{\tilde
C s}(\pe xs)\\
&\mu_{Cs}(\pe xs) \leftarrow \max_{\pe y{C}:\pe ys=\pe xs} \phi_C(\pe y{C}) \prod_{t\in C\setm \{s\}} \mu_{tC}(\pe yt)\\
\end{aligned}
\right.
\end{equation}
\end{definition}
These algorithms reduce to the original ones when
only single vertex and pair interactions exist. Let us check this with 
sum-prod.  In this case, the set $\CS$ contains all singletons $C=\{s\}$, with
associated function $\phi_s$, and all edges $\defset{s,t}$ with
associated function $\phi_{st}$. We have links between $s$ and $\{s\}$
and $s$ and $\{s,t\}\in E$. For singletons, we have
$$
m_{s\{s\}}(\pe xs) \leftarrow \prod_{t\sim s} m_{s\{s,t\}} (\pe xs) \text{
and } m_{\{s\}s}(\pe xs) \leftarrow \phi_s(\pe xs).
$$
For pairs,
$$
m_{s\{s,t\}}(\pe xs) \leftarrow \phi_s(\pe xs) \prod_{\tilde t\in\CV_s\setm \{t\}}
m_{\{s,\tilde t\}s} (\pe xs)
$$
and
$$
m_{\{s,t\}s}(\pe xs) \leftarrow \sum_{\pe yt} \phi_{st}(\pe xs,\pe yt)
m_{t\{s,t\}}(\pe yt)
$$
and, combining the last two assignments, it becomes clear that we retrieve the initial algorithm with
$m_{\{s,t\}s}$ taking the role of what we previously denoted $m_{ts}$.

The important question, obviously, is whether the algorithms
converge. The following result shows that this is true when the factor
graph is acyclic.

\begin{proposition}
\label{prop:fct.grph.cvg}
Let $G = (V\cup\CS, E)$ be a factor graph with associated
functions $\phi_C$. Assume that $G$ is acyclic. Then the sum-prod and max-prod algorithms converge
in finite time. 

After convergence, we have 
$\sig_s(\pe xs) = \prod_{C, s\in C} m_{Cs}(\pe xs)$
and 
$\rho_s(\pe xs) = \prod_{C, s\in C} \mu_{Cs}(\pe xs)$.
\end{proposition}
\begin{proof}
Let us assume that $G$ is connected, which is without loss of
generality, since the following argument can be applied to each component of $G$ separately.
Since $G$ is acyclic, we can
arbitrarily select one of its vertexes as a root to form a tree.
This
being done, we can see that the messages going upward in the tree
(from children to parent) progressively stabilize, starting with leaves.
Leaves in the factor graph indeed are either singletons, $C=\{s\}$, or vertexes
$s\in V$ that belong to only one set $C\in\CS$. In the first case, the
algorithm imposes (taking, for example, the sum-prod case)
$m_{\{s\}s}(\pe xs) = \phi_s(\pe xs)$, and in the second case 
$m_{sC}(\pe xs) = 1$. So the messages sent upward by the leaves are set at the
first step. Since the messages going from a child to its parents only
depend on the messages that it received from its other neighbors
in the acyclic graph, which are its children in the tree,  it is clear
that all upward messages progressively stabilize until the root is
reached. Once this is done, messages propagate downward 
from each parent to its children. This stabilizes as soon as all
incoming messages to the parent are stabilized, since outgoing messages
only depend on those. At the end of the upward phase, this is true
for the root, which can then send its stable message to its
children. These children now have all their incoming messages and can
now send their messages to their own children and so on down to the
leaves.

We now consider the second statement,   proceeding by induction,  assuming that the result is true for any
smaller graph than the one considered. Let $s_0$ be the selected root, and
consider all vertexes $s\neq s_0$ such that  there exists $C_s\in
\CS$ such
that $s_0$ and $s$ both belong to $C_s$. Given $s$, there cannot be more than
one such $C_s$ since this would create a loop in the graph. For each such $s$, consider the part
$G_s$ of $G$ containing all descendants of $s$. Let $V_s$ be the set
of vertexes among the descendants of $s$ and $\CC_s$ the set of $C$'s
below $s$. Define
\[
U_s(\pe x{V_s}) = \prod_{C\in\CC_s}\phi_C(\pe xC).
\]

Since the upward phase
of the algorithm does not depend on the ancestors of $s$, the messages
incoming to $s$ for the sum-prod algorithm restricted to $G_s$ are the
same as with the general algorithm, so that, using the
induction hypothesis 
\[
\sum_{\pe y{V_s}, \pe ys=\pe xs} U_s(\pe y{V_s}) = \prod_{C\in \CC_s, s\in C}
m_{Cs}(\pe xs) = m_{sC_s}(\pe xs).
\]

Now let $C_1, \ldots, C_n$ list all the sets in $\CC$ that contain $s_0$,
which must be non-intersecting (excepted at $\{s_0\}$), again not to
create loops. Write
\[
C_1 \cup \cdots \cup C_n = \{s_0, s_1, \ldots, s_q\}.
\]
 Then, we
have
\[
U(x) = \prod_{j=1}^n \phi_{C_j}(\pe x{C_j}) \prod_{i=1}^qU_{s_i}(\pe x {V_{s_i}})
\]
and letting $S = \bigcup_{j=1}^n C_j\setm \{s_0\}$, 
\begin{eqnarray*}
\sig_{s_0}(\pe x {s_0}) &=& \sum_{\pe yV: \pe y{s_0}= \pe x{s_0}} \prod_{j=1}^n \phi_{C_j}(\pe y{C_j})
\prod_{i=1}^q U_{s_i}(\pe y{V_{s_i}})\\
&=& \sum_{\pe y_S: \pe y{s_0}= \pe x{s_0}} \prod_{j=1}^n \phi_{C_j}(\pe y{C_j})
\prod_{i=1}^q m_{s_iC_{s_i}}(\pe y{s_i})\\
&=& \prod_{j=1}^n \sum_{\pe y_{C_j}: \pe y{s_0}=\pe x{s_0}} \phi_{C_j}(\pe y{C_j})
\prod_{s\in C_j\setm\{s_0\}} m_{sC_{s}}(\pe y{s})\\
&=& \prod_{j=1}^n  m_{C_js_0}(\pe x{s_0})
\end{eqnarray*}
which proves the required result (note that, when factorizing the sum,
we have used the fact that the sets $C_j\setm \{s_0\}$ are non
intersecting). An almost identical argument holds for the max-prod
algorithm.
\end{proof}

\begin{remark}
Note that these algorithms are not always feasible. For example, it is
always possible to represent a function $U$ on $\CF(V)$ with the
trivial factor graph in which $\CS=\{V\}$ and $E$ contains all
$\{s,V\}, s\in V$ (using $\phi_V=U$), but computing $m_{Vs}$ is
identical to directly computing $\sig_s$ with a sum over all
configurations on $V\setm\{s\}$ which grows exponentially.
In fact, the complexity of the sum-prod and max-prod algorithms is
exponential in the size of the largest $C$ in $\CS$ which should
therefore remain small.
\end{remark}

\begin{remark}
It is not always possible to decompose a function so that the
resulting factor graph is acyclic with small degree (maximum number of
edges per vertex). Sum-prod
and max-prod can still be used with loopy networks, sometimes with excellent
results, but without theoretical support.
\end{remark}

\begin{remark}
\label{rem:JT.example}
One can sometimes transform a given factor graph into an acyclic one
by grouping vertexes. Assume that the set $\CS \sub \twop{V}$ is
given. We will say that a partition  $\De = (D_1, \ldots D_k)$ of $V$
is $\CS$-admissible if, for any $C\in \CS$ and any $j\in \{1, \ldots,
k\}$, one has either $D_j\cap C = \emp$ or $D_j\sub C$. 

If $\De$ is $\CS$-admissible, one can define a new factor graph $\tilde G$ as follows. We first let $\tilde V = \{1, \ldots,
k\}$. To define $\tilde \CS \sub \twop{\tilde V}$ assign to each $C\in\CS$ the set $J_C$ of indexes $j$ such that $D_j \sub C$.  From the admissibility assumption,
\begin{equation}
\label{eq:ctc}
C = \bigcup_{j\in J_C} D_j,
\end{equation}
so that $C \mapsto J_C$ is one-to-one.
Let $\tilde \CS = \defset{J_C, C\in\CS}$.
Group variables using $\pe {\tilde x}k = \pe x{D_k}$, so that $\tilde F_k = \CF(D_k)$. Define $\tilde \Phi =
(\tilde\phi_{\tilde C}, \tilde C \in\tilde \CS)$ by
$\tilde\phi_{\tilde C} = \phi_C$ where $C$ is given by
\cref{eq:ctc}.

In other terms, one groups variables $(\pe xs, s\in V)$ into clusters, to
create a simpler factor graph, which may be acyclic even if the
original one was not. For example, if $V = \{a,b,c,d\}$, $\CS =\{A,
B\}$ with 
$A=\{a,b,c\}$ and $B = \{b,c,d\}$, then $(A, c, B, b)$ is a cycle in
the associated factor graph. If, however, one takes $D_1 = \{a\}$,
$D_2= \{b,c\}$ and $D_3 = \{d\}$, then $(D_1, D_2, D_3)$ is
$\CS$-admissible and  the associated factor graph is acyclic. In fact,
in such a case, the resulting factor graph, considered as a graph with
vertexes given by subsets of $V$, is a special case of a junction tree,
which is defined in the next section.
\end{remark}

\subsection{Junction trees}

\begin{definition}
\label{def:jct.tree}
Let $V$ be a finite set. A junction tree on $V$ is an undirected
acyclic graph $\mathbb G = (\CS, \mathbb E)$ where $\CS \sub \twop{V}$ is a family
of subsets of $V$ that satisfy the following property, called the running intersection constraint:
if $C, C'\in\CS$ and $s\in C\cap C'$, then all sets $C''$ in the
(unique) path connecting $C$ and $C'$ in $\mathbb G$ must also contain
$s$.
\end{definition}

\begin{remark}
\label{rem:JT.example.2}

Let us check that the clustered factor graph $\tilde G$ defined in \cref{rem:JT.example} is equivalent to a junction tree when acyclic. 
Using the same
notation, let $\hat \CS = \{D_1, \ldots, D_k\}\cup \CS$, removing if needed sets $C\in \CS$ that coincide with one of the $D_j$'s.  Place an edge between $D_j$ and $C$ if and only if $D_j \sub C$. 

Let
$(C_1, D_{i_1}, \ldots, D_{i_{n-1}}, C_n)$ be a path
in that graph. Assume that $s\in C_1\cap C_2$. Let $D_{i_n}$ be the unique $D_j$ that contains $s$. It is such that from the  the admissibility assumption, $D_{i_n} \sub C_1$ and $D_{i_n} \sub C_n$, which implies that $(C_1, D_{i_1}, \ldots, C_n, D_{i_n}, C_1)$ is a path in $\tilde G$. Since $\tilde G$ is acyclic, this path must be a union of folded paths. But it is easy to see that any folded path satisfies the running intersection constraint.  (Note that there was no loss of generality in assuming that the path started and ended with a ``$C$'', since any ``$D$'' must be contained in the $C$ that follows or precedes it.)
\end{remark}

We now consider a probability distribution written in
the form
\[
\pi(x) = \frac1Z \prod_{C\in\CS} \phi_C(\pe xC)
\]
and we  make the assumption that $\CS$ can be organized as a
junction  tree.

Belief propagation can be extended to junction
trees. Fixing a root $C_0\in \CS$, we first choose an orientation on $\mathbb G$, which induces as
usual a partial order on $\CS$. For $C\in \CS$, define $\CS_C^+$ as
the set of all $B\in\CS$ such that $B>C$. Define also 
\[
V_C^+ = \bigcup_{B\in \CS_C^+} B.
\]
We want to compute sums
\[
\sig_C(\pe xC) = \sum_{\pe y{V\setm C}} U(\pe xC \wedge \pe y{V\setm C}),
\]
where $U(x) = \prod_{C\in\CS} \phi_C(\pe xC)$.
We have
\[
\sig_C(\pe xC) = \sum_{\pe y{V\setm C}} \phi_C(\pe xC) \prod_{B\in \CS\setm
  \{C\}} \phi_B(\pe x{B\cap C} \wedge \pe y{B\setm C}).
\]
Define
\[
\sig^+_C(\pe xC) = \sum_{\pe y{V_C^+\setm C}} \prod_{B>C} \phi_B(\pe x{B\cap C} \wedge \pe y{B\setm C}).
\]
Note that we have $\sig_{C_0} = \phi_{C_0} \sig_{C_0}^+$ at the root. 
We have the recursion formula
\begin{eqnarray*}
\sig^+_C(\pe xC) &=& \sum_{\pe y{V_C^+\setm C}} \prod_{C\to B} \left(\phi_{B}(\pe x{B\cap C} \wedge \pe y{B\setm C})\prod_{B'>B} \phi_{B'}(\pe x{B'\cap C} \wedge \pe y{B'\setm C})\right)\\
&=& \prod_{C\to B} \sum_{\pe y{B\cup V_B^+\setm C}}  \phi_{B}(\pe x{B\cap C}
\wedge \pe y{B\setm C})\prod_{B'>B} \phi_{B'}(\pe x{B'\cap C} \wedge
\pe y{B'\setm C})\\
&=& \prod_{C\to B} \sum_{\pe y{B\setm C}} \phi_{B}(\pe x{B\cap C}
\wedge \pe y{B\setm C})\sig_B^+(\pe x{B\cap C}\wedge
\pe y{B\setm C}).
\end{eqnarray*}
The inversion between the sum and product in the second equation
above was possible because the sets $B\cup V_B^+ \setm C$, $C\to B$
are disjoint. Indeed, if there existed 
$B,B'$ such that $C\to B$ and $C\to B'$, and descendants $C'$ of $B'$
and $C''$ of $B''$ with a non-empty intersection, then this
intersection would have to be included in every set in the (non-oriented) path connecting $C'$
and   $C''$ in $\mathbb G$. Since this path contains $C$, the
intersection must also be included in $C$, so that the sets  $B\cup
V_B^+ \setm C$, with $C\to B$
are disjoint. 

Introduce messages
\[
m_B^+(\pe x{ C}) = \sum_{\pe y{B\setm C}} \phi_{B}(\pe x{B\cap C}
\wedge \pe y{B\setm C})\sig_B^+(\pe x{B\cap C}\wedge
\pe y{B\setm C})
\]
where $C$ is the parent of $B$. Then
\[
m_{B}^+(\pe x{C}) = \sum_{\pe y{B\setm C}} \phi_{B}(\pe x{B\cap C}
\wedge \pe y{B\setm C}) \prod_{B\to B'} m_{B'}^+(\pe x{B\cap C}\wedge
\pe y{B\setm C})
\]
with
\[
\sig^+_C(\pe xC) = \prod_{C\to B} m^+_B(\pe x{C})
\]
which provides $\sig_C$ at the root. Reinterpreting this discussion in
terms of the undirected graph, we are led to introducing messages 
$m_{BC}(\pe xC)$ for $B\sim C$ in $\mathbb G$, with the message-passing rule
\begin{equation}
\label{eq:jct.tree.mes}
m_{BC}(\pe x{C}) = \sum_{\pe y{B\setm C}} \phi_{B}(\pe x{B\cap C}
\wedge \pe y{B\setm C}) \prod_{B'\sim B, B'\neq C } m_{B'B}(\pe x{B\cap C}\wedge
\pe y{B\setm C}).
\end{equation}
Messages progressively stabilize when applied in $\mathbb G$, and at
convergence, we have
\begin{equation}
\label{eq:jct.tree.sum}
\sig_C(\pe xC) = \phi_C(\pe xC) \prod_{B\sim C} m_{BC}(\pe x{C}).
\end{equation}

Note that the complexity of the junction tree algorithm is exponential
in the cardinality of the largest $C\in \CS$. This algorithm will
therefore be unfeasible if $\CS$ contains sets that are too large.

\section{Building junction trees}

There is more than one family of set interactions with respect to
which a given probability $\pi$ can be decomposed (notice that, unlike
in the Hammersley-Clifford Theorem, we do not assume that the
interactions are normalized), and not all of them can be organized as
a junction tree. One can however extend any given family into a new one
on which one can build a junction tree.
\begin{definition}
\label{def:clique.ext}
Let $V$ be a set of vertexes, and $\CS_0\sub \twop{V}$. We say that a
set $\CS \sub \twop{V}$ is an extension of $\CS_0$ if, for any $C_0\in
\CS_0$, there exists a $C\in \CS$ such that $C_0\sub C$.

A tree $\mathbb G = (\CS, E)$ is a junction-tree extension of $\CS_0$
if $\CS$ is an extension of $\CS_0$ and $\mathbb G$ is a junction tree.   
\end{definition}

If $\Phi^0 = (\phi^0_C, C\in \CS_0)$ is a consistent family of set
interactions, and $\CS$ is an extension of $\CS_0$, one can build a
new family, $\Phi = (\phi_C, C\in \CS)$, of set interactions which
yields the same probability distribution, i.e., such that, for all
$x\in \CF(V)$,  
\[
\prod_{C\in \CS}\phi_C(\pe xC) \propto \prod_{C_0\in
  \CS_0}\phi_{C_0}^0(\pe x{C_0}).
\]
For this, it suffices to build a mapping say $T: \CS_0 \to \CS$ such
that $C_0 \sub T(C_0)$ for all $C_0\in \CS_0$, which is always
possible since $\CS$ is an extension of $\CS_0$ (for example,
arbitrarily order the elements of $\CS$ and let $T(\CS_0)$ be the
first element of $\CS$, according to this order, that contains
$C_0$).  One can then define 
\[
\phi_C(\pe xC) = \prod_{C_0: T(C_0)=C} \phi_{C_0}^0(\pe x{C_0}).
\] 

Given $\Phi^0$, our goal is to design a junction-tree extension which
is as feasible as possible. So, we are not interested by the trivial
extension $\mathbb G = (V, \emp)$, since the resulting junction-tree
algorithm is unfeasible as soon as $V$ is large.  \Cref{th:jct.tree.exist} in the next section will be the first step in the design of an algorithm that computes
junction trees on a given graph.

\subsection{Triangulated graphs}
\begin{definition}
\label{def:triangul}
Let $G = (V,E)$ be an undirected graph. Let  
$(s_1, s_2, \ldots, s_n)$ be a path in $G$. One says that this path
has a chord at $s_j$,   with $j\in
\{2, \ldots, n\}$ , if $s_{j-1} \sim s_{j+1}$, and we will refer to $(s_{j-1}, s_j, s_{j+1})$ as a chordal triangle. A path in $G$ is
achordal if it has no chord.

One says that  $G$ is triangulated (or chordal) if it has no achordal
loop.
\end{definition}
\begin{definition}
\label{def:decomposable}
The graph $G$ is decomposable if it satisfies the following recursive
condition: it is either complete, or there exists disjoint
subsets $(A,B,C)$ of $V$ such that
\begin{itemize}
\item $V = A \cup B \cup C$,  
\item $A$ and $B$ are not empty,
\item $C$ is clique in $G$, $C$ separates $A$ and $B$, 
\item the restricted
graphs, $G_{A\cup C}$ and $G_{B\cup C}$ are decomposable.
\end{itemize}
\end{definition}
These definitions are in fact equivalent, as stated in the
following proposition.

\begin{proposition}
\label{prop:trg.dec}
An undirected graph is triangulated if and only if it is decomposable
\end{proposition}
\begin{proof}
To prove the ``if'' part, we proceed by induction on $n=|V|$. Note that every graph for $n\leq3$ is both decomposable and triangulated (we leave the verification to the reader).  Assume that the statement ``decomposable $\Rightarrow$ triangulated'' holds for graphs with less than $n$ vertexes, and take $G$ with $n$ vertexes. Assume that $G$ is decomposable. If it is complete, it is obviously triangulated. Otherwise, there exists $A,B,C$ such that  $V = A\cup B \cup C$, with $A$ and $B$ non-empty such
that $G_{A\cup C}$ and $G_{B\cup C}$ are decomposable, hence triangulated from the induction hypothesis,  and such that $C$ is a
clique which separates $A$ and $B$.  Assume that $\gamma$ is an achordal loop in $G$. Since it cannot be included in $A\cup C$ or
$B\cup C$, $\gamma$ must  go from $A$ to $B$ and back, which implies that it
passes at least twice in $C$. Since $C$ is complete, the original loop can be shortcut to form subloops   in $A\cup C$ and $B\cup C$. If one of (or both) these loops has cardinality 3, this would provide $\gamma$ with a chord, which contradicts the assumption. Otherwise, the following lemma also provides a contradiction, since one of the two chords that it implies must also be a chord in the original $\gamma$. 

\begin{lemma}
Let $(s_1, \ldots, s_n, s_{n+1}=s_1)$ be a loop in a triangulated graph, with $n\geq 4$. Then the path has a chord at  two non-contiguous vertexes at least. 
\end{lemma}
To prove the lemma, assume the contrary and let $(s_1, \ldots, s_n, s_{n+1}=s_1)$ be a loop that does not satisfy the condition, with $n$ as small as possible. If $n > 4$, the loop must have a chord, say at $s_j$, and one can remove $s_j$ from the loop to still obtain a smaller loop that must satisfy the condition in the  lemma, since $n$ was as small as possible. One of the two chords must be at a vertex other than the two neighbors of $s_j$, and thus provide a second chord in the original loop, which is a contradiction. Thus $n=4$, but $G$ being triangulated implies that this 4-point loop has a diagonal, so that the condition in the lemma also holds, which provides a contradiction.

For the ``only if'' part of \cref{prop:trg.dec}, assume that $G$ is triangulated. We prove
that the graph is decomposable by induction on $|G|$.  The induction
will work if we can show that, if $G$ is triangulated, it is either
complete or there exists a clique in $G$ such that $V\setm C$ is
disconnected, i.e., there exist two elements $a, b\in V\setm C$ which are
related by no path in $V\setm C$. Indeed, we will then be able to
decompose $V = A\cup B\cup C$, where $A$ and $B$ are unions of
(distinct) connected components of $V\setm C$. Take, for example, $A$ to be the set of vertexes connected to $A$ in $G\setminus C$, and $B = V\setminus (A\cup C)$, which is not empty since it contains $b$. Note that restricted graphs from
triangulated graphs are triangulated too. 

So, assume that $G$ is triangulated, and not complete. Let $C$ be a subset
of $V$ that satisfies the property that $V\setm C$ is disconnected, and take $C$ minimal, so  that $V\setm C'$ is connected for any $C'\sub C$, $C'\neq
C$. We want to show that $C$ is a clique, so take $s$ and $t$ in $C$
and assume that they are not neighbors to reach a contradiction.

Let $A$ and $B$ be two connected components of $V\setm C$. For any
$a\in A$,  $b\in B$, and $s,t\in C$, we know that there exists a path
between $a$ and $b$ in $V\setm C\cup \{s\}$ and another one in $V\setm
C \cup \{t\}$, the first one passing by $s$ (because it would
otherwise connect $a$ and $b$ in $V\setm C$) and the second one
passing by $t$. Any point before $s$ (or $t$) in these paths must
belong to $A$, and any point after them must belong to
$B$. Concatenating these two paths, and removing multiple points if
needed, we obtain a loop passing in $A$, then by $s$, then in $B$,
then by $t$. We can recursively remove all points at which these paths
have a chord. We can also notice that we cannot remove $s$ nor $t$ in this
process, since this would imply an edge between $A$ and $B$, and that
we  must
leave at least one element in $A$ and one in $B$ because removing the
last one would require $s\sim t$. So, at the end, we obtain an achordal loop
with at least four points, which contradicts the fact that $G$ is
triangulated.
\end{proof}

We can now characterize graphs that admit junction trees over the set
of their maximal cliques.
\begin{theorem}
\label{th:jct.tree.exist}
Let $G = (V, E)$ be an undirected graph, and $\CC^*_G$ be the set of all
maximum cliques in $G$. The following two properties are equivalent. 
\begin{itemize}
\item[(i)] There exists a junction tree over $\CC^*_G$.
\item[(ii)] $G$ is triangulated/decomposable.
\end{itemize}
\end{theorem}

\begin{proof}
The proof works by induction on the number of maximal
cliques, $|\CC^*_G|$. If $G$ has only one maximal clique, then $G$ is complete, because any point not included in this clique will have to be included in another maximal clique, which leads to a contradiction. So $G$ is decomposable, and, since any single node obviously provides a junction tree, (i) is true also. 

Now, fix $G$ and assume that the theorem is true
for any graph with fewer maximal cliques.
First assume that $\CC_G^*$ has a junction tree, $\mathbb T$. Let
$C_1$ be a leaf in $\mathbb T$, connected, say, to $C_2$, and let
$\mathbb T_2$ be $\mathbb T$ restricted to $\CC_2 = \CC_G^* \setm
\{C_1\}$. 
Let $V_2$ be the unions of
maximal cliques from nodes in $\mathbb T_2$. A maximal clique $C$ in $G_{V_2}$ is a clique in $G_V$ and therefore
included in some maximal clique $C'\in \CC_V$. If $C'\in \CC_2$, then
$C'$ is also a clique in $G_{V_2}$, and for $C$ to be maximal, we need
$C= C'$. If $C' = C_1$, we note that we must also have
\[
C = \bigcup_{\tilde C\in \CC_2} C\cap \tilde C
\]
and whenever $C\cap \tilde C$ is not empty, this set must be included
in any node in the path in $\mathbb T$ that links $\tilde C$ to
$C_1$. Since this path contains $C_2$, we have $C\cap \tilde C\sub C_2$
so that $C\sub C_2$, but, since $C$ is maximal, this would imply that
$C = C_2 = C_1$ which is impossible. 

This shows that $\CC_{G_2}^* = \CC_2$. This also shows that $\mathbb
T_2$ is a junction tree over $\CC_2$. So, by the
induction hypothesis, $G_{V_2}$ is decomposable. If $s\in V_2 \cap
C_1$, then $s$ also belongs to some clique $C'\in \CC_2$, and
therefore belongs to any clique in the path between $C'$ and $C_1$,
which includes $C_2$. So $s\in C_1\cap C_2$ and $C_1\cap V_2 = C_1
\cap C_2$. So, letting $A = C_1\setm (C_1\cap C_2)$, $B = V_1 \setm
(C_1\cap C_2)$, $S = C_1\cap C_2$, we know that
$G_{A\cup S}$ and $G_{B\cup S}$ are decomposable (the first one being
complete), and that $S$ is a clique. To show that $G$ is decomposable,
it remains to show that $S$ separates $A$ from $B$.  

If a path connects
$A$ to $B$ in $G$, it must contain an edge, say $\{s, t\}$, with
$s\in V\setm S$ and $t\in S$; $\{s, t\}$ must be included in a maximal clique
in $G$. If this clique is $C_1$, we have $s\in C_1\cap V_2 = S$. The same argument shows that this is the only
possibility, because, if $\{s, t\}$ is included in some
maximal clique in $\CC_2$, then we would find $t\in C_1\cap
C_2$. So $S$ separates $A$ and $B$ in $G$.

\medskip

Let us now prove the converse statement, and assume that $G$ is
decomposable. If $G$ is complete, it has only one maximal clique and we
are done. Otherwise, there exists a partition $V = A\cup B\cup S$ such
that $G_{A\cup S}$ and $G_{B\cup S}$ are decomposable, $A$ and $B$ separated by $S$
which is complete. Let $\CC_A^*$ be the maximal cliques in $G_{A\cup S}$ and
$\CC_B^*$ the maximal cliques in $G_{B\cup S}$. By hypothesis, there exist
junction trees $\mathbb T_A$ and $\mathbb T_B$ over $\CC_A^*$ and
$\CC_B^*$. 

Let $C$ be a maximal clique in $G_{A\cup S}$. Assume that $C$  intersect
$A$; $C$ can be extended to a maximal clique, $C'$, in $G$,
but $C'$ cannot intersect $B$ (since this would imply a direct edge
between $A$ and $B$) and is therefore included in $A\cup S$,
so that $C=C'$. Similarly, all maximal cliques in $G_{B\cup S}$ that
intersect $B$ also
are maximal cliques in $G$. 

The clique $S$ is included in some maximal clique $S_A^*\in
\CC_A^*$. From the previous discussion, we have either $S^*_A = S$ or
$S^*_A \in \CC_G^*$. Similarly, $S$ can be extended to a maximal
clique $S^*_B \in \CC_B^*$, with $S^*_B=S$ or $S^*_B\in
\CC_G^*$. Notice also that at least one of $S^*_A$ or $S^*_B$ must be
a maximal clique in $G$: indeed, assume that both sets are equal to $S$,
which, as a clique, can extended to a maximal clique $S^*$ in $G$;
$S^*$ must be included either in $A\cup S$ or in
$B\cup S$, and therefore be a maximal clique in the corresponding
graph which yields $S^* = S$. Reversing the notation if needed, we
will assume that $S_A^* \in \CC_G^*$.

All elements of $\CC_G^*$ must belong either to $\CC_A^*$ or $\CC_B^*$
since any maximal clique, say $C$, in $G$ must be included in either $A\cup
S$ or $B\cup S$, and therefore also provide a maximal
clique in the related graph. So the nodes in $\mathbb T_A$ and
$\mathbb T_B$ enumerate all maximal cliques in $G$, and we can build a
tree $\mathbb T$ over $\CC^*_G$ by identifying $S^*_A$ and $S^*_B$ to $S^*$ and
merging the two trees at this node. To conclude our proof, it only
remains to show that the running intersection property is
satisfied. So consider two nodes $C,C'$ in $\mathbb T$ and take $s\in
C\cap C'$. If the path between these nodes remain in $\CC^*_A$, or in
$\CC^*_B$, then $s$ will belong to any set along that path, since the
running intersection is true on $\mathbb T_A$ and $\mathbb
T_B$. Otherwise, we must have $s\in S$, and the path must contain
$S^*$ to switch trees, and $s$ must still belong to any clique in the
path (applying the running intersection property between the beginning
of the path and $S^*$, and between $S^*$ and the end of the path).
\end{proof}

This theorem delineates a strategy in order to build a junction tree
that is adapted to a given family of local interactions $\Phi =
(\phi_C, C\in C)$. Letting $G$ be the graph induced by these
interactions, i.e., $s\sim_G t$ if and only if there exists $C\in \CC$
such that $\{s,t\}\sub C$, the method  proceeds as follows.
\begin{itemize}
\item[(JT1)] Extend $G$ by adding edges to obtain a triangulated graph $G^*$.
\item[(JT2)] Compute the set $\CC^*$ of maximal cliques in $G^*$,
  which therefore extend $\CC$.
\item[(JT3)] Build a junction tree over $\CC^*$.
\item[(JT4)] Assign interaction $\phi_C$ to a clique $C^*\in\CC^*$
  such that $C\sub C^*$.
\item[(JT5)] Run the junction-tree belief propagation algorithm to
  compute the marginal of $\pi$ (associated to $\Phi$) over each set
  $C^*\in\CC^*$.
\end{itemize}
Steps (JT4) and (JT5) have already been discussed, and we now explain
how the first three steps can be implemented. 

\subsection{Building triangulated graphs}

First consider step (JT1). To triangulate a graph $G = (V, E)$, it
suffices to order its vertexes so that $V = \{s_1, \ldots,
s_n\}$, and then run the following algorithm.

\begin{algorithm}[Graph triangulation] Initialize the algorithm with $k=n$ and
$E_k = E$. Given $E_k$, determine $E_{k-1}$ as follows:
\begin{itemize} 
\item Add an edge to any pair of neighbors of $s_k$ (unless, of
course, they are already linked). 
\item Let $E_{k-1}$ be the new set of
edges.
\end{itemize}
\end{algorithm}

 Then the graph  $G^* = (V, E_0)$ is triangulated. Indeed taking any
 achordal loop, and selecting the vertex with highest
 index in the loop, say $s_k$, brings a contradiction, since the
 neighbors of $s_k$ have been linked when building $E_{k-1}$.

However, the quality of the triangulation, which can be measured by the
number of added edges, or by the size of the maximal cliques, highly
depends on the way vertexes have been numbered. Take the simple
example of the linear graph with three vertexes $A\sim B \sim C$. If
the point of highest index is $B$, then the previous algorithm will
return the three-point loop $A\sim B\sim C\sim A$. Any other ordering
will leave the linear graph, which is already triangulated, invariant.

So, one must be careful about the order with which nodes will be
processed. Finding an optimal ordering for a given global cost is
 an NP-complete problem. However, a very simple modification of
the previous algorithm, which starts with $s_n$ having the minimal
number of neighbors, and at each step defines $s_k$ to be the one with
fewest neighbors that haven't been visited yet, provides an
efficient way for building triangulations. (It has the merit of leaving
$G$ invariant if it is a tree, for example).  Another criterion may be
preferred to the number of neighbors (for example, the number of
new edges that would be needed if $s$ is added).

If $G$ is triangulated, there exists an ordering of $V$ such that the
algorithm above leaves $G$ invariant. We now proceed to a proof of
this statement and also show that such an ordering can be computed
using an algorithm called
maximum cardinality search, which, in addition, allows one to decide
whether a graph is triangulated. We start with a definition that
formalizes the sequence of operations in the triangulation algorithm.
\begin{definition}
\label{def:node.elim}
Let $G = (V, E)$ be an undirected graph. A node elimination consists
in selecting  a vertex $s\in V$ and building the graph $G^{(s)} =
(V^{(s)}, E^{(s)})$ with $V^{(s)} = V \setm \{s\}$, and $E^{(s)}$
containing all pairs $\{t, t'\}\sub V^{(s)}$ such that either $\{t, t'\}
\in E$ or $\{t, t'\}\sub \CV_s$. 

$G^{(s)}$ is called the $s$-elimination graph of $G$.
The set of added edges, namely $E^{(s)} \setm (E\cap E^{(s)})$ is
called the deficiency set of $s$ and denoted $D(s)$ (or $D_G(s)$).
\end{definition}

So, the triangulation algorithm implements a sequence of node
eliminations, successively applied to $s_n, s_{n-1}$, etc. One says
that such an elimination process is {\em perfect} if, for all $k=1,
\ldots, n$,  the deficiency
set of $s_k$ in the graph obtained after elimination of $s_n, \ldots,
s_{k+1}$ is empty (so that no edge is added during the process). We will
also say that $(s_1, \ldots, s_n)$ provides a perfect ordering for
$G$. 
\begin{theorem}
\label{th:perf.order}
An undirected graph $G = (V, E)$ admits a perfect ordering
if and only if it is triangulated.
\end{theorem}
\begin{proof}
The ``only if'' part is obvious, since, the triangulation algorithm following a perfect ordering
  does not add any edge to $G$, which
must therefore have been triangulated to start with.

We now proceed to the ``if'' part. For this it suffices to prove that for any triangulated graph, there exists a vertex $s$ such that $D_G(s) = \emptyset$. One can then easily prove the result by induction, since, after removing this $s$, the remaining graph $G^{(s)}$ is still triangulated and would admit (by induction) a perfect ordering that completes this first step.

To prove that such an $s$ exists, we take a decomposition $V = A \cup S\cup B$, in which $S$ is complete and separates $A$ and $B$,  such that $|A\cup S|$ is minimal (or $|B|$ maximal). We claim that $A\cup S$ must be complete. Otherwise, since $A\cup S$ is still triangulated, There exists a similar decomposition $A\cup S = A'\cup S' \cup B'$. One cannot have $S\cap A'$ and $S\cap B'$ non empty simultaneously, since this would imply a direct edge from $A'$ to $B'$ ($S$ is complete). Say that $S\cap A'=\emptyset$, so that $A'\subset A$. Then the decomposition $V = A'\cup S' \cup (B' \cup B)$ is such that $S'$ separates $A'$ from $B\cup B'$. Indeed, a path from $A'$ to $b\in B\cup B'$ must pass in $S'$ if $b\in B'$, and, if $b
\in B$, it must pass in $S$ (since it links $A$ and $B$). But $S\subset S'\cup B'$  so that the path must intersect  $S'$. We therefore obtain a decomposition that enlarges $B$, which is a contradiction and shows that $A\cup S$ is complete. Given this, any element $s\in A$ can only have neighbors in $A\cup S$ and is therefore such that $D_G(s) = \emptyset$, which concludes the proof.
\end{proof}

If a graph is triangulated, there is more than one perfect ordering of
its vertexes. One of these orderings is provided the {\em maximum
  cardinality search} algorithm, which also allows one to decide
whether the graph is triangulated. We start with a
definition/notation.

\begin{definition}
\label{def:order}
If $G = (V, E)$ is an undirected graph, with $|V| = n$, any ordering $V = (s_1,
\ldots, s_n)$ can be identified with the bijection $\al: V\to \{1,
\ldots, n\}$ defined by $\al(s_k) = k$. In other terms, $\alpha(s)$ is the rank of $s$ in the ordering. 
We will refer to $\al$ as an
ordering, too.

Given an ordering $\al$, we define incremental neighborhoods
$\CV^{\al, k}_s$, for $s\in V$ and $k=1, \ldots, n$ to be the
intersections of $\CV_s$ with the sets $\al^{-1}(\{1, \ldots, k\})$,
i.e., 
\[
\CV^{\al,k}_s = \defset{t\in V, t\sim s, \al(t) \leq k}.
\]

One says that $\al$ satisfies the maximum cardinality property if, for
all $k=\{2, \ldots, n\}$ 
\begin{equation}
\label{eq:max.card}
|\CV^{\al,k-1}_{s_k}| = \max_{\al(s) \geq k} |\CV^{\al,k-1}_{s}|.
\end{equation}
where $s_k = \al^{-1}(k)$.
\end{definition}

Given this, we have the proposition:
\begin{proposition}
\label{prop:max.card}
If $G = (V, E)$ is triangulated, then any ordering that satisfies the
maximum cardinality property is perfect.
\end{proposition}

 \Cref{eq:max.card} immediately provides an algorithm that
constructs an ordering satisfying the maximum cardinality
property given a graph $G$. From  \cref{prop:max.card}, we see that, if for
some $k$, the largest set $\CV^{\al,k-1}_{s_k}$ is not a clique, then $G$ is
not triangulated. We now proceed to the proof of this proposition.
\begin{proof}
Let $G$ be triangulated, and assume that $\al$ is an ordering that
satisfies \cref{eq:max.card}. Assume that $\al$ is not proper in
order to reach a contradiction. 

Let $k$ be the first index for which
$\CV^{\al,k-1}_{s_k}$ is not a clique, so that $s_k$ has two
neighbors, say $t$ and $u$, such that $\al(t)< k$, $\al(u) < k$ and $
t\not\sim u$. Assume that $\al(t) > \al(u)$. Then $t$ must have a
neighbor that is not neighbor of $s$, say $t'$, such that $\al(t') <
\al(t)$ (otherwise, $s$ would have more neighbors than $t$ at order
less than $\al(t)$, which contradicts the maximum cardinality property). The sequence $t', t, s, u$ forms a path that is
such that $\al$ increases from $t'$ to $s$, then decreases from $s$ to
$u$, and contains no chord. Moreover, $t'$ and $u$ cannot be neighbors,
since this would yield an achordal loop and a  contradiction. The proof of 
\cref{prop:max.card} consists in showing that this construction can be
iterated until a contradiction is reached.

More precisely, assume that an achordal path $s_1, \ldots, s_k$ has been obtained, such that
$\al(s)$ is first increasing, then decreasing along the path, and such
that, at extremities one either has $\al(s_1) < \al(s_k) <
\al(s_2)$ or $\al(s_k) <\al(s_1) < \al(s_{k-1})$. In fact, one can
switch between these last two cases by
reordering the path backwards. Both paths $(u,s,t)$ and $(u,s, t, t')$
in the discussion above satisfy this property.

\begin{itemize}
\item Assume, without loss of generality, that $\al(s_1) < \al(s_k) <
\al(s_2)$ and note that, in the considered path, $s_1$ and $s_k$ cannot be
neighbors (for, if $j$ is the last index smaller than $k-1$ such that $s_j$ and $s_k$
are neighbors, then $j$ must also be smaller than $k-2$ and the loop $s_j, \ldots, s_{k-1}, s_k$ would be achordal). 
\item Since
$\al(s_2) > \al(s_k)$, and $s_1$ and $s_2$ are neighbors, $s_k$ must
have a neighbor, say $s'_k$, such
that $s'_k$ is not neighbor of $s_2$ and $\al(s'_k) < \al(s_k)$. 
\item Select the first index $j>2$ such that $s_j \sim s'_{k}$, and consider
the path $(s_1, \ldots, s_j, s'_k)$. This path is achordal, by
construction, and one cannot have $s_1 \sim s'_k$ since this would
create an achordal loop. Let us show that $\al$ first increases and then decreases
along this path. Since $s_2$ is in the path, $\al$ must first
increase, and it suffices to show that $\al(s'_k) < \al(s_j)$. If
$\al$ increases from $s_1$ to $s_j$, then $\al(s_j) > \al(s_2) >
\al(s_k) > \al(s'_k)$. If $\al$ started decreasing at some point
before $s_j$, then $\al(s_j) > \al(s_k) > \al(s'_k)$. 
\item Finally, we need
to show that the $\al$-value at one extremity is between the first two
$\al$-values on the other end of the path. If $\al(s'_k) < \al(s_1)$,
and since we have just seen that $\al(s_j) >\al(s_k) > \al(s_1)$, we
do get $\al(s'_k) < \al(s_1) < \al(s_j)$. If $\al(s'_k) > \al(s_1)$,
then, since by construction $\al(s_2) > \al(s_k) > \al(s'_k)$, we have
$\al(s_2) > \al(s'_k) > \al(s_1)$. 
\item So, we have obtained a new path
that satisfies the same property that the one we started with, but
with a maximum value at end points smaller than the initial one, i.e.,
\[
\max(\al(s_1), \al(s'_k)) < \max(\al(s_1), \al(s_k)).
\]
Since $\al$ takes a finite number of values, this process cannot be
iterated indefinitely, which yields our contradiction.
\end{itemize} 
\end{proof}

\subsection{Computing maximal cliques}
At this point, we know that a graph must be triangulated for its
maximal cliques to admit junction trees, and we have an algorithm to
decide whether a graph is triangulated, and extend it into a
triangulated one if needed. This provides the first step, (JT1), of
our description of the junction tree algorithm. The next step, (JT2),
requires computing a list of maximal cliques. Computing maximal cliques in general graph is an NP complete problem,
for which a large number of algorithms has been developed (see, for
example, \cite{pardalos1994maximum} for a review). For graphs with a
perfect ordering,
however, this problem can always be solved in a polynomial
time. 

Indeed, assume that a perfect ordering is given for $G = (V, E)$, so
that $V = \{s_1, \ldots, s_n\}$ is such that, for all $k$, 
$\CV_{s_k}':=\CV_{s_k}\cap\{s_1, \ldots, s_{k-1}\}$ is a clique. Let $G_k$ be $G$ restricted
to $\{s_1, \ldots, s_k\}$ and $\CC^*_k$ be the set of maximal cliques
in $G_k$. Then the set $C_k := \{s_k\} \cup
\CV_{s_k}'$ is the only maximal clique in $G_k$ that contains $s_k$: it is a clique because  the
ordering is perfect, and any clique that contains $s_k$ must be included in it (because its elements are either $s_k$ or neighbors of $s_k$). It follows from this that the set $\CC^*_k$ can be deduced
from $C^*_{k-1}$ by
\[
\begin{cases}
\CC^*_k = \CC^*_{k-1} \cup \{C_k\} \text{ if } \CV_k'\not\in \CC^*_{k-1}\\  
\CC^*_k = (\CC^*_{k-1} \cup \{C_k\}) \setm \{\CV'_k\} \text{ if } \CV_k'\in
\CC^*_{k-1}
\end{cases} 
\]
This allows one to enumerate all elements in $\CC^*_G=\CC^*_n$,
starting with $\CC^*_1 = \{\{s_1\}\}$.

\subsection{Characterization of junction trees}
We now discuss the last remaining point, (JT3). For this, we need to form
the {\em clique graph} of $G$, which is the undirected  graph $\mathbb
G = (\CC_G^*, \mathbb E)$ defined by
$(C, C') \in \mathbb E$ if and only if $C\cap C'\neq
\emp$. We then have the following fact:
\begin{proposition}
\label{prop:clq.conn}
The clique graph $\mathbb G$ of a connected triangulated undirected
graph $G$  is connected.
\end{proposition}
\begin{proof} 
We proceed by induction, and assume that the result is true if $|V| =
n-1$ (the proposition obviously holds if $|V|=1$).
Assume that a perfect
order on $G$ has been chosen, say $V = \{s_1, \ldots, s_n\}$. Let
$G'$ be $G$ restricted to $\{s_1, \ldots, s_{n-1}\}$,
and $\mathbb G'$ the associated clique graph. Because $\{s_n\} \cup
\CV_{s_n}$ is a clique, any path in $G$ provides a valid path in $G'$
after removing all occurrences of $s_n$ (because any two neighbors of
$s_n$ are linked). The induction hypothesis also implies that $\mathbb
G'$ is connected. 
Since $G$ is connected, $\CV_{s_n}$ is not
empty. Moreover, $C:=\{s_n\}\cup \CV_{s_n}$ must be a maximal clique in
$G$ (since we assume that the order is perfect) and it is the only
maximal clique in $G$ that contains $s_n$ (all other maximal cliques
in $G$ therefore are maximal cliques in $G'$ also). To prove that $\mathbb
G$ is connected, it 
suffices to prove that $C$ is connected to any other maximal clique, $C'$, in
$G$ by a path in $\mathbb G$. If $t\in C$, $t\neq s_n$, there exists a
maximal clique, say $C''$, in $G'$ that contains $t$, and, since
$\mathbb G'$ is connected, there exists a path
$(C_1=C', \ldots, C_q = C'')$
connecting $C'$ to $C''$ in $\mathbb G'$. Let $j$ be the first integer such that $C_j = \CV_{n}$ (take $j={q+1}$ if this never
happens). Then $(C_1, \ldots, C_{j-1}, C)$ is a path linking $C'$ and $C$
in $\mathbb G$.
\end{proof}

We hereafter assume that $G$,
and hence $\mathbb G$, is connected. This is not real loss of
generality because connected components in undirected graphs yields
independent processes that can be handled separately.
We assign weights to edges of the clique graph of $G$ by defining
$w(C, C') = |C\cap C'|$. A
subgraph $\tilde T$ of any given graph $\tilde G$ is called a spanning
tree if $\tilde T$ is a tree with set of vertexes equal to the set of
vertexes of $\tilde G$. 
If 
$\mathbb T = (\CC_G^*, \mathbb E')$  is a spaning tree of $\mathbb G$, we
define the total weight
\[
w(\mathbb T) = \sum_{\{C,C'\}\in\mathbb E'} w(C,C').
\]

We then have the
proposition:
\begin{proposition}\cite{finn1994optimal}
\label{prop:opt.jct}
If $G$ is a connected triangulated graph, the set of junction trees over $\CC_G^*$ coincides with the set of
maximizers of $w(\mathbb T)$ over all spanning trees of $\mathbb G$.
\end{proposition} 
(Notice that $\mathbb G$ being connected implies that spanning trees
over $\mathbb G$ exist.)

Before proving this proposition, we discuss some properties related to
maximal (or maximum-weight) spanning trees over an undirected graph. 
For this discussion, we let $G = (V, E)$ be any undirected graph with
weight $(w(e), e\in E)$. We will then apply these results to a clique
graph when will switch back to the general notation of this section.
Maximal spanning trees can be computed using the so-called Prim's
algorithm \cite{jarnik1930jistem,prim1957shortest,dijkstra1959note}.

\begin{algorithm}[Prim's algorithm]
Initialize the algorithm with a single-node tree $T_1 =
(\{s_1\},  \emp)$, for some arbitrary $s_1\in V$.
Let $T_{k-1} = (V_{k-1}, E_{k-1})$ be the tree obtained at step $k-1$ of the algorithm. If $k\leq n$, the next tree is built as follows.
\begin{enumerate}
\item Let
\[
V_k = \{s_k\} \cup V_{k-1}\  (s_k\not \in V_{k-1}.)
\]
\item Let $E_k = \{e_k\}\cup E_{k-1}$, such that $e_k = \{s_k, s\}$ for some
$s\in V_{k-1}$ satisfying
\begin{equation}
\label{eq:prim}
w(e_k) = \max\Big(w(\{t, t'\}), \{t, t'\}\in E, t\not\in V_{k-1}, t'\in V_{k-1}\Big).
\end{equation}
\end{enumerate}
\end{algorithm}

The ability of this algorithm to always build a maximal spanning tree
is summarized in the following proposition
\cite{gondran1983graphs,mchugh1990algorithm}.
\begin{proposition}
\label{prop:prim.alg}
If $G=(V, E, w)$ is a weighted, connected undirected graph, Prim's algorithm, as
described above, provides a sequence $T_k = (V_k, E_k)$, for $k=1,
\ldots, n$ of subtrees of $G$ such that $V_n=V$ and, for all $k$,
$T_k$ is a maximal spanning tree for the restriction $G_{V_k}$ of $G$
to $V_k$.

Moreover, any maximal spanning tree of $G$, can be realized as $T_n$,
where $(T_1, \ldots, T_n)$ is a sequence provided by Prim's algorithm.
\end{proposition}
\begin{proof}
We first prove that, for all $k$, $T_k$ is a maximal spanning tree on
the graph $G_{V_k}$. 

We will prove a slightly stronger statement, namely, that, for all $k$,
$T_k$ can be extended to form a maximal spanning tree of $G$. This is
stronger, because, if $T_k = (V_k, E_k)$ can be extended to a maximal spanning
tree $T = (V,E)$, and if $T'_k = (V_k, E'_k)$ is a spanning tree for
$G_{V_k}$ such that $w(T_k) < w(T'_k)$, then the graph $T'=(V, E')$ with
\[
E' = (E\setm E_k)\cup E'_k
\]
would be a spanning tree for $G$ with $w(T) < w(T')$, which is
impossible. (To see that $T'$ is a tree, notice that paths in $T'$ are
in one-to-one correspondence with paths in $T$ by replacing any
subpath within $T'_k$ by the unique subpath in $T_k$ that has the same
extremities.)

Clearly, $T_1$, which only has one vertex, can be extended to a maximal
spanning tree. Let $k\geq 1$ be
the last integer for which this property is true for all $j = 1, \dots, k$. 
 If $k=n$, we are done. Otherwise, take a
maximum spanning tree, $T$, that extends $T_{k}$. This tree cannot
contain the new edge added when building $T_{k+1}$, namely
$e_{k+1}=\{s_{k+1}, s\}$ as defined in Prim's algorithm, since it
would otherwise also extend $T_{k+1}$. Consider the path $\ga$ in
$T$ that links $s$ to $s_k$. This path must have an edge $e = \{t, t'\}$
such that $t\in V_{k}$ and $t'\not\in V_{k}$, and by definition of
$e_{k+1}$, we must have
$w(e) \leq w(e_{k+1})$. Notice that $e$ is uniquely defined, because a
path leaving $V_k$ cannot return in this set, since one would be
otherwise able to close it into a loop by inserting the only path in
$T_k$ that connects its extremities.

Replace $e$ by $e_{k+1}$ in $T$. The resulting graph,
say $T'$,
is still a spanning tree for $G$. From any path in $T$, one can create
a path in $T'$ with the same extremities by replacing any
occurrence of the edge, $e$, by the concatenation of the unique path
in $T$ going from $t$ to $s$, followed by $(s,s_{k+1})$, followed by the
unique path in $T$ going from $s_{k+1}$ to $t'$. This implies that $T'$ is
connected. It is also
acyclic, since any loop in $T$ would have to contain $e_{k+1}$ (since
$T$ is acyclic), but there is no
other path than $(s, s_{k+1})$ in $T'$ that links $s$ and $s_k$, because
this path would have to be in $T$, and we have removed the only
possible one from $T$ by deleting the edge $e$.

As a conclusion, $T'$ is an extension of $T_{k+1}$, and a spanning tree with total weight larger or
equal to the one of $T$, and must therefore be optimal, too. But this
contradicts the fact that $T_{k+1}$ cannot be extended to a maximal
tree, so that $k=n$ and the sequence of trees provided by Prim's
algorithm is optimal.

To prove the second statement, let $T$ be an optimal spanning
tree. Let $k$ be the largest integer such that there exists a sequence
$(T_1, \ldots, T_k)$ generated by Prim's algorithm, such that, for all
$j=1, \ldots, k$, $T_j$ is a subtree of $T$. One necessarily has
$j\geq 1$, since $T$ extends any one-vertex tree. If $k=n$, we are
done. Assuming otherwise, let $T_k = (V_k,
E_k)$ and make one more step
of Prim's algorithm, selecting an edge $e_{k+1} = (s_{k+1}, s)$
satisfying \cref{eq:prim}. By assumption, $e_{k+1}$ is not in
$T$. Take as before the unique path linking $s$ and $s_{k+1}$ in $T$
and let $e$ be the unique edge at which this path leaves
$V_k$. Replacing $e$ by $e_{k+1}$ in $T$ provides a new spanning tree, $T'$. One
must have $w(e) \geq w(e_{k+1})$ because $T$ is optimal, and
$w(e_{k+1}) \geq w(e)$ by \cref{eq:prim}. So $w(e) = w(e_{k+1})$, and
one can use $e$ instead of $e_{k+1}$ for the $(k+1)$th step of Prim's
algorithm. But this contradicts the fact that $k$ was the largest
integer in a sequence of subtrees of $T$ that is generated by Prim's
algorithm, and one therefore has $k=n$.
\end{proof}

The proof of  \cref{prop:opt.jct}, that we provide now, uses
very similar ``edge-switching'' arguments.
\begin{proof}[Proof of  \cref{prop:opt.jct}]

Let us start with a maximum weight spanning tree for $\mathbb G$, say $\mathbb T$, and
show that it is a
junction tree. Since $\mathbb T$ has maximum weight, we know that it can be
obtained via Prim's algorithm, and that there exists a sequence
$\mathbb T_1,
\ldots, \mathbb T_n=\mathbb T$ of trees constructed by this
algorithm. Let $\mathbb T_k = (\CC_k, \mathbb E_k)$.  

We proceed by contradiction.
Let $k$ be the largest index such that $\mathbb
T_k$ can
be extended to a junction tree for $\CC_G^*$, and let $\mathbb T'$ be a junction tree
extension of $\mathbb T_{k}$. Assume that $k<n$, and let $e_{k+1} = (C_{k+1}, C')$ be the edge
that has been added when building $\mathbb T_{k+1}$, with $\CC_{k+1} =
\{C_{k+1}\}\cup \CC_{k}$. This edge is not
in $\mathbb T'$, so that there  exists a unique edge $e = (B,B')$ in the path
between $C_k$ and $C'$ in $\mathbb T'$  such that $B\in \CC_{k}$ and
$B'\not\in \CC_{k}$. We must have $w(e) = |B\cap B'| \leq w(e_{k+1}) =
|C_{k+1} \cap C'|$. But, since the running intersection property is true
for $\mathbb T'$, both $B$ and $B'$ must contain $C_{k+1}\cap C'$ so that
$B\cap B' = C_{k+1}\cap C'$. This implies that, if one modifies $\mathbb
T'$ by replacing edge $e$ by edge $e_{k+1}$, yielding a new spanning tree
$\mathbb T''$, the running intersection property is still satisfied in
$\mathbb T'$. Indeed if a vertex $s\in V$ belongs to both extremities
of a path containing $B$ and $B'$ in $\mathbb T'$, then it must belong
to $B\cap B'$, and hence to $C_{k+1}\cap C'$, and therefore to any set in
the path in $\mathbb T'$ that linked $C_{k+1}$ and $C'$.  So we found
a junction tree extension of $\mathbb T_{k+1}$, which contradicts our
assumption that $k$ was the largest. We must therefore have $k=n$ and $\mathbb
T$ is a junction tree.

Let us now consider the converse statement and assume that $\mathbb T$
is a junction tree. Let $k$ be the largest integer such that there
exists a sequence of subgraphs of $\mathbb T$ that is provided by
Prim's algorithm. Denote such a sequence by $(\mathbb T_1, \ldots,
\mathbb T_k)$, with $\mathbb T_j = (\CC_j, \mathbb E_j)$. Assume (to
get a contradiction) that $k<n$, and consider a new step for Prim's
algorithm, adding a new edge $e_{k+1} = \{C_{k+1}, C'\}$ to $\mathbb
T_k$. Take as before the path in $\mathbb T$ linking $C'$ to $C_{k+1}$ in $\mathbb
T$, and select the edge $e$ at which this path leaves $\CC_k$. If $e =
(B, B')$, we must have $w(e) = |B\cap B'| \leq w(e_k) = |C_{k+1} \cap
C'|$, and the running intersection property in $\mathbb T$ implies
that $C_{k+1}\cap C' \sub B\cap B'$, which implies that $w(e)
=w(e_{k+1})$. This implies that adding $e$ instead of $e_{k+1}$ at
step $k+1$ is a valid choice for Prim's algorithm, and contradicts the
fact that $k$ was the largest number of such steps that could provide
a subtree of $\mathbb T$. So $k=n$ and $\mathbb T$ is maximal.
\end{proof}

\chapter{Bayesian Networks}
\label{chap:bayes.net}
\section{Definitions}

Bayesian networks are graphical models supported
by directed acyclic graphs (DAG), which provide them with an ordered
organization (directed graphs were introduced in  \cref{def:dir.grph}). 

We first introduce some notation. Let $G=(V,E)$ be a directed acyclic graph. The parents of $s\in V$ are
vertexes $t$ such that $(t,s)\in E$, and its children are $t$'s such
that $(s,t)\in E$. The set of parents of $s$ is denoted $\pa{s}$, and
the set of its children is $\cl{s}$, with $\CV_s = \cl{s} \cup
\pa{s}$.

Similarly to trees, the vertexes of $G$ can be
partially ordered by $s\leq_G t$ if and only if
there exists a path going from $s$ to $t$. Unlike trees, however,
there can be more than one minimal element in $V$, and we
still call roots vertexes that have no parent, denoting
$$
V_0 = \defset{s\in V: \pa{s} = \emp}.
$$
We also call leaves, or terminal nodes, vertexes that have no children. Unless otherwise
specified, we assume that all graphs are connected.

 Bayesian networks over $G$ are defined as follows. We use the same notation as with Markov random fields to represent the set of configurations $\CF(V)$ that contains collections $x = (x_s, s\in V)$ with $x_s\in F_s$.
\begin{definition}
\label{def:bn}
A random variable $X$ with values in $\CF(V)$ is a Bayesian network over a DAG $G=(V,E)$ if and only
if its distribution can be written in the form
\begin{equation}
\label{eq:bn}
P_X(x) = \prod_{s\in V_0} p_s(\pe xs) \prod_{s\in V\setm V_0} p_s(\pe x{\pa{s}}, \pe x s)
\end{equation}
where $p_s$ is, for all $s\in V$, a probability distribution with
respect to $\pe xs$.
\end{definition}
Using the convention that conditional distributions given the empty set
are just absolute distributions, we can rewrite \cref{eq:bn} as
\begin{equation}
\label{eq:bn.2}
P_X(x) = \prod_{s\in V} p_s(\pe 
x{\pa{s}}, \pe x s).
\end{equation}
One can verify that $\sum_{x\in \Om} P^X(x) =1$. Indeed, when summing over $x$, we can
start summing over all $\pe xs$ with $\cl{s}= \emp$ (the leaves). Such
$\pe xs$'s only appear in the corresponding $p_s$'s, which disappear since they sum
to 1. What remains is the sum of the product over $V$ minus the
leaves, and the argument can be iterated until the remaining sum is 1
(alternatively, work by induction on $|V|$). This fact is also a
consequence of  \cref{prop:rest} below, applied with $A=\emp$.

\section{Conditional independence graph}
\subsection{Moral graph}
Bayesian networks have a conditional independence structure which is
not exactly given by $G$, but can be deduced from it. Indeed, fixing $S\sub V$, we can see, when computing the probability of $\pe XS=\pe xs$
given $\pe X{S^c} = \pe x{S^c}$, which is
\[
\myP(\pe XS = \pe x S \mid \pe X{S^c} = \pe x{S^c} ) = \frac{1}{Z(\pe x{S^c})} \prod_{s\in V} p_s(\pe x{\pa{s}}, \pe xs),
\]
 that the only variables $\pe xt, t\not \in S$ that
can be factorized in the normalizing constant are those that are
neither parent nor children of vertexes in $S$, and do not share a
child with a vertex in $S$ (i.e., they intervene in no
$p_s(\pe x {\pa{s}}, \pe xs)$ that  involve elements of $S$). This suggests the following definition.
\begin{definition}
\label{def:g.sharp}
Let $G$ be a directed acyclic graph. We denote
$G^\sharp=(V,E^\sharp)$ the undirected graph on $V$ such that
$\defset{s,t}\in E^\sharp$ if one of the following conditions is satisfied
\begin{itemize}
\item Either $(s,t)\in E$ or $(t,s)\in E$.
\item There exists $u\in V$ such that $(s,u)\in E$ and $(t,u)\in
E$.
\end{itemize}
\end{definition}
$G^\sharp$ is sometimes called the {\em moral graph} of $G$ (because
it forces parents to marry~!). A path in $G^\sharp$ can be visualized as a path in $G^\flat$ (the undirected graph associated with $G$) which is
allowed to jump between parents of the same vertex even if they were
not connected originally.

The previous discussion implies:
\begin{proposition}
\label{prop:g.sharp}
Let $X$ be a Bayesian network on $G$. We have
$$
\cind{S}{T}{U}_{G^\sharp}\Ria \cind{\pe XS}{\pe XT}{\pe XU},
$$
i.e., $X$ is $G^\sharp$-Markov.
\end{proposition}

This proposition can be refined by noticing that the joint
distribution of $\pe XS$, $\pe XT$ and $\pe XU$ can be deduced from a Bayesian
network on a graph restricted to the ancestors of $S\cup T\cup
U$. \Cref{def:subg} for restricted graphs extends without change to directed graphs, and we repeat it below for convenience.
\begin{definition}
\label{def:rest.grph}
Let $G=(V,E)$ be a graph (directed or undirected), and $A\sub V$. The
restricted graph $G_A = (A, E_A)$ is such that the elements of $E_A$
are the edges $(s,t)$ (or $\defset{s,t}$) in $E$ such that both $s$
and $t$ belong to $A$.
\end{definition}
Moreover, for a directed acyclic graph $G$ and $s\in V$, we define
the set of ancestors of $s$ by
\begin{equation}
\label{eq:anc}
\CA_s = \defset{t\in V, t\leq_G s}
\end{equation}
for the partial order on $V$ induced by $G$. 

If $S\sub V$, we denote $\CA_S = \bigcup_{s\in S} \CA_s$. Note that,
by definition, $S\sub \CA_S$. The following proposition is true.
\begin{proposition}
\label{prop:rest}
Let $X$ be a Bayesian network on $G=(V,E)$ with distribution given by
\cref{eq:bn.2}. Let $S\sub V$ and $A = \CA_S$. Then the distribution
of $\pe XA$ is a Bayesian network over $G_A$ given by
\begin{equation}
\label{eq:bn.rest}
\myP(\pe XA = \pe xA) = \prod_{s\in A} p_s(\pe 
x{\pa{s}}, \pe xs).
\end{equation}
\end{proposition}
There is no ambiguity in the notation $\pa{s}$, since the
parents of $s\in A$ are the same in $G_A$ as in $G$. 
\begin{proof}
One needs to show that
\[
\prod_{s\in A} p_s(\pe x{\pa{s}}, \pe xs) = \sum_{x_{A^c}} \prod_{s\in V} p_s(\pe x{\pa{s}}, \pe xs).
\]
This can be done by induction on the cardinality of $V$. Assume that
the result is true for graphs of size $n$, and let $|V| = n+1$ (the
result is obvious for graphs of size 1).

If $A=V$, there is nothing to prove, so assume that $A^c$ is not
empty. Then $A^c$ must contain a leaf in $G$, since otherwise, $A$
would contain all leaves and their ancestors which would imply that
$A=V$. 

If $s\in A^c$ is a leaf in $G$, one can remove the variable $\pe xs$ from
the sum, since it only appear in $p_s$ and transition probabilities
sum to one. But one can now apply the induction assumption to the
restriction of $G$ to $V\setm \{s\}$. 
\end{proof}

Given  \cref{prop:rest},  \cref{prop:g.sharp} can
therefore be refined as follows.
\begin{proposition}
\label{prop:g.sharp.2}
Let $X$ be a Bayesian network on $G$. We have
$$
\cind{S}{T}{U}_{(G_{\CA_{S\cup T\cup U}})^\sharp}\Ria \cind{\pe XS}{\pe XT}{\pe XU}.
$$
\end{proposition}

 \Cref{prop:rest} is also used in the proof of the following
proposition.
\begin{proposition}
\label{prop:bayes.net.cond}
Let $G = (V,E)$ be a directed acyclic graph, and $X$ be a Bayesian
network over $G$. Then, for all $s\in S$
\[
\myP(\pe Xs = \pe xs \mid \pe X{\mathcal A_s\setm \{s\}} = \pe x{\mathcal A_s\setm \{s\}}) = \myP(\pe Xs = \pe xs \mid \pe X{\pa{s}} = \pe x{s^{-}}) = p_s(\pe x{\pa{s}}, \pe xs).
\]
\end{proposition}
\begin{proof} 
By  \cref{prop:rest}, we can without loss of generality
assume that $V = \mathcal A_s$. Then
\begin{eqnarray*}
\myP(\pe Xs = \pe xs \mid \pe X{\mathcal A_s\setm \{s\}} = \pe x{\mathcal A_s\setm
  \{s\}}) &\propto & \myP(\pe X{\mathcal A_s} = \pe x{\mathcal A_s})\\
&=& p_s(\pe x{\pa{s}}, \pe xs) Z(\pe x{\mathcal A_s\setm
  \{s\}})
\end{eqnarray*}
where
\[
Z(\pe x{\mathcal A_s\setm
  \{s\}}) = \prod_{t\in \mathcal A_s\setm\{s\}} p_t(\pe x{\pa{t}}, \pe xt)
\]
disappears when the conditional probability is normalized.
\end{proof}

\subsection{Reduction to d-separation}
We now want to reformulate 
\cref{prop:g.sharp.2} in terms of the unoriented graph $G^\flat$ and
specific features in $G$ called v-junctions, that we now define.
\begin{definition}
\label{def:v.j}
Let $G=(V,E)$ be a directed graph. A v-junction is a triple of
distinct vertexes,
$(s,t,u)\in V\times V \times V$ such that $\defset{s,u}\sub \pa{t}$
(i.e., $s$ and $u$ are parents of $t$). 
   
We will say that a path $(s_1, \ldots, s_N)$ in $G^\flat$ passes at $s=s_k$ with a
v-junction if $(s_{k-1}, s_k, s_{k+1})$ is a v-junction in $G$.
\end{definition}

We have the lemma:
\begin{lemma}
\label{lem:d.sep}
Two vertexes $s$ and $t$ in $G$ are
separated by a set $U$ in $(G_{\CA_{\{s,t\}\cup U}})^\sharp$ if and
only if any path between $s$ and $t$ in $G^\flat$ must either
\begin{itemize}
\item[(1)] Pass at a vertex in $U$ without a v-junction.
\item[(2)] Pass in  $V \setm \CA_{\{s,t\}\cup U}$ at a v-junction.
\end{itemize}
\end{lemma}
\begin{proof}
\

{\noindent \it Step 1.} We first note that the v-junction clause is
redundant in (2). It can be removed without affecting the
condition. Indeed, if a path in $G^\flat$ passes in $V \setm
\CA_{\{s,t\}\cup U}$ one can follow this path downward (i.e.,
following the orientation in $G$) until a v-junction is met. This has
to happen before reaching the extremities of the path, since $u$ would
be an ancestor of $s$ or $t$ otherwise.  We can therefore work with the weaker
condition (that we will denote (2)')
in the rest of proof.

\bigskip
{\noindent \em Step 2.} 
Assume that $U$ separates $s$ and $t$ in $(G_{\CA_{\{s,t\}\cup
U}})^\sharp$. Take a path $\ga$ between $s$ and $t$ in $G^\flat$. We need to
show that the path satisfies (1) or (2)'. So assume that (2)' is false
(otherwise we are done) so that $\gamma$ is included in
$\CA_{\{s,t\}\cup U}$. We can modify $\ga$ by
removing all the central nodes in v-junctions and still keep a valid path in $(G_{\CA_{\{s,t\}\cup
U}})^\sharp$ (since parents are connected in the moral graph). The remaining path must intersect $U$ by assumption, and
this cannot be at a v-junction in $\ga$ since we have removed them. So
(1) is true.

\bigskip
\noindent {\em Step 3.}
Conversely, assume that (1) or (2) is true for any path in
$G^\flat$. Consider a path $\ga$ in $(G_{\CA_{\{s,t\}\cup U}})^\sharp$
between $s$ and $t$. Any edge in $\ga$ that is not in $G^\flat$ must
involve parents of a common child in $\CA_{\{s,t\}\cup
  U}$. Insert this child between the parents every time this occurs,
resulting in a v-junction added to $\gamma$. Since the added
vertexes are still in $\CA_{\{s,t\}\cup
  U}$, the new path still has no intersection with $V\setm \CA_{\{s,t\}\cup
  U}$ and must therefore satisfy (1). So there must be an intersection
with $U$ without a v-junction, and since the new additions are all
at v-junctions, the intersection must have been originally in $\ga$,
which therefore passes in $U$. This shows that $U$ separates $s$ and
$t$ in $(G_{\CA_{\{s,t\}\cup U}})^\sharp$.
\end{proof}

Condition (2) can be further restricted to provide the notion of
$d$-separation.
\begin{definition}
\label{def:d.sep}
One says that two vertexes $s$ and $t$ in $G$ are
$d$-separated by a set $U$ if and
only if any path between $s$ and $t$ in $G^\flat$ must either
\begin{itemize}
\item[(D1)] Pass at a vertex in $U$ without a v-junction.
\item[(D2)] Pass in  $V \setm \CA_{U}$ with a v-junction.
\end{itemize}
\end{definition}

Then we have:
\begin{theorem}
\label{th:d.sep}
Two vertexes $s$ and $t$ in $G$ are
separated by a set $U$ in $(G_{\CA_{\{s,t\}\cup U}})^\sharp$ if and
only if they are $d$-separated by $U$.
\end{theorem}
\begin{proof}
It suffices to show that if condition ((D1) or (D2)) holds for any path between
$s$ and $t$ in $G^\flat$, then so does  ((1) or (2)).  So take a path between
$s$ and $t$: if (D1) is true for this path, the
conclusion is obvious, since (D1) and (1) are the same. So assume that
(D1) (and therefore (1)) is false and that (D2) is true. Let $u$ be a
vertex in $V\setm \mathcal A_U$ at which $\ga$ passes with a v-junction.

Assume that (2) is false. Then $u$ must be an
ancestor of either $s$ or $t$. Say it is an ancestor of $s$: there is
a path in $G$ going from $u$ to $s$  without passing by $U$ (otherwise
$u$ would be an ancestor of $U$); one can replace the portion
of the old path between $s$ and $u$ by this new one, which does not
pass by $u$ with a v-junction anymore. So the new path still does
not satisfy (D1) and must satisfy (D2). Keep on removing all
intersections with ancestors of $s$ and $t$ that have v-junctions to
finally obtain a path that satisfies neither (D1) or (D2) and a
contradiction to the
fact that $s$ and $t$ are $d$-separated by $U$.
\end{proof}

\subsection{Chain-graph representation}
The $d$-separability property involves both unoriented and oriented
edges. It is in fact a property of the hybrid graph in which the
orientation is removed from the edges that are not involved in a
v-junction, and retained otherwise. Such graphs are particular
instances of chain graphs.

\begin{definition}
\label{def:ch.grph}
A chain graph $G = (V,E, \tilde E)$ is composed with a finite set $V$
of vertexes, a set $E\sub {\boldsymbol \CP}_2(V)$ of unoriented edges and a set
$\tE\sub E\ti E \setm \defset{(t,t), t\in E}$ of oriented edges with
the property that $E \cap \tE^\flat=\emp$, i.e., two vertexes cannot
be linked by both an oriented and an unoriented edge.

A path in a chain graph is a a sequence of vertexes $s_0,
\ldots, s_N$ such that for all $k\geq 1$, $s_{k-1}$ and $s_k$ form an
edge, which means that either $\defset{s_{k-1}, s_k}\in E$ or
$(s_{k-1}, s_k) \in \tE$. 

A chain graph is acyclic if it contains no loop. It is semi-acyclic if
it contains no loop containing oriented edges.
\end{definition}
 
We start with the following equivalence relation within vertexes in a semi-acyclic
chain graph.

\begin{proposition}
Let $G = (V,E, \tilde E)$ be a semi-acyclic chain graph. Define the
relation $s\,\CR\, t$ if and only if
there exists a path in the unoriented subgraph $(V,E)$ that links $s$
and $t$. Then $\CR$ is an equivalence relation.
\end{proposition}

The proposition is obvious. This relation partitions $V$ in
equivalence classes, the set
of which being denoted $V_\CR$. If $S\in V_\CR$, then any pair $s,
t$ in $S$ is related by an unoriented path, and if $S\neq S'\in
V_\CR$, no elements $s\in S$ and $t\in S'$ can be related by such a
path.

Moreover, no path in $G$ between two elements of $S\in
V_\CR$, can contain a directed edge,  since these elements must also be related by an undirected
path, and this would create a loop in $G$ containing an undirected edge. So the restriction of $G$
to $S$ is an undirected graph. 

One can define a directed graph over equivalence classes as follows. Let $G_\CR =
(V_\CR, E_\CR)$ be such that
$(S,S') \in E_\CR$ if and only if there exists $s\in S$ and $t\in
S'$ such that $(s,t)\in\tE$. The graph $G_\CR$ is acyclic: any loop in $G_\CR$
would induce a loop in $G$ containing at least one oriented edge.

We now can formally define a probability distribution on a semi-acyclic chain graph.
\begin{definition}
\label{def:acg}
Let $G = (V, E, \tE)$ be a semi-acyclic chain graph. One says that a random variable
$X$ decomposes on $G$ if and only if: $(\pe XS, S\in V_\CR)$ is
a Bayesian network on $G_\CR$ and the conditional distribution of
$\pe XS$ given $\pe X{S'}, S'\in \pa{S}$ is $G_S$-Markov, such that, for $s\in S$,
$P(\pe Xs = \pe xs\mid \pe Xt, t\in S, X_{S'}, S'\in \pa{S})$ only depends on $\pe xt$
with $\defset{s,t}\in E$ or $(t,s) \in \tilde E$.
\end{definition}

Returning to our discussion on Bayesian networks, we have the
following.  Associate to a DAG $G = (V,E)$ the chain graph
$G^\dagger = (V, E^\dagger, \tE^\dagger)$ defined by: $\defset{s,t}\in
E^\dagger$ if and only if $(s,t)$ or $(t,s)\in E$ and is not involved
in a v-junction, and $(s,t)\in \tE^\dagger$ if $(s,t)\in E$ and is
involved in a v-junction. This graph is acyclic; indeed, take any loop
in $G^\dagger$: when its edges are given their original 
orientations in $E$, the sequence cannot contain a v-junction, since
the orientation in v-junctions are kept in $G^\dagger$; the path therefore
constitutes a loop in $G$ which is a contradiction. 

All, excepted at most one, vertexes in an
equivalence class $S\in G^\dagger_\CR$ have all their parents in
$S$. Indeed, assume that two vertexes,  $s$ and $t$, in $S$ have
parents outside of
$S$. There exists an unoriented path, $s_0=s, s_1, \ldots, s_N = t$,
in $G^\dagger$ connecting them,  since they belong to the same
equivalence class. The edge
at $s$ must be oriented from $s$ to $s_1$ in $G$, since otherwise
$s_1$ would be a second parent to $s$ in $G$, creating a v-junction,
and the edge would have remained
oriented in $G^\dagger$. Similarly, the last edge in the path must be oriented from $t$ to
$s_{N-1}$ in $G$. But this implies that there exists a v-junction in
the original orientation along the path, which cannot be constituted
with only unoriented edges in $G^\dagger$. So we get a contradiction.

Thus,  random variables that decompose on $G^\dagger$ are ``Bayesian
networks'' of acyclic graphs, or trees since we know these are
equivalent. The root of each tree must have multiple (vertex) parents
in the parent tree in $G_\CR$. The following theorem states that
all Bayesian networks are equivalent to such a process.

\begin{theorem}
\label{th:acg}
Let $G = (V,E)$ be a DAG. The random variable $X$ is a Bayesian network on
$G$ if and only if it decomposes over $G^\dagger$.
\end{theorem}
\begin{proof}
Assume that $X$ is a Bayesian network on $G$. We can obviously rewrite the
probability distribution of $X$ in the form
$$
\pi(x) = \prod_{S\in G^\dagger_\CR}\prod_{s\in S} p_s(\pe x{\pa{s}}, \pe xs).
$$
Since every vertex in $S$ has its parents in $S$ or in $\bigcup_{T\in \pa{S}} T$,
this {\em a fortiori} takes the form
$$
\pi(x) = \prod_{S\in G^\dagger_\CR} p_S((\pe xT, T\in S^{-}), \pe xs ).
$$
So $\pe XS, S\in V_\CR$ is a Bayesian network. Moreover, 
$$
p_S((\pe xT, T\in S^{-}), \pe xs ) = \prod_{s\in S} p_s(\pe x{\pa{s}}, \pe xs)
$$
is a tree distribution with the required form of the individual
conditional distributions. 

Now assume that $X$ decomposes on $G^\dagger$. Then the conditional
distribution of $\pe XS$ given $\pe XT, T\in \pa{S}$ is Markov for the acyclic
undirected graph $G_S$, and can therefore be expressed as a tree
distribution consistent with the orientation of $G$. 
\end{proof}

\subsection{Markov equivalence}

While the previous discussion provides a rather simple description of
Bayesian networks in terms of chain graphs, it does not go all the way
in reducing the number of oriented edges in the definition of  a Bayesian
network. The issue is, in some way, addressed by the notion of Markov
equivalence, which is defined as follows.

\begin{definition}
\label{def:mar.eq}
Two directed acyclic graphs on the same set of vertexes $G = (V,E)$
and $\tilde G = (V, \tilde E)$ are Markov-equivalent
if any family of random variables that decomposes as a (positive) Bayesian network
over one of them also decomposes as a Bayesian network over the other.
\end{definition}

The notion of Markov equivalence is exactly described by
d-separation. This is stated in the following theorem, due to Geiger
and Pearl \citep{Geiger1990,geiger1990logic}, that we state without proof.
\begin{theorem}
\label{th:geig.p}
$G$ and $\tilde G$ are Markov equivalent if and only if, whenever two
vertexes are d-separated by a set in one of them, the same separation
is true with the other. 
\end{theorem}

This property can be expressed in a strikingly simple
condition. One says that a v-junction $(s,t,u)$ in a DAG is {\em unlinked}
if $s$ and $u$ are not neighbors.   
\begin{theorem}
\label{th:mar.eq}
$G$ and $\tilde G$ are Markov equivalent if and only if $G^\flat =
\tilde G^\flat$ and $G$ and $\tilde G$ have the same unlinked v-junctions.
\end{theorem}
\begin{proof}

\begin{enumerate}[label={\it Step \arabic*.}]
\item
We first show that
a given pair of vertexes in a DAG is unlinked if and only if it can
be d-separated by some set in the graph. Clearly, if they are linked,
they cannot be d-separated (which is the ``if'' part), so what really needs to be proved is that
unlinked vertexes can be d-separated. Let $s$ and $t$ be these vertexes
and let $U = \CA_{\{s,t\}}\setm \{s,t\}$. Then $U$ d-separates $s$ and
$t$ since any path between $s$ and $t$ in $(G_{\CA_{\{s,t\}\cup
U}})^\sharp = (G_{\CA_{\{s,t\}}})^\sharp$ must obviously pass in $U$.

\item  We now prove the only-if part of 
\cref{th:mar.eq} and
therefore assume that $G$ and $\tilde G$ are Markov equivalent, or, as
stated in  \cref{th:geig.p}, that
d-separation coincides in $G$ and $\tilde G$. We want to prove that
$G^\flat = \tilde G^\flat$ and unlinked v-junctions are the same.

\begin{enumerate}[label={\it Step 2.\arabic*.}]
\item The first statement is obvious from Step 1: d-separation determines
the existence of a link, so if d-separation coincides in the two
graphs, then the same holds for  links and $G^\flat = \tilde G^\flat$. 

\item So let us proceed to the second statement and
let $(s,t,u)$ be an unlinked
v-junction in $G$. We want to show that it is also a v-junction in
$\tilde G$ (obviously unlinked since links coincide). 

We will denote by $\tilde \CA_S$ the ancestors of some set $S\sub V$ in $\tilde G$ (while $\CA_S$
still denotes its ancestors in $G$). Let  $U =
\CA_{\{s,u\}}\setm \{s,u\}$.  Then,
as we have shown in Step 1, $U$ d-separates $s$ and $u$ in $G$,
so that, by assumption it also d-separates  them in $\tilde G$. 

We
know that $t\not \in U$, because it cannot be both a child and an
ancestor of $\{s, u\}$ in $G$ (this would induce a loop). The path
$(s,t,u)$ links $s$ and $u$ and does not pass in $U$, which is
only possible (since $U$ d-separates $s$ and $t$ in $\tilde G$) if it passes in
$V-\tilde \CA_U$ at a v-junction: so $(s,t,u)$ is a v-junction in
$\tilde G$, which is what we wanted to prove.
\end{enumerate}

\item We now consider the converse statement and assume that $G^\flat = \tilde G^\flat$
and unlinked v-junctions coincide. We want to show that d-separation
is the same in $G$ and $\tilde G$. So, we assume that $U$ d-separates $s$ and
$t$ in $G$, and we want to show that the same is true in $\tilde
G$. Thus, what we need to prove is:

{\em Claim 1.} Consider a path $\ga$ between $s$ and $t$ in $\tilde
G^\flat= G^\flat$. Then $\ga$ either  (D1)
passes in $U$ without a v-junction in $\tG$, or (D2) in $V\setm
\tilde\CA_U$ with a v-junction in $\tilde G$. 

We will prove Claim 1 using a series of lemmas. We say that $\ga$ has a three-point loop at
$u$ if $(v, u, w)$ are three consecutive points in $\ga$ such that
$v$ and $w$ are linked. So $(v, u, w, v)$ forms a loop in the
undirected graph.

\begin{lemma}
\label{lem:me.1}
If
$\gamma$ is a path between $s$ and $t$ that does not satisfy (D2) for $G$
and passes in $U$ without three-point loops, then $\ga$ satisfies
(D1) for $\tilde G$. 
\end{lemma}
The proof is easy: since $\ga$ does not satisfy (D2) in $G$,
it satisfies (D1) and passes in $U$ without a v-junction in
$G$. But this intersection cannot be a v-junction in $\tilde G$
since it would otherwise have to be linked and constitute a three-point
loop in $\ga$, which proves that (D1) is true for $\ga$ in $\tG$.

The next step is to remove the three-point loop condition in 
\cref{lem:me.1}. This will be done using the next two results.
\begin{lemma}
\label{lem:me.2}
Let $\ga$ be a path with a three-point loop at
$u\in U$ for $G$. Assume that $\ga\setm u$ (which is a valid path in
$G^\flat$) satisfies (D1) or (D2) in
$\tilde G$. Then $\ga$ satisfies (D1) or (D2) in $\tilde G$.
\end{lemma}
To prove the lemma, let $v$ and $w$ be the predecessor and
successor of $u$ in $\ga$. First assume that $\ga\setm u$ satisfies (D1) in
$\tilde G$. If
this does not happen at $v$ or at $w$, then this will apply also to
$\ga$ and we are done, so let us assume that $v\in U$ and that $(v', v,w)$ is not a
v-junction in $\tilde G$, where $v'$ is the predecessor of $v$.  If $(v',v,u)$ is
not a v-junction in $\tilde G$, then (D1) is true for $\ga$ in $\tilde
G$. If it is
a v-junction, then $(v,u,w)$ is not and (D1) is true too.

Assume now that (D2) is true for $\ga\setm u$ in $\tG$. Again, there
is no problem if (D2) occurs for some point other than $v$ or $w$, so
let us consider the case for which it happens at $v$. This means that 
$v\not\in \tilde \CA_U$ and $(v',v,w)$ is a v-junction. But, since
$u\in U$, the link between $u$ and $v$ must be from $u$ to $v$ in
$\tilde G$ so that
there is no v-junction at $u$ and (D1) is true in $\tilde G$. This proves 
\cref{lem:me.2}.

\begin{lemma}
\label{lem:me.3}
Let $\ga$ be a path with a three-point loop at
$u\in U$ for $G$. Assume that $\ga$ does not satisfy (D2) in $G$. Then
$\ga\setm u$ does not satisfy this property either.
\end{lemma}
Let us assume that $\ga\setm u$ satisfies (D2) and reach a
contradiction. Letting $(v,u,w)$ be the three-point loop, (D2) can only
happen in $\ga\setm u$ at $v$ or $w$, and let us assume that this
happens at $v$, so that, $v'$ being the predecessor of $v$, $(v',v,w)$
is a v-junction in $G$ with $v\not\in \CA_U$. Since $v\not\in \CA_U$,
the link between $u$ and $v$ in $G$ must be from $u$ to $v$, but this
implies that $(v',v,u)$ is a v-junction in $G$ with $v \not \in \CA_U$
which is a contradiction: this proves  \cref{lem:me.3}.

The previous three lemmas directly imply the next one.
\begin{lemma}
\label{lem:me.4}
If
$\gamma$ is a path between $s$ and $t$ that does not satisfy (D2) for
$G$, then $\ga$ satisfies
(D1) or (D2) for $\tilde G$. 
\end{lemma}
Indeed, if we start with $\ga$ that does not satisfy (D2) for $G$, 
 \cref{lem:me.3} allows us to progressively remove three-point loops from
$\ga$ until none remains with a final path that satisfies the
assumptions of  \cref{lem:me.1} and therefore satisfies (D1) in $\tilde G$,
and  \cref{lem:me.2} allows us to add the points that we have
removed in reverse order while always satisfying (D1) or (D2) in $\tG$.

We now partially relax the hypothesis that (D2) is not satisfied with the next lemma.
\begin{lemma}
\label{lem:me.5}
If
$\gamma$ is a path between $s$ and $t$ that does not pass in $V\setm
\CA_U$ at a linked v-junction for $G$, then $\ga$ satisfies
(D1) or (D2) for $\tilde G$. 
\end{lemma}
Assume that $\ga$  does not satisfy (D2) for
$\tilde G$ (otherwise the result is proved). By  \cref{lem:me.4},  $\ga$ must satisfy
(D2) for $G$. So, take an intersection of $\ga$ with $V\setm \CA_U$ that
occurs at a v-junction in $G$, that we will denote $(v, u, w)$. This is
still a v-junction in $\tilde G$ since we assume it to be
unlinked. Since (D2) is false in $\tilde G$, we must have
$u\in \tilde \CA_U$, and there is an oriented path, $\tau$, from
$u$ to $U$ in $\tilde G$. 

We can assume that $\tau$ has no v-junction in $G$. If a v-junction
exists in $\tau$,
then this v-junction must be linked (otherwise this would also be a
v-junction in $\tilde G$ and contradict the fact that $\tau$ is
consistently oriented in $\tilde G$), and this link must be oriented
from $u$ to $U$ in $\tilde G$ to avoid creating a loop in this
graph. This implies that we can bypass the v-junction while keeping a
consistently oriented path in $\tilde G$, and iterate this until
$\tau$ has no v-junction in $G$. But this implies that $\tau$ is
consistently oriented in $G$, necessarily from $U$ to $u$ since
$u \not\in \CA_U$.

Denote $\tau = (u_0=u, v_1, \ldots,
u_n\in U)$. We now prove by induction that each $(v, u_k, w)$ is an unlinked
v-junction. This is true when $k=0$, and let us assume that it is true
for $k-1$. Then $(u_{k}, u_{k-1}, v)$ is a v-junction in $G$ but not
in $\tilde G$: so it must be linked and there exists an edge between
$v$ and $u_k$. In $\tilde G$, this edge must be oriented from $v$
to $u_k$, since $(v, u_{k-1}, u_k, v)$ would form a loop
otherwise. For the same reason, there must be an edge in $\tilde G$
from $w$ to $u_k$ so that $(v, u_k, w)$ is an unlinked
v-junction. 

Since this is true for $k=n$, we can replace $u$ by $u_n$ in $\ga$
and still obtain a valid path. This can be done for all 
intersections of $\ga$ with $V\setm \CA_U$ that occur at
v-junctions. This finally yields a path (denote it
$\bar\ga$) which does not satisfy (D2) in $G$ anymore, and therefore
satisfies (D1) or (D2) in $\tG$: so  $\bar\ga$
must either pass in $U$ without a v-junction or in $V\setm \tilde A_U$
at a v-junction. None of the  nodes that were modified can satisfy any of these
conditions, since they were all in $U$ with a v-junction, so that the
result is true for the original $\ga$ also. This proves 
\cref{lem:me.5}.

So the only unsolved case is when $\ga$ is allowed to pass in $V\setm
\CA_U$ at linked v-junctions. We define an algorithm that removes them
as follows. Let $\ga_0=\ga$ and let $\ga_k$ be the path after step $k$ of
the algorithm. One passes from $\ga_k$ to $\ga_{k+1}$ as follows.
\begin{itemize}
\item If $\ga_k$ has no linked v-junctions in $V\setm \CA_U$ for $G$,
  stop.
\item Otherwise, pick such a v-junction and let $(v,u,w)$ be the
  three nodes involved in it.
\begin{enumerate}[label=(\roman*),labelindent=0.5cm, leftmargin=.5cm]
\item If $v\in U, v'\not\in U$ and $(v', v, u)$ is a v-junction in $\tG$, remove
  $v$ from $\ga_k$ to define $\ga_{k+1}$.
\item Otherwise,  if  $w\in U, w'\not\in U$ and $(u,w, w')$ is a v-junction in $\tG$, remove
  $w$ from $\ga_k$ to define $\ga_{k+1}$.
\item Otherwise, remove $u$ from $\ga_k$ to define $\ga_{k+1}$.
\end{enumerate}
\end{itemize}

None of the considered cases can disconnect the path. This is clear
for case (iii)  since $v$ and $w$ are linked. For case (i), note that,
in $G$, $(v',v, u)$ cannot be a v-junction since $(v,u,w)$ is
one. This implies that the v-junction in $\tG$ must be linked and that
$v'$ and $u$ are connected.

The algorithm will stop at some point with some $\ga_n$ that does not
have any linked v-junction in $V\setm \CA_U$ anymore, which implies
that (D1) or (D2) is true in $\tG$ for $\ga_n$. To prove that this
statement holds for $\ga$, it suffices to show that if (D1) or (D2) is true
in $\tG$ with
$\ga_{k+1}$, it must have been true with $\ga_k$ at each step of the
algorithm. So let's assume that $\ga_{k+1}$ satisfies (D1) or (D2) in $\tG$. 

First assume
that we passed from $\ga_k$ to $\ga_{k+1}$ via case (iii). Assume that
(D2) is true for $\ga_{k+1}$, with as usual the only interesting case
being when this occurs at $v$ or $w$. Assume it occurs at $v$ so that
$(v',v,w)$ is a v-junction and $v\not\in \tilde\CA_U$. If $(v',v,u)$
is a v-junction, then (D2) is true with $\ga_k$. Otherwise, there is an
edge from $v$ to $u$ in $\tG$ which also implies an edge from $w$ to
$u$ since $(v,u,w,v)$ would be a loop otherwise.  So $(v,u,w)$ is a
v-junction in $\tG$, and $u$ cannot be in $\tilde\CA_U$ since its
parent, $v$ would be in that set also. So (D2) is true in $\tG$. Now,
assume that 
(D1) is true at $v$, so that $(v',v,w)$ is not a v-junction and $v\in
U$. If $(v',v,u)$ is not a v-junction either, we are done, so assume
the contrary. If $v'\in U$, then we cannot have a v-junction at $v'$
and (D1) is true. But $v'\not\in U$ is not possible since this leads
to case (i).

Now assume that we passed from $\ga_{k}$ to $\ga_{k+1}$ via case
(i). Assume that (D1) is true for $\ga_k$: this cannot be at $v'$
since $v'\not\in U$, neither at $u$ since $u\not\in \CA_U$, so it will
also be true for $\ga_{k+1}$. The same statement holds with (D2) since
$(v',v,u)$ is a v-junction in $\tG$ with $v\in U$ which implies that
both $v'$ and $u$ are in $\tilde\CA_U$. Case (ii) is obviously
addressed similarly.
\end{enumerate}
With this,  the proof
of  \cref{th:mar.eq} is complete.
\end{proof}

\subsection{Probabilistic inference: Sum-prod algorithm}

We now discuss the issue of using the sum-prod algorithm to compute marginal probabilities,
$\myP(\pe Xs = \pe xs)$ for $s\in V$ when $X$ is a Bayesian network on $G =
(V,E)$. By definition, $\myP(X = x)$ can be written in the form
$$
\myP(X=x) = \prod_{C\in\CC} \phi_C(\pe xC)
$$
where $\CC$ contains all subsets $C_s := \defset{s} \cup \pa{s}$, $s\in
V$. Marginal probabilities can therefore be computed easily when the
factor graph associated to $\CC$ is acyclic, according to 
\cref{prop:fct.grph.cvg}.  However, because of the specific form of the $\phi_C$'s (they are
conditional probabilities), the sum-prod algorithm can be analyzed
in more detail, and provide correct results even when the factor graph
is not acyclic.

The general rules for the sum-prod algorithm are
\[
\left\{
\begin{aligned}
&m_{sC}(\pe xs) \leftarrow \prod_{\tilde C, s\in \tilde C,  \tilde C\neq C} m_{\tilde C s}(\pe xs)\\
&m_{Cs}(\pe xs) \leftarrow \sum_{\pe y{C}: \pe ys = \pe xs} \phi_C(\pe y{C}) \prod_{t\in C\setm \{s\}} m_{tC}(\pe yt)
\end{aligned}
\right.
\]

They take a particular form for Bayesian networks, using the fact that a vertex $s$ belongs to $C_s$, and to
all $C_t$ for $t\in \cl{s}$.
\begin{eqnarray*}
m_{sC_s}(\pe xs) &\leftarrow& \prod_{t\in \cl{s}} m_{C_ts}(\pe xs),\\
m_{sC_t}(\pe xs) &\leftarrow& m_{C_ss}(\pe xs) \prod_{u\in \cl{s}, u\neq t}
m_{C_us}(\pe xs), \text{ for } t\in \cl{s},\\
&&\\
m_{C_ss}(\pe xs) &\leftarrow& \sum_{\pe y{C_s}, \pe ys = \pe xs} p_s(\pe y{\pa{s}}, \pe xs) \prod_{t\in
\pa{s}} m_{tC_s}(\pe yt),\\
m_{C_ts}(\pe xs) &\leftarrow& \sum_{\pe y{C_t}, \pe ys = \pe xs} p_t(\pe xs\wedge \pe y{\pa{s}\setminus\{t\}}, \pe yt) m_{tC_t}(\pe yt) \prod_{u\in
\pa{s}, u\neq t} m_{uC_t}(\pe yu), 
\\ && \text{ for } t\in \cl{s}.
\end{eqnarray*}

These relations imply that, if $\pa{s}=\emp$ ($s$ is a root), then
$m_{C_ss} = p_s(\pe xs)$. Also, if $\cl{s} = \emp$ ($s$ is a leaf) then
$m_{sC_s} = 1$. The following proposition shows that many of the
messages become constant over time.
\begin{proposition}
\label{prop:bp.bay.1}
All upward messages, $m_{sC_s}$ and $m_{C_ts}$ with $t\in \cl{s}$ become
constant (independent from $\pe xs$) in finite time.
\end{proposition}
\begin{proof}
This can be shown recursively as follows. Assume that, for a given
$s$,  $m_{tC_t}$ is
constant for all $t\in \cl{s}$ (this is true if $s$ is a leaf). Then,
\begin{eqnarray*}
m_{C_ts}(\pe xs) &\leftarrow& \sum_{pe y{C_t}, \pe ys = \pe xs} 
p_t(\pe xs\wedge \pe y{\pa{s}\setminus\{t\}}, \pe yt) m_{tC_t}(\pe yt) \prod_{u\in
\pa{s}, u\neq t} m_{uC_t}(\pe yu), \\
&=& m_{tC_t} \sum_{\pe y{C_t}, \pe ys = \pe xs} p_t(\pe xs\wedge \pe y{\pa{s}\setminus\{t\}}, \pe yt) \prod_{u\in
\pa{s}, u\neq t} m_{uC_t}(\pe yu)\\
&=& m_{tC_t} \sum_{\pe y{C_t\setm \{t\}}, \pe ys = \pe xs} \prod_{u\in
\pa{s}, u\neq t} m_{uC_t}(\pe yu)\\
&=& m_{tC_t} \prod_{u\in \pa{s}, u\neq t} \sum_{\pe yu} m_{uC_t}(\pe yu)
\end{eqnarray*}
which is constant. Now
$$
m_{sC_s}(\pe xs) \leftarrow \prod_{t\in \cl{s}} m_{C_ts}(\pe xs)
$$
is also constant. This proves that all $m_{sC_s}$ progressively become
constant, and, as we have just seen, this implies the same property
for $m_{C_ts}$, $t\in \cl{s}$.
\end{proof}

This proposition implies that, if initialized with constant messages
(or after a finite time), the sum-prod algorithm iterates
\begin{eqnarray*}
m_{sC_s} &\leftarrow& \prod_{t\in \cl{s}} m_{C_ts}\\
m_{C_ss}(\pe xs) &\leftarrow& \sum_{\pe y{C_s}, \pe ys = \pe xs} p_s(\pe y_{\pa{s}}, \pe xs) \prod_{t\in
\pa{s}} m_{tC_s}(\pe yt)\\
m_{sC_t}(\pe xs)  &\leftarrow& m_{C_ss}(\pe xs) \prod_{u\in \cl{s}, u\neq t}
m_{C_us}, \  \ t\in \cl{s}\\
m_{C_ts} &\leftarrow& m_{sC_s} \prod_{u\in \pa{t}, u\neq s}
\sum_{\pe yu} m_{uC_s}(\pe yu), \  \ t\in \cl{s}.
\end{eqnarray*}
From this expression, we can conclude
\begin{proposition}
\label{prop:bp.bay.2}
If the previous algorithm is first initialized with upward messages,
$m_{sC_s} = m_{C_ts}$ all equal to 1, and if  downward messages are
computed top down from the roots to the leaves, the obtained
configuration of messages is invariant for the sum-prod algorithm.
\end{proposition}
\begin{proof}
If all upward messages are equal to 1, then clearly, the downward
messages sum to 1 once they are updated from roots to leaves, and this
implies that the upward messages will remain equal to 1 for the next
round.  The obtained configuration is invariant since the downward
messages are recursively uniquely defined by their value at the roots.
\end{proof}
The downward messages, under the
previous assumptions, satisfy $m_{sC_t}(\pe xs)  = m_{C_ss}(\pe xs)$ for all
$t\in \cl{s}$ and therefore
\begin{equation}
\label{eq:mes.bay}
m_{C_ss}(\pe xs) = \sum_{\pe y{C_s}, \pe ys = \pe xs} \pi(\pe y{\pa{s}}, \pe xs) \prod_{t\in
\pa{s}} m_{C_tt}(\pe yt).
\end{equation}
Note that the associated ``marginals'' inferred by the sum-prod algorithm are 
$$
\sig_s(\pe xs) = \prod_{C, s\in C} m_{Cs}(\pe xs) = m_{C_ss}(\pe xs) 
$$
since $m_{C_ts}(\pe xs) = 1$ when $t\in \cl{s}$.

Although the sum-prod algorithm initialized with unit messages
converges to a stable configuration if run top-down, the obtained
$\sig_s$'s do not necessarily provide the correct single site marginals. There is a
situation for which this is true, however, which is when the initial
directed graph is singly connected, as we will see below. Before this,
let us analyze the complexity resulting from an iterative computation of
the marginal probabilities, similar to what we have done with trees.

We define the
depth of a vertex in $G$ as follows.
\begin{definition}
\label{def:depth}
Let $G=(V,E)$ be a DAG. The depth of a vertex $s$ in $V$ is defined
recursively by
\begin{itemize}[labelindent=0.5cm]
\item[-] $\mathrm{depth}(s) = 0$ if $s$ has no parent.
\item[-] $\mathrm{depth}(s) = 1 + \max\big(\mathrm{depth}(t), t\in
\pa{s}\big)$ otherwise.
\end{itemize}
\end{definition}

The recursive computation of marginal distributions is made possible
(although not always feasible)
with the following remark.
\begin{lemma}
\label{lem:c.i.dag}
Let $X$ be a Bayesian network on the DAG $G=(V,E)$, and $S\sub V$,
such that all elements in $S$ have the same depth. Let
$\pa{S}$ be the set of parents of elements in $S$, and $T =
\mathrm{depth}^-(S)$ the set of vertexes in $V$ with depth strictly smaller
than the depth of $S$. Then 
$\cind{\pe XS}{\pe X{T\setm \pa{S}}}{\pe X{\pa{S}}}$ and the variables $\pe Xs, s\in S$
are conditionally independent given $\pe X{\pa{S}}$. 
\end{lemma}
\begin{proof}
It suffices to show that vertexes in $S$ are separated from  $T\setm
\pa{S}$ and from other elements of $S$ by $\pa{S}$ for the
graph $(G_{S\cup T})^\sharp$. Any path starting at $s\in S$ must either
pass by a parent of $s$ (which is what we want), or by one of its children, or by another
vertex that shares a child with $s$ in $G_{S\cup T}$. But $s$ cannot have any
child in $G_{S\cup T}$, since this child cannot have a smaller depth
than $s$,
and it cannot be in $S$ either since all elements in $S$ have the same
depth.
\end{proof}
This lemma allows us to work recursively as follows. Assume that we can
compute marginal distributions over sets $S$ with maximal depth no
larger than $d$. Take a set $S$ of maximal depth $d+1$, and let $S_0$ be the
set of elements of depth $d+1$ in $S$. Then, letting
$T=\mathrm{depth}^-(S) = \mathrm{depth}^-(S_0)$, and $S_1 = S\setm S_0$,
\begin{eqnarray}
\nonumber
\myP(\pe XS = \pe xS) &=& \sum_{\pe y{T\setm S_1}} \myP(\pe X{S_0} = \pe x{S_0} \mid \pe XT = \pe y{T\setm S_1}\wedge
\pe x{S_1}) P(\pe X{T\cup S_1} =  \pe y{T\setm S_1}\wedge \pe x{S_1})\\
\label{eq:rec.inf}
&=& \sum_{\pe y{\pa{S}\setm S_1}} \prod_{s\in S_0} p_s(\pe {(y\wedge x)}{\pa{s}}\wedge
\pe x{S_1}, \pe xs) \myP(\pe X{\pa{S_0}\cup S_1} = \pe  y{\pa{S_0}\setm S_1}\wedge \pe x{S_1})
\end{eqnarray}
Since $\pa{S}\cup S_1$ has maximal depth strictly smaller than the
maximal depth of $S$, this indeed provides a recursive formula for the
computation of marginal over subsets of $V$ with increasing maximal
depths. However, because one needs to add parents to the
considered set when reducing the depth, one may end up having to
compute marginals over very large sets, which becomes intractable
without further assumptions.

A way to reduce the complexity is to assume that the graph $G$ is
singly connected, as defined below.

\begin{definition}
\label{def:sin.com}
A DAG $G$ is singly connected if there exists at most one path in $G$ that connects
any two vertexes. 
\end{definition}
Such a property is true for a tree, but also
holds for some networks with multiple parents. We have the
following nice property in this case.

\begin{proposition}
\label{prop:sin.con}
Let $G$ be a singly connected DAG and $X$ a Bayesian network on
$G$. If $s$ is a vertex in $G$, the variables $(\pe Xt, t\in \pa{s})$ are
mutually independent.
\end{proposition}
\begin{proof}
We have, using  \cref{prop:rest},
$$
\myP(\pe X{\pa{s}} = \pe x{\pa{s}})  = \sum_{\pe y{\CA_{\pa{s}}}, \pe y{\pa{s}} = \pe x{\pa{s}}}
\prod_{u\in \CA_{\pa{s}}} p_u(\pe y{\pa{u}}, \pe yu).
$$
Because the graph is singly connected, two parents of $s$ cannot have a
common ancestor (since there would then be two paths from this
ancestor to 
$S$). So $\CA_{\pa{s}}$ is the disjoint union of the $\CA_{t}$'s for $t\in
\pa{s}$ and we can write
\begin{eqnarray*}
\myP(\pe X{\pa{s}} = \pe x{\pa{s}})  &=& \sum_{\pe y{\CA_{\pa{s}}}, \pe y{\pa{s}} = \pe x{\pa{s}}}
 \prod_{t\in \pa{s}}  \prod_{u\in\CA_t} p_u(\pe y{\pa{u}}, \pe yu)\\
&=& \prod_{t\in \pa{s}} \sum_{\pe y{\CA_{t}}, \pe y{t} = \pe x{t}}
   \prod_{u\in\CA_t} p_u(\pe y{\pa{u}}, \pe yu)\\
&=& \prod_{t\in \pa{s}} \myP(\pe Xt = \pe xt)
\end{eqnarray*}
This proves the lemma.
\end{proof}

 \Cref{eq:rec.inf} can be simplified under the assumption of
a singly connected graph, at least for the computation of single
vertex marginals; we have, if $s\in V$ and $G$ is singly connected
\begin{equation}
\label{eq:sing.con}
\myP(\pe Xs = \pe xs) = \sum_{\pe y{\pa{s}}}  p_s(\pe y{\pa{s}}, \pe xs) \prod_{t\in \pa{s}}
\myP(\pe X{t} = \pe yt).
\end{equation}
This is now recursive in single vertex marginal probabilities. It
moreover coincides with the recursive equation that defines the
messages $m_{C_ss}$ in \cref{eq:mes.bay}, which shows that the
sum-prod algorithm provides the correct answer in this case.

\subsection{Conditional probabilities and interventions}
One of the main interests of graphical models is to provide an
ability to infer the behavior of hidden variables of interest given
other, observed, variables. When dealing with oriented graphs the
way this should be analyzed is, however, ambiguous. 

Let's consider an example, provided by the graph in  \cref{fig:snow}.
\begin{figure}[h]
\centering
		\begin{tikzpicture}[mynode/.style={ellipse, draw=black!60, fill=gray!5, ultra thick, , minimum size=10mm},]
				\node[mynode](H1){No school};
				\node[mynode](H2)[above left =1.5cm and -.5cm of H1]{Bad weather};
				\node[mynode](H3)[above right=1.5cm and -.5cm of H1]{Broken HVAC};

		\draw[black, ultra thick,->] (H2.south) -- (H1.north);
		\draw[black, ultra thick,->] (H3.south) -- (H1.north);
		\end{tikzpicture}
\caption{\label{fig:snow} Example of causal graph.}
\end{figure}
The Bayesian network interpretation of this graph is that both
events (which may be true or false) ``Bad weather'' and ``Broken HVAC'' happen first, and that they are
independent. Then, given their observation, the ``No school'' event may
occur, probably more likely if the weather is bad or the HVAC is broken or snow, and even more likely if both happened at the same time.

Now consider the following passive observation: you wake up, you
haven't checked the weather yet or the news yet, and someone tells you that there is no
school today. Then you may infer that there is more chances than usual
for bad weather or the HVAC broken at school. Conditionally to this information, these two events become
correlated, even if they were initially independent. So, even if the
``No school'' event is considered as a probabilistic consequence of its
parents, observing it influences our knowledge on them.

Now, here is an intervention, or manipulation: the school superintendent has declared
that he has given enough snow days for the year and declared that there
would be school today whatever happens. So you know that the
``no-school'' event will not happen. Does it change the risk of bad weather of broken HVAC? Obviously not: an intervention on a node does not affect the
distribution of the parents. 

Manipulation and passive observation are two very different ways of
affecting unobserved variables in Bayesian networks. Both of them may
be relevant in applications. Of the two, the simplest to analyze is
intervention, since it merely consists in clamping one of the
variables while letting the rest of the network dynamics
unchanged. This leads to the following formal definition of
manipulation.
\begin{definition}
\label{def:manip}
Let $G = (V,E)$ be a directed acyclic graph and $X$ a Bayesian network
on $G$. Let $S$ be a subset of $G$
and $\pe xS\in F_S$ a given configuration on $S$. Then the manipulated distribution
of $X$ with fixed values $\pe xS$ on $S$ is the Bayesian network on the
restricted graph $G_S$, with the same conditional probabilities, using
the value $\pe xs$ every time a vertex $s\in S$ is a parent of $t\in
V\setm S$ in $G$.
\end{definition}
So, if the distribution of $X$ is given by \cref{eq:bn.2}, then its
distribution after manipulation on $S$ is 
$$
\tilde \pi(\pe y{V\setm S}) = \prod_{t\in V\setm S} p_t(\pe y{\pa{t}}, \pe yt)
$$
where $\pa{t}$ is the set of parents of $t$ in $G$, and $\pe ys=\pe xs$
whenever $s\in \pa{t}\cap S$.

The distribution of a Bayesian network $X$ after passive observation
$\pe XS = \pe xS$ is
not so easily described. It is obviously the conditional distribution
$P(\pe X{V\setm S} = \pe y{V\setm S}\mid \pe XS = \pe xS)$ and therefore requires using
the conditional dependency structure, involving the moral
graph and/or d-separation.

Let us discuss this first in the simpler case of trees, for which the
moral graph is the undirected acyclic graph underlying the tree, and
d-separation is simple separation on this acyclic graph. We can then
use  \cref{prop:cond.dist.grph} to understand the new
structure after conditioning: it is a $G^\flat_{V\setm S}$-Markov
random field, and, for $t\in V\setm S$, the conditional distribution
of $\pe Xt = \pe yt$ given its neighbors is the same as before, using the
value $\pe xs$ when $s\in S$. But note that when doing this (passing to $G^\flat$), we broke the
causality relation between the variables. We can however always go
back to a tree (or forest, since connectedness may have been broken) with
the same edge orientation as they initially were, but this requires
reconstituting the edge joint probabilities from the new acyclic
graph, and therefore using (acyclic) belief propagation.

With general Bayesian networks, we know that the moral graph can be
loopy and therefore a source of  difficulties. The following
proposition states that the damage is circumscribed to the ancestors of
$S$. 

\begin{proposition}
\label{prop:inf.cond}
Let $G=(V,E)$ be a directed acyclic graph, $X$ a Bayesian network on
$G$, $S\sub V$ and $\pe x{\CA_S}\in \CF(\CA_S)$. Then the conditional
distribution of $\pe X{\CA_S^c}$ given by $\pe X{\CA_S} = \pe x{\CA_S}$
coincides with the manipulated distribution in 
\cref{def:manip}.
\end{proposition}
\begin{proof} 
The conditional distribution is proportional to 
$$
\prod_{s\in V} p(\pe y{\pa{s}}, \pe ys)
$$
with $\pe yt = \pe xt$ if $t\in \CA_S$. Since $s\in\CA_S$ implies $\pa{s}\sub
\CA_S$, all terms with $s\in \CA_S$ are constant in the sum and can be
factored out after normalization. So the conditional distribution is
proportional to 
$$
\prod_{s\in \CA_S^c} p(\pe y{\pa{s}}, \pe ys)
$$
with $\pe yt = \pe xt$ if $t\in \CA_S$. But we know that such products sum
to 1, so that the conditional distribution is equal to this expression
and therefore provides a Bayesian network on $G_{\CA_S^c}$.
\end{proof}

\section{Structural equation models}
\label{sec:sem}

Structural equation models (SEM's) provides an alternative (and essentially equivalent) formulation of Bayesian networks, which may be more convenient to use, especially when dealing with variables taking values in general state spaces.

Let $G=(V, E)$ be a directed acyclic graph.  SEMs are associated to families of functions $\Phi_s: \CF(\pa{s}) \times \CB_s\to F_s$ and random variables $\bfxi_s: \Omega \to \CB_s$ (where $\CB_s$ is some measurable set), for $s\in V$. The random field $X: \Omega \to \CF(V)$ associated to the SEM satisfies the equations
\begin{equation}
\label{eq:sem}
X_s =\pe \Phi s(\pe X{s-}, \pe \bfxi s).
\end{equation}
Because of the DAG structure, these equations uniquely define $X$ once $\bfxi$ is specified. As a consequence, there exists a function $\boldsymbol\Psi$ such that $X = \boldsymbol \Psi(\bfxi)$.

The model is therefore fully specified by the functions $\pe \Phi s$ and the probability distributions of the variables $\pe \bfxi s$. We will assume that they have a density, denoted $\pe g s, s\in V$, with respect to some measure $\mu_s$ on $\CB_s$. 
They are typically chosen as uniform distributions on $\CB_s$ (continuous and compact, or discrete) or as standard Gaussian when $\CB_s = \mR^{d_s}$ for some $d_s$. One also generally assumes that the variables $(\pe \bfxi s, s\in V)$ are jointly independent, and we make this assumption below.


Let $V_k$, $k\geq 0$, be the set of vertexes in $V$ with depth $k$ (c.f. \cref{def:depth}) and $V_{<k} = V_0 \cup \cdots\cup V_{k-1}$. Then (using the independence of $(\pe\bfxi s, s\in V)$, for $s\in V_k$,  the conditional distribution of $\pe Xs$ given $\pe X{V_{<k}} = \pe x{V_{<k}}$ is the distribution  of $\pe\Phi s(\pe x{s-}, \pe\bfxi s)$. Formally this is given by 
\[
\pe\Phi s(\pe x{s-}, \cdot)_{\sharp} (\pe gs\mu_s),
\]
the pushforward of the distribution of $\pe \bfxi s$ by $\pe\Phi s(\pe x{s-}, \cdot)$.

More concretely, assume that $\bfxi_s$ follows a uniform distribution on $\CB_s = [0, 1]^h$ for some $h$, and assume that $F_s$ is finite for all $s$. Then,
\[
P(\pe Xs = \pe xs \mid \pe X{V_{<k}} = \pe x{V_{<k}}) = \mathrm{Volume}(U_s(  \pe x{\pa{s}}, \pe xs)) \defeq p_s( \pe x{\pa{s}}, \pe xs)
\]
where 
\[
U_s(\pe x{\pa{s}}, \pe xs) = \left\{\xi\in [0,1]^h :  \pe\Phi s(\pe x{s-}, \xi) = \pe xs\right\}.
\] 
Since variables $\pe Xs$, $s\in V_k$ are conditionally independent given $\pe X {V<k}$, we find that $X$ decomposes as a Bayesian network over $G$, 
\[
P(X=x) = \prod_{s\in V}  p_s( \pe x{\pa{s}}, \pe xs).
\]
Similarly, if $F_s = \CB_s = \mR^{d_s}$, $\pe \xi s\sim\CN(0, \Id[d_s])$, and $\pe \bfxi s \mapsto \pe{\Phi_\theta}s(\pe x{\pa{s}}, \pe \xi s)$ is invertible, with $C^1$ inverse $\pe xs \mapsto \pe{\Psi_\theta} s(\pe x{\pa{s}}, \pe xs)$, then $X$ is a Bayesian network, with continuous variables, and, using the change of variable formula, the conditional distribution of $\pe Xs$ given $\pe X{\pa{s}} = \pe x{s-}$ has  p.d.f.
\[
 p_s( \pe x{\pa{s}}, \pe xs) = \frac{1}{(2\pi)^{d_s/2}} \exp\left(-\frac12 |x_s - 
\pe{\Psi_\theta}s(\pe x{\pa{s}}, \pe xs)|^2\right) \left|\det (\partial_{\pe xs}\pe{\Psi_\theta}s(\pe x{\pa{s}}, \pe xs))\right|.
\]
A simple and commonly used special case for this example are linear SEMs, with
\[
\pe Xs = a_s + b_s^T\pe Xs + \sigma_s \pe\xi s.
\] 
In this case, the inverse mapping is immediate and the Jacobian determinant in the change of variables is $1/\sigma_s^{d_s}$. 

\chapter{Latent Variables and Variational Methods}
\label{chap:var.bayes}

\section{Introduction}

We will describe, in the next chapters, methods that fit a parametric model to the observation while introducing  unobserved, or ``latent,'' components in their models. Such latent components typically attach  interpretable information or structure  to the data. We have seen one such example in the form of the mixture of Gaussian in \cref{chap:intro}, that we will revisit in \cref{chap:clustering}. We now provide a  presentation of the variational Bayes paradigm that provides a general strategy to address latent variable problems \citep{neal1998view,jaakkola1997variational,
attias1999variational,jordan1999introduction}. 

The general framework is as follows. Variables in the model  are divided in two groups: the observable part, that we denote  $X$, and the latent part, denoted $Z$. In many models, $Z$ represents some unobservable structure, such that $X$ conditional to $Z$ has some relatively simple distribution (in a Bayesian estimation context, $Z$ often contains model parameters). The quantity of interest, however, is the conditional distribution of $Z$ given $X$ (also called the ``posterior distribution''), which allows one to infer the latent structure from the observations, and will also have an important role in maximum likelihood parametric estimation, as we will see below. This conditional distribution is not always easy to compute or simulate, and variational Bayes provides a framework under which it can be approximated.

\section{Variational principle}
\label{sec:var.principle}
We  consider a pair of random variables $X$ and $Z$, where $X$ is considered as ``observed'' and $Z$ is hidden, or ``latent''. We will use $U = (X,Z)$ to denote the two variables taken together. We denote as usual by $P_U$ the probability law of $U$, defined on $\CR_U = \CR_X \times \CR_Z$ by $P_U(A)= \myP(U\in A)$. We will also assume that there exists a measure $\mu$ on $\CR_U$ that decomposes as a product measure $\mu = \mu_X \times \mu_Z$ (where $\mu_X$ and $\mu_Z$ are measures on $\CR_X$ and $\CR_Z$), such that  $P_U \ll \mu$ ($P_U$ is absolutely continuous with respect to $\mu$). This implies that $P_U$ has a density with respect to $\mu$ that we will denote $f_U$. If both $\CR_X$ and $\CR_Z$ are discrete, $\mu$ is typically the counting measure, and if they are both Euclidean spaces, $\mu$ can be the Lebesgue measure on the product.\footnote{The reader unfamiliar with measure theory may want to read this discussion by replacing $d\mu_X$ by $dx$, $d\mu_Z$ by dz and $d\mu_U$ by $dx\,dz$, i.e., in the context of continuous probability distributions having p.d.f.'s with respect to the Lebesgue's measure.}

The variables $X$ and $Z$ then have probability density functions with respect to $\mu_X$ ad $\mu_Z$, given by
\[
f_X(x) = \int_{\CR_Z} f_U(x,z) \mu_Z(dz) \quad \text{ and } \quad f_Z(z) = \int_{\CR_X} f_U(x,z) \mu_X(dx).
\]

The conditional distribution of $X$ given $Z=z$, denoted $P_X(\,\cdot\mid Z=z)$, has density $f_X(x\mid z) = f_U(x,z)/f_Z(z)$ with respect to $\mu_X$ and that of $Z$ given $X=x$, denoted  $P_Z(\,\cdot\mid X=x)$, has density $f_Z(z\mid x) = f_U(x,z) / f_X(x)$ with respect to $\mu_Z$. 
We will be mainly interested by approximations of $P_Z(\,\cdot\mid X=x)$, assuming that $P_Z$ and $P_X(\,\cdot\mid Z=z)$ (and hence $P_U$) are easy to compute or simulate.

We will use the Kullback-Liebler divergence to quantify the accuracy of the approximation. As stated in \cref{prop:kl}, we have
\[
P_Z(\,\cdot\mid X=x) = \argmin_{\nu\in  \mathcal M_1(\CR_Z)} \KL(\nu\ \|\ P_Z(\cdot\mid X=x))
\]
where $\mathcal M_1(\CR_Z)$ denotes the set of all probability distributions on $\CR_Z$. Note that all distributions $\nu$ for which $\KL(\nu\ \|\ P_Z(\cdot\mid X=x))$ is finite must be absolutely continuous with respect to $\mu_Z$ and therefore take the form $\nu = g \mu_Z$. One   has
\begin{align}
\nonumber
\KL(g\mu_Z\ \|\ P_Z(\cdot|X=x)) &= \int_{\CR_Z} \log \frac{g(z)}{f_Z(z|x)} g(z) \mu_Z(dz) \\
\label{eq:variational.b}
&= \int_{\CR_Z} \log \frac{g(z)}{f_U(x,z)} g(z) \mu_Z(dz) + \log f_X(x).
\end{align}
We will denote by $\CP(\mu_Z)$, or just $\CP$ when there is no ambiguity, the set of all p.d.f.'s $g$ with respect to $\mu_Z$, i.e., the set of all non-negative measurable functions on $\CR_Z$ with $\int_{\CR_Z} g(z) \mu_Z(dz) = 1$.

The basic principle of variational Bayes methods is to replace  $\CP$ by a subset $\hCP$ and to define the approximation
\[
\widehat P_Z(\ccdot \mid X=x) = \argmin_{g \in \hCP} \KL(g \mu_Z\ \|\ P_Z(\ccdot|X=x)).
\]
For the approximation to be practical, the set $\hCP$ must obviously be chosen so that the computation of $\widehat P_Z(\ccdot \mid X=x)$ is computationally feasible. We now review a few examples, before passing to the EM algorithm and its  approximations.

\section{Examples}
\label{sec:var.bayes.expl}
\subsection{Mode approximation}
\label{sec:mode.approx}
Assume that $\CR_Z$ is discrete and $\mu_Z$ is the counting measure so that 
\[
\KL(g\mu_Z\ \|\ P_Z(\ccdot\mid X=x))  - \log f_X(x) = \sum_{z\in \CR_Z} \log \frac{g(z)}{f_U(x,z)} g(z),
\]
the sum being infinite if there exists $z$ such that $\nu(z) > 0$ and $f_U(x, z) = 0$.
Take 
\[
\hCP = \defset{\bfone_z: z\in \CR_Z},
\]
the family of all Dirac functions on $\CR_Z$. Then,
\[
\KL(\bfone_z\mu_Z\ \|\ P_Z(\ccdot|X=x)) - \log f_X(x) =  -\log f_U(x,z).
\]
The variational approximation of $P_Z(\ccdot\mid X=x)$ over $\hCP$ therefore is the Dirac measure at point(s) $z\in \CR_Z$ at which $f_U(x, z)$ is largest, i.e., the mode(s) of the posterior distribution. This approximation is often called the MAP approximation (for maximum a posteriori).

If $\CR_Z$ is, say, $\mR^q$ and $\mu_Z=dz$ is Lebesgue's measure, then the previous construction does not work because $\bfone_z$ is not a p.d.f. with respect to $\mu_Z$. Instead of Dirac functions, one can however use constant functions  on small balls. Let $B(z, \ep)$ denote the open ball with radius $\epsilon$, and let $|B(z, \ep)|$ denote its volume. Let $\mathfrak u_{z,\ep} = \bfone_{B(z,\ep)}/|B(z, \ep)|$. Fixing $\epsilon$, we can consider the set 
\[
\hCP = \defset{\mathfrak u_{z, \ep}: z\in \mR^q}.
\]
Now, one has (leaving the computation to the reader)
\[
\KL(\mathfrak u_{z, \ep} dz\ \|\ P_Z(\ccdot\mid X=x)) - \log f_X(x) =  -\log \left(\frac{1}{|B(z, \ep)}\int_{B(z, \ep)} f_U(x, z') dz' \right).
\]
The limit for small $\epsilon$  (assuming that $f_U(x, \cdot)$ is continuous at $z$, or defining the limit up to sets of measure zero) is $ -\log f_U(x,z)$, justifying again choosing the mode of the posterior distribution of $Z$ for the approximation. 

The mode approximation has some limitations. First, it is in general a very crude approximation of the posterior distribution. Second, even with the assumption that  $f_U$ has closed form,  this p.d.f. is often difficult to maximize (for example when defining models over large discrete sets). In such cases, the mode approximation has limited practical use.

\subsection{Gaussian approximation}

Let us still assume that $\CR_Z = \mR^q$ and that $\mu_Z = dz$.  Let $\hCP$ be the family of all Gaussian distributions $\CN(m, \Sig)$ on $\mR^q$. Then, denoting by $\phi(\ccdot;m,\Sigma)$ the density  of  $\CN(m, \Sig)$,
\begin{multline*}
\KL(\phi(\ccdot;\,m , \Sig)\ \|\ P_Z(\ccdot\mid X=x)) - \log f_X(x) =-\frac{q}2 \log 2\pi - \frac q2 - \frac12 \log \det(\Sigma) \\
- \int_{\mR^q} \log f_U(x,z) \phi(z;m,\Sigma) dz.
\end{multline*}

In order to provide the best approximation, $m$ and $\Sigma$ must therefore maximize
\begin{equation}
\int_{\mR^q} \log f_U(x,z) \phi(z;m,\Sigma) dz + \frac12 \log \det(\Sigma).
\label{eq:laplace.approx}
\end{equation}
The resulting optimization problem does not have a closed form solution in general (see \cref{sec:vae} for an example in which stochastic gradient methods are used to solve this problem). Another approach that is commonly used in practice is to push  the approximation further by replacing $\log f_U(x,z)$ by its second-order expansion around its maximum as a function of $z$. Let $m(x)$ be the posterior mode, i.e., the value of $z$ at which $x \mapsto \log f_U(x, z)$ is maximal, that we will assume to be unique. Let $H(x)$ denote the $q\times q$ Hessian matrix formed by the second partial derivatives of  $-\log f_U(x, z)$ (with respect to $z$) at $z = m(x)$. This matrix is positive semidefinite according to the choice made for  $m(x)$, and we will assume that it is positive definite. Since the first derivatives of $\log f_U(x, z)$ at $m(x)$ must vanish, we have the expansion:
\[
 \log f_U(x,z) = \log f_U(x,m(x)) - \frac12 (z-m(x))^T H(x) (z-m(x)) + \cdots
\]
 Plugging the expansion into the integral in \cref{eq:laplace.approx} yields
\[
-\frac12 \trace(H(x) \Sigma) - \frac12  (m-m(x))^T H(x) (m-m(x)) + \frac12 \log\det\Sig.
\]
To maximize this expression, one must clearly take $m = m(x)$. Moreover, 
\[
\prt_{\Sigma}\left(-\trace(H(x) \Sigma) + \log\det\Sig\right) = -H(x)^T + (\Sigma^T)^{-1} = -H(x) + \Sigma^{-1},
\]
and we see that  one must take $\Sigma = H(x)^{-1}$. This provides the Laplace approximation \citep{dieudonne1971infinitesimal} of the posterior, $\CN(m(x), H(x)^{-1})$, which is practical when the mode and corresponding second derivatives are feasible to compute.

\subsection{Mean-field approximation}
\label{sec:mean.field}

This section generalizes the approach discussed in \cref{prop:mean.f} for Markov random fields.
Assume that $\CR_Z$ can be decomposed into several components $\CR_Z^{[1]}, \ldots, \CR_Z^{[K]}$, writing $z = (z^{[1]}, \ldots, z^{[K]})$ (for example, taking $K=q$ and $z^{[i]} = z^{(i)}$, the $i$th coordinate of $z$ if $\CR_Z= \mR^q$).  Also assume that $\mu_Z$  splits into a product measure $\mu_Z^{[1]} \otimes\cdots\otimes \mu_Z^{[K]}$. 
Mean-field approximation consists in assuming that probabilities $\nu$ in $\hCP$ split into independent components, i.e.,  their densities $g$ take the form:
\[
g(z) = g^{[1]}(z^{[1]}) \cdots g^{[K]}(z^{[K]}).
\]
Then,
\begin{multline}
\label{eq:var.b.mean.field}
\KL(\nu\ \|\ P_Z(\ccdot\mid X=x))  - \log f_X(x) = \sum_{j=1}^K \int_{\CR_Z^{[j]}} \log g^{[j]}(z^{[j]}) g^{[j]}(z^{[j]}) \mu_Z^{[j]}(dz^{[j]}) \\
- \int_{\CR_Z} \log f_U(x,z) \prod_{j=1}^q g^{[j]}(z^{[j]}) \mu_{Z}(dz).
\end{multline}
The mean-field approximation may be feasible when $\log f_U(x,z)$ can be written as a sum of products of functions of each $z^{[j]}$. Indeed, assume that 
\begin{equation}
\label{eq:mean.field.decomp}
\log f_U(x,z) = \sum_{\alpha\in A} \prod_{j=1}^K \psi_{\alpha,j}(z^{[j]}, x)
\end{equation}
where $A$ is a finite set. To shorten notation, let us denote by $\langle \psi\rangle$ the expectation of a function $\psi$
with respect to the product p.d.f. $g$.
Then, \cref{eq:var.b.mean.field} can be written as
\[
\KL(\nu\ \|\ P_Z(\ccdot\mid X=x))  - \log f_X(x) = \sum_{j=1}^K \langle \log g^{(j)}(z^{[j]}) \rangle 
- \sum_{\al\in A} \prod_{j=1}^K \langle \psi_{\alpha,j}(z^{[j]}, x) \rangle.
\]

The following lemma will allow us to identify the form taken by the optimal p.d.f. $g^{[j]}$. 
\begin{lemma}
\label{lem:max.ent}
Let $Q$ be  a set equipped with a positive measure $\mu$. Let $\psi: Q \to \mR$ be a measurable function such that 
\[
C_\psi \defeq \int_Q \exp(\psi(q)) \mu(dq) < \infty.
\]
Let 
\[
g_\psi(q) = \frac{1}{C_\psi} \exp(\psi(q)).
\]
Let $g$ be any p.d.f. with respect to $\mu$, and define
\[
F(g) = \int_Q (\log g(q) - \psi(q)) g(q) \mu(dq).
\]
Then $F(g_\psi) \leq F(g)$. 
\end{lemma}
\begin{proof}
We note that $g_\psi > 0$, and that 
\[
\KL(g\ \|\ g_\psi) = F(g) + \log C_\psi = F(g) - F(g_\psi),
\]
which proves the result, since KL divergences are always non-negative.
\end{proof}

Applying this lemma separately to each function $g^{[j]}$ implies that any optimal $g$ must be such that
\[
g^{[j]}(z^{[j]}) \propto \exp\left(\sum_{\al\in A} M_{\alpha,j}\psi_{\alpha,j}(z^{[j]}, x)\right)
\]
with
\[
M_{\alpha,j} = \prod_{j'=1, j'\neq j}^K \langle \psi_{\alpha,j'}(z^{[j']}, x) \rangle.
\]
We therefore have
\begin{equation}
\label{eq:mean.field.cons}
\langle \psi_{\alpha,j}(z^{[j]}, x) \rangle = \frac{\int_{\CR_Z^{[j]}} \psi_{\alpha,j}(z^{[j]}, x) \exp\left(\sum_{\al'\in A} M_{\alpha',j}\psi_{\alpha',j}(z^{[j]}, x)\right) \mu_Z^{[j]}(dz^{[j]})}{\int_{\CR_Z^{[j]}} \exp\left(\sum_{\al'\in A} M_{\alpha',j}\psi_{\alpha',j}(z^{[j]}, x)\right) \mu_Z^{[j]}(dz^{[j]})}
\end{equation}

This specifies a relationship expressing $\langle \psi_{\alpha,j}(z^{[j]}, x) \rangle$ as a function of the other expectations $\langle \psi_{\alpha',j'}(z^{(j')}, x) \rangle$ for $j\neq j'$. These equations put together are called the {\em mean-field consistency equations}. When these equations can be written explicitly, i.e., when the integrals in \cref{eq:mean.field.cons} can be evaluated analytically (which is generally the case when the p.d.f.'s $g^{[j]}$ can be associated with standard distributions), one obtains an algorithm that iterates \cref{eq:mean.field.cons} over all $\alpha$ and $j$ until stabilization (each step reducing the objective function in \cref{eq:var.b.mean.field}).

Let us retrieve the result obtained in \cref{prop:mean.f} using the current formalism. Assume that  $\CR_X$ finite and $\CR_Z = \{0,1\}^L$, where $L$ can be a large number, with
\[
f_U(x, z) = \frac1C \exp\left( \sum_{j=1}^L \alpha_j(x) z^{(j)} + \sum_{i,j=1, i<j}^L \beta_{ij}(x) z^{(i)} z^{(j)}\right).
\]
Take $K=L$, $z^{[j]} = z^{(j)}$. Applying the previous discussion, we see that $g^{[j]}$ must take the form
\[
g^{[j]}(z^{(j)}) = \frac{\exp\left(\alpha_j(x) z^{(j)}  + \sum_{i\neq j}\beta_{ij}(x) \langle z^{(i)} \rangle z^{(j)}\right) }{1 + \exp\left(\alpha_j(x)   + \sum_{i\neq j}\beta_{ij}(x) \langle z^{(i)} \rangle\right)}
\]
In particular
\[
\langle z^{(j)} \rangle =  \frac{\exp\left(\alpha_j(x)  + \sum_{i\neq j}\beta_{ij}(x) \langle z^{(i)} \rangle \right) }{1 + \exp\left(\alpha_j(x)  + \sum_{i\neq j}\beta_{ij}(x) \langle z^{(i)} \rangle\right)}
\]
providing the mean-field consistency equations. 

In this special case, it is also possible to express the objective function as a simple function of the expectations $\langle z^{(j)}\rangle$'s. We indeed have, letting $\rho_j = \langle z^{(j)}\rangle$,
\[
\sum_{z\in \CR_Z} \log f_U(x,z) \prod_{j=1}^L g^{[j]}(z^{(j)})  =
-\log C + \sum_{j=1}^L \alpha_j(x) \rho_j + \sum_{i,j=1, i<j}^L \beta_{ij}(x) \rho_i\rho_j.
\]
The values of $\rho_1, \ldots, \rho_L$ are then obtained  by maximizing
\[
\sum_{j=1}^L \alpha_j(x) \rho_j + \sum_{i,j=1, i<j}^L \beta_{ij}(x) \rho_i\rho_j - \sum_{j=1}^L \big(\rho_j\log\rho_j + (1-\rho_j)\log(1-\rho_j)\big).
\]
The consistency equations express the fact that the derivatives of this expression with respect to each $\rho_j$ vanish.

\section{Maximum likelihood estimation}

%
%

\subsection{The EM algorithm}
We now consider maximum likelihood estimation with latent variables and use the notation of \cref{sec:var.principle}. The main tool is the following obvious consequence of \cref{eq:variational.b}.
 \begin{proposition}
 \label{prop:var.bayes}
 One has
 \[
\log f_X(x) = \max_{g\in \CP(\mu_Z)} \int_{\CR_Z} \log \left(\frac{f_U(x,
z)}{g(z)}\right) g(z) d\mu_Z(z)
\]  
and the maximum is achieved for $g(z) = f_Z(z\mid x)$, the conditional p.d.f. of $Z$ given $X=x$.
 \end{proposition}
 \begin{proof}
 \Cref{eq:variational.b} implies that 
 \[
 \int_{\CR_Z} \log \left(\frac{f_U(x,
z)}{g(z)}\right) g(z) d\mu_Z(z) = \log f_X(x) - \KL(g\,\mu_Z\ \|\ P_Z(\ccdot|X=x))
\]
and the r.h.s. is indeed maximum when the Kullback-Liebler divergence vanishes, that is, when $g$ is the p.d.f. of  $P_Z(\ccdot\mid X=x)$.
\end{proof}

We will use this proposition for the derivation of the expectation-maxi\-mi\-zation (or EM) algorithm for maximum likelihood with latent variables. We now assume that  
 $P_U$, and therefore $f_U$, is parametrized by $\th\in \Theta$, and that a training set  $T = (x_1, \ldots, x_N)$ of realizations of $X$ is observed. To indicate the dependence in $\theta$, we will write $f_U(x,z\,;\,\theta)$, or $f_Z(z\mid x\,;\,\theta)$.
  The maximum likelihood estimator (m.l.e.) then maximizes
 \[
 \ell(\th) = \sum_{k=1}^N \log f_X(x_k\,;\,\theta)\,.
 \]
 The EM algorithm is useful when the
computation of the m.l.e. for complete observations, i.e., the maximization
of 
\[
 \log f_U(x, z\,;\,\theta)
\]
when both $x$ and $z$ are given, is easy, whereas the same problem with the marginal distribution
is hard.
   
From \cref{prop:var.bayes}, we have:
\[
\sum_{k=1}^N\log f_X(x_k\,;\,\theta)= \sum_{k=1}^N \max_{g_k\in \CP(\mu_Z)} \int_{\CR_Z} \log \left(\frac{f_U(x_k,
z\,;\,\theta)}{g_x(z)}\right) g_k(z) \mu_Z(dz)  
\]
Therefore the maximum likelihood requires to compute
\begin{equation}
\label{eq:lik.var}
\max_{\th, g_1, \ldots,  g_N} \sum_{k=1}^N \int_{\CR_Z} \log \left(\frac{f_U(x_k,
z\,;\,\theta)}{g_{k}(z)}\right) g_{k}(z) \mu_Z(dz)  .
\end{equation}

The maximization can therefore be done by iterating the following  two steps.
\begin{enumerate}[label=\arabic*., wide=0.5cm]
\item Given $\th_n$,  compute 
\[
\argmax_{g_1, \ldots, g_N} \sum_{k=1}^N \int_{\CR_Z} \log \left(\frac{f_U(x_k,
z\,;\,\theta)}{g_k(z)}\right) g_k(z) \mu_Z(dz) .
\]
\item Given $g_x, x\in T$,  compute 
\begin{align*}
&\argmax_{\th}  \sum_{k=1}^N \int_{\CR_Z} \log \left(\frac{f_U(x_k,
z\,;\,\theta)}{g_k(z)}\right) g_k(z) \mu_Z(dz) \\
&=  \argmax_{\th} 
\sum_{k=1}^N \int_{\CR_Z} \log \left(f_U(x_k,
z\,;\,\theta)\right) g_k(z) \mu_Z(dz).
\end{align*}
\end{enumerate}

Step 1. is explicit and its solution is $g_k(z) = f_Z(z\mid x_k\,;\,\theta)$. 
Using this, both steps can be grouped together, yielding the EM algorithm.
\begin{algorithm}[EM algorithm]
\label{alg:EM}
Let a statistical model with density $f_U(x,z\,;\,\theta)$ modeling an observable variable $X$ and a latent variable $Z$ be given, and a training set $T = (x_1, \ldots, x_N)$ be observed. Starting with an initial guess of the parameter, $\theta(0)$, the EM algorithm iterates the following equation until numerical stabilization. 
\begin{equation}
\label{eq:EM}
\th_{n+1} = \argmax_{\th'} \sum_{k=1}^N \int_{\CR_Z} \log \left(f_U(x_k,
z\,;\,\theta')\right) f_Z(z\mid x_k\,;\,\theta_n) \mu_Z(dz).
\end{equation}
\end{algorithm}
\Cref{eq:EM} maximizes (in $\th'$) a function defined as an expectation (for $\th_n$), justifying the name "Expectation-Maximization."

\subsection{Application: Mixtures of Gaussian}
A mixture of Gaussian (MoG) model was introduced in \cref{chap:intro} (equation \cref{eq:MoG.model}). We now reinterpret it (in a slightly generalized version) as a model with partial observations and show how the EM algorithm can be applied. Let $\phi(x\,;\, m, \Sigma)$ denote the p.d.f. of the $d$-dimensional multivariate Gaussian distribution with mean $m$ and covariance matrix $\Sigma$. We model $f_X(x\,;\,\theta)$ as
\[
f_X(x\,;\,\theta) = \sum_{j=1}^p \alpha_j \phi(x,;\, c_j, \Sigma_j).
\]
Here, $\theta$ contains all sequences $\alpha_1, \ldots, \alpha_p$ (non-negative numbers that sum to one), $c_1, \ldots, c_p\in \mR^d$ and $\Sigma_1, \ldots, \Sigma_p$ ($d\times d$ positive definite matrices).

Using the previous notation, we therefore have $\CR_X = \mR^d$, and $\mu_X$ the Lebesgue measure on that space. 
The variable $Z$ will take values in $\CR_Z = \{1, \ldots, p\}$, with $\mu_Z$ being the counting measure. We model the joint density function for $(X,Z)$ as
\begin{equation}
\label{eq:mog.lik}
f_U(x,z\,;\,\theta) = \alpha_z \phi(x;\, c_z, \Sigma_z).
\end{equation}

Clearly $f_X$ is the marginal p.d.f. of $f_U$. One can therefore consider $Z$ as a latent variable, and  estimate $\th$ using the EM algorithm.

We now make \cref{eq:EM} explicit for mixtures of Gaussian.
For given $\th$ and $\th'$ and $x\in\CR$, let
\begin{align*}
U_x(\th, \th') &= \frac d2 \log 2\pi + \int_{\CR_Z} \log \left(f_U(x,
z\,;\,\theta')\right) f_Z(z\mid x\,;\,\theta) d\mu_Z(z)  \\
& = \sum_{z=1}^p \left(\log\alpha'_z - \frac{1}{2}
\log\det\Sig'_z -   \frac12 (x-c'_z)^T{\Sig'_z}^{-1} (x-c'_z)\right) f_Z(z\mid x\,;\,\theta) 
\end{align*}
with
\[
f_Z(z\mid x\,;\,\theta) = \frac{(\det \Sig_z)^{-\frac12} \alpha_z e^{- \frac12
(x-c_z)^T\Sig_z^{-1} (x-c_z)}}{\sum_{j=1}^p (\det \Sig_j)^{-\frac12} \alpha_j e^{-\frac12
(x-c_j)^T\Sig_j^{-1} (x-c_j)}}.
\]

If $\th_n$ is the current parameter in the EM, the next one, $\th_{n+1}$ must  maximize
$\sum_{k=1}^N U_{x}(\th_n, \th')$. This can be solved in closed form. To compute $\al'_1, \ldots, \al'_p$, one must maximize 
\[
\sum_{k=1}^N\sum_{z=1}^p (\log\al'_z) f_Z(z\mid x\,;\,\theta)
\]
subject to the constraint that $\sum_z \al'_z = 1$. This yields
\[
\al'_z = \sum_{k=1}^N \Lfrac{f_Z(z\mid x_k\,;\,\theta)}{\sum_{j=1}^p\sum_{k=1}^N f_Z(j\mid x_j\,;\,\theta)} = \lfrac{\zeta_z}{N}
\]
with
$\zeta_z = \sum_{k=1}^N f_Z(z\mid x_k\,;\,\theta).$

The centers $c'_1, \ldots, c'_p$ must minimize $\sum_{k=1}^N (x_k-c'_z)^T{\Sig'_z}^{-1}
(x_k-c'_z)f_Z(z|x\,;\,\theta)$, which yields
\[
c'_z = \frac{1}{\zeta_z}\sum_{k=1}^N x f_Z(z\mid x_k\,;\,\theta).
\]
Finally, $\Sig'_z$ must minimize
\[
\frac{\zeta_z}{2}
\log\det\Sig'_z  + \frac12   \sum_{k=1}^N (x_k-c'_z)^T{\Sig'_z}^{-1} (x_k-c'_z) f_Z(z\mid x_k\,;\,\theta),
\]
which yields
\[
\Sig'_z = \frac{1}{\zeta_z} \sum_{k=1}^N (x_k - c'_z)(x_k-c'_z)^T f_Z(z\mid x_k\,;\,\theta).
\]
We can now summarize the algorithm.
\begin{algorithm}[EM for Mixture of Gaussian distributions]
\label{alg:mix.gauss}
\begin{enumerate}[label=\arabic*., wide=0.5cm]
\item Initialize the  parameter $\th(0) = (\al(0), c(0), \Sig(0))$. Choose a small constant $\ep$ and a maximal number of iterations $M$.
\item At step $n$ of the algorithm, let $\th = \th_n$ be the current parameter, writing for short $\theta = (\alpha, c, \Sigma)$. 
\item Compute, for $k=1, \ldots, N$ and $z=1, \ldots, p$
\[
f_Z(z\mid x_k\,;\,\theta) = \frac{(\det \Sig_z)^{-\frac12} \al_i e^{- \frac12
(x_k-c_z)^T\Sig_z^{-1} (x_k-c_z)}}{\sum_{j=1}^p (\det \Sig_j)^{-\frac12} \al_j e^{-\frac12
(x_k-c_j)^T\Sig_j^{-1} (x_k-c_j)}}
\]
and let $\zeta_z = \sum_{k=1}^N f_Z(z\mid x_k\,;\,\theta)$, $z=1, \ldots, p$.
\item Let $\al'_z = \zeta_z/N$.
\item For $z=1, \ldots, p$, let 
\[
c'_z = \frac{1}{\zeta_z}\sum_{k=1}^N x_k f_Z(z\mid x_k\,;\,\theta).
\]
\item For $z=1, \ldots, p$, let 
\[
\Sig'_z = \frac{1}{\zeta_z} \sum_{k=1}^N (x_k - c'_z)(x_k-c'_z)^T f_Z(z\mid x_k\,;\,\theta).
\]
\item Let $\th' = (\mu', c', \Sig')$. If $|\th' - \th| <\ep$ or $n+1=M$: return $\theta'$ and exit the algorithm.
\item Otherwise, set $\th_{n+1}=\th'$ and return to step 2. 
\end{enumerate}
\end{algorithm}

\begin{remark}
\label{rem:mog}
\Cref{alg:mix.gauss} can be simplified by making restrictions on the model. Here are some examples.
\begin{enumerate}[label=(\roman*)]
\item One may restrict to $\Sig_z = \sig^2_z \Id[d]$ to reduce the number of free parameters. Then, step 7 of the algorithm needs to be replaced   by:
\[
(\sig'_z)^2 = \frac{1}{d\zeta_i} \sum_{k=1}^N |x_k - c'_z|^2 f_Z(z\mid x_k\,;\,\theta).
\]
\item Alternatively, the model may be simplified by assuming that all covariance matrices coincide: $\Sig_z = \Sig$ for $z=1, \ldots, p$. Then, step 7 becomes
\[
\Sig' = \frac{1}{N} \sum_{z=1}^p \sum_{k=1}^N (x_k - c'_z)(x_k-c'_z)^T f_Z(z\mid x_k\,;\,\theta).
\]
\item Finally, one may assume that $\Sig$ is known and fixed in the algorithm (usually in the form $\Sig = \sig^2\Id[d]$ for some $\sig>0$) so that step 7 of the algorithm can be removed.
\item One may also assume also that the (prior) class probabilities are known, typically set to $\al_z = 1/p$ for all i, so that step 4 can be skipped.

\end{enumerate}
\end{remark}

\subsection{Stochastic approximation EM}
\label{sec:saem}
The stochastic approximation EM (or SAEM) algorithm has been proposed by \citet{delyon1999convergence} (see this reference for convergence results) to address the situation in which the expectations for the posterior distribution cannot be computed in closed form, but can be estimated using Monte-Carlo simulations.  SAEM uses a special form of stochastic approximation, different from the SGD algorithm described in \cref{sec:sgd}. It updates, at each step $n$, an approximate objective function that we will denote $\lambda_n$ and a current parameter $\theta(n)$.  It implements the following iterations:
\begin{equation}
\left\{
\begin{aligned}
\xi^{(k)}_{n+1} & \sim P_Z(\ccdot\mid X=x_k\,;\,\theta_n),\quad  k=1, \ldots, N\\
\la_{n+1}(\theta') &= \Big(1-\frac1{n+1}\Big)\la_n(\theta') + \frac1{n+1}\Big( \sum_{k=1}^N \log f_U(x_k, \xi^{(k)}_{n+1}\,;\,\theta') - \la_n(\theta')\Big),\  \theta'\in \Theta\\
\theta_{n+1} &= \argmax_{\theta'}  \la_{n+1}(\theta')
\end{aligned}
\right.
\label{eq:saem.1}
\end{equation}
The second step means that 
\[
\la_{n}(\theta') = \sum_{k=1}^N \left(\frac{1}{n} \sum_{j=1}^n \log f_U(x, \xi^{(k)}_j\,;\,\theta')\right).
\]
Given that $\xi^{(k)}_{n+1} \sim P_Z(\ccdot\mid X=x_k\,;\,\theta_n)$, one expects this expression to approximate
\[
\sum_{k=1}^N \int_{\CR_Z} \log \left(f_U(x_k,
z\,;\,\theta')\right) f_Z(z\mid x_k\,;\,\theta) d\mu_Z(z)
\]
so that the third step of \cref{eq:saem.1} can be seen as an approximation of \cref{eq:EM}. Sufficient conditions under which this actually happens (and $\theta(n)$ converges to a local maximizer of the likelihood) are provided in \citet{delyon1999convergence} (see also \citet{kuhn2004coupling} for a convergence result under more general hypotheses on how $\xi$ is simulated).

To be able to run this algorithm efficiently, one needs the simulation of  the posterior distribution to be feasible. Importantly, one also needs to be able to update efficiently the function $\la_n$. This can be achieved when the considered model belongs to an exponential family, which corresponds to assuming that the p.d.f. of $U$ takes the form
\[
f_U(x,z\,;\,\theta) = \frac{1}{C(\theta)} \exp\big(\psi(\theta)^TH(x,z)\big)
\]
for some  functions $\psi$ and $H$. For example, the MoG model of equation \cref{eq:MoG.model} takes this form, with
\begin{align*}
\psi(\th)^T &=\Big (&&\log \al_1 -\frac12 m_1^T\Sigma_1^{-1} m_1 -\frac12 \log\det\Sigma_1, \ldots, \log\al_p - \frac12 m_p^T\Sigma_p^{-1} m_p-\frac12 \log\det\Sigma_p,\\
 &&&\Sigma_1^{-1}m_1, \ldots, \Sigma_p^{-1} m_p,\\
 &&& \Sigma_1^{-1}, \ldots, \Sigma_p^{-1}\Big),\\
H(x,z)^T &= \Big(&&\bfone_{z=1}, \ldots, \bfone_{z=p},\\
&&&x\bfone_{z=1}, \ldots, x\bfone_{z=p},\\
&&& -\frac12 xx^T\bfone_{z=1}, \ldots,-\frac12 xx^T\bfone_{z=p} \Big)
\end{align*}
and
$C(\theta) = (2\pi)^{pd/2}$.

For such a model, we can replace the algorithm in \cref{eq:saem.1} by the more manageable one:
\begin{equation}
\left\{
\begin{aligned}
\xi^{(k)}_{n+1} & \sim P_Z(\ccdot\mid X=x_k\,;\,\theta_n),\  k=1, \ldots, N\\
\eta^{(k)}_{n+1} &= \Big(1-\frac1{n+1}\Big)\eta^{(k)}_n + \frac1{n+1} (H(x_k, \xi^{(k)}_{n+1}) - \eta^{(k)}_n)\\
\la_{n+1}(\theta') &= \psi(\theta')^T \Big(\sum_{k=1}^N \eta^{(k)}_{n+1}\Big) - \log C(\theta')\\
\theta_{n+1} &= \argmax_{\theta'}  \la_{n+1}(\theta')
\end{aligned}
\right.
\label{eq:saem.2}
\end{equation}
We leave to the reader the  computation leading to the implementation of this algorithm for mixtures of Gaussian.

\subsection{Variational approximation}
\label{sec:var.approx}
Returning to \cref{prop:var.bayes} and equation \cref{eq:lik.var}, we see that one can make a variational approximation of the maximum likelihood by computing
\begin{equation}
\max_{\th\in \Theta, g_1, \ldots, g_N \in \hCP} \sum_{k=1}^N \int_{\CR_Z} \log \left(\frac{f_U(x_k,
z\,;\,\theta)}{g_k(z)}\right) g_k(z) \mu_Z(dz) ,
\label{eq:em.var}
\end{equation}
where $\hCP\sub \CP$ is a class of p.d.f. with respect to $\mu_Z$. The resulting algorithm is then implemented by iterating the computation of $g_1, \ldots, g_N$, using approximations similar to those provided in \cref{sec:var.bayes.expl}, and maximization in $\theta$ for given $g_1, \ldots, g_N$.  
 This variational approximation of the maximum likelihood estimator is therefore  provided by the following algorithm.
\begin{algorithm}[Variational Bayes approximation of the m.l.e.]
\label{alg:var.EM}
Let a statistical model with density $f_U(x,z\,;\,\theta)$ modeling an observable variable $X$ and a latent variable $Z$ be given, and a training set $T = (x_1, \ldots, x_N)$ be observed. Let $\hCP$ be a set of p.d.f.'s on $\CR_Z$ and define 
\[
\hat g(\ccdot;\,x, \theta) = \argmin_{g\in \hCP} \int_{\CR_Z} \log \left(\frac{g(z)}{f_U(x,
z\,;\,\theta)}\right) g(z) \mu_Z(dz)
\]
(assuming that this minimizer is uniquely defined).

Starting with an initial guess of the parameter, $\theta_0$,  iterate the following equation until numerical stabilization:
\begin{equation}
\label{eq:var.EM}
\th(n+1) = \argmax_{\th'} \sum_{k=1}^N\int_{\CR_Z} \log \left(f_U(x_k,
z\,;\,\theta')\right) \hat g(z\, ;\, x_k,\,\theta(n)) \mu_Z(dz).
\end{equation}
\end{algorithm}

Assume that the distributions in $\hCP$ are also parametrized, denoting their parameter by $\eta$, belonging to some Euclidean domain $V$. Let $g(\ccdot;\, \eta)$ denote the p.d.f. in $\hCP$ with parameter $\eta$. Letting $\boldsymbol \eta  = (\eta_1, \ldots, \eta_N)$ denote an element of $V^T$ (parameters in $V$ indexed by elements of the training set), \cref{eq:em.var} can then be written as the maximization of  
\begin{equation}
F(\theta, \bfeta) = \sum_{k=1}^N \int_{\CR_Z} \log \left(\frac{f_U(x_k,
z\,;\,\theta)}{g(z;\eta_k)}\right) g(z;\eta_k) \mu_Z(dz) .
\label{eq:em.var.par}
\end{equation}

This expression is amenable to a stochastic gradient ascent implementation \cite{paisley2012variational}. We have 
\[
\partial_\theta \int_{\CR_Z} \log \left(\frac{f_U(x_k,
z\,;\,\theta)}{g(z;\eta_k)}\right) g(z;\eta_k) \mu_Z(dz) = \int_{\CR_Z} \partial_\theta \log f_U(x_k,
z\,;\,\theta) g(z;\eta_k) \mu_Z(dz)
\]
and 
\begin{align*}
&\partial_{\eta} \int_{\CR_Z} \log \left(\frac{f_U(x_k,
z\,;\,\theta)}{g(z;\eta_k)}\right) g(z;\eta_k) \mu_Z(dz) \\
&= 
  \int_{\CR_Z} \left(-\partial_\eta \log g(z;\eta_k) g(z;\eta_k) + \log \left(\frac{f_U(x_k,
z\,;\,\theta)}{g(z;\eta_k)}\right) \partial_\eta g(z;\eta_k)\right) \mu_Z(dz)\\
 &= 
  \int_{\CR_Z}\log \left(\frac{f_U(x,
z\,;\,\theta)}{g(z;\eta_k)}\right)\partial_\eta \log g(z;\eta_k) g(z;\eta_k)  \mu_Z(dz)
\end{align*}
Here, we have used the fact that, for all $\eta$, 
\[
  \int_{\CR_Z} \partial_\eta \log g(z;\eta) g(z;\eta)\,\mu_Z(dz) =   \int_{\CR_Z} \partial_\eta  g(z;\eta)\mu_Z(dz) = 0
  \]
  since $\int_{\CR_Z} g(x, \eta)\mu_Z(dz) = 1$.
  
Denote by $\pi_{\bfeta}$ the probability distribution of the random variable $\bfZ$ taking values in $\CR_Z^{N}$ obtained by  sampling $\bfZ= (Z_1, \ldots, Z_N)$ such that the components $Z_k$ are independent and with p.d.f. $g(\ccdot;\, \eta_k)$ with respect to $\mu_Z$. Define
\[
\Phi_1(\theta, \bfz) =  \sum_{k=1}^N \partial_\theta \log f_U(x_k, z_k\,;\,\theta)
\]
and
\[
\Phi_2(\theta, \bfeta, \bfz) = \sum_{k=1}^N\log \left(\frac{f_U(x_k,
z_k\,;\,\theta)}{g(z_k;\eta_k)}\right)\partial_\eta \log g(z_k;\eta_k).
\]
Then, following \cref{sec:sgd}, one can maximize \cref{eq:em.var.par} using the algorithm
\begin{equation}
\label{eq:sga.var.1}
\left\{
\begin{aligned}
\theta_{n+1} &= \theta_n + \gamma_{n+1} \Phi_1(\theta_n, \bfZ_{n+1})\\
\bfeta_{n+1} &= \bfeta_n + \gamma_{n+1} \Phi_2(\theta_n, \bfeta_n, \bfZ_{n+1})
\end{aligned}
\right.
\end{equation}
where $\bfZ_{n+1} \sim \pi_{\bfeta_n}$. 

Alternatively (for example when $N$ is large), one can also sample from minibatches at each update, or, as illustrated below, using a single training sample. 
This requires redefining $\pi_{\bfeta}$ as the distribution on $\{1, \ldots, N\}\times \CR_Z$ with p.d.f. $\phi_{\bfeta}(k, z) = g(z;\eta_k) / N$. One can now use
\[
\Phi_1(\theta, k, z) =  \partial_\theta \log f_U(x_k, z\,;\,\theta)
\]
and
\[
\Phi_2(\theta, \eta, k, z) = \log \left(\frac{f_U(x_k,
z\,;\,\theta)}{g(z;\eta)}\right)\partial_\eta \log g(z;\eta).
\]
This gives the iterations
\begin{equation}
\label{eq:sga.var.2}
\left\{
\begin{aligned}
\theta_{n+1} &= \theta_n + \gamma_{n+1} \partial_\theta \log f_U(x_{J_{n+1}}, Z_{n+1}\,;\,\theta_n)\\
\eta_{n+1,J_{n+1}} &= \eta_{n, x_{J_{n+1}}} + \gamma_{n+1} \log \left(\frac{f_U(x_{J_{n+1}},
Z_{n+1}\,;\,\theta_n)}{g(Z_{n+1};\eta_{n, J_{n+1}})}\right)\partial_\eta \log g(Z_{n+1};\eta_{n, J_{n+1}})
\end{aligned}
\right.
\end{equation}
with $(J_{n+1}, Z_{n+1}) \sim \pi_{\bfeta_{n}}$. 

\section{Remarks}
\subsection{Variations on the EM}
Based on the formulation of the EM as the solution of \cref{eq:lik.var}, it should be clear  that solving \cref{eq:EM} at each step can be replaced by any update of the parameter that increases \cref{eq:lik.var}.
For example, \cref{eq:EM} can be replaced by a partial run of a gradient ascent algorithm, stopped before convergence. One can also use a coordinate ascent strategy. Assume that $\theta$ can be split into several components, say two, so that $\theta = (\theta^{(1)}, \theta^{(2)})$. Then, \cref{eq:EM} may then be split into 
\begin{align*}
\th^{(1)}_{n+1} & = \argmax_{\th^{(1)}} \sum_{x\in T} \int_{\CR_Z} \log \left(f_U(x,
z\,;\,\theta^{(1)}, \theta^{(2)}_{n})\right) f_Z(z\mid x\,;\,\theta(n)) \mu_Z(dz)\\
\th^{(2)}_{n+1} & = \argmax_{\th^{(2)}} \sum_{x\in T} \int_{\CR_Z} \log \left(f_U(x,
z\,;\,\theta^{(1)}_{n+1}, \theta^{(2)})\right) f_Z(z\mid x\,;\,\theta(n)) \mu_Z(dz).
\end{align*}
Doing so is, in particular, useful when both these steps are explicit, but not \cref{eq:EM}.
 
\subsection{Direct minimization}
\label{sec:direct.min.incomplete}
While the EM algorithm is widely used in the context of partial observations, it is also possible to make explicit the derivative of 
\[
\log f_X(x\,;\,\theta) = \log \int_{\CR_Z} f_U(x, z\,;\,\theta) \mu_Z(dz)
\]
with respect to the parameter $\theta$. Indeed, differentiating the integral and writing $\prt_\th f_U = f_U \prt_\th \log f_U$, we have
\begin{align*}
\prt_\theta \log f_X(x\,;\,\theta) &= \int_{\CR_Z} \prt_\th \log f_U(x, z\,;\,\theta) \frac{f_U(x, z\,;\,\theta)}{f_X(x\,;\,\theta)}  \mu_Z(dz) \\
&= \int_{\CR_Z} \prt_\th \log f_U(x, z\,;\,\theta) f_Z(z\,|\,x, \theta)  \mu_Z(dz).
\end{align*}
In other terms, the derivative of the log-likelihood of the observed data is the conditional expectation of the derivative of the log-likelihood of the full data given the observed data. When computable, this expression can be used with standard gradient-based optimization methods, such as those described in \cref{chap:optim}. This expression is also amenable to a stochastic gradient ascent algorithm, namely
\begin{equation}
\label{eq:direct.min.incomplete}
\theta_{n+1} = \theta_n + \gamma_{n+1} \sum_{x\in T} \partial_\theta f_U(x, Z_{n+1, x}, \theta_n)
\end{equation}
where $Z_{n+1, x}$ follows the distribution with density $f_Z(\cdot\,|\,x, \theta_n)$ with respect to $\mu_Z$. An alternative SGA implementation can use the discussion in \cref{sec:var.approx}, with the density $g_{\eta_x}$ replaced by $f_Z(\ccdot\mid x, \eta_x)$, which leads to 
\[
\left\{
\begin{aligned}
\theta_{n+1} &= \theta_n + \gamma_{n+1} \sum_{x\in T} \partial_\theta \log f_U(x, Z_{n+1, x}, \theta_n)\\
\eta_{n+1,x} &= \eta_{n,x} - \gamma_{n+1} \partial_{\eta_x} \log f_Z(Z_{n+1,x}\mid x, \eta_x), \quad x\in T
\end{aligned}
\right.
\]
where $Z_{n+1, x}$ follows the distribution with density $f_Z(\cdot\mid x, \eta_{n,x})$.

\subsection{Variational approximations}
Let  
\[
\mathbf f(T, \bfz; \theta)= \prod_{k=1}^N f_U(x_k, z_k; \theta)
\]
for $\bfz = (z_1, \ldots, z_N)\in \CR_Z^N$. One can write \cref{alg:var.EM} in the slightly different form
\[
\max_{\th\in \Theta, \bfg \in \hCP} \int_{\CR^N_Z} \log \left(\frac{\mathbf f(T, \bfz; \theta)}{\bfg(\bfz)}\right) \bfg(\bfz) \mu_Z^{\otimes N} (d\bfz) ,
\]
where $\hCP$ now denotes a subset of $\CP_{\mu_Z^{\otimes N}}$, i.e., a set of densities with respect to $\mu_Z^{\otimes N}$. Obviously, if $\hCP = \CP_{\mu_Z^{\otimes N}}$, the optimal $\bfg$ is the product
\[
\bfg(\bfz) = \prod_{k=1}^N f(z_k\mid x_k; \theta).
\]
Writing the approximation in this form provides more latitude in the choice of the set $\hCP$ and its parametrization. To take a simple example, one can specify a single functional form $(x, z) \mapsto g(x, z; \eta)$ of transition density, parametrized by $\eta$, and let $\hCP$ be the set of all densities $\bfg$ in the form
\[
\bfg(\bfz; \eta) = \prod_{k=1}^N g(x_k, z_k; \eta),
\]
in which $\eta$ is taken independent of $k$. Using this model, \cref{eq:sga.var.2} (as an example) 
can then be modified to yield
\begin{equation}
\label{eq:sga.var.3}
\left\{
\begin{aligned}
\theta_{n+1} &= \theta_n + \gamma_{n+1} \partial_\theta \log f_U(x_{J_{n+1}}, Z_{n+1}\,;\,\theta_n)\\
\eta_{n+1} &= \eta_{n} + \gamma_{n+1} \log \left(\frac{f_U(x_{J_{n+1}},
Z_{n+1}\,;\,\theta_n)}{g(Z_{n+1};\eta_n)}\right)\partial_\eta \log g(Z_{n+1};\eta_{n}).
\end{aligned}
\right.
\end{equation}


\subsection{Product measure assumption}
We have worked, in this chapter, under the assumption that $P_U$ was absolutely continuous with respect to a product measure $\mu_U = \mu_X \otimes \mu_Z$. This is not a mild assumption, as it fails to include some important cases, for example when $X$ and $Z$ have some deterministic relationship, the simplest instance being when $X = F(Z)$ for some function $F$. In many cases, however, one can make simple transformations on the model that will make it satisfy this working assumption.   For example, if $X = F(Z)$, one can generally split $Z$ into $Z = (Z^{(1)}, Z^{(2)})$ so that the equation $X = F(Z)$ is equivalent to $Z^{(2)} = G(X, Z^{(1)})$ for some function $G$. One can then apply the discussion above to $U = (X,Z^{(1)})$ instead of $U = (X,Z)$. 

Using more advanced measure theory, however, one can see that this product decomposition assumption was in fact unnecessary. Indeed, one can assume that the measure $\mu_U$ can ``disintegrate'' in the following sense: there exists a measure $\mu_X$ on $\CR_X$ and, for all $x\in \CR_X$, a measure $\mu_Z(\ccdot\mid x)$ on $\CR_Z$ such that, for all functions $\psi$ defined on $\CR_U$, 
\[
\int_{\CR_U} \psi(x,z) \mu_U(dx,dz) = \int_{\CR_X} \int_{\CR_Z} \psi(x,z) \mu_Z(dz\mid x) \mu_X(dx).
\]
This is now a mild assumption, which is true \citep{bogachev2007measure} as soon as one assumes that $\mu_U(\CR)$ is finite (which is not a real loss of generality as one can  reduce to this case by replacing if needed $\mu_U$ by an equivalent probability distribution). 

With this assumption, the marginal distribution of $X$ had a p.d.f. with respect to $\mu_X$ given by
\[
f_X(x) = \int_{\CR_Z} f_U(x,z) \mu_Z(dz\mid x)
\]
and the conditional distributions $P_Z(\ccdot\mid x)$ have a p.d.f. relative to $\mu_Z(\ccdot\mid x)$ given by
\[
f_Z(z\mid x) = \frac{f_U(x,z)}{f_X(x)}. 
\]
The computations and approximations made earlier in this chapter can then be applied with essentially no modification. 


\chapter{Learning Graphical Models}
\label{chap:learning.graphical}
We discuss, in this chapter, several methods designed to learn parameters of graphical models, starting with the somewhat simpler case of Bayesian networks, than passing to Markov random fields on loopy graphs.

\section{Learning Bayesian networks}
\label{sec:graph.estim}

\subsection{Learning a Single Probability}
Since Bayesian networks are specified by probabilities and conditional
probabilities of configurations of variables, we start with a
discussion of the basic problem of estimating discrete
probability distributions.

The obvious way to estimate the probability of an event $A$ based on a
series of $N$ independent experiments is by using relative
frequencies
$$
f_A = \frac{\# \{A \text{ occurs}\}}{N}.
$$
This estimation is unbiased ($\myE(f_A) = \myP(A)$) and its variance is
$\myP(A)(1-\myP(A))/N$. This implies that the relative error
$\de_A = f_A/\myP(A) -1$ has zero mean and variance
$$
\sig^2 = \frac{1-\myP(A)}{N\myP(A)}\,.
$$

This number can clearly become very large when $\myP(A) \simeq 0$. In particular,
when $\myP(A)$ is small compared to $1/N$, the relative frequency 
will often be $f_A = 0$, leading to the false conclusion that $A$ is not just rare, but
impossible. 
If there are reasons to expect beforehand that $A$ is indeed possible, it is
important to inject this prior belief in the procedure, which suggest using  Bayesian estimation methods.

The main assumption for these methods is to consider the unknown
probability, $p=\myP(A)$, as a random variable, yielding a generative
process in which a random probability is first obtained, and then $N$
instances of $A$ or not-$A$ are generated using this probability.

Assume that the ``prior distribution'' of $p$ (which determines a prior belief) has a p.d.f. $q$ (with respect to Lebesgue's measure) on the unit interval. Given on
$N$ independent observations of occurrences of $A$, each following a
Bernoulli distribution $b(p)$, the joint likelihood of all involved variables is given
by
\[
 \binom{N}{k} p^k(1-p)^{N-k} q(p),
\]
where $k$ is the number of times the event $A$ has been observed.

The conditional density of $p$ given the observation ($k$ occurrences
of $A$) is called the {\em posterior distribution}. Here, it is given
by 
$$
q(p\mid k) = \frac{q(p)}{C_k} p^k(1-p)^{N-k}
$$
where $C_k$ is a normalizing constant.
If there was no specific prior knowledge on $p$ (so that $q(p)=1$),
the resulting distribution is a beta distribution with parameters $k+1$
and $N-k+1$, the beta distribution being defined as follows.

\begin{definition}
\label{def:beta}
The beta distribution with parameters $a$ and $b$ (abbreviated
$\beta(a,b)$) has density with respect to Lebesgue's measure
$$
\rho(t) = \frac{\Ga(a+b)}{\Ga(a)\Ga(b)} t^{a-1}(1-t)^{b-1} \text{ if } t\in [0,1]
$$
and $\rho(t) = 0$ otherwise, with
$$
\Ga(x) = \int_0^\infty t^{x-1} e^{-t} dt.
$$
\end{definition}

From the definition of a beta distribution, it is clear also that, if
we choose the prior to be $\be(a+1,\nu-a+1)$ then the
posterior is $\be(k+a+1, N+\nu -(k+a) +1)$. The posterior therefore
belongs to the same family of distributions as the prior, and one says
that the beta distribution is a {\em conjugate prior} for
the binomial distribution. The mode of the posterior distribution
(which is the maximum a posteriori (MAP) estimator) is given by
$$
\hat p = \frac{k+a}{N+\nu}.
$$

This estimator now provides a positive value even if $k=0$. By
selecting $a$ and $\nu$, one therefore includes the prior belief
that $p$ is positive.

\subsection{Learning a Finite Probability Distribution}

Now assume that $F$ is a finite space and that we want to estimate a
probability distribution $p = (p(x), x\in F)$ using a Bayesian approach as
above. We cannot use the previous approach to estimate each $p(x)$ separately,
since these probabilities are linked by the fact that they sum to 1. We
can however come up with a good (conjugate) prior, identified, as done above, by computing
the posterior associated to a uniform prior distribution.

Letting $N_x$ be the number of times $x\in F$ is observed among $N$
independent samples of a random variable $X$ with distribution $P_X(\cdot)=p(\cdot)$,
the joint distribution of $(N_x,x\in F)$ is multinomial, given by
$$
\myP(N_x, x\in F\mid p(\cdot)) = \frac{N!}{\prod_{x\in F} N_x!} \prod_{x\in F}
p(x)^{N_x}.
$$
The posterior distribution of $p(\cdot)$ given the observations with a
uniform prior is proportional to $\prod_{x\in F}
p(x)^{N_x}$. It belongs to the family of {\em Dirichlet distributions},
described in the following definition.
\begin{definition}
\label{def:dirich}
Let $F$ be a finite set and $\CS_F$ be the simplex defined by
$$
\CS_F = \defset{(p(x), x\in F): p(x)\geq 0, x\in F \text{ and } \sum_{x\in F}
p(x) = 1}.
$$
The Dirichlet distribution with parameters $a = (a(x), x\in F)$
(abbreviated $\mathrm{Dir}(a)$) has density
 $$
\rho(p(\cdot)) = \frac{\Ga(\nu)}{\prod_{x\in F} \Ga(a(x))} \prod_{x\in F}
p(x)^{a(x)-1}, \text{ if } x\in \CS_F
$$
and 0 otherwise, with $\nu = \sum_{x\in F} a(x)$.
\end{definition}
Note that, if $F$ has cardinality 2, the Dirichlet distribution
coincides with the beta distribution. Similarly to the beta for the
binomial, and almost by construction, the Dirichlet distribution is a
conjugate prior for the multinomial. More precisely, if the prior
distribution for $p(\cdot)$ is $\mathrm{Dir}(1+a(x), x\in F)$, then the
posterior after $N$ observations of $X$ is $\mathrm{Dir}(1+N_x+a(x),
x\in F)$, and the MAP estimator is given by
$$
\hat p(x) = \frac{N_x+a(x)}{N+\nu}
$$
with $\nu= \sum_{x\in F} a(x)$. 

\subsection{Conjugate Prior for Bayesian Networks}
We now consider a Bayesian network on the set $\CF(V)$ containing configurations $x = (\pe xs, s\in V)$ with $\pe xs\in F_s$.
We want to estimate the conditional probabilities
in the representation
$$
\myP(X=x) = \prod_{s\in V} p_s(\pe  x {\pa{s}}, \pe xs).
$$
Assume that $N$ independent observations of $X$ have been made. Define
the counts $N_s(\pe xs,  \pe x{\pa{s}})$ to be the number of times the observation
$\pe x{\{s\}\cup \pa{s}}$ has been made. Then, it is straightforward to see
that, assuming a uniform prior for the $p_s$, their posterior
distribution is proportional to
$$
\prod_{s\in V} \prod_{\pe x{\pa{s}}\in F_{\pa{s}}} \prod_{\pe xs\in F_s} p_s(
\pe x{\pa{s}}, \pe xs)^{N_s(\pe xs, \pe x{s^{-}})}.
$$

This implies that, for the posterior distribution, the conditional
probabilities $p_s(\pe x{\pa{s}}, \cdot)$ are independent and follow a Dirichlet
distribution with parameters $1+N_s(\pe xs, \pe x{\pa{s}})$, $\pe xs\in F_s$. 

So, independent Dirichlet distributions indexed by configurations of
parents of nodes provide a conjugate prior for the general Bayesian network
model. This prior is specified by a family of positive numbers
\begin{equation}
\label{eq:bayes.dir.all}
\left(a_s(\pe xs, \pe x{\pa{s}}), s\in V, \pe xs\in F_s, \pe x{\pa{s}}\in \CF(\pa{s})\right),
\end{equation}
yielding a prior probability proportional to 
$$
\prod_{s\in V} \prod_{\pe x{\pa{s}}\in F_{\pa{s}}} \prod_{\pe xs\in F_s} p_s(\pe
x{\pa{s}}, \pe xs)^{a_s(\pe xs, \pe x{s^{-}})-1}.
$$
and a MAP estimator
\begin{equation}
\label{eq:map.bn}
\hat p_s(\pe x{\pa{s}}, \pe xs) = \frac{N_s(\pe xs, \pe x{\pa{s}}) + a_s(\pe xs,
\pe x{s^{-}})}{N_s(\pe x {s^{-}}) + \nu_s(\pe x{s^{-}})}
\end{equation}
where $N_s(\pe x{\pa{s}}) = \sum_{\pe xs\in F_s} N_s(\pe xs, \pe x{\pa{s}})$ and
$\nu_s(\pe x{\pa{s}}) = \sum_{\pe xs\in F_s} a_s(\pe xs, \pe x{\pa{s}})$.

One can restrict the huge class of coefficients described by
\cref{eq:bayes.dir.all} to a smaller class by imposing the following
 condition. 
\begin{definition}
\label{def:a.consist}
One says that the family of coefficients
\[
a = (a_s(\pe xs, \pe x{\pa{s}}), s\in V, \pe xs\in F_s, \pe x{\pa{s}}\in \CF(\pa{s})),
\]
is consistent if there exists a positive scalar $\nu$ and a probability
distribution $P'$ on $\CF(V)$ such that
\[
a_s(\pe xs, \pe x{\pa{s}})= \nu P'_{\{s\}\cup \pa{s}}(\pe x{\{s\}\cup \pa{s}}).
\]
\end{definition}
The class of products of Dirichlet distributions with consistent families of coefficients still provides a
conjugate prior for Bayesian networks (the proof being left to the
reader). Within this class, the simplest choice (and most natural in the
absence of additional information) is to assume
that $P'$ is uniform, so that
\begin{equation}
\label{eq:prior}
a_s(\pe xs, \pe x{\pa{s}}) = \frac{\nu'}{|\CF(\{s\}\cup \pa{s})|}.
\end{equation}
With this choice, $\nu'$ is the only parameter that needs to be
specified. It is often called the equivalent sample size for the prior
distribution.

We can see from \cref{eq:map.bn} that using a prior distribution is
quite important for Bayesian networks, since, when the number of
parents increases, some configurations on $\CF(\pa{s})$ may not be
observed, resulting in an undetermined value for the ratio 
\[
N_s(\pe xs, \pe x{\pa{s}})/N_s(\pe x{s^{-}}), 
\]
even though, for the estimated model, the
probability of observing $\pe x{\pa{s}}$ may not be zero.

\subsection{Structure Scoring}
Given a prior defined as a family of Dirichlet distributions associated to $a = (a_s(\pe xs, \pe x{\pa{s}})$ for $s\in
V, \pe xs\in F_s, \pe x{\pa{s}}\in \CF(\pa{s})$, the joint density of the
observations and parameters is given by
\[
P( x, \th) = \prod_{s, \pe x{\pa{s}}} \mathcal D(a_s(\cdot, \pe x{\pa{s}}))
\prod_{s, \pe xs, \pe x{\pa{s}}} p(\pe x{\pa{s}}, \pe xs)^{N_s(\pe xs, \pe x{\pa{s}}) +
  a_s(\pe xs, \pe x{\pa{s}}) -1}
\] 
with 
\[
\mathcal D(a(\la), \la\in F) = \frac{\Ga(\nu)}{\prod_\la \Ga(a(\la))}
\]
and $\nu = \sum_\la a(\la)$. Here, $\th$ represents the parameters of the model, i.e., the conditional distributions
that specify the Bayesian network. Note that $P(x, \theta)$ is a density over the product space $\CF(V) \times \Theta$ where $\Theta$ is the space of all these conditional distributions.
The marginal of this likelihood over all
possible parameters, i.e., 
\[
P( x) = \int P(\mathbf x, \th) d\th
\]
provides the expected likelihood of the sample relative to the
distribution of the parameters, and only depends on the structure of
the network. In our case, integrating with respect to $\th$ yields
\[
\ln P( x) = \sum_{s, x_{\pa{s}}} \ln \frac {\mathcal D(a_s(\cdot,
\pe x{\pa{s}}))}{\mathcal D(a_s(\cdot,
\pe x{\pa{s}}) + N_s(\cdot, \pe x{\pa{s}}))}.
\]
Letting 
\[
\ga(s, \pa{s}) = \sum_{\pe x{\pa{s}}} \ln \frac {\mathcal D(a_s(\cdot,
\pe x{\pa{s}}))}{\mathcal D(a_s(\cdot,
\pe x{\pa{s}}) + N_s(\cdot, \pe x{\pa{s}}))},
\]
the decomposition
\[
\ln P( x) = \sum_{s\in V} \ga(s, \pa{s})
\]
expresses this likelihood as a sum of ``scores'' (associated to each node
and its parents), which depends on the observed sample. The scores
that are computed above are often called {\em Bayesian scores} because
they derive from a Bayesian construction. One can also consider 
simpler scores, such as penalized likelihood:
\[
\ga(s, \pa{s}) = - \sum_{\pe x{\pa{s}}} \hat {\CH}(\pe X s\mid \pe X{\pa{s}})\, |\CF(\pa{s})| - \rho |\pa{s}|,
\]  
where $\hat {\CH}$ is the conditional entropy for the empirical
distribution based on observed samples. Structure learning algorithms \citep{neapolitan2004learning,koller2009probabilistic} are designed to optimize such scores.

\subsection{Reducing the Parametric Dimension}

In the previous section, we estimated all  conditional
probabilities intervening in the network. This is obviously a lot of
parameters and, even with a regularizing prior, the estimated values
are likely to be be inaccurate for small sample sizes. It then becomes desirable to simplify the parametric complexity of the
model. 

When the sets $F_s$ are not too large, which is common in
practice, the parametric explosion is due to the multiplicity of
parents, since the number of conditional probabilities $p_s(\pe x{\pa{s}}, \cdot)$
grows exponentially with $|\pa{s}|$. One way to simplify this is to assume that the conditional
probability at $s$ only depends on $\pe x{\pa{s}}$ via some ``global-effect''
statistic $g_s(\pe x{\pa{s}})$. The idea, of course, is that the number of
values taken by $g_s$ should remain small, even if the number of parents is
large.

Examples of some functions $g_s$ can be $\max(\pe xt, t\in \pa{s})$, or the
min, or some simple (quantized) function of the sum. With binary
variables ($F_s=\{0,1\}$), logical operators are also available (``and'',
``or'', ``xor''), as well as combinations of them. The choice made for the
functions $g_s$ is part of building the model, and would rely on
the specific context and prior information on the process, which
is always important to account for, in any statistical problem.

Once the $g_s$'s are fixed, learning the network distribution, which is
now given by
$$
\pi(x) = \prod_{s\in V} p_s(g_s(\pe x{\pa{s}}), \pe xs)
$$
can be done exactly as before, the parameters being all $p_s(w, \lambda),
\la\in F_s, w\in W_s$, where $W_s$ is the range of $g_s$, and
Dirichlet priors can be associated to each $p_s(w, \cdot)$ for $s\in V$ and
$w\in W_s$. The counts provided in \cref{eq:prior} now can be chosen
as
\begin{equation}
\label{eq:prior2}
a_s(x_s, w) =  \frac{\nu'}{|F|\,|g_s^{-1}(w)|}.
\end{equation}

\section{Learning Loopy Markov Random Fields}

Like everything else, parameter estimation for loopy networks is much
harder than with trees or Bayesian networks.  There is usually no
closed form expression for the estimators, and their computation relies
on more or less tractable numerical procedures.

\subsection{Maximum Likelihood with Exponential Models}
In this section, we consider a parametrized model for a Gibbs
distribution 
\begin{equation}
\label{eq:exp.m}
\pi_\th(x) = \frac{1}{Z_\th} \exp(-\th^TU(x))
\end{equation}
where $\th$ is a $d$-dimensional parameter and $U$ is a function from
$\CF(V)$ to $\mR^d$. For example, if $\pi$ is an Ising model with
\[
\pi(x) = \frac{1}{Z} \exp\big( \al\sum_{s\in V} \pe xs + \be \sum_{s\sim t}
\pe xs \pe xt\big),
\]
then $\th = (\al, \be)$ and $U(x) = - (\sum_s \pe xs,
\sum_{s\sim t} \pe xs \pe xt)$. Most of the Markov random fields models that
are used in practice can be put in this form. The constant $Z_\th$ in \cref{eq:exp.m} is 
$$
Z_\th  = \sum_{x\in \CF(V)} \exp(-\th^TU(x))
$$
and is usually not computable.

Now, assume that an $N$-sample, $x_1, \ldots, x_N$, is observed for
this distribution. The maximum likelihood estimator maximizes
$$
\ell(\th) = \frac{1}{N} \sum_{k=1}^N \ln \pi_\th(x_k) = -\th^T \bar
U_N - \ln Z_\th
$$
with $\bar U_N = (U(x_1)+\cdots+U(x_N))/N$. 

We have the following proposition, which is a well-known property
of exponential families of probabilities.
\begin{proposition}
\label{prop:ml.exp}
The log-likelihood, $\ell$, is a concave function of $\th$, with
\begin{equation}
\label{eq:ml.exp.1}
\nabla{\ell}(\th) = E_\th(U) - \bar U_N
\end{equation}
and
\begin{equation}
\label{eq:ml.exp.2}
\nabla^2\ell(\th) = - \text{Var}_\th(U)
\end{equation}
where $E_\th$ denotes the expectation with respect to $\pi_\th$ and
$\text{Var}_\th$ the covariance matrix under the same distribution.
\end{proposition}

We skip the proof, which is just computation. 
This proposition implies that a local  maximum of $\th \mapsto
\ell(\th)$ must also be global. Any such maximum must be a solution of
$$
E_\th(U) = \bar U_N(x_0)
$$
and conversely.
There are some situations in which the maximum does not exist, or is
not unique. Let us first discuss the second case.

If several solutions exist, the log-likelihood cannot be strictly
concave: there must exist at least one $\th$ for which
$\text{Var}_\th(U)$ is not definite. This implies that there exists a nonzero
vector $u$ such that
$\text{var}_\th(u^TU)= u^T \text{Var}_\th(U)u=0$. This is only
possible when $u^TU(x) = \text{cst}$ for all $x\in F_V$. Conversely, if this
is true, $\text{Var}_\th(U)$ is degenerate {\em for all $\th$}.

So, the non-uniqueness of the solutions is only possible when a
deterministic affine relation exists between the components of $U$,
i.e., when the model is over-dimen\-sioned. Such situations are usually
easily dealt with by removing some parameters. In all other cases, there exists
at most one maximum.

For a concave function like $\ell$ to have no maximum, there must exist what is
called a {direction of recession} \citep{rockafellar1970convex}, which is a  direction
$\alpha\in\mR^d$ such that, for all $\th$, the function
$t\mapsto \ell(\th+t\al)$ is increasing. In this case the maximum
is attained ``at infinity''. Denoting $U_\al(x) = \al^T U(x)$, the
derivative in $t$ of $\ell(\th+t\al)$ is
\[
E_{\th+t\al}(U_\al) - \bar U_\al
\]
where $\bar U_\alpha = \alpha^T \bar U_N$.
This derivative is positive for all $t$ if and only if
\begin{equation}
\label{eq:recess}
\bar U_\al = U_\al^* := \min\{U_\al(x), x\in \CF(V)\}
\end{equation}
and $U_\al$ is not constant. 
To prove this, assume that the derivative is positive. Then $U_\al$ is
not constant (otherwise, the derivative would be zero). Let $\CF_\al^*\sub \CF(V)$ be the set of configurations $x$
for which $U_\al(x) = U_\al^*$. Then
\begin{align*}
&E_{\th+t\al}(U_\al)\\
 &= \frac{\sum_{x\in \CF(V)} U_\al(x)
\exp(-\th^TU(x) - tU_\al(x))} {\sum_{x\in \CF(V)}
\exp(-\th^T U(x) - tU_\al(x))}\\
&= \frac{\sum_{x\in \CF(V)} U_\al(x)
\exp(- \th^T U(x) - t(U_\al(x)-U_\al^*))} {\sum_{x\in \CF(V)}
\exp(- \th^TU(x) - t(U_\al(x)-U_\al^*))}\\
&= \frac{U_\al^* \sum_{x\in \CF_\al^*} \exp(- \th^TU(x))
+ \sum_{x\not\in \CF_\al^*} U_\al(x) \exp(- \th^TU(x) -
t(U_\al(x)-U_\al^*))} { \sum_{x\in \CF_\al^*} \exp(- \th^TU(x)) +
\sum_{x\not\in \CF_\al^*} \exp(-\th^TU(x) - t(U_\al(x)-U_\al^*))}.
\end{align*}
When $t$ tends to $+\infty$, the sums over $x\not\in
\CF_\al^*$ tend to 0, which implies that 
$E_{\th+t\al}(U_\al)$ tends to $U_\al^*$. So, if  
$E_{\th+t\al}(U_\al) - \bar U_\al > 0$ for all $t$, then $\bar U_\al =
U_\al^*$ and $U_\al$ is not constant. The converse statement is obvious.

As a conclusion, the function $\ell$ has a finite maximum if and only
if there is no direction $\alpha\in \mR^d$ such that $\al^T(U(x) -
\bar U_N) \leq 0$ for all $x\in \CF(V)$. Equivalently, $\bar U_N$ must
belong to the interior of the convex hull of the finite set
$$
\defset{U(x), x\in \CF(V)}\sub \mR^d.
$$

In such a case, that we hereafter assume, computing the maximum likelihood estimator boils down to
solving the equation
$$
E_\th (U) = \bar U_N.
$$
Because the maximization problem is concave, we know that numerical
algorithms such as gradient ascent, 
\begin{equation}
\label{eq:grd.des}
\th(t+1) = \th(t) + \ep (E_{\th(t)} (U) - \bar U_N),
\end{equation}
converge to the optimal parameter. Unfortunately, the computation of
the expectations and covariance matrices can only be made explicitly
for acyclic models, for which parameter estimation is not a problem
anyway. For general loopy graphical models, the expectation can be
estimated iteratively using Monte-Carlo methods. It turns out that
this estimation can be synchronized with gradient descent to obtain a
consistent algorithm, which is described in the next section.

\subsection{Maximum likelihood with stochastic gradient ascent}
\label{sec:stoc.grad}

As remarked above, for fixed $\th$, we have designed, in \cref{chap:inf.mrf}, Markov chain Monte
Carlo algorithms that asymptotically sample form $\pi_\th$. Select one
of these algorithms, and let $ p_\th$ be the corresponding
 transition probabilities for a given $\th$, so that $p_\theta(x,y) = \myP(X_{n+1} = y \mid X_n=x)$ for the sampling chain. Then, define the iterative
algorithm, initialized with arbitrary $\th_0$ and $x_0 \in \CF(V)$,
that loops over the following two steps.
\begin{enumerate}[label=(SG\arabic*)]
\item Sample from the distribution $ p_{\th_t}(x_t,
\cdot)$ to obtain a new configuration $x_{t+1}$.
\item Update the parameter using
\begin{equation}
\label{eq:stoc.grad}
\th_{t+1} = \th_t + \ga_{t+1} (U(x_{t+1}) - \bar U_N).
\end{equation}
\end{enumerate}

This algorithm differs from the situation considered in \cref{sec:sgd} in that the distribution of the sampled variable $x_{t+1}$ depends on both the current parameter $\theta_t$ and on the current variable $x_t$. Convergence requires additional constraints on the size of the gains $\gamma(t)$ and we have the following theorem \citep{you88}.
\begin{theorem}
\label{th:stoc.grad}
If $p_\th$ corresponds to the Gibbs sampler or Metropolis
algorithm, and $\ga_{t+1} = \ep/(t+1)$ for small enough $\ep$, the
algorithm that iterates (SG1) and (SG2) converges almost surely to the maximum
likelihood estimator.
\end{theorem}

The speed of convergence of such algorithms depends both on the speed
of convergence of the Monte-Carlo sampling and of the original
gradient ascent. The latter can be improved somewhat with variants similar to those discussed in \cref{sec:sgd}, for example by choosing data-adaptive gains as in the ADAM algorithm. 

\subsection{Relation with Maximum Entropy}

The maximum likelihood estimator is closely related to what is called
the maximum entropy extension of a set of constraints. Let the
function $U$ from $\CF(V)$ to $\mR^d$ be given. An element $u\in\mR^d$ is
said to be a consistent assignment for $U$ if there exists a
probability distribution $\pi$ on $\CF(V)$ such that $E_\pi(U) = u$. An
example of consistent assignment is any empirical average $\bar U$
based on a sample $(x^{(1)}, \ldots, x^{(N)})$, since $\bar U =
E_\pi(U)$ for
\[
\pi = \frac1N\sum_{k=1}^N \de_{x^{(k)}}.
\]

Given $U$ and a consistent assignment, $u$,   the associated maximum
entropy extension is defined as a probability distribution $\pi$
maximizing the entropy, $\CH(\pi)$, subject to the constraint $E_\pi(U)
= u$. This is a convex optimization problem, with constraints
\begin{equation}
\label{eq:max.ent}
\left\{
\begin{aligned}
&\sum_{x\in \CF(V)} \pi(x) = 1 \\
&\sum_{x\in \CF(V)} U_j(x) \pi(x) = u_j, j=1, \ldots, d\\
&\pi(x) \geq 0, x\in \CF(V)
\end{aligned}\right. 
\end{equation}
Because the entropy is strictly convex, there is a unique solution to
this problem. We first discuss non-positive solutions, i.e., solutions
for which $\pi(x) = 0$ for some $x$. An important fact is that, if,
for a given $x$, there exists $\pi_1$ such that $E_{\pi_1}(U) = u$ and
$\pi_1(x) > 0$, then the optimal $\pi$ must also satisfy $\pi(x) >
0$. This is because, if $\pi(x) = 0$, then, letting $\pi_\ep = (1-\ep)
\pi + \ep \pi_1$, we have $E_{\pi_\ep}(U) = u$ since this constraint
is linear, $\pi_\ep(x) > 0$ and
\begin{eqnarray*}
H(\pi_\ep) - H(\pi) &=& -\sum_{y, \pi(y) > 0} (\pi_\ep(y) \ln\pi_\ep(y)
- \pi(y) \ln\pi(y)) \\
&&-\sum_{y, \pi(y) = 0} \ep \pi_1(y) (\ln(\ep) + \ln \pi_1(y)) \\
&=& -\ep \ln\ep \sum_{y, \pi(y) =0} \pi_1(y) + O(\ep)
\end{eqnarray*}
which is positive for small enough $\ep$, contradicting the fact that
$\pi$ is a maximizer. 

Introduce the set $\mathcal N_u$ containing all configurations $x\in
\CF(V)$ such that $\pi(x)=0$ for all $\pi$ such that $E_\pi(U) = u$. Then
we know that the maximum entropy extension satisfies $\pi(x) > 0$
if $x\not\in \mathcal N_u$. Introduce Lagrange multipliers $\th_0, \th_1, \ldots, \th_d$ for the
$d+1$ equality constraints in \cref{eq:max.ent}, and the Lagrangian
\[
L = H(\pi) +  \sum_{x\in \CF(V)\setm \CN_u} (\th_0 +
\th^TU(x))\pi(x) 
\]
in which we have set $\th = (\th_1, \ldots, \th_d)$, we find that the
optimal $\pi$ must satisfy
\[
\left\{
\begin{aligned}
\ln \pi(x) = -\th_0 - 1 -\th^TU(x)\\
\sum_x \pi(x) = 1 \\
E_{\pi}(U) = \bar u\\
\end{aligned}
\right.
\]
In other terms, the maximum entropy extension is characterized by
\[
\pi(x) = \frac{1}{Z_\th} \exp(-\th^T U(x))\bfone_{\CN_u^c}(x)
\]
and $E_\pi(U) = u$. 

In particular, if $\CN_u = \emp$, then the maximum entropy extension
is positive. If, in addition, $u = \bar U$ for some observed sample,
then it coincides with the maximum likelihood estimator for
\cref{eq:exp.m}. Notice that, in this case,  the condition $\CN_u\neq\emp$ coincide
with the  condition that there exists $\al$ such that $\al^T U(x) \geq
\al^T u$ for all $x$, with $\al^T U(x)$ not constant. Indeed, assume
that the latter condition is true. Then, if $E_\pi(U) = u$, then $E_\pi(\al^T U) =
\al^T u$, which is only possible if $\pi(x) = 0$ for all $x$ such
that $\al^T U(x) < \al^T u$. Such $x$'s exist by assumption, and
therefore $\mathcal N_{u}\neq \emp$. Conversely, assume $\CN_u\neq
\emp$. If condition \cref{eq:recess} is not satisfied, then we have
shown when discussing maximum likelihood that an optimal parameter for
the exponential model would exist, leading to a positive distribution
for which $E_\pi(U) = u$, which is a contradiction.  
 
\subsection{Iterative Scaling}
\label{sec:iter:scal}

Iterative scaling is a method that is well-adapted to learning
distributions given by \cref{eq:exp.m}, when $U$ can be interpreted
as a random histogram, or a collection of them. 
More precisely, assume that for all $x\in \CF(V)$, one has 
\[
U(x) =
(U_1(x), \ldots, U_q(x))\]
 with 
\[
\sum_{j=1}^q U_j(x) = 1 \text{ and } U_j(x) \geq 0.
\]
Let the parameter be given by $\th = (\th_1, \ldots, \th_q)$. Assume
that $x_1, \ldots, x_N$ have been observed, and let $u\in
\mR^d$ be a consistent assignment for $U$, with $u_j>0$ for $j=1,
\ldots, d$ and such that $\CN_u = \emp$. Iterative scaling computes the
maximum entropy extension of $E_\pi(U) = u$, that we will denote
$\pi^*$. It is supported by the
following lemma. 
\begin{lemma}
\label{lem:iter.sca}
Let $\pi$ be a probability on $\CF(V)$ with $\pi >0$ and define
\[
\pi'(x) = \frac{\pi(x)}{\ze} \prod_{j=1}^d \left(\frac{u_j}{E_\pi(
    U_j)}\right)^{U_j(x)}
\]
where $\ze$ is chosen so that $\pi'$ is a probability. Then $\pi' > 0$ and
\begin{equation}
\KL(\pi^*\|\pi') - \KL(\pi^*\|\pi) \leq -\KL(u\|E_\pi(U))\leq 0
\label{eq:lem.iter.sca}
\end{equation}
\end{lemma}
\begin{proof}
Note that, since $\pi >0$,  $E_\pi(U_j)$ must also be positive for all
$j$, since $E_\pi(U_j) = 0$ would otherwise imply $U_j=0$ and $u_j =
0$ for $u$ to be consistent. So, $\pi'$ is well defined and obviously
positive.

We have
\begin{eqnarray*}
\KL(\pi^*\|\pi') - \KL(\pi^*\|\pi) &=& \ln\ze - \sum_{x\in \CF(V)}
\pi^*(x) \sum_{j=1}^d U_j(x) \ln \frac{u_j}{E_\pi(
    U_j)} \\
&=&  \ln\ze - \sum_{j=1}^d u_j \ln \frac{u_j}{E_\pi(
    U_j)} \\
&=& \ln\ze -  \KL(u\|E_\pi(U)).
\end{eqnarray*}
(We have used the identity $E_{\pi^*}(U) = u$.)
So it suffices to prove that $\ze \leq 1$. We have
\begin{eqnarray*}
\ze &=& \sum_{x\in \CF(V)} \pi(x) \prod_{j=1}^d \left(\frac{u_j}{E_\pi(
    U_j)}\right)^{U_j(x)} \\
&\leq & \sum_{j=1}^d \sum_{x\in\CF(V)} \pi(x) U_j(x) \frac{u_j}{E_\pi (U_{j})}\\
&= & \sum_{j=1}^d E_\pi(U_{j}) \frac{u_j}{E_\pi(U_j)} = 1,
\end{eqnarray*}
which proves the lemma (we have used the fact that, for $x_i$, $w_i$ positive numbers with $\sum_i w_i=1$, one has $\prod_i x_i^{w_i} \leq \sum_i w_i x_i$, which is a consequence of the concavity of the logarithm).
\end{proof}

Consider the iterative algorithm
\[
\pi_{n+1}(x) = \frac{\pi_n(x)}{\ze_n} \prod_{j=1}^d \left(\frac{u_j}{E_{\pi_n}(
    U_j)}\right)^{U_j(x)}
\]
initialized with a uniform
distribution. Equivalently, using the exponential formulation, define,
for $j=1, \ldots, d$,
\begin{equation}
\label{eq:iter.sca.th}
\th_{n+1,j} = \th_{n,j} + \ln \frac{E_{\th_n} (U_j)}{u_j} + \KL(u\|E_{\th_n}(U)),
\end{equation}
with $\pi_\th$ given by \cref{eq:exp.m}, initialized with  $\th_0 =
0$.  Note that adding a term that is independent of $j$ to $\th$
does not change the value of $\pi_\th$, because the $U_j$'s sum to
1. The model is in fact overparametrized, and the addition of the KL
divergence in \cref{eq:iter.sca.th} ensures that $\sum_{i=1}^d u_i\th_i
= 0$ at all steps. 

This algorithm always 
reduces the Kullback-Leibler distance to the maximum entropy
extension. This distance  being always positive, it therefore converges to
a limit, which, still according to  \cref{lem:iter.sca}, is only
possible if $\KL(u\|E_{\pi_n}(U))$ also tends to 0, that is
$E_{\pi_n}(U) \to u$. Since the
space of probability distributions is compact, the Heine-Borel theorem
implies that the sequence $\pi_{\th_n}$ has at least one accumulation
point, that we now identify. If $\pi$ is such a point, one must have $E_{\pi}(U) =
u$. Moreover, we have $\pi >0$, since otherwise $\KL(\pi^*\|\pi) =
+\infty$. To prove that $\pi = \pi^*$ (and therefore the limit of the
sequence), it remains to show that it can be put in the form
\cref{eq:exp.m}. For this, define the vector space $\CV$ of functions $v:
\CF(V) \to \mR$ which can be written in the form
\[
v(x) = \al_0 + \sum_{j=1}^g \al_j U_j(x).
\]
Since $\ln\pi_{\th_n}\in \CV$ for all $n$, so is its limit, and this proves that $\ln \pi$ belongs to
$\CV$. We have obtained the following proposition.

\begin{proposition}
\label{prop:iter.sca}
Assume that for all $x\in \CF(V)$, one has $U(x) =
(U_1(x), \ldots, U_d(x))$ with 
\[
\sum_{j=1}^d U_j(x) = 1 \text{ and } U_j(x) \geq 0.
\]
Let $u$ be a consistent assignment for the expectation of $U$ such
that$\CN_u = \emp$. Then, the algorithm described in  \cref{eq:iter.sca.th} converges
to the maximum entropy extension of $u$.
\end{proposition}

This is the iterative scaling algorithm. This method can be extended
in a straightforward  way to handle the maximum entropy extension for
a family of functions $U^{(1)}, \ldots, U^{(K)}$, such that, for all
$x$ and for all $k$, $U^{(k)}(x)$ is a $d_k$-dimensional vector such
that
\[
\sum_{j=1}^{d_k} U^{(k)}_j(x) = 1.
\]
The maximum entropy extension takes the form
\[
\pi_\th(x) = \frac1{Z_\th} \exp\Big(-\sum_{k=1}^K
(\th^{(k)})^T U^{(k)}(x)\Big),
\]
where $\th^{(k)}$ is $d_k$-dimensional, and iterative scaling can then
be implemented by updating only one of these vectors at a time, using
\cref{eq:iter.sca.th} with $U = U^{(k)}$.

The restriction to $U(x)$ providing a discrete probability distribution
for all $x$ is, in fact, no loss of generality. This is because adding
a constant to $U$ does not change the resulting exponential model in
\cref{eq:exp.m}, and multiplying $U$ by a constant can be also
compensated by dividing $\th$ by the same constant in the same
model. So, if $u_-$ is a lower bound for $\min_{j, x} U_{j}(x)$, one
can replace $U$
by $(U-u_-)$, and therefore assume that $U \geq 0$, and if $u_+$ is an
upper bound for $\sum_j U_j(x)$, we can replace $U$ by $U/u_+$ and
therefore assume that $\sum_j U_j(x) \leq 1$. Define 
\[
U_{d+1}(x) = 1 - \sum_{j=1}^d U_j(x) \geq 0.
\]
Then, the maximum entropy extension for $(U_1, \ldots, U_d)$ with
assignment $(u_1, \ldots, u_d)$ is obviously also the extension for
$(U_1, \ldots, U_{d+1})$, with assignment $(u_1, \ldots, u_{d+1})$,
where
\[
u_{d+1} = 1 - \sum_{j=1}^d u_j,
\]
and the latter is in the form required in  \cref{prop:iter.sca}. Note that iterative scaling requires to compute the expectation of $U_1, \ldots, U_d$ before each update. These are not necessarily available in closed form and may have to be estimated using Monte-Carlo sampling.

\subsection{Pseudo likelihood}
\label{sec:pseudo.lik}
Maximum likelihood estimation is a special case of {\em minimal
contrast estimators}. These estimators are based on the definition of
a measure of dissimilarity, say $C(\pi \| \tilde \pi)$, between two probability distributions $\pi$
and $\tilde \pi$. The usual assumptions on $C$ are
that $C(\pi \| \tilde \pi) \geq 0$, with equality if and only if $\pi=
\tilde \pi$, and that $C$ is --- at least --- continuous in $\pi$ and
$\tilde\pi$. 
Minimal contrast estimators approximate the problem of
minimizing $\th \mapsto C(\pi_{\mathrm{true}}\| \pi_\th)$ over a parameter $\th\in \Theta$,
(which is not feasible, since $\pi_{\mathrm{true}}$, the true
distribution of the data, is unknown) by the minimization of
$\th\mapsto C(\hat \pi\| \pi_\th)$ where $\hat \pi$ is the empirical
distribution computed from  observed data. Under mild conditions on
$C$, these estimators are generally consistent when $N$ tends to
infinity, which means that the estimated parameter asymptotically (in
the sample size $N$) provides the
best (according to $C$) approximation of $\pi_{\mathrm{true}}$ by the family $\pi_\th, \theta\in\Theta$.

The contrast that is associated with maximum likelihood is the
Kullback-Leibler divergence.
Indeed, given a sample $x_1, \ldots, x_N$, we have
\begin{eqnarray*}
\KL(\hat \pi\|\pi_\th) &=& E_{\hat \pi}\ln\hat \pi - E_{\hat \pi} \ln \pi_\th\\
&=& E_{\hat \pi}\ln\hat \pi - \sum_{k=1}^N \ln \pi_\th(x_k).
\end{eqnarray*}
Since $E_{\hat \pi}\ln\hat \pi$ does not depend on $\th$, minimizing
$\KL(\hat \pi\| \pi_\th)$ is equivalent to maximizing $\sum_{k=1}^N \ln
\pi_\th(x_k)$ which is the log-likelihood.

Maximum pseudo-likelihood estimators form another class of minimal
contrast estimators for graphical models. Given a distribution $\pi$
on $\CF(V)$, define the
local specifications
$\pi_{s}(\pe xs\mid \pe xt, t\neq s)$ to be conditional distributions at one
vertex given the others, and the contrast
\[
C(\pi\| \tilde \pi)  = \sum_{s\in V} E_\pi(\ln \frac{\pi_s}{\tilde \pi_s}).
\]
Because we can write, using standard properties of conditional expectations,
\[
C(\pi\| \tilde \pi) = \sum_{s\in V} E_\pi\left(E_{\pi_s}(\ln
\frac{\pi_s}{\tilde \pi_s})\right) 
= \sum_{s\in V} E(\KL(\pi_s(\cdot\mid \pe Xt,
t\neq s)\|\tilde \pi_s(\cdot \mid \pe Xt, t\neq s)),
\]
we see that $C(\pi, \tilde\pi)$ is always positive, and vanishes (under the assumption
of positive $\pi$) only if all the local specifications for $\pi$ and
$\tilde\pi$ coincide, and this can be shown to imply that
$\pi=\tilde\pi$. Indeed, for any $x, y\in \CF(V)$, and choosing some
order $V = \{s_1, \ldots, s_n\}$ on $V$, one can write
\[
\frac{\pi(x)}{\pi(y)} = \prod_{k=1}^n \frac{\pi(\pe x{s_k}|\pe x{s_1},
  \ldots, \pe x{s_{k-1}}, \pe y{s_{k+1}}, \ldots, \pe y{s_n})}
{\pi(\pe x{s_k}|\pe x{s_1},
  \ldots, \pe x{s_{k-1}}, \pe y{s_{k+1}}, \ldots, \pe y{s_n})}
\]
and the ratios $\pi(x)/\pi(y)$, for $x\in \CF(V)$, combined with the constraint that $\sum_x\pi(x)=1$ uniquely
define $\pi$.

So $C$ is a valid contrast and 
\[
C(\hat\pi\| \pi_\th) = \sum_{s\in V} E_{\hat\pi} \ln \hat \pi_s -
\sum_{s\in V} \sum_{k=1}^N \ln \pi_{\th,s} (x_k^{(s)}|x_k^{(t)},
t\neq s).
\]

This yields the {\em maximum pseudo-likelihood estimator} (or
pseudo maximum likelihood) defined as
a maximizer of the function (called log-pseudo-likelihood)
$$
\th \mapsto \sum_{s\in V} \sum_{k=1}^N \ln \pi_{\th,s} (x_k^{(s)}|x_k^{(s)},
t\neq s).
$$

Although maximum likelihood is known to provide the most accurate
approximations in many cases, maximum of pseudo likelihood has the
important advantage to be, most of the time, computationally
feasible. This is because, for a model like \cref{eq:exp.m}, local
specifications are given by 
$$
\pi_{\th,s}(\pe xs\mid \pe xt, t\neq s) = \frac{\exp(-\th^TU(x))}{\sum_{\pe ys\in
F_s} \exp(-\theta^T U(\pe ys\wedge \pe x{V\setm s}))}.
$$
and therefore include no intractable normalizing constant. Maximum of
pseudo-likelihood estimators can be computed using standard
maximization algorithms. For exponential
models such as \cref{eq:exp.m}, the log-pseudo-likelihood is, like the
log-likelihood, a concave function.

\subsection{Continuous  variables and score matching}
\label{sec:score.matching}
The methods that were presented so far for discrete variables formally generalize to more general state spaces, even though consistency or convergence issues in non-compact cases can be significantly harder to address. Score matching is a parameter estimation method that was introduced in \cite{hyvarinen2005estimation} and was designed, in its original version, to estimate parameters for statistical models taking the form
\[
\pi_\theta(x) = \frac{1}{C(\theta)}\exp\left(- F(x, \theta)\right)
\]
with $x\in \mR^d$. We assume below suitable integrability and differentiability conditions, in order to justify differentiation under integrals whenever they are needed.
The ``score function'' is defined as
\[
s(x, \theta) = - \nabla_x \log \pi_\theta(x) = \nabla_x F(x, \theta)
\]
where $\nabla_x$ denotes the gradient with respect to the $x$ variable. Letting $\pi_{\mathrm{true}}$ denote the p.d.f. of the true data distribution (not necessarily part of the statistical model), score matching minimizes
\[
f(\theta) = \int_{\mR^d} |s(x, \theta) - s_{\mathrm{true}}(x)|^2 \pi_{\mathrm{true}}(x) dx
\]
where $s_{\mathrm{true}} = - \nabla \log \pi_{\mathrm{true}}$.  This integral can be restricted to the support of $\pi_{\mathrm{true}}$, if we don't want to assume that $\pi_{\mathrm{true}}$ is non-vanishing. Note, however that $f(\theta) = 0$ implies that $\log \pi_\theta (\cdot, \theta) = \log \pi_{\mathrm{true}}$, $\pi_{\mathrm{true}}$-almost everywhere, so that $\pi_\theta(x) = c \pi_{\mathrm{true}}(x)$ for some constant $c$ and $x$ in the support of $\pi_{\mathrm{true}}$. Only if $\pi_{\mathrm{true}}(x) > 0$ for all $x\in \mR^d$, can we conclude that this requires $\pi_\theta = \pi_{\mathrm{true}}$. 

Expanding the squared norm and applying the divergence theorem yield
\begin{align*}
f(\theta) &=  \int_{\mR^d} |\nabla_x \log \pi_\theta(x)|^2\pi_{\mathrm{true}}(x) dx  - 2\int_{\mR^d} \nabla_x \log \pi_\theta(x)^T \nabla \pi_{\mathrm{true}}(x) dx \\
&+ \int_{\mR^d} | s_{\mathrm{true}}(x)|^2 \pi_{\mathrm{true}}(x) dx\\
 &=  \int_{\mR^d} |\nabla_x \log \pi_\theta(x)|^2\pi_{\mathrm{true}}(x)dx  + 2\int_{\mR^d} \Delta \log \pi_\theta(x)^T  \pi_{\mathrm{true}}(x) dx + \int_{\mR^d} | s_{\mathrm{true}}(x)|^2 dx
 \end{align*}
 To justify the use of the divergence theorem, one needs to assume two derivatives in the log-likelihoods with sufficient decay at infinity (see \citet{hyvarinen2005estimation} for details).  
 This shows that minimizing $f$ is equivalent to minimizing 
 \begin{align*}
 g(\theta) &= \int_{\mR^d} |\nabla_x \log \pi_\theta(x)|^2\pi_{\mathrm{true}}(x)dx  + 2\int_{\mR^d} \Delta \log \pi_\theta(x)^T  \pi_{\mathrm{true}}(x) dx\\
 & = \myE(|\nabla_x \log \pi_\theta(X)|^2  + 2\Delta \log \pi_\theta(X)).
 \end{align*}
 In this form, the objective function can be approximated by a sample average, so that, given observed data $x_1, \ldots, x_N$, one can define the score-matching estimator as a minimizer of 
 \begin{equation}
 \label{eq:scorematch.obj}
 \sum_{k=1}^N \left(|\nabla_x \log \pi_\theta(x_k)|^2  + 2\Delta \log \pi_\theta(x_k)\right).
 \end{equation}
 
 \begin{remark}
 \label{rem:score.discrete}
The method can be adapted to deal with discrete variables replacing derivatives with differences. Let $X$ take values in a finite set, $\CR_X$, on which a graph structure can be defined, writing $x\sim y$ if  $x$ and $y$ are connected by an edge. For example, if $X$ is itself a Markov random field on a graph $\CG = (V, E)$, so that $\CR_X = \CF(V)$, one can define $x\sim y$ if and only if $\pe x s = \pe y s$ for all but one $s\in V$. One can then define the score function
\[
s_\theta(x, y) = 1 - \frac{\pi_\theta (y)}{\pi_\theta(x)}
\]
defined over all $x, y\in \CR_X$ such that $x\sim y$. Now the score matching functional is  
\begin{align*}
f(\theta) &= \sum_{x\in \CR_X} \left(\sum_{y\sim x} |s_\theta(x,y) - s_{\mathrm{true}}(x,y)|^2\right) \pi_{\mathrm{true}}(x)\\
& = \sum_{x\in \CR_X} \sum_{y\sim x} s_\theta(x,y)^2 \pi_{\mathrm{true}}(x) - 2 \sum_{x\in \CR_X} \sum_{y\sim x} s_\theta(x,y) s_{\mathrm{true}}(x,y) \pi_{\mathrm{true}}(x) + \cdot \\
&= \sum_{x\in \CR_X} \sum_{y\sim x} s_\theta(x,y)^2 \pi_{\mathrm{true}}(x) -  2 \sum_{x\in \CR_X} \sum_{y\sim x} s_\theta(x,y) \pi_{\mathrm{true}}(x) + 2 \sum_{x\in \CR_X} \sum_{y\sim x} s_\theta(x,y) \pi_{\mathrm{true}}(y) + \cdots\\
&= \sum_{x\in \CR_X} \sum_{y\sim x} (s_\theta(x,y)-1)^2 \pi_{\mathrm{true}}(x) + 2 \sum_{x\in \CR_X} \sum_{y\sim x} s_\theta(y,x) \pi_{\mathrm{true}}(x) + \cdots\\
\end{align*}
where ``$\cdots$'' denote terms that do not depend on $\theta$. 
Based on training data, a discrete score matching estimator can therefore be defined as a minimizer of
 \begin{equation}
 \label{eq:scorematch.obj.disc}
\sum_{k=1}^N \sum_{y\sim x_k} (s_\theta(x_k,y)-1)^2 + 2 \sum_{k=1}^N \sum_{y\sim x_k} s_\theta(y,x_k) 
\end{equation}

 \end{remark}

\section{Incomplete observations for graphical models}
\subsection{The EM Algorithm}

 Missing variable sin the context of graphical models may
correspond to real processes that cannot be measured, which is common, for example, with biological data. They may be more conceptual
objects that are interpretable but are not parts of the data
acquisition process, like phonemes in speech recognition, or edges and
labels in image processing and object recognition. They may also be
variables that have been added to the model to increase its parametric
dimension without increasing the complexity of the graph. However, as we will see, dealing with incomplete or imperfect observations  brings the parameter estimation problem to a new level
of difficulty.

Since it is the most common approach to address incomplete or noisy observations, we start with a description of how the EM algorithm (\cref{alg:EM}) applies to graphical models, and of its limitations.  
We assume a graphical model on an undirected graph $G = (V, E)$, in which we assume that $V$ is separated in two non-intersecting subsets, $V = S\cup H$. Letting $X$ be a $G$-Markov random field, the part $\pe XS$ is assumed to be observable, and $\pe XH$ is hidden.

We assume that $X$ takes values in $\CF(V)$, where we still denote by $F_s$ the sets in which $X_s$ takes values for $s\in V$. We let the model distribution belong to an exponential family, with  
\begin{equation}
\label{eq:exp.fam}
\pi_\th(x) = \frac{1}{Z(\th)} \exp\big( -\th^TU(x)\big), \ x\in \CF(V).
\end{equation}
Assume that an $N$-sample $\pe {x_1} S, \ldots, \pe {x_N}S$ is observed over $S$.
Since 
\[
\log \pi_\th(x) = - \log Z(\th) - \th^TU(x) ,
\]
the transition from $\th_n$ to $\th_{n+1}$ in \cref{alg:EM} is done by maximizing
\begin{equation}
\label{eq:em.exp.1}
-\log Z(\th) -  \th^T\bar U_n
\end{equation}
where
\begin{equation}
\label{eq:em.exp.2}
\bar U_n = \frac1N\sum_{k=1}^N E_{\th_n}(U(X)\mid \pe XS=\pe {x_k}S).
\end{equation}
So, the M-step of the EM, which maximizes \cref{eq:em.exp.1},
coincides with the complete-data maximum-likelihood problem
for which the empirical average of $U$ is replaced by the average of its conditional
expectations given the observations, as given in
\cref{eq:em.exp.2}, which constitutes the E-step. As a consequence, a strict application of the EM algorithm for graphical models is unfeasible, since each step requires running an algorithm of similar complexity maximum likelihood for complete data, that we already identified as a challenging, computationally costly problem. The same remark holds for the SAEM algorithm of \cref{sec:saem}, which also requires solving a maximum likelihood problem at each iteration.

\subsection{Stochastic gradient ascent}

The stochastic gradient ascent described in \cref{sec:stoc.grad} can be extended to partial observations \citep{you89}, even though it loses the global convergence guarantee that resulted from the concavity of the log-likelihood for complete observations. Indeed, applying the computation of \cref{sec:direct.min.incomplete}, to a model given by \cref{eq:exp.fam}, we get using  \cref{prop:ml.exp},
\[
\prt_\th \ln \psi_\th = E_\th(E_\th(U) - U \mid \pe XS =\pe xS) = E_\th(U) -
E_\th(U \mid \pe XS=\pe xS)
\]
where we $\psi_\theta(\pe xS)$ denotes the marginal distribution of $\pi_\theta$ on $S$.

%


Let $\pi_\theta(\pe xH \mid \pe xS)$ denotes the conditional probability $P(\pe XH = \pe xH\mid \pe XS =\pe sS)$ for the distribution $\pi_\theta$, therefore taking the form
\[
\pi_\theta(\pe xH\mid \pe xS) = \frac{1}{\tilde Z(\theta, \pe xS)}\exp\left(-\theta^TU(\pe xS \wedge \pe xH)\right).
\]
Assume given an ergodic
transition probability $p_\th$ on $\CF(V)$, and a family of
ergodic transition probabilities
$p_\th^{\pe xS}$, $\pe xS\in \CF(S)$, such that the invariant
distribution of $p_\th$ is $\pi_\th$, and the one of $p_\th^{\pe xS}$ is
$\pi_\th(\cdot\mid \pe xS)$. Then the following SGA algorithm can be used to estimate $\theta$

\begin{algorithm}
\label{alg:mrf.sga.incomplete}
Start the algorithm with an initial parameter
$\th(0)$ and initial configurations $x(0)$ and
$\pe {x_k}H (0)$, $k=1, \ldots, N$. Then, at step $n$, 
\begin{enumerate}[label=(SGH\arabic*)]
\item Sample from the distribution $ p_{\th(n)}(x(n) ,
\cdot)$ to obtain new configurations $x(n+1) \in \CF(V)$.
\item For $k=1, \ldots, N$, sample from the distribution $ p_{\th(n)}^{\pe {x_k}S}(\pe{x_k}H(n) ,
\cdot)$ to obtain a new configuration $\pe {x_k}^H(n+1)$ over the hidden vertexes.
\item Update the parameter using
\begin{equation}
\label{eq:stoc.grad.h}
\th(n+1) = \th(n) + \ga(n+1) \left(U(x(n+1)) - \frac1N \sum_{k=1}^N
U(\pe{x_k}S\wedge \pe{x_k}H(n+1))\right).
\end{equation}
\end{enumerate}
\end{algorithm}

\subsection{Pseudo-EM Algorithm}

The EM update
\[
\th_{n+1} = \argmax_\th \Big(\sum_{k=1}^N E_{\th_n}\left(\ln\pi_\th(X)\mid \pe XS = \pe {x_k}S\right)\Big).
\]
being challenging for Markov random fields, it is tempting to replace the log-likelihood in the expectation by an other contrast, such as the log-pseudo-likelihood. A similar approach to that described here was introduced in \citet{chalmond1989iterative}, for situations  when the  conditional distribution
of $\pe XS$ given $\pe XH$ is ``simple enough'' (for example, if the variables $X_s, s\in S$ are conditionally  independent given $\pe XH$) and when the cardinality of the sets $F_s$, $s\in H$ is small (binary, or ternary, variables). 

The algorithm has the following variational interpretation. Fix $\pe xS \in \CF(S)$ and $s\in H$. Also denote $\mu_s = 1/|\CF(H\setminus \{s\})|$. If $q$ is a transition probability from $\CF(H\setminus \{s\})$ to $F_s$, let
\begin{equation}
\label{eq:d.xi.2}
 \Delta^{(s)}_\th(q, \pe xS) = \sum_{y\in \CF(H)}
\left(
\log \left(
\frac{\pi_{\th, s}(\pe ys \wedge \pe xS \mid \pe y{H\setminus \{s\}})}{q(\pe y{H\setminus \{s\}}, \pe ys)\mu_s}
\right) 
q(\pe y{H\setminus \{s\}}, \pe ys) \mu_s 
\right).
\end{equation}
This function is concave in $q$, since its first partial derivative with respect to $q(\pe y{H\setminus \{s\}}, \pe ys)$ (for each $y\in \CF(H)$) is given by
\[
\mu_s\log 
\pi_{\th, s}(\pe ys \wedge \pe xS \mid \pe y{H\setminus \{s\}}) \mu_s(\pe y{H\setminus \{s\}})
- \mu_s \log( q(\pe y{H\setminus \{s\}}, \pe ys) \mu_s) - \mu_s
\]
so that its Hessian is the  diagonal matrix with negative entries $-\mu_s/ q(\pe y{H\setminus \{s\}}, \pe ys)$. Using Lagrange multipliers to express the constraints
$\sum_{\pe ys \in F_s} q(\pe y{H\setminus \{s\}}, \pe ys) = 1$ for all $\pe  y{H\setminus \{s\}}$, we find  that $\Delta^{(s)}_\th(q, \pe xS)$ is maximized when $q(\pe y{H\setminus \{s\}}, \pe ys)$ is proportional to 
$\pi_{\th, s}(\pe ys \wedge \pe xS \mid \pe y{H\setminus \{s\}})$, yielding
\[
q(\pe y{H\setminus \{s\}}, \pe ys) = \pi_{\th, s}(\pe ys \mid \pe xS \wedge \pe y{H\setminus \{s\}}).
\]

Now, consider the problem of maximizing 
\begin{equation}
\label{eq:pseudo.em.obj}
\sum_{k=1}^N \sum_{s\in H} \Delta^{(n)}_\theta(q_k^{(s)}, \pe {x_k}s)
\end{equation}
with respect to $\theta$ and $\pe{q_k}s$, $k=1, \ldots, N$, $s\in H$. Consider an iterative  maximization scheme in which, from a current parameter $\theta_n$, one first, maximizes \cref{eq:pseudo.em.obj} with respect to transition probabilities $\pe {q_k}s$, then with respect to $\theta$ to obtain $\theta_{n+1}$. This scheme provides the iteration
\begin{multline*}
\theta_{n+1} = \\
\argmax_\theta \sum_{k=1}^N \sum_{s\in H} 
\sum_{y\in \CF(H)}
\left(\log \pi_{\th, s}(\pe ys \wedge \pe {x_k}S \mid \pe y{H\setminus \{s\}})
\right) \pi_{\th_n, s}(\pe ys \mid \pe {x_k}S \wedge \pe y{H\setminus \{s\}})\mu_s.
\end{multline*}

\subsection{Partially-observed Bayesian networks on trees}

We now consider the situation in which the joint distribution of
$X = \pe XS\wedge \pe XH$ is a Bayesian network over a directed acyclic graph
$G=(V,E)$.  

Assume that $x_1^{(S)}, \ldots, x^{(S)}_N$ are observed. The
parameter $\th$ is the collection of all $p(\pe x{\pa{s}}, \pe xs)$ for $s\in
V$. Define the random variables $I_{s, x}(y)$ equal to one if
$\pe y{\{s\}\cup \pa{s}} = \pe x{\{s\}\cup \pa{s}}$ and zero otherwise. We can write
\[
\ln \pi(y) = \sum_{s\in S} \ln p_s(\pe y{\pa{s}}, \pe ys)
= \sum_{s\in S}\sum_{\pe x{\{s\}\cup \pa{s}}\in \CF(\{s\}\cup \pa{s})} \ln
p_s(\pe x{\pa{s}}, \pe xs) I_{s,x}(y)
\]
This implies that 
\begin{align*}
& \sum_{k=1}^N E_{\th_n}\left(\ln\pi(\pe {x_k}S,
\pe XH)\mid  \pe XS = \pe {x_k}S\right)  \\
&= \sum_{\pe x{\{s\}\cup \pa{s}}\in \CF(\{s\}\cup \pa{s})} \ln
p_s(\pe x{\pa{s}}, \pe xs)  \sum_{k=1}^N E_{\th_n}(I_{s,x}(X)\mid  \pe XS = \pe {x_k}S)
\\
&= \sum_{\pe x{\{s\}\cup \pa{s}}\in \CF(\{s\}\cup \pa{s})} \ln
p_s(\pe x{\pa{s}}, \pe xs)  \sum_{k=1}^N \pi_{\th_n}(\pe x{\{s\}\cup \pa{s}}\mid  \pe XS = \pe {x_k}S).
\end{align*}
The EM iteration at step $n$ then is
$$
p^{(n+1)}_s(\pe x{\pa{s}}, \pe xs) = \frac{1}{Z_s(\pe x{s^{-}})} \sum_{k=1}^N
\pi_{\th_n}(\pe x{\{s\}\cup \pa{s}}\mid  \pe XS = \pe {x_k}S)
$$
with 
$$
\pi_{\th_n}(x) = \prod_{s\in V} p^{(n)}(\pe x{\pa{s}}, \pe xs),
$$
$Z_s$ being a normalization constant.

If the estimation is solved with a Dirichlet prior $\text{Dir}(1+a_s(\pe xs,
\pe x{\pa{s}}))$, the update formula becomes
\begin{equation}
\label{eq:em.tree}
p^{(n+1)}_s(\pe x{\pa{s}}, \pe xs) =  \frac{1}{Z_s(\pe x{s^{-}})} \left(a_s(\pe xs, \pe x{\pa{s}})
+ \sum_{k=1}^N
\pi_{\th_n}(\pe x{\{s\}\cup \pa{s}} \mid \pe XS = \pe {x_k}S)\right).
\end{equation}

This algorithm is very simple when the conditional distributions
$\pi_{\th_n}(\pe x{s\cup \pa{s}}\mid  \pe XS = \pe {x_k}S)$ can be easily
computed, which is not always the case for a
general Bayesian network, since  conditional distributions do not
always have a structure of Bayesian network. The computation is simple
enough for trees, however, since conditional tree distributions are
still trees (or forests). More
precisely, the conditional distribution given the observed variables
can be written in the form
$$
\pi(\pe yH\mid \pe xS) = \frac{1}{Z(\pe xS)} \prod_{s\in H} \phi_{s,x}(\pe ys)
\prod_{t\sim s, \{s,t\} \sub H} \phi_{st}(\pe ys, \pe yt)
$$
with  $\phi_{s,\pa{s}}(\pe ys, \pe y{\pa{s}}) = p_s(\pe y{\pa{s}}, \pe ys)$ and, letting
$\phi_s(\pe ys) = p_s(\pe ys)$ if $\pa{s} = \emp$ and 1 otherwise,
$$
\phi_{s,x}(\pe ys) =\phi_s(\pe ys) \prod_{t\sim s, t\in S}
\phi_{st}(\pe ys, \pe xt).
$$
So, the marginal joint distribution of a vertex
and its parents are directly given by  belief propagation, using the
just defined interactions. This training algorithm is summarized below.

\begin{algorithm}[Learning tree distributions with hidden variables]
Start with some initial
guess of the conditional probabilities (for example, those given by
the prior). The iterate the following two steps providing the transition from  $\theta_n$ to step $\theta_{n+1}$.

\begin{itemize}
\item[(1)] For $k=1, \ldots, N$, use belief propagation (or sum-prod) to compute 
all $\pi_{\th_n}(\pe x{\{s\}\cup \pa{s}}\mid  \pe XS = x_k^{(S)})$. Note that these
probabilities can be 0 or 1 when $s\in S$ and/or $\pa{s}\subset S$.
\item[(2)] Use \cref{eq:em.tree} to compute the next set of
parameters.
\end{itemize}
\end{algorithm}

The tree case includes the important example of hidden Markov models,
which are defined as follows. $S$ and $H$ are ordered, with same
cardinality, say $S=\{s_1, \ldots, s_q\}$ and $H=\{h_1, \ldots,
h_q\}$. Edges are $(h_1,h_2), \ldots, (h_{q-1}, h_q)$ and $(h_1,s_1),
\ldots, (h_q, s_q)$. The interpretation generally is that the hidden
variables, $h_s$, are the variables of interest, and behave like a
Markov chain, and that the observations, $x_s$, are either noisy or
transformed versions of them. A major application is in speech
recognition, where the $h_s$'s are labels that represent specific
phonemes (little pieces of spoken words) and the $x_s$'s are measured
signals. The transitions between hidden variables then describe how phonemes are likely
to appear in sequence for a given language, and those between hidden and observed variables
describe how each phoneme is likely to be pronounced and heard.

\subsection{General Bayesian networks}

The algorithm in the general case can move from tractable to
intractable depending on the situation. This must generally be
handled in a case by case basis, by analyzing the conditional
structure, for a given model, knowing the observations.

In practice, it is always possible to use loopy belief propagation to
obtain some approximation of the conditional probabilities, even if it
is not sure that the algorithm will converge to the correct
marginals. When feasible, junction trees can be used, too.  
 Monte-Carlo sampling is also an option, although quite
computational.

\chapter{Deep Generative Methods}
\label{chap:generative}

\section{Normalizing flows}
\label{sec:norm.flows}
\subsection{General concepts}
We develop, in this chapter, methods that model stochastic processes using  a feed-forward approach that generates complex random variables using non-linear transformations of simpler ones. Many of these methods can be seen as instances of structural equation models (SEMs), described in \cref{sec:sem}, with, for deep-learning implementations, high-dimensional parametrizations of \cref{eq:sem}. 

With start with the formally simple case where the modeled variable takes values in $\mathbb R^d$ and is modeled as
\[
X = g(Z)
\]
where $Z$ also takes values in $\mR^d$, with a known distribution, and $g$ is $C^1$, invertible, with a $C^1$ inverse on $\mathbb R^d$, i.e., is a {\em diffeomorphism} of $\mR^d$. Let us denote by $h$ the inverse of $g$.

If $Z$ has a p.d.f. $f_Z$ with respect to Lebesgue's measure, then, using the change of variable formula, the p.d.f. of $X$ is 
\[
f_X(x) = f_Z(h(x))\ |\det{\partial_x h(x)}|.
\]

Now, given a training set $T = (x_1, \ldots, x_N)$, the log-likelihood, considered as a function of $h$, is given by
\begin{equation}
\label{eq:norm.flow.lik}
\ell(h) = \sum_{k=1}^N \log f_Z(h(x_k)) + \sum_{k=1}^N \log |\det{\partial_x h(x_k)}|\,.
\end{equation}
This expression should then be maximized with respect to $h$, subject to some restrictions or constraints to avoid overfitting.


\subsection{A greedy computation}
One can define a rich class of diffeomorphisms through iterative compositions of simple transformations. This framework was introduced in \citep{tabak2010density}, where a greedy approach was suggested to build such compositions. The method was termed ``normalizing flows,'' since it create a discrete flow of diffeomorphisms that transform the data into a sample of a normal distribution.

We quickly describe the basic principles of the algorithm. One starts with a 
parametrized family, say $(\psi_\alpha, \alpha \in A)$ of diffeomorphisms of $\mR$. Such families are relatively easy to design, one example proposed in \citep{tabak2010density} being a smoothed version of the piecewise linear function
\[
u \mapsto v_0 + (1-\sigma) u + \gamma |(1-\sigma) u - u_0|
\]
which is increasing as soon as $0\leq \max(\sigma, \gamma) < 1$. The smoothed version has an additional parameter, $\epsilon$, and takes the form
\[
u \mapsto v_0 + (1-\sigma) u + \gamma \sqrt{\epsilon^2 + ((1-\sigma) u - u_0)^2}.
\]
This transformation is parametrized by $\alpha = (v_0, \sigma, \gamma, u_0, \epsilon)$. 
Other families of parametrized transformations can be designed.

A multivariate transformation $\phi_{\boldsymbol\alpha, U}: \mR^d\to\mR^d$ can then be associated to families $\boldsymbol\alpha = (\alpha_1, \ldots, \alpha_d)$ and orthogonal matrices $U$  by taking
\[
\phi_{\boldsymbol\alpha, U}(x) = \begin{pmatrix}
\psi_{\alpha_1}(\pe y1)\\ \vdots \\ \psi_{\alpha_d}( \pe yd)
\end{pmatrix}
\]
with $y = Ux$.

The algorithm in \citep{tabak2010density} is initialized with $h_0 = \id[d]$ and updates the transformation at step $n$ according to 
\[
h_{n} = \phi_{\boldsymbol\alpha_n, U_n}\circ h_{n-1}.
\] 

In this update, $U_n$ is generated as a random rotation matrix, and $\boldsymbol\alpha_n$ is determined as a gradient ascent update (starting from $\boldsymbol\alpha = 0$) for the maximization of 
\[
\boldsymbol\alpha \mapsto \ell(\phi_{\boldsymbol\alpha, U_n}\circ h_{n-1}).
\]
(Here, the current value $h_{n-1}$ is not revisited, therefore providing a ``greedy'' optimization method.)

Letting $z_{n,k} = h_n(x_k)$, the chain rule implies that   
\begin{multline*}
\ell(\phi_{\boldsymbol\alpha, U_n}\circ h_{n-1})) = \sum_{k=1}^N \log f_Z(\phi_{\boldsymbol\alpha, U_n}(z_{n-1,k})) + \sum_{k=1}^N \log |\det{\phi_{\boldsymbol\alpha, U_n}(z_{n-1,k})}| \\+ \sum_{k=1}^N \log |\det{\partial_x h_{n-1}(x_k)}|\,.
\end{multline*}
Since the last term does not depend on $\boldsymbol\alpha$, we see that it suffices to keep track of the ``particle'' locations,  $z_{n-1, k}$ to be able to compute $\boldsymbol\alpha_n$. Note also that these locations are easily updated with
$z_{n, k} = \phi_{\boldsymbol\alpha_n, U_n}(z_{n-1, k})$.

\subsection{Neural implementation}
This iterated composition of diffeomorphisms obviously provides a neural architecture  similar to those discussed in \cref{chap:neural.nets}. Fixing the number of iterations to be, say, $m$, one can consider families of diffeomorphisms $(\phi_\theta)$ indexed by a parameter $w$ (we had $w = (\boldsymbol\alpha, U)$ in the previous discussion), and optimize 
\cref{eq:norm.flow.lik} over all functions $h$ taking the form $h = \phi_{w_m} \circ \cdots \circ \phi_{w_1}$. Letting $z_{j,k} =  \phi_{w_j} \circ \cdots \circ \phi_{w_1}(x_k)$ for $j\leq m$ (with $z_{0,k} = x_k$), we can write
\[
\ell(h) = \sum_{k=1}^N \log f_Z(z_{m,k}) + \sum_{k=1}^N \sum_{j=1}^m \log |\det{\partial_x \phi_{w_j}(z_{j-1,k})}|.
\]

Normalizing flows in this form are described in \citep{rezende2015variational,kobyzev2020normalizing,
papamakarios2021normalizing}. The gradient of $\ell$ with respect to the parameters $w_1, \ldots, w_m$ can be computed by backpropagation. We note however that, unlike typical neural implementations, the parameters may come with specific constraints, such as $U\in \CO_d(\mR)$ when $w = (\boldsymbol\alpha, U)$, so that the gradient and associated displacement may have to be adapted compared to standard gradient ascent implementations (see \cref{sec:grad.orthogonal} for a discussion of first-order implementations of gradient methods for functions of orthogonal matrices, and \citep{absil2008optimization} for more general methods on optimization over matrix groups).

\subsection{Time-continuous version}
In \cref{sec:continuous.nn}, we described how diffeomorphisms could be generated as flows of differential equations, and this remark can be used to provide a time-continuous version of normalizing flows. Using \cref{eq:neural.ode}, one generates trajectories $z(\cdot)$ by solving over, say, $[0, T]$
\[
\partial_t z(t) = \psi_{w(t)}(z(t))
\]
with $z(0) = x$ for some function $w: t\mapsto w(t)$. Letting $z(t) = h_{w}(t, x)$ (which defines $h_w$), we know that, under suitable assumptions on $\psi$, the mapping $x \mapsto h_w(t, x)$ is a diffeomorphism of $\mR^d$. 
One can then maximize  
\[
\ell(h_w(T, \cdot)) = \sum_{k=1}^N \log f_Z(h_w(T, x_k)) + \sum_{k=1}^N \log |\det{\partial_x h_w(T, x_k)}|
\]
with respect to the function $w$. Let $z_k(t) = h_w(t, x_k)$ and $J_k(t) = \log |\det{\partial_x h_w(t, x_k)}|$. We have, by definition 
\[
\partial_t z_k(t) = \psi_{w(t)}(z_k(t))
\]
with $z_k(0) = x_k$. One can also show that
\[
\partial_t J_k(t) = \nabla\cdot \psi_{w(t)}(z_k(t))
\]
with $J_k(0) = 0$,
where the r.h.s. is the divergence of $\psi_{w(t)}$ evaluated at $z_k(t)$. We provide a quick (and formal) justification of this fact. First note that differentiating $\partial_t h_w(t,x) = \psi_{w(t)}(h_w(t,x))$ with respect to $x$ yields
\[
\partial_t \partial_x h_w(t,x) = \partial_x \psi_{w(t)}(h_w(t,x)) \partial_x h_w(t,x).
\] 
The mapping $\mathcal J: A \mapsto \log |\det(A)|$ is differentiable on the set of invertible matrices and is such that $d\mathcal J(A) H = \trace(A^{-1}H)$. Applying the chain rule, we find
\begin{align*}
\partial_t \log|\det{\partial_x h_w(t,x)}| &= \trace(\partial_x h_w(t,x)^{-1} \partial_x \psi_{w(t)}(h_w(t,x)) \partial_x h_w(t,x))\\
& = 
\trace(\partial_x \psi_{w(t)}(h_w(t,x))) = \nabla\cdot \psi_{w(t)}(h_w(t,x)).
\end{align*}

From this, it follows that the time-continuous normalizing flow problem can be reformulated as maximizing
\[
\sum_{k=1}^N \log f_Z(z_k(T)) + \sum_{k=1}^N J_k(T)
\]
subject to $\partial_t z_k(t) = \psi_{w(t)}(z_k(t))$, 
$\partial_t J_k(t) = \nabla\cdot \psi_{w(t)}(z_k(t))$, $z_k(0)=x_k$, $J_k(0) = 0$. This is an optimal control problem, whose analysis can be done similarly to that made in \cref{sec:neural.ode}, provided that $\nabla\cdot \psi_{w(t)}$ can be expressed in closed form.

Note that the inverse of $h_w(T, \cdot)$, which provides the generative model going from $Z$ to $X$ can also be obtained as the solution of an ODE. Namely, if one solves the differential equation
\[
\partial_t x(t) = - \psi_{w(T-t)}(x(t))
\]
with initial condition $x(0) = z$, then $x(T)$ solves the equation $h_w(T, \cdot) = z$.

\section{Non-diffeomorphic models and variational autoencoders}
\label{sec:gen.nondiff}
\subsection{General framework}
The previous discussion addressed the situation  $X = g(Z)$ when $g$ is a diffeomorphism, which required, in particular, that $X$ and $Z$ are real vectors with identical dimensions. This may not always be desirable, as one may prefer a small-dimensional variable $Z$ (in the spirit of the factor analysis methods discussed in \cref{chap:dim.red}), or a high-dimensional $Z$ to increase, for example the modeling power. In addition, the observation variables may be discrete, which precludes the use of the change of variables formula. In such cases, $Z$ has to be treated as a hidden variable using one of the methods discussed in \cref{chap:var.bayes}.  

It will convenient  to model the generative process in the form of a conditional distribution of $X$ given $Z$ rather than a deterministic function. We place ourselves in the framework of \cref{chap:var.bayes} (with slightly modified notation) and let $\CR_X$ and $\CR_Z$ denote the measured spaces over where $X$ and $Z$ take their values, with measures $\mu_X$ and $\mu_Z$, and assume that the conditional distribution of $X$ given $Z=z$ has density $f_X(x\mid z, \theta)$ with respect to $\mu_X$, for some parameter $\theta$. We also assume that $Z$ has a distribution with density $f_Z$ with respect to $\mu_Z$, that we assume given and unparametrized. One can then directly apply the algorithms provided in \cref{chap:var.bayes}, and in particular the variational methods described in \cref{sec:var.approx} with an appropriate definition of the approximation of the conditional density of $Z$ given $X$. An important example in this context is provided by variational autoencoders (VAEs) that we now present.

\subsection{Generative model for VAEs}
\label{sec:vae}
VAEs \citep{kingma2014auto-encoding,kingma2019introduction} model $X\in \mR^d$ as $X = g(Z, \theta) + \ep$ where $\ep$ is a centered Gaussian noise with covariance matrix $Q$. The function $g$ is typically nonlinear, and VAEs have been introduced with this function modeled as a deep neural network (see \cref{chap:neural.nets}). Letting $\phi_\CN(\ccdot;\,0, Q)$ denote the p.d.f. of the Gaussian distribution $\CN(0, Q)$, the conditional distribution of $X$ given $Z=z$ has density
\[
f_X(x\mid z, \theta) = \phi_\CN(x - g(z, \theta))\,;\, 0, Q)
\]
with respect to Lebesgue's measure on $\mR^d$.

Following the procedure in \cref{sec:var.approx}, we define an approximation of the conditional distribution of $Z$ given $X$. Assuming that $Z\in \mR^q$, we let this distribution be $\CN(\mu(x, w), \Sigma(x, w))$ for some functions $\mu$ and $\Sigma$, $w$ being a parameter. To ensure that $\Sigma \succeq 0$, we will represent it in the form $\Sigma(x, w) = S(x, w)^2$ where $S$ is a symmetric matrix. In \citep{kingma2014auto-encoding}, both functions $\mu$ and $S$ are represented as neural networks parametrized by $w$. The joint density of $X$ and $Z$ is such that
\begin{align*}
\log f_{X,Z}(x,z\,;\,\theta, Q) &= \log \phi_\CN(x - g(z, \theta))\,;\, 0, Q) + \log f_Z(z)\\
&= -\frac12 (x - g(z, \theta))^T Q^{-1} (x - g(z, \theta)) - \frac12 \log \det Q - \frac d2 \log 2\pi + \log f_Z(z).
\end{align*}
We also have
\[
\log \phi_\CN(z\,;\, \mu(x, w), S(x, w)^2) = -\frac12 (z - \mu(x,w))^T S(x, w)^{-2} (z - \mu(x,w)) - \log\det S(x, w) - \frac q2 \log 2\pi.
\]

We can then rewrite the algorithm in \cref{eq:sga.var.2} as
\begin{equation}
\label{eq:sga.vae}
\left\{
\begin{aligned}
\theta_{n+1} &= \theta_n +\gamma_{n+1} \partial_\theta \log f_{X,Z}(X_{n+1},Z_{n+1};\theta_n, Q_n)
\\
Q_{n+1} &= Q_n +\gamma_{n+1} \partial_Q \log f_{X,Z}(X_{n+1},Z_{n+1};\theta_n, Q_n)
\\
w_{n+1} &= w_n + \gamma_{n+1} \log \left( \frac{f_{X,Z}(X_{n+1},Z_{n+1};\theta_n, Q_n)}{\phi_\CN(X_{n+1}\,;\, \mu(X_{n+1}, w_n), S(X_{n+1}, w_n)^2)}\right) \\
& \qquad\qquad \times \partial_w \log \phi_\CN(X_{n+1}\,;\, \mu(X_{n+1}, w_n), S(X_{n+1}, w_n)^2)
\end{aligned}
\right.
\end{equation}
where $X_{n+1}$ is drawn uniformly from the training data and 
\[
Z_{n+1} \sim \CN(\mu(X_{n+1}, w_n), S(X_{n+1}, w_n)^2).
\]
The derivatives in this system can be computed from those of $g, \mu$ and $S$ (typically involving back-propagation) and the  expression of the derivatives of the  determinant and inverse of a matrix provided in \cref{eq:det.der,eq:inv.der}. 

The computations can be simplified if one  assumes that $f_Z$ is the p.d.f. of a standard Gaussian, i.e., $f_Z = \phi_\CN(\cdot;0, \Id[q])$. Indeed, in that case, the integral in \cref{eq:em.var}, which is, using the current notation,
\begin{equation}
\label{eq:var.vae}
\int_{\mR^p}\log \frac{\phi_\CN(x-g(z,\theta)\,;\, 0, Q)
\phi_\CN(z\,;\,0, \Id[p])}{\phi_\CN(z\,;\, \mu(x, w), S(x,w)^2)}
 \phi_\CN(z\,;\, \mu(x, w), S(x,w)^2) dz,
 \end{equation}
can be partially computed. For any two $q$-dimensional Gaussian p.d.f.'s, one has
\begin{multline}
\label{eq:gauss.kl}
\int_{\mR^p} \log \phi_\CN(z\,;\,\mu_1, \Sigma_1)\ \phi_\CN(z\,;\,\mu_2, \Sigma_2)\, dz = -\frac12\trace(\Sigma_1^{-1} \Sigma_2) - \frac{1}{2} (\mu_2-\mu_1)^T \Sigma_1^{-1} (\mu_2-\mu_1)\\ - \frac12 \log \det(\Sigma_1) - \frac{q}{2} \log(2\pi).  
\end{multline}
As a consequence, \cref{eq:var.vae} becomes
\begin{multline}
\label{eq:var.vae.2}
-\frac{1}{2} \myE_w\left((X-g(Z, \theta))^T Q^{-1}(X-g(Z, \theta))\right) - \frac12 \log \det Q - \frac{d}2 \log 2\pi \\ - \myE_w\left(\frac12 \trace(S(X,w)^2) + \frac{1}{2} |\mu(X,w)|^2 - \log \det(S(X,w))\right) + \frac q2,
\end{multline}
where $\myE_w$ denotes the expectation for the random variable $(X,Z)$ where $X$ follows a uniform distribution over training data and the  conditional distribution of $Z$ given $X=x$ is $\CN(\mu(x,w)\,,\,S(x,w)^2)$.

The algorithm proposed in \citet{kingma2014auto-encoding} introduces a change of variable $Z = \mu(X, w) + S(X,w) U$ where $U\sim\CN(0, \Id[q])$, rewriting \cref{eq:var.vae.2} as
\begin{multline}
\label{eq:var.vae.3}
-\frac{1}{2} \myE\left((X-g(\mu(X,w) + S(X,w)U, \theta))^T Q^{-1}(X-g(\mu(X,w) + S(X,w)U, \theta))\right) \\  - \myE_w\left(\frac12 \trace(S(X,w)^2) + \frac{1}{2} |\mu(X,w)|^2   - \log \det(S(X,w))\right)\\
- \frac12 \log \det Q - \frac{d}2 \log 2\pi + \frac q2,
\end{multline}
with a modified version of \cref{eq:sga.vae}. Letting
\begin{multline*}
F(\theta, Q, w, x, u) = -\frac{1}{2} (x-g(\mu(x,w) - S(x,w)U, \theta))^T Q^{-1}(x-g(\mu(x,w) - S(x,w)U, \theta))\\ - \frac12 \log \det Q  - \frac12 \trace(S(x,w)^2) - \frac{1}{2} |\mu(x,w)|^2   +  \log \det(S(x,w))
\end{multline*}
the resulting algorithm is
\begin{equation}
\label{eq:sga.vae.4}
\left\{
\begin{aligned}
\theta_{n+1} &= \theta_n +\gamma_{n+1} \partial_\theta F(\theta_n, Q_n, w_n, X_{n+1},U_{n+1})\\
Q_{n+1} &= Q_n +\gamma_{n+1} \partial_Q F(\theta_n, Q_n, w_n, X_{n+1},U_{n+1})
\\
w_{n+1} &= w_n - \gamma_{n+1} \partial_w F(\theta_n, Q_n, w_n, X_{n+1},U_{n+1})
\end{aligned}
\right.
\end{equation}
where $X_{n+1}$ is drawn uniformly from the training data and 
$U_{n+1} \sim \CN(0, \Id[q])$.

\subsection{Discrete data}
This framework can be  adapted to situations in which the observations are discrete. Assume, as an example, that $X$ takes values in $\{0,1\}^V$, where $V$ is a set of vertexes, i.e., $X$ is a binary Markov random field on $V$. Assume, as a generative model, that  conditionally to the latent variable $Z\in \mR^p$, the variables $\pe Xs, s\in V$ are independent and $\pe Xs$ follows a Bernoulli distribution with parameter $\pe gs(z, \theta)$, where $g(\cdot, \theta): \mR^q \to [0,1]^V$. Assume also that $Z \sim \CN(0, \Id[q])$, and define, as above, an approximation of the conditional distribution of $Z$ given $X=x$ as a Gaussian with mean $\mu(x,w)$ and covariance matrix $S(x, w)^2$. Then, the joint density of $X$ and $Z$ (with respect to the product of the counting measure on $\{0,1\}^V$ and Lebesgue's measure on $\mR^q$) is
\[
\log f_{X,Z}(x,z\,;\,\theta) = \sum_{s\in V} (\pe xs \log \pe gs(z, \theta)+ (1-\pe xs)\log(1-\pe gs(z,\theta))) + \log \phi_\CN(z\,;\, 0, \Id[q])
\]
and \cref{eq:sga.vae} becomes
\begin{equation}
\label{eq:sga.vae.disc}
\left\{
\begin{aligned}
\theta_{n+1} &= \theta_n +\gamma_{n+1} \partial_\theta \log f_{X,Z}(X_{n+1},Z_{n+1};\theta_n, Q_n)
\\
\\
w_{n+1} &= w_n + \gamma_{n+1} \log \left( \frac{f_{X,Z}(X_{n+1},Z_{n+1};\theta_n)}{\phi_\CN(X_{n+1}\,;\, \mu(X_{n+1}, w_n), S(X_{n+1}, w_n)^2)}\right) \\
& \qquad\qquad \times \partial_w \log \phi_\CN(X_{n+1}\,;\, \mu(X_{n+1}, w_n), S(X_{n+1}, w_n)^2)
\end{aligned}
\right.
\end{equation}

\section{Generative Adversarial Networks (GAN)}
\label{sec:gan}

\subsection{Basic principles}
Similarly to the methods discussed so far, GAN's \citep{goodfellow2014generative}, use a one-step nonlinear generator
$X = g(Z, \theta)$, with $\theta\in \mR^K$, to model observed data (we here switch back to a deterministic relation), where $Z$ has a known distribution, with p.d.f. $f_Z$, for example $Z\sim \CN(0, \Id[q])$. However, unlike the exact or approximate likelihood maximization that were discussed in \cref{sec:norm.flows,sec:gen.nondiff}, GANs us a different criterion for estimating the parameter $\theta$ by minimizing metrics that can be approximated by optimizing a classifier.
 The classifier is a function $x \mapsto f(x, w)$, parametrized by $w\in \mR^M$, whose goal is to separate simulated samples from real ones: it takes values in $[0,1]$ and estimates the (posterior) probability that its input $x$ is real.  The adversarial paradigm in GAN's consists in estimating $\theta$ and $w$ together so that generated data, using $\theta$, are indistinguishable from real ones using the optimal $w$.
Their basic structure is summarized in Figure \ref{fig:gan.1}.

\begin{figure}[h]

	\begin{center}
		\begin{tikzpicture}[roundnode/.style={rectangle, draw=green!60, fill=green!5, thin, minimum size=20mm}, squarednode/.style={rectangle, draw=black!60, fill=gray!5, very thick},]
		\node[squarednode, minimum height=1cm] (C) {Classifier} ;
		\node[squarednode, minimum height= 1cm, minimum width = 1cm] (P) [above=.5cm of C] {Prediction} ;
		\node[squarednode, minimum height= 1cm, minimum width = 1cm] (W) [below=.5cm of C] {$W$} ;
		\node[squarednode, minimum height= 1cm, minimum width = 2cm] (T) [above right=-.4cm and .75cm of C] {Data} ;
		\node[squarednode, minimum height= 1cm, minimum width = 2cm] (S) [below right=-.4cm and .75cm of C] {Simulation} ;
		\node[squarednode, minimum height= 1cm, minimum width = 1cm] (G) [above right=-.4cm and .75cm of S] {Generator} ;
		\node[squarednode, minimum height= 1cm, minimum width = 1cm] (TH) [below=.5cm of G] {$\theta$} ;
		\node[squarednode, minimum height= 1cm, minimum width = 1cm] (N) [right=.75cm of G] {Noise} ;
		\draw[blue, thick,-{Latex[length=2mm]}] (C.north) -- (P.south) node[midway, above] {};
		\draw[red, thick,-{Latex[length=2mm]}] (W.north) -- (C.south) node[midway, above] {};
		\draw[blue, thick,-{Latex[length=2mm]}] (T.west) -- (C.east) node[midway, above] {};
		\draw[blue, thick,-{Latex[length=2mm]}] (S.west) -- (C.east) node[midway, above] {};
		\draw[blue, thick,-{Latex[length=2mm]}] (G.west) -- (S.east) node[midway, above] {};
		\draw[blue, thick,-{Latex[length=2mm]}] (N.west) -- (G.east) node[midway, above] {};
		\draw[red, thick,-{Latex[length=2mm]}] (TH.north) -- (G.south) node[midway, above] {};
		\end{tikzpicture}
\caption{\label{fig:gan.1}
Basic structure of		
GAN's: $W$ is optimized to improve the prediction problem: ``real data'' vs. ``simulation''. Given $W$, $\th$ is optimized to worsen the prediction.}
		\end{center}
\end{figure}

\subsection{Objective function}
 Let $P_\th$ denote the distribution of $g(Z, \th)$, and $P_{\mathrm{true}}$ the target distribution of real data, represented by the variable $X$. 
One can formalize the ``real data'' vs. ``simulation'' problem with a  pair of random variables $(X_\theta, Y)$ where $Y$ follows a Bernoulli distribution with parameter $1/2$, and $P(X_\theta \in A \mid Y=y)$ is $P_{\mathrm{true}}(A)$ when $y=1$ and $P_\th(A)$ when $y=0$. 
 Given a loss function $r: \{0,1\} \times [0,1] \to [0, +\infty)$, one can define 
\[
U(\th, w) = \myE(r(Y, f(X_\theta, w)))
\]
and 
\[
U^*(\th) = \min_{w\in \mR^M} U(\th, w).
\]
 We want to maximize $U^*$ or, equivalently,  solve the optimization problem
\[
\th^* = \argmax_{\th \in \mR^K} \min_{w\in \mR^M} U(\th, w).
\]

Note that
\[
2 U(\th, w) = \myE(r(1, f(X, w))) + \myE(r(0, f(X_\theta, w)))
\]
so that choosing the cost requires to specify the two functions $t\mapsto r(1, t)$ and $t\mapsto r(0, t)$. 
 In \citet{goodfellow2014generative}, they are:
\begin{equation}
\label{eq:gan.original.cost}
\begin{aligned}
r(1,t) &=  -\log t\\
r(0,t) & = -\log (1-t).
\end{aligned}
\end{equation}

\subsection{Algorithm}
Using costs in \cref{eq:gan.original.cost}, one must compute
\[
\begin{aligned}
\th^* &= \argmin_{\th \in \mR^K} \max_{w\in \mR^M} \Big(\myE(\log f(X, w)) + \myE(\log(1-f(X_\theta, w)))\Big)\\
 &= \argmin_{\th \in \mR^K} \max_{w\in \mR^M} \Big(\myE(\log f(X, w)) + \myE(\log(1-f(g(Z, \th), w)))\Big).
 \end{aligned}
\]

Such min-max, or saddle-point problem are numerically challenging. The following algorithm was proposed in \citet{goodfellow2014generative}, and also includes a stochastic approximation component. Indeed,  in practice, the true distribution is only known through the observation of training data, say $x_1, \ldots, x_N$. Moreover, the distribution of $X_\theta$ is only accessible through Monte-Carlo simulation, so that both  expectations can only be approximated through finite-sample averaging.

\begin{algorithm}[GAN training algorithm]
\label{alg:gan}
\begin{enumerate}[label = \arabic*., wide=0.5cm]
\item Extract a batch of $m$ examples from training data, simulate $m$ samples according to $P_\th$ and run a few (stochastic) gradient ascent steps with fixed $\th$ to update $w$, replacing expectations by averages.
\item Generate $m$ new samples of $Z$ and update $\th$ with fixed $w$ by iterating a few steps of (stochastic) gradient descent.
\end{enumerate}
\end{algorithm}
\bigskip

\subsection{Associated probability metric and Wasserstein GANs}
Let $\CF$ be the family of all measurable functions: $f: \mR^d\to [0,1]$.
Given two  random variables $X_1, X_2:\Omega\to \mR^d$, with respective distributions $P_1, P_2$ (so that $\myP(X_i\in A) = P_i(A)$, consider the function
\[
D(P_1, P_2) = 2\log 2 + \max_{f\in\CF} \Big(\myE(\log f(X_1)) + \myE(\log(1-f(X_2)))\Big)
\]
 Assume that $X_1$ (resp. $X_2$) has a p.d.f. $g_1$ (resp. $g_2$) with respect to some measure $\mu$.
 Then
\[
\myE(\log f(X_1)) + \myE(\log(1-f(X_2))) = \int_{\mR^d} (g_1\log f + g_2\log(1-f)) d\mu
\]
which is maximal at $f^* = g_1/(g_1+g_2)$.
 For this $f^*$, 
\[
\begin{aligned}
2\log 2 + \myE(\log f^*(X_1)) + \myE(\log(1-f^*(X_2))) 
&= \int_{\mR^d} g_1 \log \frac{2g_1}{g_1+g_2} dx + \int_{\mR^d} g_2 \log \frac{2g_2}{g_1+g_2} d\mu\\
&= \KL\left(g_1\ \big\|\ \frac{g_1+g_2}2\right) + \KL\left(g_2\ \big\|\ \frac{g_1+g_2}2\right).
\end{aligned}
\]

 This last expression is the {\em Jensen-Shannon divergence} between $g_1$ and $g_2$ (cf. \cref{sec:divergence}).   One can then define
\[
\what D(P_1, P_2) = 
\max_{w\in \mR^M} \Big(\myE(\log f(X_1, w)) + \myE(\log(1-f(X_2, w)))\Big)
\]
as an approximation of $D$ in which the set of all possible functions with values in $[0,1]$ is replaced by those arising from the GAN classification network, parametrized by $w$.
 This approximation is useful when $g_1, g_2$ are only observable through random sampling or simulation.
 With this interpretation, GANs minimize $\what D(P_{\mathrm{true}}, P_\th)$.

 This discussion suggests that new types of GAN's may be designed using other discrepancy functions between probability distributions, provided that they can be expressed in terms of the maximization of some quantity over some space of functions.
  Consider, for example, the norm in total variation, defined by (for discrete distributions)
\[
D_{\mathrm{var}}(P_1,P_2) = \frac 12 \sum_{x} |P_1(x) - P_2(x)|.
\]
or, in the general case $D_{\mathrm{var}}(P_1,P_2)  = \max_A(P_1(A) - P_2(A))$.

 If $\CF$ is the space of continuous functions $f: \mR^d \to [0,1]$, then we also have (cf. \cref{prop:var.dist})
\[
D_{\mathrm{var}}(P_1,P_2)  = \max_{f\in\CF} (\myE(f(X_1)) - \myE(f(X_2))).
\]
 Since neural nets typically generate continuous functions with values in $[0,1]$, one could train GANs by maximizing 
\[
\what D_{\mathrm{var}}(P_1, P_2) = 
\max_{w\in \mR^M} \Big(\myE(f(X_1, w)) - \myE(f(X_2, w))\Big).
\]
However, the total variation distance can be too crude as a way to compare probability distributions, especially when these distributions have atoms (points $x$ such that $P_i(\{x\})>0$. For example, the total variation distance between two Dirac distributions at, say,  $x_1$ and $x_2$ in $\mR^d$ is always 1, unless $x_1=x_2$. As a consequence, if $x_n$ converges to $x$, with $x_n\neq x$, then $D_{\mathrm{var}}(\delta_{x_n}, \delta_x)\not\to 0$.

 The  Monge-Kantorovich distance (introduced in \cref{sec:monge}) is better adapted. Using \cref{eq:monge} with $\rho(x,y) = |x-y|$ and \cref{th:kant.rub}, we have
\begin{equation}
\label{eq:gan.MK}
D_{\mathrm{MK}}(P_1,P_2)  = \max_{f\in\CF} (\myE(f(X_1)) - \myE(f(X_2)))
\end{equation}
where $\CF$ is now the space of contractive (or 1-Lipschitz) functions. 
 Using the fact that a neural network with all weights bounded by a constant $K$ generates a function whose Lipschitz constant is controlled solely by $K$, one can then approximate (up to a  multiplicative constant) the Wasserstein distance by
\[
\what D_{\mathrm{MK}}(P_1,P_2)  = \max_{w\in \mathbb W} \Big(\myE(f(X_1, w)) - \myE(f(X_2, w))\Big)
\]
where $\mathbb W$ is the set of all weights bounded by a fixed constant. Wasserstein GANs (WGANs \citep{arjovsky2017wasserstein}) must then solve the saddle-point problem 
\[
U(\th, w) = \max_{w\in \mathbb W} \Big(\myE(f(X, w)) - \myE(f(X_\theta, w))\Big)
\]
and 
\[
U^*(\th) = \min_{w\in \mR^M} U(\th, w), 
\]
with an algorithm similar to that described earlier.

\begin{remark}
\label{rem:improved.gan}
As a final reference, we note the ``improved WGAN'' algorithm introduced in \citet{gulrajani2017improved} in which the boundedness constraint in the weights is replaced by an explicit control of the derivative in $x$ of the function $f$. Given independent $X_1$ and $X_2$ with respective distributions $P_1$, $P_2$, define a random variable $Z = (1-U) X_1 + U X_2$ where $U$ is uniformly distributed over $[0, 1]$ and $(U, X_1, X_2)$ are independent. Then, \citet{gulrajani2017improved} use the following approximation of the Wasserstein distance between $P_1$ and $P_\th$:
\[
\what D_{\mathit{MK}}(P_{\mathrm{true}},P_\th)  = \max_{w\in \mathbb W} \Big(E_{\mathrm{true}}(f(X, w)) - E_{\th}(f(X, w)) - \tilde E_\th((|\prt_z f(Z, w)| - 1)^2) 
\Big).
\]
This approximation is justified by the fact that optimal solutions of \cref{eq:gan.MK} satisfy (almost surely for the optimal coupling)
\[
f(y) - f(x) = |x-y|
\]
(see \citet{villani2009optimal}, theorem 5.10 and \citet{gulrajani2017improved}).
\end{remark}

\section{Reversed Markov chain models}

\subsection{General principles}
The discussions in \cref{sec:gen.nondiff,sec:gan} can be applied to sequences of structural equations (describing finite Markov chains) in the form
\[
\left\{
\begin{aligned}
Z_0 &= \bfxi_0\\
Z_{k+1} &= g(Z_k, \bfxi_k; \theta_k), \ k=0, \ldots, m-1\\
X &= Z_m
\end{aligned}
\right.
\]
where $\bfxi_0, \ldots, \bfxi_{m-1}$ are random variables with fixed distribution. Indeed, letting $\tilde Z = (\bfxi_0, \ldots, \bfxi_{n-1})$ and $\tilde \theta = (\theta_0, \ldots, \theta_{m-1})$ the whole system can be considered as a function $X = G(\tilde Z, \tilde{\theta})$ as considered in these sections. This representation, however, includes a large number of hidden variables, and it is unclear whether much improvement can be added to the case $m=1$ to justify the additional computational load. 

While direct Markov chain modeling may have a limited appeal, reversed Markov chains use a different generative approach in that they first model a forward Markov chain $Z_n, n\geq 0$ which is ergodic with known (and easy to sample from) limit distribution $Q_\infty$, and initial distribution $Q_{\mathrm{true}}$, the true distribution of the data. If one fixes  a large enough number of steps, say, $\tau$, then it is reasonable to assume that $Z_\tau$ approximately follows the limit distribution, $Q_\infty$. One can then (approximately) sample from $Q_{\mathrm{true}}$ by sampling $\tilde Z_0$ according to $Q_\infty$ and then applying $\tau$ steps of the time-reversed Markov chain.  

Reversed chains were discussed in \cref{sec:mcmc.reversible}. Assuming that $Q_{\mathrm{true}}$ and $P(z, \, \cdot\, )$ have a density with respect to a fixed measure $\mu$ on $\CR_Z$,  we found that $\tilde Z_k = Z_{\tau-k}$ is a non-homogeneous Markov chain whose transition probability $\tilde P_k(x, A) = P(\tilde Z_{k+1} \in A\mid \tilde Z_k = x)$ has density
\[
\tilde p_k(x,y) =  \frac{p(y,x)q_{\tau-k-1}(y)}{q_{\tau-k}(x)}
\]
with respect to $\mu$, where $q_n$ is the p.d.f. of $Q_n = Q_{\mathrm{true}} P^n$, the distribution of $Z_n$.

The distributions $Q_n, n\geq 0$ are unknown, since they depend on the data distribution $P_{\mathrm{true}}$, and the transition probabilities above must be estimated from data to provide a sampling algorithm from the reversed Markov chain. While, at first glance, this does not seem like a simplification of the problem, because one now has to sample from a potentially large number ($\tau$) of distributions instead of one, this leads, with proper modeling and some intensive learning, to powerful generative models.

To make this approach more efficient, the forward chain should be making small changes to the current configuration at each step (e.g., adding a small amount of noise). This ensures that the reversed transition probabilities $\tilde p_k(x, \cdot)$ are close to Dirac distributions and are therefore likely to be well approximated by simple unimodal distributions such as Gaussians. Importantly, the estimation problem does not have hidden data: given an observed sample, one can simulate $\tau$ steps of the forward chain to obtain, after reversing the order, a full observation of the reversed chain. Moreover, in some cases, analytical considerations can lead to partial computations that facilitate the modeling of the reversed transitions.

\subsection{Binary model}
We now take some examples, starting with a discrete case. Let $Q_{\mathrm{true}}$ be the distribution of a binary random field with state space $\{0, 1\}$ over a set of vertexes $V$, i.e., with the notation of \cref{sec:random.fields}, $\CR_X = \CF(V)$ with $F = \{0,1\}$. Fix a small $\ep>0$ and define the transition probability $p(x,y)$ for $x,y \in \CF(V)$ by
\[
p(x,y) = \prod_{s\in V} \left((1-\epsilon) \bfone_{\pe y s = \pe x s} + \epsilon\bfone_{\pe y s = 1 - \pe x s}\right).
\]
Since $p(x,y) >0$ for all $x$ and $y$, the chain converges (uniformly geometrically) to its invariant probability $Q_\infty$ and one easily checks that this probability is such that all variables are independent Bernoulli random variables with success probability $1/2$. Assuming that $\tau$ is large enough so that $Q_\tau \simeq Q_\infty$, the sampling algorithm initializes the reversed chain as independent Bernoulli$(1/2)$ variables and runs $\tau$ steps using the transitions $\tilde p_k$, which must be learned from data. The following proposition provides a first-order approximation of $\tilde p_k$. We  write $x\sim_s y$ if $\pe ys = 1 - \pe x s$ and $\pe yt = \pe xt$ for $s\neq t$, and write $x\sim y$ if $x\sim_s y$ for some $s$.

\begin{proposition}
\label{prop:reversed.mc.bin}
For the model above, let $\pe{\sigma_k}s(x) = \frac{q_{\tau-k}(y)}{q_{\tau-k}(x)}$. Then, one has $\tilde p_k(x,y) = \hat p_k(x,y) + O(\epsilon^2)$ with
\[
\hat p_k(x,y) = \prod_{s\in V} \left((1-\epsilon \pe{\sigma_k}s(x)) \bfone_{\pe y s = \pe x s} + \epsilon \pe{\sigma_k}s(x)\bfone_{\pe y s = 1 - \pe x s}\right)\,.
\]
\end{proposition}
\begin{proof}
We have 
\[
q_k(x) = \sum_{y\in \CF(V)} q_{k-1}(y) p(y,x).
\]
The probability of flipping two or more values of $y$ using $p(y,x)$ is 
\[
1 - (1-\epsilon)^N - N\epsilon (1-\epsilon)^{N-1}  = \frac{N(N-1)}2 \epsilon^2 + o(\epsilon^2)
\]
with $N = |V|$. As a consequence, we have
\[
q_k(x) = (1-N\epsilon) q_{k-1}(x) + \epsilon \sum_{y: y\sim x} q_{k-1}(y) + O(\epsilon^2)
\]

Since it implies that $q_k(x) = q_{k-1}(x) + O(\epsilon)$, this expression can be reversed by writing
\[
q_{k-1}(x) = \frac{1}{1-N\epsilon} q_k(x) - \frac{\epsilon}{1-N\epsilon} \sum_{y:y\sim x} (q_k(y) + O(\epsilon)) + O(\epsilon^2)
\]
which yields
\[
q_{k-1}(y) = (1+N\epsilon) q_{k}(y) - \epsilon \sum_{x: x\sim y} q_{k}(x) + O(\epsilon^2)
\]
Similarly, we have
\[
p(y,x) = (1-N\epsilon) \bfone_{x=y} + \epsilon \bfone_{x \sim y } + O(\epsilon^2).
\]
This gives
\[
p(y,x)q_{k-1}(y) =  q_{k}(x) \bfone_{x=y} - \epsilon \bfone_{x=y}  \sum_{x': x'\sim y} q_{k}(x' ) + \epsilon
q_k(y) \bfone_{x\sim y } + O(\epsilon^2),
\]
and we finally get
\[
\tilde p_k(x,y) = \left(1 - \epsilon   \sum_{x': x'\sim y} \frac{ q_{\tau-k}(x' )}{q_{\tau-k}(x)}\right) \bfone_{x=y} + \epsilon
\frac{q_{\tau-k}(y)}{q_{\tau-k}(x)} \bfone_{x\sim y } + O(\epsilon^2)
\]

If one lets $\pe{\sigma_k}s(x) = \frac{q_{\tau-k}(y)}{q_{\tau-k}(x)}$ with $y\sim_s x$, and defines
\[
\hat p_k(x,y) = \prod_{s\in V} \left((1-\epsilon \pe{\sigma_k}s(x)) \bfone_{\pe y s = \pe x s} + \epsilon \pe{\sigma_k}s(x)\bfone_{\pe y s = 1 - \pe x s}\right),
\]
one then checks easily that $\hat p_k(x,y) = \tilde p_k(x,y) + O(\epsilon^2)$.
\end{proof}

 This proposition suggests modeling the reversed chain using transitions $\hat p_k$, for which the mapping $x\mapsto (\pe{\sigma_k}s(x), s\in V)$ needs to be learned from data (for example using a deep neural network).   The approximation of the reversed chain, that we still denote $\hat p$ takes the form, with a parameter $\theta$,
\[
\hat p_k(x,y, \theta) = \prod_{s\in V} \left((1-\epsilon \pe{f}s(k, x, \theta)) \bfone_{\pe y s = \pe x s} + \epsilon \pe{f}s(t, x, \theta)\bfone_{\pe y s = 1 - \pe x s}\right),
\]
Given training $T = (x_1, \ldots, x_N)$ one can now define the random variable $Z = (X(\tau),\ldots, X(0))$ which reverses the forward Markov chain with $\mu_0$ equal to the uniform distribution on $T$ ($X(0)$ sampled uniformly at random among $x_1, \ldots, x_N$) and generate an arbitrary large number of independent samples of $Z$, say, $T' = (z_1, \ldots, z_M)$. The log-likelihood of this sample is given by
\begin{align*}
&= \sum_{j=1}^M \sum_{k=0}^{\tau-1} \log \hat p_k(z_j(\tau-k), z_j(\tau-k-1), \theta) \\
&= \sum_{j=1}^M \sum_{k=0}^{\tau-1} \sum_{s\in V} \left(\log(1-\epsilon \pe{f}s(k, x, \theta)) \bfone_{\pe y s = \pe x s} + \log(\epsilon \pe{f}s(t, x, \theta)) \bfone_{\pe y s = 1 - \pe x s}\right)
\end{align*}

 Note that $1 - \sigma_k(x)$ is precisely the score function introduced for discrete distributions in \cref{rem:score.discrete}.

\subsection{Model with continuous variables}
We now switch to an example with vector-valued variables, $\CR_X = \mR^d$, and assume that the forward Markov chain is such that, conditionally to $X_n=x$,  
\[
X_{n+1} \sim \CN(x + h f(x), \sqrt h\Id[d]),
\]
 where $f$ is $C^1$. We saw in \cref{sec:mcmc.rd} that, when $f = -\nabla H/2$ for a $C^2$ function $H$ such that $\exp(-H)$ is integrable, this chain converges (approximately for small $h$) to a limit distribution with p.d.f. (with respect to Lebesgue's measure) proportional to $\exp(-H)$.  In the linear case, in which $f(x) = -Ax/2$ for some positive definite symmetric matrix $A$, so that $H(x) = \frac12 x^TAx$, the limit distribution can be identified exactly as $\CN(0, \Sigma_h)$ where $\Sigma_h$ satisfies the equation
\[
A\Sigma_h + \Sigma_h A - \frac{h}{2} A^2 - 2\Id[d] = 0
\]  
whose solution is $\Sigma_h = (A - hA^2/4)^{-1}$
(details being left to the reader). This implies that this limit distribution can be easily sampled from, for any choice of $A$. 

We now return to general $f$'s and make, like in the discrete case, a first-order approximation  of the reversed chain. 
\begin{proposition}
\label{prop:reversed.mc.gauss}
Let $q_n$ denote the p.d.f. of the $X_n$, defined above. 
Assume that the second derivative of $q_n$ is bounded by a constant that does not depend on $n$. Then, for any smooth function $\gamma: \mR^d\to \mR$ with bounded first and second derivatives, one has
\begin{equation}
\label{eq:reversec.mc.gauss}
\myE(\gamma(X_{k-1}) \mid X_{k} = x) = \gamma(x) - h \nabla \gamma(x)^T f(x) - h\nabla \gamma(x)^T s_{k-1} (x)  + \frac{h}{2} \Delta \gamma(x) + o(h)
\end{equation}
with $s_{k-1}(x) = - \nabla \log q_{k-1}(x)$.
\end{proposition}
\begin{proof}

Considering the reversed chain, and letting $q_k$ denote the p.d.f. of $X_k$ for the forward chain, we have
\begin{align*}
\myE(\gamma(X_{k-1}) \mid X_{k} = x) & = \int_{\mR^d} \gamma(y) \tilde p_{k}(x,y) dy\\
& = \int_{\mR^d} \gamma(y)  p(y,x)  \frac{q_{k-1}(y)}{q_k(x)}dy\\
& = \frac{1}{(2\pi h)^{d/2}}\int_{\mR^d} \gamma(y)    \frac{q_{k-1}(y)}{q_k(x)} e^{-\frac1{2h} |x - y - h f(y)|^2} dy\\
& = \frac{1}{(2\pi)^{d/2}}\int_{\mR^d} \gamma(x - \sqrt h u)    \frac{q_{k-1}(x - \sqrt h u)}{q_k(x)} e^{-\frac1{2} |u - \sqrt h f(x - \sqrt h u)|^2} du,
\end{align*}
with the change of variable $u = (x - y)/\sqrt h$. We make a first-order expansion of the terms in this integral, with
\begin{align*}
\gamma(x - \sqrt h u)  q_{k-1}(x - \sqrt h u) &= \gamma(x)  q_{k-1}(x) - \sqrt h \nabla (\gamma q_{k-1})(x)^T u + \frac h2 u^T \nabla^2(\gamma q_{k-1})(x) u \\
& + O( h^{\frac32} |u|^3).
\end{align*}
and
\begin{align*}
e^{-\frac1{2} |u - \sqrt h f(x - \sqrt h u)|^2} &=  e^{-\frac1{2}|u|^2} e^{ \sqrt h u^T f(x) - h u^Td f(x) u - \frac12 |f(x)|^2 + o(h)}\\
&=  e^{-\frac1{2}|u|^2} \left(1 + \sqrt h u^T f(x) - h u^Td f(x) u - \frac h2 |f(x)|^2 + \frac h2  |u^T f(x)|^2 + |u|^3 O(h^{\frac32})\right).
\end{align*}
Taking products
\begin{align*}
&\gamma(x - \sqrt h u)  q_{k-1}(x - \sqrt h u) e^{-\frac1{2} |u - \sqrt h f(x - \sqrt h u)|^2} \\
&=  e^{-\frac1{2}|u|^2} \gamma(x)  q_{k-1}(x) \left(1 + \sqrt h u^T f(x) - h u^Td f(x) u - \frac h2 |f(x)|^2 + \frac h2  |u^T f(x)|^2\right)\\
& + e^{-\frac1{2}|u|^2}\left(-\sqrt h \nabla (\gamma q_{k-1})(x)^T u - h (\nabla (\gamma q_{k-1})(x)^T u)(f(x)^Tu) + \frac h2 u^T \nabla^2(\gamma q_{k-1})(x) u \right) + |u|^3 O(h^{\frac32})
\end{align*}
We now take the integral with respect to $u$ (recall that $\myE(U^T A U) = \trace(A)$ if $A$ is any square matrix and $U$ is standard Gaussian), so that
\begin{align*}
&\frac{1}{(2\pi)^{d/2}}\int_{\mR^d}\gamma(x - \sqrt h u)  q_{k-1}(x - \sqrt h u) e^{-\frac1{2} |u - \sqrt h f(x - \sqrt h u)|^2} du\\
&=  \gamma(x)  q_{k-1}(x) + h \left( -\gamma(x)  q_{k-1}(x) \nabla \cdot f(x) 
- \nabla (\gamma q_{k-1})(x)^T f(x) + \frac 12 \Delta(\gamma q_{k-1})(x)\right) + O(h^{\frac32})\\
&=  q_{k-1}(x) \left(\gamma(x)  + h \left( -\gamma(x)  \nabla \cdot f(x) 
- \left(\frac{\nabla (\gamma q_{k-1})(x)}{q_{k-1}(x)}\right)^T f(x) + \frac 12 \frac{\Delta(\gamma q_{k-1})(x)}{q_{k-1}(x)}\right)\right) + O(h^{\frac32}),
\end{align*}
where $\nabla\cdot f$ is the divergence of $f$.

To compute an expansion of  $q_k(x)$, it suffices to take $\gamma = 1$ above, so that
\[
q_k(x) =   q_{k-1}(x) \left(1  + h \left( - \nabla \cdot f(x) 
- \left(\frac{\nabla q_{k-1}(x)}{q_{k-1}(x)}\right)^T f(x) + \frac 12 \frac{\Delta q_{k-1}(x)}{q_{k-1}(x)}\right)\right) + O(h^{\frac32}).
\]
It remains to take the first-order expansion of the ratio, removing terms that cancel, getting 
\[
\myE(\gamma(X_{k-1}) \mid X_{k} = x) = \gamma(x) - h \nabla \gamma(x)^T f(x) + h\nabla \gamma(x)^T\left(\frac{\nabla q_{k-1}(x)}{q_{k-1}(x)} \right) + \frac{h}{2} \Delta \gamma(x) + O(h^{\frac32}).
\]
\end{proof}

We note that, for the forward chain,
\[
\myE(\gamma(X_{k+1})\mid X_k = x) = \myE(\gamma(x + h f(x) + \sqrt hU))
\]
where $U \sim \CN(0, \Id[d])$. Making the second order expansion
\begin{align*}
\gamma(x + h f(x) + \sqrt hU) &= \gamma(x)  + \sqrt h \nabla \gamma(x)^T U + h \nabla \gamma(x)^Tf(x) + \frac{h}{2} U^T \nabla^2\gamma(x) U + |U|^3 O(h^{\frac32})
\end{align*}
and taking the expectation gives
\begin{equation}
\label{eq:forward.markov}
\myE(\gamma(X_{n+1})\mid X_n = x) = \gamma(x) + h \nabla \gamma(x)^Tf(x) +\frac{h}{2} \Delta \gamma(x) + O(h^{\frac32}).
\end{equation}

Comparing with \cref{eq:reversec.mc.gauss}, we find that  $\tilde X_k = X_{\tau-k}$ behaves, for small $h$, like the non-homogeneous Markov chain
such that the conditional distribution of $\tilde X_{k+1}$ given $\tilde X_{k} = x$ is $\CN(x - h f(x) - h s_{\tau-k-1}(x), \sqrt h \Id[d])$, with $s_{\tau-k-1}(x) = - \nabla \log q_{\tau-k-1}$, the score function introduced in \cref{sec:score.matching}. One can therefore use score-matching methods from that section to estimate this distribution from observations of the forward chain initialized with training data. 

\subsection{Continuous-time limit}
\label{sec:reversed.diffusion}

The forward schemes described in the previous examples can be interpreted as time discretizations of continuous-time processes over discrete or continuous variables. In the latter case, the example $X_{k+1} \sim \CN(x + h f(x), \sqrt h \Id[d])$ conditionally to $X_k = x$ is a discretization of the stochastic differential equation 
\[
dx_t = f(x_t) dt + dw_t
\]
(see \cref{rem:langevin}), where $w_t$ is a Brownian motion and the diffusion is initialized with $Q_{\mathrm{true}}$. We found that going backward meant (at first order and conditionally to $X_k=x$)
\[
X_{k-1} \sim \CN(x - h f(x) - h s_{k-1}(x), \sqrt h \Id)
\]
that we can rewrite as
\[
x_\tau - X_{k-1} \sim \CN(x_\tau - x + h f(x) + h s_{k-1}(x), \sqrt h \Id).
\]
Following the definition in \citet{anderson1982reverse}, this corresponds to a first-order discretization of the {\em reverse diffusion}
\[
dx_t = (f(x_t) + s_t(x_t)) dt + d\tilde w_t,\ t\leq \tau
\]
where $\tilde w_t$ is also a Brownian motion. This reverse diffusion with $X_\tau \sim Q_\infty$ will therefore approximately sample from $Q_{\mathrm{true}}$. (With this terminology, forward and reverse diffusions have similar differential notation, but mean different things.) Note that, in the continuous-time limit, the reverse Markov process follows the distribution of the reversed diffusion exactly.

\subsection{Differential of neural functions}

As we have seen in the previous two examples, estimating the reversed Markov chain requires computing the score functions of the forward probabilities. In the case of continuous variables, this score function is typically parametrized as  a neural network, so that the function $s_k(x) = -\nabla \log q_k(x)$ is computed as $s_k(x) = F(x;\CW_k)$, with the usual definition $F(x, \CW_k) = z_{m+1}$ with $z_{j+1} = \phi_j(z_j, w_{jk})$, $z_0 = x$ and $\CW_k =  (w_{0k}, \ldots, w_{mk})$. 

Assume that a training set $T$ is observed. Running the forward Markov chain initialized with elements of $T$ generates a new training set at each time step, that we will denote $T_k$ at step $k$. We have seen in \cref{sec:score.matching} that the score function $s_k$ could be estimated by minimizing, with respect to $\CW$
\[
 \sum_{x\in T_k} \left(|F(x,\CW) |^2  - 2\nabla \cdot  F(x,\CW)\right).
\]
This term involves the differential of $F$, which is defined recursively by (simply taking the derivative at each step)
\[
dF(x, \CW) = \zeta_{m+1}, \ \zeta_{j+1} = d\phi_j(z_j, w_{j})\zeta_j,
\]
 with $\zeta_0 = \Id[d]$. From this recursive definition, back-propagation can be applied, in principle, to compute the derivative of $dF(x, \CW)$ with respect to $\CW$.  The feasibility of this computation, however, is limited when $d$ is large ($d$ could be tens of thousands if one models images), since computing the $d\times d$ matrix $dF(x, \CW)$ is then intractable. 
 
One can note that, for any $h\in \mR^d$, the vector $dF(x, \CW) h$ also satisfies the recursion
\[
dF(x, \CW)h = \zeta_{m+1}h, \ \zeta_{j+1}h = d\phi_j(z_j, w_{j})\zeta_jh,
\]
with $\zeta_0h = h$ and 
\[
\nabla \cdot  F(x,\CW) = \sum_{i=1}^d \mathfrak e_i^T dF(x,\CW) \mathfrak e_i
\]
where $\mathfrak e_1, \ldots, \mathfrak e_d$ is the canonical basis of $\mR^d$. Putting the divergence of $F$ in this form does not reduce the computation cost (which is, roughly $d^2 m$, assuming that all $z_j$'s have the same dimension), but expresses the divergence term in a form that is amenable to stochastic gradient descent (which is typically already used to approximate the sum over $x$). Indeed, if $\xi$ follows any distribution with zero mean and covariance matrix equal to the identity (such as a standard Gaussian, or the uniform distribution on the unit sphere), then
\[
\nabla \cdot  F(x,\CW) = \myE(\xi^T dF(x,\CW) \xi)
\]
so that $\xi$ can be generated in minibatches in SGD implementations (see \citep{song2020sliced}, where this approach is called ``sliced score matching'').


\chapter{Clustering}
\label{chap:clustering}
\label{sec:clustering}
\section{Introduction}
We now describe a collection of methods designed to divide a training set into homogeneous subsets, or clusters. This grouping operation is a key problem in many applications for which it is important to categorize the data in order to obtain improved understanding of the sampled phenomenon, and sometimes to be able to apply a different approach to subsequent processing or analysis adapted to each cluster. 

We will assume that the variables of interest belong a  set $\CR= \CR_X$ where $\CR$ is equipped with a discrepancy function $\al: \CR\times \CR \to [0, +\infty)$. Often, $\al$ is derived from a distance $\rho$ on $\CR$, but this is not always the case. We will assume that the data results from a  training set  $T = (x_1, \ldots, x_N)$. However, it may happen that 
only the discrepancy matrix $A = (\al(x, y), x,y\in T)$ is observed, while a coordinate representation of the elements of $T$ is not  available.
 
Let us consider a few examples.
\begin{enumerate}[label=(\roman*), wide=.5cm]
\item The simplest case is when $\CR = \mR^d$ with the standard Euclidean metric.  Slightly more generally, a metric may be defined by $\rho^2(x,y) = \|h(x)-h(y)\|_H^2$, where $H$ is an inner-product space and the feature function $h:\CR \mapsto H$ may be unknown, while its associated ``kernel'', $K(x,y) = \scp{h(x)}{h(y)}_H$ is known (this is a metric if $h$ is one-to-one). In this case
\[
\rho^2(x,y) = K(x,x) - 2K(x,y) + K(y,y).
\]
Typically, one then takes $\al = \rho$ or $\al = \rho^2$.
\item
Very often, however, the data is not Euclidean, and the distance does not correspond to a feature space representation. This is the case, for example, for data belonging to ``curved spaces'' (manifolds), for which one may use the intrinsic distance provided by the length of shortest paths linking two points (assuming of course that this notion can be given a rigorous meaning). The simplest example is data on the unit sphere, where the distance $\rho(x,y)$ between two points $x$ and $y$ is the length of the shortest large circle that connects them, satisfying
\[
|x-y|^2 = 2 - 2\cos\rho(x,y).
\]
Once again, $\al = \rho$ or $\rho^2$ is a typical choice.
\item
 A more complex example is provided by $\CR$ being the space of symmetric positive-definite matrices on $\mR^d$, for which one defines the length of a differentiable curve $(S(t), t\in[a,b])$ in this space by
 \[
 \int_a^b \sqrt{\trace((S(t)^{-1}\prt_t S)(S(t)^{-1}\prt_t S)^T)} dt
 \]
 and for which
 \[
 \rho^2(S_1, S_2) = \sum_{i=1}^d (\log \la_i)^2 
 \]
 where $\la_1, \ldots, \la_d$ are the eigenvalues of $S_1^{-1/2} S_2 S_1^{-1/2}$ or, equivalently, solutions of the generalized eigenvalue problem $S_2 u = \la S_1 u$ (see, for example, \cite{fletcher2004principal}). 
  \item Another common assumption is that the elements of $\CR$ are vertices of a weighted graph of which $T$ is a subgraph; $\rho$ may then be, e.g., the geodesic distance on the graph.
 \end{enumerate}
 

\section{Hierarchical clustering and dendograms}

\subsection{Partition trees}
This method builds clusters by organizing them in a binary hierarchy in which the data is divided into subsets, starting with the full training set, and iteratively splitting each subset into two parts until reaching singletons. This results in a binary tree structure, called a {\em dendogram}, or partition tree, which is defined as follows.
\begin{definition}
\label{def:dendogram}
A partition tree of a finite set $A$ is a finite collection of nodes $\CT$ with the following properties.
\begin{enumerate}[label=(\roman*), wide=.5cm]
\item Each node has either zero  or exactly two children. (We will use the notation $v\to v'$ to indicate that $v'$ is a child of $v$. 
\item All nodes but one have exactly one parent. The node without parent is the {\em root} of the tree.  
\item To each node $v\in\CT$ is associated a subset $A_v\sub A$.
\item If $v'$ and $v''$ are the children of $v$, then $(A_{v'}, A_{v''})$ forms a partition of $A_v$.
\end{enumerate}
Nodes without children are called leaves, or  terminal nodes.
We will say that the hierarchy is complete  if $A_v=A$ if $v$ is the root, and  $|A_v|=1$ for all terminal nodes. 
\end{definition}
An example of partition tree is provided in \cref{fig:dendogram}.

\begin{figure}
\centering
\begin{tikzpicture}
\node (1) {1: $\{a,b,c,d,e,f\}$};
\node (2) [below left=.5cm and -.1cm of 1] {2: $\{a,c,f\}$};
\node (3) [below right=.5cm and -.1cm of 1] {3: $\{b,d,e\}$};
\node (4) [below left=.5cm and -.1cm of 2] {4: $\{a,f\}$};
\node (5) [below right=.5cm and -.1cm of 2] {5: $\{c\}$};
\node (6) [below left=.5cm and -.1cm of 3] {6: $\{d\}$};
\node (7) [below right=.5cm and -.1cm of 3] {7: $\{b,e\}$};
\node (8) [below left=.5cm and -.1cm of 4] {8: $\{a\}$};
\node (9) [below right=.5cm and -.1cm of 4] {9: $\{f\}$};
\node (10) [below left=.5cm and -.1cm of 7] {10: $\{b\}$};
\node (11) [below right=.5cm and -.1cm of 7] {11: $\{e\}$};

\draw[->] (1.south) -- (2.north east);
\draw[->] (1.south) -- (3.north west);
\draw[->] (2.south) -- (4.north east);
\draw[->] (2.south) -- (5.north west);
\draw[->] (4.south) -- (8.north east);
\draw[->] (4.south) -- (9.north west);
\draw[->] (3.south) -- (6.north east);
\draw[->] (3.south) -- (7.north west);
\draw[->] (7.south) -- (10.north east);
\draw[->] (7.south) -- (11.north west);
\end{tikzpicture}
\caption{\label{fig:dendogram}
A partition tree of the set $\{a,b,c,d,e,f\}$.}
\end{figure}

The construction of the tree can follow two directions, the first one being bottom-up, or agglomerative, in which the algorithm starts with the collection of all singletons and merges subsets one pair at a time until everything is merged into the full dataset. The second approach is top-down, or divisive, and initializes the algorithm with the full training set which is recursively split until singletons are reached. The first approach, on which we now focus, is more common, and computationally simpler. 

We let $T$ denote the training set and assume that a matrix of  dissimilarities 
\[
(\al(x,y),\, x,y\in T)
\]
 is given. We will make the abuse of notation of considering that $T$ is a set even though some of its elements may be repeated. This is no loss of generality, since $T = (x_1, \ldots, x_N)$ can always be replaced by the subset $\{ (k,x_k), k=1, \ldots, N\}$ of $\mN \times \CR$.

\subsection{Bottom-up construction}

We will extend $\al$ to a dissimilarity measure between subsets $A, A'\sub T$ that we will denote  $(A, A') \mapsto \phi(A, A')$. Once $\phi$ is defined, agglomeration works along the following  algorithm.
\begin{algorithm}
\label{alg:hierarchical.bup}
\begin{enumerate}[label=\arabic*., wide=0.5cm]
\item Start with the collection $\CT_1, \ldots, \CT_N$ of all single-node trees associated to each element of $T$. Let $n=0$ and $m=N$.
\item Assume that, at  step $n$ of the algorithm, one has a collection of partition trees $\CT_1, \ldots, \CT_m$ with root nodes $r_1, \ldots, r_{m}$ associated with subsets $A_{r_1}, \ldots, A_{r_m}$ of $T$. Let the total collection of nodes be indexed as $\CV_n = \{v_1, \ldots, v_{N+n}\}$, so that $\{r_1, \ldots, r_m\} \subset \CV_n$.
\item If $m=1$, stop the algorithm. 
\item Select indices $i,j\in \{1, \ldots, m\}$ such that $\phi(A_{r_i}, A_{r_j})$ is minimal, and merge the corresponding trees by creating a new node $v_{n+1+N}$ with the root nodes of $\CT_i$  and $\CT_j$ as children (so that $v_{n+1+N}$ is associated with $A_{r_i}\cup A_{r_j}$). Add $v_{n+1+N}$ to the collection of root nodes, and remove $r_i$ and $r_j$.
\item Set $n\to n+1$ and $m\to m-1$ and return to step 2.
\end{enumerate}
\end{algorithm}

Clearly,  the specification of the extended dissimilarity measure ($\phi$) is a key element of the method.
Some of  most commonly used extensions are:
\begin{enumerate}[label=$\bullet$,wide=0.5cm]
\item Minimum gap: $\phi_{\min}(A, A') = \min(\al(x,x'): x\in A, x'\in A')$.
\item Maximum dissimilarity: $\phi_{\max}(A, A') = \max(\al(x,x'): x\in A, x'\in A')$.
\item Sum of dissimilarities: \[
\phi_{\text{sum}}(A,A') =  \sum_{x\in A} \sum_{x'\in A'} \al(x,x')
\]
\item Average dissimilarity: \[
\phi_{\text{avg}}(A,A') = \frac{1}{|A|\,|A'|} \sum_{x\in A} \sum_{x'\in A'} \al(x,x').
\]
\end{enumerate}

As shown in the next two propositions, the maximum distance favors clusters with small diameters, while 
using minimum gaps tends to favor connected clusters.
\begin{proposition}
\label{prop:aggl.max}
Let $\diam(A) = \max(\al(x,y), x,y\in A)$.
The agglomerative algorithm using $\phi_{\max}$ is identical to that using $\phi(A,A') = \diam(A\cup A')$.
\end{proposition}
\begin{proof}
Call Algorithm 1 the agglomerative algorithm using $\phi_{\max}$, and Algorithm 2 the one using $\phi$. 
At initialization, we have (because all sets are singletons), 
\begin{equation}
\label{eq:aggl.max}
\phi_{\max}(A_k, A_l) = \diam(A_k\cup A_l) \text{ for all }1 \leq k\neq l \leq m. 
\end{equation}

 We show that this property remains true at all steps of the algorithms. Proceeding by induction, assume that, up to the step $n$, Algorithms 1 and 2 have been identical and result in sets $(A_1, \ldots, A_m)$ satisfy \cref{eq:aggl.max}. Then the next steps of the two algorithms coincide and 
 assume, without loss of generality, that this next step merges $A_{m-1}$ with $A_m$.
 Let $A'_{m-1} = A_{m-1} \cup A_m$ so that 
$\diam(A'_{m-1}) \leq \diam(A_i\cup A_j)$ for all $1\leq i\neq j\leq m$. 

We need to show that the new partition satisfies \cref{eq:aggl.max}, which requires that 
\[
\phi_{\max}(A'_{m-1}, A_k) = \diam(A'_{m-1} \cup A_k)\]
 for $k=1, \ldots, m-2$. 

 We have 
\[
\diam(A'_{m-1}\cup A_k) = \max(\diam(A'_{m-1}), \diam(A_k), \phi_{\max}(A'_{m-1}, A_k)),
\]
so that we must show that 
\[
\max(\diam(A'_{m-1}), \diam(A_k)) \leq \phi_{\max}(A'_{m-1}, A_k).
\] 
 Write
\begin{align*}
\phi_{\max}(A'_{m-1}, A_k) &= \max(\phi_{\max}(A_m, A_k), \phi_{\max}(A_{m-1}, A_k)) \\
& = \max(\diam(A_m\cup A_k), \diam(A_{m-1}\cup A_k))
\end{align*}
where the last identity results from the induction hypothesis. 

 The fact that 
\[
\diam(A_k) \leq \max(\diam(A_m\cup A_k), \diam(A_{m-1}\cup A_k))
\]
is obvious, and the inequality 
\[
\diam(A'_{m-1}) \leq \max(\diam(A_m\cup A_k), \diam(A_{m-1}\cup A_k))
\]
results from the fact that $A_m$ and $A_{m-1}$ was an optimal  pair. This shows that the induction hypothesis remains true at the next step and concludes the proof of the proposition.
\end{proof}

\bigskip
We now analyze $\phi_{\min}$ and, more specifically, the equivalence between the resulting algorithm and the one using the following measure of connectedness.  For a given set $A$ and $x,y\in A$, let 
\begin{multline*}
\tilde \al_A(x,y) = \inf\big\{\ep: \exists n>0, \exists (x=x_0, x_1, \ldots, x_{n-1}, x_n=y) \in A^{n+1}:\\ \al(x_i, x_{i-1})\leq \ep \text{ for } 1\leq i \leq n\big\}.
\end{multline*}
So $\tilde \al_A$ is the smallest $\ep$ such that there exists a sequence of steps of size less than $\ep$ in $A$ going from $x$ to $y$. The function 
\[
\mathrm{conn}(A) = \max\{\tilde\al_A(x,y):x,y\in A\}
\]
measures how well the set $A$ is connected relative to the dissimilarity measure $\al$. and we have: 
\begin{proposition}
\label{prop:aggl.min}
The agglomerative algorithm using $\phi_{\min}$ is identical to that using $\phi(A,A') = \mathrm{conn}(A\cup A')$.
\end{proposition}
\begin{proof}
The proof is similar to that of \cref{prop:aggl.max}. Indeed one can note  that 
\[
\mathrm{conn}(A\cup A') = \max(\mathrm{conn}(A), \mathrm{conn}(A'), \phi_{\min}(A,A'))\,.
\]
Given this we can proceed by induction and prove that, if the current decomposition is   $A_1, \ldots, A_m$ such that $\psi(A_k \cup A_l)  = \phi_{\min}(A_k,  A_l)$ for all $1 \leq k\neq l \leq m$, then this property is still true after merging using $\phi_{\min}$ and $\phi$.

Assuming again that $A_{m-1}$ and $A_m$ are merged, and letting $A'_{m-1} = A_m \cup A_{m-1}$, we need to show that $\mathrm{conn}(A_k \cup A'_{m-1})  = \phi_{\min}(A_k,  A'_{m-1})$ for all $k=1, \ldots, m-2$, which is the same as showing that:
\[
\max(\mathrm{conn}(A_k), \mathrm{conn}(A'_{m-1})) \leq \phi_{\min}(A_k,  A'_{m-1}) = \min(\phi_{\min}(A_k,  A_{m-1}), \phi_{\min}(A_k,  A_{m})).
\]
From the induction hypothesis, we have
\[
\min(\phi_{\min}(A_k,  A_{m-1}), \phi_{\min}(A_k,  A_{m})) = \min(\mathrm{conn}(A_k \cup A_{m-1}), \mathrm{conn}(A_k \cup  A_{m}))
\]
and both terms in the right-hand side are larger than $\mathrm{conn}(A_k)$ and also larger than $\mathrm{conn}(A'_{m-1})$ which was a minimizer.
\end{proof}


\subsection{Top-down construction}

The agglomerative method is the most common way to build dendograms, mostly because of the simplicity of the construction algorithm. The divisive approach is more complex, because the division step, which  requires, given a set $A$, to optimize a splitting criterion over all two-partitions of $A$, may be significantly more expensive than the merging steps in the agglomerative algorithm. 
The top-down construction therefore requires the specification of a ``splitting algorithm'' $\sig: A \mapsto (A', A'')$ such that $(A', A'')$ is a partition of $A$.  We assume that, if $|A|>1$, then the partition $A,A''$ is not trivial, i.e., neither set is empty.

Given $\sig$, the top-down construction is as follows. 
\begin{algorithm}
\label{alg:hierarchical.tdown}
\begin{enumerate}[label=\arabic*., wide=0.5cm]
\item  Start with the one-node partition tree   $\CT_0= (T)$.
\item Assume that at a given step of the algorithm, the current partition is  $\CT$.
\item If $\CT$ is complete, stop the algorithm.
\item For each terminal node $v$ in $\CT$ such that $|A_v| > 1$, compute $(A'_v, A''_v) = \sig(A_v)$ and add two children $v'$ and $v''$ to $v$ with $A_{v'} = A'_v$ and $A_{v''} = A''_v$.
\item Return to step 2.
\end{enumerate}
\end{algorithm}
The division of a set into two parts is itself a clustering algorithm, and one may apply any of those described in the rest of this chapter.

\subsection{Thresholding}
Once a complete hierarchy is built, it provides a complete binary partition tree $\CT$. This tree provides in turn a collection of partitions of $\CV$, each of them obtained through pruning. We now formalize this operation.

Let $\CV_T$ denote the set of terminal nodes in $\CT$ and $\CV_0 = \CV \setminus \CV_T$ contain the interior nodes. Define a {\em pruning set} to be a subset  $\CD\sub \CV_0 $ that contains no pair of nodes $v,v'$ such that $v'$ is a descendant of $v$. To any pruning set $\CD$,  one can associate the pruned subtree $\CT(\CD)$ of $\CT$ consisting of $\CT$ from which  all the vertices that are descendants of elements of $\CD$ are removed. From any such pruned subtree, one obtain a partition $S(\CD)$ of $T$ formed by the collection of sets $A_v$ for $v$ in the terminal nodes of $\CT(\CD)$. Between the extreme case $S(v_0) = \{\CV\}$ (where $v_0$ is the root of $\CT$) and $S(\emp) = (\{x\}, x\in \CV_T)$, there exists a huge number of possible partitions obtained in this way.

It is often convenient to organize these partitions according to the level sets of  a well-chosen score function $v \mapsto h(v)$ defined over $\CV_0$. For $\CD \sub \CV$, we denote by $\max(\CD)$ the set of its deepest elements, i.e., the set formed by those $v\in\CD$ that have no descendant in $\CD$. 
Then, for any $\la \in\mR$, one can define $\CD^+_\la = \max\defset{v: h(v) \geq \la}$ (resp. $\CD^-_\la = \max \defset{v: h(v) \leq \la}$) and the associated partition $S(\CD^+_\la)$ (resp. $S(\CD^-_\la)$). 
The score function $h$ can be  linked to the construction algorithm. For example, if one uses a bottom-up construction using an extended dissimilarity $\phi$, one can associate to each node $v$ with $v\in \CV_0$ the value of $\phi(A_{v'}, A_{v''})$ where $v'$ and $v''$ are the children of $v$. 

Another way to define such scores functions is by assigning weights to edges in $\CT$. Indeed, given a collection $w$ of positive numbers $w(v,v')$ for $v\to v'$ in $\CT$, one can define a score $h_w$  recursively by letting $h_w(v_0) = 0$  and $h_w(v') = h_w(v) + w(v,v')$ if $v'$ is a child of $v$. The choice $w(v,v') = 1$ for all $v,v'$ provide the usual notion of depth in the tree. 

Scores can also be built bottom-up,  letting $h(v) = 0$ for terminal nodes and,  for $v\in \CV_0$,
\[
h_w(v) = \max(h_w(v') + w(v,v'), h_w(v'')+ w(v,v''))
\]
where $v', v''$ are the children of $v$
Here, taking $w=1$ provides the height of each node.

\section{K-medoids and K-mean}
\subsection{K-medoids}
One of the limitations of  hierarchical clustering  is that it is a greedy approach that does not optimize a global quality measure associated with the partition. Such quality measures can indeed be defined based on the  heuristic that clusters should be homogeneous (for some criterion) and far apart from each other. 

In centroid-based methods, the homogeneity criterion is the minimum, over all possible points in $\CR$, of the sum of dissimilarities between elements of the cluster and that  point. More precisely, for any $A\subset \CR$, and any dissimilarity measure $\al$,  define the {\em central dispersion} index
\begin{equation}
\label{eq:v.alpha}
V_\al(A) = \inf\left\{\sum_{x\in A} \al(x, c): c\in \CR\right\}.
\end{equation}
If $c$ achieves the minimum in the definition of $V_\alpha$, it is called a {\em centroid} of $A$ for the dissimilarity $\alpha$.

The most common choice is $\al = \rho^2$, where $\rho$ is a metric on $\CR$, and in this case, we will just use $V$ in place of $V_{\rho^2}$. Note also that it is always possible to limit $\CR$ to the training set $T$, in which case the optimization in \cref{eq:v.alpha} is over a finite number of centers. This makes centroid-based methods also applicable to the situation when the matrix of dissimilarities is the only input provided to the algorithm, or when the set $\CR$ and the function $\al$ are too complex for the optimization in \cref{eq:v.alpha} to be feasible.

A centroid, $c$, in \cref{eq:v.alpha} may not always exists, and when it exists it may not always be unique. For $\al = \rho^2$, a point $c$ such that 
\[
V(A) = \sum_{x\in A} \rho^2(x, c)
\]
is called a Fr\'echet mean of the set $A$. Returning to the examples provided in the beginning of this chapter, two antipodal points on the sphere (whose distance is $\pi$) have an infinity of Fr\'echet means (or midpoints in this case) provided by every point in the equator between them. In contrast, the example provided with symmetric matrices provides a so-called Hadamard space \cite{burago2001course} and the Fr\'echet mean in that case is unique.  Of course, for Euclidean metrics, the Fr\'echet mean is just the usual one.

Returning to our general discussion, the K-medoids method optimizes the sum of central dispersions with a fixed number of clusters. Note that the letter K in K-medoids originally refers to this number of clusters, but this notation conflicts with other notation in this book (e.g., reproducing kernels) and we shall denote by $p$ this target number\footnote{We  still call the method K-medoids rather than $p$-medoids, to keep the name universally used in the literature.}. So the K-medoids method minimizes
\[
W_\al(A_1, \ldots, A_p) = \sum_{i=1}^p V_\al(A_i)
\] 
over all partitions $A_1, \ldots, A_p$ of the training set $T$. Equivalently, it minimizes
\begin{equation}
\label{eq:kmed.2}
\mW_\al(A_1, \ldots, A_p, c_1, \ldots, c_p) = \sum_{i=1}^p \sum_{x\in A_i} \al(x, c_i)
\end{equation}
over all partitions of $T$ and $c_1, \ldots, c_p\in\CR$. Finally, taking first the minimum with respect to $A_i$, which corresponds to associating each $x$ to the subset with closest center, K-medoids, an equivalent formulation  minimizes
\[
\tilde W_\al(c_1, \ldots, c_p) = \sum_{x\in T} \min\big\{\al(x, c_i), i=1, \ldots, p\big\}.
\] 

The standard implementation of K-medoids solves this problem using an alternate minimization, as defined in the following algorithm.
\begin{algorithm}[K-medoids] 
\label{alg:kmed}
Let $T\subset \CR$ be the training set.
 Start with an initial choice of $c_1, \ldots, c_p\in \CR$ and iterate over the following two steps until stabilization:
\begin{enumerate}[leftmargin=1cm]
\item For $i = 1, \ldots, p$, let $A_i$ contain points $x\in T$ such that $\al(x, c_i) = \min\{\al(x, c_j), j=1, \ldots, p\}$. In case of a tie in this minimum, assign $x$ to only one of the tied sets (e.g., at random) to ensure that $A_1, \ldots, A_p$ is a partition.
\item For $i=1, \ldots, p$, let $c_i$ be a minimizer of $\sum_{x\in A_i} \al(x, c_i)$ if $A_i$ is not empty, or $c_i$ be a random point in $T$ otherwise.
\end{enumerate}
\end{algorithm}

It should be clear that each step reduces the total cost $\mW_\al$ and that this cost should stabilize at some point (which provides the stopping criterion) because there is only a finite number of possible partitions of $T$. However, there can be many possible limit points that are stable under the previous iterations, and some may correspond to poor ``local minima'' of the objective function. Since the end-point of the algorithm depends on the initialization, this step requires extra care. One may design {\it ad-hoc} heuristics in order to start the algorithm with a good initial point that is likely to provide a good solution at the end. These heuristics may depend on the problem at hand, or use a generic strategy. As a common example of the latter, one may ensure that the initial centers are sufficiently far apart by picking $c_1$ at random, $c_2$ as far as possible from $c_1$, $c_3$ maximizing the sum of distances to $c_1$ and $c_2$ etc. 
One also typically runs the algorithm several times with random initial conditions and select the best solution over these multiple runs. 

The second step of \cref{alg:kmed} can  be computationally challenging depending on the set $\CR$ and the dissimilarity measure $\al$. When $\CR = \mR^d$ and $\al = \rho^2$ is the square Euclidean distance, the solution is explicit and $c_i$ is simply the average of all points in $A_i$. The resulting algorithm is the original incarnation of K-medoids, and called K-means \cite{steinhaus1956division,lloyd1982least,macqueen1967some}. K-means is probably  the most popular clustering method and is often a step in  more advanced approaches, as we will discuss later. The two steps of \cref{alg:kmed} are then simplified as follows.
\begin{algorithm}[K-means] 
\label{alg:kmeans}
Let $T\subset \mR^d$ be the training set.
 Start with an initial choice of $c_1, \ldots, c_p\in \mR^d$ and iterate over the following two steps until stabilization:
\begin{enumerate}[leftmargin=1cm]
\item For $i = 1, \ldots, p$, let $A_i$ contain points $x\in T$ such that $|x - c_i|^2 = \min\{|x - c_j|^2, j=1, \ldots, p\}$. In case of tie in this minimum, assign $x$ to only one of the tied sets (e.g., at random) to ensure that $A_1, \ldots, A_p$ is a partition.
\item For $i=1, \ldots, p$, let 
\[
c_i = \frac1{|A_i|} \sum_{x\in A_i} x
\]
if $A_i$ is not empty, or $c_i$ be a random point in $T$ otherwise.
\end{enumerate}
\end{algorithm}

\subsection{Mixtures of Gaussian and deterministic annealing}
Mixtures of Gaussian (MoG) 
were discusssed in \cref{chap:var.bayes} and in \cref{alg:mix.gauss}.  
Recall that they model the observed data $X$ together with a latent class variable $Z\in \{1, \ldots, p\}$ with joint distribution
\[
f(x,z;\theta) =  (2\pi)^{-\frac d 2} (\det \Sig_z)^{-\frac12} \al_z e^{-\frac12 
(x-c_z)^T\Sig_z^{-1} (x-c_z)}
\]
where $\th$ contains the weights, $\al_1, \ldots,
\al_p$, the means, $c_1, \ldots,
c_p$ and the covariance matrices $\Sig_1, \ldots, \Sig_p$ (we create, hopefully without risk of confusion, a short-lived conflict of notation between the weights and the dissimilarity function). The posterior class probabilities 
\[
f_Z(i\mid x\,;\,\theta) = \frac{(\det \Sig_i)^{-\frac12} \al_i e^{- \frac12
(x-c_i)^T\Sig_i^{-1} (x-c_i)}}{\sum_{j=1}^p (\det \Sig_j)^{-\frac12} \al_j e^{-\frac12
(x-c_j)^T\Sig_j^{-1} (x-c_j)}}, \quad
i=1, \ldots, p,
\]
which are computed in step 3 of \cref{alg:mix.gauss} can be interpreted as a likelihood that observation $x$ belongs to group $i$.  As a consequence, the mixture of Gaussian algorithm can also be seen as a clustering method, in which one assigns each $x\in T$ to cluster $i$ when $i = \argmax\{f_Z(j\mid x, \theta): j=1, \ldots, p\}$, making an arbitrary decision in case of a tie.

In the special case in which all variances are fixed and equal to  $\sig^2\Id[d]$, and all prior class probabilities are equal to $1/p$ (see \cref{rem:mog}), the EM algorithm for mixtures of Gaussian is also called ``soft K-means'', because it replaces the ``hard'' cluster assignments in K-means by ``soft'' ones represented by the update of the posterior distribution. We repeat its definition here for completeness (where $\th = (c_1, \ldots, c_p)$).
\begin{algorithm}[Soft K-means]
\label{alg:soft.kmeans}
\begin{enumerate}[label=\arabic*., wide=0.5cm]
\item Choose a number $\sig^2>0$, a small constant $\ep$ and a maximal number of iterations $M$. Initialize the  centers  $c = (c_1, \ldots, c_p)$. 
\item At step $n$ of the algorithm, let $c$ be the current centers. 
\item Compute, for $x\in T$ and $i=1, \ldots, p$
\[
f_Z(i\mid x, \theta) = \frac{e^{- \frac1{2\sig^2} 
|x-c_i|^2}}{\sum_{j=1}^p e^{-\frac1{2\sig^2}
|x-c_j|^2}}
\]
and let $\zeta_i = \sum_{k=1}^N f_Z(i|x, \th)$, $i=1, \ldots, p$.
\item For $i=1, \ldots, p$, let 
\[
c'_i = \frac{1}{\zeta_i}\sum_{x\in T} x f_Z(i|x, \th).
\]
\item If $|c' - c| <\ep$ or $n=M$: stop the algorithm.
\item Replace $c$ by $c'$ and $n$ by $n+1$ and return to step 2. 
\end{enumerate}
\end{algorithm}

When $\sig^2\to 0$, $f_Z(\ccdot|x_k, \theta)$ converges to the uniform probability on indexes $j$ such that $c_j$ is closest to $x_k$, which is a Dirac measure unless there are ties. Class allocation and center
updating become then asymptotically identical to the K-means algorithm. 
A variant of soft K-means, called deterministic annealing \cite{rose1990deterministic}, applies \cref{alg:soft.kmeans} while letting $\sig$ slowly tend to 0. This new algorithm is experimentally more robust than K-means, in that it is less likely to be trapped in bad local minimums.

\begin{remark}
\label{rem:soft.kmeans}
The soft K-means algorithm can also be defined directly as an alternate minimization method for the objective function
\[
F(c, f_Z) = \frac12 \sum_{x\in T} \sum_{j=1}^p f_Z(j|x) |x- c_j|^2 + \sig^2 \sum_{x\in T}
\sum_{j=1}^p f_Z(j|x)\log f_Z(j|x),
\]
with the constraints $f_Z(j|x) \geq 0$ for all $j$ and $x$ and $\sum_{j=1}^p f_Z(j|x) = 1$. One can check (we leave this as an exercise) that Step 3 in \cref{alg:soft.kmeans} provides the optimal $f_Z$ for $F$ when $c$ is fixed, and that Step 4 gives the optimal $c$ when $f_Z$ is fixed (see \cref{prob:soft.kmeans}).
\end{remark}

\begin{remark}
\label{rem:assign.new}
We note that, if a K-means, soft K-means or MoG algorithm has been trained on a training set $T$, it is then easy to assign a new sample $\tilde x$ to one of the clusters. Indeed, for K-means, it suffices to determine the center closest to $\tilde x$, and for the other methods to maximize $f_Z(j|\tilde x, \theta)$, which is computable given the model parameters. In contrast, there was no direct way to do so using hierarchical clustering. 
\end{remark}

\subsection{Kernel (soft) K-means}

We now consider the soft K-means algorithm in feature space, and introduce
features $h_k = h(x_k)$ in an inner product space $H$ such that $\scp{h_k}{h_l}_H = K(x_k, x_l)$ for
some positive definite kernel. As usual, the underlying assumption is that the computation of $h(x)$ does not need to be feasible, while evaluations of $K(x,y)$ are easy. Let us consider the minimization of
$$
\frac12 \sum_{x\in T} \sum_{j=1}^p f_Z(j\mid x) \|h(x)- c_j\|_H^2 + \sig^2 \sum_{x\in T}
\sum_{j=1}^p f_Z(j|x)\log f_Z(j|x)
$$
for some $\sig^2>0$ (kernel K-means corresponds to taking the limit $\sig^2\to 0$).
Given $f_Z$, the optimal centers are 
\[
c_j=\frac{1} \zeta_j \sum_{x\in T} f_Z(j\mid x) h(x)
\] 
with $\zeta = \sum_{x\in T} f_Z(j\mid x)$.
They belong to the feature space, $H$, and are therefore  not computable in general. However,
the distance between them and a point $h(y)\in H$ is explicit and
given by
\[
\|h(y) - c_j\|_H^2 =  K(y,y) - \frac{2}{\zeta_j}
\sum_{x\in T} f_Z(j|x) K(y,x) + \frac{1}{\zeta_j^2} \sum_{x,x'\in T}
f_Z(j|x)f_Z(j|x') K(x, x').
\]
 The class probabilities at each iteration can therefore be updated using
$$
f_Z(j|x) = \frac{e^{- \lfrac{\|h(x) - c_j\|_H^2}{2\sig^2}}}{\sum_{j'=1}^p e^{- \lfrac{\|h(y) - c_{j'}\|_H^2}{2\sig^2}}}\,.
$$
This yields the soft kernel K-means algorithm, that we repeat below.
\begin{algorithm}[Kernel soft K-means]
\label{alg:clust.mog.2}
Let $T\subset \mR^d$ be the training set.
Initialize the algorithm with some choice for $f_Z(j|x)$, $j=1, \ldots, p$, $x\in T$ (for example: $f_Z(j|x) = 1/p$ for all $j$ and $x$).
\begin{enumerate}[leftmargin=1cm]
\item For $j=1, \ldots, p$ and $x\in T$ compute
\[
\|h(x) - c_j\|_H^2= K(x,x) - \frac{2}{\zeta_j}
\sum_{x'\in T} f_Z(j|x') K(x,x') + \frac{1}{\ze_j^2} \sum_{x',x''\in T}
f_Z(j|x')f_Z(j|x'') K(x', x'')
\]
with $\ze_j = \sum_{x'\in T} f_Z(j|x')$.
\item Compute, for $x\in T$ and $j=1, \ldots, p$,
\[
f_Z(j|x) = \frac{e^{- \|h(x) - c_j\|_H^2/2\sig^2}}{\sum_{j'=1}^p e^{- \|h(y) - c_{j'}\|_H^2/2\sig^2}}\,.
\]
\item If the variation of $f_Z$ compared to the previous iteration is small, or if a maximum number of iterations has been reached, exit the algorithm.
\item Return to step 1.
\end{enumerate}
After convergence, the clusters are computed by assigning $x$ to $A_i$ when $i = \argmax\{f_Z(j|x): j=1, \ldots, p\}$, making an arbitrary decision in case of a tie.
\end{algorithm}
For ``hard'' K-means (with $\sig^2\to 0$), step 2 simply updates $f_Z(j|x)$ as the uniform probability on the set of indexes $j$ at which $\|h(x) - c_j\|_H^2$ is minimal.  

\subsection{Convex relaxation}
\label{sec:kmeans.relax}
We return to the initial formulation of K-means for Euclidean data, as a minimization, over all partitions  $\mathcal A = \{A_1, \ldots, A_K\}$ of $\{1, \ldots, N\}$ of 
\[
W(\mathcal A) = \sum_{j=1}^K \sum_{k\in A_j} |x_k - c_j|^2
\]
where $c_j$ is the average of the points $x_j$ such that $j\in A_j$. 
We start with a simple transformation expressing this function in terms of the matrix $S_\alpha$ of square distances $\alpha(x_k,x_l) = |x_k - x_l|^2$. Indeed, we have 
\begin{align*}
\sum_{k\in A_j} |x_k - c_j|^2 &= \sum_{k\in A_j} |x_k|^2 - \frac{1}{|A_j|}\left|\sum_{k\in A_j} x_k \right|^2\\
&= \sum_{k\in A_j} |x_k|^2 - \frac{1}{|A_j|}\sum_{k,l\in A_j} x_k^T x_l \\ 
&= \frac1{2|A_j|} \sum_{k,l\in A_j} (|x_k|^2 + |x_l|^2 - 2 x_k^T x_l) \\ 
&= \frac1{2|A_j|} \sum_{k,l\in A_j} |x_k - x_l|^2  
\end{align*}
Introduce the vector $u_j\in \mR^N$ with coordinates $\pe{u_j}k = 1/\sqrt{|A_j|}$ for $k\in A_j$ and 0 otherwise. Then 
\begin{equation}
\label{eq:dispersion}
\frac1{2|A_j|} \sum_{k,l\in A_j} |x_k - x_l|^2 = \frac12 u_j^T S_\alpha u_J =  \frac{1}{2} \trace(S_\alpha u_j u_j^T)\,.
\end{equation}
Let
\[
Z(\mathcal A) = \sum_{j=1}^p u_ju_j^T,
\]
so that $Z(\CA)$ has entries $Z^{(k,l)}(\CA) = 1/|A_j|$ for $k,l\in A_j$, $j= 1, \ldots p$ and 0  for all other $k,l$. Summing \cref{eq:dispersion} over $j$,
we get
\[
W(\CA) = \frac{1}{2} \trace(S_\alpha Z(\CA)).
\]

The matrix $Z(\CA)$ is symmetric, has non-negative entries. It moreover satisfies $Z(\CA) \dsone_N = \dsone_N$ and $Z(\CA)^2 = Z(\CA)$. 
Interestingly, these properties characterize matrices $Z$ associated with partitions, as stated in the next proposition \citep{peng2005new,peng2007approximating}.
\begin{proposition}
\label{prop:mat.partition}
Let $Z\in \CM_N(\mR)$ be a symmetric matrix with non-negative entries satisfying $Z\dsone_N = \dsone_N$ and $Z^2 = Z$. Then there exists a partition $\CA$ of $\{1, \ldots, N\}$ such that $Z = Z(\CA)$.
\end{proposition}
\begin{proof}
Note that $Z$ being symmetric and satisfying $Z^2 = Z$ imply that it is an orthogonal projection with eigenvalues 0 and 1. In particular $Z$ is positive semidefinite. This implies that, for all $i,j\in \{1, \ldots, N\}$, one has
\[
Z(i,j)^2 \leq Z(i,i), Z(j,j).
\]
This inequality combined with $\sum_{j=1}^N Z(k, j) = 1$ (expressing $Z\dsone_N = \dsone_N$) shows that all diagonal entries of $Z$ are positive.

Define on $\{1,\ldots, N\}$ the relation $k\sim j$ if and only if $Z(j,k) >0$. The relation is symmetric and we just checked that $k\sim k$ for all $k$. It is also transitive, from the relation (deriving from $Z^2=Z$)
\[
Z(k,j) = \sum_{i=1}^N Z(k,i) Z(i,j)
\]
which shows (since all terms in the sum are non-negative) that $k\sim i$ and $j\sim i$ imply $k\sim j$.

Let $\CA = \{A_1, \ldots, A_q\}$ be the partition of $\{1, \ldots, N\}$ formed by the equivalence classes for this relation. We now show that $Z = Z(\CA)$.

We have, for all $k,j\in \{1, \ldots, N\}$
\begin{align*}
\sum_{i=1}^N Z(k, i) (Z(k, j) - Z(i, j)) & =Z(k,j)  \sum_{i=1}^N Z(k, i) - \sum_{i=1}^N Z(k, i) Z(i, j)\\
\nonumber
&= Z(k,j) - \sum_{i=1}^N Z(k, i) Z(i, j) = 0
\end{align*}
Now, if $k,j \in A_s$ for some $s$, the identity reduces to
\begin{equation}
\label{eq:Z.identity}
\sum_{i\in A_s} Z(k, i) (Z(k, j) - Z(i, j)) = 0.
\end{equation}
Choose $k$ such that $Z(k,k) = \max\{Z(i,i): i \in A_s\}$. Then, for all $i,j\in A_s$,
$Z(i,j) \leq \sqrt{Z(i,i)Z(j,j)} \leq Z(k,k)$ and \cref{eq:Z.identity} for $j=k$ yields
\[
\sum_{i\in A_s} Z(k, i) (Z(k, k) - Z(k,i)) = 0,
\]
which is only possible (since all $Z(k,i)$ are positive) if $Z(k,i) = Z(k,k)$ for all $i\in A_s$. From $Z(k,i) \leq \sqrt{Z(i,i)Z(k,k)}$, we get $Z(i,i) = Z(k,k)$ for all $i$, and therefore (reapplying what we just found to $i$ instead of $k$) $Z(i,j) = Z(i,i) = Z(k,k)$ for all $i,j\in A_s$. Finally, we have
\[
1 = \sum_{i\in A_s} Z(k,i) = |A_s| Z(k,k)
\]
showing that $Z(k,k) = 1/|A_s|$ and completing the proof that $Z = Z(\CA)$.
\end{proof}

Note that the number of clusters, $|\CA|$, is equal to the trace of $Z(\CA)$.
This shows that minimizing $W(\CA)$ over partitions with $p$ clusters is equivalent to the constrained optimization problem minimizing
\begin{equation}
\label{eq:kmeans.matrix}
G(Z) = \trace(S_\alpha Z)
\end{equation}
over all matrices $Z$ such that $Z\geq 0$, $Z^T = Z$, $Z \dsone_N = \dsone_N$, $\trace(Z) = p$ and $Z^2 = Z$. This is still a difficult problem, since it is equivalent to K-means, which is NP hard. Seeing the problem in this form, however, is more amenable to approximations and, in particular, convex relaxations.

In \citep{peng2007approximating}, it is proposed to use a semidefinite program (SDP) as a relaxation. The conditions $Z = Z^T$ and $Z^2=Z$ require that all eigenvalues of $Z$ are either 0 or 1, and a direct relaxation is to replace these constraints by $Z^T = Z$, $0 \preceq Z$, and $Z \preceq \Id[N]$.  The last inequality is however redundant if we add the conditions
\footnote{Recall that $Z\succeq 0$ means that $Z$ is positive definite, while $Z\geq 0$ indicates that all its entries are non-negative.}
 $Z \geq 0$ and $Z\dsone = \dsone$. This is a consequence of the Perron-Frobenius theorem which states that a matrix $\tilde Z$ with positive entries has a largest (in modulus) real eigenvalue, which has multiplicity one and is associated with an eigenvector with positive coordinates, the latter eigenvector being (up to multiplication by a constant) the unique eigenvector of $\tilde Z$ with positive coordinates. So, if a matrix $\tilde Z$ is symmetric, satisfies $\tilde Z > 0$ and $\tilde Z \dsone_N = \dsone_N$, then $\tilde Z \preceq \Id[N]$. Applying this result to $\tilde Z = (1-\epsilon) Z + (\epsilon/N) \dsone_N\dsone_N^T$ and letting $\epsilon$ tend to 0 shows that any matrix $Z$ with non-negative entries satisfying $Z\dsone_N = \dsone_N$ also satisfies $Z\preceq \Id[N]$.
 
This provides the following SDP relaxation of K-means \citep{peng2007approximating}: minimize  
\begin{equation}
\label{eq:kmeans.matrix.2}
G(Z) = \trace(S_\alpha Z)
\end{equation}
subject to $Z^T = Z$, $Z \dsone_N = \dsone_N$, $\trace(Z) = p$, $Z\geq 0$, $Z\succeq 0$.

Clusters can be immediately inferred from the columns of the matrix $Z(\CA)$, since they are identical for two indices in the same cluster, and orthogonal to each other for two indices in different clusters. Let $z_1(\CA), \ldots, z_N(\CA)$ denote the columns of $Z(\CA)$ and  $\bar z_k(\CA) = z_k(\CA)/|z_k(\CA)|$. One has $|\bar z_k(\CA) - \bar z_l(\CA)|= 0$ if $k$ and $l$ belong to the same cluster and $\sqrt 2$ otherwise. 

These properties will not necessarily be satisfied by a
solution, say, $Z^*$, of the SDP relaxation, but, assuming that the approximation is good enough, one may still consider the  normalized columns of $Z^*$ and expect them to be similar for indices in the same cluster,  and away from each other otherwise. Denoting by $\bar z^*_1, \ldots, \bar z^*_N$ these normalized columns, one can then run on them the standard K-means algorithm, or a spectral clustering method such as those described in the next sections, to infer clusters.

\begin{remark}
\label{rem:sdp.kmeans}
Clearly, one can use any  symmetric matrix $S$ in the definition of $G$ in \cref{eq:kmeans.matrix,eq:kmeans.matrix.2}. The method is equivalent to, or to a relaxation of, K-means only when $S$ is formed with squared norms in inner-product spaces, which does include  kernel K-means, for which
\[
\alpha(x_k, x_l) = K(x_k,x_k) - 2 K(x_k,x_l) + K(x_l, x_l).
\]
If $\alpha$ is an arbitrary discrepancy measure, the minimization of $G(Z)$ still makes sense, since it is equivalent to minimizing
\[
G(Z(\CA)) = \sum_{j=1}^p D_\al(A_j).
\]
where
\begin{equation}
\label{eq:vv.alpha}
D_\al(A) = \frac1{|A|} \sum_{x,y\in A} \al(x, y)\,.
\end{equation}
is a (normalized) measure of size, that we will call the {\em $\al$-dispersion} of a finite set $A$.
\end{remark}

\begin{remark}
Instead of using dissimilarities, some algorithms are more naturally defined in terms of similarities. Given such a similarity measure, say, $\beta$, one must maximize rather than minimize the index $\Delta_\beta$ (which becomes, rather than a measure of dispersion, a measure of concentration).

One passes from a dissimilarity $\alpha$ to a similarity $\beta$  by applying a decreasing function to the former, a common choice being
\[
 \beta(x,x') = \exp(-\alpha(x,x')/\tau)
 \]
for some $\tau>0$.

Alternatively, one can fix an element $x_0\in \CR$ and  let 
\[
\be(x,y) = \al(x,x_0) + \al(y,x_0) - \al(x,y) - \al(x_0,x_0),
\]
(note that the last term, $\al(x_0,x_0)$ is generally equal to 0).  For example, if $\al(x,y) = |x-y|^2$, then $\be(x,y) = 2(x-x_0)^T(y-x_0)$ (for which it is natural to take $x_0=0$). If $\al$ is a distance (not squared!), then $\be \geq 0$ by the triangular inequality. In this case, we have 
\begin{align*}
\De_\be(A_1, \ldots, A_p) &=
\sum_{k=1}^n D_\be(A_k) \\
&= \sum_{k=1}^p \frac{1}{|A_p|} \sum_{x,y\in A_k} \al(x,x_0) + \sum_{k=1}^p \frac{1}{|A_p|} \sum_{x,y\in A_k} \al(y,x_0)\\
&- \sum_{k=1}^p \frac{1}{|A_p|} \sum_{x,y\in A_k} \al(x_0,x_0) - \sum_{k=1}^p \frac{1}{|A_p|} \sum_{x,y\in A_k} \al(x,x_0)\\
&= 2 \sum_{k=1}^p \sum_{x\in A_k} \al(x,x_0) - \sum_{k=1}^p |A_k| \al(x_0,x_0) - \De_\al(A_1, \ldots, A_p) \\
&= 2 \sum_{x\in T} \al(x,x_0) - |T| \al(x_0,x_0) - \De_\al(A_1, \ldots, A_p) 
\end{align*}
so that minimizing $\De_\al$ is equivalent to maximizing $\De_\be$. 

\end{remark}

%

\section{Spectral clustering}
\label{sec:min.disp}
 
\subsection{Spectral approximation of minimum discrepancy}
\label{sec:spectral.clus}

One refers to spectral methods algorithms that rely on computing eigenvectors and eigenvalues (the spectrum) of data-dependent matrices. In the case of minimizing discrepancies, they can be obtained by further simplifying \cref{eq:kmeans.matrix.2}, essentially by removing some constraints.

One indeed gets a simpler problem if the non-negativity constraint, $Z\geq 0$, is removed. Doing so, one cannot guarantee anymore that $Z\preceq \Id[N]$, so we need to reinstate this constraint. We will first make the further simplification to remove the constraint $Z\dsone_N = \dsone_N$, the problem becoming minimizing $\trace(S_\alpha Z)$ over all $Z\in \CS_N^+(\mR)$ such that $0\preceq Z \preceq \Id[N]$ and $\trace(Z) = p$. Decomposing $Z$ in an eigenbasis, i.e., looking for it in the form
\[
Z = \sum_{j=1}^N \xi_j e_je_j^T, 
\]
this is equivalent to minimizing
\begin{equation}
\sum_{j=1}^N \xi_j e_j^TS_\alpha e_j
\label{eq:spectral.1}
\end{equation}
subject to $0\leq \xi_j \leq 1$, $\sum_{j=1}^N \xi_j = p$ and $e_1, \ldots, e_N$ orthonormal basis of $\mR^N$. First consider  minimization with respect to the basis, fixing $\xi$. 
There is obviously no loss of generality in requiring that $\xi_1 \leq \xi_2 \leq \cdots \leq \xi_N$, and using \cref{cor:pca.base} (modified to minimizing \cref{eq:spectral.1} rather than maximizing it) we know that an optimal basis is given by the eigenvectors of $S_\alpha$, ordered with non-decreasing eigenvalues. Letting $\lambda_1 \leq \cdots \leq \lambda_N$ denote these eigenvalues, we find that $\xi_1, \ldots, \xi_N$ must be a non-decreasing sequence minimizing
\[
\sum_{j=1}^N \lambda_j \xi_j
\]
subject to $0\leq \xi_k \leq 1$ and $\sum_{j=1}^N \xi_j = p$. The optimal solution is obtained by taking $\xi_1 = \cdots = \xi_p = 1$, since, for any other solution
\begin{align*}
\label{eq:spectral.2}
\sum_{j=1}^N \lambda_j \xi_j - \sum_{j=1}^p \lambda_j &\geq 
\lambda_{p+1} \sum_{j=p+1}^N \xi_j + \sum_{j=1}^p \lambda_j(\xi_j - 1)\\
&=  \lambda_{p+1} \sum_{j=1}^p (1-\xi_j) + \sum_{j=1}^p \lambda_j(\xi_k - 1)\\
&= \sum_{j=1}^p (\lambda_{p+1} - \lambda_j)(1-\xi_j) 
\\ &\geq 0.
\end{align*}

The following algorithm (similar to that discussed in \citep{drineas2004clustering}) summarizes this discussion.
\begin{algorithm}[Spectral clustering: version 1]
\label{alg:sc.v1}
Let $S_\alpha$ be an $N\times N$ discrepancy matrix. Let $p$ denote the number of clusters.
\begin{enumerate}
\item Compute the eigenvectors of $S_\alpha$ associated with the $p$ smallest eigenvalues.
\item Denoting these eigenvectors by $e_1, \ldots, e_p$, define $y_1, \ldots, y_N \in \mR^p$ by $\pe{y_k}j = \pe{e_j}k$.
\item Run K-means on $(y_1, \ldots, y_N)$ to determine a partition.
\end{enumerate}
\end{algorithm}

This algorithm needs to be slightly modified if one also wants $Z$ to satisfy $Z\dsone_N = \dsone_N$. In that case, $\dsone_N$ is one of the eigenvectors (with eigenvalue 1), and the others are orthogonal to it. As a consequence, one now looks for $Z$ in the form
\[
Z = \sum_{k=1}^{N-1} \xi_j e_je_j^T + \frac1N \dsone_N\dsone_N^T
\]
leading to the minimization of
\[
\sum_{j=1}^{N-1} \xi_j e_j^T S_\alpha e_j  + \frac1N \dsone_N^TS_\alpha\dsone_N
\]
over all $\xi_1, \ldots, \xi_{N-1}$ such that $0\leq \xi_j \leq 1$ and $\sum_{j=1}^N \xi_j = p-1$, and over all $e_1, \ldots, e_{N-1}$ such that $e_1, \ldots, e_{N-1}, \dsone_N/\sqrt N$ form an orthonormal basis. The main difference with the previous problem is that we now need to ensure that all $e_j$'s are perpendicular to $\dsone$.

To achieve this, introduce the projection matrix $P = \Id[N] - \dsone_N\dsone_N^T/N$ and let $\tilde S_\alpha = PS_\alpha P$. Then, since $u^T\dsone = 0$ implies $u^T\tilde S_\alpha u = u^TS_\alpha u$, it is equivalent to minimize
\[
\sum_{j=1}^{N-1} \xi_j e_j^T \tilde S_\alpha e_j 
\]
over all $\xi_1, \ldots, \xi_{N-1}$ such that $0\leq \xi_j \leq 1$ and $\sum_{j=1}^N \xi_j = p-1$, and over all $e_1, \ldots, e_{N-1}$ such that $e_1, \ldots, e_{N-1}, \dsone/\sqrt N$ form an orthonormal basis. Because $\tilde S_\alpha \dsone = 0$, we know that $\tilde S_\alpha$ can be diagonalized in an orthonormal basis $(e_1, \ldots, e_{N-1}, \dsone_N/\sqrt N)$, and we obtain an optimal solution by selecting the $p-1$ vectors associated with smallest eigenvalues, with associated $\xi_j = 1$. We therefore get a modified version of the spectral clustering algorithm.
\begin{algorithm}[Spectral clustering: version 2]
\label{alg:sc.v2}
Let $S_\alpha$ be an $N\times N$ discrepancy matrix. Let $p$ denote the number of clusters. Let $P = \Id[N] - \dsone_N\dsone_N^T/N$.
\begin{enumerate}
\item Compute $\tilde S_\al = PS_\alpha P$
\item Compute the eigenvectors of $\tilde S_\alpha$ associated with the $p-1$ smallest eigenvalues.
\item Denoting these eigenvectors by $e_1, \ldots, e_{p-1}$, define $y_1, \ldots, y_N \in \mR^{p-1}$ by $\pe{y_k}j = \pe{e_j}k$.
\item Run K-means on $(y_1, \ldots, y_N)$ to determine a partition.
\end{enumerate}
\end{algorithm}

\section{Graph partitioning}
\label{sec:graph.cluster}

Similarity measures are often associated with graph structures, with a goal of finding a partition of their set of vertices. So, let $T$ denote the set of these vertices and assume that to all pairs $x,y\in T$, one attribute  a weight given by  $\be(x,y)$, where $\be$ is assumed to be non-negative. We define $\be$ for all $x,y\in T$, but we interpret $\be(x,y) = 0$ as marking the absence of an edge between $x$ and $y$.

  Let $V$ denote the vector space of all functions $f: T\to \mR$ (we have $\dim(V) = |T|$). This space can be equipped with the standard Euclidean norm, that we will call in this section the $L^2$ norm (by analogy with general spaces of square integrable functions), letting,
\[
|f|_2^2 = \sum_{x\in T} f(x)^2.
\]
One can also associate a measure of smoothness for a function $f\in V$ by computing the discrete ``$H^1$'' semi-norm,
\[
|f|^2_{H^1} = \sum_{x,y\in T} \be(x,y) (f(x)-f(y))^2.
\]
With this definition, ``smooth functions'' tend to have similar values at points $x,y$ in $T$ such that $\be(x,y)$ is large while there is  less constraint when $\be(x,y)$ is small. In particular, $|f|_{H^1} = 0$ if and only if $f$ is constant on connected components of the graph.\footnote{Two nodes $x$ and $y$ are connected in the graph if there is a sequence $z_0, \ldots, z_n$ in $T$ such that $z_0=x$, $z_n = y$ and $\be(z_i, z_{i-1}) > 0$ for $i=1, \ldots, n$. This provides an equivalence relation and equivalent classes are called connected components. }

The notion of connected components, combined with thresholding, can be used to build a hierarchical family of partitions of the graph. Define, for all $t>0$, the thresholded weights $\be^{(t)}(x,y) = \max(\be(x,y) - t, 0)$. The set of connected components associated with the pair $(V, \be^{(t)})$ forms a partition, say, $\CA^{(t)}$, of $T$. The resulting set of partitions is {\em nested} in the sense that, if $s < t$, the sets forming the partition $\CA^{(s)}$ are unions of sets forming $\CA^{(t)}$.
 This thresholding procedure is not always satisfactory, however, because there does not always exist a fixed value of $t$ that produces a good quality cluster decomposition.

If there exist $p$ connected components, then the subspace of all functions $f\in V$ such that $|f|_{H_1} = 0$ has dimension $p$. If $C_1, \ldots, C_p$ are the connected components, this space is generated by the functions $\de_{C_k}$, $k=1, \ldots, p$, with $\de_{C_k}(x) = 1$ if $x\in C_k$ and 0 otherwise. These functions form, in addition, an orthogonal system for the Euclidean inner product: $\scp{\de_{C_k}}{\de_{C_l}}_2 = 0$ if $k\neq l$. 

One can write $\frac12 |f|^2_{H^1} = f^T L f$ where $L$, called the {\em Laplacian operator} associated to the considered graph, is defined by
\[
Lf(x) = \sum_{y\in T} L(x,y) f(y)
\]
and 
\begin{equation}
\label{eq:laplacian.graph}
L(x,y) =\left( \sum_{z\in T} \be(x,z)\right) \bfone_{x=y} - \be(x,y).
\end{equation}
The vectors $\de_{C_k}$, $k=1, \ldots, p$ are then an orthogonal basis of the null space of $L$. Conversely, let $(e_1, \ldots, e_p)$ be any  basis of this null space. Then, there exists an invertible matrix $A = (a_{ij}, i,j=1, \ldots, p)$ such that
\[
e_i(x) = \sum_{j=1}^p a_{ij} \de_{C_j}(x).
\]
Associate to each $x\in T$ the vector $e(x) = \begin{pmatrix} e_1(x)\\ \vdots\\ e_p(x)\end{pmatrix}\in\mR^p$. Then, for any $x,y\in T$, we have $e(x) = e(y)$ if and only if $\de_{C_j}(x) = \de_{C_j}(y)$ for all $j=1, \ldots, p$ (because $A$ is invertible), that it, if and only if $x$ and $y$ belong to the same connected component. So, given any basis of the null space of $L$,  the function $x\mapsto e(x)$ determines these connected components.
 So, a---not very efficient---way of determining the connected components of the graph can be to diagonalize the operator $L$ (written as an $N$ by $N$ matrix, where $N=|T|$), extract the $p$ eigenvectors $e_1, \ldots, e_p$ associated with eigenvalue zero and deduce from the function $e(x)$ above the set of connected components.
 
\begin{figure}
\includegraphics[width=0.3\textwidth]{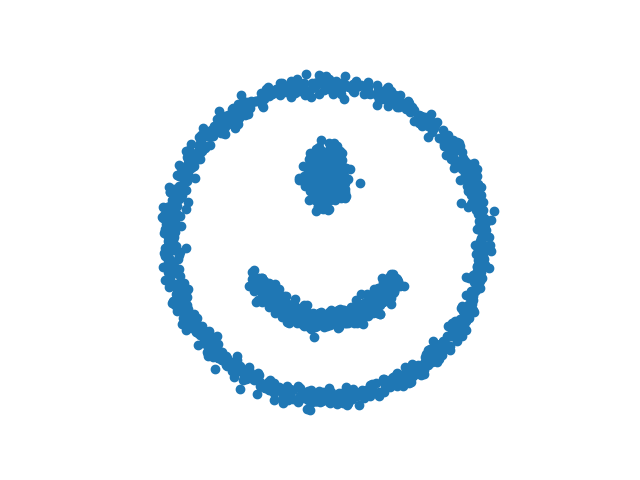}
\includegraphics[width=0.3\textwidth]{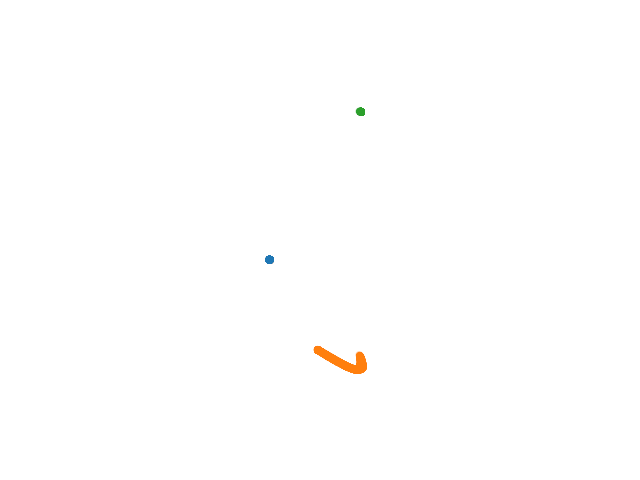}
\includegraphics[width=0.3\textwidth]{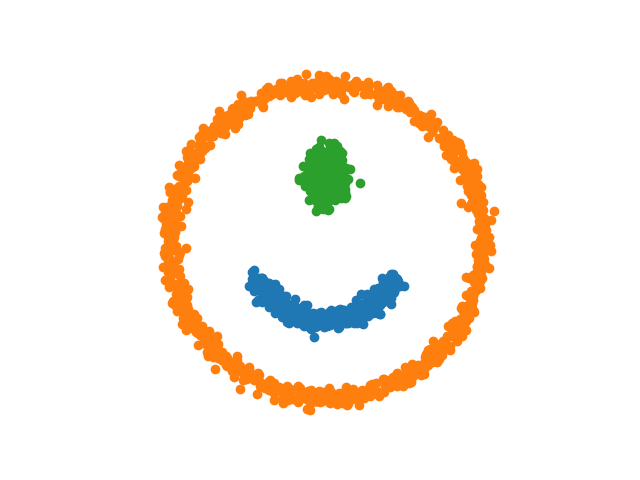}
\caption{\label{fig:graph.laplacian} Example of data transformed using the eigenvectors of the graph Laplacian. Left: Original data. Center: Result of a Kmeans algorithm with three clusters applied to the transformed data (2D projection). Right: Visualization of the cluster labels on the original data.  }
\end{figure}

Now, in practice, the graph associated to $T$ and $\be$ will not separate nicely into connected components in order to cluster the training set. Most of the time, because of noise or some weak connections, there will be only one such component, or in any case much less than what one would expect when clustering the data. The previous discussion suggests, however, that in the presence of moderate noise in the connection weights, one may expect that the eigenvectors associated to the $p$ smallest eigenvalues of $L$ provide vectors $e(x), x\in T$ such that $e(x)$ and $e(y)$ have similar values if $x$ and $y$ belong to the same cluster (see \ref{fig:graph.laplacian}). In such cases, these clusters should be easy to determine using, say,  K-means on the transformed dataset $\tilde T = (e(x), x\in T)$. This is summarized in the following algorithm.

\begin{algorithm}[Spectral Graph Partitioning]
\label{alg:graph.part}
Let $T\subset \CR$ be the training set and $(x,y) \mapsto \be(x,y)$ a similarity measure defined on $T\times T$. Let $p$ be the desired number of clusters.
\begin{enumerate}[leftmargin=1cm]
\item Form the Laplacian operator described in \cref{eq:laplacian.graph} and let $e_1, \ldots, e_p$ be its eigenvectors associated to the $p$ lowest eigenvalues. For $x\in T$, let $e(x)\in \mR^p$ be given by
\[
e(x) = (e_1(x), \ldots, e_p(x))^T\in \mR^p.
\]
\item Apply the K-means algorithm (or one of its variants) with $p$ clusters to $\tilde T = (e(x), x\in T)$.
\end{enumerate}
\end{algorithm}

\section{Deciding the number of clusters}

\subsection{Detecting elbows}
The number, $p$, of subsets with respect to which the population should be partitioned is rarely known a priori, and several methods have been introduced in the literature in order to assess the ideal number of clusters. We now review some of these methods, and denote, for this purpose, by
$\CL^*(p)$ the minimized cost function obtained with $p$ clusters, e.g., using \cref{eq:kmed.2},
\[
\CL^*(p) = \min\{\mW_\al(A_1, \ldots, A_p, c_1, \ldots, c_p): A_1, \ldots, A_p \text{ partition of } T, c_1, \ldots, c_p\in \CR\},
\]
in the case of K-medoids (this definition is algorithm dependent).
It is clear that $\CL^*$ is a decreasing function of $p$. It is also natural to expect that $\CL^*$ should decrease significantly when $p$ is smaller than the correct number of clusters, while the variation should be more marginal when $p$ is overestimated, because the cost in putting together two sets of points that are far apart (which happens when $p$ is too small) is typically larger than the gain in splitting a homogeneous region in two.

\begin{figure}[h]
\centering
\includegraphics[width=0.4\textwidth]{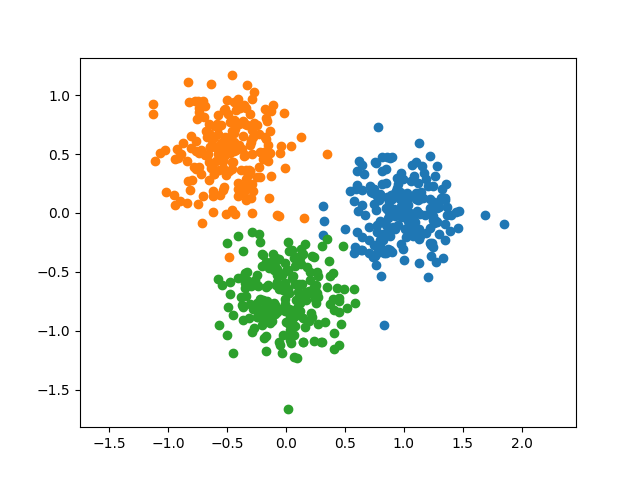}
\includegraphics[width=0.4\textwidth]{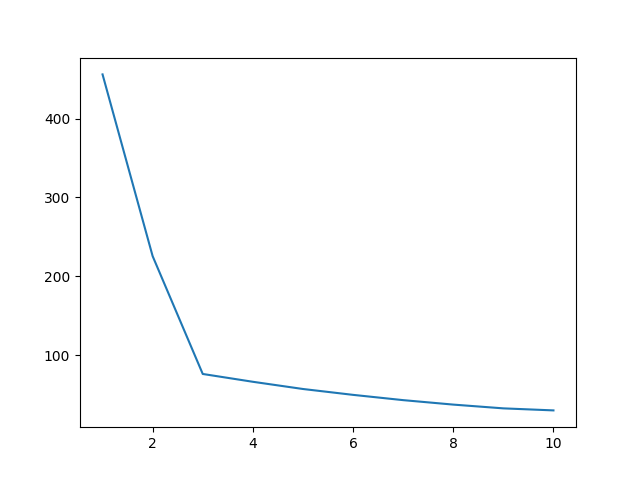}\\
\includegraphics[width=0.4\textwidth]{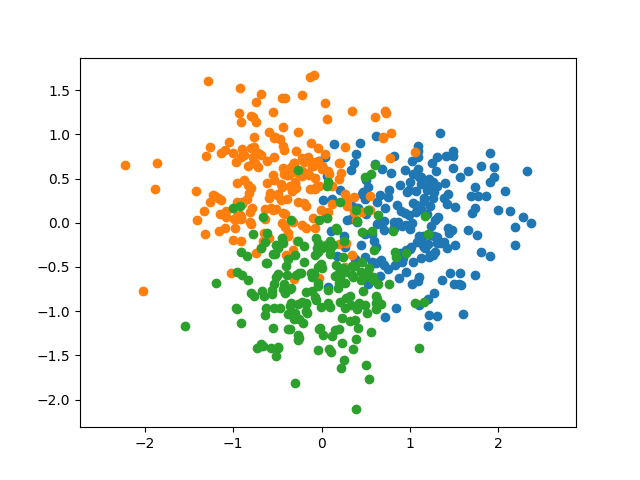}
\includegraphics[width=0.4\textwidth]{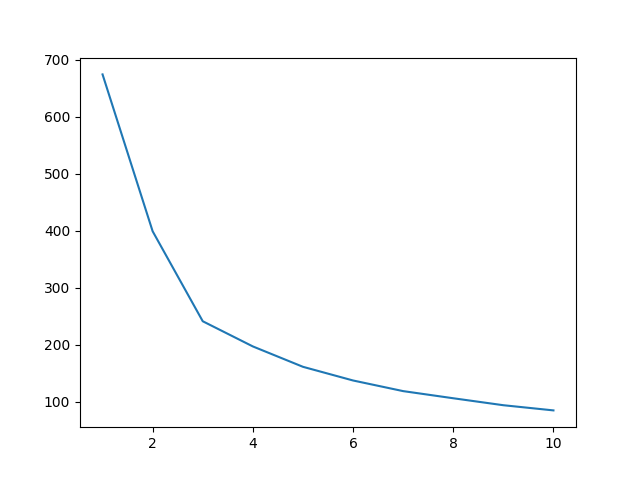}
\caption{\label{fig:clusters.elbow} Elbow graphs for K-means clustering for two populations generated as mixtures of Gaussian.}
\end{figure}

The simplest approach in this context is to visualize $\CL^*(p)$ as a function of $p$ and try to locate at which value the resulting curve makes an ``elbow,'' i.e., switches from a sharply decreasing slope to a milder one. \Cref{fig:clusters.elbow} provides an illustration of this visualization when the true number of clusters is three (the data in each cluster following a normal distribution). When the clusters are well separated, an elbow clearly appears on the graph of $\Ga_\al^*$, but this situation is harder to observe when  clusters overlap with each other.

One can measure the ``curvature'' at the elbow using the distance between each point in the graph of $(p, W_\al^*(p))$ and 
the line between its predecessor and successor. The result gives the criterion
\[
C(p) = \frac{\CL^*(p+1) + \CL^*(p-1) - 2 \CL^*(p)}{\sqrt{(\CL^*(p+1) - \CL^*(p-1))^2 + 4}},
\]
specifying the elbow point as the value of  $p$ at  which $C$ attains its maximum. For both examples in \cref{fig:clusters.elbow}, this method returns the correct number of clusters (3).

\subsection{The Cali\'nski and Harabasz index}
Several other criteria have been introduced in the literature. \citet{calinski1974dendrite}  propose to minimize the ratio of normalized between-group and within-groups sums of squares associated with K-means. For a given $p$, let $c_1, \ldots, c_p$ denote the optimal centers, and $A_1, \ldots, A_p$ the optimal partition, with $N_k = |A_k|$. The normalized between-group sum of squares is 
\[
h_\al(p) = \frac1{p-1} \sum_{k=1}^p N_k |c_k - \bx|^2
\]
and the normalized within-group sum of squares is   
\[
w_\al(p) = \frac1{N-p} \sum_{k=1}^p \sum_{x\in A_k} |x - c_k|^2
\]
\citet{calinski1974dendrite} suggest to maximize $\ga_{CH}(p) = h_\al(p)/w_\al(p)$. 

This criterion can be extended to other types of cluster analysis. We have seen in \cref{sec:min.disp} that, when $\al(x,y) = |x-y|^2$,
\[
\frac12 \sum_{k=1}^p \sum_{x,y\in A_k} \al(x,y)/N_k = \sum_{k=1}^p \sum_{x\in A_k} |x - c_k|^2.
\]
We also have
\[
\sum_{x\in T} |x-\bx|^2 = \sum_{k=1}^p \sum_{x\in A_k} |x - c_k|^2 + \sum_{k=1}^p N_k |c_k - \bx|^2
\]
and the left-hand side is also equal to 
\[
\frac1{2N} \sum_{x,y\in T} \al(x,y).
\]
It follows that, when  $\al(x,y) = |x-y|^2$,
\[
h_\al(p) =  \frac1{2(p-1)} \left(\frac1N \sum_{x,y\in T} \al(x,y) - \sum_{k=1}^p \sum_{x,y\in A_k} \al(x,y)/N_k\right)
\]
and 
\[
w_\al(p) = \frac1{2(N-p)} \sum_{k=1}^p \sum_{x,y\in A_k} \al(x,y)/N_k.
\]
These expressions can obviously be applied to any dissimilarity measure, extending $\gamma_{CH}$ to general clustering problems. 

\subsection{The ``silhouette'' index}
For $x\in T$, let 
\[
d_\al(x, A_k) = \frac1{N_k} \sum_{y\in A_k} \al(x,y).
\]
Let $a_\al(x, p) = d_\al(x, A(x))$ and $b(x, p) = \min\{d_\al(x, A_k): A_k\neq A(x)\}$. Define the  silhouette index of $x$ in the segmentation \citep{rousseeuw1987silhouettes}by
\[
s_\al(x, p) = \frac{b_\al(x,p) - a_\al(x,p)}{\max(b_\al(x,p), a_\al(x,p))} \in [-1,1].
\]
This index measures how well $x$ is classified in the partitioning. It is large when the mean distance between $x$ and other objects in its class is small compared to the minimum mean distance between $x$ and any other class. In order to estimate the best number of clusters with this criterion, one then can maximize the average index:
\[
\ga_{R}(p) = \frac1N \sum_{x\in T} s_\al(x,p).
\]
 
\begin{remark}
One can rewrite the Cali\'nski and Harabasz index using the notation introduced for the silhouette index.
Indeed, let $A(x)$ be the cluster $A_k$ to which $x$ belongs. Then 
\[
h_{\al}(p) =\frac1{2(p-1)} \sum_{x\in T} \sum_{k=1}^p \frac{N_k}{N} (d_\al(x, A_k) - d_\al(x, A(x)))
\] 
and
\[
w_\al(p) = \frac1{2(N-p)} \sum_{k=1}^p \sum_{x\in A_k} d_\al(x, A_k).
\]
\end{remark}
 
\subsection{Comparing to homogeneous data} 
Several selection methods choose $p$ based on the comparison of the data to a ``null hypothesis'' of no cluster. 
For example, assume that K-means is applied to a training set $T$ where samples are drawn uniformly according to the uniform distribution on $[0,1]^d$. Given  centers, $c_1, \ldots, c_p$, let $\bar A_k$ be the set of points in $[0,1]^d$ that are closer to $c_k$ than to any other point. Then the segmentation of $T$ is formed by the sets $A_k = \{x\in T: x\in \bar A_k\}$ and, for large enough $N$, we can approximate $|A_k|/N$ (by the Law of Large Numbers) by the volume of the set $\bar A_k$, that we will denote by $\vol(\bar A_k)$.

\begin{figure}
\centering
\includegraphics[width=0.75\textwidth]{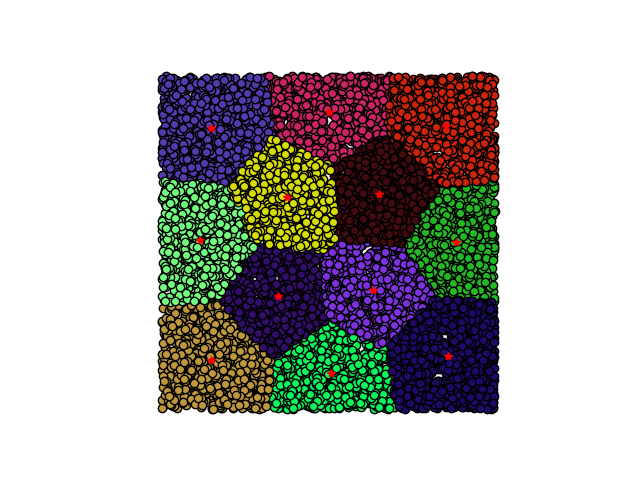}
\caption{\label{fig:unif.clust} Division of the unit square into clusters for uniformly distributed data.}
\end{figure}

Let us assume that $c_1, \ldots, c_p$ are uniformly spaced, so that the sets $\bar A_k$ have similar volumes (close to $1/p$) and have roughly spherical shapes (see \cref{fig:unif.clust}). This implies that
\[
\int_{\bar A_k} |x- c_k|^2 dx \simeq \vol(A_k) \frac{r_p^2d}{d+2}
\]
where $r_p$ is the radius of a sphere of volume $1/p$, i.e., $pr_p^d \simeq d/\Gamma_{d-1}$ where $\Gamma_{d-1}$ is the surface area of the unit sphere in $\mR^d$. So, we should have, for some constant $C$ that only depends on $d$,
\[
\sum_{x\in A_k} |x-c_k|^2 \simeq N_k \int_{\bar A_k} |x- c_k|^2 dx  \simeq C(d) (pN)  p^{-2/d-1} = C(d) N p^{-2/d}.
\]
This suggests that, for fixed $N$ and $d$, $p^{2/d} \CL^*(p)$ should vary slowly when $p$ overestimate the number of clusters (assuming that this operation divides an homogeneous cluster). Based on this analysis, \citet{krzanowski1988criterion} introduced the difference-ratio criterion, namely,
\[
\ga_{KL}(p) = \left|\frac{(p-1)^{\frac{2}{d}} \CL^*(p-1) - p^{\frac{2}{d}} \CL^*(p)}{p^{\frac{2}{d}} \CL^*(p) - (p+1)^{\frac{2}{d}} \CL^*(p+1)}\right|,
\]
and estimate the  number of clusters by taking $p$ maximizing $\ga_{KL}$.
\bigskip

Another similar approach, introduced by \citet{sugar2003finding}, is based on an analysis of mixtures of Gaussian, namely assuming an underlying model with $p_0$ groups, where data in group $k$ follow a Gaussian distribution $\CN(\mu_k, \Id)$ (possibly after standardizing the covariance matrix). In that work, the authors show that, if $d$ (the dimension) tends to infinity, with the minimal distance between centers growing proportionally to $\sqrt{d}$, then $\CL^*(p)/d$ tends to infinity when $p < p_0$. They also show that, with similar assumptions, $\CL^*(p)/d$ behaves like $p^{-2/d}$ for $p\geq p_0$, still for large dimensions. Based on this, they suggest using the criterion
\[
\ga_{SJ}(p) = \left(\frac{\CL^*(p)}{d}\right)^{-\nu} - \left(\frac{\CL^*(p-1)}{d}\right)^{-\nu}
\]
(with the convention that $\CL^*(0) = 0$) for some positive number $\nu$ and select the value of $p$ that maximizes $\ga_{SJ}$. Indeed, in the case of Gaussian mixtures, the choice $\nu = d/2$ ensures that, in large dimensions,  $\ga_{SJ}(p)$ is small for $p<p_0$, that it is close to 1 for $p>p_0$ and close to $p_0$ for $p=p_0$.

\bigskip
A more computational approach, based on Monte-Carlo simulations has been introduced in \citet{tibshirani2001estimating}, defining the {\em gap index}
\[
\ga_{TWH}(p)  =  E(\CL^*(p, \mT^\sharp)) - \CL^*(p, T)
\]
where the $\CL^*(p, T)$ denotes the optimal value of the optimized cost with $p$ clusters for a training set $T$. The notation 
$\mT^\sharp$ represent a random training set, with same size and dimension as $T$, generated using an unclustered probability distribution used as a reference.
In \citet{tibshirani2001estimating}, this distribution is taken as uniform (over the smallest hypercube containing the observed data), or uniform on the coefficients of a principal component decomposition of the data (see \cref{chap:dim.red}). The expectation $ E(\CL^*(p, \mT^\sharp))$ is computed by Monte-Carlo simulation, by sampling many realizations of the training set $\mT$, running the clustering algorithm for each of them and averaging the optimal costs.  

One can expect $\CL^*(p, T)$ (for observed data) to decrease much faster (when adding a cluster) than its expectation for homogeneous data when $p<p_0$, and the decrease of both terms to be comparable when $p\geq p_0$. 
 So the number of clusters can in principle be estimated by detecting an elbow in the graph of  $\ga_{TWH}(p)$ as a function of $p$. The procedure suggested in \citet{tibshirani2001estimating} in order to detect this elbow if to look for the first index $p$ such that
\[
\ga_{TWH}(p+1) \leq \ga_{TWH}(p) + \sig(p+1)
\]
where $\sig(p+1)$ is the standard deviation of $\CL^*(p+1, \mT^\sharp)$ for homogeneous data, also estimated via Monte-Carlo simulation. 
\bigskip

\begin{figure}
 \centering
\includegraphics[width=\textwidth]{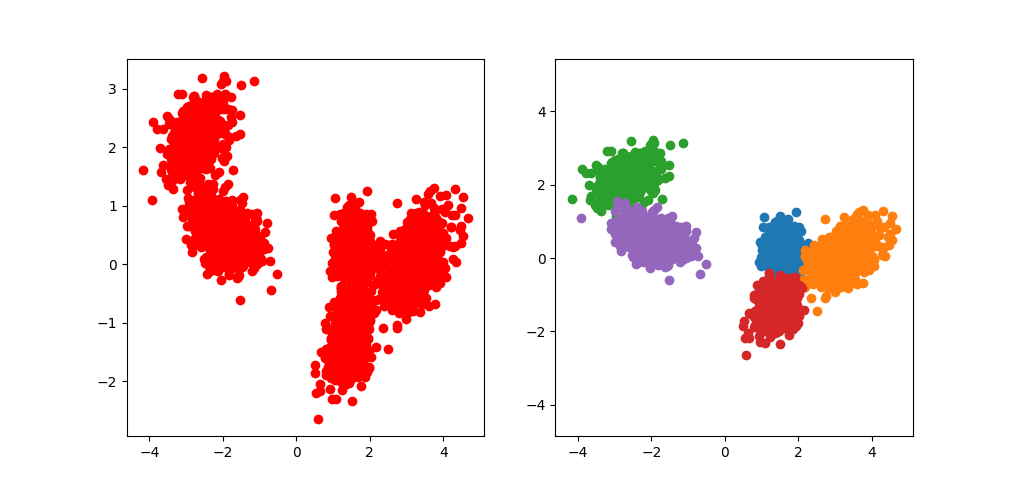}
\includegraphics[width=\textwidth]{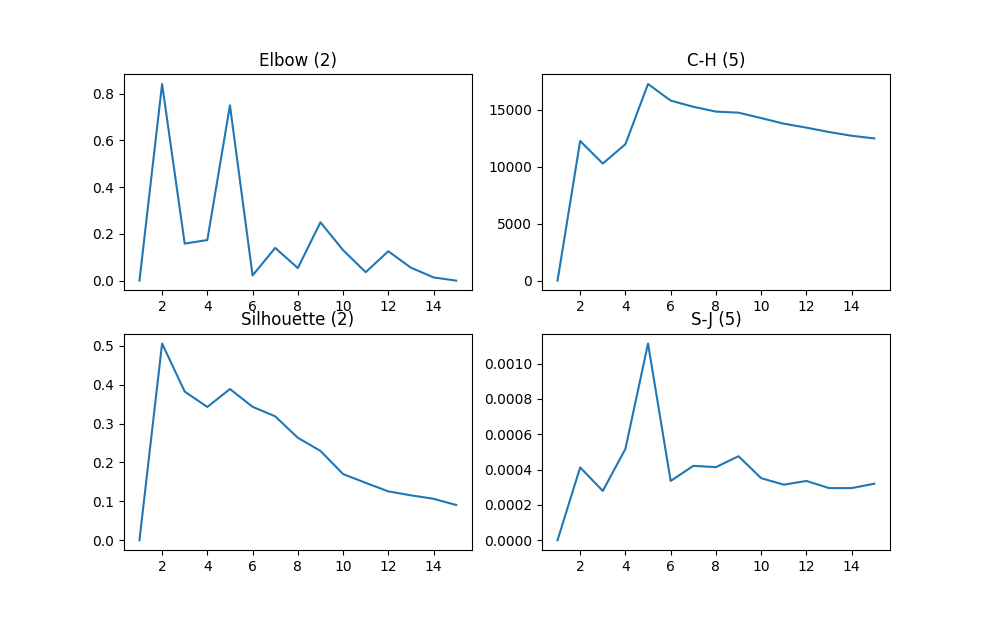}
\caption{\label{fig:clust.indx.1} Comparison of cluster indices for Gaussian clusters. First row: original data and ground truth. Second panel: plots of four indices as functions of $p$ (Elbow; Cali\'nski and Harabasz; silhouette; Sugar and James)}
\end{figure}

\begin{figure}
 \centering
\includegraphics[width=\textwidth]{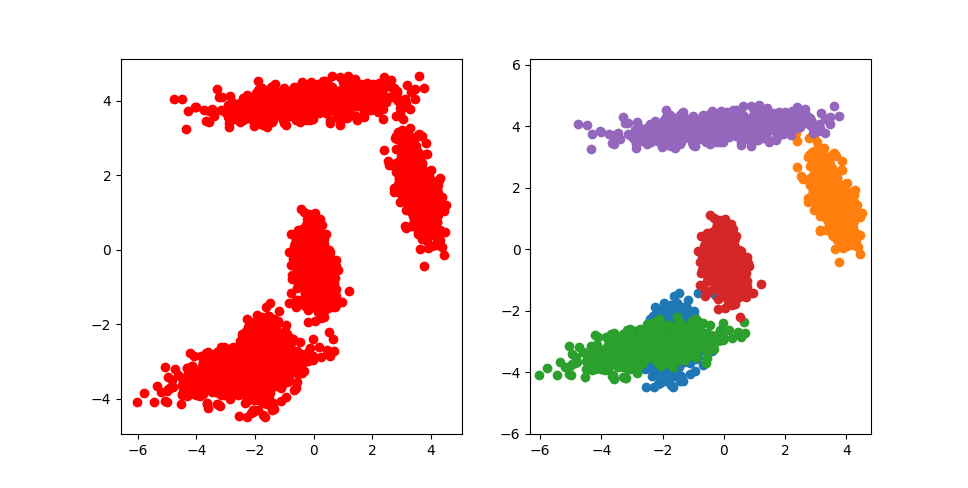}
\includegraphics[width=\textwidth]{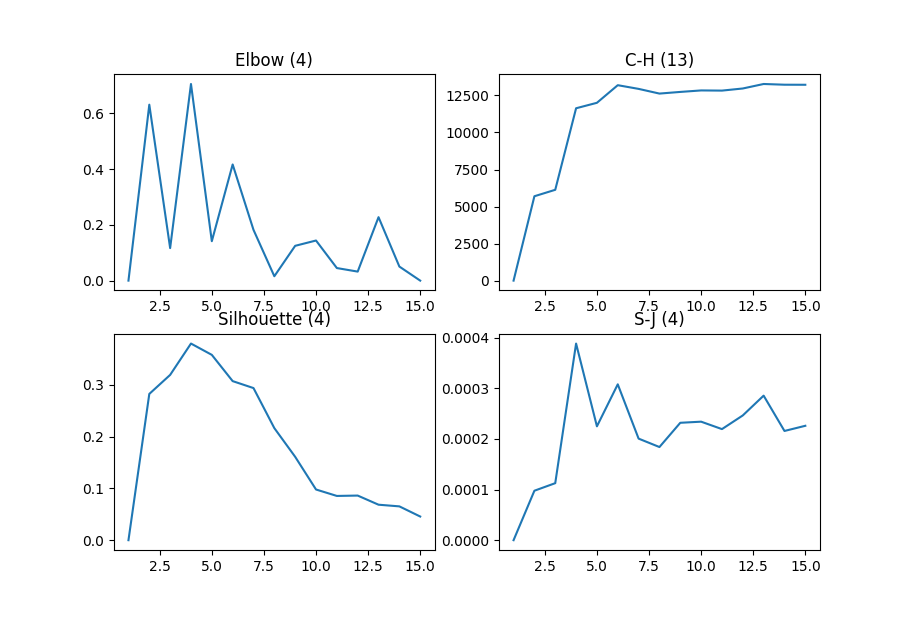}
\caption{\label{fig:clust.indx.2} Comparison of cluster indices for Gaussian clusters. First row: original data and ground truth. Second panel: plots of four indices as functions of $p$ (Elbow; Cali\'nski and Harabasz; silhouette; Sugar and James).}
\end{figure}

\begin{figure}
 \centering
\includegraphics[width=\textwidth]{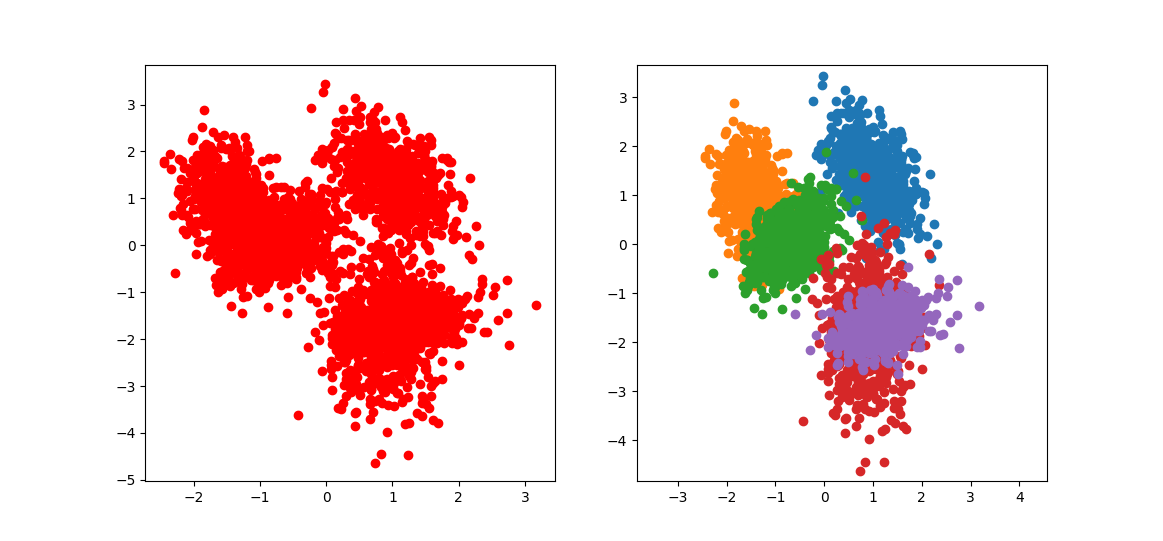}
\includegraphics[width=\textwidth]{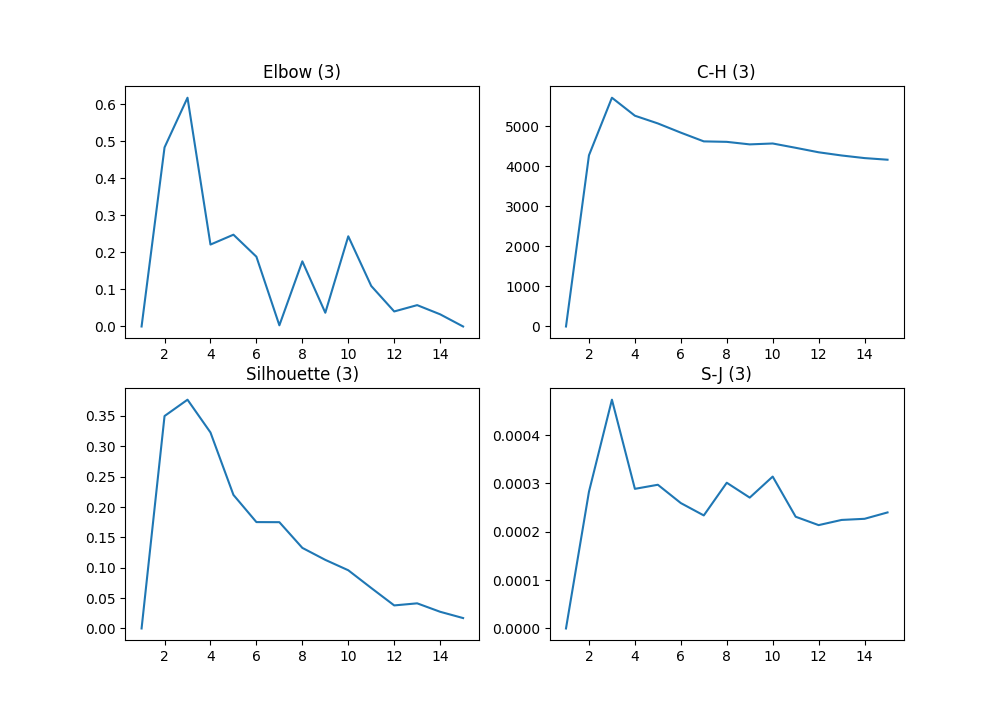}
\caption{\label{fig:clust.indx.3} Comparison of cluster indices for Gaussian clusters. First row: original data and ground truth. Second panel: plots of four indices as functions of $p$ (Elbow; Cali\'nski and Harabasz; silhouette; Sugar and James).}

 \end{figure}

Figures \crefrange{fig:clust.indx.1}{fig:clust.indx.3} provide a comparative illustration of some of these indexes.

\section{Bayesian Clustering}

\subsection{Introduction}
We have seen an example of model-based clustering with mixtures of Gaussian distributions. The main parameters in this model were the number of classes, $p$, and the probabilities $\al_j$ associated to each cluster,  and the parameter of the conditional distribution (e.g., $\CN(c_j, \sig^2\Id_{\mR^d})$) of $X$  conditionally to being in the $j$th cluster. In the approach we described, these parameters were estimated from data using maximum likelihood (through the EM algorithm) and probabilities $f_Z(j|x)$ were then estimated in order to compute the most likely clustering.We interpreted $f_Z(j|x)$  as the conditional probability $P(Z=z|X=x)$, where $Z\in \{1, \ldots, p\}$ represents the group variable. The natural generative order is $Z \to X$: first decide to which group the observation belongs to, then sample the value of $X$ conditional to this group. Clustering is in this case reversing the order, i.e., computing the posterior distribution of $Z$ given $X$.

In a Bayesian approach, the parameters $p, \usal, \usc$ and $\sig^2$ are also considered as random variables, so that  (letting $\underline \th$ denote the vector formed by these parameters), the generative random sequence becomes $\underline{\th} \to Z \to X$. Importantly, $\underline \th$ is assumed to be generated once for all, even if several samples of $X$ are observed, yielding the generative sequence for an $N$-sample,
\[
\underline{\th} \to (Z_1, \ldots, Z_N) \to (X_1, \ldots, X_N).
\]
We use below underlined letters to denote configurations of points, $\usZ = (Z_1, \ldots, Z_N)$, $\usX = (X_1, \ldots, X_N)$, etc. We also use capital letters or boldface letters (for Greek symbols) to differentiate random variable from realizations.

Clusters are still evaluated based on the conditional distribution of $\usZ$ given $\usX$, but this distribution must be evaluated by averaging the conditional distribution of $\usZ$ and $\underline\theta$ given $\usX$ with respect to $\underline \th$, formally\footnote{The symbol $\propto$ means ``equal up to a multiplicative constant''.},
\begin{align*}
P(\usz|\usx) &= \int P(\usz, \underline\th|\usx) P(\underline\th) d\underline\th\\
 &\propto \int \prod_{k=1}^N P(x_k|z_k, \underline\th) P(z_k|\underline\th) P(\underline\th)d\underline\th.
\end{align*}
In this expression, $P(\underline\th)d\underline\th$ implies an integration with respect to the prior distribution of the parameters. This distribution is part of the design of the method, but one usually chooses it so that it leads to simple computations, using so-called {\em conjugate priors}, which are such that posterior distributions belong to the same parametric family as the prior.  For example, the conjugate prior for the mean of a Gaussian distribution (such as $c_i$ in our model) is also a Gaussian distribution. The conjugate prior for a  scalar variance is the inverse gamma distribution, with density  
\[
\frac{v^u}{\Gamma(u)} s^{-u-1}\exp(-v/s)
\]
for some parameters $u,v$. A conjugate prior for the class probabilities $\underline \alpha = (\al_1, \ldots, \al_p)$ is the Dirichlet distribution, with density
\[
D(\al_1, \ldots, \al_p) = \frac{\Gamma(a_1+\cdots + a_p)}{\Ga(a_1) \cdots \Ga(a_p)} \prod_{j=1}^p \al_j^{a_j-1}
\]
on the simplex 
\[
\mathfrak S_p = \{(\al_1, \ldots, \al_p)\in \mR^p: \al_i \geq 0, \al_1+\cdots+\al_p=1\}.
\]

Note that these conjugate priors have the same form (up to normalization) as the parametric model densities when considered as functions of the parameters. 

\subsection{Model with a bounded number of clusters}
We first discuss the Bayesian approach assuming that the number of clusters is smaller than a fixed number, $p$. 
In this example, we assume  that $c_1, \ldots, c_p$ are modeled as independent Gaussian variables $\CN(0, \tau^2 \Id_{\mR^d})$, $\sig^2$ with an inverse gamma distribution with parameters $u$ and $v$ and $(\al_1, \ldots, \al_p)$ using a Dirichlet distribution with parameters $(a, \ldots, a)$. 

\paragraph{Analytical example.}
The joint probability density of $(\usX, \usZ)$ and $\underline{\boldsymbol \th}$ is proportional to
\begin{multline*}
(\sig^2)^{-u-1} e^{-v/\sig^2} e^{-\sum_{j=1}^p |c_j|^2/2\tau^2} \prod_{j=1}^p \al_j^{a-1}\prod_{k=1}^N \frac{e^{-|x_k - c_{z_k}|^2/2\sig^2}}{(\sig^2)^{d/2}} \prod_{k=1}^N \al_{z_k} \\
= (\sig^2)^{-u-dN/2 -1} \exp\left(- (v + \frac12 \sum_{k=1}^N|x_k-c_{z_k}|^2)/\sig^2\right)\prod_{j=1}^p \al_j^{a+N_j-1} . 
\end{multline*}
One can explicitly integrate this last expression with respect to $\sig^2$ and $\underline\alpha$, using the expressions of the normalizing constants in  the inverse gamma and Dirichlet distributions, yielding (after integration and ignoring constant terms)
\begin{multline*}
\frac{\Ga(a+N_1) \cdots \Ga(a+N_p)}{(v + \frac12 \sum_{k=1}^N|x_k-c_{z_k}|^2)^{u+dN/2}} \exp\left(- \sum_{j=1}^p |c_j|^2/2\tau^2\right)\\
= \frac{\Ga(a+N_1) \cdots \Ga(a+N_p)}{(v + \frac12 S_w + \frac12 \sum_{j=1}^p N_j |c_{j} - \bar x_j|^2)^{u+dN/2}} \exp\left(- \sum_{j=1}^p |c_j|^2/2\tau^2\right)
\end{multline*}
where $S_w = \sum_{k=1}^N |x_k - \bar x_{z_k}|^2$ is the within group sum of squares. Note that this sum of squares depends on $\usx$ and $\usz$, and that $(N_1, \ldots, N_p)$, the group sizes, depend on $\usz$. 

Let us assume a ``non-informative prior'' on the centers, which corresponds to letting $\tau$ tend to infinity and neglecting the last exponential. The remaining expression can now be integrated with respect to $c_1, \ldots, c_p$ by making a change of variables $\mu_j = \sqrt{N_j/(2v+S_k)} (c_j - \bx _j)$ and using the fact that
\begin{multline*}
\int_{(\mR^d)^p} \frac{dc_1\ldots dc_p}{(v + \frac12 S_w + \frac12 \sum_{j=1}^p N_j |c_{j} - \bar x_j|^2)^{u+dN/2}} = \\
(2v+S_w)^{(p-N)d/2 - u)}  \prod_{j=1}^p N_j^{-d/2}\int_{(\mR^d)^p} \frac{d\mu_1\ldots d\mu_p}{(\frac12+ \frac12 \sum_{j=1}^p |\mu_j|^2)^{u+dN/2}}
\end{multline*}
and the final integral does not depend on $\usx$ or  $\usz$. It follows from this that the conditional distribution of $\usZ$ given $\usx$ takes the form
\[
P(\usz\mid \usx) = C(\usx)\frac{\prod_{j=1}^p \Ga(a+N_j)}{(2v+S_w)^{(N-p)d/2 + u)}  \prod_{j=1}^p N_j^{d/2}}
\]
where $C(\usx)$ is a normalization constant ensuring that the right-hand side is a probability distribution over configurations $\usz = (z_1, \ldots, z_N) \in \{1, \ldots, p\}^N$. In order to obtain the most likely configuration for this posterior distribution, one should therefore minimize in $\usz$ the function
\[
((N-p)\frac d2 +u)\log (2v+S_w) + \frac d2 \sum_{j=1}^p \log N_j  -\sum_{j=1}^p \log \Ga(a+N_j).
\] 
This final optimization problem cannot be solved in closed form, but this can be performed numerically.  One can simplify it a little by only keeping the main order terms in the last two sums (using Stirling formula for the Gamma function) and minimize
\[
((N-p)\frac d2 +u)\log (2v+S_w)  - \sum_{j=1}^p (a+N_j) \log (a+N_j).
\] 
This expression has a nice interpretation, since
the first term minimizes the within-group sum of squares, the same objective function as in K-means, and the second one is an entropy term that favors clusters with similar sizes.

\paragraph{Monte-Carlo simulation.}
An alternative  to this analytical  approach is to use Monte-Carlo simulations to estimate some properties of the posterior distribution numerically. While they are often computationally demanding, Monte-Carlo methods are more flexible and can be used in situations when analytic computations are in\-trac\-table. In order to sample from the distribution of $\usZ$ given $\usx$, it is actually easier to sample from the joint distribution of $(\usZ, \underline{\boldsymbol \th})$ given $\usx$, because this distribution has a simpler form. Of course, if the pair $(\usZ, \underline{\boldsymbol \th})$ is sampled from the conditional distribution given $\usx$, the first component, $\usZ$ will follow the posterior distribution we are interested in. 

In the context of the discussed example, this reduces to sampling from a distribution proportional to 
\begin{equation}
\label{eq:MoG.posterior}
(\sig^2)^{-u-1} e^{-v/\sig^2} e^{-\sum_{j=1}^p |c_j|^2/2\tau^2} \prod_{j=1}^p \al_j^{a-1}\prod_{k=1}^N \frac{e^{-|x_k - c_{z_k}|^2/2\sig^2}}{(\sig^2)^{d/2}} \prod_{k=1}^N \al_{z_k}\,.
\end{equation}
Sampling from all these variables at once is  not tractable, but it is easy to sample from them in sub-groups, conditionally to the rest of the variables. We can, for example, deduce from the expression above the following conditional distributions.
\begin{enumerate}[label=(\roman*), wide=0.5cm]
\item Given $(\underline\alpha, \usc, \usz)$, $\sig^2$ follows an inverse gamma distribution with parameters $u+dN/2$ and $v + \frac12 \sum_{k=1}^N|x_k-c_{z_k}|^2$. 
\item Given $(\usz, \usz, \sig^2)$, $\underline\alpha$ follows a Dirichlet distribution with parameters $a+N_1, \ldots, a+N_p$.
\item Given $(\usz, \sig^2, \underline\alpha)$, $c_1, \ldots, c_p$ are independent and follow a Gaussian distribution, respectively with mean $(1 + \sig^2/(N_j\tau^2))^{-1} \bar x_j$ and variance $(N_j/\sig^2 + 1/\tau^2)^{-1}$.
\item Given $(\sig^2, \underline\alpha, \usc)$, $z_1, \ldots, z_N$ are independent and 
\[
P(z_k=j|\sig^2, \usal, \usc, \usx) \propto \al_j e^{-|x_k-c_j|^2/2\sig^2}.
\]
\end{enumerate}

\begin{algorithm}[Gibbs sampling for mixture of Gaussian (Bayesian case)]
\label{alg:gs.MoG}
\begin{enumerate}[label=(\arabic*)]
\item Initialize with variables $\underline\alpha, \usc, \sig$ and $\usz$, for example generated according to the prior distribution.
\item Loop a large number of times over the following steps.
\begin{enumerate}[label=(\roman*),wide=1cm]
\item Simulate  a new value of $\sig^2$ according to an inverse gamma distribution with parameters $u+dN/2$ and $v + \frac12 \sum_{k=1}^N|x_k-c_{z_k}|^2$. 
\item Simulate new values for $\al_1, \ldots, \al_p$ according to a Dirichlet distribution with parameters $a+N_1, \ldots, a+N_p$.
\item Simulate new values for  $c_1, \ldots, c_p$ independently, sampling $c_i$ according to a Gaussian distribution with mean $(1 + \sig^2/(N_j\tau^2))^{-1} \bar x_j$ and variance $(N_j/\sig^2 + 1/\tau^2)^{-1}$.
\item Simulate new values of $z_1, \ldots, z_N$ independently such that
\[
P(z_k=j|\sig^2,\usal, \usc, \usx) \propto \al_j e^{-|x_k-c_j|^2/2\sig^2}.
\]
\end{enumerate}
\end{enumerate}
\end{algorithm}

Note that this algorithm is only asymptotically providing a sample of the posterior distribution (it has to be stopped at some point, of course). Note also that, at each step, the labels $z_1, \ldots, z_N$ provide a random partition of the set $\{1, \ldots, N\}$, and this partition changes at every step. 

To estimate one single partition out of this simulation, several strategies are possible. Using the simulation, one can estimate the probability $w_{kl}$ that $x_k$ and $x_l$ belong to the same cluster. This can be dome by averaging the number of times that $z_k = z_l$ was observed along the Gibbs sampling iterations (from which one usually excludes a few early ``burn-in'' iterations). These weights, $w_{kl}$ can then be used as similarity measures in a clustering algorithm.

Alternatively, one can average for each $k$, the values of the class center $c_{z_k}$ associated to $k$, still along the Gibbs sampling iterations. These average values can then be used as input of, say, a K-means algorithm to estimate final clusters.

\paragraph{Mean-field approximation.}
We conclude this section with a variational Bayes approximation of the posterior distribution. We will make a mean-field approximation, in which all parameters and latent variables are independent, therefore approximating the distribution in \cref{eq:MoG.posterior} by a product distribution taking the form
\[ g(\sigma^2, \underline\alpha, \underline c, \underline z) = 
g^{(\sigma^2)}(\sigma^2) g^{(\alpha)}(\underline \alpha) \prod_{j=1}^p g^{(c)}_j(c_j)\prod_{k=1}^N g^{(z)}_k(z_k).
\]
Here $\underline c = (c_1, \ldots, c_p)$, $\underline z = (z_1, \ldots, z_N)$ and $\underline\alpha = (\alpha_1, \ldots, \alpha_p)$. We have $\sigma^2 \in (0, +\infty)$, $\underline c \in (\mR^{d})^p$, $\underline\alpha\in \mathcal S$, the set of all non-negative $\al_1, \ldots, \alpha_p$ that sum to one, and $\underline z\in \{1, \ldots, p\}^N$ (so that $g_k^{(x)}$ is a p.m.f. on $\{1, \ldots, p\}$.  We will use the discussion in \cref{sec:mean.field} and \cref{lem:max.ent}, and use the notation introduced in that section to denote as  $\avg{\boldsymbol\phi}$ the expectation a variable $\boldsymbol\varphi$ of the variables above for the p.d.f. $g$.

The log-likelihood for a mixture of Gaussian takes the form (ignoring contant terms)
\begin{align*}
\ell(\sigma^2, \underline \alpha, \underline c, \underline z ) =& 
  -(u+1) \log \sigma^2 - v\sigma^{-2}  
- \frac1{2\tau^2} \sum_{k=1}^p |c_j|^2 + \sum_{j=1}^p (a-1)\log \alpha_j  \\ 
&- \frac{Nd}{2} \log \sigma^2
- \frac12 \sigma^{-2}  \sum_{k=1}^N |x_k-c_{z_k}|^2  + \sum_{k=1}^N \log\alpha_{z_k}\\
= & -(u+1) \log \sigma^2 - v\sigma^{-2} - \frac1{2\tau^2} \sum_{k=1}^p |c_j|^2 + \sum_{k=1}^p (a-1)\log \alpha_j  \\ 
&- \frac{Nd}{2} \log \sigma^2
- \frac12 \sigma^{-2}  \sum_{k=1}^N \sum_{j=1}^p |x_k-c_{j}|^2 \bfone_{z_k=j}  + \sum_{k=1}^N \sum_{j=1}^p \log\alpha_{j} \bfone_{z_j=k}
\end{align*}
and can therefore be decomposed as  a sum of products of functions of single variables, as assumed in \cref{sec:mean.field}. Using \cref{lem:max.ent}, we can identify each of the distributions composing $g$, namely:
\begin{itemize}
\item $g^{(\sigma^2)}$ is the p.d.f. of an inverse gamma with parameters $\tilde u = u+Nd/2$ and
\[
\tilde v = \nu + \frac12   \sum_{k=1}^N \sum_{j=1}^p \avg{|x_k-C_j|^2}\, \avg{Z_k=j}.
\]
\item  $g_j^{(c)}$ is the p.d.f. of a Gaussian, with parameters  $\CN(\tilde m_j, \tilde \sigma_j^2\Id[d])$, with, letting
\[
\tilde \zeta(j) = \sum_{k=1}^N \avg{Z_k=j} = \sum_{k=1}^N g^{(z)}_k(j),
\]
$\tilde \sigma^2_j =  \left(\frac1{\tau^2} + \avg{\boldsymbol\sigma^{-2}} \tilde\zeta(j) \right)^{-1}$ and
$\tilde m_i = \avg{\boldsymbol\sigma^{-2}} \tilde \sigma^2_j  \sum_{k=1}^N \avg{Z_k=j} x_k$.
\item 
 $g^{(\alpha)}$ of a Dirichlet distribution, with parameters $\tilde a_1, \ldots, \tilde a_k$, with $\tilde a_i = a + \tilde \zeta(j) $.
 \item Finally $g_k^{(z)}$ is a p.m.f. on $\{1, \ldots, p\}$ with
 \[
 g_k^{(z)}(j) \propto \exp\left(- \frac12 \avg{\boldsymbol\sigma^{-2}} \avg{|x_k-C_j|^2} +\avg{\log \boldsymbol\alpha_j}\right).
\]
\end{itemize}

To complete the consistency equations, it now suffices to evaluate the expectations in the formula above as functions of the other parameters. We leave to  the reader the verification of the following statements.
\begin{itemize}
\item If $\boldsymbol\sigma^2$ follows an inverse gamma distribution with parameters $\tilde{u}$ and $\tilde{v}$, then  $\avg{\sigma^{-2}} = \tilde{u}/\tilde{v}$.
\item If $C_j \sim \CN(\tilde m_j, \tilde{\sigma}^2_j \Id[d])$, then $\avg{|x_k-C_j|^2} = |x_k-\tilde m_j|^2 + d\tilde{\sigma}^2_j$. 
\item  If $\underline{\boldsymbol\alpha}$ follows a Dirichlet distribution with parameters $\tilde a_1, \ldots, \tilde a_p$, then $\avg{\log\boldsymbol\alpha_j} = \psi(\tilde a_j) - \psi(\tilde a_1+\cdots + \tilde a_p)$ where $\psi$ is the {\em digamma} function (derivative of the logarithm of the gamma function).
\end{itemize}

Combining these facts with the expression of the mean-field parameters, we can now formulate a mean-field estimation algorithm for mixtures of Gaussian that iteratively applies the consistency equations.

\begin{algorithm}[Mean-field algorithm for mixtures of Gaussian]
\label{alg:mean.field.mog}
\begin{enumerate}[label=(\arabic*), wide=0.5cm]
\item: Input: training set $(x_1, \ldots, x_N)$, number of clusters $p$, prior parameters $u, v, \tau^2$ and $a$ .
\item Initialize variables $\tilde \sigma^2_1, \ldots, \tilde \sigma^2_p$, $\tilde m_1, \ldots, \tilde m_p$, $\tilde a_1, \ldots, \tilde a_p$, $\tilde g_k(j)$, $k=1, \ldots, N$, $j=1, \ldots, p$. 
\item Let  $\tilde \zeta(j) = \sum_{k=1}^N \tilde g_k(j)$, $j=1, \ldots, p$.
\item Let 
\[
\tilde \rho^2 = \frac{1}{u+Nd/2} \left(v + \frac12 \sum_{k=1}^N \sum_{j=1}^p \tilde g_k(j) |x_k-\tilde m_j|^2 + \frac{d}2 \sum_{j=1}^p \tilde\sigma_j^2\tilde\zeta(j)\right).
\]
\item For $j=1, \ldots, p$, let $\tilde \sigma^2_i =  \left(\frac1{\tau^2} +\frac{\tilde\zeta(j)}{\tilde \rho^2} \right)^{-1}$ and $\tilde m_i = \frac{\tilde\sigma_j^2}{\tilde \rho^2} \sum_{k=1}^N \tilde g_k(j) x_k$.
\item Let $\tilde a_i = a + \tilde \zeta(j) $, $j=1, \ldots, p$.
\item For $k=1, \ldots, N$, $j=1, \ldots, p$, let
\[
\tilde g_k(j) \propto \exp\left(- \frac1{2\tilde\rho^2}  \left(|x_k-\tilde m_j|^2 + d \tilde\sigma_j^2\right) + \psi(\tilde a_j) \right).
\]
\item Compare the updated variables with their previous values and stop if the difference is below a tolerance level. Otherwise, return to (3).
\end{enumerate}
\end{algorithm}
After convergence $g_k^{(z)}$ provides the mean-field approximation of the posterior probability of classes for observation $k$ and can be used to determine clusters.

\subsection{Non-parametric priors}
\label{sec:cluster.npb}
\paragraph{The Polya urn}
In the previous model with $p$ clusters or less, the joint distribution of $Z_1, \ldots, Z_N$ is given by
\[
\pi(z_1, \ldots, z_N) = \frac{\Ga(pa)}{\Ga(a)^p} \int_{\mathfrak S_p} \prod_{j=1}^p \al_j^{a+N_j -1} d\al = \frac{\Ga(pa)}{\Ga(pa+N)} \prod_{j=1}^p \frac{\Ga(a+N_j)}{\Ga(a)}.
\]
Conditional to $z_1, \ldots, z_N$, the data model was completed by sampling $p$ sets of parameters, say, $\th_1, \ldots, \th_p$, each belonging to a parameter space $\Th$  and following a prior probability distribution with density, say, $\psi$ and variables $X_1, \ldots, X_N$, where $X_k\in \CR$ was drawn according to a law dependent on its cluster, that we will denote $\phi(\,\cdot\mid\th_{z_k})$. The complete likelihood of the data is now
\[
L(\usz, \underline\th, \usx) = \frac{\Ga(pa)}{\Ga(pa+N)} \prod_{j=1}^p \frac{\Ga(a+N_j)}{\Ga(a)} \prod_{j=1}^p \psi(\th_j) \prod_{k=1}^N \phi(x_k|\th_{z_k}).
\]

Note that the right-hand side does not change if one relabels the values of $z_1, \ldots, z_N$, i.e., if one replaces each $z_k$ by $s(z_k)$ where $s$ is a permutation of $\{1, \ldots, p\}$, creating a new configuration denoted $s\cdot \usz$. Let $[\usz]$ denote the equivalence class of $\usz$, containing all $\usz' = s \cdot \usz, s\in \mathfrak S_N$: all the labelings in $[\usz]$ provide the same partition of $\{1, \ldots, N\}$ and can therefore be identified. One defines a probability distribution $\bar\pi$ over these equivalence classes by letting
\[
\bar \pi([\usz]) = |[\usz]|\frac{\Ga(pa)}{\Ga(pa+N)} \prod_{j=1}^p \frac{\Ga(a+N_j)}{\Ga(a)}.
\]
The first term on the right-hand side is  the number of elements in the equivalence class of $[\usz]$. To compute it, 
let $p_0 = p_0(\usz)$ denote the number of different values  taken by $z_1, \ldots, z_N$, i.e., the ``true'' number of  clusters (ignoring the empty ones), which now is a function of $\usz$. Let $A_1, \ldots, A_{p_0}$ denote the partition associated with $\usz$.  New labelings equivalent to $\usz$ can be obtained by assigning any index $i_1\in \{1, \ldots, p\}$ to elements of $A_1$, then any index $i_2\neq i_1$ to elements of $A_2$, etc., so that there are  $|[\usz]| = p!/(p-p_0)!$ choices. We therefore find:
\[
\bar \pi([\usz]) = \frac{p!}{(p-p_0)!} \frac{\Ga(pa)}{\Ga(pa+N)} \prod_{j=1}^p \frac{\Ga(a+N_j)}{\Ga(a)}.
\]
Letting $\la = pa$ and using the formula $\Ga(x+1) = x\Ga(x)$, this can be rewritten as
\[
\bar \pi([\usz]) = \frac{p(p-1)\cdots(p-p_0+1)}{\la (\la+1) \ldots (\la+N-1)} \prod_{j=1}^p \prod_{i=0}^{N_j-1} (\la/p+i).
\]

Now, the class $[\usz]$ contains exactly one element $\hat \usz$ with the following properties
\begin{enumerate}[label=$\bullet$, wide=0.5cm]
\item $\hat z_1 = 1$,
\item $\hat z_k \leq \max(z_j, j<k) + 1$ for all $k>1$.
\end{enumerate}
This means that the $k$th label is either one of those already appearing in $(\hat z_1, \ldots, \hat z_{k-1})$ or the next integer in the enumeration. We will call such a $\hat \usz$ admissible. 
If we assume that $\usz$ is admissible in the expression of $\bar \pi$, we can write
\[
\bar \pi([\usz]) = \frac{\prod_{j=1}^{p_0} \left(\la (1-j/p) \prod_{i=1}^{N_j-1} (\la/p + i)\right)}{\la (\la+1) \ldots (\la+N-1)}.
\]
If one takes the limit $p\to\infty$ in this expression, one still gets a probability distribution on admissible labelings, namely
\begin{equation}
\label{eq:dp.classes}
\bar \pi([\usz]) = \frac{\la^{p_0} \prod_{j=1}^{p_0} (N_j-1)! }{\la (\la+1) \ldots (\la+N-1)}.
\end{equation}
Recall that, in this equation, $p_0$ is a function of $\usz$, equal, for admissible labelings, to the largest $j$ such that $N_j > 0$.

The probability $\bar \pi$ is generated by the following sampling scheme, called the Polya urn process simulating admissible labelings.

\begin{algorithm}[Polya Urn]
\label{alg:polya.urn}
\begin{enumerate}[label=\arabic*, wide=0.5cm]
\item Initialize $k=1$, $z_1=1$, $j=1$. Let $N_1 = 1$
\item At step $k$, assume that $z_1, \ldots,  z_k$ have been generated, with associated number of clusters equal to $j$ and $N_1, \dots, N_j$ elements per cluster. Generate $z_{k+1}$ such that
\begin{equation}
\label{eq:polya.urn}
z_{k+1} = 
\left\{
\begin{aligned}
&i&& \text{ with probability } && \frac{N_i}{\la + k}, \text{ for } i=1, \ldots, j\\
&j+1&& \text{ with probability } && \frac{\la}{\la + k}
\end{aligned}
\right.
\end{equation}
\item If $z_{k+1} = i \leq j$, then replace $N_i$ by $N_{i}+1$, $k$ by $k+1$.
\item If $z_{k+1} = j+1$, let $N_{j+1} = 1$, replace $j$ by $j+1$ and $k$ by $k+1$.
\item If $k<N$, return to step 2, otherwise, stop.
\end{enumerate}
\end{algorithm}

Using this prior, the complete model for the distribution of the observed data is
\[
L(\usz, \underline \th, \usx) = \frac{\la^{p_0} \prod_{j=1}^{p_0} (N_j-1)! }{\la (\la+1) \ldots (\la+N-1)} \prod_{j=1}^{p_0} \psi(\th_j) \prod_{k=1}^N \phi(x_k|\th_{z_k})
\]
Recall that, in this expression, $\usz$ is restricted to the set of admissible labelings. We also note that admissible labelings are in one-to-one correspondence with the partitions of $\{1, \ldots, N\}$, so that the latent variable $\usz$ in this expression can also be interpreted as representing a random partition of this set.

\paragraph{Dirichlet processes.}

As we will see later, the expression of the global likelihood and the Polya urn model will suffice for us to develop  non-parametric clustering methods for a set of observations  $x_1, \ldots, x_N$. However, this model is also associated to an important class of random probability distributions (i.e., random variables taking values in some set of probability distributions) called Dirichlet processes for which we provide a brief description.

The distribution in \cref{eq:dp.classes} was obtained by passing to the limit from a  model 
that first generates $p$ numbers $\al_1, \ldots, \al_p$, then generates the labels $z_1, \ldots, z_N \in \{1, \ldots, p\}$ identified modulo relabeling. This distribution  can also be defined directly, by first defining an infinity of positive numbers $(\al_j, j\geq 1)$ such that $\sum_{i=1}^\infty \al_i = 1$,  followed by the generation of random labels $Z_1, \ldots, Z_N$ such that $P(Z_k = j) = \al_j$, followed once again with an identification up to relabeling.

 The distribution of $\underline{\boldsymbol \al}$ that leads to the Polya urn is called the {\em stick breaking process}. This process is such that
\[
\bfal_j = U_j \prod_{i=1}^{j-1}(1-U_i)
\] 
where $U_1, U_2, \ldots$ is a sequence of i.i.d. variables following a $\mathrm{Beta}(1, \la)$ distribution, i.e., with p.d.f. $\la (1-u)^{\la-1}$ for $u\in [0,1]$. The stick breaking interpretation comes from the way $\bfal_1, \bfal_2, \ldots$ can be simulated: let $\bfal_1 \sim \mathrm{Beta}(1, \la)$; given $\bfal_1, \ldots, \bfal_{j-1}$, let $\bfal_j = (1-\bfal_1 - \cdots - \bfal_{j-1})U_j $ where $U_j \sim \mathrm{Beta}(1, \la)$ and is independent from the past. Each step can be thought of as breaking the remaining length, $(1-\bfal_1 - \cdots - \bfal_{j-1})$, of an original stick of length 1 using a beta-distributed variable, $U_j$. This process leads to the distribution \cref{eq:dp.classes} over admissible distributions, i.e., if $\al$ is generated according to the stick breaking process, and $Z_1, \ldots, Z_N$ are independent, each such that $P(Z_k=j) = \al_j$, then the probability that $(Z_1, \ldots, Z_N)$ is identical, after relabeling, to the admissible configuration $z$ is given by \cref{eq:dp.classes}. (We skip the proof of this result, which is not straightforward.)

Now, take a realization $\usal = (\al_1, \al_2, \ldots)$ of the stick-breaking process, and independent realizations $\useta = (\eta_1,  \eta_1, \ldots)$ drawn according to the p.d.f. $\psi$. Define
\begin{equation}
\label{eq:dir.proc}
\rho = \sum_{j=1}^\infty \al_j \de_{\eta_j}\,.
\end{equation}
For any realization of $\bfal$ and of $\bfeta$, $\rho$ is a probability distribution on the parameter space $\Th$ (in which one chooses $\eta_i$ with probability $\al_i$). Since $\bfal$ and $\bfeta$ are both random variables, this defines a random variable $\boldsymbol\rho$  with values in the space of probability measures on $\Th$.

This process has the following characteristic property. For any family $V_1, \ldots, V_k\sub \Th$ forming a partition of that set,  the random variable $(\boldsymbol\rho(U_1), \ldots, \boldsymbol\rho(U_k))$  follows a Dirichlet distribution with parameters 
\[
\left(\la \int_{U_1} \psi\,d\eta , \ldots, \la \int_{U_1} \psi\,d\eta \right).
\] 
This is the definition of  a Dirichlet process with parameters $(\la, \psi)$, or, simply, with parameter $\la\psi$. Conversely, one can also show that any Dirichlet process can be decomposed as in \cref{eq:dir.proc} where $\bfal$ is a stick-breaking process and $\bfeta$ independent realizations of $\psi$.

\paragraph{Monte-Carlo simulation.}
The joint distribution of labels, parameters and observed variables can also be deduced from \cref{eq:dp.classes}, with a joint p.d.f. given by
\begin{equation}
\label{eq:dp.mix}
\frac{\la^{p_0-1} \prod_{j=1}^{p_0} (N_j-1)!}{(\la+1) \cdots (\la+N-1)} \prod_{j=1}^{p_0} \psi(\eta_j) \prod_{k=1}^N \phi(x_k|\eta_{z_k}).
\end{equation}
The forward simulation of this distribution is a straightforward extension of \cref{alg:polya.urn}, namely:
\begin{algorithm}
\label{alg:polya.urn.complete}
\begin{enumerate}[label=\arabic*, wide=0.5cm]
\item Initialize $k=1$, $z_1=1$, $j=1$. Let $N_1 = 1$.
\item Sample $\eta_1\sim \psi$ and $x_1 \sim \phi(\cdot|\eta_1)$.
\item At step $k$, assume that $z_1, \ldots,  z_k$ has been generated, with associated number of clusters equal to $j$ and $N_1, \dots, N_j$ elements per cluster. Generate $z_{k+1}$ such that
\[
z_{k+1} = 
\left\{
\begin{aligned}
&i&& \text{ with probability } && \frac{N_i}{\la + k}, \text{ for } i=1, \ldots, j\\
&j+1&& \text{ with probability } && \frac{\la}{\la + k}
\end{aligned}
\right.
\]
\item If $z_{k+1} = i \leq j$, sample $x_{k+1} \sim \phi(\,\cdot\, | \eta_i)$. Replace $N_i$ by $N_{i}+1$, $k$ by $k+1$.
\item If $z_{k+1} = j+1$, let $N_{j+1} = 1$, sample $\eta_{j+1} \sim \psi$ and $x_{k+1} \sim \phi(\,\cdot\, | \eta_{j+1})$. Replace $j$ by $j+1$ and $k$ by $k+1$.
\item If $k<N$, return to step 2, otherwise, stop.
\end{enumerate}
\end{algorithm}

This algorithm cannot be used, of course, to sample from the conditional distribution of $Z$ and $\bfeta$ given $X=x$, and Markov-chain Monte-Carlo must be used for this purpose.  In order  to describe how Gibbs sampling may be applied to this problem, we  use the fact that, as previously remarked, using admissible labelings $z$ is equivalent to using partitions $\CA = (A_1, \ldots, A_{p_0})$ of $\{1, \ldots, N\}$, and we will use the latter formalism to describe the algorithm. We will also use the notation $\eta_A$ to denote the parameter associated to $A\in \CA$ so our new notation for the variables is $(\CA, \useta)$ where $\CA$ is a partition of $\{1, \ldots, N\}$ and $\useta$ is a collection $(\eta_A, A\in \CA)$ with $\eta_A \in \Th$.
Given this, we want to sample from a conditional p.d.f.
\begin{equation}
\label{eq:dp.mix.cond}
\Phi(\CA, \useta | x) \propto \frac{\la^{|\CA|-1} \prod_{A\in\CA} (|A|-1)!}{(\la+1) \cdots (\la+N-1)} \prod_{A \in \CA} \psi(\eta_A) \prod_{k\in A}  \phi(x_k|\eta_A). 
\end{equation}
As an additional notation, given a partition $\CA$ and an index $k\in \{1. \ldots, N\}$, we let $\CA_k$ denote the set $A$ in $\CA$ that contains $k$.

The following points are relevant for the design of the sampling algorithm. 
\begin{enumerate}[label=(\arabic*), wide=0.5cm]
\item The conditional distribution of $\underline \bfeta$ given $\CA$ and the training data is proportional to 
\[
\prod_{A\in\CA} \left(\psi(\useta_A) \prod_{k\in A}  \phi(x_k|\useta_A)\right) 
\]
This shows that the parameters $\eta_A, A\in \CA$  are independent of each other, with $\eta_A$ following a distribution proportional to
\[
\eta \mapsto \psi(\eta) \prod_{k\in A_j}  \phi(x_k|\eta).
\]
Sampling from this distribution generally offers no special difficulty, especially if the prior $\psi$ is conjugate to $\phi$. Importantly, one does not need to sample exactly from $\eta_A$, and it is often more convenient to separate $\eta_A$ into several components (such as mean and variance for mixtures of Gaussian) and sample from them alternatively, creating another level of Gibbs sampling.

\item We now consider  the issue of updating $\CA$. We will use for this purpose the formalism of \cref{alg:gibbs.sampling}. In particular,  for each $k\in \{1, \ldots, N\}$, we associate to the variable $ (\CA, \useta)$ the pair $(\CA^{(k)}, \useta^{(k)})$, where $\CA^{(k)}$ is the partition  of $\{1, \ldots, N\} \setminus\{k\}$ formed by the sets $A^{(k)}  = A \setminus \{k\}$  and $\eta_A^{(k)} = \eta_A$, unless $A = \{k\}$, in which case the set and the corresponding $\eta_A$ are dropped. 

We can write $\Phi(\CA, \useta|\usx)$ in the form
\begin{equation}
\label{eq:dp.gs.1}
\Phi(\CA, \useta|\usx) \propto q(\CA_k, \eta_{\CA_k}) \phi(x_k|\eta_{\CA_k})\frac{\la^{|\CA^{(k)}|-1} \prod_{B \in A^{(k)}} (|B|-1)!}{(\la+1) \cdots (\la+N-1)} \prod_{B \in \CA^{(k)}} \psi(\eta_B) \prod_{l \in B}  \phi(x_l|\eta_B)
\end{equation}
with
\[
q(A, \th) = \sum_{B\in \CA^{(k)}} |B| \bfone_{A = B\cup \{k\}} + \la \psi(\th) \bfone_{A = \{k\}}
\]
Partitions $\CA'$ that are consistent with $\CA^{(k)}$ allocate $k$ to one of the clusters in $\CA^{(k)}$ or create a new cluster with a new parameter $\eta'_k$. If one replaces $(\CA, \bfeta)$ by $(\CA', \bfeta')$, only the first two terms in \cref{eq:dp.gs.1} will be affected, so that the conditional probability of $\CA'$ given $\CA^{(k)}$ is proportional to $q(\CA'_k, \eta_{\CA'_k})\phi(x_k|\eta_{\CA'_k})$ and  given by
\[
\left\{ 
\begin{aligned}
&\frac{|B| \phi(x_k|\eta_B)}{\mathfrak C_1 + \la\mathfrak C_2} \text{ if } \CA'_k = B\cup \{k\}, \eta'_B = \eta_B, B\in \CA^{(k)}\\
& \frac{\la \phi(x_k|\eta'_k) \psi(\eta'_k)}{\mathfrak C_1 + \la\mathfrak C_2} \text{ if } \CA'_k = \{k\}, 
\end{aligned}
\right.
\]
where 
\[
\mathfrak C_1 = \sum_{B \in \CA^{k}} |B| \phi(x_k|\eta_B) \quad \text{ and }\quad  \mathfrak C_2 = \int_{\Th} \phi(x_k|\th) \psi(\th) d\th.
\]
Concretely, this means that one first decides to allocate $k$ to a set $B$ in $\CA^{(k)}$ with probability $|B| \phi(x_k|\eta_B)/(\mathfrak C_1 + \lambda\mathfrak C_2)$ and to create a new set with probability $\la \mathfrak C_2 / (\mathfrak C_1 + \la\mathfrak C_2)$. If a new set is created, then the associated parameter $\eta'_{\{k\}}$ is sampled according to the p.d.f. $\phi(x_k|\th) \psi(\th/\mathfrak C_2$.

\item However, sampling using this conditional probability  requires the computation of the integral $\mathfrak C_2$, which can represent a significant computational burden, since this has to be done many times in a Gibbs sampling algorithm.
A modification of this algorithm, introduced in \citet{neal2000markov}, avoids this computation by adding new auxiliary variables at each step of the computation. These  variables are $m$ parameters $\eta_1^*, \ldots, \eta_m^*\in \Th$ where $m$ is a fixed integer. To define the joint distribution of $\CA, \useta, \useta^*$, one lets the marginal distribution of $(\CA, \useta)$ be given by \cref{eq:dp.mix.cond} and conditionally to $\CA, \useta$, let $\bfeta^*_1, \ldots, \bfeta^*_m$ be:
\begin{enumerate}[label = $(\roman*)$, wide=0.5cm]
\item independent with density $\psi$ if $|\CA_k| > 1$;
\item such that $\eta^*_j = \eta_{\CA_k}$ and the other $m-1$ starred parameters are independent with distribution $\psi$, where $j$ is randomly chosen in $\{1, \ldots, m\}$ if $\CA_k = \{k\}$.
\end{enumerate} 

With this definition, the joint conditional distribution of $(\CA, \useta, \useta^*)$ takes the form
\begin{multline}
\label{eq:dp.gs.2}
\widehat \Phi(\CA, \useta, \useta^*|\usx) \propto \hat q(\CA_k, \eta_{\CA_k}, \useta^*) \phi(x_k|\eta_{\CA_k})
\\\frac{\la^{|\CA^{(k)}|-1} \prod_{B \in A^{(k)}} (|B|-1)!}{(\la+1) \cdots (\la+N-1)} \prod_{B \in \CA^{(k)}} \psi(\eta_B) \prod_{l \in B}  \phi(x_l|\eta_B)
\end{multline}
with
\[
\hat q (A, \th, \eta^*_1, \ldots, \eta_m^*) = \sum_{B\in \CA^{(k)}} |B| \bfone_{\th = \eta_B, A = B\cup \{k\}} \prod_{j=1}^m \psi(\eta^*_j) + \frac{\la}{m} \sum_{j=1}^m\bfone_{\th = \eta^*_j, A = \{k\}} \psi(\th) \prod_{i=1, i\neq j}^m \psi(\eta^*_i).
\]
Note that $\widehat\Phi$ depends on $k$, so that the definition of the auxiliary variables will change at each step of Gibbs sampling.
The conditional distribution, for $\widehat\Phi$, of $\CA', \useta'$ given $\CA^{(k)}, \useta^{(k)}, \useta^*$
is such that
\begin{enumerate}[label = $\bullet$, wide=0.5cm]
\item $\CA'_k=B \cup\{k\}$ and $\eta'_{\CA'_k} = \eta_B$ with probability $|B| \phi(x_k|\eta_B)/\mathfrak C$, for $B\in \CA^{(k)}$.
\item $\CA'_k = \{k\}$ and $\eta_{\CA'_k} = \eta^*_j$ with probability $(\la/m) \phi(x_k|\eta^*_j)/\mathfrak C$, $j=1, \ldots, m$.
\end{enumerate}
The constant $\mathfrak C$ is given by
\[
\mathfrak C = \sum_{B \in \CA^{k}} |B| \phi(x_k|\eta_B) + \frac{\la}m  \sum_{j=1}^m \phi(x_k|\eta^*_j)
\]
and is therefore easy to compute.
\end{enumerate}

We can now summarize this discussion with Neal's version of the Gibbs sampling algorithm. 
\begin{algorithm}[Neal]
\label{alg:dpm.posterior.neal}
Initialize the algorithm with some arbitrary partition and parameters $(\CA, \useta)$ (for example, generated using the Dirichlet prior). Use the same notation to denote these variables at the end of the previous iteration of the algorithm. The next iteration is then run as follows.
\begin{enumerate}[label=(\arabic*)]
\item For $k=1, \ldots, N$, reallocate $k$ to a cluster as follows.   
\begin{enumerate}[label=(\roman*), wide=0.5cm]
\item Form the new family of sets $\CA^{(k)}$ and labels $\useta^{(k)}$ by removing $k$ from the partition $\CA$.
\item If $|\CA_k| > 1$, generate $m$ variables $\eta^*_1, \ldots, \eta^*_m$ according to $\psi$. If $\CA_k= \{k\}$, generate only $m-1$ such variables and let the last one be equal to $\eta_{\CA_k}$.
\item Allocate $k$ to a new cluster $A'$ with parameter $\eta'_{A'}$ according to probabilities proportional to  
\[
\left\{
\begin{aligned}
& |B|  \phi(x_k|\eta^{(k)}_B) \text{ if } A' = B \cup \{k\} \text{ and } \eta'_{A'} = \eta^{(k)}_B\\
&\frac{\la}{m}  \phi(x_k|\eta^*_j) \text{ if } A = \{k\}  \text{ and }   \eta'_{A'}  = \eta^*_j, j=1, \ldots, m
\end{aligned}
\right.
\]
\end{enumerate}
\item For $A\in \CA$, update $\eta_A, A\in \CA$ according to the distribution proportional to
\[
\psi(\eta) \prod_{k\in A} \phi(x_k|\eta)
\]
either directly, or via one step of Gibbs sampling visiting each of the variables that constitute $\eta_A$.
\item Loop a sufficient number of times over the previous two steps.
\end{enumerate}
\end{algorithm}

After running this algorithm, the set of clusters should be finalized by using statistics computed along the simulation, as discussed after \cref{alg:gs.MoG}.
\bigskip

\paragraph{Full example: Mixture of Gaussian.}
To conclude this section, we summarize the Monte-Carlo sampling algorithm for mixtures of Gaussian using a non-parametric Bayesian prior. Here, $\eta\in \Th$ is the center $c \in \mR^d $, with  prior distribution $\psi =  \CN(0, \tau^2\Id[d])$. The previous algorithm must be modified because an additional parameter $\sig^2$ is shared by all classes, with prior given by an inverse gamma distribution with parameters $u$ and $v$. The conditional distribution of the data is $\phi(x|c, \sig) \sim \CN(c, \sig^2\Id[d])$.

\begin{algorithm}[Gibbs sampling for non-parametric mixture of Gaussian]
\label{alg:gs.MoG.full}
\begin{enumerate}[label=(\arabic*)]
\item Initialize the algorithm with some arbitrary partition and parameters $(\CA, \useta)$.
\item For $k=1, \ldots, N$, reallocate $k$ to a cluster as follows.   
\begin{enumerate}[label=(\roman*), wide=0.5cm]
\item Form the new family of sets $\CA^{(k)}$ and labels $\bfeta^{(k)}$ by removing $k$ from the partition $\CA$.
\item If $|\CA_k|  > 1$, generate $m$ variables $c^*_i$, $i=1, \ldots, m$ independently with $c^*_i \sim \CN(0, \tau^2\Id[d])$. If $\CA_k= \{k\}$, generate only $m-1$ such pairs of variables and let the last one be equal to $c_{\CA_k} $.
\item Allocate $k$ to a new cluster $A'$ with parameter $c'_{A'}$ according to probabilities proportional to  
\[
\left\{
\begin{aligned}
& |B|  \exp\big(-\frac{|x_k - c^{(k)}_B|}{2\sig^2}\big) \text{ if } A' = B \cup \{k\} \text{ and } c'_{A'} = c^{(k)}_B\\
&\frac{\la}{m}  \exp\big(-\frac{|x_k - c^*_B|}{2\sig^2}\big) \text{ if } A = \{k\}  \text{ and }   c'_{A'}  = c^*_j, j=1, \ldots, m.
\end{aligned}
\right.
\]
\end{enumerate}

\item Simulate  a new value of $\sig^2$ according to an inverse gamma distribution with parameters $u+dN/2$ and $v + \frac12 \sum_{k=1}^N|x_k-c_{\CA_k}|^2$. 
\item Simulate new values for  $c_A, A\in\CA$ independently, sampling $c_A$ according to a Gaussian distribution with mean $(1 + \sig^2/(N_j\tau^2))^{-1} \bar x_A$ and variance $(|A|/\sig^2 + 1/\tau^2)^{-1}$, where
\[
\bar x_A = \frac1{|A|} \sum_{k\in A} x_k.
\] 
\end{enumerate}
\end{algorithm}

\newpage
\markright{PROBLEMS}
\section*{Problems}
\addcontentsline{toc}{section}{Problems}
\input Problems_Clustering


\chapter{Dimension Reduction and Factor Analysis}
\label{chap:unsupervised}
\label{chap:dim.red}

\section{Principal component analysis}
\label{sec:pca}

\subsection{General Framework}

Factor analysis aims at representing potentially high-dimensional data as functions of a (generally) small number of ``factors,'' with a representation taking the general form
\begin{equation}
\label{eq:fact.anal.gen}
X = \Phi(Y, \theta) + \text{residual},
\end{equation}
where $X$ is the observation, $Y$ provide the factors and $\Phi$ is a function parametrized by $\theta$.
A factor analysis model must therefore specify $\Phi$ (often, a linear function of $Y$), add hypotheses on $Y$ (such as its dimension, or properties of its distribution) and on the residuals.
The transformation $\Phi$ is estimated from training data, but, ideally, the method should also provide an algorithm that infers $Y$ from a new observation of $X$. Most of the time,  $Y$ is small dimensional so that the model also implies a reduction of dimension.

Assuming that $N$ independent realizations of $X$, denoted $x_1, \ldots, x_N$, are observed, forming our training set $T$, our goal is to obtain a small-dimensional representation of these data, while loosing a minimal amount of relevant information. Principal component analysis (or PCA), with which we start our discussion, is the simplest and most commonly used approach developed for this purpose.
In the following, we assume that the random variable $X$ takes values in a finite- or
infinite-dimensional inner-product space $H$. We will denote, as usual, by
$\scp{.}{.}_H$ the product in this space.

If $V$ is a finite-dimensional subspace of $H$, we denote by $P_V(y)$  the orthogonal projection of $y\in H$ on $V$, i.e., the element $\xi\in V$ such that $\|y-\xi\|^2_H$ is minimal (see \cref{sec:orth.proj}).  Recall that this orthogonal projection if characterized by the two properties: (i) $P_V(y) \in V$ and (ii) $(y-P_V(y)) \perp V$.

Given a target dimension $p$, PCA determines a $p$-dimensional subspace of $H$, say, $V$, and a point $c\in H$, such that, letting
\[
R_k = x_k - c - P_V(x_k-c)
\]
for $k = 1, \ldots, N$, the residual sum of squares
\begin{equation}
\label{eq:pca.res}
S = \sum_{k=1}^N \|R_k\|^2_H
\end{equation}
is as small as possible. 
One can immediately notice that an optimal choice for $c$ is 
$c = \bx = \sum_{k=1}^N x_k/N$.
Indeed, using the linearity of the orthogonal projection, we have
\begin{align*}
S &= \sum_{k=1}^N \|x_k - P_V(x_k) - (c-P_V(c))\|^2_H \\
& = \sum_{k=1}^N \|x_k - P_V(x_k) - (\bx-P_V(\bx))\|^2_H + N  \|\bx - P_V(\bx) - (c-P_V(c))\|^2_H.
\end{align*}
Given this, there would be no loss of generality in assuming that all $x_k$'s
have been replaced by $x_k-\bx$ and in taking $c=0$. While this is often done in the literature, there are some advantages (especially when discussing kernel methods) in  keeping the average explicit in the computation, as we will continue to do. 

Introducing an orthonormal basis $(e_1, \ldots, e_p)$ of $V$, one has 
\[
P_V(x_k - \bx) = \sum_{i=1}^p \rho_{k}(i)
e_i
\]
with 
$\rho_{ki} = \scp{x_k-\bx}{e_i}_H.$
One can then reformulate the PCA problem in terms of $(e_1, \ldots, e_p)$, which  must minimize
\begin{eqnarray*}
S &=& \sum_{k=1}^N \|x_k-\bx - \sum_{i=1}^p \scp{x_k-\bx}{e_i} e_i\|_H^2\\
& =& \sum_{k=1}^N \|x_k-\bx\|_H^2 - \sum_{i=1}^p \sum_{k=1}^N
\scp{x_k-\bx}{e_i}^2_H.
\end{eqnarray*}

For $u, v \in H$, define
$$
\scp{u}{v}_T = \frac{1}{N} \sum_{k=1}^N \scp{x_k-\bx}{u}_H\scp{x_k-\bx}{v}_H
$$
and $\norm{u}_T = \scp{u}{u}_T^{1/2}$ (the index $T$ refers to the
fact that this norm is associated with the training set). This provides a new quadratic
form on $H$. The formula above shows that minimizing $S$  is equivalent
to maximizing
$$
\sum_{i=1}^p \|e_i\|^2_T
$$ 
subject to the constraint that $(e_1, \ldots, e_p)$ is orthonormal in $H$. Note that $\scp{u}{v}_T = \scp{A_T u}{v}_H$ with
\begin{equation}
\label{eq:pca.AT}
A_T u = \frac1N \sum_{k=1} \scp{x_k-\bar x}{u}_H (x_k-\bar x).
\end{equation}
Letting $W = \vspan(x_1-\bx, \ldots, x_N -\bx)$, we have $\mathrm{Range}(A_T) = W$. The inclusion $\mathrm{Range}(A_T) \subset W$ is obvious, and the converse can be proved by showing that $\mathrm{Range}(A_T)^\perp \subset W^\perp$. Taking $u \in \mathrm{Range}(A_T)^\perp$, we have
\[
0 = \scp{A_Tu}{u}_H = 0 = \frac1N \sum_{k=1}^N \scp{u}{x_k - \bx}_H^2
\]
showing that $\scp{u}{x_k - \bx}_H=0$ for all $k$ and $u \in W^\perp$.

Let us consider a slightly more general problem. If $H$ is a separable Hilbert space\footnote{A Hilbert space is an inner-product space which is complete for its norm. A separable Hilbert space must have a dense countable subset, which, in particular, implies that it has orthonormal bases.}
 and $\mu$ is a square-integrable probability measure on $H$, such that 
\[
\int_H \|x\|^2_H\, d\mu(x)<\infty,
\]
one can define $m = \int_H x d\mu(x)$ and $\sig_\mu^2 = \int_H \|x-m\|^2_H d\mu$. 
One can then define the {\em covariance bilinear form}
\[
\Ga_\mu(u,v) = \int_H \scp{u}{x-m}_H\, \scp{v}{x-m}_H\, d\mu(x),
\]
which satisfies $\Ga_\mu(u,v) \leq \sigma_\mu^2 \|u\|_H\,\|v\|_H$. 

With this notation, we have
\[
\scp{u}{v}_T = \Ga_{\hat\mu_T}(u,v),
\]
where $\hat\mu_T = (1/N) \sum_{k=1}^N \de_{x_k}$ is the empirical measure (and in that case $m = \bar x$). We can therefore generalize the PCA problem by considering the  maximization of 
\begin{equation}
\label{eq:pca.hilb}
\sum_{k=1}^p \Ga_\mu(e_k, e_k)
\end{equation}
over all orthonormal families $(e_1, \ldots, e_p)$ in $H$.

Because $\mu$ is square integrable, the associated operator, $A_\mu$ defined by 
\begin{equation}
\label{eq:cov.operator}
\scp{u}{A_\mu v}_H = \Ga_\mu(u,v)
\end{equation} 
for all $u,v\in H$, is a {\em Hilbert-Schmidt operator} \citep{yos70}. Such an operator can, in particular, be diagonalized in an orthonormal basis of $H$, i.e., there exists an orthonormal
basis $(f_1, f_2, \ldots)$ of $H$ such that $A_\mu f_i = \la^2_i f_i$ for a non-increasing sequence of eigenvalues (with $\la_1\geq \la_2 \geq \cdots \geq 0$) such that
\[
\sig_\mu^2 = \sum_{k=1}^\infty \la^2_i.
\]

%
%

The main statement of the following result is in finite dimensions, a simple application of \cref{cor:pca.base}. We here give a direct proof that also works in infinite dimensions.
 \begin{theorem}
\label{th:rayleigh}
Let $(f_1, f_2, \ldots)$ be an orthonormal basis of eigenvectors of  $A_\mu$ with associated eigenvalues $\la_1^2\geq \la_2^2 \geq \cdots \geq 0$.
Then an orthonormal family $(e_1, \ldots, e_p)$ in $H$ maximizes \cref{eq:pca.hilb} if and only if,
\begin{equation}
\label{eq:rayleigh}
\vspan(f_j: \la^2_j>\la^2_p) \subset \vspan(e_1, \ldots,
e_p) \subset \vspan(f_j: \la^2_j \geq \la^2_p).
\end{equation}
In particular $f_1, \ldots, f_p$ always provide a solution and $\vspan(e_1, \ldots, e_p) = \vspan(f_1, \ldots, f_p)$ for any other solution as soon as $\la^2_p > \la^2_{p+1}$. 
\end{theorem}

\begin{definition}
\label{def:princ.comp}
When $\mu = \hat \mu_T$, the vectors $(f_1, \ldots, f_p)$ are
called (with some abuse when eigenvalues coincide) the first $p$ principal components of the training set $(x_1, \ldots, x_N)$. 
\end{definition}
\begin{proof}
If $(e_1, \ldots, e_p)$ is an orthonormal family in $H$, let 
\[
F(e_1, \ldots, e_p) = \sum_{k=1}^p \Ga_\mu(e_k, e_k)\,.
\]
Note that $F(f_1, \ldots, f_p) = \la^2_1 + \cdots + \la^2_p$.
Write $e_k = \sum_{j=1}^\infty \al_{k}^{(j)} f_j$ (so that $\al_{k}^{(j)} = \scp{f_j}{e_k}_H$). These coefficients satisfy $\sum_{j=1}^\infty \al_{k}^{(j)}\al_{l}^{(j)} = 1$ if $k=l$ and 0 otherwise. Then
\[
\Ga_\mu(e_k, e_k) = \sum_{j=1}^\infty \la^2_j (\al_{k}^{(j)})^2.
\]
We have
\begin{align*}
F(e_1, \ldots, e_p) &= \sum_{k=1}^p \sum_{j=1}^\infty \la^2_j (\al_{k}^{(j)})^2
= \sum_{k=1}^p \sum_{j=1}^p \la^2_j (\al_{k}^{(j)})^2 + \sum_{k=1}^p \sum_{j=p+1}^\infty \la^2_j (\al_{k}^{(j)})^2
\\& 
\leq \sum_{k=1}^p \sum_{j=1}^p \la^2_j (\al_{k}^{(j)})^2 + \sum_{k=1}^p \sum_{j=p+1}^\infty \la^2_{p+1} (\al_{k}^{(j)})^2
= \sum_{j=1}^p (\la^2_j-\la^2_{p+1}) \sum_{k=1}^p (\al_{k}^{(j)})^2 + p\la^2_{p+1}.
\end{align*}
Let $P$ denote the orthogonal projection operator from $H$ to $\vspan(e_1, \ldots, e_p)$. We have, for any $h\in H$, $\|Ph\|_H^2 \leq \|h\|_H^2$ with equality if and only if $h\in\vspan(e_1, \ldots, e_p)$. Applying this to $h=f_j$, with $P(f_j) = \sum_{k=1}^p \al_{k}^{(j)} e_k$, we get $\sum_{k=1}^p  (\al_{k}^{(j)})^2 \leq 1$ with equality if and only if $f_j\in \vspan(e_1, \ldots, e_p)$. 

As a consequence, the previous upper bound on $F(e_1, \ldots, e_p)$ implies 
\[
F(e_1, \ldots, e_p) \leq  \sum_{j=1}^p \la^2_j.
\]
This upper bound is attained at $(e_1, \ldots, e_p) = (f_1, \ldots, f_p)$, which is therefore a maximizer. Also,  inspecting the argument above, we see that $F(e_1, \ldots, e_p) < \la^2_1 + \cdots + \la^2_p$ unless
\begin{enumerate}[label=(\alph*)]
\item for all $k\leq p$ and $j\geq p+1$: $\al_{k}^{(j)} = 0$ if $\la^2_j > \la^2_{p+1}$, and
\item for all $j\leq p$: $\sum_{k=1}^p (\al_{k}^{(j)})^2 = 1$ unless $\la^2_j = \la^2_{p+1}$.
\end{enumerate}
Condition (a) implies that $\vspan(e_1, \ldots, e_p) \subset \vspan(f_j: \la^2_j \leq \la^2_{p+1})$. If $\la^2_p = \la^2_{p+1}$, the inclusion 
$\vspan(e_1, \ldots,
e_p) \subset \text{span}(f_j: \la^2_j \leq \la^2_p)$ therefore holds. 
If $\la^2_p < \la^2_{p+1}$, condition (b) requires $\sum_{k=1}^p (\al_{k}^{(j)})^2 = 1$ for all $j\leq p$, which implies $f_j \in\vspan(e_1, \ldots, e_p)$ for $j\leq p$, so that $\vspan(e_1, \ldots, e_p) = \vspan(f_1, \ldots, f_p)$ and  the inclusion also hold.

Condition (b) always requires  $\sum_{k=1}^p (\al_{k}^{(j)})^2 = 1$ (hence $f_j\in \vspan(f_1, \ldots, f_p)$) when $\la_j < \la_p$, showing that $\vspan(f_j: \la^2_j<\la^2_p) \subset \vspan(e_1, \ldots,
e_p)$. \Cref{eq:rayleigh} therefore always holds for $(e_1, \ldots, e_p)$ such that $F(e_1, \ldots, e_p) = \la^2_1 +\cdots +\la^2_p$. Furthermore, conditions (a) and (b) always hold for any orthonormal family that satisfy \cref{eq:rayleigh}, showing that any such solution is optimal.
\end{proof}

Notice that the optimal  $S$ in \cref{eq:pca.res} is such that
$$
S = N\sum_{i>p}\la^2_i.
$$ 

\begin{remark}
The interest of discussing PCA associated with a covariance operator for a square integrable measure (in which case it is often called a Karhunen-Loeve (KL) expansion) is that this setting is often important when discussing infinite-dimensional random processes (such as Gaussian random fields). Moreover, these operators quite naturally provide asymptotic versions of sample-based PCA. Interesting issues, that are part of {\em functional data analysis} \citep{rs97}, address the design of proper estimation procedures to obtain converging estimators of KL expansions based on finite samples for stochastic processes in infinite-dimensional spaces.
\end{remark}

\subsection{Computation of the principal components}
\label{sec:pca.comp}
\paragraph{Small dimension.}
Assume that $H$ has finite dimension, $d$, i.e.,  $H= \mR^d$, and represent $x_1, \ldots,
x_N\in \mR^d$ as
column vectors. Let the inner product on $H$ be associated
to a positive-definite symmetric matrix $Q$:
\[
\scp{u}{v}_H = u^T Q v.
\]  
Introduce the covariance matrix of the data
\[
\Sig_T = \frac{1}{N} \sum_{k=1}^N (x_k - \bx)(x_k-\bx)^T,
\]
Write $A_T = A_{\hat \mu_T}$, for short,  in \cref{eq:cov.operator}. We have:
\begin{align*}
\scp{u}{A_T v}_H & = \frac1N \sum_{k=1}^N (u^TQ(x_k-\bx))(v^TQ(x_k-\bx))\\
&= \frac1N \sum_{k=1}^N u^TQ(x_k-\bx)(x_k-\bx)^TQv\\
&= \scp{u}{\Sig_T Qv}_H\,,
\end{align*}
so that $A_T = \Sig_T Q$.

The eigenvectors, $f$, of $A_T$ are such that $Q^{1/2}f$ are eigenvectors of the symmetric matrix $Q^{1/2} \Sig_T Q^{1/2}$, which shows that they form an orthogonal system in $H$, which will be orthonormal if the eigenvectors are normalized so that $f^T Q f =1$. Equivalently, they solve the generalized eigenvalue problem $Q\Sig_T Q f = \la^2 Q f$, which may be preferred numerically to  diagonalizing the non-symmetric matrix $\Sig_T Q$.

\begin{remark}
If $Q^{-1}$ is easy to compute, then one can directly solve the generalized eigenvalue problem  $\Sig_T \tilde f = \la^2 Q^{-1} \tilde f$ and set $f = Q^{-1} \tilde f$. The normalization $f^T Q f = 1$ is then obtained by normalizing $\tf$ so that  $\tilde f^T Q^{-1}\tf=1$. 
\end{remark}

\begin{remark}
The ``standard'' version of PCA applies this computation using the Euclidean inner product, with $Q = \Id_{\mR^d}$, and the principal components are the eigenvectors of the covariance matrix of $T$ associated with the largest eigenvalues.
\end{remark}

\paragraph{Large dimension.}
It often happens that the dimension of $H$ is much larger than the
number of observations, $N$. In such a case, the previous approach is
quite inefficient (especially when the dimension of $H$ is infinite!)  and one should proceed as follows. 

Returning to the original problem, let $W =\mathrm{span}\{x_1-\bx, \ldots, x_N-\bx\}$ be the vector space generated by the centered data. Assuming that $p\leq \mathrm{dim}(W)$ (since $p=\mathrm{dim}(W)$ already guarantees exact reconstruction), one can see that any optimal $V$ must be a subspace of $W$. This is a consequence of \cref{th:rayleigh}, since $W$ is generated by the eigenvectors of $A_T$ associated with positive eigenvalues.

Indeed, letting $V' = P_W(V)$ (the projection of $V$ on $W$), we have, for $\xi \in W$,
\begin{align*}
\|\xi - P_V \xi\|^2_H &= \|\xi\|^2_H - 2\scp{\xi}{P_V \xi}_H + \|P_V \xi\|_H^2 \\
& = \|\xi\|^2_H - 2\scp{P_W \xi}{P_V \xi}_H + \|P_V \xi\|_H^2 \\
& = \|\xi\|^2_H - 2\scp{\xi}{P_WP_V \xi}_H + \|P_V \xi\|_H^2 \\
& \geq \|\xi\|^2_H - 2\scp{\xi}{P_WP_V x}_H + \|P_W P_V \xi\|_H^2 \\
& =  \|\xi - P_W P_V \xi\|_H^2 \\
& \geq  \|\xi - P_{V'} \xi\|_H^2 \,.
\end{align*}
In this computation, we have used the facts that $P_W \xi = \xi$ (since $\xi\in W$), that $\|P_WP_V\xi\|_H \leq \|P_V\xi\|_H$, that $P_WP_V\xi\in V'$ and that $P_{V'}(\xi)$ is the best approximation of $\xi$ by an element of $V'$.
This shows that (since $x_k - \bx\in W$ for all $k$)
\[
\sum_{k=1}^N \|x_k - \bx - P_{V} (x_k -\bx)\|_H^2 \geq \sum_{k=1}^N \|x_k-\bx - P_{V'} (x_k-\bx)\|_H^2
\]
with $V'$ a subspace of $W$ of dimension less than $p$, proving the result. This computation also shows that no improvement in PCA can be obtained by looking for spaces of dimension $p\geq \mathrm{dim}(W)$ (with $\mathrm{dim}(W)\leq N-1$ because the data is centered). 
\bigskip

It therefore suffices to look for
$f_1, \ldots, f_{p}$ in the form
$$
f_i = \sum_{k=1}^N \alpha_{k}^{(i)} (x_k-\bx).
$$
for some $\al_{k}^{(i)}$, $1\leq k\leq N, 1\leq i\leq p$.

With this notation, we have $\scp{f_i}{f_j}_H = \sum_{k,l=1}^N
\alpha_{k}^{(i)}\alpha_{l}^{(j)} \scp{x_k-\bx}{x_l-\bx}_H$ and
\begin{multline*}
\scp{f_i}{f_j}_T = \frac1N \sum_{l=1}^N \scp{f_i}{x_l-\bx}_H\scp{f_j}{x_l-\bx}_H \\
= \frac1N \sum_{k,k'=1}^N
\al_{k}^{(i)}\al_{k'}^{(j)} \sum_{l=1}^N \scp{x_k-\bx}{x_l-\bx}_H\scp{x_{k'}-\bx}{x_{l}-\bx}_H.
\end{multline*}
Let $S$ be the Gram matrix of the centered data, i.e., 
\[
S = (\scp{x_k-\bx}{x_l-\bx}_H,  k,l=1, \ldots, N).
\]
Let $\al^{(i)}$ be the column vector with coordinates $\al_{k}^{(i)}$, $k=1, \ldots, N$. We
have
$\scp{f_i}{f_j}_H = (\al^{(i)})^TS\pe\al j$ and $\scp{f_i}{f_j}_T = (\pe \al i)^T S^2
\pe \al j/N$, which implies that, in this representation, the operator $A_T$ is  given by $S/N$. Thus, the previous simultaneous orthogonalization problem can
be solved in terms of the $\al$'s by diagonalizing $S$ and taking the first eigenvectors, normalized so that $(\pe \al i)^T S\pe \al i=1$. Let $\la^2_j$, $j=1, \ldots, N$ be the eigenvalues of  $S/N$ (of which only the first $\min(d, N-1)$ may be non-zero). In this representation, the decomposition of the projection of $x_k$ on the PCA basis is given by
\[
x_k = \sum_{j=1}^p \pe {\be_{k}} j f_j
\]
with
\[
\pe {\be_{k}} j = \scp{x_k-\bx}{f_j}_H = \sum_{l=1}^N \alpha_{l}^{(j)} \scp{x_l-\bx}{x_k-\bx}_H = N\la_j^2 \al_{k}^{(j)}\,.
\]

\section{Kernel PCA}
\label{sec:kernel.pca}
Since the previous computation only depended on the inner products
$\scp{x_k-\bx}{x_l-\bx}_H$, PCA can be performed in reproducing kernel Hilbert spaces, and the resulting method is called
kernel PCA. In this framework, $X$ may take values in any set $\CR$ with a representation $h: \CR \to H$. The associated  kernel, $K(x,x')
=\scp{h(x)}{h(x')}_H$,  provides a closed form expression of  the inner products in terms of the original variables. The
feature function itself is most of the time unnecessary.

The kernel version of PCA consists in replacing $x_k - \bx$ with $h(x_k) - \bar h$ where $\bar h$ is the average feature. This leads to defining 
a ``centered kernel:''
\begin{eqnarray*}
K_c(x,x') &=&\scp{h(x) - \bar h}{h(x')-\bar h}_H\\
&=&\scp{h(x)}{h(x')}_H - \scp{h(x)+h(x')}{\bar h} + \norm{\bar h}_H^2\\
&=& K(x_k, x_l) -\frac{1}{N} \sum_{k=1}^N (K(x,x_k)+K(x',x_k)) +  \frac{1}{N^2}\sum_{k,l=1}^N K(x_k, x_l).
\end{eqnarray*}
Then the Gram matrix in feature space is  $S$ with $s_{kl} = K_c(x_k, x_l)$ and the computation described in the previous section can be applied. Note that, if one denotes, as usual $\CK = \CK(x_1, \ldots, x_N)$ the matrix formed by kernel evaluations $K(x_k, x_l)$, and if one lets $P = \Id[N] - \dsone_N\dsone_N/N$, then we have the simple matrix expression $S = P\CK P$.

Letting $\pe \al 1, \ldots, \pe \al p \in \mR^N$ be the first $p$ eigenvectors of $S$, normalized so that $(\pe \al i)^TS\pe \al i = 1$,
the principal
directions are vectors in feature space given by (using the notation in the previous section in which the $k$th coordinate of $\pe \alpha i$ is $\alpha_k^{(i)}$)
$$
f_i = \sum_{k=1}^N \al_{k}^{(i)} (h(x_k) - \bar h)\,,
$$
and they are not computable when the features not known explicitly. However, a few geometric features associated with these directions can be characterized using the kernel only.

 Consider the line in
feature space $D_i = \defset{\bar h + \la f_i, \la\in \mR}$. Let $\Om_i$ denote the points $x\in \CR$ such that $h(x) \in D_i$. Then $x\in \Om_i$ if and only if  $h(x)$  coincides with its orthogonal projection on $D_i$, which
is equivalent to
$$
\scp{h(x)-\bar h}{f_i}_H^2 = \norm{h(x) - \bar h}_H^2,
$$
which can be expressed with the kernel as
\begin{equation}
\label{eq:kernel.component}
K_c(x,x) - \left(\sum_{k=1}^N \pe{\al_{k}}i K_c(x,x_k)\right)^2 = 0\,.
\end{equation}
This provides a  nonlinear equation in $x$. In particular, $\Om_i$ is generally nonlinear, possibly with several connected components. Note that, by definition, the difference in \cref{eq:kernel.component} is always non-negative, so that a way to visualize $\Omega_i$ is to compute its sub-level sets, i.e., the set of all $x$ such that
\[
K_c(x,x) - \left(\sum_{k=1}^N \pe{\al_{k}}i K_c(x,x_k)\right)^2 \leq \epsilon
\]
for small $\epsilon$.

Similarly, the feature vector $h(x) - \bar h$ belongs to the space generated by the first $p$ components if and only if
\[
\sum_{i=1}^p \scp{h(x)-\bar h}{f_i}_H^2 = \norm{h(x) - \bar h}_H^2 
\]
i.e., 
$$
\sum_{i=1}^p \left(\sum_{k=1}^N \al_{k}^{(i)} K_c(x,x_k)\right)^2 = K_c(x,x).
$$

\bigskip

One can also compute the finite-dimensional coordinates
of $h(x)$ in the PCA basis, and this computation is easier. The representation is
\[
x \mapsto (u_1(x), \ldots, u_p(x))
\]
with 
\[
u_i = \scp{h(x) - \bar h}{f_i}_H =  \sum_{k=1}^N \al_{k}^{(i)} K_c(x,
x_k) \,.
\]
This provides an explicit nonlinear transformation that maps each data point $x$ into a $p$-dimensional point.
This representation allows one to easily exploit the reduction of dimension.

\section{Statistical interpretation and probabilistic PCA}

There is a simple probabilistic interpretation of linear PCA. Assume that $H=\mR^d$ with the standard inner product and 
that $X$ is a centered  random vector with covariance
matrix $\Sig$. Consider the problem that consists in finding a
factor decomposition 
\[
X = \sum_{i=1}^p Y^{(i)} e_i + R
\]
 where $Y = (Y^{(1)}, \ldots, Y^{(p)})^T$ forms
a $p$-dimensional centered  vector, $e_1, \ldots,
e_p$ is an orthonormal system, and $R$ is a random vector,
independent of $Y$ and as small as possible, in the
sense that $E(|R|^2)$ is minimal. 
One can see that, in an optimal
decomposition, one needs $R^T{e_i} = 0$ for all $i$, because one
can always write
\[
\sum_{i=1}^p Y^{(i)} e_i + R = \sum_{i=1}^p (Y^{(i)} + {R^T}{e_i}) e_i +
R - \sum_{i=1}^p {R^T}{e_i} e_i\,.
\]
If $R$ is centered, then so is $R - \sum_{i=1}^p {R^T}{e_i} e_i$ 
and the latter provides a better solution since $|R - \sum_{i=1}^p {R^T}{e_i} e_i| \leq |R|$. Also, there is no loss of generality in requiring that $(Y^{(1)}, \ldots, Y^{(p)})$ are uncorrelated, as this can always be obtained after  a change of
basis in $\text{span}(e_1, \ldots, e_p)$. 

Assuming this, we can write
$$
E(|X|^2) = \sum_{i=1}^p E((Y^{(i)})^2) + E(|R|^2)
$$
with $Y^{(i)} = {e_i^T}{X}$. So, to minimize $E(|R|^2)$, one needs to
maximize 
$$\sum_{i=1}^p E(({e_i^T}{X})^2)$$
 which is equal to (letting $\Sig$ be the covariance matrix of $X$)
$$
\sum_{i=1}^p e_i^T \Sig e_i.
$$
The solution of this problem is given by the first $p$
eigenvectors of $\Sig$. PCA (with a Euclidean metric) exactly applies this procedure, with
$\Sig$ replaced by the empirical covariance. 

``Probabilistic PCA'' is based on a slightly different statistical model in which it is
assumed that $X$ can be decomposed as 
\[
X = \sum_{i=1}^p \la_i Y^{(i)} e_i + \sig R,
\]
 where $R$ is a $d$ dimensional standard
Gaussian vector and $Y = (Y^{(1)}, \ldots, Y^{(p)})^T$ a $p$-dimensional
standard Gaussian vector, independent of $R$. The main difference with standard PCA is that the total variance of the residual, here $d\sig^2$, is a model parameter and not a quantity to minimize. 

In addition to $\sig^2$, the model is parametrized by the coordinates of $e_1,
\ldots, e_p$ and the values of $\la_1, \ldots, \la_p$. Introduce the $d\ti p$ matrix 
\[
W = [\la_1 e_1, \ldots, \la_p e_p].
\]
We can  rewrite this
model in the form
$$X = WY + \sig^2 R$$
where the parameters are $W$ and $\sig^2$, with the constraint that
$W^TW$ is a diagonal matrix. As a linear combination of independent Gaussian
random variables, $X$ is Gaussian with covariance matrix $WW^T +
\sig^2 \Id$. The log-likelihood of the observations $x_1, \ldots, x_N$ therefore is
\begin{equation}
\label{eq:p.pca.lik}
L(W, \sig) = -  \frac{N}{2} \Big(d \log 2\pi + \log \det(WW^T +
\sig^2 \Id) + \trace((WW^T + \sig^2 \Id)^{-1}\Sig_T)\Big)
\end{equation}
where $\Sig_Y$ is the empirical covariance matrix of $x_1, \ldots,
x_N$. This function can be maximized explicitly in $W$ and $\sig$, as stated in the following proposition. 
\begin{proposition}
\label{prop:p.pca}
Assume that the matrix $\Sig_T$ is invertible.
The log-likelihood in \cref{eq:p.pca.lik} is maximized by taking 
\begin{enumerate}[label = (\roman*)]
\item $W = [\la_1e_1, \ldots, \la_pe_p]$ where $e_1, \ldots,
e_p$ are the eigenvectors of $\Sig_T$  associated to the $p$ largest eigenvalues, and  $\la_i = \sqrt{\de_i^2
- \sig^2}$, where $\de_i^2$ is the eigenvalue of $\Sig$ associated to
$e_i$; 
\item and 
$$
\sig^2 = \frac{1}{d-p} \sum_{i=p+1}^d \de_i^2.
$$
\end{enumerate}
\end{proposition}
\begin{proof}
We make the following change of variables: let $\rho^2 = 1/\sig^2$ and 
\[
\mu_i^2 = \frac{1}{\sig^2} - \frac{1}{\la_i^2 + \sig^2}.
\]
Let $Q = [\mu_1e_1, \ldots, \mu_pe_p]$. We have
\[
(WW^T + \sig^2\Id)^{-1} = \rho^2\Id - QQ^T.
\]
To see this, complete $(e_1, \ldots, e_p)$ into an orthonormal basis of $\mR^d$, letting $e_{p+1}, \ldots, e_d$ denote the added vectors. Then
\[
WW^T + \sig^2\Id = \sum_{i=1}^p (\la_i^2 + \sig^2) e_ie_i^T + \sum_{i=p+1}^d \sig^2 e_ie_i^T
\]
so that
\[
(WW^T + \sig^2\Id)^{-1} = \sum_{i=1}^p (\la_i^2 + \sig^2)^{-1} e_ie_i^T + \sum_{i=p+1}^d \sig^{-2} e_ie_i^T = \rho^2\Id - QQ^T.
\]
Using the new  variables, we can reformulate the problem as the minimization of
\[
-\sum_{i=1}^p \log(\rho^2 - \mu_i^2) - (d-p)\log\rho^2 + \rho^2\trace(\Sig) - \sum_{j=1}^p \mu^2_j e_j^T\Sig e_j.
\]
From \cref{th:pca.base}, we have
\[
\sum_{j=1}^p \mu^2_j e_j^T\Sig e_j \leq \sum_{j=1}^p \mu_j^2\de_j^2
\]
and this upper bound is attained by letting $e_1, \ldots, e_p$ be the first $p$ eigenvectors of $\Sig$. 
Using this, we see that $\sig^2, \mu_1^2, \ldots, \mu_p^2$ must minimize
\[
-\sum_{i=1}^p \log(\rho^2 - \mu_i^2) - (d-p)\log\rho^2 + \rho^2\sum_{j=1}^d \de_j^2 - \sum_{j=1}^p \mu_j^2\de_j^2\,.
\]
Computing the solution is elementary and left to the reader. It yields, when expressed as functions of $\sig^2, \la_1^2, \ldots, \la_p^2$, the expressions given in the statement of the theorem.
\end{proof}

\section{Generalized PCA}

\begin{figure}[h]
\begin{center}
\includegraphics[width = 0.45\textwidth]{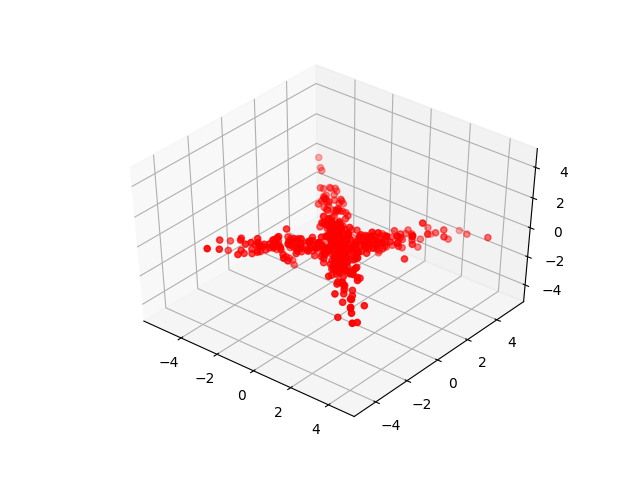}
\includegraphics[width = 0.45\textwidth]{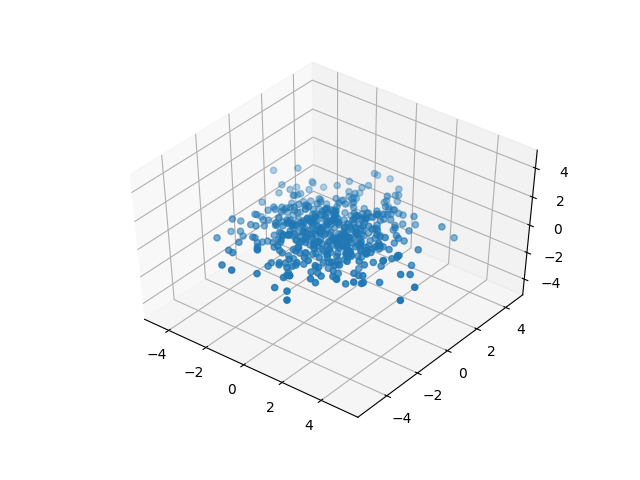}
\end{center}
\caption{\label{fig:gpca} PCA cannot distinguish between the situations depicted in the two datasets.}
\end{figure}

We now discuss a dimension reduction method called generalized PCA (GPCA) \citep{vidal2005generalized} which,
instead of looking for the best linear approximation of the training
set by one specific subspace,  provides an
approximation by a finite union of such spaces. 
As a motivation, consider the situation in \cref{fig:gpca} in which  part of the data is
aligned along one direction in space, and another part along another
direction. Then, the only information that PCA can retrieve (provided that
the two directions intersect) is the plane generated by the two
directions, which will be captured by the two principal
components. PCA will not be able to determine the individual directions. GPCA addresses this type of situation as follows.

For simplicity, assume that we want to decompose the data  along unions of  hyperplanes in $\mR^d$.  Such
hyperplanes have equations of the form
$u^T\tilde x = 0$ where $\tilde x$ is our notation for the vector $(1,
x^T)^T$. If we have two hyperplanes, specified by $u_1$ and $u_2$ and all the
training samples approximately  belong to one of them, then one has, for all $k=1, \ldots, N$:
$$
(u^T_1\tx_k)(u^T_2\tx_k) = \tx_k^T u_1u_2^T \tx_k \simeq 0.
$$
Similarly, for $n$ hyperplanes, the identity is, for $k=1, \ldots, N$:
$$
\prod_{j=1}^n (u^T_j \tx_k) \simeq 0.
$$

Define
\[
F(x) = 
\prod_{j=1}^n (u^T_j x) = \sum_{1\leq i_1, \ldots,  i_n \leq
d} u_1(i_1) \cdots u_n(i_n) x^{(i_1)}\cdots x^{(i_n)}.
\]
Regrouping the terms associated with the same powers of $x$, one can express $F$ in the form
\begin{equation}
\label{eq:F}
F(x) = \sum_{p_1 + \ldots + p_d = n} q_{p_1\ldots p_d}\,  (x^{(1)})^{p_1} \ldots
(x^{(d)})^{p_d}\,.
\end{equation}
The collection of $\binom{n+d-1}{n}$ numbers $ Q = (q_{p_1\ldots p_n}, p_1 + \cdots + p_d = n)$ takes a specific form (that we will not need to make explicit) as a function of the unknown $u_1, \ldots, u_n$, but the first step of GPCA ignores this constraint and estimates $Q$ minimizing 
$$
\sum_{k=1}^N \left(\sum_{p_1 + \ldots + p_d = n} q_{p_1\ldots p_d} \,  (x_k^{(1)})^{p_1} \ldots
(x_k^{(d)})^{p_d}\right)^2
$$
under the constraint $\sum q_{p_1 \ldots p_n}^2 = 1$ (to avoid trivial solutions). Choosing an ordering on the set of indices $(p_1, \ldots,p_d)$ such that $p_1+ \cdots +p_d=n$, one can stack the coefficients in $Q$ and the monomials $ (x_k^{(1)})^{p_1} \ldots
(x_k^{(d)})^{p_d}$ to form two vectors denoted $Q$ (with some abuse of notation) and $V(x_k)$. One can then rewrite the problem of determining $Q$ as minimizing $Q^T\Sig Q$ subject to $|Q|^2 = 1$, where 
\[
\Sig = \sum_{k=1}^N V(x_k)V(x_k)^T.
\]
The solution is given by the eigenvector associated with the smallest eigenvalue of $\Sig$. If the model is exact, this eigenvalue should be zero, and if only one decomposition of the data in a set of distinct hyperplanes exists (i.e., if $n$ is not chosen too large), then $Q$ is the unique solution up to a multiplicative constant. 

Once $Q$ is found, it remains to identify the vectors $u_1, \ldots, u_n$. This identification can be obtained by inspecting the gradient of $F$ on the union of hyperplanes. Indeed, one has, for $x\in \mR^d$,
\[
\nabla F(x) = \sum_{j=1}^n\left(\prod_{j'\neq j} u_{j'}^Tx\right) u_j.
\]
However, if $x$ belong in one and only one of the hyperplanes, say $x^Tu_j = 0$, then all terms in the sum vanish but one and $\nabla F(x)$ is proportional to $u_j$. So, if the model is exact, one has, for each $k=1, \ldots, N$, either $\nabla F(x_k) = 0$ (if $x_k$ belongs to the intersection of two hyperplanes) or $\nabla F(x_k)/|\nabla F(x_k)| = \pm u_j$ for some $j$, and the sign ambiguity can be removed by ensuring, for example, that the first non-vanishing coordinate of  $u_j$ is positive. (The gradient of $F$ can be computed from $Q$ using \cref{eq:F}.) The computation of $\nabla F$ on training data therefore allows for an exact computation of the hyperplanes.

In practice, when noise is present, one cannot expect this computation to be exact.
The vectors $u_1, \ldots, u_n$ can be estimated by clustering the collection of non-vanishing gradients $\nabla F(x_k)$, $k=1, \ldots, N$. For example, one can compute a dissimilarity matrix such as  
 $d_{kl} = 1 - \cos^2(\th_{kl})$, where
$\th_{kl}$ is the angle between $\nabla F(x_k)$ and $\nabla F(x_l)$, and apply one of the methods discussed in   \cref{sec:spectral.clus}.

This analysis provides a decomposition of the training set into $n$
(or fewer) hyperplanes. The computation can then
be recursively refined in order to obtain smaller dimensional subspaces by applying the same method separately to each
hyperplane.

\section{Nuclear norm minimization and robust PCA}
\label{sec:rpca}

\subsection{Low-rank approximation}
One can also interpret PCA in terms of low-rank matrix approximations. Let $\CX_c$ be the $N$ by $d$ matrix $(x_1- \bx, \ldots, x_N-\bx)^T$, which, in generic situations, has rank $d-1$. Let $Q \in \CS^{++}_d$ be a positive definite matrix of size $d$and define, for an $(N, d)$ matrix $A$, the norm
\[
|A|_Q^2 = \trace(AQA^T).
\]

We show that PCA in $H= \mR^d$ with inner product $\scp{x}{y}_H = x^TQy$ is equivalent to a low-rank approximation of $\CX_c$ according to  $|\cdot|_Q$, i.e., to the minimization,  over all $N$ by $d$ matrices $\CZ$ of rank $p$, of the norm of the difference
\begin{equation}
\label{eq:low.rank.pca}
|\CX_c - \CZ|_Q^2 = \trace((\CX_c-\CZ)Q(\CX_c-\CZ)^T)\,.
\end{equation}
To justify this, we can use the results of \cref{sec:low.rank} to approximate $\tilde{\CX}_c = \CX_c Q^{1/2}$ by some matrix $\tilde {\CZ}$ and then let $\CZ = \tilde\CZ Q^{-1/2}$. Using \cref{prop:rank.p.approx}, we see that one can take 
\[
\CZ = \CX_c Q^{1/2} V_{\lceil d, p\rceil} V_{\lceil d, p\rceil}^TQ^{-1/2}
= \CX_c Q WW^T
\]
with $W =  Q^{-1/2}V_{\lceil d, p\rceil}$ and $\CX_cQ^{1/2} =  UDV^T$ is a singular-values decomposition of $\tilde \CX_c$. Noting that $\CX_c^T\CX_c = N\Sigma_T$, the first $p$ columns of $W$ are the $p$ principal components of $T$, as derived in  \cref{sec:pca.comp}. Moreover, letting $W = [f_1, \ldots, f_p]$, one sees that the $k$th row of $\CZ$ is given by
\[
\sum_{j=1}^p \scp{x_k - \bx}{f_j}_H f_j
\]
which is the PCA approximation of  $x_k - \bx$.

In the rest of this discussion, we restrict to the Euclidean case, in which $Q = \Id[d]$ and we simply write $|A|$ rather than $|A|_Q$.
In the next section, we will explore variations on PCA in which the minimization of  $|\CX_c - \CZ|^2$ is completed with a penalty that depends on the singular values of the matrix $\CZ$. As a first example, 
one can modify PCA by adding  a penalty on the rank (i.e., on the number of non-zero singular values), minimizing:
\[
{\ga} |\CX_c - \CZ|^2+  \rank(\CZ)
\]
for some parameter $\ga>0$. However, the solution to this problem is a small variation of that of standard PCA. It is indeed given  by standard PCA with $p$ components where $p$ minimizes 
\[
N\ga \sum_{k=p+1}^d \la_k^2 +  p = N\ga \sum_{k=p+1}^d (\la_k^2 - (N\ga)^{-1}) + d,
\]
i.e.,  $p$ is the index of the last eigenvalue that is larger than $(N\ga)^{-1}$.

\subsection{The nuclear norm}
Based on the fact that $\rank(\CZ)$ is the number of non-zero singular values of $\CZ$, one can use the same heuristics as in the development of the lasso, and replace the number of non-zero values by the sum of the absolute values of the singular values, which is just the sum of singular values since they  are non-negative. This provides the nuclear norm of $A$, defined in \cref{sec:matrix.norms} by
\[
|A|_* = \sum_{k=1}^d \sig_k
\]
where $\sigma_1, \ldots, \sigma_d$ are the singular values of $A$.  We will consider below the problem of minimizing
\begin{equation}
\label{eq:robust.pca.2}
{\ga} |\CX_c - \CZ|^2+  |\CZ|_*
\end{equation}
and show that its solution is once again similar to PCA.

We recall the characterization of the nuclear norm from \cref{prop:nuclear}. If $A$ is an  $N$ by $d$ matrix,
\[
|A|_* = \max\Big\{\trace(UAV^T): U \text{ is }N\times N \text{ and } U^TU = \Id,  V \text{ is } d\times d \text{ and } V^TV = \Id\Big\}\,.
\]

In \citet{cai2010singular}, the authors consider the minimization of \cref{eq:robust.pca.2} 
and prove the following result. Recall that we have defined the shrinkage function 
\[
S_\tau: t \mapsto \sign(t)\max(|t|-\tau, 0)
\]
 (with $\tau \geq 0$), using the same notation $S_\tau(X)$ when applying $S_\tau$ to every entry of a vector or matrix $X$.   Following \citet{cai2010singular}, we define the {\em singular value thresholding operator} $A \mapsto \mS_\tau(A)$, where $A$ is any rectangular matrix, by
\[
\mS_\tau(A) = US_\tau(\Delta) V^T
\]
when $A = U\Delta V^T$ is a singular value decomposition of $A$.
\begin{proposition}
\label{prop:sh.svd}
Let us assume without loss of generality that $N\geq d$. 
The function $\CZ \mapsto {\ga} |\CX_c - \CZ|^2+  |\CZ|_*$ is minimized by $\CZ = \mS_{1/2\ga}(\CX)$.
\end{proposition}
\begin{proof}
Representing $\CZ$ by its singular value decomposition, we have the equivalent formulation of minimizing 
\begin{align*}
F(U, V, D) &= \ga|\CX_c - UDV^T|^2 +  |D|_*\\
&= \ga|\CX_c|^2 - 2\ga\trace(\CX_c^TUDV^T) + \ga|D|^2 + |D|_*
\end{align*}
over all orthonormal matrices $U$ and $V$ and diagonal matrices with non-negative coefficients $D$. From \cref{th:trace.ineq}, we know that $\trace(\CX_c^TUDV^T)$ is less than the sum of the products of the non-increasingly ordered singular values of $\CX_c$ and $D$ and this upper bound is attained by taking $U= \bar U$ and $V=\bar V$ where $\bar U$ and $\bar V$ are  the matrices providing the SVD of $\CX_c$, i.e., such that $\CX_c = \bar U\Delta \bar V^T$ where $\Delta$ is diagonal with non-decreasing coefficients along the diagonal. So, letting $\la_1 \geq \cdots \geq \la_d \geq 0$ and $\mu_1 \geq \cdots \geq \mu_d\geq 0$ be the singular values of $\CX_c$ and $\CZ$, we have just proved that, for any $D$,
\[
F(U, V, D) \geq  F(\bar U, \bar V, D) = -2\ga \sum_{i=1}^d \mu_i\la_i + \ga\sum_{i=1}^d \mu_i^2 + \sum_{i=1}^d\mu_i\,.
\]
The lower bound is minimized when  $\mu_i = \max(\la_i - 1/2\ga, 0)$. This proves the proposition.
\end{proof}

\subsection{Robust PCA}
As a consequence, the nuclear norm penalty provides the same principal directions (after replacing $\ga$ by $2\ga$) as the rank penalty, but applies a shrinking operation rather than thresholding on the singular values. The difference is however more fundamental if, in addition to using the nuclear norm as a penalty, on replaces the squared Frobenius norm on the approximation error by the $\ell^1$ norm, where, for an $n$ by $m$ matrix $A$ with coefficients $(a(i,j))$,
\[
|A|_{\ell_1} = \sum_{i,j} |a(i,j)|\,.
\]
This is the formulation of  robust PCA \citep{candes2011robust}, which minimizes
\begin{equation}
\label{eq:r.pca}
{\ga} |\CX_c - \CZ|_{\ell^1}+  |\CZ|_*
\end{equation}
with respect to $\CZ$.

Robust PCA (which was initially named Principal Component Pursuit by the authors in \citet{candes2011robust}) is designed for situations in which $\CX_c$ can be decomposed as the sum of a low-rank matrix $\CZ$ and of a sparse residual $S$. Some theoretical justification was provided in the original paper, stating that if $\CX_c = \CZ + S$, with $\CZ = UDV^T$ (its singular value decomposition) such that $U$ and $V$ are sufficiently ``diffuse'' and $\rank(\CZ)$ is small enough, with the residual's sparsity pattern taken uniformly at random over the subsets of entries of $S$ with a sufficiently  small cardinality, then robust PCA is able to reconstruct the decomposition exactly with high probability (relative to the random selection of the sparsity pattern of $S$). We refer to \citet{candes2011robust} for the long proof that justifies this statement.

Robust PCA can be solved using the ADMM algorithm (\cref{sec:proximal}) after reformulating the problem as the minimization of 
\[
\ga |R|_{\ell_1} +  |\CZ|_*
\]
subject to $R + \CZ = \CX_c$.  The algorithm therefore iterates over the following steps.
\begin{equation}
\label{eq:pcp}
\left\{
\begin{aligned}
\CZ^{(k+1)} &=  \argmin_\CZ \Big( |\CZ|_* + \frac{1}{2\al} |\CZ + R^{(k)} - \CX_x + U^{(k)}|^2\Big)\\
R^{(k+1)} &= \argmin_R \Big(\ga |R|_{\ell^1} + \frac1{2\al} |\CZ^{(k+1)} + R - \CX_c + U^{(k)}|^2\Big) \\
U^{(k+1)} &= U^{(k)} + \CZ^{(k+1)} + R^{(k+1)} - \CX_c
\end{aligned}
\right.
\end{equation} 
The first minimization is covered by  \cref{prop:nuclear} and yields
\[
\CZ^{(k+1)} = \mS_\al(\CX_c - R^{(k)} - U^{(k)}).
\]
The second minimization is solved by a standard shrinking operation, i.e., 
\[
R^{(k+1)} = S_{\ga\al} (\CX_c - \CZ^{(k+1)} - U^{(k)}). 
\]
Using this, we can rewrite the robust PCA algorithm as the sequence of fairly simple iterations.
\begin{algorithm}
\label{alg:robust.pca}
\begin{enumerate}
\item Choose a small enough constant $\alpha$ and a very small tolerance level $\epsilon$.
\item Initialize the algorithm with $N$ by $d$ matrices $R^{(0)}$ and $U^{(0)}$ (e.g., equal to zero).
\item At step $n$, apply the iteration: 
\begin{equation}
\label{eq:pcp.2}
\left\{
\begin{aligned}
\CZ^{(k+1)} &= \mS_\al(\CX_c - R^{(k)} - U^{(k)}) \\
R^{(k+1)} &=S_{\ga\al} (\CX_c - \CZ^{(k+1)} - U^{(k)}) \\
U^{(k+1)} &= U^{(k)} + \CZ^{(k+1)} + R^{(k+1)} - \CX_c
\end{aligned}
\right.
\end{equation} 
\item Stop the algorithm is the variation compared to variables at the previous step is below the tolerance level. Otherwise, apply step $n+1$.
\end{enumerate}
\end{algorithm}

\section{Independent component analysis}

Independent component analysis (ICA) is a factor analysis method that represents  a $d$-dimensional random variable $X$  in the form
$X = A Y$ where $A$ is a fixed $d\ti d$ invertible matrix and
$Y$ is a $d$-dimensional random vector with independent components. 
There are two main approaches in this setting. The first one optimizes the matrix $W = A^{-1}$ so that  the components of $WX$ are ``as independent as possible'' according to a suitable criterion. The second one is  model-based, where a statistical model is assumed for $Y$, and its parameters, together with the entries of the matrix $A$, are estimated via maximum likelihood. Before describing each of these methods, we first discuss the extent to which the coefficients of $A$ are identifiable. 

\subsection{Identifiability}

A statistical model is identifiable if its parameters (which could be finite- of infinite-dimensional) are uniquely defined by the distribution of the observable variables. In the case of ICA, this question boils down to deciding whether $AY \sim A'Y'$ (i.e., they have the same probability distribution) implies that $A = A'$ (where $Y$ and $Y'$ are two random vectors with independent components).

It should be clear that the answer to this question is negative, because there are trivial transformations of the matrix $A$ that do not break the ICA model. One can, for example, take any invertible diagonal matrix, $D$, and let $A' = AD^{-1}$ and $Y' = DY$. The same statement can be made if $D$ is replaced by a permutation matrix, $P$, which reorders the components of $Y$. So we know that $AY \sim A'Y'$ is possible already when $A' = ADP$ where $D$ is diagonal and invertible and $P$ is a permutation matrix. Note that iterating such matrices (i.e., letting $A' = ADPD'P'$) does not extend the class of transformations because one  has $DP = PP^{-1}DP$ and one can easily check that  $P^{-1}DP$ is diagonal, so that one can rewrite any product of permutations and diagonal matrices as a single diagonal matrix multiplied by a single permutation. 

It is interesting, and fundamental for the well-posedness of ICA, that, under one  important additional assumption, the indeterminacy in the identification of $A$ stops at these transformations. The additional assumption is that at most one of the components of $Y$ follows a Gaussian distribution. That such a restriction is needed is clear from the fact that one can transform any Gaussian vector $Y$ with independent components into another, $BY$, one as soon as $BB^T$ is diagonal. If two or more components of $Y$ are Gaussian, one can restrict these matrices $B$ to only affect those components. If only one of them is Gaussian, such an operation has no effect. 

The following theorem is formally stated in \citet{comon1994independent}, and is a rephrasing of the Darmois-Skitovitch theorem \citep{darmois1953analyse,skitovich1954linear}. The proof of this theorem relies on complex analysis arguments on characteristic functions and is beyond the scope of these notes (see \citet{kagan1973characterization} for more details).

\begin{theorem}
\label{th:ica.charac}
Assume that $Y$ is a random vector with independent components, such that at most one of its components is Gaussian. Let $A$ be an invertible linear transformation and $\tilde Y = CY$. Then the following statements are independent. 
\begin{enumerate}[label=(\roman*)]
\item For all $i\neq j$, the components $\pe{\tilde Y} i , \tilde Y^{(j)}$ are independent.
\item $\tilde Y^{(1)}, \ldots, \tilde Y^{(d)}$ are mutually independent. 
\item $C = DP$ is the product on a diagonal matrix and of a permutation.
\end{enumerate}
\end{theorem}

The equivalence of (ii) and (iii) implies that the ICA model is identifiable up to multiplication on the right by a permutation and a diagonal matrix. Indeed, if $X = AY = A'Y'$ are two decompositions, then it suffices to apply the theorem  to $C = (A')^{-1} A$ to conclude. The equivalence of (i) and (ii) is striking, and has the important consequence that, if the data satisfies the ICA model, then, in order to identify $A$ (up to the listed indeterminacy), it suffices to look for $Y = A^{-1}X$ with pairwise independent components, which is a much lesser constraint than full mutual independence. 

As a final remark on the Gaussian indeterminacy, we point out that, if the mean ($m$) and covariance matrix ($\Sig$) of $X$ are known (or estimated from data), the ICA problem can be reduced to looking for orthogonal transformations $A$. Indeed, assuming $X = AY$ and letting $\tX = \Sig^{-1/2} (X - m)$ and $\tY = D^{-1/2}(Y - A^{-1} m)$, where $D$ is the (diagonal) covariance matrix of $Y$, we have
\[
\tX = \Sig^{-1/2}(AY - m) = \Sig^{-1/2}A D^{1/2} \tY.
\]
Letting $\tilde A = \Sig^{-1/2}A D^{1/2}$, we have $\Id_{\mR^d} = E(\tX \tX^T) = \tilde A \tilde A^T$ so that $\tilde A$ is orthogonal. This shows that  the ICA problem for $\tX$ in the form $\tX = \tilde A \tY$ with the restriction that $\tilde A$ is orthogonal has a solution, and also provides a solution of the original ICA problem by letting $A = \Sig^{1/2} \tilde A$ and $Y = \tY - \tilde A^{-1} \Sig^{-1/2} m$. Therefore, the indeterminacy associated with Gaussian vectors is as general as possible up to a normalization of first and second moments.

\subsection{Measuring independence and non-Gaussianity}

Independence between $d$ variables is a very strong property and its complete characterization is computationally challenging. The fact that the joint p.d.f. of the $d$ variables  (we will restrict, to simplify our discussion, to variables that are absolutely continuous) factorizes into the product of the marginal p.d.f.'s of each variable can be measured by computing the mutual information between the variables, defined by (letting $\phi_Z$ denote the p.d.f. of a variable $Z$)
\[
I(Y) = \int \log \frac{\phi_Y(y)}{\prod_{i=1}^d \phi_{\pe Y i}(\pe y i)} \phi_Y(y) dy.
\]
The mutual information is always non-negative and vanishes only if the components of $Y$ are mutually independent.
Therefore, one can represent ICA as an optimization problem minimizing $I(WX)$ with respect to all invertible matrices $W$ (so that $W = A^{-1}$). Letting
\[
h(Y) = - \int \log \phi_Y(y)\, \phi_Y(y) dy
\]
denote the ``differential entropy'' of $Y$, we can write
\[
I(Y) = \sum_{i=1}^d h(\pe Y i) - h(Y).
\]

If $Z = WX$, then $\phi_Z(z) = \phi_X(W^{-1}x) |\det(W)|^{-1}$. Using this expression in $h(Z)$ and making a change of variables yields $h(Z) = h(X) + \log |\det W|$ and 
\[
I(WX) = \sum_{i=1}^d h(\pe Z i) - \log |\det(W)| - h(X).
\]
This shows that the optimal $W$ can be obtained by minimizing
\[
F(W) =  \sum_{i=1}^d h(W^{(i)}X) - \log |\det(W)|
\]
where $W^{(i)}$ is the $i$th row of $W$. This brings a notable simplification, since this expression only involves differential entropies of scalar variables, but still remains a  challenging problem. 

In \citet{comon1994independent}, it is proposed to use cumulant expansions of the entropy around that of a Gaussian with identical mean and variance to approximate the differential entropy. If $\xi \sim \CN(m, \sig^2)$ , then
\[
h(\xi) = \frac12 + \frac12 \log(2\pi\sig^2).
\]
Define, for a general random variable $U$ with standard deviation $\sigma_U$, the non-Gaussian entropy, or negentropy, defined by 
\[
\nu(U) = \frac12 + \frac12 \log(2\pi\sig_U^2) - h(U)\,.
\]
One can shows that $\nu(U) \geq 0$ and is equal to 0 if and only if $U$ is Gaussian. One can rewrite $F(W)$ as 
\[
F(W) =  \frac d2 + \frac d2 \log(2\pi) + \sum_{i=1}^d \log(\sig_{W^{(i)}X}^2) - \sum_{i=1}^d \nu(W^{(i)}X) - \log |\det(W)|
\]

As we remarked earlier, if we replace $X$ by $\Sig^{-1/2} (X-m)$ (after estimating the covariance matrix of $X$), there is no loss of generality in requiring that $W$ is an orthogonal matrix, in which case both $\sig_{W^{(i)}X}^2$ and $|\det W|$ are equal to 1. Assuming such a reduction is done, we see that the problem now requires to maximize
\begin{equation}
\label{eq:negentropy}
\sum_{i=1}^d \nu(W^{(i)}X)
\end{equation}
among all orthogonal matrices $W$. Still in \citet{comon1994independent}, an approximation of the negentropy $\nu(U)$ is provided as a function of the third and fourth cumulants of the distribution of $U$. These are given by
\[
\ka_3 = E((U-E(U))^3)
\]
and
\[
\ka_4 = E((U-E(U))^4) - 3\sig_U^4.
\]
In particular, when $U$ is normalized, i.e.,  $E(U) = 0$ and $\sig^2_U = 1$, we have $\ka_3 = E(U^3)$ and $\ka_4 = E(U^4) - 3$. The approximation is then
\[
\nu(U) \simeq \frac{\ka_3^2}{12} + \frac{\ka_4^2}{48} + \frac{7\ka_3^4}{48} - \frac{\ka_3^2\ka_4}{8}\,.
\]
This approximation was derived from an Edgeworth expansion (which can be thought of as a Taylor expansion around a Gaussian distribution) of the p.d.f. of $U$. Plugging this expression into \cref{eq:negentropy} provides an expression that can be maximized in $W$ where the cumulants are replaced by their sample estimates. However, the maximized function involves high-degree polynomials in the unknown coefficients of $W$, and this simplified problem still presents numerical challenges. 

\bigskip

An alternative approximation of the negentropy has been proposed in \citet{hyvarinen1998new} relying on the maximum entropy principle, described in the following theorem. Associate to any random variable $Y:\CG \to \mR$ the differential entropy
\[
h_\mu(Y) = - \int_{\CG} \log \phi_Y(x) \phi_Y(x) d\mu(x)
\]
if the distribution of $Y$ has a density, denoted $\phi_Y$, with respect to $\mu$ and $h_\mu(Y) = -\infty$ otherwise.
Use also the same notation
\[
h_\mu(\phi) = - \int_{\CG} \log \phi(x) \phi(x) d\mu(x)
\]
for a p.d.f. $\phi$ with respect to $\mu$ (i.e., such that $\phi$ is non-negative and has integral 1). Then, the following is true.
\begin{theorem}
\label{th:max.ent.principle}
Let $\bfg = (\pe g 1, \ldots, \pe g p)^T$ be a function defined on a measurable space $\CG$, taking values in $\mR^p$, and let $\mu$ be a measure on $\CG$. Let $\Gamma_\mu$ be the set of all   
$\boldsymbol\lambda = (\pe \lambda 1, \ldots, \pe \lambda p)\in \mR^p$ such that 
\begin{equation}
\label{eq:max.ent.principle.2}
\int_\CG \exp\left(\boldsymbol\la^T \bfg(y)\right) d\mu(y) < \infty.
\end{equation}
Then 
\begin{equation}
\label{eq:max.ent.principle.0}
h_\mu(Y) \leq \mathrm{inf}\left\{- \boldsymbol\la^T E(\bfg(Y)) + \log \int_\CG \exp\left(\boldsymbol\la^T \bfg(y)\right) d\mu(y): \boldsymbol\lambda \in \Gamma_\mu \right\}.
\end{equation}

Define, for $\boldsymbol\lambda\in \Gamma_\mu$, 
\begin{equation}
\label{eq:max.ent.exp}
\psi_{\boldsymbol \lambda}(x) = \frac{\exp\left(\boldsymbol\la^T \bfg(x)\right) d\mu(x)}{\int_\CG \exp\left(\boldsymbol\la^T \bfg(y)\right) d\mu(y)}.
\end{equation}

Assume that the infimum in \cref{eq:max.ent.principle.0} is attained at an interior point $\boldsymbol \lambda^*$ of $\Gamma_\mu$. Then 
\begin{equation}
\label{eq:max.ent.principle.3}
h(\phi_{\boldsymbol\lambda^*}) = \max\{h(\tilde Y): E_{\tilde Y}(\bfg) = E_Y(\bfg), i=1, \ldots, p\}.
\end{equation}
\end{theorem}

\begin{proof}
Assume that $Y$ has a p.d.f. $\phi_Y$ with respect to $\mu$ (otherwise the lower bound in \cref{eq:max.ent.principle.0} is $-\infty$). 
Then
\[
h_\mu(Y) + \boldsymbol\la E(\bfg(Y)) - \log \int \exp\left(\boldsymbol\la \bfg(y)\right) d\mu(y) = - \int_\CG \phi_Y(x) \log \frac{\phi_Y(x)}{\psi_{\boldsymbol\lambda}(x)} d\mu(x) \leq 0
\]
since $\int_\CG \phi_Y(x) \log \frac{\phi_Y(x)}{\psi_{\boldsymbol\lambda}(x)} d\mu(x)$ is a KL divergence and is always non-negative.

Assume that $\boldsymbol\lambda$ is in $\mathring \Gamma_\mu$. Then, there exists $\epsilon > 0$ such that, for any $\bfu\in \mR^p$, $|\bfu|=1$, $\boldsymbol\lambda + \epsilon \bfu\in \Gamma_\mu$. Using the fact that $e^\beta \geq e^\alpha + (\beta-\alpha)e^\alpha$, we can write
\begin{align*}
\epsilon \bfu^T \bfg e^{\lambda^T \bfg} &\leq e^{(\lambda+\epsilon u)^T\bfg} - e^{\lambda^T \bfg}\\
-\epsilon \bfu^T \bfg e^{\lambda^T \bfg} &\leq e^{(\lambda-\epsilon u)^T\bfg} - e^{\lambda^T \bfg}
\end{align*}
yielding
\[
\epsilon |\bfu^T \bfg| e^{\lambda^T \bfg} \leq \max(e^{(\lambda+\epsilon u)^T\bfg}, e^{(\lambda-\epsilon u)^T\bfg}) - e^{\lambda^T \bfg} \leq e^{(\lambda+\epsilon u)^T\bfg} + e^{(\lambda-\epsilon u)^T\bfg} - e^{\lambda^T \bfg}.
\]
Since the upper-bound is integrable with respect to $\mu$, so is the lower bound, showing that (taking $\boldsymbol u$ in the canonical basis of $\mR^p$)
\[
\int_\CG |\pe g i(y)| e^{\lambda^T \bfg(y)}d\mu(y) <\infty
\]
for all $i$, or
\[
\int_\CG |\boldsymbol g(y)| e^{\lambda^T \bfg(y)}d\mu(y) <\infty.
\]
Let $\bfc = E(\bfg(Y))$ and define
\begin{equation}
\Psi_{\boldsymbol c}(\boldsymbol\la) = - {\boldsymbol c}^T\boldsymbol\la +  \log \int \exp(\boldsymbol\la^T\boldsymbol g(y)) dy.
\label{eq:max.ent.psi}
\end{equation} 
Then
\[
\prt_{\boldsymbol\lambda} \Psi_c = - c^T + \frac{\int \boldsymbol g(x)^T\exp(\boldsymbol\la^T\boldsymbol g(x)) dx}{\int \exp(\boldsymbol\la^T\boldsymbol g(y)) dy}
= -c^T + \int_{\CG} \boldsymbol g(x)^T \psi_{\boldsymbol\lambda}(x) dx\,.
\]
Since $\boldsymbol \la^*$ is a minimizer, we find that,  if $\tilde Y$ is a random variable with p.d.f. $\boldsymbol\lambda^*$, then $E(\tilde Y) = c = E(Y)$. In that case, the upper-bound in \cref{eq:max.ent.principle.0} is $h_\mu(\tilde Y)$, proving \cref{eq:max.ent.principle.3}.
\end{proof}
\begin{remark}
The previous theorem is typically applied with $\mu$ equal to Lebes\-gue's measure on $\CG = \mR^d$ or to a counting measure with $\CG$  finite. To rewrite the statement of \cref{th:max.ent.principle} in those cases, it suffices to replace $d\mu(x)$ by $dx$ for the former, and integrals by sums over $\CG$ for the latter. In the rest of the discussion, we restrict to the case when $\mu$ is Lebesgue's measure,  using $h(Y)$ instead of $h_\mu(Y)$.
\end{remark}

\begin{remark}
This principle justifies, in particular, that the negentropy is always non-negative since it implies that a distribution that maximizes the entropy given its first and second moments must be Gaussian.  
\end{remark}

The right-hand side of \cref{eq:max.ent.principle.0} provides a variational approximation of the entropy. If one uses this approximation when minimizing 
$h(W^{(1)} X) + \cdots + h(W^{(d)} X)$, the resulting problem can be expressed as a minimization, with respect to
$W$ and $\boldsymbol\lambda_1, \ldots, \boldsymbol\lambda_d\in \mR^p$ of
\[
- \sum_{j=1}^d \boldsymbol\la^T E(\bfg(W^{(j)}X)) + \sum_{j=1}^d \log \int \exp\left(\boldsymbol\la^T \bfg(y)\right)\,dy\,.
\]

While it is possible to solve this optimization problem directly,  a further approximation of the upper bound can be developed leading to a simpler procedure. We have seen in the previous proof that, defining $\Psi_{\bfc} $ by \cref{eq:max.ent.psi} and denoting by  $E_{\boldsymbol\la}$ the expectation with respect to $\phi_{\boldsymbol\lambda}$, one has
\[
\nabla \Psi_\bfc(\boldsymbol\la) = - \bfc + E_{\boldsymbol\la}(\bfg)\,.
\]
Taking the second derivative, one finds
\[
\nabla^2 \Psi_\bfc(\boldsymbol\la) = E_{\boldsymbol\la}((\bfg-E_{\boldsymbol\la}(\bfg))(\bfg-E_{\boldsymbol\la}(\bfg))^T).
\]
Now choose $\bfc_{0}$ such that a maximizer of $\Psi_{\bfc_{0}}(\boldsymbol\la)$, say, $\boldsymbol\la^{\bfc_{0}}$, is known. If $\bfc$ is close to $\bfc_{0}$, a first order expansion indicates that, for $\boldsymbol\la^{\bfc}$ maximizing $\Psi_\bfc$, one should have
\[
\boldsymbol\la^{\bfc} \simeq \boldsymbol\la^{\bfc_{0}} + \nabla^2 \Psi_\bfc(\boldsymbol\la^{\bfc_{0}})^{-1} (\bfc - \bfc_{0})
\]
with
\[
\Psi_\bfc(\boldsymbol\la^c)\simeq \Psi_\bfc(\boldsymbol\la^{\bfc_{0}}) - (\bfc-\bfc_{0})^T\nabla^2 \Psi_\bfc(\boldsymbol\la^{\bfc_{0}})^{-1} (\bfc - \bfc_{0}). 
\]
One can then use the right-hand side as an approximation of the optimal entropy. 

This leads to simple computations under the following assumptions. First, assume that  the first two functions $\pe g1$ and $\pe g2$ are $u$ and $u^2/\sqrt 3$. 
Let $\phi_0$ be the p.d.f. of a standard Gaussian. Assume that the functions $\pe g j$ are chosen so that
\[
\int \pe g i(u) \pe g j(u) \phi_0(y) dy = \de_{ij}
\]
for $i,j=1, \ldots, p$ and such that $\int \pe g i(u) \phi_0(y) dy = 0$ for $i\neq 2$. Take
\[
\bfc_0 = \int \bfg \phi_0(u) du
\]
so that $c^{(1)}_0=0$, $c^{(2)}_0 = 1/\sqrt 3$ and $c^{(i)}_0 = 0$ for $i\geq 2$.

Then $\la^{\bfc_{0}}$ provides, by construction, the distribution $\phi_0$ and  for any $\bfc$, $\nabla^2\Psi_\bfc(\boldsymbol\la^{\bfc_{0}}) = \Id[p]$. With these assumptions, the approximation is 
\begin{align*}
\Psi_\bfc(\boldsymbol\la) &= h(\phi_0) - |\bfc-\bfc_{0}|^2\\
&= \frac12(1+ \log 2\pi) - \sum_{j\geq 3} (\pe c j)^2 
\end{align*}
(assuming that the data is centered and normalized so that $\pe c 1 = 0$ and $\pe c 2 = 1/\sqrt 3$). The ICA problem can then be solved by maximizing
\begin{equation}
\label{eq:ica.easy}
\sum_{j=1}^d \sum_{i=1}^p E(\pe g i(W^{(j)} X))^2
\end{equation}
over orthogonal matrices $W$. 

\begin{remark}
Without the assumption made on the functions $\pe g j$, one needs to compute $S = \Cov(g(U))^{-1}$ where $U\sim \CN(0,1)$ and maximize
\[
\sum_{j=1}^d (E(\bfg(W^{(j)} X)) - E(\bfg(U)))^TS(E(\bfg(W^{(j)} X)) - E(\bfg(U))).
\]
Clearly, this expression can be reduced to \cref{eq:ica.easy} by replacing $\bfg$ by $S^{-1/2}(\bfg- E(\bfg(U)))$. Note also that we retrieve here a similar idea to the negentropy, maximizing a deviation to a Gaussian.
\end{remark}

\subsection{Maximization over orthogonal matrices}
\label{sec:grad.orthogonal}
In the previous discussion, we reached a few times a formulation of ICA which required optimizing a function $W \mapsto F(W)$ over all orthogonal matrices. We now discuss how such a problem may be implemented. 

In all the examples that were considered, there would have been no loss of generality in requiring that $W$ is a rotation, i.e., $\det(W) = 1$. This is because one can change the sign of this determinant by simply changing the sign of one of the independent components, which is always possible. (In fact, the indeterminacy in $W$ is by right multiplication by the product of a permutation matrix and a diagonal matrix with $\pm1$ entries.)

Let us assume that $F(W)$ is actually defined and differentiable over all invertible matrices, which form an open subset of the linear space $\CM_d(\mR)$ of $d$ by $d$ matrices.  Our optimization problem can therefore be considered as the minimization of $F$ with the constraint that $WW^T = \Id[d]$. 

Gradient descent derives from the analysis that a direction of descent should be a matrix $H$ such that $F(W + \ep H) < F(W)$ for small enough $\ep>0$ and on the remark that $H = -\nabla F(W)$ provides such a direction. This analysis does not apply to the constrained optimization setting because, unless the constraints are linear, $W+\ep H$ will generally stop to satisfy them when $\ep >0$, requiring the use of more complex procedures, as discussed in \cref{sec:opt.constr}. In our case, however, one can take advantage of the fact that orthogonal matrices form a group to replace the perturbation $W \mapsto W + \ep H$ by $W \mapsto We^{\ep H}$ (using the matrix exponential) where $H$ is moreover required to be skew symmetric ($H+H^T=0)$, which guarantees that $e^{\ep H}$ is an orthogonal matrix with determinant 1 (and all rotation matrices can be represented in this way). Now, using the fact that $e^{\ep H} = \Id + \ep H + o(\ep)$, we can write
\[
F(We^{\ep H}) = F(W) + \ep \trace(\nabla F(W)^T WH) + o(\ep)\,.
\]
Let $\nabla^s F(W)$ be the skew symmetric part of $W^T\nabla F(W)$, i.e., 
\[
\nabla^s F(W) = \frac12 (W^T\nabla F(W) - \nabla F(W)^TW).
\]
Then, if $H$ is skew symmetric, 
\begin{align*}
\trace(\nabla^s F(W)^T H) &= \frac12 \trace(\nabla F(W)^T WH) - \frac12 \trace(W^T\nabla F(W) H) \\
&=  \frac12 \trace(\nabla F(W)^T WH) + \frac12 \trace(W^T\nabla F(W) H^T)\\
& = \trace(\nabla F(W)^T WH) 
\end{align*}
so that 
\[
F(We^{\ep H}) = F(W) + \ep \trace(\nabla^s F(W)^T H) + o(\ep)\,.
\]
This show that $H = - \nabla^s F(W)$ provides a direction of descent in the orthogonal group, in the sense that, if $\nabla^s F(W) \neq 0$, 
then $F(We^{-\ep\nabla^s F(W)}) < F(W)$
for small enough $\ep > 0$. As a consequence, the algorithm 
\begin{equation}
\label{eq:opt.expon}
W_{n+1} = W_n  e^{-\ep_n\nabla^s F(W_n)}
\end{equation}
combined with a line search for $\ep_n$ implements gradient descent in the group of orthogonal matrices, and therefore converges to a local minimizer of $F$. 

If one linearizes the r.h.s. of \cref{eq:opt.expon} as a function of $\ep$, one gets
\begin{align*}
W_n  e^{-\ep_n\nabla^s F(W_n)} &=  W_n + \frac{\ep_n} 2 W_n( (W_n)^T\nabla F(W_n) - \nabla F(W_n)^TW_n) + o(\ep)\\
& = W_n + \frac{\ep_n} 2 ( \nabla F(W_n) - W_n\nabla F(W_n)^TW_n) + o(\ep).
\end{align*}
As already argued, this linearized version cannot be used when optimizing over the orthogonal group. However, if one denotes by $\om(A)$ the unitary part of the polar decomposition of $A$, i.e., $\om(A) = (AA^T)^{-1/2}A$, then the algorithm
\[
W_{n+1} = \om\left(W_n + \frac{\ep_n} 2 ( \nabla F(W_n) - W_n\nabla F(W_n)^TW_n)\right)
\]
also provides a valid gradient descent algorithm.
\bigskip

\subsection{Parametric ICA}
We now describe a parametric version of ICA in which a model is chosen for the independent components of $Y$. The simplest version of to assume that all $Y^{(j)}$ are i.i.d. with some prescribed p.d.f., say, $\psi$, and a typical example for $\psi$ is a logistic distribution with
\[
\psi(t) = \frac{2}{(e^t+
e^{-t})^2}.
\]
If $y$ is a vector in $\mR^d$, we will use, as usual, the notation $\psi(y)  = (\psi(y^{(1)}), \ldots, \psi(y^{(d)}))^T$ for $\psi$ applied to each component of $y$.

This provides a statistical  model whose  parameter is the matrix $A$, or preferably $W = A^{-1}$, which may be estimated using maximum likelihood. Indeed, the p.d.f. of $X$ is
$$
f_X(x) = |\det W|\, \prod_{j=1}^d \psi(W^{(j)}x)
$$
where $W^{(j)}$ is the $j$th row of $W$, so that $W$ can be estimated by maximizing
\[
\ell(W) = N \log\abs{\text{det}(W)} +  \sum_{k=1}^N \sum_{j=1}^d \log \psi(W^{(j)}x_k)\,.
\]
If we denote by $\Ga(W)$ the matrix with coefficients 
\[
\ga_{ij}(W) = \sum_{k=1}^N x_k(i) \frac{\psi'(W^{(j)}x_k)}{\psi(W^{(j)}x_k)}
\]
and use the fact that the gradient of $W \mapsto \log |\det W|$ is $W^{-T}$ (the inverse transpose of $W$, see \cref{sec:diff.examples}), we can write
\[
\nabla \ell(W) = NW^{-T} + \Ga(W).
\] 

We need however the maximization to operate on sets of invertible matrices, and it is more natural to move in this set through multiplication than through addition, because the product of two invertible matrices is always invertible, but not necessarily their sum. So, similarly to the previous section, we will look for small variations in the form  $W \mapsto We^{\ep H}$, or simply, in this case, $W \mapsto W(\Id[d] + \ep H)$. In both cases, the first order expansion of the log-likelihood gives
\[
\ell(W) + \ep \trace((NW^{-T} + \Ga(W))^TWH)
\]
which suggests taking 
\[
H = W^T(NW^{-T} + \Ga(W)) = N\Id + W^T\Ga(W).
\]

Dividing $H$ by $N$, we obtain the following variant of gradient ascent for maximum likelihood
\[
W_{n+1} = (1 + \ep_n) W_n +  \ep_n W_nW_n^T\Ga(W_n)\,.
\]
This algorithm numerically performs much better than standard gradient ascent. It moreover presents the advantage of avoiding computing the inverse of $W$ at each step.

\subsection{Probabilistic ICA}

Note that the algorithms that we discussed concerning ICA were all formulated in terms of the matrix $W = A^{-1}$, which ``filters'' the data into independent components. As a result, ICA requires as many independent components 
 as the dimension of  $X$. Moreover, because the components are typically normalized to have equal variance, there is no obvious way to perform dimension reduction using this method. Indeed, ICA is typically run after the data is preprocessed using PCA, this preprocessing step providing the reduction of dimension.
It is however possible to define a model similar to probabilistic PCA, assuming a limited number of components to which a Gaussian noise is added, in the form
$$
X  = \sum_{j=1}^p a_j Y^{(j)} + \sig R 
$$
with $p< d$, $a_1, \ldots, a_p\in \mR^d$,  $Y^{(1)},
\ldots, Y^{(p)}$ independent variables as before, and $R \sim \CN(0, \Id_{\mR^d})$.  This model is identifiable (up to permutation and scalar multiplication of the components) as soon as none of the variables $Y^{(j)}$  is Gaussian.

 Let us assume a parametric setting similar to that of the previous section, so that $Y^{(1)}, \ldots, Y^{(p)}$ are explicitly modeled as independent variables with p.d.f. $\psi$. Introduce the matrix $A = [a_1, \ldots, a_p]\in \CM_{d,p}(\mR)$, so that the model 
can also be written $X = AY +\sig R$, where  $A$ and 
$\sig^2$ are the unknown model parameters.

The p.d.f. of $X$ is now given by
\[
f_X(x; A, \sig^2) = \frac{1}{(2\pi\sig^2)^{d/2}} \int_{\mR^p} e^{-\frac{|x - Ay|^2}{2\sig^2}} \left(\prod_{i=1}^p \psi(y^{(i)})\right)  dy^{(1)}\ldots dy^{(p)},
\]
which is definitely not a closed form. Since we are in  a situation in which
the pair of random variables is imperfectly observed through $X$, using the
EM algorithm (\cref{chap:var.bayes}) is an option, but it may, as we  see below,  lead to  heavy computation. The
basic step of the EM is, given current parameters $A_0, \sig_0$, to
maximize the conditional expectation (knowing $X$, for the current parameters) of
the joint log-likelihood of $(X,Y)$ with respect to the new
parameters. In this context,  the joint distribution of $(X,Y)$ has density
$$
f_{X,Y}(x,y;A, \sig^2) = \frac{e^{-\frac{\abs{x- Ay}^2}{2\sig^2} } }{(2\pi\sig^2)^{d/2}}\prod_{i=1}^p \psi(y^{(i)}),
$$
so that, the conditional joint likelihood over the training set is
\begin{multline*}
- \frac{Nd}{2}\log (2\pi\sig^2)  - \frac{1}{2\sig^2} \sum_{k=1}^N E_{A_0,
\sig_0}( \abs{x_k- AY}^2 \mid X=x_k) \\
- \sum_{k=1}^N \sum_{j=1}^p  E_{A_0, \sig_0^2}
(\log \psi(Y^{(j)}) \mid X=x_k).
\end{multline*}

Notice that the last term does not depend on $A$ or $\sig^2$, and that, given
$A$, the optimal value of $\sig^2$ is given by
$$
\sig^2 = \frac{1}{Nd} \sum_{k=1}^N E_{A_0,
\sig_0}( \abs{x_k- AY}^2 \mid X=x_k)
$$
The minimization of
$$
\sum_{k=1}^N E_{A_0,
\sig_0}( \abs{x_k- AY}^2 \mid X=x_k)
$$
with respect to $A$ is a least squares problem.
Letting $\pe {b_{k}} j = E_{A_0, \sig_0}(Y^{(j)} \mid X=x_k)$ and $s_k(i,j) = E_{A_0,\sig_0}(Y^{(i)}Y^{(j)}|X=x_k)$, the
gradient of the previous term is
\[
-2\sum_{k=1}^N E_{A_0, \sig^2_0}((x_k - AY)Y^T | X=x_k) =
-2\sum_{k=1}^N (x_k b_k^T - A S_k),
\]
$b_k$ being the column vector with coefficients $\pe {b_k}j$ and $S_k$ the
matrix with coefficients $s_k(i,j)$. The result therefore is 
$$ A = \left(\sum_{k=1}^N x_k b_k^T\right)\left(\sum_{k=1}^N S_k\right)^{-1} .$$

Unfortunately, the computation of the moments of the conditional
distribution of $Y$ given $x_k$ (needed in $b_k $ and $S_k$) is a
difficult task. The conditional density of   $Y$ given $X=x_k$ is
$$
g(y|x_k) =  \psi(y) e^{-\frac{|A_0y - x|^2}{2\sig_0^2} }/Z(A_0,
\sig_0)
$$
from which moments cannot be  computed analytically in general. Monte-Carlo
sampling algorithms can be used however to approximate these moments,
but they are computationally demanding. And they must be run at
every step of the EM.

In place of the exact EM, one may use a mode approximation (\cref{sec:mode.approx}), which  replaces the conditional
likelihood of $Y$ given $X=x_k$
by a Dirac distribution at the mode:
\[
\hat y_{A_0, \sig_0}(x_k) = \text{argmax}_y \left( \psi(y)
e^{-\frac{|A_0y - x_k|^2}{2\sig_0^2} }\right).
\]
The maximization step then reduces to maximizing
in $A, \sig^2$
\begin{equation}
\label{eq:mode.ica}
- \frac{Nd}{2}\log (2\pi\sig^2)  - \frac{1}{2\sig^2} \sum_{k=1}^N
\abs{x_k- A\hat y_{A_0, \sig_0}(x_k)}^2.
\end{equation}
This therefore provides a  two-step procedure.
\begin{algorithm}[Probabilistic ICA: mode approximation]
\label{alg:ica.mode}
\begin{enumerate}[label=(\arabic*),wide]
\item Initialize the algorithm with $A_0, \sigma_0$. 
\item At step $n$:
\begin{enumerate}[label=(\roman*),wide=1cm]
\item For $k=1, \ldots, N$, maximize $ \prod_{i=1}^p \psi(y^{(i)})
e^{-\frac{|A_ny - x_k|^2}{2\sig_n^2} }$ to obtain $\hat y_{A_n,\sig_n}(x_k)$.
This requires a
numerical optimization procedure, such as gradient ascent. The problem (after taking the logarithm) is
concave when $\log \psi$ is concave. 
\item Minimize \cref{eq:mode.ica} with
respect to $A$, $\sig^2$, yielding
$$ A_{n+1} = \left(\sum_{k=1}^N x_k b_k^T\right)\left(\sum_{k=1}^N S_k\right)^{-1}$$
with $b_{k} = \hat y_{A_n, \sig_n}(x_k)$, $S_k = \hat y_{A_n,
\sig_n}(x_k)\hat y_{A_n, \sig_n}(x_k)^T$, and
$$
\sig_{n+1}^2 = \frac{1}{Nd} \sum_{k=1}^N \abs{x_k- A\hat y_{A_n, \sig_n}}^2.
$$
\end{enumerate}
\item Stop if the variation of the parameter is below a tolerance level. Otherwise, iterate to the next step.
\end{enumerate}
\end{algorithm}

Once $A$ and $\sig^2$ have been estimated, the independent components  associated to a new observation $x$ can be estimated by $\hat
y_{A,\sig}(x)$, therefore minimizing
$$
\frac{1}{2\sig^2} \abs{x_k- Ay}^2 + \sum_{j=1}^p  \log \psi(y^{(j)}),
$$
yielding the map estimate, the same convex optimization problem as in
step (1) of \cref{alg:ica.mode}. Now we can see how the method takes from both  PCA and
ICA: the columns of $A$, $a_1, \ldots, a_p$ can be considered as  $p$ 
principal directions, and are fixed after learning; they are not
orthonormal, and do not satisfy the nesting properties of PCA (that those $p$
contain those for $p-1$). The coordinates of $x$ with respect to this basis
is not a projection, as would be provided by PCA, but the result of a
penalized estimation problem. The penalty associated to the logistic
case is 
$$
\log \psi(y^{(j)}) =  \log 2 - 2 \log (e^{y^{(j)}} + e^{-y^{(j)}}).
$$
This distribution with ``exponential tails'' has the interest of
allowing large values of $y^{(j)}$, which generally entails {\em sparse
decompositions}, in which $y$ has a few large coefficients, and many
zeros.

As an alternative to the mode approximation of the EM,  which may lead to biased estimators, one may use the SAEM algorithm (\cref{sec:saem}), as proposed in \citet{allassonniere2012stochastic}. Recall that the EM algorithm replaces the parameters $A_0, \sig_0^2$ by minimizers of 
\begin{align*}
&\frac{Nd}{2}\log (\sig^2)  + \frac{1}{2\sig^2} \sum_{k=1}^N E_{A_0,
\sig_0}( \abs{x_k- AY}^2 | X=x_k) \\
&= \frac{Nd}{2}\log (\sig^2) + \frac{1}{2\sig^2} \sum_{k=1}^N |x_k|^2 - \frac{1}{\sig^2} \sum_{k=1}^N x_k^T Ab_k + \frac{1}{2\sig^2} \sum_{k=1}^N \trace(A^TAS_k),
\end{align*}
where the computation of $b_{k}^{(j)} = E_{A_0, \sig_0}(Y^{(j)} | X=x_k)$ and $s_k(i,j) = E_{A_0,\sig_0}(Y^{(i)}Y^{(j)}|X=x_k)$ was the challenging issue. In the SAEM algorithm, the statistics $b_k$ and $S_k$ are part of a stochastic approximation scheme, and are estimated in parallel with EM updates as follows.
\begin{algorithm}[SAEM for probabilistic ICA]
\label{alg:ica.saem}
Initialize the algorithm with parameters $A$, $\sig^2$. Define a sequence of decreasing steps, $\gamma_t$.

Let, for $k=1, \ldots, N$, $b_k = 0$ and $S_k = \Id$. Iterate the following steps.
\begin{enumerate}
\item For $k=1, \ldots, N$, sample $y_k$ according to the conditional distribution of $Y$ given $X=x_k$, using the current parameters $A$ and $\sig^2$.
\item Update $b_k$ and $S_k$, letting (assuming step $t$ of the algorithm)
\[
\left\{
\begin{aligned}
b_k &\to b_k +  \ga_{t} (Y_k - b_k)\\
S_k &\to S_k +  \ga_{t} (Y_kY_k^T - S_k)
\end{aligned}
\right.
\]
\item Replace $A$ and $\sig^2$ by
\[
A = \left(\sum_{k=1}^N x_k b_k^T\right)\left(\sum_{k=1}^N S_k\right)^{-1} 
\]
and 
\[
\sig^2 = \frac{1}{Nd} \sum_{k=1}^N \abs{x_k- A\hat y_{A_0, \sig_0}}^2.
\]
\end{enumerate}
\end{algorithm}

The parameter $\ga_t$ should be decreasing with $t$, typically so that $\sum_t \ga_t = +\infty$ and $\sum_t\ga_t^2 < \infty$ (e.g., $\ga_t \propto 1/t$). One way to sample from $Y_k$ is to use a rejection scheme, iterating the procedure which samples $y$ according to the prior and accepts the result with probability  $M\exp(- |x_k - Ay|^2/2\sig^2)$ until acceptance. Here $M$ must be chosen so that
$M \max_y \exp(- |x_k - Ay|^2/2\sig^2) \leq 1$ (e.g., $M=1$). 
This method will work for small $p$, but for large $p$, the probability of acceptance may be very small. In such cases, $Y_k$ can be sampled changing one component at a time using a Metropolis-Hastings scheme (\cref{sec:met.hast}). If component $j$ is updated, this scheme  samples a new value of $y$ (call it $y'$) by changing only $\pe y j$ according to the prior distribution $\psi$ and accept the change with probability
\[
    \min\left(1, \frac{\exp(- |x_k - Ay'|^2/2\sig^2)}{\exp(- |x_k - Ay|^2/2\sig^2)}\right).
\]

\section{Non-negative matrix factorization}

In this section, we consider factor analysis methods  that approximate a random variable $X$ in the form $X = \sum_{j=1}^p \pe a j \pe Y j$ with the constraint that the scalars $\pe a 1$, \ldots, $\pe a p \in \mR$ and the vectors $\pe Y 1, \ldots, \pe Y p \in \mR^d$   are respectively non-negative and with non-negative entries. This model makes sense, for example, when $X$ represents the total multivariate production (e.g., in terms of number of molecules of various types) resulting of several chemical reactions that operate together. Another application is when $X$ is a list of preference scores associated with a person for, say, books or movies, and  each person is modeled as a positive linear combination of $p$ ``typical scorers,'' represented by the vector $\pe Y j$ for $j=1, \ldots, p$. 

When training data $(x_1, \ldots, x_N)$ is observed and stacked in an $N$ by $d$ matrix $\CX$,  the decomposition can be summarized for all observations together in the matrix form
\[
\CX = \CA Y^T
\]
where $\CA$ is $N$ by $p$ and provides the coefficients $\pe {a_k}j$ associated with each observation and  $Y = [\pe y 1, \ldots, \pe y p]$ is $d$ by $p$ and provides the $p$ typical profiles. 
The matrices $\CA$ and $Y$ are unknown and their estimation subject to the constraint of having non-negative components represent the non-negative matrix factorization (NMF) problem. 

NMF is often implemented by solving the constrained optimization problem of minimizing 
$|\CX - \CA Y^T|^2$ subject to $\CA$ and $Y$ having non-negative entries. 
 This problem is non-convex in general but the sub-problems of optimizing either $\CA$ or $Y$ when the other matrix is fixed are simple quadratic programs (minimizing a quadratic function of the variable with linear constraints). This suggests using an alternating minimization method,  iterating steps in which $\CA$ is updated with $Y$ fixed, followed by an update of $Y$ with $\CA$ fixed. 
However,  solving a full quadratic program at each step would be computationally prohibitive with large datasets, and simpler update rules have been suggested, updating each matrix in turn with a guarantee  of reducing the objective function.

If $Y$ is considered as fixed and $\CA$ is the free variable, we have
\begin{align*}
|\CX - \CA Y^T|^2 &= |\CX|^2 - 2 \trace(\CX^T\CA Y^T) + \trace(\CA Y^TY\CA^T)\\
&= \trace(\CA^T \CA Y^T Y) - 2\trace(\CA^T(\CX Y)) + |\CX|^2\,.
\end{align*}
The next lemma will provide update steps for $\CA$.

\begin{lemma}
\label{lem:nmf}
Let $M$ be an $n$ by $n$ symmetric matrix and $b \in \mR^n$, both assumed to have non-negative entries. 
Let $u\in \mR^n$, also with non-negative coefficients, and let
\[
\pe v i = \pe u i \left(\frac{\pe b i}{\sum_{j=1}^d m(i,j) \pe u j}\right).
\]
Then
\[
v^TMv - 2b^Tv \leq u^TMu - 2b^Tu\,.
\]
Moreover, $v=u$ if and only if $u$ minimizes $u^TMu - 2b^Tu$ subject to $\pe u i\geq 0$, $i=1, \ldots, n$.

\end{lemma}
\begin{proof}
Let $F(u) = u^TMu - 2b^Tu$.
We look for $\pe v i = \pe \beta i \pe u i$ with $\pe \beta i \geq 0$ such that $F(v) \leq F(u)$. We have
\begin{align*}
F(v) &= \sum_{i,j=1}^n \pe \beta i \pe \beta j \pe u i\pe u j m(i,j) - 2\sum_{i=1}^n \pe b i \pe \beta i \\
&\leq \frac12 \sum_{i,j=1}^n ((\pe \beta i)^2 + (\pe \beta j)^2) \pe u i\pe u j m(i,j) - 2\sum_{i=1}^n \pe b i \pe \beta i \pe u i\\
& =   \sum_{i,j=1}^n (\pe \beta i)^2 \pe u i \pe u j m(i,j) - 2\sum_{i=1}^n \pe b i \pe \beta i \pe u i
\end{align*}
When $\beta = \dsone_n$, the upper-bound is equal to  $F(u)$. So, if we choose  $\beta$ minimizing the upper-bound, we will indeed find $v$ such that $F(v) \leq F(u)$. Rewriting the upper-bound as
\[
\sum_{i=1}^n \pe u i  \left((\pe \beta i)^2 \left(\sum_{j=1}^n m(i,j) \pe u j\right) - 2 \pe b i \pe \beta i\right)
\]
we see that $\pe \beta i = \pe b i /  \sum_{j=1}^n m(i,j) \pe u j$ provides such a minimizer, which proves  the first statement of the lemma. For the second statement, we have $\pe v i = \pe u i$ if and only if $\pe u i = 0$ or  $\sum_{j=1}^n m(i,j) \pe u j = \pe b i$, and one directly checks that these are exactly the KKT conditions for a minimizer of $F$ over vectors with non-negative entries.
%
%
\end{proof}

 To apply the lemma to the minimization in $\CA$, let $M: \CA \mapsto \CA Y^TY$ and $b=\CX Y$ (we are working in the linear space of $N$ by $p$ matrices). 
 Then  the update
\[
\pe{a_k}{i} \mapsto a_k^{(i)} \frac{(\CX Y)(i,k)}{(\CA Y^TY)(i. k)}
\]
decreases the objective function. 

Similarly, applying the lemma with the operator $Y \mapsto Y\CA^T\CA$ and $b = \CX^T\CA$ gives the update for $Y$, namely
\[
y_j^{(i)} \mapsto y_j^{(i)} \frac{(\CX^T \CA)(i,j)}{(Y \CA^T\CA)(i,j)}.
\]

We have therefore obtained the following algorithm.
\begin{algorithm}[NMF, quadratic cost]
\label{alg:nmf.quadratic}
\begin{enumerate}[label=\arabic*., wide=0.5cm]
\item Fix $p>0$ and let $\CX$ be the $N$ by $d$ matrix containing the observed data. Initialize the procedure with matrices $\CA$ and $Y$, respectively of size $N$ by $p$ and $d$ by $p$, with positive coefficients.
At a given stage of the algorithm, let $\CA$ and $Y$ be the current matrices providing an approximate decomposition of $\CX$. 
\item For the next step, let $\tilde{\CA}$ be the matrix with coefficients
\[
\pe{\tilde a_k} i  = \pe{a_k} i \frac{(\CX Y)(i,k)}{(\CA Y^TY)(i,k)}
\]
and $\tilde Y$ the matrix with coefficients
\[
\pe{\tilde y_j} i = \pe{y_j} i \frac{(\CX^T \tilde{\CA})(i,j)}{(Y \tilde{\CA}^T\tilde{\CA})(i,j)}.
\]
\item
Replace $\CA$ by $\tilde \CA$ and $Y$ by $\tilde Y$, iterating until numerical convergence.
\end{enumerate}
\end{algorithm}

\begin{remark}
\label{rem:als}
Without the positivity constraints, the minimization of $|\CX - \CA Y^T|^2$ can be done by alternatively minimizing this expression in $\CA$ and in $Y$ until convergence. Each of these steps is an easy-to-solve least-squares problem, and the method is called ``alternating least squares'' \cite{van1990non,breiman1985estimating}. 
\end{remark}

An alternative version of the method has been proposed, where the objective function is $\Phi(\CA Y^T)$, where, for an $N$ by $d$ matrix $\CZ = [z_1, \ldots, z_N]^T$,
\[
\Phi(\CZ) = \sum_{k=1}^N \sum_{i=1}^d (\pe {z_k} i - \pe{x_k} i \log \pe {z_k} i)
\]
which is indeed minimal for $\CZ = \CX$.   We state and prove a second lemma that will allow us to address this problem.
\begin{lemma}
\label{lem:nmf.2}
Let $M$ be an $n$ by $q$ matrix and $x \in \mR^n$, $b\in \mR^q$, all assumed to have positive entries. 
For $u\in (0, +\infty)^q$, define
\[
F(u) = \sum_{j=1}^q \pe b j \pe u j - \sum_{i=1}^n \pe x i  \log \sum_{j=1}^q m(i,j) \pe u j.
\]
Define $v \in (0, +\infty)^q$ by
\[
\pe v j = \pe u j\left(\frac{\sum_{i=1}^n m(i,j) \pe x i / \pe\al i}{\pe b j}\right)
\]
with $\pe \al i = \sum_{k=1}^q m(i,k) \pe u k $.
Then
$
F(v) \leq F(u)$. Moreover, $v=u$ if and only if $u$ minimizes $F$ on $(0, +\infty)^q$.

\end{lemma}
\begin{proof}
Introduce a variable $\pe \be j >0$ for $j=1, \ldots, q$ an let $\pe w j = \pe u j  \pe \beta j$. Then
\begin{align*}
F(w) &= \sum_{j=1}^q \pe b j  \pe u j \pe \be j - \sum_{i=1}^n \pe x i \log \sum_{j=1}^q m(i,j) \pe u j \pe \be j \\
&= \sum_{j=1}^q \pe b j  \pe u j \pe \be j - \sum_{i=1}^n \pe x i \log \frac{\sum_{j=1}^q m(i,j) \pe u j \pe \be j}{\sum_{j=1}^q m(i,j) \pe u j} - \sum_{i=1}^n \pe x i \log \sum_{j=1}^q m(i,j) \pe u j
\end{align*}
Let $\rho(i,j) = m(i,j) \pe u j / \pe \al i$. Since the logarithm is concave, we have 
\[
\log \sum_{j=1}^q \rho(i,j) \pe \be j \geq \sum_{j=1}^q \rho(i,j) \log\pe \be j
\]
so that 
\[
F(w) \leq \sum_{j=1}^q \pe b j \pe u( j \pe \be j - \sum_{i=1}^n \sum_{j=1}^q \pe x i \rho(i,j) \log \pe \be j - \sum_{i=1}^n \pe x i \log \sum_{j=1}^q m(i,j) \pe u j.
\]
The upper bound with $\pe \be j \equiv 1$ gives $F(u)$, so minimizing this expression in $\be$ will give $F(w) \leq F(u)$. This minimization is straightforward and gives
\[
\pe \be j = \frac{\sum_{i=1}^n \pe x i \rho(i,j)}{\pe b j \pe u j} = \frac{\sum_{i=1}^n m(i,j) \pe x i / \pe\al i}{\pe b j}
\]
and the optimal $w$ is the vector $v$ provided in the lemma. Finally, one checks  that $F$ is a convex function $v=u$ if and only if $\nabla F(u) = 0$, i.e.,  $u$ is a minimizer.
\end{proof}

We can now apply this lemma to derive update rules for $\CY$ and $\CA$, where the objective is
\[
\sum_{k=1}^N \sum_{i=1}^d \sum_{j=1}^p \pe {y_j} i \pe{a_k}j - \sum_{k=1}^N\sum_{i=1}^d \pe{x_k} i \log\sum_{j=1}^p \pe{y_j} i \pe {a_k} j.
\]
Starting with the minimization in $\CA$, we  apply the lemma to each  index $k$ separately,  taking $n=d$ and $q=p$, with $\pe b j = \sum_{i=1}^d \pe{y_j} i$  and $m(i,j) = \pe{y_i} j$. 
 Then the update is 
\[
a_k(j) \mapsto a_k(j) \frac{\sum_{i=1}^d \pe {x_k} i \pe{y_j} i/\pe{\al_k} i}{\sum_{i=1}^d \pe{y_j}i}
\]
with $\pe{\al_k}i = \sum_{j=1}^p \pe{y_j}i \pe{a_k} j$.

For $Y$, we can work with fixed $i$ and apply the lemma with $n=N$, $q=p$, $\pe b j = \sum_{k=1}^N \pe{a_k} j$ and $m(k,j) = \pe{a_k} j$. 
 This gives the update:
\[
\pe{y_j} i \mapsto \pe{y_j} i  \frac{\sum_{k=1}^N \pe {x_k} i \pe{a_k} j/\pe{\al_k} i}{\sum_{k=1}^N \pe{a_k} j},
\]
still with $\pe{\al_k} i = \sum_{j=1}^p \pe{y_j} i \pe{a_k} j$.

We summarize this in our second algorithm for NMF.
\begin{algorithm}[NMF, logarithmic cost]
\label{alg:nmf.log}
\begin{enumerate}[label=\arabic*., wide=0.5cm]
\item Fix $p>0$ and let $\CX$ be the $N$ by $d$ matrix containing the observed data. 
\item Initialize the procedure with matrices $Y$ and $\CA$, respectively of size $N$ by $p$ and $d$ by $p$, with positive coefficients.
At a given stage of the algorithm, let $\CA$ and $Y$ be the current matrices decomposing $\CX$. 
\item Let $\tilde{\CA}$ be the matrix with coefficients
\[
\pe{\tilde a_k} j = \pe{a_k}j \frac{\sum_{i=1}^d \pe{x_k} i \pe{y_j}i/\pe{\al_k} i}{\sum_{i=1}^d \pe {y_j} i}
\]
with $\pe{\al_k} i = \sum_{j=1}^p \pe{y_j} i \pe{a_k} j$.
\item Let $\tilde Y$ the matrix with coefficients
\[
\pe{\tilde y_j} i = \pe{y_j} i \frac{\sum_{k=1}^N \pe{x_k} i \pe{\tilde a_k} j/\pe{\tilde \al_k} i}{\sum_{j=1}^p \pe{\tilde a_k} j}
\]
with $\pe{\tilde \al_k} i = \sum_{j=1}^p \pe{y_j} i \pe{\tilde a_k}j$.
\item Replace $\CA$ by $\tilde \CA$ and $Y$ by $\tilde Y$, iterating until numerical convergence.
\end{enumerate}
\end{algorithm}

\section{Variational Autoencoders}
\label{sec:vae.2}

Variational autoencoders, which were described in \cref{sec:vae}, can be interpreted as non-linear factor models in which $X = g(\th, Y) + \ep$, where $\ep$ is a centered Gaussian noise with covariance matrix $Q$ and $Y\in \mR^p$ has a known probability distribution, such as $Y\sim \CN(0, \Id[p])$. In this framework, the conditional distribution of $Y$ given $X=x$ was approximated as a Gaussian distribution with mean $\mu(x, w)$ and covariance matrix $S(x, w)^2$. The implementation in \citet{kingma2014auto-encoding,kingma2019introduction} use neural networks for the three functions $g$, $\mu$ and $S$.

\section{Bayesian factor analysis and Poisson point processes}

\subsection{A feature selection model}
\label{sec:b.fact}
The expectation in many factor models is that individual observations are obtained by mixing pure categories, or topics, and represented as a weighted sum or linear combination of a small number of uncorrelated or independent variables. Let us denote by  $p$ the number of possible categories, which, in this section, can be assumed to be quite large. 

We will assume that each observation randomly selects a small number among these categories before combining them.  Let us consider (as an example) the situation in which observations $X_1, \ldots, X_N$ take the form of a probabilistic ICA model
\[
X_k = \sum_{j=1}^p {a_k}(j) b_k(j) \pe Y j  + \sig R_k, 
\]
where:
\begin{enumerate}[label=\textbullet]
\item $R_k$ follows a standard Gaussian distribution,
\item $a_k(1), \ldots, a_k(p)$ are independent with $a_k(j) \sim \CN(m_j, \tau_j^2)$,
\item $b_k(1), \ldots, b_k(p)$ are independent and follow a Bernoulli distribution with parameter $\pi_j$,
\item $\pe Y 1, \ldots, \pe Y p$ are independent standard Gaussian random variables,
\item $\sig^2$ follows an inverse gamma distribution with parameters $\al_0, \be_0$,
\item $\tau_1^2, \ldots, \tau_p^2$ follow independent inverse gamma distributions with parameters $\al_1, \be_1$,
\item $m_j$ follow a Gaussian $\CN(0, \rho^2)$ and,
\item $\pi_j$ follow a beta distribution with parameters $(u,v)$.
\end{enumerate}
 The priors are, as usually, chosen so that the computation of posterior distributions is easy, i.e., they are conjugate priors. The observed data is therefore obtained by  selecting  components $Y_j$ with probability $\pi_j$ and weighted with a Gaussian random coefficient, then added before introducing noise.

Let $n_j = \sum_{k=1}^N b_k(j)$. Ignoring constant factors, the joint likelihood of all variables together is proportional to:
\begin{align*}
L \propto& \sig^{-Nd}\exp\left(-\frac{1}{2\sig^2} \sum_{k=1}^N |X_k  - \sum_{j=1}^p a_k(j) b_k(j) \pe Y j|^2\right) \\
& \prod_{j=1}^p \left(\tau_j^{-N}\exp\left(-\frac{1}{2\tau_j^2} \sum_{k=1}^N (a_k(j)-m_j)^2\right)\right) 
\ \exp\left(-\frac{1}{2\rho^2} \sum_{j=1}^p m_j^2\right)\\
& \prod_{j=1}^p \left(\pi_j^{n_j} (1-\pi_j)^{N-n_j}\right)
\prod_{j=1}^p \left((\tau^2_j)^{-\al_1-1} \exp(-\be_1/\tau_j^2) \right)\\
&(\sig^2)^{\al_0-1} \exp(-\be_0/\sig^2)  \prod_{j=1}^p \left(\pi_j^{u-1} (1-\pi_j)^{v-1}\right) \  \exp\left(- \frac12 \sum_{i=1}^p |\pe Y i|^2\right)
\end{align*}

In spite of the complexity of this expression, it is relatively straightforward (by considering each variable in isolation) to see that 
\begin{enumerate}[label=$\bullet$]
\item The conditional distribution of $\sig^2, \tau_1^2, \ldots, \tau_p^2$ given all other variables remains a product of inverse gamma distributions. 
\item The conditional distribution of $\pe Y 1, \ldots, \pe Y p$ given the other variables is Gaussian. 
\item The conditional distribution of $\pi_1, \ldots, \pi_p$ given the other variables is a product of beta distributions. 
\item The conditional distribution of $m_1, \ldots, m_p$ given the other variables remain independent Gaussian.
\item The posterior distribution of $a_1, \ldots, a_N$ (considered as $p$-dimensional vectors) given the other variables is a product of independent Gaussian (but the components $a_k(j)$, $j=1, \ldots, p$ are correlated). 
\item For the posterior distribution  given the other variables, $b_1, \ldots, b_N$ (considered as $p$-dimensional vectors) are independent. The components of each $b_k$ are not independent but each $b_k(j)$ being a binary variable follows a Bernoulli distribution given the other ones. 
\end{enumerate}
These remarks provide a basis of a Gibbs sampling algorithm for the simulation of the posterior distribution of all unobserved variables (the computation of the parameters of each of the conditional distribution above requires some additional work, of course). This simulation does not explicitly provide a matrix factorization of the data (in the sense of a single matrix $\mathcal A$ such that $\CX=\mathcal AY$, as considered in the previous section), but a probability distribution on such matrices, expressed as $\mathcal A(k,j) = a_k(j) b_k(j)$. One can however use the average of the matrices obtained through the simulation for this purpose. Additional information can be obtained through this simulation. For example, the expectation of $b_k(j)$ provides a measure of proximity of observation $k$ to category $j$.

\subsection{Non-negative and count variables}
\label{sec:b.count}

\paragraph{Poisson factor analysis.}
Many variations can be made on the previous construction. When the observations are non-negative, for example, an additive Gaussian noise may not be well adapted. In such cases, one should model the conditional distribution of $X$ given $a$, $b$ and $Y$ as a distribution over non-negative numbers with mean $(a\odot b)^TY$ (for example a gamma distribution with appropriate parameters). However,  posterior sampling generally is more challenging with such choices because simple conjugate priors are not always available.

An important special case is when $X$ is a count variable taking values in the set of non-negative integers. In this case (starting with a model without feature selection), modeling $X$ as a Poisson variable with mean $a(1) \pe Y 1 + \cdots + a(p) \pe Y p$ leads to tractable computations, once it is noticed that $X$ can be seen as a sum of random variables $Z^{[1]}, \ldots, Z^{[p]}$ where $Z^{[i]}$ follows a Poisson distribution with parameter $a(i) \pe Y i$. This suggests introducing new latent variables $(Z^{[1]}, \ldots, Z^{[p]})$, which are not observed but follow, conditionally to their sum, which is $X$ and is observed, a multinomial distribution with parameters $X, q_1, \ldots, q_p$, with $q_i = a(i) \pe Y i /(\sum_{j=1}^p a(j)\pe Y j)$. 

This provides what is referred to as a Poisson factor analysis (PFA). As an example, consider a Bayesian approach where, for the prior distribution,  $a(1), \ldots, a(p)$ are independent and follow a gamma distribution with parameters $\al_0$ and $\be_0$,  and $\pe Y 1, \ldots, \pe Y p$ are independent, exponentially distributed with parameter $1$.  The joint likelihood of all  data then is (up to constant factors):
\begin{align*}
L \propto & \exp\left(-\sum_{k=1}^N (a_k(1) \pe Y 1 + \cdots + a_k(p) \pe Y p)\right) \left(\prod_{k=1}^N \prod_{i=1}^p \frac{(a_k(i) \pe Y i)^{z_k^{[i]}}}{z^{[i]}_k!}\right)\\
& \left(\prod_{k=1}^N \prod_{i=1}^p a_k(i)^{\al-1}\right) \exp\left(-\be \sum_{k=1}^N\sum_{i=1}^p a_k(i)\right) \exp\left(- \sum_{i=1}^p \pe Y i\right) \,.
\end{align*}
This is the GaP (For Gamma-Poisson) model introduced in \citet{canny2004gap}. The conditional distribution of the variables $(a_k(i))$ given $(Z_k^{[i]})$ and $(Y_i)$ are independent and gamma-distributed, and so are $(\pe Y i)$ given the other variables. Finally, for each $k$, the family $(Z_k^{[1]}, \ldots, Z_k^{[p]})$ follows a multinomial distribution conditionally to their sum, $X_k$, and the rest of the variables, and these variables are conditionally independent across $k$.

\paragraph{GaP with feature selection}
One can include a feature selection step in this model by introducing binary variables $b(1), \ldots, b(p)$, with selection probabilities $\pi_1, \ldots, \pi_p$, with a $\mathrm{Beta}(u,v)$ prior distribution on $\pi_i$. Doing so, the likelihood of the extended model is:
\begin{align*}
L \propto & \exp\left(-\sum_{k=1}^N (a_k(1)b_k(1) \pe Y 1 + \cdots + a_k(p) b_k(p)\pe Y p)\right) \left(\prod_{k=1}^N \prod_{i=1}^p \frac{(a_k(i) b_k(i) \pe Y i)^{z_k^{[i]}}}{z^{[i]}_k!}\right)\\
& \left(\prod_{k=1}^N \prod_{i=1}^p a_k(i)^{\al-1}\right) \exp\left(-\be \sum_{k=1}^N\sum_{i=1}^p a_k(i)\right) \exp\left(- \sum_{i=1}^p \pe Y i\right) \\
& \prod_{j=1}^p \left(\pi_j^{n_j} (1-\pi_j)^{N-n_j}\right)  \prod_{j=1}^p \left(\pi_j^{u-1} (1-\pi_j)^{v-1}\right) \,.
\end{align*}
where, as before, $n_j = \sum_{k=1}^N b_k(j)$. The conditional distribution of $\pi_1, \ldots, \pi_p$ given the other variables is therefore still that of a family of independent beta-distributed variables. The binary variables $b_k(1), \ldots, b_k(p)$ are also conditionally independent  given the other variables, with $b_k(i) = 1$ with probability one if $z_k^{[i]} > 0$ and with probability $\pi_j \exp(- a_k(j) \pe Y j)$ if $z_k^{[j]} = 0$.

\subsection{Feature assignment model}
\label{sec:indian.buffet}
The previous models assumed that $p$ features were available, modeled as $p$ random variables with some prior distribution, and that each observation picks a subset of them, drawing feature $j$ with probability $\pi_j$. We denoted by $b_k(j)$ the binary variable indicating whether feature $j$ was selected for observation $k$, and $n_j$ was the number of times that feature was selected. Finally, we modeled $\pi_j$ as a beta variable with parameters $u$ and $v$.

One can compute, using this model, the probability distribution of of the feature selection variables, $\bfb = (b_k(j), j=1, \ldots, p, k=1, \ldots, N)$. From the model definition, the probability of observing such a configuration is given by
\begin{align*}
Q(\bfb) &= \frac{\Ga(u+v)^p}{\Ga(u)^p\Ga(v)^p} \int \prod_{j=1}^p \pi_j^{n_j + u-1} (1-\pi_j)^{N-n_j + v-1} d\pi_1 \ldots d\pi_p\\
&= \prod_{j=1}^p  \frac{\Ga(u+v) \Ga(u+n_j) \Ga(v+N-n_j)}{\Ga(u)\Ga(v)\Ga(u+v+N)}\\
&= \prod_{j=1}^p \frac{u(u+1) \cdots (u+n_j-1)v(v+1)\cdots(v+N-n_j-1)}{(u+v)(u+v+1) \cdots (u+v + N-1)}
\end{align*}

Denote by $n_{jk} = \sum_{l=1}^{k-1} b_l(j)$ the number of observations with index less than $k$ that pick feature $j$. Using this notation, we can write, using the fact that
\[
u(u+1) \cdots (u+n_j-1) = \prod_{k=1}^N(u+n_{jk})^{b_k(j)}
\]
and a similar identity for  $v(v+1)\cdots(v+N-n_j-1)$, 
\begin{align*}
Q(\bfb) &= \prod_{j=1}^p \prod_{k=1}^N \frac{(u + n_{jk})^{b_k(j)} (v+k-1-n_{jk})^{1-b_k(j)}}{u+v+k-1}\\
&= \prod_{k=1}^N \prod_{j=1}^p \left(\frac{u + n_{jk}}{u+v+k-1}\right)^{b_k(j)} \left(\frac{v+k-1-n_{jk}}{u+v+k-1}\right)^{1-b_k(j)}.
\end{align*}
Using this last equation, we can interpret the probability $Q$ as resulting from a progressive feature assignment process. The first observation, $k=1$, for which $n_{jk} = 0$ for all $j$, chooses each feature with probability $u/(u+v)$. When reaching observation $k$, feature $j$ is chosen with probability $(u+n_{jk})/(u+v+k-1)$. At all steps, features are chosen independently from each other.


Let $F_k$ be the set of features assigned to observation $k$, i.e., $F_k = \{j: b_k(j) = 1\}$ and
\[
G_{k} = F_k \setminus \bigcup_{l=1}^{k-1} F_{l}
\]
be the set of features used in observation $k$ but in no previous observation.  Let $C_k = F_k \setminus G_k$ and $U_k = G_1\cap \cdots \cap G_{k-1}$.  Instead of considering configurations $\bfb = (b_k(j), i=1, \ldots, d, k=1, \ldots, N)$ we may alternatively consider the family of sets
$\bfS = (G_k, C_{k}, 1\leq  k \leq N)$.  Such a family must satisfy the property that the sets $G_{k}$ and $C_{k}$ are non-intersecting, 
$C_k \subset U_k$ and $G_l\cap G_k = \emptyset$ for $l<k$. It
provide a unique configuration $\bfb$ by letting $b_k(j) = 1$ if and only if $j\in G_k\cup C_k$. 
Letting $q_k = |G_k|$ and $p_k = |U_k|$, the probability $Q(\bfb)$ can be re-expressed in terms of $\bfS$, letting (with some abuse of notation)
\begin{multline*}
Q(\bfS) = \prod_{k=1}^N \Bigg(
 \left(\frac{u}{u+v+k-1}\right)^{q_k} \left(\frac{v+k-1}{u+v+k-1}\right)^{p-p_{k+1}} \\
\prod_{j\in U_k} \left(\frac{u + n_{jk}}{u+v+k-1}\right)^{\bfone_{j\in C_k}} \left(\frac{v+k-1-n_{jk}}{u+v+k-1}\right)^{\bfone_{j\not\in C_k}} \Bigg).
\end{multline*}
Let $\bfS_k = (G_l, C_l, l\leq k)$. Then the expression of $Q$ shows that, conditionally to $\bfS_{k-1}$, $G_k$ and $C_k$ are independent. Elements in $C_k$ are chosen independently for each feature $j\in U_k$ with probability $(u + n_{jk})/(u+v+k-1)$.  Moreover, the conditional distribution of $q_k$ given $\bfS_{k-1}$ is proportional to
\[
 \left(\frac{u}{u+v+k-1}\right)^{q_k} \left(\frac{v+k-1}{u+v+k-1}\right)^{p-p_{k} - q_k}
 \]
i.e., it is a binomial distribution with parameters $p-p_k$ and $u/(u+v+k-1)$. Finally, given $\bfs_{k-1}$ and $q_k$, the distribution of $G_k$ is uniform among all  $\binom{p-p_k}{q_k}$ subsets of
\[
\{1, \ldots, p\} \setminus (G_1\cup \cdots\cup G_{k-1})
\]
with cardinality $q_k$.

If there is no special meaning in the feature label, which is the case in our discussion of prior models in which all features are sampled independently with the same distribution, we may identify configurations that can be deduced from each other by relabeling (note that relabeling features does not change the value of $Q$). 

Call a configuration normal if $G_k = \{p_k+1, \ldots, p_{k+1}\}$.  Given $\bfS$, it is always possible to relabel the features with a permutation $\sigma$ so that, for each $k$,  $\sigma(G_k) = \{p_k+1, \ldots, p_{k+1}\}$. There are, in fact,  $q_1! \ldots q_N!$ such permutations. We can complete the process generating $\bfS$ by adding at the end a transformation into a normal configuration (picking uniformly at random one of the possible ones). The probability of a normal configuration $\bfS$ obtained through this process is (using a simple counting argument)
 \begin{multline*}
Q(\bfS) = \prod_{k=1}^N \Bigg(\binom{p-p_k}{q_k}
 \left(\frac{u}{u+v+k-1}\right)^{q_k} \left(\frac{v+k-1}{u+v+k-1}\right)^{p-p_{k+1}} \\
\prod_{j\in U_k} \left(\frac{u + n_{jk}}{u+v+k-1}\right)^{\bfone_{j\in C_k}} \left(\frac{v+k-1-n_{jk}}{u+v+k-1}\right)^{\bfone_{j\not\in C_k}} \Bigg),
\end{multline*}


This provides a new incremental procedure that directly samples normalized assignments. First let $q_1$ follow a binomial distribution $\mathrm{bin}(p, u/(u+v))$ and assign the first observation to features $1$ to $q_1$. Assume that $p_k$ labels have been created before step $k$. Then select for observation $k$ some of the already labeled features, each label $j$ being selected with probability $(u+n_{jk})/(u+v+k-1)$ as above. Finally, add $q_k$ new features where $q_k$ follows a binomial distribution $\mathrm{bin}(p-p_k, u/(u+v+k-1))$.

This discussion is clearly reminiscent of the one that was made in \cref{sec:cluster.npb} leading to the Polya urn process, and we want here also to let $p$ tend to infinity (with fixed $N$) with proper choices of $u$ and $v$ as functions of $p$ in the expression above. Choose two positive numbers $c$ and $\ga$ and let $u = c\ga/p$ and $v = c - u$. Note that, with the incremental simulation process that we just described, the conditional expectation of the next number of labels, $p_{k+1}$, given the current one, $p_k$, is
\[
E(p_{k+1}\mid p_k) = \frac{(p-p_k)u}{u+v+k-1} + p_k = \frac{c\ga}{c+k-1} + \left(1- \frac{c\ga}{p(c+k-1)}\right) p_k\leq \frac{c\ga}{c+k-1} + p_k .
\]
Taking expectations on both sides, we get 
\[
E(p_{k+1}) \leq \sum_{l=1}^k \frac{c\ga}{c+l-1} \leq  \sum_{l=1}^N \frac{c\ga}{c+l-1}
\]
so that this expectation is bounded independently of $k$. This shows in particular that $p_k/p$ tends to 0 in probability (just applying Markov's inequality) and that the binomial distribution $\mathrm{bin}(p-p_k, u/(u+v+k-1))$ can be approximated by a Poisson distribution with parameter $c\ga / (c+k-1)$. 

So, when $p\to \infty$, we obtain the following incremental simulation process for the feature labels, that we combine with the actual simulation of the features, assumed to follow a prior distribution with p.d.f. $\psi$. This process is called the Indian buffet process in the literature, the analogy being that a buffet offers an infinite variety of dishes, and each observation is a customer who tastes a finite number of them. 

\begin{algorithm}[Indian buffet process]
\label{alg:indian.buffet}
\begin{enumerate}[label=\arabic*., wide=0.25cm]
\item Initialization:
\begin{enumerate}[label=(\roman*), wide=1cm]
\item Sample an integer $q_1$ according to a Poisson distribution with parameter $\ga$.
\item Sample features $\pe y 1, \ldots, \pe y {q_1}$ according to $\psi$.
\item Assign these features to observation 1, and let $n_{2,j} = 1$ for $j=1, \ldots, q_1$.
\end{enumerate}
\item Assume that observations 1 to $k-1$ have been obtained, with $p_k$ features $\pe y 1, \ldots, \pe y {p_k}$ such that the $j$th feature has been chosen $n_{k,j}$ times.
\begin{enumerate}[label=(\roman*), wide=1cm]
\item For $j=1, \ldots, p_k$, assign feature $j$ to sample $k$ with probability $\frac{n_{k,j}}{c+k-1}$. If $j$ is selected, let $n_{k+1,j} = n_{k,j} +1$, otherwise let $n_{k+1,j} = n_{k,j}$.
\item Sample an integer $q_k$ according to a Poisson distribution with parameter $\frac{c\ga}{c+k-1}$ and let $p_{k+1} = p_k + q_k$.
\item Sample features $\pe y {p_k+1}, \ldots, \pe y {p_{k+1}}$ according to $\psi$.
\item Assign these features to observation k, and let $n_{k+1,j} = 1$ for $j=p_k+1, \ldots, p_{k+1}$.
\end{enumerate}
\item If $k=N$, stop, otherwise replace $k$ by $k+1$ and return to Step 2.
\end{enumerate}
\end{algorithm}

\section{Point processes and random measures}

\subsection{Poisson processes}

If $\CZ$ is a set, we will denote by $\CP_c(\CZ)$ the set composed with all finite or countable subsets of $\CZ$. 
A point process over $\CZ$ is a random variable $S: \Om \to \CP_c(\CZ)$, i.e., a variable that provides a countable random subset of $\CZ$. If $B\subset \CZ$ one can then define the counting function $\nu_S(B) = |S \cap B| \in \mZ\cup\{+\infty\}$.

A proper definition of such point processes requires some measure theory. Equip $\CZ$ with a $\sig$-algebra $\mA$ and consider the set $\CN_0$ of integer-valued measures $\mu$ on $(\CZ, \mA)$ such that $\mu(\CZ) <\infty$. Let $\CN$ be the set formed with all countable sums of measures in $\CN_0$. Then a general point process is a mapping $\nu: \Om \to \CN$ such that for all $k\in  \mN\cup\{+\infty\}$ and  all $B\in \mA$, the event $\{\nu(B) = k\}$ is measurable. Recall that, for each $B\in \mA$, $\nu(B)$ is itself a random variable, that we may denote $\om \mapsto \nu_\om(B)$. One then define the intensity of the process as the the function $\mu : B \mapsto \myE(\nu(B))$.

The following proposition provides an important identity satisfied by such models.
\begin{theorem}[Campbell identity]
\label{th:campbell}
Let $\nu$ be a point process with intensity $\mu$. For $\om\in \Om$, let $X_\om: \Om' \to \CZ$ be a random variable with distribution $\nu_\om$ (defined, if needed, on a different probability space $(\Om', \myP')$). Then, for any $\mu$-integrable function $f$:
\begin{equation}
\label{eq:campbell}
\myE(f(X)) = \int_{\CZ} f(z) d\mu(z).
\end{equation}
\end{theorem}
Here, the expectation of $f(X)$ is over both spaces $\Om$ and $\Om'$ and corresponds to the average of $f$. 
The identity is an immediate consequence of Fubini's theorem.

We will be mainly interested in the family of Poisson point processes. These processes are themselves parametrized by a measure, say $\mu$, on $\CZ$ such that $\mu$ is $\sig$-finite and $\mu(B) = 0$ if $B$ is a singleton. A Poisson process with intensity measure $\mu$ is a point process $\nu$ such that: 
\begin{enumerate}[label = (\roman*)]
\item If $B_1, \ldots , B_n$ are non-intersecting pairwise, then $\nu(B_1), \ldots, \nu(B_n)$ are mutually independent.
\item for all $B$, $\nu(B) \sim \mathit{Poisson}(\mu(B))$. 
\end{enumerate}
We take the convention that $\nu(B) = 0$ (resp. $=\infty$)  almost surely if $\mu(B)=0$ (resp. $=\infty$). Note that  property (i) also implies that if $g_1, \ldots, g_n$ are measurable functions from $\CZ$ to $(0, +\infty)$ such that $g_ig_j = 0$ for $i\neq j$, then the variables $\nu(g_i) = \int_{\CZ} g_i(z) d\nu(z)$ are independent.

If $\mu(\CZ)<\infty$ (i.e., $\mu$ is finite), one can represent the distribution of a Poisson point process as
\[
\nu = \sum_{k=1}^{\nu(\CZ)} \de_{X_k}
\]
with
 $\nu(\CZ)\sim \mathit{Poisson}(\mu(\CZ))$ and, conditionally to $\nu(\CZ) = N$, $X_1, \ldots, X_N$ are i.i.d. and follow the probability distribution $\bar\mu = \mu /\mu(\CZ)$. This measure can also be identified with the random set $S = \{X_1, \ldots, X_{\nu(\CZ)}\}$. The assumption that $\mu(\{z\}) = 0$ for any singleton implies that $\nu(\{z\}) = 0$ almost surely. It also ensures that the points $X_1, \ldots, X_N$ are distinct with probability one.
 
 If $\mu$ is $\sig$-finite, then (by definition), it is a countable sum of finite measures $\mu_1, \mu_2, \ldots$. Then $\nu$ can  be generated as the sum of independent $\nu_1, \nu_2, \ldots$, where $\nu_i$ is a Poisson process with intensity $\mu_i$. It can moreover be identified with the countable random set
 $ S = \bigcup_{i=1}^\infty S_i$, where $S_i$ is the random set associated with $\nu_i$. Note that, in this construction, one can always assume that the measures $\mu_1, \mu_2, \ldots$ are mutually singular (i.e., $\mu_i(B) > 0$ for some $i$ implies that $\mu_j(B) = 0$ for $j\neq i$).

By considering Poisson processes on $(0, +\infty) \times \CZ$, one can define weighted random measures.  Indeed, such a point process takes values in the collection of all sets of the form $\{(w_k, z_k), k\in I \}$ where $I$ is finite or countable. These subsets can be represented as the  sum of weighted Dirac masses, 
\[
\xi = \sum_{k\in I} w_k \de_{z_k}\,.
\]
To ensure that the points $(z_k, k\in I)$ generated by this process are all different, we need to assume that the intensity $\mu$ of this random process is such that $\mu((0, +\infty) \times \{z\}) = 0$ for all $z\in \CZ$. We will refer to $\xi$ as a weighted Poisson process. 


 In the following, we will consider this class of random measures, with the small addition of allowing for an extra term including a measure supported by a fixed set. More precisely, given a (deterministic) countable subset  $\CI \sub \CZ$, a family of independent random variables $(\rho_z, z\in\CI)$ and a $\sig$-finite measure $\mu_o$ such that  $\mu_o((0, +\infty) \times \{z\}) = 0$ for all $z\in \CZ$, one can define the random measure
 \[
 \xi = \xi_f + \xi_o
 \]
 where $\xi_o$ is  a weighted Poisson process with intensity $\mu_o$, assumed independent of $(\rho_z, z\in \CI)$ and
 \[
 \xi_f = \sum_{z\in \CI} \rho_z \de_z.
 \]
 The subscripts $o$ and $f$ come from the terminology introduced in \citet{kingman1967completely}, which studies  ``completely random measures,'' which are a random measures 
that satisfy point (i) in the definition of a Poisson process. Under mild assumptions, such measures can be decomposed as a sum of a weighted Poisson process (here, $\xi_o$, the ordinary part), of a process with fixed support, (here, $\xi_f$, the fixed part) and of a deterministic measure (which is here taken to be 0).

Let us rapidly check that $\xi$ satisfies property (i). Let $B_1, \ldots, B_n$ be non-overlapping elements of $\mA$. Get $g_i(w, z) = w \bfone_{B_i}(z)$. Then
\[
\xi(B_i) = \xi_f(B_i) + \nu_o(g_i) 
\]
where $\nu_o$ is a Poisson process with intensity $\mu_o$. Since the sets do not overlap, the variables $(\xi_f(B_i), i=1, \ldots, n)$ are independent, and so are $(\nu_o(g_i), i=1, \ldots, n)$ since $g_ig_j = 0$ for $i\neq j$. Since $\xi_f$ and $\nu_o$ are, in addition independent, we see that $(\xi(B_i), i=1, \ldots, n)$ are independent.

 The intensity measure of such a process is still defined by 
\[
\eta(B) = E(\xi(B)) =  \sum_{z\in \CI} P( (\rho_z, z) \in B) +  \int_{(0, +\infty)\times B} w d\mu_o(w, z)
\]
where the last term is an application of Campbell's inequality to the Poisson process $\nu_o$ and the function $g(w,x) = w\bfone_B(x)$.

\subsection{The gamma process}
The main example of such processes in factor analysis is the beta process that will be discussed in the next section. We start, however, with a first example that is closely related with the Dirichlet process, called the gamma process.

In this process, one fixes a finite measure $\pi_0$ on $\CZ$ and defines  $\mu$ on $(0, +\infty) \times \CZ$ by
\[
\mu(dw, dz) = cw^{-1} e^{-cw} \pi_0(dz) dw.
\]
Because $\mu$ is $\sig$-finite but not finite (the integral over $t$ diverges at $t=0$), every realization of $\xi$ is an infinite sum
\[
\xi = \sum_{k=1}^\infty w_k \de_{z_k}.
\]
The intensity measure of $\xi$ is 
\[
\eta(B) = c \pi_0(B) \int_0^{+\infty} e^{-cw}dw = \pi_0(B).
\]
In particular, 
\[
\sum_{k=1}^\infty w_k = \eta(\CZ) = \pi_0(\CZ) < \infty.
\]

For fixed $B$, the variable $\xi(B)$ follows a Gamma distribution. This can be proved by computing the Laplace transform of $\xi$,
$E(e^{-\la \xi(B)})$, and identify it to that of a Gamma. To make this computation, consider the point process $\nu_J$ restricted to a interval $J\subset (0, +\infty)$ with $\min(J)>0$, and $\xi_J$ the corresponding weighted process. Let $m_J(t) = \int_J w^{-1} ce^{-(c+t)w}\, dw$. Then a realization of $\nu_J$ can be obtained by first sampling $N$ from a Poisson distribution with parameter $\mu(J\times \CZ)= m_J(0) \pi_0(\CZ)$ and then sampling $N$ points $(w_i, z_i)$ independently from the distribution $\mu/(m_J(0)\pi_0(\CZ))$. This implies that
\begin{align*}
E(e^{-t\xi_J(B)}) &= \sum_{n=0}^\infty e^{-m_J(0) \pi_0(\CZ)} \frac{(m_J(0) \pi_0(\CZ))^n}{n!} \left(\frac{\int_0^\infty e^{-tw\bfone_B(z)} w^{-1} ce^{-cw} dw d\pi_0(z)}{m_J(0)\pi_0(\CZ)}\right)^n\\
&= \sum_{n=0}^\infty \frac{e^{-m_J(0) \pi_0(\CZ)}}{n!} \left(\pi_0(B) m_J(t) + (\pi_0(\CZ) -\pi_0(B))m_J(0)\right)^n\\
&= e^{\pi_0(B) (m_J(t) - m_J(0))}\,.
\end{align*}
Now, 
\[
m_J(t) - m_J(0) = c\int_J e^{cw} \frac{e^{-tw}-1}{w} dw
\]
is finite even when $J= (0, +\infty)$. With a little more work justifying passing to the limit, one finds that, for $J= (0, +\infty)$,
\[
E(e^{-t\xi_J(B)}) = \exp\left(\pi_0(B) \int_0^{+\infty} e^{-cw} \frac{e^{-tw}-1}{w} dw\right).
\]
Finally, write
\begin{align*}
c\int_0^{+\infty} e^{-cw} \frac{e^{-tw}-1}{w} dw &= -c\int_0^{+\infty} e^{-cw} \int_0^t e^{-sw} ds dw \\
&= -c\int_0^{t} \int_{0}^{+\infty} e^{-(s+c)w} dw ds\\
&= -\int_0^{t} c(s+c)^{-1} ds = -c\log(1+\frac{t}{c}).
\end{align*}
This shows that 
\[
E(e^{-t\xi_J(B)}) = \left(1+\frac{t}{c}\right)^{-c\pi_0(B)}
\]
which is the Laplace transform of a Gamma distribution with parameters $c\pi_0(B)$ and $c$, i.e., with density proportional to $w^{c\pi_0(B)-1} e^{-cw}$.

As a consequence, the normalized process $\delta = \xi/\xi(\CZ)$ is a Dirichlet process with intensity $c\pi_0$. Indeed, if $B_1, \ldots, B_n$ is a partition of $\CZ$ the family $(\delta(B_1), \ldots, \delta(B_n))$ is the ratio of $n$ independent gamma variables to their sum, which provides a Dirichlet distribution, and this property characterizes Dirichlet processes.

\subsection{The beta process}

The definition of the beta process parallels that of the gamma process, with weights taking this time values in $(0,1)$. Fix again a finite measure $\pi_0$ on $\CZ$ and let  $\mu_o$ on $(0, +\infty) \times \CZ$ be defined by
\[
\mu_o(dw, dz) = cw^{-1} (1-w)^{c-1}  \pi_0(dz) dw.
\]
The associated weighted Poisson process can therefore be represented as a sum
\[
\xi_o = \sum_{k=1}^\infty w_k \de_{z_k},
\]
and its intensity measure is
\[
\eta_o(B) = c \pi_0(B) \int_0^{1} (1-t)^{c-1} dw = \pi_0(B).
\]
In particular, since $\pi_0$ is finite, we have $\sum_{k=1}^\infty w_k <\infty$ almost surely. A beta process is the sum of the process $\xi_o$ and of a fixed set process
\[
\xi_f = \sum_{z\in \CI} w_z \de_z
\]
where $\CI$ is a fixed finite set and $(w_z, z\in\CI)$ are independent and follow a beta distribution with parameters $(a(z), b(z))$.

If $\CZ$ is a space of features, such a process provides a prior distribution on feature selections. It indeed provides, in addition to the deterministic set $\CI$, a random countable set $\CJ\sub \CZ$, with a set of random weights $w_z$, $z\in \CF:= \CI\cup \CJ$. Given this, one defines the feature process as the selection of a subset $A \sub \CF$ where each feature $z$ is selected with probability $w_z$. Because $E(|A|) = \sum_{z\in \CF} w_z$ is finite, $A$ is finite with probability 1.

In the same way the Polya urn could be used to sample from a realization of a Dirichlet process  without actually sampling the whole process, there exists an algorithm that samples a sequence of feature sets $(A_1, \ldots, A_n)$ from this feature selection process without needing the infinite collection of weights and features associated with a beta process. We assume in the following that the prior process has an empty fixed set. (Non-empty fixed sets will appear in the posterior.)

The first set of features, $A_1$, is obtained as follows according to a Poisson process with intensity $\pi_0$: choose the number $N$ of  features in $A_1$ according to a Poisson distribution with parameter $\pi_0(\CZ)$; then sample $N$ features independently according to the distribution $\pi_0/\pi_0(\CZ)$. 

Now assume that $n$ sets of features $A_1, \ldots, A_{n}$ have been obtained and we want to sample a new set $A_{n+1}$ conditionally to their observation. Let $\CJ_{n}$ be the union of all random features obtained up to this point and $n(z)$, for $z\in \CJ_{n}$ the number of times this feature was observed in $A_1, \ldots, A_{n}$. 
Then the conditional distribution of the beta process $\xi$ given this observation is still a beta process, with fixed set given by
$\CI = \CJ_{n}$,  $(a(z), b(z)) = (n(z), c+n - n(z))$ for $z\in \CJ_{n-1}$ and base measure $\pi_{n} = c\pi_0/(c+n)$. This implies that the next set $A_{n+1}$ can be obtained by sampling from the associated feature process. To do this, one first selects features $z\in \CJ_n$ with probability $n(z)/(c+n)$, then selects additional features $z_1, \ldots, z_N$ independently with distribution  $\pi_0/\pi_0(\CZ)$ where $N$ follows a Poisson distribution with parameter $c\pi_0(\CZ)/(c+n)$. 
This is the Indian buffet process, described in  \cref{alg:indian.buffet} (taking  $\pi_0 = \ga \psi$).

\subsection{Beta process and feature selection}
The beta process can be used as a prior for feature selection within a factor analysis model, as described in the previous paragraph. It is however easier to approximate it with a model with almost surely finite support. Indeed, letting, for $\ep >0$
\[
\mu(dw, dz) = \frac{\Gamma(c + 1)}{\Gamma(\ep+1)\Gamma(c-\ep)} w^{\ep - 1} (1-w)^{c-\ep} \pi_0(dz) dw,
\]
one obtains a finite measure since
\[
\int_0^{+\infty} \int_{\CZ} \mu_o(dw, dz) = \frac{c\ga}{\ep}
\]
where $\ga = \pi_0(\CZ)$. Note that $\mu$ is normalized so that $E(\xi(B)) = \pi_0(B)$ for $B\sub \CZ$. 

In this case, the prior generates features by first sampling their number, $p$, randomly according to  a Poisson distribution with mean $c\ga/\ep$, then select $p$ probabilities  $w_1, \ldots, w_p$ independently using a beta distribution with parameters $\ep$ and $c-\ep$, and finally attach to each $i$ a feature $z_i$ with  distribution $\pi_0/\gamma$. The features associated with a given sample are then obtained by selecting each $z_i$ with probability $w_i$.

The model described in \cref{sec:indian.buffet} provides an approximation of this prior using a  finite number of features. With our notation here, this corresponds to taking $p\gg 1$ and $\ep = c\ga/p$.

\newpage
\markright{PROBLEMS}
\section*{Problems}
\addcontentsline{toc}{section}{Problems}
\input Problems_Factor_Analysis


\chapter{Data Visualization and Manifold Learning}
\label{chap:manifold.learning}
The methods described in this chapter aim at computing low-dimensional representations of possibly high-dimensional, allowing for its visual exploration by summarizing its structure in a user-accessible interface. Although similar in purpose to factor analysis methods, the accent here is to provide an interpretable representation of the data, rather than a generative model. (In encoder/decoder language, factor analysis is more about decoding, which data visualization is about encoding, with some approaches providing both.)

Data visualization methods are in general optimized for a given dataset, and do not always provide a direct mechanism for adding new data into the representation. 
The methods presented in this chapter take as input similarity or dissimilarity matrices between data points and do not require, say, Euclidean coordinates. This is in particular the case for multidimensional scaling, described in the next section.

\section{Multidimensional scaling}

Let $D = (d_{kl}, \,k,l=1, \ldots, N)$ be the given dissimilarity matrix. The goal of multidimensional scaling (MDS) is to determine a small-dimensional Euclidean representation, say $y_1, \ldots, y_N\in \mR^p$, such that $\abs{y_k - y_l}^2  \simeq
d_{kl}^2$. We review below two versions of this algorithm, referred to as ``similarity'' and ``dissimilarity'' matching.

\subsection{Similarity matching (Euclidean case)}
A standard hypothesis of  MDS assumes that the distances $d_{kl}$ derive from a representation in feature space, so that $d_{kl}^2 = \|h_k - h_l\|_H^2$ for some inner-product space $H$ and (typically unknown) features $h_1, \ldots, h_N$. Making this assumption, we note that, since the Euclidean distance is invariant by translation, there is no loss of generality in taking $h_1 + \cdots + h_N=0$, which will be done in the following.

We look for a $p$-dimensional representation in the form  $y_k = \Phi h_k$ where $\Phi$ is a linear transformation (and we want $y_k$ to be computable directly from dissimilarities, since we do not assume that $h_k$ is known). Since we are only interested in a transformation of  $h_1, \ldots, h_N$, it suffices to compute $\Phi$ in the vector space generated by them, so that we let
\[
\Phi: \mathrm{span}(h_1, \ldots, h_N) \to \mR^p\,,
\]
and we want $\Phi$ to
(approximately) conserve the norm, i.e., be close to being an isometry. 

Because isometries are one-to-one and onto, the existence of an exact isometry would require $V \defeq \mathrm{span}(h_1, \ldots, h_N)$ to be $p$-dimensional. The mapping $\Phi$ could then  be defined as $\Phi(h) = (\scp{h}{e_1}_H, \ldots, \scp{h}{e_p}_H)$ where $e_1, \ldots, e_p$ is any orthonormal basis of $V$. In the general case, however, where the dimension of $V$ is larger than $p$, the problem is  to find a best $p$-dimensional approximation of the training data, leading to a problem similar to PCA in feature space (\cref{sec:kernel.pca}).

Indeed, as we have seen in  \cref{sec:pca.comp}, this best approximation can be obtained by diagonalizing the Gram matrix $S$ of $h_1, \ldots, h_N$, which is such that  $s_{kl} = \scp{h_k}{h_l}_H$. (Recall that we assume that $\bar h = 0$, so we do not center the data.) Using the notation in \cref{sec:pca.comp}, let $\pe \al 1, \ldots, \pe \al p$ denote the eigenvectors associated with the $p$ largest eigenvalues of $S$, normalized so that $(\pe \al i)^TS\pe \al i = 1$ for $i=1, \ldots, p$. One can then take
\[
e_i = \sum_{l=1}^N \pe{\al_{l}}i h_l
\]
and, for $k=1, \ldots, N$, $j=1, \ldots, p$:
\begin{equation}
\label{eq:mds.sol}
\pe {y_{k}} i = \la_i^2 \pe{\al_{k}} i
\end{equation}
where $\la_i^2$ is the eigenvalue associated with $\pe \al i$.

This does not entirely address the original problem, since the inner products $s_{kl}$ are not given, but only the  distances $d_{kl}$, which satisfy
\begin{equation}
\label{eq:s.to.d}
 d_{kl }^2 = - 2 s_{kl} + s_{kk} +s_{ll}\,.
\end{equation}
One therefore needs to solve a linear system of equations in the unknown $s_{kl}$.  
This system is under-determined, because  $D$ is invariant by any transformation $h_k \mapsto h_k + h_0$ (for a fixed $h_0$), and $S$ is not. However, the assumption $h_1+\cdots+h_N=0$ provides the additional constraint needed to provide a unique solution. 
Summing  \cref{eq:s.to.d} over $l$, we then get 
\begin{equation}
\label{eq:s.to.d.2}
\sum_{l=1}^N d_{kl}^2 = Ns_{kk} + \sum_{l=1}^N s_{ll}\,.
\end{equation}
Summing this equation over $k$, we find 
\[
\sum_{k,l=1}^N d_{kl}^2 = 2N \sum_{l=1}^N s_{ll}.
\]
Using this in \cref{eq:s.to.d.2}, we get 
\[
s_{kk} = \frac1N\sum_{l=1}^N d_{kl}^2 - \frac1{2N^2} \sum_{k,l=1}^N d_{kl}^2,
\]
and, from \cref{eq:s.to.d}
\[
s_{kl} = -\frac12\left(d_{kl}^2 - \frac1N\sum_{k'=1}^N d_{k'l}^2 - \frac1N\sum_{l'=1}^N d_{kl'}^2 + \frac1{N^2} \sum_{k',l'=1}^N d_{k'l'}^2\right).
\]
If we denote by $D^{\odot 2}$ the matrix formed with the squared distances $d^2_{kl}$, this identity can we rewritten in the simpler form  
\begin{equation}
\label{eq:dissim.sim}
S = -\frac12 P D^{\odot 2} P
\end{equation}
with $P = \Id[N] -\dsone_N \dsone_N^T/N$.

We now show that this PCA-based approach to MDS is equivalent to the problem of minimizing
\begin{equation}
\label{eq:sim.match}
F(y) = \sum_{k,l=1}^N (y_k^Ty_l - s_{kl})^2 
\end{equation}
over all $y_1, \ldots, y_N\in \mR^p$ such that $y_1 + \cdots + y_N = 0$, 
which can be interpreted as matching ``similarities'' $s_{kl}$ rather than distances. Indeed, letting $\CY$ denote the $N$ by $p$ matrix with rows $y_1^T, \ldots, y_N^T$, we have
\[
F(y) = \trace( (\CY\CY^T - S)^2).
\]
Finding $\CY$ is equivalent to finding a symmetric matrix $M$ of rank $p$ minimizing $\trace((M-S)^2)$.
 We have, using the trace inequality (\cref{th:trace.ineq}), and letting $\la_1^2 \geq \cdots \geq \la_N^2$  (resp. $\mu_1^2 \geq \cdots \geq \mu_p^2$) denote the eigenvalues of $S$ (resp. $M$),
\begin{align*}
\trace((M-S)^2) & = \trace(M^2) - 2\trace(MS) + \trace(S^2) \\
&= \sum_{k=1}^p \mu_k^4 - 2\trace(MS) + \sum_{k=1}^N \la_k^4\\
&\geq \sum_{k=1}^p \mu_k^4 - 2\sum_{k=1}^p \la^2_k\mu^2_k + \sum_{k=1}^N \la_k^2\\
&= \sum_{k=1}^p (\la^2_k-\mu^2_k)^2 + \sum_{k=p+1}^N \la_k^4\\
&\geq  \sum_{k=p+1}^N \la_k^4
\end{align*}
This lower bound is attained when $M$ and $S$ can be diagonalized in the same orthonormal basis with $\la^2_k=\mu^2_k$ for $k=1, \ldots, p$. So, letting $S = UDU^T$, where $U$ is orthogonal and $D$ is diagonal with decreasing numbers on the diagonal, an optimal $M$ is given by $M = U_{\lceil d,p\rceil} D_{\lceil p,p\rceil} U_{\lceil d,p\rceil}^T$, where $U_{\lceil d,p\rceil}$ is formed with the first $p$ columns of $U$ and $D_{\lceil d,p\rceil}$ is the first $p\times p$ block of $D$. This shows that the matrix $\CY = U_{\lceil d,p\rceil} D_{\lceil p,p\rceil}^{1/2}$ provides a minimizer of $F$.
The matrix $U = [\pe u 1, \ldots, \pe u N]$ differs from the matrix $A = [\pe \al 1, \ldots, \pe \al N]$ above through the normalization of its column vectors: we have $S\pe \al i = \la^2_i \pe \al i$ with $(\pe \al i)^T S \pe \al i = 1$ while $S\pe u i = \la^2_i \pe u i$ with $(\pe u i)^T S \pe u i = \la^2_i$ showing that $\pe \al i = \la_i^{-1} \pe u i$. This shows that $A_{\lceil d,p\rceil} = U_{\lceil d,p\rceil} D_{\lceil p,p\rceil}^{-1/2}$ so that $\CY$ can also be rewritten as $\CY = A_{\lceil d,p\rceil} D_{\lceil p,p\rceil}$, i.e., $\pe {y_{k}} i =  \la_i^2 \pe {\al_{k}} i$ This gives the same expression as the one that was obtained before.

The minimization of $F$ is called {\em similarity matching}. Clearly, this method can be applied when one starts directly with a matrix of dissimilarities $S$, provided it satisfies $\sum_{l=1}^N s_{kl}=0$ for all $k$. If this is not the case, then interpreting $s_{kl}$ as an inner product $h_k^Th_l$, it is natural to replace $s_{kl}$ by what would give $(h_k - \bar h)^T(h_l - \bar h)$, namely, by 
\[
s'_{kl} = s_{kl} - \frac1N \sum_{l'=1}^N s_{kl'} - \frac1N \sum_{ k'=1}^N s_{k' l} + \frac1{N^2} \sum_{k', l'=1}^N s_{k'l'}.
\]  
Interestingly, this last discussion provided us with yet another interpretation of PCA.

\subsection{Dissimilarity matching}
While the minimization of \cref{eq:sim.match} did not provide us with a new way of analyzing the data (since it was equivalent to PCA), the direct comparison of dissimilarities, that is, the minimization of 
\[
G(y) = \sum_{k,l=1}^N (|y_k-y_l| - d_{kl})^2
\]
over $y_1, \ldots, y_N\in \mR^p$, provides a different approach.  Because this may be useful in practice and does not bring in much additional difficulty, we will allow for the possibility of weighting the differences in $G$ and consider the minimization of 
\[
G(y) = \sum_{k,l=1}^N w_{kl}(|y_k-y_l| - d_{kl})^2
\]
where $W = (w_{kl})$ is a symmetric matrix of non-negative weights. The only additional complexity resulting from adding weights is that the indeterminacy on $y_1, \ldots, y_N$ is that $G(y)=G(y')$ as soon as $y-y'$ is constant on every connected component of the graph associated with the weight matrix $W$, so that the constraint on $y$ should be replaced by
\[
\sum_{k\in \Ga} y_k = 0
\]
for any connected component $\Ga$ of this graph. (If all weights are positive, then the only connected component is $\{1, \ldots, N\}$ and we retrieve our previous constraint  $\sum_{k=1}^N y_k = 0$.)

Standard nonlinear optimization methods, such as projected gradient descent, may be used to minimize $G$, but the preferred algorithm for MDS uses a stepwise procedure resulting from the addition of an auxiliary variable.
Rewrite
\[
G(y) = \sum_{k,l=1}^N w_{kl} |y_k-y_l|^2 - 2 \sum_{k,l=1}^N w_{kl} d_{kl} |y_k-y_l| + \sum_{k,l=1}^N d_{kl}^2\,.
\]
We have,  for $u\in \mR^p$: 
\[
|u| = \max\{z^T u: z\in \mR^p, |z| = \bfone_{u\neq 0}\}\,.
\]
Using this identity, we can  introduce auxiliary variables $z_{kl}$, $k,l=1, \ldots, N$ in $\mR^p$, with $|z_{kl}| = 1$ if $y_k\neq y_l$, and  define
\[
\hat G(y,z) = \sum_{k,l=1}^N w_{kl} |y_k-y_l|^2 - 2 \sum_{k,l=1}^N w_{kl} d_{kl} (y_k-y_l)^T z_{kl} + \sum_{k,l=1}^N d_{kl}^2\,.
\]
We then have
\[
G(y) = \min_{z: |z_{kl}|=1 \text{ if } y_k\neq y_k} \hat G(y,z).
\]
As a consequence,  minimizing $G$ in $y$ can be achieved by minimizing $\hat G$ in $y$ and $z$ and discarding $z$ when this is done. One can minimize $\hat G$ iteratively, alternating minimization in $y$ given $z$ and in $z$ given $y$, both steps being elementary. In order to describe these steps, we introduce some matrix notation. 

Let $L$ denote the Laplacian matrix of the weighted graph on $\{1, \ldots, N\}$ associated with the weight matrix $W$, namely $L = (\ell_{kl}, k,l=1, \ldots, N)$ with $\ell_{kk} = \sum_{k=1}^N w_{kl} - w_{kk}$ and $\ell_{kl} = -w_{kl}$ when $k\neq l$. Then, 
\[
\sum_{k,l=1}^N w_{kl} |y_k-y_l|^2 = 2 \trace( \CY^T L \CY).
\]
Defining $u_k\in \mR^p$ by
\[
u_k = \sum_{l=1}^N w_{kl} d_{kl} (z_{kl}-z_{lk}),
\]
and $\CU = \begin{pmatrix} u_1^T\\ \vdots \\ u_N^T\end{pmatrix}$, we have
\[
\sum_{k,l=1}^N w_{kl} d_{kl} (y_k-y_l)^T z_{kl} = \trace (\CU^T \CY).
\]
With this notation, the optimal matrix $\CY$ must minimize
\[
2 \trace( \CY^T L \CY) - 2 \trace(\CU^T \CY). 
\]

Let $m$ be the number of connected components of the weighted graph.
Recall that the matrix $L$ is positive semi-definite and that an orthonormal basis of its null space is provided by vectors, say $e_1, \ldots, e_m$, that are constant on each of the $m$ connected components of the graph, so that the constraint on $\CY$ can be written as $e_j^T\CY = 0$ for $j=1, \ldots, m$. Introduce the matrix
\[
\hat L = L + \sum_{k=1}^m e_ke_k^T
\]
which is positive definite. Our minimization problem is then equivalent to minimizing
\[
2 \trace( \CY^T \hat L \CY) - 2 \trace(\CU^T \CY), 
\]
subject to $e_j^T\CY = 0$ for $j=1, \ldots, m$. The derivative of this function is 
$4 \hat L\CY - 2 \CU$
so that an optimal $\CY$ must satisfy
\[
4 \hat L\CY - 2 \CU + \sum_{j=1}^m  e_j\mu_j^T = 0
\]
with Lagrange multipliers $\mu_1, \ldots, \mu_m\in \mR^p$. This shows that
\[
\CY = \frac12 \hat L^{-1} \Big(\CU - \frac12 \sum_{j=1}^m  e_j\mu_j^T\Big) =  \frac12 \hat L^{-1} \CU - \frac14 \sum_{j=1}^m  e_j\mu_j^T
\]
where we have used the fact that $\hat L^{-1} e_j = e_j$. We can now identify $\mu_j$ since
\[
0 = e_j^T\CY = \frac12 e_j^T \hat L^{-1} \CU - \frac14 \sum_{j'=1}^m e_j^T e_{j'} \mu_{j'}^T = \frac12 e_j^T \CU - \frac14 \mu_j^T
\]
so that $\mu_j^T = 2 e_j^T\CU$ and the optimal $\CY$ is
\[
\CY = \frac12 \hat L^{-1} \CU - \frac12 \sum_{j=1}^m e_j e_j^T \CU.
\]
Note that this expression can be rewritten as
\[
\CY = \frac12  P_L \hat L^{-1} \CU
\]
where $P_L = \Id[N]  - \sum_{k=1}^N e_je_j^T$ is the projection onto the space perpendicular to the null space of $L$ (i.e., the range of $L$).
In the case where the graph has a single connected component, one has  $m=1$ and $e_1 = \dsone_N/\sqrt N$ yielding
\[
P_L = \Id[N] - \frac1N \dsone_N\dsone_N^T.
\]

The minimization in $z$ given $y$ is straightforward: if $y_k\neq y_k$, then $z_{kl} = (y_k-y_l)/|y_k-y_l|$. If $y_k=y_l$, then one can take any value for $z_{kl}$ and the simplest is of course $z_{kl}=0$. Using the previous computation, we can summarize a training algorithm for multi-dimensional scaling, called SMACOF for ``Scaling by Maximizing a Convex Function'' (see, e.g., \citet{borg2005modern} for more details and references).

\begin{algorithm}[SMACOF]
Assume that a symmetric matrix of dissimilarities $(d_{kl}, k,l=1, \ldots, N)$ is given, together with a matrix of weights $(w_{kl}, k,l=1, \ldots, N)$. Fix a target dimension, $p$, and a tolerance constant, $\ep$.
\begin{enumerate}[label=\arabic*., wide=0.5cm]
\item Compute the Laplacian matrix $L$ of the graph associated with the weights, the projection matrix $P_L$ onto the range of $L$ and the matrix $M = (L + \Id[N] - P_L)^{-1}$.
\item Initialize the algorithm with some family $y_1, \ldots, y_N \in \mR^p$ and let $\CY = \begin{pmatrix} y_1^T\\ \vdots \\ y_N^T\end{pmatrix}$.
\item At a given step of the algorithm, let $\CY$ be the current solution and compute, for $k=1, \ldots, N$:
\[
u_{k} = 2 \sum_{l=1}^N w_{kl} d_{kl} \frac{y_{k}-y_{l}}{|y_k-y_l|} \bfone_{y_k\neq y_l}
\]
to form the matrix $\CU = \begin{pmatrix} u_1^T\\ \vdots \\ u_N^T\end{pmatrix}$.
\item Compute $\CY' = \frac12  P_L M \CU$.
\item If $|\CY -\CY'| \leq \ep$, exit and return $\CY'$.
\item Return to step 3.
\end{enumerate}
\end{algorithm}

\section{Manifold learning}

The goal of MDS is to map a full matrix of distances into a low-dimensional Euclidean space. Such a representation, however, cannot address the possibility that the data is supported by a low-dimensional, albeit nonlinear, space.  For example, people living on Earth belong, for all purposes, to a two-dimensional structure (a sphere), but any faithful Euclidean representation of the world population needs to use the three spatial dimensions. One may also argue that the relevant distance between points on Earth is not the Euclidean one  (because one would never travel {\em through}  Earth to go from one place to another), but the distance associated to the shortest path on the sphere, which is measured along great circles.

\begin{figure}
\includegraphics[trim=1cm 0cm 1cm 1cm, clip, width=0.45\textwidth]{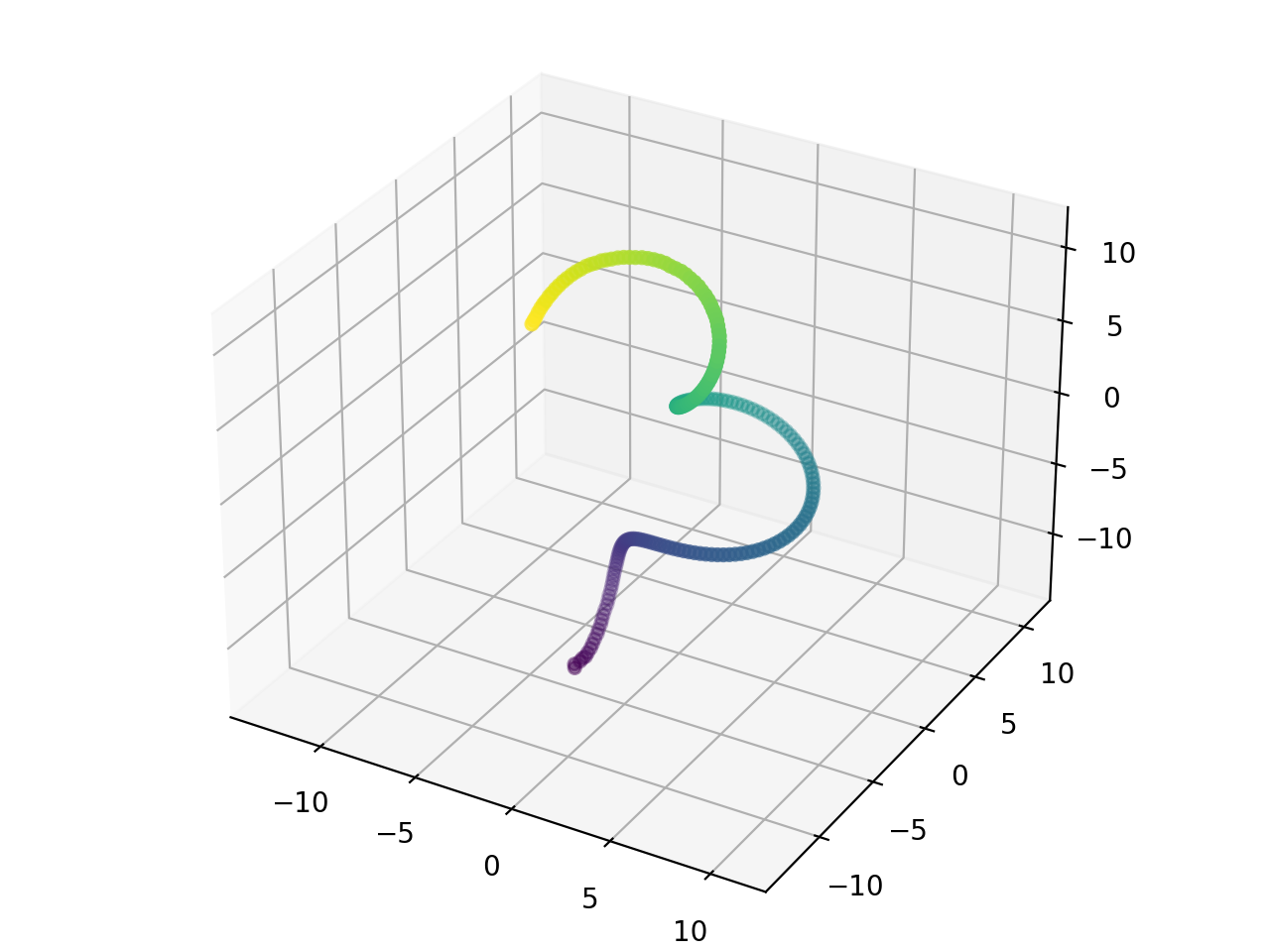}
\includegraphics[trim=1cm 0cm 1cm 1cm, clip, width=0.45\textwidth]{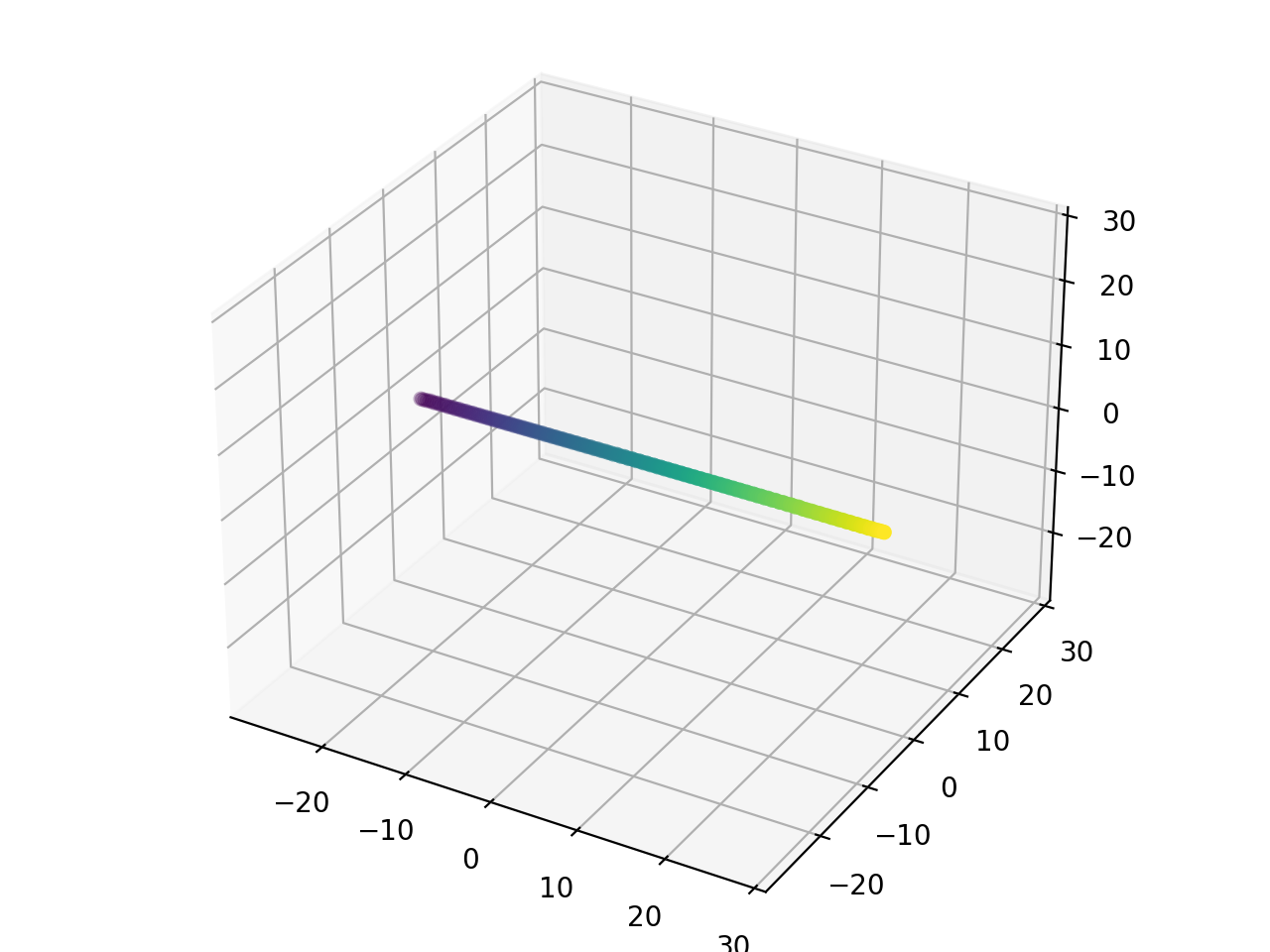}
\caption{\label{fig:isomap.curve} Left: Multidimensional scaling applied to a 3D curve embedded in a 10-dimensional space retrieves the Euclidean structure. Right: Isomap, in contrasts, identifies the one-dimensional nature of the data.} 
\end{figure}

To take another example, the left panel in \cref{fig:isomap.curve} provides the result of applying MDS to a ten-dimensional dataset obtained by applying a random ten-dimensional rotation to a curve supported by a three-dimensional torus. MDS indeed retrieves the correct curve structure in space, which is  three dimensional. However, for a person ``living'' on the curve, the data is one-dimensional, a fact that is captured by the Isomap method that we now describe.

\subsection{Isomap}

Let us return to the example of people living on the spherical Earth. One can define the distance between two points on Earth either as the shortest length a person would have to travel (say, by plane) to go from one point to the other (that we can call the {\em intrinsic distance}), or simply the {\em chordal distance} in 3D space between the two points. The first one is obviously the most relevant to the spherical structure of the Earth, but the second one is  easier to compute given the locations of the points in space.  

For typical datasets, the geometric structure of the data (a sphere in the Earth example) is unknown, and the only information that is available is their chordal distance in an ambient space (which can be very large). An important remark, however is that, when  points are close to each other, the two distances can be expected to be similar, if we assume that the geometry of the set supporting the data is locally linear (e.g., that it is, like the sphere, a ``submanifold'' of the ambient space, with small neighborhoods of any data point well approximated, at first order, by points on a  tangent space).
Isomap uses this property, only trusting small distances in the matrix $D$, and infers large distances by adding the costs resulting from traveling from data points to nearby data points.

To formalize this, fix an integer $c$.  
Given $D$, the $c$-nearest neighbor graph on $\CV = \{1, \ldots, N\}$ places an edge between $k$ and $l$ if and only if $d_{kl}$ is among the $c$ smallest values in $\{d_{kl'}, l'\neq k\}$  or $x_l$ among the $c$ smallest values in  $\{d_{k'l}, k'\neq l\}$. We will write $k\sim_c l$ to indicate that there exists an edge between $k$ and $l$ in this graph.
 One then defines the geodesic distance on the graph as
\[
\pe{d_{kl}} * = \min\left\{ \sum_{j=1}^m d_{k_{j-1}k_{j}}: k_0, \ldots, k_m \in \{1, \ldots, N\}, k_0=k \sim_c k_1\sim_c \cdots \sim_c k_{m-1} \sim_c k_m=l, m\geq 0\right\}\,.
\]

This geodesic distance can be computed incrementally as follows. First define $d^{(1)}_{kl} = |x_k-x_l|$ if $k\sim_c l$ and   $ d^{(1)}_{kl} = +\infty$ otherwise (and also let $d^{(1)}_{kk}=0$). Then, given $d^{(n-1)}$, define
\[
d^{(n)}_{kl}  = \min\left\{d^{(n-1)}_{kl'} + d^{(1)}_{ll'}\: l'=1, \ldots, N\right\}
\]
until the entries stabilize, i.e., $d^{(n+1)} = d^{(n)}$, in which case one has $\pe d *=d^{(n)}$. The validity of the last statement can be easily proved by checking that 
\[
d^{(n)}_{kl} = \min\left\{ \sum_{j=1}^n d^{(1)}_{k_{j-1}k_{j}}: k_0, \ldots, k_n \in \{1, \ldots, N\}, k_0=k, k_n=l\right\}\,,
\]
which can be done by induction, the details being left to the reader. It should also be clear that the procedure will stabilize after no more than $N$ steps.

\begin{figure}
\includegraphics[trim=1cm 0cm 1cm 1cm, clip, width=0.45\textwidth]{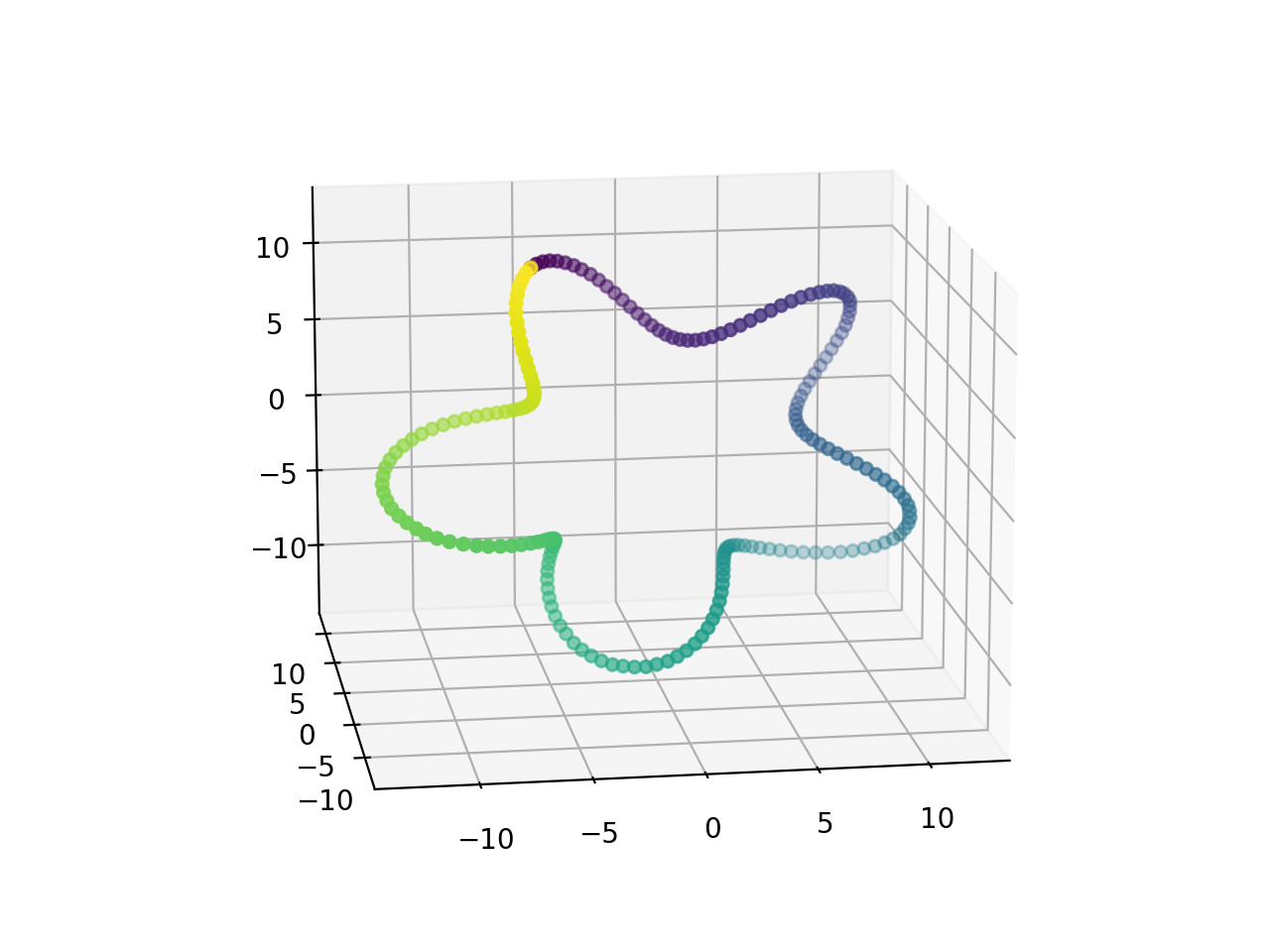}
\includegraphics[trim=1cm 0cm 1cm 1cm, clip, width=0.45\textwidth]{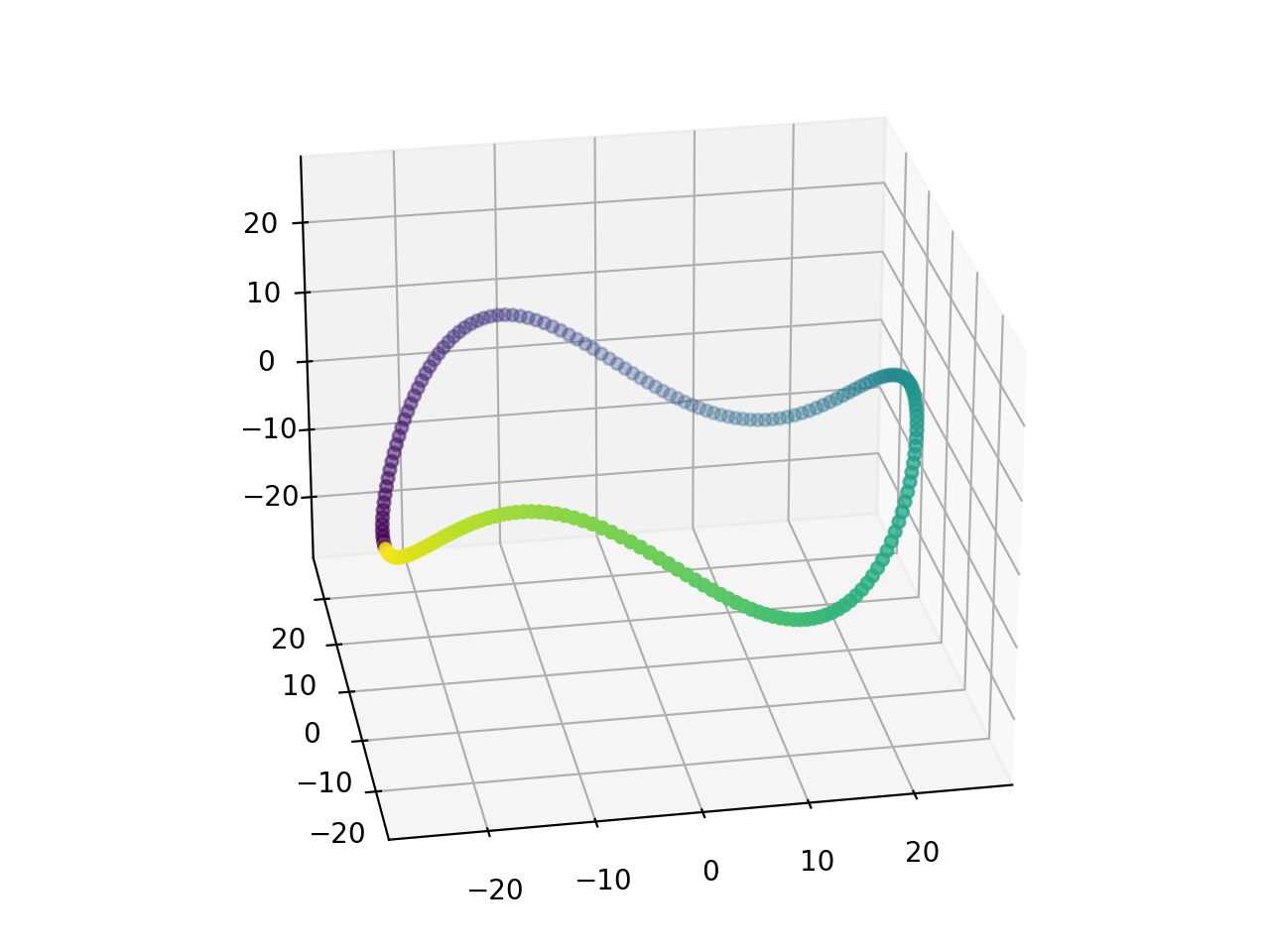}
\caption{\label{fig:isomap.curve.closed} Left: Multidimensional scaling applied to a closed 3D curve embedded in a 10-dimensional space retrieves the Euclidean structure. Right: Isomap, applied to the same data.} 
\end{figure}

Once the distance is computed, Isomap then  applies standard MDS, resulting in a straightened representation of the data like in \cref{fig:isomap.curve}. Another example is provided in \cref{fig:isomap.curve.closed}, where, this time, the input curve is closed and cannot therefore be represented as a one-dimensional structure. One can note, however, that, even in this case, Isomap still provides some simplification of the initial shape of the data.

\subsection{Local Linear Embedding}
\label{sec:lle}
Local linear embedding (LLE) exploits in a different way the fact that manifolds are locally well approximated by linear spaces. Like Isomap, it starts also with building a $c$-nearest-neighbor graph on $\{1, \ldots, k\}$. Assume, for the sake of the discussion, that the distance matrix is computed for possibly unobserved data $T = (x_1, \ldots, x_N)$. 
Letting $\CN_k$ denote the indices of the nearest neighbors of $k$ (excluding $k$ itself), the basic assumption is that $x_k$ should approximately lie in the affine space generated by $x_l$, $l\in \CN_k$. Expressed in barycentric coordinates, this space is defined by
\[
\CT_k = \defset{ \sum_{l\in\CN_k} \pe \rho l x_l : \rho\in \mR^{\CN_k}, \sum_{l\in \CN_k} \pe \rho l = 1},
\]
and $\CT_k$ can be interpreted as an approximation of the tangent space at $x_k$ to the data manifold. Optimal coefficients $(\pe {\rho_k} l, k=1, \ldots, N, l\in\CN_k)$ providing the representation of $x_k$ in that space can be estimated by minimizing, for all $k$
\[
\left |x_k - \sum_{\l\in\CN_k}\pe {\rho_{k}} l x_l \right|^2
\]  
subject to $\sum_{l\in \CN_k} \pe {\rho_{k}} l = 1$. This is a simple least-square program. Let $c_k = |\CN_k|$ ($c_k=c$ in the absence of ties).
Order the elements of $\CN_k$ to represent $\pe{\rho_{k}} l$, $l\in\CN_k$ as a vector denoted $\bfrho_k\in \mR^{c_k}$. Similarly, let $S_k$ be the Gram matrix associated with $x_l$, $l\in\CN_k$ formed with all inner products $x_{l'}^T x_l$, $l,l'=1, \ldots, N$ and let $\bfr_k$ be the vector composed with products $x_k^Tx_l$, $l\in\CN_k$. Assume that $S_k$ is invertible, which is generally true if $c < d$, unless the neighbors are exactly linearly aligned. Then, the optimal $\bfrho_k$ and the Lagrange multiplier $\la$ for the constraint are given by
\begin{equation}
\label{eq:lle.1}
\begin{pmatrix} \bfrho_k\\ \la\end{pmatrix} = 
\begin{pmatrix} S_k & \dsone_{c_k}\\
\dsone_{c_k}^T & 0
\end{pmatrix}^{-1}
\begin{pmatrix} \bfr_k\\ 1\end{pmatrix}.
\end{equation}
If  $S_k$ is not invertible, the problem is under-constrained and one of its solutions can be obtained by  replacing the inverse above by a pseudo-inverse.

The low-dimensional representation of the data, still denoted $(y_1, \ldots, y_N)$ with $y_k\in\mR^p$ is then estimated so that the relative position of $y_k$ to its neighbors is the same as that of $x_k$, i.e., so that
\[
y_k \simeq \sum_{l\in\CN_k} \pe {\rho_{k}} l y_l.
\]
These vectors are estimated by minimizing
\[
F(y) = \sum_{k=1}^N \left| y_k - \sum_{l\in \CN_k} \pe{\rho_{k}}l y_l\right|^2.
\]
Some additional constraints are needed to avoid, for example, the trivial solution $y_k = 0$ for all $k$. One can also note that replacing all $y_k$'s by $y_k' = Ry_k + b$ where $R$ is an orthogonal transformation in $\mR^p$ and $b$ is a translation does not change the value of $F$, so that there is no loss of generality in assuming that $\sum_{k=1}^N y_k = 0$ and that $\sum_{k=1}^N y_ky_k^T = D_0$, a diagonal matrix. However, if one lets $y'_k = Dy_k$ where $D$ is diagonal, then
\[
F(y) = \sum_{i=1}^p D_{ii}^2 \sum_{k=1}^N \left( \pe{y_k}i - \sum_{l\in \CN_k} \pe{\rho_{k}}l \pe{y_l}i\right)^2.
\]
This shows that one should not allow the diagonal coefficients of $D_0$ to be chosen freely, since otherwise the minimizing solution would require to take this coefficient to $0$. So, $D_0$ should be a fixed matrix, and by symmetry, it is natural to take $D_0 = \Id[p]$. (Any other solution---for a different $D_0$---can then be obtained by rescaling independently the coordinates of $y_1, \ldots, y_N$.)

Extend $\pe{\rho_{k}}l$ to an $N$-dimensional vector by taking $\pe{\rho_{k}}k = -1$ and $\pe{\rho_{k}}l=0$ if $l\neq k$ and $l\not \in \CN_k$. We can write 
\[
F(y) = \sum_{k=1}^N \left |\sum_{l=1}^N \pe{\rho_{k}}l y_l \right|^2.
\]
Expanding the square, this is
\[
F(y) = \sum_{l,l'=1}^N w_{ll'} y_l^Ty_{l'}
\]
with $w_{ll'} = \sum_{k=1}^N \pe{\rho_{k}}l \pe{\rho_{k}}{l'}$. Introducing the matrix $\CW$ with entries $w_{kl}$ and the $N\times p$ matrix $\CY = \begin{pmatrix}
y_1^T \\\vdots \\y_N^T
\end{pmatrix}
$, we have the simple expression
\[
F(y) =  \trace(\CY^T \CW \CY)\,.
\]
The constraints are $\CY^T\CY = \Id[p]$ and $\CY^T \dsone_N = 0$. Without this last constraint, we know that an optimal solution is provided by $\CY = [e_1, \ldots, e_p]$ where $e_1,\ldots, e_p$ provide an orthonormal family of eigenvectors associated to the $p$ smallest eigenvalues of $\CW$ (this is a consequence of \cref{cor:pca.base}).  To handle the additional constraint, it suffices to note that $\CW \dsone_N = 0$, so that $\dsone_N$ is a zero eigenvector. Given this, it suffices to compute   $p+1$ eigenvectors associated to smallest eigenvalues of $\CW$, $e_1,\ldots, e_{p+1}$, with the condition that $e_1 = \pm \dsone_N/\sqrt N$ (which is automatically satisfied unless 0 is a multiple eigenvalue of $\CW$) and let 
\[
\CY = [e_2, \ldots, e_{p+1}].
\]
Note that $e_2, \ldots, e_{p+1}$ are also the $p$ smallest eigenvectors of $\CW + \la \dsone\dsone^T$ for any large enough $\la$, e.g., $\la > \trace(\CW)/N$.

\begin{figure}
\centering
\includegraphics[trim=1cm 0cm 1cm 1cm, clip, width=0.45\textwidth]{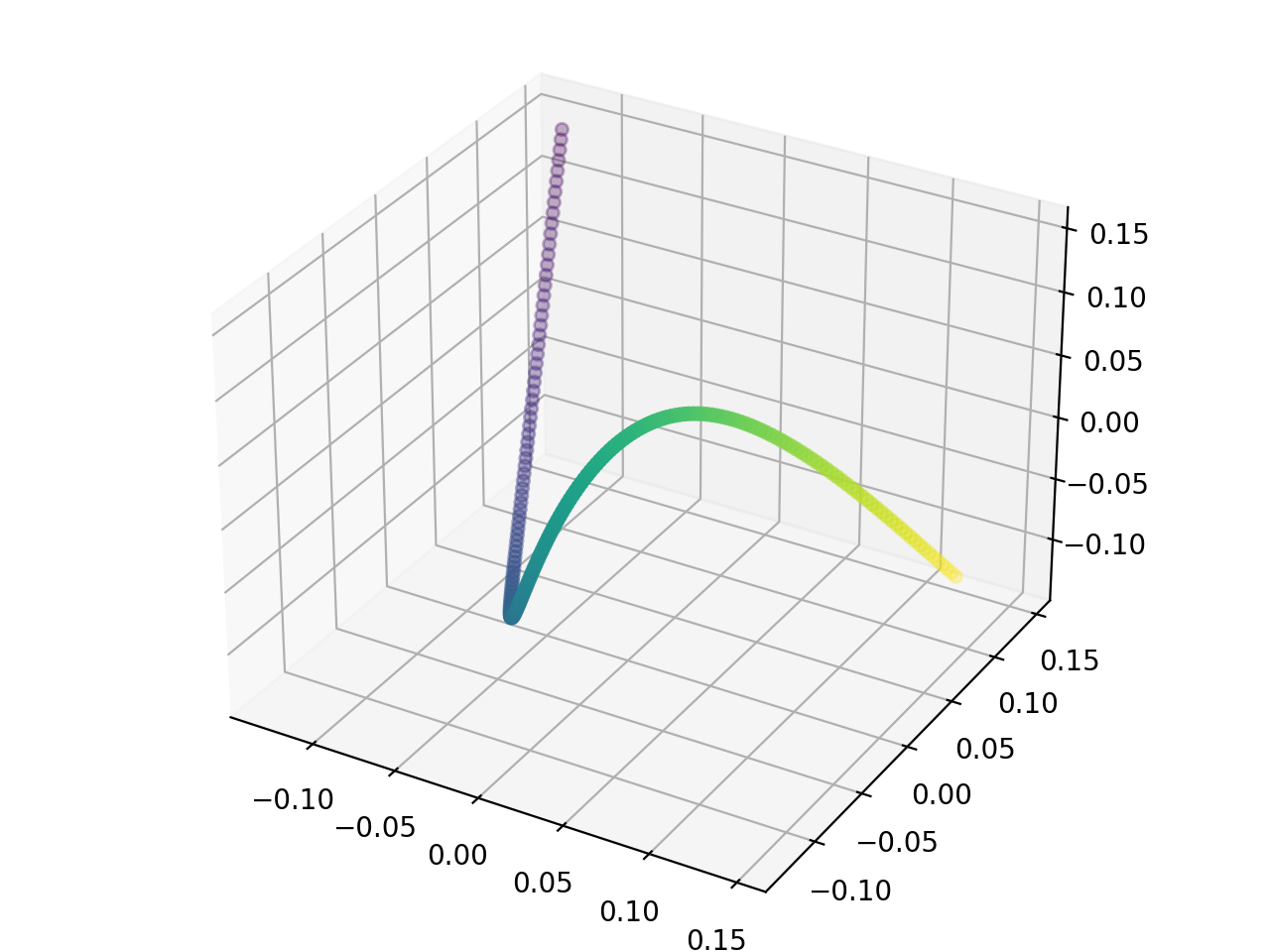}
\includegraphics[trim=1cm 0cm 1cm 1cm, clip, width=0.45\textwidth]{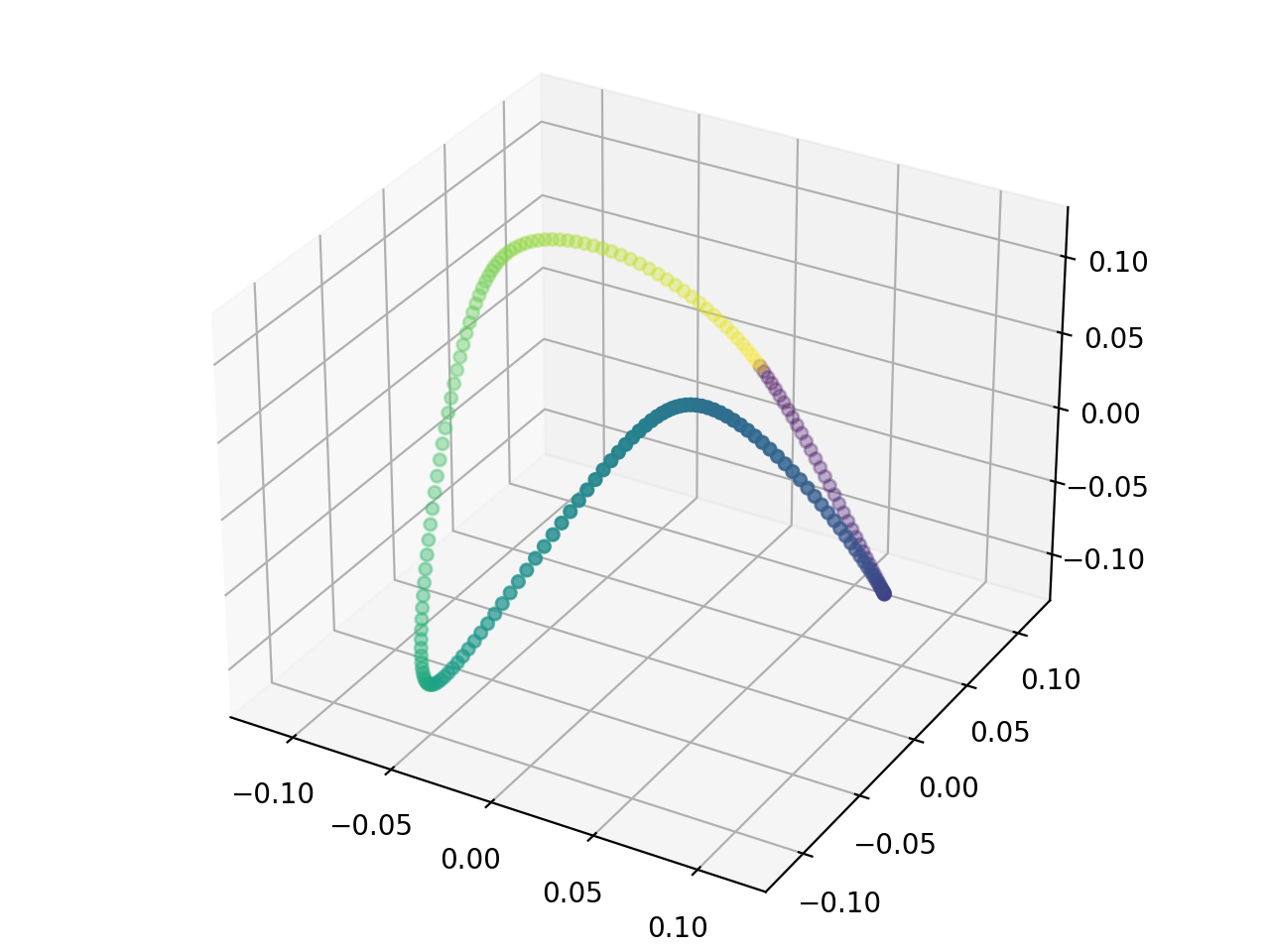}
\caption{\label{fig:lle.curve} Local linear embedding with target dimension $3$ applied to the data in \cref{fig:isomap.curve} and \cref{fig:isomap.curve.closed}.} 
\end{figure}

LLE is summarized in the following algorithm.
\begin{algorithm}[Local linear embedding]
\label{alg:lle}
The input of the algorithm is 
\begin{enumerate}[label= (\roman*),wide=1cm]
\item Either a training set $T = (x_1, \ldots, x_N)$, or its Gram matrix $S$ containing all inner products $x_k^T x_l$ (or more generally inner products in feature space), or a dissimilarity matrix $D = (d_{kl})$. 
\item An integer $c$ for the graph construction.
\item An integer $p$ for the target dimension.
\end{enumerate}
\begin{enumerate}
\item If not provided in input, compute the Gram matrix $S$ and distance matrix $D$ (using \cref{eq:s.to.d} and \cref{eq:dissim.sim}).
\item Build the $c$-nearest-neighbor graph associated with the distances. Let $\CN_k$ be the set of neighbors of $k$, with $c_k = |\CN_k|$.
\item For $k=1, \ldots, N$, let $S_k$ be the sub-matrix of $S$  associated with $(x_l, l\in \CN_k)$ and compute coefficients $(\pe {\rho_{k}}l, l\in \CN_k)$ stacked in a vector $\bfrho_k\in \mR^{c_k}$ by solving \cref{eq:lle.1}.
\item  Form the matrix $\CW$ with entries  $w_{ll'} = \sum_{k=1}^N \pe{\rho_{k}}l \pe{\rho_{k}}{l'}$ with $\rho$ extended so that $\pe{\rho_{k}}k = -1$ and $\pe{\rho_{k}}l=0$ if $l\neq k$ and $l\not \in \CN_k$.
\item Compute the first $p+1$ eigenvectors, $e_1, \ldots, e_{p+1}$, of $\CW$ (associated with smallest eigenvalues) arranging for $e_1$ to be proportional to $\dsone_N$.
\item Set $\pe{y_k}i = \pe{e_{i+1}}k$ for $i=1, \ldots, p$ and $k=1, \ldots, N$.
\end{enumerate}
\end{algorithm}
The results of LLE applied to the datasets described in \cref{fig:isomap.curve} and \cref{fig:isomap.curve.closed} are provided in \cref{fig:lle.curve}.

\begin{remark}
For both Isomap and LLE, the $c$-nearest-neighbors graph can be replaced by the graph formed with edges between all pairs of points that are at distance less than $\ep$ from each other, for a chosen $\ep>0$, with no change in the algorithms.

These parameters ($c$ or $\ep$) must be chosen carefully and may have an important impact on the output of the algorithm. Choosing them too small would not allow for a correct estimation of distances in Isomap (with possibly some of them being infinite if the graph has more than one connected component), or of the linear approximations in LLE. However, choosing them too large
may break the  hypothesis that the data is locally Euclidean or linear that forms the basic principle of these algorithms.
\end{remark}

\subsection{Graph Embedding}

Both Isomap and LLE  are based on the construction of a nearest-neighbor graph using dissimilarity data and assuming the conservation of some of its geometric features when deriving a small-dimensional representation. For LLE, a weight matrix $\CW$ was first estimated based on optimal linear approximations of $x_k$ by its neighbors, and the representation was computed by estimating the eigenvectors associated with the smallest eigenvalues of $\CW$ (excluding the eigenvector proportional to $\dsone$). However, both methods were motivated by the intuition that the dataset was supported by a continuous small-dimensional manifold.
We now discuss methods that are solely motivated by the discrete geometry of a graph, for which we use tools that are similar to our discussion of graph clustering in \cref{sec:graph.cluster}. 

Adapting the notation in that section to the present one, we start with a graph with $N$ vertices and weights $\be_{kl}$ between these vertices (such that $\be_{ll}=0$) and we form the Laplacian operator defined by, for any vector $u\in \mR^N$:
\[
\frac12 \|u\|_{H^1}^2 = \frac12 \sum_{l,l'=1}^N \be_{ll'} (\pe u l - \pe u{l'})^2 = u^TLu, 
\]
so that $L$ is identified as the matrix with coefficients $\ell_{ll'} = - \be_{ll'}$ for $l\neq l'$ and $\ell_{ll} = \sum_{l'=1}^N \be_{ll'}$. The matrix $\CW$ that was obtained for LLE coincides with this graph Laplacian if one lets $\be_{ll'} = - w_{ll'}$ for $l\neq l'$, since we have $\sum_{l'=1}^N w_{ll'} = 0$. The usual requirement that weights are non-negative is no real loss of generality, because in LLE (and in the Graph embedding method above), one is only interested in eigenvectors of $\CW$ (or $L$ below) that are perpendicular to $\dsone$, and those remain the same if one replaces $\CW$ by
\[
\CW - a \dsone_N\dsone_N^T + Na \Id[N]
\]
which has negative off-diagonal coefficients $\tilde w_{ll'} = w_{ll'} - a$ for large enough $a$.

In graph (or Laplacian) embedding, the starting point is a weighted graph on $\{1, \ldots, N\}$ with edge weights $\be_{ll'}$ interpreted as similarities between vertexes. These weights may or may  not have been deduced from measures of dissimilarity $(d_{ll'}, k,l=1, \ldots, N)$ which themselves may or  may not have been computed as distances between training data $x_1, \ldots, x_N$. If one starts with dissimilarities, it is typical to use simple transformations to compute edge weights, and one the most commonly used is
\[
\be_{ll'} = \exp(-d_{ll'}^2/2\tau^2)
\]
for some constant $\tau$. These weights are usually truncated, replacing small values by zeros (or the computation is restricted to nearest neighbors), to ensure that the resulting graph is sparse. This speeds up the computation of eigenvectors for large datasets. 

Given a target dimension $p$, the graph is then represented as a collection of points $y_1, \ldots, y_N\in \mR^p$, where $y_k$ is associated to vertex $k$. For this purpose, one needs to compute the first $p+1$ eigenvectors, $e_1, \ldots, e_{p+1}$, of the graph Laplacian, with the requirement that $e_1 = \pm \dsone_N/\sqrt{N}$. (This is always possible and  can be achieved numerically by computing eigenvectors of $L + c\dsone\dsone^T$ for large enough $c$.) The graph representation is then given by  $\pe {y_k} i = \pe{e_{i+1}}k$ for $i=1, \ldots, p$ and $k=1, \ldots, N$. Note that these are exactly the same operations as those described in steps 4 and 5 of the LLE algorithm. This construction can be further interpreted in several ways, as listed below.

\begin{enumerate}[label={\bf (\arabic*)}]
\item
One way to interpret this construction is that $e_2, \ldots, e_{p+1}$ (the coordinate functions for the representation $y_1, \ldots, y_N$) minimize
\[
\sum_{j=1}^p \|e_i\|^2_{H_1}
\]
subject to $e_2, \ldots, e_{p+1}$ being perpendicular to each other and perpendicular to the constant functions (these constraints being justified for the same reasons as those discussed for LLE). Small $H^1$ semi-norms being associated with smoothness on the graph, we see that we are looking for the smoothest zero-mean representation of the data.

\item
Based on our discussion of LLE, we can make an alternative interpretation by introducing a symmetric square root $R$ of the Laplacian matrix $L$ or any matrix such that $RR^T = L$. Writing $R = [\bfrho_1, \ldots, \bfrho_N]$, one has
\[
L = \sum_{k=1}^N \bfrho_k \bfrho_k^T
\]
and $\sum_{k=1}^N  \bfrho_k = 0$. With this notation, we can interpret Laplacian embedding as the minimization of
\[
\sum_{k=1}^N \left| \sum_{l=1}^N \pe{\rho_{k}}l y_l\right|^2
\]
(subject to previous orthogonality constraints). In other terms, $y_1, \ldots, y_N$ are determined so that the linear relationships
\[
\pe {\rho_k}k y_k = -\sum_{l\neq k} \pe{\rho_{k}}l y_l
\]
are satisfied, which is similar to the LLE condition, without the requirement that $\pe {\rho_k} k = 1$. 

\item
An alternate requirement that could have been made for LLE is that $\sum_{l=1}^N (\pe{\rho_{k}}l)^2 = 1$ for all $k$. Instead of having to solve a linear system in step 2 of \cref{alg:lle}, one would then compute an eigenvector with smallest eigenvalue of $S_k$. For graph embedding, this constraint can be enforced by modifying the Laplacian matrix, since $\sum_{l=1}^N (\pe{\rho_{k}}l)^2$ is just the $(k,k)$ coefficient of $RR^T$. Given this, let $D$ be the diagonal matrix formed by the diagonal elements of $L$, and define the so-called ``symmetric Laplacian'' $\tilde L = D^{-1/2} L D^{-1/2}$. One obtain an alternative, and popular, graph embedding method by replacing $e_1, \ldots, e_{p+1}$ above by the first $p$ eigenvectors of $\tilde L$.

\item
Another interpretation of this representation can be based on the random walk associated with the graph structure. Consider the Markov process $t \mapsto \bfq(t)$ defined as follows. The initial position, $\bfq(0)$ is selected according to some arbitrary distribution, say $\pi_0$. Conditional to $\bfq(t) = k$, the next position is determined by setting random waiting times $\tau_{kl}$, each distributed as an exponential distribution with rate $\be_{kl}$ (or expectation $1/\be_{kl}$), and the process moves to the position $l$ for which $\tau_{kl}$ is smallest after waiting for that time. Let  $P(t)$ be the matrix with coefficients $P(t, k, l) = P(\bfq(t+s) = l\mid \bfq(s) = k)$. Then, one shows that
\[
P(t) = e^{-t L},
\]
where the right-hand side is the matrix exponential. If $\la_1 = 0 \leq \lambda_2 \leq\cdots \leq \la_N$ are the eigenvalues of $L$ with corresponding eigenvectors $e_1, \ldots, e_N$, then
\[
P(t) = \sum_{i=1}^N e^{-t\la_i} e_ie_i^T
\]
In particular, restricting the first eigenvectors of $L$ provides an approximation of this stochastic process, i.e., 
\[
P(t) \simeq \frac{\dsone\dsone^T}N + \sum_{i=1}^p e^{-t\la_{i+1}} y(i)y(i)^T.
\] 

\item
One can also consider the discrete-time version of the walk, for which, considering integer times $t\in \mN$, 
\[
P(\bfq(t+1) = l \mid \bfq(t) = k) =\left\{
\begin{aligned}
 \frac{\be_{kl}}{\sum_{l'=1, l'\neq k}^N \be_{kl'}} \text{ if } l\neq k\\
 0 \text{ if } l=k
 \end{aligned}
 \right.
\]
Introducing the matrix $B$ of similarities $\be_{kl}$ (with zero on the diagonal) and the diagonal matrix $D$ with coefficients 
$d_{kk} = \sum_{l =1, l\neq k}^N \be_{kl}$, the r.h.s. of the previous equation is the $k,l$ entry of the matrix $\tilde P = D^{-1} B$. Then, for any integer $s$, $P(\bfq(t+s) = l \mid \bfq(t) = k)$ is the $k,l$ entry if $\tilde P^s = D^{-1/2} (D^{-1/2} B D^{-1/2})^s D^{1/2}$.

The Laplacian matrix $L$ is given by $L = D-B$. The normalized Laplacian is 
\[
\bar L = D^{-1/2} L D^{-1/2} = \Id[N] - D^{-1/2} B D^{-1/2}
\]
so that
\[
\tilde P^s = D^{-1/2} (\Id[N] - \bar L)^s D^{1/2}.
\]
If one introduces the eigenvectors $\bar e_1, \ldots, \bar e_N$ of the normalized Laplacian, still associated with non-decreasing eigenvalues $\bar\la_1=0, \ldots, \bar\la_N$, and arranges without loss of generality that $\bar e_1 \propto D^{1/2} \dsone_N$, then 
\[
\tilde P^s = D^{-1/2} \left(\sum_{i=1}^N (1 - \bar \la_i)^s \bar e_i\bar e_i^T\right) D^{1/2}.
\]
This shows that, for $s$ large enough, the transitions of this Markov chain are well approximated by its first terms, suggesting using the alternative representation based on the normalized Laplacian:
\[
\bar y_k(i) = \bar e_{i+1}(k).
\]

\end{enumerate} 

\subsection{Stochastic neighbor embedding}

\subsubsection{General algorithm}
Stochastic neighbor embedding (SNE, \citet{hinton2002stochastic}), and its variant (t-SNE, \citet{maaten2008visualizing}) have become a popular tool for the visualization of high-dimensional data based on dissimilarity matrices. One of the key contributions of this algorithm is to introduce a local data rescaling step, that allows for visualization of more homogeneous point clouds. 

Assume that dissimilarities $D = (d_{kl}, k,l=1, \ldots, N)$ are observed. The basic principle in SNE is to deduce from the dissimilarities a family of $N$ probability distributions on $\{1, \ldots, N\}$, that we will denote $\pi_k$, $k=1, \ldots, N$, with the property that $\pi_k(k) = 0$. The computation of these probabilities determine the local normalization step, and we will return to this later. Given these $\pi_k$'s, one then estimate low-dimensional representations $\bfy = (y_1, \ldots, y_N)$ such that $\pi_k \simeq \psi_k$ where $\psi_k$ is given by
\[
\psi_k(l;\bfy) = \frac{\exp\big(-\beta\big(|y_k-y_l|^2\big)\big)}{\sum_{l'=1, l'\neq k}^N \exp\big(-\beta\big(|y_k-y_{l'}|^2\big)\big)} \bfone_{l\neq k}.
\] 
Here, $\beta: [0, +\infty) \to [0, +\infty)$ is an increasing differentiable function that tends to $+\infty$ at infinity. The  derivative is denoted $\prt\be$. The original version of SNE \citep{hinton2002stochastic} uses $\beta(t) = t$ and t-SNE \citep{maaten2008visualizing} takes $\beta(t) = \log(1+t)$.  


The determination of the representation can then be performed by minimizing a measure of discrepancy between the probabilities $\pi_k$ and $\psi_k$. In \citet{hinton2002stochastic}, it is suggested to minimize the sum of Kullback-Liebler divergences, namely
\[
\sum_{k=1}^N \KL(\pi_k\ \|\ \psi_k(\cdot;\bfy)) 
\]
or, equivalently, to maximize
\begin{align*}
F(\bfy) &= \sum_{k,l=1}^N \pi_k(l) \log \psi_k(l ; \bfy) \\
&= - \sum_{k,l=1}^N \be(|y_k-y_l|^2) \pi_k(l) + \sum_{k=1}^N \log \left(\sum_{l=1, l\neq k}^N \exp(-\be(|y_k-y_l|^2))\right)
\end{align*}
The gradient of this function can be computed by evaluating the derivative at $\ep=0$ of $f: \ep \mapsto F( \bfy+\ep \bfh)$. This computation gives
\begin{align*}
f'(0) =& -2 \sum_{k,l=1}^N \prt\be(|y_k-y_l|^2) (y_k-y_l)^T(h_k-h_l) \pi_k(l) \\
&+ 2\sum_{k=1}^N \sum_{l=1}^N \prt \be(|y_k-y_l|^2) (y_k-y_l)^T(h_k-h_l)\psi_k(l;\bfy)\\
 =& - 2 \sum_{k=1}^N h^T_k \sum_{l=1}^N \prt\be(|y_k-y_l|^2)(y_k - y_l) (\pi_k(l) + \pi_l(k) - \psi_k(l;\bfy) - \psi_l(k;\bfy))
\end{align*}
This shows that
\[
\prt_{y_k} F(\bfy) = -2  \sum_{l=1}^N \be(|y_k-y_l|^2)(y_k - y_l) (\pi_k(l) + \pi_l(k) - \psi_k(l;\bfy) - \psi_l(k;\bfy)).
\]

This is a rather simple expression that can be used with any first-order optimization algorithm to maximize $F$. The algorithm in
\citet{hinton2002stochastic} uses gradient ascent with momentum, namely iterating 
\[
\bfy^{(n+1)} = \bfy^{(n)} + \gamma \nabla F(\bfy^{(n)}) + \alpha^{(n)} (y^{(n)} - y^{(n-1)}).
\]
Choosing $\alpha^{(n)} = 0$ provides standard gradient ascent with fixed gain $\gamma$ (of course,  other optimization methods may be used).

A variant of the algorithm replaces the node-dependent probabilities $\pi_k$ by a single, symmetric, joint distribution $\bar\pi$ on $\{1, \ldots, N\}^2$, $(k,l) \mapsto \bar\pi(k,l)$, satisfying $\bar\pi(k,k) = 0$ and $\bar\pi(k,l) = \bar\pi(l,k)$.  The target distribution $\bar\psi$ then becomes 
\[
\bar\psi(k,l;\bfy) =    \frac{\exp(-\be(|y_k-y_l|^2))}{\sum_{k'\neq l'=1}^N \exp(-\be(|y_{k'} - y_{l'}|^2))}.
\]
With such a choice, the objective function has a simpler form,  namely minimizing $\KL(\bar\pi\ \|\ \bar\psi(\cdot, y))$ or maximizing the expected likelihood
\[
\bar F(\bfy) = \sum_{k,l=1}^N \bar \pi(k,l) \log\bar \psi(k,l; y) = -\sum_{k,l=1}^N \be(|y_k-y_l|^2) \bar\pi(k,l) + \log \left(\sum_{k\neq l=1}^N \exp(-\be(|y_k-y_l|^2))\right).
\] 
The gradient of this symmetric version of $F$ can be computed similarly to the previous one and is given by
\[
\prt_{y_k} \bar F(\bfy) = -4  \sum_{l=1}^N \prt\be(|y_k-y_l|^2) (y_k - y_l) (\bar\pi_(k, l) - \bar\psi(k, l;\bfy)).
\]

\subsubsection{Setting initial probabilities}

The probabilities $\pi_k(l)$ or $\bar\pi(k,l)$ are deduced from the dissimilarities as
\[
\pi_k(l) = \frac{e^{-d_{kl}^2/2\sigma^2_k}}{\sum_{l'=1, l'\neq k}^Ne^{-d_{kl'}^2/2\sigma^2_k}}
\]
for $l\neq k$ and 
\[
\bar\pi(k,l) = \frac{\pi_k(l) + \pi_l(k)}{2n}.
\]

The coefficients $\sigma^2_k$, $k=1, \ldots, N$ operate the local normalization, justifying, in particular, the parameter-free expression chosen for $\psi$ and $\bar\psi$. These coefficients are estimated
so as to adjust the entropies of all $\pi_k$ to a fixed value, which is a parameter of the algorithm. Note that, letting $t = 1/2\sigma_k^2$
and $H(\pi_k) = -\sum_{l=1}^N \pi_k(l) \log\pi_k(l)$, 
\[
\prt_t H(\pi_k) = - \sum_{l=1}^N \prt_t \pi_k(l) \log\pi_k(l) - \sum_{l=1}^N \prt_t \pi_k(l) 
= - \sum_{l=1}^N \prt_t \pi_k(l) \log\pi_k(l).
\]
Now
\[
\prt_t \log \pi_k(l) = - d_{kl}^2 + \bar d^2_k 
\]
with $\bar d_k^2 = \sum_{l'=1}^N d_{kl'}^2 \pi_k(l')$.
Writing $\prt_t \pi_k(l) = \pi_k(l) \prt_t\log\pi_k(l)$, we have
\[
\prt_t H(\pi_k) = \sum_{l=1}^N (d_{kl} \log\pi_k(l)) \pi_k(l)  - \bar d_k \sum_{l=1}^N \pi_k(l) \log\pi_k(l).
\]
Using Schwartz inequality, we see that $\prt_t H(\pi_k) \leq 0$ so that $H(\pi_k)$ is decreasing as a function of $t$, i.e., increasing as a function of $\sigma^2_k$. When $\sigma_k^2\to 0$, $\pi_k$ converges to the uniform distribution on the set of nearest neighbors of $k$ (the indexes $l\neq k$ such that $d_{kl}^2$ is minimal) and, letting $\nu_k$ denote their number, which is typically equal to 1, $H(\pi_k)$ converges to $\log\nu_k$. When $\sigma^2_k$ tends to infinity, $\pi_k$ converges to the uniform distribution over indexes $l\neq k$, whose entropy is $\log (N-1)$. This shows that $e^{H(\pi_k)}$, which is called the {\em perplexity} of $\pi_k$ can take any value between $\nu_k$ and $N-1$. The common target value of the perplexity can therefore be taken anywhere between $\max_k\nu_k$ and $N-1$. In \citet{maaten2008visualizing}, it is recommended to choose a value between 5 and 50.

\begin{remark}
The complexity of the computation of the gradient of the objective function (either $F$ or $\bar F$ ) scales like the square of the size of the training set, which may be prohibitive when $N$ is large. In \citet{van2014accelerating}, an accelerated procedure, that involves an approximation of the gradient is proposed. (This procedure is however limited to representations in dimensions 2 or 3.)
\end{remark}

\subsection{Uniform manifold approximation and projection (UMAP)}

UMAP is similar in spirit to t-SNE, with a few important differences that result in a simpler optimization problem and faster algorithms. Like Isomap, the approach is based on matching distances between the high-dimensional data and the low-dimensional representation. But while Isomap estimates a unique distance on the whole training set (the geodesic distance on the nearest-neighbor graph), UMAP estimates as many ``local distances'' as observations before ``patching'' them to form the final representation. 

The idea of transporting possibly non-homogeneous locally defined objects on initial data to a homogeneous low-dimensional visualization is what makes UMAP similar to t-SNE. The difference is that t-SNE transports local probability distributions, while UMAP transports metric spaces. More precisely, given distances $(d_{kl}, k,l=1, \ldots, N)$ and an integer $m$ provided as input, the algorithm builds, for each $k=1, \ldots, N$ a (pseudo-)metric $\delta_k$ on the associated data graph by letting
\[
\delta^{(k)}(k,l) = \delta^{(k)}(l,k) = \frac{1}{\sigma_k}\left(d_{kl} - \min_{l'\neq   k} d_{kl'}\right)
\]
if $l$ is among the $m$ nearest neighbors of $k$, where $m$ is a parameter of the algorithm, with all other values of $\pe \delta k$ being infinite. The normalization parameter $\sigma_k$ has a role similar to that of the same parameter in t-SNE in that it   tends to make the representation  homogeneous.  Here, it is computed such that
\[
\sum_l \exp(-\delta^{(k)}(l,l')) = \log_2 m\,.
\]
   
Each such metric provides a weighted graph structure on $\{1, \ldots, N\}$ by defining weights $w^{(k)}_{ll'} = \exp(-\delta^{(k)}(l,l'))$. 
In UMAP \citep{zadeh1996fuzzy}, these weights are interpreted in the framework of {\em fuzzy sets}, where a fuzzy set is defined by a pair $(A, \mu)$ where $A$ is a set and $\mu$ a function $\mu: A\to [0,1]$. The function $\mu$ is called the membership function and $\mu(x)$ for $x\in A$ is the membership strength of $x$ to $A$.  Letting $\CV = \{1, \ldots, N\}$ and $\CE = \CV \times \CV$, one then interprets the weights as defining the membership strength of edges to the graph, i.e., one defines the ``fuzzy graph'' $\CG^{(k)} = (\CV, \CE, \mu^{(k)})$ where $\mu^{(k)}(l,l') = w^{(k)}_{ll'}$ is the membership strength of edge $(l,l')$ to  $\CG^{(k)}$.

This is, of course, just a reinterpretation of weighted graphs in terms of fuzzy sets, but allows one to combine the collection $(\CG^{(k)}, k=1, \ldots, N)$ using simple fuzzy sets operations, namely, defining the combined (fuzzy) graph $\CG  = (\CV, \CE, \mu)$ with 
\[
(\CE, \mu) = \bigcup_{k=1}^N (\CE, \mu^{(k)})
\]
being the fuzzy union of the edge sets. There are, in fuzzy logic, multiple ways to define set unions \citep{gupta1991theory}, and the one selected for UMAP define $(A, \mu) \cup (A', \mu') = (A\cup A', \nu)$ with
$\nu(x) = \mu(x) + \mu'(x) - \mu(x)\mu'(x)$ ($\mu(x)$ and $\mu'(x)$ being defined as 0 is $x\not\in A$ or $x\not\in A'$ respectively).   In UMAP, each edge $\mu^{(k)}(l,l')$ is non-zero only is $k=l$ or $l'$ so that
\[
\mu(l,l') = w^{(l)}_{ll'} +  w^{(l')}_{ll'} -  w^{(l)}_{ll'} w^{(l')}_{ll'}.
\]  

This defines an input fuzzy graph structure on $\{1, \ldots, N\}$ that serves as target for an optimized similar structure associated with the representation $\bfy = (y_1, \ldots, y_N)$. This representation, since it is designed as a  homogeneous representation of the data, provides a unique fuzzy graph $\CH(\bfy) = (\CV, \CE, \nu(\cdot; \bfy))$ and the edge membership function is defined by $\nu(l,l';\bfy) = \phi_{a,b}(y_l, y_{l'})$ with
\[
\phi_{a,b}(y,y') = \frac{1}{1 + a |y-y'|^b}.
\]
The parameters $a$ and $b$ are adjusted so that $\phi_{a,b}$ provides a differentiable approximation of the function
\[
\psi_{\rho_0}(y,y') = \exp(- \max(0, |y-y'| - \rho_0))
\]
where $\rho_0$ is an input parameter of the algorithm. This function $\psi_{\rho_0}$ takes the same form as the membership function defined for local graphs $\CG^{(k)}$, and  its replacement by $\phi_{a,b}$ makes possible the use of gradient-based methods for the determination of the optimal $\bfy$ ($\psi_{\rho_0}$ is not differentiable everywhere).

The representation $\bfy$ is optimized by minimizing the ``fuzzy set cross-entropy''
\[
C(\mu\|\nu(\cdot, \bfy)) = \sum_{(k,l) \in \CE} \left(\mu(k,l) \log\frac{\mu(k,l)}{\nu(k,l|\bfy)} +(1- \mu(k,l)) \log\frac{1-\mu(k,l)}{1-\nu(k,l|\bfy)}\right)
\]
or, equivalently, maximizing (using, for short, $\phi = \phi_{a,b}$)
\begin{align*}
F(\bfy) &= \sum_{(k,l) \in \CE} \left(\mu(k,l) \log\nu(k,l|\bfy) +(1- \mu(k,l)) \log(1-\nu(k,l|\bfy))\right)\\
&= \sum_{(k,l) \in \CE} \left(\mu(k,l) \log\phi(y_k, y_{l}) +(1- \mu(k,l)) \log(1-\phi(y_k,y_{l}))\right).
\end{align*}

Note the important simplification compared to the similar function $F$ is t-SNE, in that the logarithm of a potentially large sum is avoided. We have
\begin{align*}
\prt_{y_k} F(\bfy) =& 2 \sum_{l=1}^N \mu(k,l) \prt_{y_k} \log\phi(y_k, y_{l}) + 2 \sum_{l=1}^N (1-\mu(k,l)) \prt_{y_k} \log(1-\phi(y_k, y_{l}))\\
=& 2 \sum_{l=1}^N \mu(k,l) \prt_{y_k} \log\frac{\phi(y_k, y_{l})}{1-\phi(y_k, y_{l})} + 2 \sum_{l=1}^N \prt_{y_k} \log(1-\phi(y_k, y_{l})).
\end{align*}
The optimization can be implemented using stochastic gradient ascent. Introduce random variables $\xi_{kl}$ and $\xi'_{kl}$ both taking value in $\{0,1\}$, all independent of each other and such that $P(\xi_{kl} = 1) = \mu_{kl}$ and $P(\xi'_{kl} = 1) = \epsilon$. Define
\[
H_k(\bfy, \xi, \xi') = 2 \sum_{l=1}^N \xi_{kl} \prt_{y_k} \log\frac{\phi(y_k, y_{l})}{1-\phi(y_k, y_{l})} + 2c_k  \sum_{l=1}^N\sum_{l'=1}^N \xi_{kl}\xi'_{kl'} \prt_{y_k} \log(1-\phi(y_k, y_{l'})).
\]
Then, if one takes $c_k = 1/(\epsilon\sum_{l} \mu(k,l))$ one has
\[
E(H_k(\bfy, \xi, \xi')) = \prt_{y_k} F(\bfy).
\]

 This corresponds to SGA iterations in which: 
\begin{enumerate}
\item Each edge $(k,l)$ is selected with probability $\mu(k,l)$ (which are zero for unless $k$ and $l$ are neighbors); 
\item If $(k,l)$ is selected, one selects an additional edges $(k,l')$ each with probability $\ep$.
\end{enumerate} 
Letting $l_1, \ldots, l_m$ be the number of edges selected,  $y_k$ is updated according to
\[
y_k \leftarrow y_k + 2\gamma \left(\prt_{y_k} \log\frac{\phi(y_k, y_{l})}{1-\phi(y_k, y_{l})} + c_k \sum_{j=1}^m 
\prt_{y_k} \log(1-\phi(y_k, y_{l'}))\right).
\]

\begin{remark}
If one prefers using probability rather than fuzzy set theory, the graphs $\CG^{(k)}$ may also be interpreted as random graphs in which edges are added independently from each other and each edge $(l,l')$  is drawn with probability $\mu^{(k)}(l,l')$. The combined graph $\CG$ is then the random graph in which $(l,l')$ is present if and only if it is in at least one of the $\CG^{(k)}$ and the objective function $C$ coincides with the KL divergence between this random graph and the random graph similarly defined for $\bfy$.

However, this fuzzy/random graph formulation of UMAP---which corresponds to current practical implementations---is only a special case of the theoretical construction made in \citet{mcinnes2020umap} which builds on the theory of (fuzzy) simplicial sets and their representation of metric spaces.  We refer the interested reader to this reference, which requires a mathematical background beyond the scope of these notes. 
\end{remark}

\newpage
\markright{PROBLEMS}
\section*{Problems}
\addcontentsline{toc}{section}{Problems}
\input Problems_Manifold_learning

\chapter{Generalization Bounds}
\label{chap:ms}

We provide, in this chapter, an introduction to some theoretical aspects of  statistical learning, mostly focusing on the derivation of generalization bounds that provide high-probability guarantees on the generalization error of predictors using training data. While these bounds are not always of practical  use, because  making them small
in realistic situations would require an enormous amount of training
data, their derivations and the form they take for specific model classes bring important insight on the structure of the learning problem, and help understand why some methods may perform well while others do not. 

\section{Notation}

We here recall some notation  introduced in \cref{chap:general}. We consider a pair of random variables $(X,Y)$, with $X:\Om\to\CR_X$ and $Y:\Om \to \CR_Y$. Regression problems correspond to $\CR_Y = \mR$ (or $\mR^q$ if multivariate) and classification to $\CR_Y$ being a finite set. A predictor is a function $f: \CR_X\to\CR_Y$. The general prediction problem is to find such a predictor within a class of functions, denoted $\CF$, minimizing the prediction (or generalization) error
\[
R(f) = \myE(r(Y, f(X))),
\]
where $r: \CR_Y\times\CR_Y \to [0. +\infty)$ is a risk function. 

A training set is a family $T = ((x_1, y_1), \ldots, (x_N, y_N)) \in (\CR_X\times \CR_Y)^N$, the set $\CT$ of all possible training sets  therefore being the set of all finite sequences in $\CR_X\times \CR_Y$. A training algorithm can then be seen as a function $\CA: \CT \to \CF$ which associates to each training set $T$ a function $\CA(T) = \hat f_T$.

Given $T \in \CT$, The training set error associated to a function $f\in \CF$ is
\[
\hat R_T(f) = \frac{1}{|T|} \sum_{(x,y)\in T} r(y, f(x)))
\]
and the in-sample error associated to a learning algorithm is the function $T \mapsto \CE_T \defeq \hat R_T(\hat f_T)$.  Fixing the size ($N$) of $T$, one also considers the random variable $\mathbb T$ with values in $\CT$ distributed as an $N$-sample of the distribution of $(X,Y)$. 

A good learning algorithm should be such that the generalization error $R(\hat f_T)$ is small, at least in average (i.e., $E(R(\hf_{\mathbb T}))$ is small). Our main goal in this chapter is to describe generalization bounds trying to find upper-bounds for $R(\hat f_T)$ based on $\CE_T$ and properties of the function class $\CF$. These bounds will reflect the bias-variance trade-off, in that, even though large function classes provide smaller in-sample errors, they will also induce a large additive term in the upper-bound, accounting for the ``variance'' associated to the class. 

\begin{figure}
\centering
\includegraphics[width=0.4\textwidth]{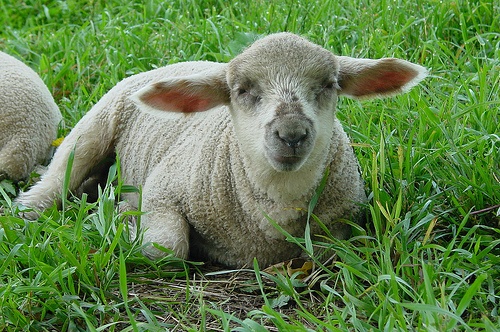}
\includegraphics[width=0.4\textwidth]{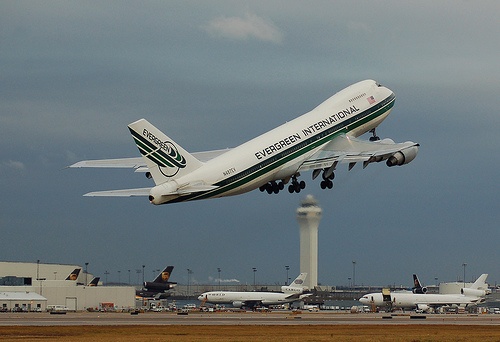}\\
\includegraphics[width=0.4\textwidth]{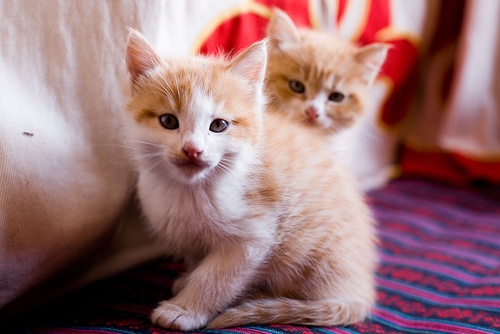}
\includegraphics[width=0.4\textwidth]{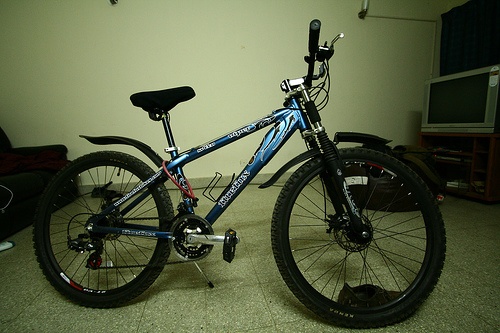}
\caption{\label{fig:pipe} Images extracted from the PASCAL challenge 2007 dataset \citep{everingham2010pascal}, in which categories must be associated with images. There is little ambiguity on correct answers  based on observing the image, i.e., little randomness in the variable $Y$.}
\end{figure}

\begin{remark}
Both variables $X$ and $Y$ are assumed to be random in the previous setting,
but there are often situations when one of them is ``more random'' than the
other. Randomness in $Y$ is associated to measurement errors, or
ambiguity in the decision.  Randomness in $X$ more generally relates to
the issue of sampling a dataset in a large dimensional space. In some
cases, $Y$ is not random at all: for example, in object recognition,
the question of assigning categories for images such as those depicted in  \cref{fig:pipe} has a quasi-deterministic answer. Sometimes, it is $X$ who is not random, for example when observing noisy signals where $X$ is a deterministic discretization of a time interval and $Y$ is some function of $X$ perturbed by noise.  
\end{remark}

\section{Penalty-based Methods and Minimum Description Length}
\label{sec:is.err}

\subsection{Akaike's information criterion}


We make a computation under the following assumptions. We assume a regression model $Y = f_\theta(X) + \ep$ where $\epsilon \sim \CN(0, \sigma^2)$ and $f$ is some function parametrized by $\theta \in \mR^m$. We also assume that the true distribution is actually covered by this model and represented by a parameter $\theta_0$. Let $\hth_T$ denote the parameter estimated by least squares using a training set $T$, and denote for short $\hf_T = f_{\hth_T}$. 

The in-sample error is
\[
\CE_T = \frac1N \sum_{k=1}^N (y_k - \hf_T(x_k))^2.
\]
We  want to compare the training-set-averaged prediction error and the average in-sample error, namely compute the error bias
\[
\Delta_N = \myE(R(f_\mT)) - \myE(\CE_\mT)\,.
\]
Write 
\[
\Delta_N = \myE(R(f_\mT)) - R(f_{\th_0}) + R(f_{\th_0}) - E(\CE_\mT).
\]

We  make a heuristic argument to evaluate $\Delta_N$.
We can use the fact that $\hth_T$ minimizes the empirical error and write
\[
\frac1N \sum_{k=1}^N (Y_k-f_{\th_0}(X_k))^2 = \CE_\mT +  \sigma^2 (\hth_\mT - \th_0)^T J_{\mT} (\hth_\mT - \th_0) + o(|\hth_\mT - \th_0|^2)
\]
with 
\[
J_T = \frac1{2\sigma^2 N} \sum_{k=1}^N \prt_\th^2 ((y_k - f_\th(x_k))^2)_{{\th=\hth_T}},
\]
which is an $m$ by $m$ symmetric matrix.

Now, using the fact that $\th_0$ minimizes the mean square error (since $f_{\th_0}(x) = \myE(Y\mid X=x)$), we can write, for any $T$:
\[
R(f_T) = R(f_{\th_0}) +  \sigma^2 (\hth_T - \th_0)^T I  (\hth_T - \th_0) + o(|\hth_T - \th_0|^2)
\]
with 
\[
I = \frac1{2\sigma^2}  \myE(\prt_\th^2 (Y-f_{\th}(X))^2_{{\th=\th_0}}).
\]

As a consequence, we can write (taking expectations in both Taylor expansions)
\[
\Delta_N = \sigma^2 \myE \left((\hth_\mT - \th_0)^T J_{\mT} (\hth_\mT - \th_0) \right)+ \sigma^2 \myE\left((\hth_\mT - \th_0)^T I  (\hth_\mT - \th_0)\right) + o(\myE(|\hth_\mT - \th_0|^2)).
\]
(We skip hypotheses and justification for the analysis of the residual term.)

We now note that, because we are assuming a Gaussian noise, and that the true data distribution belongs to the parametrized family, the least-square estimator is also a maximum likelihood estimator. Indeed, the likelihood of the data is
\[
\frac{1}{(2\pi\sigma^2)^{N/2}} \exp\left(- \frac1{2\sigma^2}\sum_{k=1}^N (Y_k - f_{\th}(X_k))^2\right) \prod_{k=1}^N \phi_X(X_k)
\]
where $\phi_X$ is the p.d.f.  of $X$ and does not depend on the unknown parameter. We can therefore apply classical  results from mathematical statistics \citep{van2000asymptotic}. Under some mild smoothness assumptions on the mapping $\th \mapsto f_{\th}$,  $\hth_\mT$ converges to $\th_0$ in probability when $N$ tends to infinity, the matrix $J_\mT$ converges to $I$, which is the model's Fisher information matrix, and $\sqrt N (\hth_\mT - \th_0)$ converges in distribution to a Gaussian $\CN(0, I^{-1})$ . This implies that both $N(\hth_\mT - \th_0)^T J_{\mT} (\hth_\mT - \th_0)$ and $N(\hth_\mT - \th_0)^T I  (\hth_\mT - \th_0)$ converge to a chi-square distribution with $m$ degrees of freedom, whose expectation is $m$, which indicates that $\Delta_N$ has order $2\sigma^2m/N$.

This analysis can be used to develop model selection rules, in which one chooses between models of dimensions $k_1 <k_2 < \cdots < k_q = m$ (e.g., by truncating the last coordinates of $X$).  The rule suggested by the previous computation is to  select $j$ minimizing  
\[
\CE^{(j)}_T(\hf_T) + \frac{2\sig^2 k_j}N,
\]
  where $\CE^{(j)}$ is the in-sample error computed using the $k_j$-dimensional model.
This is an example of a penalty-based method, using the so-called Akaike's
information criterion (AIC) \citep{akaike1973information}.

\subsection{Bayesian information criterion and minimum description length}
Other penalty-based methods are more size-averse and replace  the
constant, $2$, in AIC by  a function of $N = |T|$, for example
$\log N$. Such a change can be justified by a Bayesian
analysis, yielding the Bayesian information criterion (BIC) \citep{schwarz1978estimating}. The approach in this case is not based on an evaluation of the error, but on an asymptotic estimation of the posterior distribution resulting from a Bayesian model selection principle.  Like in the previous section, we content ourselves with a heuristic discussion.

Let us consider a statistical model parametrized by $\th\in \Th$, where $\Th$ is an open convex subset of $\mR^m$ with p.d.f. given by
\[
f(z;\theta) = \exp(\theta^TU(z) - C(\th))\,,
\]
with $U: \mR^d \to \mR^m$and $z = (x,y)$. We are given a family of sub-models represented by $\CM_1, \ldots, \CM_q$, where, for each $j$, $\CM_j$ is the intersection of $\Theta$ with a $k_j$-dimensional affine subspace of $\mR^m$. We are also given a prior distribution for $\th$ in which a sub-model is first chosen, with probabilities $\al_1, \ldots, \al_q$, and given that, say, $\CM_j$ is selected, $\th\in \CM_j$ is chosen with a probability distribution with density $\phi_j$ with respect to Lebesgue's measure on $\CM_j$ (denoted $dm_j$). Given training data $T = (z_1, \ldots, z_N)$, Bayesian model selection consists in choosing the model $\CM_j$ where $j$ maximizes the posterior log-likelihood
\[
\mu(\CM_j|T) = \log\int_{\mR^m} \al_j e^{N(\th^T \bar U_T - C(\th))} \phi_j dm_j(\th)
\]
where $\bar U_T = (U(z_1) + \cdots + U(z_N))/N$. 

Consider the maximum likelihood estimator $\hat \th_j$ within  $\CM_j$, maximizing $\ell(\theta, \bar U_T) = \th^T \bar U_T - C(\th)$ over $\CM_j$. Then one has
\[
\ell(\th, \bar U_T) = \ell(\hat\th_j, \bar U_T) + \frac12 (\th - \hat \th_j)^T \prt^2_\th\ell(\hth_j, \bar U_T)(\th-\hth_j) + R_j(\th, \hth_j) |\th-\hth_j|^3
\]
Note that the first derivative of $\ell$ is $\prt_\th \ell  = \bar U - E_\th(U)$ where $E_\th$ is the expectation for $f(\cdot, \th)$. The second derivative is $-\var_\th(U)$ (showing that $\ell$ is concave) and the third derivative involves third-order moments of $U$ for $E_\th$ and (like the second derivative) does not depend on $\bar U_T$. In particular, we can assume that, for any $M>0$, there exists a constant $C_M$ such  that whenever $\max(|\th|, |\hth_j|) \leq M$, we have $R_j(\th, \hth_j) \leq C_M$.

The law of large numbers implies that $\bar U_T$ converges to a limit when $N$ tends to infinity, and our assumptions imply that $\hth_j$ converges to the parameter providing the best approximation of the distribution of $Z$ for the Kullback-Leibler divergence. In particular, with probability 1, there exists an $N$ such that $\hth_j$ belongs to any large enough, but fixed, compact set. Moreover, the second derivative $\ell(\hth_j)$ will also converge to a limit, $-\Sig_j$. 

For any $\ep>0$, write
\begin{multline*}
\int_{\mR^m} \al_j e^{N(\th^T \bar U_T - C(\th))} \phi_j dm_j(\th) \\
= \int_{|\th - \hth_j| \leq \ep} e^{N(\th^T \bar U_T - C(\th))} \phi_j dm_j(\th) + \int_{|\th - \hth_j| \geq \ep} e^{N(\th^T \bar U_T - C(\th))} \phi_j dm_j(\th)\,.
\end{multline*}
The second integral  converges to $0$ exponentially fast when $N$ tends to $\infty$. The first one  behaves essentially like 
\[
\int_{\CM_j} e^{- \frac12 N (\th - \hat \th_j)^T \Sig_j^{-1} (\th - \hat \th_j) + \log\phi_j(\th)} dm_j(\th)\,.
\]
Neglecting $\log\phi_j(\th)$, this integral behaves like $(2\pi\det (\Sig_j/N))^{-1/2}$, whose logarithm is $(-k_j(\log N)/2)$ plus constant terms. As a consequence, we find that 
\[
\mu(\CM_j\mid T) = \max_{\th\in \CM_j} \ell(\th) - \frac {k_j}2 \log N + \text{ bounded terms}.
\]

Consider, as an example, linear regression  with $Y = \be_0 + b^T x + \sig^2 \nu$ where $\nu$ is a standard Gaussian random variable. Assume that the distribution of $X$ is known, or, preferably, make the previous discussion conditional to $X_1, \ldots, X_N$. 
Let sub-models $\CM_j$ correspond to the assumption that all but the first $k_j-1$ coefficients of $b$ vanish. 
Then, up to bounded terms, the Bayesian estimator must minimize (over such parameters $b$)
\[
\frac1{2\sig^2} \sum_{k=1}^N (y_k - \be_0 - b^T x_k)^2 + \frac {k_j} 2 \log N\,.
\]
or
\[
\CE^{(j)}_T + \frac{k_j\sig^2}N \log N\,.
\]

We now turn to another
interesting point of view, which provides the same penalty, based
on maximum description length principle (MDL; \citet{rissanen1989stochastic})  measuring the coding efficiency of a model.

Let us fix some notation. We assume that one has $q$ competing models
for predicting $Y$ from $X$, for example, linear
regression models based on different subsets of the explanatory
variables. Denote these models $\CM_1, \ldots, \CM_q$. Each model will
be seen, not as an assumption on the true joint distribution of $X$ and
$Y$, but rather as a tool to efficiently encode the training set
$((x_1, y_1), \ldots, (x_N, y_N))$. To describe MDL, which selects the model that provides the most efficient code, we need to
reintroduce a few basic concepts of information theory.

The entropy of a discrete probability $P$ over a set $\Om$ is 
$$
\CH_2(P) = - \sum_{x\in \Om} p_x \log_2p_x.
$$
(The logarithm in base 2 is used because of the tradition of coding with
bits in information theory.)

For a discrete random variable $X$, the entropy $\CH_2(X)$ is $\CH_2(P_X)$
where $P_X$ is the probability distribution of $X$. The relation
between the entropy and coding theory is as follows: a code is a
function which associates to any element $\om\in \Om$ a string of bits
$c(\om)$. The associated code-length is denoted $l_c(\om)$, which is simply
the number of bits in $c(\om)$. When $P$ is a probability on $\Om$,
the efficiency of a code is measured by the average code-length:
$$
E_P(l_c) = \sum_{\om\in\Om} l_c(\om) P(\om).
$$

Shannon's theorem \citep{shannon1948mathematical, cover2012elements} states that, under some conditions on the code
(ensuring that any sequence of words can be recognized as soon as it is observed: one says that it is instantaneously decodable) the average
code length can never be larger than the entropy of $P$. Moreover, it states that
there exists codes that achieve this lower bound with no more than one
bit loss,  such that for all $\om$, $l_c(\om) \leq -
\log_2(P(\om)) + 1$. These optimal codes, such as the Huffman code \citep{cover2012elements}, can completely be determined from the
knowledge of $P$. This allows one to interpret a probability $P$ on $\Om$
as a tool for designing codes with code-lengths essentially equal to   ($-\log_2
P$).

This statement can be generalized to continuous random
variables (replacing the discrete probability $P$ by a probability density function, say $\phi$) if one introduces a coding precision level, denoted $\de_0$, meaning that the decoded values may differ by no more than $\de_0$ from the encoded ones. The result is that
the optimal code-length at precision $\de_0$ can be estimated (up to one extra bit) by $- \log_2 \phi - \log_2\de_0$. 

In our context, each model of the conditional distribution of $Y$ given $X$, with conditional density 
 $\phi(y|x)$, provides a way to encode the training
set with a total code length, for $(y_1, \ldots, y_N)$, of
$$
- \sum_{k=1}^N \log_2 \phi(y_k\mid x_k) - N\log_2\de_0
$$
(working, as before, conditionally to $x_1, \ldots, x_N$).
We assume that the precision at which the data is encoded is fixed,
which implies 
that the last term does not affect the model choice. Now, assume a sequence of $m$  parametrized model classes, $\CM_1, \ldots, \CM_m$ and let $\phi(y\mid x,\th, \CM_j)$ denote the conditional distribution with parameter $\th$ in the class $\CM_j$. 
Within model $\CM_j$, the optimal code length corresponds to the
maximum likelihood:
$$
- \sum_{k=1}^N \log_2 \phi(x, y\mid \hat \th_j, \CM_j) = - \max_{\th}\left(\sum_{k=1}^N \log_2 \phi(x, y; \th, \CM_j)\right).
$$

If the models are nested, which is often the case,  the most efficient  will always be the
largest model, since the maximization is on a larger set. However,  the minimum description length (MDL) principle uses the fact that, in
order to decode the compressed data, the model, including its optimal
parameters, has to be known, so that the complete code  needs to
include a model description. The decoding algorithm will then
be: decode the model, then use it to decode the data. 

So assume that a model (one of the $\CM_j$'s) has a $k_j$-dimensional
parameter $\th$. Also assume that a probability distribution, $\pi(\th\mid \CM_j)$, is used to
encode $\th$. Also choose a precision level, $\de_{ij}$, for each coordinate in
$\th$, $i=1, \ldots, k_j$. (Previously, we could consider the precision of
the $y_k$, $\de_0$, as fixed, but now, the precision level for parameters is a variable that will be optimized.) The total description length using this model now becomes
$$
- \sum_{k=1}^N \log_2 \phi(y_k\mid x_k; \th, \CM_j) - \log_2\pi(\th\mid \CM_j) -
\sum_{i=1}^{k_j} \log_2(\de_{ij}).
$$

Let $\hth^{(j)}$ be the parameter that maximizes
$$
L(\th\mid \CM_j)  = \sum_{k=1}^N \log_2 \phi(y_k\mid x_k; \th, \CM_j) + \log_2\pi(\th\mid \CM_j)$$ 

If $\pi$ is interpreted as a prior distribution of the parameters,
$\hth^{(j)}$ is the {\em maximum a posteriori} Bayes estimator. We now take
the correction caused by 
$(\de_{ij}, i=1, \ldots, k_j)$ into account, by assuming that the $i$th coordinate in $\hth^{(j)}$
is truncated to  $-\log_2 \de_i$ bits. Let
$\bth^{(j)}$ denote this approximation. A second-order expansion of
$L(\th|\CM_j)$ around $\hth^{(j)}$ yields (assuming sufficient differentiability) 
$$
L(\bth^{(j)}\mid \CM_j) = L(\hth^{(j)}\mid \CM_j) + \frac{1}{2} (\bth^{(j)} - \hth^{(j)})^T S_{\hth^{(j)}}
(\bth^{(j)} - \hth^{(j)}) + o(|\bth^{(j)}- \hth^{(j)}|^2)
$$
where $S_\th$ is the matrix of second derivatives of  $L(\cdot\mid \CM_j)$ at
$\th$. Approximating $\bth^{(j)} - \hth^{(j)}$ by $\de^{(j)}$ (the
$k_j$-dimensional vector with coordinates $\de_{ij}$, $i=1, \ldots, k_j$), we see that the precision
should maximize
$$
\frac{1}{2} (\de^{(j)})^T S_{\hth^{(j)}} \de^{(j)} + \sum_{i=1}^{k_j} \log_2\de_{ij}.
$$
Note that $S_{\hth^{(j)}}$ must be negative semi-definite, since $\hth$ is a
local maximum. Assuming it is non-singular, the previous
expression can be maximized and yields
\begin{equation}
\label{max.riss}
S_{\hth^{(j)}} \de^{(j)} = -\frac{1}{\log 2} \frac{1}{\de^{(j)}}
\end{equation}
where $1/\de^{(j)}$ is the vector with coordinates $(1/\de_{ij})$.

Let us now make an asymptotic evaluation. Because  $L(\th\mid \CM_j)$
includes a sum over $N$ independent terms, it is reasonable to assume that $S_{\hth^{(j)}}$ has order
$N$, and more precisely, that $S_{\hth^{(j)}}/N$ has a limit. Rewrite
 \cref{max.riss} as
$$
\frac{S_{\hth^{(j)}}}{N} {\sqrt{N}\de^{(j)}} = -\frac{1}{\log 2} \frac{1}{\sqrt{N}\de^{(j)}}\,.
$$
This implies that $\sqrt{N}\de^{(j)}$ is the solution of an equation which
stabilizes with $N$, and it is therefore reasonable to assume that the optimal $\de_{ij}$ takes the form $\de_{ij} =
c_i(N\mid \CM_j)/\sqrt{N}$, with $c_i(N\mid \CM_j)$ converging to some limit when $N$ tends
to infinity. The total cost can therefore be estimated by
$$
- L(\hth^{(j)}\mid \CM_j) + \frac{k_j}{2} \log_2 N - \frac{k_j}{2} - \sum_{i=1}^{k_j} \log_2 c_i(N\mid \CM_j)
$$
The last two terms are O(1), and can be neglected, at least when $N$
is large compared to $k_j$. The final criterion becomes the
penalized likelihood
$$
l_d(\th\mid \CM_j) = L(\th|\CM_j) - \frac{k_j}{2} \log_2 N
$$
in which we see that the dimension of the model appears with a factor
$\log_2N$ as announced (one needs to normalize both terms by $N$ to
compare with the previous paragraph). 


\section{Concentration inequalities}
\label{sec:ci}

The discussion of the AIC was a first attempt at evaluating a prediction error.
It was however done under very specific parametric assumptions, including the fact that the true distribution of the data was within the considered model class.
It was, in addition, a bias evaluation, i.e., we estimated how much, in average, the in-sample error was less than the generalization error.
We would like to obtain upper bounds to the generalization error that hold with high probability, and rely as little as possible on assumptions on the true data distribution. 

One of the main tools used in this context are concentration inequalities, which provide upper bounds on the various probabilities of events involving a large number of random variables. The current section provides a review of some of these inequalities.


\subsection{Cram\'er's theorem}
If $X_1, X_2, \ldots$ are independent, integrable random variables with identical distributions (to that of a random variable $X$), the law of large numbers tells us that the empirical mean $\bar X_N = (X_1 + \cdots + X_N)/N$ converges with probability one to $m=E(X)$. When the variables are square integrable, Chebychev's inequality provides an easy proof of the weak  law of large numbers. Indeed,
\[
\myP\big(|\bar X_n - m| > \ep\big) \leq \frac{1}{\ep^2} \myE\big((\bar X_N - m)^2\bfone_{|\bar X_n - m| > \ep}\big) \leq \frac{\mathrm{var}(\bar X_N)}{\ep^2} = \frac{\mathrm{var}(X)}{N\ep^2}.
\]

A stronger assumption on the moments of $X$ yields a stronger inequality. One says that $X$ has exponential moments if there exists $\la_0 > 0$ such that $\myE(e^{\la_0|X|}) < \infty$. In this case, the cumulant-generating function, defined, for $\la\in\mR$, by
\begin{equation}
\label{eq:laplace}
M_X(\la) = \log \myE(e^{\la X}) \in [0, +\infty], 
\end{equation}
is finite for $\la \in [-\la_0, \la_0]$. 

Here are a few straightforward properties of the cumulant-generating function. 
\begin{enumerate}[label=(\roman*)]
\item One has $M_X(0) = 0$. 
\item For any $a\in \mR$, one has
$M_{aX}(\la) = M_X(a\la)$. 
\item If $X_1$ and $X_2$ are independent variables, one also has
\[
M_{X_1+X_2}(\la) = M_{X_1}(\la) + M_{X_2}(\la).
\]
In particular, $M_{X+a}(\la) = M_X(\la) + \la a$, so that $M_{X-E(X)}(\la)  = M_X(\la) - \la E(X)$.
\item
Finally, Markov's inequality (which states that, for any non-negative variable $Y$, $P(Y> t) \leq E(Y)/t$) applied to $Y = e^{\la X}$ for $\la> 0$ yields
\begin{equation}
\label{eq:markov.exp}
\myP(X>t) = \myP(e^{\lambda X} > e^{\lambda t}) \leq e^{M_X(\la) - \la t}\,.
\end{equation}
(Note that this inequality is trivially true for $\lambda=0$.)
\end{enumerate}

From these properties, one can easily derive a concentration inequality for the mean of independent random variables. 
We have $M_{\bar X_N}(\la) = N M_X(\la/N)$ and applying \eqref{eq:markov.exp} we get, for any $\la \geq 0$ and $t>0$
\[
\myP(\bar X_N - m > t) \leq e^{-\la(m+t) + M_{\bar X_N}(\la)} = e^{-N \left(\frac{\la(m+t)}N - M_X\left(\frac{\la}N\right)\right)}
\]
where the right-hand side may be infinite. Because this inequality is true for any $\la$, we have
\[
\myP(\bar X_N - m > t) \leq e^{-N M_{X,+}^*(m+t)}
\]
where $M_{X,+}^*(u) = \sup_{\la\geq 0} (\la u - M_X(\la))$, which is non-negative since the maximized quantity vanishes for $\lambda=0$. A symmetric computation yields   
\[
P(\bar X_N - m < -t) \leq e^{-N M_{X,-}^*(m-t)}
\]
where $M_{X,-}^*(t) = \sup_{\la\leq 0} (\la t - M_X(\la))$, which is also non-negative. 

Let 
\begin{equation}
\label{eq:cramer}
M_X^*(t) = \sup_{\la\in \mR} (\la t - M_X(\la)) \geq 0
\end{equation}
(this is the Fenchel-Legendre transform of the cumulant generating function, sometimes called the Cram\'er transform of $X$). One  has $M_X^*(m+t) = M^*_{X,+}(m+t)$ for $t>0$. Indeed, because $x\mapsto e^{\la x}$ is convex, Jensen's inequality implies that 
\[
\myE(e^{\la X}) \geq e^{\la m}
\]
so that $\la (m+t) - M_X(\la) \leq \la t < 0$ if $\la<0$.  Similarly,  $M_X^*(m-t) = M^*_{X,-}(m-t)$ for $t >0$. We therefore have the following result.
\begin{theorem}
\label{th:cramer.inf}
Let $X_1, \ldots, X_N$ be independent and identically distributed random variables. Assume that these variables are integrable and let $m = \myE(X_1)$. Then, for all $t>0$,
\[
\myP(\bar X_N - m > t) \leq e^{-N M_{X}^*(m+t)}
\]
and
\[
\myP(|\bar X_N - m| > t) \leq 2e^{-N \min(M_X^*(m+t), M_X^*(m-t))}
\]
\end{theorem}
The last inequality derives from
\[
\myP(|\bar X_N - m| > t)  = \myP(\bar X_N - m > t) + \,uP(\bar X_N - m < -t)\,.
\]

This is our first example of concentration inequality that shows that, when 
\[
\min(M_X^*(m+t), M_X^*(m-t))>0,
\]
 the probability of a deviation by $t$ at least of $\bar X_n$ from its mean decays exponentially fast.  The derivation of the inequality above was quite easy: apply Markov's inequality in a parametrized form and optimize over the parameter. It is therefore surprising that this inequality is sharp, in the sense that a similar lower bound also holds. Even though we are not going to use it in the rest of this chapter, it is worth sketching the argument leading to this lower bound, which involves an interesting step making a change of measure. 

Assume (without loss of generality) that $m=0$ and consider $\myP(\bar X_n > t)$. Assume, to simplify the discussion, that the supremum of $\la \mapsto \ep\la - M_X(\la)$ is attained at some $\la_t$. We have 
\[
\prt_\la M_X(\la) = \frac{\myE(Xe^{\la X})}{\myE(e^{\la X})}.
\]
Let $q_\la(x) = \frac{e^{\la x}}{\myE(e^{\la X})}$ and $\myP_\la$ (with expectation $\myE_\lambda$) the probability distribution on $\Omega$ with density $q_\la(X)$ with respect to $\myP$, so that $\prt_\la M_X(\la) = \myE_\la(X)$. We have, since $\la_t$ is a maximizer, $\myE_{\la_t}(X) = t$. Moreover, fixing $\de >0$, 
\begin{align*}
\myP(\bar X_N > t) &= \myE(\bfone_{\bar X_N > t}) \\
& \geq   \myE(\bfone_{ |\bar X_n - t-\de| < \de}) \\
& \geq  \myE\big(\bfone_{ |\bar X_n - t -\de| < \de} e^{N\la \bar X_N - Nt - 2N\de}\big) \\
& = e^{-N(t+2\de)}  M_X(\la)^N \myP_\la ( |\bar X_N - t-\de | < \de)
\end{align*}
If one takes $\la = \la_{t+\de}$, this implies that
\[
\myP(\bar X_N > t) \geq e^{-N M_X^*(t+\de)} e^{-N\de} \myP_{\la_{t+\de}} ( |\bar X_N - t-\de| < \de)\,.
\]
By the law of large numbers (applied to $\myP_{\la_{t+\de}}$), $\myP_{\la_{t+\de}} ( |\bar X_N - t-\de| < \de)$ tends to 1 when $N$ tends to infinity. This implies that the logarithmic rate of convergence to 0 of $\myP(\bar X_N > t)$ is larger than $N(M_X^*(t+\de) + \de)$, for any $\de>0$, to be compared with the rate $NM_X^*(t)$ for the upper bound.
In Large Deviation theory,  the upper and lower bounds are often simplified by considering the limit of 
$\log \myP(\bar X_N > t)/N$, which, in this case, is  $M_X^*(t)$ (and this result is called Cram\'er's therorem).  

 While Cram\'er's upper bound is sharp, its computation requires an exact knowledge of the distribution of $X$, which is not a common situation. 
The following sections optimize the upper bound in situations where only partial information on the variable is known, such as its moments or its range. As a first example, we consider concentration of the mean for sub-Gaussian variables.

\subsection{Sub-Gaussian variables}

If $X$ has exponential moments, then, (applying again Markov's inequality)
\[
P(|X| > x) \leq C e^{-\la x}
\]
for some positive constants $C$ and $\la$. Reducing if needed the value of $\la$, one can assume that $C$ takes some predetermined (larger than 1) value, say, $C=2$, the simple argument being left to the reader. A random variable such that, for some $\la>0$ 
\[
\myP(|X| > x) \leq 2 e^{-\la x}
\]
is called sub-exponential (and this property is equivalent to $X$ having exponential moments). Similarly, one says that $X$ is sub-Gaussian if, some $\sig >0$, 
\begin{equation}
\label{eq:subgauss}
\myP(|X| > x) \leq 2 e^{- \frac{x^2}{2\sig^2}}.
\end{equation}
Sub-Gaussian random variables are such that $M(\la) < \infty$ for all $\la\in \mR$. Indeed, for $\la>0$
\begin{align*}
\myE(e^{\la |X|}) &= \int_0^\infty \myP(e^{\la |X|} > z) dz \\
&= 1 + \int_1^\infty \myP(|X| > \la^{-1} \log z) dz\\
& \leq 1 + 2 \int_1^\infty e^{-\frac{(\log z)^2}{2\sig^2\la^2}} dz\\
& \leq 1 + 2 \int_1^\infty e^{x - \frac{x^2}{2\la^2\sig^2}} dx\\
&\leq 1 +  2\sqrt{2\pi} \la \sig e^{\frac{\la^2\sig^2}2}\,.
\end{align*}

\begin{proposition}
\label{prop:subg}
Assume that $X$ is sub-Gaussian, so that \eqref{eq:subgauss} holds for some $\sig^2>0$. Then, for any $t>0$, we have
\[
\myP(\bar X_n - E(X) > t) \leq \left(1 + \frac{4t^2}{\sig^2}\right)^N e^{-\frac{Nt^2}{2\sig^2}}\,.
\]
\end{proposition}
\begin{proof}
Let us assume, without loss of generality, that $\myE(X) = 0$. For  $\la>0$, we then have
\[
\myE(e^{\la X}) = 1 + E(e^{\la X} - \la X - 1)\,.
\]
Let $\phi(t) = e^t - t - 1$. We have $\phi(t) \geq 0$ for all $t$, $\phi(0) = 0$ and, for $z>0$, the equation $z = \phi(t)$ has two solutions, one positive and one negative that we will denote $g_+(z) > 0 > g_-(z)$. We have
\begin{align*}
\myE(\phi(\la X)) &= \int_0^\infty \myP(\phi(\la X) > z) dz \\
&= \int_0^\infty \myP(\la X > g_+(z)) dz + \int_0^\infty \myP(\la X < g_-(z)) dz 
\end{align*}
The change of variable $u = g_+(z)$ in the first integral is equivalent to $u>0$, $\phi(u) = z$ with $dz = (e^u-1) du$. Similarly, $u=-g_-(z)$ in the second integral gives $u>0$, $\phi(-u) = z$ and $dz = (1-e^{-u})du$ so that
\begin{align*}
\myE(\phi(\la X)) &= \int_0^\infty \myP(\la X > u) (e^u-1)du + \int_0^\infty \myP(\la X < -u) (1-e^{-u}) du\\
& \leq \int_0^\infty \myP(\la |X| > u) (e^u-e^{-u}) du.
\end{align*}
(Using the fact that $\max(\myP(\la X > u), \myP(\la X < -u)) \leq \myP(\la|X| > u)$.) We have
\begin{align*} 
\int_0^\infty \myP(\la |X| > u) (e^u-e^{-u}) du &\leq 2 \int_0^{+\infty}  (e^{u} - e^{-u}) e^{- \frac{u^2}{2\la^2\sig^2}} du \\
&= 2\la\sig \int_0^{+\infty} (e^{\la\sig v} - e^{-\la\sig v}) e^{- \frac{v^2}{2}} dv\\
&= 2\la\sig e^{\frac{\la^2\sig^2}2} \sqrt{2\pi} (\Phi(- \sig\la) - \Phi(\sig\la))\\
&\leq 4\la^2\sig^2 e^{\frac{\la^2\sig^2}2}
\end{align*}
where $\Phi$ is the cumulative distribution function of the standard Gaussian and we have used $\Phi(-t) - \Phi(t) \leq 2t/\sqrt{2\pi}$.
We therefore have
\[
M_X(\la) \leq \log\left(1 + 4\la^2\sig^2 e^{\frac{\la^2\sig^2}2}\right) 
\leq \frac{\la^2\sig^2}2 + \log(1 + 4\la^2\sig^2)\,.
\]
This implies
\[
M_X^*(t) =   \sup_{\la>0} (\la t - M_X(\la)) \geq \frac{t^2}{\sig^2} - M_X(t/\sig^2) \geq \frac{t^2}{2\sig^2} - \log(1 + \frac{4t^2}{\sig^2})
\]
so that 
\[
\myP(\bar X_n > t) \leq \left(1 + \frac{4t^2}{\sig^2}\right)^N e^{-\frac{Nt^2}{2\sig^2}}\,.
\]
\end{proof}

The following result allows one to control the expectation of a non-negative sub-Gaussian random variable. 
\begin{proposition}
\label{prop:subg.exp}
Let $X$ be a non-negative random variable such that 
\[
\myP(X>t) \leq Ce^{-t^2/2\sig^2}
\]
for some constants $C$ and $\sig^2$. Then, 
\[
\myE(X) \leq 3 \sig \sqrt{\log C}.
\]
\end{proposition}
\begin{proof}
For any $\al\in (1, C]$, one has
\[
\min(1,  Ce^{-t^2/2\sig^2}) \leq \al e^{-\frac{t^2\log\al}{2\sig^2\log C}},
\]
which implies that 
\[
\myE(X) = \int_0^{+\infty} \myP(X>t) dt \leq \frac{\al}{2\log\al} \sqrt{2\pi} \sig \sqrt{\log C}
\]
Taking $\al = \sqrt{e}$ gives
\[
\myE(X) \leq \sqrt{\pi e} \sig \sqrt{\log C} \leq 3 \sig \sqrt{\log C}.
\]
\end{proof}

\subsection{Bennett's inequality}
The following proposition (see \citep{bennett1962probability}) provides an upper bound for $M_X(\la)$ as a function of $\myE(X)$ and $\var(X)$ under the additional assumption that $X$ is bounded from above.
\begin{proposition}
\label{prop:bennett}
Let $m = \myE(X)$ and assume that for some constant $b$, one has  $X\leq b$ with probability one. Then, for any $\sig^2>0$ such that $\var(X) \leq \sig^2$, one has
\begin{equation}
\label{eq:bennett}
\myE(e^{\la X}) \leq e^{\la m} \left(\frac{(b-m)^2}{(b-m)^2 + \sig^2} e^{- \frac{\la\sig^2}{(b-m)}} + \frac{\sig^2}{(b-m)^2+\sig^2} e^{\la(b-m)}\right)
\end{equation}
for any $\lambda\geq 0$.
\end{proposition}
\begin{proof}
There is no loss of generality in assuming that $m=0$ and $\la = 1$, in which case one must show that 
\begin{equation}
\label{eq:bennett.2}
\myE(e^{X}) \leq \frac{b^2}{b^2 + \sig^2} e^{- \frac{\sig^2}{b}} + \frac{\sig^2}{b^2+\sig^2} e^{b}
\end{equation}
if $X<b$ and $E(X^2) \leq \sig^2$.
Indeed, if this inequality is true for $m=0$ and $\la=1$, \eqref{eq:bennett} in the general case will result from letting $X = Y/\la + m$ and applying the special case to $Y$.

The right-hand side of \eqref{eq:bennett.2} is exactly $\myE(e^X)$ when $X$ follows the discrete distribution $P_0$ supported by two points $x_0$ and $b$, and such that  $E(X)=0$ and $\myE(X^2) = \sig^2$, which requires $x_0 = -\sig^2/b$ and $P(X=x_0) = b^2/(\sig^2 + b^2)$.

Now consider the quadratic function $v(x) = \al x^2 + \be x + \ga$ which intersects $x\mapsto e^x$ at $x=x_0$ and $x=b$, and is tangent to it at $x=x_0$, i.e.,  $v(b) = e^b$ and $v(x_0) = v'(x_0) = e^{x_0}$ (this uniquely defines $v$). Then $e^x \leq v(x)$ for $x<b$, yielding
\[
\myE(e^X) \leq \al\sig^2 + \ga.
\]
However, since $v(X) = e^X$ almost surely when $X\sim P_0$,  this upper bound is  attained and equal to that provided in \eqref{eq:bennett.2}.
\end{proof}

If $F(\la)$ denotes the right-hand side of \eqref{eq:bennett},  we have, for $m\leq u < b$, 
\[
M_X^*(t) \geq \sup_{\la\geq 0} (\la u - \log F(\la))
\]
and we now estimate this lower bound. Maximizing $\la y - \log F(\la)$ is equivalent to minimizing
\[
\la \mapsto \frac{(b-m)^2 e^{-\frac{\la(\sig^2+(u-m))}{b-m}} + \sig^2 e^{\la (b-u)}}{(b-m)^2 + \sig^2}.
\]
Introduce the notation $\rho = \sig^2/(b-m)^2$, $\mu = \la(b-m)$ and $x = (u-m)/(b-m)$, so that the function to minimize is
\[
\mu\mapsto 
\frac{e^{-\mu(\rho+x)} + \rho e^{\mu (1-x)}}{1+\rho}.
\]
Computing the derivative in $\mu$ and equating it to 0 gives 
\[
\mu = \frac{1}{1+\rho}\log \frac{\rho+x}{\rho(1-x)},
\]
which is non-negative since $\rho + x - \rho(1-x) = (1+\rho) x$. 
For this value of $\mu$, we have 
\begin{align*}
\frac{e^{-\mu(\rho+x)} + \rho e^{\mu (1-x)}}{\rho+1} & = e^{-\mu(\rho+x)} \frac{1 + \rho e^{\mu (1+\rho)}}{\rho+1}\\
&= e^{-\mu(\rho+x)} \frac{1 + \rho\frac{\rho+x}{\rho(1-x)}}{\rho+1}\\
&= \frac{e^{-\mu(\rho+x)}}{1-x}
\end{align*} 
and
\begin{align*}
-\log\frac{e^{-\mu(\rho+x)} + \rho e^{\mu (1-x)}}{\rho+1} & = \mu(\rho+x) + \log(1-x)\\
&= \frac{\rho+x}{1+\rho}\log \frac{\rho+x}{\rho(1-x)} + \log(1-x) \\
&= \frac{\rho+x}{1+\rho}\log \frac{\rho+x}{\rho} + \frac{1-x}{1+\rho}\log(1-x)\,.
\end{align*}
This provides a lower bound for $M_X^*(m + (b-m)x)$, and yields the following corollary.
\begin{corollary}
\label{coro:bennett}
Assume that  $X$ satisfy the conditions of \cref{prop:bennett}. Then
\begin{equation}
\label{eq:bennett.bound}
\myP(\bar X_N > m+t) \leq \exp\left(-N\left(\frac{\rho+x}{1+\rho}\log \frac{\rho+x}{\rho} + \frac{1-x}{1+\rho}\log(1-x)\right)\right)
\end{equation}
with $x = t/(b-m)$ and $\rho = \sig^2/(b-m)^2$. 
\end{corollary}

Bennett's inequality is sometimes stated in a slightly weaker, but simpler form \citep{massart2007concentration}. Returning to the proof of  \cref{prop:bennett} and using the fact that $\log u \leq u-1$,  equation \eqref{eq:bennett.2} implies
\begin{align*}
\log \myE(e^{X}) &\leq \frac{b^2}{b^2 + \sig^2} e^{- \frac{\sig^2}{b}} + \frac{\sig^2}{b^2+\sig^2} e^{b} - 1\\
& = \frac{b^2}{b^2 + \sig^2} (e^{- \frac{\sig^2}{b}} + \frac{\sig^2}{b} - 1) + \frac{\sig^2}{b^2+\sig^2} (e^{b} - b - 1).
\end{align*}
We will use the following lemma.
\begin{lemma}
\label{lem:phi.inc}
The function $\phi: u\mapsto (e^u-u-1)/u^2$ is non-decreasing.
\end{lemma}
\begin{proof}
We have $\phi'(u) = \psi(u)/u^3$ where $\psi(u) = ue^u - 2e^u + u + 2$, yielding $\psi'(u) = u e^u - e^u + 1$, $\psi''(u) = u e^u$. Therefore,  $\psi'$ is has its minimum at $u=0$ with $\psi'(0)=0$ so that $\psi$ is increasing. Since $\psi(0)=0$, we have $\psi(u)/u^3 \geq 0$. 
\end{proof} 
We therefore have
\begin{align*}
\log \myE(e^{X}) &\leq \frac{b^2}{b^2 + \sig^2} (e^{- \frac{\sig^2}{b}} + \frac{\sig^2}{b} - 1) + \frac{\sig^2}{b^2+\sig^2} (e^{b} - b - 1)\\
& = \frac{b^2}{b^2 + \sig^2} \frac{\sig^4}{b^2} \phi(-\sig^2/b) + \frac{\sig^2}{b^2+\sig^2} b^2\phi(b)\\
&\leq \left(\frac{\sig^4}{b^2 + \sig^2} + \frac{\sig^2b^2}{b^2+\sig^2}\right)\phi(b)\\
&=\frac{\sig^2}{b^2} (e^b - b - 1)
\end{align*}
This shows that 
\[
\log \myE(e^{\la X}) \leq \frac{\sig^2}{b^2} (e^{\la b} - \la  b - 1)
\] 
and 
\[
M_X^*(t) \geq \frac{\sig^2}{b^2} \max_\la (\la b^2 t/\sig^2 - e^{\la b} + \la  b + 1) = \frac{\sig^2}{b^2} h(bt/\sig^2)
\]
where $h(u) = (1+u)\log(1+u) - u$.

We summarize this in the following corollary.
\begin{corollary}
\label{coro:bennett.2}
Assume that  $X$ satisfy the conditions of \cref{prop:bennett}. Then, for $t>0$, 
\begin{equation}
\label{eq:bennett.bound.2}
\myP(\bar X_N > m+t) \leq \exp\bigg(-\frac{N\sig^2}{(b-m)^2} h\Big(\frac{(b-m)t}{\sig^2}\Big)\bigg)
\end{equation}
where $h(u) = (1+u)\log(1+u) - u$.
\end{corollary}

This estimate can be further simplified as follows. Let $g$ be such that  $g''(u) = (1+u/3)^{-3}$ and $g(0) = g'(0) = 0$, which gives
$g(u) = u^2/(2 + 2u/3)$.
Noting that $h''(u) = (1+u)^{-1}$ and that $(1+u)^{-1} \geq (1+u/3)^{-3}$, for $u\geq 0$ we find, integrating twice, that $h(u) \geq g(u)$ for $u\geq 0$.   This shows that the following upper-bound is also true:
\begin{equation}
\label{eq:bennett.bound.3}
\myP(\bar X_N > m+t) \leq \exp\bigg(- \frac{Nt^2}{2\sig^2 + 2t(b-m)/3}\bigg)\,.
\end{equation}
This upper bound is known as Bernstein's inequality.

\begin{remark}
It should be clear that, in the previous discussion, one may relax the assumption that $X_1, \ldots, X_N$ are identically distributed as long as there is a common function $M$ such that $M_{X_k}(\la) \leq m_k + M(\la)$ for all $k$, with $m_k = \myE(X_k)$. We have in this case
\[
\myP(\bar X_N > \bar m_N +t) \leq \exp(-N M^*(t))
\]
with $\bar m_N = (m_1 + \cdots + m_N)/N$ and $M^*(t) = \sup_\la (\la t - M(\la))$. This remark can be, in particular, applied to the situation in which $X_1, \ldots, X_N$ satisfy the conditions of  \cref{prop:bennett} with the same constants $b$ and $\sig^2$, yielding the same upper bound as in  equation \eqref{eq:bennett.bound}. 
\end{remark}


\subsection{Hoeffding's inequality}

We now consider the case in which the random variables $X_1, \ldots, X_N$ are bounded from above and from below,  and start with the following consequence of  \cref{prop:bennett}.
\begin{proposition}
\label{prop:ci.bounded}
Let $X$ be a random variable taking values in the interval $[a,b]$. Let $m=\myE(X)$. Then
\begin{equation}
\label{eq:ci.bounded}
\myE(e^{\la X}) \leq \frac{b-m}{b-a}e^{\la a} + \frac{m-a}{b-a}e^{\la b} \leq e^{\la m} e^{\frac{\la^2(b-a)^2}8}
\end{equation}
for all $\lambda\in\mR$.
\end{proposition}
\begin{proof}
We first note that, if $X$ takes values in $[a,b]$, then
$\var(X) \leq (b-m)(m-a)$ (using $\sig^2 = (b-m)(m-a)$ in \cref{eq:bennett}). To prove the upper bound on the variance, introduce the function $g(x) = (x-a)(x-b)$ so that $g(x) \leq 0$ on $[a,b]$. Noting that one can write $g(x) = (x-m)^2 + (2m-a-b)(x-m) + (a-m)(b-m)$, we have
\[
\myE(g(X)) = \var(X) - (b-m)(m-a) \leq 0,
\]
which proves the inequality.

This shows that, if $\la\geq 0$, we can apply  \cref{prop:bennett} with $\sigma^2 = (b-m)(m-a)$, which provides the first inequality in \cref{eq:ci.bounded}. To handle the case $\lambda  \leq 0$, it suffices to apply this inequality with $\tilde \lambda = -\lambda$, $\tilde X = -X$, $\tilde a=-b$, $\tilde b = -a$ and $\tilde m = -m$.

The second inequality, namely 
\[
\left(\frac{b-m}{b-a}e^{\la a} + \frac{m-a}{b-a}e^{\la b}\right) \leq e^{\la m} e^{\frac{\la^2(b-a)^2}8}
\]
requires a little additional work. Letting $u = (m-a)/(b-a)$, $\al = \la (b-a)$ and taking logarithms, we need to prove that
\[
\log(1 -u + u e^\al)  - u\al \leq \frac{\al^2}8
\]
Let $f(\al)$ denote the difference between the right-hand side and  left-hand side. Then $f(0) =  0$,
\[
f'(\al) = \frac{\al}4 - \frac{ue^\al }{1 - u +ue^\al} + u,
\]
(so that $f'(0) = 0$) and
\[
f''(\al) = \frac14 - \frac{u(1-u)e^\al}{(1 - u + ue^\al)^2}\,.
\]
For positive numbers $x=1-u$ and $y = ue^\al$, one has $(x+y)^2 \geq 4xy$, which shows that $f''(\al) \geq 0$. This proves that $f'$ is non-decreasing with $f'(0) = 0$, proving that $f$ is minimized at $\al=0$, so that $f(\al) \geq 0$ as needed. 
\end{proof}

We can then deduce the following theorem \citep{hoeffding1994probability}. 
\begin{corollary}[Hoeffding Inequality]
\label{coro:hoeffding}
If $X_1, \ldots, X_N$ are independent, taking values, respectively, in intervals of length, $c_1, \ldots, c_N$  and
$Y = X_1 + \cdots + X_N$, then
\begin{equation}
\label{eq:hoeffding}
\myP(Y > \myE(Y)+ t) \leq \exp\left(-\frac{2t^2}{|c|^2}\right)
\end{equation}
and
\begin{equation}
\label{eq:hoeffding.sym}
\myP(Y < \myE(Y)- t) \leq \exp\left(-\frac{2t^2}{|c|^2}\right)
\end{equation}
where $|c|^2 = \sum_{k=1}^N c_k^2$. 
\end{corollary}
\begin{proof}
We have, by  \cref{prop:ci.bounded}, for any $\la>0$
\[
\myP(Y > \myE(Y) + t) \leq e^{- \big(\la t -\sum_{k=1}^N M_{X_k}(\la )\big)} \leq e^{-(\la t -\frac{\la^2}8 |c|^2)}
\]
The upper bound is minimized for $\la = 4t/|c|^2$, yielding \cref{eq:hoeffding}.  \Cref{eq:hoeffding.sym} is obtained by applying \cref{eq:hoeffding} to $-X$.
\end{proof}

An important special case of this inequality is when $X_1, \ldots, X_N$ are i.i.d. taking values in an interval of length $\delta$. Then
\begin{equation}
\label{eq:hoeffding.2}
\myP(\bar X_N > \myE(X)+ t) \leq \exp\left(-\frac{2Nt^2}{\delta^2}\right).
\end{equation}
This inequality is obtained after applying Hoeffding's inequality to $X_1/N, \ldots, X_N/N$, therefore taking $c_1=\cdots=c_N = \delta/N$ and $|c|^2 = \delta^2/N$.

\subsection{McDiarmid's inequality}

One can  relax the assumption that the random variables $X_1, \ldots, X_N$ are independent and only assume that these variables behave like ``martingale increments,'' as stated in the following proposition \citep{dembo2011large}.
\begin{proposition}
\label{prop:martingale.0}
Let $X_1, \ldots, X_N$, $Z_1, \ldots, Z_N$ be two sequences of $N$ random variables such that 
\[
\myE(Z_k\mid X_1,Z_1, \ldots, X_{k-1}, Z_{k-1}) = m_k
\]
is constant and  $|Z_k - m_k| \leq c_k$ for some constants $c_1, \ldots, c_N$. Then
\[
\myP(Y > \myE(Y) +t) \leq e^{- 2t^2/|c|^2}
\]
with $Y=Z_1+\cdots + Z_N$ and $|c|^2 = \sum_{k=1}^N c_k^2$.

\end{proposition}
\begin{proof}
 \Cref{prop:ci.bounded} applied to the conditional distribution implies that, for $\la\geq 0$:
\[
\log \myE(e^{\la (Z_k-m_k)}\mid X_1,Z_1, \ldots, X_{k-1}, Z_{k-1}) \leq \log \myE(e^{\la |Z_k-m_k|}\mid X_1,Z_1 \ldots, X_{k-1}, Z_{k-1})\leq  \frac{\la^2 c_k^2}8\,.
\]
Let $S_k = \sum_{j=1}^k (Z_j- m_j)$. Then
\[
\myE(e^{\la S_k}) = \myE(e^{\la S_{k-1}}E(e^{\la (Z_k-m_k)}\mid X_1,Z_1, \ldots, X_{k-1}, Z_{k-1})) \leq e^{\frac{\la^2 c_k^2}8} \myE(e^{\la S_{k-1}})
\]
so that 
\[
\myE(e^{\la S_N})\leq e^{\frac{\la^2}8 \sum_{k=1}^N c_k^2}
\]
and the result follows from Markov's inequality optimized over $\la$.
\end{proof}

We will use this proposition to prove the ``bounded difference,'' or McDiarmid's inequality.
\begin{theorem}[McDiarmid's inequality]
\label{th:mdd}
Let $X_1, \ldots, X_N$ be independent random variables and $g: \mR^N \to \mR$ a function such that there exists $c_1, \ldots, c_N$ such that
\begin{equation}
\label{eq:mdd}
|g(x_1, \ldots, x_{k-1}, x_k, x_{k+1}, \ldots, x_N) - g(x_1, \ldots, x_{k-1}, \tilde x_k, x_{k+1}, \ldots, x_N)| \leq c_k
\end{equation}
for all $k=1, \ldots, N$ and $x_1, \ldots, x_{k-1}, x_k,\tilde x_k, x_{k+1}, \ldots, x_N$. 
Then
\[
\myP\left(g(X_1, \ldots, X_N) > \myE(g(X_1, \ldots, X_N))+ t\right) \leq e^{-2t^2/|c|^2}
\]
with $|c|^2 = c_1^2 + \cdots + c_N^2$.
\end{theorem}
\begin{proof}
Let $m = \myE(g(X_1, \ldots, X_N))$. Let $Z_0=0$, 
\[
Y_k = \myE(g(X_1, \ldots, X_N)\mid X_1, \ldots, X_k) - m
\]
 and $Z_k = Y_k - Y_{k-1}$. Note that $Z_k$ is a function of $X_1, \ldots, X_k$ and can therefore be omitted from the conditional expectation given $(X_1, Z_1, \ldots, X_{k-1}, Z_{k-1})$.
  
We have $\myE(Y_k) = 0$ and $\myE(Y_k\mid X_1, \ldots, X_{k-1}) = Y_{k-1}$ so that $\myE(Z_k\mid X_1, \ldots, X_{k-1}) = 0$. Because the variables are independent,  we have, letting $\tilde X_1, \ldots, \tilde X_N$ be independent copies of $X_1, \ldots, X_N$,
\begin{multline*}
Z_k = \myE(g(X_1, \ldots,X_{k-1}, X_k, \tilde X_{k+1}, \ldots,  \tilde X_N)\mid X_1, \ldots, X_k) \\
- \myE(g(X_1, \ldots, X_{k-2}, \tilde X_{k-1}, \tilde X_k,\ldots,  \tilde X_N)\mid X_1, \ldots, X_{k-1})\,.
\end{multline*}
For fixed $X_1, \ldots, X_{k-1}$,  \cref{eq:mdd} implies that $Z_k$ varies in an interval of length $c_k$ at most (whose bounds depend on $X_1, \ldots, X_{k-1}$) so that $|Z_k - E(Z_k)| \leq c_k$.  \Cref{prop:martingale.0} implies that
\[
\myP(Z_1+\cdots+Z_N \geq t) \leq  e^{-2t^2/|c|^2},
\]
which concludes the proof since 
\[
Z_1+\cdots+Z_N = g(X_1, \ldots, X_N) - \myE(g(X_1, \ldots, X_N)).
\]
\end{proof}

\subsection{Boucheron-Lugosi-Massart inequality}
The following result \citep{boucheron2000sharp}, that we state without proof, extends on the same idea.
\begin{theorem}
\label{th:blm}
Let $X_1, \ldots, X_N$ be independent random variables. Let 
\[
Z = g(X_1, \ldots, X_N)
\]
 with $g: \mR^N \to [0, +\infty)$ and for $k=1, \ldots, N$, 
\[
Z_k = g_k(X_1, \ldots, X_{k-1}, X_{k+1}, \ldots, X_N)
\]
with $g_k: \mR^{N-1} \to \mR$. Assume that, for all $k=1, \ldots, N$, one has $0\leq Z-Z_k \leq 1$ and that 
\[
\sum_{k=1}^N (Z-Z_k) \leq Z.
\]
Then 
\[
\myP(Z - \myE(Z) > t) \leq \exp(-\myE(Z) h(t/\myE(Z))) \leq \exp\left(-\frac{t^2}{2\myE(Z) +2t/3}\right)
\]
where $h(u) = (1+u) \log(1+u) - u$. Moreover, for $t < \myE(Z)$,
\[
\myP(Z - \myE(Z) < -t) \leq \exp(-\myE(Z) h(-t/\myE(Z))) \leq \exp(- t^2/2\myE(Z))\,.
\]

Finally, for all $\la\in \mR$
\begin{equation}
\label{eq:blm}
\log \myE(e^{\la(Z-\myE(Z))}) \leq \myE(Z) (e^\la - \la - 1).
\end{equation}
\end{theorem}

\section{Bounding the empirical error with the VC-dimension}

\subsection{Introduction}
 \Cref{sec:ci} provides some of the most important inequalities used to evaluate the deviation of various combinations of independent random variables (e.g., their empirical mean) from their expectations (the reader may refer to \citet{ledoux1991probability,dgl96,talagrand2006generic,dembo2011large,vershynin2018high} and other textbooks on the subject for further developments). 

We now return to   the problem of estimating the generalization error based on training data.
 For a given predictor $f$, concentration bounds allow us to control the probability
\[
\myP(R(f) - \hat R_\mT(f) > t)
\]
where
\[
R(f) = \myE(r(Y, f(X))
\]
and
\[
\hat R_T(f) = \frac{1}{N} \sum_{k=1}^N r(y_k, f(x_k))
\]
for a training set  $T = (x_1,y_1, \ldots, x_N, y_N)$.

If this probability is small, then $R(f) \leq \hat R_{\mT}(f) + t$ with high probability, providing a likely upper bound to the generalization error of $f$. For example, if $r$ is the 0--1 loss in a classification problem, Hoeffding's inequality implies, for training sets of size $N$, 
\[
\myP(R(f) - \hat R_\mT(f) > t) \leq e^{-2Nt^2}\,.
\]

Now  \cref{coro:hoeffding} does not hold if we replace $f$ by $\hf_\mT$, i.e., if $f$ is estimated from the training set $\mT$, which is, unfortunately, the situation we are interested in. Before addressing this problem, we point out that this inequality does apply to the case in which $f = \hf_{\mT_0}$ where $\mT_0$ is another training set, independent from $\mT$, so that
\[
\myP(R(\hf_{\mT_0}) - \hat R_\mT(\hf_{\mT_0}) > t) \leq e^{-2Nt^2}\,, 
\]
which is proved by writing
\[
\myP( R(\hf_{\mT_0}) - \hat R_\mT(\hf_{\mT_0}) > t) = \myE(\myP(R(\hf_{T})-\hat R_\mT(\hf_{T})\mid > t|\mT_0 = T))\,.
\]
In this situation, the empirical risk is computed on a test or validation set ($\mT$) independent of the set used to estimate $f$ ($\mT_0$).

If one does not have a test set, and $\hf_\mT$ is optimized over a set $\CF$ of possible predictors, one can rarely do much better than starting from  a variation of the trivial upper bound 
\[
\myP(R(\hf_\mT) - \CE_\mT > t) \leq \myP\Big(\sup_{f\in \CF} (R(f) - \CE_\mT(f)) > t\Big)
\]
(with $\CE_\mT = \hat R_\mT(\hf_\mT)$) and the concentration inequalities discussed in  \cref{sec:ci} need to be extended to provide upper bounds to the right-hand side. 
\begin{remark}
\label{rem:measurability}
 Computing supremums of functions over non countable sets may bring some issues regarding measurability.
To avoid complications, we will always assume, when computing supremums over infinite 
 sets, that such supremums can be reduced to maximizations over finite sets, i.e., 
when considering $\sup_{f\in\CF} \Phi(f)$ for some function $\Phi$, we will assume that there exists a nested sequence of finite subsets $\CF_n\sub\CF$ such that
\begin{equation}
\label{eq:measurability}
\sup\{\Phi(f): f\in \CF\}  = \lim_{n\to\infty} \sup\{\Phi(f): f\in \CF_n\}\,.
\end{equation}
This is true, for example, when $\CF$ has a topology that admits a countable dense subset, with respect to which $\Phi$ is continuous.
\end{remark}

When $\CF$ is a finite set, one can use a ''union bound'' with
\[
\myP(\sup_{f\in \CF} (R(f) - \CE_\mT(f)) > t) \leq \sum_{f\in\CF} \myP(R(f) - \CE_\mT(f) > t) \leq |\CF| \max_{f\in \CF} \myP(R(f) - \CE_\mT(f) > t).
\]
Such bounds cannot be applied to the typical case in which $\CF$ is infinite, and is likely to provide very poor estimates even when $\CF$ is finite, but $|\CF|$ is large. 
However, all proofs of concentration inequalities applied to such supremums require using a union bound at some point, often after considerable preparatory work. Union bounds will in particular appear in conjunction with  the Vapnik-Chervonenkis dimension that we now discuss.

\subsection{Vapnik's theorem}

We consider a classification problem with two classes, 0 and 1, and therefore let $\CF$ be a set of binary functions, i.e., taking values in $\{0,1\}$. We also assume that the risk function $r$ takes values in the interval $[0,1]$ (using, for example, the 0--\!1 loss).
Let
\begin{equation}
\label{eq:ut}
U(t) = \myP\Big(\sup_{f\in\CF} (R(f) - \CE_\mT(f)) > t\Big).
\end{equation}

A fundamental theorem of Vapnik  provides an estimate of $U(t)$ based on the
number of possible ways to split a training set of $2N$ points into two
classes using functions in $\CF$. The rest of this section is devoted to a discussion of this result and related notions. 

If $A$ is a finite subset of $\CR$, we let $\CF(A)$ denote the set $\{f_{|_A}: f \in \CF\}$ of restrictions of elements of $\CF$ to the set $A$. As a convention, we let $\CF(\emptyset) = \{f_\emptyset\}$, containing the so-called empty function.
Since $\CF$ only contains binary functions, we have $|\CF(A)| \leq 2^{|A|}$. If $x_1, \ldots, x_M\in \CR$, we let, with a slight abuse of notation,
\[
\CF(x_1, \ldots, x_M) = \CF(A)
 \]
 where $A = \{x_i, i=1, \ldots, M\}$.
 This provides the number of possible splits of a training set  $T = (x_1, \ldots, x_M)$ using classifiers in $\CF$. Fixing in this section a random variable $X$, we let
\[
S_\CF(M) = E(|\CF(X_1, \ldots, X_M)|)
\]
where
 the expectation is taken over all
$M$ i.i.d. realizations  from $X$. 
We also let
\[
S_\CF^*(M) = \max\{ |\CF(A)|: A\sub\CR, |A| \leq M\}.
\]

The following theorem controls $U$ in \cref{eq:ut} in terms of $S_\CF$.
\begin{theorem}[Vapnik]
\label{th:vc}
With the notation above, one has, for $t \geq \sqrt{2/N}$: 
\begin{equation}
\label{eq:vc}
\myP\Big(\sup_{f\in\CF} (R(f) - \CE_\mT(f)) > t\Big) \leq 2 S_\CF(2N) e^{-Nt^2/8},
\end{equation}
which implies that, with probability at least $1-\de$, we have
\begin{equation}
\label{eq:vc.delta}
\forall f \in\CF: R(f) \leq  \CE_\mT(f)) + \sqrt{\frac8N\left(\log S_\CF(N) + \log \frac2\de\right)}
\end{equation} 
\end{theorem}
(The requirement that $t \geq\sqrt{2/N}$ does not really reduce the range of applicability of \cref{eq:vc}, since, for $t\leq \sqrt{2/N}$,  the upper bound in that equation is typically much larger than 1.)
\begin{proof}
We first show that the problem can be symmetrized with the inequality, valid  if $Nt^2 \geq 2$, 
\begin{equation}
\label{eq:vc.symm}
\myP\Big(\sup_{f\in\CF} (R(f) - \CE_\mT(f)) \geq t\Big) \leq 2\myP\Big(\sup_{f\in\CF} (\CE_{\mT'}(f) - \CE_{\mT}(f)) \geq \frac t 2\Big)
\end{equation}
in which $\mT'$ is a second training set (independent of $\mT$) with $N$ samples also. In view of assumption \cref{eq:measurability}, there is no loss of generality in assuming that $\CF$ is finite. Associate to  any training set $T$, a classifier $f_T\in \CF$ maximizing  $R(f_T) - \CE(f_T) $. One then has
\begin{align*}
\myP\left(\sup_{f\in\CF} (\CE_{\mT'}(f) - \CE_{\mT}(f)) \geq \frac t 2\right) & \geq \myP\left((\CE_{\mT'}(f_\mT) - \CE_{\mT}(f_\mT)) \geq \frac t 2\right)\\
& \geq \myP\left((R(f_\mT) - \CE_{\mT'}(f_{\mT}) \leq \frac t 2\text{ and } R(f_\mT) - \CE_{\mT}(f_T)) \geq t\right)\\
& = \myE\left( \bfone_{R(f_\mT) - \CE_{\mT}(f_\mT)) \geq  t} \myP\Big(R(f_\mT) - \CE_{\mT'}(f_{\mT}) \leq \frac t 2\ \Big|\ \mT\Big) \right)
\end{align*}
Conditional to $\mT$, $\CE_{\mT'}(f_{\mT})$ is the average of $M$ i.i.d. Bernoulli random variables, with variance bounded from above by $1/4$ and 
\[
\myP\Big(R(f_\mT) - \CE_{\mT'}(f_{\mT}) \leq \frac t 2\ \Big|\ \mT\Big) \geq 
 1 - \frac{1/4}{Nt^2/4} \geq \frac12\,.
\]
It follows that 
\begin{multline*}
\myP\Big(\sup_{f\in\CF} (\CE_{\mT'}(f) - \CE_{\mT}(f)) > \frac t2\Big) \\
\geq \frac12 \myP\Big(R(f_\mT) - \CE_{\mT}(f_\mT)) \geq  t\Big) = \frac12 \myP\Big(\sup_{f\in\CF} (R(f) - \CE_T(f)) \geq t\Big).
\end{multline*}
This justifies \cref{eq:vc.symm}.

%

Now consider a family of independent Rademacher random variables $\bfxi_1, \ldots, \bfxi_N$, also independent of $\mT$ and $\mT'$, taking values $-1$ and $+1$ with equal probability. By symmetry, 
\[
\sup_{f\in\CF}  (\CE_{\mT'}(f) - \CE_{\mT}(f)) = \sup_{f\in\CF}  \sum_{k=1}^N (r(Y_k, f(X_k)) - r(Y'_k, f(X'_k)))/N
\]
has the same distribution as 
\[
\sup_{f\in\CF}  \sum_{k=1}^N \bfxi_k (r(Y_k, f(X_k)) - r(Y'_k, f(X'_k)))/N\,.
\]
Now, there are at most $|\CF(X_1, \ldots, X_N, X'_1, \ldots, X'_N)|$ different sets of coefficients in front of $\xi_1, \ldots, \xi_N$ in the above sum when $f$ varies in $\CF$, so that, conditioning on $\mT, \mT'$ and taking a union bound , we have
\begin{align*}
&\myP\Big(\sup_{f\in\CF}  \sum_{k=1}^N \bfxi_k (r(Y_k, f(X_k)) - r(Y'_k, f(X'_k)))/N \geq t/2\ \Big|\ \mT, \mT'\Big)\\
& \leq 
 |\CF(X_1, \ldots, X_N, X'_1, \ldots, X'_N)|  \\
 & \qquad\qquad \sup_{f\in\CF}\myP\Big(\sum_{k=1}^N \bfxi_k (r(Y_k, f(X_k)) - r(Y'_k, f(X'_k)))/N \geq t/2\ \Big|\ \mT, \mT'\Big)
\end{align*}

The variables $\bfxi_k(r(Y_k, f(X_k)) - r(Y'_k, f(X'_k))$ are centered and belong to the interval $[-1,1]$, which has length 2, so that
 Hoeffding's inequality implies
\[
\myP\Big(\sum_{k=1}^N \bfxi_k (r(Y_k, f(X_k)) - r(Y'_k, f(X'_k)))/N \geq t/2\ \Big|\  \mT, \mT'\Big) \leq e^{-2N(t/2)^2/4} = e^{-Nt^2/8}
\]
and taking expectation over $\mT$ and $\mT'$ yields
\[
\myP\left(\sup_{f\in\CF} (R(f) - \CE_\mT(f)) \geq t\right) = 2 S_\CF(2N)e^{-Nt^2/8}.
\]

\Cref{eq:vc.delta} is then obtained from letting $\de = 2 S_\CF(2N)e^{-Nt^2/8}$ so that 
$t = \sqrt{\frac8N \log \frac{2 S_\CF(2N)}{\delta}}$ with $R(f) \leq \CE_{\mT}(f) + t$ for all $f$ with probability $1-\delta$ or more.
\end{proof}
 
\subsection{VC dimension}

To obtain a practical bound, the quantity $S_\CF(2N)$, or its upper-bound $S_\CF^*(2N)$, needs to be estimated.
We prove below an important property
of $S_\CF^*$, namely that, either $S_\CF^*(M) = 2^M$ for all $M$, or there
exists an $M_0$ for which $S_\CF^*(M_0) < 2^{M_0}$, and taking $M_0$ to be
the largest one for which an equality occurs, $S_\CF^*(M)$ has order $M^{M_0}$ for all $M\geq
M_0$. This motivates the following definition of the VC-dimension of the model class.
\begin{definition}
\label{def:vc.dim}
The Vapnik-Chervonenkis dimension (or VC dimension) of  the model class $\CF$ is 
\[
\VC(\CF)  = \max\{M: S^*_{\CF}(M) = 2^M\}\,.
\]
(where the infimum of an empty set is $+\infty$).
\end{definition}

\begin{remark}
\label{rem:vc}
If, for a finite set $A \subset \CR$, one has $|\CF(A)| = 2^{|A|}$, one says that {\em $A$ is shattered by $\CF$}. So $\VC(\CF)$ is the largest integer $M$ such that there exists a set of cardinality $M$ in $\CR$ that is shattered by $\CF$.
\end{remark}

We now evaluate the growth of $S^*_\CF(M)$ in terms of the VC-dimension, starting with the following lemma, which states that, if $A$ is a finite subset of $\CR$, there are at least $|\CF(A)|$ subsets of $A$ that are shattered by $\CF$. 
\begin{lemma}[Pajor]
\label{lem:pajor}
Let $A$ be a finite subset of $\CR$. Then
\[
|\CF(A)| \leq |\{B\sub A: |\CF(B)| = 2^B\}|  \,.
\]
\end{lemma}
\begin{proof}
 The statement holds for $A = \emptyset$, for which $|\CF_\emptyset|=1 = 2^0$. 
 For $|A|=1$,  the upper-bound is either $1$ if $|\CF(A)| = 1$, or 2 if $|\CF(A)| = 2$, and the collection of sets  $B\sub A$ such that $|\CF(B)| = 2^B$ is $\{\emptyset\}$ in the first case and $\{\emptyset, A\}$ in the second one. 
 So, the statement is true for $|A|=0$ or 1.

Proceeding by induction, assume that the result is true if $|A| \leq N$, and consider a set $A'$ with $|A'| = N+1$. Assume that $|\CF(A')| \geq 2$ (otherwise there is nothing to prove), which implies that there exists $x\in A'$ such that $|\CF(x)|=2$.  Take such an $x$ and write $A' = A \cup \{x\}$ with $x\not \in A$. Let
\[
\CF_0 = \defset{f\in \CF: f(x) = 0} \text{ and } \CF_1 = \defset{f\in \CF: f(x) = 1}.
\]
Since $\CF_0 \cap \CF_1 = \emptyset$, we have 
\[
|\CF(A')| = |\CF_0(A')| + |\CF_1(A')|.
\]
Since $f(x)$ is constant on $\CF_0$ (resp. $\CF_1$), we have  $|\CF_0(A')| = |\CF_0(A)|$ (resp.  $|\CF_1(A')| = |\CF_1(A)|$), and the induction hypothesis implies
\begin{align*}
|\CF(A')| & \leq |\{B\sub A: |\CF_0(B)| = 2^B\}| + |\{B\sub A: |\CF_1(B)| = 2^B\}| \\
& = |\{B\sub A: |\CF_0(B)| = 2^B \text{ or } |\CF_0(B)| = 2^B\}| \\
&+ |\{B\sub A: |\CF_0(B)| =|\CF_1(B)| = 2^B\}| .
\end{align*}
If $B\sub A$ is shattered by $\CF_0$ or $\CF_1$, it is obviously shattered by $\CF$. Moreover, if $B$ is shattered by both, then $B\cup \{x\}$ is shattered by  $\CF$. The upper bound in the equation above is therefore less than the total number of sets shattered by $\CF$, which proves the lemma.
\end{proof}

From this lemma, it results that if $\VC(\CF) = D<\infty$, then $S^*_\CF(M)$ is bounded by the total number of subsets of cardinality $D$ or less in a set of cardinality $M$. This provides the following result, which implies that the term in front of the exponential in \cref{eq:ut} grows polynomially in $N$ if $\CF$ have finite VC-dimension.
\begin{proposition}[Sauer-Shelah's lemma]
\label{prop:sauer}
If $D$ is the VC-dimension of $\CF$, then, for $N\geq D$,
\[
S^*_\CF(N) \leq \left(\frac{eN}{D}\right)^D\,.
\]
\end{proposition}
\begin{proof}
Pajor's lemma implies that 
\[
S^*_\CF(N) \leq \sum_{k=0}^D \binom{N}{k}
\]
and the statement of the proposition derives from the standard upper bound
\[
\sum_{k=0}^D \binom{N}{k} \leq \left(\frac{eN}{D}\right)^D
\]
that we now justify for completeness. We have
\[
\binom{N}{k} = \frac{N!}{(N-k)!k!} \leq \frac{N^k}{k!} \leq \frac{N^D}{D^D}  \frac{D^k}{k!}
\]
if $k\leq D \leq N$. This yields
\[
\sum_{k=0}^D \binom{N}{k} \leq \frac{N^D}{D^D}  \sum_{k=0}^D \frac{D^k}{k!} \leq \frac{N^De^D}{D^D} 
\]
as required.
\end{proof}

We can therefore state a corollary to  \cref{th:vc} for model classes with finite VC-dimension.
\begin{corollary}
\label{corr:vc}
Assume that $\VC(\CF) = D < \infty$. Then, for $t\geq \sqrt{2/N}$ and $N\geq D$, 
\begin{equation}
\label{eq:vc.corr}
\myP\Big(\sup_{f\in\CF} (R(f) - \CE_\mT(f)) > t\Big) \leq 2 \left(\frac{2eN}D\right)^D e^{-Nt^2/8}.
\end{equation}
and
\begin{equation}
\label{eq:vc.corr.2}
\myP\Big(\sup_{f\in\CF} (R(f) - \CE_\mT(f)) \leq \sqrt{\frac{8}N}\sqrt{D\log\frac{eN}D + \log\frac2\delta } \Big) \geq 1 - \delta.
\end{equation}
\end{corollary}

\subsection{Examples}

The following result provides the VC-dimension of the collection of linear classifiers.
\begin{proposition}
\label{prop:vc.lin}
Let $\CR = \mR^d$ and $\CF = \defset{x \mapsto \sign(a_0 + b^Tx): \be_0\in \mR, b \in \mR^d}$. Then 
\[
\VC(\CF) = d+1.
\]
\end{proposition}
\begin{proof}
Let us show that no set of $d+2$ points can be shattered by $\CF$. Use the notation $\tilde x = (1, x^T)^T$ and $\be = (a_0, b^T)^T$, and consider $d+2$ points $x_1, \ldots, x_{d+2}$. Then $\tx_{1}, \ldots, \tx_{d+2}$ are linearly dependent and one of them, say, $\tx_{d+2}$ can be expressed as a linear combination of the others. Write 
\[
 \tx_{d+2} = \sum_{k=1}^{d+1} \alpha_k \tx_{k}\,.
 \]
 Then there is no function $f\in \CF$ (taking the form $\tilde x \mapsto \sign(\be \tilde x)$) that maps $(x_{1}, \ldots, x_{d+2})$ to $(\sign(\alpha_1), \ldots, \sign(\alpha_{d+1}), -1)$ (where the definition of $\sign(0) = \pm 1$ is indifferent), since any such function satisfies
\[
\be^T \tx_{d+2} = \sum_{k=1}^{d+1} \alpha_k \be^T \tx_{k} > 0\,.
 \]
 This proves $\VC(\CF) < d+2$. To prove that $\VC(\CF) = d+1$, it suffices to exhibit a set of $d+1$ vectors in $\mR^d$ that can be shattered by $\CF$. Choose $x_{1}, \ldots, x_{d+1}$ such that  $\tx_{1}, \ldots, \tx_{d+1}$ are linearly independent (for example $x_{i} = \sum_{k=1}^{i-1} e_i$, where $(e_1, \ldots, e_d)$ is the canonical basis of $\mR^d$). This linear independence implies that, for any vector $\alpha = (\alpha_1, \ldots, \alpha_{d+1})^T\in \mR^{d+1}$, there exists a vector $\be\in \mR^{d+1}$ such that $\tilde x_i^T\be = \alpha_i$ for all $i=1, \ldots, d+1$. This shows that any combination of signs for $\tilde x_i^T\be$ can be achieved, so that $(x_1, \ldots, x_{d+1})$ is shattered.
\end{proof}

%
Upper-bounds on VC dimensions of more complex models have also been proposed in the literature. As an example, the following theorem, that we provide without proof, considers feed-forward neural networks with piecewise  linear units (such as ReLU, see \cref{chap:neural.nets}). This theorem is a special case of Theorem 7 in \citet{bartlett2019nearly}, in which the more general case of networks with piecewise polynomial units is provided. Given integers $L$, $U_1, \ldots, U_L$ and $W_1, \ldots, W_L$, define the function class 
\[
\CF(L,(U_i), (W_i), p)
\]
 that consists of feed-forward neural networks with $L$ layers, $U_i$ piecewise linear computational units with less than $p$ pieces in the $i$th layer,  and such that the total number of parameters involved in layers $1,2, \ldots, j$ is less than $W_j$. 
 
\begin{theorem}
\label{th:vc.nn}
\[
\VC(\CF(L,(U_i), (W_i), p)) = O(\bar L W_L \log(pU)).
\]
where $U = U_1 + \cdots + U_L$ and
\[
\bar L = \frac1{W_L} \sum_{j=1}^L W_j.
\]
\end{theorem}
Note that $p=2$ for ReLU networks. Theorem 7 in \citet{bartlett2019nearly} also provides a more explicit upper bound, namely
\[
\VC(\CF(L,(U_i), (W), p)) \leq L + \bar L W_L \log_2\left(4ep \sum_{i=1}^L i U_i \log_2\left(\sum_{i=1}^L(2epi U_i)\right)\right).
\]

\subsection{Data-based estimates}
Approximations of the shattering numbers  can be computed using training
data. One can, in particular, prove a concentration inequality \citep{boucheron2000sharp} on $\log S_\CF(X_1, \ldots, X_N)$, which may in turn be used to estimate $\log(S_\CF(2N))$. In the following, we let
$\CH_{\mathit{VC}}(N, \CF)$ denote the expectation of $\log S_\CF(X_1, \ldots, X_N)$. It is often referred to as the VC entropy of $\CF$. 
\begin{theorem}
\label{th:vc.conc}
One has, letting $\CH_{\mathit{VC}} = \CH_{\mathit{VC}}(N, \CF)$:
\[
\myP(\log S_\CF(X_1, \ldots, X_N) \geq \CH_{\mathit{VC}} + t) \leq \exp\left(- \frac{t^2}{2\CH_{\mathit{VC}} + 2t/3}\right)
\]
and
\[
\myP(\log S_\CF(X_1, \ldots, X_N) \leq \CH_{\mathit{VC}} - t) \leq \exp\left(- \frac{t^2}{2\CH_{\mathit{VC}}}\right)
\]
\end{theorem}
\begin{proof}
We show that the random variable $Z = \log_2 S_\CF(X_1, \ldots, X_N)$ satisfies the assumptions of  \cref{th:blm}, with
\[
Z_k = \log_2 S_\CF(X_1, \ldots,X_{k-1}, X_{k+1}, \ldots,  X_N).
\]
Clearly, $0\leq Z$, $0 \leq Z - Z_k \leq 1$, because one can do no more than double $S_{\CF}$ by adding one point. We need to show that
\begin{equation}
\label{eq:blm.z}
\sum_{k=1}^N (Z-Z_k) \leq Z. 
\end{equation}
Note that $Z$ is the base-two entropy of the uniform distribution, $\pi$, on the set 
\[
\CF(X_1, \ldots, X_N)\sub\{-1, 1\}^N.
\]

We will use the following lemma.
\begin{lemma}
Let $A$ be  a finite set and $\psi$ a probability distribution on $A^N$. Let $\psi_k$ be its marginal when the $k$th variable is removed. Then: 
\label{lem:han}
\begin{equation}
\label{eq:han}
\sum_{k=1}^N  \CH_2(\psi_k) - (N-1) \CH_2(\psi) \geq 0.
\end{equation}
\end{lemma}
This lemma is a special case of a collection of results on non-negative entropy measures developed in \citet{han1978nonnegative}, and we provide a direct proof below for completeness. 

 Given the lemma,  let $\pi_k$ denote the marginal distribution of $\pi$ when the $k$th variable is removed, i.e., 
\begin{multline*}
\pi_k(\ep_1, \ldots  \ep_{k-1}, \ep_{k+1}, \ldots, \ep_N)\\
 =  \pi(\ep_1, \ldots  \ep_{k-1}, -1, \ep_{k+1}, \ldots, \ep_N)  + \pi(\ep_1, \ldots  \ep_{k-1}, 1, \ep_{k+1}, \ldots, \ep_N).
\end{multline*}
We have:
\[
\sum_{k=1}^N (\CH_2(\pi)- \CH_2(\pi_k)) \leq H(\pi) 
\]
from which \cref{eq:blm.z} derives since $Z = \CH_2(\pi)$ and $Z_k \geq \CH_2(\pi_k)$. The result then follows from   \cref{th:blm}.

We now prove  \cref{lem:han} by induction (this proof requires some basic notions of information theory). For convenience, introduce random variables $(\xi_1, \ldots, \xi_N)$ such that $\xi_k \in A$, with joint probability distribution given by $\psi$. Let $Y = (\xi_1, \ldots, \xi_N)$, $Y^{(k)}$ the $(N-1)$-tuple formed from $Y$ by removing $\xi_k$, $Y^{(k,l)}$ the $(N-2)$-tuple obtained by removing $\xi_k$ and $\xi_l$, etc. 
Inequality \eqref{eq:han} can then be rewritten
\[
\sum_{k=1}^N  \CH_2(Y^{(k)}) - (N-1) \CH_2(Y) \geq 0.
\]
This inequality is obviously true for $N=1$, and it is true also for $N=2$ since it gives in this case the well-known inequality $\CH_2(Y_1, Y_2) \leq \CH_2(Y_1) + \CH_2(Y_2)$. 
Fix $M>2$ and assume that the lemma is  true for any $N<M$. To prove the statement for $N = M$, we will use the following inequality, which holds for any three random variables $U_1, U_2, U_3$:
\[
\CH_2(U_1, U_3) + \CH_2(U_2, U_3) \geq \CH_2(U_1, U_2, U_3) + \CH_2(U_3)\,.
\]
This inequality is equivalent to the statement on conditional entropies that $\CH_2(U_1, U_2\mid U_3) \leq \CH_2(U_1\mid U_3) + \CH_2(U_2\mid U_3)$. We apply it, for given $k\neq l$, to $U_1 = Y_l$, $U_2 = Y_k$, $U_3 = Y^{(k,l)}$, yielding 
\[
\CH_2(Y^{(k)}) + \CH_2(Y^{(l)}) \geq \CH_2(Y) + \CH_2(Y^{(k,l)}).
\]
We now sum over all pairs $k\neq l$, yielding
\[
2(N-1) \sum_{k=1}^N \CH_2(Y^{(k)}) \geq N(N-1)\CH_2(Y) + \sum_{k\neq l} \CH_2(Y^{(k,l)}).
\]
 We finally use the induction hypothesis to write that, for all $k$
 \[
 \sum_{l\neq k} \CH_2(Y^{(k,l)}) \geq (N-2) \CH_2(Y^{(k)})
 \]
  and obtain
\[
2(N-1) \sum_{k=1}^N \CH_2(Y^{(k)}) \geq N(N-1)\CH_2(Y) + (N-2) \sum_{k=1}^N \CH_2(Y^{(k)}), 
\]
which provides the desired result after rearranging the terms.
\end{proof}

Note that  \cref{th:vc} involves $S_\CF(2N)$, with:
\[
\log_2(S_\CF(2N)) = \log_2 \myE(S_\CF(X_1, \ldots, X_{2N})) \geq \CH_{\mathit{VC}}(2N, \CF)
\]
from Jensen's inequality. This implies that the high-probability upper bound on $\CH_{\mathit{VC}}(2N, \CF)$ that results from the previous theorem is not necessarily an upper bound on $\log(S_\CF(2N))$. It is however proved in \citet{boucheron2000sharp} that 
\[
\log_2 \myE(S_\CF(X_1, \ldots, X_{2N})) \leq \frac{1}{\log 2} \CH_{\mathit{VC}}(2N, \CF)
\]
also holds (as a consequence of  \cref{eq:blm}). A little more work (see \citet{boucheron2000sharp}) combining  \cref{th:vc} and \cref{th:vc.conc} implies the following bound, which holds with probability $1-\de$ at least:
\[
\forall f \in \CF: R(f) \leq \CE(f) + \sqrt{\frac{6\log S_\CF(X_1, \ldots, X_N)}N} + 4 \sqrt{\frac{\log(2/\de)}N}.
\]

\section{Covering numbers and chaining}
The upper bounds using the VC dimension relied on the number of different values taken by a set of functions when evaluated on a finite set, this number being used to apply a union bound. A different point of view may be applied when one relies on some notion of continuity of the family of functions on which a uniform concentration bound is needed, with respect to a given metric. This viewpoint is furthermore applicable when the sets $\CF(X_1, \ldots, X_N)$ are infinite. To develop these tools, we will need some new concepts measuring the size of sets in a metric space.

\subsection{Covering, packing and entropy numbers}
\label{sec:covering}

\begin{definition}
\label{def:covering.number}
Let $(\CG, \rho)$ be a metric space and let $\ep>0$. The $\ep$-covering number of $(\CG, \rho)$. denoted $\CN(\CG, \rho, \ep)$, is the smallest integer $n$ such that there exists a subset $G\sub\CG$ such that $|G|=n$ and $\max_{g\in \CG} \rho(g, G) \leq \ep$. 

Let $\gamma>0$. The $\gamma$-packing number $\CM(\CG, \rho, \ga)$, is the largest number $n$ such that there exists a subset $A\sub\CG$ with cardinality $n$ such that any two distinct elements of $A$ are at distance strictly larger than $\ga$ (such sets are called $\ga$-nets). 

When $\CG$ and $\rho$ are well understood from the context, we will write simply $\CN(\epsilon)$ and $\CM(\gamma)$.
\end{definition}

\begin{proposition}
\label{prop:pack.cov}
One has, for any $\gamma>0$:
\[
\CM(\CG, \rho, 2\ga)\leq \CN(\CG, \rho, \ga)\leq \CM(\CG, \rho, \ga).
\]
\end{proposition}
\begin{proof}
Let $A$ be a maximal $\ga$-net. Then, for all $x\in \CG$, there exists $y\in A$ such that $\rho(x,y) \leq \ga$: otherwise $A\cup \{x\}$ would also be a $\gamma-net$. This shows that $\max(\rho(x, A), x\in\CG) \leq \ga$ and $\CN(\CG, \rho, \ga) \leq |A|$. 

Conversely, let $A$ be a $2\ga$-net. Let $G$ be an optimal $\ga$-covering. Associate to each $y\in A$ a point $x\in G$ at distance less than $\ga$: at least one exists because $G$ is a covering. This defines a function $f: A\to G$, which is necessarily one-to-one, because if two points in $A$ map to the same point in $G$, the distance between these two points would be less than or equal to $2\ga$. This shows that $\CM(\CG, \rho, 2\ga)\leq \CN(\CG, \rho, \ga)$.
\end{proof}

The entropy numbers of $(\CG, \rho)$, denoted, for an integer $N$, $e(\CG, \rho, N)$ (or just $e(N)$) represent the best accuracy that can be achieved by subsets of $\CG$ of size $N$, namely
\begin{equation}
\label{eq:cov.num}
e(\CG, \rho, N) = \min_{G\sub\CG, |G| = N} \max\{\rho(g, G): g\in\CG\}.
\end{equation}

We have:
\begin{subequations}
\begin{equation}
\label{eq:e.N}
e(\CG, \rho, N) = \inf\{\ep: N(\CG, \rho, \ep) \leq N\}
\end{equation}
and
\begin{equation}
\label{eq:N.e}
N(\CG, \rho, \ep) = \min\{N: e(\CG, \rho, N) \leq \ep\}.
\end{equation}
\end{subequations}

\subsection{A first union bound}
Let $Z$ be a random variable $Z:\Om\to \CZ$. We will consider a space $\CG$ of functions $g: \CZ \to \mR$, such that (to simplify the discussion) $\myE(g(Z)) = 0$ for all $g\in \CG$. In this section, we assume that functions in $\CG$ are bounded and let 
\[
\rho_\infty(g,g') = \sup_{z\in\CZ} |g(z) - g'(z)|\,.
\]

Assume that $\CN(\CG, \rho_\infty, \ep)<\infty$, for all $\ep>0$ (which requires the set $\CG$ to be precompact for the $\rho_\infty$ metric). Take $t>0$, $0<\ep<t$  and choose a set  $G\sub \CG$ such that $|G| = \CN(\CG, \rho_\infty, \ep)$. Then, using a union bound,
\begin{align}
\label{eq:chaining.intro.0}
\myP(\sup_{g\in \CG} g(Z) \geq t) &\leq \myP(\sup_{g\in G} g(Z) \geq t-\ep)\\
\nonumber
& \leq \CN(\CG, \rho_\infty, \ep)\, \sup_{g\in \CG} \myP(g(Z) \geq t-\ep).
\end{align}
Now, if each function in $\CG$ satisfies a concentration inequality, say,
\[
\myP(g(Z) \geq u) \leq e^{-\frac{u^2}{2\mu(g)}}
\]
for some $\mu(g) >0$, then, assuming that  $\mu(\CG) \defeq \max_{g\in \CG} \mu(g)$ is finite, we find that, for $0<\ep<t$,
\[
\myP(\sup_{g\in \CG} g(Z) \geq t) \leq \CN(\CG, \rho_\infty, \ep)\,
e^{-\frac{(t-\ep)^2}{2\mu(\CG)}}\,.
\]

We now apply this inequality to  the case of  binary classification, where a binary variable $Y$ is predicted by an input variable $X$, with a model class of classifiers $\CF$ and the 0--1 loss function. 
If $A$ is a finite family of elements of $\CR$, we define, for $f, f'\in \CF$ 
\[
\rho_A(f, f') = \frac1{|A|} \sum_{x\in A} \bfone_{f(x)\neq f'(x)}.
\]
Let
\[
\bar \CN(\CF, \ep, N) = E\left(\CN(\CF, \rho_{\{X_1, \ldots, X_N\}}, \ep)\right)
\]
where $X_1, \ldots, X_N$ is an i.i.d. sample of $X$. We then have the following proposition.
\begin{proposition}
\label{prop:entropy.first}
For all $\ep>0$, one has
\begin{equation}
\label{eq:chaining.intro}
\myP\Big(\sup_{f\in\CF} (R(f) - \CE_\mT(f)) \geq t\Big) \leq 2 \bar \CN(\CF, \ep/2, N) e^{-\frac{N(t/2-\ep)^2}{4}}\,.
\end{equation}
\end{proposition}
\begin{proof}
A key step in the proof of  \cref{th:vc}, was to show that 
\begin{equation}
\label{eq:chaining.intro.00}
\myP\Big(\sup_{f\in\CF} (R(f) - \CE_\mT(f)) \geq t\Big) \leq 2 \myP\Big(\sup_{f\in\CF}  \sum_{k=1}^N \bfxi_k (r(Y'_k, f(X'_k)) - r(Y_k, f(X_k))) \geq Nt/2\Big).
\end{equation}
where $\bfxi_1, \ldots, \bfxi_N$ are Rademacher random variables and $\mathit{\mT}, \mT'$ are two independent training sets of size $N$. We start from this inequality and  bound the conditional expectation 
\begin{equation}
\label{eq:chaining.intro:2}
\myP\Big(\sup_{f\in\CF}  \sum_{k=1}^N \bfxi_k (r(Y'_k, f(X'_k)) - r(Y_k, f(X_k))) \geq Nt/2\ \Big|\  \mT, \mT'\Big)
\end{equation}
and therefore consider $r(Y'_k, f(X'_k)) - r(Y_k, f(X_k))$ as constants that we will denote $c_k(f)$. Since we are using a 0--1 loss, we have $c_k(f)\in \{-1, 0,1\}$ and, for $f,f'\in\CF$, 
\begin{equation}
\label{eq:cck}
|c_k(f) - c_k(f')| \leq \bfone_{f(X_k) \neq f'(X_k)} + \bfone_{f(X'_k) \neq f'(X'_k)}\,.
\end{equation}
Consider the random variable $Z = (\bfxi_1, \ldots, \bfxi_N)$, and let 
\[
\CG = \defset{g_f, f\in \CF}
\]
with
\[
g_f(\xi_1, \ldots, \xi_N) = \frac1N \sum_{k=1}^N c_k(f) \xi_k\,. 
\]
We have
\[
\rho_\infty(g_f, g_{f'}) = \frac1N \sum_{k=1}^N |c_k(f) - c_k(f')| \,.
\]
Applying Hoeffding's inequality, we have, for $u>0$ and using the fact that $c_k\in [-1,1]$
\[
\myP(g_{f}(Z) > u \mid \mT, \mT') \leq e^{-\frac{2Nu^2}{4}} = e^{-\frac{Nu^2}{2}}
\]
and the  discussion preceding the theorem yields the fact that, for any $\ep>0$:
\begin{equation}
\label{eq:chaining.intro.1}
\myP(\sup_{f\in\CF} g_f(Z) > t/2\mid  \mT, \mT') \leq \CN(\CG, \ep, \rho_\infty) e^{-\frac{N(t/2-\ep)^2}{2}}\,.
\end{equation}
Let $A = (X_1, \ldots, X_N, X'_1, \ldots, X'_N)$ so that 
\[
\rho_{A}(f,f') = \frac1{2N} \sum_{k=1}^{N} \left(\bfone_{f(X_k) \neq f'(X_k)} + \bfone_{f(X'_k) \neq f'(X'_k)}\right).
\] 
Using \cref{eq:cck}, we have $\rho_\infty(g_f, g_{f'}) \leq 2\rho_{A}(f,f')$, which implies
\[
\CN(\CG, \ep, \rho_\infty) \leq \CN(\CF, \ep /2, \rho_{A})\,.
\]
Using this in \cref{eq:chaining.intro.1} and taking the expectation in \cref{eq:chaining.intro:2}, we get
\begin{equation}
\myP\Big(\sup_{f\in\CF} (R(f) - \CE_\mT(f)) \geq t\Big) \leq 2 \bar \CN(\CF, \ep/2, N) e^{-\frac{N(t/2-\ep)^2}{2}}
\end{equation}
which is valid for all $\ep>0$. 
\end{proof}
One can retrieve the bound obtained in  \cref{th:vc} using the obvious fact that 
\[
\CN(\CF, \ep, \rho_{A}) \leq  |\CF(A)|,
\]
for any $A\subset \CR$, so that 
\[
\myP\Big(\sup_{f\in\CF} (R(f) - \CE_\mT(f)) \geq t\Big) \leq 2  \CS(\CF, 2N) e^{-\frac{N(t/2-\ep)^2}{2}}
\]
for any $\ep>0$, and letting $\ep$ go to zero, 
\[
\myP\Big(\sup_{f\in\CF} (R(f) - \CE_\mT(f)) \geq t\Big) \leq 2  \CS(\CF, 2N) e^{-\frac{Nt^2}{8}}\,.
\]
So \eqref{eq:chaining.intro} provides a family of equations that depend on a parameter $\ep$ which, in the limit $\ep\to 0$, includes  \cref{th:vc} as a particular case. For a given $N$, optimizing \cref{eq:chaining.intro} over $\ep$ may give a better upper bound, provided one has a good way to estimate $\bar \CN(\CF, \ep/2, N)$ (which is, of course,  far from obvious).

\subsection{Evaluating covering numbers}
Covering numbers can be evaluated in some simple situations. The following proposition provides an example in finite dimensions. 
\begin{proposition}
\label{prop:ent.numb.param}
Assume that  $\CG$ is a parametric family of functions, so that $\CG = \defset{g_\th, \th \in \Th}$ where $\Th \subset \mR^m$. Assume also that, for some constant $C$, $\rho_\infty(g_\th, g_{\th'}) \leq C |\th - \th'|$ for all $\th, \th'\in \Th$. 
Let $\pe{\CG}M = \defset{g_\th: \th \in \Th, |\th| \leq M}$. Then
\[
\CN(\CG, \rho_\infty, \ep) \leq \left(1 + \frac{2CM}{\ep}\right)^m
\]
\end{proposition}
\begin{proof}
Letting $\rho$ denote the Euclidean distance in $\mR^m$, our hypotheses imply that $\CN(\pe{\CG} M, \rho_\infty, \ep)$ is bounded by $\CN(B_M, \rho, \ep/C)$ where $B_M$ is the ball with radius $M$ in $\mR^m$. Now, if $\th_1, \dots, \th_n$ is an $\al$-covering of $B_M$, then $\th_1/M, \dots, \th_n/M$ is an $(\al/M)$-covering of $B_1$, which shows (together with a symmetric argument) that $\CN(B_M, \rho, \al) = \CN(B_1, \rho, \al/M)$ and we get
\[
\CN(\pe {\CG}M, \rho_\infty, \ep) \leq \CN(B_1, \rho, \ep/MC)
\]
and we only need to evaluate $\CN(B_1, \rho, \al)$ for $\al > 0$. Using  \cref{prop:pack.cov}, one can instead evaluate $\CM(B_1, \rho, \al)$. So let $A$ be an $\al$-net in $B_1$. Then
\[
\bigcup_{x\in A} B_\rho(x, \al/2) \subset B_\rho(0, 1+\al/2)
\]
and, since the sets in the union are disjoint, 
\[
\sum_{x\in A} \mathrm{volume}(B_\rho(x, \al/2)) = |A| \mathrm{volume}(B_\rho(0, \al/2)) \leq \mathrm{volume}(B_\rho(0, 1+\al/2))\,.
\]
Letting $C_m$ denote the volume of the unit ball in $\mR^m$, this shows
\[
|A| C_m \left(\frac\al 2\right)^m \leq C_m \left(1 + \frac\al 2\right)^m
\]
and 
\[
|A| \leq \left(1 + \frac2 \al\right)^m\,,
\]
which concludes the proof.
\end{proof}

One can also obtain entropy number estimates in infinite dimensions. Here, we quote a result applicable to spaces of smooth functions, referring to \citet{vander1996Weak} for a proof.
\begin{theorem}
\label{th:entropy.continuous}
Let $\CZ$ be a bounded convex subset of $\mR^d$ with non-empty interior. For $p\geq 1$ and $f\in C^p(\CZ)$, let 
\[
\|f\|_{p, \infty} = \max\defset{ |D^k(f(x)|: k=0, \ldots, p, x\in \CZ}.
\]
Let $\CG$ be the unit ball for this norm, 
\[
\CG = \defset{f\in C^p(\CZ): \|f\|_{p, \infty} \leq 1}.
\]
Let $\CZ^{(1)}$ be the set of all $x\in \mR^d$ at distance less than $1$ from $\CR$.

Then there exists a constant $K$ depending only on $p$ and $d$ such that
\[
\log \CN(\ep, \CG, \rho_\infty) \leq K \mathrm{volume}(\CZ^{(1)})\left(\frac1\ep\right)^{d/p}
\]
\end{theorem}

\subsection{Chaining}
The distance $\rho_\infty$ may not always be the best one to analyze the set of functions, $\CG$. For example, if $\CG$ is a class of functions with values in $\{-1,1\}$, then $\rho_{\infty}(g,g') = 2$ unless $g=g'$.  In such contexts, it is often preferable to use distances that compute average discrepancies, such as
\begin{equation}
\label{eq:rho.p}
\rho_p(g,g') = E(|g(Z) - g'(Z)|^p)^{1/p}\,,
\end{equation}
for some random variable $Z$.
Such distances, by definition, do not provide uniform bounds on differences between functions (that we used to write \cref{eq:chaining.intro.0}), but can rather be used in upper-bounds on the probabilities of deviations from zero, which have to be handled somewhat differently. We here summarize a general approach called ``chaining,''  following for this purpose the presentation made in \citet{talagrand2014upper} (see also \citet{audibert2007combining}). From now on,  we assume that
 $(\CG, \rho)$ is a (pseudo-)metric space of functions $g: \CZ \to \CR$ and $Z$ a random variable taking values in $\CZ$. 
We will make the basic assumption that, for all $g,g'\in\CG$ and $t>0$,
\[
\myP(|g(Z) - g'(Z)| > t)\leq 2 e^{-\frac{t^2}{2\rho(g,g')^2}}.
\]
Note that this assumption includes cases in which
\[
\myP(|g(Z) - g'(Z)| > t)\leq 2 e^{-\frac{t^2}{2\rho(g,g')^\al}}.
\]
for some $\al\in (0, 2]$, because, if $\rho$ is a distance, then so is $\rho^{\al/2}$ if $\al\leq 2$.
We will also assume that $\myE(g(Z)) = 0$ in order to avoid centering the variables at every step. 

We are interested in upper bounds for $\myP(\sup_{g\in\CG} g(Z) > t)$. To build a chaining argument, consider a family $(G_0, G_1, \ldots)$ of subsets  of $\CG$. Assume that  $|G_k| \leq N_k$ with  $N_k$ chosen, for future simplicity, so that $N_{k-1}N_k \leq N_{k+1}$. For $g\in \CG$, let $\pi_k(g)$ denote a closest point to $g$ in $G_k$. Also assume that $G_0 = \{g_0\}$ is a singleton, so that $\pi_0(g)= g_0$ for all $g\in\CG$. (One can generally assume without harm that $0\in\CG$, in which case one should choose $g_0=0$ in the following discussion.)
For $g\in G_n$, we therefore have
\[
g - g_0 = \sum_{k=1}^n (\pi_k(g) - \pi_{k-1}(g))\,.
\]
Let $(t_1, t_2, \ldots)$ be a sequence of numbers that will be determined later. Let
\begin{equation}
\label{eq:chaining.sn}
S_n = \max_{g\in G_n} \sum_{k=1}^n t_k \rho(\pi_k(g), \pi_{k-1}(g)).
\end{equation}
Then, for any $t$, 
\begin{align*}
&\myP(\sup_{g\in G_n} g(Z) - g_0(Z) > tS_n) \\
& \leq \myP(\exists g \in G_n, \exists k \leq n:   \pi_k(g)(Z) - \pi_{k-1}(g)(Z) >  tt_k \rho(\pi_k(g), \pi_{k-1}(g)))\\
&\leq \myP(\exists k \leq n,  \exists g\in G_k, g'\in G_{k-1}:  g(Z) - g'(Z) >  tt_k \rho(g, g'))\\
&\leq \sum_{k=1}^n N_k N_{k-1} \sup_{g\in G_k, g'\in G_{k-1}} \myP(g(Z) - g'(Z) >  tt_k \rho(g, g'))\\
&\leq 2\sum_{k=1}^n N_{k+1} e^{-\frac{t^2t_k^2}2}
\end{align*}
If one takes $N_k = 2^{2^k}$, which satisfies $N_kN_{k-1} = 2^{2^k + 2^{k-1}} \leq N_{k+1}$, and $t_k = 2^{k/2}$, one finds that
\[
\myP(\sup_{g\in G_n} g(Z) - g_0(Z) > tS_n) \leq 2 \sum_{k=1}^n 2^{2^{k+1}} e^{-{2^{k-1} t^2}}.
\]
The upper bound converges (as a function of $n$) as soon as $t > 2 \sqrt{\log 2}$. Moreover, one has
\[
2 \sum_{k=1}^n 2^{2^{k+1}} e^{-{2^{k-1} t^2}} = 2e^{-\frac{t^2}2} \sum_{k=1}^n e^{-2^{k-2} (t^2-8\log2)} \leq 2e^{-\frac{t^2}2}\sum_{k=1}^\infty e^{-2^{k-2}}
\]
when $t > \sqrt{1 + 8\log 2}$. This provides a concentration bound for $\myP(\sup_{g\in G_n} g(Z) - g_0(Z) > tS_n)$, that we may rewrite as
\begin{equation}
\label{eq:chaining.1}
\myP(\sup_{g\in G_n} g(Z) - g_0(Z) > t) \leq C e^{-\frac{t^2}{2S_n^2}}
\end{equation}
for $t> 2 S_n\sqrt{\log 2}$, $C = 2 \sum_{k=1}^\infty e^{-2^{k-2}}$ and $S_n$ given by \cref{eq:chaining.sn}, with $t_k = 2^{k/2}$. Moreover, we have 
\begin{align*}
S_n &= \max_{g\in G_n} \sum_{k=1}^n 2^{k/2} \rho(\pi_k(g), \pi_{k-1}(g))\\
&\leq  \max_{g\in G_n} \sum_{k=1}^n 2^{k/2} (\rho(g, G_k) + \rho(g, G_{k-1})) \\
&\leq  2 \max_{g\in G_n} \sum_{k=0}^n 2^{k/2} \rho(g, G_k)
\end{align*} 
and this simpler upper bound can be used in \cref{eq:chaining.1}. 

We haven't made many assumptions so far on the sequence $G_0, G_1, \ldots$, beyond bounding their cardinality, but it is natural to require that they are built in order to behave like a dense subset of $\CG$, so that 
\begin{equation}
\label{eq:chain.assump.1}
\lim_{n\to\infty} \max_{g\in \CG} \rho(x, G_n) =0.
\end{equation}
Note that this requires that the set $\CG$ is precompact for the distance $\rho$. We will also assume that 
\begin{equation}
\label{eq:chain.assump.2}
\lim_{n\to\infty} \sup_{g\in G_n} g(x) = \sup_{g\in \CG} g(x).
\end{equation}
Then, we have proved the following result \citep{talagrand2006generic}.
\begin{theorem}
\label{th:gen.chain}
Let $G_0, G_1, \ldots$ be a family of subsets of $\CG$ satisfying \cref{eq:chain.assump.1} and \cref{eq:chain.assump.2} and such that $G_0 = \{g_0\}$ and $|G_n| \leq 2^{2^n}$ for $n\geq 0$. Let 
\begin{equation}
\label{eq:gen.chain.S}
S = 2 \sup_{g\in \CG} \sum_{n=0}^\infty 2^{n/2} \rho(g, G_n)
\end{equation}
Then, for $t>  S \sqrt{1 + 8\log 2}$, 
\begin{equation}
\label{eq:gen.chain}
\myP(\sup_{g\in \CG} g(Z) - g_0(Z) > t) \leq C e^{-\frac{t^2}{2S^2}}
\end{equation}
with 
$C = 2 \sum_{k=1}^\infty e^{-2^{k-2}}$.
\end{theorem}

The exponential rate of convergence in the right-hand side of \cref{eq:gen.chain} is the quantity $S$, and the upper bound will be improved when building the sequence $(G_0, G_1, \ldots)$ so that $S$ is as small as possible. Such an optimization for a given family of functions is however a formidable problem. It is however interesting to see (still following \citep{talagrand2006generic}) that  \cref{th:gen.chain} implies a classical inequality in terms of what is called the metric entropy of the metric space $(\CG, \rho)$.

\subsection{Metric entropy}

If $S$ is given by \cref{eq:gen.chain.S}, we have 
\[
S = 2 \sup\Big(\sum_{n=0}^\infty 2^{n/2} \rho(g, G_n): g\in\CG\Big) \leq 2 \sum_{n=0}^\infty 2^{n/2} \sup\{ \rho(g, G_n) : g\in\CG\}
\]
Take $G_n$ achieving the minimum in the entropy number $e(\CG, \rho, 2^{2^n})$. Then, \cref{eq:gen.chain} holds with $S$ replaced by
\[
\hat S = 2 \sum_{n=0}^\infty 2^{n/2} e(\CG, \rho, 2^{2^n})\,.
\]

Consider the function
\begin{equation}
\label{eq:met.ent}
h(\CG, \rho) = \int_0^\infty \sqrt{\log \CN(\CG, \rho, \ep)} d\ep ,
\end{equation}
which is known as {\em Dudley's metric entropy} of the space $(\CG, \rho)$.
We have 
\[
h(\CG, \rho) = \int_0^{e(2)} \sqrt{\log \CN(\ep)} d\ep +  \sum_{n=1}^\infty \int_{e(2^{2^{n-1}})}^{e(2^{2^n})} \sqrt{\log \CN(\ep)} d\ep .
\]
If $\ep \in [e(2^{2^{n-1}}), e(2^{2^n}))$, we have $\CN(\ep) > 2^{2^n}$ so that 
\begin{align*}
h(\CG, \rho) &\geq e(2) \sqrt{\log 3} + \sum_{n=1}^\infty 2^{n/2} (e(2^{2^n}) - e(2^{2^{n-1}})) \\
&\geq \big(1 - \frac{\sqrt2}2\big) \sum_{n=1}^\infty 2^{n/2} e(2^{2^n}).
\end{align*}
Therefore,
\[
\hat S \leq \frac{4}{2-\sqrt2} h(\CG, \rho) \leq 7 h(\CG, \rho)
\]
and this upper bound can also be used to obtain a simpler (but weaker) form of  \cref{th:gen.chain}.

\begin{remark}
\label{rem:entropy.vc}
The covering numbers of a class $\CG$ of binary functions $g$ with values in $\{-1, 1\}$ can be controlled by the VC dimension of the class. Here, we consider $\rho(g,g') = \myP(g\neq g') = \rho_1(g,g')/2$. Then, the following theorem holds.
\begin{theorem}
\label{th:vc.entropy}
Let $\CG$ be a class of binary functions such that $D = \VC(\CG)< \infty$. Then, there is a universal constant $K$ such that, for any $\ep \in (0,1)$,
\[
\CN(\CG, \rho, \ep) \leq KD(4e)^D \left(\frac1\ep\right)^{D-1}
\]
with $\rho(g,g') = \myP(g\neq g')$.
\end{theorem}
We refer to \citet{vander1996Weak}, Theorem 2.6.4 for a proof, which is rather long and technical. 
\end{remark}

\subsection{Application}
We quickly show how this discussion can be turned into results applicable to the classification problem.
If $\CF$ is a function class of binary classifiers and $r$ is the risk function, one can consider the class
\[
\CG = \{(x,y) \mapsto r(y, f(x)): f\in\CF\}\,.
\]
If $r$ is the 0--1 loss, we have $\VC(\CG) \leq \VC(\CF)$. Indeed, if one considers $N$ points in $\CR\times \{-1,1\}$, say $(x_1, y_1, \ldots, x_N, y_N)$, then
\begin{multline*}
\CG(x_1, y_1, \ldots, x_N, y_N) \\
= \{r(1, f(x_k)): k=1, \ldots, N, y_k=1\} \cup \{r(-1, f(x_k)): k=1, \ldots, N, y_k=-1\}.
\end{multline*}
If the two sets in the right-hand side are not empty, i.e., the numbers $N_{(1)}$ and $N_{(-1)}$ of $k$'s such that $y_k=1$ or $y_k=-1$ are not zero, then
\[
|\CG(x_1, y_1, \ldots, x_N, y_N)| \leq 2^{N_{(1)}} + 2^{N_{(-1)}},
\]
which is less that $2^N$ as soon as $N>2$. So, taking $N>2$, for $(x_1, y_1, \ldots, x_N, y_N)$ to be shattered by $\CG$, we need $N_{(1)}=N$ or $N_{(-1)} = N$ and in this case, the inequality:
\[
|\CG(x_1, y_1, \ldots, x_N, y_N)| \leq |\CF(x_1, \ldots, x_N)|
\]
is obvious.
The same inequality will be true for some $x_1, \ldots, x_N$ with $N=2$, except in the uninteresting case where $f(x) = 1$ (or $-1$) for every $x\in \CR$. 

A similar inequality holds for entropy numbers with the $\rho_1$ distance (cf. \cref{eq:rho.p}) because 
\[
\myE(|r(Y,f(X)) - r(Y,f'(X))|) \leq \myP(f(X) \neq f'(X))
\]
whenever $r$ takes values in $[0,1]$, which implies that
\[
\CN(\CG, \rho_1, \ep) \leq \CN(\CF, \rho_1, \ep)
\]
for all $\ep>0$. Note however that evaluating this upper bound may still be challenging and would rely on strong assumptions on the distribution of $X$ allowing to control $\myP(f(X) \neq f'(X))$.

We now assume  that functions in $\CF$ define ``posterior probabilities'' on $\CG$.
 More precisely, given $\la \in \mR$ we can define the probability $\pi_\la$ on $\{-1,1\}$ by
\[
\pi_\la(y) = \frac{e^{\la y}}{e^{-\la} + e^{\la}}.
\]
Now, if $\CF$ is a class of {\em real-valued} functions, we can define the risk function
\[
r(y, f(x)) = \log \frac1{\pi_{f(x)}(y)}\,.
\]

Since $|\prt_\la \log \pi_\la(y)| = |y - \tanh \la| \leq 2$ for $y\in \{-1, 1\}$, we have
\[
|r(y, f(x)) - r(y, f'(x))| \leq 2 |f(x) - f'(x)|
\]
so that entropy numbers in $\CG$ can be estimated from entropy numbers in $\CF$. As an example, let $\CF$ be a space of affine functions $x \mapsto a_0 + b^T x$, $x\in \mR^d$. Assume that the random variable $X$ is bounded, so that one can take $\CR$ to be an open ball centered at 0 with radius, say, $U$. For $M>0$, let
\[
\CF_M = \{f:x \mapsto a_0 + b^T x: |b| \leq M, |a_0| \leq UM\}\,.
\]
The restriction $|b|\leq M$ is equivalent to using a penalty method, such as, for example, ridge logistic regression. Moreover, if $|b| \leq M$, it is natural to assume that $|a_0| \leq UM$ because otherwise $f$ would have  a constant sign on $\CR$. In this case, we get
\[
\rho_\infty(r(y, f(x)), r(y, f'(x))) \leq |a_0 - a'_0| + U |b-b'|
\]
and a small modification of the proof of  \cref{prop:ent.numb.param} shows that 
\[
\CN(\CF, \rho_\infty, \ep) \leq \left(1 + \frac{4CU}{\ep}\right)^{d+1}
\]

\section{Other complexity measures}
\subsection{Fat-shattering and margins}
VC-dimension and metric entropy are measures that control the complexity of a model class, and can therefore be evaluated a priori without observing any data. These bounds can be improved, in general, by using information derived from the training set, and, particular the classification margin that has been obtained \citep{bartlett1999generalization}. 
%

For this discussion, we need to  return to the definition of covering numbers. If $\CF$ is a function class, $\rho_\infty$ the supremum metric on $\CF$, $\ep>0$ and $N$ is an integer, we let 
\[
\CN(\CF, \rho_\infty, \ep, N) = \max\{\CN(\CF(A), \rho_\infty, \ep): A\subset \CR, |A| = N\}
\]
that we will abbreviate in $\CN_\infty(\ep, N)$ when $\CF$ is known from the context. We will assume that functions in $\CF$ take values values in $[-1, 1]$, and we define for $\ga \geq 0$, $y\in\{0,1\}$, $u\in \mR$:
\[
r_\ga(y, u) = \left\{
\begin{aligned}
0 & \text{ if } u< - \ga \text{ and } y= 0  \\
0 & \text{ if } u> \ga \text{ and } y= 1\\
1 & \text{ otherwise}
\end{aligned}
\right.
\]  
So, $r_\gamma(y, f(x))$ is equal to 0 if $f(x)$ correctly predicts $y$ with margin $\gamma$ and to 1 otherwise. 
We then define the classification error with margin $\ga$ as
\[
R_\ga(f) = E(r_\ga(Y, f(X)))
\]
and, given a training set $T$ of size $N$
\[
\CE_{\ga, T} = \frac1N \sum_{k=1}^N r_\ga(y_k, f(x_k)).
\]
We then have the following theorem \citep{anthony2009neural}.
\begin{theorem}
\label{th:margin}
If $t \geq \sqrt{2/N}$
\begin{equation}
\label{eq:margin}
\myP(\sup_{f\in\CF} (R_0(f) -  \CE_{\ga, \mT}(f)) > t) \leq 2 \CN_\infty({\ga}/2, 2N) e^{-{Nt^2}/8},
\end{equation}
or, equivalently, with probability larger than $1-\de$, one has, for all $f\in \CF$,
\begin{equation}
\label{eq:margin.2}
R_0(f) -  \CE_{\ga, \mT}(f)) \leq \sqrt{\frac8N \left(\log \CN_\infty({\ga}/2, 2N) + \log\frac{2}{\de}\right)}\,.
\end{equation}

\end{theorem}
\begin{proof}
We first note that, for $Nt^2 > 2$, 
\[
\myP\Big(\sup_{f\in\CF} (R_0(f) -  \CE_{\ga, \mT}(f)) > t\Big) \leq 2 \myP\Big(\sup_{f\in\CF} (\CE_{\mT'}(f) - \CE_{\ga, \mT}(f)) > \frac\ep 2\Big),
\]
which is proved exactly the same way as \cref{eq:vc.symm} in  \cref{th:vc}, and we skip the argument. 

We have  
\[
\CE_{\mT'}(f) - \CE_{\ga, \mT}(f) = \frac1N \sum_{k=1}^N (r_0(Y'_k, f(X'_k)) - r_\ga(Y_k, f(X_k)))
\]
and because $(X_k, Y_k)$ and $(X'_k, Y'_k)$ have the same distribution, $\sup_{f\in\CF} (\CE_{\mT'}(f) - \CE_{\ga, \mT}(f))$ has the same distribution as
\begin{multline*}
\Delta_{\mT, \mT'}(\xi_1, \ldots, \xi_N) = \\
\sup_{f\in\CF} \frac1N \sum_{k=1}^N \big((r_0(Y'_k, f(X'_k)) - r_\ga(Y_k, f(X_k)))\bfxi_k + (r_0(Y_k, f(X_k)) - r_\ga(Y'_k, f(X'_k)))(1-\bfxi_k)\big)
\end{multline*}
where $\bfxi_1, \ldots, \bfxi_N$ is a sequence of Bernoulli random variables with parameter $1/2$. 

We now estimate $\myP(\Delta_{\mT, \mT'}(\bfxi_1, \ldots, \bfxi_N) > t/2\mid \mT, \mT')$ and we therefore consider $\mT$ and $\mT'$ as fixed.  Let $F$ be a subset of $\CF$, with cardinality $\CN_\infty(\ga/2, 2N)$, such that for all $f\in\CF$ there exists an $f'\in F$ such that $|f(x) - f'(x)| \leq \ga/2$ for all $x\in \{X_1, \ldots, X_N, X'_1, \ldots, X'_N\}$. Then  we claim that 
\[
\Delta_{\mT, \mT'}(\xi_1, \ldots, \xi_N) \leq \Delta'_{\mT, \mT'}(\xi_1, \ldots, \xi_N)
\]
where 
\[
\Delta'_{\mT, \mT'}(\xi_1, \ldots, \xi_N) = 
\max_{f\in F} \frac1N \sum_{k=1}^N (2\xi_k-1) \big(r_{\frac{\ga}2}(Y'_k, f(X'_k)) - r_{\frac{\ga}2}(Y_k, f(X_k))\big)\,.
\]
This is because, for any $(x,y)\in \CR\times\{0,1\}$, and $f, f'$ such that $|f(x) -  f'(x)| < \ga/2$, we have
$r_0(y, f(x)) \leq r_{\ga/2}(y, f'(x))$ and $r_{\ga/2}(y, f'(x)) \leq r_\ga(y, f(x))$: if an example is misclassified by $f$ (resp. $f'$) at a given margin, it must be misclassified by $f'$ (resp. $f$) at this margin plus $\ga/2$.    

Now, 
\begin{multline*}
\myP(\Delta'_{\mT, \mT'}(\bfxi_1, \ldots, \bfxi_N)>\frac t2) \\
\leq |F| \max_{f\in F} \myP\Big(\frac1N \sum_{k=1}^N (2\bfxi_k-1) (r_{\frac{\ga}2}(Y'_k, f(X'_k)) - r_{\frac{\ga}2}(Y_k, f(X_k))) >\frac t 2\Big)
\end{multline*}
to which we can apply Hoeffding's inequality, yielding
\[
\myP\Big(\Delta'_{\mT, \mT'}(\bfxi_1, \ldots, \bfxi_N)>\frac t2\Big) \leq |F| e^{-{Nt^2}/8},
\]
which concludes the proof, since, by \cref{prop:pack.cov}, $|F|\leq \CN_\infty({\ga}/2, 2N)$.

\end{proof}

In order to evaluate the covering numbers $\CN_\infty(\ep, N)$ using quantities similar to  VC-dimensions, a different type of set decomposition and shattering has been proposed. Following \citet{alon1997scale}, we introduce the following notions. Recall that a family of functions $\CF: \CR \to \{0,1\}$ shatters a finite set $A\sub\CR$ if and only if $|\CF(A)| = 2^{|A|}$. The following definitions are adapted to functions taking values in a continuous set.  
\begin{definition}
\label{def:all.shatter}
Let $\CF$ be a family of functions $f:\CR\to [-1,1]$ and $A$ a finite subset of $\CR$. 
\begin{enumerate}[label=(\roman*)]
\item One says that $\CF$ $P$-shatters $A$ if there exists a function $g_A:\CR\to \mR$ such that, for each $B\sub A$, there exists a  function $f\in \CF$ such that $f(x) \geq g_A(x)$ if $x\in B$ and $f(x) < g_A(x)$ if $x\in A\setminus B$.
\item Let $\ga$ be a positive number. One says that $\CF$ $P_\ga$-shatters $A$ if there exists a function $g_A:\CR\to \mR$ such that, for each $B\sub A$, there exists a  function $f\in \CF$ such that $f(x) \geq g_A(x)+\ga$ if $x\in B$ and $f(x) \leq g_A(x)-\ga$ if $x\in A\setminus B$.
\end{enumerate}
\end{definition}
Note that only the restriction of $g_A$ to $A$ matters in this definition.
This function acts as a threshold for binary classification. More precisely, given a function $g: A \to \mR$, one can associate to every $f\in \CF$ the binary function $f_g$ with $f_g(x)$ equal to 1 if $f(x) \geq g(x)$ and to 0 otherwise. Letting $\CF_g = \{f_g: f\in \CF\}$ we see that $\CF$ P-shatters $A$ if there exists a function $g_A$ such that $\CF_{g_A}$ shatters $A$. The definition of $P_\gamma$-shattering introduces a margin in the definition of $f_g$ (with $f_g(x)$ equal to 1 if $f(x) \geq g(x)+\gamma$, to 0 if $f(x) \leq g(x)-\gamma$ and is ambiguous otherwise), and $A$ is $P_\gamma$-shattered by $\CF$ if, for some $g_A$, the corresponding $\CF_{g_A}$ shatters $A$ without ambiguities.  

\begin{definition}
\label{def:P.dim}
One then defines the $P$-dimension of $\CF$ by
\[
\Pdim(\CF) = \max\{|A|: A\sub \CR, \CF\ P\text{-shatters } A\},
\]
and similarly the $P_\ga$-dimension of $\CF$ is
\[
\Pdim[\ga](\CF) = \max\{|A|: A\sub \CR, \CF\ P_\ga\text{-shatters } A\}.
\]
\end{definition}

The $P_\ga$-dimension of $\CF$ will replace the VC-dimension in order to control the covering numbers. More precisely, we have the following theorem \citep{alon1997scale}.
\begin{theorem}
\label{th:p.gamma}
Let $\ga>0$ and assume that $\CF$ has $P_{\ga/4}$-dimension $D<\infty$. Then,
\[
\CN_\infty(\ga, N) \leq 2 \left(\frac{16N}{\ga^2}\right)^{D\log(4eN/(D\gamma))}\,.
\] 
\end{theorem}
\begin{proof}
The proof is quite technical and relies on a combinatorial argument in which $\CF$ is first assumed to take integer values before addressing the continuous case.  

\noindent{\it Step 1}. We first assume that functions in $\CF$ take values in the finite set $\{1, \ldots, r\}$ where $r$ is an integer. For the time of this proof, we introduce yet another notion of shattering called S-shattering (for strong shattering) which is essentially the same as $P_1$-shattering, except that functions $g$ are restricted to take values in $\{1, \ldots, r\}$. Let $A$ be  a finite subset of $\CR$. 
Given a function $g: \CR \to \{1, \ldots, r\}$, we say that $(\CF, g)$ S-shatters $A$ if, for any $B\sub A$, there exist $f\in \CF$ satisfying $f(x) \geq g(x) + 1$ for $x\in B$ and $f(x) \leq g(x)-1$ if $x\in A\setminus B$. We say that $\CF$ S-shatters $A$ if $(\CF,g)$ S-shatters $A$ for some $g$.
The S-dimension of $\CF$ is the cardinality of the largest subset of $\CR$ that can be S-shattered and will be denoted $\Sdim(\CF)$. The first, and most difficult, part of the proof is to show that, if $\Sdim(\CF) = D$, then
\[
\CM(\CF(A), \rho_\infty, 2) \leq 2(|A|r^2)^{\lceil\log_2 y\rceil}
\]
with 
\[
y = \sum_{k=1}^D \binom{|A|}{k} r^k
\]
and  $\lceil i \rceil$ denotes the smallest integer larger than $u\in \mR$.
Here, $\CM$ is the packing number defined in \cref{sec:covering}.

To prove this, we can assume that $r\geq 3$, since, for $r\leq 2$, $\CM(\CF(A), \rho_\infty, 2)=1$ (the diameter of $\CF$ for the $\rho_\infty$ distance is 0 or 1). Let $\CG(A) =  \{1, \ldots, r\}^A$ be the set of all functions $f: A \to \{1, \ldots, r\}$ and let
\[
\CU_A = \defset{F\sub \CG(A): \forall f,f'\in F, \exists x\in A \text{ with } |f(x) - f'(x)| \geq 2}\,.
\]
For $F\in \CU_A$, let
\[
\CS_A(F) = \{(B,g): B\sub A, B\neq \emptyset, g:B\to \{1, \ldots, r\}, (F,g) \text{ S-shatters } B\}.
\]
Let $t_A(h) = \min\{|\CS_A(F)|: F\in\CU_A, |F| = h\}$ (where the minimum of the empty set is $+\infty$). Since we are considering in $\CU_A$ all possible functions from $A$ to $\{1, \ldots, r\}$, it is clear that $t_A(h)$ only depends on $|A|$, and we will also denote it by $t(h, |A|)$. 

Note that, by definition, if $(B,g)\in \CS_A(F)$, and $F\sub\CF$, then $|B| \leq D$. So, the number of elements in $\CS_A(F)$ for such an $F$ is less or equal than the number of possible such pairs $(B,g)$, which is strictly less than
$y = \sum_{k=1}^D \binom{|A|}{k} r^k$. So, if $t(h, |A|) \geq y$, then there cannot be any $F\sub\CF$ in the set $\CU_A$ and $\CM(\CF(A), \rho_\infty, 2) < h$. The rest of the proof consists in showing that  $t(h, |A|) \geq y$. 
\medskip

For any $n\geq 1$, we have $t(2,n) = 1$: fix $x\in A$, and $F = \{f_1, f_2\}\in \CG$ such that $f_1(x) = 1$, $f_2(x) = 3$ and $f_1(y) = f_2(y)$ if $y\neq x$. Then only $(\{x\}, g)$ is S-shattered by $F$, with $g$ such that $g(x) = 2$.

Now, assume that, for some integer $m$, $t(2mnr^2, n) <\infty$, so that there exists $F\in\CU_A$ such that $|F| = 2mnr^2$. Arrange the elements of $F$ into $mnr^2$ pairs $\{f_i, f'_i\}$. For each such pair, there exists $x_i\in A$ such that $|f_i(x_i) - f'_i(x_i)| > 1$. Since there are at most $n$ selected $x_i$, one of them must be appearing at least $mr^2$ times. Call it $x$ and keep (and reindex) the corresponding  $mr^2$ pairs, still denoted $\{f_i, f'_i\}$. Now, there are at most $r(r-1)/2$ possible distinct values for the unordered pairs $\{f_i(x), f'_i(x)\}$, so that one of them must be appearing at least $2mr^2/r(r-1) > 2m$ times. Select these functions, reindex them  and exchange the role of $f_i$ and $f'_i$ if needed to obtain $2m$ pairs $\{f_i, f'_i\}$ such that $f_i(x) = k$ and $f'_i(x) = l$ for all $i$ and fixed $k,l\in \{1, \ldots, r\}$ such that $k+1<l$. 
Let $F_1 = \{f_1, \ldots, f_{2m}\}$ and $F'_1 = \{f'_1, \ldots, f'_{2m}\}$. Let $A' = A\setminus\{x\}$. Then both $F_1$ and $F'_1$ belong to $\CU_{A'}$, which implies that both $\CS_{A'}(F_1)$ and $\CS_{A'}(F'_1)$ have cardinality at least $t(2m, n-1)$. Moreover, both sets are included in $\CS_A(F)$, and if $(B,g) \in \CS_{A'}(F_1)\cap\CS_{A'}(F'_1)$, then $(B\cup\{x\}, g')\in \CS_A(F)$, with $g'(y) = g(y)$ for $y\in B$ and $g'(x) = k+1$. This provides $2t(2m, n-1)$ elements in $\CS_A(F)$ and shows the key inequality (which is obviously true when the left-hand side is infinite)
\[
t(2mnr^2, n) \geq 2 t(2m, n-1)\,.
\]

This inequality can now be used to prove by induction that for all $0\leq k < n$, one has $t(2(nr^2)^k, n) \geq 2^k$, since
\[
t(2((n+1)r^2)^{k+1}, n+1) \geq 2 t(2((n+1)r^2)^{k}, n) \geq 2  t(2(nr^2)^k, n).
\]
For $k\geq n$, one has $2(nr^2)^k > r^n$, where $r^n$ is the number of functions in $\CG(A)$, so that $t(2(nr^2)^k, n) = +\infty$. So, $t(2(nr^2)^k, n) \geq 2^k$ is valid for all $k$ and  it suffices to take $k = \lceil \log_2y\rceil$ to obtain the desired result.

\noindent {\it Step 2.} The next step uses a discretization scheme to extend the previous result to functions with values in $[-1,1]$. More precisely, given $f: \CR\to [0,1]$, and $\eta>0$, let
\[
f^\eta(x) = \max\{k\in\mN: 2k\eta -1 \leq f(x)\}
\]
which takes values in $\{0, \ldots, r\}$ for $r = \lfloor \eta^{-1}\rfloor$. If $\CF$ is a class of functions with values in $[-1,1]$, define $\CF^\eta = \{f^\eta: f\in\CF\}$. With this notation, the following holds.
\begin{enumerate}[label= (\alph*)]
\item For all $\ga \leq \eta$: $\Sdim(\CF^\eta) \leq \Pdim[\ga](\CF)$
\item For all $\ep\geq 4\eta$ and $A\sub \CR$: $\CM(\CF(A), \rho_\infty, \ep) \leq \CM_\infty(\CF^\eta(A), \rho_\infty, 2)$.
\end{enumerate}

To prove (a), assume that $\CF^\eta$ S-shatters $A$, so that there exists $g$ such that, for all $B\sub A$, there exists $f\in \CF$ such that $f^\eta(x) \geq g(x) + 1$ for $x\in B$ and $f^\eta(x) \leq g(x) - 1$ for $x \in A \setminus B$. Using the fact that $2\eta f^\eta(x) -1 \leq f(x) < 2 \eta f^\eta(x) + 2\eta - 1$, we get $f(x) \geq 2\eta g(x) +2\eta-1$ for $x\in B$ and $f(x) \leq 2\eta g(x)-1 $ for $x\in A\setminus B$. So taking $\tilde g(x) = 2\eta g(x) + \eta - 1$ as threshold function (which does not depend on $B$), we see that $\CF$ $P_\ga$-shatters $A$ if $\ga\leq \eta$.

For (b), we deduce from the definition of $f^\eta$ that $|f^\eta(x) - \tilde f^\eta(x)| > (2\eta)^{-1}|f(x) - \tilde f(x)| - 1$ so that, if $\ep = 4\eta$,  $|f(x)-\tilde f(x)| \geq \ep$    implies $|f^\eta(x) - \tilde f^\eta(x)| > 1$, or, equivalently $|f^\eta(x) - \tilde f^\eta(x)| \geq 2$.

\noindent {\it Step 3.} We can now conclude. Taking $\ga>0$, we have, if $|A| = N$
\[
\CN(\CF(A), \rho_\infty, \gamma) \leq \CM(\CF(A), \rho_\infty, \ga) \leq \CM(\CF^{\ga/4}(A), \rho_\infty, 2) \leq 2\left(\frac{16N}{\ga^2}\right)^{\lceil\log y\rceil}
\]
with 
\[
y = \sum_{k=1}^D \binom{N}{k} (\ga/4)^{-k} \leq (\ga/4)^{-D}\sum_{k=1}^D \binom{N}{k} \leq \left(\frac{4Ne}{D\ga}\right)^D.
\]
Since the maximum of $\CN(\CF(A), \rho_\infty, \gamma)$ over $A$ with cardinality $N$ is $\CN_\infty(\gamma, N)$, the proof is complete.
\end{proof}

One can use this result to evaluate margin bounds on linear classifiers with bounded data. Let $\CR$ be the ball with radius $\La$ in $\mR^d$ and consider the model class containing all functions $f(x) = a_0 + b^T x$ with $a_0\in [-\Lambda, \Lambda]$ and $b\in \mR^d$, $|b|\leq 1$. Let $A = \{x_1, \ldots, x_N\}$ be a finite subset of $\CR$. Then, $\CF$ $P_\ga$-shatters $A$ if and only if there exists $g_1, \ldots, g_N\in \mR$ such that, for any sequences $ \xi = (\xi_1, \ldots, \xi_N)\in \{-1,1\}^N$ , there exists $a^{\xi}_0\in [-\Lambda,\Lambda]$ and $b^{\xi}\in \mR^d$, $|b^\xi|\leq 1$ with
$\xi_k(a^{\xi}_0  + (b^\xi)^T x_k - g_k) \geq \ga$ for $k=1, \ldots, N$. Summing over $N$, we find that
\[
N\ga + \sum_{k=1}^N g_k\xi_k \leq a^{\xi}_0\sum_{k=1}^N\xi_k + (b^{\xi})^T \sum_{k=1}^N \xi_k x_k\,.
\]
This shows that, for any sequence $\xi_1, \ldots, \xi_N$, 
\[
N\ga + \sum_{k=1}^N g_k\xi_k \leq \La\Big|\sum_{k=1}^N\xi_k\Big| + \Big|\sum_{k=1}^N \xi_k x_k\Big|
\]
Applying the same inequality after changing the signs of $\xi_1, \ldots, \xi_N$ yields
\[
N\ga \leq N\ga + \Big|\sum_{k=1}^N g_k\xi_k\Big| \leq \La\Big|\sum_{k=1}^N\xi_k\Big| + \Big|\sum_{k=1}^N \xi_k x_k\Big|\,.
\]
This shows, in particular, that (letting $\bfxi_1, \ldots, \bfxi_N$ be independent Rademacher random variables)
\[
\myP\left(\La\Big|\sum_{k=1}^N\bfxi_k\Big| + \Big|\sum_{k=1}^N \bfxi_k x_k \Big| \geq N\ga\right) = 1.
\]
However, using the identity $(A+B)^2 \leq 2A^2 + 2 B^2$, we have 
\[
\myP\left(\La\Big|\sum_{k=1}^N\bfxi_k\Big| + \Big|\sum_{k=1}^N \bfxi_k x_k\Big| \geq N\ga\right) 
\leq P\left(2\La^2\Big|\sum_{k=1}^N\bfxi_k\Big|^2 + 2\Big|\sum_{k=1}^N \bfxi_k x_k\Big|^2 \geq N^2\ga^2\right).
\]
Since
\[
\myE\left(2\La^2\Big|\sum_{k=1}^N\bfxi_k\Big|^2 + 2\Big|\sum_{k=1}^N \bfxi_k x_k\Big|^2\right) = 2N\La^2 + 2\sum_{k=1}^n |x_k|^2 \leq 4N\La^2,
\]
Markov's inequality implies 
\[
\myP\left(2\La^2\Big|\sum_{k=1}^N\bfxi_k\Big|^2 + 2\Big|\sum_{k=1}^N \bfxi_k x_k\Big|^2 \geq N^2\ga^2\right)
\leq \frac{4\La^2}{N\ga^2}\,.
\]
We get a contradiction unless $N \leq 4\La^2/\ga^2$, which shows that
$\Pdim[\ga](\CF) \leq 4\La^2/\ga^2$.   \Cref{th:p.gamma} then implies that 
\[
\CN_\infty(\ga, N) \leq 2 \left(\frac{16N}{\ga^2}\right)^{\frac{63\La^2}{\ga^2} \log\left(\frac{16eN\gamma}{\La^2}\right)}
\]
and this upper bound can then be plugged into equations \cref{eq:margin} or \cref{eq:margin.2}
to estimate the generalization error.

Beyond the explicit expression of the upper bound, the important point in the previous argument is that the $P_\ga$-dimension is bounded independently from the dimension $d$ of $X$ (and therefore also applies in the infinite-dimensional case). 
This should be compared to what we found for the VC-dimension of separating hyperplanes, which was $d+1$ (cf.  \cref{prop:vc.lin}).

\begin{remark}
\label{rem:multiple.bounds}
Note that the upper-bound obtained in  \cref{th:margin} depends on a parameter ($\ga$) and the result is true for any choice of this parameter. It is tempting at this point to optimize the bound with respect to $\ga$, but this would be a mistake since a family of events being likely does not imply that their intersection is likely too. However, with a little work, one can ensure that an intersection of slightly weaker inequalities holds. Indeed, assume that an estimate similar to \cref{eq:margin} holds, in the form
\[
\myP(R_0(\hf_\mT) > U_\mT(\ga) + t) \leq C(\ga) e^{-mt^2/2}
\]
or, equivalently
\[
\myP\Big(R_0(\hf_\mT) > U_\mT(\ga) + \sqrt{t^2 + 2\log C(\ga)}\Big) \leq e^{-mt^2/2}\,,
\]
where $U_\mT(\ga)$ depends on data and is increasing (as a function of $\ga$), and $C(\ga)$ is a decreasing function of $\ga$. Consider a decreasing sequence $(\gamma_k)$ that converges to 0 (for example $\ga_k = L2^{-k}$). Choose also an increasing function $\ep(\ga)$. Then
\begin{multline*}
\myP\Big(R_0(\hf_\mT) > \min\{U_\mT(\ga) + \sqrt{t^2+ 2\log C(\ga)+ \ep^2(\ga)}: 0 \leq \ga \leq L\}\Big)\\
 \leq 
\myP\Big(R_0(\hf_\mT) > \min\{U_\mT(\ga_k) + \sqrt{t^2+ 2\log C(\ga_{k-1}) + \ep^2(\ga_k)}: k\geq 1\}\Big)\,.
\end{multline*}
Moreover
\begin{align*}
&\myP\Big(R_0(\hf_\mT) > \min\{U_\mT(\ga_k) + \sqrt{t^2+  2\log C(\ga_{k-1}) + \ep^2(\ga_k)}: k\geq 1\}\Big) \\
&\quad \leq \sum_{k=0}^\infty \myP\Big(R_0(\hf_\mT) >  U_\mT(\ga_k) + \sqrt{t^2+  2\log C(\ga_{k-1}) + \ep(\ga_k)}\Big)\\
&\quad \leq \sum_{k=0}^\infty \frac{C(\ga_k)}{C(\ga_{k-1})}  e^{-m\ep^2(\ga_k)/2 - mt^2/2}\,.
\end{align*}
So, it suffices to choose $\ep(\ga)$ so that
\[
C_0 = \sum_{k=1}^\infty \frac{C(\ga_k)}{C(\ga_{k-1})} e^{-m\ep^2(\ga_k)/2} < \infty
\]
to ensure that
\[
\myP\Big(R_0(\hf_\mT) > \min\{U_\mT(\ga) + \sqrt{t^2+ 2\log C(\ga)+\ep^2(\ga)}: \ga_0 \leq \ga \leq L\}\Big) \leq C_0 e^{-mt^2/2}.
\]
For example, if $\ga_k = L2^{-k}$, one can take 
\[
\ep(\ga) = \sqrt{\frac{2}{m} \left(\log\frac{C(\ga)}{C(\ga/2)} + \log \ga^{-1}\right)}
\]
which yields $C_0 \leq L$. 
\end{remark}

\subsection{Maximum discrepancy}
\label{sec:max.disc}
Let $T$ be a training set and let  $T_1$ and $T_2$ form a fixed partition of the training set in two
equal parts. Assume, for simplicity, that $N$ is even and that the method for selecting the two parts is deterministic, e.g., place the first half of $T$ in $T_1$ and second one in $T_2$. Following \citet{bbl02}, 
one can then define the maximum
discrepancy on $T$ by
$$
C_{T} = \sup_{f\in\CF} (\CE_{T_1}(f) - \CE_{T_2}(f))
$$
 This discrepancy measures the extent to which estimators may differ when trained on
two independent half-sized training sets. 
For a binary classification problem, the estimation of $C_{T}$ can
be made with the same algorithm as the initial classifier, since
$\CE_{T_1}(f) - \CE_{T_2}(f)$ is, up to a constant, exactly the classification error for
the training set in which the class labels are flipped for the data in
$T_2$.

Following \citep{bbl02}, we now discuss concentration bounds that rely on  $C_{T}$ and start with the following Lemma.
\begin{lemma}
\label{lem:max.disc}
Introduce the function
\Eq{
\Phi(T) = \sup_{f\in\CF} (R(f) - \CE_T(f)) - \sup_{f\in\CF} (\CE_{T_1}(f) - \CE_{T_2}(f)).
}
Then $\myE(\Phi(\mT)) \leq 0$. 
\end{lemma}
 \begin{proof}
Note that, if $\mT'$ is a training set, independent of $\mT$ with identical distribution, then, for any $f_0\in \CF$, 
\[ 
R(f_0) - \CE_\mT(f_0) = \myE(\CE_{\mT'}(f_0) - \CE_{\mT}(f_0)\mid\mT)  \leq \myE(\sup_{f\in\CF}( \CE_{\mT'}(f) - \CE_{\mT}(f))\mid \mT)
\]
so that 
\[
\myE(\sup_{f\in\CF} (R(f) - \CE_T(f))) \leq \myE(\sup_{f\in\CF} (\CE_{\mT'}(f) - \CE_{\mT}(f))).
\]
Now, for a given $f$, we have $\CE_{\mT}(f) = \frac12 (\CE_{\mT_1}(f) + \CE_{\mT_2}(f))$ and splitting $\mT'$ the same way, we have $\CE_{\mT'}(f) = \frac12 (\CE_{\mT'_1}(f) + \CE_{\mT'_2}(f))$.

We can therefore  write
\begin{align*}
\myE(\sup_{f\in\CF} (R(f) - \CE_T(f))) & \leq \frac12 \myE(\sup_{f\in\CF} (\CE_{\mT'_1}(f) - \CE_{\mT_1}(f)) + (\CE_{\mT'_2}(f) - \CE_{\mT_2}(f)))\\
& \leq \frac12 \left(\myE(\sup_{f\in\CF} (\CE_{\mT'_1}(f) - \CE_{\mT_1}(f))) + \myE(\sup_{f\in\CF} (\CE_{\mT'_2}(f) - \CE_{\mT_2}(f))\right)\\
&= \myE(\sup_{f\in\CF} (\CE_{\mT_1}(f) - \CE_{\mT_2}(f)))
\end{align*}
where we have used the fact that both $(\mT'_1, \mT_1)$ and $(\mT'_2, \mT_2)$ form random training sets with identical distribution to $(\mT_1, \mT_2)$.

This proves that $\myE(\Phi(\mT)) \leq 0$.
\end{proof}

Using the lemma, one can write
\[
\myP(\sup_{f\in\CF} (R(f) - \CE_\mT(f))\geq C_\mT + \ep) = \myP(\Phi(\mT) \geq \ep) \leq \myP(\Phi(\mT) - \myE(\Phi(\mT)) \geq \ep).
\]
One can then use McDiarmid's inequality (\cref{th:mdd}) after noticing that, letting $z_k = (x_k, y_k)$ for $k=1, \ldots, N$, 
\[
\max_{z_1, \ldots, z_N, z'_k} \big| \Phi(z_1, \ldots, z_N) - \Phi(z_1, \ldots, z_{k-1}, z'_k, z_{k+1}, \ldots, z_N)\big| \leq \frac3N
\]
yielding
\[
\myP(\sup_{f\in\CF} (R(f) - \CE_\mT(f))\geq C_\mT + \ep) \leq e^{-\frac{2N\ep^2}9}.
\]

\subsection{Rademacher complexity}
We now extend the previous definition by computing discrepancies over random two-set partitions of the training set, which have equal size in average.
This leads to the  empirical
Rademacher complexity of the function class. Let $\bfxi_1, \ldots,
\bfxi_N$ be a sequence of Rademacher random variables (equal to -1 and
+1 with equal probability $1/2$). Then, the  (empirical) Rademacher
complexity of the training set $T$ for the model class $\CF$ is
$$
\mathrm{rad}(T) = \myE\Big(\sup_{f\in\CF} \frac{1}{N} \sum_{k=1}^N \bfxi_k
r(Y_k, f(X_k))\ \Big|\ \mT=T\Big).
$$

The mean Rademacher complexity is then the expectation of this
quantity  over the training set distribution. The Rademacher complexity can be computed with a---costly---Monte-Carlo simulation, in which the best estimator is
computed with randomly flipped labels corresponding to the values of $k$ such that  $\xi_k = -1$. 

This measure of complexity was introduced to the machine learning framework in \citet{koltchinskii2002empirical,bartlett2002rademacher}, and Rademacher sums have been extensively studied in relation to empirical processes (cf. \citet{ledoux1991probability}, chapter 4). 

One can  bound the Rademacher complexity in terms of VC dimension. 
\begin{proposition}
\label{prop:rad.vc}
Let $\CF$ be a function class such that  $D= \VC(\CF) <\infty$.
Then
\[
\mathrm{rad}(T) \leq \frac{3}{\sqrt N} \sqrt{2D \log(eN/D) }\,.
\]
\end{proposition}
\begin{proof}
One has, using Hoeffding's inequality
\[
\myP\Big(\sup_{f\in\CF} \frac{1}{N} \sum_{k=1}^N \bfxi_k
r(y_k, f_k) > t\Big) \leq  |\CF(T)| \sup_{f\in \CF} \myP\Big(\frac{1}{N} \sum_{k=1}^N \bfxi_k
r(y_k, f_k) > t\Big) \leq |\CF(T)| e^{-Nt^2/2}.
\]

This implies that 
\[
\myP\Big(\sup_{f\in\CF} \Big|\frac{1}{N} \sum_{k=1}^N \bfxi_k
r(y_k, f_k)\Big| > t\Big) \leq 2 |\CF(T)| e^{-Nt^2/2}
\]
and  \cref{prop:subg.exp} implies 
\[
\mathrm{rad}(T) \leq \frac{3 \sqrt{2 |\CF(T)|}}{\sqrt N}\,.
\]
Therefore if $D= \VC(\CF) <\infty$,  \cref{prop:sauer} implies
\[
\mathrm{rad}(T) \leq \frac{3}{\sqrt N} \sqrt{2D \log(eN/D) }\,.
\]
\end{proof}

We now discuss generalization bounds using Rademacher's complexity.
While we still consider binary classification problems (with $\CR_Y = \{-1, 1\}$), we will assume  that $\CF$ contains functions that can take arbitrary scalar values, and the 0--1 loss function becomes $r(y,y') = \bfone_{yy' \leq 0}$ with $y\in \{-1,1\}$ and $y'\in \mR$. We will also consider functions that dominate this loss, i.e., functions $\rho: \CR_Y\times \CR \to [0,1]$ such that
\[
r(y, y') \leq \rho(y, y')
\]
for all $y\in\CR_Y$, $y'\in \mR$. Some  examples are the margin loss $\rho^*_h(y,y') = \bfone_{yy'\leq h}$ for $h \geq 0$, or the piecewise linear function
\[
\rho_h(y,y') = \left\{
\begin{aligned}
&1 &\text{ if } yy' \leq 0\\
1 &- yy'/h& \text{ if } 0\leq yy' \leq h \\
&0& \text{ if } yy'\geq h
\end{aligned}
\right.
\]

If $\CG$ is a class of functions $g: \CZ \to \mR$, we will denote
\[
\Rad_{\CG}(z_1, \ldots, z_N) = \frac1N E\left(\sup_{g\in\CG} \sum_{k=1}^N \bfxi_k g(z_k)\right)
\]
and
\[
\overline{\Rad}_{\CG}(N) = \myE\big(\Rad_{\CG}(Z_1, \ldots, Z_N)\big)\,.
\]
Our previous notation can then be rewritten as $\mathrm{rad}(T) = \Rad_{\CG}(z_1, \ldots, z_n)$ where $z_i = (x_i,y_i)$ and $\CG$ is the space of functions: $g: (x,y) \mapsto r(y, f(x))$ for $f\in\CF$. 
The following theorem is proved in \citet{koltchinskii2002empirical,bartlett2002rademacher}.  
\begin{theorem}
\label{th:rademacher}
Let $\rho$ be a function dominating the risk function $r(y,y') = \bfone_{yy' \leq 0}$. Let 
\[
\CG^\rho = \{(x,y)\mapsto \rho(y, f(x)) - 1: f\in \CF\}\,
\]
and
\[
\CE_{T}^\rho(f) = \frac1N \sum_{k=1}^N \rho(y_k, f(x_k)).
\]
Then 
\[
\myP(\sup_{f\in\CF} R(f) \geq  \CE_{T}^\rho(f) + 2\overline{\Rad}_{\CG^\rho}(N) + t) \leq e^{-Nt^2/2}
\]
\end{theorem}
\begin{proof}
For $f\in\CF$, we have 
\[
R(f) - \CE^\rho(f) \leq \myE(\rho(Y, f(X))) - \CE^\rho(f)
\leq \Phi(Z_1, \ldots, Z_N)
\]
where
\[
\Phi(Z_1, \ldots, Z_N) = \sup_{g\in\CG^\rho} \Big(\myE(g(Z)) - \frac{1}N \sum_{k=1}^N g(Z_k)\Big)\,.
\]
Since changing one variable among $Z_1, \ldots, Z_N$ changes $\Phi$ by at most $2/N$, McDiarmid's inequality implies that
\[
\myP(\Phi(Z_1, \ldots, Z_N) - \myE(\Phi(Z_1, \ldots, Z_N)) \geq t) \leq e^{-Nt^2/2}.
\]
Now we have
\begin{align*}
\myE(\Phi(Z_1, \ldots, Z_N)) &= \myE\Big(\sup_{g\in\CG^\rho} \myE\Big(\frac1N \sum_{k=1}^N g(Z'_k) - \frac{1}N \sum_{k=1}^N g(Z_k)\, \Big|\, Z_1, \ldots, Z_N\Big)\Big)\\
&\leq  \myE\Big(\sup_{g\in\CG^\rho} \Big(\frac1N \sum_{k=1}^N g(Z'_k) - \frac{1}N \sum_{k=1}^N g(Z_k)\Big)\Big)\\
&\leq  \myE\Big(\myE\Big(\sup_{g\in\CG^\rho} \Big(\frac1N \sum_{k=1}^N \bfxi_k(g(Z'_k) - g(Z_k))\Big)\, \big |\, Z, Z'\Big)\Big)\\
&\leq  2\myE\Big(\myE\Big(\sup_{g\in\CG^\rho} \Big(\frac1N \sum_{k=1}^N \xi_kg(Z_k)\Big)\, \big|\,Z\Big)\Big)\\
&\leq 2\overline{\Rad}_{\CG^\rho}(N)\,,
\end{align*}
of which the statement of the theorem is a direct consequence.
\end{proof}

%

\subsection{Algorithmic Stability}
Another result using McDiarmid's inequality is pro\-ved in \citet{bousquet2002stability}, and is based on the stability of a  classifier when one removes a single example from the training set. As before, we consider training sets $\mT$ of size $N$, where $\mT$ is a random variable.

For $k\in \{1, \ldots, N\}$, and a training set $T = (x_1, y_1, \ldots, x_N, y_N)$, we let $T^{(k)}$ be the training set with sample $(x_k, y_k)$ removed. One says that the predictor $(T\mapsto \hf_T)$ has uniform stability $\be_N$ for the loss function $r$ if, for all $T$ of size $N$, all $k\in \{1, \ldots, N\}$, and all $x,y$:
\begin{equation}
\label{eq:unif.stab}
|r(\hf_T(x), y) - r(\hf_{T^{(k)}} (x), y)| \leq \be_N\,.
\end{equation}

With this definition, the following theorem holds.
\begin{theorem}[\citet{bousquet2002stability}]
Assume that $\hf_T$ has uniform stability $\be_N$ for training sets of size $N$ and that the loss function $r(Y, f(X))$  is almost surely bounded by $M>0$. Then, for all $\ep > 2\be_N$, one has
\[
\myP(R(\hf_\mT) \geq \CE_\mT(\hf_\mT) + \ep) \leq e^{-2N \left(\frac{\ep - 2\be_N}{4N\be_N +M}\right)^2}.
\]
\end{theorem}
Of course, this theorem is interesting only when $\be_N$ is small as a function of $N$, i.e., when $N\be_N$ is bounded.
\begin{proof}
Let $Z_i = (X_i, Y_i)$ and $F(Z_1, \dots, Z_N) = R(\hf_\mT) - \CE_\mT(\hf_\mT)$. We want to apply McDiarmid inequality (\cref{th:mdd}) to $F$, and therefore estimate 
\[
\de_k(F) \defeq  \max_{z_1, \ldots, z_N, z'_k} \big| F(z_1, \ldots, z_N) - F(z_1, \ldots, z_{k-1}, z'_k, z_{k+1}, \ldots, z_N)\big|\,.
\]

Introduce a training set $\tilde T_k$ in which the variable $z_k$ is replaced by $z'_k=(x'_k, y'_k)$. Because $\tilde T_k^{(k)} = T^{(k)}$, we have 
\begin{eqnarray*}
|R(\hf_T) - R(\hf_{\tilde T_k})| &\leq& E(|r(Y, \hf_T(X)) - r(Y, \hf_{\tilde T_k}(X)))|\\
&\leq& E(|r(Y, \hf_T(X)) - r(Y, \hf_{T^{(k)}}(X))|) \\
&&+ E(|r(Y, \hf_{\tilde T_k}(X)) - E(r(Y, \hf_{\tilde T_k^{(k)}}(X)))|)\\
& \leq& 2\beta_N
\end{eqnarray*}

Similarly, we have
\begin{eqnarray*}
|\CE_T(\hf_T) - \CE_{\tilde T_k})(\hf_{\tilde T_k})| &\leq& \frac 1N \sum_{l\neq k} |r(y_l, \hf_T(x_l),) - r(y_l, \hf_{\tilde T_k}(x_l))| \\
&& + \frac1 N |r(y_k, \hf_T(x_k)) - r(y'_k, \hf_{\tilde T_k}(x'_k))|\\
&\leq& \frac 1N \sum_{l\neq k} |r(y_l, \hf_T(x_l)) - r(y_l, \hf_{T^{(k)}}(x_l))| \\
&&+ \frac 1N \sum_{l\neq k} |r(y_l, \hf_{\tilde T_k}(x_l)) - r(y_l, \hf_{\tilde T^{(k)}_k}(x_l))| + \frac M N\\
&\leq & 2\be_N + \frac MN
\end{eqnarray*}
Collecting these results, we find that  $\de_k(F) \leq 4\be_N + \frac MN$, so that, by  \cref{th:mdd}, 
\[
\myP\left(R(\hf_\mT) \geq \CE_\mT(\hf_\mT) + \myE(R(\hf_\mT) - \CE_\mT(\hf_\mT))\right) + \ep) \leq \exp\left(- \frac{2N\ep^2}{(4N\be_N + M)^2}\right).
\]

It remains to evaluate the expectation in this formula. Introducing as above variables $Z'_1, \ldots, Z'_N$ and using the same notation for $\tilde T_k$, we have
\[
\myE(R(\hf_\mT)) = \myE(r(Y'_k, f_\mT(X'_k))) = \myE(r(Y_k, f_{\tilde \mT_k}(X_k)))\,.
\]
Using this, we have
\begin{eqnarray*}
\myE(R(\hf_\mT) - \CE_\mT(\hf_\mT)) &=& \frac1N \sum_{k=1}^N \myE(r(Y_k, f_{\tilde \mT_k}(X_k))-r(Y_k, f_{\mT}(X_k)))\\
&=& \frac1N \sum_{k=1}^N \myE(r(Y_k, f_{\tilde \mT_k}(X_k))-r(Y_k, f_{\tilde \mT_k^{(k)}}(X_k)))\\
&& + \frac1N \sum_{k=1}^N \myE(r(Y_k, f_{\mT^{(k)}_k}(X_k))-r(Y_k, f_{\mT}(X_k))) 
\end{eqnarray*}
from which one deduces that
\[
|\myE(R(\hf_\mT) - \CE_\mT(\hf_\mT))| \leq 2\be_N.
\]
We therefore obtain  
\[
\myP\left(R(\hf_\mT) \geq \CE_\mT(\hf_\mT) + \ep + 2\be_N\right) \leq \exp\left(- \frac{2N\ep^2}{(4N\be_N + M)^2}\right).
\]
as required.
\end{proof}

\subsection{PAC-Bayesian bounds}
\label{sec:pacb}
Our final discussion of concentration bounds for the empirical error uses a slightly different paradigm from that discussed so far. The main difference is that, instead of computing one predictor $\hf_T$ from a training set $T$, it would return a random variable with values in $\CF$, or, equivalently, a probability distribution on $\CF$ (therefore assuming that this space is measurable) that we will denote $\hat\mu_T$. The training set error is now defined by:
\[
\bar \CE_T(\mu) = \int \CE_T(f)  d\mu(f)\,,
\]
for any probability distribution $\mu$ on $\CF$, while the generalization error is:
\[
\bar R(\mu) = \int_{\CF} R(f) d\mu(f)\,.
\]
Our goal is to obtain upper bounds on $\bar R(\mu_\mT) - \bar \CE_\mT(\mu_\mT)$ that hold with high probability. In this framework, we have the following result, in which $\CQ$ denotes the space of probability distributions on $\CF$. 

Assume that the loss function $r$ takes its values in $[0,1]$.
Recall that $\KL(\mu\|\pi)$ is the Kullback-Leibler divergence from $\mu$ to $\pi$, defined by
\[
\KL(\mu\|\pi) = \int_{\CF} \log(\phi(f)) \phi(f) d\pi(f)
\]
if $\mu$ has a density $\phi$ with respect to $\pi$ and $+\infty$ otherwise. Then, the following theorem holds.

\begin{theorem}[\citet{mcallester1999pac}]
\label{th:pacb.1}
With the notation above, for any fixed probability distribution $\pi\in\CQ$, 
\begin{equation}
\label{eq:pacb.1}
\myP\Big(\sup_{\mu\in\CQ} (\bar R(\mu) - \bar\CE_{\mT}(\mu)) > \sqrt{t + \frac{\KL(\mu\|\pi)}{2N}}\Big) \leq 2Ne^{-Nt}.
\end{equation}
\end{theorem}
Taking $t = \log(2N/\de)/2N$, the theorem is equivalent to the statement that, with probability $1-\de$, one has
\begin{equation}
\label{eq:pacb.2}
\bar R(\mu) - \bar\CE_{\mT}(\mu) \leq \sqrt{\frac{\log{2N/\de} + \KL(\mu\|\pi)}{2N}}.
\end{equation}
\begin{proof}
We first show that, for any probability distributions $\pi,\mu$ on $\CF$, and any function $H$ on $\CF$, 
\[
\int_{\CF} H(f) d\mu - \log \int_{\CF} e^{H(f)} d\pi \leq \KL(\mu\|\pi)\,.
\]
Indeed, assume that $\mu$ has a density $\phi$ with respect to $\pi$ (otherwise the upper bound is infinite) and let 
\[
\phi_H = \frac{e^{H}}{\int_{\CF} e^{H(f)} d\pi}\,.
\]
Then 
\begin{align*}
\KL(\mu\|\pi) - \int_{\CF} H(f) d\mu + \log \int_{\CF} e^{H(f)} d\pi &= \int_{\CF} \phi\log\phi d\pi - \int_\CF \phi \log\phi_H d\pi\\
& = \int_{\CF} \frac{\phi}{\phi_H} \left(\log \frac{\phi}{\phi_H} \right) \phi_H d\pi\\
&= \KL(\mu\|\phi_H \pi) \geq 0,
\end{align*}
which proves the result (and also shows that one can only have equality when $\phi = \phi_H$ $\pi$-almost surely.)

Let $\chi(u) = \max(u,0)^2$. We can use this inequality to show that, for any probability $Q\in \CQ$ and $\la>0$,
\[
\la \chi(\bar R(\mu) - \bar\CE_{\mT}(\mu)) \leq \lambda\int_{\CF} \chi(R(f) - \CE_\mT(f)) d\mu(f) \leq KL(\mu\|\pi) + \log \int_{\CF} e^{\la \chi(R(f)-\CE_\mT(f))} d\pi
\]
where we have applied Jensen's inequality to the convex function $\chi$.
This yields
\[
e^{\la \chi(\bar R(\mu) - \bar\CE_{\mT}(Q))} \leq e^{KL(\mu\|\pi)} \int_{\CF} e^{\la \chi(R(f)-\CE_\mT(f))} d\pi.
\]
Hoeffding's inequality implies that, for all $f\in\CF$ and $t\geq 0$
\[
\myP(\chi(R(f)-\CE_\mT(f)) > t) =  \myP(R(f)-\CE_\mT(f) > \sqrt t) \leq e^{-2Nt}
\]
so that 
\begin{align*}
\myE\big(e^{\la \chi(R(f)-\CE_\mT(f))}\big) &=\int_0^\infty \myP(\la \chi(R(f)-\CE_\mT(f)) > \log t) dt \\
&\leq 1 + \int_1^{e^\la} e^{-\frac{2N\log t}{\la}} dt\\
& = 1 + \int_0^{\la} e^{-\frac{2Nu}{\la} +u } du\\
&= 1 +\la \frac{e^{\la - 2N}-1}{\la - 2N}\,.
\end{align*}

From this and Markov's inequality, we get, for any $\la >0$: 
\[
\myP(\sup_{\mu\in\CQ} \chi(\bar R(\mu) - \bar\CE_{\mT}(\mu)) > t + \KL(\mu\|\pi)/\la ) \leq e^{- \la t} \left(1 +\la \frac{e^{\la - 2N}-1}{\la - 2N}\right).
\]
Taking $\la = 2N$ yields
\[
\myP(\sup_{\mu\in\CQ} \chi(\bar R(\mu) - \bar\CE_{\mT}(\mu)) > t + \KL(\mu\|\pi)/2N ) \leq 2Ne^{- 2Nt}, 
\]
which implies
\[
\myP\Big(\sup_{\mu\in\CQ} \bar R(\mu) - \bar\CE_{\mT}(\mu) > \sqrt{t + \KL(\mu\|\pi)/2N} \Big) \leq 2Ne^{- 2Nt}, 
\]
concluding the proof.
%
\end{proof}
\begin{remark}
\label{rem:conc.multiple.bayesian}
Note that the proof, which follows that given in \citet{audibert2007combining}, provides a family of inequalities obtained by taking $\la = 2N/c$ in the final step, with $c>1$. In this case
\[
1 +\la \frac{e^{\la - 2N}-1}{\la - 2N} \leq 1 + \frac{\la}{2N-\la} = \frac{c}{c-1}
\]
and one gets
\[
\myP\Big(\sup_{\mu\in\CQ} \bar R(\mu) - \bar\CE_{\mT}(\mu) > \sqrt{t + c \KL(\mu\|\pi)/2N} \Big) \leq \frac{c}{c-1} 2Ne^{- 2Nt}.
\]
\end{remark}

\begin{remark}
\label{rem:conbayesian}
One special case of  \cref{th:pacb.1} is when $\pi$ is a discrete probability measure supported by a subset $\CF_0$ of $\CF$ and $\mu$ corresponds to a deterministic predictor optimized over $\CF_0$, and is therefore a Dirac measure on some element $f\in \CF_0$. Because $\de_f$ has density $\phi(g) = 1/\pi(g)$ if $g=f$ and 0 otherwise  with respect to $\pi$, we have 
$\KL(\de_f\|\pi) = - \log \pi(f)$ and  \cref{th:pacb.1} implies that, with probability larger than $1-\de$,
\[
R(f) - \CE_{\mT}(f) \leq \sqrt{\frac{\log{2N/\de} - \log\pi(f)}{2N}}.
\]

The term $\log2N$ is however superfluous in this simple context, because one can write, for any $t>0$
\[
\myP\Big(\sup_{f\in\CF_0} R(f) - \CE_\mT(f) \geq \sqrt{t- \frac{\log(\pi(f))}{2N}}\Big) \leq 
\sum_{f\in\CF_0} e^{-2N(t\frac{\log(\pi(f))}{2N})}
= e^{-2Nt}
\]
so that, with probability $1-\de$ (letting $t = \log(1/\de)/2N$), for all $f\in \CF_0$:
\[
R(f) - \CE_\mT(f) \leq \sqrt{\frac{- \log{\de} - \log\pi(f)}{2N}}\,.
\]
\end{remark}

\section{Application to model selection}
We now describe how the previous results can, in principle, be applied to model selection \citep{bbl02}.
We assume that we have a countable family of nested
models classes $(\CF^{(j)}, j\in\CJ)$. 
Denote, as usual, by $\CE_T(f)$  the empirical prediction error in the training
set for a given function $f$. We will denote by $\hf^{(j)}_T$ a minimizer of the in-sample error for $\CF^{(j)}$, such that
\[
\CE_T(\hf^{(j)}_T) = \min_{f\in \CF^{(j)}} \CE_T(f).
\]
In the model selection problem, 
one would like to determine the best model class, $j=j(T)$, such that the prediction error $R(\hat f^{(j)}_T)$ is minimal, or, more realistically, determine $j^*$ such that $R(\hat f^{(j_*)}_T)$ is not too far from the optimal one.

We will consider penalty-based methods in which one minimizes $\tilde \CE_T(f) = \CE_T(f) + C_T(j)$ to determine $j(T)$. The penalty, $C_T$, may also be data-dependent, and will therefore be a random variable. The previous concentration inequalities provided highly probable upper-bounds for
$R(\hf^{(j)}_\mT)$, each exhibiting a random variable $\Gamma_\mT^{(j)}$ that is larger than
$R(\hf^{(j)}_\mT)$ with probability close to one. More precisely, we obtained inequalities taking the form (when applied to $\CF^{(j)}$)
\begin{equation}
\label{eq:basic}
\myP(R_\mT(\hf^{(j)}) \geq \Gamma_\mT^{(j)} + t) \leq c_j e^{-mt^2}
\end{equation}
for some known constants $c_j$ and $m$. For example, the VC-dimension bounds have $\Gamma_\mT^{(j)} = \CE_\mT(\hf^{(j)}_\mT)$, $c_j = 2\CS_{\CF^{(j)}}(2N)$ and $m = N/8$.  

Given such inequalities, one can develop a model selection strategy that relies on {\it a priori} weights, provided by a sequence $\pi_j$ of positive numbers such that $\sum_{j\in\CJ} \pi_j = 1$. Define
\[
\tilde \pi_j = \frac{\pi_j /c_j}{\sum_{j'=1}^\infty \pi_{j'}/c_{j'}},
\] 
and let
$$
C_T^{(j)} = \Gamma_T^{(j)} - \CE_T(\hf^{(j)}_T) + \sqrt{-\frac{\log \tilde \pi_j}{m}}
$$
yielding a penalty-based method that requires the minimization of 
\[
\tilde \CE_T(f) =  (\CE_T(f) - \CE_T(\hf^{(j)}_T)) + \Gamma_T^{(j)} + \sqrt{-\frac{\log \tilde \pi_j}{m}}\,.
\]
The selected model class is then $\CF^{(j^*)}$ where $j^*$ minimizes  $\Gamma_T^{(j)} + \sqrt{-\frac{\log \tilde \pi_j}{2m}}$. 

The same proof as that provided at the end of  \cref{sec:pacb} justifies this procedure. Indeed, for $t>0$,
\begin{align*}
\myP\left(R(\hf_\mT) - \tilde \CE_\mT(\hf_\mT) \geq t\right) & \leq \myP\left(\max_{j} (R(\hf^{(j)}_\mT) - \tilde \CE_\mT(\hf^{(j)}_\mT)) \geq t\right)\\
& \leq \myP\left(\max_{j} (R(\hf^{(j)}_\mT)  \geq R^*_j + t + \sqrt{-\frac{\log \tilde \pi_j}{m}}\right)\\
&\leq  \tilde c \sum_j \pi_j e^{-mt^2} \\
&\leq \tilde ce^{-mt^2}
\end{align*}
with $\tilde c = \sum_{j=1}^\infty \pi_j/c_j$.
\newpage
\markright{PROBLEMS}
\section*{Problems}
\addcontentsline{toc}{section}{Problems}
\input Problems_Concentration

\bibliography{machine_learning}

\end{document}

\chapter{Problems with Solutions}
\setcounter{chapter}{1}
\solutionstrue
\uselabelfalse

\chapter{Problems with solutions}
\markright{PROBLEMS WITH SOLUTIONS}
\section*{Basic concepts}
\stepcounter{chapter}
\input Problems_Bayes_Rule
\section*{Matrix Analysis}
\stepcounter{chapter}
\section*{Reproducing Kernels}
\stepcounter{chapter}
\input Problems_Kernels
\section*{Optimization}
\stepcounter{chapter}
\input Problems_Optimization
\section*{Linear Regression}
\stepcounter{chapter}
\input Problems_Linear_Regression
\section*{Linear Classification}
\stepcounter{chapter}
\input Problems_Classification
\section*{Nearest Neighbors}
\stepcounter{chapter}
\section*{Decision Trees}
\stepcounter{chapter}
\section*{Neural Networks}
\stepcounter{chapter}
\input Problems_Neural_Networks
\section*{Posterior Distributions}
\stepcounter{chapter}
\section*{Clustering}
\stepcounter{chapter}
\input Problems_Clustering
\section*{Factor Analysis}
\stepcounter{chapter}
\input Problems_Factor_Analysis
\section*{Manifold Learning}
\stepcounter{chapter}
\input Problems_Manifold_learning
\section*{Generalization Bounds}
\stepcounter{chapter}
\input Problems_Concentration

\end{document}